\documentclass[conference]{IEEEtran}
\def\IEEEtitletopspace{18pt}
\IEEEoverridecommandlockouts
\usepackage{cite}
\usepackage{amsmath,amssymb,amsfonts}
\usepackage{graphicx}
\usepackage{textcomp}

\def\BibTeX{{\rm B\kern-.05em{\sc i\kern-.025em b}\kern-.08em
    T\kern-.1667em\lower.7ex\hbox{E}\kern-.125emX}}

\usepackage{times}
\usepackage{epsfig}
\usepackage{graphicx}
\usepackage{float}
\usepackage{wrapfig}
\usepackage{amsmath,amssymb,amsthm}
\usepackage{algorithm,algorithmicx,algpseudocode}
\usepackage{bm,xspace}
\usepackage{comment}
\usepackage{verbatim}
\usepackage{multirow}
\usepackage{balance}
\usepackage{url}
\usepackage{booktabs}
\usepackage{etoolbox}
\usepackage{siunitx}
\usepackage{calc}
\usepackage{pifont,hologo}
\usepackage[usenames, dvipsnames]{xcolor}
\usepackage{nicefrac}
\usepackage[normalem]{ulem}

\usepackage[super]{nth}
\usepackage{nicefrac}
\robustify\bfseries
\robustify\uline
\def\myuline#1{#1\llap{\uline{\phantom{#1}}}}

\DeclareMathSymbol{@}{\mathord}{letters}{"3B}

\newcommand\timess{\mathbin{\!\times\!}}

\newcommand\mypara[1]{\vspace{1mm}\noindent\textbf{#1}}

\def\rot#1{\rotatebox{90}{#1}}

\def\latex/{\LaTeX}
\def\bibtex/{\hologo{BibTeX}}

\newcommand{\fg}[1]{\textcolor{blue}{$(#1\%)$}}

\begin{document}

\title{Monocular Visual-Inertial Depth Estimation\\
}

\author{Diana Wofk$^{1}$, Ren{\'e} Ranftl$^{1}$, Matthias M{\"u}ller$^{1}$, and Vladlen Koltun$^{1,2}$
\thanks{$^{1}$This work was done at Intel Labs.}%
\thanks{$^{2}$Currently affiliated with Apple.}%
}

\maketitle

\begin{abstract}

We present a visual-inertial depth estimation pipeline that integrates monocular depth estimation and visual-inertial odometry to produce dense depth estimates with metric scale. Our approach performs global scale and shift alignment against sparse metric depth, followed by learning-based dense alignment. We evaluate on the TartanAir and VOID datasets, observing up to 30\% reduction in inverse RMSE with dense scale alignment relative to performing just global alignment alone. Our approach is especially competitive at low density; with just 150 sparse metric depth points, our dense-to-dense depth alignment method achieves over 50\% lower iRMSE over sparse-to-dense depth completion by KBNet, currently the state of the art on VOID. We demonstrate successful zero-shot transfer from synthetic TartanAir to real-world VOID data and perform generalization tests on NYUv2 and VCU-RVI. Our approach is modular and is compatible with a variety of monocular depth estimation models. 

\end{abstract}


\section{Introduction}
\label{sec:intro}

Depth perception is fundamental to visual navigation, where correctly estimating distances can help plan motion and avoid obstacles. Accurate depth estimation can also aid scene reconstruction, mapping, and object manipulation. Some applications of estimated depth benefit when it is \textit{metrically accurate}---when every depth value is provided in absolute metric units and represents physical distance.

Algorithms for dense depth estimation can be broadly grouped into several categories. Stereo-based approaches rely on two or more cameras that capture different views. Structure-from-motion (SfM) tries to estimate scene geometry from a sequence of images taken by a moving camera, but it is difficult to recover depth with absolute scale since the relative pose of the camera across images is not known. Monocular approaches require just one camera and try to estimate depth from a single image. Such approaches are appealing since simple RGB cameras are compact and ubiquitous. However, monocular approaches that rely solely on visual data still exhibit scale ambiguity.

\begin{figure}[t]
\centering
  \begin{tabular}{@{}*{4}{c@{\hspace{0.5mm}}}c@{}}
    \vspace{-0.75mm}
    \includegraphics[width=0.24\linewidth]{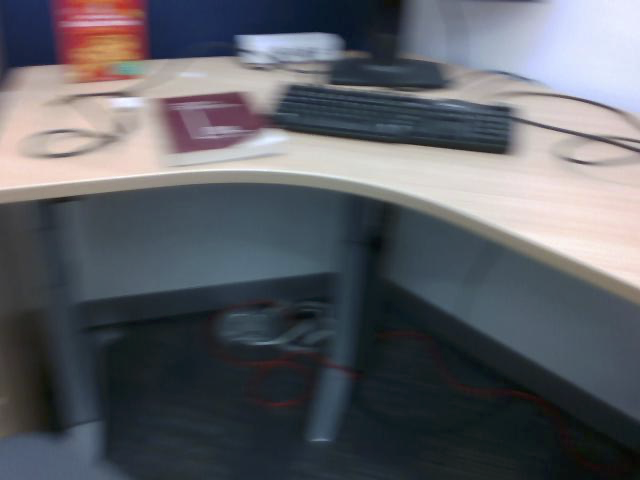}&
    \includegraphics[width=0.24\linewidth]{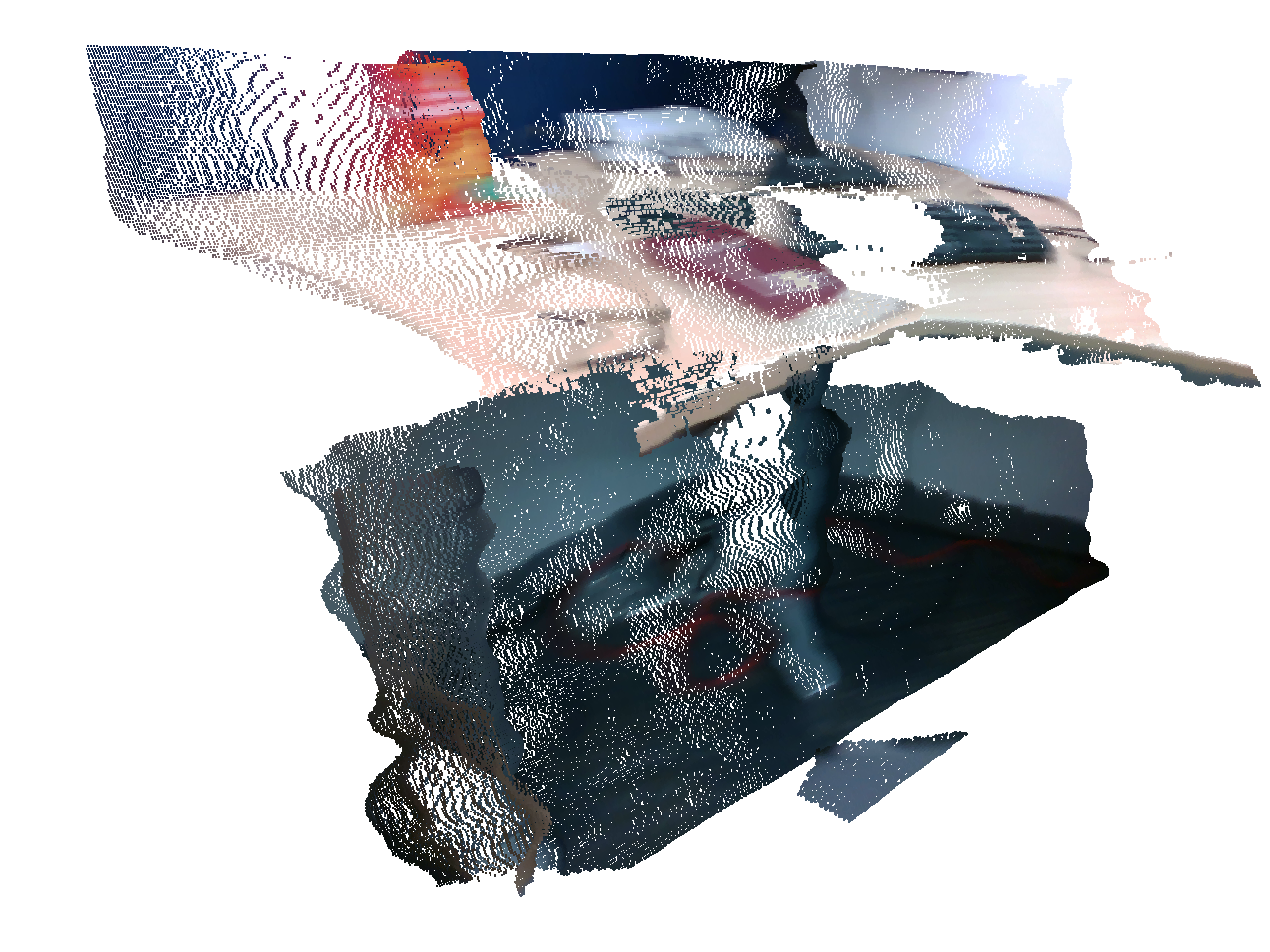}&
    \includegraphics[width=0.24\linewidth]{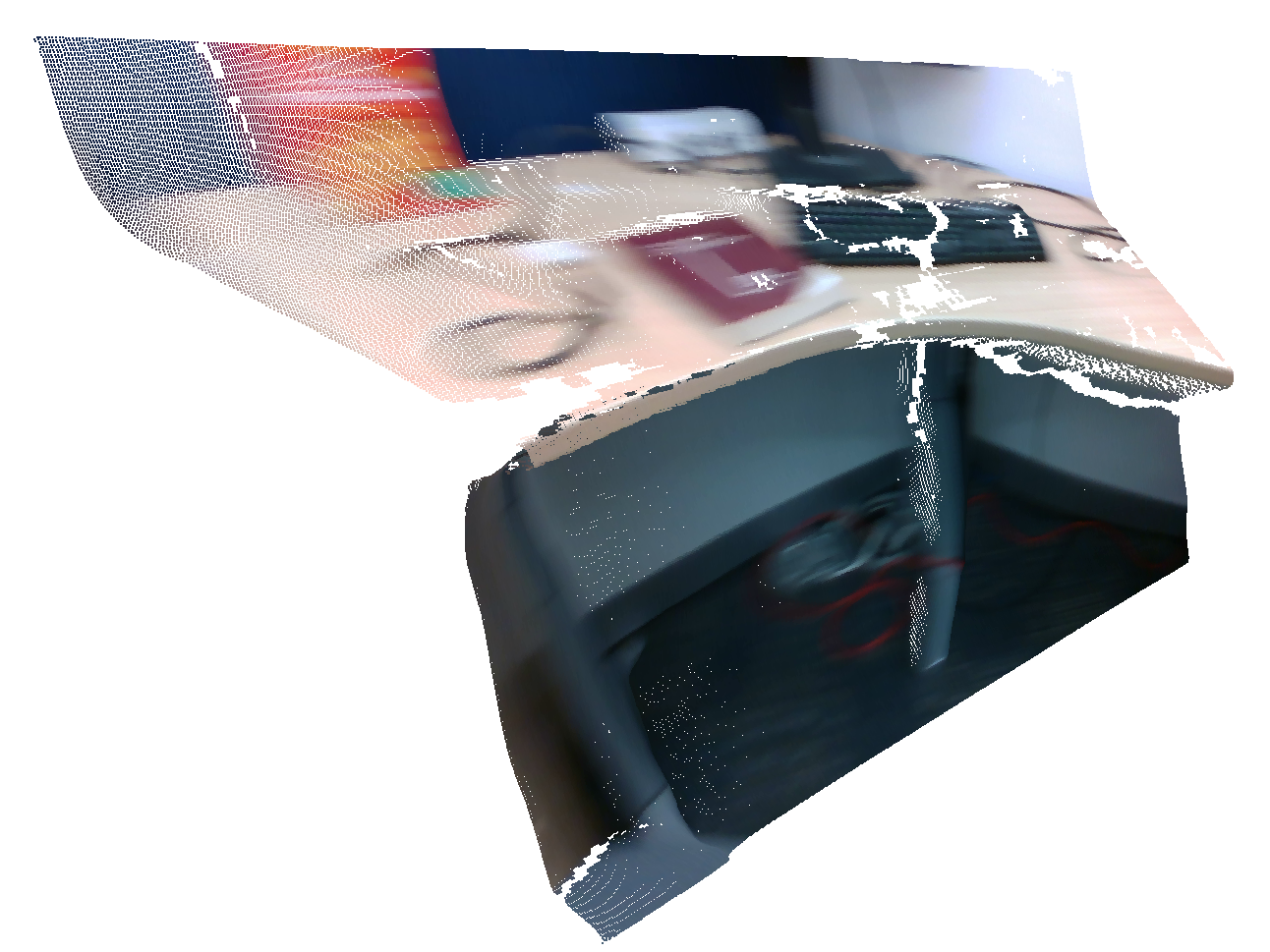}&
    \includegraphics[width=0.24\linewidth]{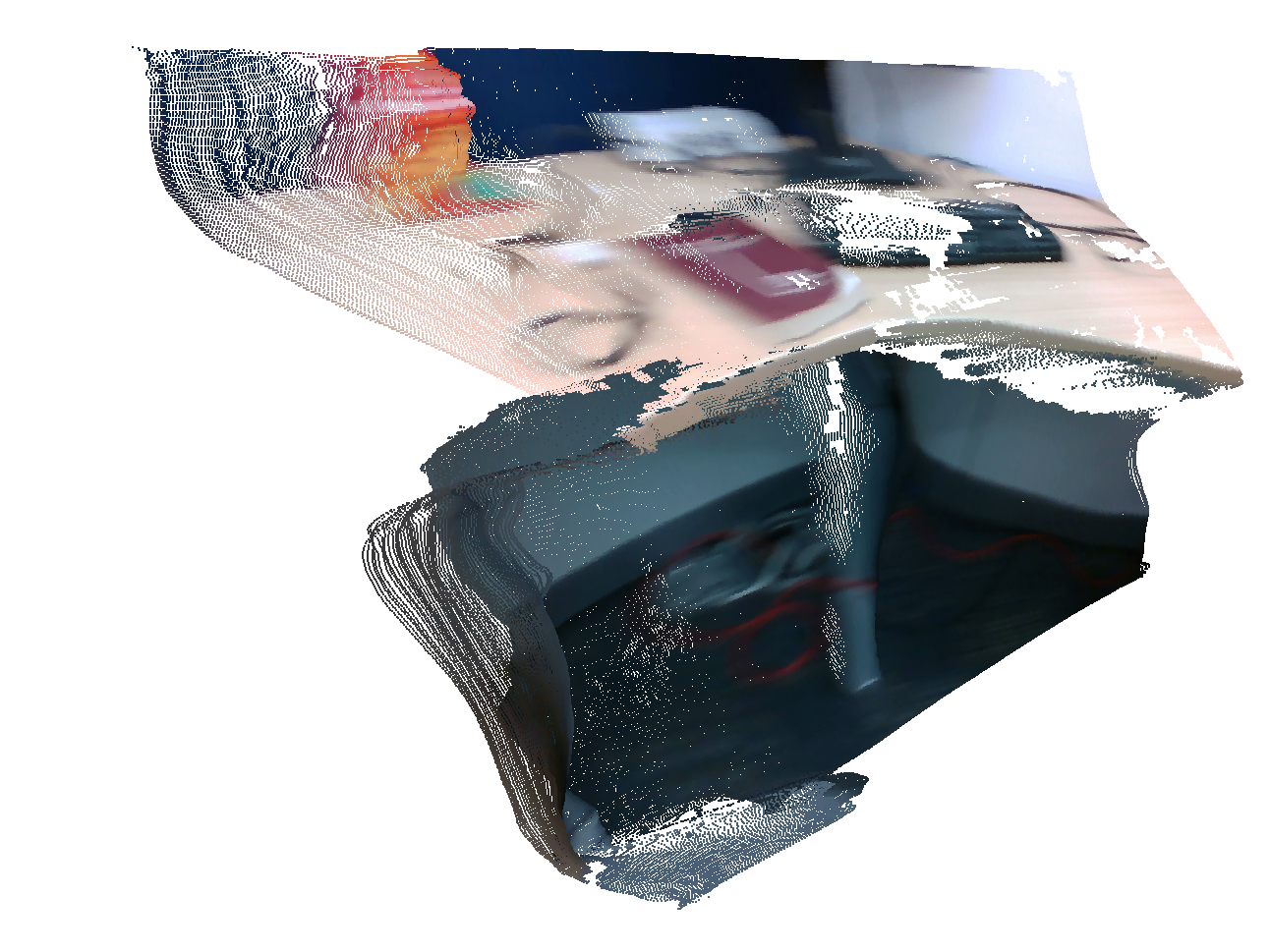}\\
    \scriptsize RGB & \scriptsize ground truth & \scriptsize GA output & \scriptsize GA+SML output\\
  \end{tabular}
  \caption{We integrate visual-inertial odometry and monocular depth estimation to produce dense depth with metric scale. Global alignment (GA) determines appropriate global scale, while dense scale alignment (SML) operates locally and pushes or pulls regions towards correct metric depth. Here, with GA+SML, objects are aligned more accurately, the center desk leg is straightened, and the top of the desk is pulled forward.}
  \label{fig:intro_teaser}
  \vspace{-12pt}
\end{figure}

Incorporating inertial data can help resolve scale ambiguity, and most mobile devices already contain inertial measurement units (IMUs). Simultaneous localization and mapping (SLAM) systems \cite{Engel2014LSDSLAMLD,MurArtal2015ORBSLAMAV,MurArtal2017ORBSLAM2AO} use visual or visual-inertial data to track scene landmarks under camera motion, compute the camera trajectory, and map the traversed environment. However, SLAM systems typically only track on the order of hundreds to thousands of sparse feature points, resulting in metric depth measurements that are only semi-dense at best. Our work explores how to use inertial data in conjunction with monocular visual data to produce fully-dense metrically accurate depth predictions as in Figure \ref{fig:intro_teaser}.

Recent advances in supervised learning-based monocular depth estimation \cite{Ranftl2020,Ranftl2021} provide high generality but do not resolve absolute metric scale and only predict relative depth. Works that use inertial data to inform metric scale typically perform depth completion given a set of known sparse metric depth points and tend to be self-supervised in nature due to a lack of visual-inertial datasets \cite{Wong2020void,Wong2021kbnet}. We seek to bridge these approaches by leveraging monocular depth estimation models trained on diverse datasets and recovering metric scale for individual depth estimates.

Our approach performs least-squares fitting of monocular depth estimates against sparse metric depth, followed by learned local per-pixel adjustment. This combination of global and dense (local) depth alignment successfully rectifies metric scale, with dense alignment consistently outperforming a purely global alignment baseline. Alignment succeeds with just 150 metric depth anchors and is robust to zero-shot cross-dataset transfer. Our pipeline is modular and is agnostic to the monocular depth estimation model and VIO system being used; it should thus benefit from continual improvement in these modules.

\section{Related Work}

Monocular depth estimation is inherently an ill-posed problem facing challenges like scale ambiguity. A common approach to handling this in supervised training has been to limit training data to particular datasets with desired environments, e.g., indoor or outdoor scenes. This encourages the supervised network to memorize a metric scale that may be globally inconsistent, results in overfitting to specific depth ranges, and hurts generalizability across environments. Recent work on dataset mixing and training loss construction \cite{Ranftl2020} has enabled robust affine-invariant monocular depth estimation across a variety of datasets. However, recovering absolute metric scale in these depth estimates remains a challenge.

\mypara{Using inertial and pose information.} Incorporating inertial data is being explored as a means of improving metric depth accuracy in self-supervised depth estimation approaches. Fei et al.~\cite{Fei2019geo} propose using global orientation from inertial measurements to regularize depth regression at training time, with an expanded loss function that penalizes planarity deviation based on gravity vectors estimated through VIO. SelfVIO \cite{Almalioglu2019SelfVIOSD} combines learning-based VIO and depth estimation to develop an adversarially trained architecture that jointly estimates ego-motion and dense depth from input RGB and IMU readings. A number of additional works incorporate pose into supervised and unsupervised approaches \cite{Patil2020DontFT,Teed2020DeepV2D,Liu2019NeuralRS,Xie2020VideoDE}, often as part of pose consistency and reprojection terms, or as a pose estimation task that is performed jointly with depth estimation. In the latter case, replacing pose networks with pose estimation from VIO/SLAM is known to improve performance~\cite{Fei2019geo,Wong2020void}.

\mypara{Depth completion from sparse depth.} Sparse depth maps or sparse point clouds, e.g., obtained with LiDAR or through VIO tracking, commonly serve as input to metric depth completion. In VOICED \cite{Wong2020void}, sparse depth from VIO is used as a depth scaffold that is refined to minimize photometric, pose, and depth consistency losses. KBNet~\cite{Wong2021kbnet} adds camera calibration and connects sparse depth and RGB encoders with backprojection layers. Other recent works also explore visual-inertial depth completion~\cite{Sartipi2020DeepDE,Merrill2021RobustMV}, although they rely on depth completion networks that are trained primarily on indoor data, thus limiting generality.

\mypara{Video depth estimation.} In the absence of inertial data, given an ordered sequence of images, temporal correlation can be used to improve scale consistency of monocular depth estimates, though still without absolute scale. CVD \cite{Luo2020cvd} leverages SfM \cite{schoenberger2016sfm} to estimate camera parameters and define geometric constraints that help resolve global scale consistency across per-frame depth maps predicted from monocular video input. Since SfM may fail under challenging motion, Robust CVD \cite{Kopf2021rcvd} replaces it with pose estimation and optimization done jointly with depth scale realignment based on a bilinear spline. In both methods, absolute metric scale remains unknown.

Our work aims to resolve scale ambiguity by performing global and local depth alignment in absolute metric space, given an off-the-shelf monocular depth estimation model and VIO system. Instead of designing novel depth estimation architectures and training procedures, we build upon existing monocular depth models and realign their output depth estimates. We do not perform depth completion~\cite{Wong2020void,Luo2020cvd}, but rather align an already-dense depth map to absolute metric scale. This is a more versatile approach as it can incorporate arbitrary monocular depth estimation models.

\begin{figure*}[t]
  \centering
   \includegraphics[width=0.80\linewidth]{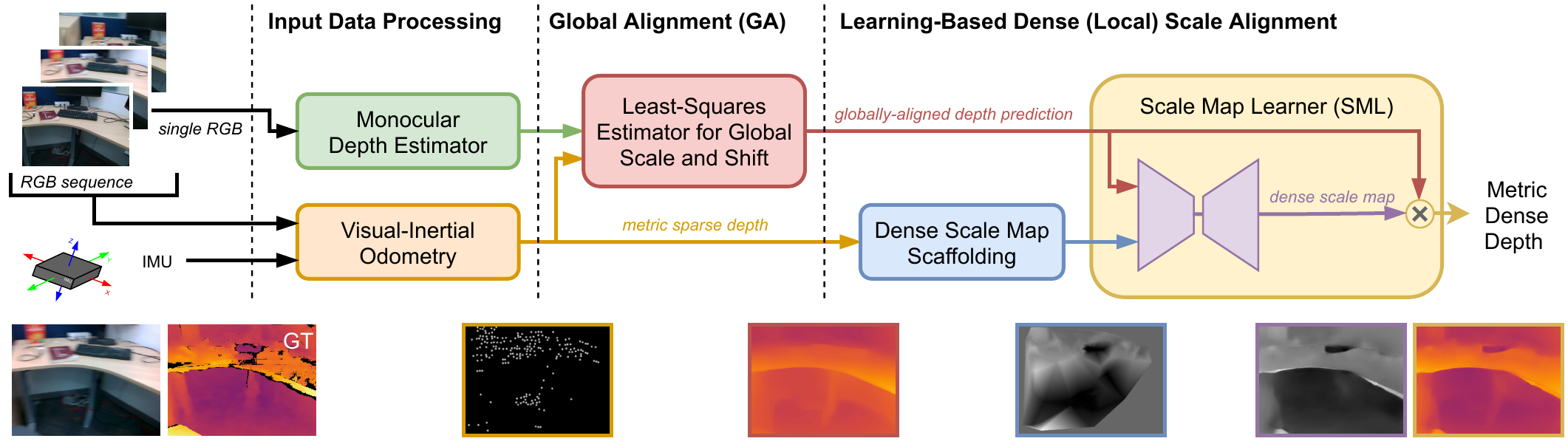}
   \caption{Our visual-inertial depth estimation pipeline. There are three stages: (1) input processing, where RGB and IMU data feed into monocular depth estimation alongside visual-inertial odometry, (2) global scale and shift alignment, where monocular depth estimates are fitted to sparse depth from VIO in a least-squares manner, and (3) learning-based dense scale alignment, where globally-aligned depth is locally realigned using a dense scale map regressed by the ScaleMapLearner (SML). The row of images at the bottom illustrate a VOID \cite{Wong2020void} sample being processed through the pipeline; from left to right: the input RGB, ground truth depth, sparse depth from VIO, globally-aligned depth, scale map scaffolding, dense scale map regressed by SML, final depth output.}
   \label{fig:methodology}
  \vspace{-12pt}
\end{figure*}

\section{Method}

We develop a modular three-stage pipeline for visual-inertial depth estimation. Its structure is illustrated in Figure \ref{fig:methodology}.

\mypara{Monocular depth estimation.} The \textit{visual} branch of our pipeline predicts depth from a single image. This is done using a pretrained model that takes in a single RGB image and produces a dense depth map up to some scale.
Monocular processing is appealing as it allows for low-complexity architectures that do not carry large computational costs. 

Our approach is compatible with  traditional convolutional models as well as newer architectures. We select DPT-Hybrid \cite{Ranftl2021} as our depth estimator; this is a transformer-based model trained on a large meta-dataset using scale- and shift-invariant losses. While it achieves high generalizability, its output measures depth relations between pixels, and depth values do not carry metric meaning. Our alignment pipeline aims to recover metric scale for every pixel in this output depth map.

\mypara{Visual-inertial odometry.} The \textit{inertial} branch of our pipeline uses IMU data together with visual data to determine metric scale. Given a sequence of RGB images with synchronized IMU data, we run VINS-Mono \cite{Qin2018} to compute the camera trajectory and yield a set of 3D world coordinates of features tracked throughout the sequence. In a reasonably textured environment, we can expect tens of tracked features per frame. By projecting feature coordinates to image space, we obtain a sequence of sparse maps containing metric depth values. These sparse depth maps serve as inputs to later alignment tasks, thereby propagating metric scale through our pipeline. 

\mypara{Global scale and shift alignment (GA).} Let $\mathbf{z}$ refer to unit-less affine-invariant inverse depth that is output by a monocular depth estimation model such as DPT-Hybrid. To reintroduce metric scale into depth, we align monocular depth estimates to sparse metric depth obtained through VIO. This global alignment is performed in inverse depth space based on a least-squares criterion \cite{Ranftl2020}. The result is a per-frame global scale $s_g$ and global shift $t_g$ that are applied to $\mathbf{z}$ as a linear transformation. Applying global scale can be interpreted as bringing depth values to a correct order of magnitude, while applying global shift can help undo potential bias or offset in the original prediction. The resulting globally-aligned depth estimates are $\tilde{\mathbf{z}} = s_g\mathbf{z}+t_g$. 

\mypara{Dense (local) scale alignment.} Due to its coarse nature, global alignment will not adequately resolve metric scale in all regions of a depth map. To address this, we propose a learning-based approach for determining dense (per-pixel) scale factors that are applied to globally-aligned depth estimates. Using MiDaS-small~\cite{Ranftl2020}, we construct a network that is trained to realign individual pixels in a depth map to improve their metric accuracy. We call this network the ScaleMapLearner (SML) and feed it an input of two concatenated data channels: the \textit{globally-aligned depth} $\tilde{\mathbf{z}}$, and a \textit{scaffolding for a dense scale map}, where $n$ locations of known sparse depth values $\mathbf{v}$ from VIO define $n$ scale anchor points $\mathbf{v}_{i}/\tilde{\mathbf{z}}_{i}$, $i\in\{1...n\}$. The region within the convex hull defined by the anchors is filled via linear interpolation of anchor values. The region outside the convex hull is filled with an identity scale value of 1.

SML regresses a dense scale residual map $\mathbf{r}$ where values are allowed to be negative. We compute the resulting scale map as $\text{ReLU}(1+\mathbf{r})$ and apply it to the input depth $\tilde{\mathbf{z}}$ to produce the output depth $\hat{\mathbf{z}} = \text{ReLU}(1+\mathbf{r})\tilde{\mathbf{z}}$. 

\mypara{Loss function.} During training, the SML network is supervised on metric ground truth $\mathbf{z}^{*}$ in inverse depth space. Let $M$ define the number of pixels with valid ground truth. Our loss function comprises two terms: an L1 loss on depth,
\begin{equation}
    \mathcal{L}_{depth}(\hat{\mathbf{z}},\mathbf{z}^{*}) = \frac{1}{M}\sum_{i=1}^{M}|\mathbf{z}^{*}_{i}-\hat{\mathbf{z}}_{i}|
\end{equation}
and a multiscale gradient matching term \cite{MegaDepthLi2018} that biases discontinuities to coincide with discontinuities in ground truth,
\begin{equation}
    \mathcal{L}_{grad}(\hat{\mathbf{z}},\mathbf{z}^{*}) = \frac{1}{K}\sum_{k=1}^{K}\frac{1}{M}\sum_{i=1}^{M}(|\nabla_{x}R_{i}^{k}|+|\nabla_{y}R_{i}^{k}|)
\end{equation}
where $R_{i}=\mathbf{z}^{*}_{i}-\hat{\mathbf{z}}_{i}$ and $R^{k}$ denotes error at different resolutions. We use $K=3$ levels, halving the spatial resolution at each level. Our final loss is $\mathcal{L} = \mathcal{L}_{depth} + 0.5\mathcal{L}_{grad}$. 

\mypara{Decoupling visual and inertial data.} Our pipeline runs monocular depth estimation and VIO in parallel and independently of each other. The intermediate outputs from these steps are then fused together to generate inputs to SML. This design choice is made to better leverage ongoing advances in monocular depth and VIO systems; newly developed modules can be easily integrated within our pipeline, and SML can be quickly retrained to benefit from the improved performance of those modules. We contrast this with designing a single unified network that learns metric depth directly from a joint RGB-IMU input. A sufficiently large corpus of RGB-D datasets containing IMU data to train such a network and have it generalize well does not exist. We still face a data challenge when training SML; however, by decoupling RGB-to-depth and VIO at the input, we provide SML with an intermediate data representation that simplifies what it needs to learn to perform metric depth alignment. In this setting, a smaller amount of training data is sufficient.
\section{Datasets and  Experiments}

A key challenge in acquiring training data for the SML network is the lack of RGB-D+IMU datasets. In our pipeline, IMU data is needed to run VIO to generate sparse metric depth. While simulators allow recording synchronized RGB-D and IMU data~\cite{AirSim2017}, manually gathering sufficient training data is difficult. We select TartanAir \cite{TartanAir2020} for its large size and variety of outdoor and indoor sequences. IMU data is not provided in this dataset. To proxy sparse depth map generation, we run the VINS-Mono feature tracker front-end~\cite{Lusk2018} to obtain sparse feature locations and then sample ground truth depth at those locations. We use a 70\%-30\% train-test split for TartanAir, with 172K training and 73K test samples taken from both easy and hard sequences.


In addition to the synthetic TartanAir dataset, we benchmark on VOID \cite{Wong2020void}, which offers real-world data collected using an Intel RealSense D435i camera and the VIO system XIVO \cite{Fei2019xivo}. This dataset is smaller than TartanAir, with only 47K training and 800 test samples. We use the published train-test split. 

\mypara{Setup.} We use MiDaS-small \cite{Ranftl2020} to construct our SML network. The encoder backbone is initialized with pretrained ImageNet \cite{Deng2009imagenet} weights, while other layers are initialized randomly. We use AdamW \cite{Loshchilov2019} with $\beta_{1}=0.9$, $\beta_{2}=0.999$, and $\lambda=0.001$. We set an initial learning rate of $5\times10^{-4}$ when training on TartanAir and $3\times10^{-4}$ on VOID. We additionally use a step-based scheduler that halves the learning rate after 5 epochs on TartanAir and after 8 epochs on VOID. We train for 20 epochs on a node with 8 GeForce RTX 2080 Ti GPUs, with a batch size of 256, and with mixed-precision training enabled. Input data is resized and cropped to a training resolution of $384\timess384$. Training takes up to 4 hours.

\mypara{Metrics.} We mainly evaluate in inverse depth space $\mathbf{z}=1/\mathbf{d}$ (in km$^{-1}$), as doing so penalizes error at closer depth ranges more significantly. We compute inverse mean absolute error iMAE $= \frac{1}{M}\sum_{i=1}^{M}|\mathbf{z}^{*}_{i}-\hat{\mathbf{z}}_{i}|$, inverse root mean squared error iRMSE $= [\frac{1}{M}\sum_{i=1}^{M}|\mathbf{z}^{*}_{i}-\hat{\mathbf{z}}_{i}|^{2}]^\frac{1}{2}$, and inverse absolute relative error iAbsRel $= \frac{1}{M}\sum_{i=1}^{M}|\mathbf{z}^{*}_{i}-\hat{\mathbf{z}}_{i}|/\mathbf{z}^{*}_{i}$. On VOID, we also compute MAE and RMSE in regular depth space $\mathbf{d}$ (in mm). 

We follow the VOID evaluation protocol of Wong et al.~\cite{Wong2020void,Wong2021kbnet} and consider ground truth depth to be valid between 0.2 and 5.0 meters. The minimum and maximum depth prediction values in these works are set to 0.1 and 8.0 meters, respectively. We clamp depth predictions, both after global alignment and after applying regressed dense scale maps, to this range. In contrast to the mostly-indoor scenes in VOID, outdoor scenes in TartanAir exhibit larger depth ranges. For TartanAir, we define ground truth depth to be valid between 0.2 and 50 meters and clamp predictions between 0.1 and 80 meters.

\subsection{Evaluation on TartanAir}

We first evaluate on \emph{synthetic} samples from the TartanAir dataset, where inertial data is unknown. To proxy sparse depth generation from VIO, we preprocess TartanAir data with a sparsifier that samples depth from the ground truth at locations determined via the VINS-Mono-based feature tracker implemented in \cite{Lusk2018}. We run monocular depth estimation with DPT-Hybrid, perform global alignment against metric sparse depth, and generate a scale map scaffolding for every sample prior to SML training. We define our baseline as global alignment only and show that performing dense scale alignment with SML improves metric depth accuracy. Table \ref{tab:tartanair_eval} shows that SML achieves 30\%, 17\%, and 26\% reduction in iMAE, iRMSE, and iAbsRel, respectively. Metrics are aggregated across a set of 690 samples taken from our TartanAir test split.

\begin{table}
  \footnotesize
  \centering
    \caption{Evaluation on TartanAir. Lower is better for all metrics.} 
    \label{tab:tartanair_eval}
    \vspace{-4pt}
    \begin{tabular}{@{}l@{\hspace{5mm}}c@{\hspace{5mm}}*{2}{S[table-format=2.2]@{\hspace{2mm}}}*{1}{S[table-format=1.3]@{\hspace{0mm}}}@{}}
    \toprule
    Method  & Depth Model                   & {iMAE}          & {iRMSE}         & {iAbsRel}       \\
    \midrule
    GA only & \multirow{2}{*}{DPT-Hybrid}   &           22.94 &           35.49 &           0.126 \\
    GA+SML  &                               & \bfseries 16.11 & \bfseries 29.48 & \bfseries 0.093 \\
    \midrule
    GA only & \multirow{2}{*}{MiDaS v2.0}   & 58.11           & 79.34           & 0.299 \\
    GA+SML  &                               & \bfseries 28.79 & \bfseries 46.67 & \bfseries 0.156 \\
    \bottomrule
  \end{tabular}
  \vspace{-12pt}
\end{table}

\begin{figure}
\centering
  \begin{tabular}{@{}*{6}{c@{\hspace{0.5mm}}}c@{}}
    {\scriptsize RGB Image} & {\scriptsize GA Depth} & {\scriptsize SML Depth} & {\scriptsize GT Depth} & {\scriptsize GA Error} & {\scriptsize SML Error}\\
    \vspace{-0.75mm}
    \includegraphics[width=0.155\linewidth]{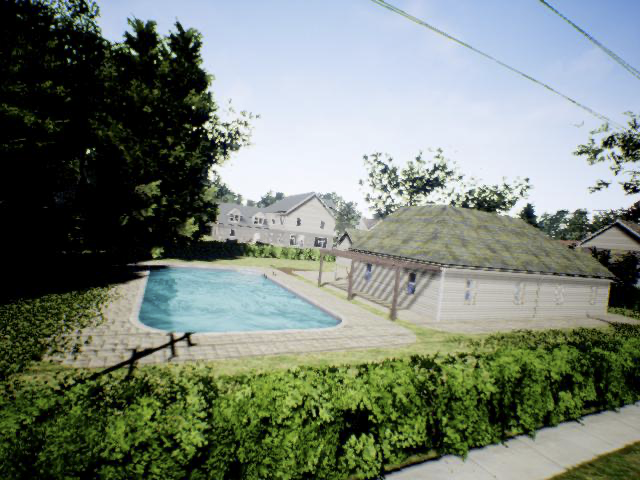}&
    \includegraphics[width=0.155\linewidth]{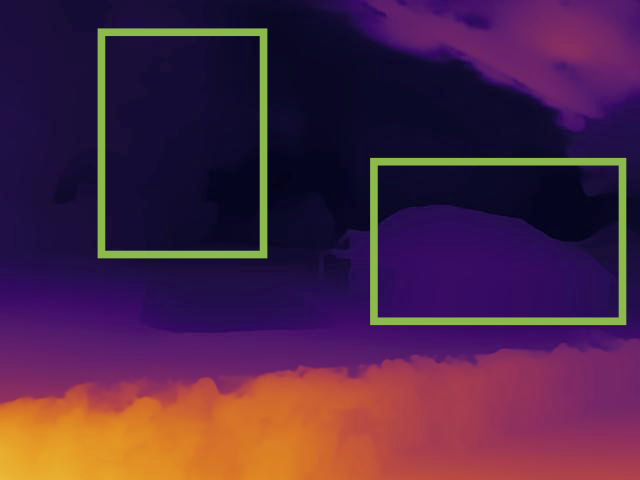}&
    \includegraphics[width=0.155\linewidth]{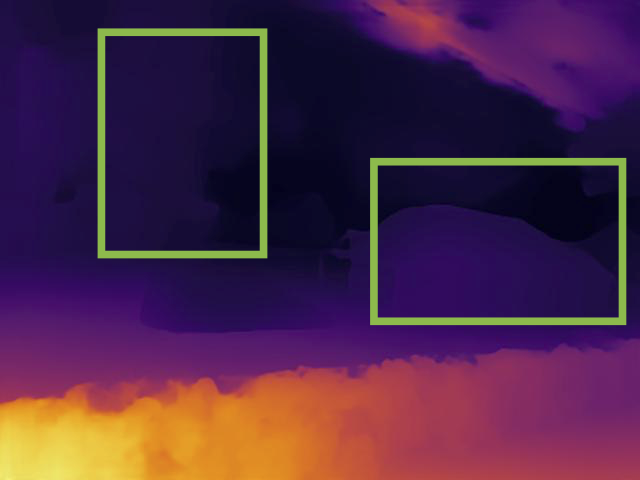}&
    \includegraphics[width=0.155\linewidth]{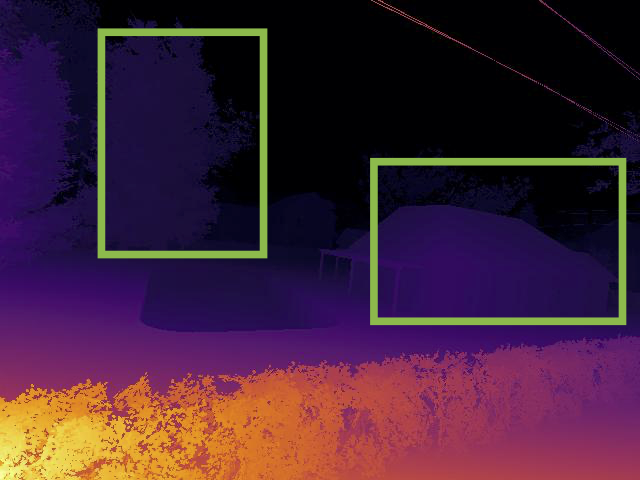}&
    \includegraphics[width=0.155\linewidth]{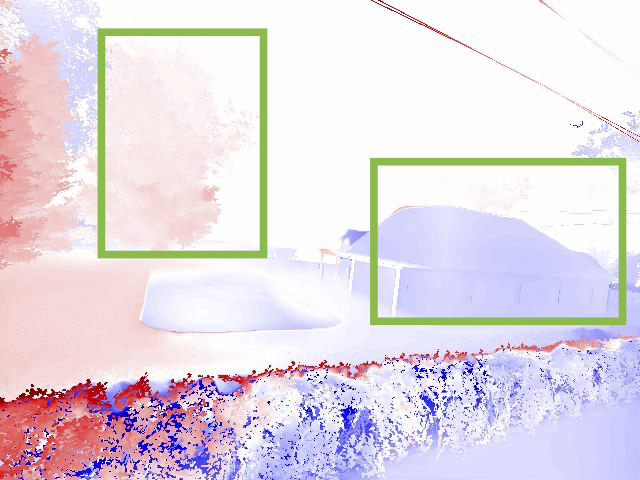}&
    \includegraphics[width=0.155\linewidth]{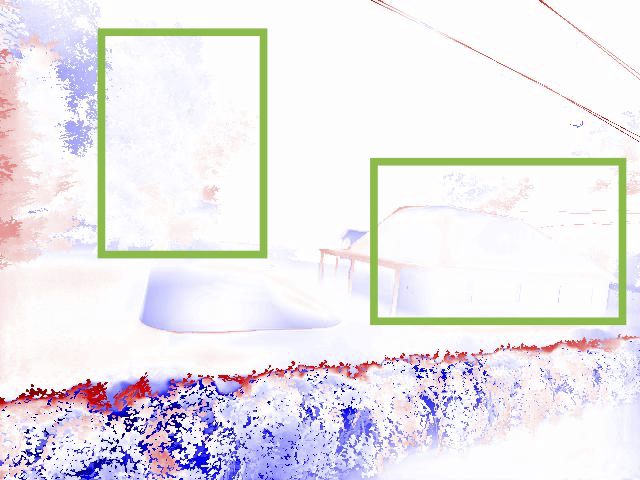}&
    \includegraphics[width=0.035\linewidth]{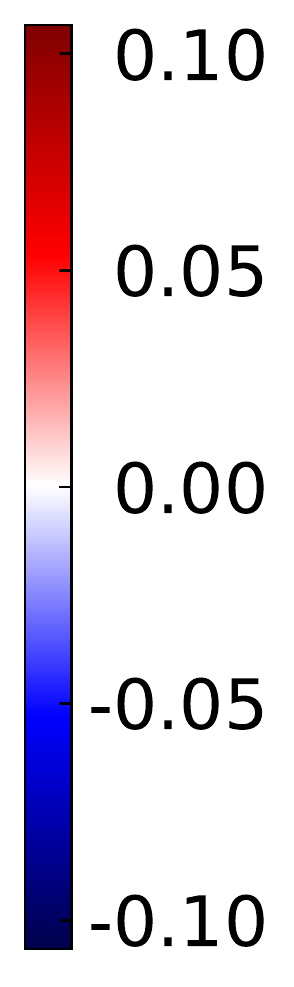}\\
    \vspace{-0.75mm}
    \includegraphics[width=0.155\linewidth]{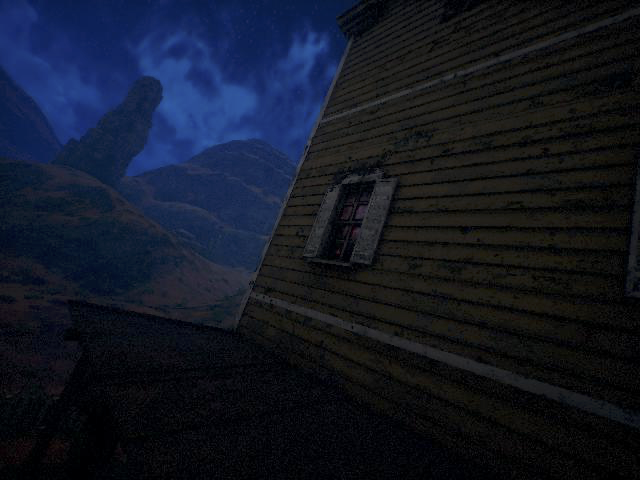}&
    \includegraphics[width=0.155\linewidth]{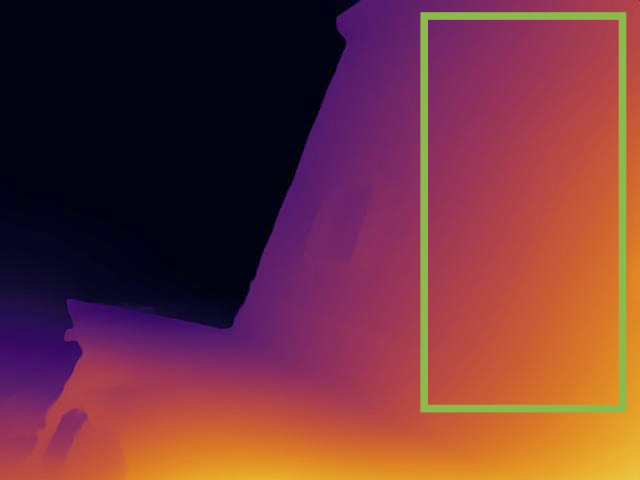}&
    \includegraphics[width=0.155\linewidth]{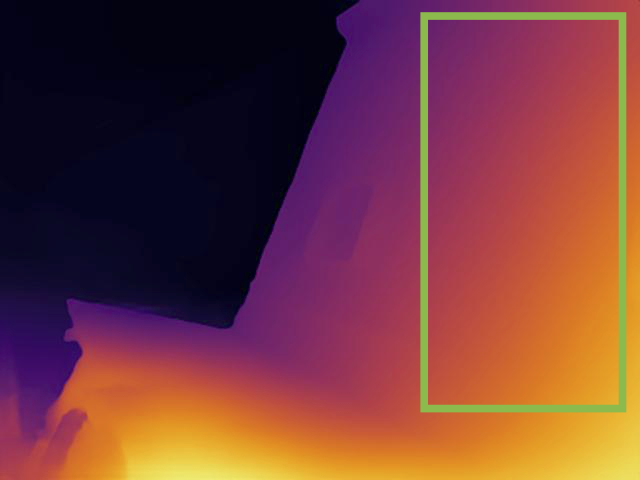}&
    \includegraphics[width=0.155\linewidth]{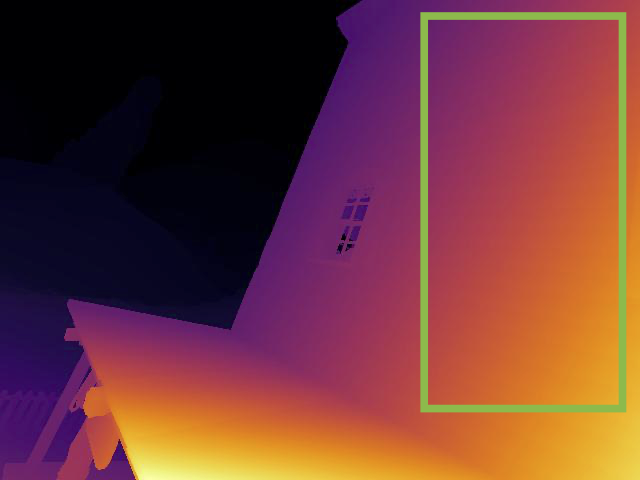}&
    \includegraphics[width=0.155\linewidth]{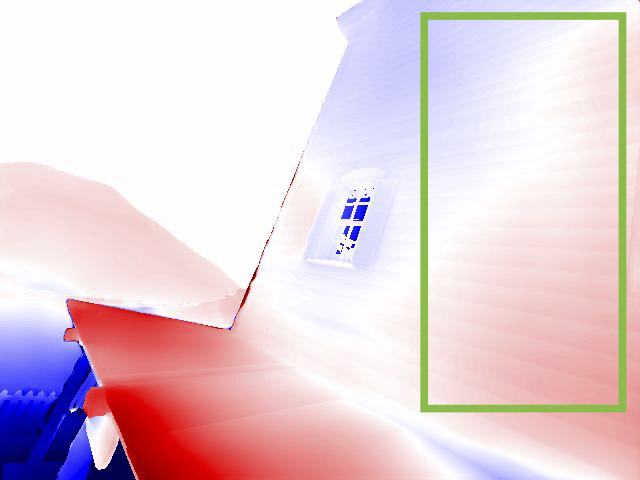}&
    \includegraphics[width=0.155\linewidth]{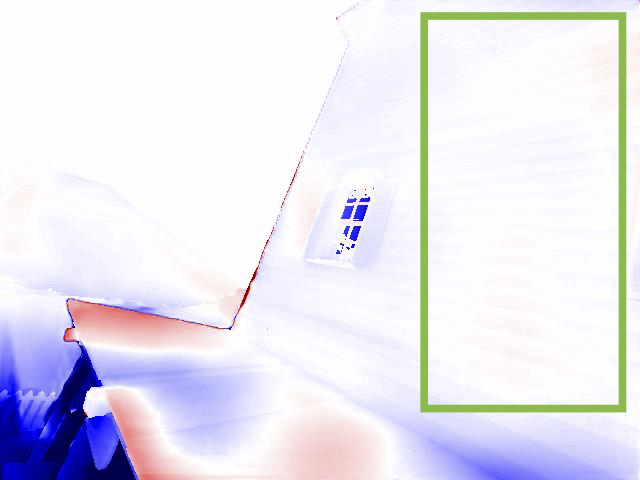}&
    \includegraphics[width=0.035\linewidth]{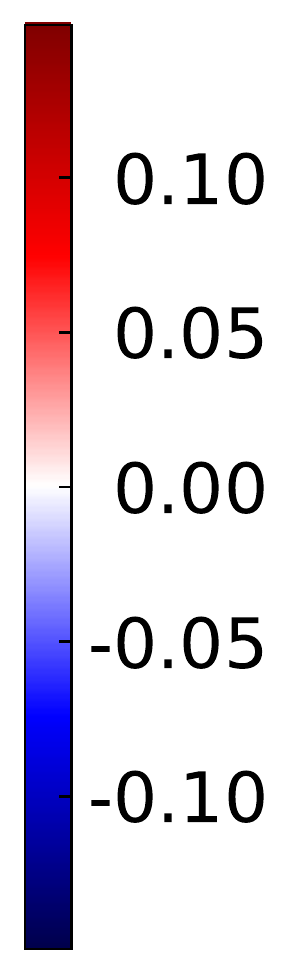}\\
    \vspace{-0.75mm}
    \includegraphics[width=0.155\linewidth]{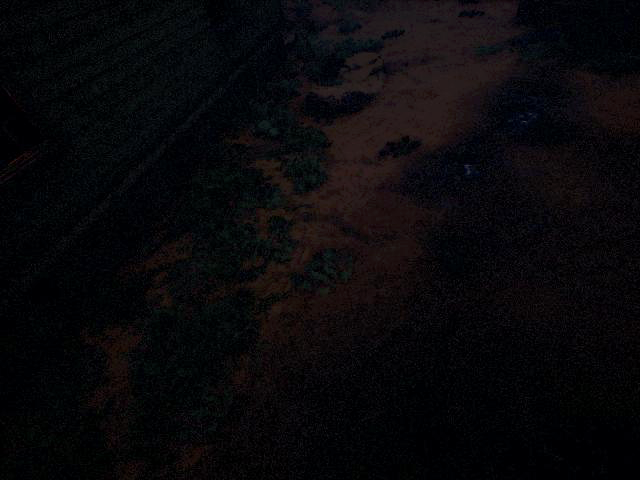}&
    \includegraphics[width=0.155\linewidth]{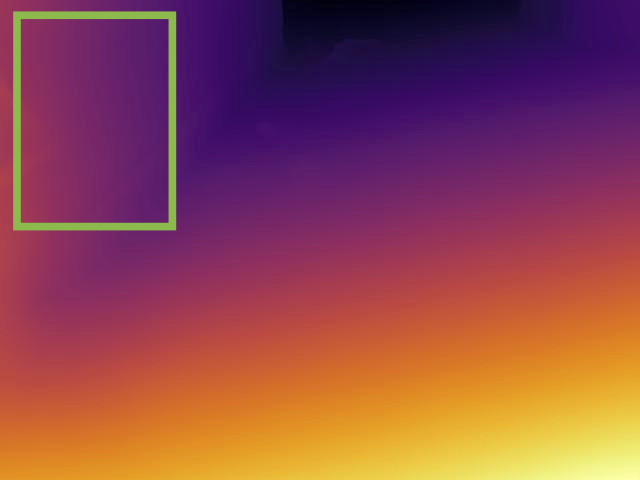}&
    \includegraphics[width=0.155\linewidth]{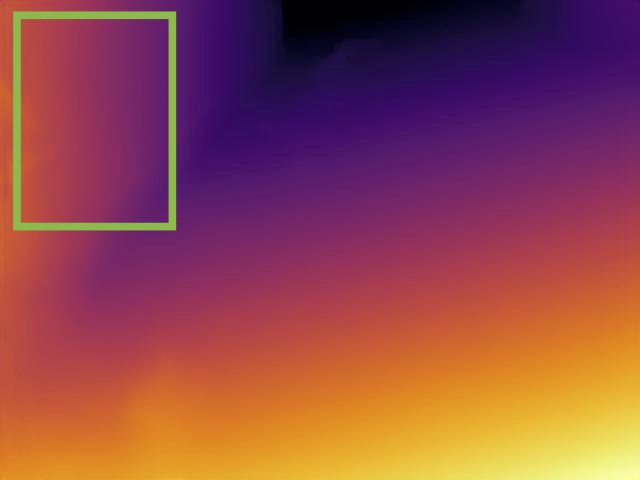}&
    \includegraphics[width=0.155\linewidth]{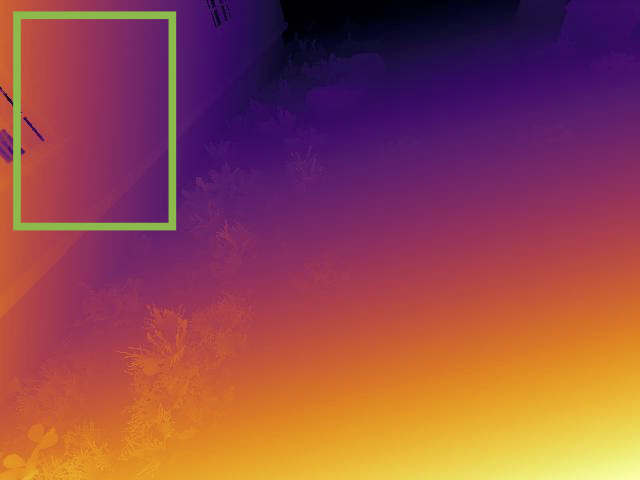}&
    \includegraphics[width=0.155\linewidth]{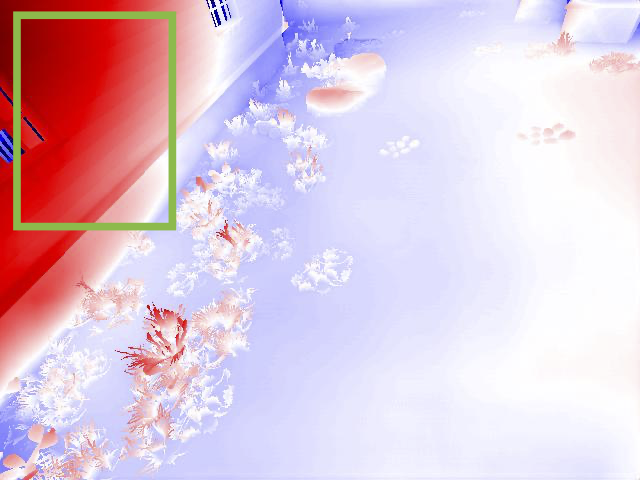}&
    \includegraphics[width=0.155\linewidth]{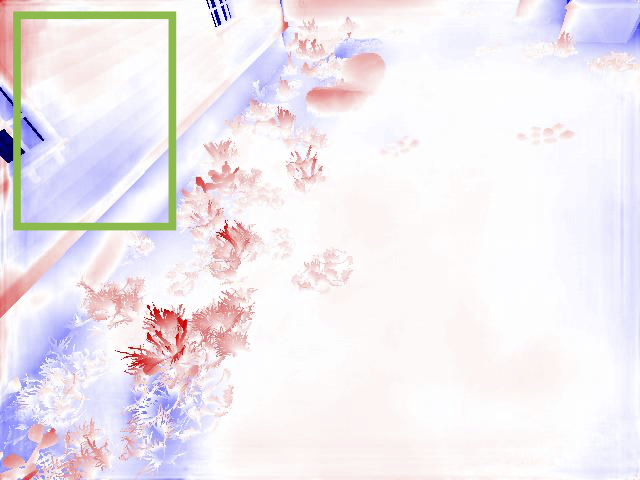}&
    \includegraphics[width=0.035\linewidth]{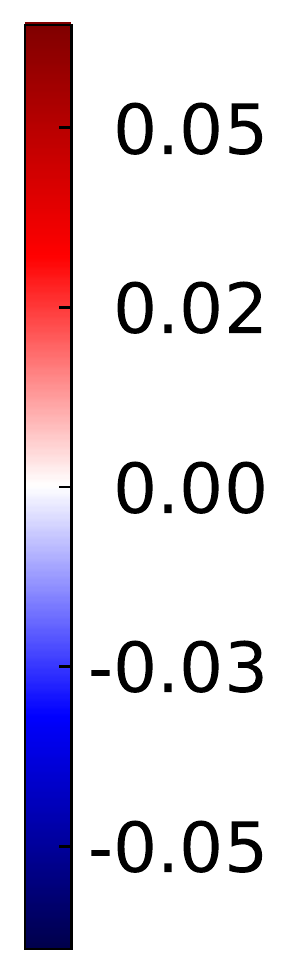}\\
  \end{tabular}
  \vspace{-8pt}
  \caption{Our method tested on TartanAir samples. In depth maps, brighter is closer and darker is farther. In error maps, red is positive inverse depth error (farther than ground truth GT) and blue is negative inverse depth error (closer than GT). Whiter regions in error indicate improved metric depth accuracy.} 
  \label{fig:vis_tartanair}
  \vspace{-12pt}
\end{figure}

Figure \ref{fig:vis_tartanair} provides a visualization of our approach on several TartanAir samples. Performance is qualitatively evaluated by comparing metric depth error for globally-aligned depth (GA error) versus densely-scaled depth (SML error). A whiter region in the error map indicates that SML improved metric depth accuracy there. The first sample depicts a neighborhood scene where the building towards center-right is pushed further back under dense scale alignment; this is confirmed by a reduction in negative (blue) error in inverse depth. The tree behind the pool is brought closer, as shown by the reduction in positive (red) error. The latter two samples depict significantly more challenging scenes due to low light as well as proximity to walls and the ground. In both, the SML still aligns surfaces towards the correct metric depth.

We note that DPT-Hybrid was trained on a large mixed dataset containing TartanAir. To remove any potential bias this contributes to SML evaluation on TartanAir, we swap in MiDaS v2.0~\cite{Ranftl2020} that has not seen any TartanAir data during training. Table \ref{tab:tartanair_eval} shows that MiDaS v2.0 still yields the same trends as DPT-Hybrid, with SML improving all metrics. 

\subsection{Evaluation on VOID}

We additionally evaluate on real-world data from the VOID dataset. We preprocess VOID data in the same manner as the TartanAir data, but using the sparse depth provided in the published dataset \cite{Wong2020void}. The first two rows of Table \ref{tab:void_eval} summarize our results when training SML directly on VOID. SML again improves over global alignment, with a 38\%, 30\%, and 39\% reduction in iMAE, iRMSE, and iAbsRel, respectively.

\begin{table}
  \footnotesize
  \setlength{\tabcolsep}{1pt} 
  \renewcommand{\arraystretch}{1} 
  \centering
  \caption{Evaluation on VOID. All methods use DPT-H as the depth model and 150 sparse depth points. Lower is better.}
  \label{tab:void_eval}
  \vspace{-4pt}
  \begin{tabular}{@{}l@{\hspace{2mm}}l@{\hspace{2mm}}*{2}{S[table-format=3.2]@{\hspace{1.5mm}}}*{1}{S[table-format=2.2]@{\hspace{1mm}}}*{1}{S[table-format=3.2]@{\hspace{1mm}}}*{1}{S[table-format=1.3]@{\hspace{0mm}}}@{}} 
    \toprule
    Method  & Training Set                  & {MAE}             & {RMSE}            & {iMAE}           & {iRMSE}          & {iAbsRel} \\
    \midrule
    GA only &                               &           165.33  &           243.11  &           75.74  &          106.37  & 0.103 \\
    GA+SML  & VOID                     &   \myuline{97.03} &  \myuline{167.82} &           46.62  &           74.67  & 0.063 \\
    GA+SML  & TA (zero-shot)    &            98.49  &           175.04  &  \myuline{45.55} &  \myuline{74.28} & \myuline{0.062} \\
    GA+SML  & TA + VOID     &  \bfseries 82.65  & \bfseries 153.51  & \bfseries 38.56  & \bfseries 66.23  & \bfseries 0.051 \\
    \bottomrule
  \end{tabular}
  \vspace{-12pt}
\end{table}

\mypara{TartanAir-to-VOID transfer.} We investigate the performance of SML when trained on TartanAir and evaluated on VOID without any finetuning (i.e., zero-shot cross-dataset transfer). This can be interpreted as a sim-to-real transfer experiment, since TartanAir consists solely of synthetic data and VOID contains real-world data samples. We observe that zero-shot testing on VOID achieves very similar error as when training directly on VOID. If evaluating in inverse depth space, zero-shot transfer even slightly outperforms direct training on VOID. This is particularly notable since it demonstrates that training on a large quantity of diverse synthetic data can indeed translate to improved real-world performance. It also shows the generalizability of our pipeline. DPT-Hybrid is already known to generalize well after having been trained on a massive mixed dataset with scale- and shift-invariant loss functions. The SML network is trained using metric loss terms; however, some metric information is provided to SML via the globally-aligned depth and scale map scaffolding inputs. Since SML only needs to learn to refine this scaffolding, it is less likely to memorize or overfit to a specific metric scale.

\begin{figure}
\centering
  \begin{tabular}{@{}*{6}{c@{\hspace{0.5mm}}}c@{}}
    {\scriptsize RGB Image} & {\scriptsize GA Depth} & {\scriptsize SML Depth} & {\scriptsize GT Depth} & {\scriptsize GA Error} & {\scriptsize SML Error}\\
    \vspace{-0.75mm}
    \includegraphics[width=0.155\linewidth]{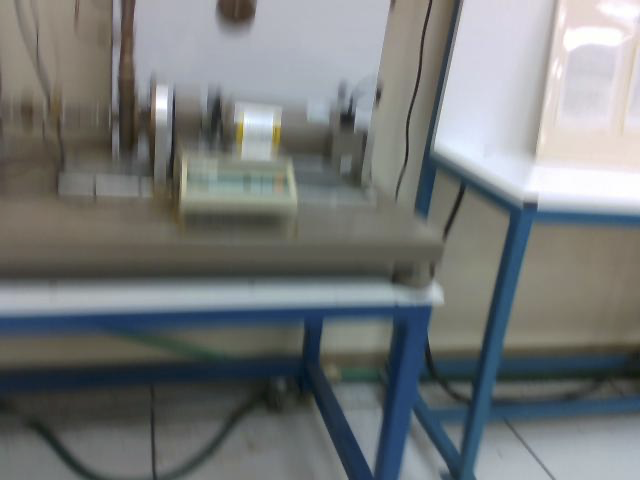}&
    \includegraphics[width=0.155\linewidth]{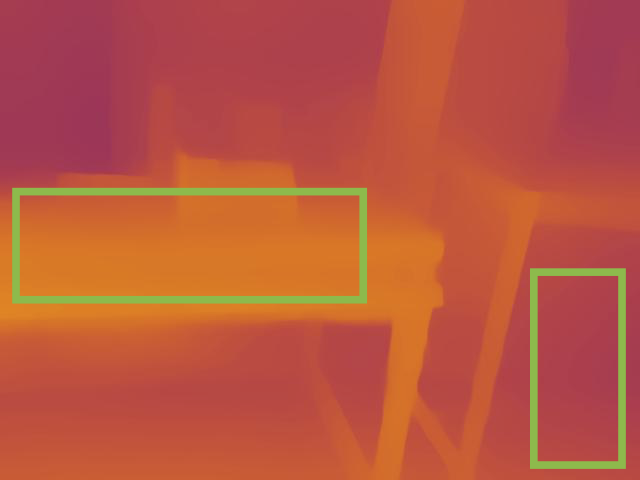}&
    \includegraphics[width=0.155\linewidth]{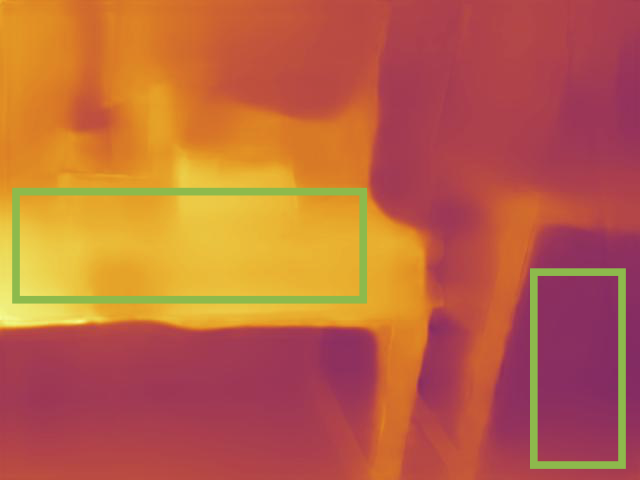}&
    \includegraphics[width=0.155\linewidth]{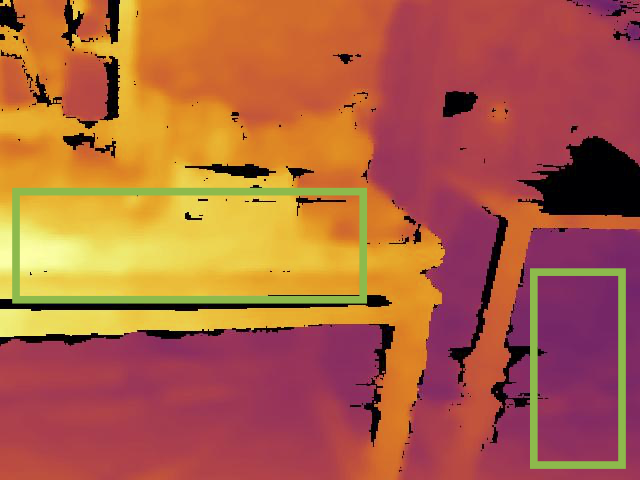}&
    \includegraphics[width=0.155\linewidth]{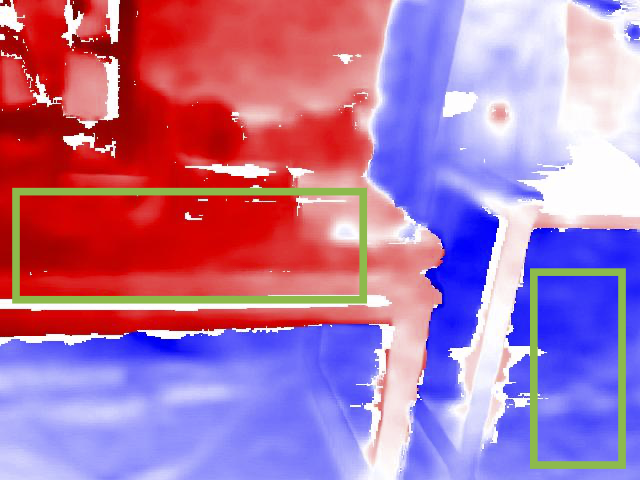}&
    \includegraphics[width=0.155\linewidth]{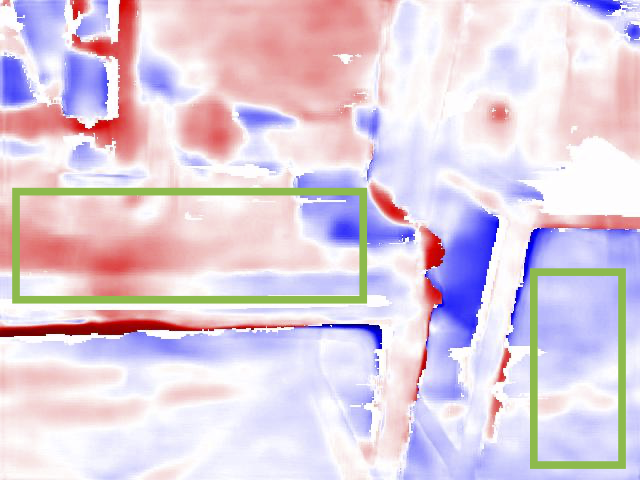}&
    \includegraphics[width=0.035\linewidth]{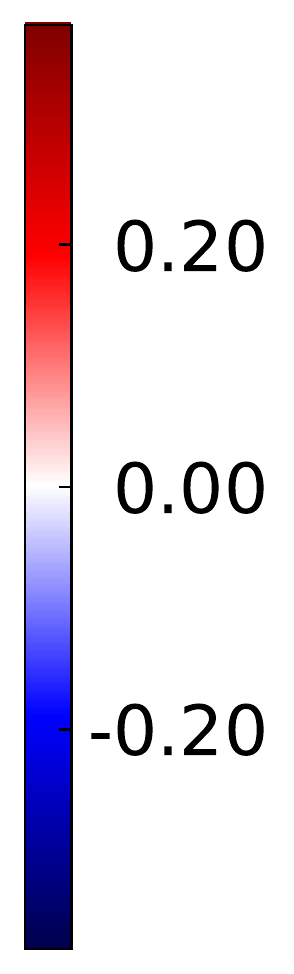}\\
    \vspace{-0.75mm}
    \includegraphics[width=0.155\linewidth]{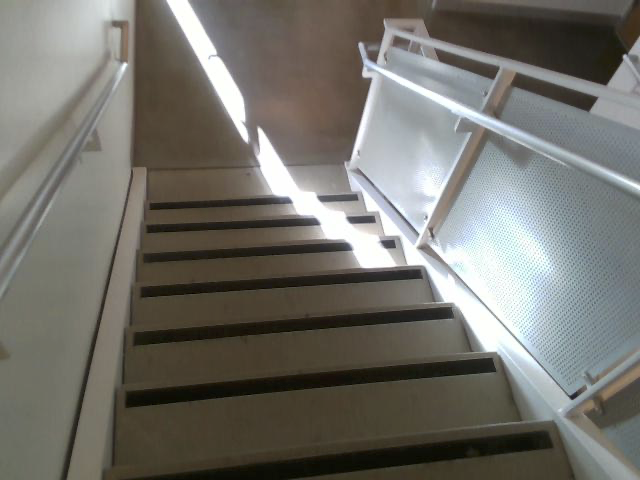}&
    \includegraphics[width=0.155\linewidth]{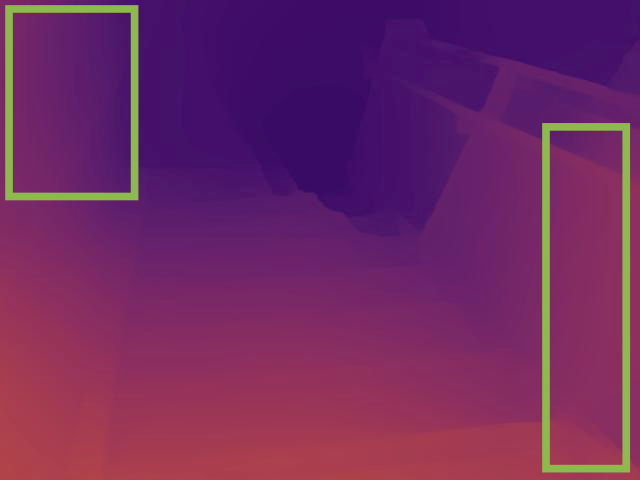}&
    \includegraphics[width=0.155\linewidth]{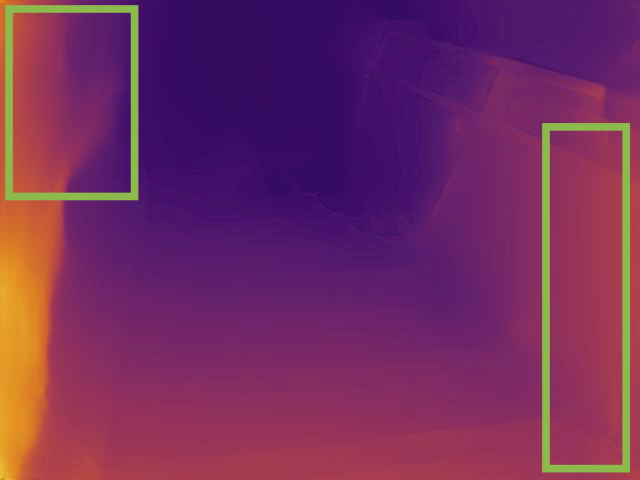}&
    \includegraphics[width=0.155\linewidth]{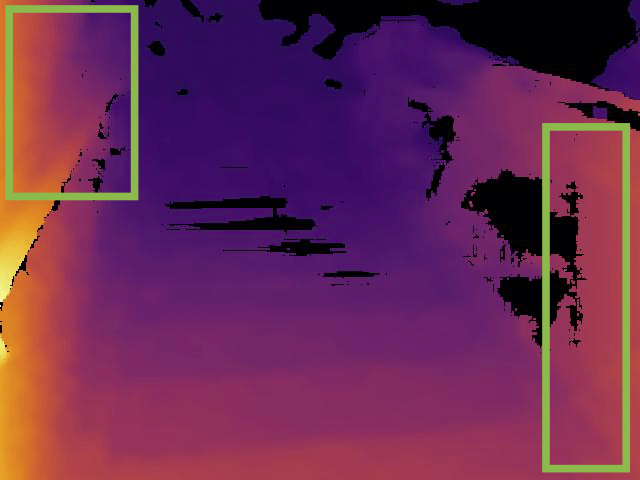}&
    \includegraphics[width=0.155\linewidth]{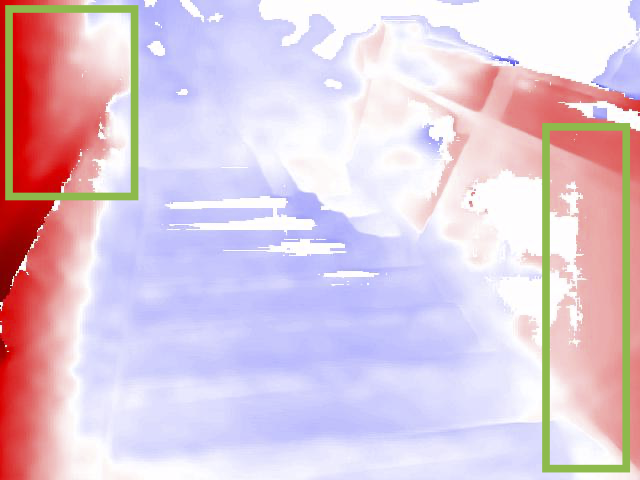}&
    \includegraphics[width=0.155\linewidth]{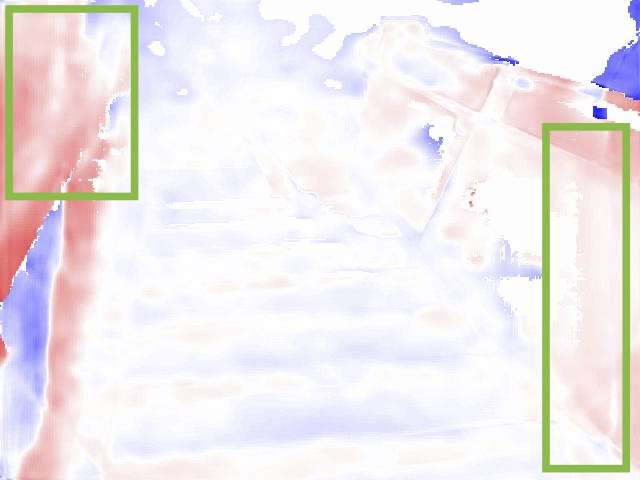}&
    \includegraphics[width=0.035\linewidth]{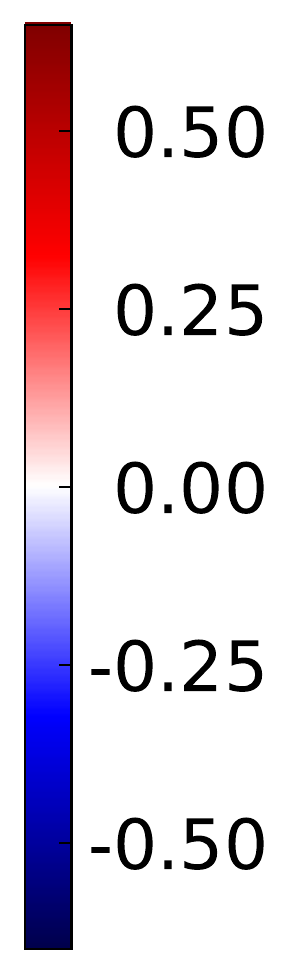}\\
    \vspace{-0.75mm}
    \includegraphics[width=0.155\linewidth]{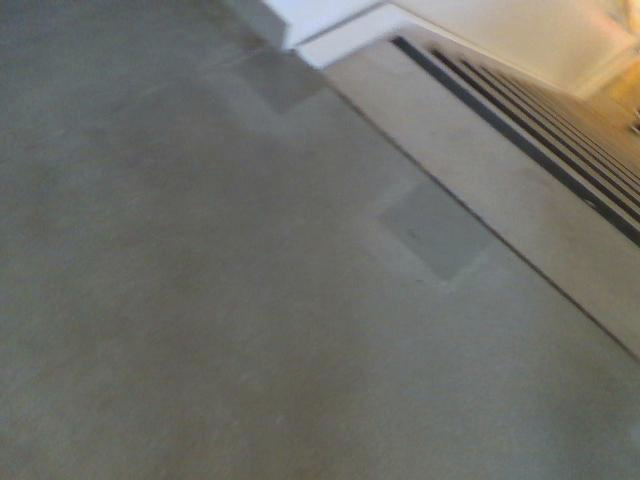}&
    \includegraphics[width=0.155\linewidth]{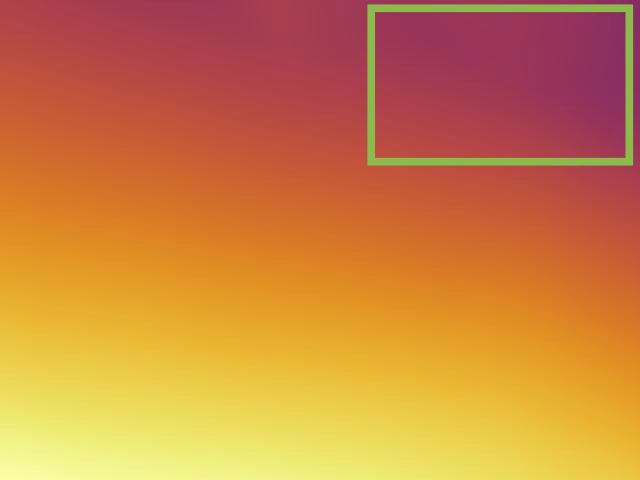}&
    \includegraphics[width=0.155\linewidth]{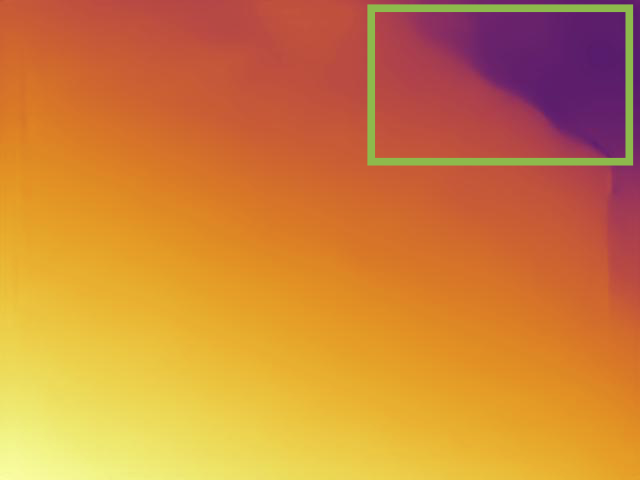}&
    \includegraphics[width=0.155\linewidth]{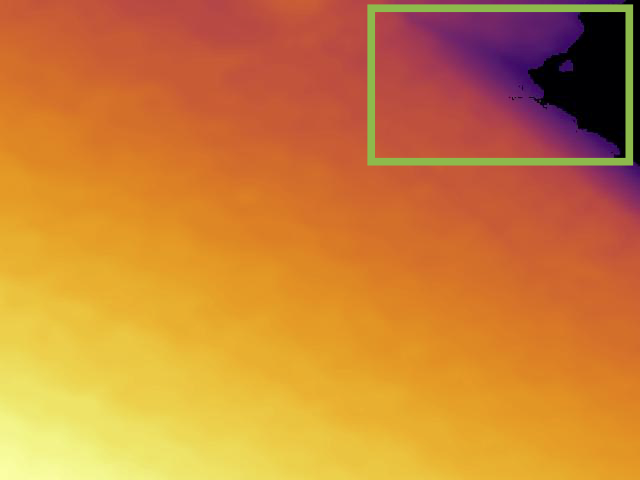}&
    \includegraphics[width=0.155\linewidth]{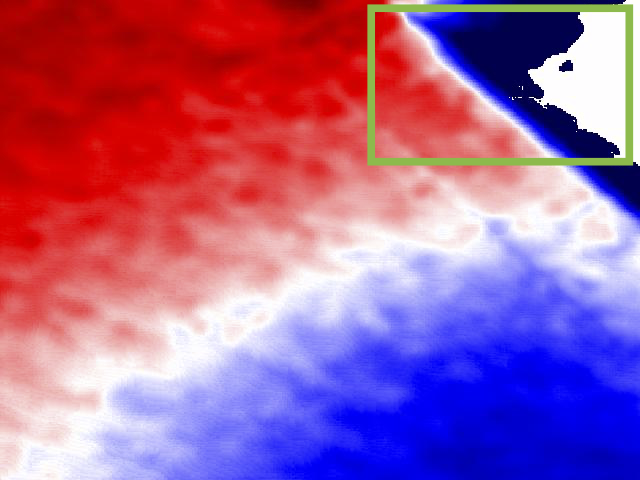}&
    \includegraphics[width=0.155\linewidth]{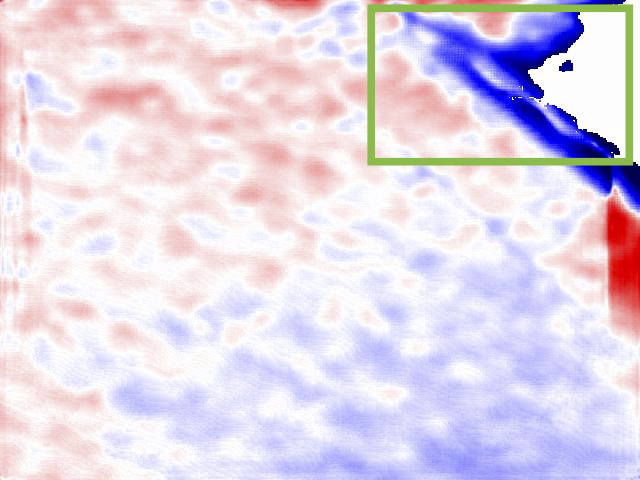}&
    \includegraphics[width=0.035\linewidth]{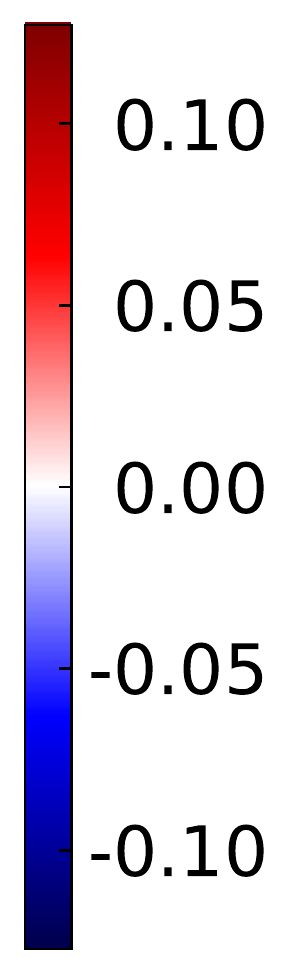}\\
  \end{tabular}
  \vspace{-8pt}
  \caption{Our method tested on VOID samples. SML is pretrained on TartanAir and fine-tuned on VOID. Color coding is the same as in Figure \ref{fig:vis_tartanair}.}
  \label{fig:vis_void}
  \vspace{-8pt}
\end{figure}

\mypara{Pretraining.} Pretraining on TartanAir and fine-tuning on VOID yields the lowest error across all metrics. We use this combination to produce the results visualized for samples in Figure \ref{fig:vis_void}. The first sample suffers from blurriness in the RGB input and depicts a cluttered scene. With global alignment only, depth predictions appear flattened: the table is aligned to be farther than ground truth (red error), while background surfaces such as walls and the floor are aligned to be closer than ground truth (blue error). Dense scale alignment with SML helps to rectify this, with noticeable reduction (whiter regions) throughout the error map. The second sample shows a staircase; in addition to reducing depth error on the steps, SML is able to correctly realign the handrail on the left. This is impressive as pixels near the image boundary fall outside the convex hull of known sparse depth points, and the scale map scaffolding that we input to SML signals no information at pixels outside the convex hull. The last sample shows a challenging viewpoint of the floor leading to a staircase in the top right corner. Global alignment alone misjudges the depth gradient at the staircase edge. SML corrects this and also reduces depth error elsewhere on the floor surface.

\begin{table}
  \centering
  \footnotesize
  \caption{Comparison on VOID. Lower is better for all metrics.}
  \label{tab:void_comparison}
  \vspace{-4pt}
  \begin{tabular}{@{}
  l@{\hspace{1mm}}|
  l@{\hspace{2mm}}
  S[table-format=3.2]@{\hspace{2mm}} 
  S[table-format=3.2]@{\hspace{2mm}} 
  S[table-format=2.2]@{\hspace{2mm}} 
  S[table-format=3.2]@{\hspace{0mm}} @{}}
    \toprule
    & Method               & {MAE} & {RMSE}  & {iMAE} & {iRMSE} \\
    \midrule
    \multirow{5}{*}{\rot{150 points}} & VOICED-S \cite{Wong2020void}    & 174.04 & 253.14 & 87.39 & 126.30 \\
    & KBNet \cite{Wong2021kbnet}      & 131.54 & 263.54 & 66.84 & 128.29 \\
    & GA+SML (DPT-BEiT-Large)      & \bfseries 76.95 & \bfseries 142.85 & \bfseries 34.25 & \bfseries 57.13 \\
    & GA+SML (DPT-Hybrid)      & \myuline{97.03} & \myuline{167.82} & \myuline{46.62} & \myuline{74.67} \\
    & GA+SML (MiDaS-small)    & 113.27 & 193.38 & 53.86 &  84.82 \\
    \midrule
    \multirow{5}{*}{\rot{500 points}} & VOICED-S \cite{Wong2020void}    & 118.01 & 195.32 & 59.29 & 101.72 \\
    & KBNet \cite{Wong2021kbnet}  & \myuline{77.70} & 172.49 & 38.87 &  85.59 \\
    & GA+SML (DPT-BEiT-Large)      & \bfseries 66.14 & \bfseries 126.44 & \bfseries 28.92 & \bfseries 49.85 \\
    & GA+SML (DPT-Hybrid)       &  81.30 & \myuline{146.16} & \myuline{37.35} &  \myuline{60.92} \\
    & GA+SML (MiDaS-small)    &  94.81 & 164.36 & 43.19 &  69.25 \\
    \bottomrule
  \end{tabular}
  \vspace{-12pt}
\end{table}

\begin{figure}[h]
\centering
  \begin{tabular}{@{}*{6}{c@{\hspace{0.5mm}}}@{}}
    \vspace{-1mm}
    &  &  & {\scriptsize GA+SML} & {\scriptsize GA+SML} & {\scriptsize GA+SML}\\
    {\scriptsize RGB Image} & {\scriptsize GT Depth} & {\scriptsize KBNet~\cite{Wong2021kbnet}} & {\scriptsize DPT-BEiT-L} & {\scriptsize DPT-H} & {\scriptsize MiDaS-s}\\
    \vspace{-0.75mm}
    \includegraphics[width=0.155\linewidth]{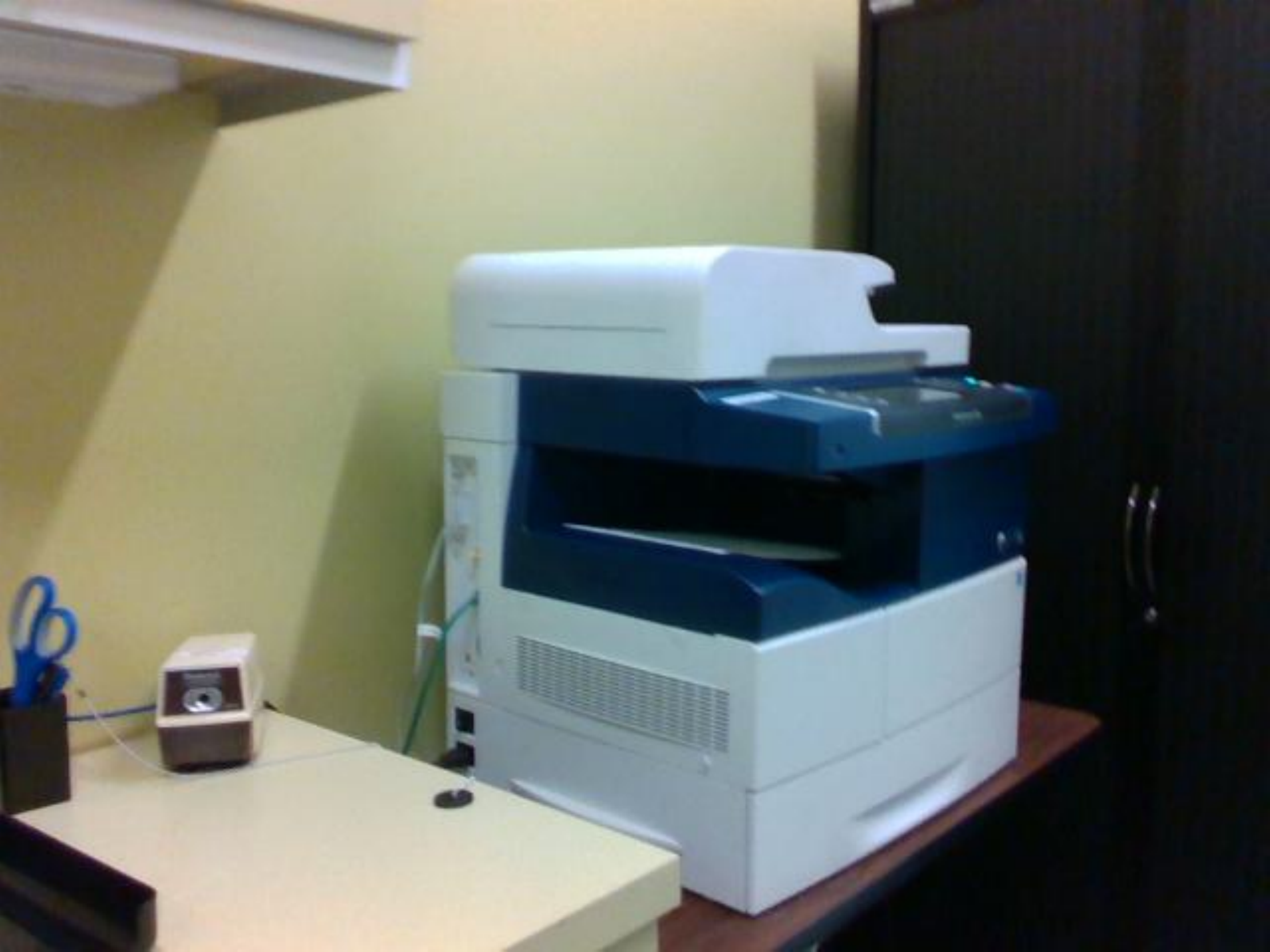}&
    \includegraphics[width=0.155\linewidth]{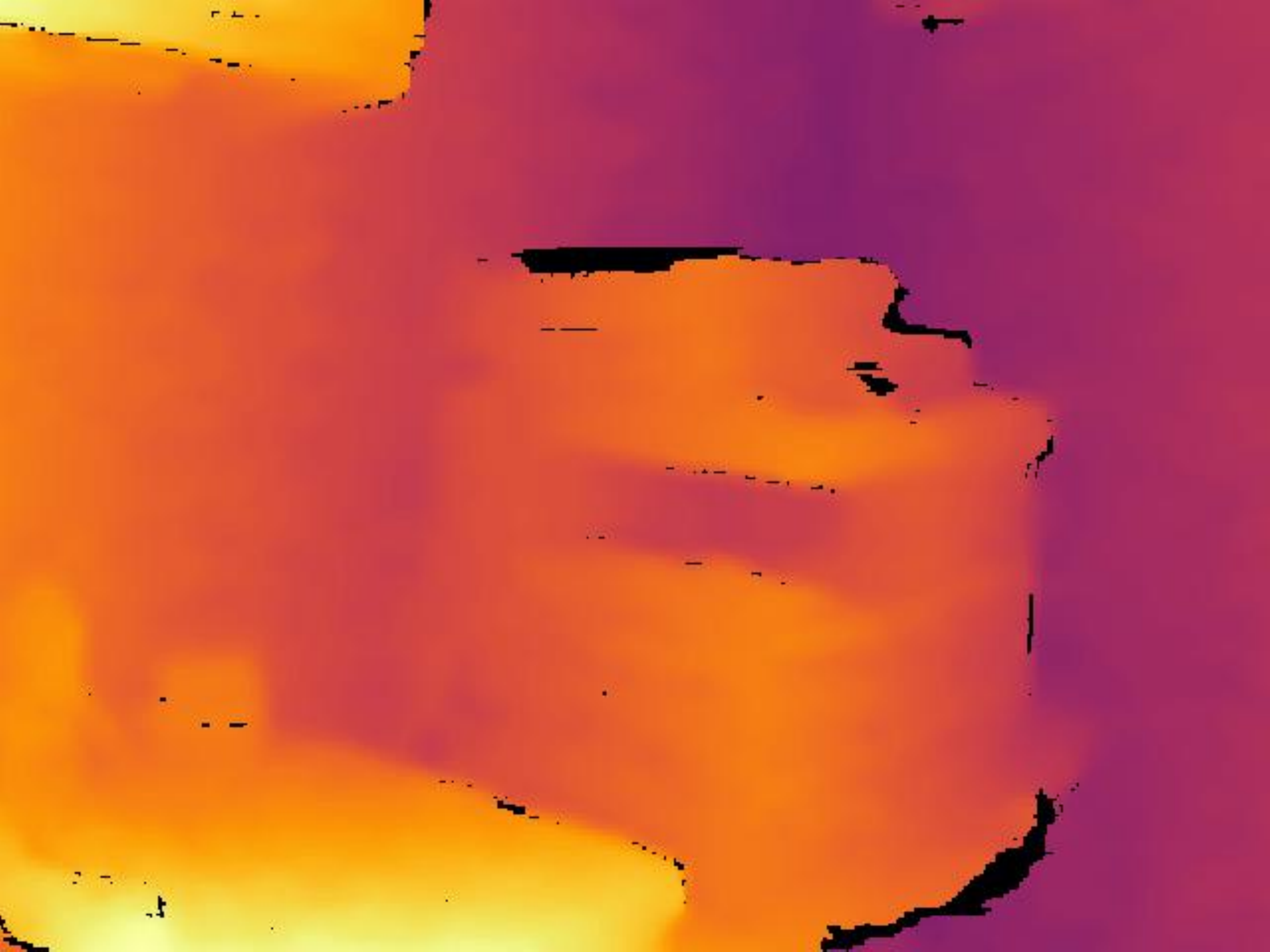}&
    \includegraphics[width=0.155\linewidth]{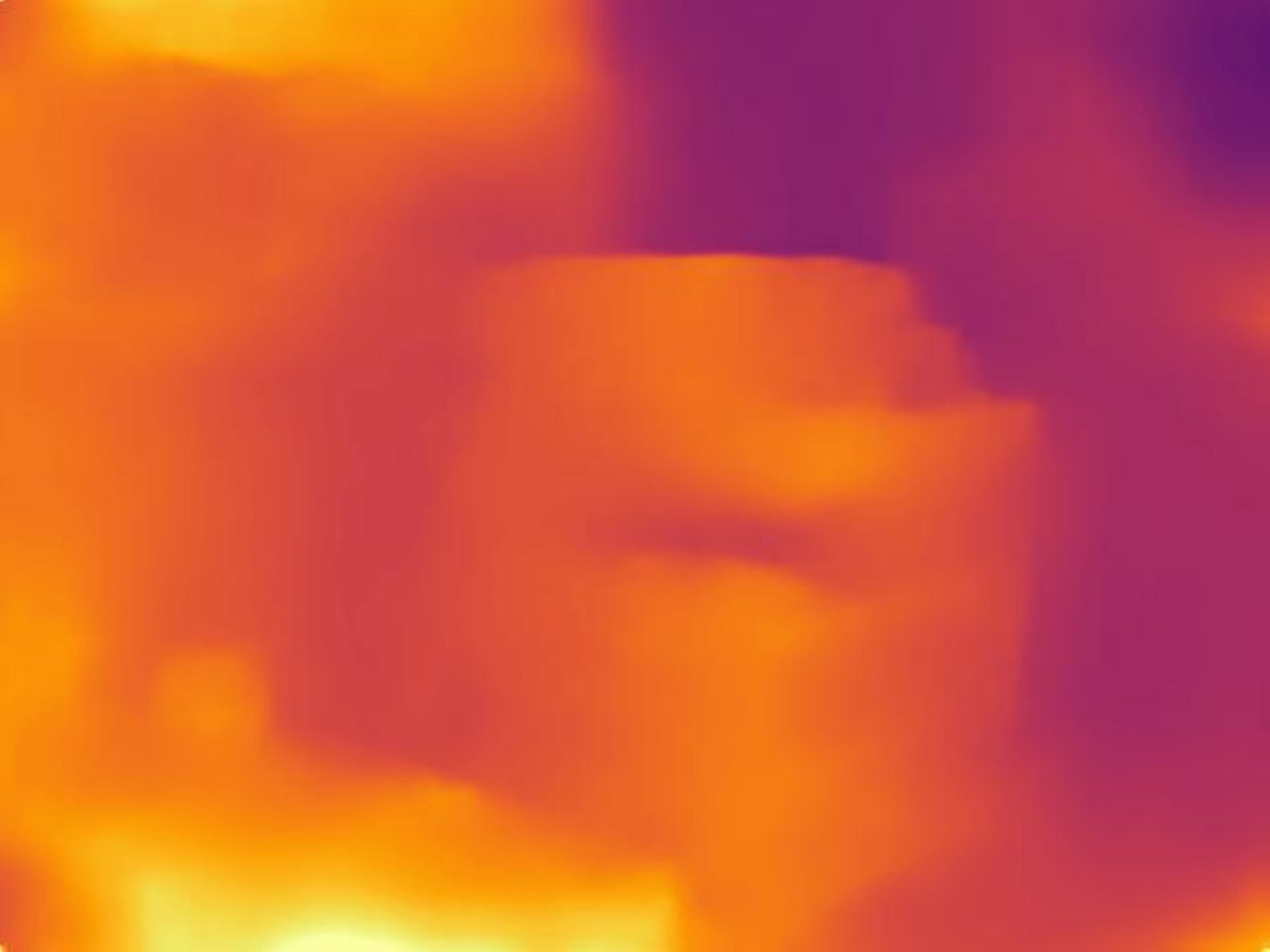}&
    \includegraphics[width=0.155\linewidth]{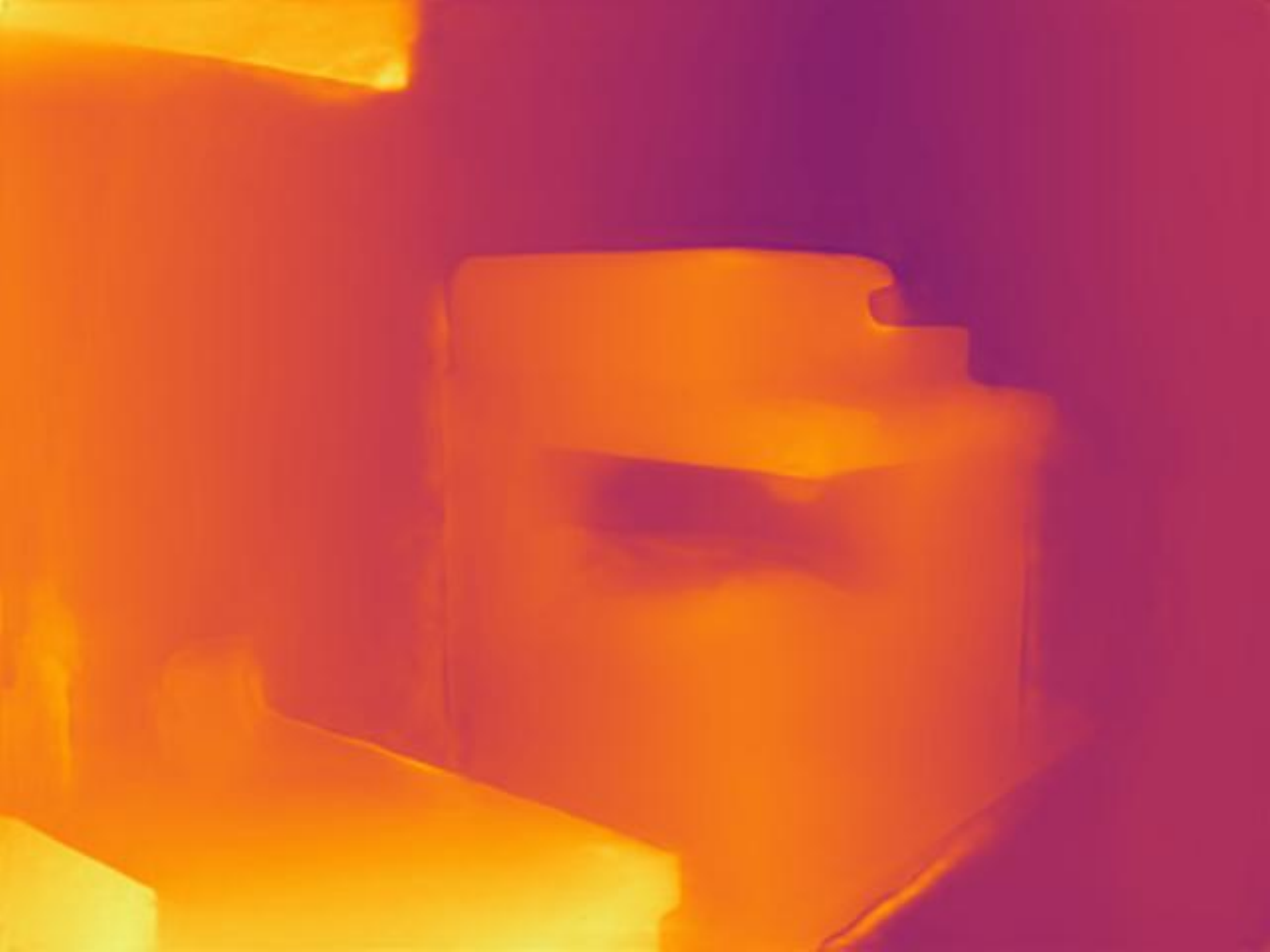}&
    \includegraphics[width=0.155\linewidth]{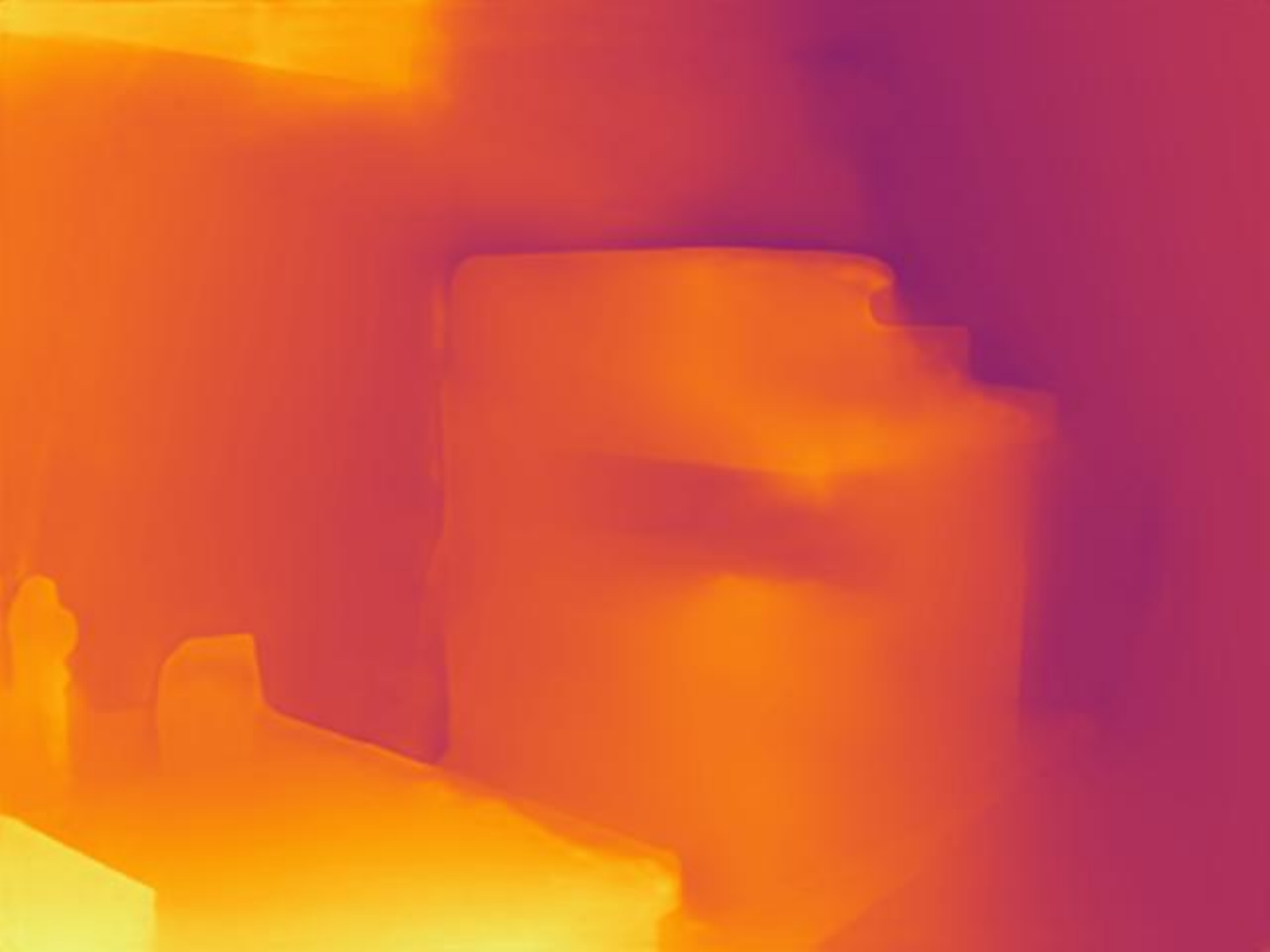}&
    \includegraphics[width=0.155\linewidth]{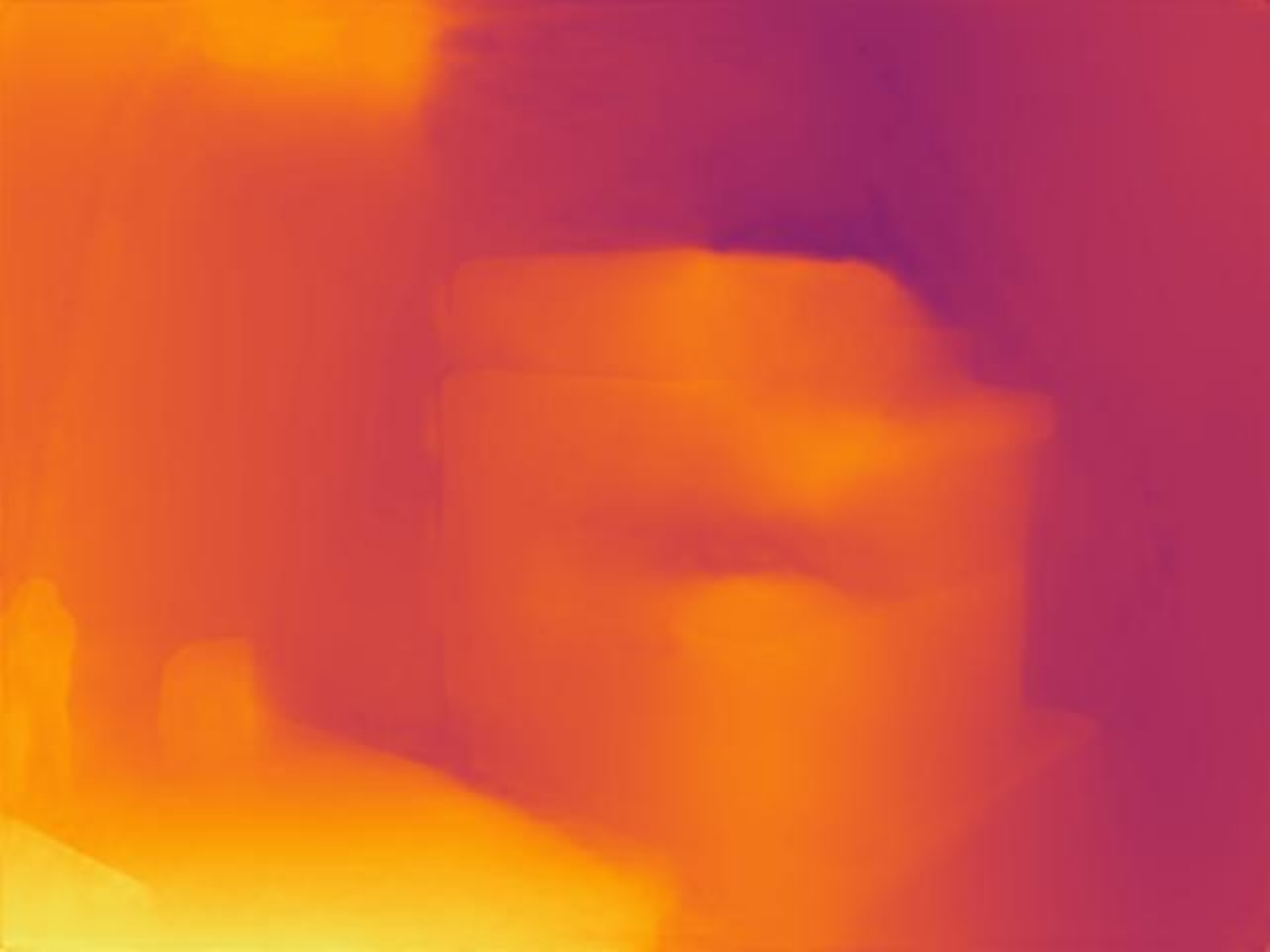}\\
    \vspace{-0.75mm}
    \includegraphics[width=0.155\linewidth]{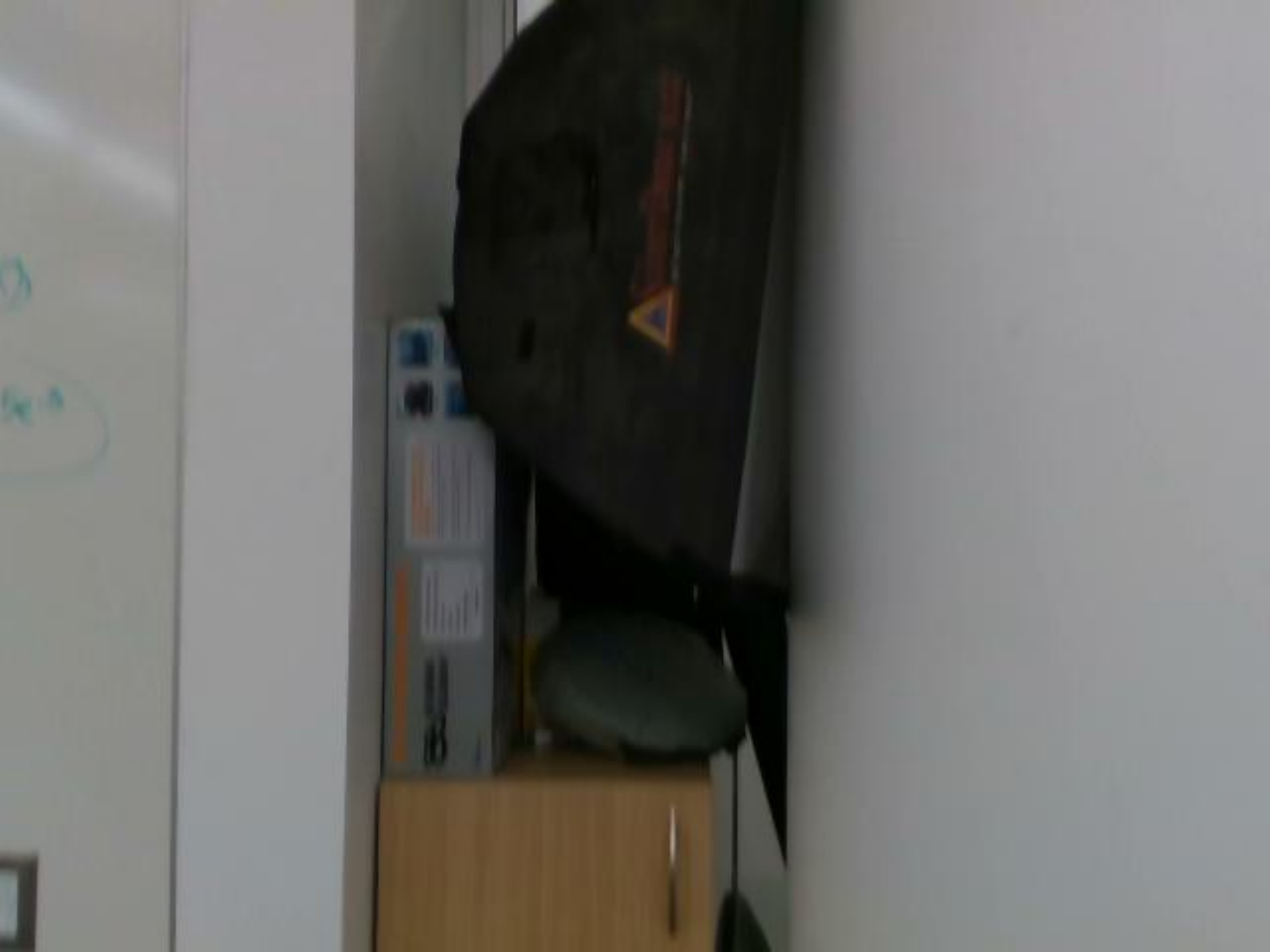}&
    \includegraphics[width=0.155\linewidth]{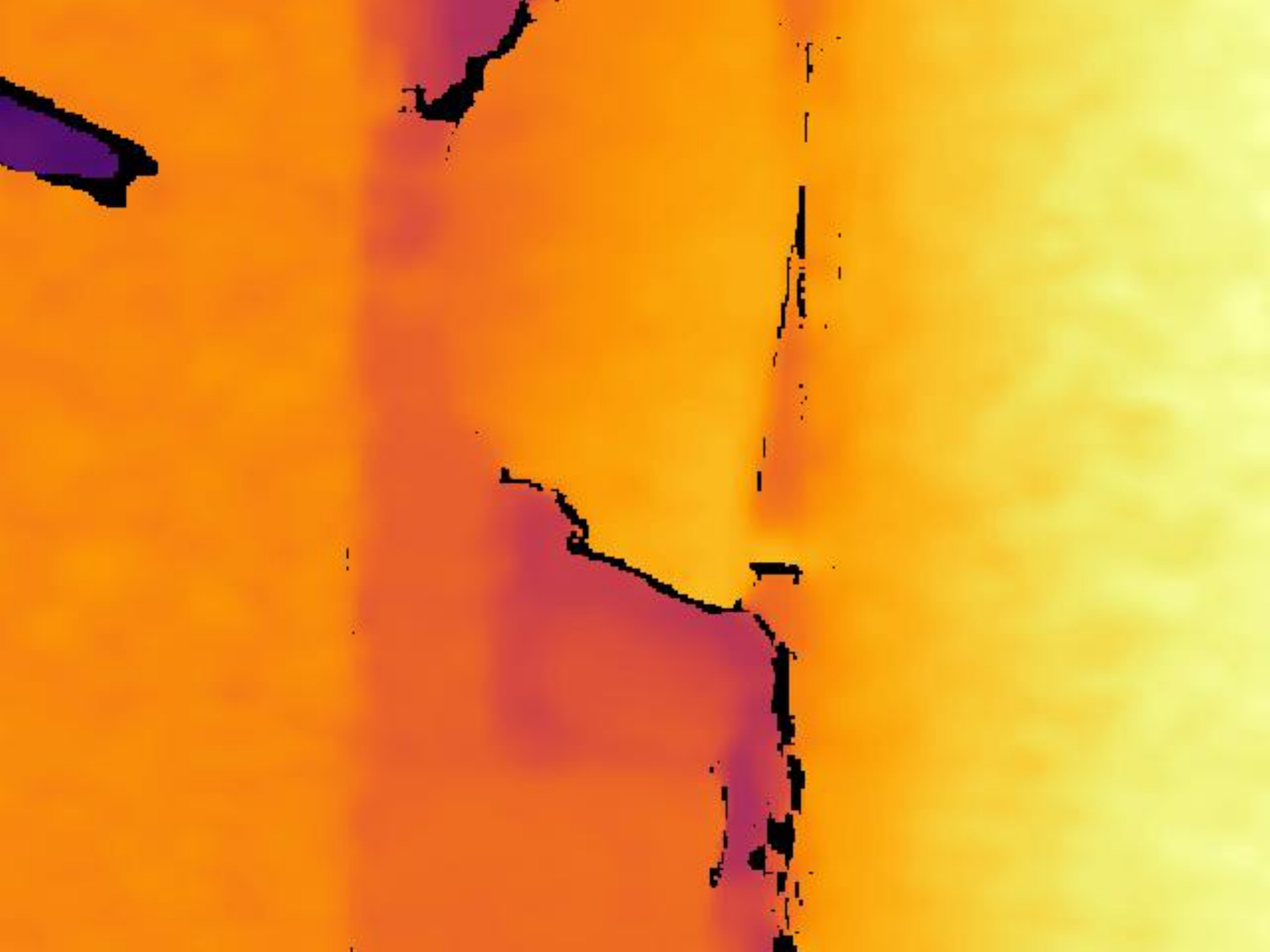}&
    \includegraphics[width=0.155\linewidth]{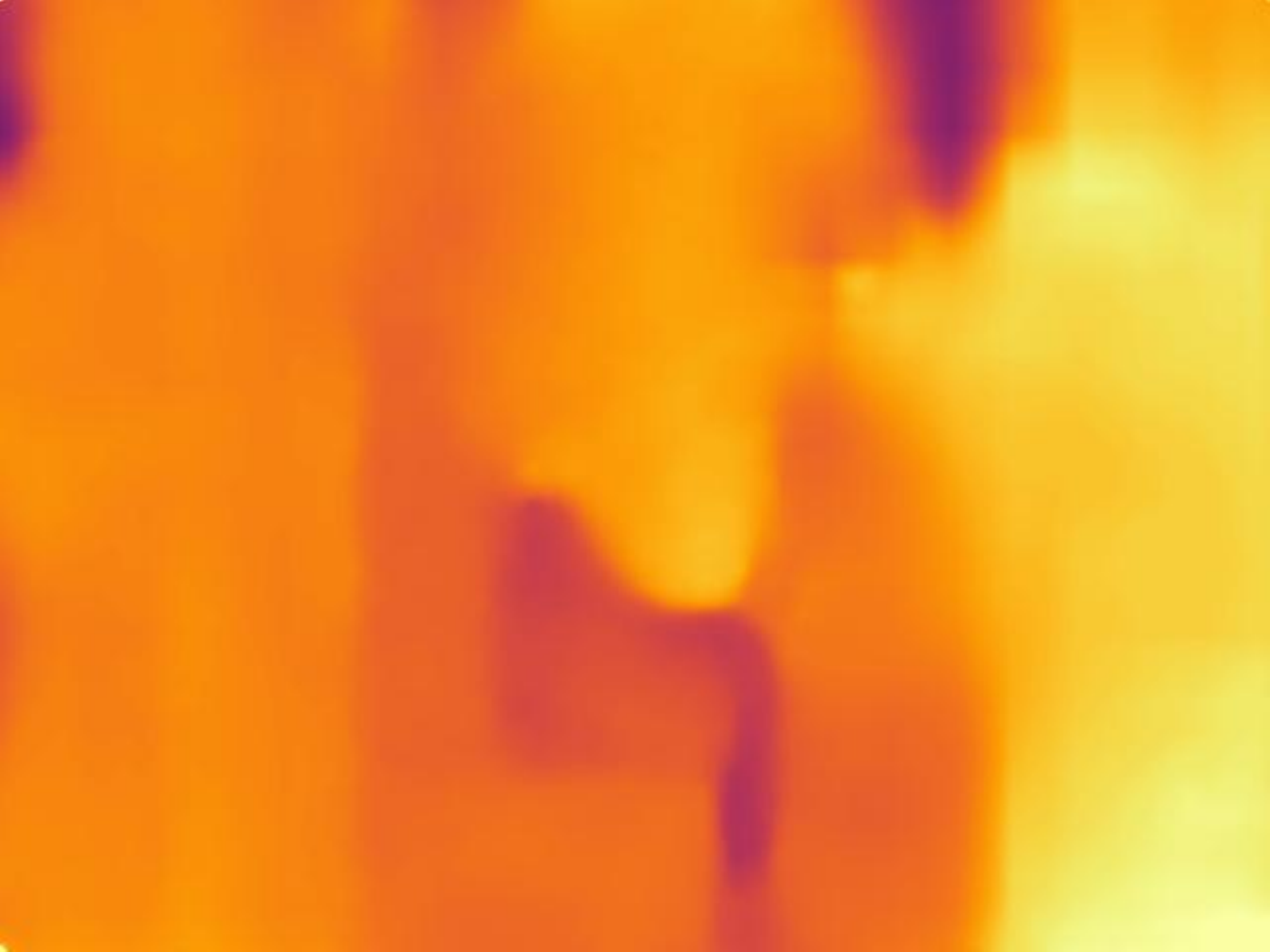}&
    \includegraphics[width=0.155\linewidth]{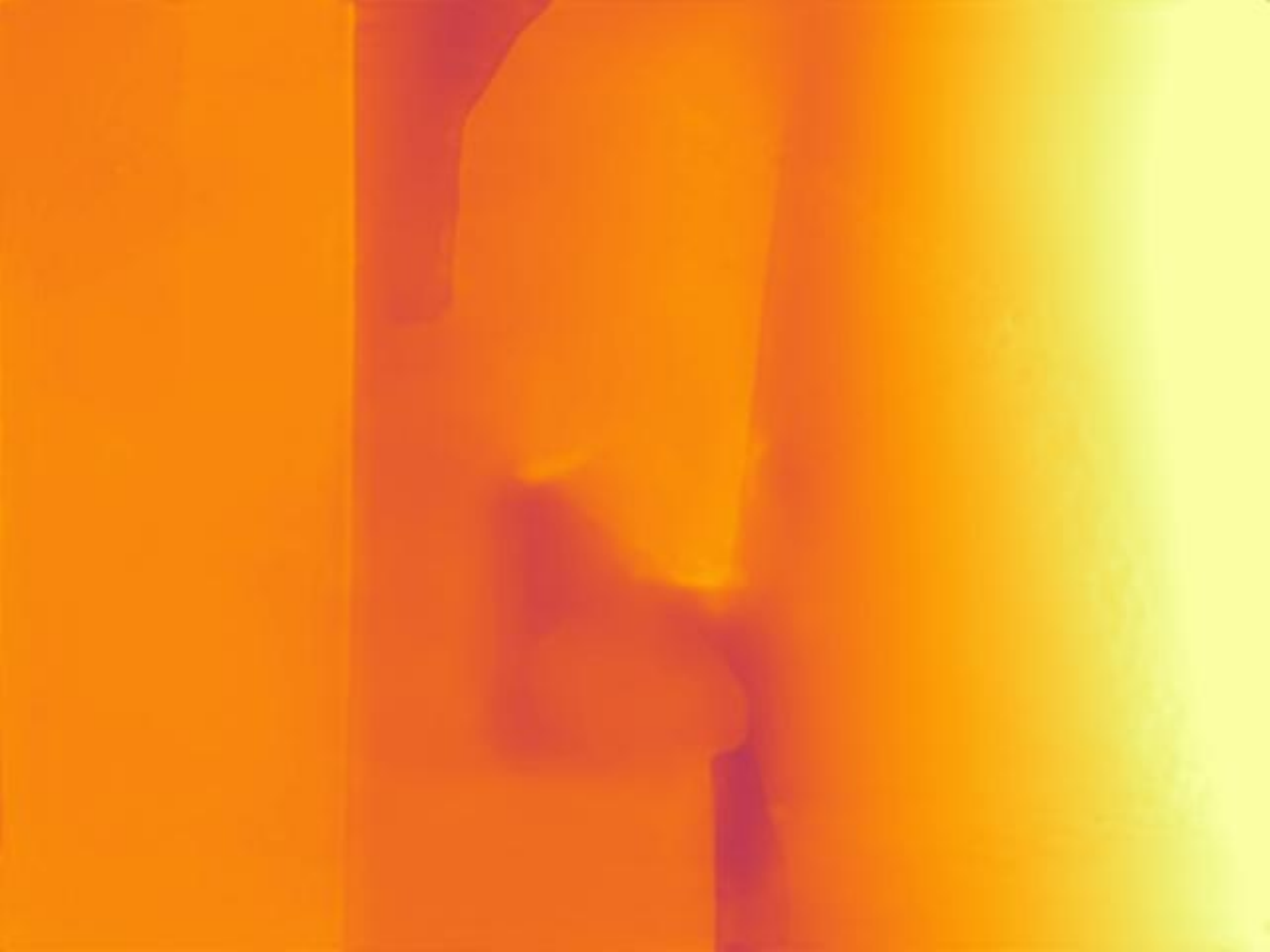}&
    \includegraphics[width=0.155\linewidth]{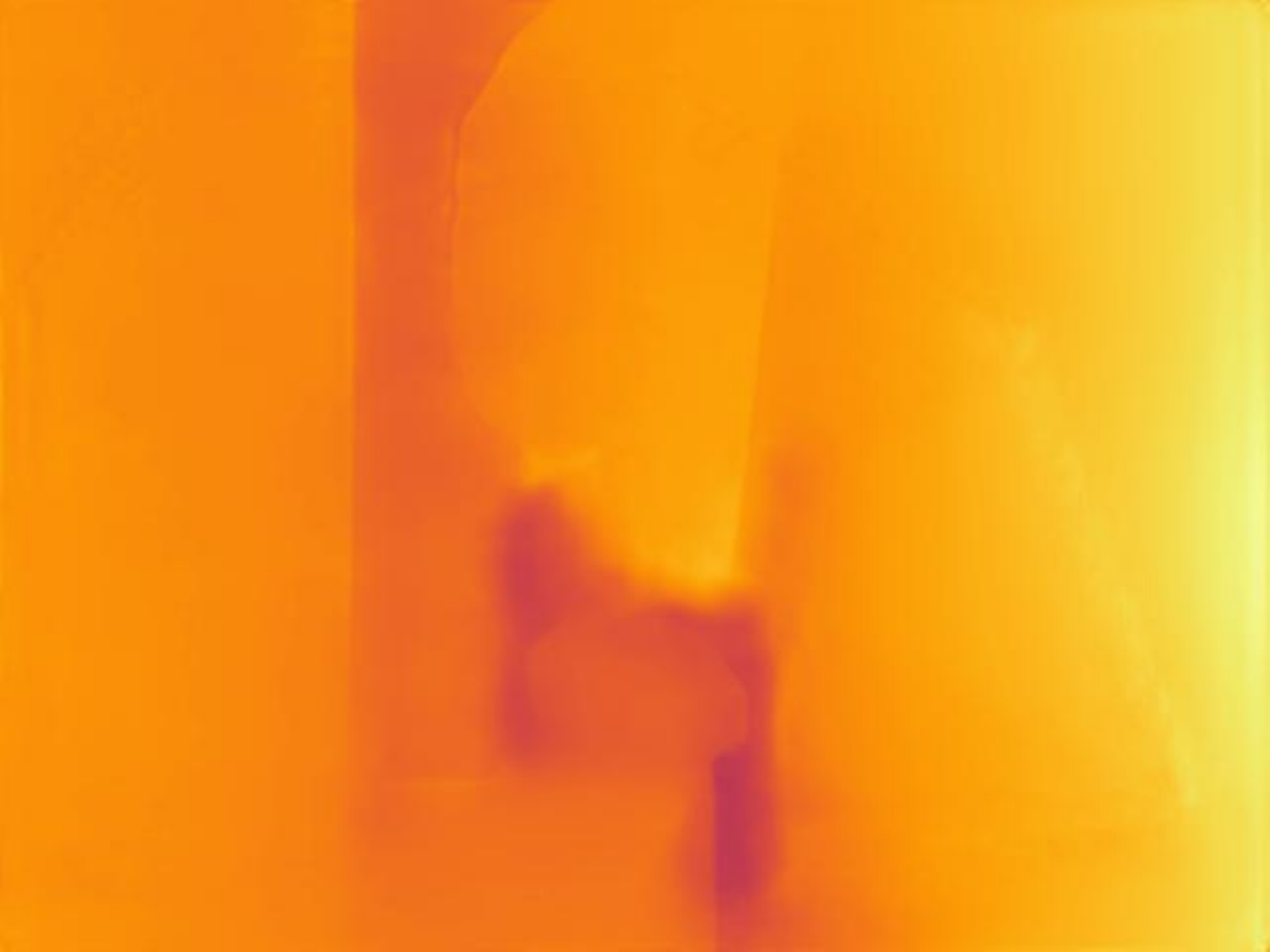}&
    \includegraphics[width=0.155\linewidth]{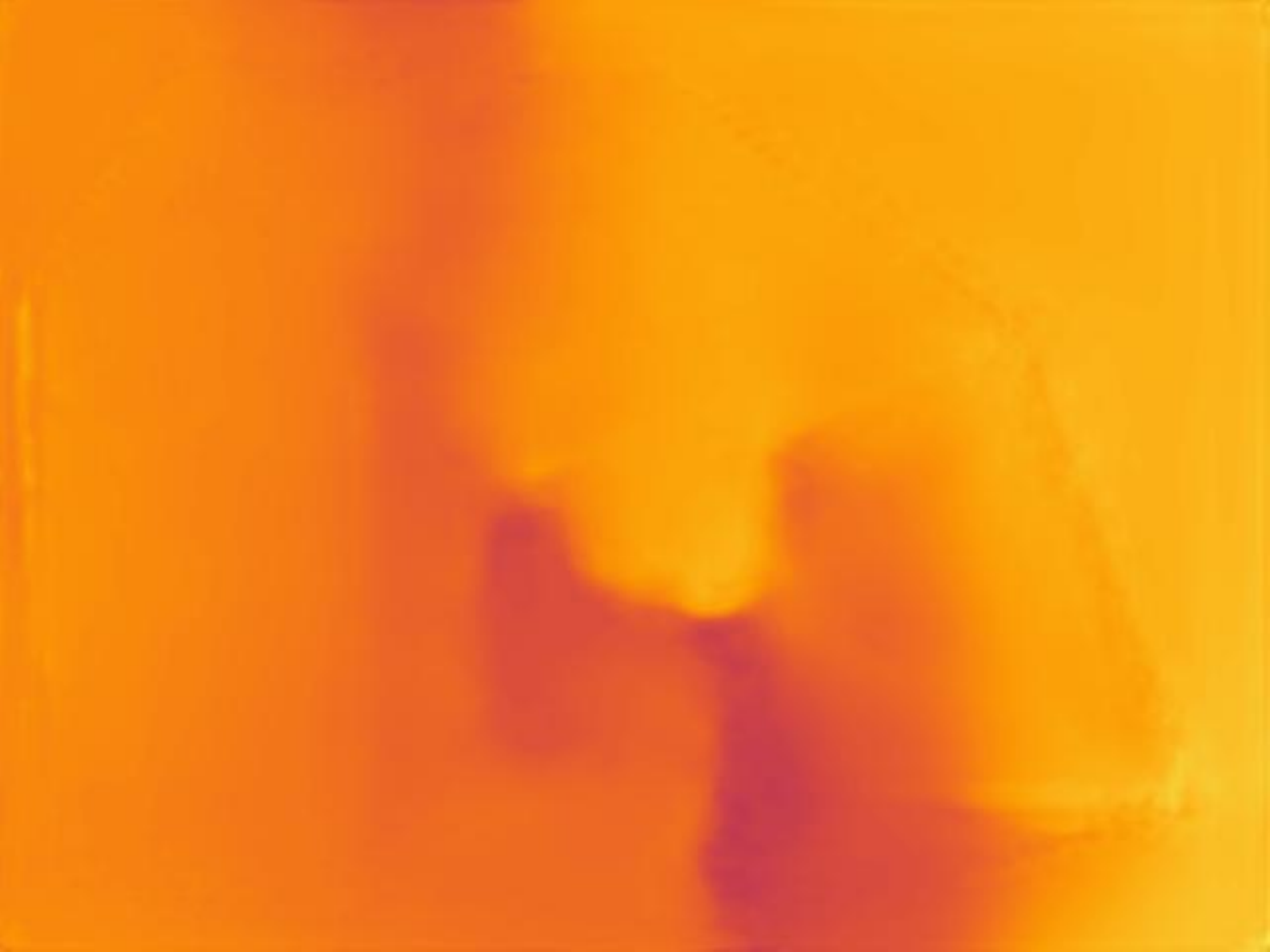}\\
    \vspace{-0.75mm}
    \includegraphics[width=0.155\linewidth]{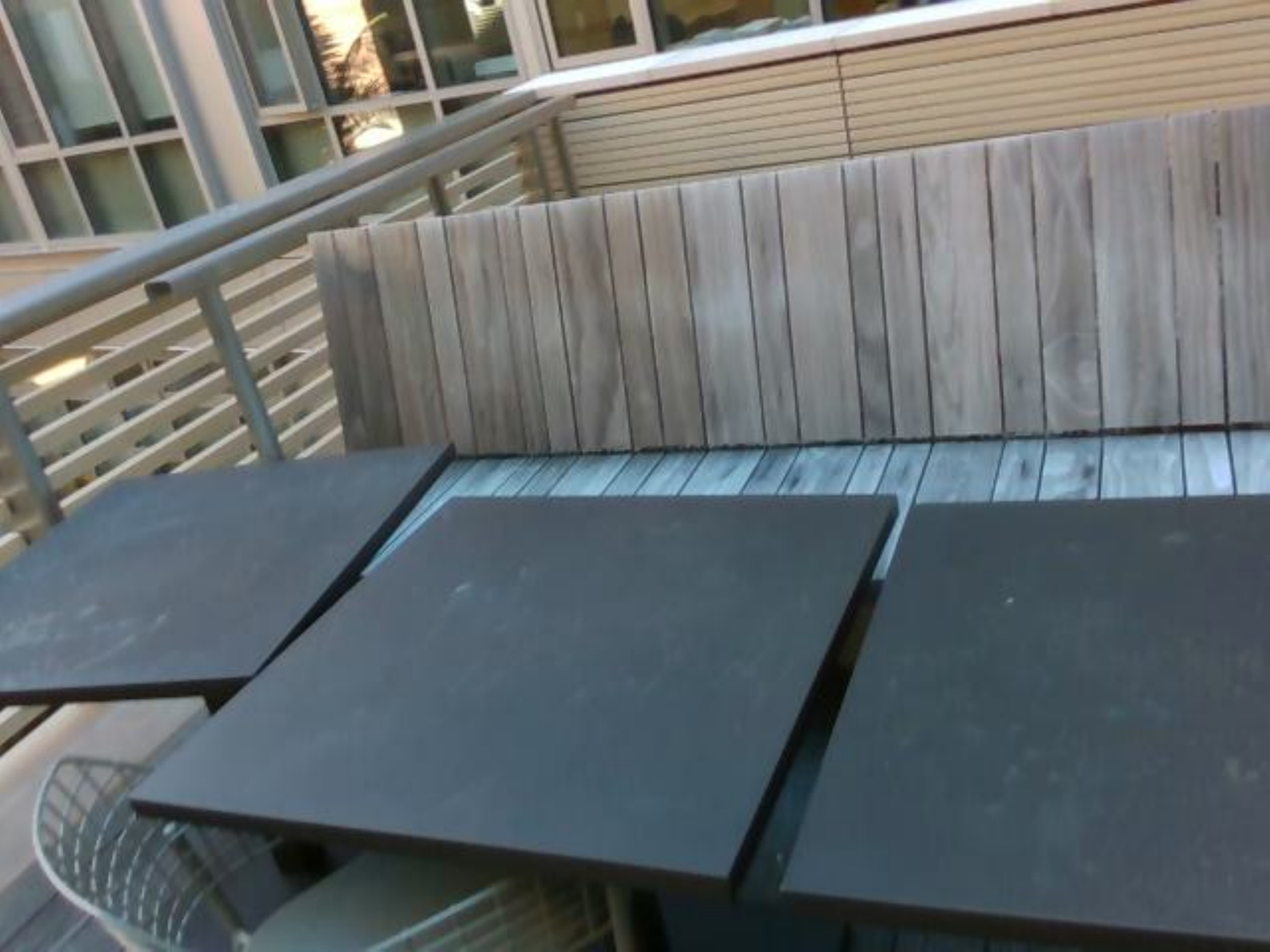}&
    \includegraphics[width=0.155\linewidth]{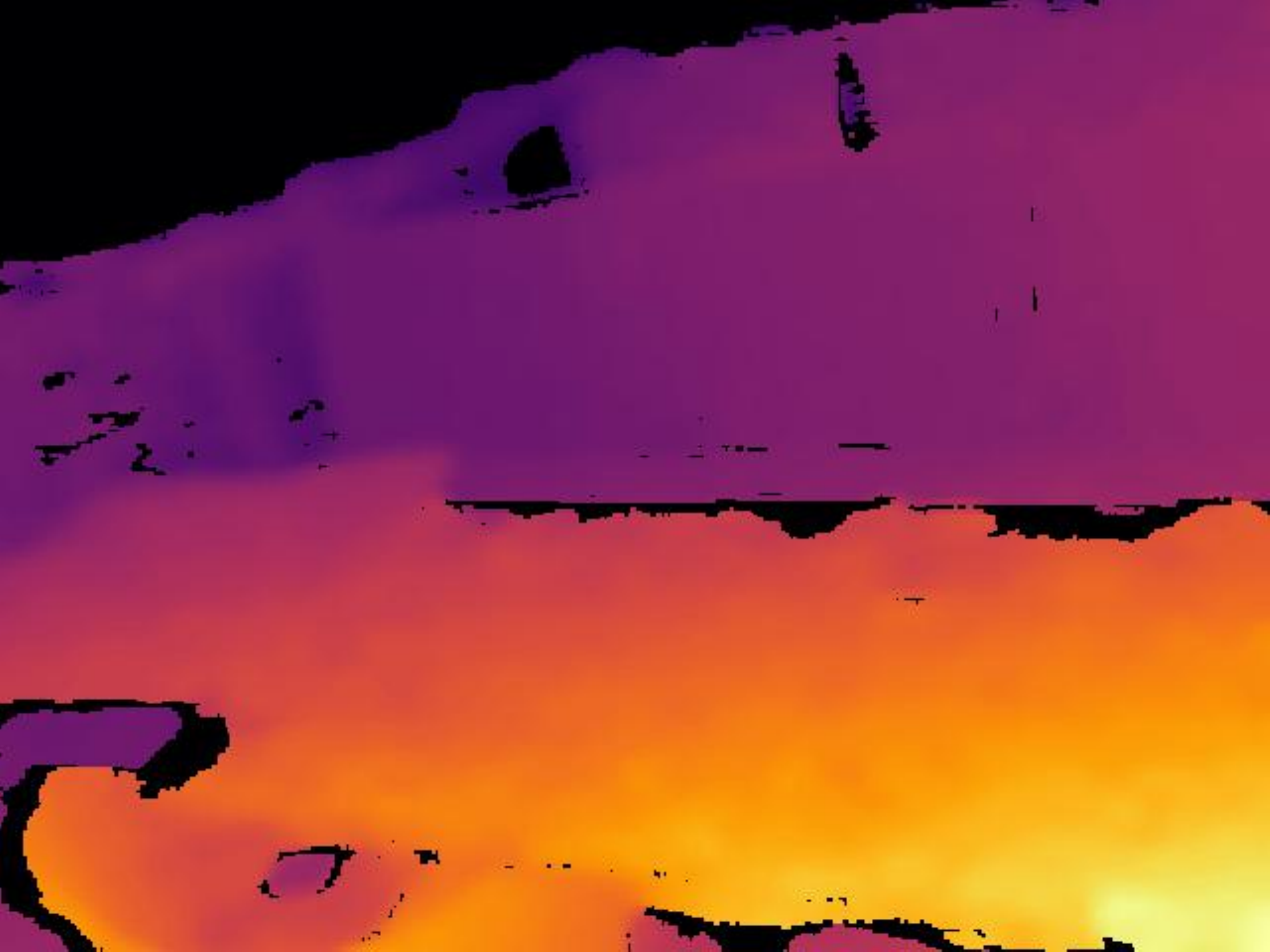}&
    \includegraphics[width=0.155\linewidth]{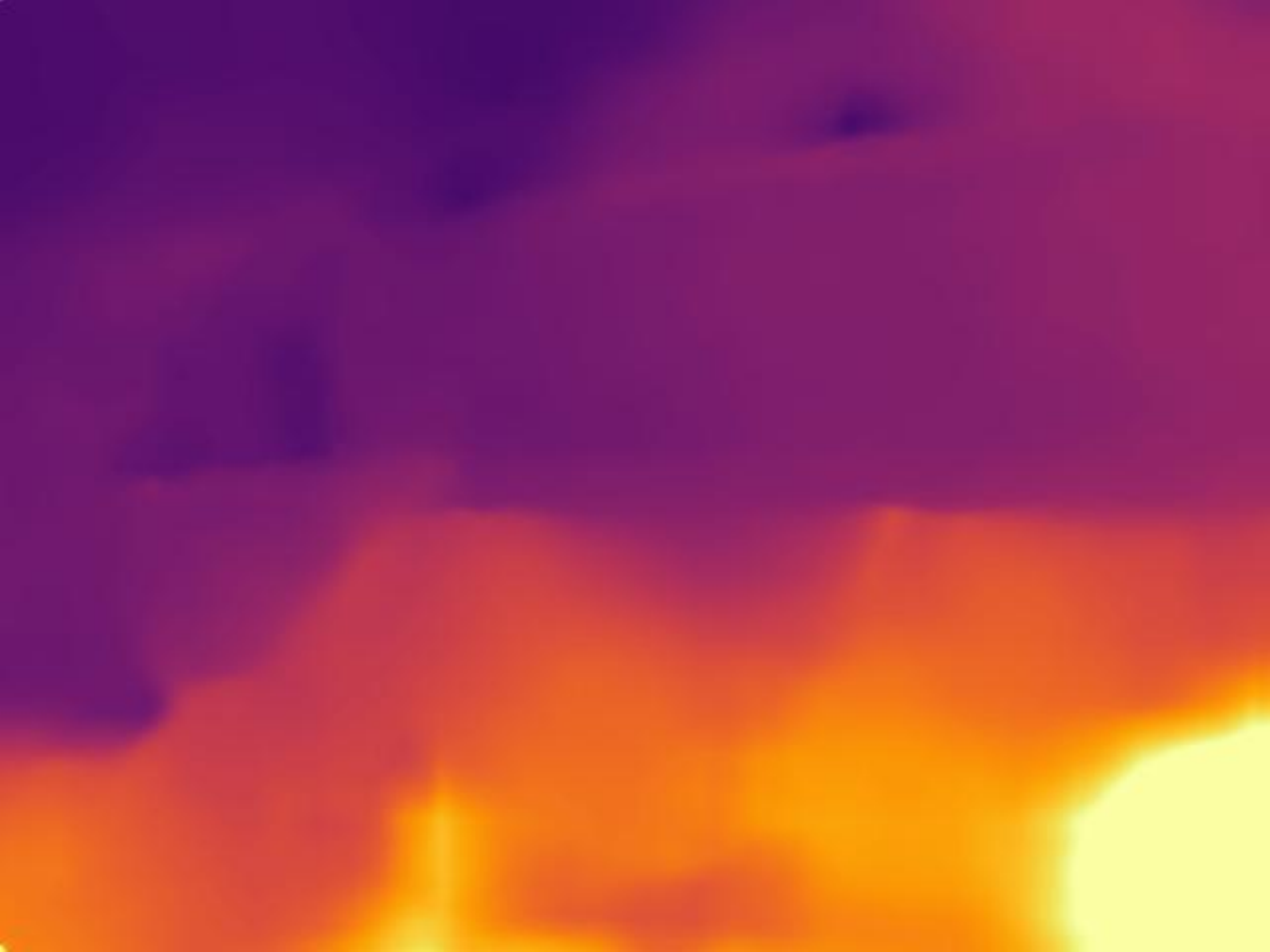}&
    \includegraphics[width=0.155\linewidth]{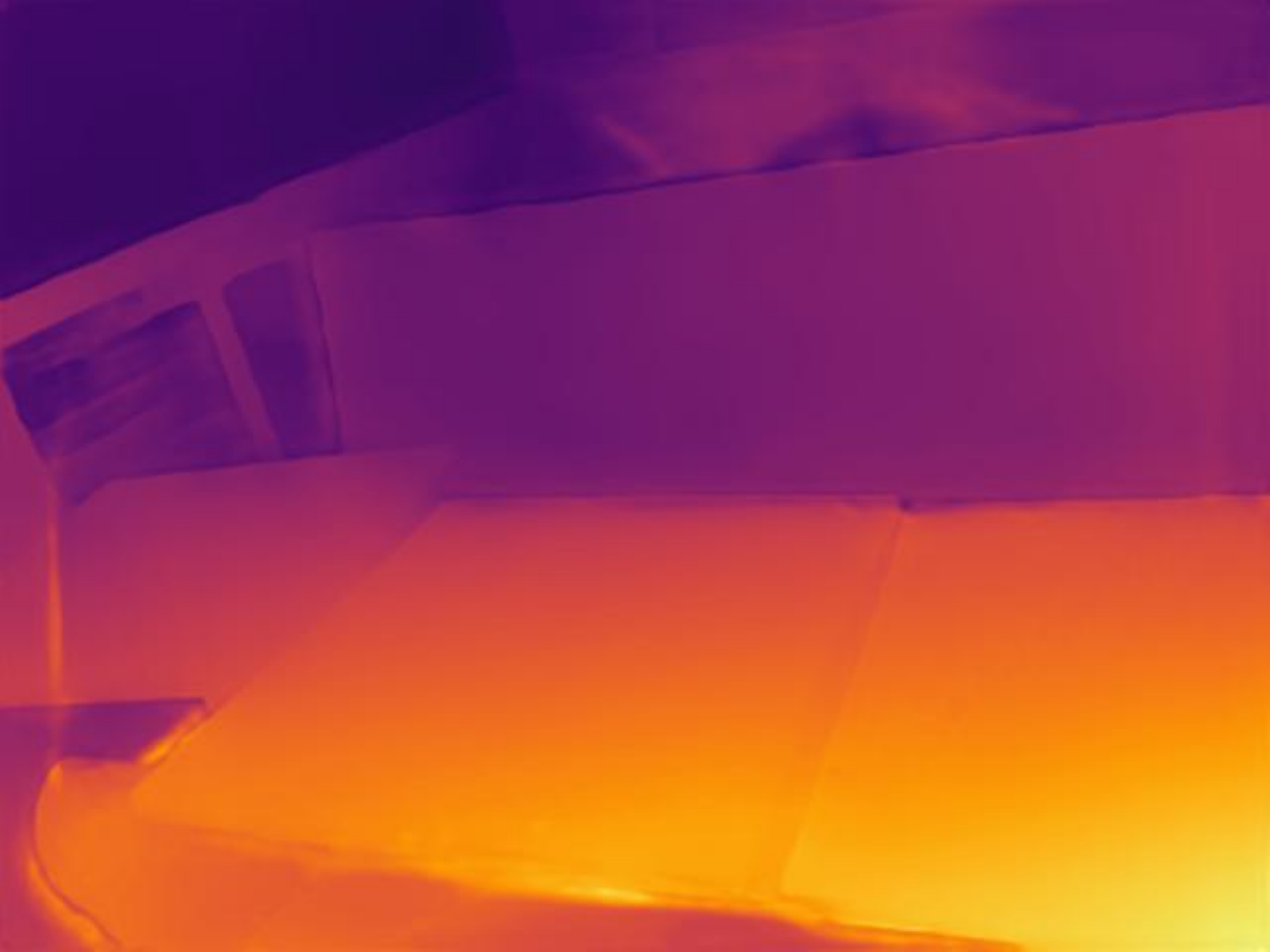}&
    \includegraphics[width=0.155\linewidth]{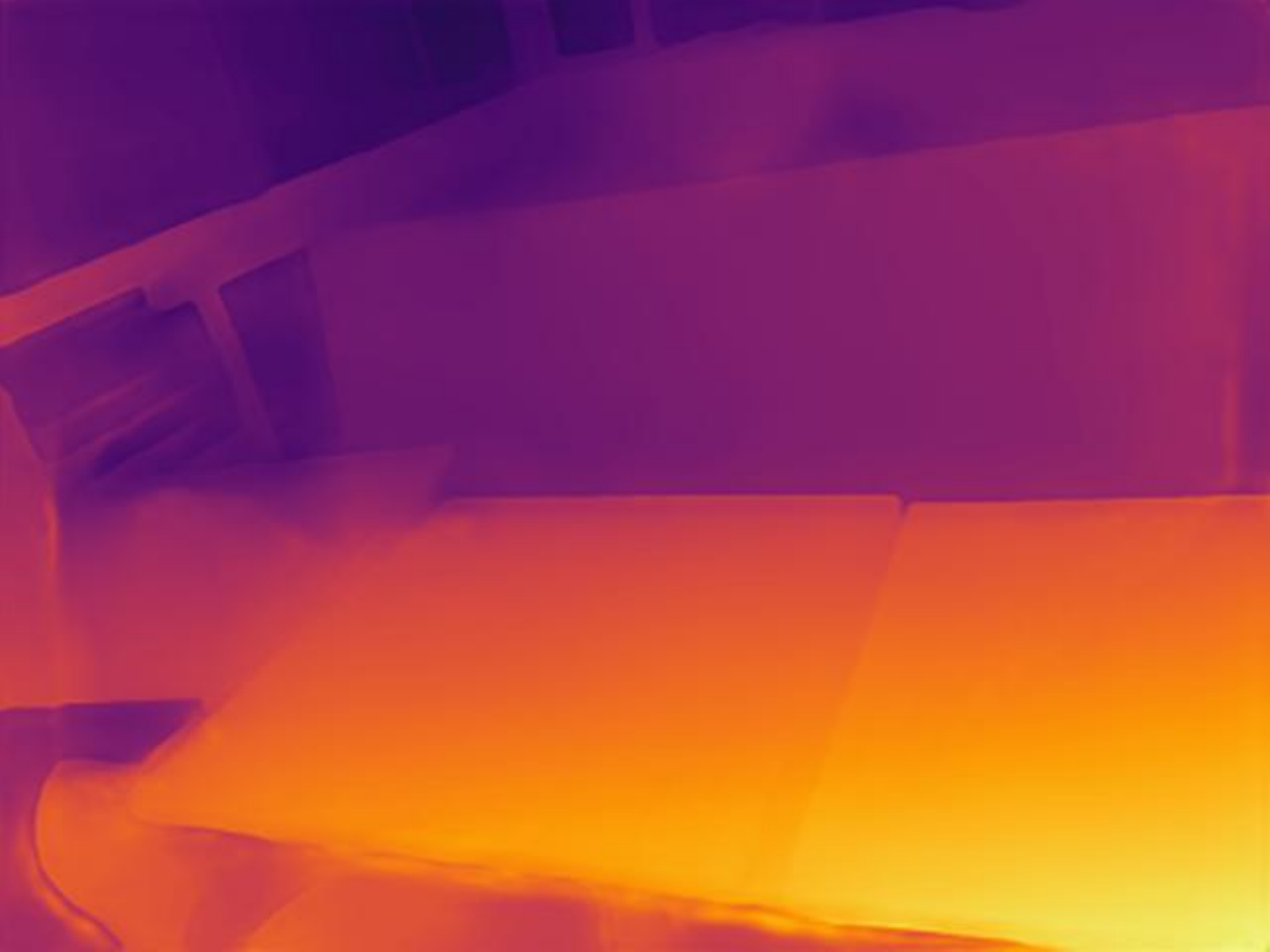}&
    \includegraphics[width=0.155\linewidth]{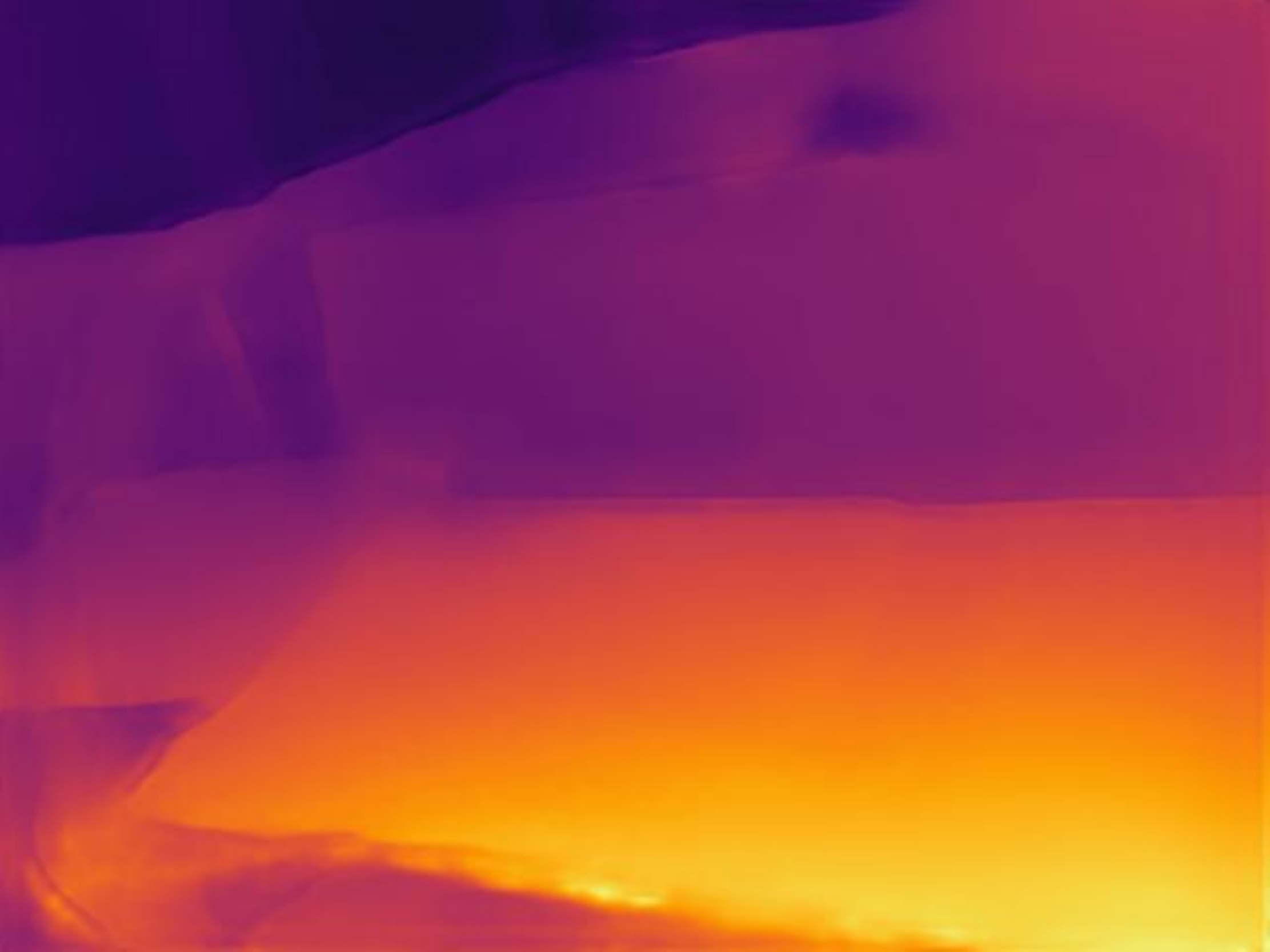}\\
    \vspace{-0.75mm}
  \end{tabular}
  \vspace{-8pt}
  \caption{Qualitative comparison of our approach against state-of-the-art KBNet on the VOID 150 dataset. SML is trained only on VOID.}
  \label{fig:vis-void-comparison}
  \vspace{-8pt}
\end{figure}

\mypara{Comparison to related work.} Our evaluation thus far has compared the impact of SML relative to global alignment only. We now compare to related work on VOID. Table \ref{tab:void_comparison} lists VOICED \cite{Wong2020void} and state-of-the-art KBNet \cite{Wong2021kbnet} alongside our approach (GA+SML). Figure~\ref{fig:vis-void-comparison} shows a qualitative comparison.

In addition to using DPT-Hybrid as the depth model in our pipeline, we try DPT-BEiT-Large for its higher accuracy and MiDaS-small for its computational efficiency. With just 150 sparse depth points, our approach GA+SML outperforms KBNet across all metrics, regardless of what depth estimator we use; improvement in iRMSE ranges from 34\% to 55\%. From Table \ref{tab:void_eval}, we see that even with zero-shot transfer, our method outperforms KBNet by 42\% in iRMSE. At a higher density of 500 points, our pipeline with DPT-BEiT-Large continues to outperform KBNet across all metrics.

\vspace{-12pt}

\subsection{Generalizability and Deployability}

We test zero-shot generalization on NYU Depth v2~\cite{Silberman:ECCV12} and VOID, comparing against NLSPN~\cite{park2020nlspn} (state of the art on NYUv2) and KBNet (state of the art on VOID). These models, having been trained on a single dataset as is commonplace with depth completion tasks, underperform when run on a different dataset. Table~\ref{tab:nyu_and_void} shows that our approach consistently achieves better generalization performance.

\begin{table}
  \centering
  \footnotesize
    \caption{Testing zero-shot generalizability on NYUv2 and VOID. DPT-Hybrid is used as the depth predictor for GA+SML.}
    \label{tab:nyu_and_void}
    \vspace{-4pt}
    \begin{tabular}{@{}
    c@{\hspace{1mm}}|
    l@{\hspace{4mm}}
    S[table-format=3.1]@{\hspace{2mm}}
    S[table-format=3.1]@{\hspace{2mm}}|
    l@{\hspace{4mm}}
    S[table-format=3.1]@{\hspace{1mm}}
    S[table-format=3.1]@{\hspace{0mm}}
    @{}}
    \toprule
     & \multicolumn{3}{c|}{\textit{NYUv2 (train) $\rightarrow$ VOID (test)}} & \multicolumn{3}{c}{\textit{VOID (train) $\rightarrow$ NYUv2 (test)}} \\
    & Method & {iMAE} & {iRMSE} & Method & {iMAE} & {iRMSE}\\
    \midrule
    150 & NLSPN \cite{park2020nlspn} & 143.0 & 238.1 & KBNet \cite{Wong2021kbnet} & 35.2 & 67.8 \\
    pts &  GA+SML & \bfseries 55.9 & \bfseries 85.2 &  GA+SML & \bfseries 30.2 & \bfseries 48.9\\
    \midrule
    500 & NLSPN \cite{park2020nlspn} & 87.9 & 174.7 & KBNet \cite{Wong2021kbnet} & 28.0 & 57.2\\
    pts &  GA+SML & \bfseries 43.9 & \bfseries 69.5 & GA+SML & \bfseries 26.8 & \bfseries 44.5 \\
    \bottomrule
  \end{tabular}
  \vspace{-6pt}
\end{table}

\begin{figure}
\centering
\footnotesize
  \begin{tabular}{@{}*{6}{c@{\hspace{0.5mm}}}c@{}}
    {\scriptsize RGB} & {\scriptsize Sparse} & {\scriptsize Scaffold} & {\scriptsize Regressed} & {\scriptsize GT Depth} & {\scriptsize GA Depth} & {\scriptsize SML Depth}\\
    \vspace{-0.75mm}
    \includegraphics[width=0.135\linewidth]{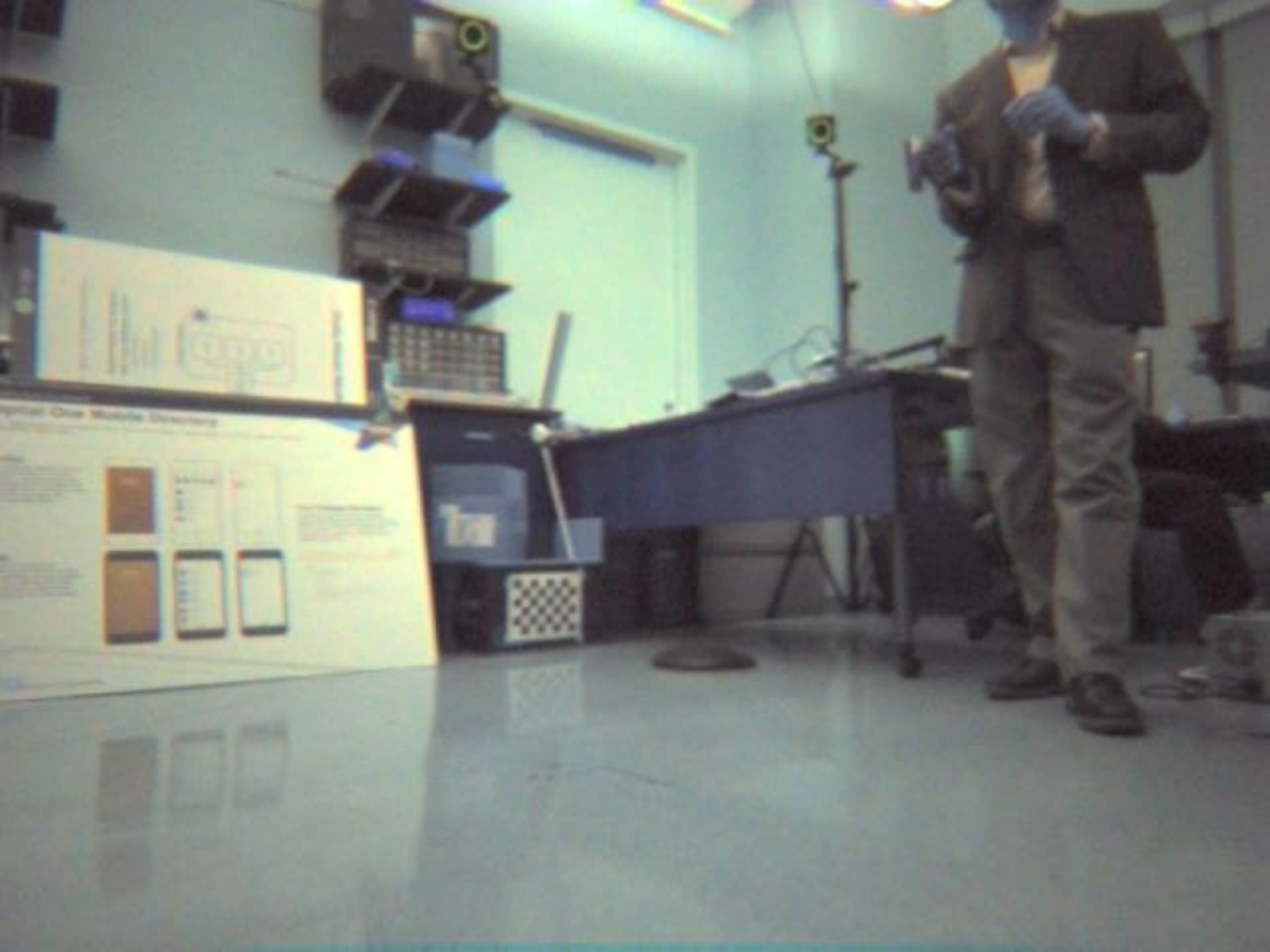}&
    \includegraphics[width=0.135\linewidth]{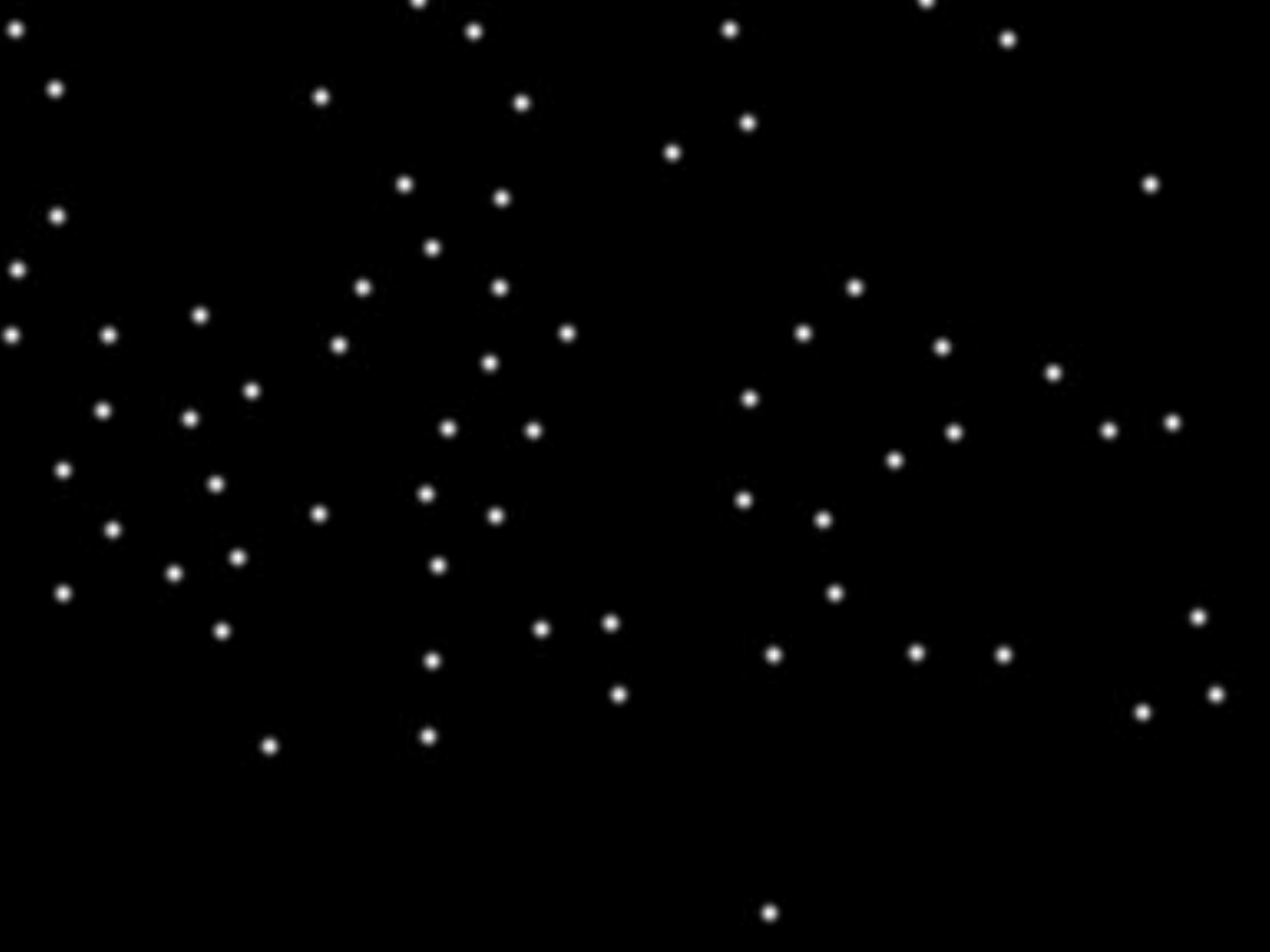}&
    \includegraphics[width=0.135\linewidth]{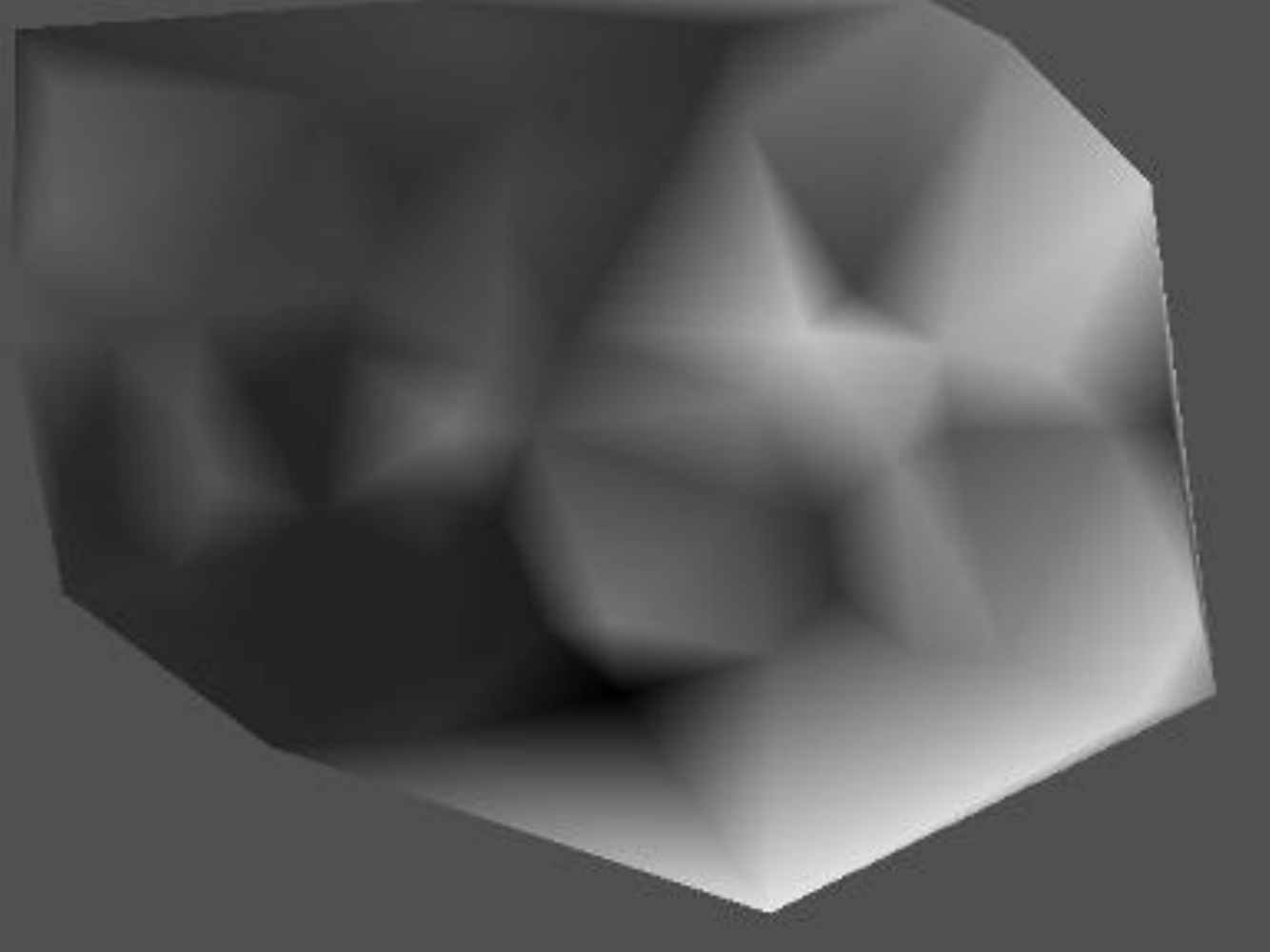}&
    \includegraphics[width=0.135\linewidth]{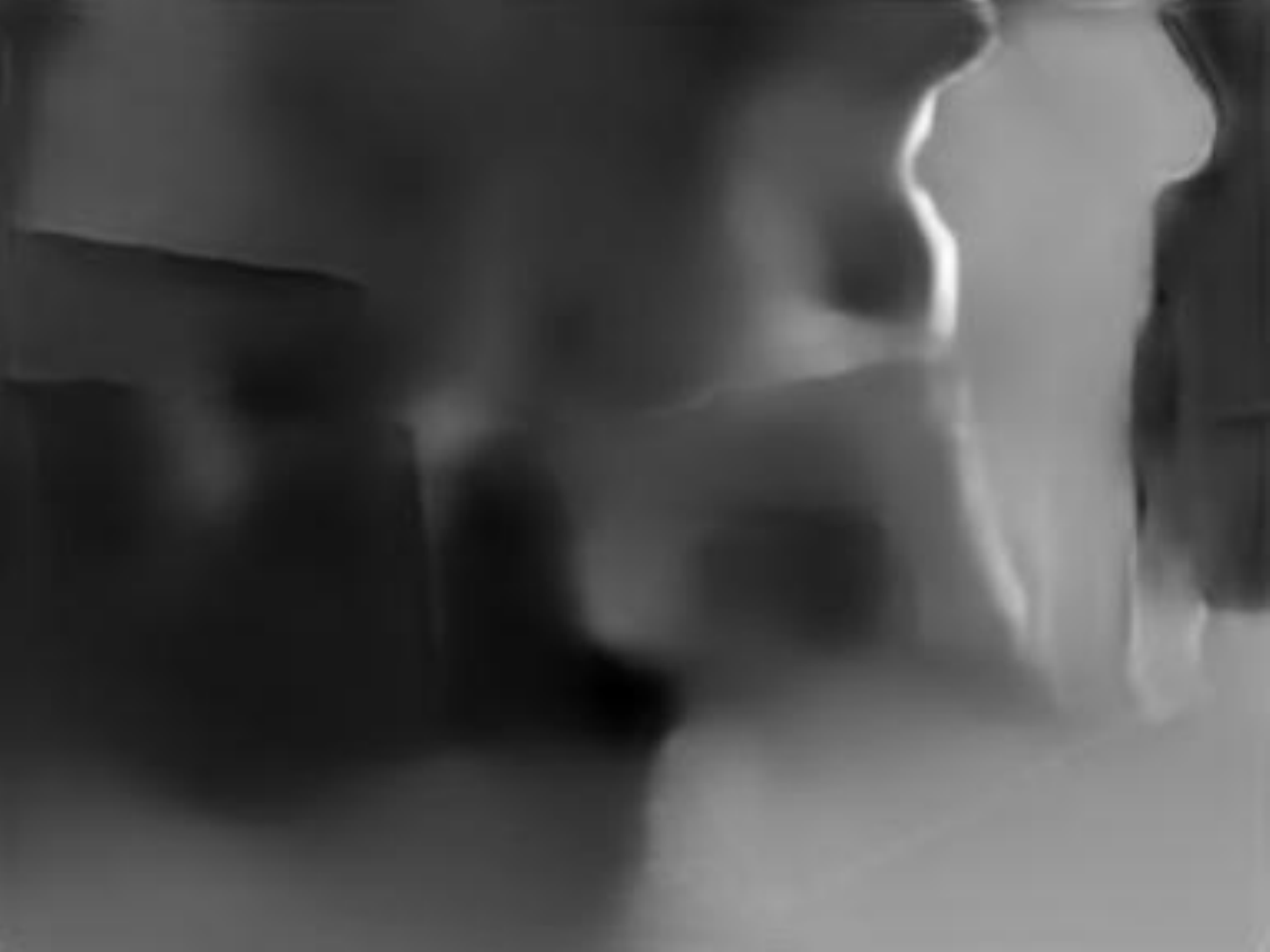}&
    \includegraphics[width=0.135\linewidth]{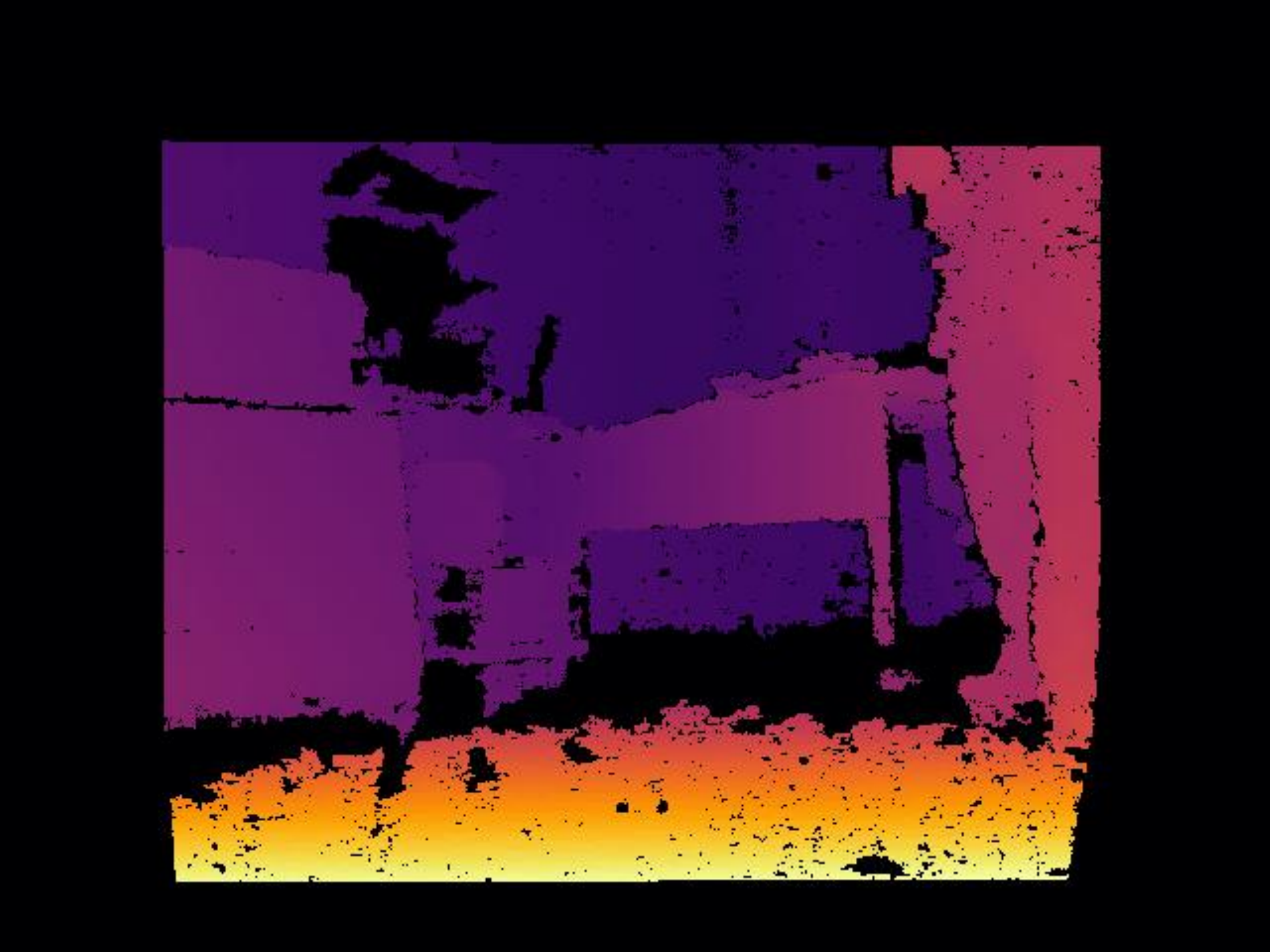}&
    \includegraphics[width=0.135\linewidth]{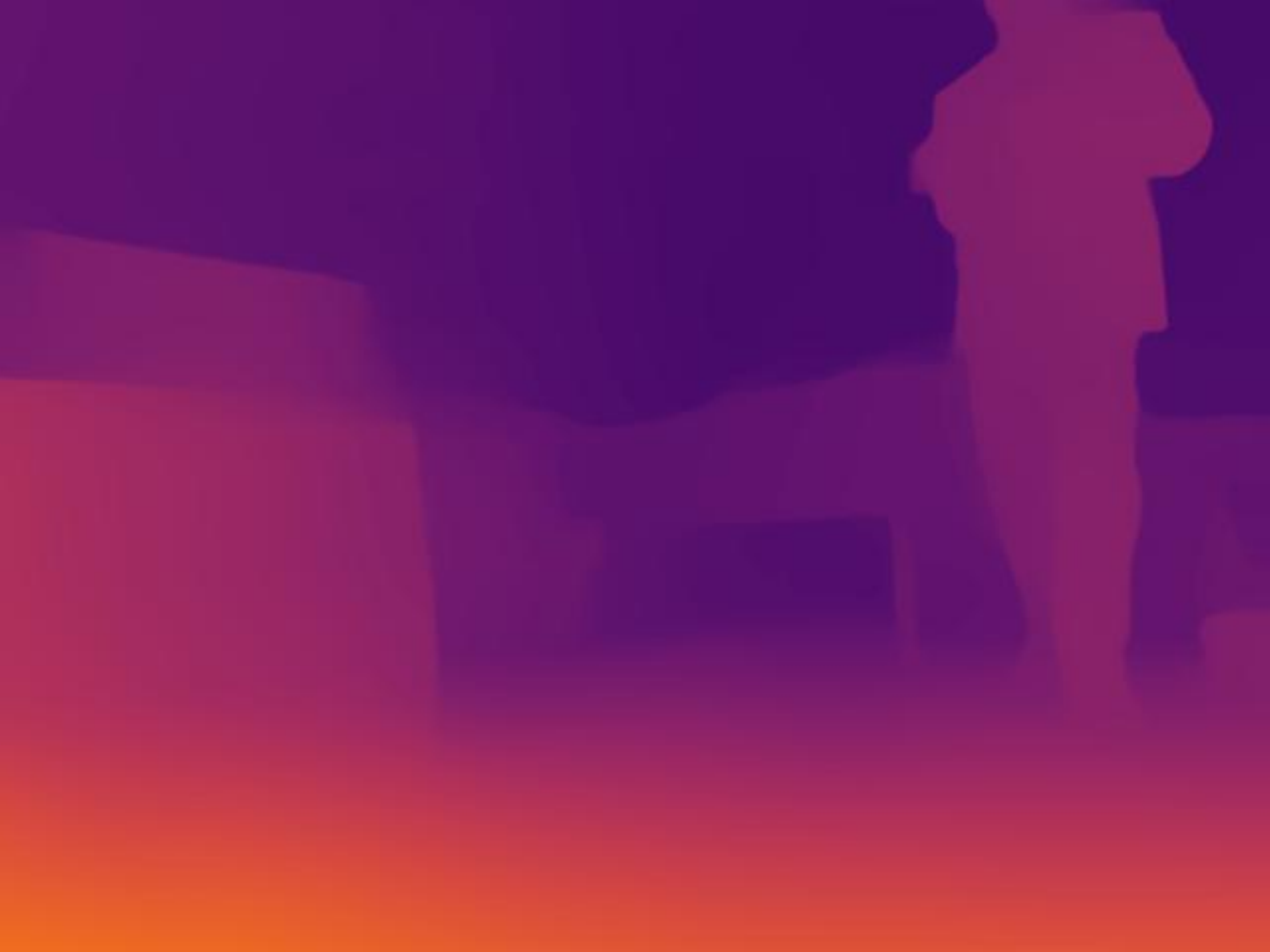}&
    \includegraphics[width=0.135\linewidth]{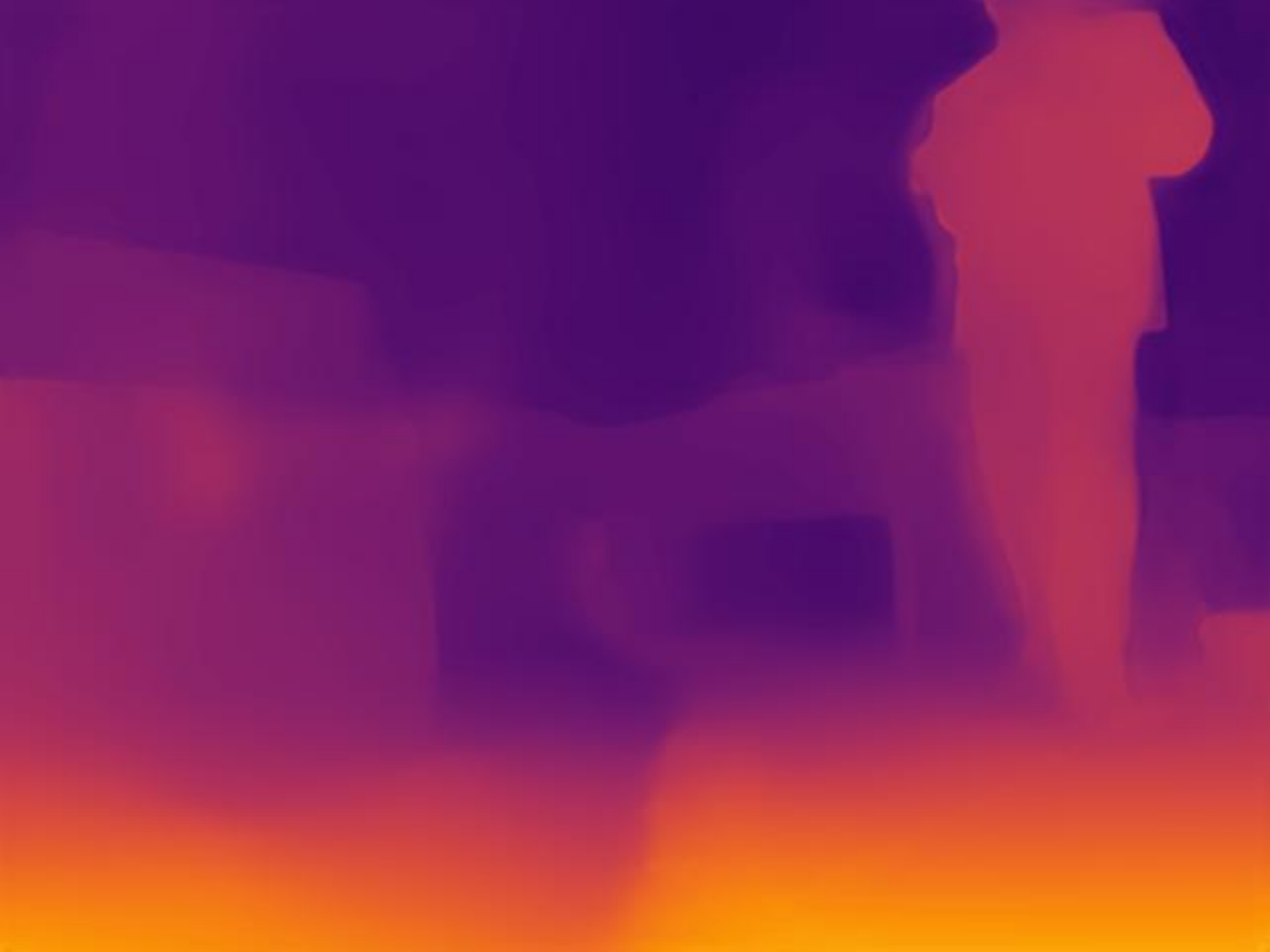}\\
    & \scriptsize 66 points & & & \multicolumn{3}{c}{\hspace{8pt}\scriptsize{iRMSE = 177.9 $\rightarrow$ 139.2}} \\
    \vspace{-0.75mm}
    \includegraphics[width=0.135\linewidth]{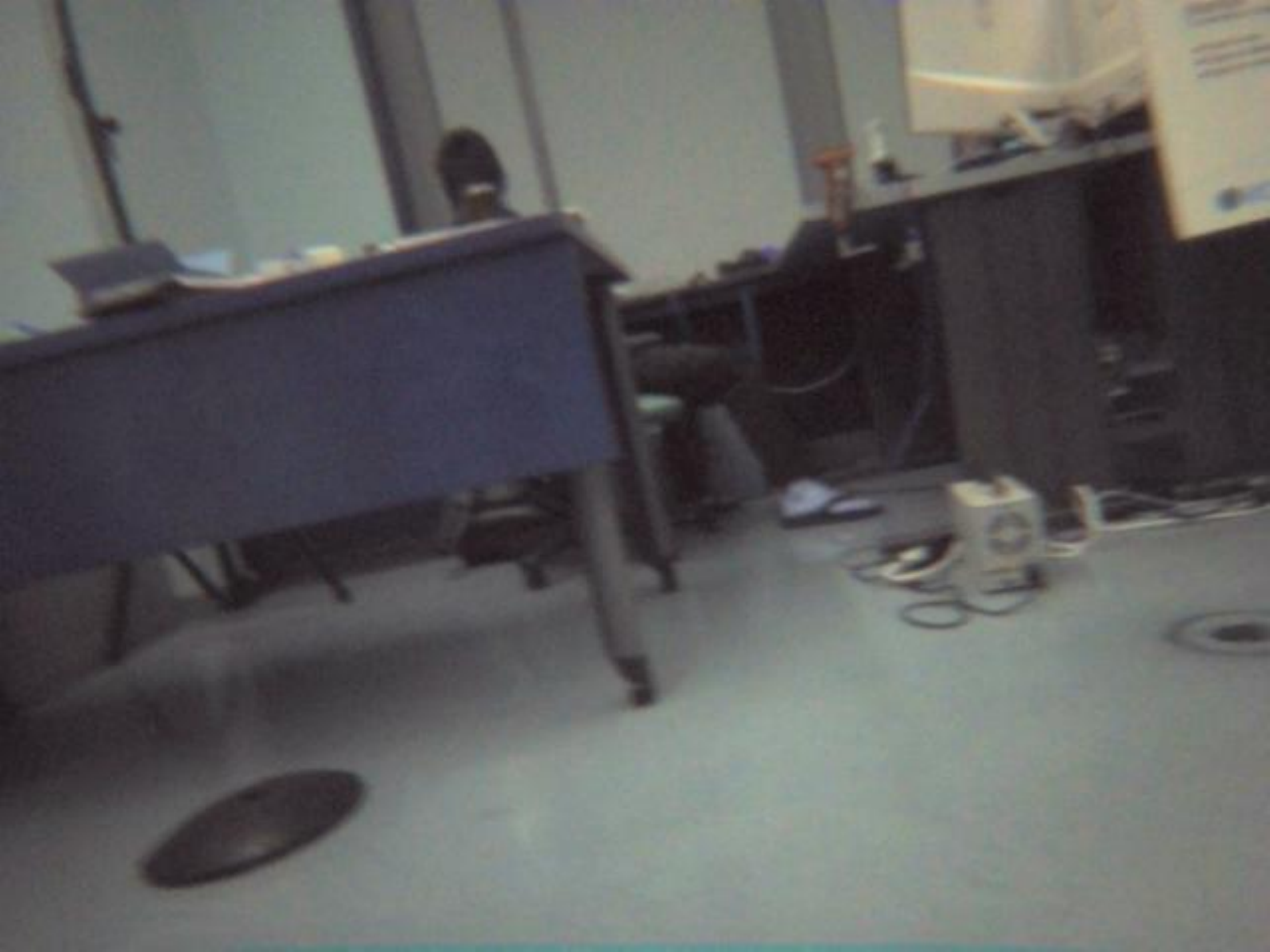}&
    \includegraphics[width=0.135\linewidth]{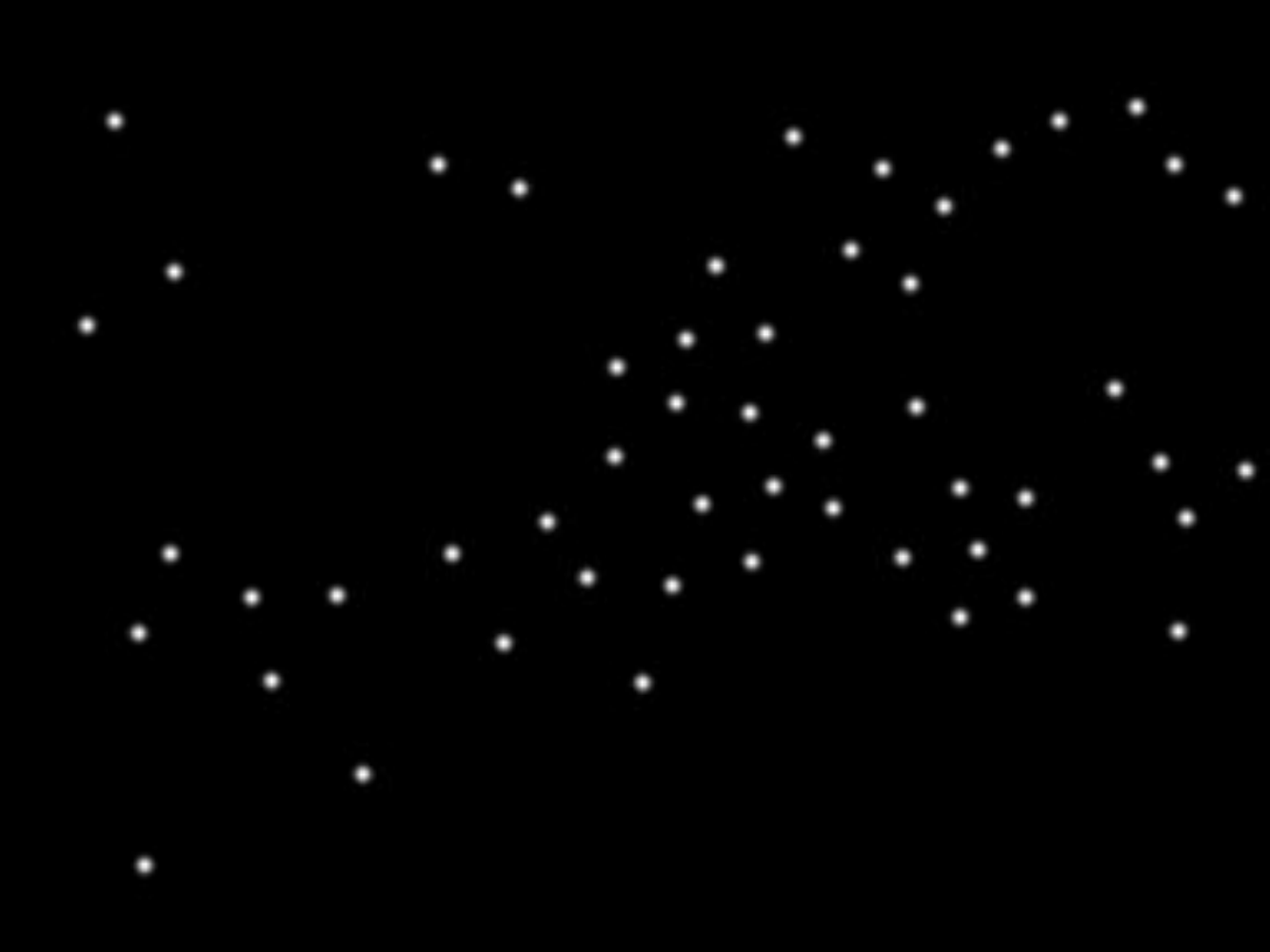}&
    \includegraphics[width=0.135\linewidth]{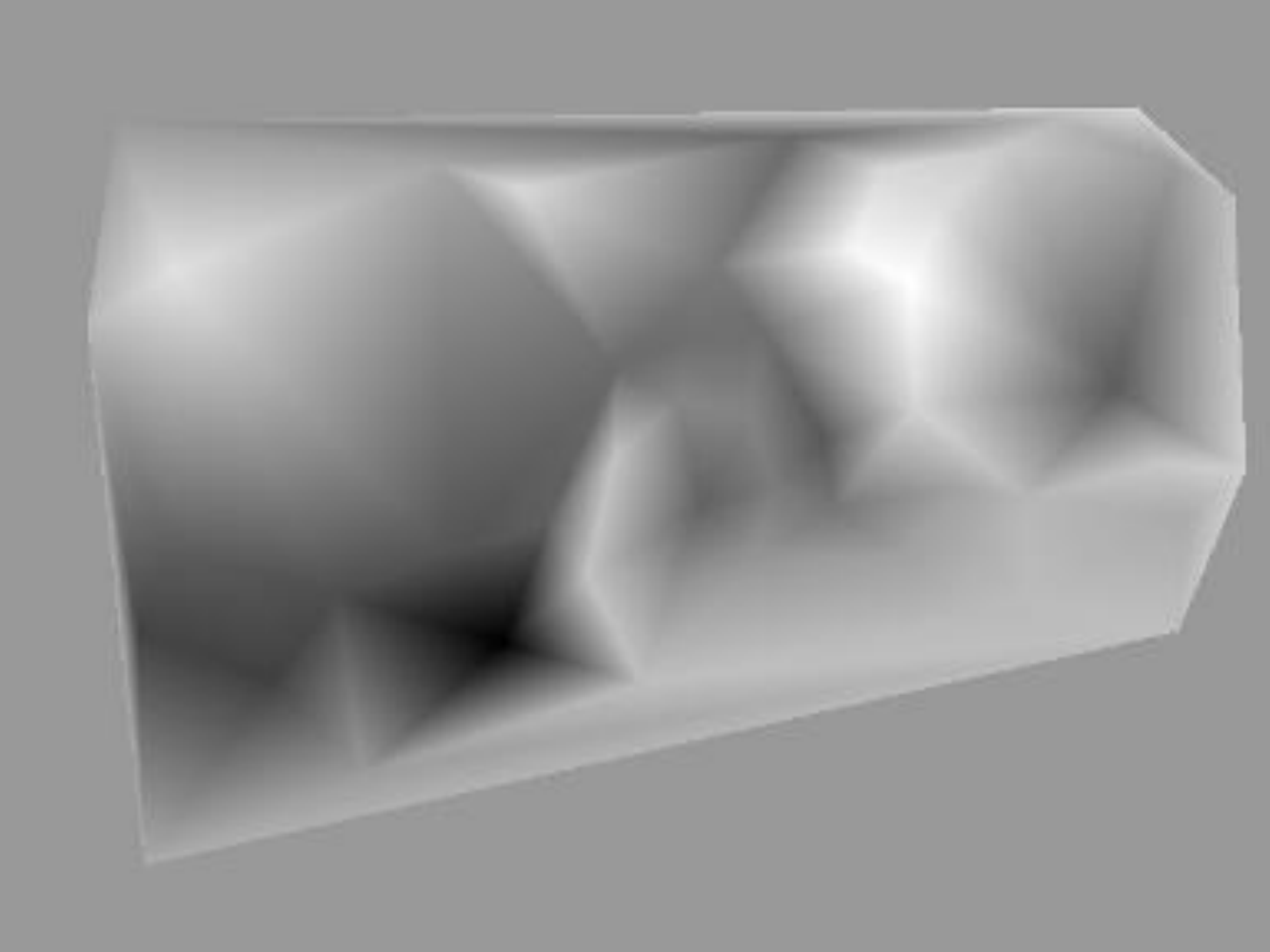}&
    \includegraphics[width=0.135\linewidth]{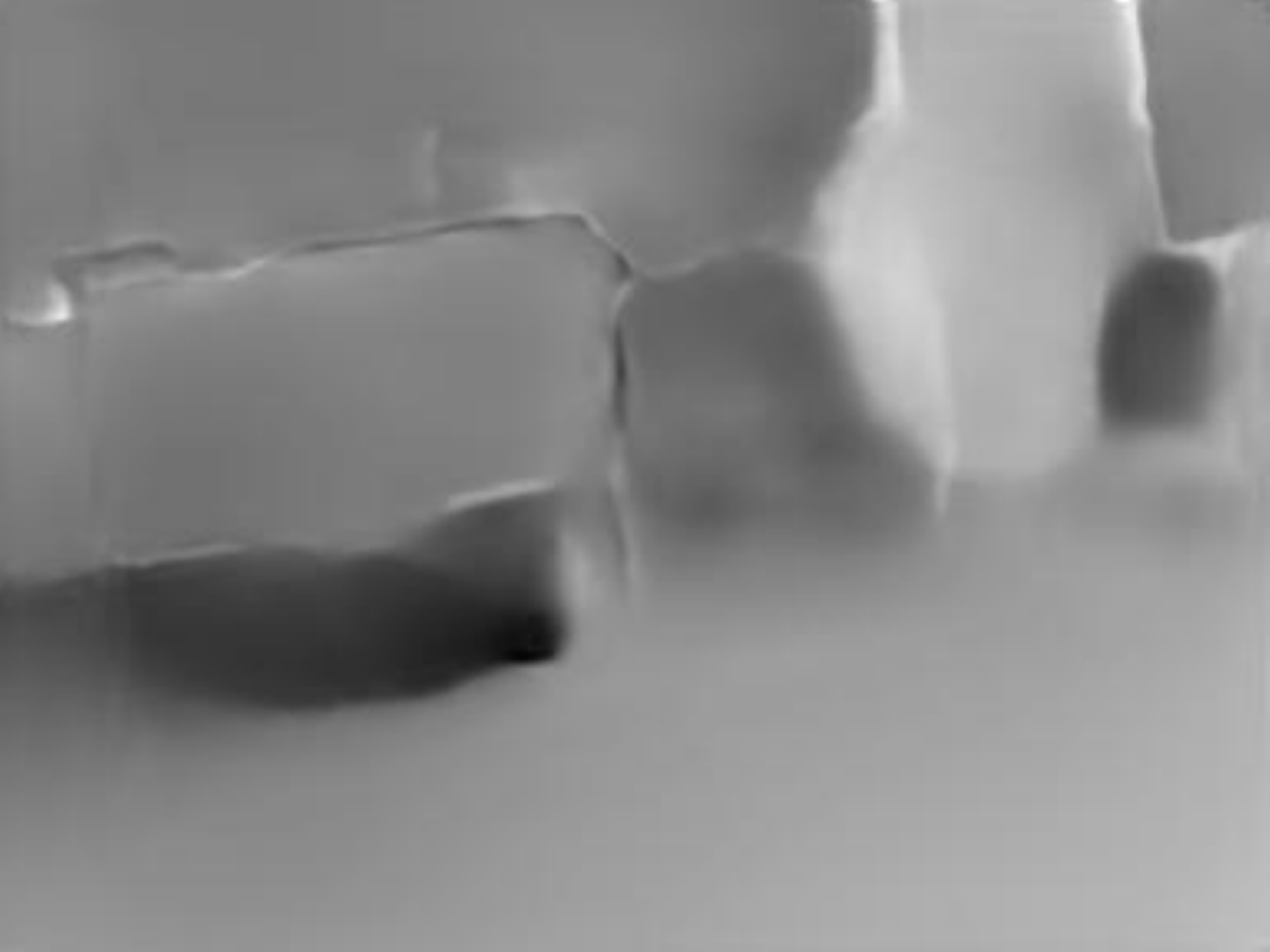}&
    \includegraphics[width=0.135\linewidth]{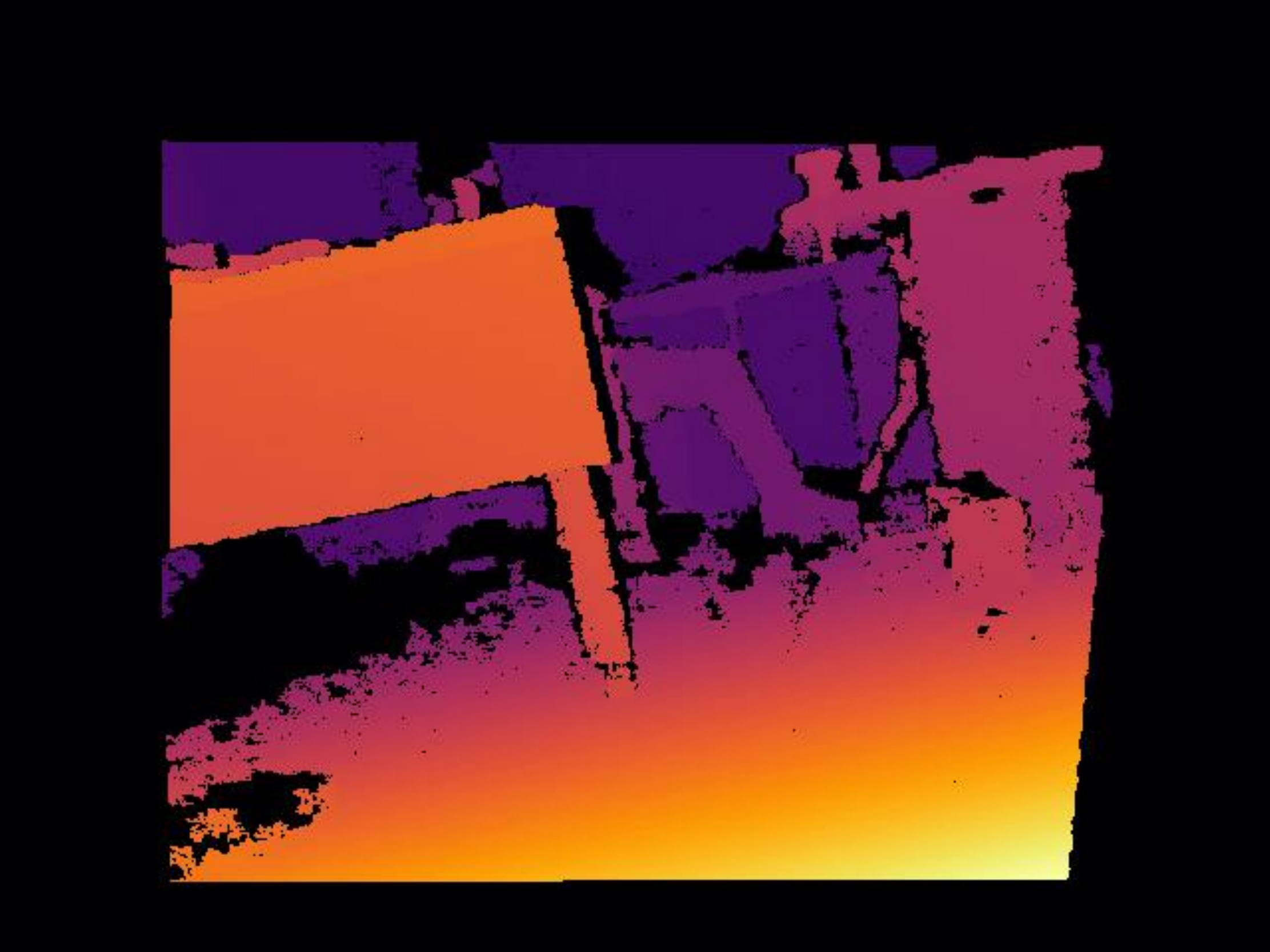}&
    \includegraphics[width=0.135\linewidth]{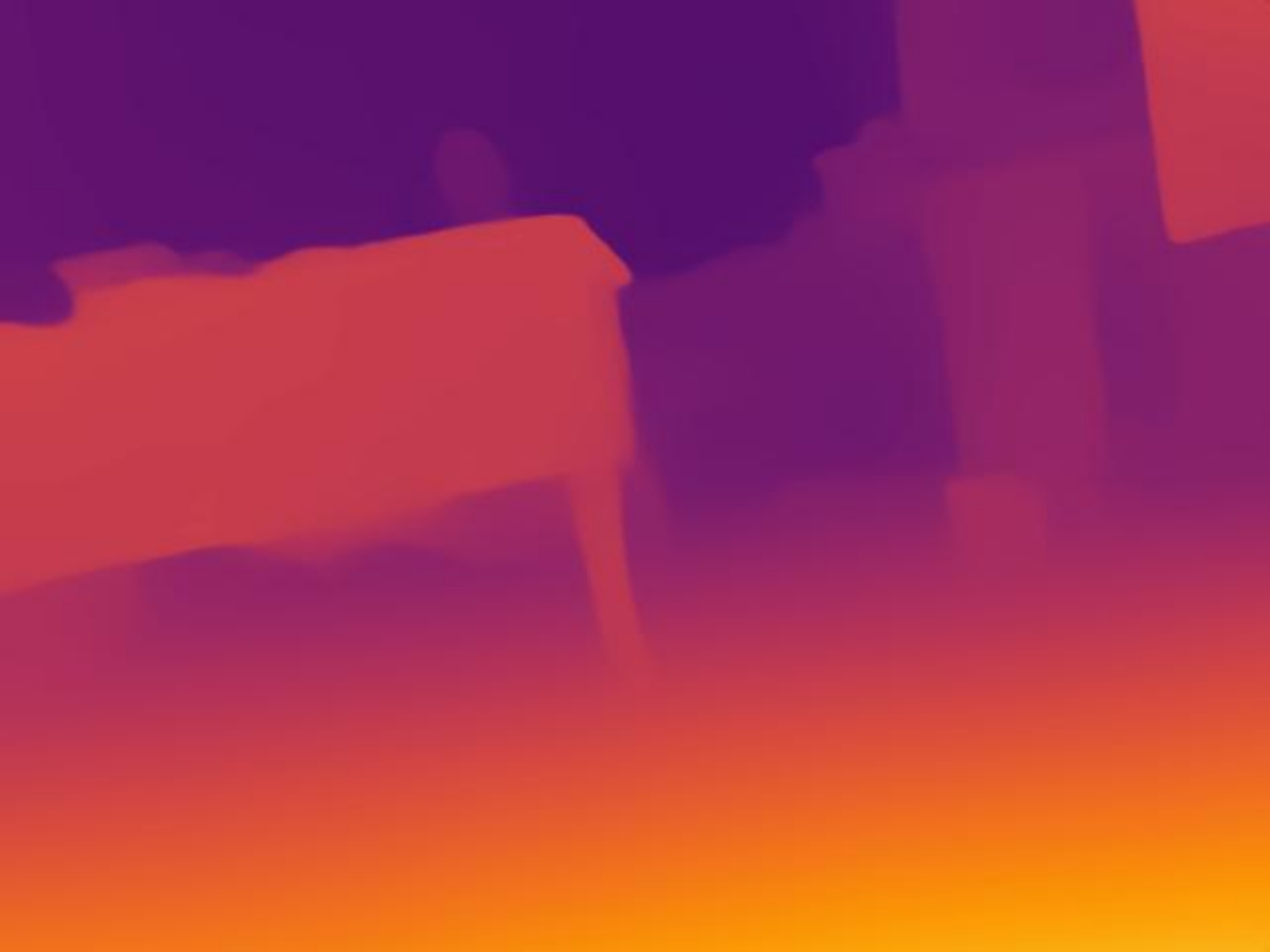}&
    \includegraphics[width=0.135\linewidth]{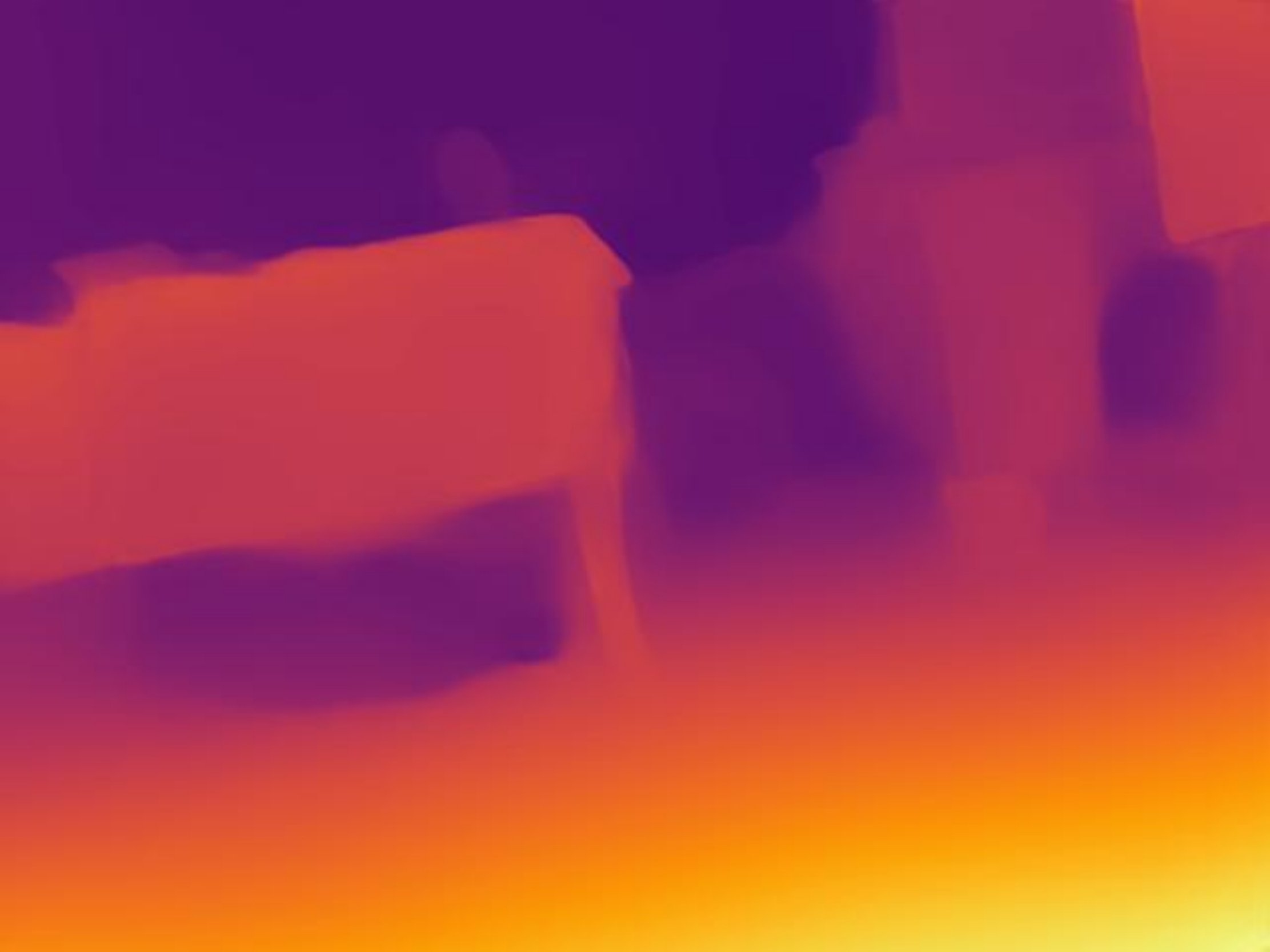}\\
    & \scriptsize 52 points & & & \multicolumn{3}{c}{\hspace{5pt}\scriptsize{iRMSE = 118.7 $\rightarrow$ 93.4}} \\
    \vspace{-0.75mm}
    \includegraphics[width=0.135\linewidth]{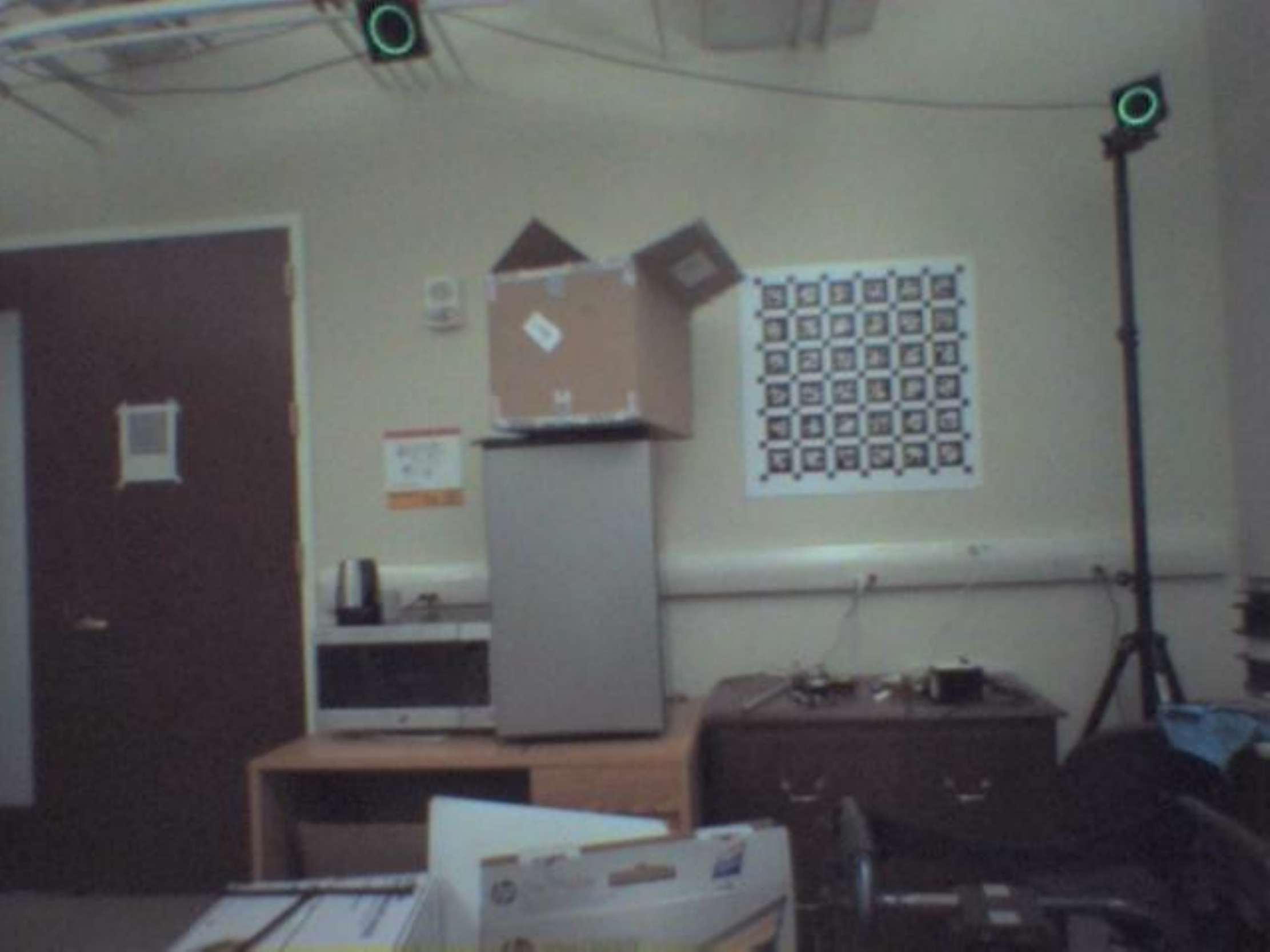}&
    \includegraphics[width=0.135\linewidth]{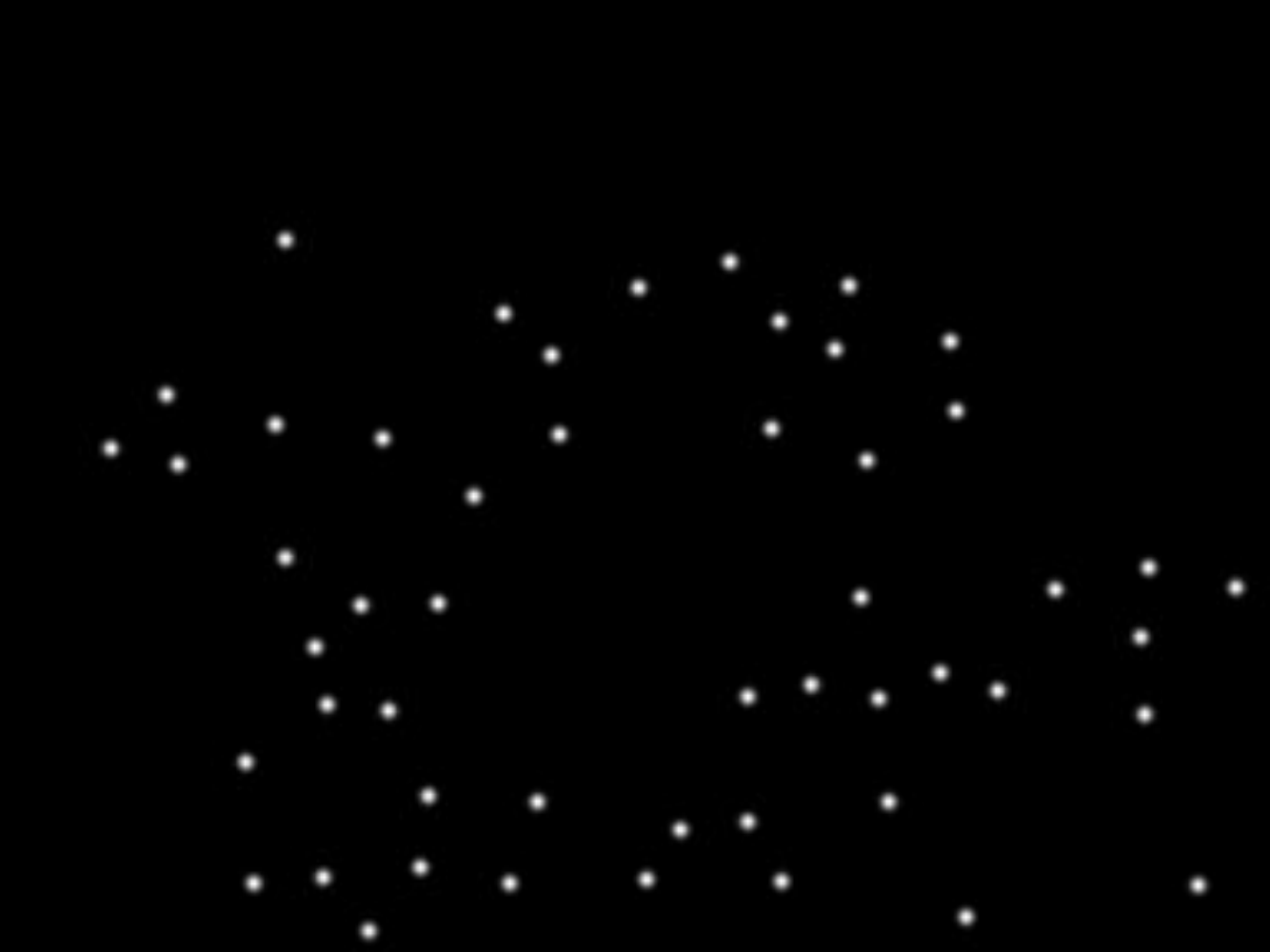}&
    \includegraphics[width=0.135\linewidth]{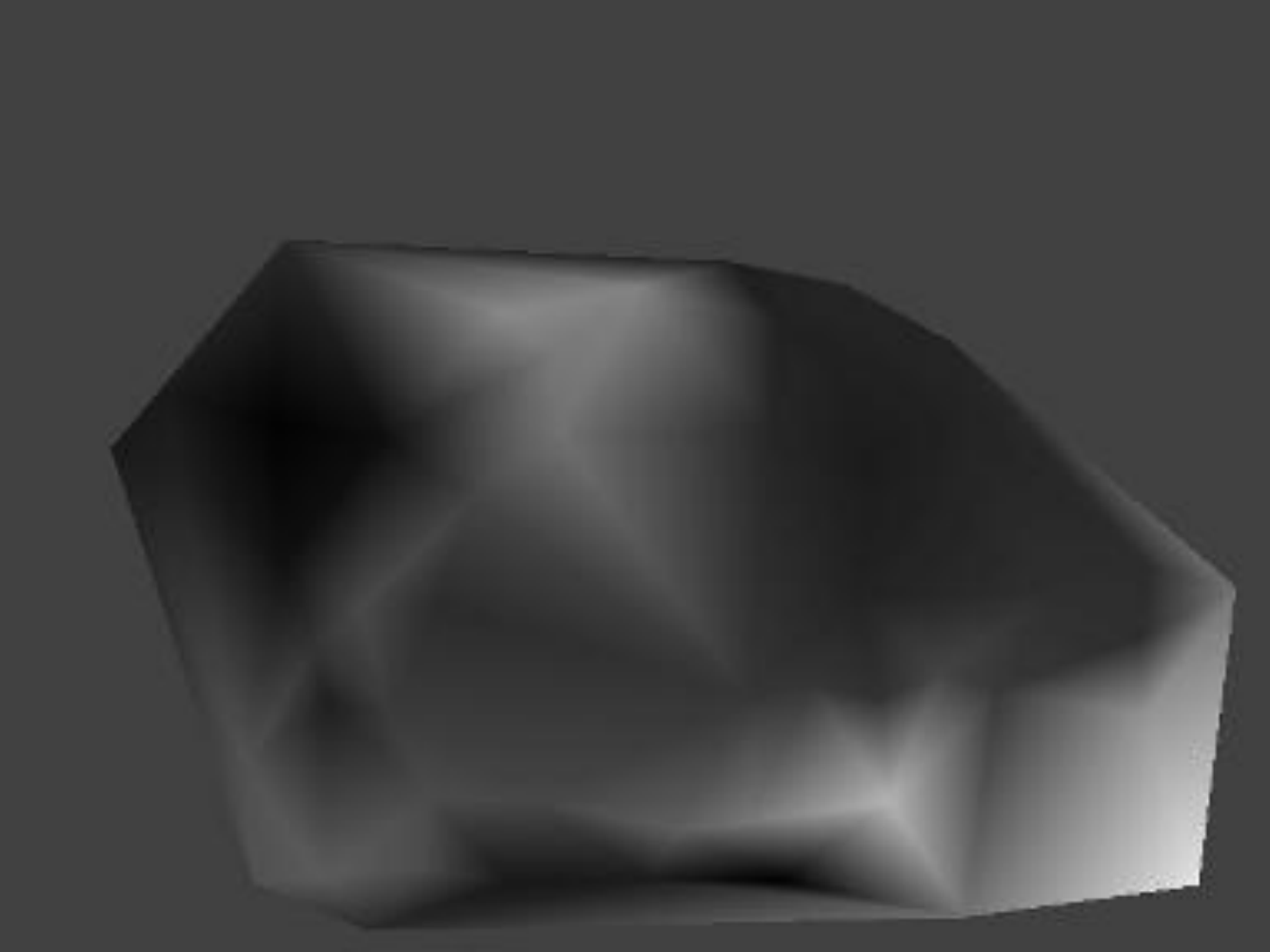}&
    \includegraphics[width=0.135\linewidth]{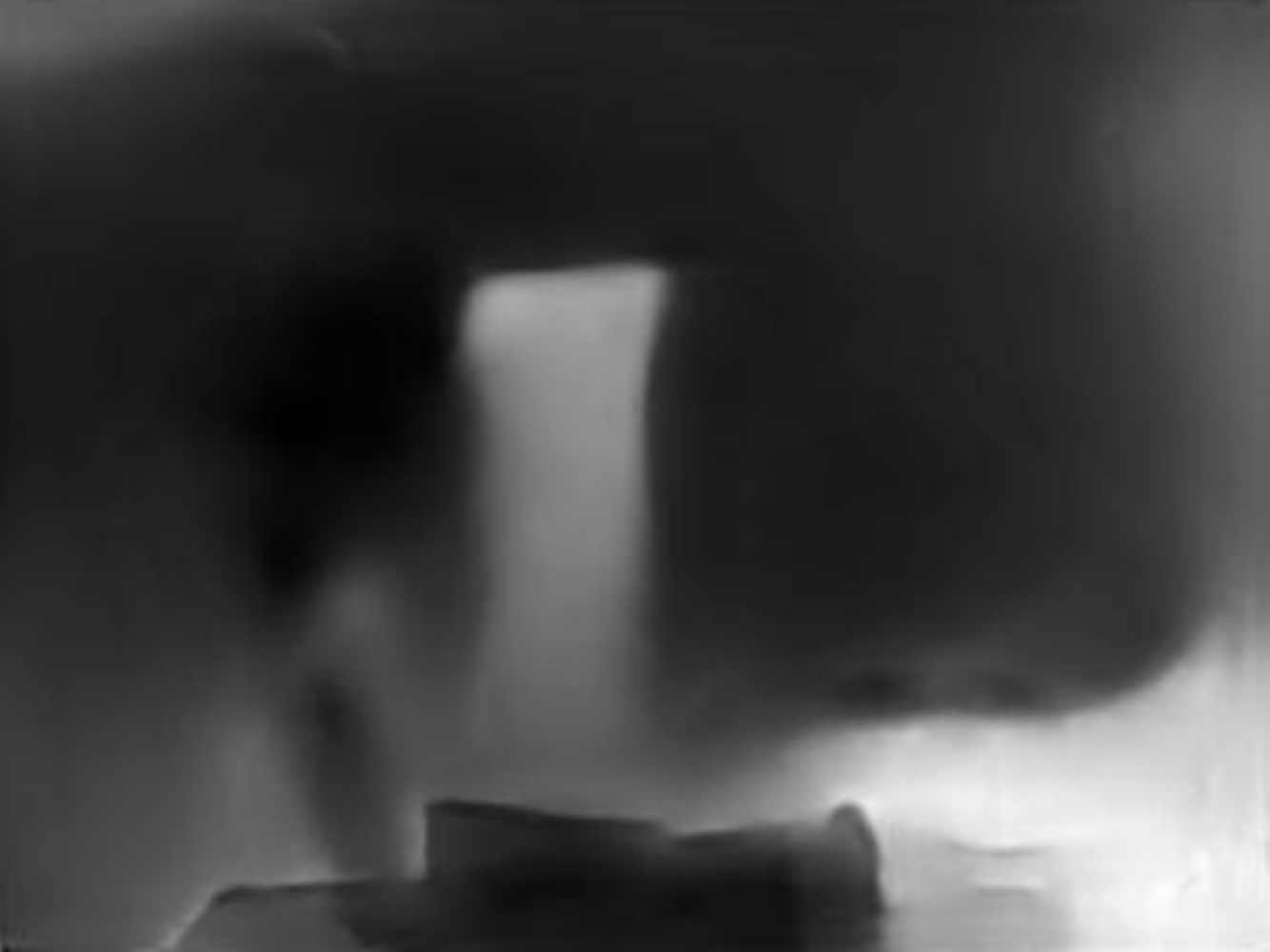}&
    \includegraphics[width=0.135\linewidth]{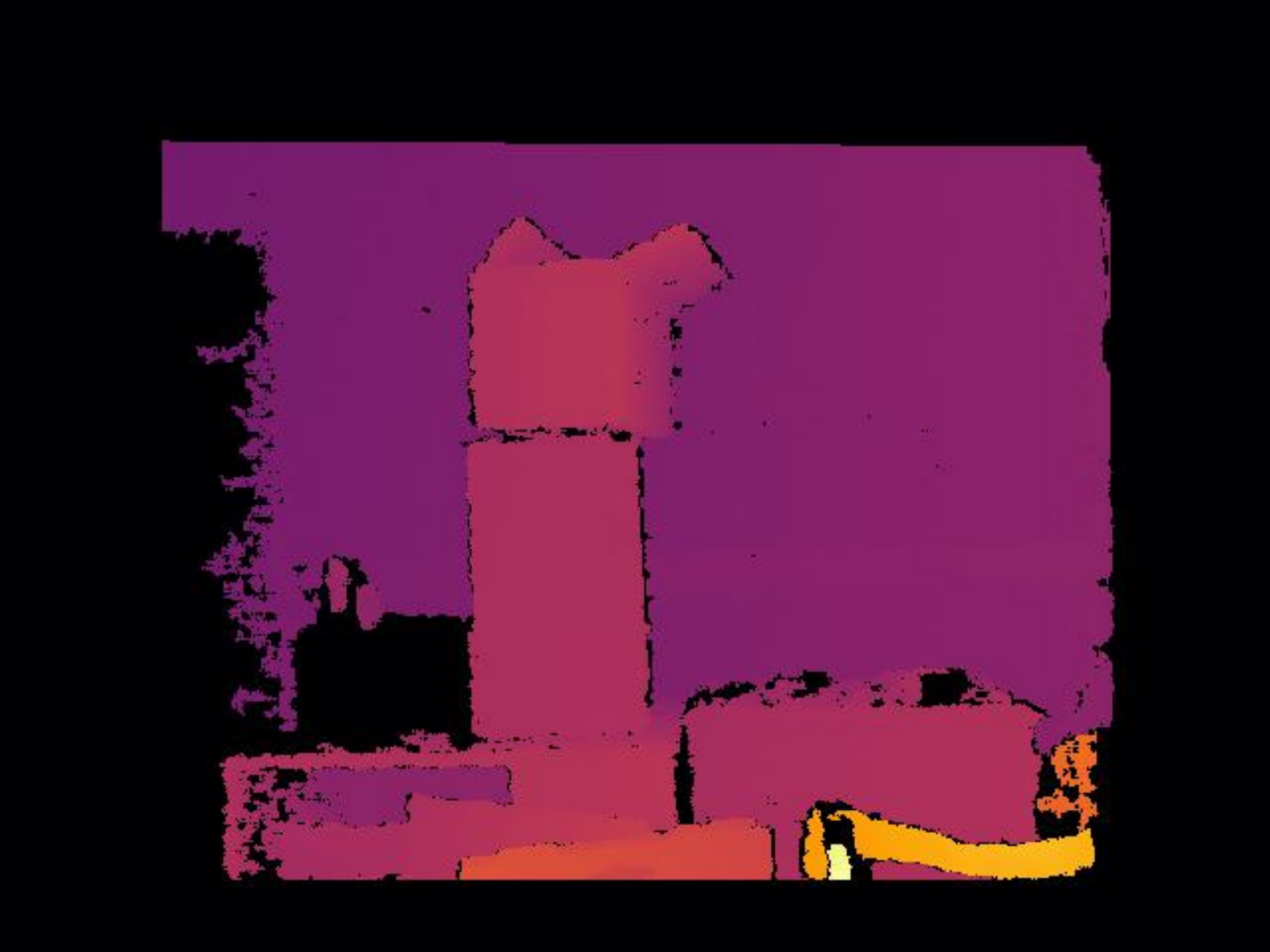}&
    \includegraphics[width=0.135\linewidth]{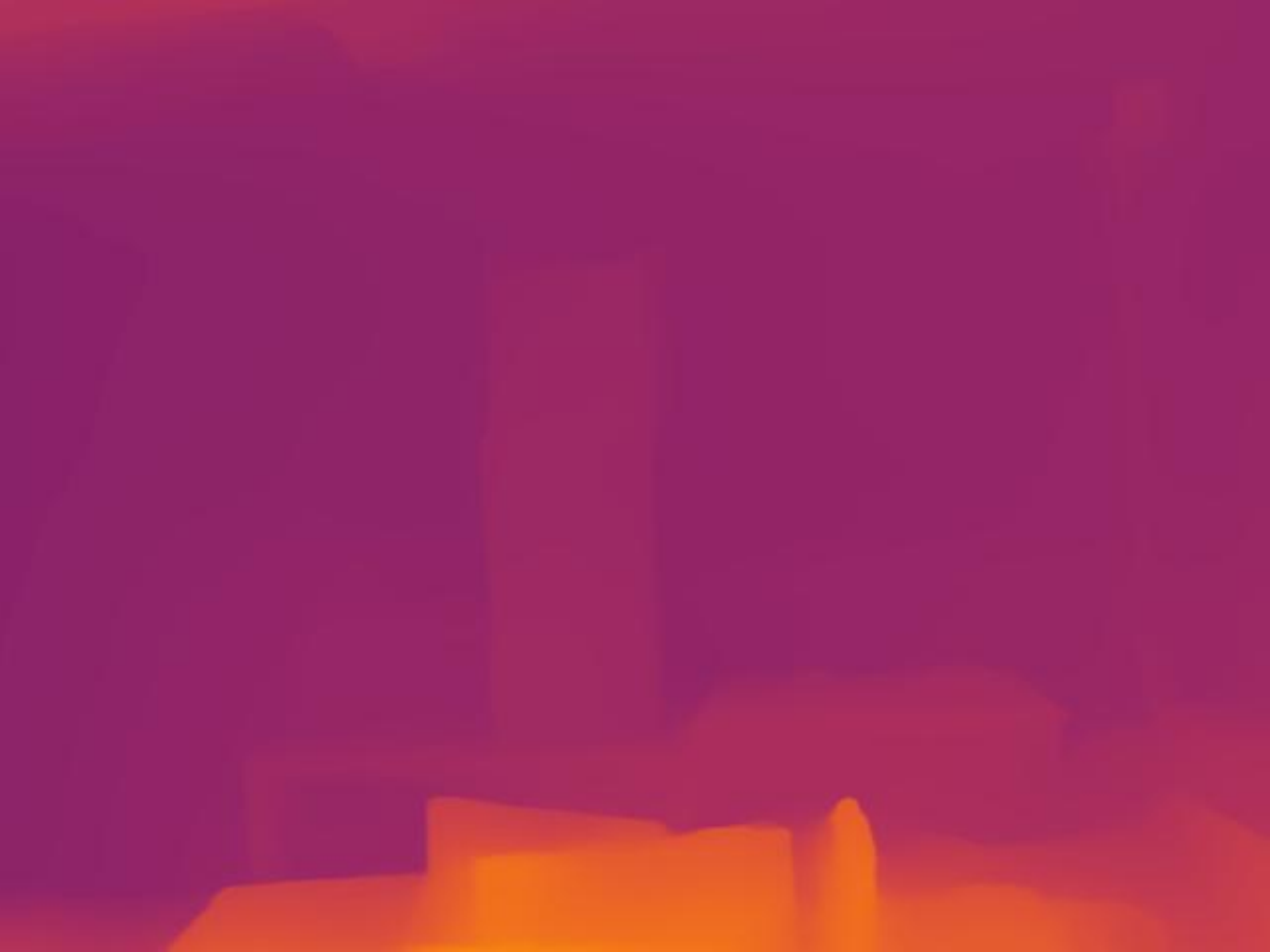}&
    \includegraphics[width=0.135\linewidth]{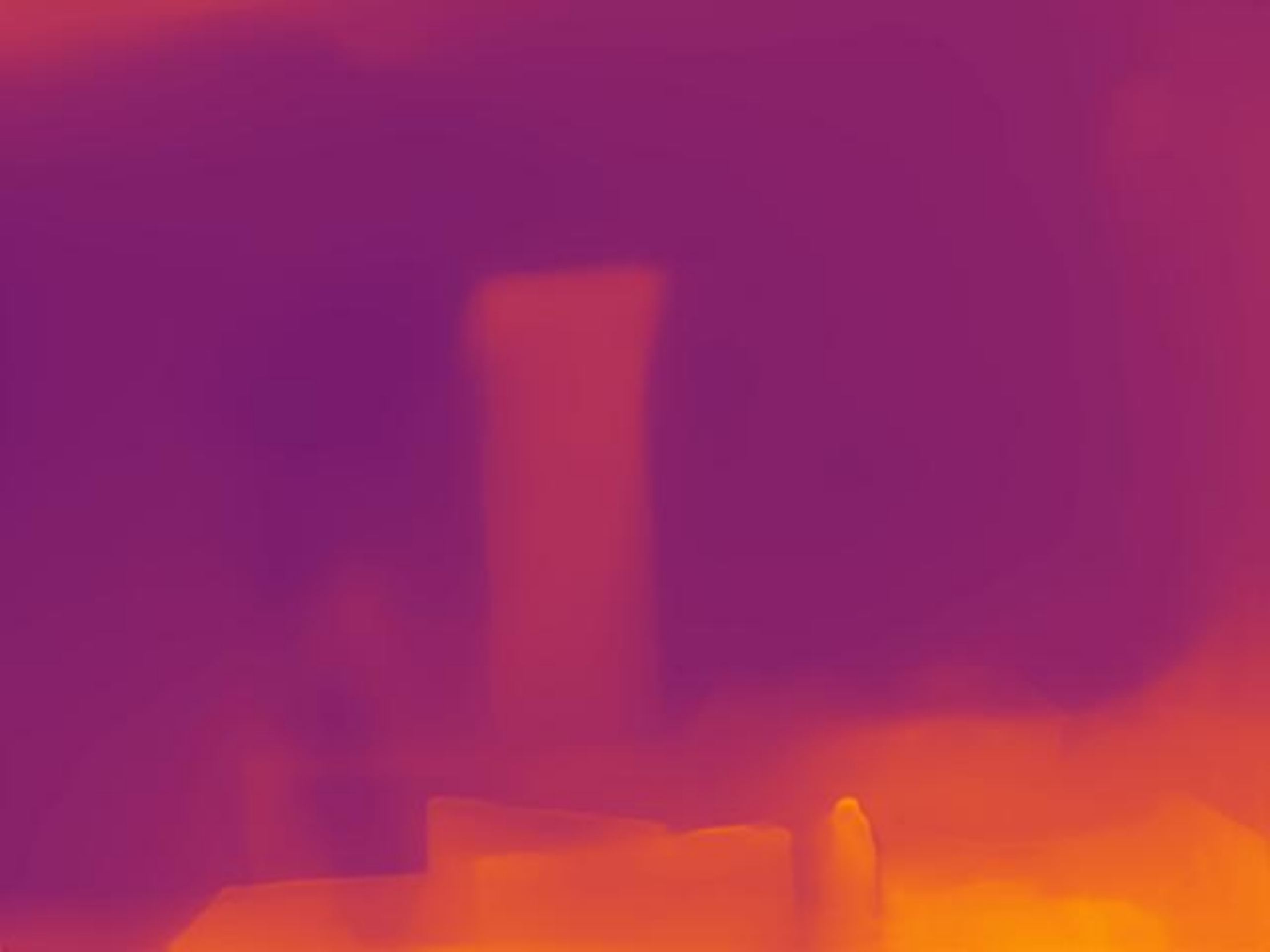}\\
    & \scriptsize 51 points & & & \multicolumn{3}{c}{\hspace{8pt}\scriptsize{iRMSE = 51.98 $\rightarrow$ 37.85}} \\
  \end{tabular}
  \vspace{-2pt}
  \caption{Our method tested on lab and corridor samples from the VCU-RVI dataset. RGB and sparse metric depth come from published rosbag data.}
  \label{fig:vcu_samples}
  \vspace{-12pt}
\end{figure}

We also test on rosbags from an entirely new dataset, VCU-RVI~\cite{vcurvi}. Figure~\ref{fig:vcu_samples} shows samples where available sparse metric depth is much lower in quantity than the 150+ points we have so far trained and tested with. Our pipeline still succeeds in resolving metric scale, with SML reducing depth error.

To demonstrate deployability, we benchmark performance on the NVIDIA Jetson AGX Orin platform and show a breakdown of component runtime in Table~\ref{tab:performance_tx2}. Measurements are averaged over 100 runs after 20 warmup runs. With acceleration via TensorRT, our depth alignment pipeline, in conjunction with a lightweight depth predictor like MiDaS-small, is viable for on-device metric depth estimation. Scale map scaffolding is one bottleneck as interpolation within the convex hull presently runs on the CPU. Data movement between the GPU and host CPU is another bottleneck that we expect can be reduced with additional engineering effort.

\begin{table}[h]
  \centering
  \footnotesize
  \caption{Runtime [ms] on Jetson AGX Orin in MAX-N mode. All pipeline variants are tested with 150 sparse metric depth points.}
  \label{tab:performance_tx2}
  \vspace{-4pt}
  \begin{tabular}{@{}
    l@{\hspace{4mm}}
    S[table-format=3.1]@{\hspace{2mm}}
    S[table-format=3.1]@{\hspace{2mm}}
    S[table-format=3.1]@{\hspace{2mm}}
    S[table-format=3.1]@{\hspace{0mm}}
    @{}}
    \toprule
    Depth predictor & {DPT-BEiT-L} & {DPT-H} & {MiDaS-s} & {MiDaS-s-TRT} \\
    Inference resolution & {384$\times$384} & {384$\times$384} & {256$\times$256} & {256$\times$256}\\
    \midrule
    Depth inference         & 144.8     & 53.9      & 29.2      & 1.5   \\
    D2H copy depth map      & 12.8      & 18.4      & 0.6       & 5.2   \\
    Global alignment        & 2.6       & 2.5       & 1.3       & 1.3   \\
    Scale map scaffolding   & 12.2      & 12.1      & 6.7       & 6.6   \\
    H2D copy SML inputs     & 3.3       & 3.3       & 2.4       & 2.2   \\
    SML-TRT inference       & 2.2       & 2.2       & 1.7       & 1.7   \\
    \midrule
    Total                   & 177.9     & 92.5      & 41.9      & 18.5  \\
    \bottomrule
  \end{tabular}
  \vspace{-6pt}
\end{table}

\begin{table*}[ht]
  \centering
  \footnotesize
  \caption{Experiments with input and regressed modalities in SML. Lower is better for all metrics.}
  \label{tab:ablations}
  \vspace{-4pt}
  \begin{tabular}{@{}
  *{6}{c@{\hspace{2mm}}}|
  *{2}{c@{\hspace{2mm}}}|
  *{2}{S[table-format=2.2]@{\hspace{3mm}}}
  *{1}{S[table-format=1.3]@{\hspace{3mm}}}|
  *{2}{S[table-format=3.2]@{\hspace{3mm}}}
  *{1}{S[table-format=1.3]@{\hspace{0mm}}}
  @{}}
    \toprule
    \multicolumn{6}{c|}{Input Modality Combinations} 
    & \multicolumn{2}{c|}{Regressing} 
    & \multicolumn{3}{c|}{On TartanAir} 
    & \multicolumn{3}{c}{On VOID (zero-shot)} \\
    {GA Depth}    & {Scale Scaff.}    & {Confidence}    & {Gradients}    & {Grayscale}     & {RGB}      & {Scale}    & {Shift}    &{iMAE} &{iRMSE}&{iAbsRel} & {iMAE} &{iRMSE} & {iAbsRel} \\
    \midrule
    \multicolumn{6}{@{}c|}{\textit{baseline (global alignment without SML)}} &       &            & 22.94 & 35.49 & 0.126    &  75.74 & 106.37 & 0.103 \\
    \midrule
    \checkmark & 		    & 			 &            & 		   & 			& \checkmark &            & 22.63 & 35.30 & 0.125    & 111.55 & 159.83 & 0.212 \\
    \checkmark &            & \checkmark & 			  & 		   & 			& \checkmark &            & 22.51 & 35.09 & 0.124    & 122.77 & 179.09 & 0.238 \\
    \checkmark & \checkmark &            & 			  & 		   & 			& \checkmark &            & \bfseries 16.11 & \bfseries 29.48 & \myuline{0.093} & \bfseries 45.55 & \bfseries 74.28 & \bfseries 0.062 \\
    \checkmark & \checkmark & \checkmark & 			  & 		   & 			& \checkmark &            & 16.90 & 30.63 & 0.094    &  63.55 &  94.78 & 0.092 \\
    \checkmark & \checkmark & \checkmark & \checkmark &            & 			& \checkmark &            & 17.07 & 30.14 & 0.098    &  \myuline{50.19} & \myuline{79.88} & \myuline{0.069} \\
    \checkmark & \checkmark & \checkmark &            & \checkmark & 			& \checkmark &            & 16.87 & 30.15 & 0.096    &  57.25 &  87.83 & 0.083 \\
    \checkmark & \checkmark & \checkmark &            &            & \checkmark & \checkmark &            & 18.03 & 31.64 & 0.102    &  59.08 &  91.00 & 0.080 \\
    \midrule
    \checkmark & \checkmark &            & 			  & 		   & 			&            & \checkmark & 17.03 & 30.12 & 0.096    &  62.34 &  90.91 & 0.086 \\
    \checkmark & \checkmark &            & 			  & 		   & 			& \checkmark & \checkmark & \myuline{16.28} & \myuline{29.53} & \bfseries 0.092 &  50.95 &  80.96 & 0.071 \\
    \bottomrule
  \end{tabular}
  \vspace{-12pt}
\end{table*}

\subsection{Ablations and Analysis}

We experiment with a number of input and regressed data modalities when designing the SML network.

\mypara{Input data modalities.} SML takes in two inputs: \textit{globally-aligned inverse depth} $\tilde{\mathbf{z}}$ and a \textit{scale map scaffolding}. We experiment with four additional inputs: (1) a \textit{confidence map} derived from a binary map pinpointing known sparse depth locations, first dilated with a $7\timess7$ circular kernel and then blurred with a $5\timess5$ Gaussian kernel to mimic confidence spread around a fixed known point; (2) a \textit{gradient map} computed using the Scharr operator; (3) a \textit{grayscale} conversion of the original RGB image; (4) and the RGB image itself. Inputs are concatenated in the channel dimension and fed into SML as a single tensor. Table \ref{tab:ablations} reports the impact of different input combinations on the metric accuracy of SML depth.

Globally-aligned depth alone is not sufficient for the network to learn dense scale regression well. An input scale map scaffolding is necessary. Conceptually, this acts as an initial guess at the dense scale map that the network is learning to regress. Without an accompanying scale map input, the confidence map negligibly improves SML learning; however, using both slightly underperforms compared to using only scale scaffolding. This is surprising, as the confidence map is meant to signal which regions in the input depth and scale scaffolding are more trustworthy. It may be that our representation of confidence is not being parsed well by SML, or that the scale map scaffolding encodes similar information, e.g., boundaries of the convex hull and approximate positions of interpolation anchors corresponding to known sparse metric depth. Incorporating edge representations in the form of gradient maps, grayscale, or RGB images, does not appear to be beneficial. This can be partly attributed to the high quality of depth predictions output by DPT, as those depth maps already exhibit clear edges and surfaces. RGB input actually worsens performance, implying that color cues are not particularly useful in dense metric scale regression.

Since we are also interested in cross-dataset transfer, we evaluate zero-shot performance of every input combination on VOID and report the results in Table \ref{tab:ablations}. Combined depth and scale scaffolding result in noticeably lower error; we therefore select this input combination for SML. 

\vspace{12pt}
\mypara{Regressing scale and shift.} SML learns dense (per-pixel) scale factors by which to multiply input depth estimates $\tilde{\mathbf{z}}$, such that the output depth $\hat{\mathbf{z}}$ achieves higher metric accuracy. The network is allowed to regress negative values as scale residuals $\mathbf{r}$, such that the output depth is $\hat{\mathbf{z}}=\text{ReLU}(1+\mathbf{r})\tilde{\mathbf{z}}$.
Our design choice to regress scale is motivated by scale factors having a more intuitive interpretation in projective geometry. Scaling a depth value at a pixel location can be interpreted as zooming in (pulling closer) or zooming out (pushing further) the object at that location in 3D space. It is more difficult to intuit the impact of shifting depth at individual pixels. We conduct two experiments that involve shift, listed in the bottom two rows of Table \ref{tab:ablations}. We regress only dense shift $\mathbf{t}$, such that the output prediction $\hat{\mathbf{z}}=\tilde{\mathbf{z}}+\mathbf{t}$. We also regress shift $\mathbf{t}$ alongside scale residuals $\mathbf{r}$, where $\hat{\mathbf{z}}=\text{ReLU}(1+\mathbf{r})\tilde{\mathbf{z}}+\mathbf{t}$. For the latter, we add a second output head to the SML network, while the encoder and decoder layers remain common to both regression tasks. When training with shift regression, our default learning rate of $5\timess10^{-4}$ prohibits loss convergence and necessitates a slightly lower one of $4\timess10^{-4}$. Overall, regressing shift does not significantly impact performance on TartanAir, and zero-shot testing on VOID indicates that regressing scale only is the most robust choice for cross-dataset transfer.
\section{Conclusion}

Combining metric accuracy and high generalizability is a key challenge in learning-based depth estimation. We propose incorporating inertial data into the visual depth estimation pipeline---not through sparse-to-dense depth completion, but rather through dense-to-dense depth alignment using estimated and learned scale factors. Inertial measurements inform and propagate metric scale through global and local depth alignment. We show improved error reduction with learning-based local alignment over least-squares global alignment alone, and demonstrate successful zero-shot cross-dataset transfer from synthetic training data to real-world test data. Our modular approach supports direct integration of existing and future monocular depth estimation and visual-inertial odometry systems. It succeeds in resolving metric scale for metrically-ambiguous monocular depth estimates, and we hope that it will assist in the deployment of robust and general monocular depth estimation models.

\bibliographystyle{IEEEtran}
\bibliography{paper}

\clearpage
\begin{appendices}

\section{Network Architectures}

In our proposed visual-inertial depth estimation pipeline, the ScaleMapLearner (SML) network that performs dense scale alignment on globally-aligned metric depth maps is based on the MiDaS-small architecture. This is one of the more mobile-friendly models within the robust and generalizable MiDaS\cite{Ranftl2020} family of monocular depth estimation models. 

\begin{figure*}[!htb]
  \centering
  \includegraphics[width=1.0\linewidth]{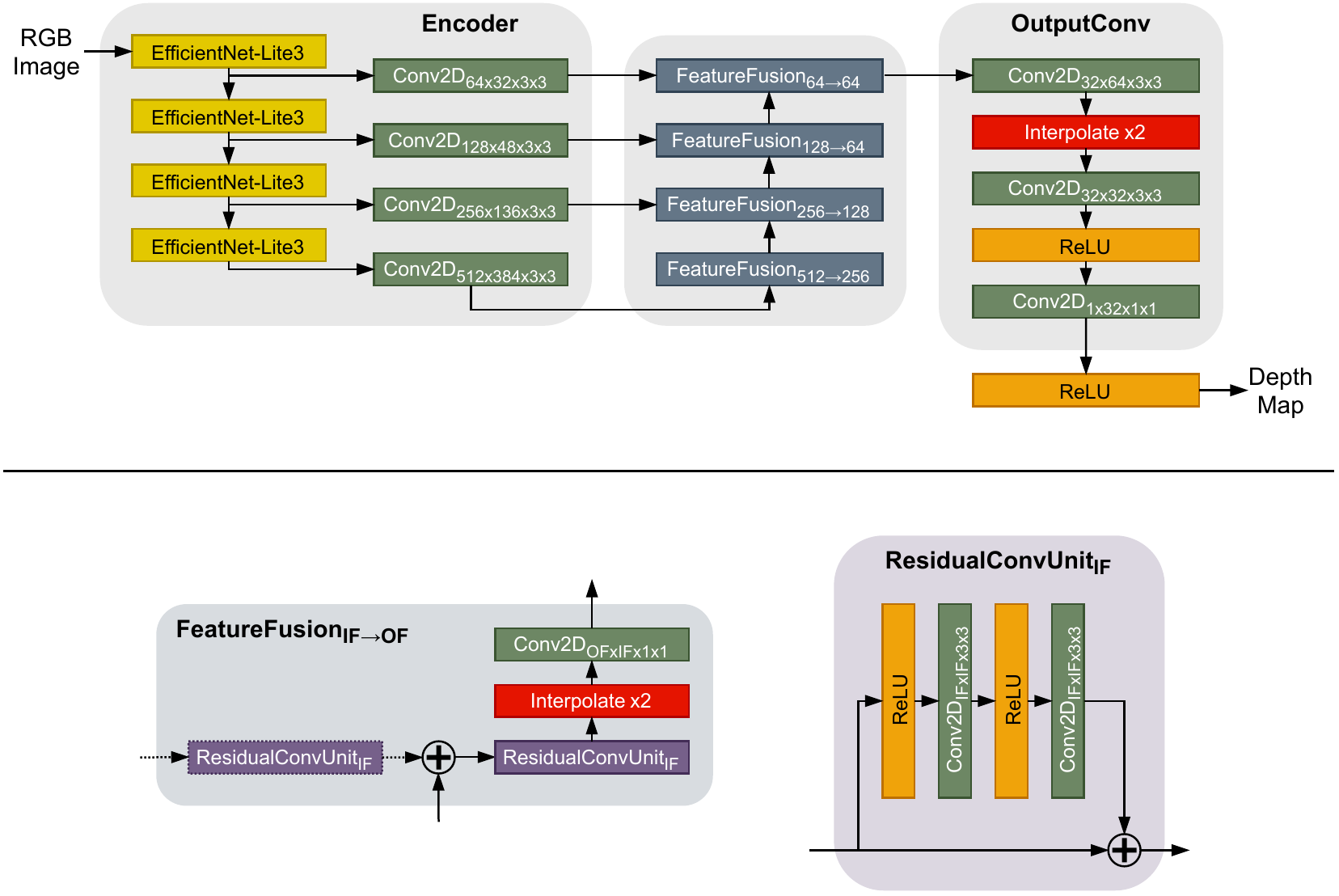}
  \caption{\textit{Top:} MiDaS-small\cite{Ranftl2020} architecture, designed for monocular depth estimation. \textit{Bottom:} Diagrams of the FeatureFusion block structure and the ResidualConvUnit used within it. These blocks are parametrized by the number of input (IF) and output (OF) features.}
  \vspace{18pt}
  \label{fig:midas-small}
\end{figure*}

\mypara{MiDaS-small.} Figure~\ref{fig:midas-small} shows an architecture diagram for the MiDaS-small network. The encoder incorporates an EfficientNet-Lite3~\cite{Tan2019EfficientNetRM} backbone, with skip connections propagating out features at four levels. The decoder consists of four FeatureFusion blocks that progressively upsample and merge features from the encoder and the skip connections.

\begin{figure*}[!htb]
  \centering
  \includegraphics[width=1.0\linewidth]{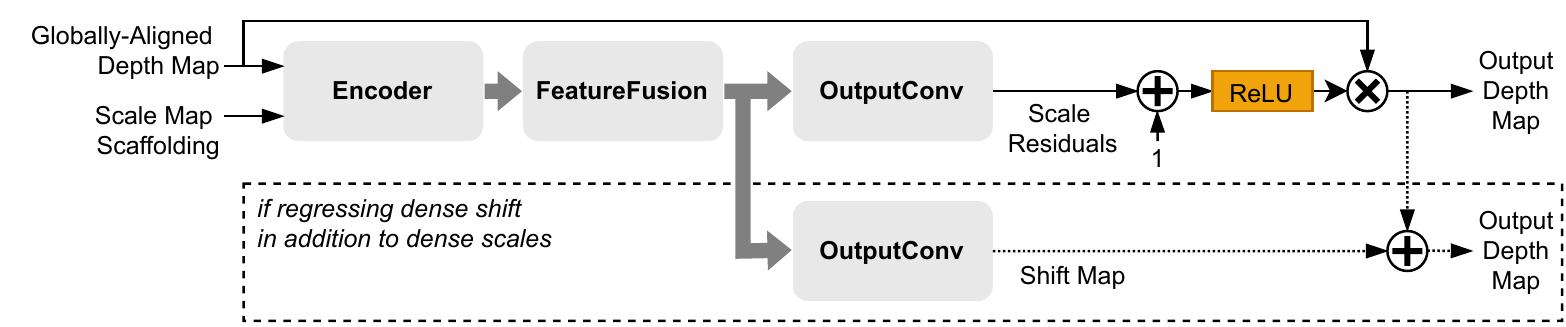}
  \caption{Our ScaleMapLearner (SML) network using MiDaS-small architecture blocks.}
  \vspace{18pt}
  \label{fig:sml-network}
\end{figure*}

\mypara{ScaleMapLearner (SML).} Figure~\ref{fig:sml-network} shows the SML architecture using MiDaS-small blocks. While MiDaS-small outputs affine-invariant depth maps, our SML network outputs metric depth maps. By default, SML regresses only scale residuals with a single OutputConv head. For ablation experiments where we regress dense shift in addition to scale residuals, a second identical OutputConv head is used; the encoder and feature fusion blocks remain common to both regression tasks.

\section{More on Sparsifiers}

In order to synthetically generate metric sparse depth for TartanAir, we use a sparsifier that samples ground truth at locations determined via feature tracking. This is a realistic sparsifier to consider, as it simulates how a real VIO system would be used to generate sparse metric depth. We target low densities of sparse points---around 150 points---in line with the quantity that would be tracked by the frontend of VINS-Mono. While it is possible to track higher densities, doing so would require more computation and would be less suitable for real-time applications.

\begin{figure}[!htb]
\centering
  \begin{tabular}{@{}*{3}{c@{\hspace{1mm}}}c@{}}
    \scriptsize{RGB Image} & \scriptsize{Sparse Depth Locations} & \scriptsize{Scale Map Scaffolding} \\
    \includegraphics[width=0.33\linewidth]{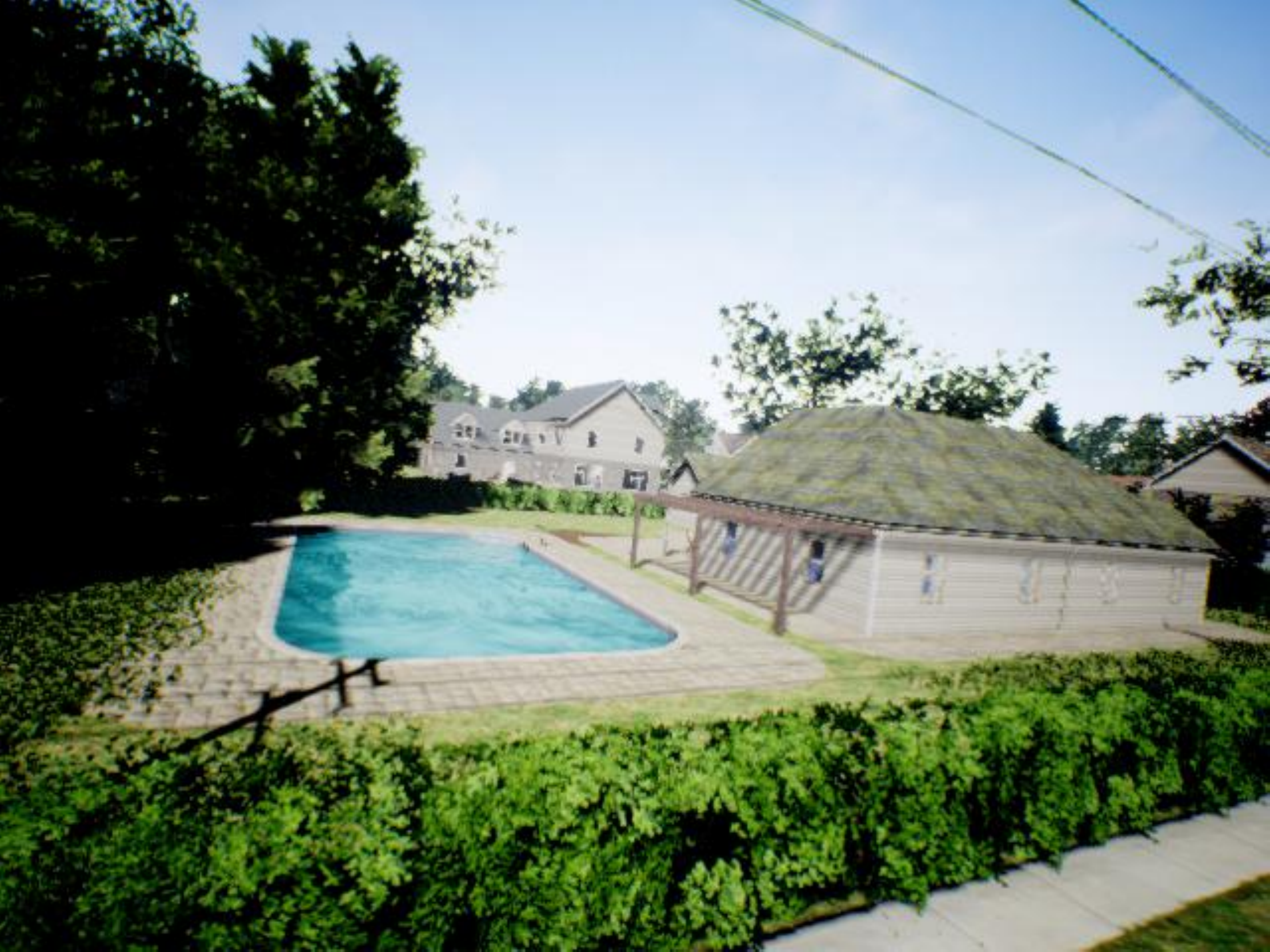}&
    \includegraphics[width=0.33\linewidth]{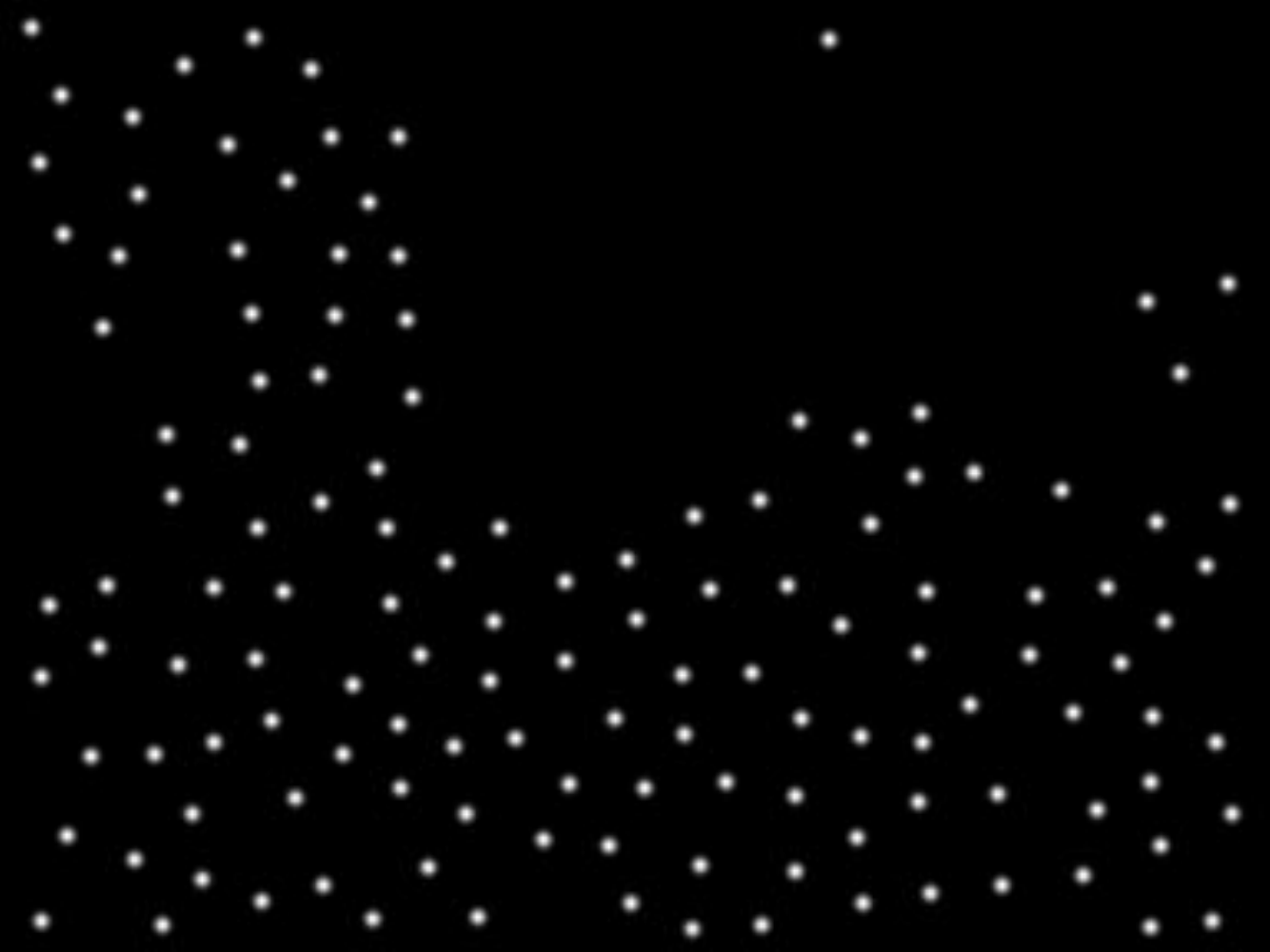}&
    \includegraphics[width=0.33\linewidth]{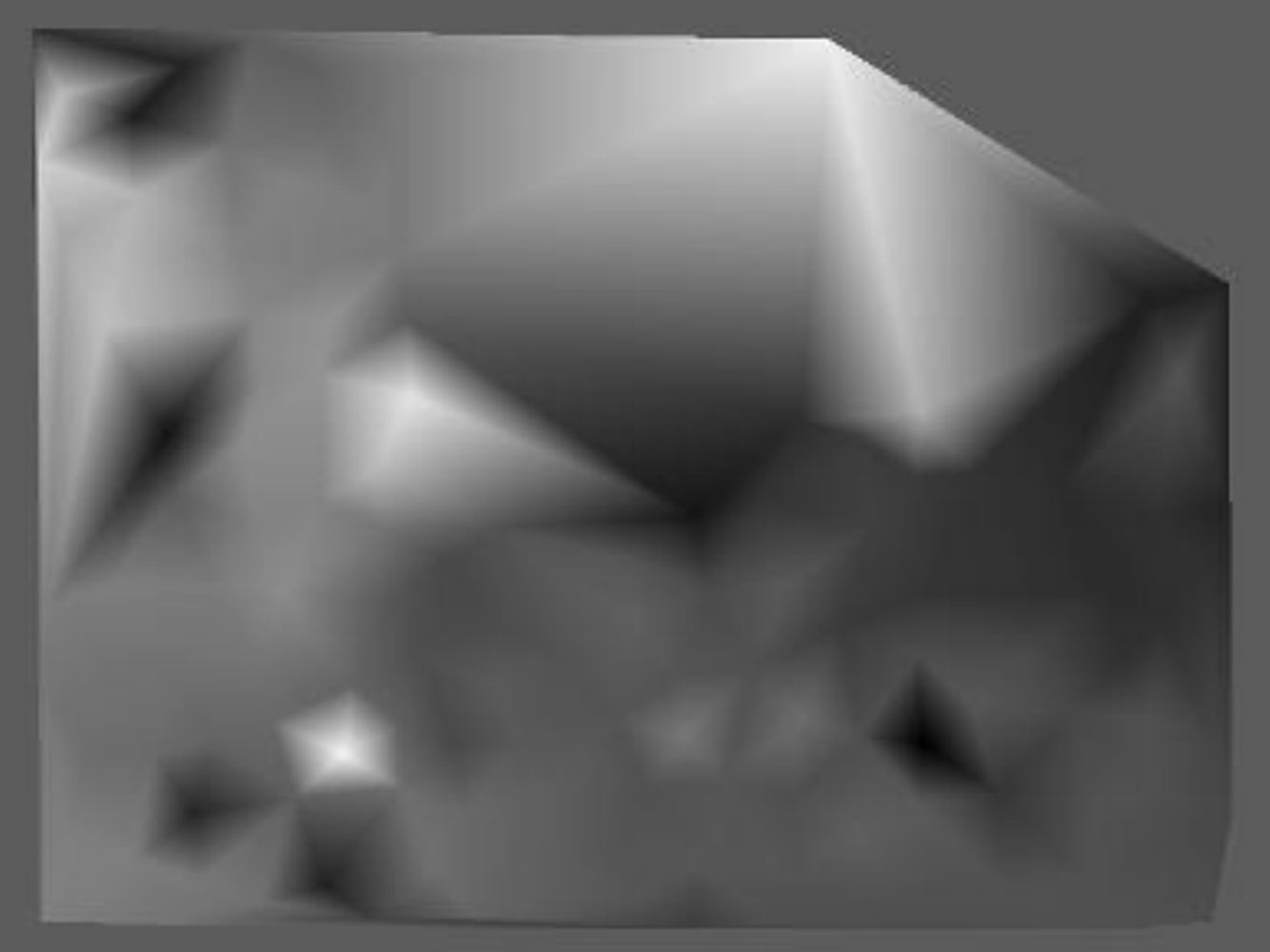}\\
    \includegraphics[width=0.33\linewidth]{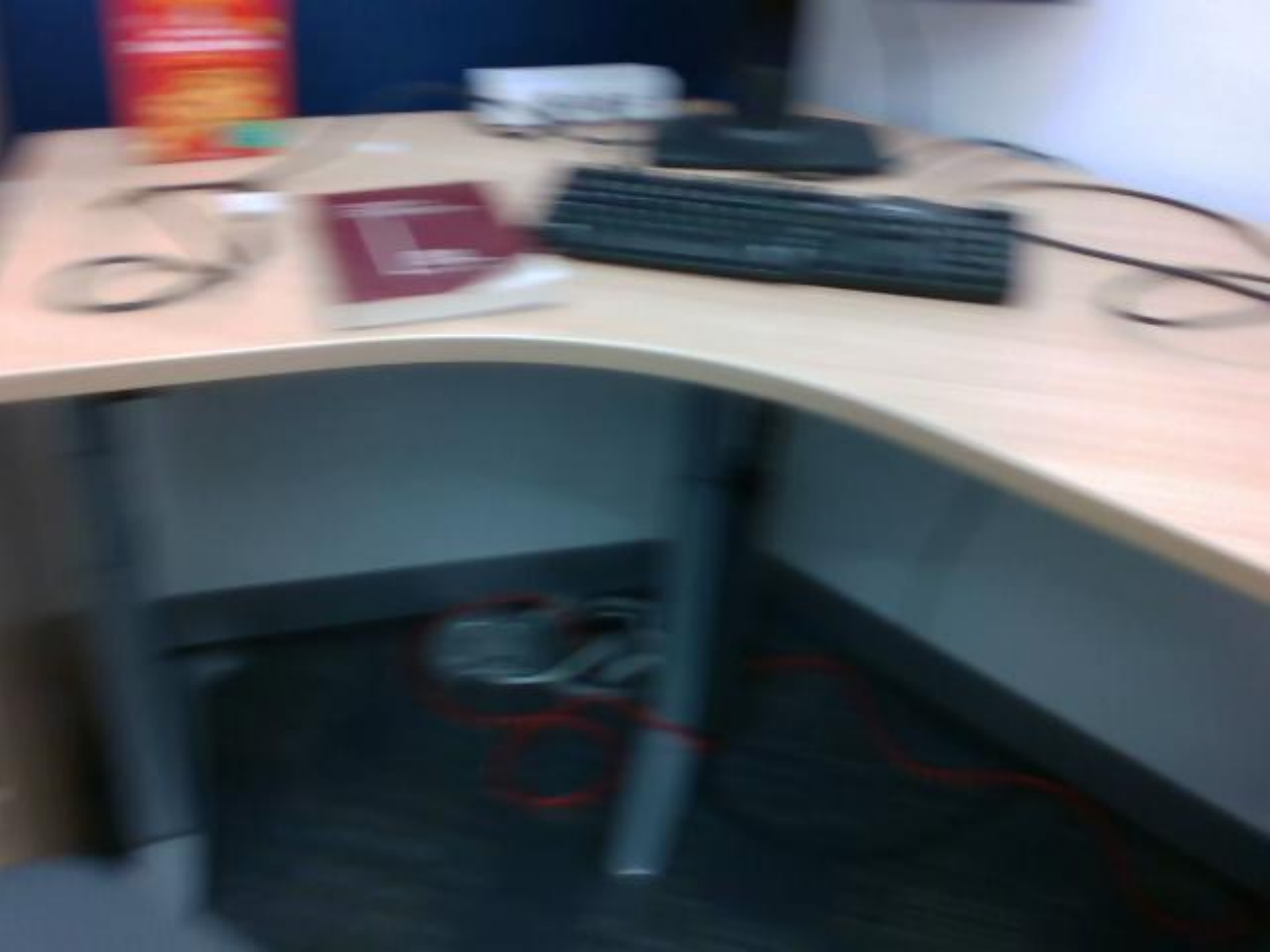}&
    \includegraphics[width=0.33\linewidth]{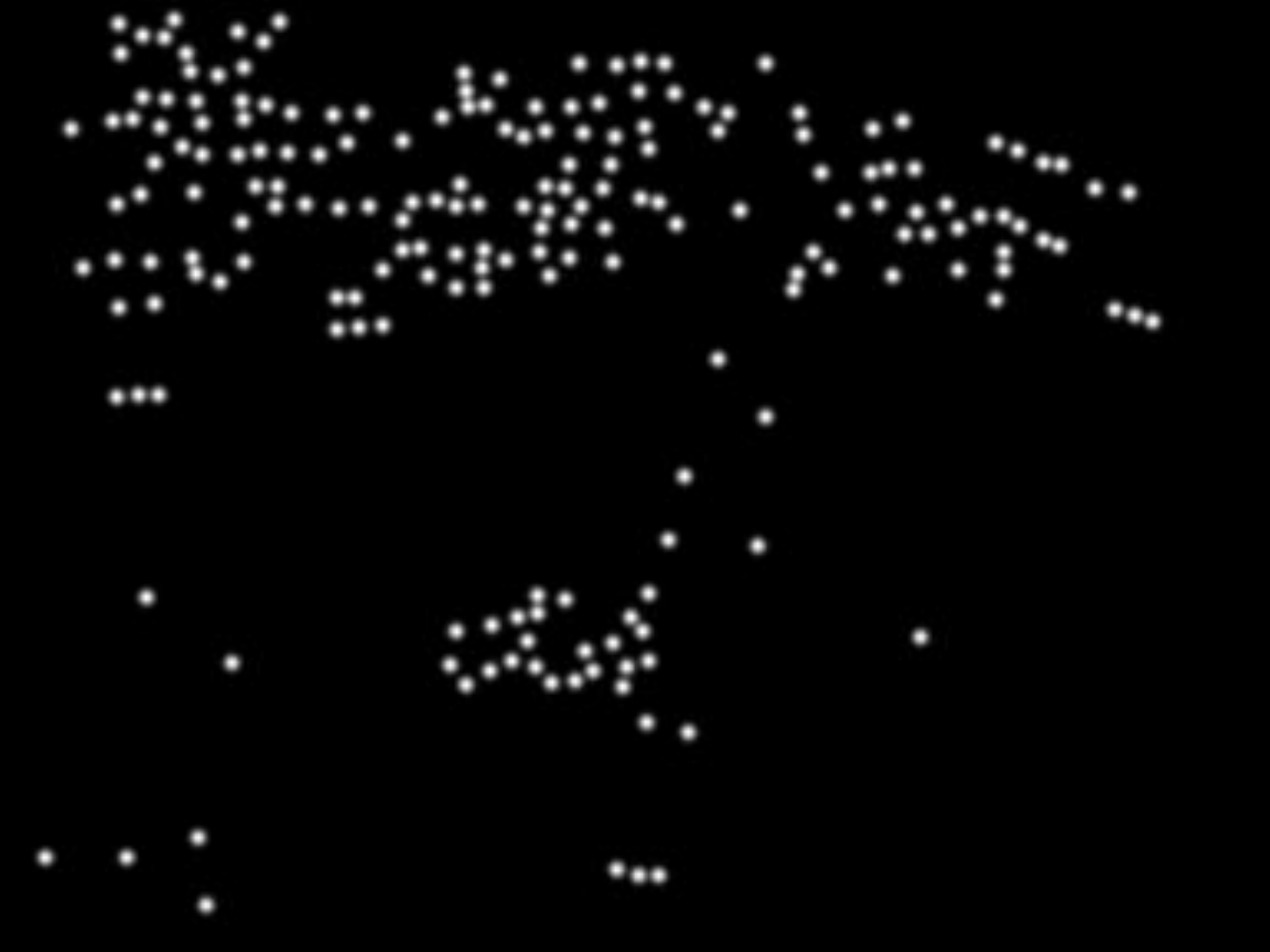}&
    \includegraphics[width=0.33\linewidth]{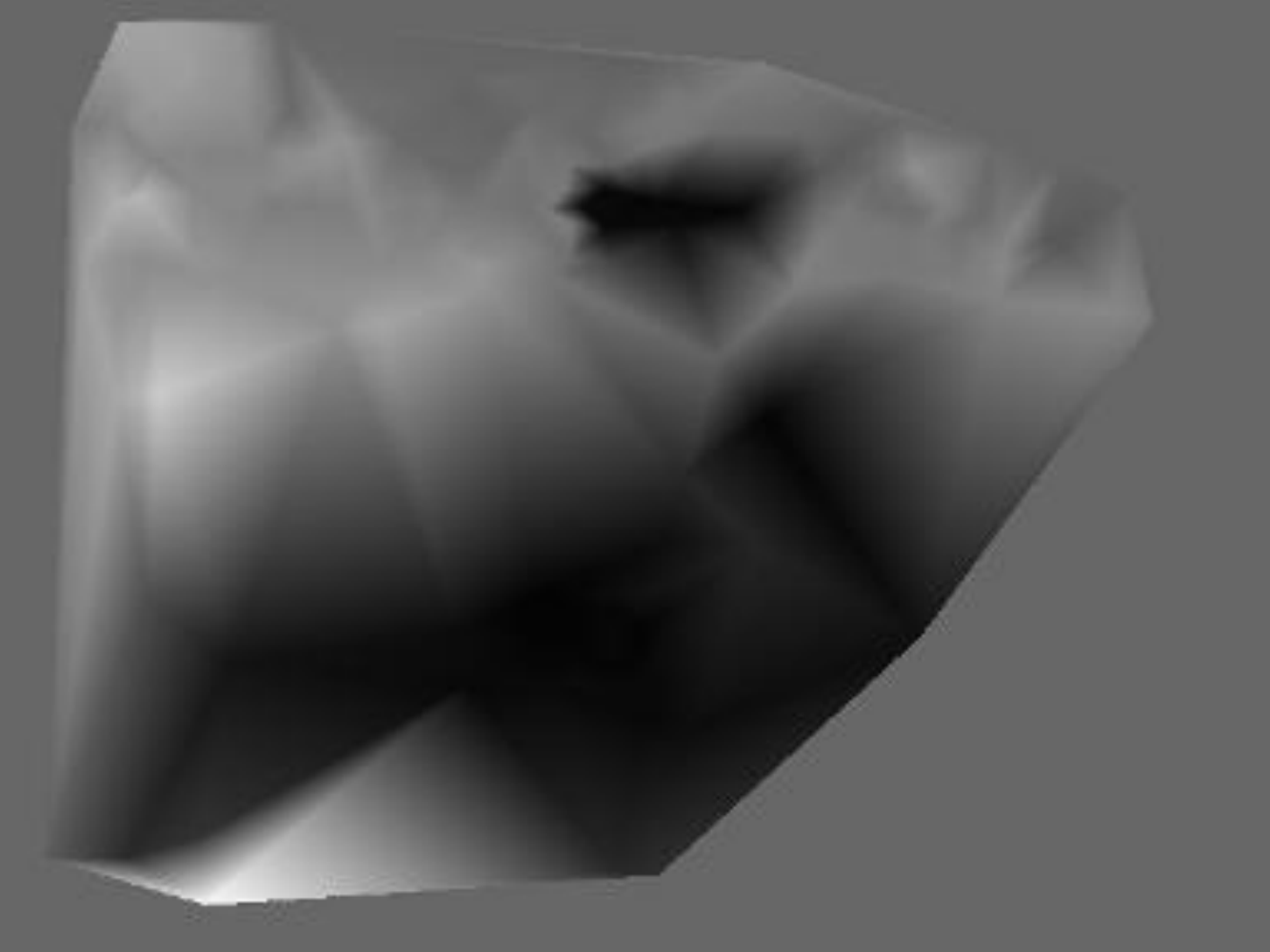}\\
  \end{tabular}
  \caption{Differences in sparsity and coverage of known metric sparse depth between TartanAir (top) and VOID (bottom) samples, shown alongside the interpolated scale map scaffolding. For TartanAir, sparse depth locations are determined through feature tracking via a VINS-Mono frontend. For VOID, sparse depth is obtained using XIVO~\cite{Fei2019xivo}.}
  \label{fig:sparsity_coverage}
\end{figure}

The feature tracker \cite{Lusk2018} we use with TartanAir enforces a minimum distance between landmarks, which leads to high coverage in the sparsity map given a sufficiently textured scene. In contrast, the sparse points provided in the VOID dataset tend to be more clustered, as illustrated in Figure \ref{fig:sparsity_coverage}. Clustering leads to regions of smoother interpolation within the convex hull of the scale map scaffolding, which makes learning dense scale easier and improves metric depth prediction. This is supported by our observation that experiments with SML see a greater reduction in error on VOID samples when compared to TartanAir samples: e.g., a 39\% reduction in iAbsRel on VOID versus a 26\% reduction in iAbsRel on TartanAir. Given these differences in sparsity patterns and coverage, our demonstration of successful zero-shot transfer from TartanAir to VOID is particularly impressive and highlights the robustness of our scale map learning approach.

\section{Expanded Evaluation and Visualizations}

\begin{figure*}[p]
\centering
  \begin{tabular}{@{}l@{\hspace{0.5mm}}*{9}{c@{\hspace{0.5mm}}}c@{}}
    & {\scriptsize RGB Image} & {\scriptsize Sparse Depth} & {\scriptsize Scales Scaffold.} & {\scriptsize Regressed Scales} & {\scriptsize GA Depth} & {\scriptsize SML Depth} & {\scriptsize Ground Truth} & {\scriptsize GA Error} & {\scriptsize SML Error}\\
    \vspace{-0.75mm}
    \scriptsize a. &
    \includegraphics[width=0.104\linewidth]{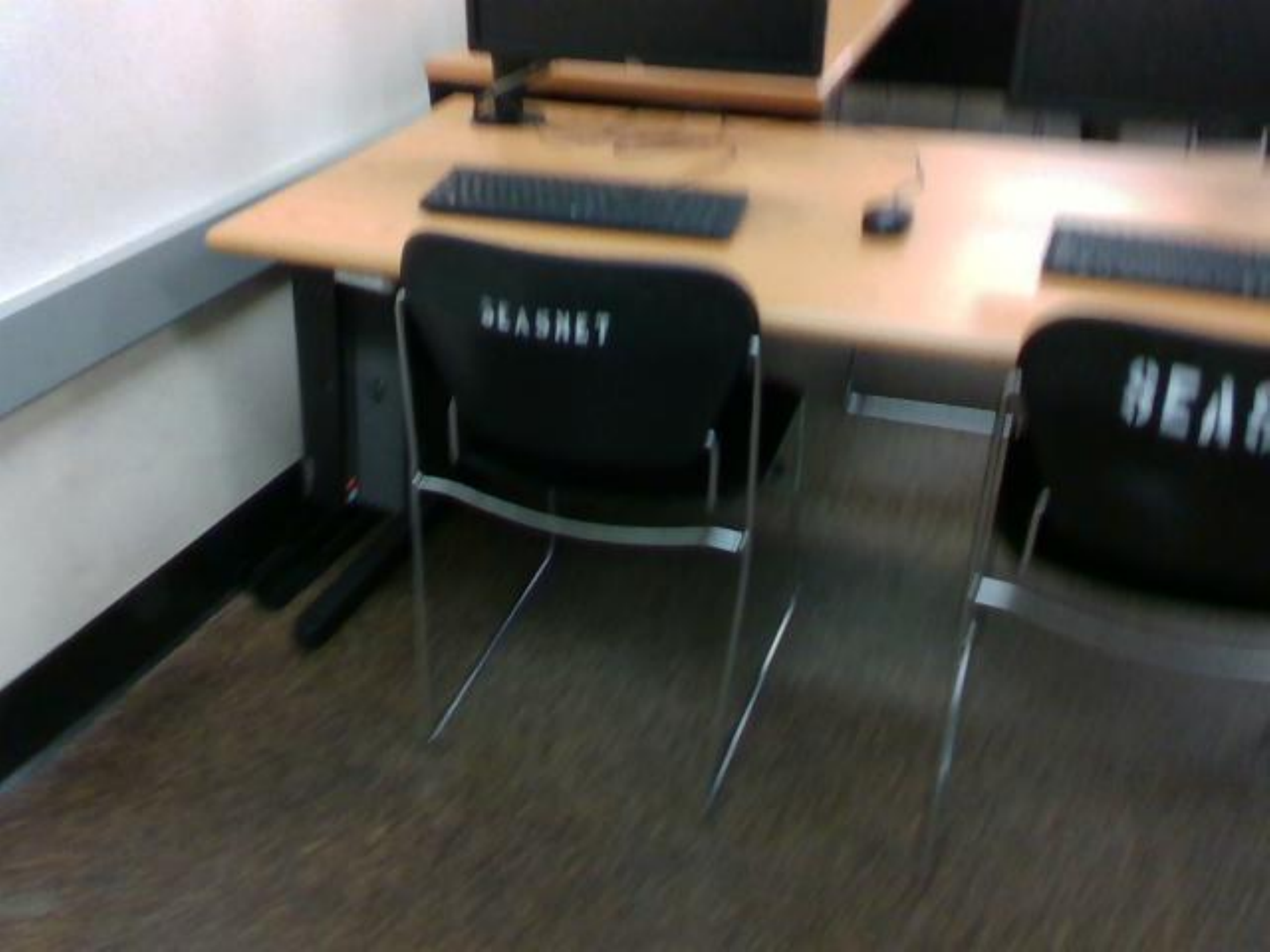}&
    \includegraphics[width=0.104\linewidth]{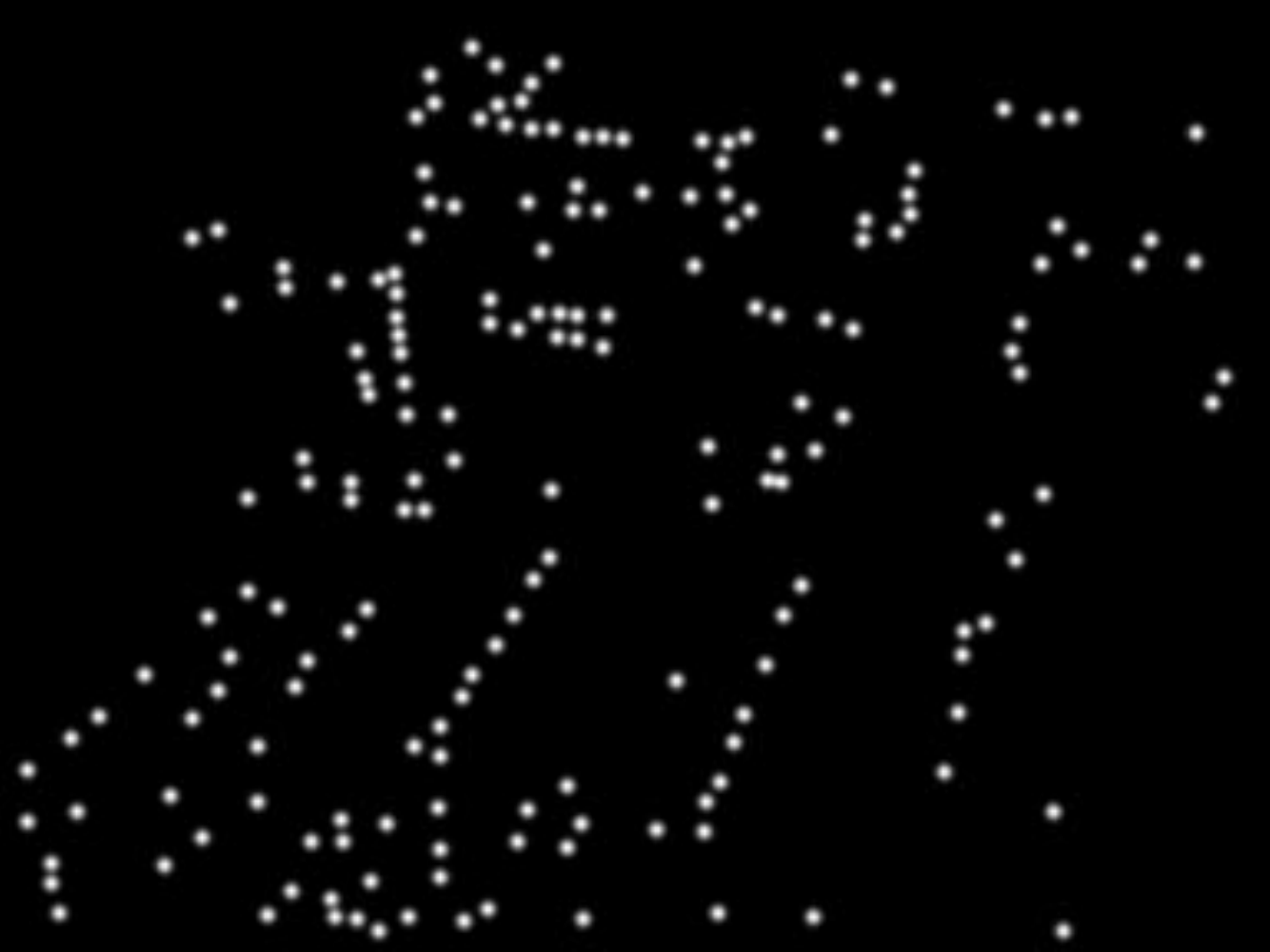}&
    \includegraphics[width=0.104\linewidth]{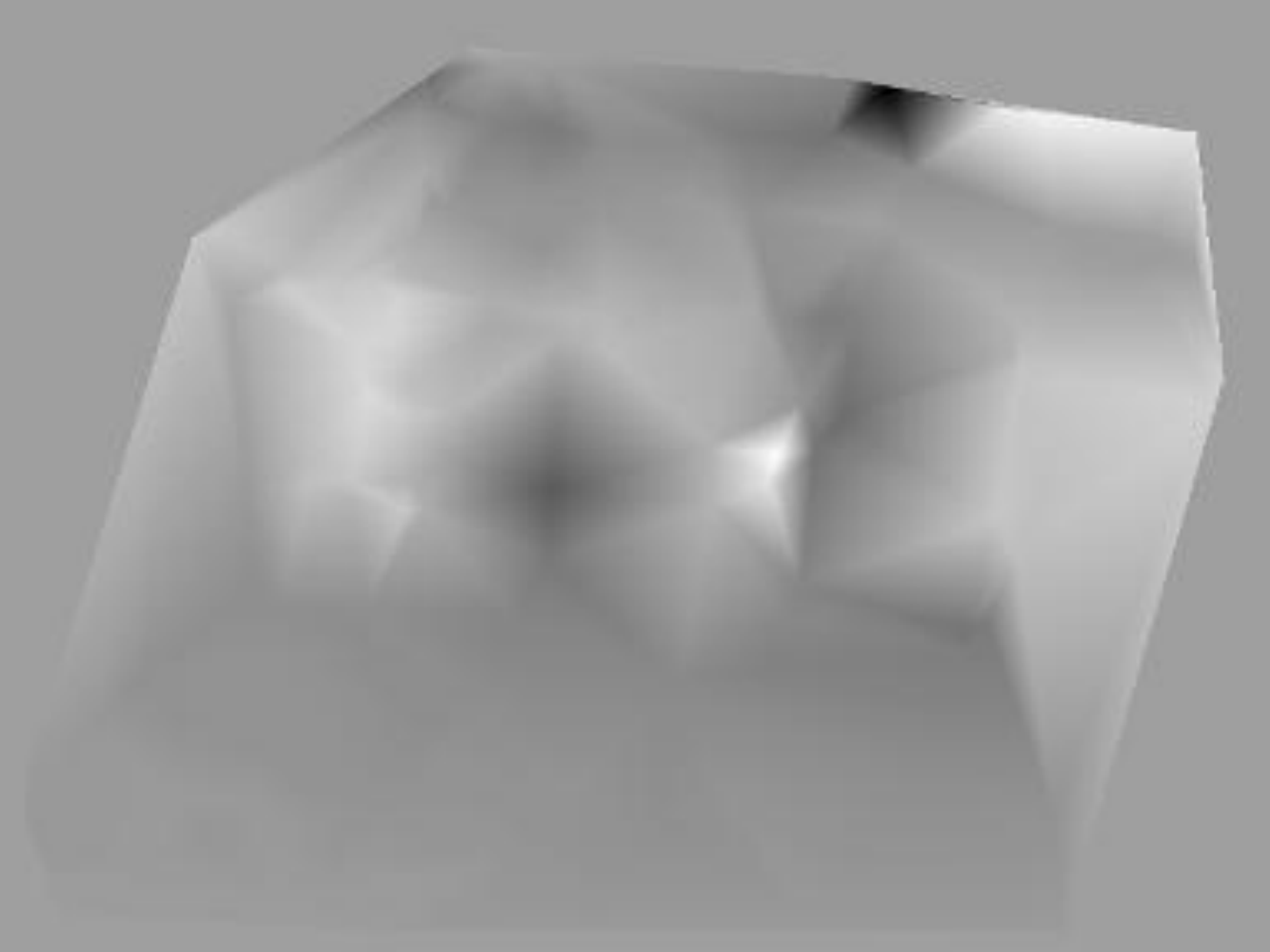}&
    \includegraphics[width=0.104\linewidth]{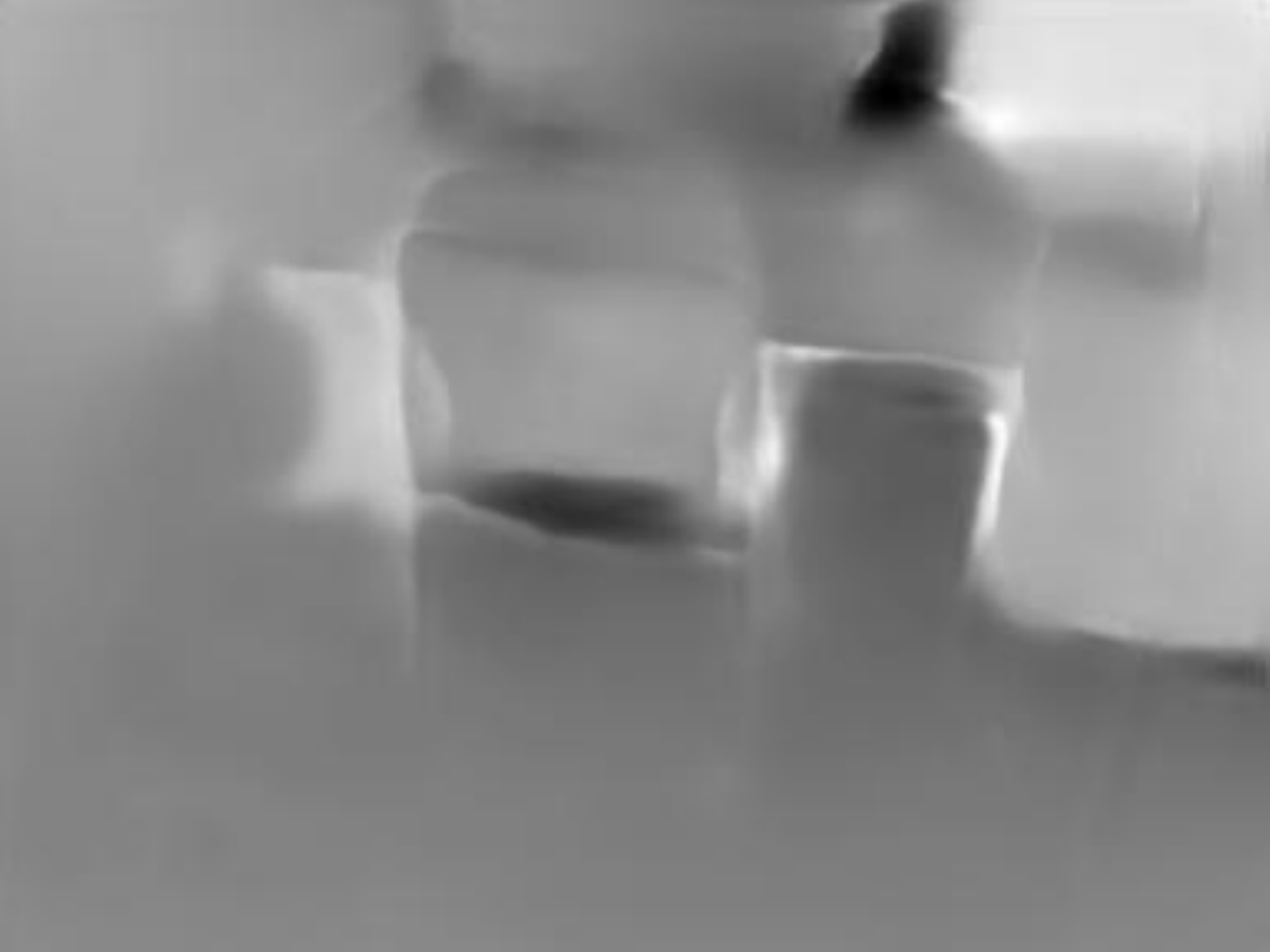}&
    \includegraphics[width=0.104\linewidth]{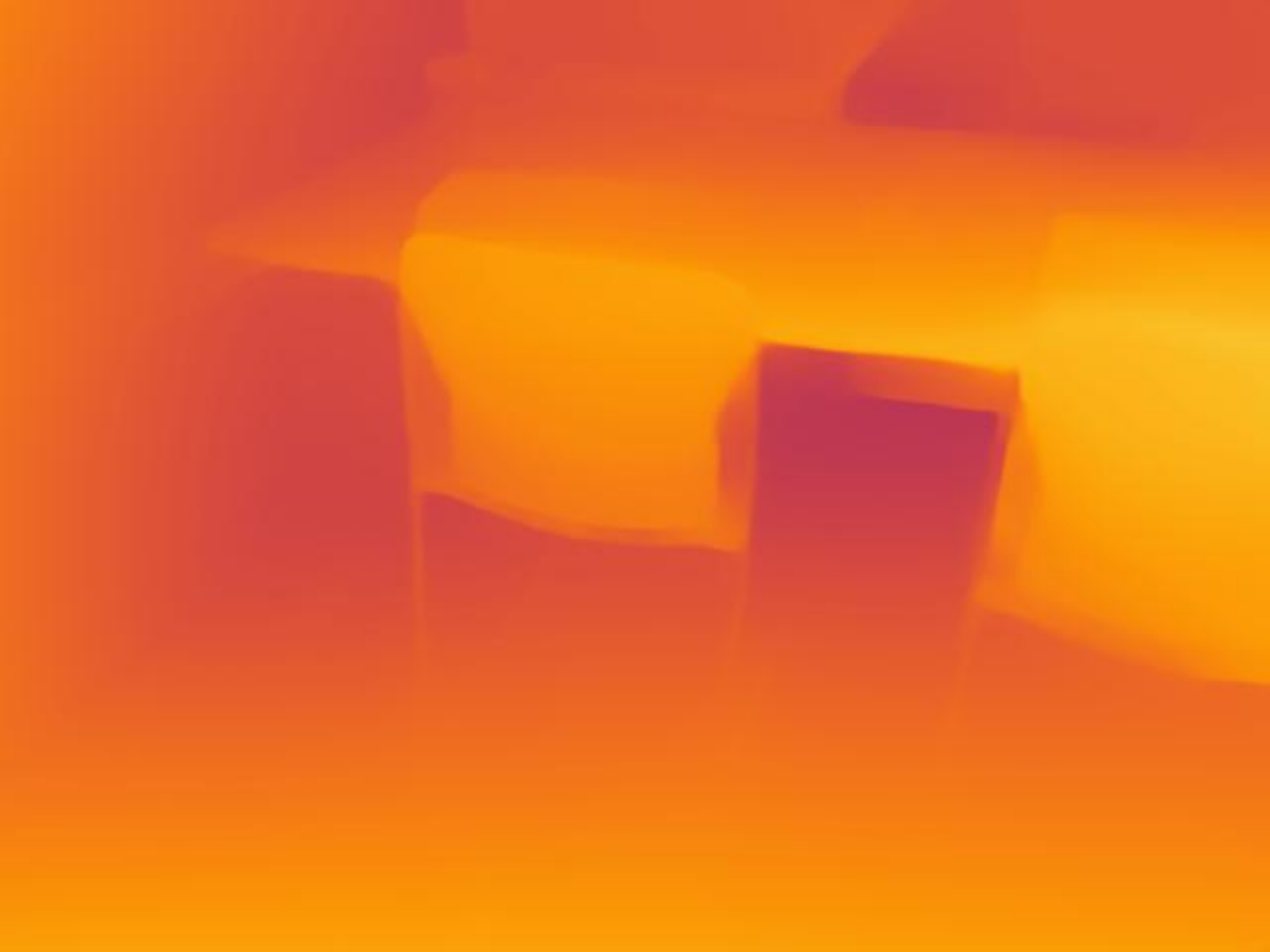}&
    \includegraphics[width=0.104\linewidth]{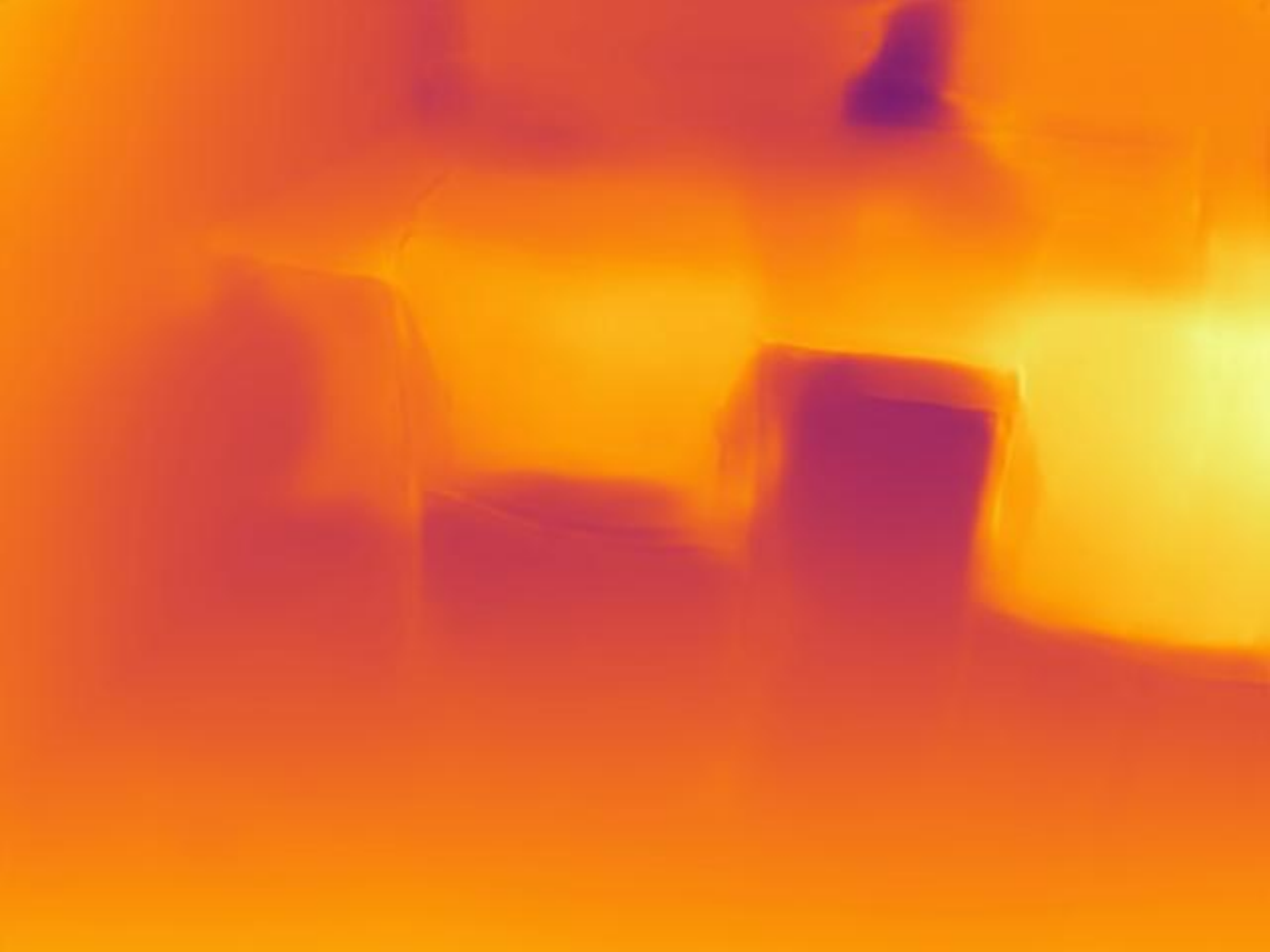}&
    \includegraphics[width=0.104\linewidth]{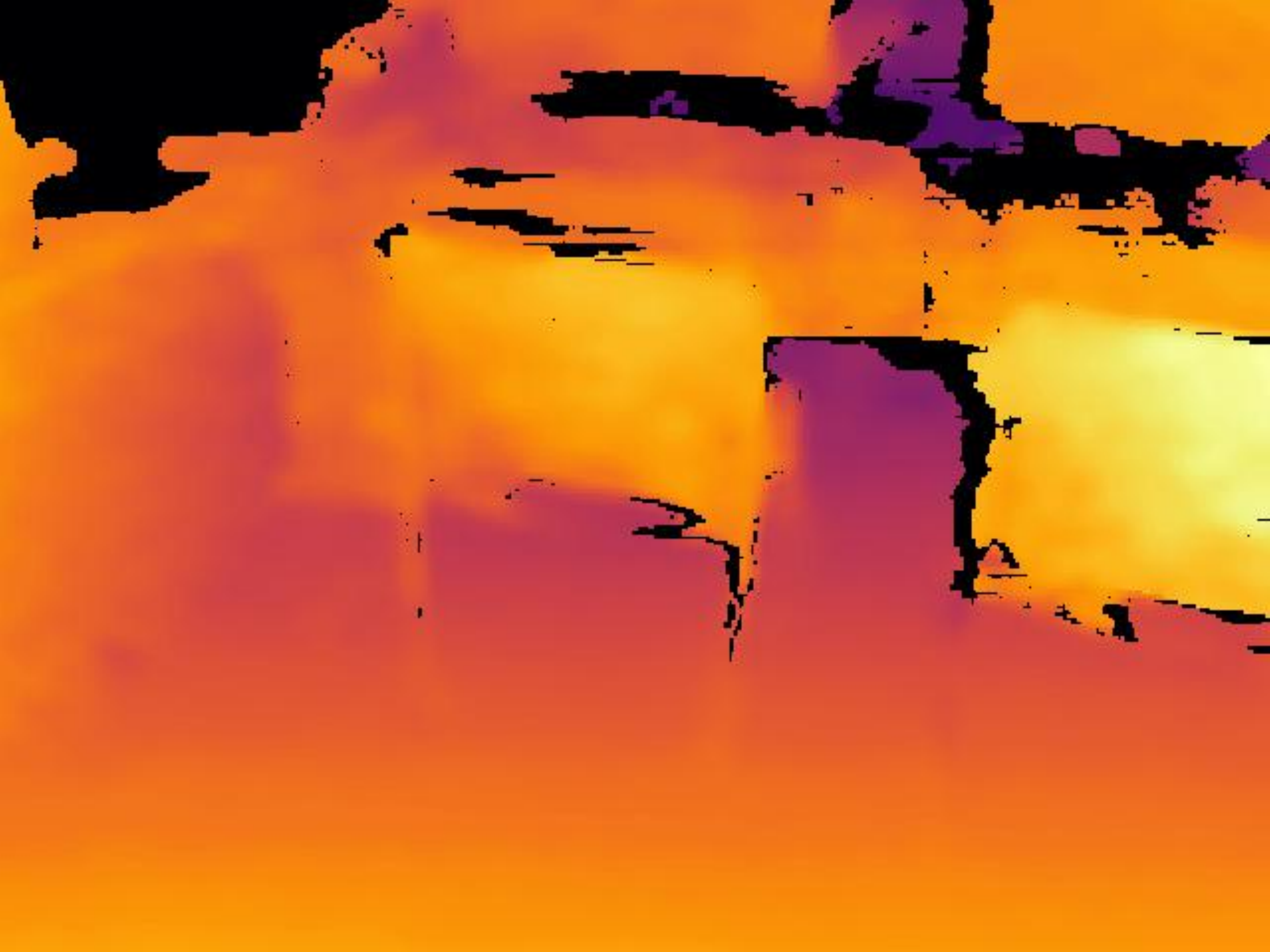}&
    \includegraphics[width=0.104\linewidth]{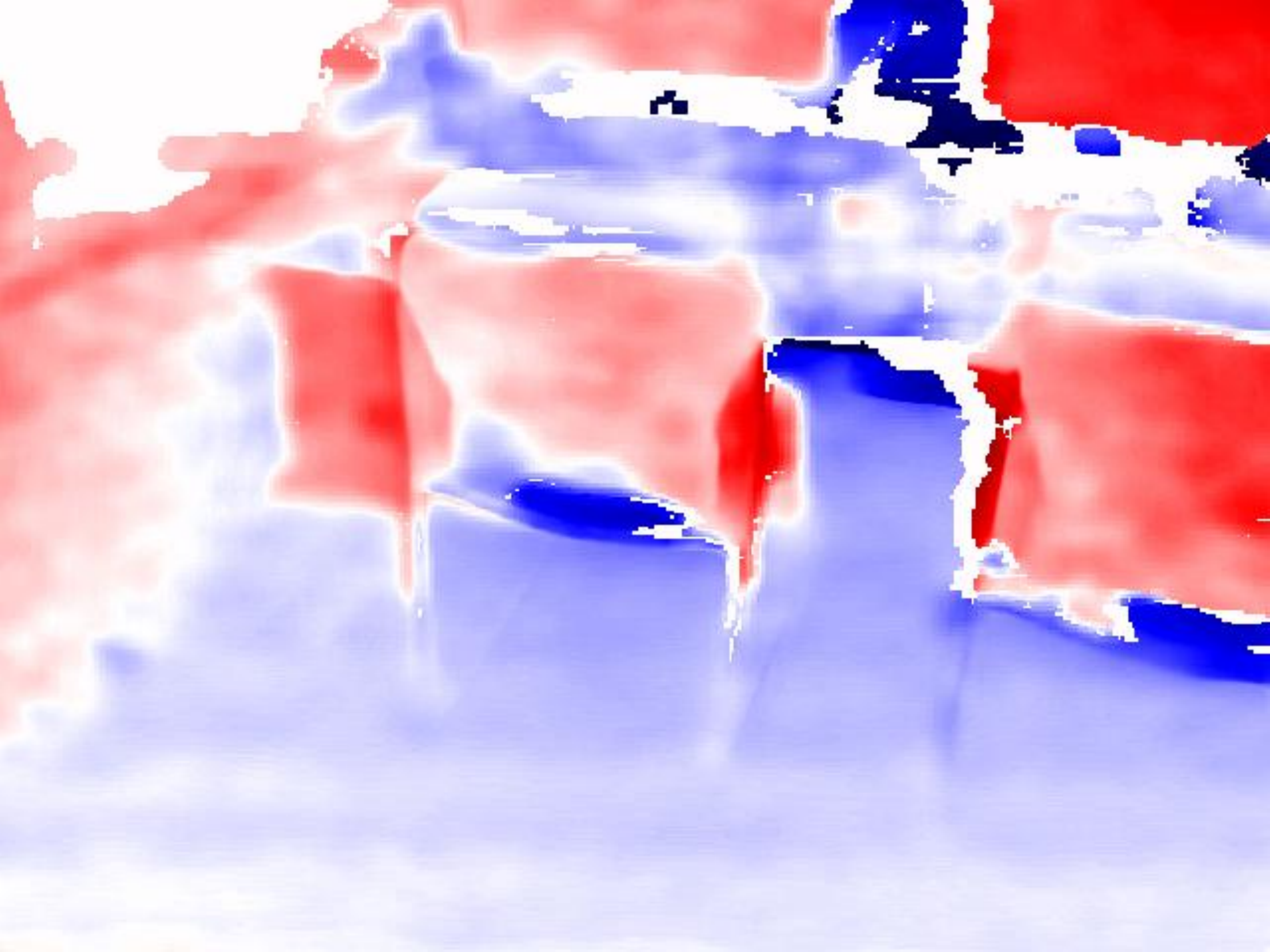}&
    \includegraphics[width=0.104\linewidth]{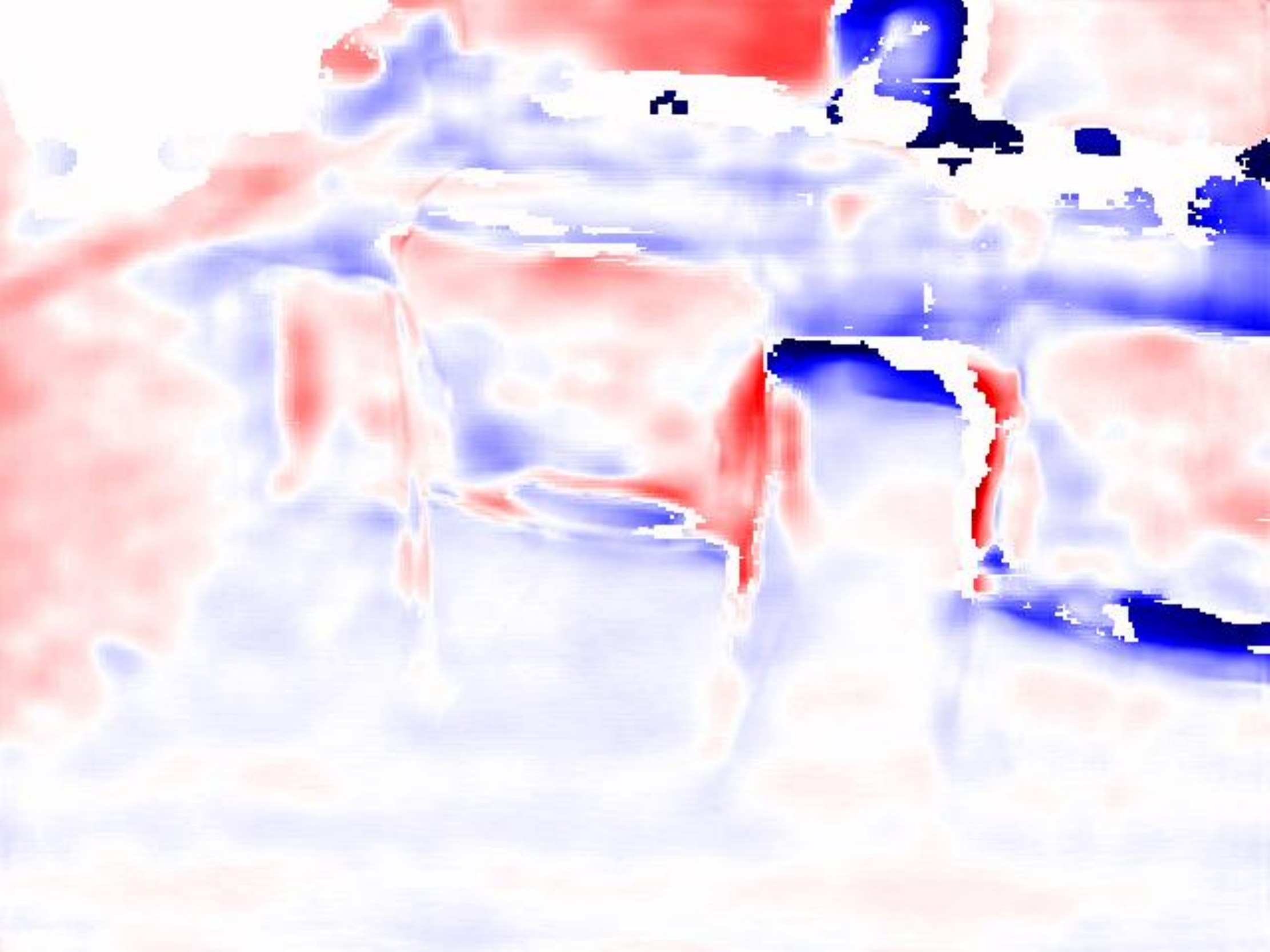}&
    \includegraphics[width=0.024\linewidth]{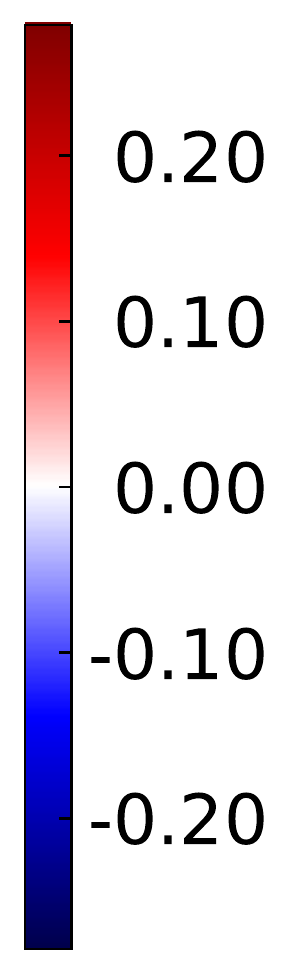}\\
    \vspace{-0.75mm}
    \scriptsize b. &
    \includegraphics[width=0.104\linewidth]{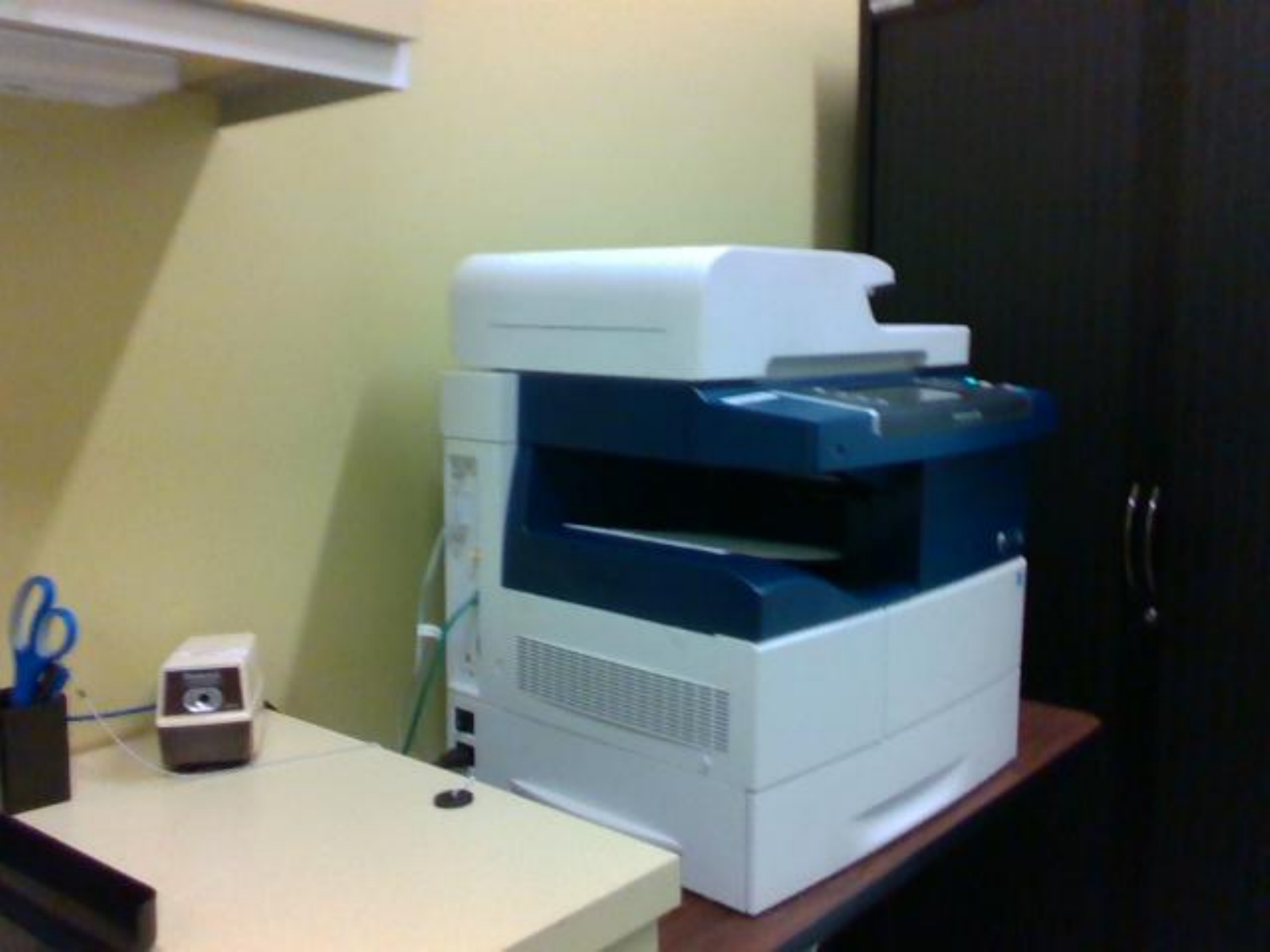}&
    \includegraphics[width=0.104\linewidth]{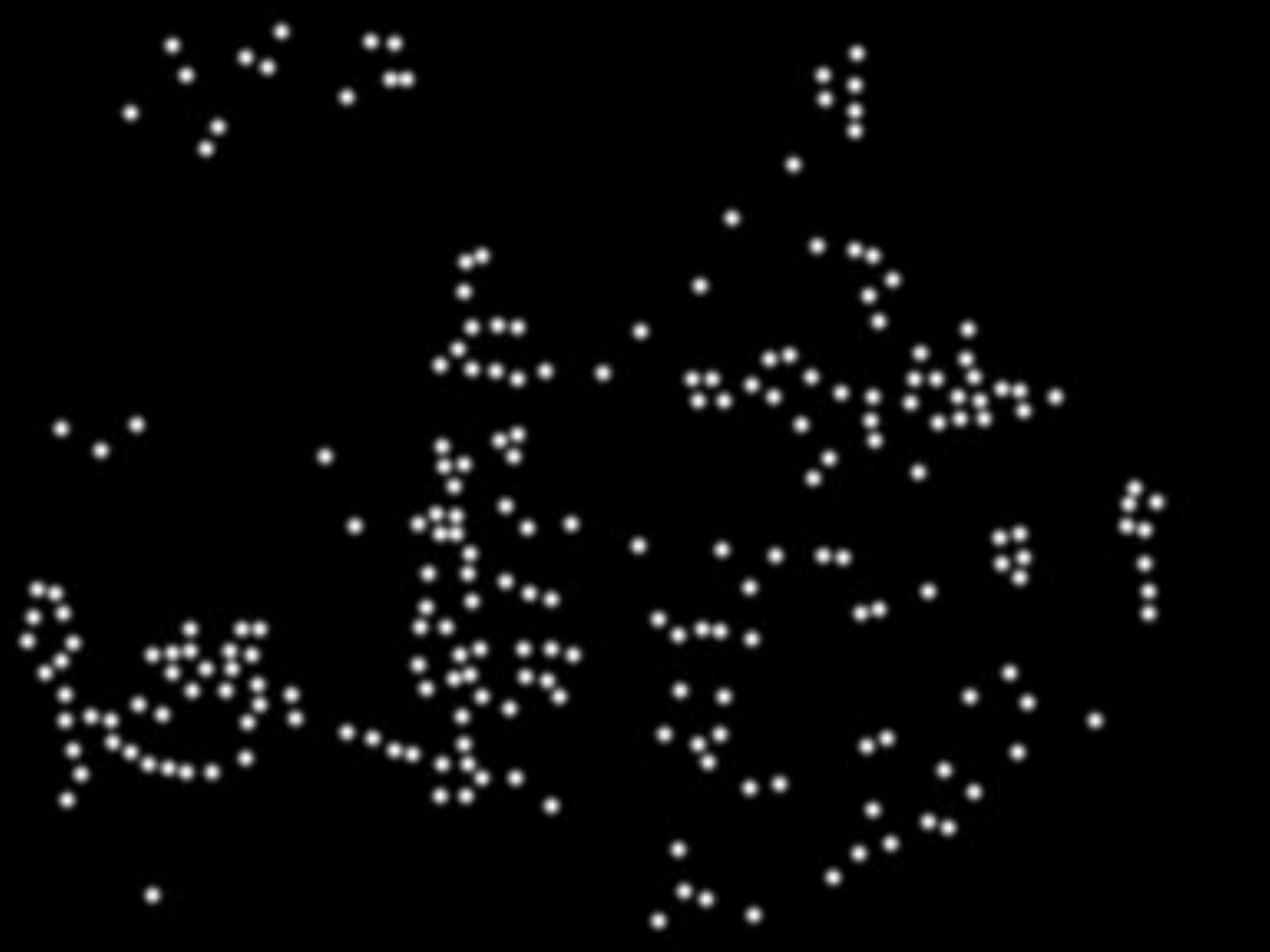}&
    \includegraphics[width=0.104\linewidth]{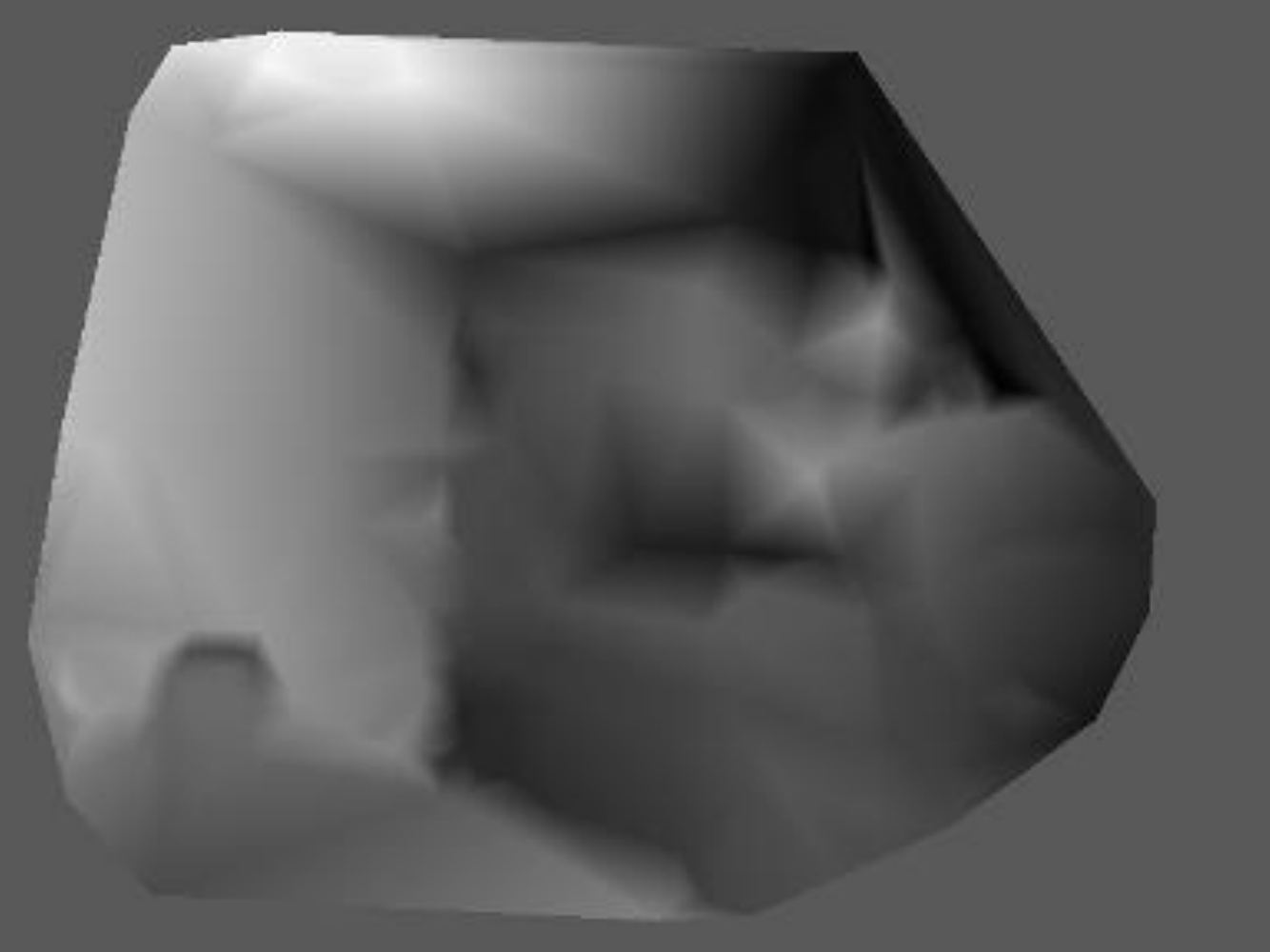}&
    \includegraphics[width=0.104\linewidth]{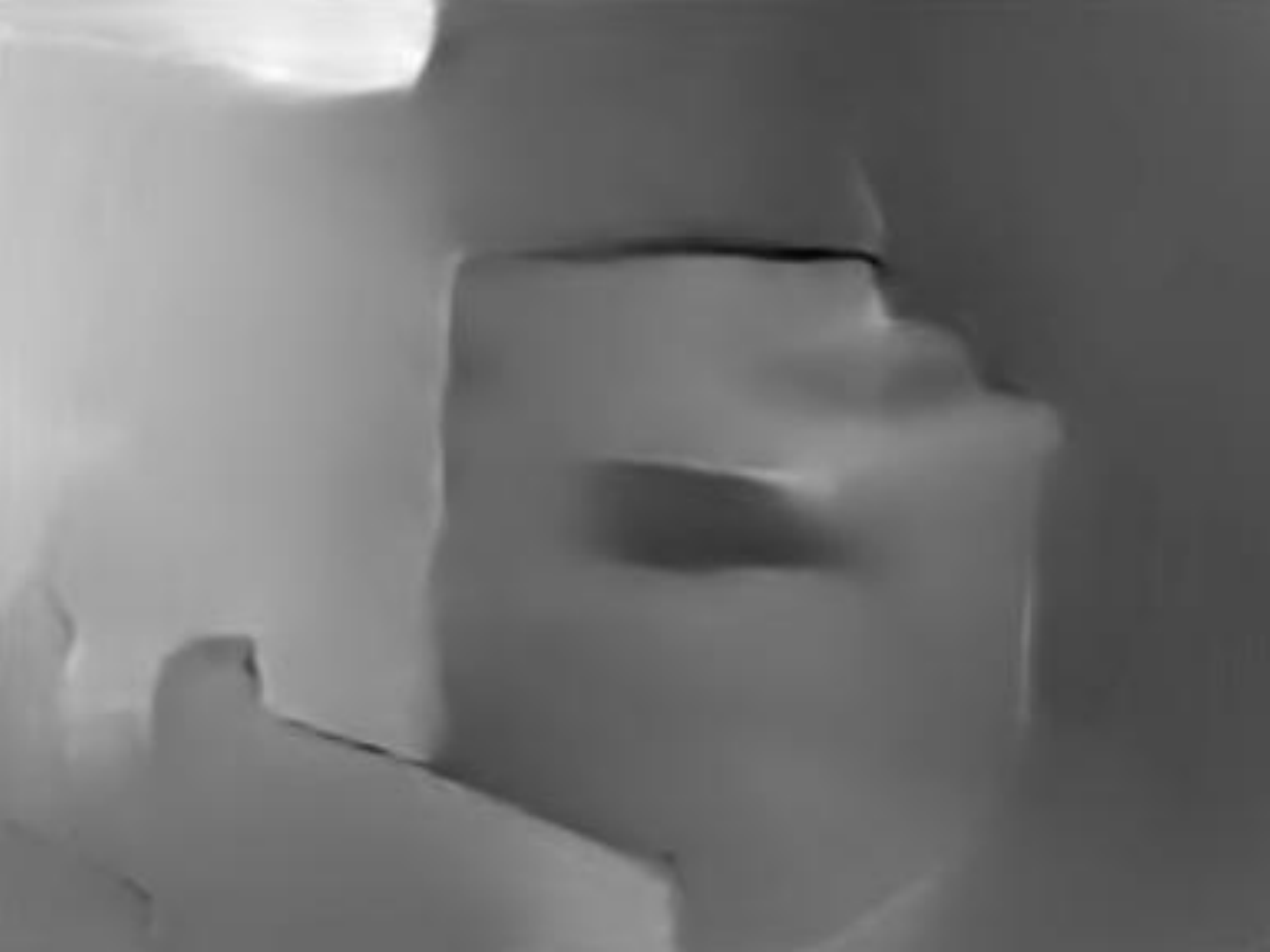}&
    \includegraphics[width=0.104\linewidth]{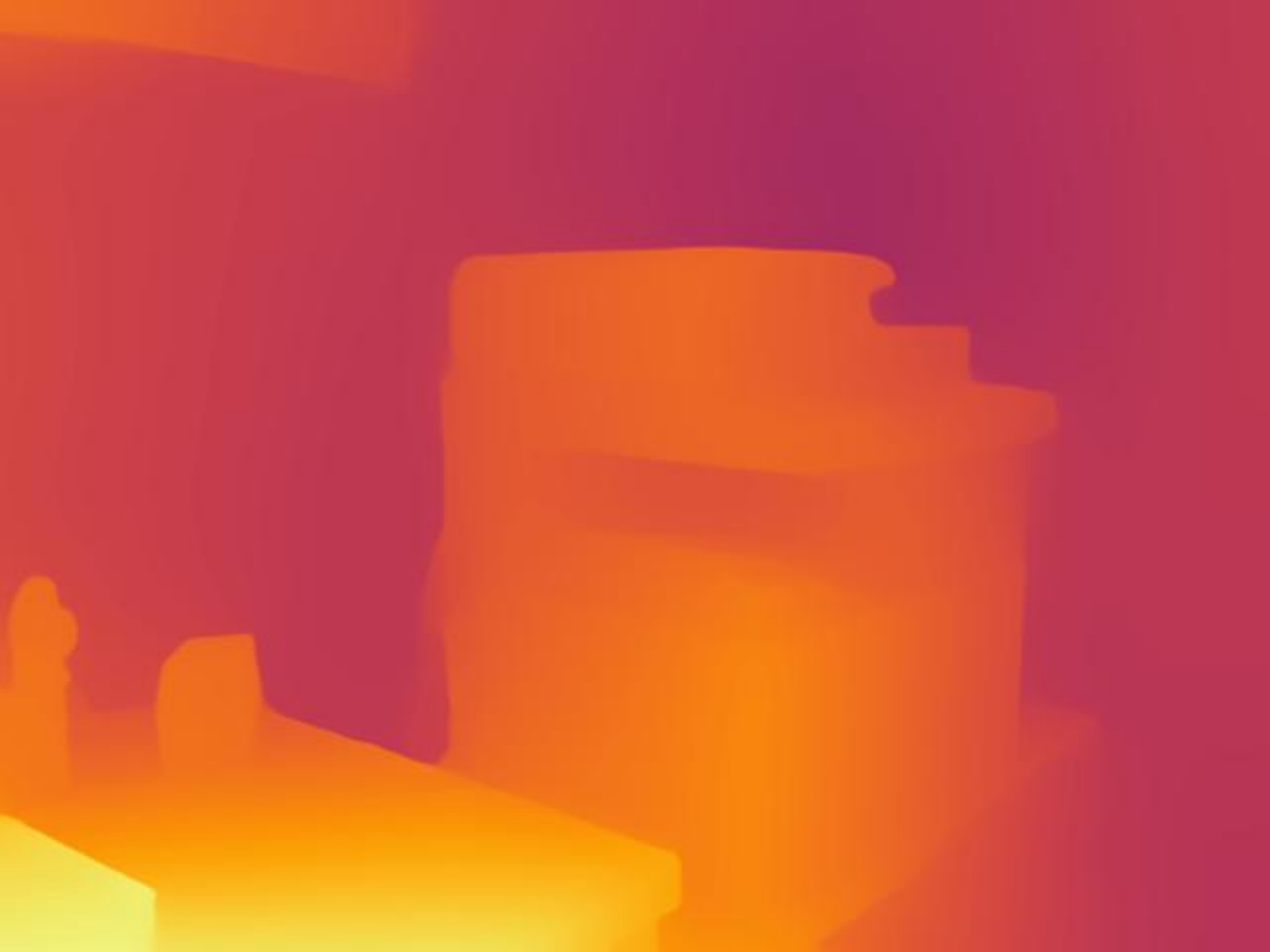}&
    \includegraphics[width=0.104\linewidth]{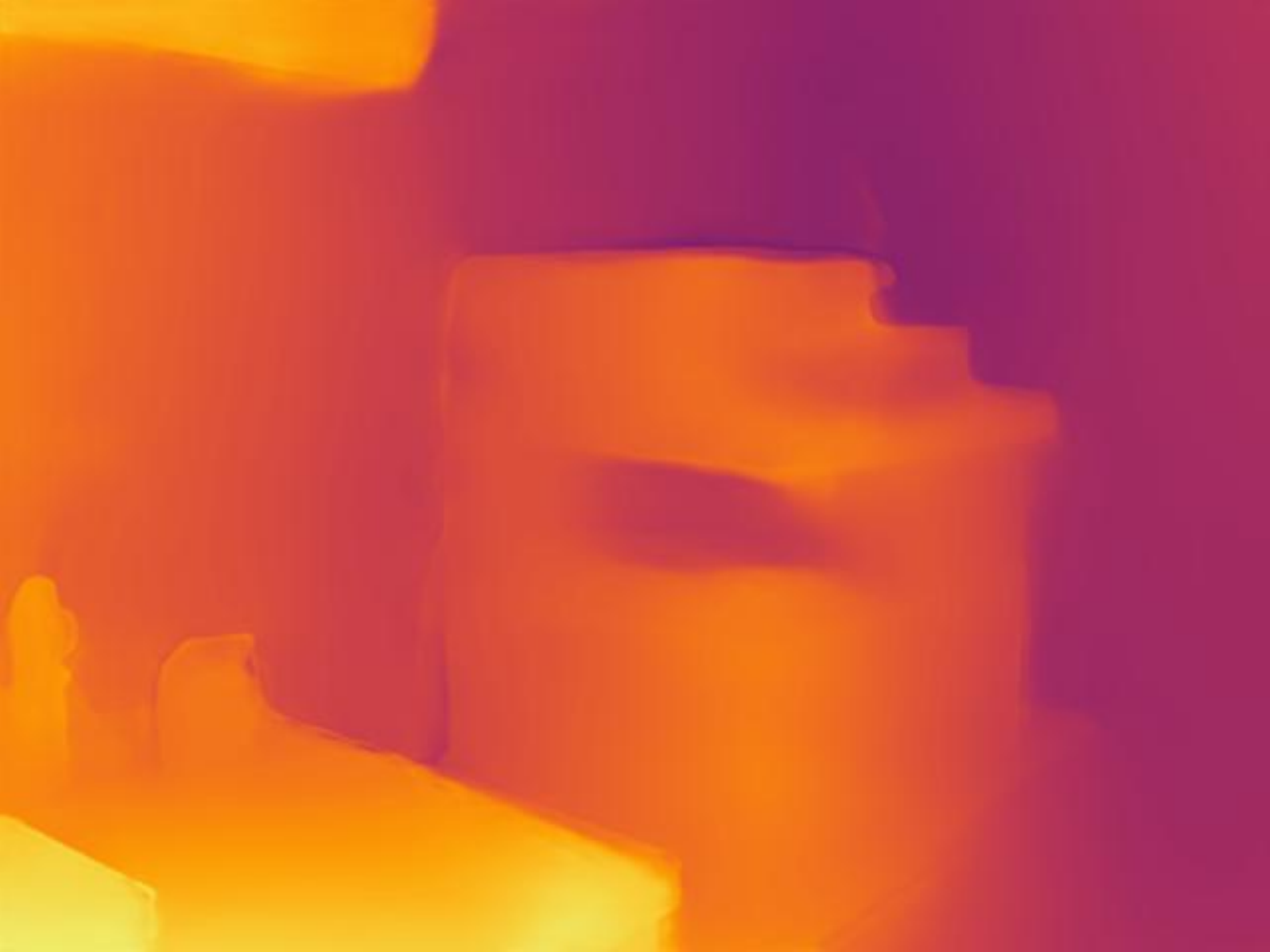}&
    \includegraphics[width=0.104\linewidth]{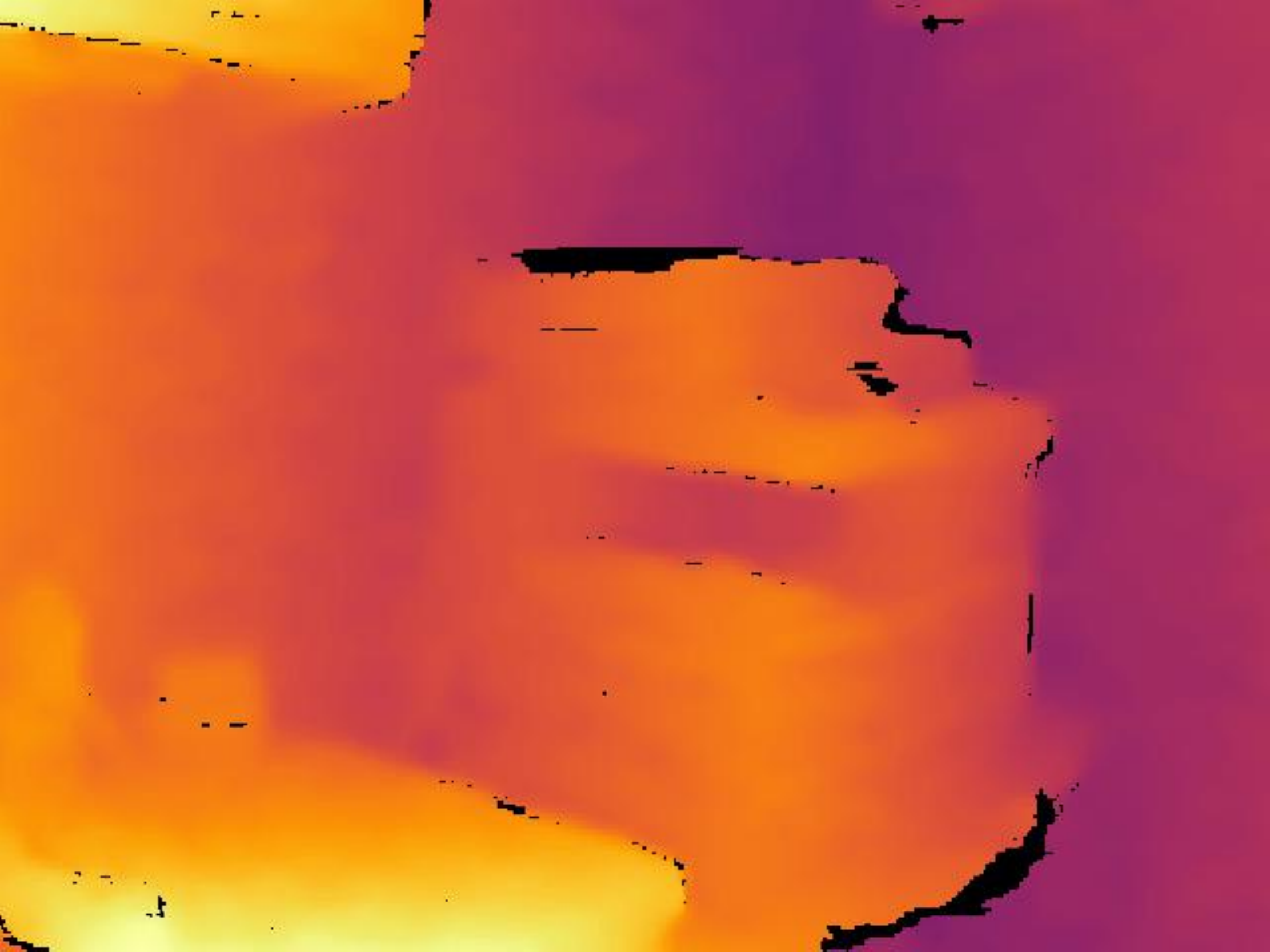}&
    \includegraphics[width=0.104\linewidth]{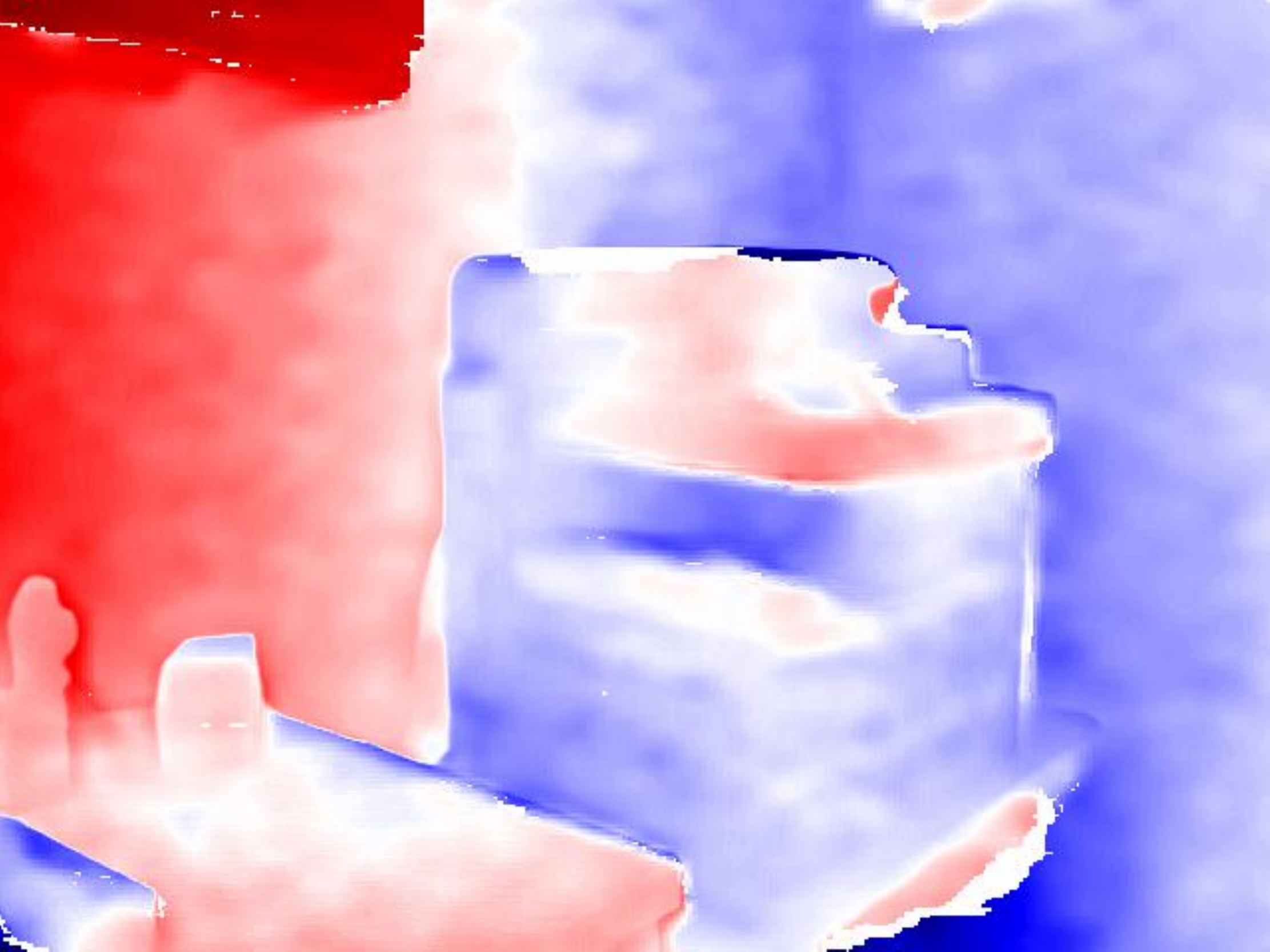}&
    \includegraphics[width=0.104\linewidth]{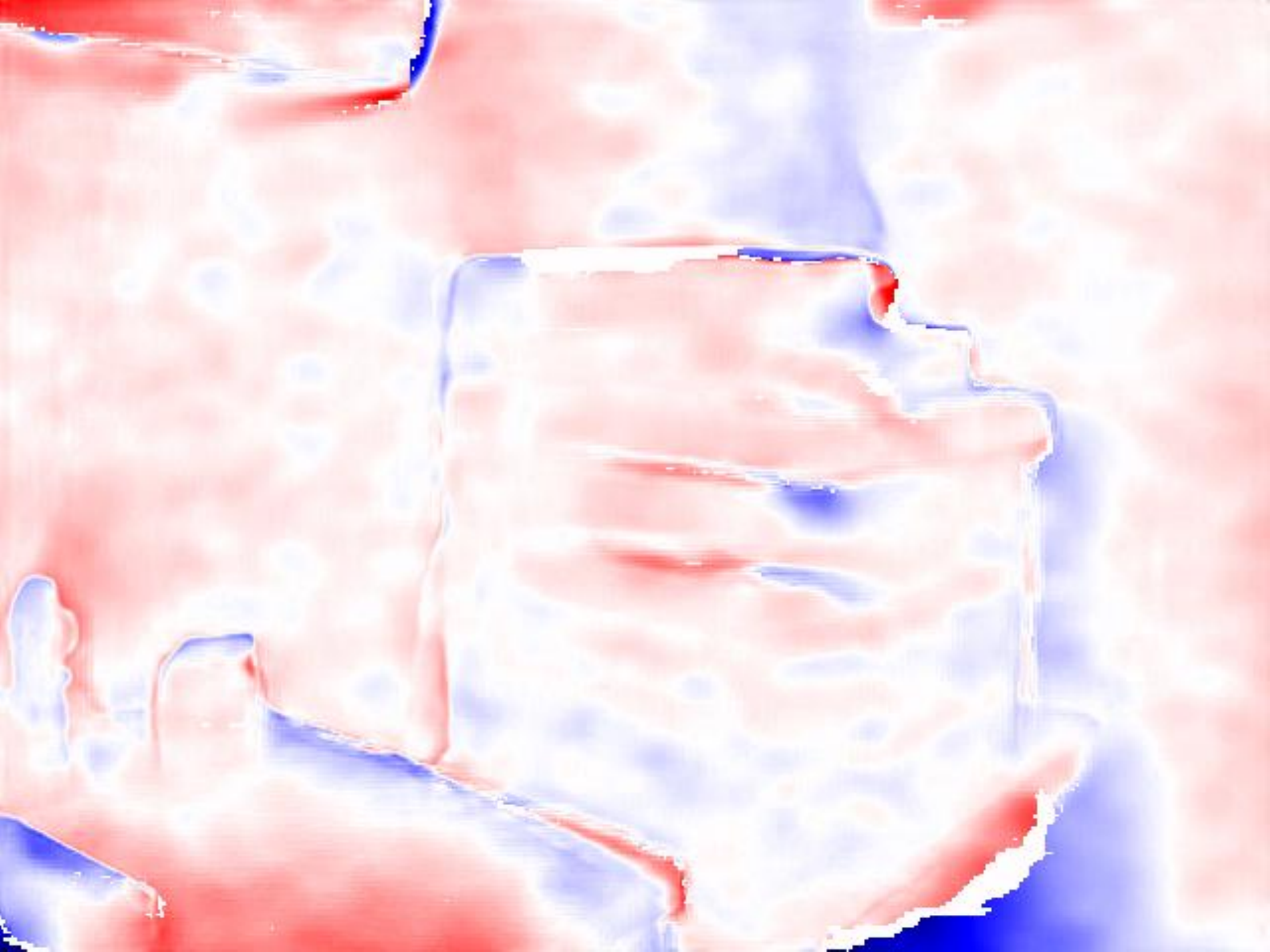}&
    \includegraphics[width=0.024\linewidth]{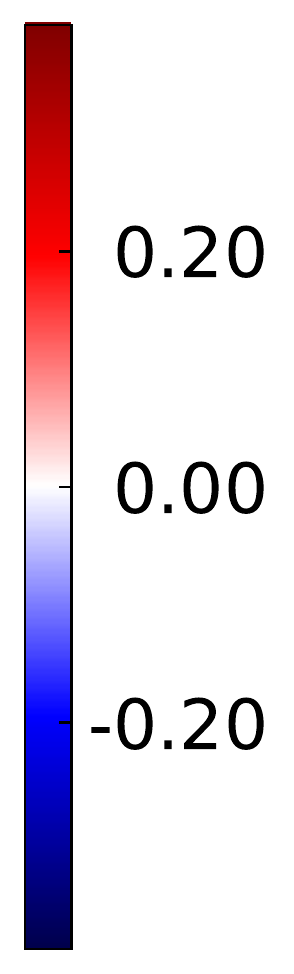}\\
    \vspace{-0.75mm}
    \scriptsize c. &
    \includegraphics[width=0.104\linewidth]{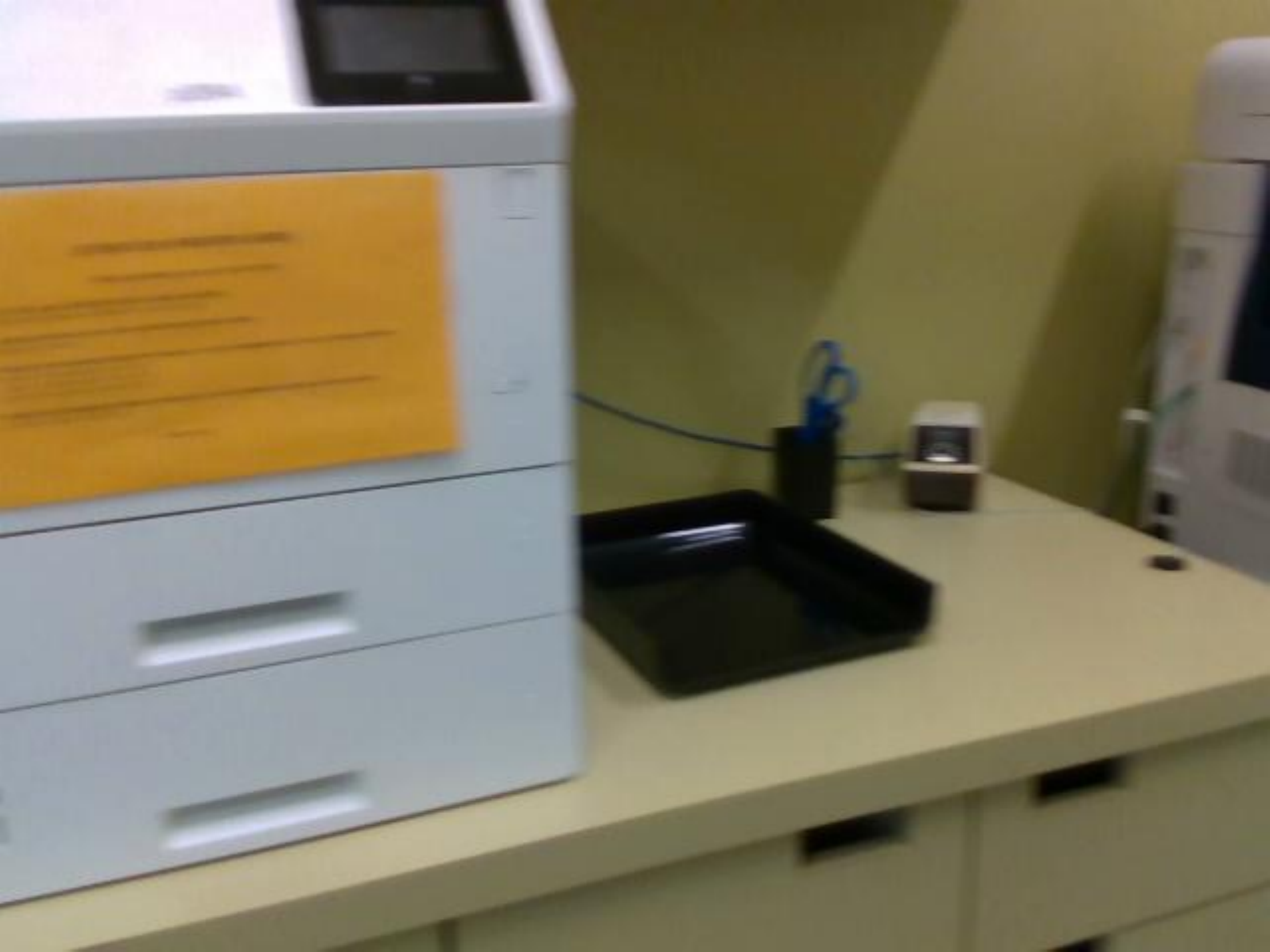}&
    \includegraphics[width=0.104\linewidth]{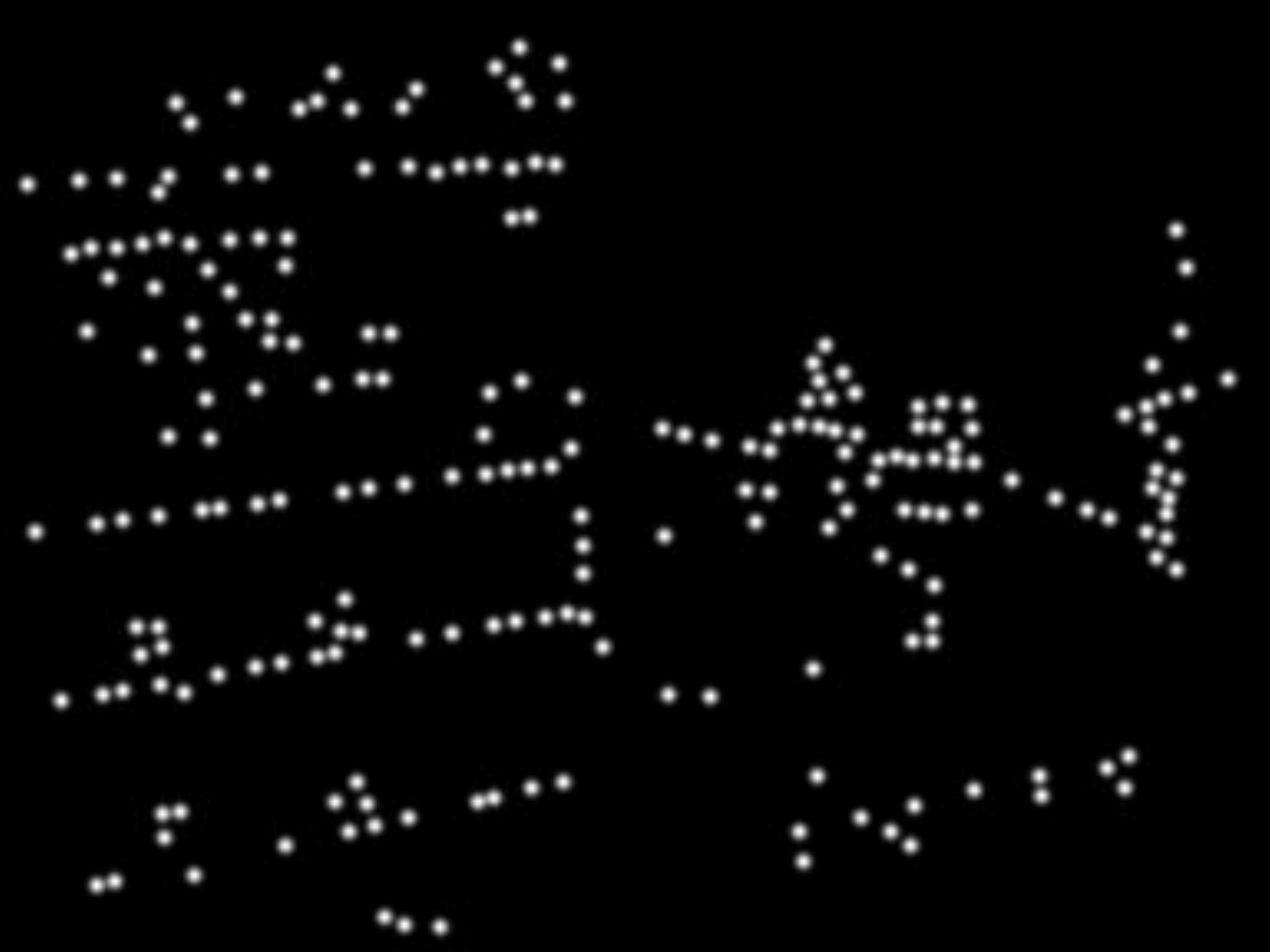}&
    \includegraphics[width=0.104\linewidth]{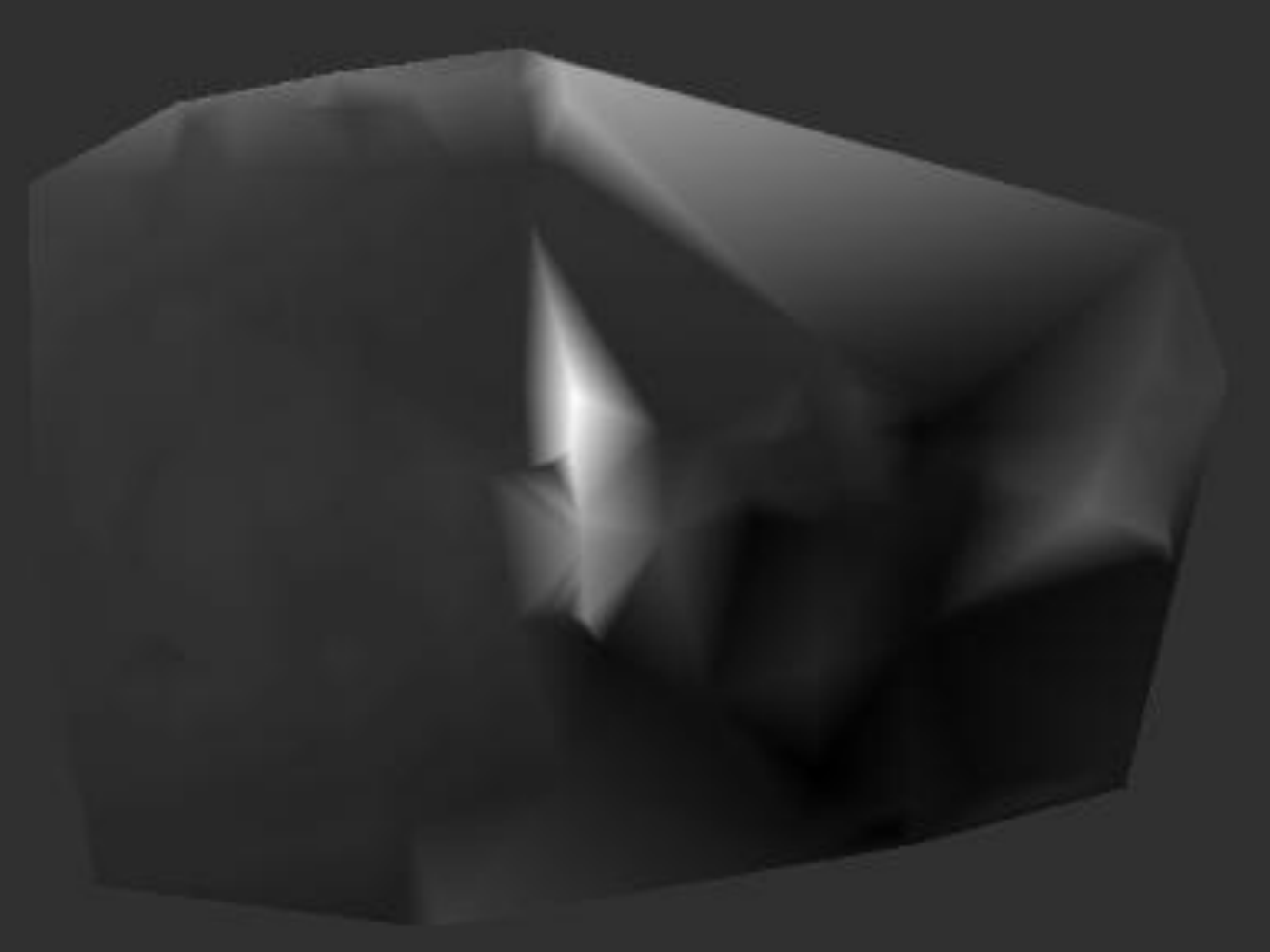}&
    \includegraphics[width=0.104\linewidth]{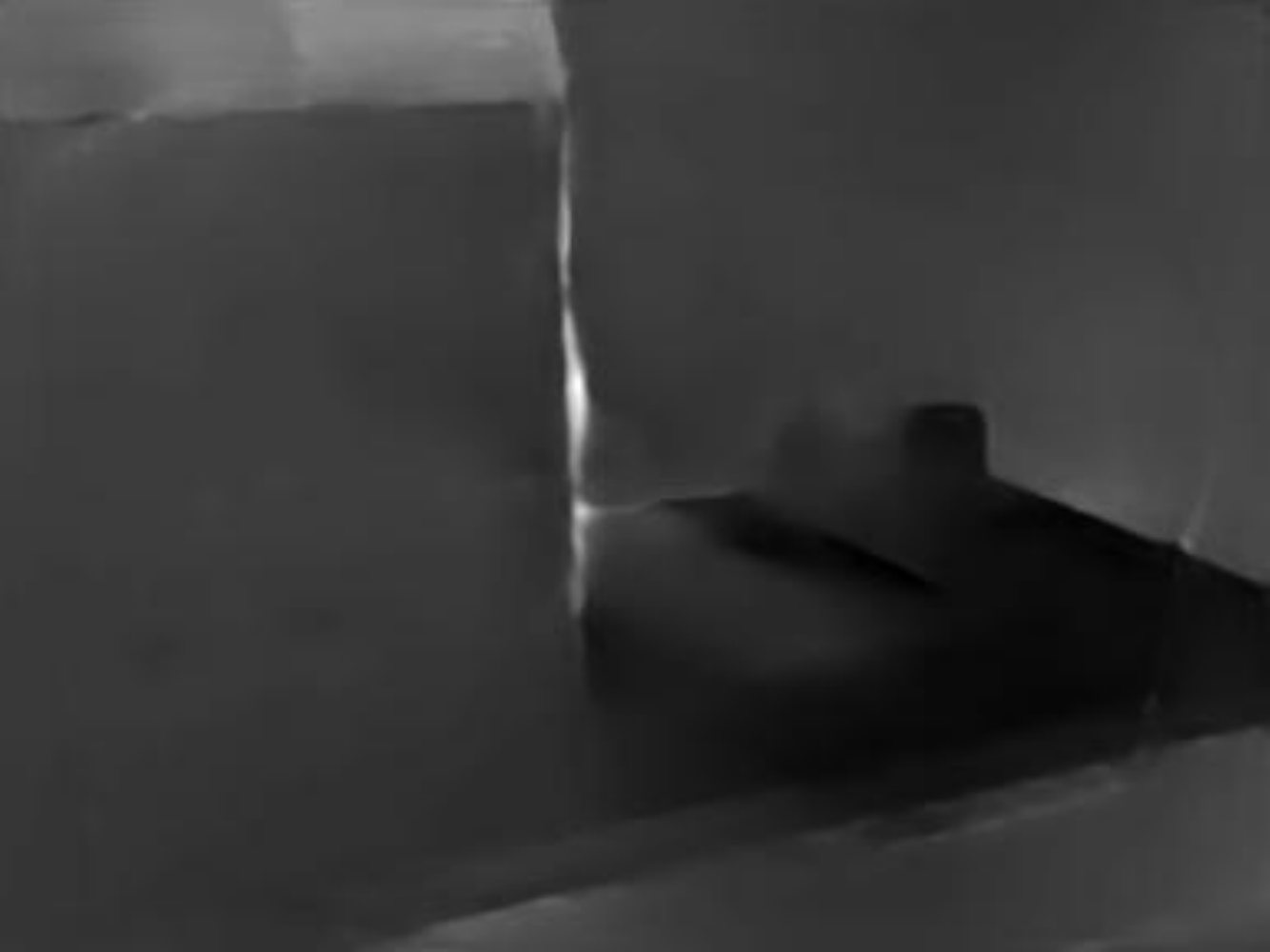}&
    \includegraphics[width=0.104\linewidth]{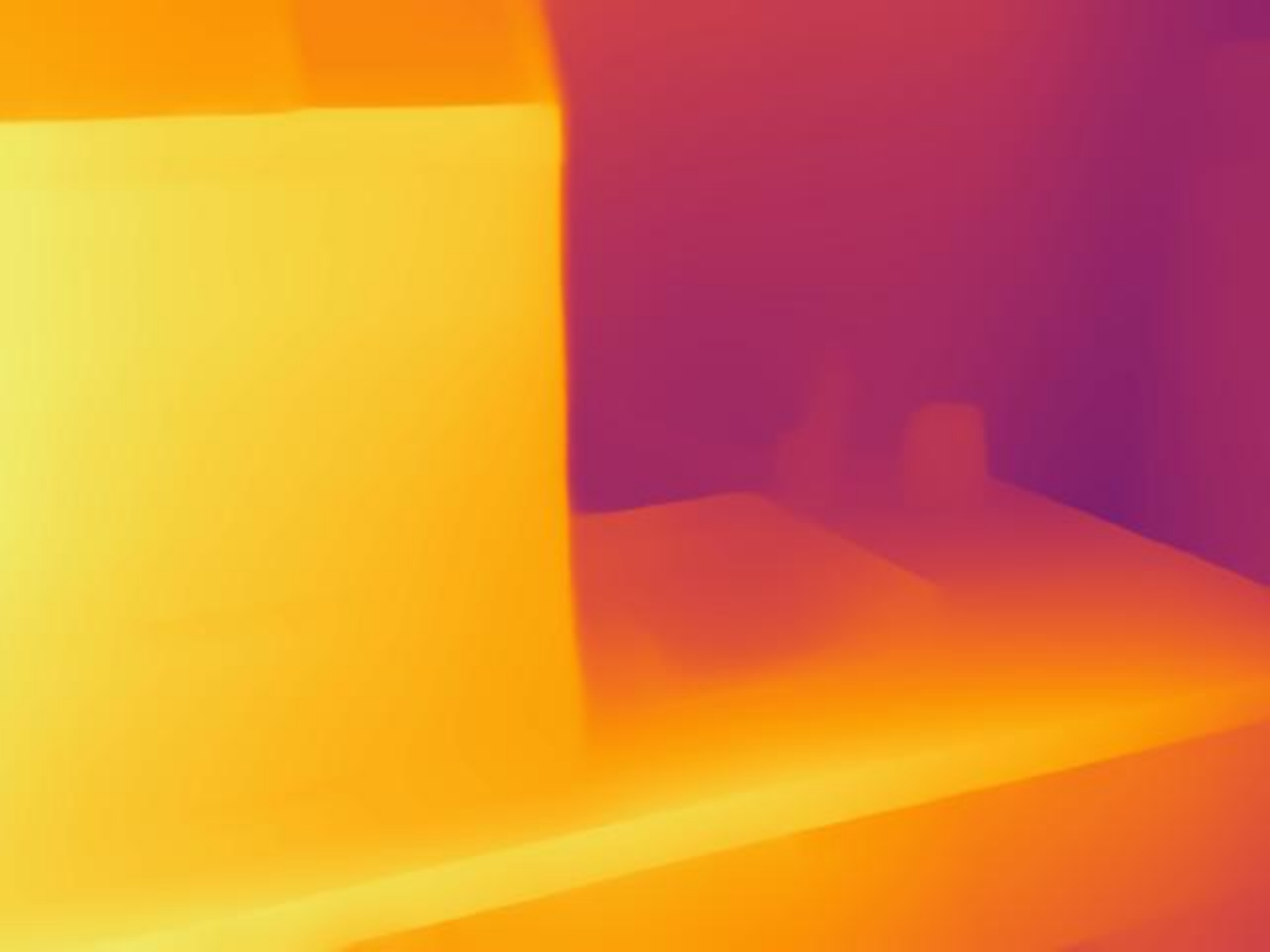}&
    \includegraphics[width=0.104\linewidth]{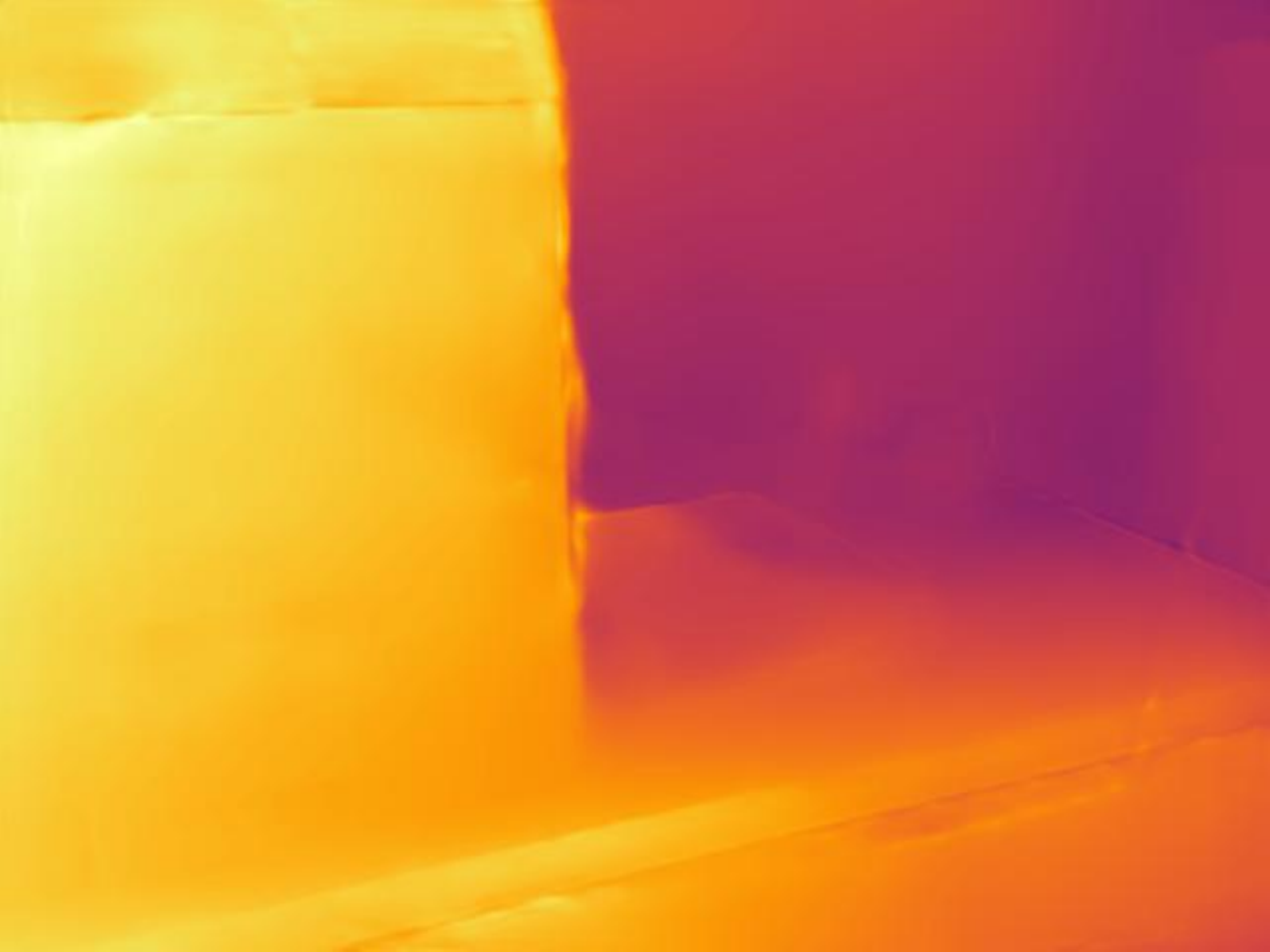}&
    \includegraphics[width=0.104\linewidth]{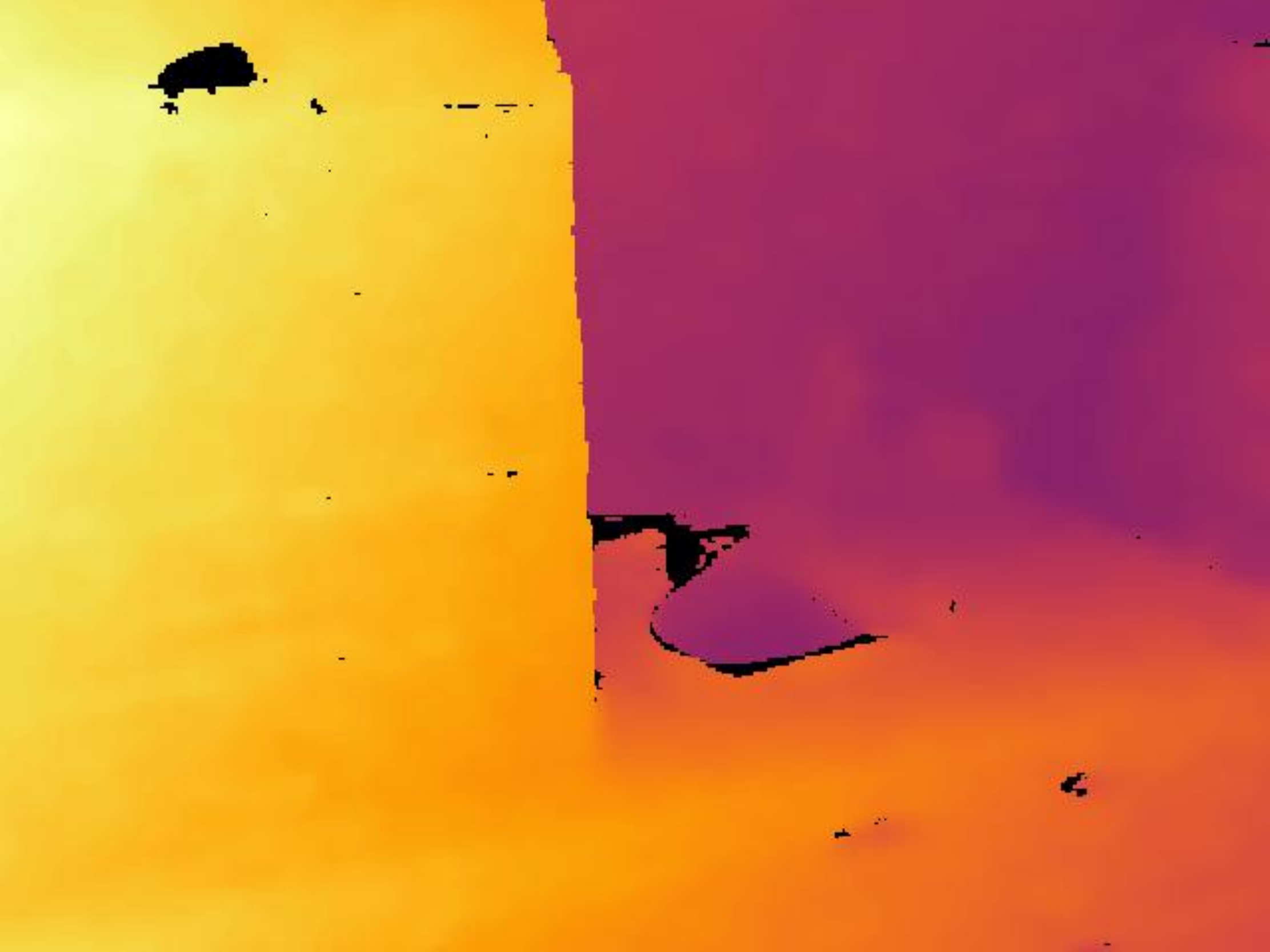}&
    \includegraphics[width=0.104\linewidth]{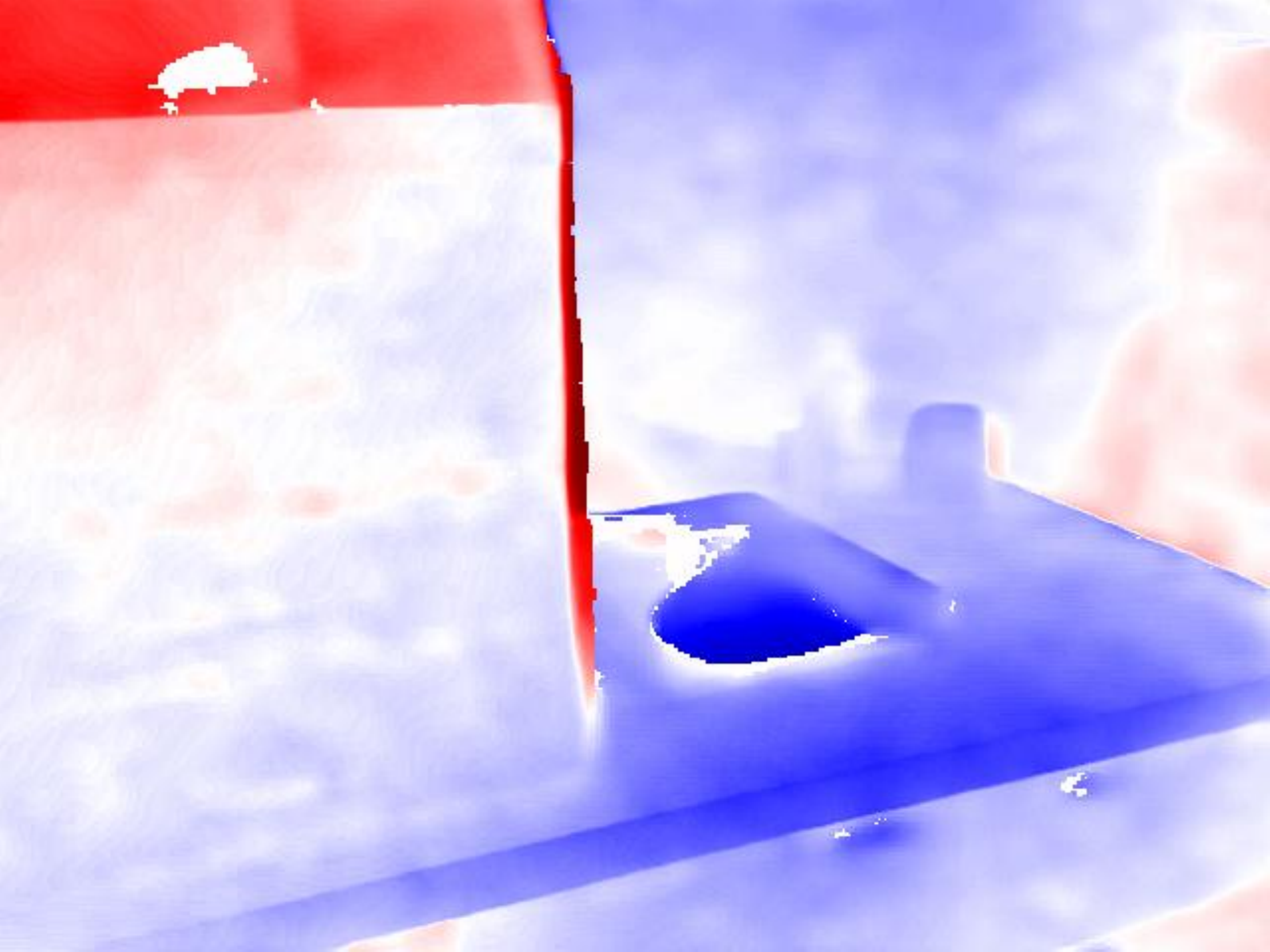}&
    \includegraphics[width=0.104\linewidth]{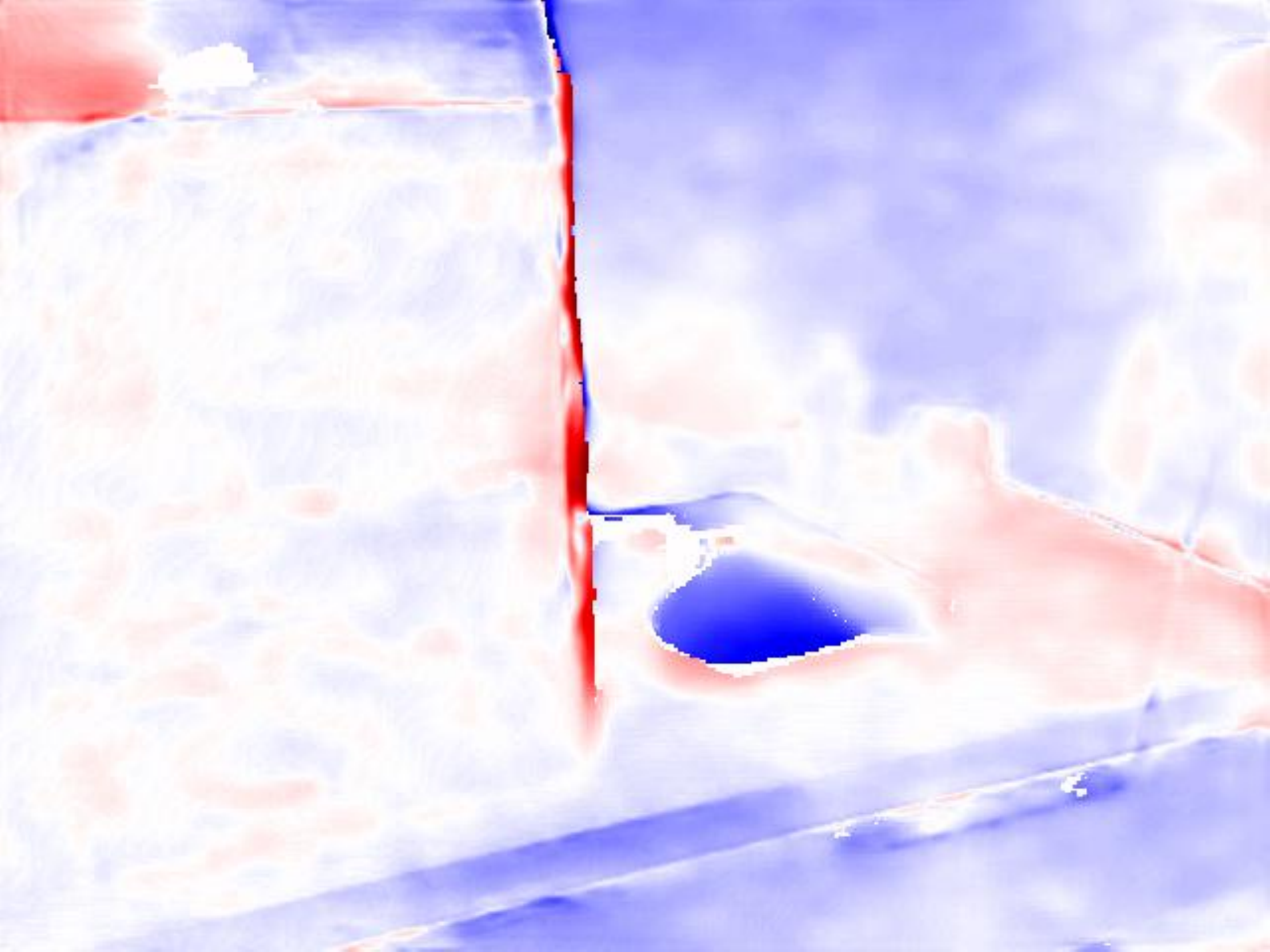}&
    \includegraphics[width=0.024\linewidth]{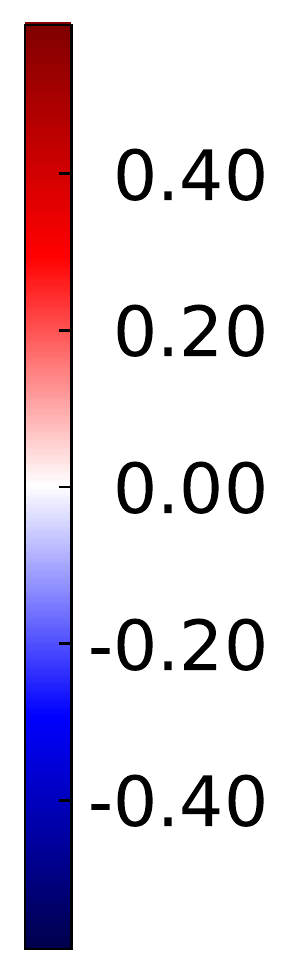}\\
    \vspace{-0.75mm}
    \scriptsize d. &
    \includegraphics[width=0.104\linewidth]{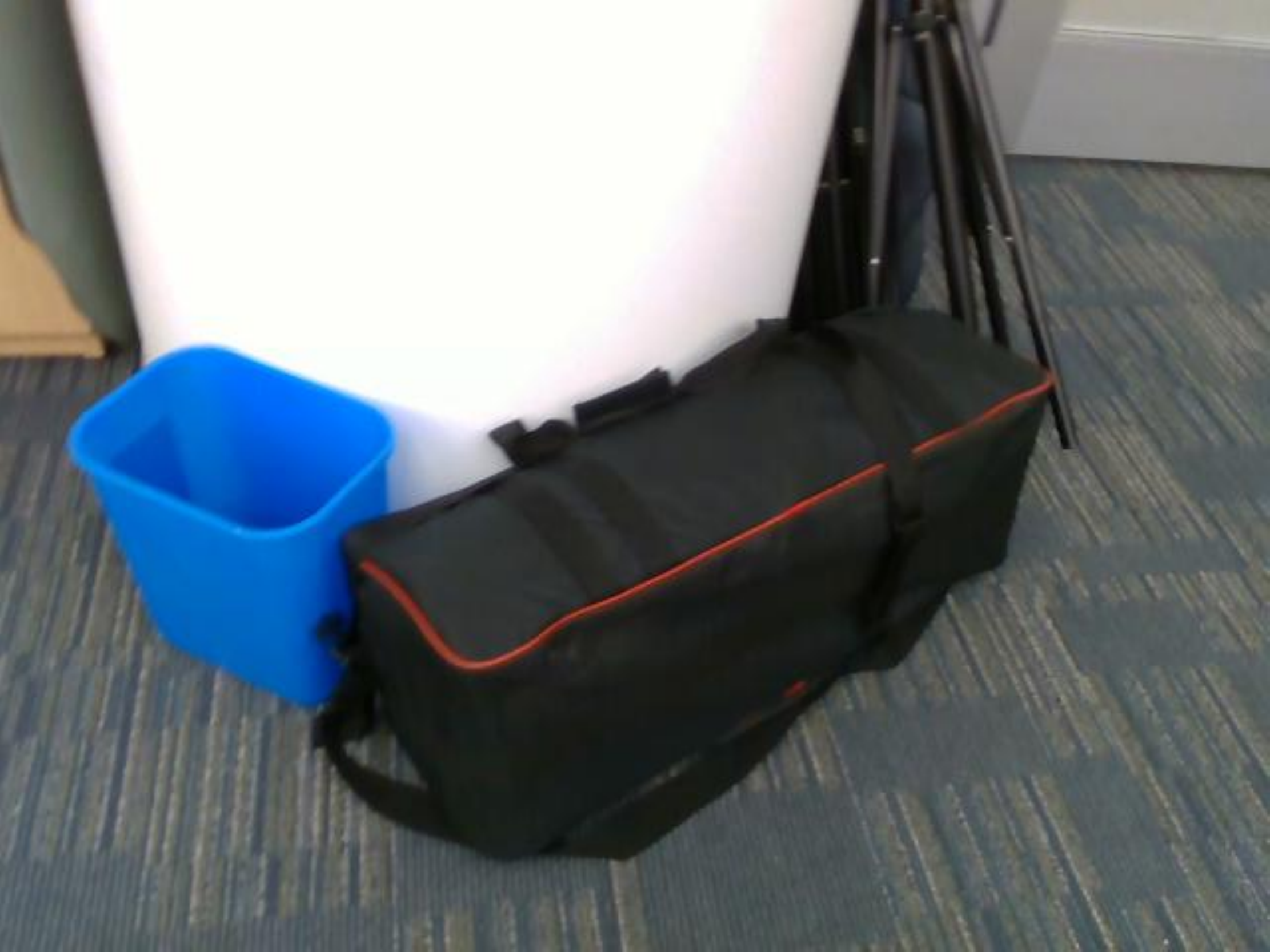}&
    \includegraphics[width=0.104\linewidth]{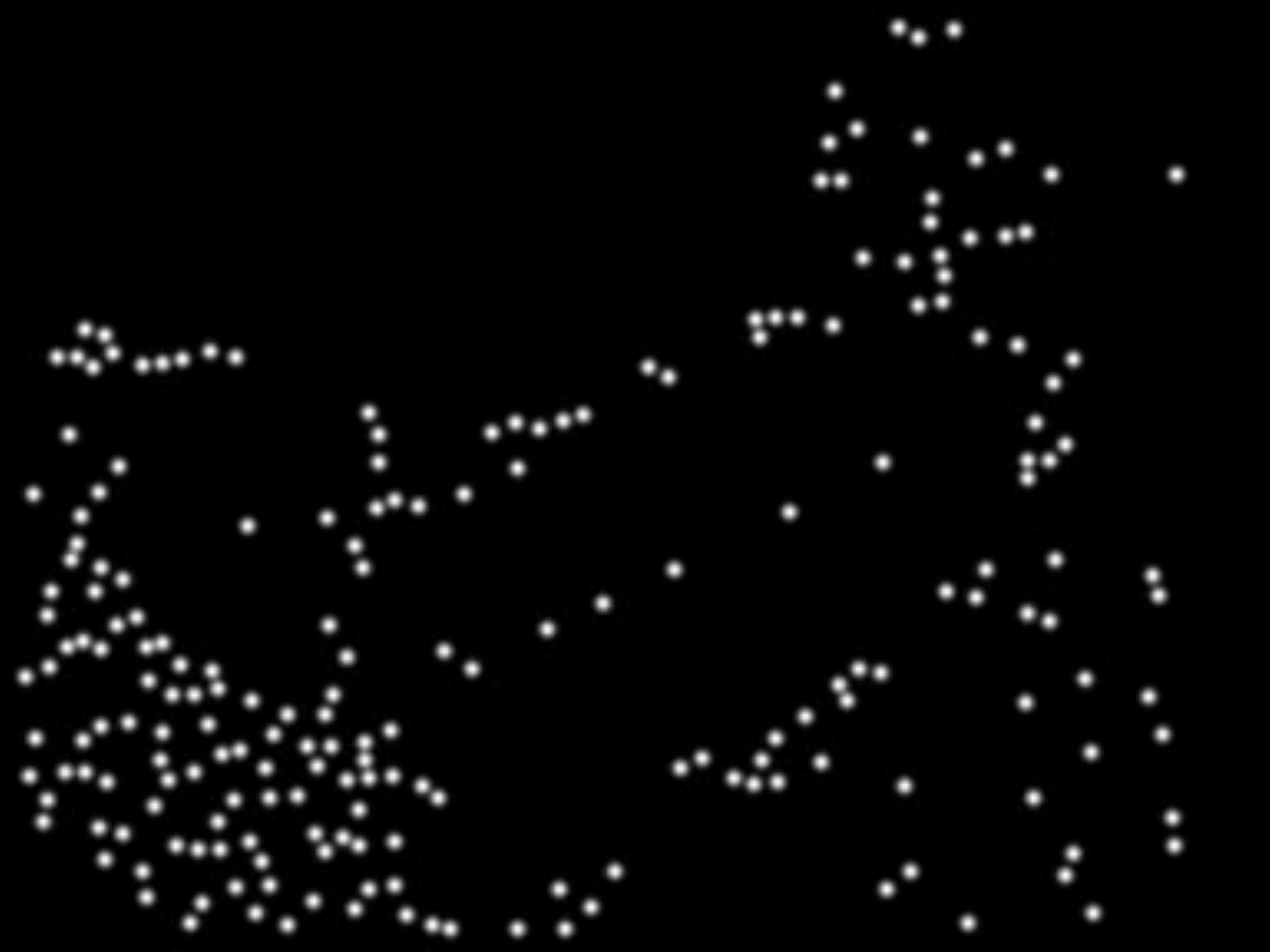}&
    \includegraphics[width=0.104\linewidth]{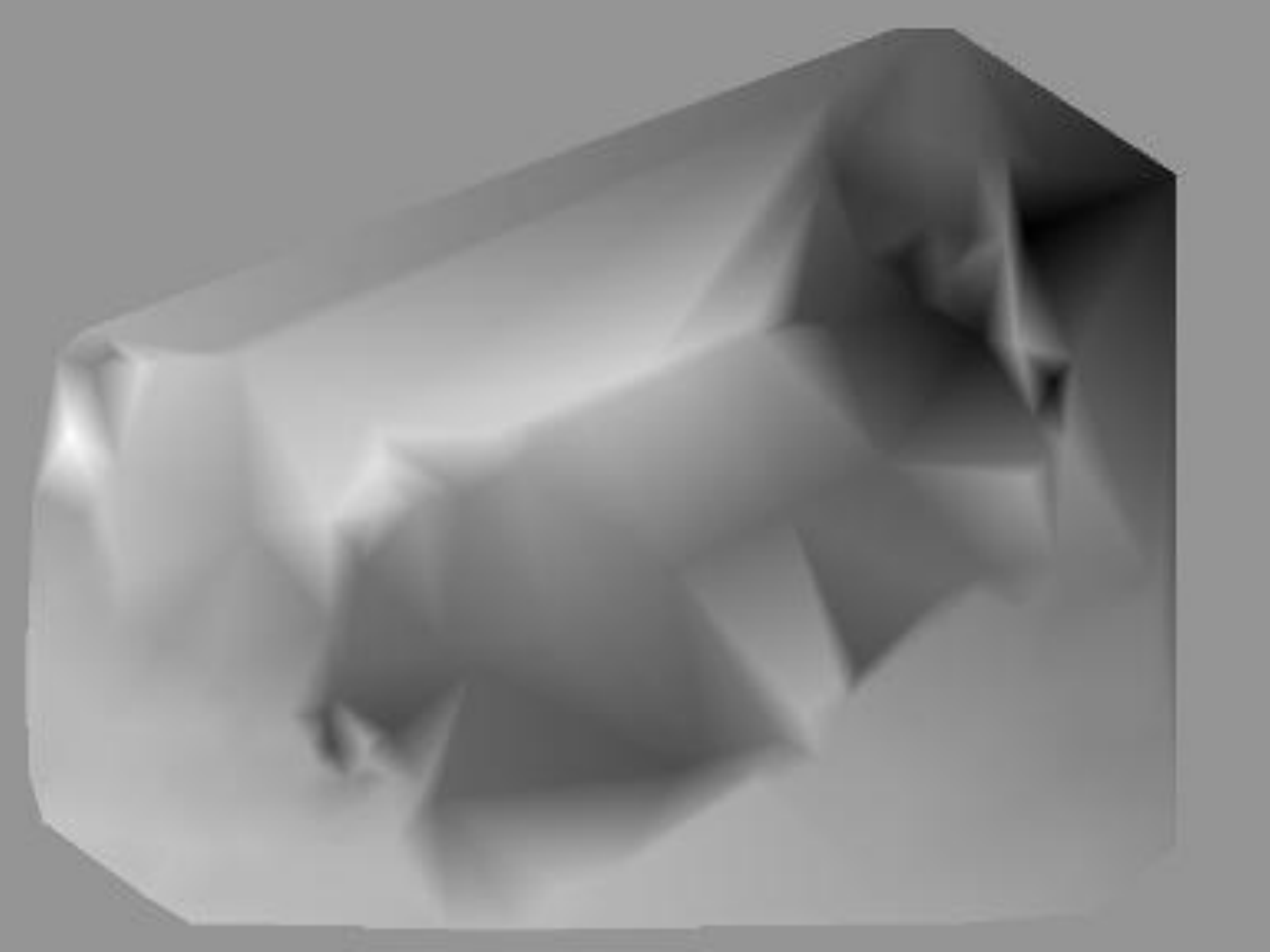}&
    \includegraphics[width=0.104\linewidth]{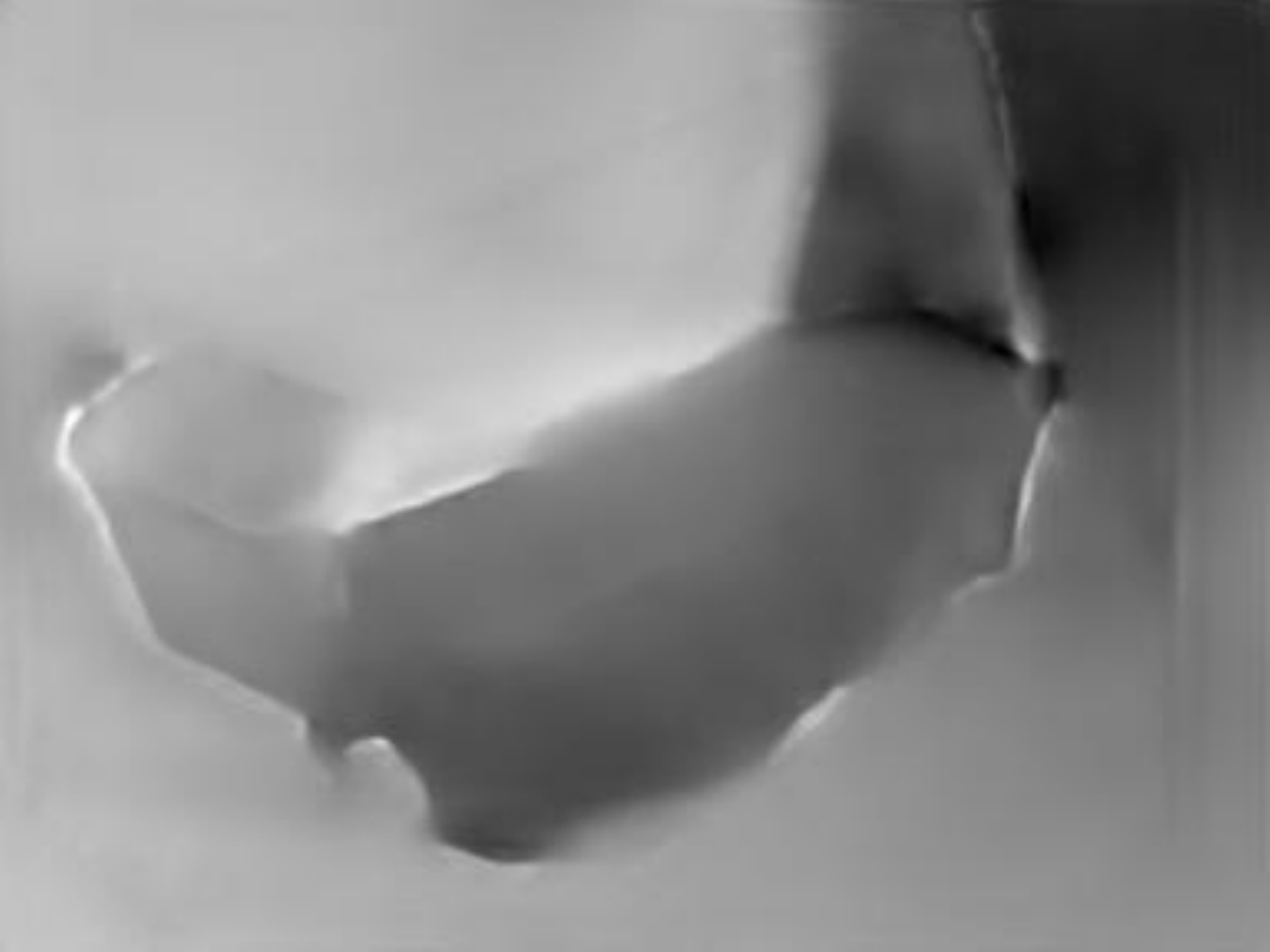}&
    \includegraphics[width=0.104\linewidth]{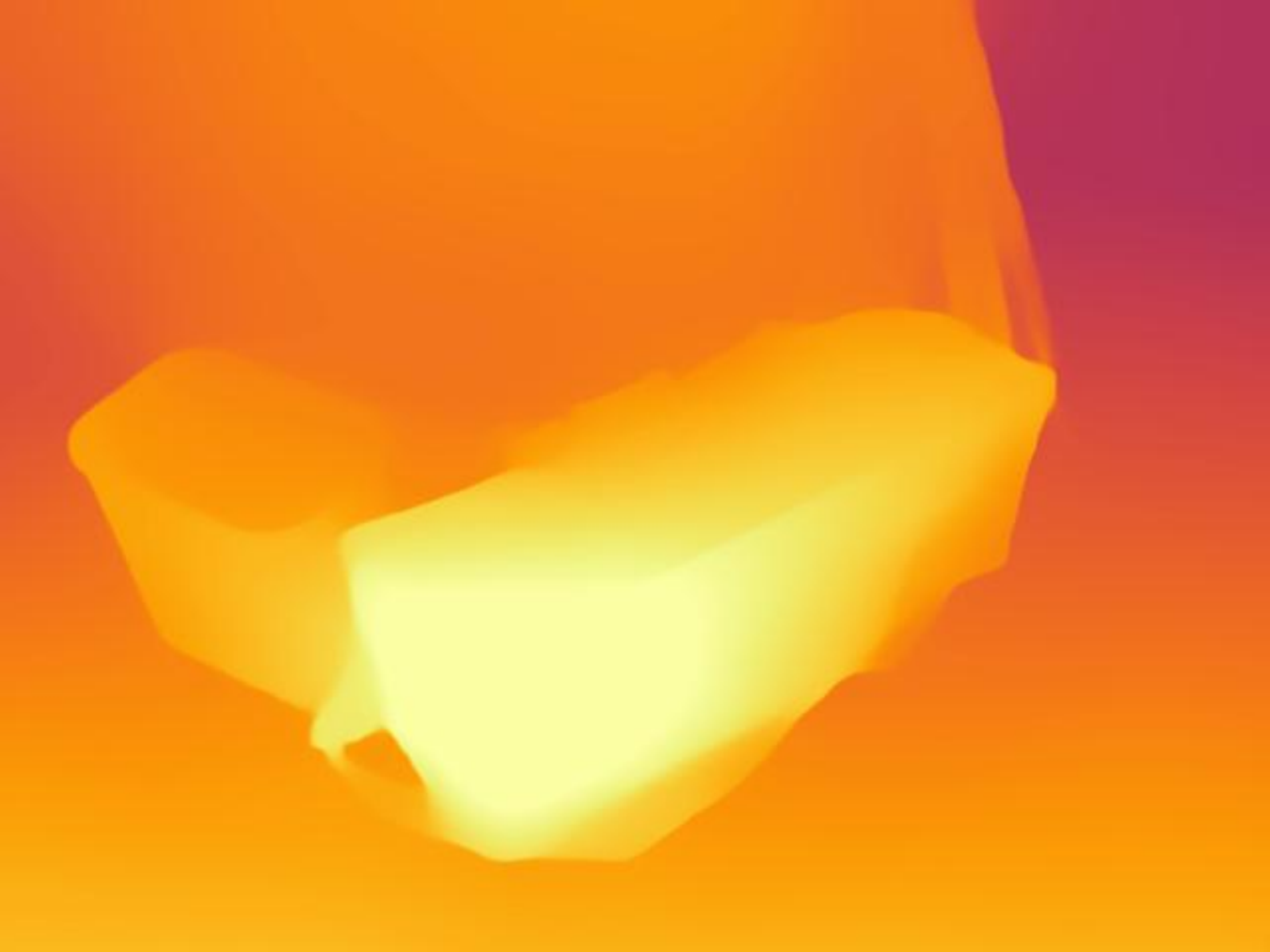}&
    \includegraphics[width=0.104\linewidth]{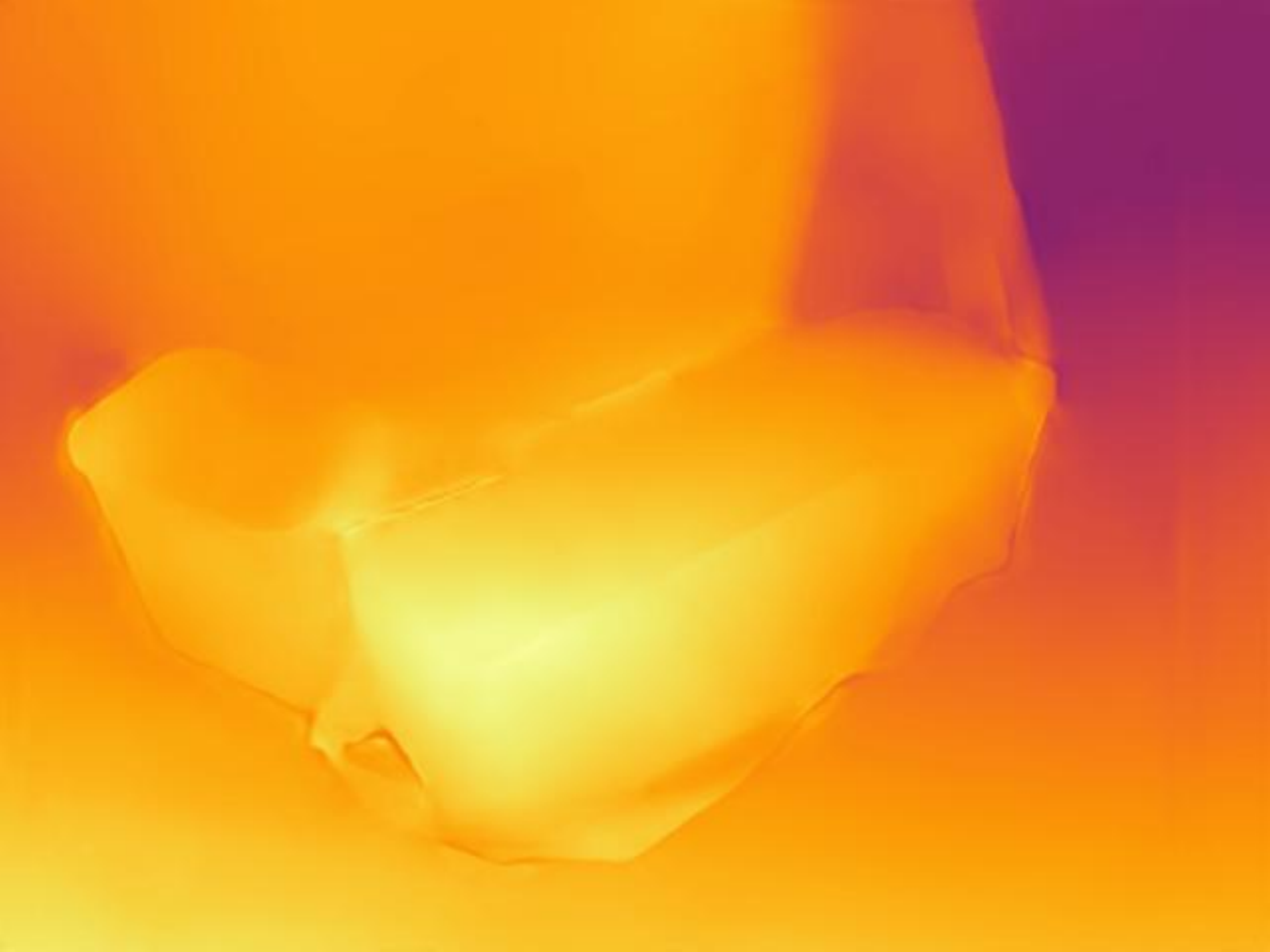}&
    \includegraphics[width=0.104\linewidth]{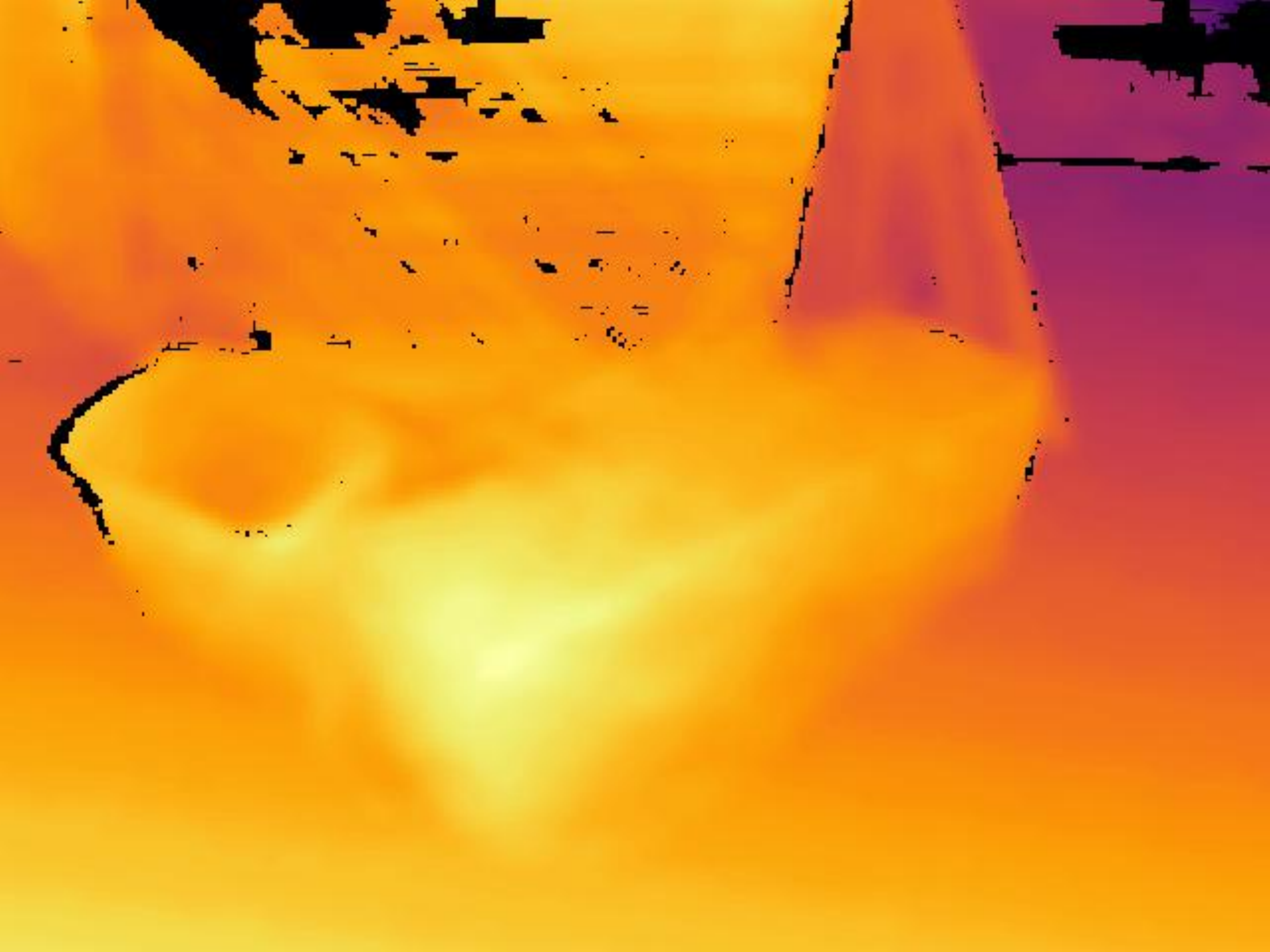}&
    \includegraphics[width=0.104\linewidth]{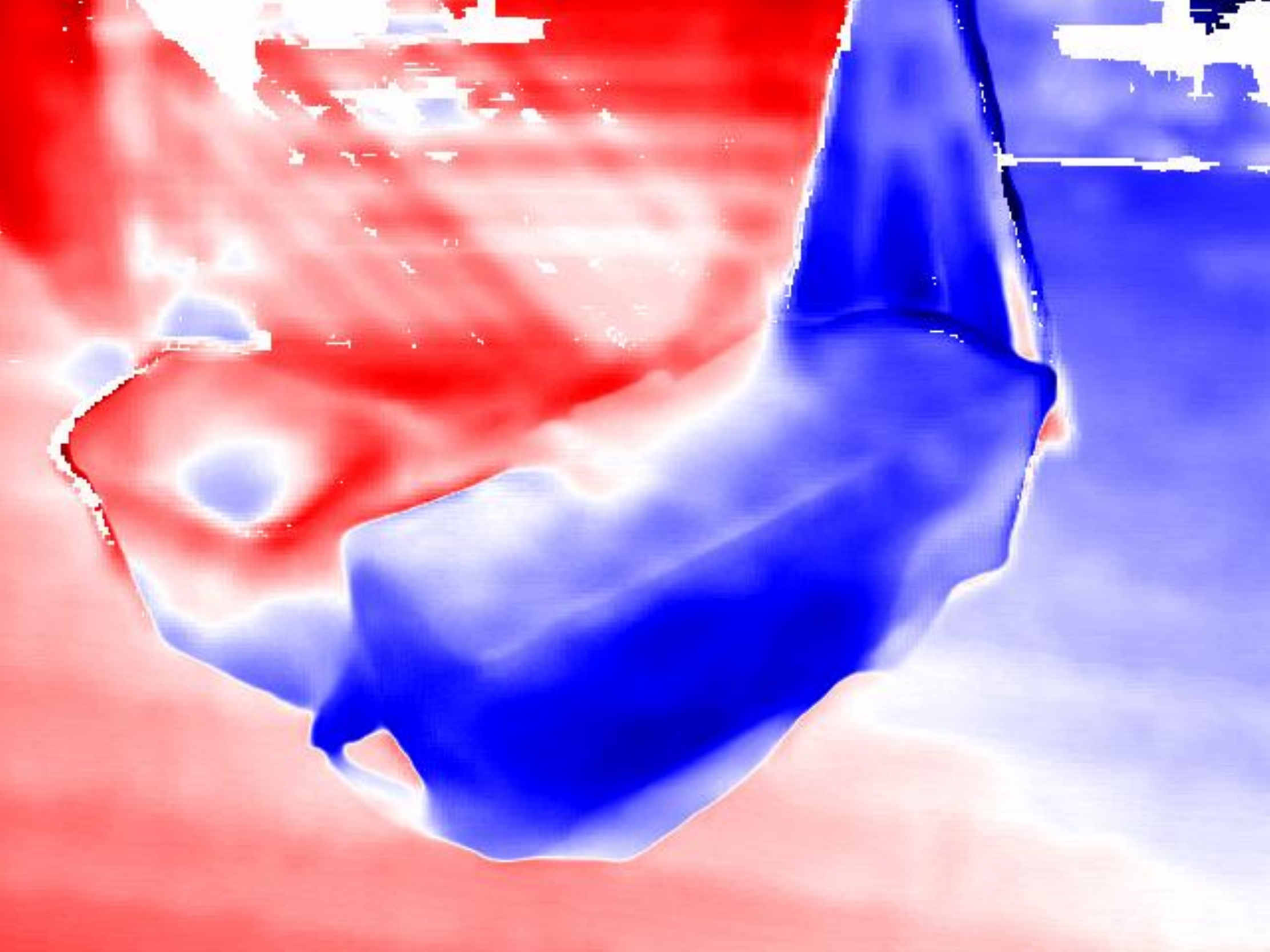}&
    \includegraphics[width=0.104\linewidth]{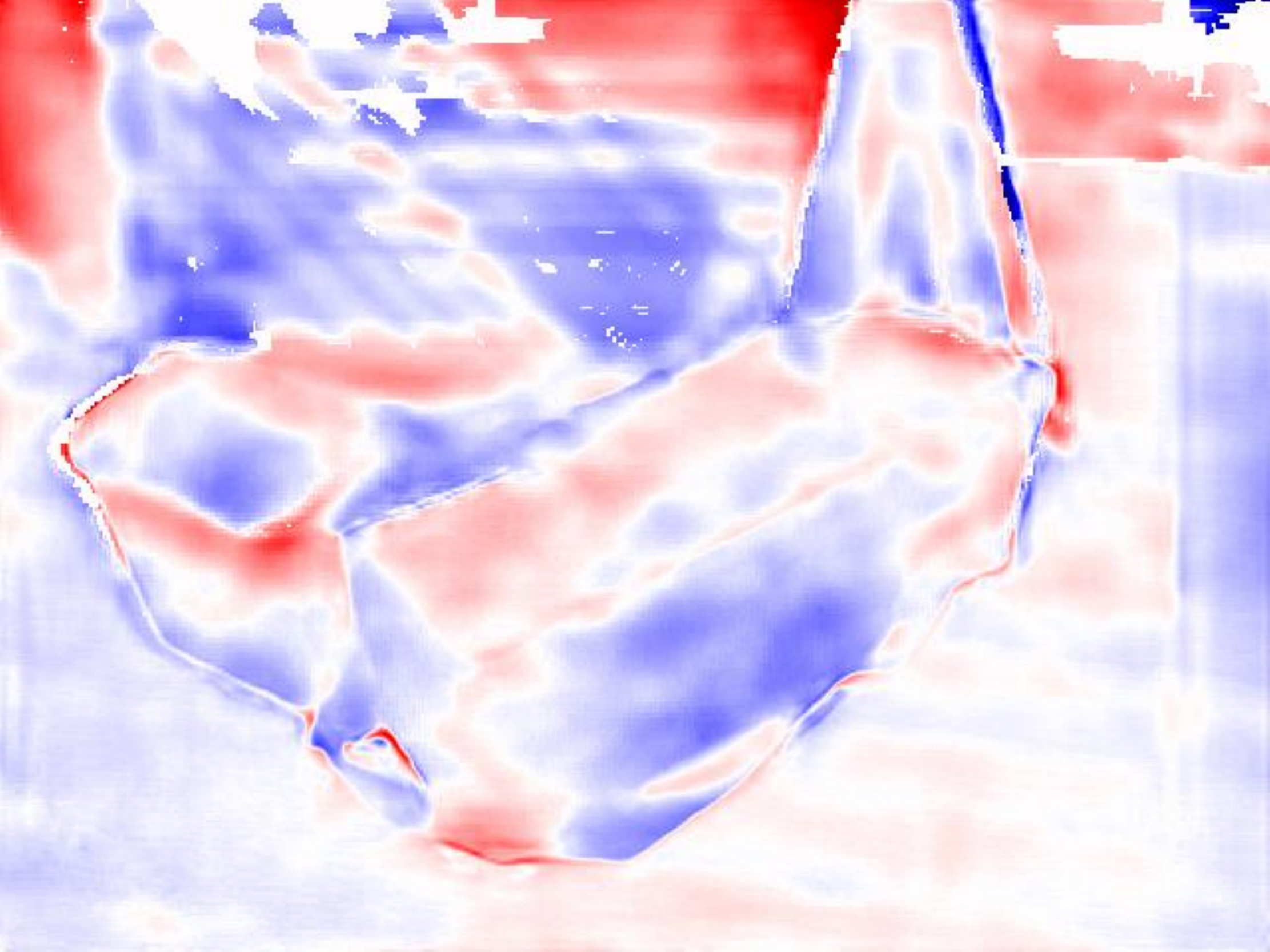}&
    \includegraphics[width=0.024\linewidth]{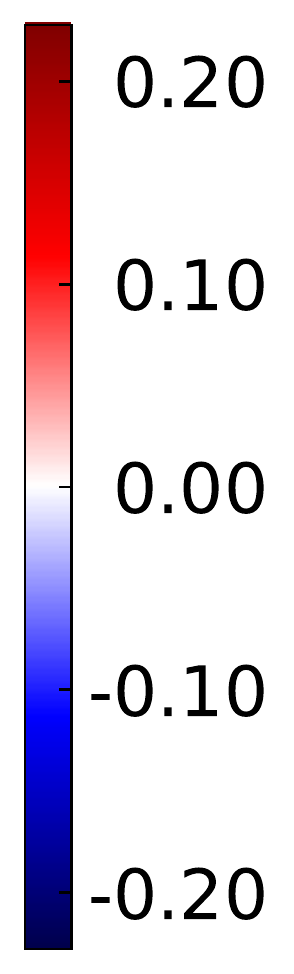}\\
    \vspace{-0.75mm}
    \scriptsize e. &
    \includegraphics[width=0.104\linewidth]{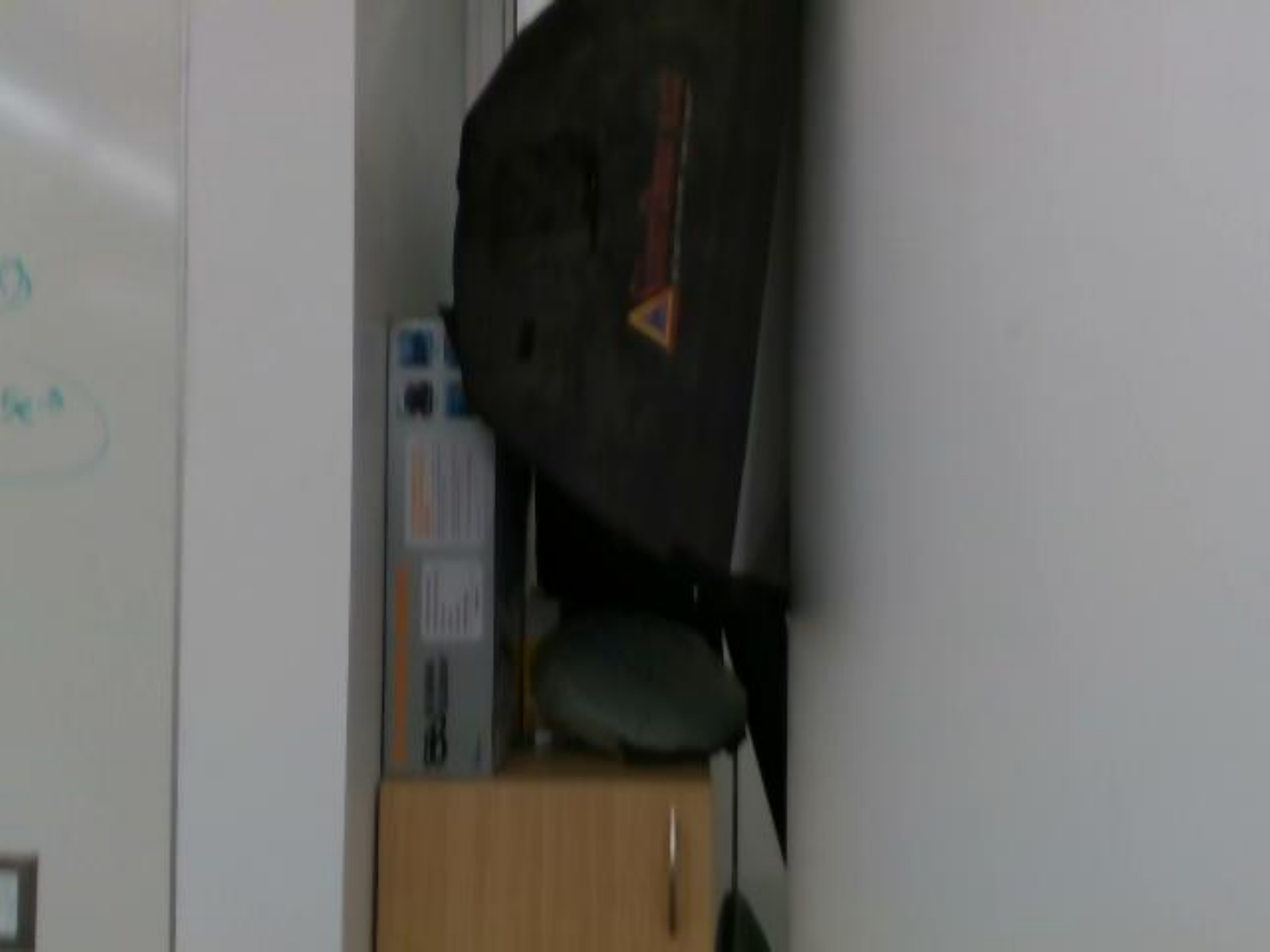}&
    \includegraphics[width=0.104\linewidth]{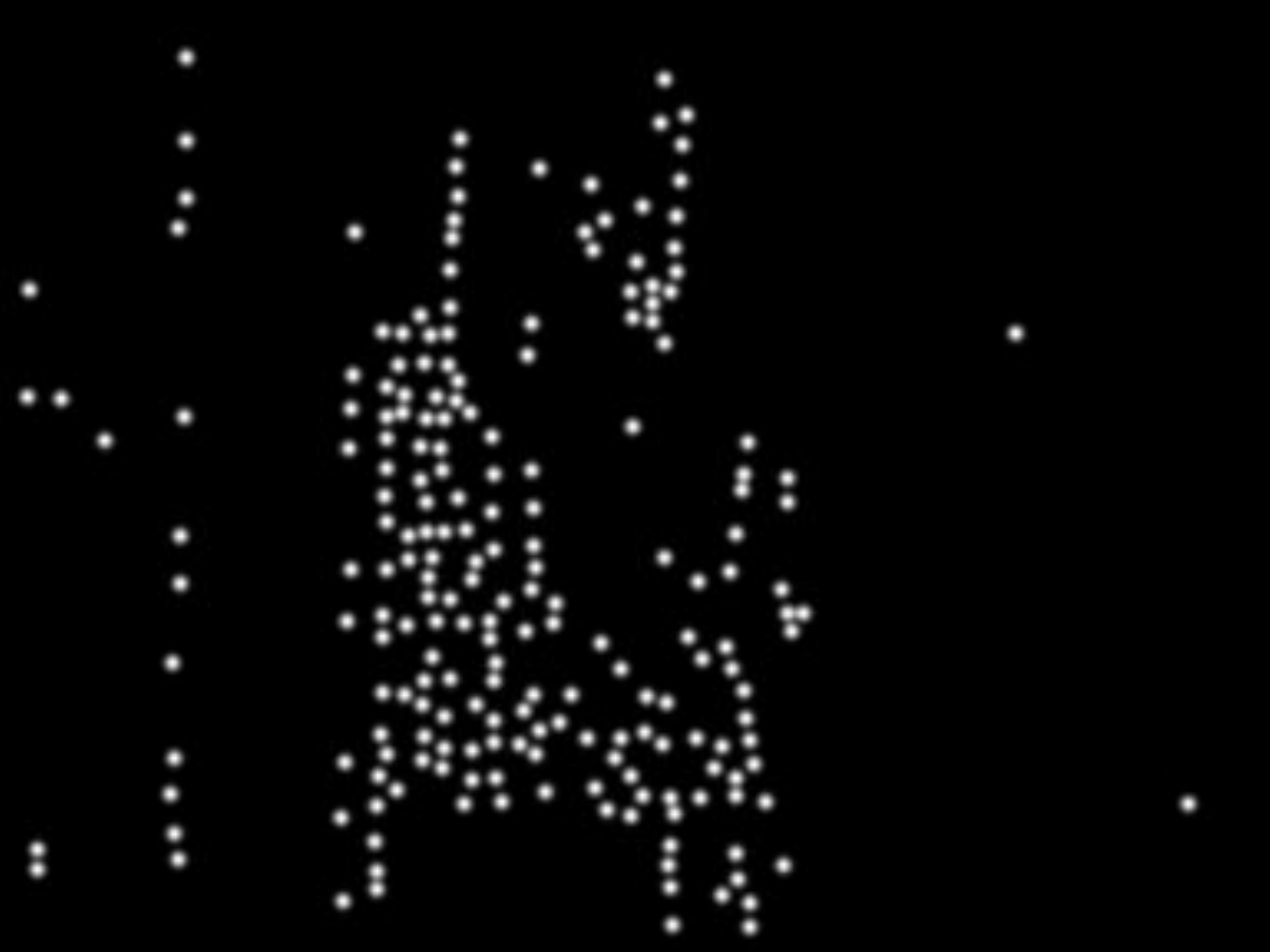}&
    \includegraphics[width=0.104\linewidth]{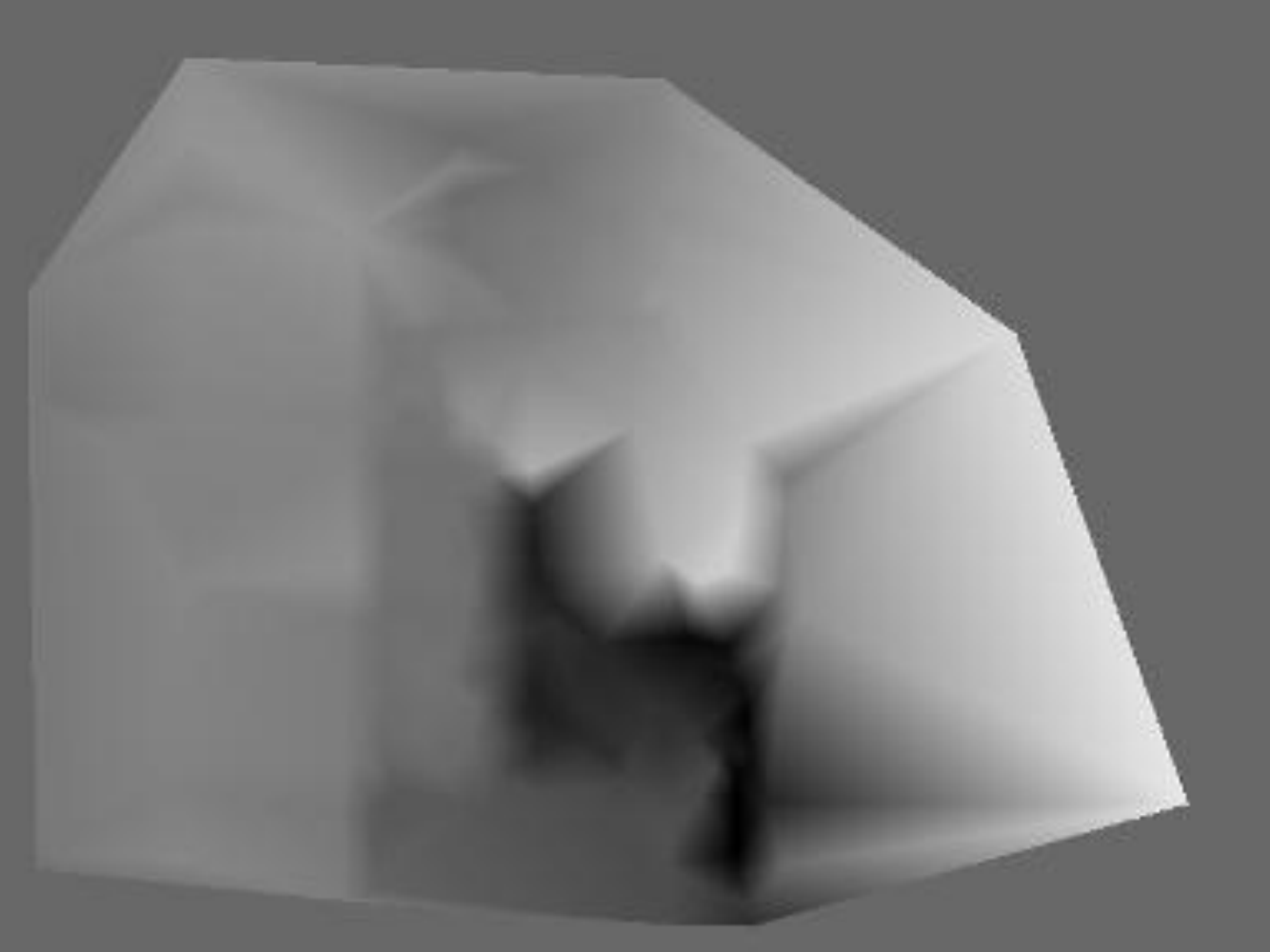}&
    \includegraphics[width=0.104\linewidth]{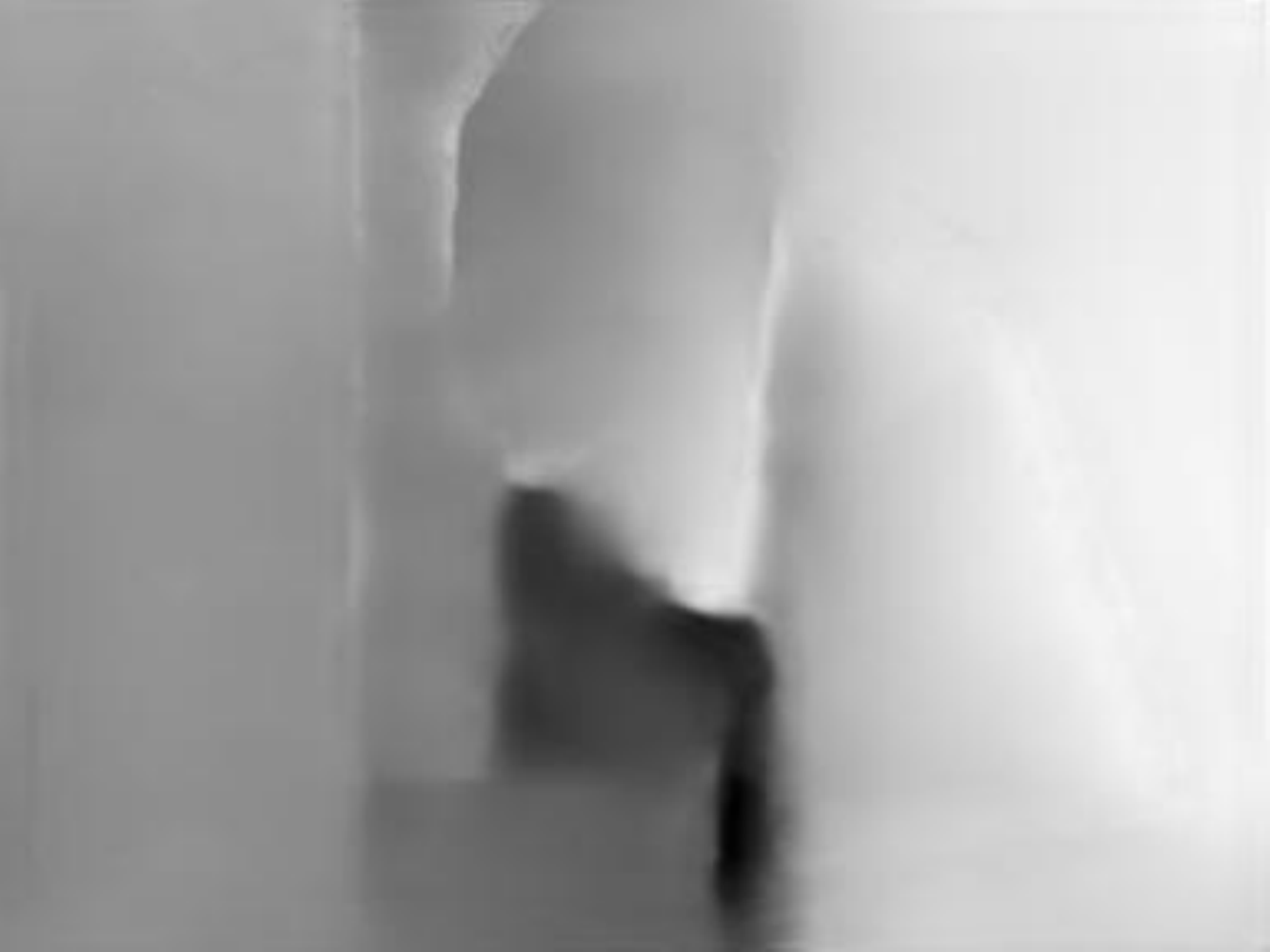}&
    \includegraphics[width=0.104\linewidth]{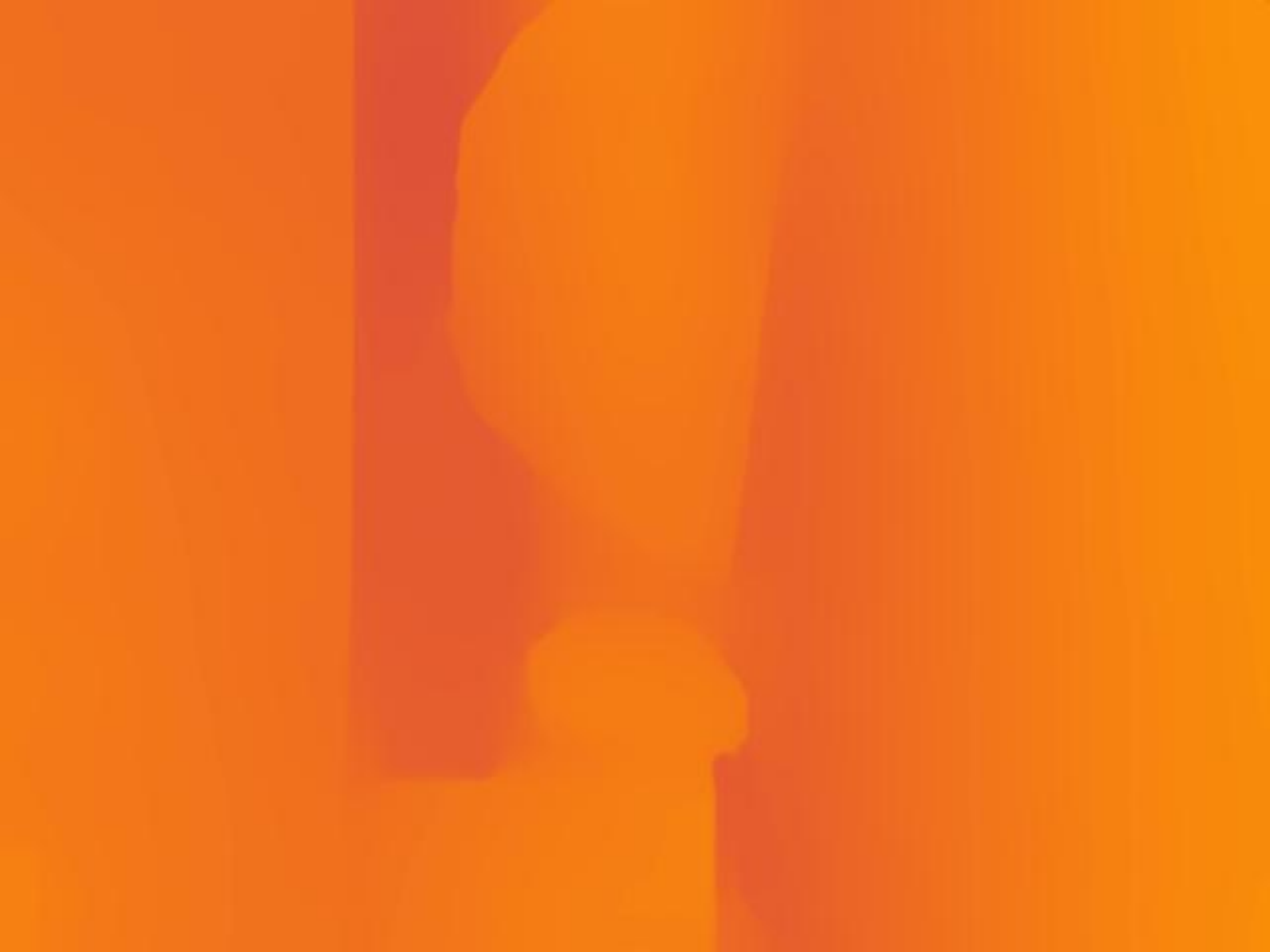}&
    \includegraphics[width=0.104\linewidth]{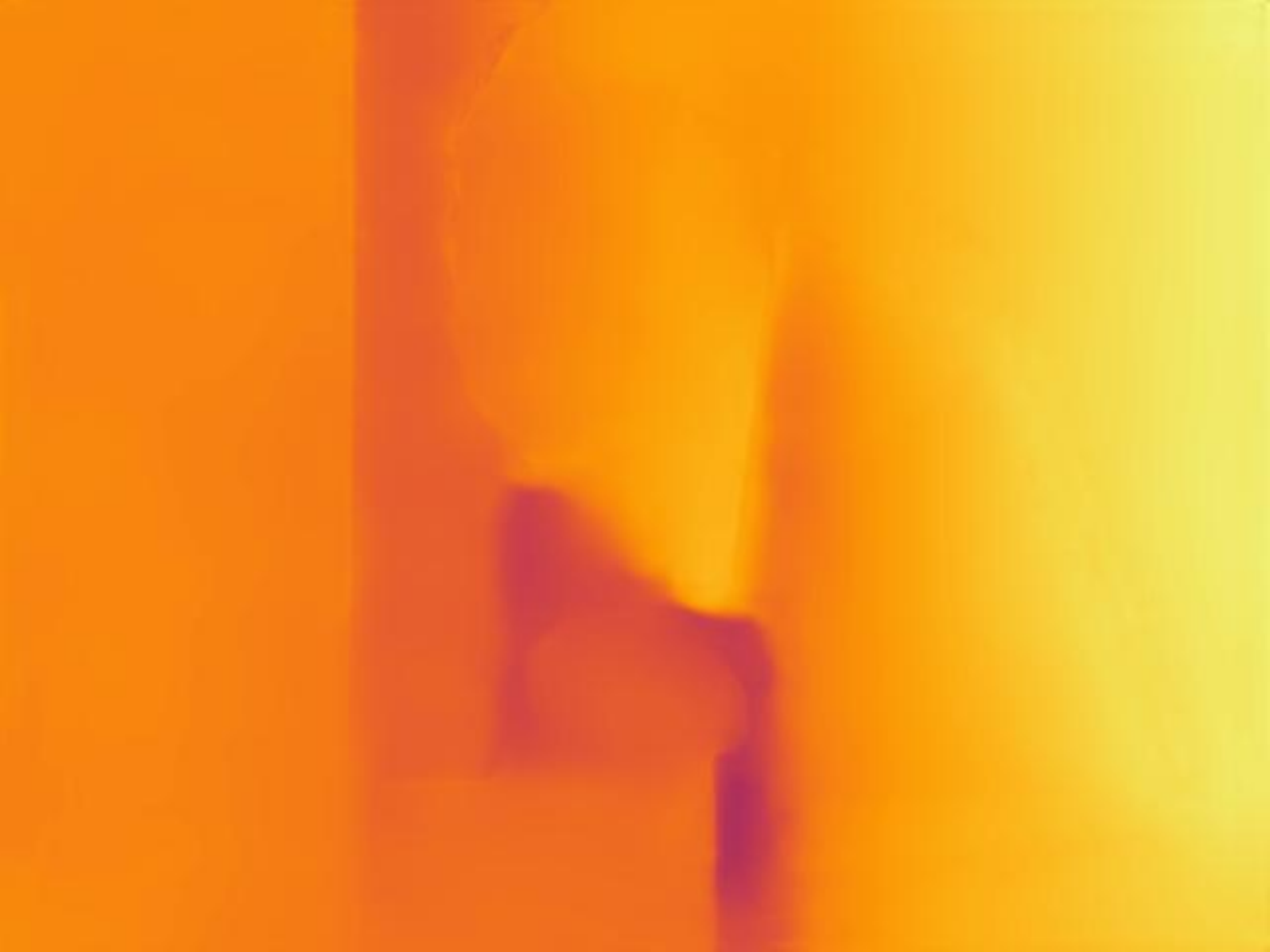}&
    \includegraphics[width=0.104\linewidth]{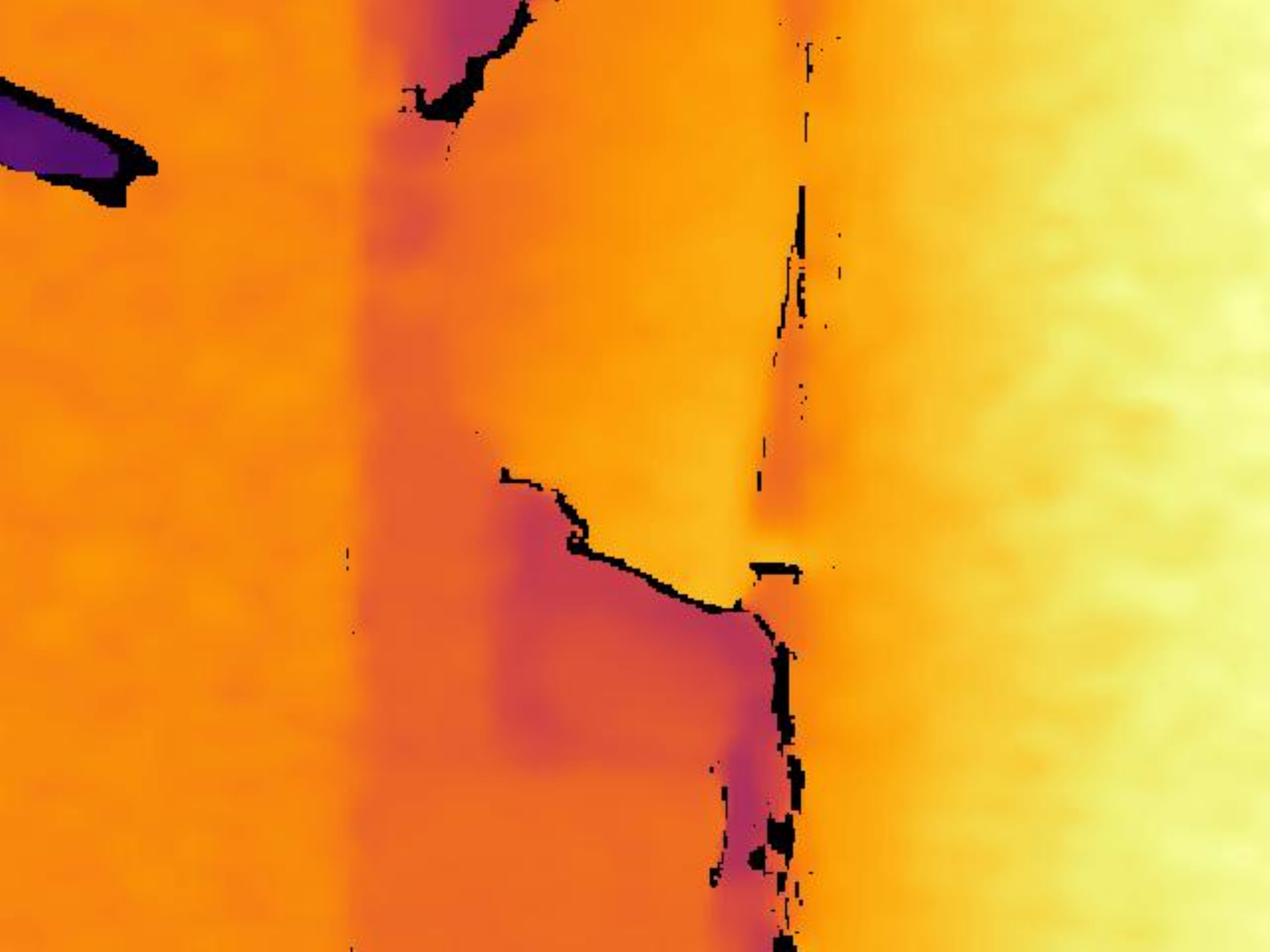}&
    \includegraphics[width=0.104\linewidth]{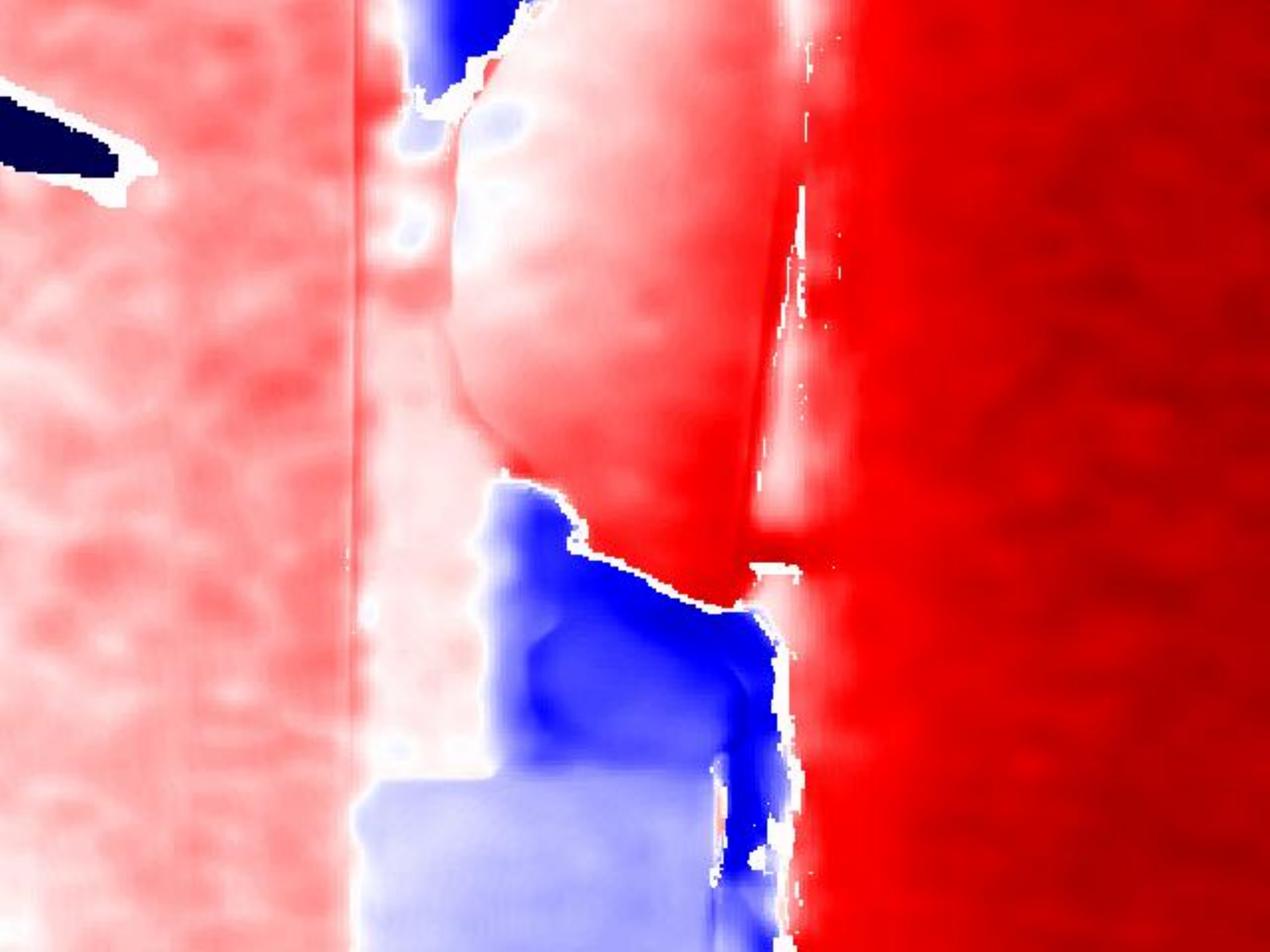}&
    \includegraphics[width=0.104\linewidth]{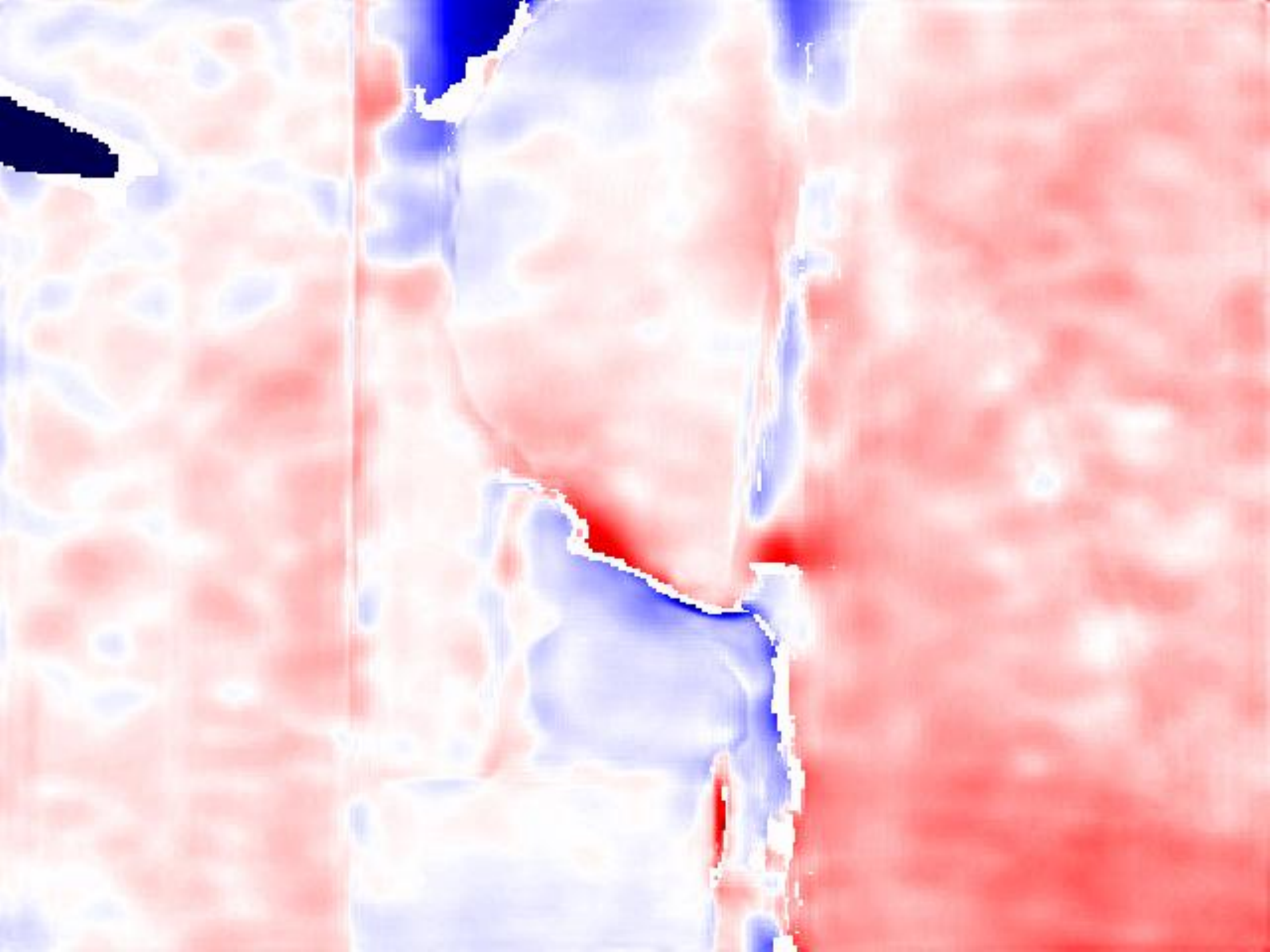}&
    \includegraphics[width=0.024\linewidth]{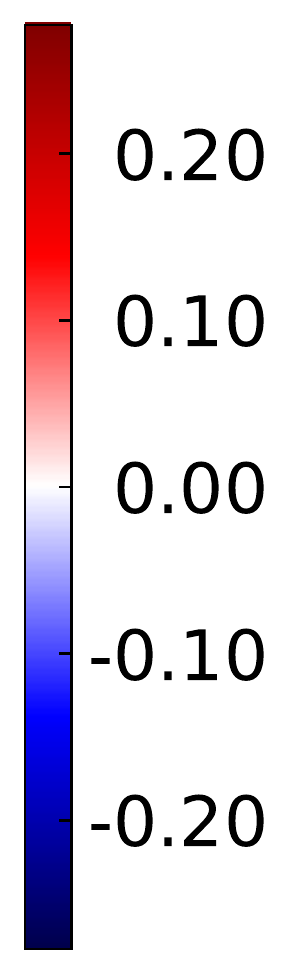}\\
    \vspace{-0.75mm}
    \scriptsize f. &
    \includegraphics[width=0.104\linewidth]{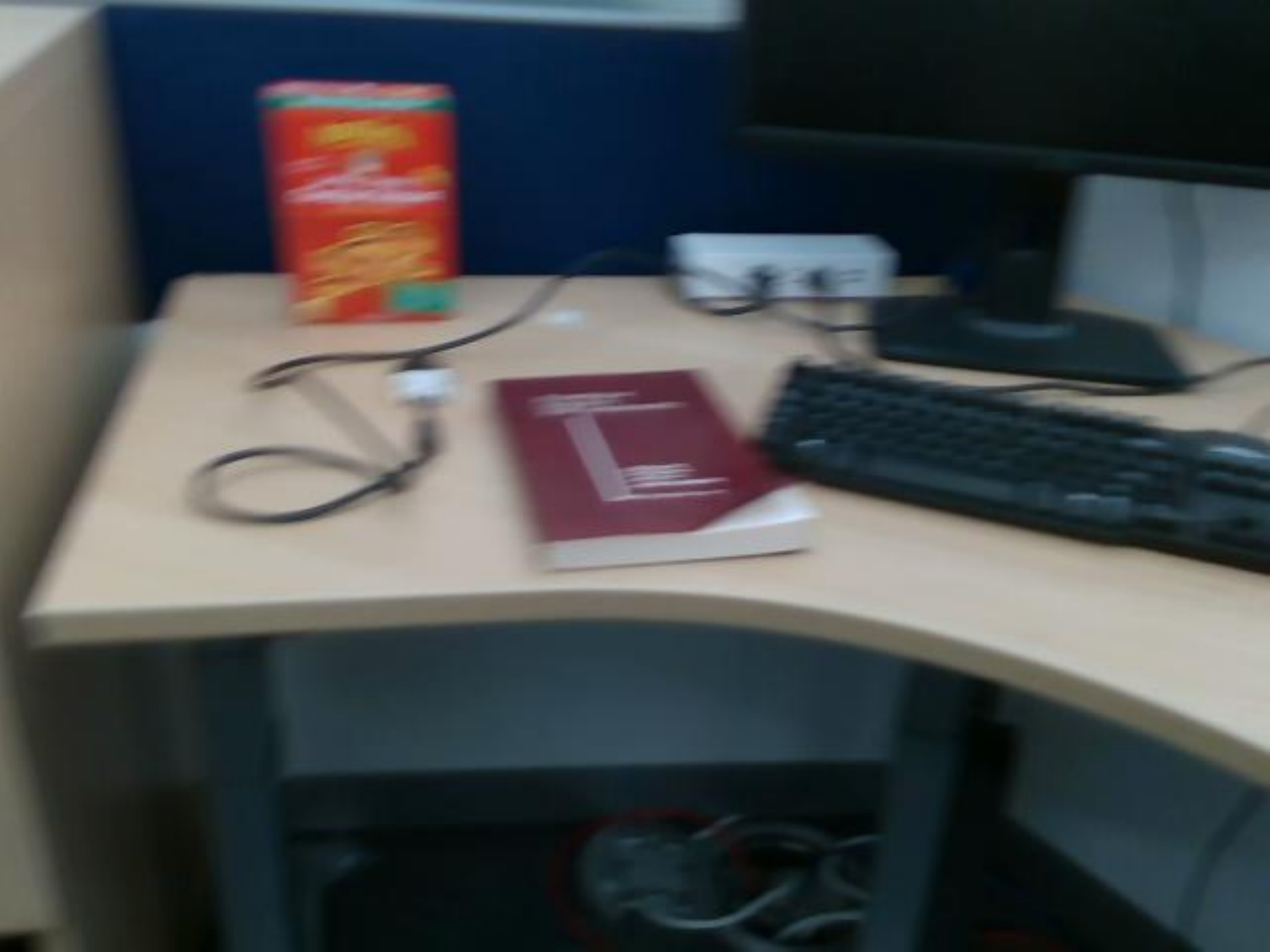}&
    \includegraphics[width=0.104\linewidth]{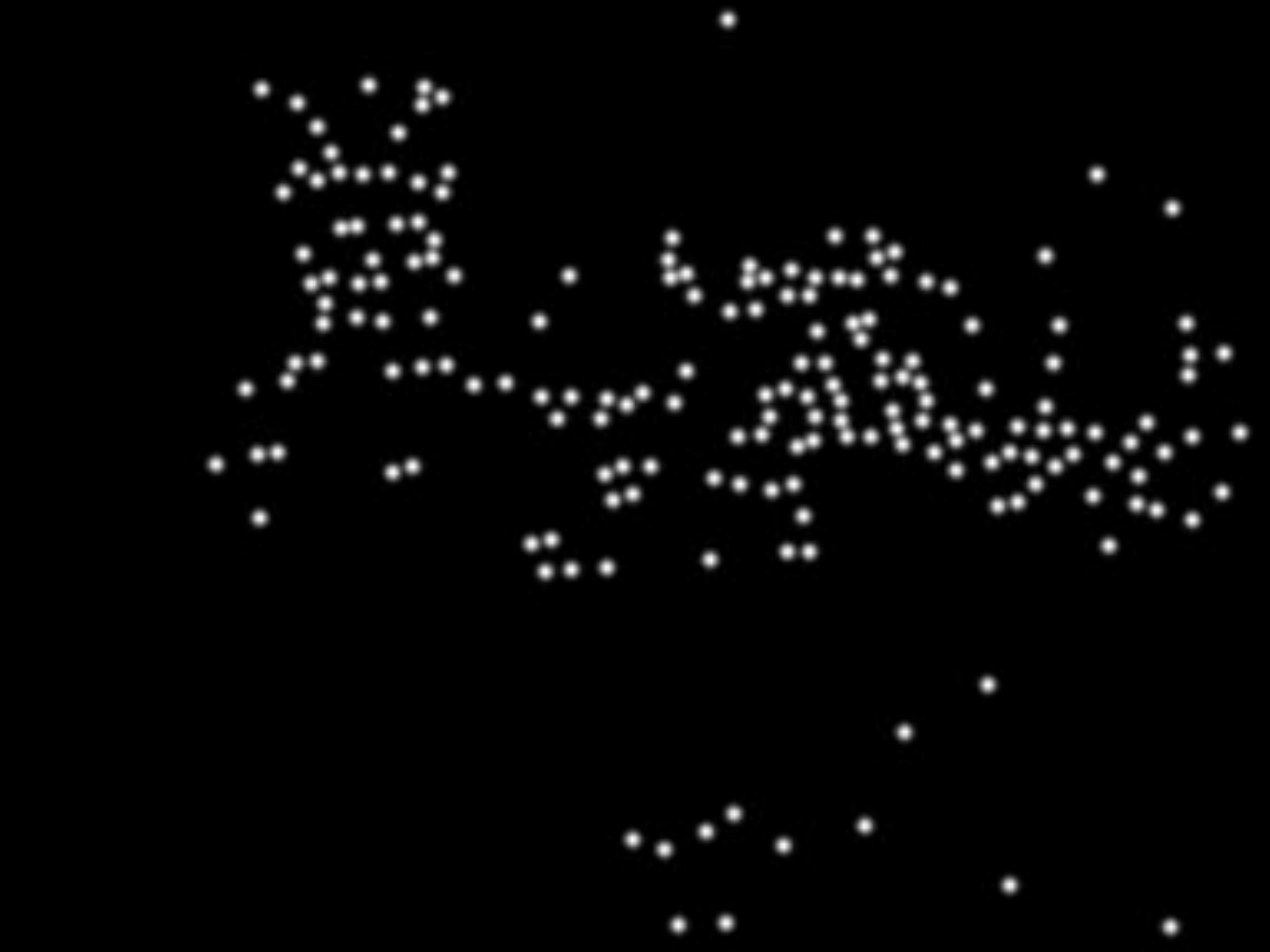}&
    \includegraphics[width=0.104\linewidth]{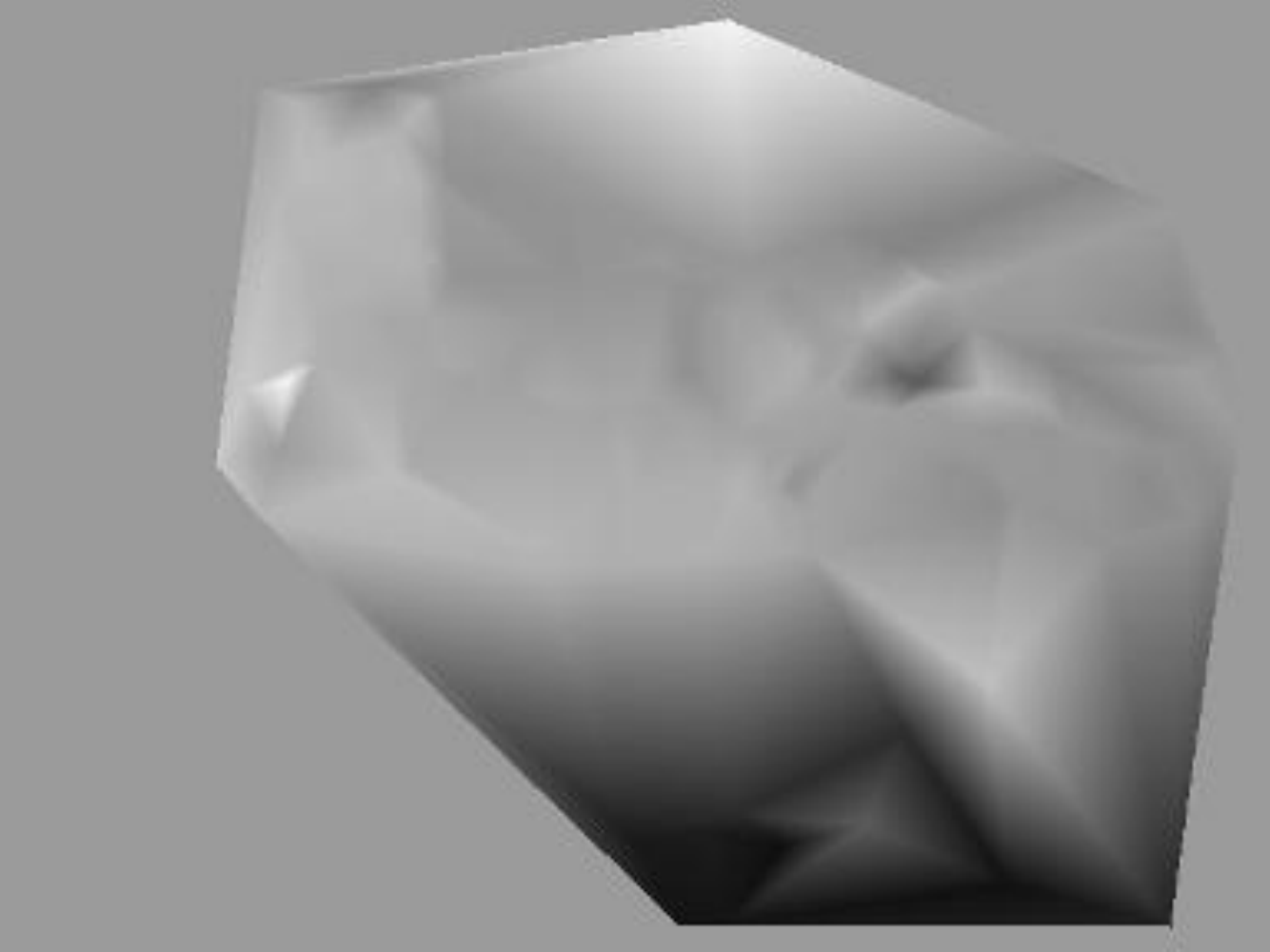}&
    \includegraphics[width=0.104\linewidth]{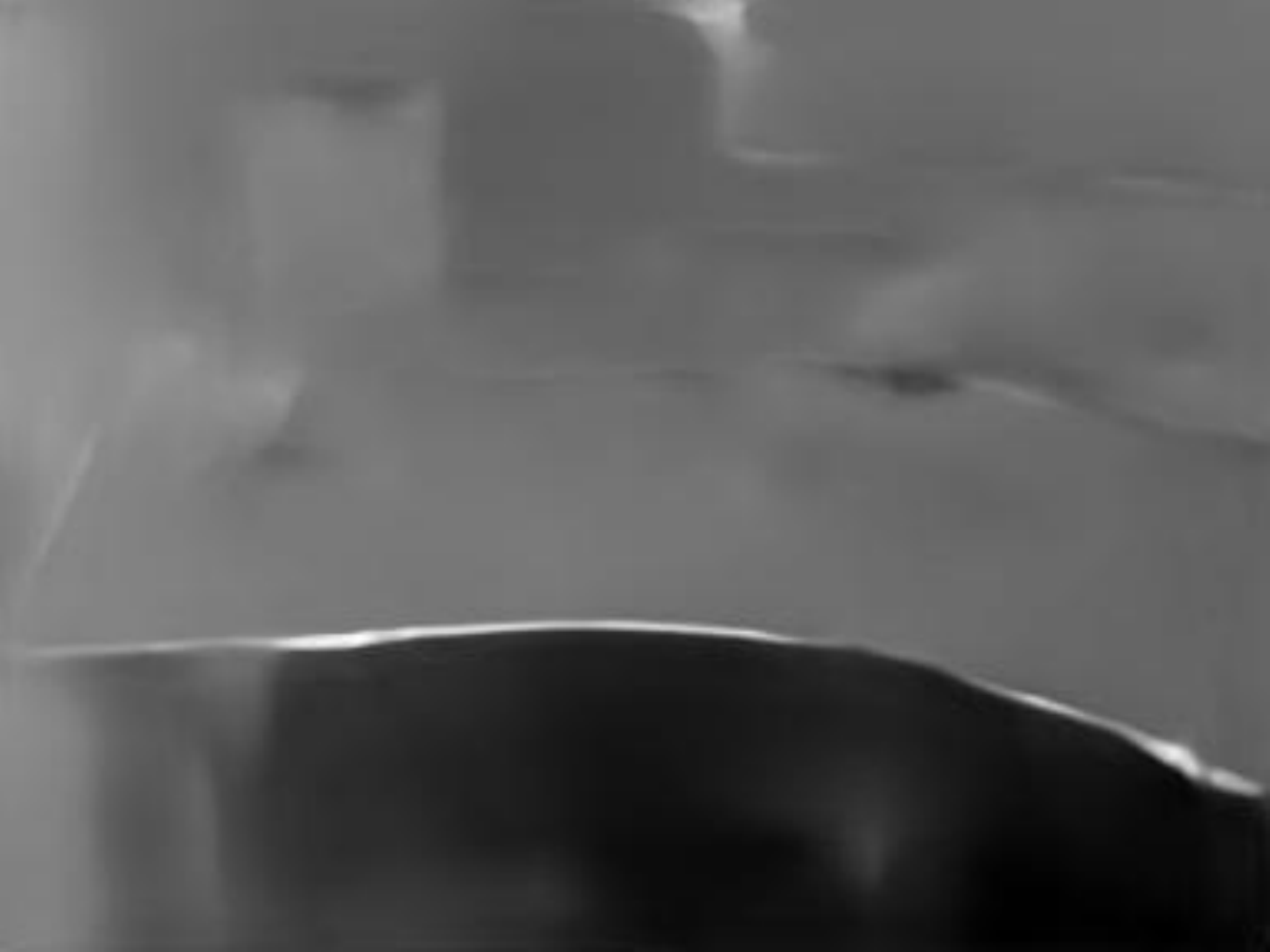}&
    \includegraphics[width=0.104\linewidth]{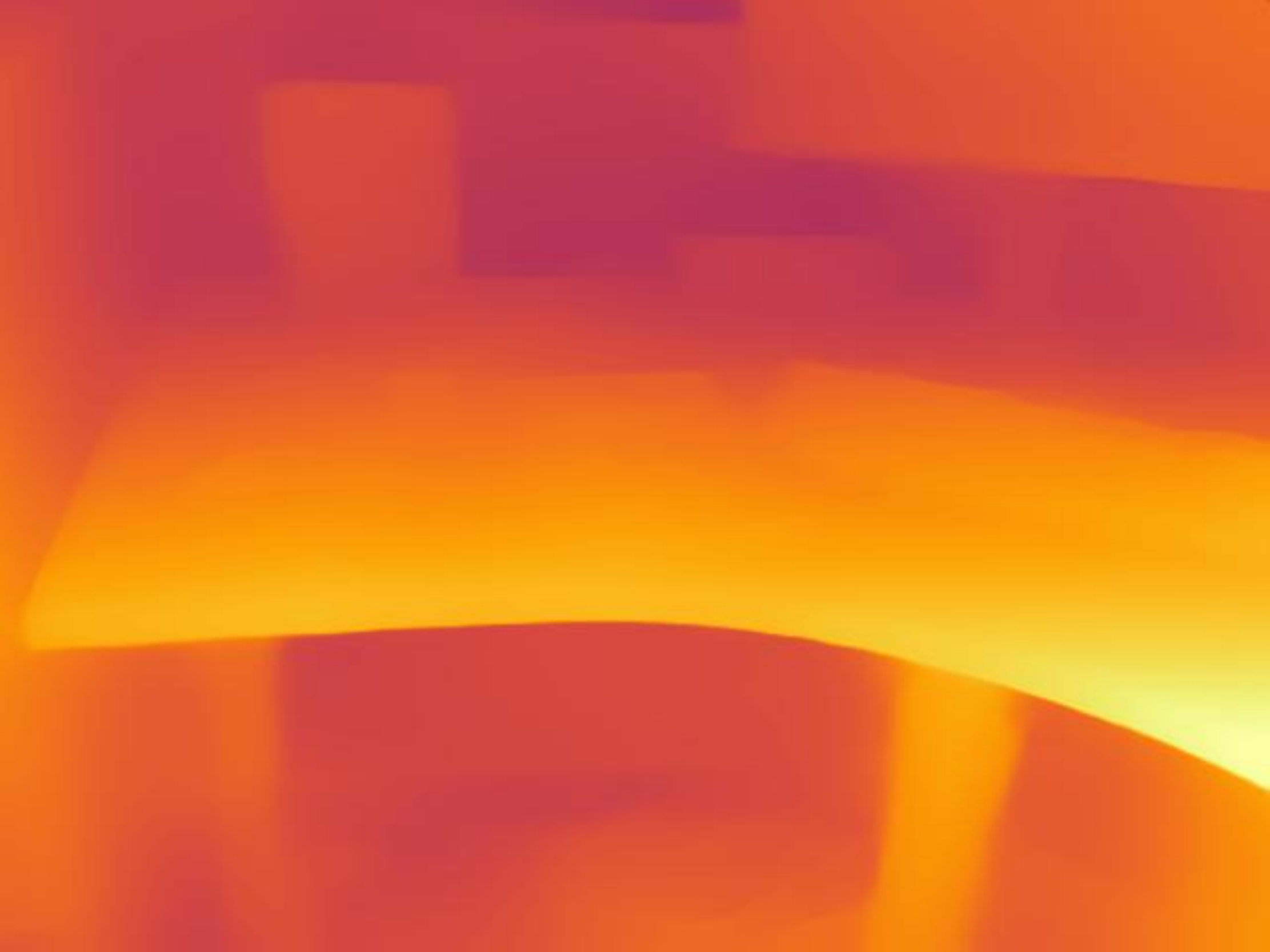}&
    \includegraphics[width=0.104\linewidth]{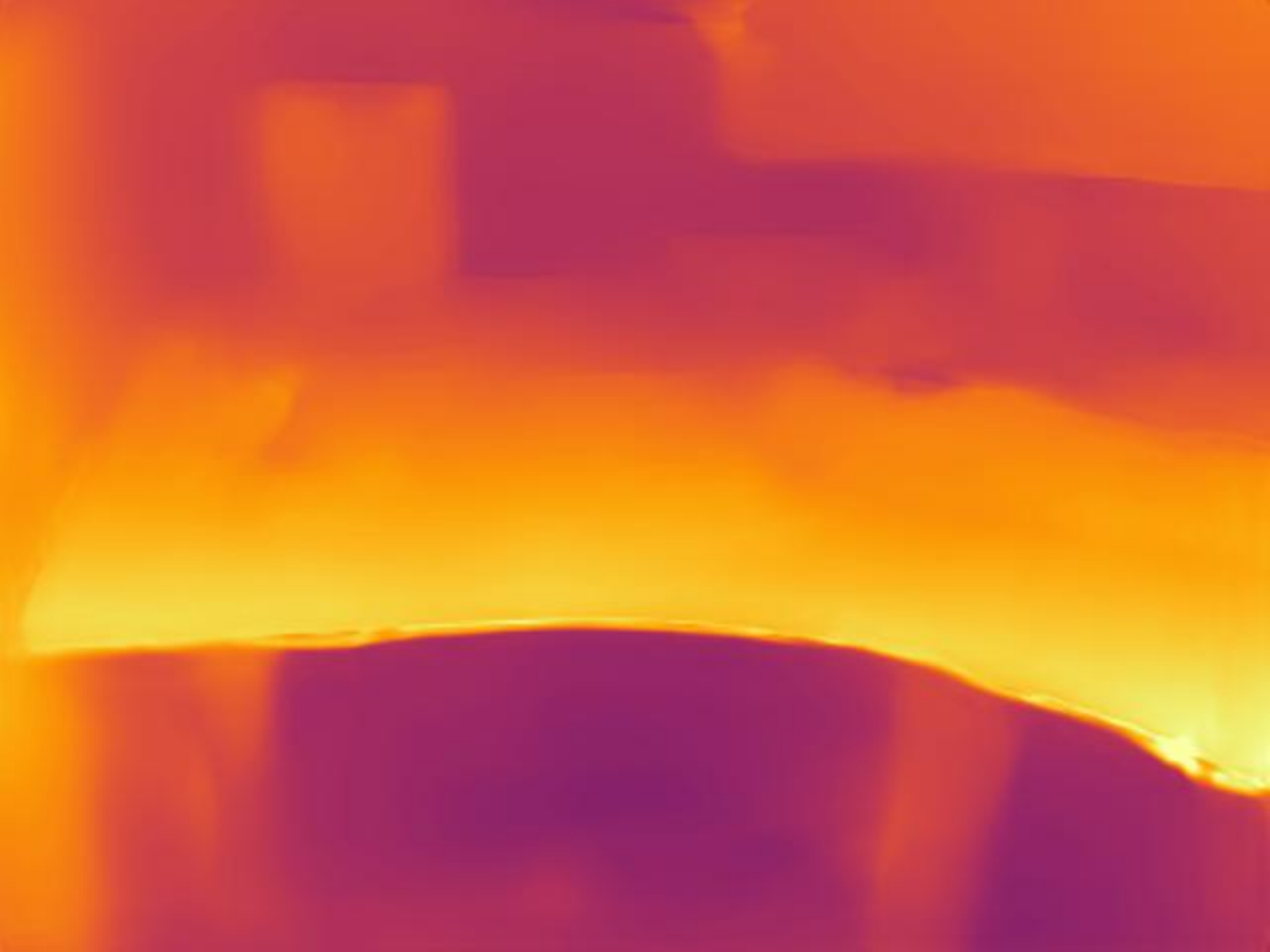}&
    \includegraphics[width=0.104\linewidth]{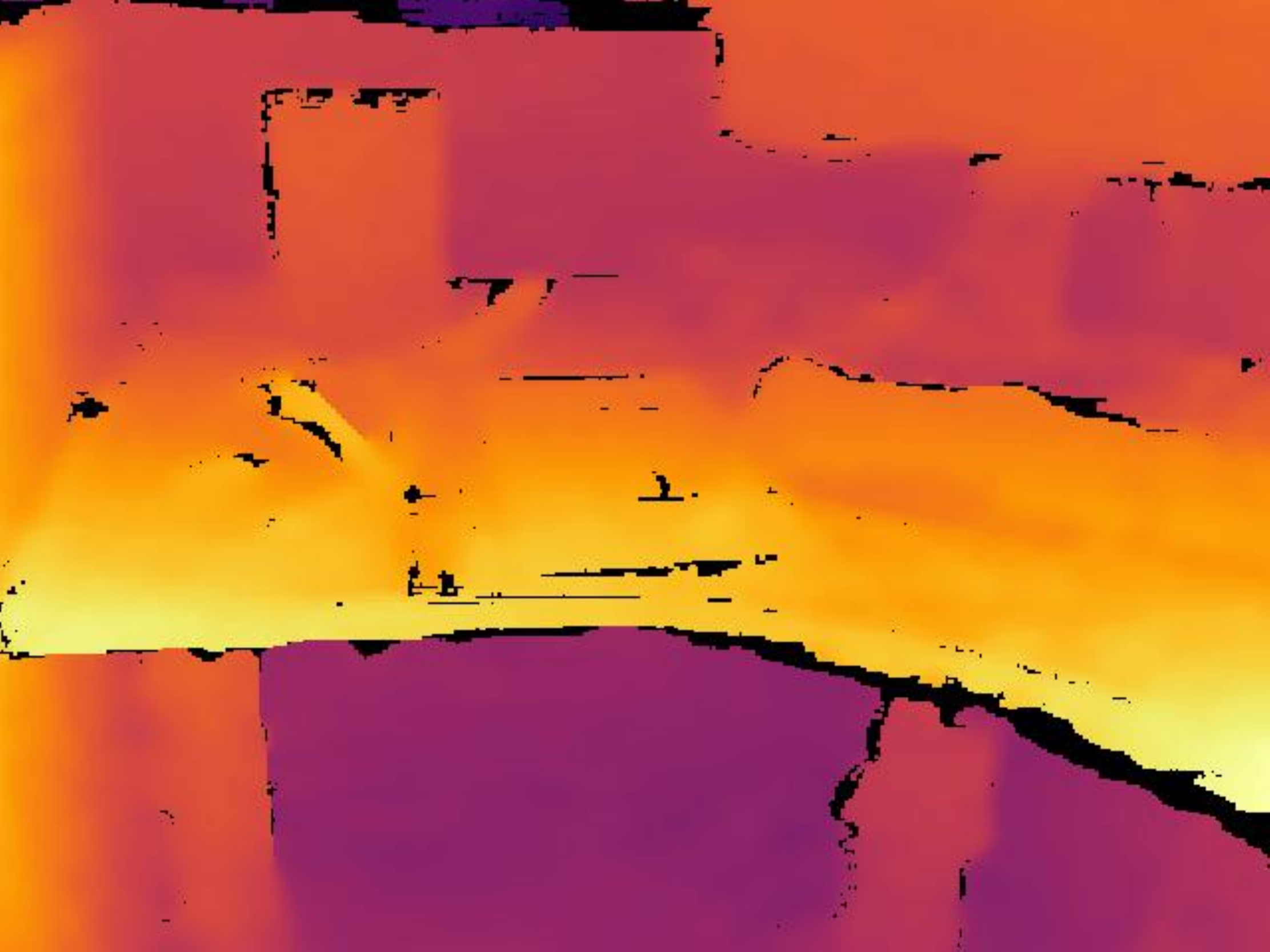}&
    \includegraphics[width=0.104\linewidth]{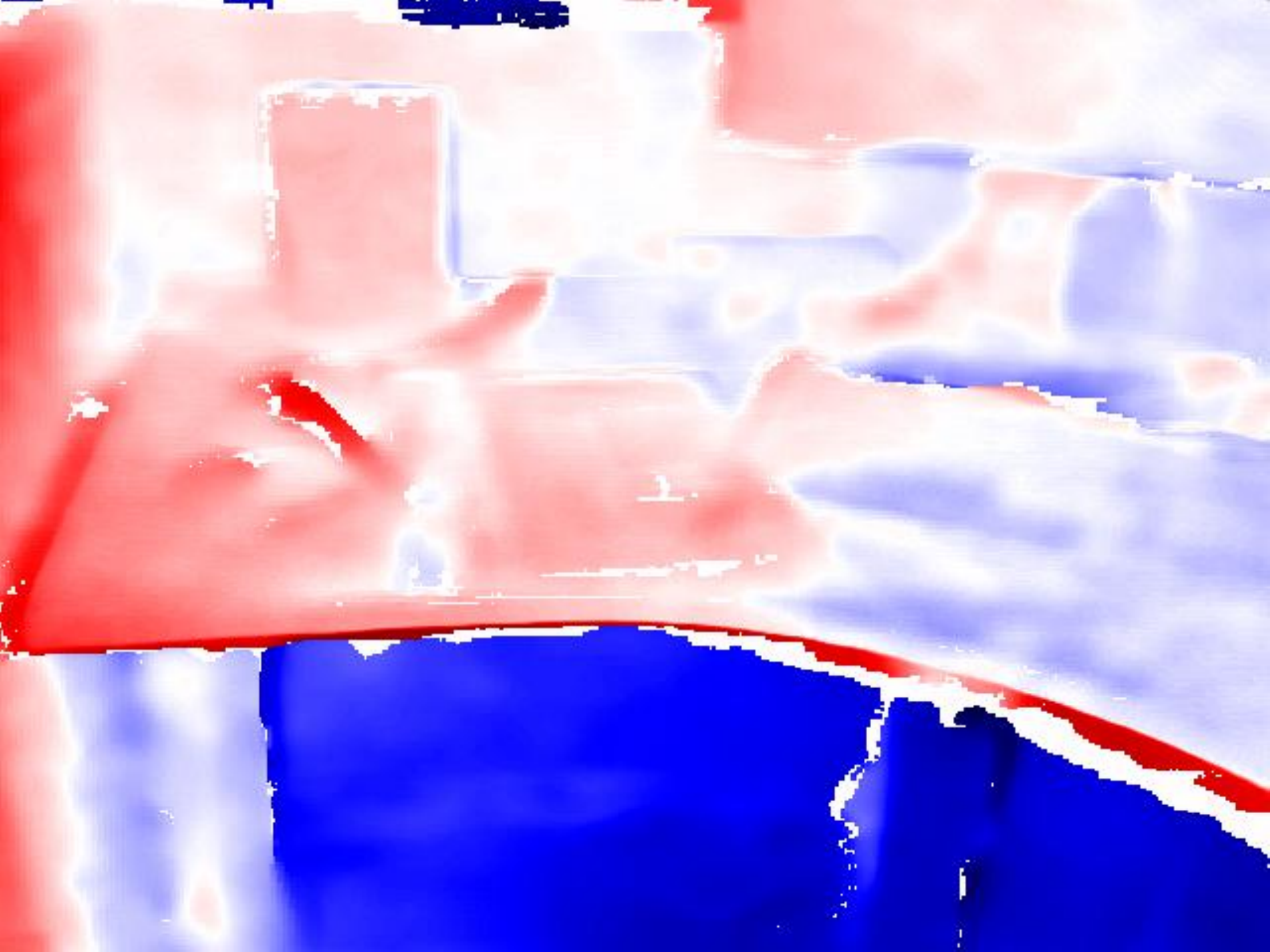}&
    \includegraphics[width=0.104\linewidth]{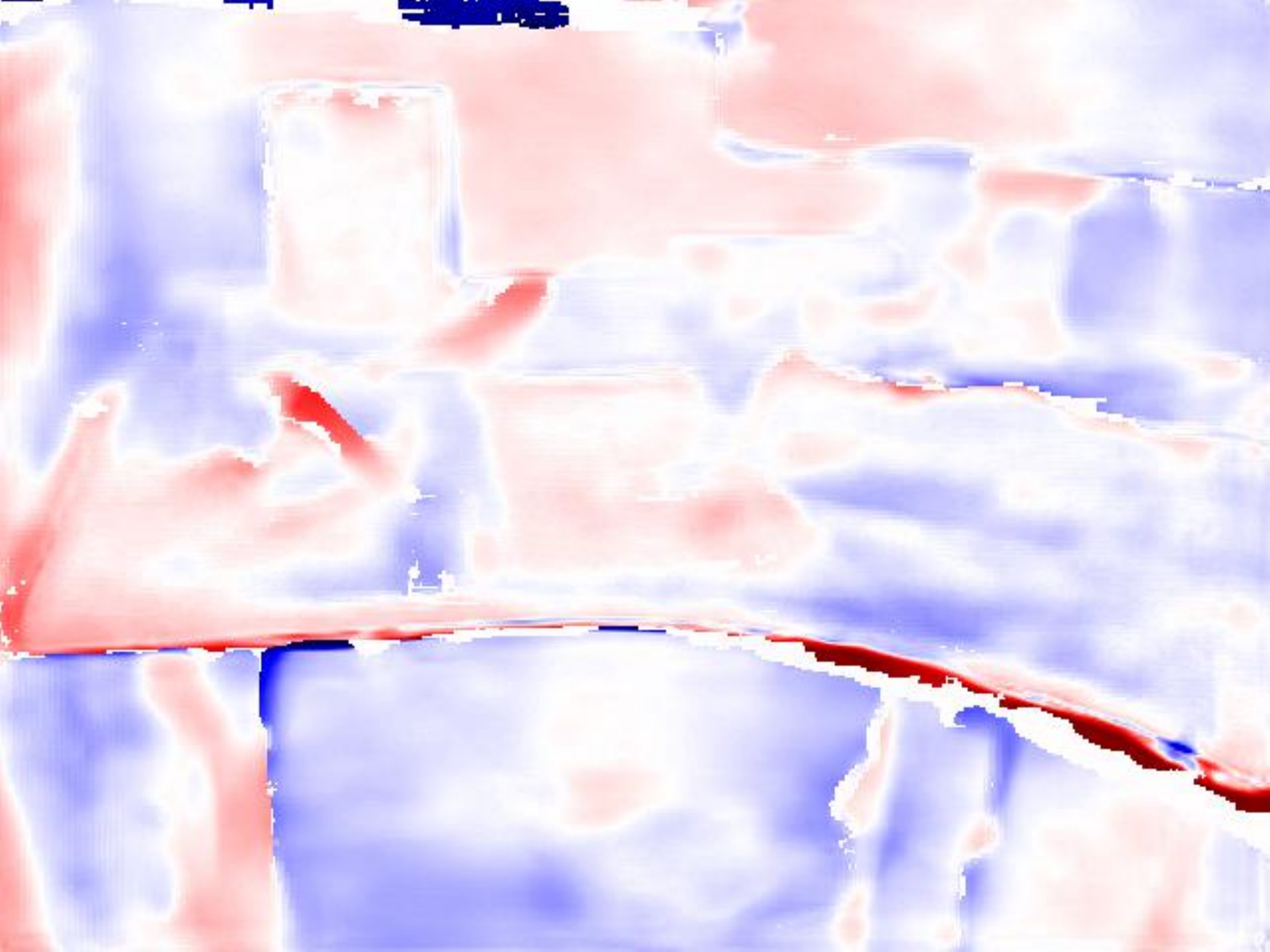}&
    \includegraphics[width=0.024\linewidth]{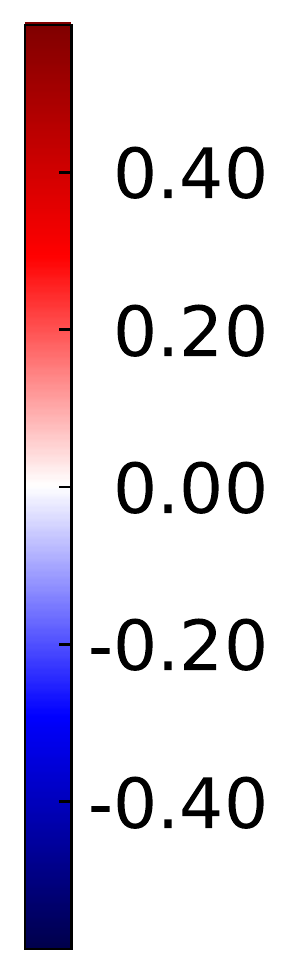}\\
    \vspace{-0.75mm}
    \scriptsize g. &
    \includegraphics[width=0.104\linewidth]{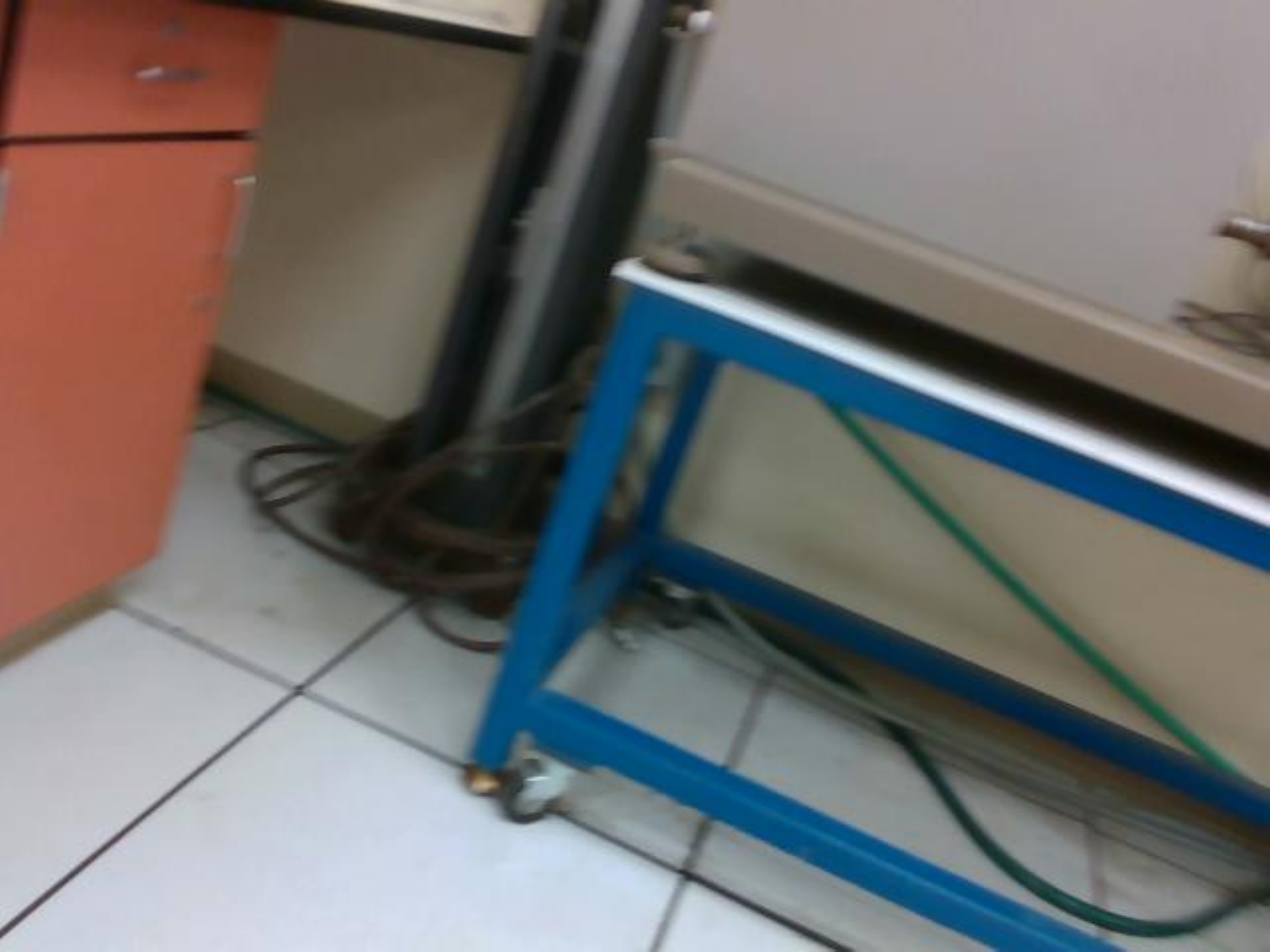}&
    \includegraphics[width=0.104\linewidth]{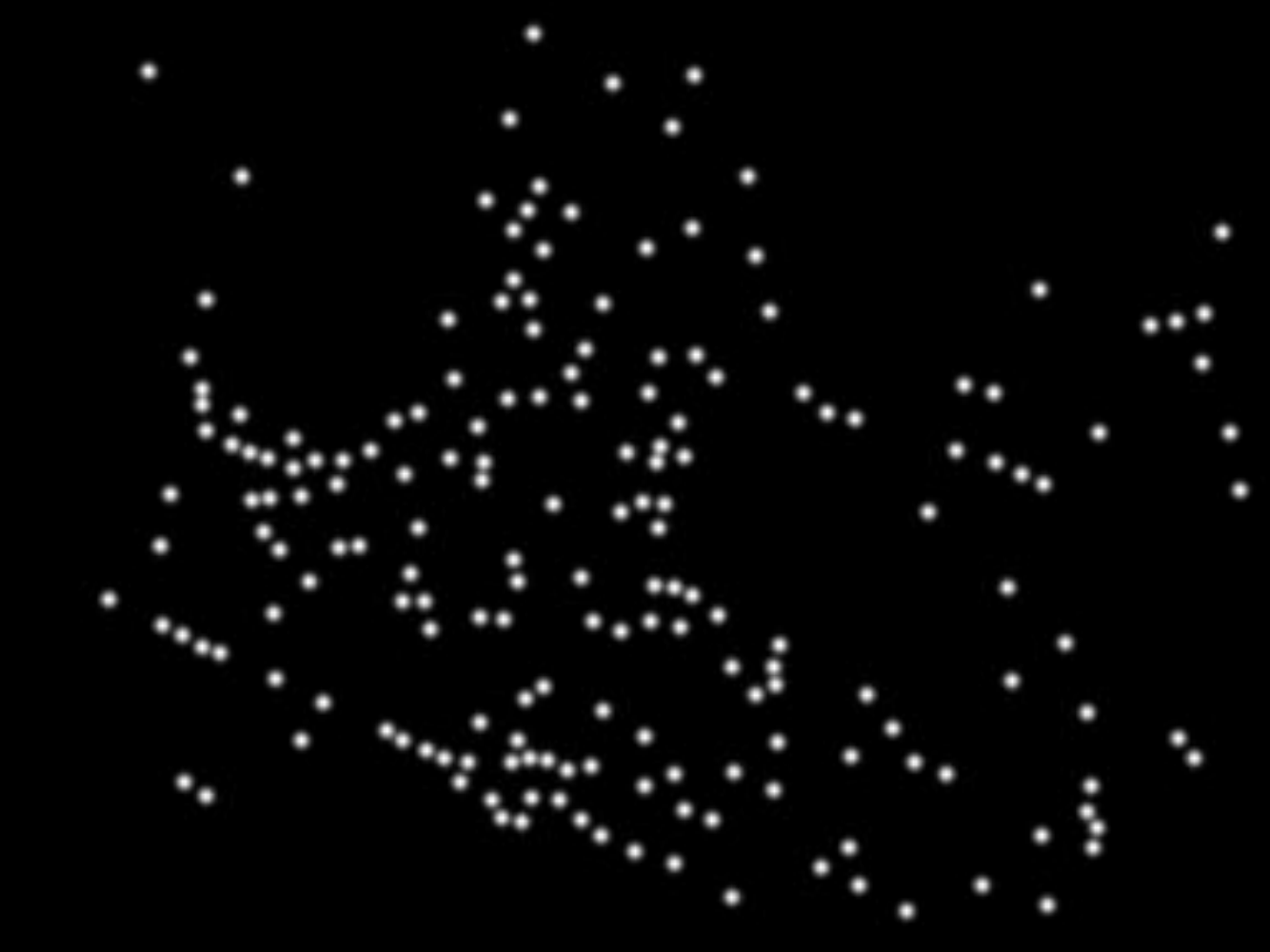}&
    \includegraphics[width=0.104\linewidth]{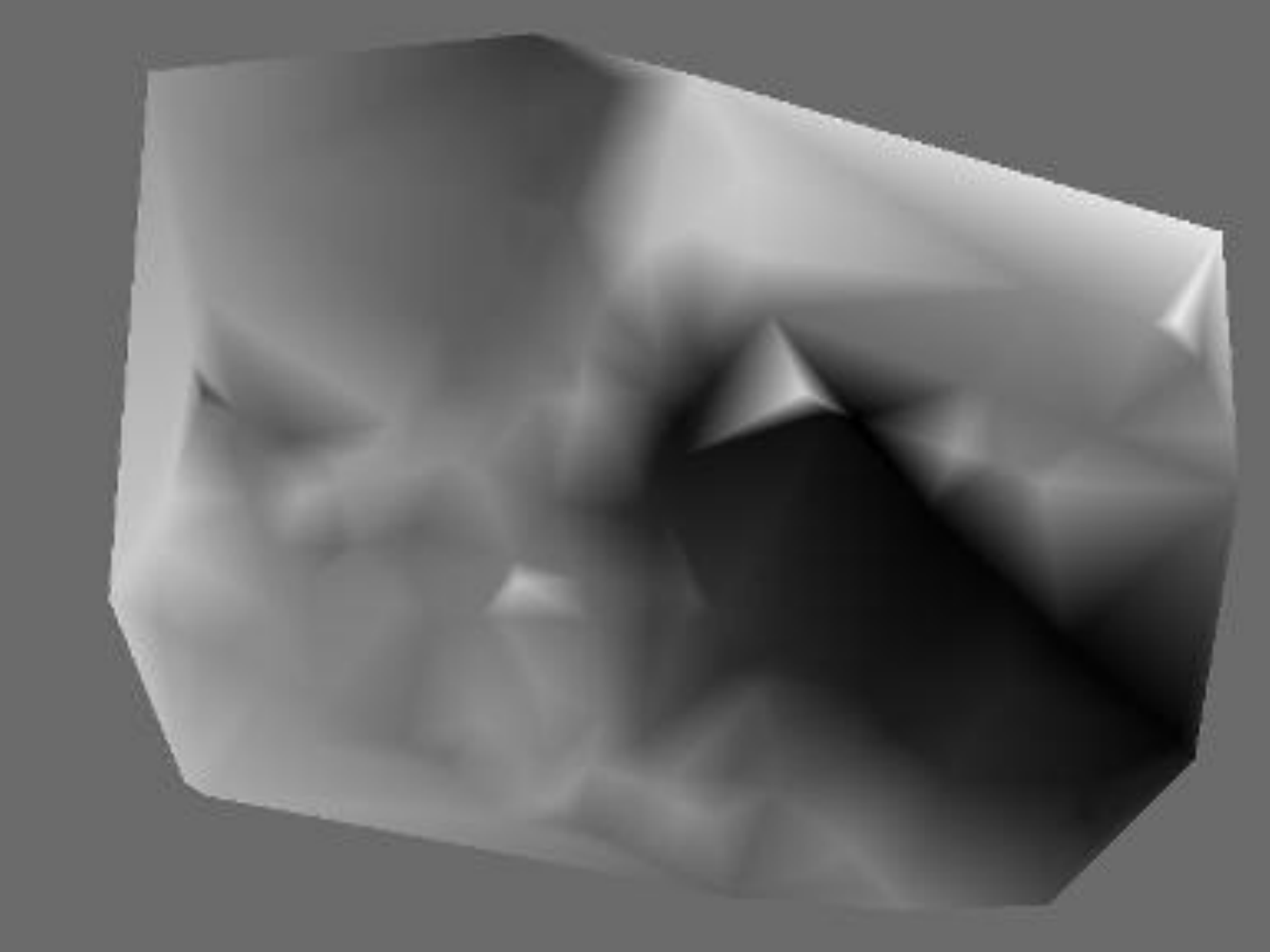}&
    \includegraphics[width=0.104\linewidth]{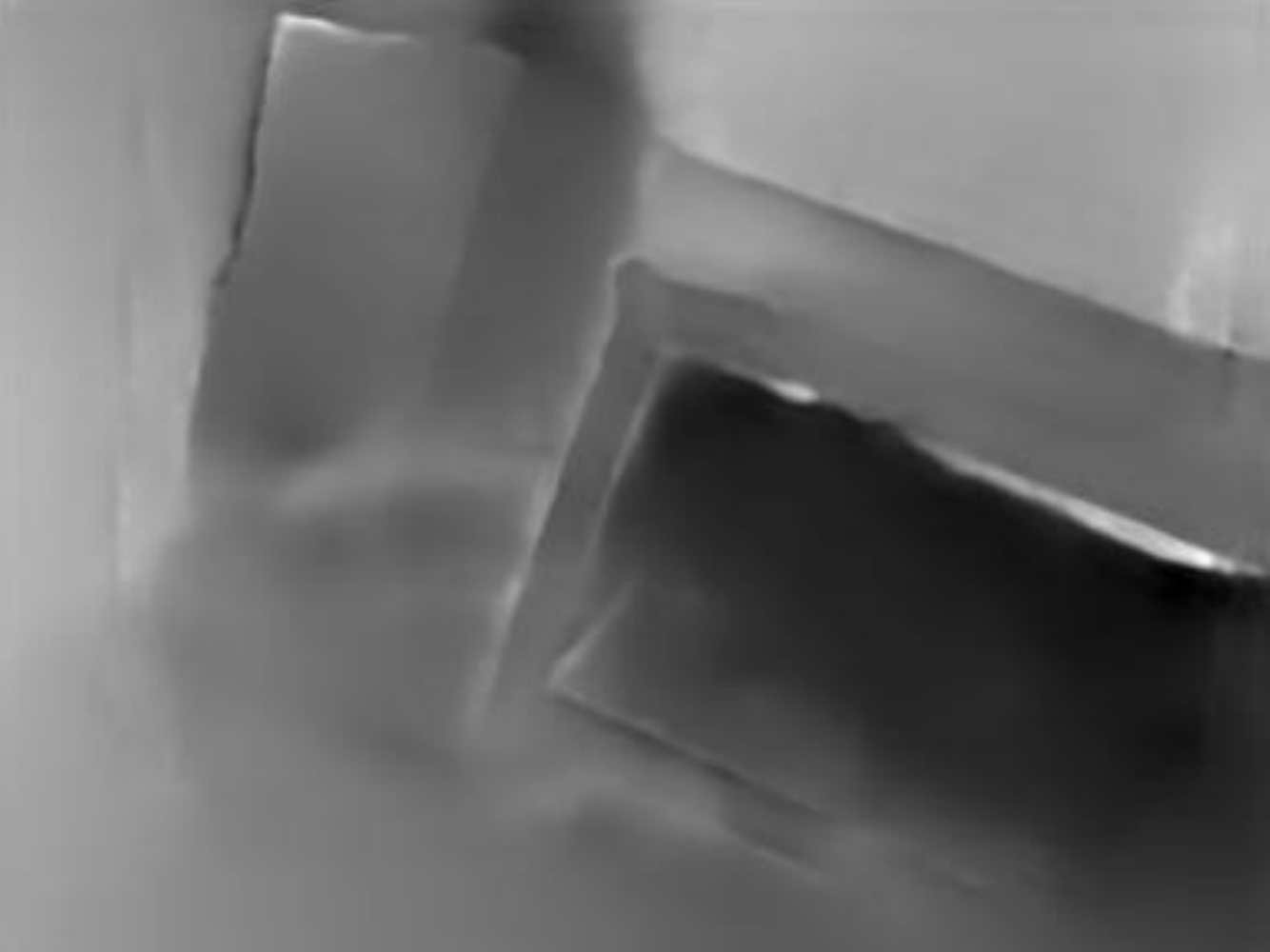}&
    \includegraphics[width=0.104\linewidth]{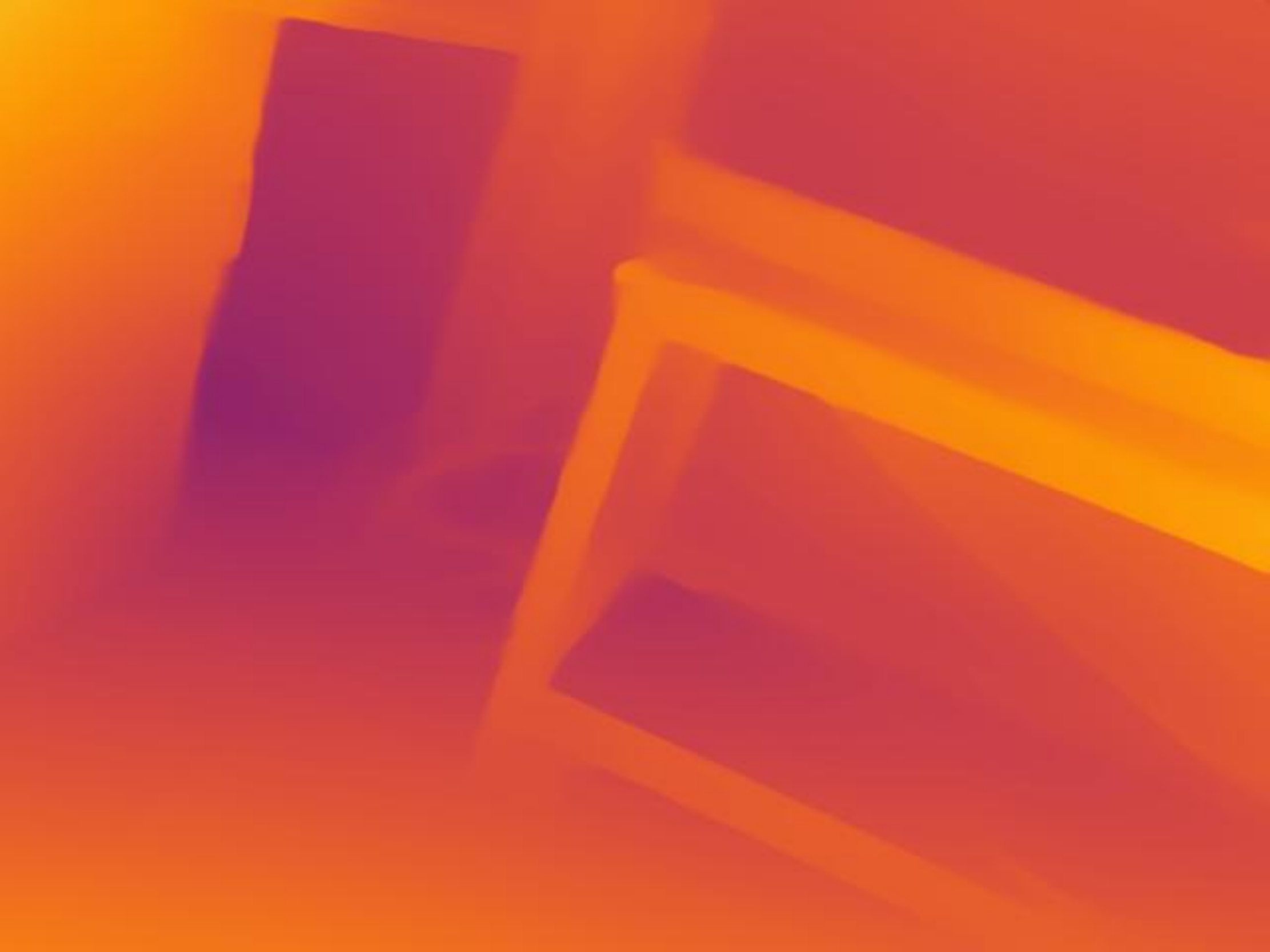}&
    \includegraphics[width=0.104\linewidth]{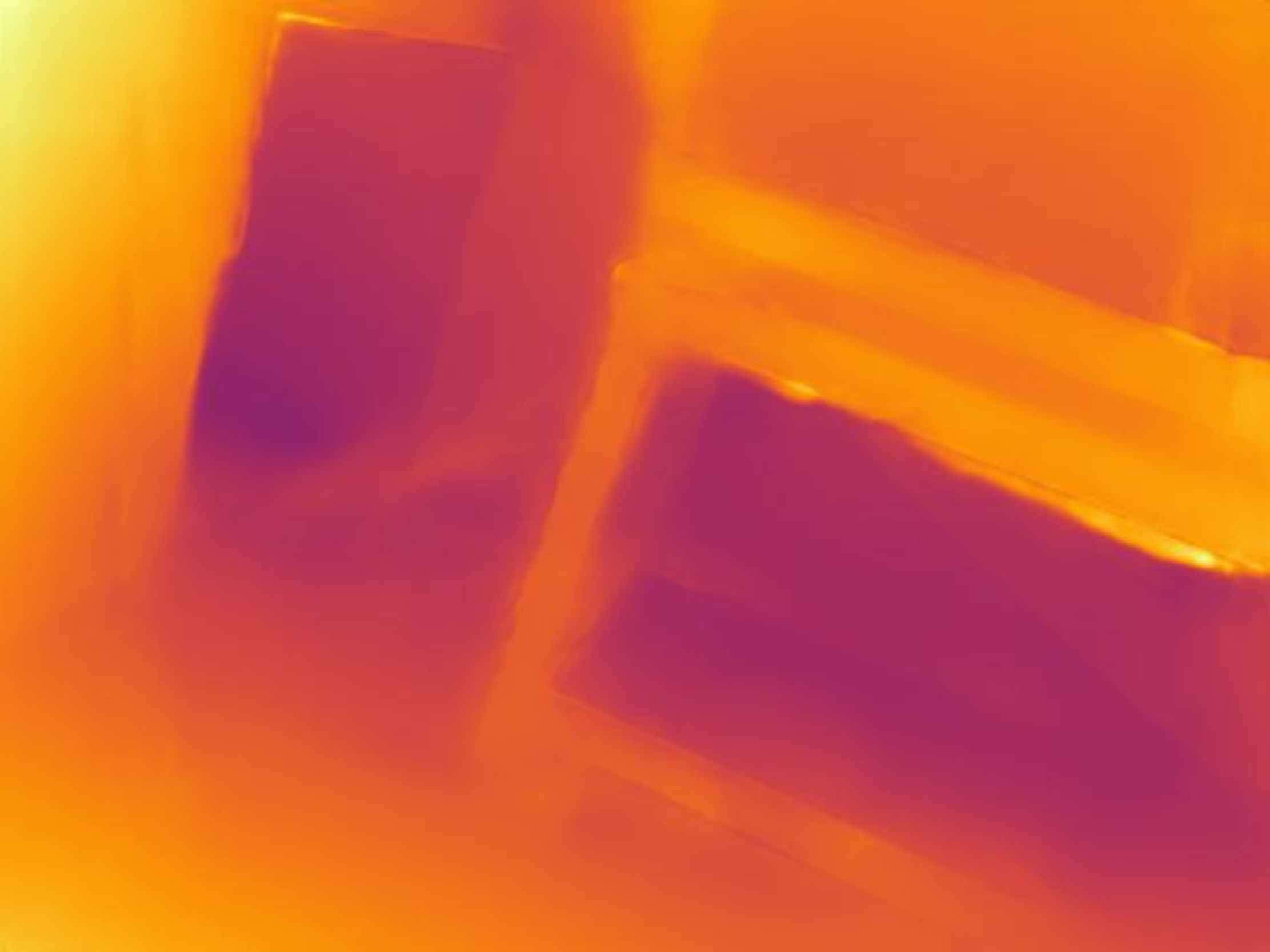}&
    \includegraphics[width=0.104\linewidth]{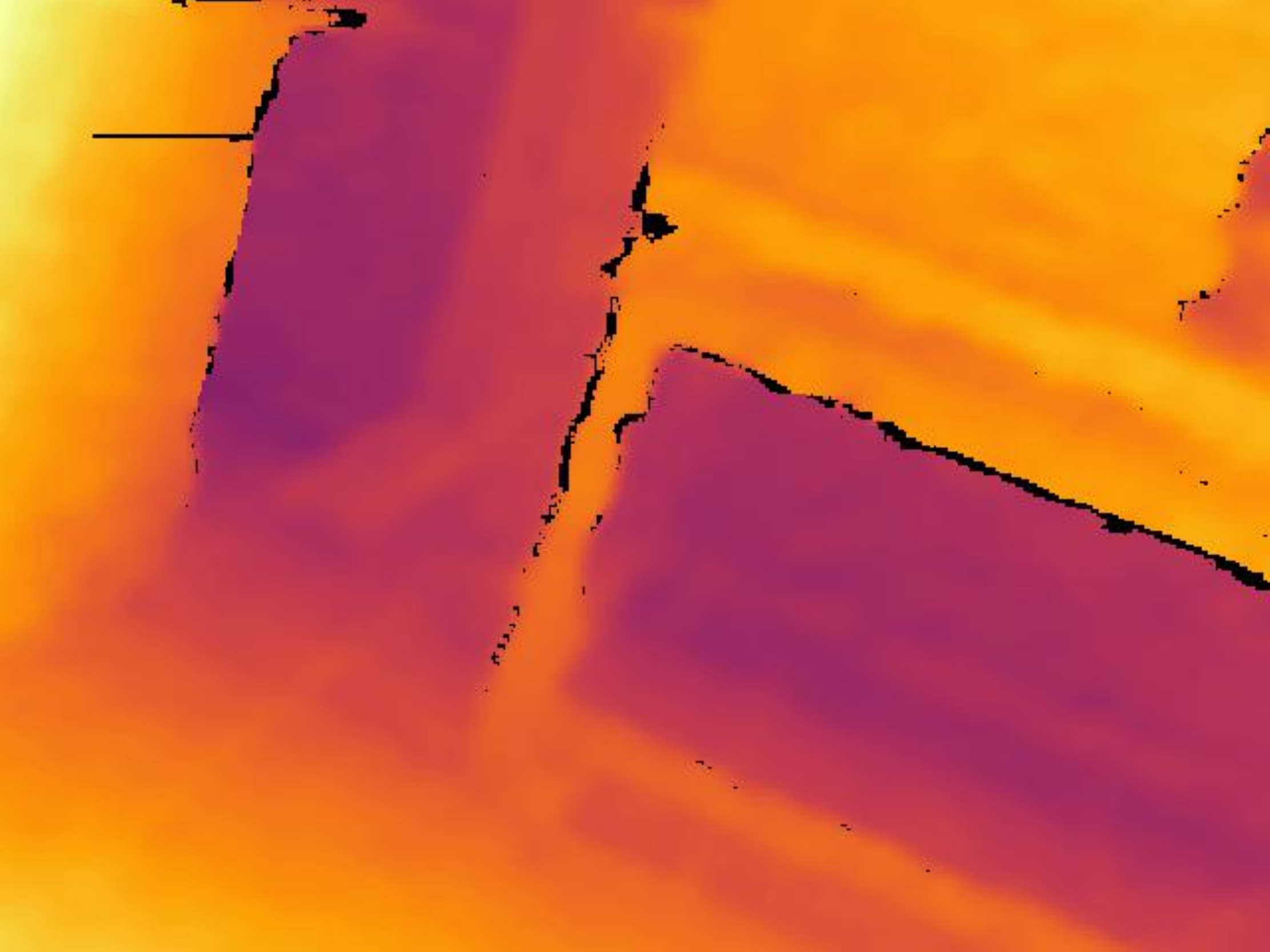}&
    \includegraphics[width=0.104\linewidth]{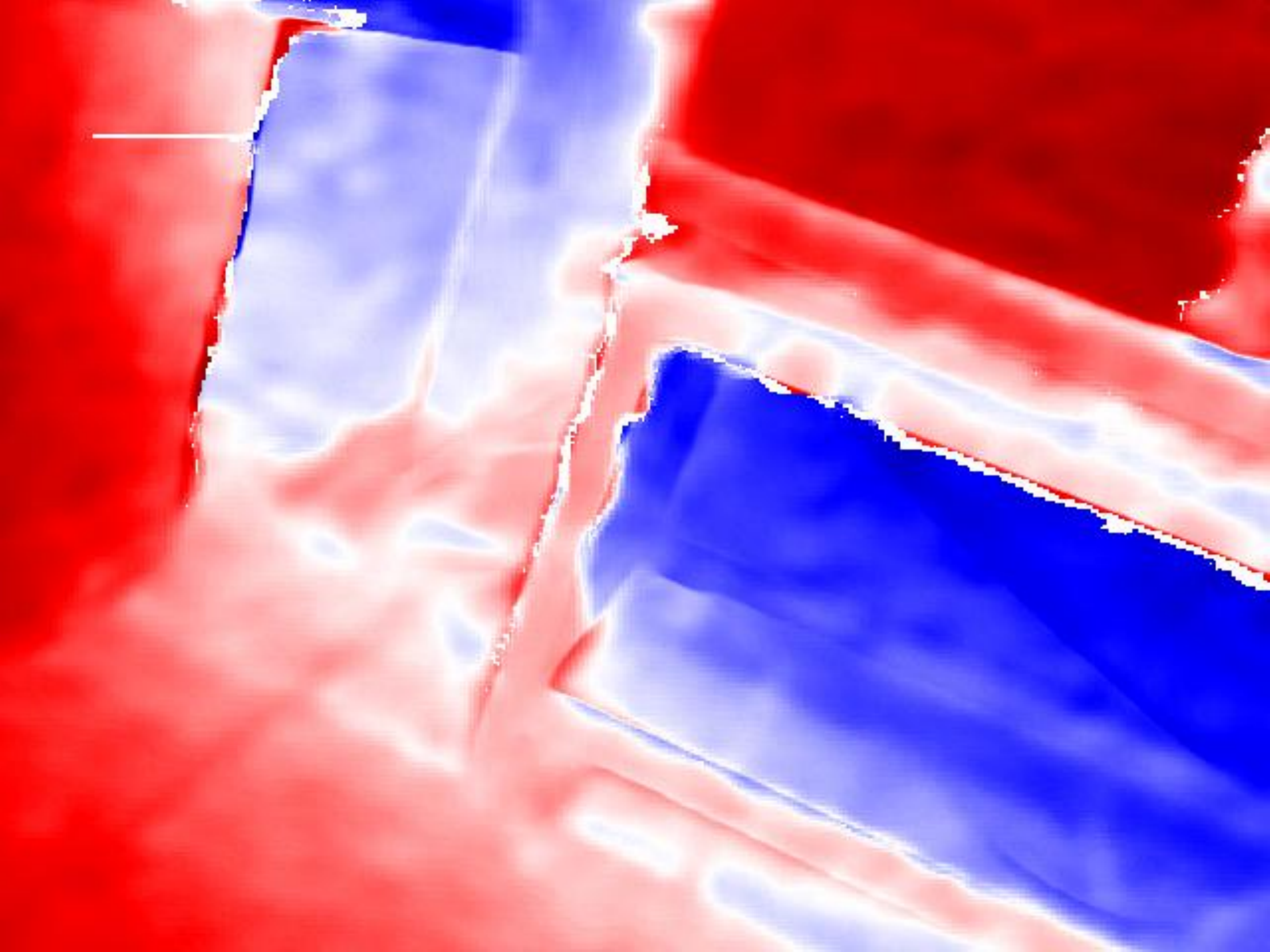}&
    \includegraphics[width=0.104\linewidth]{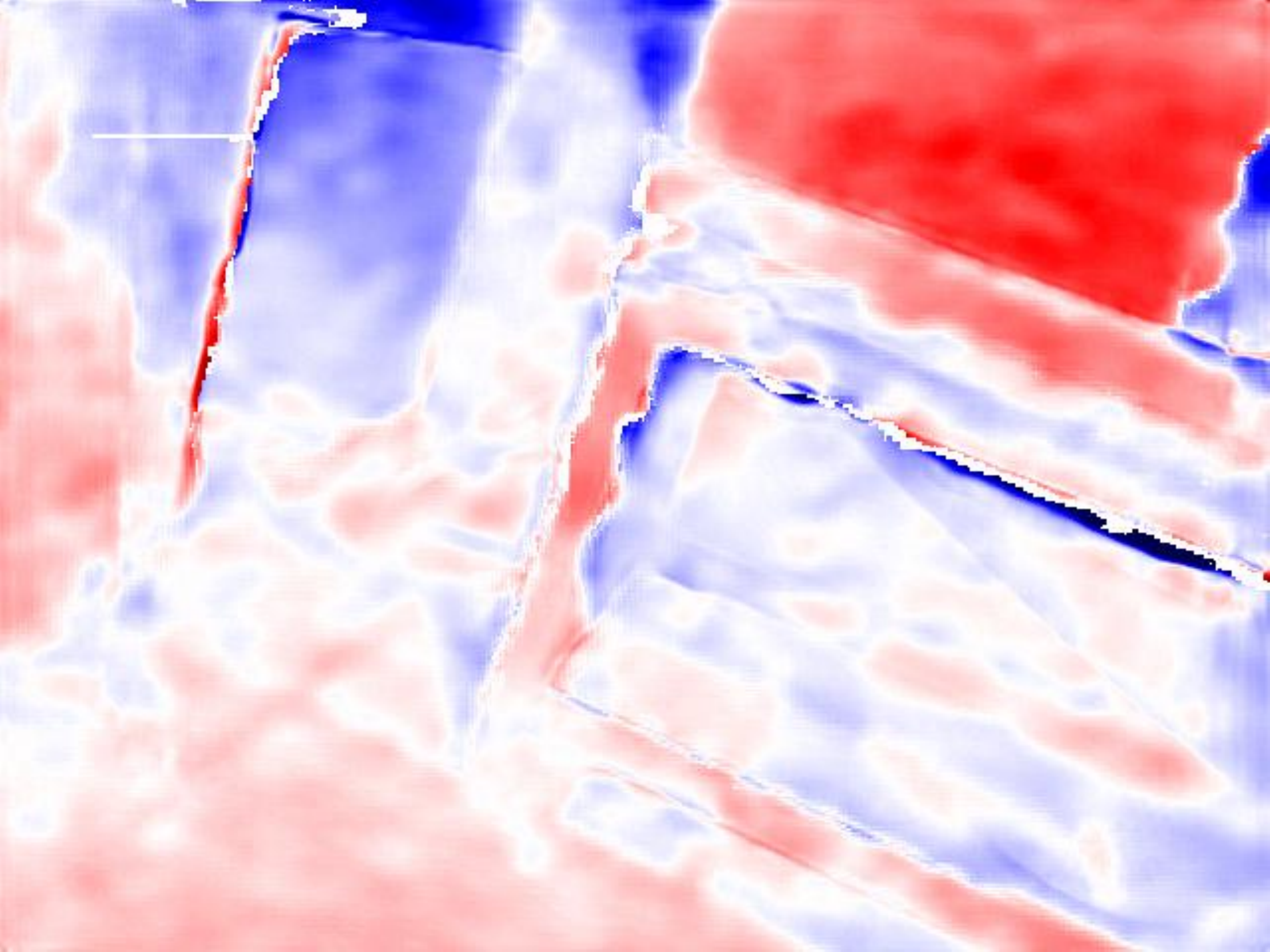}&
    \includegraphics[width=0.024\linewidth]{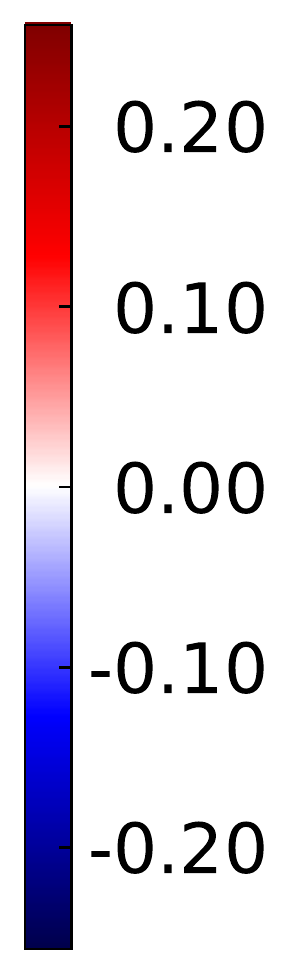}\\
    \vspace{-0.75mm}
    \scriptsize h. &
    \includegraphics[width=0.104\linewidth]{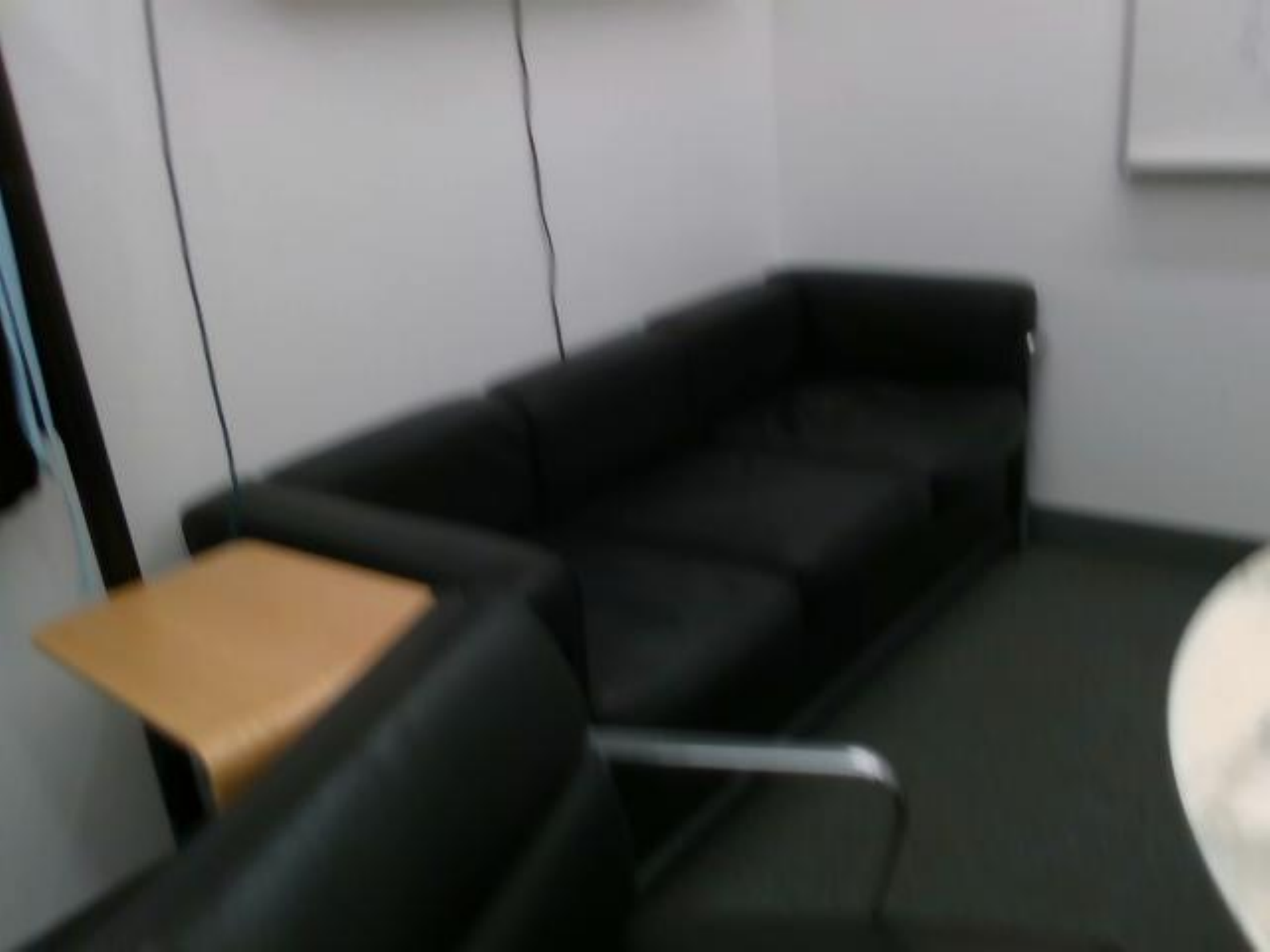}&
    \includegraphics[width=0.104\linewidth]{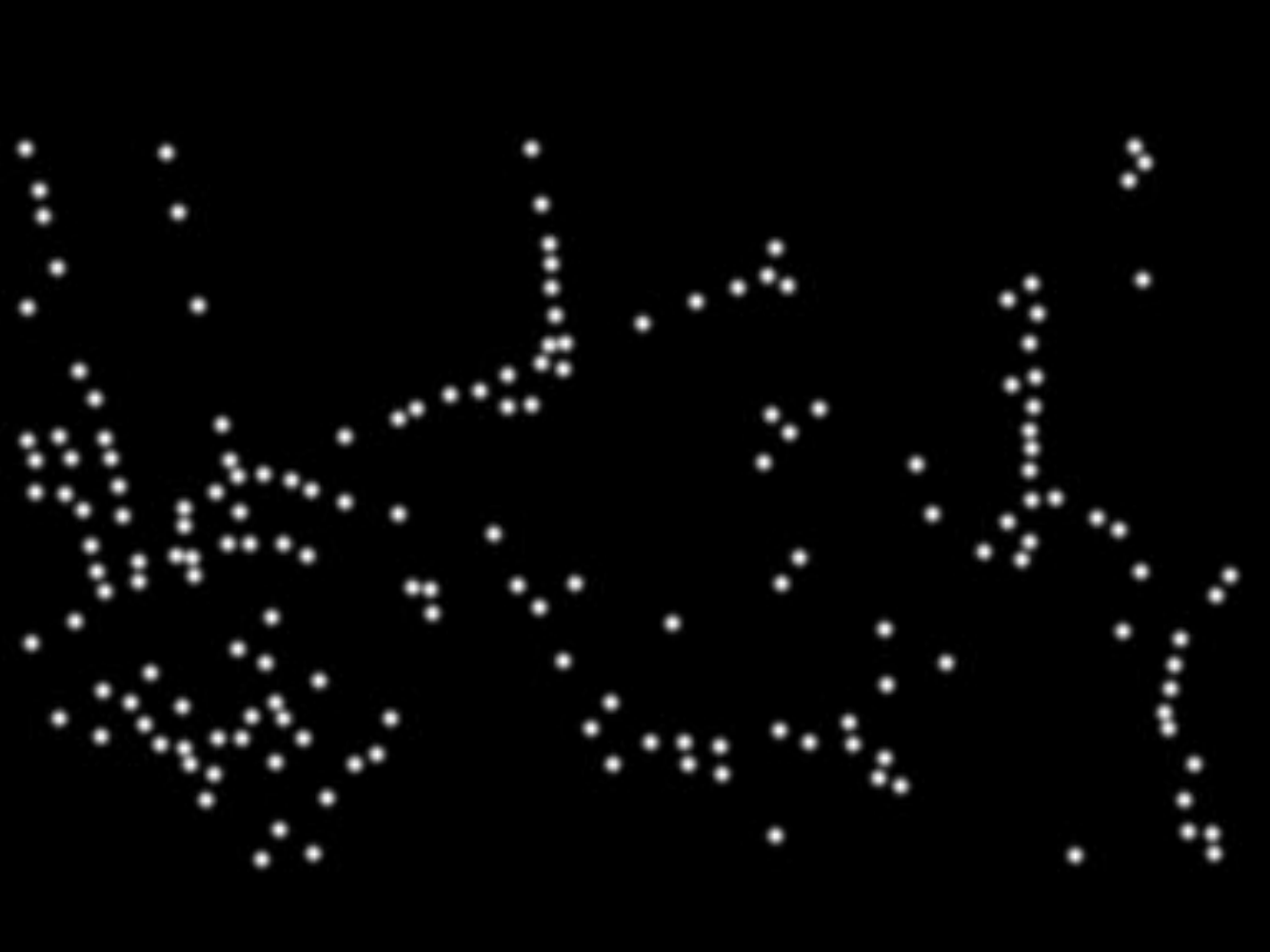}&
    \includegraphics[width=0.104\linewidth]{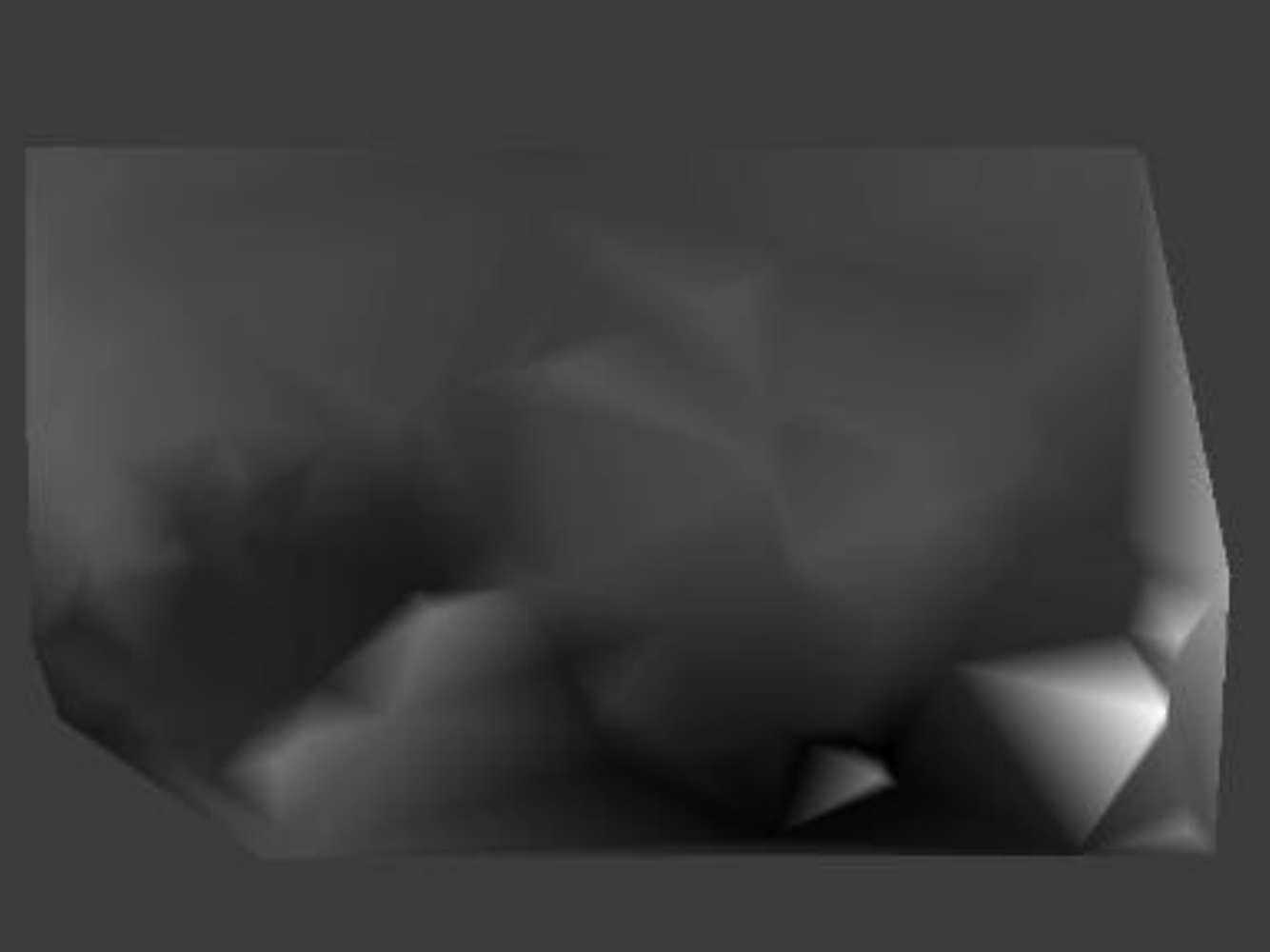}&
    \includegraphics[width=0.104\linewidth]{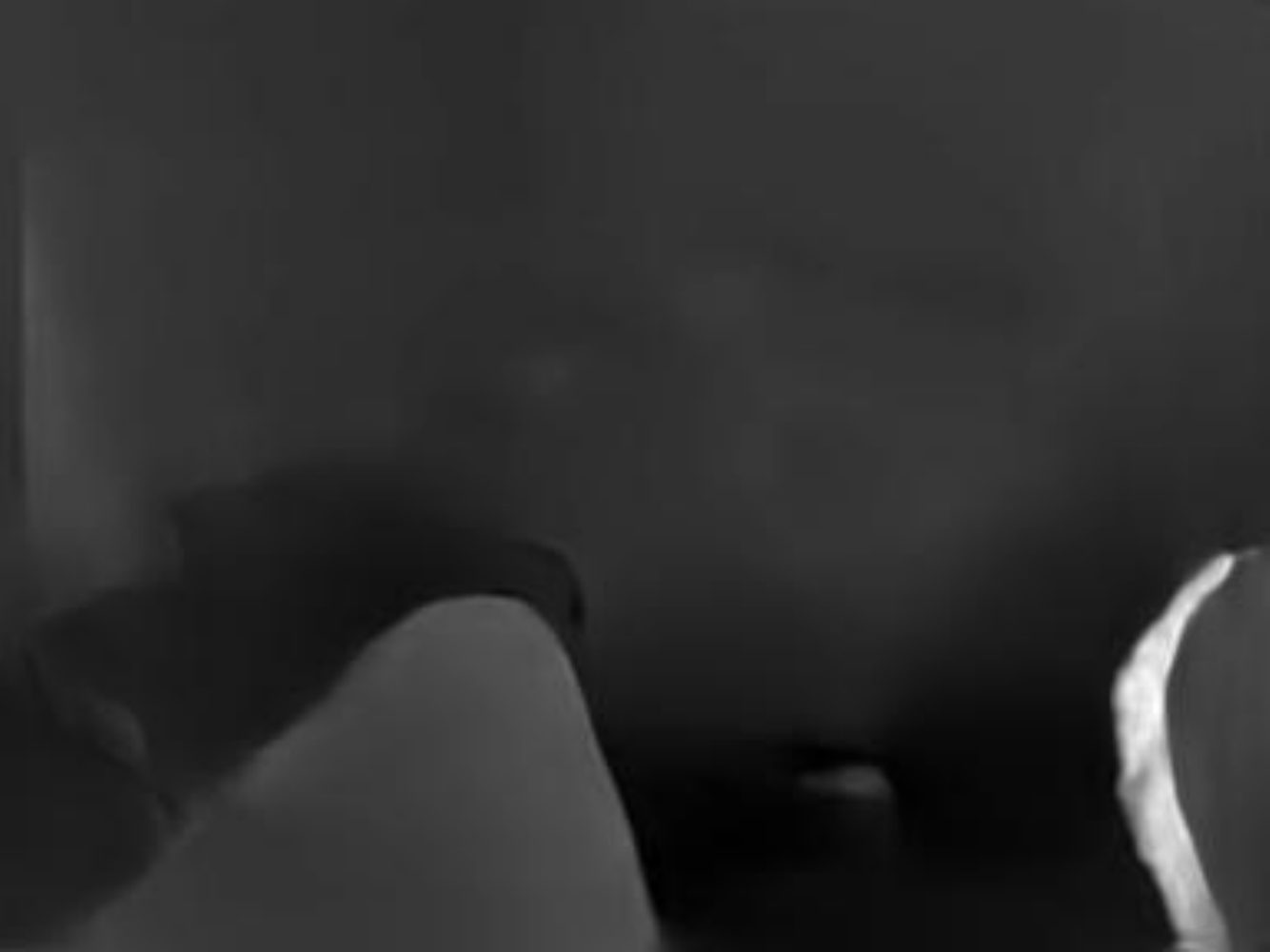}&
    \includegraphics[width=0.104\linewidth]{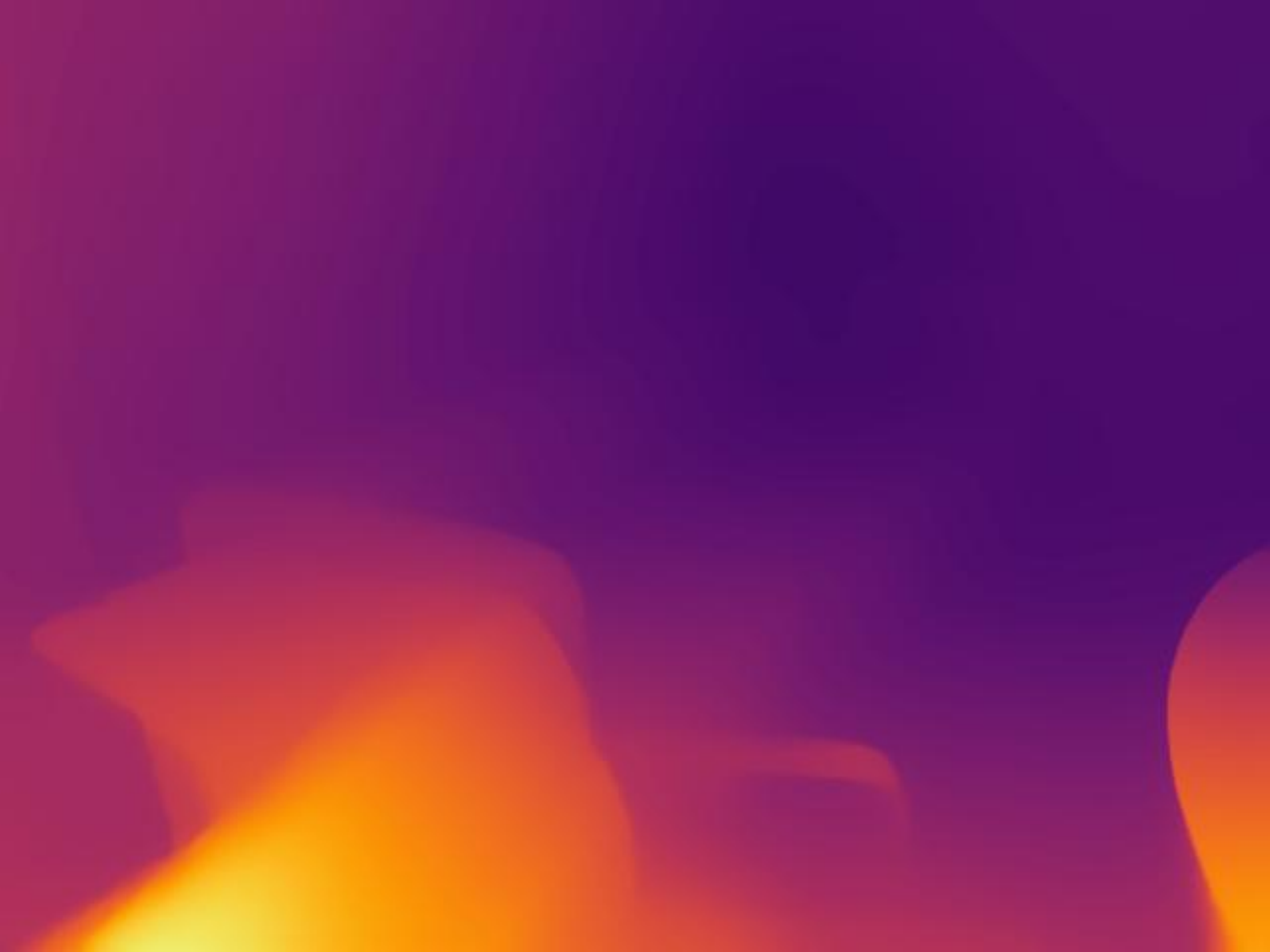}&
    \includegraphics[width=0.104\linewidth]{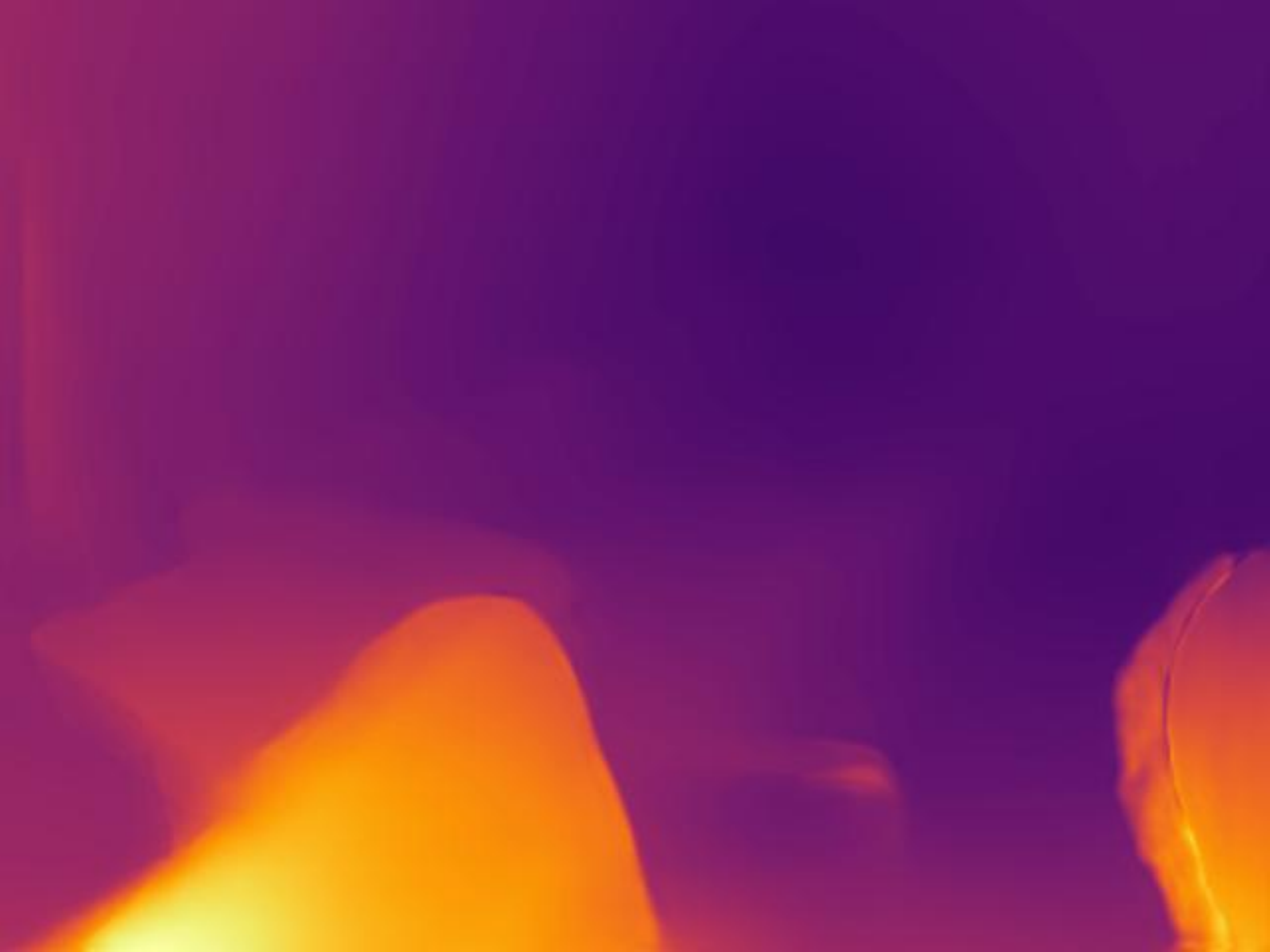}&
    \includegraphics[width=0.104\linewidth]{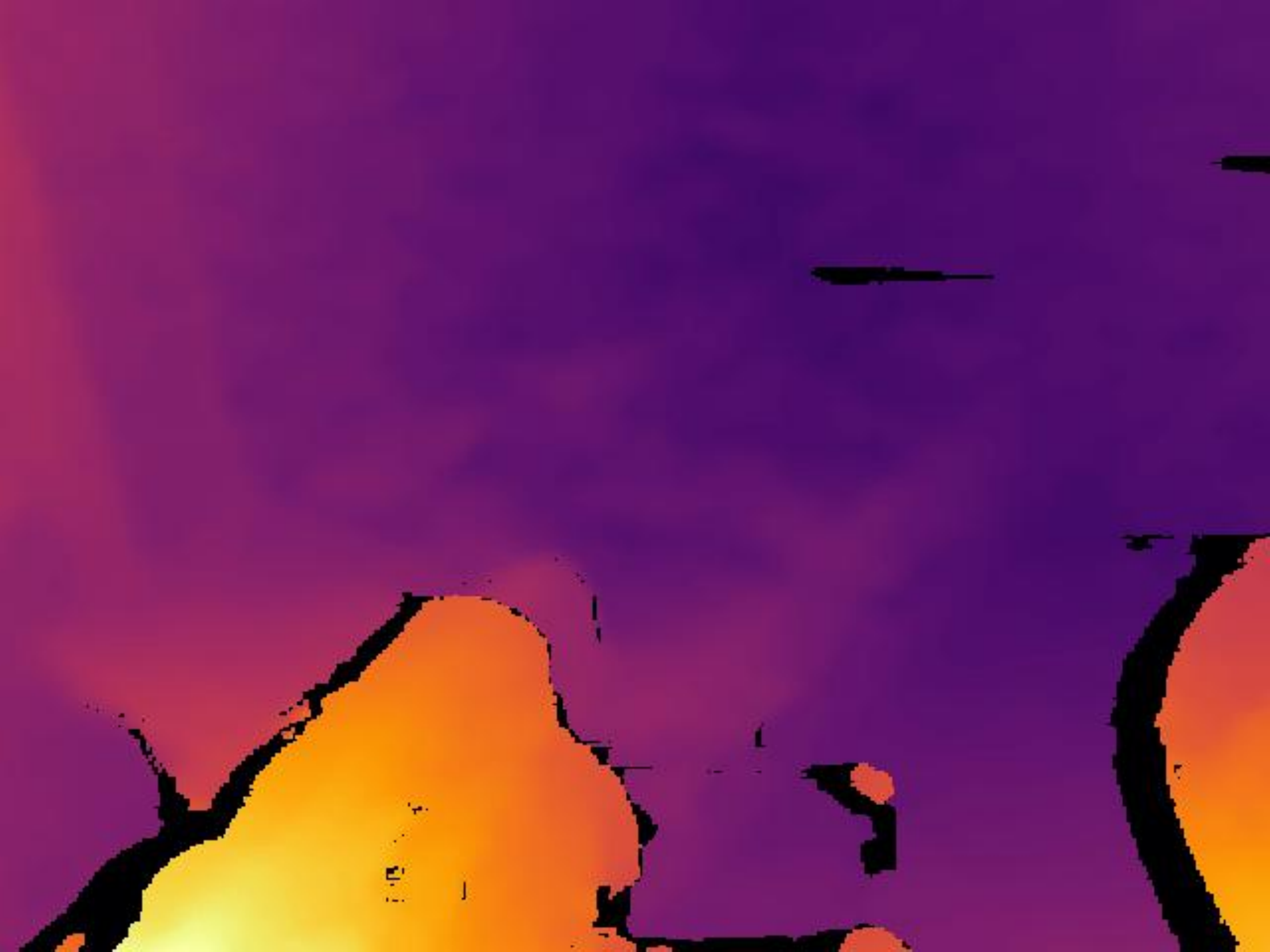}&
    \includegraphics[width=0.104\linewidth]{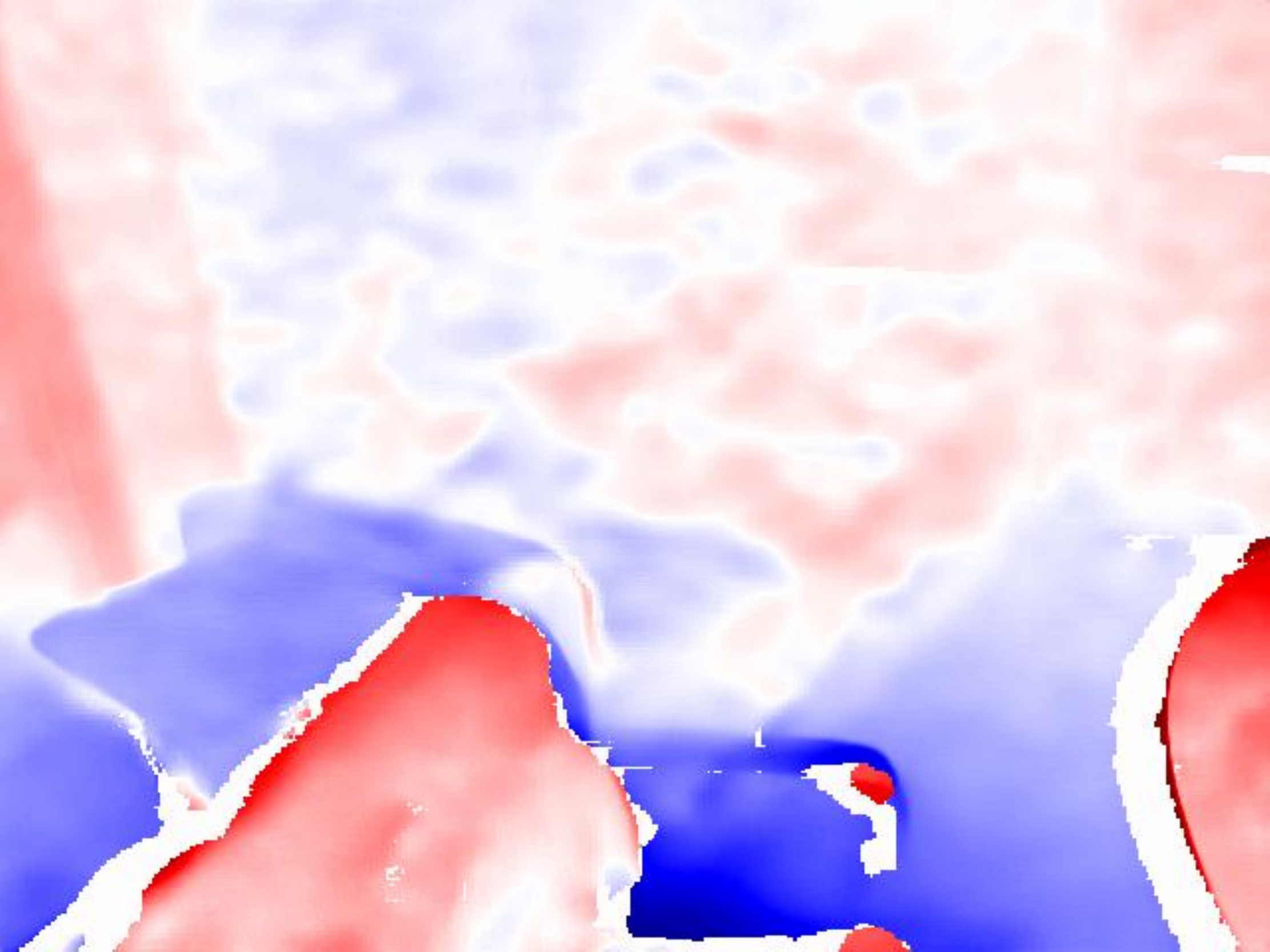}&
    \includegraphics[width=0.104\linewidth]{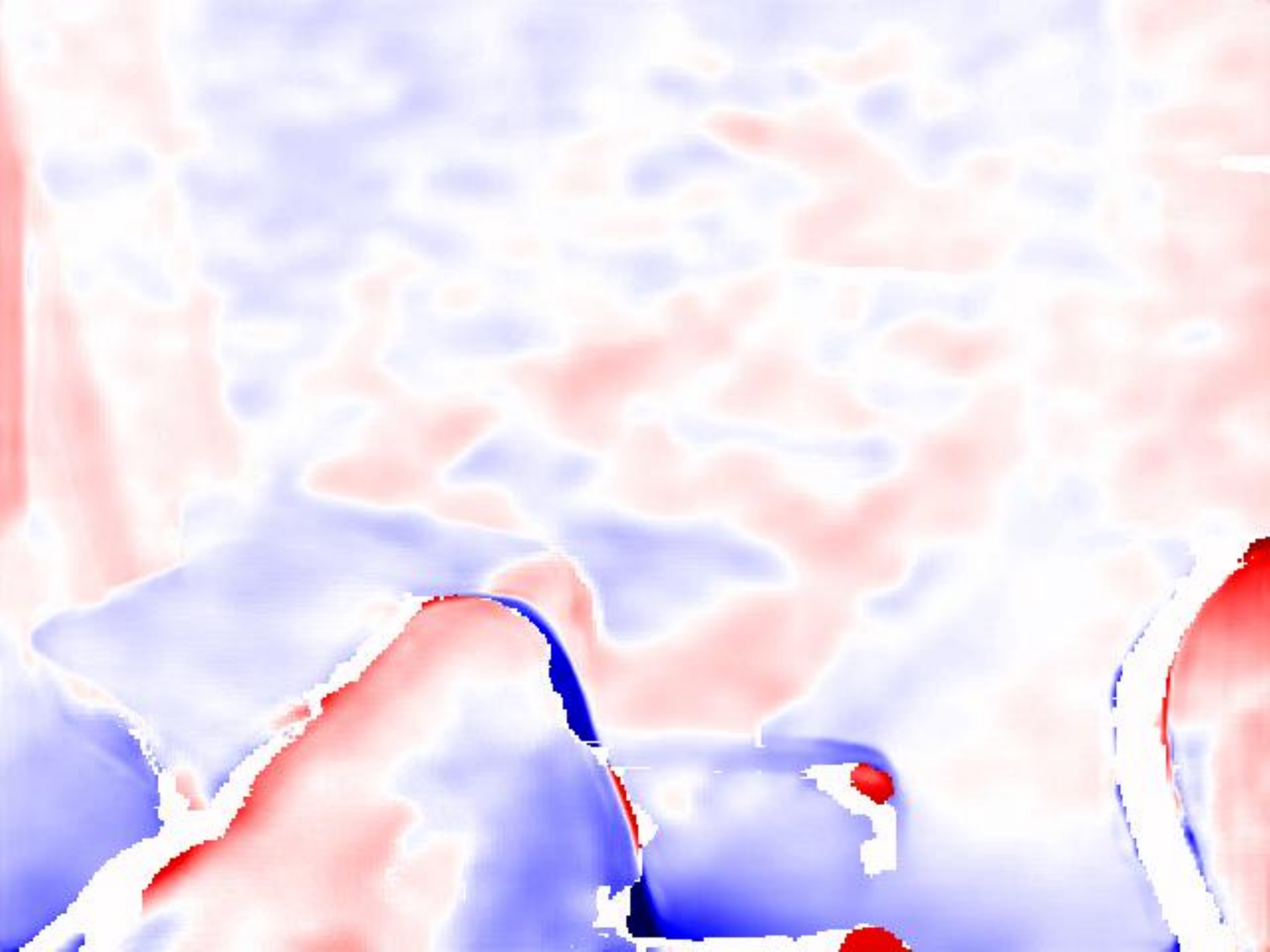}&
    \includegraphics[width=0.024\linewidth]{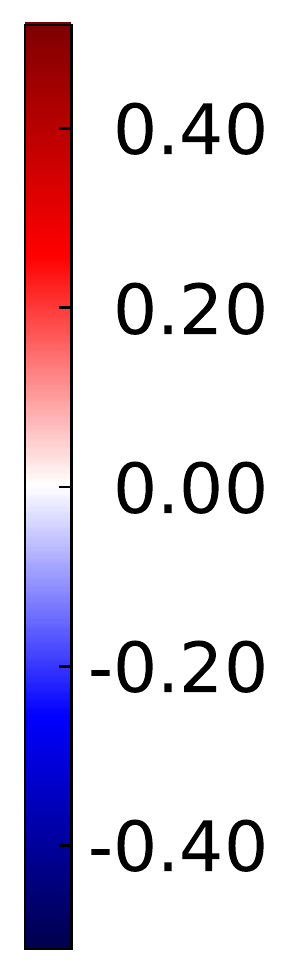}\\
    \vspace{-0.75mm}
    \scriptsize i. &
    \includegraphics[width=0.104\linewidth]{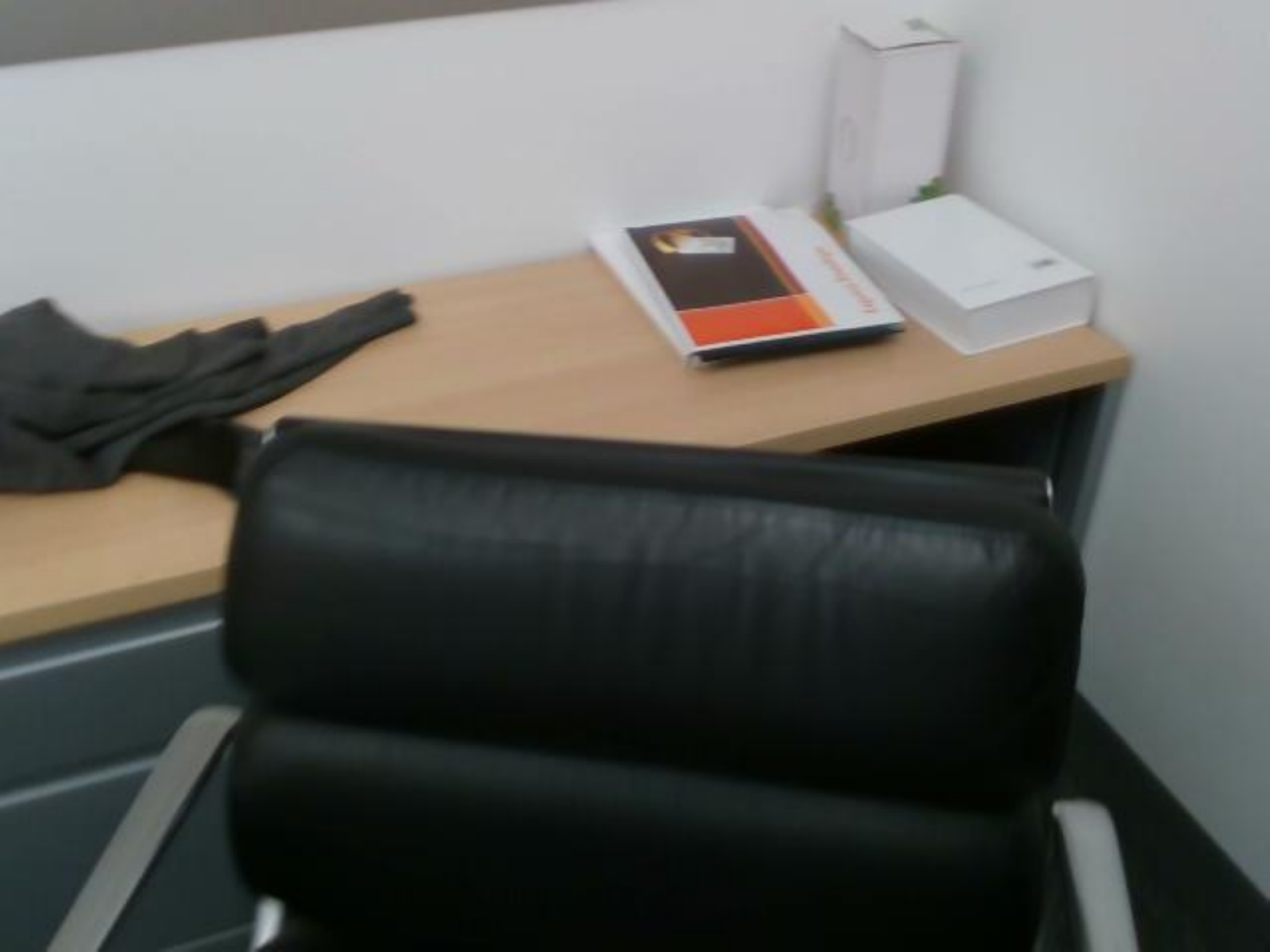}&
    \includegraphics[width=0.104\linewidth]{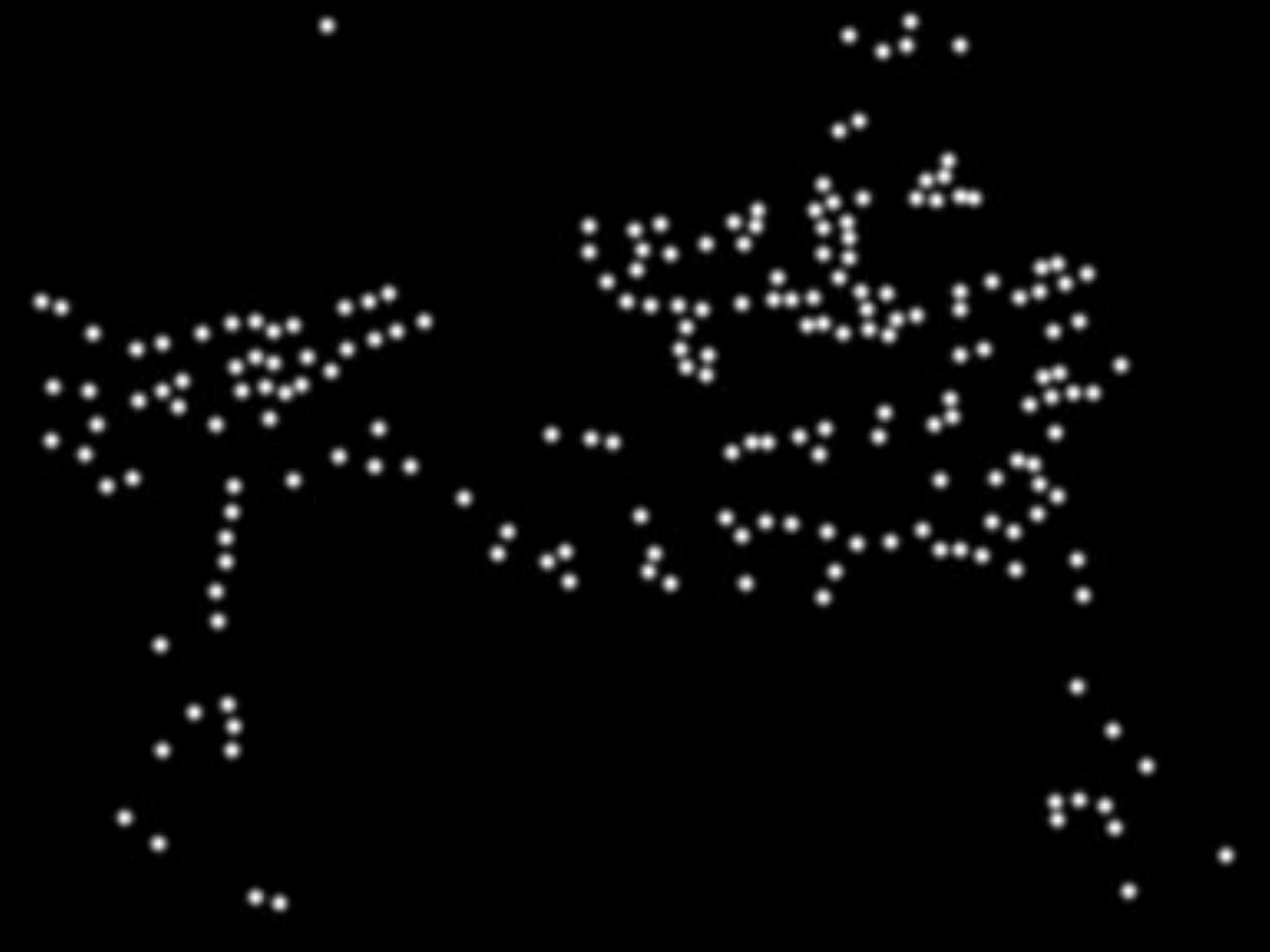}&
    \includegraphics[width=0.104\linewidth]{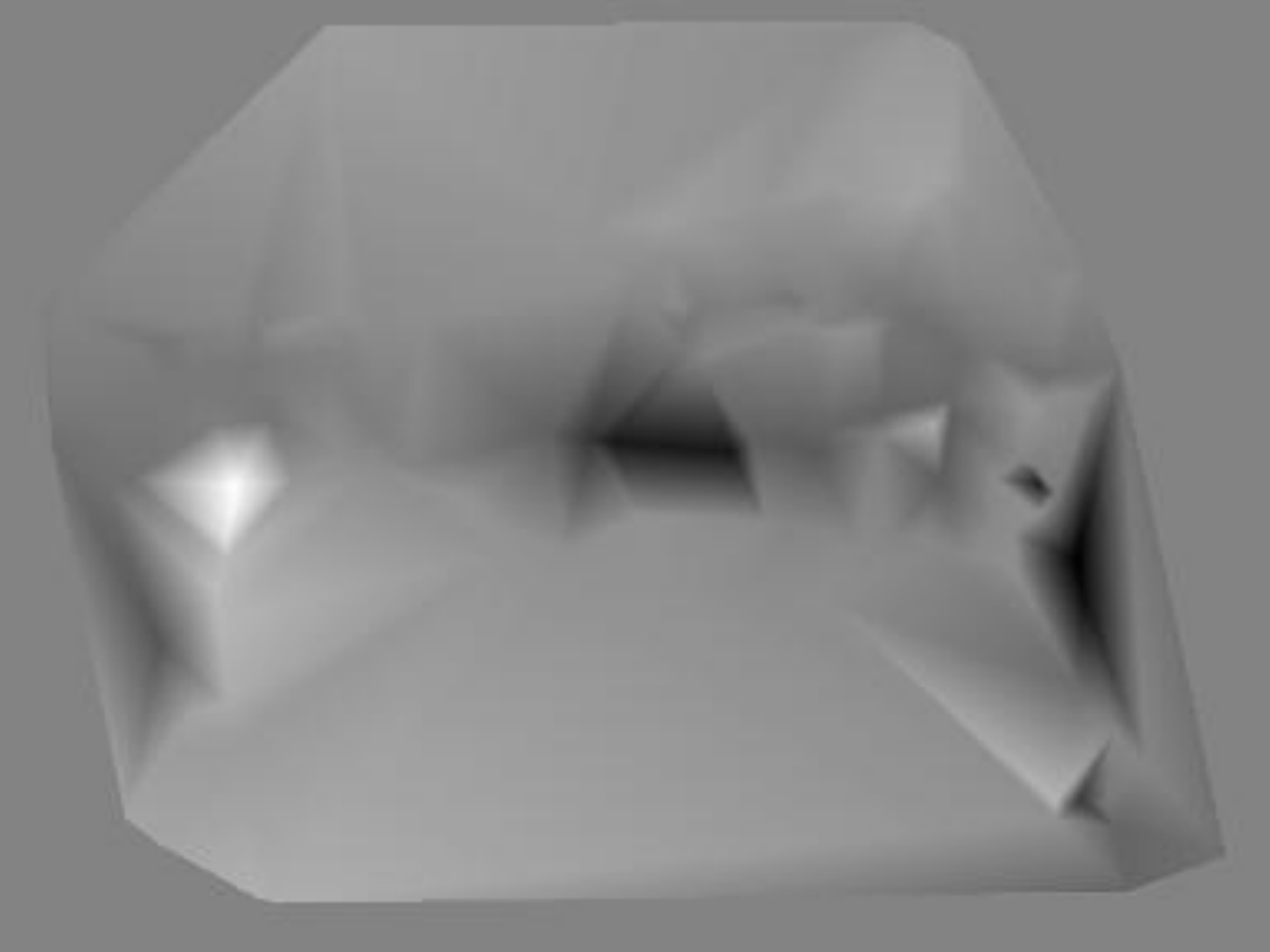}&
    \includegraphics[width=0.104\linewidth]{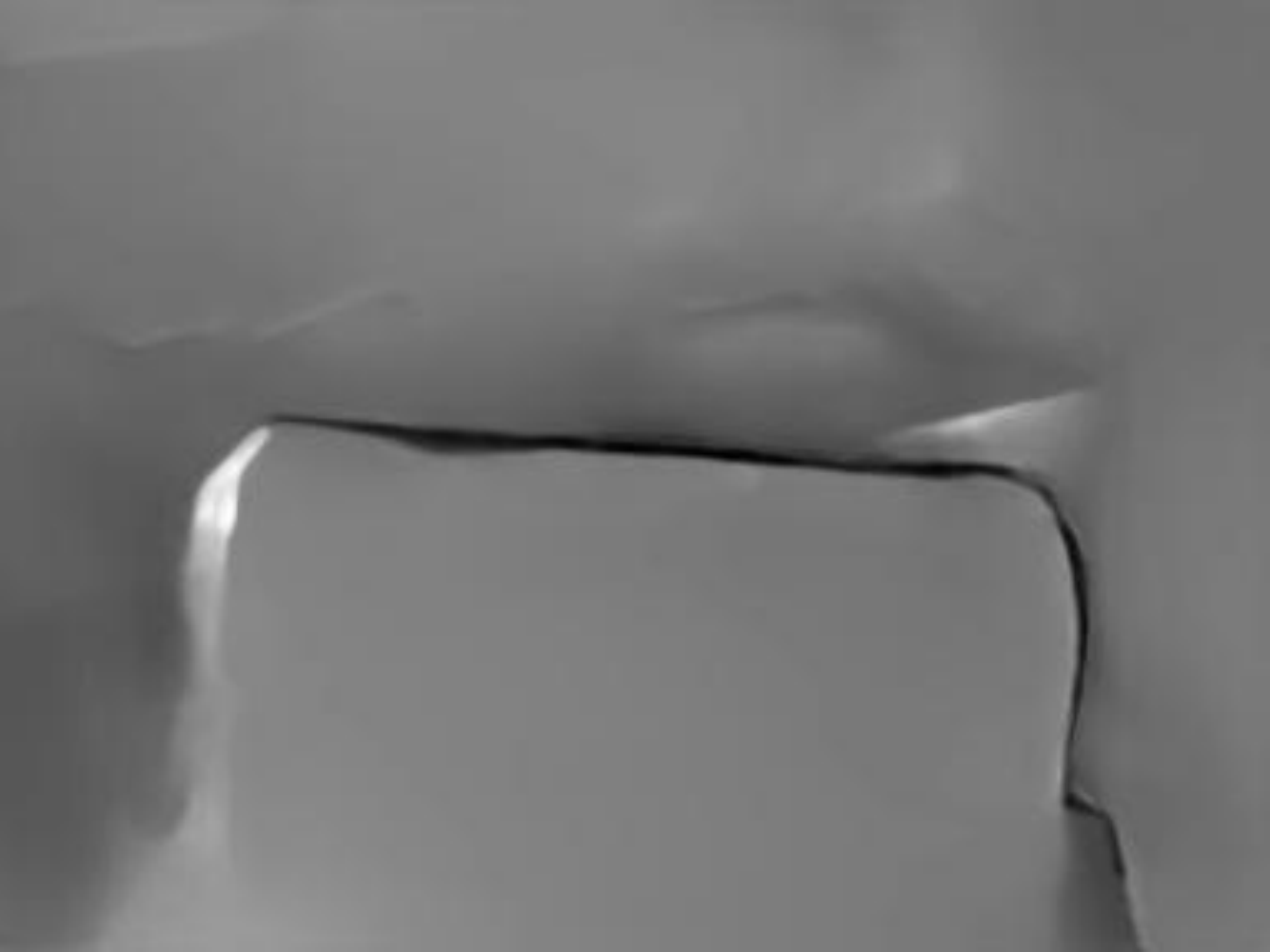}&
    \includegraphics[width=0.104\linewidth]{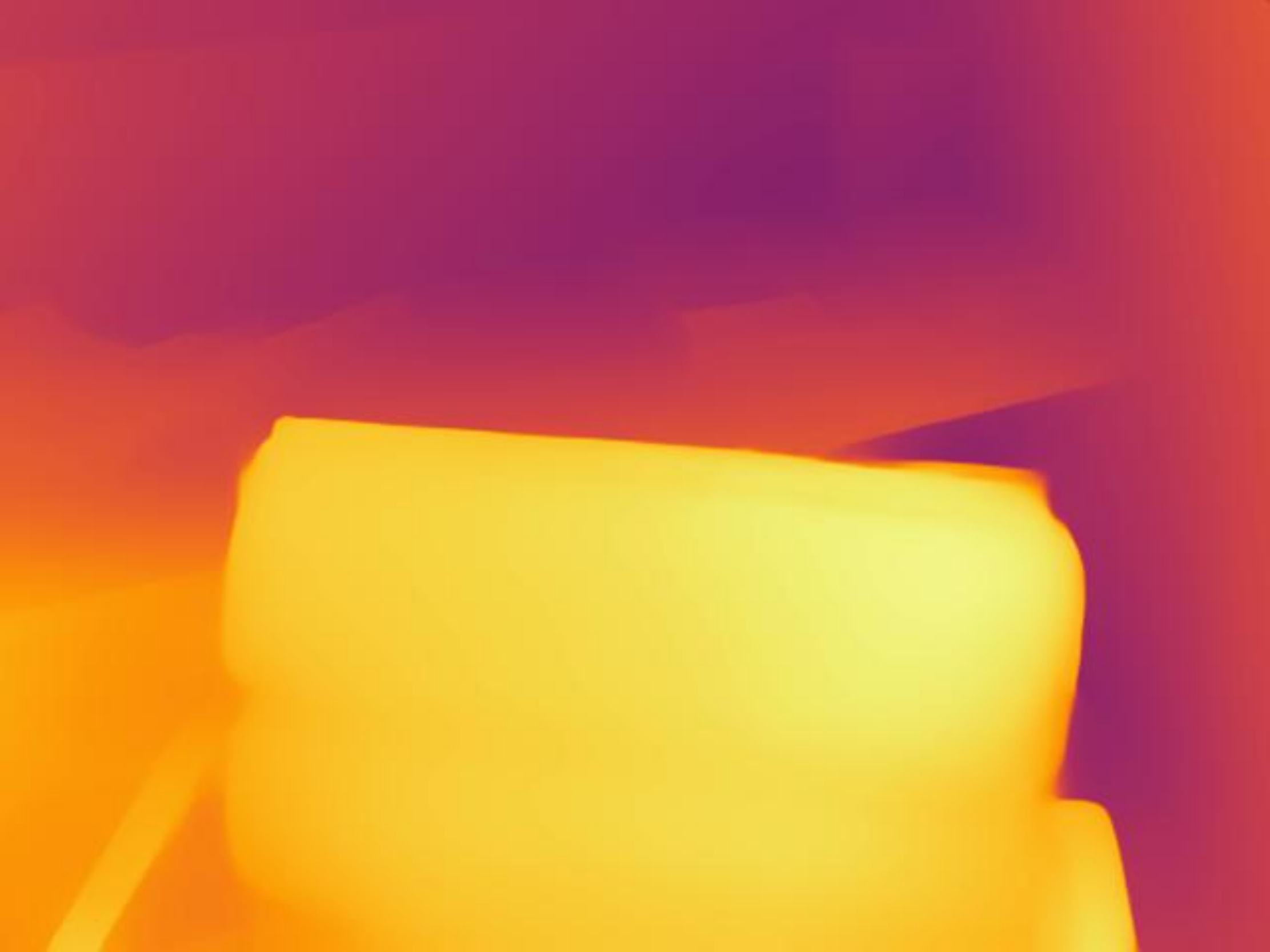}&
    \includegraphics[width=0.104\linewidth]{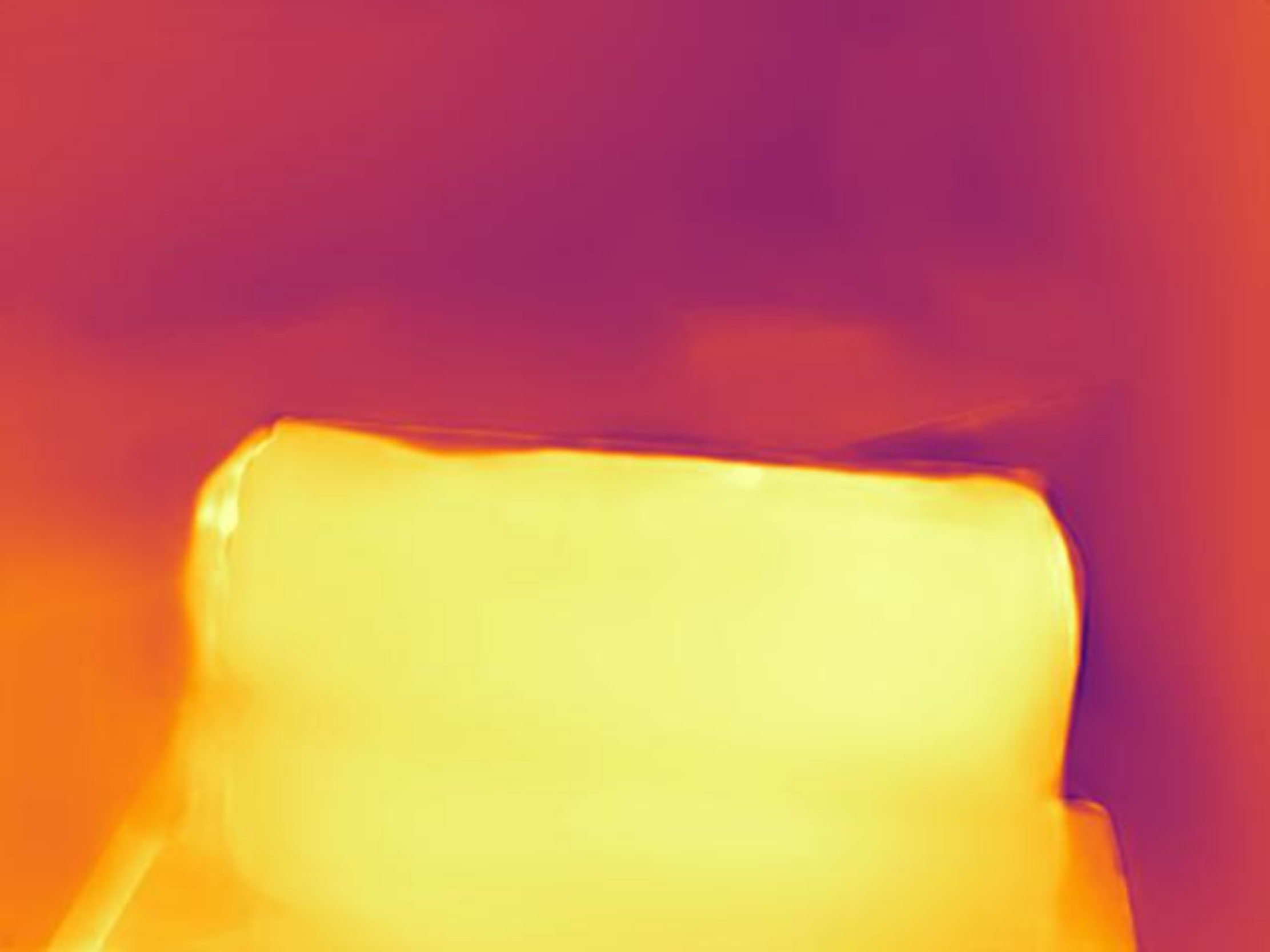}&
    \includegraphics[width=0.104\linewidth]{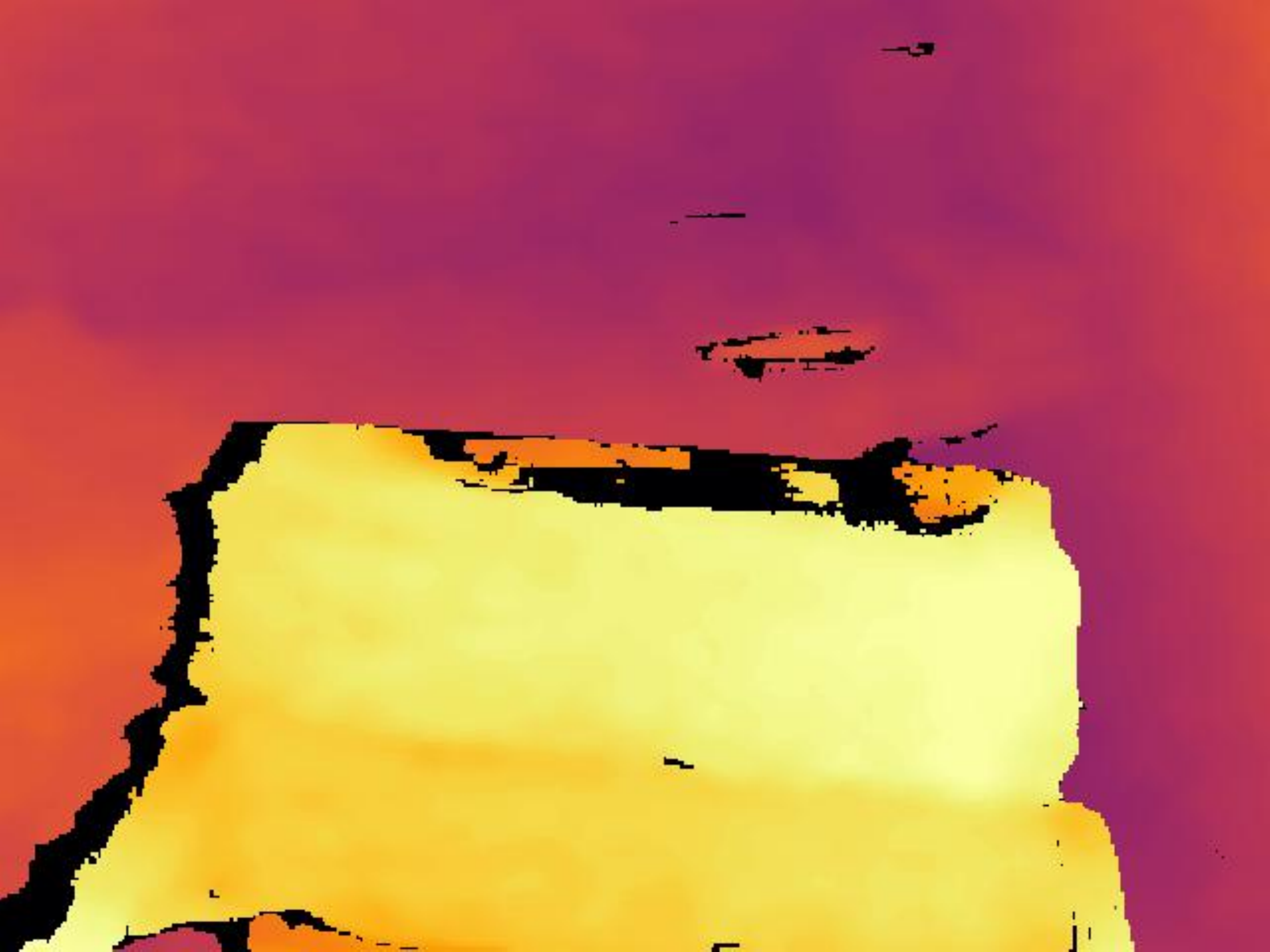}&
    \includegraphics[width=0.104\linewidth]{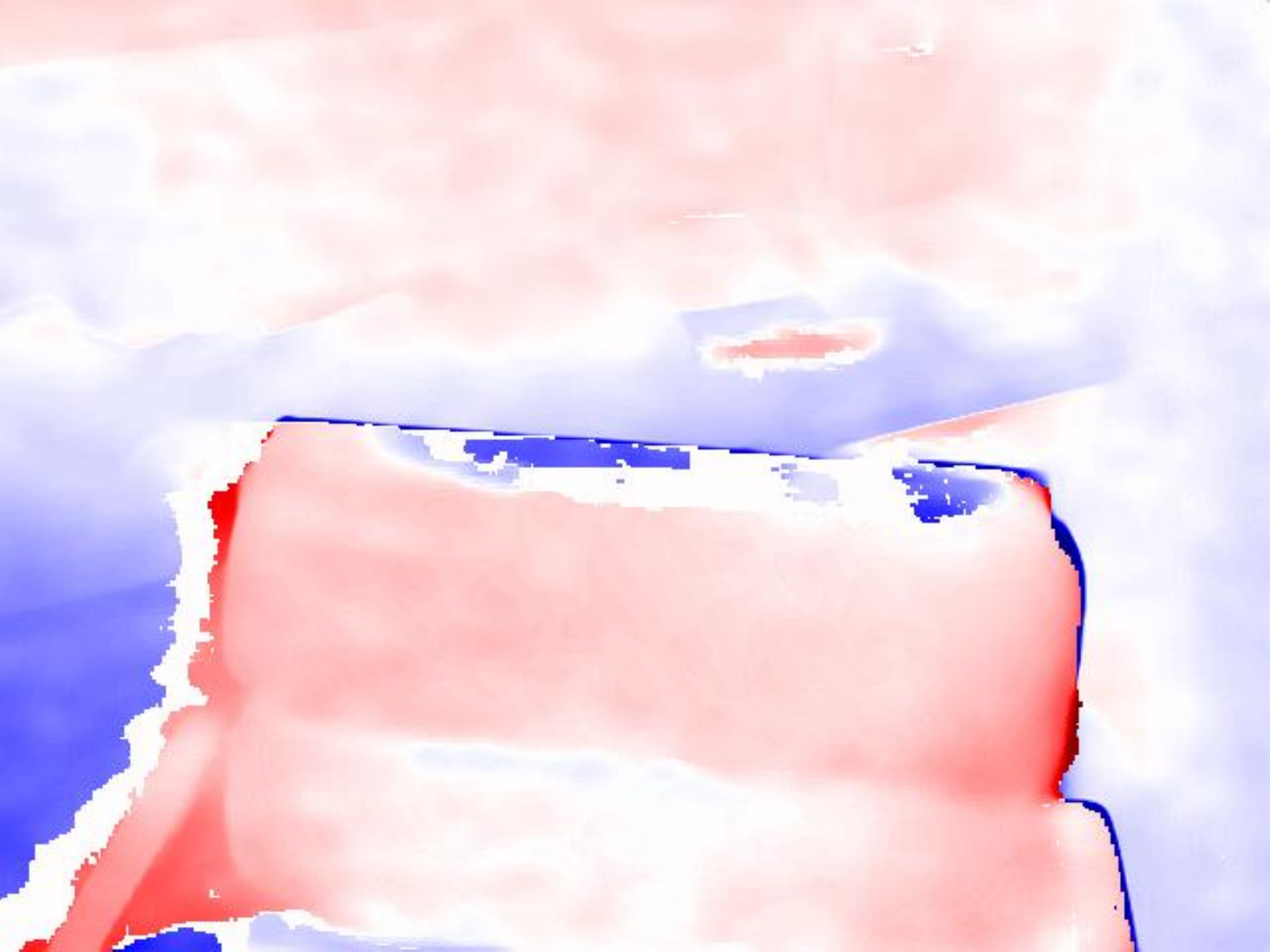}&
    \includegraphics[width=0.104\linewidth]{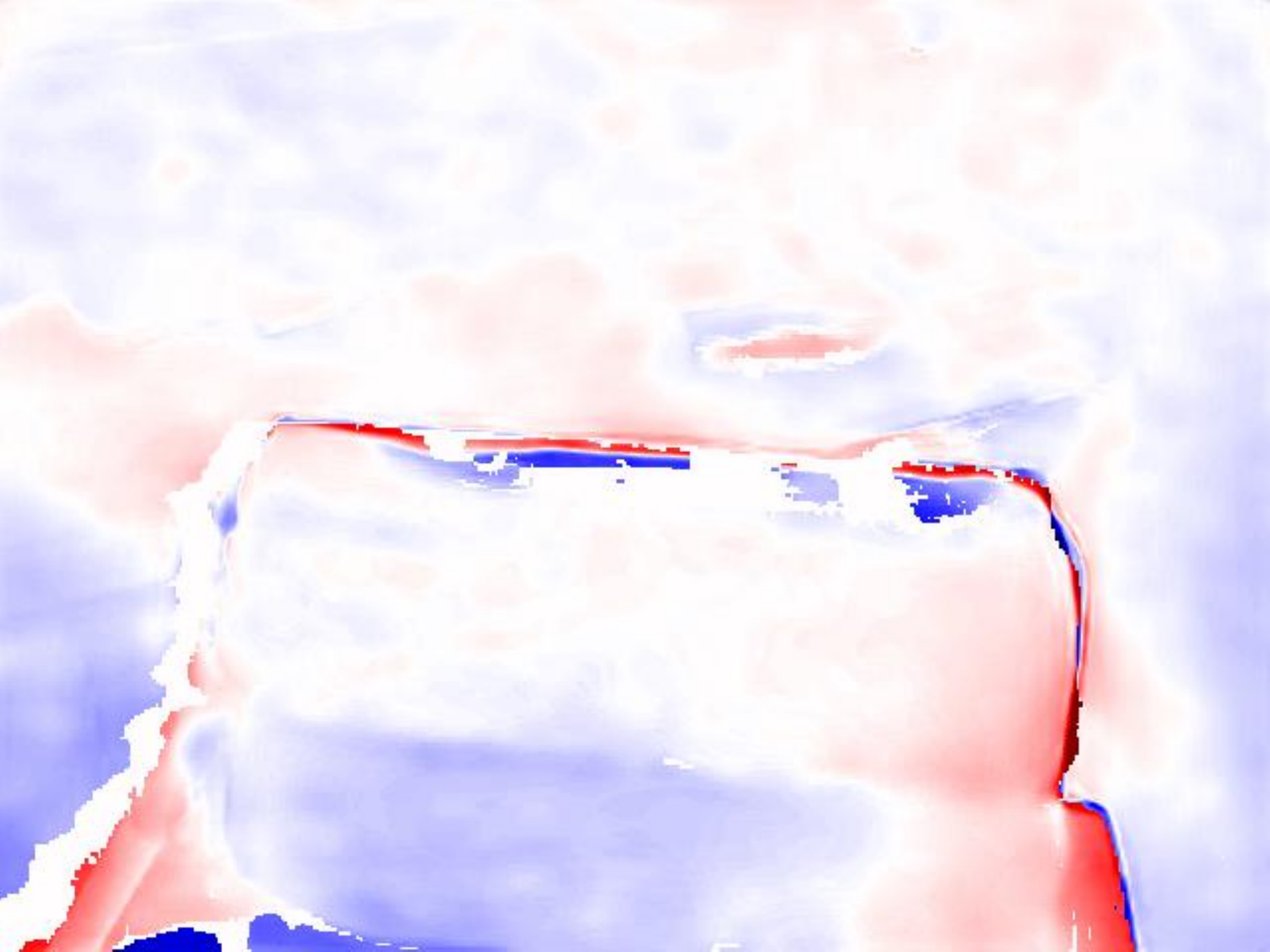}&
    \includegraphics[width=0.024\linewidth]{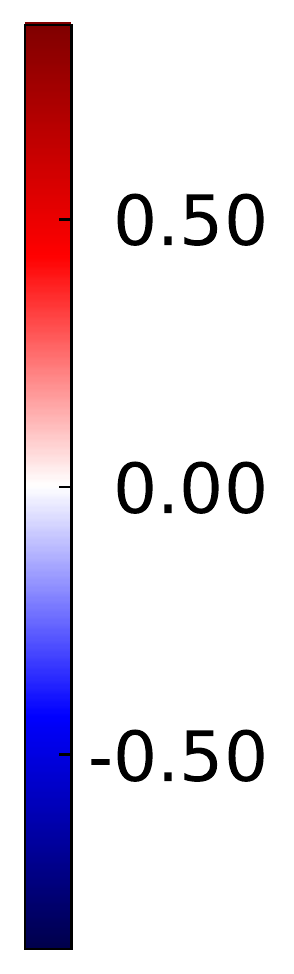}\\
    \vspace{-0.75mm}
    \scriptsize j. &
    \includegraphics[width=0.104\linewidth]{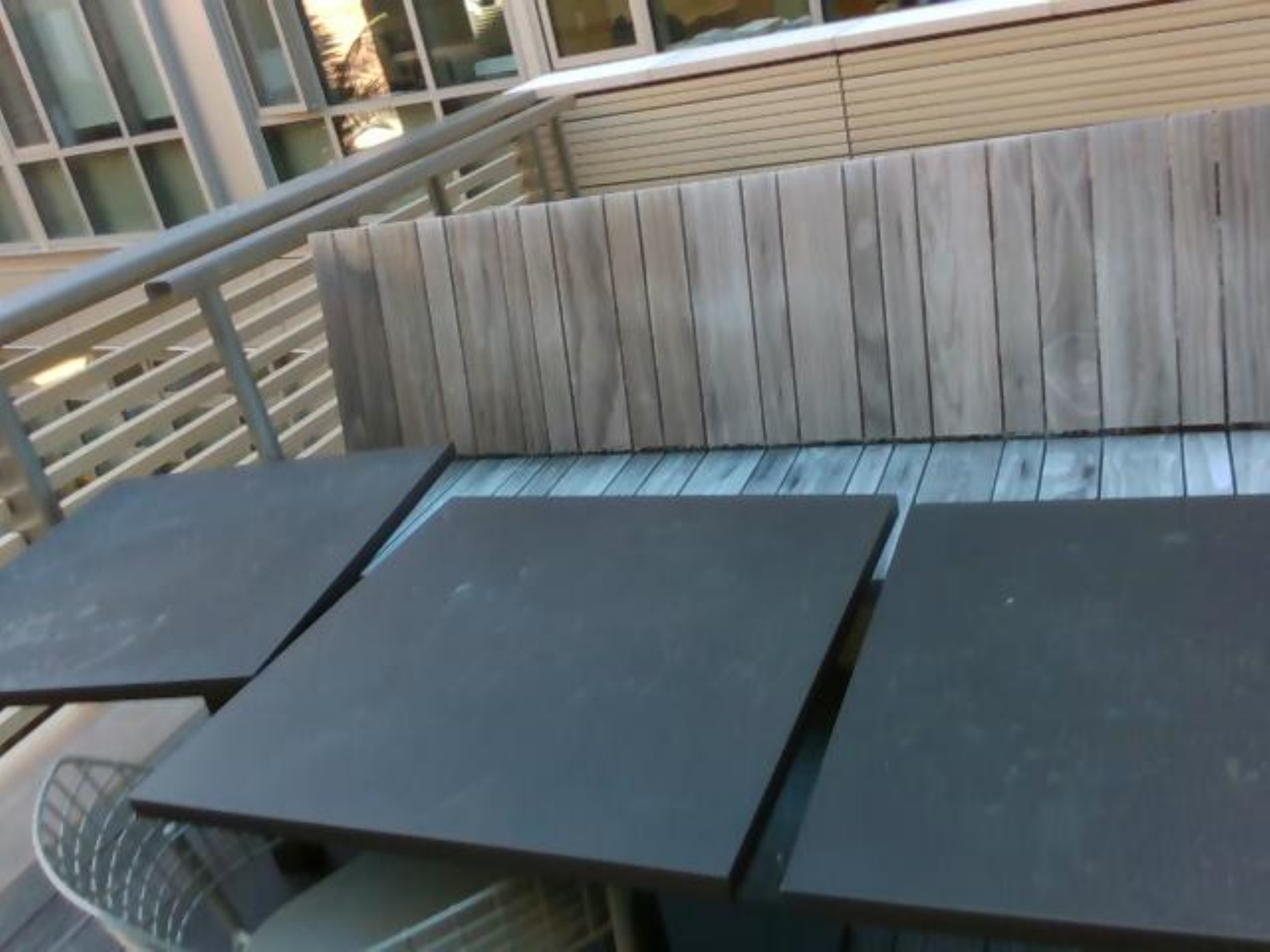}&
    \includegraphics[width=0.104\linewidth]{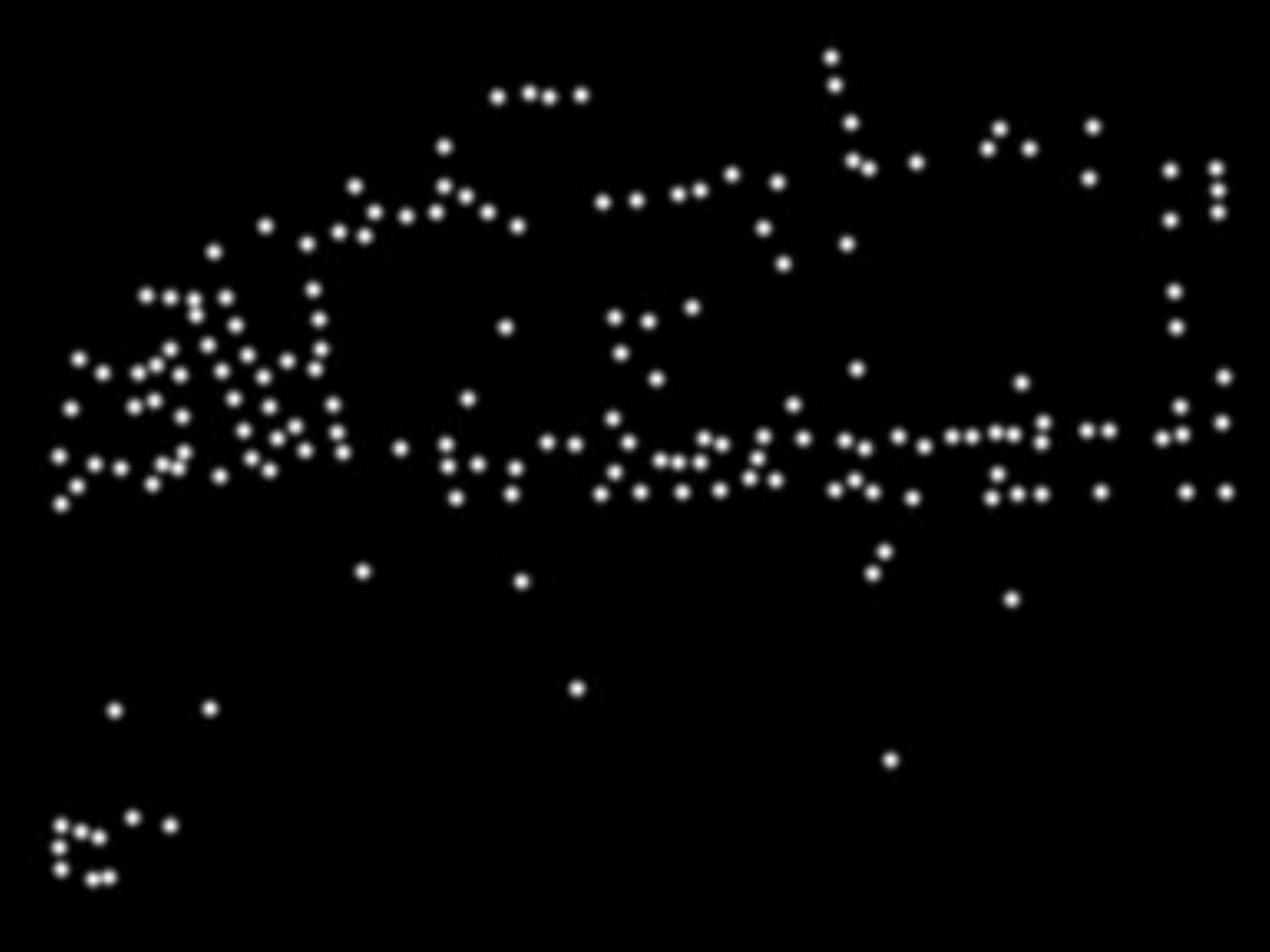}&
    \includegraphics[width=0.104\linewidth]{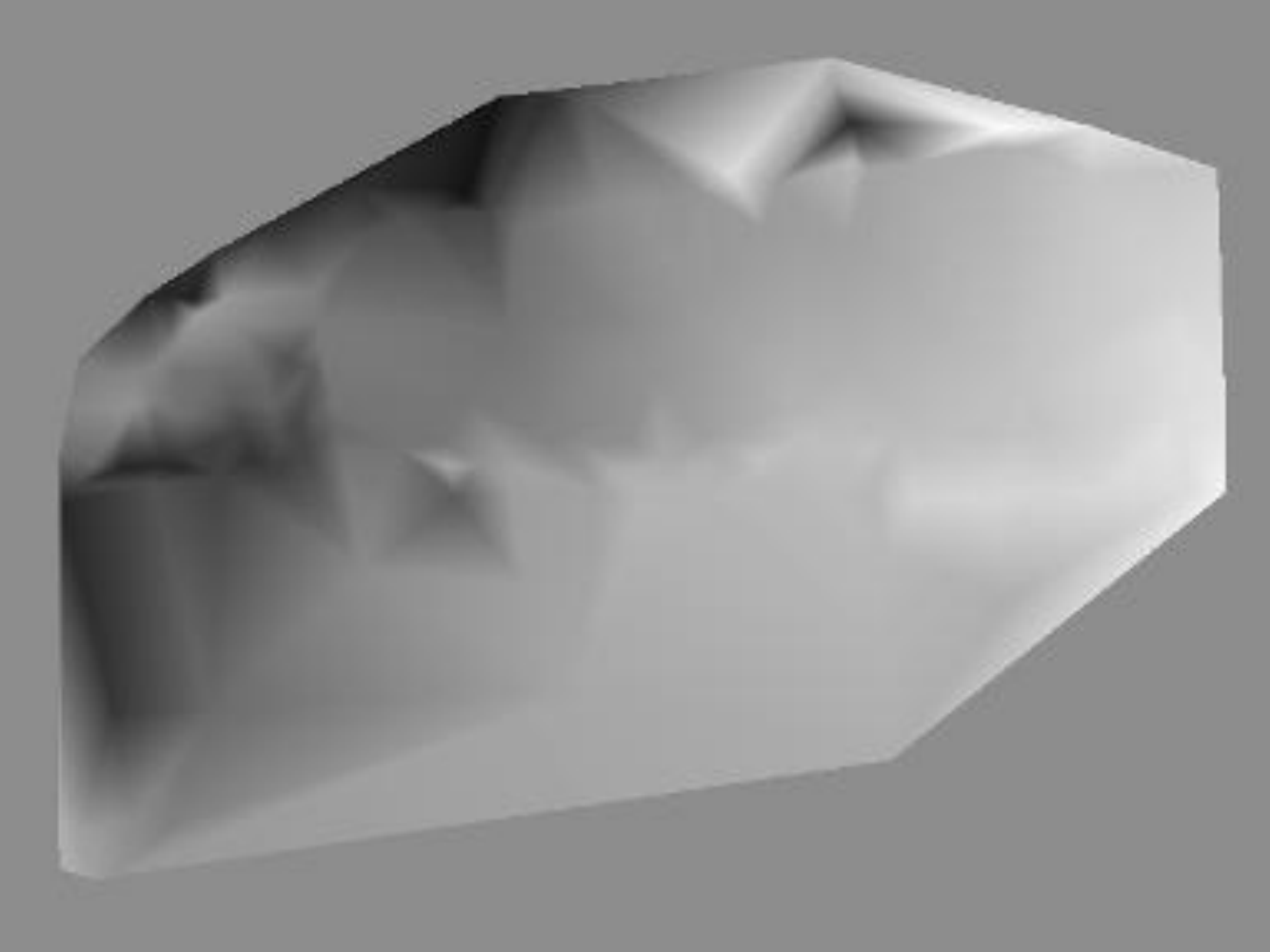}&
    \includegraphics[width=0.104\linewidth]{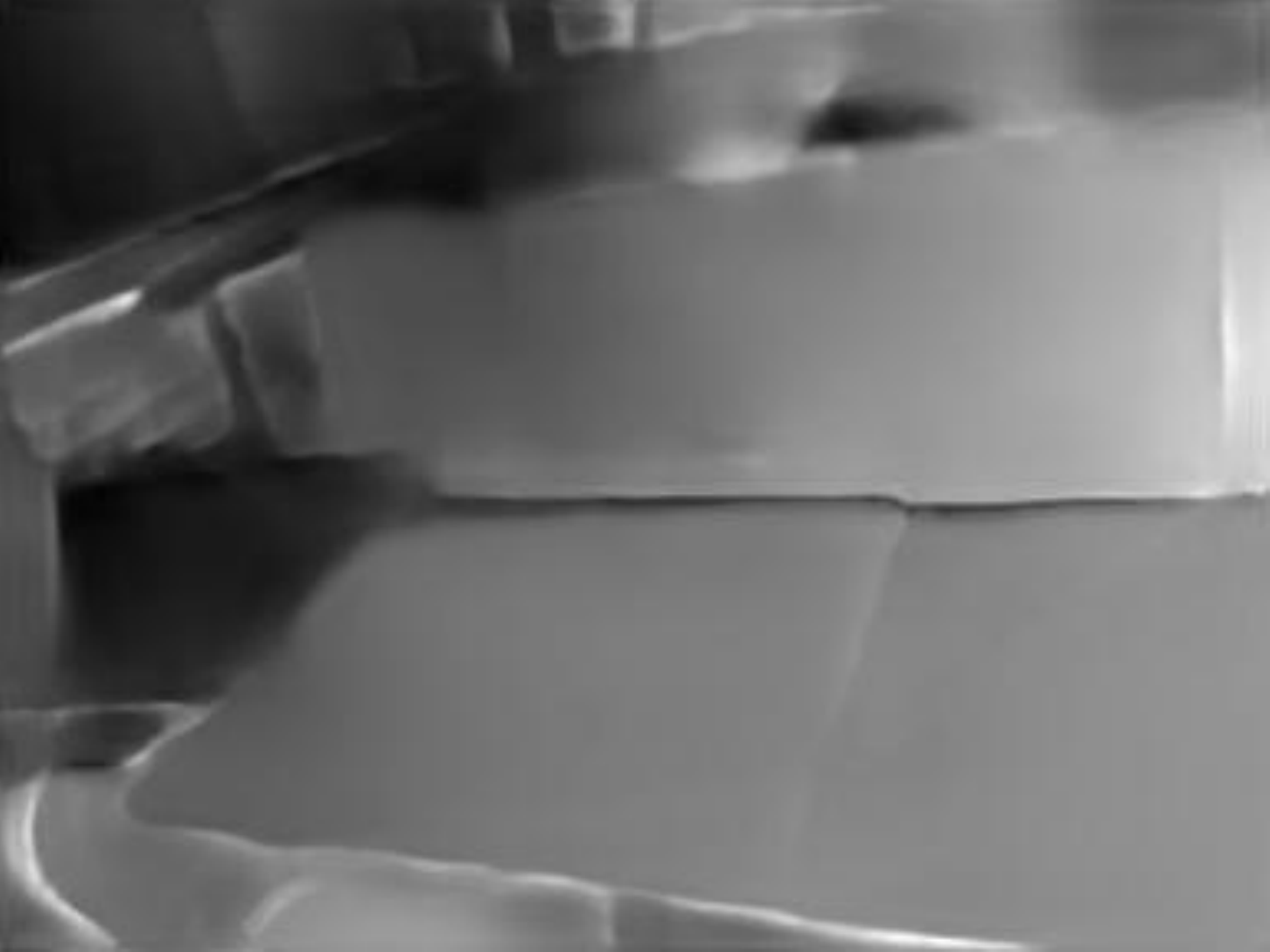}&
    \includegraphics[width=0.104\linewidth]{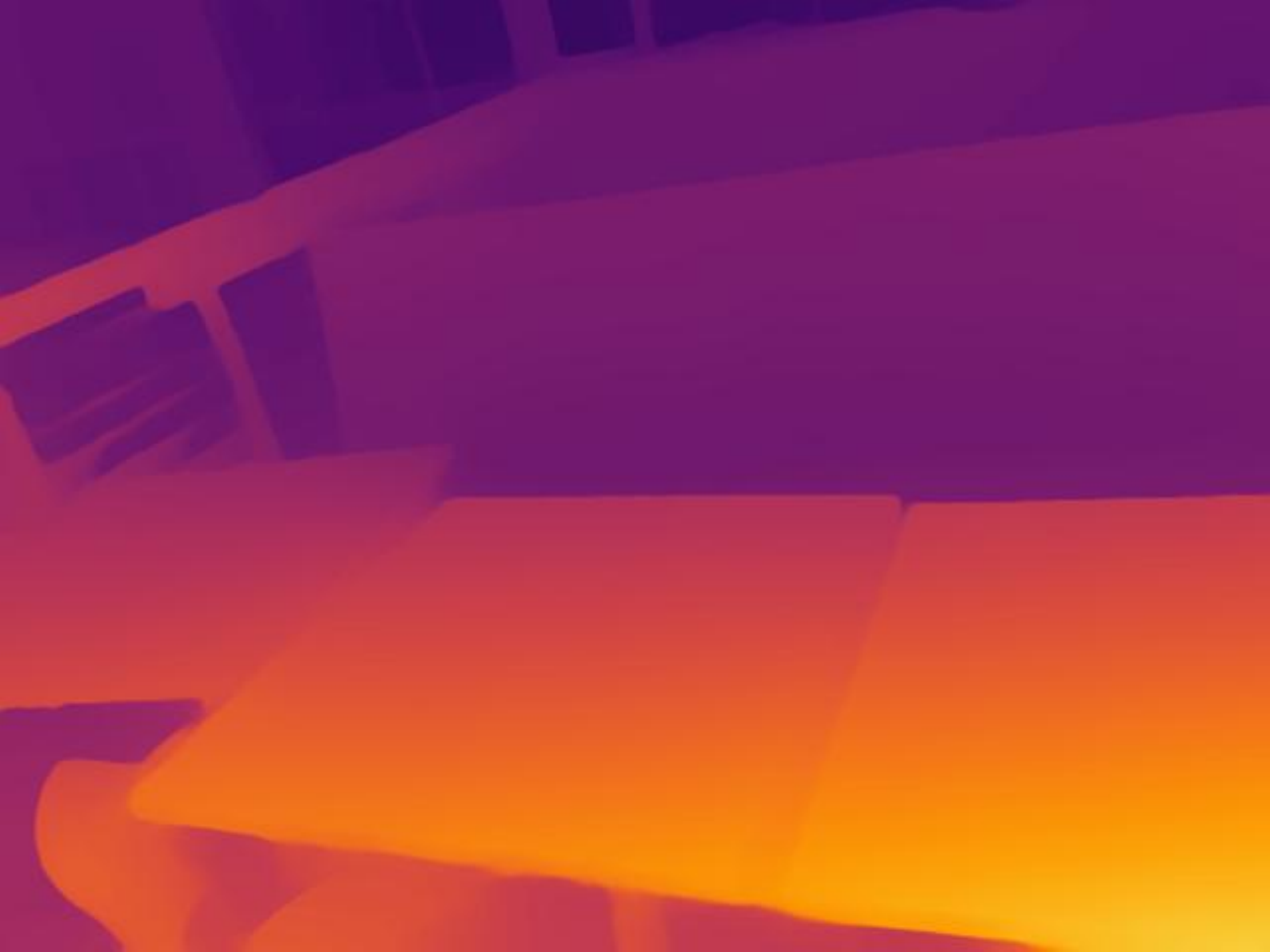}&
    \includegraphics[width=0.104\linewidth]{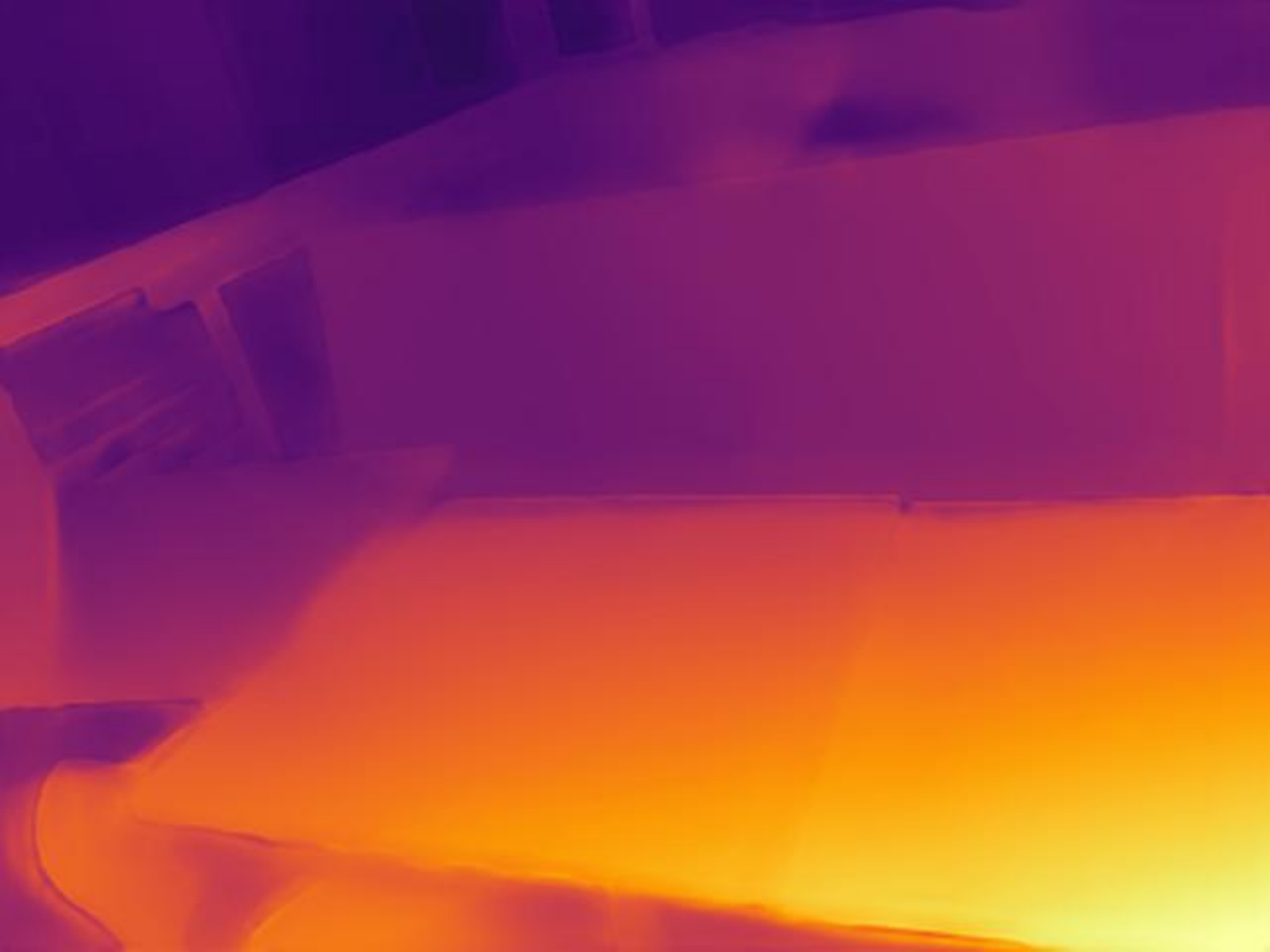}&
    \includegraphics[width=0.104\linewidth]{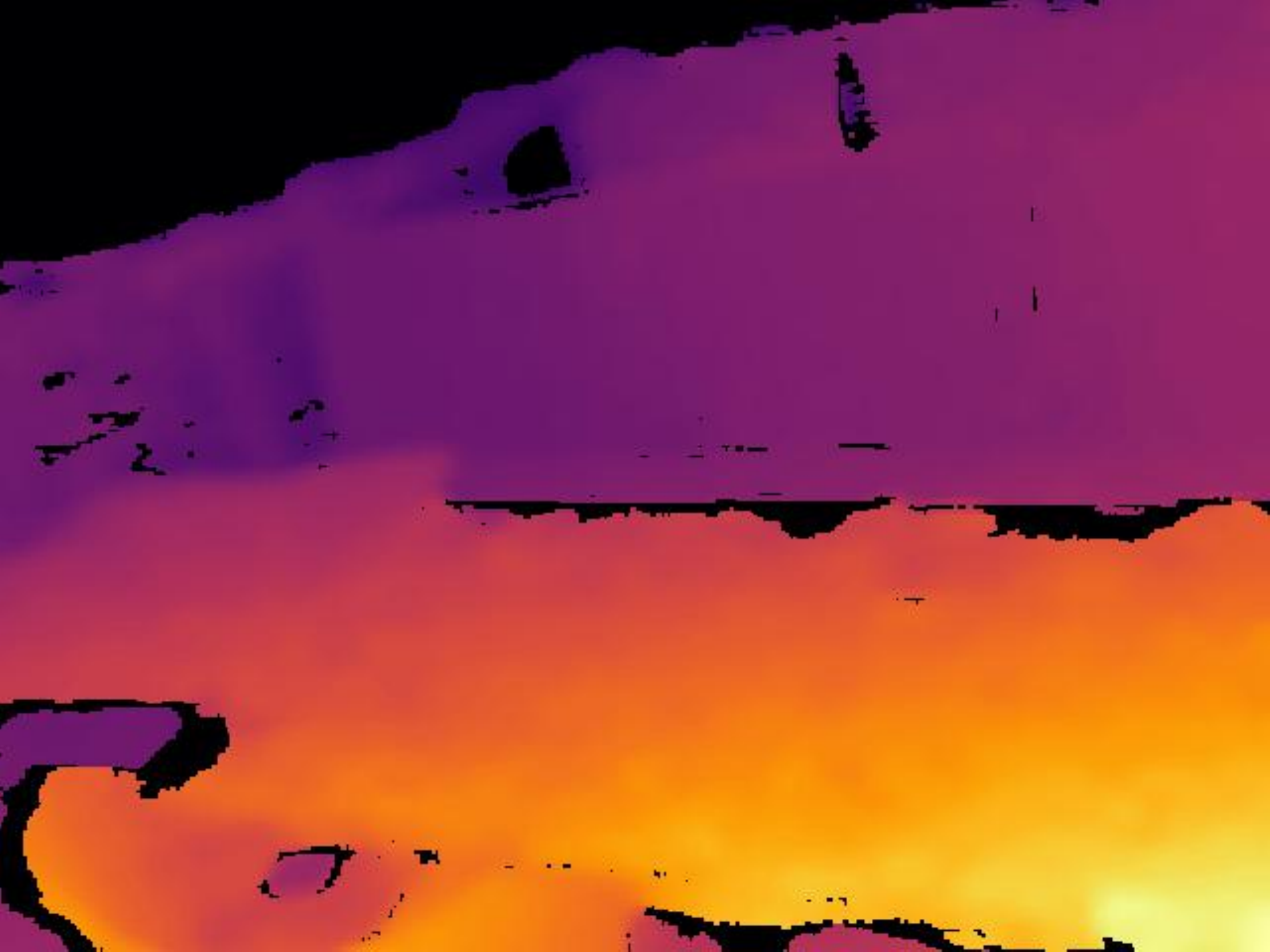}&
    \includegraphics[width=0.104\linewidth]{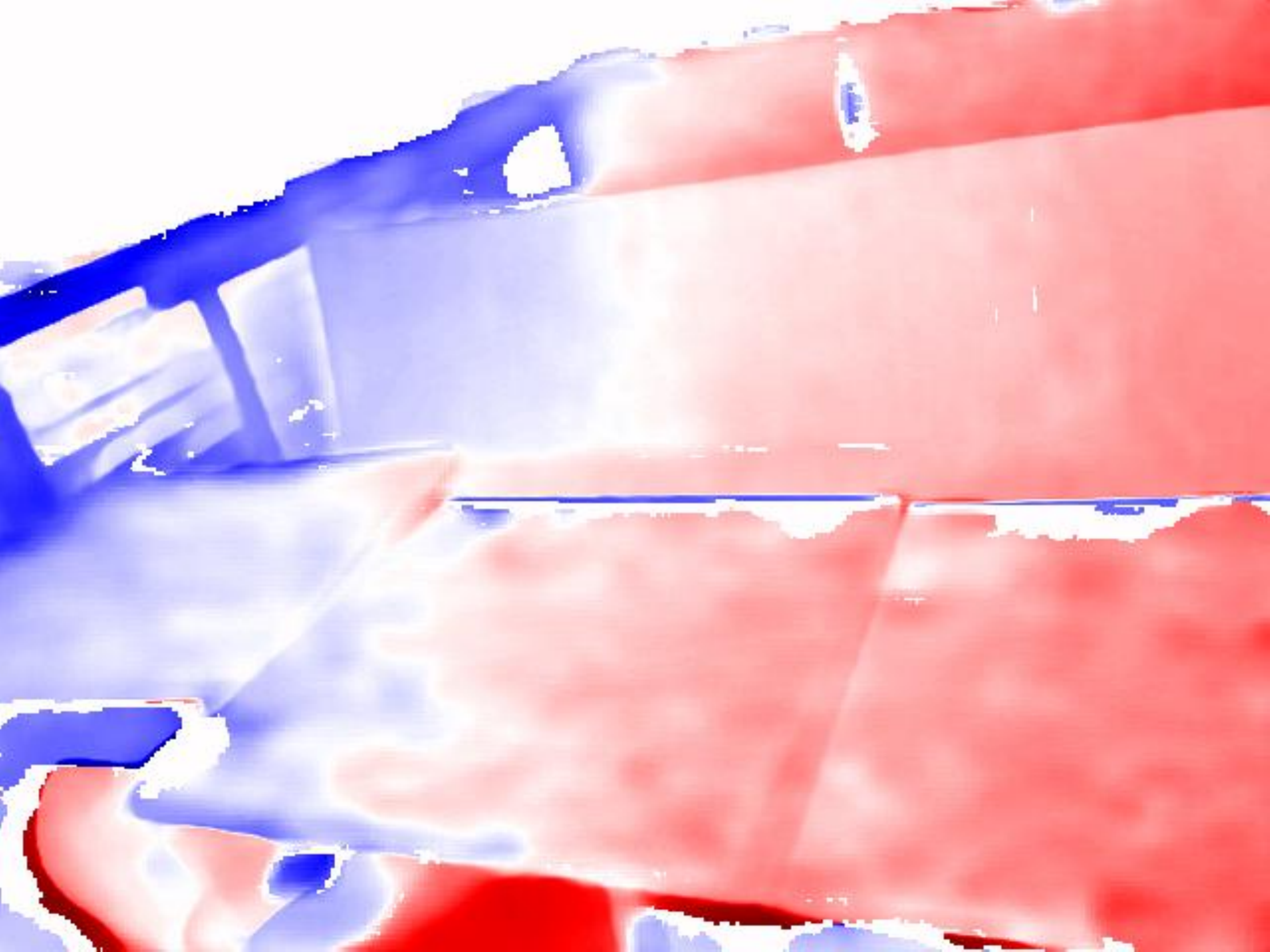}&
    \includegraphics[width=0.104\linewidth]{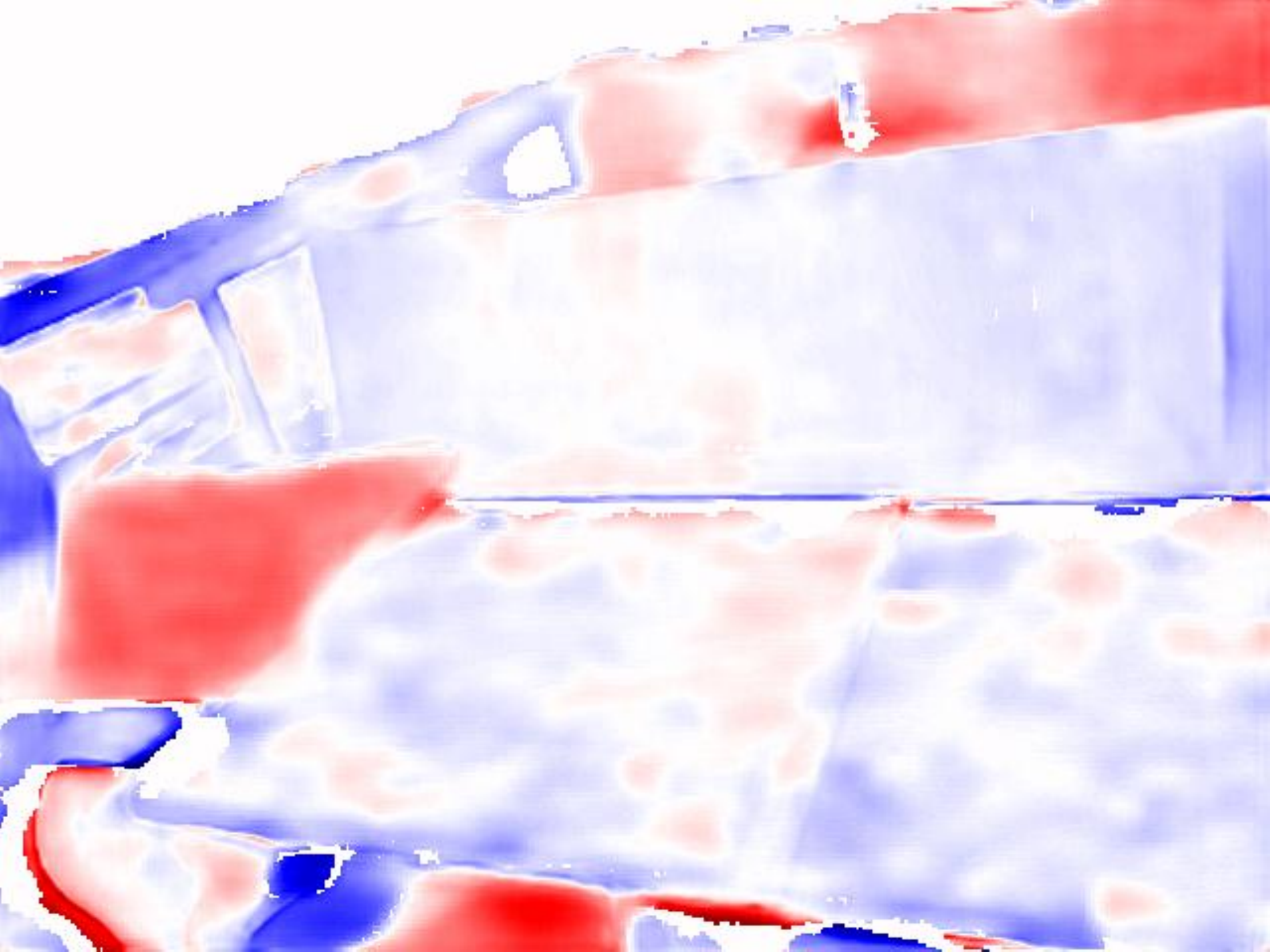}&
    \includegraphics[width=0.024\linewidth]{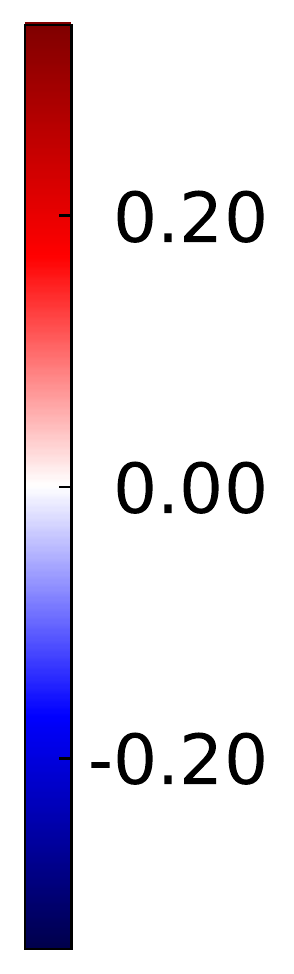}\\
    \vspace{-0.75mm}
    \scriptsize k. &
    \includegraphics[width=0.104\linewidth]{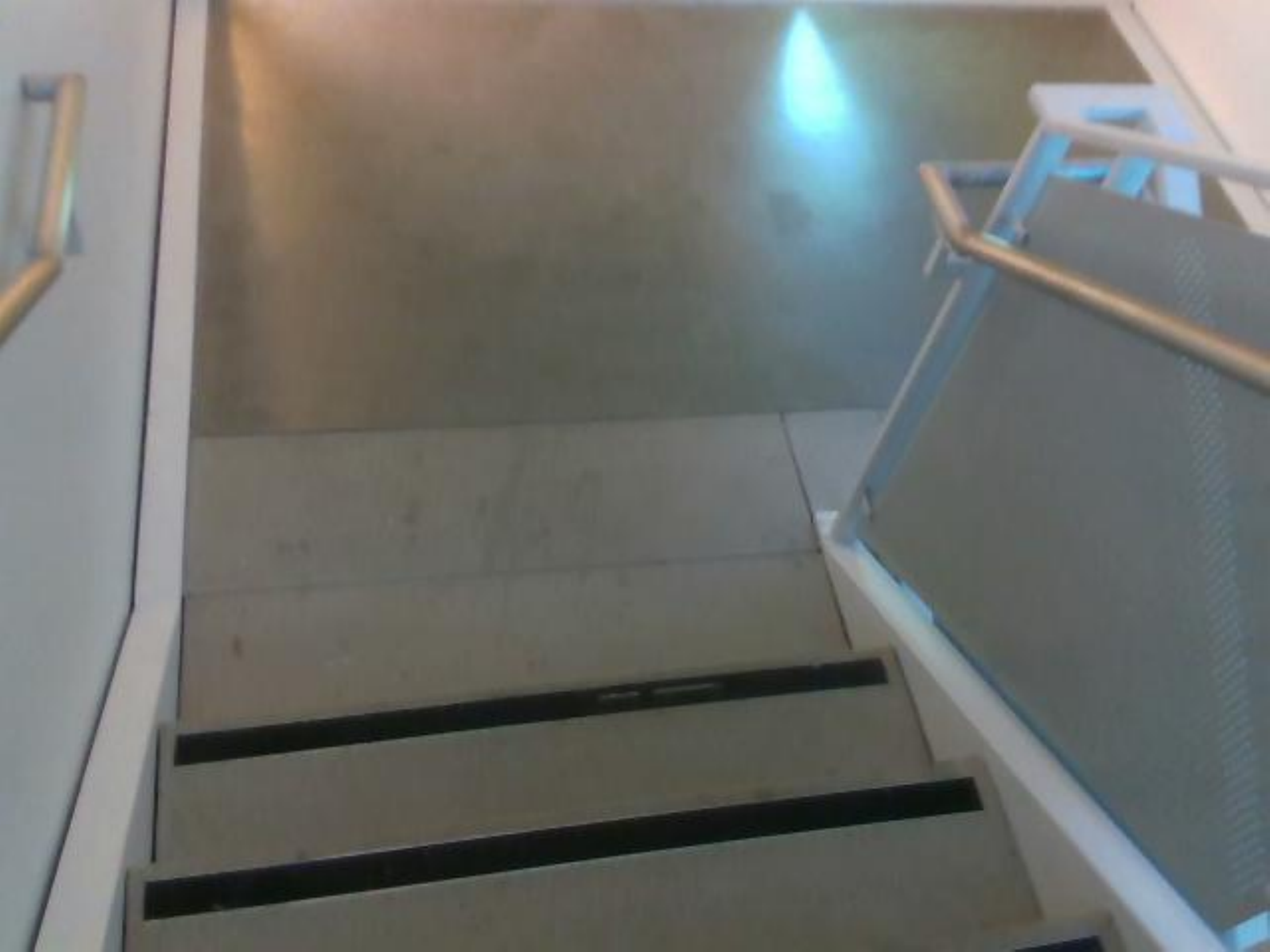}&
    \includegraphics[width=0.104\linewidth]{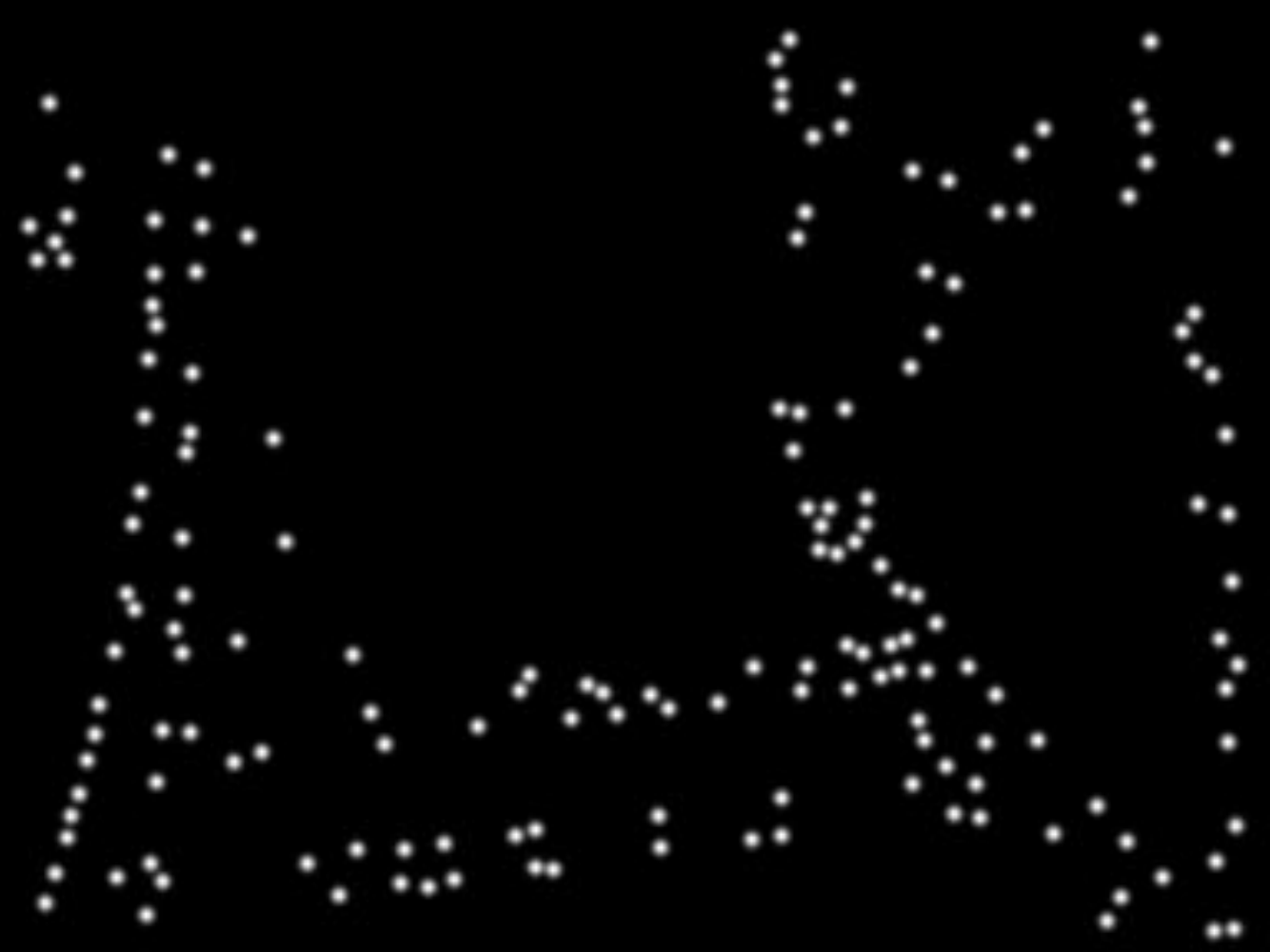}&
    \includegraphics[width=0.104\linewidth]{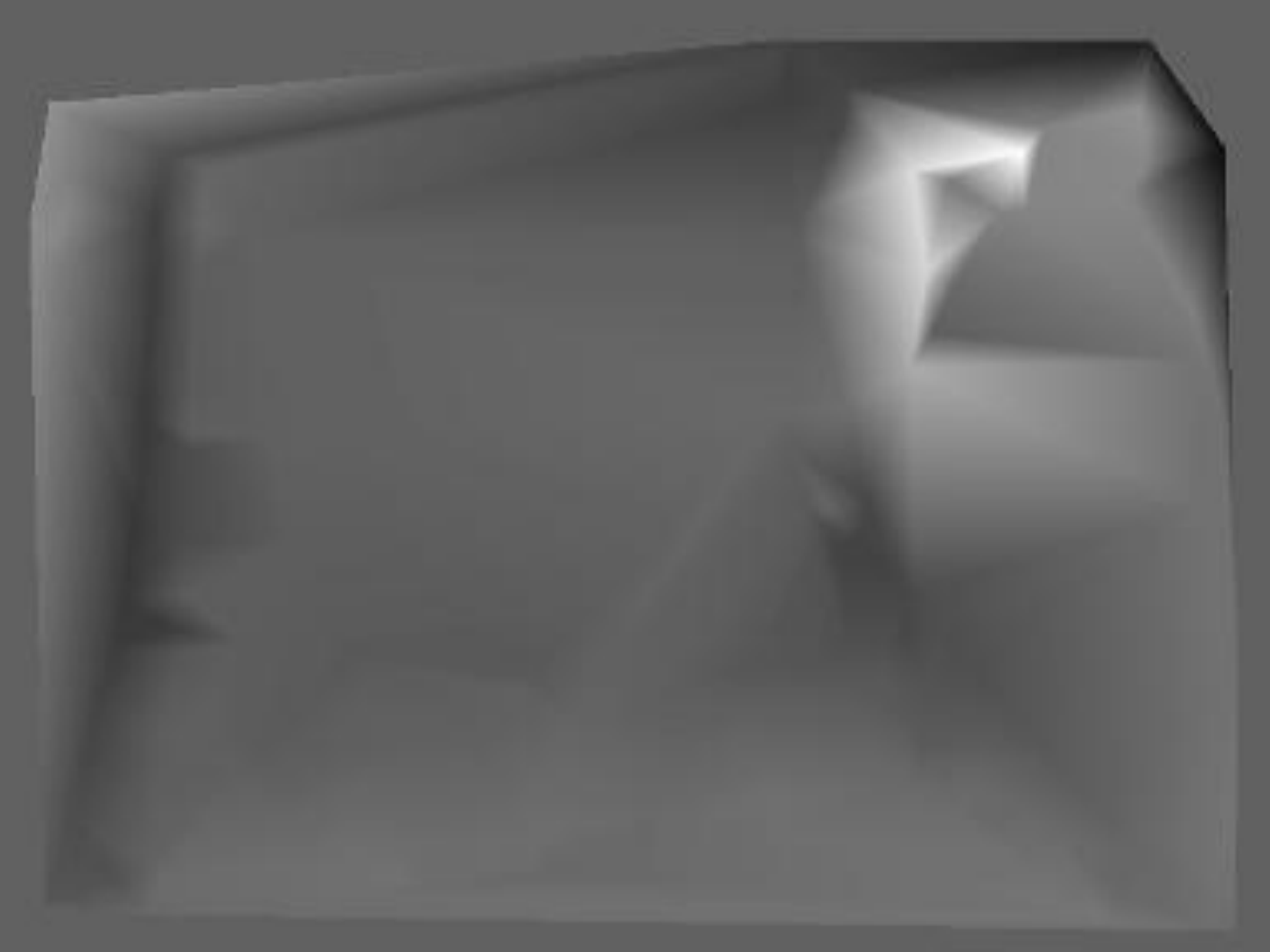}&
    \includegraphics[width=0.104\linewidth]{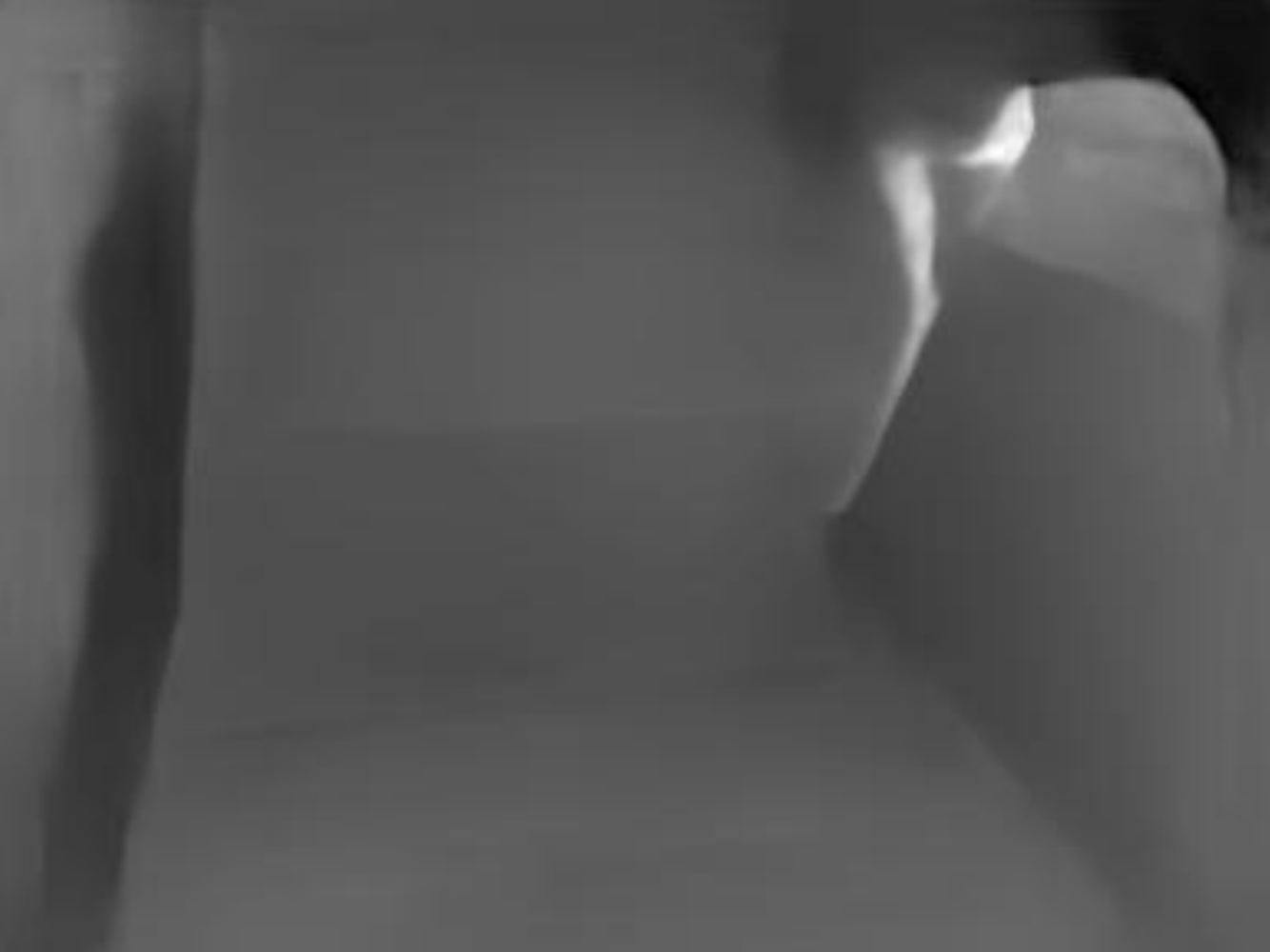}&
    \includegraphics[width=0.104\linewidth]{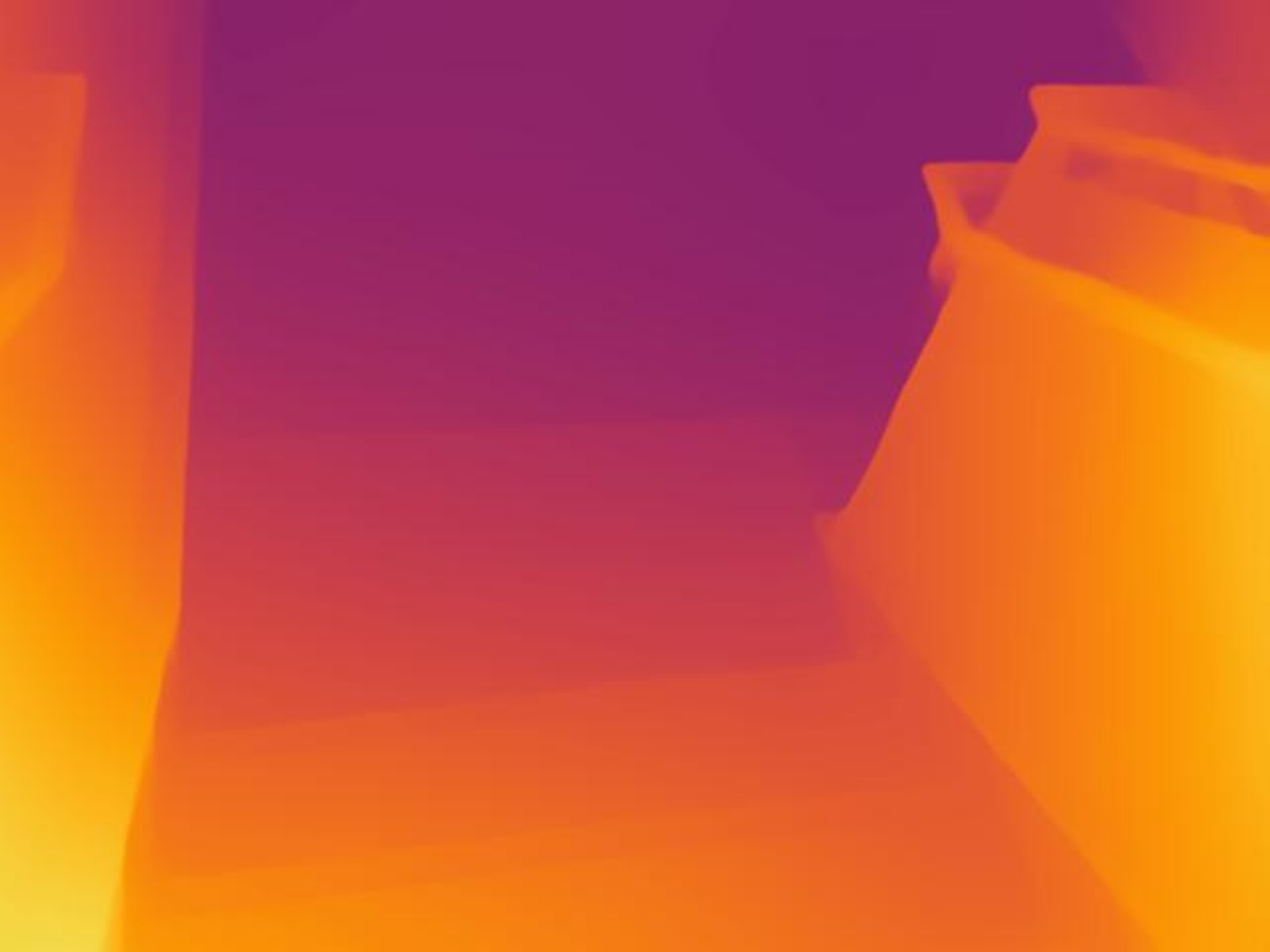}&
    \includegraphics[width=0.104\linewidth]{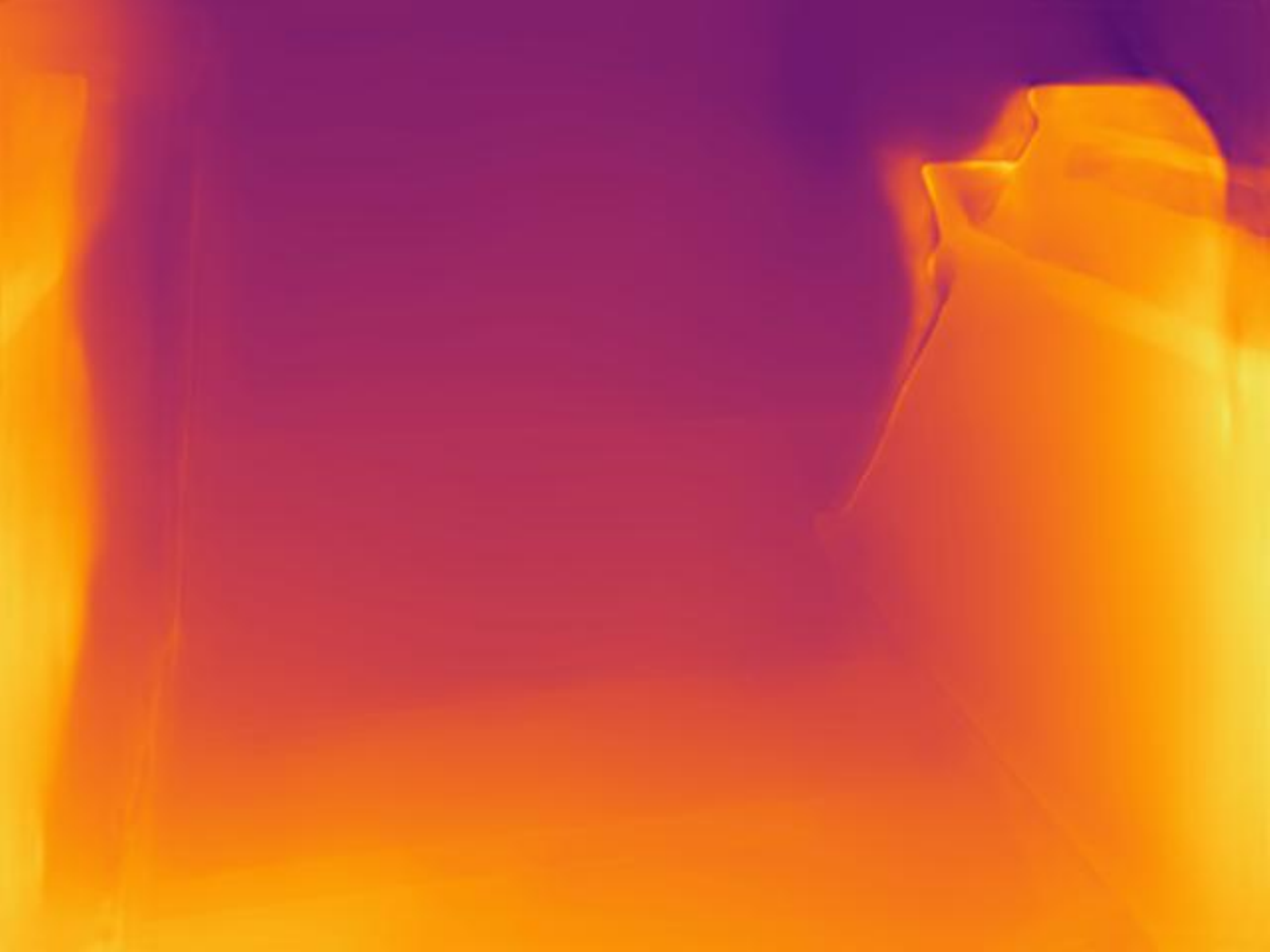}&
    \includegraphics[width=0.104\linewidth]{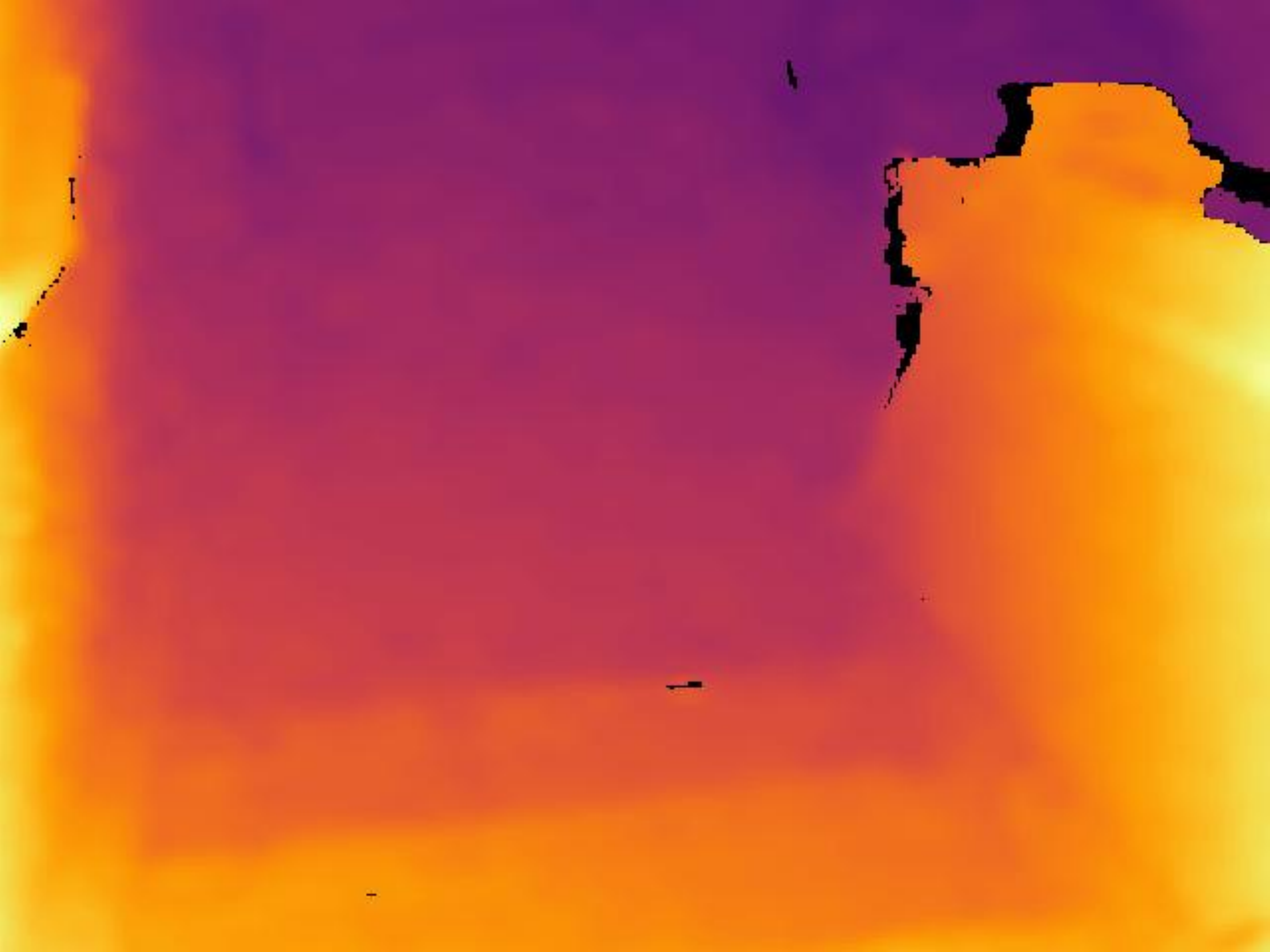}&
    \includegraphics[width=0.104\linewidth]{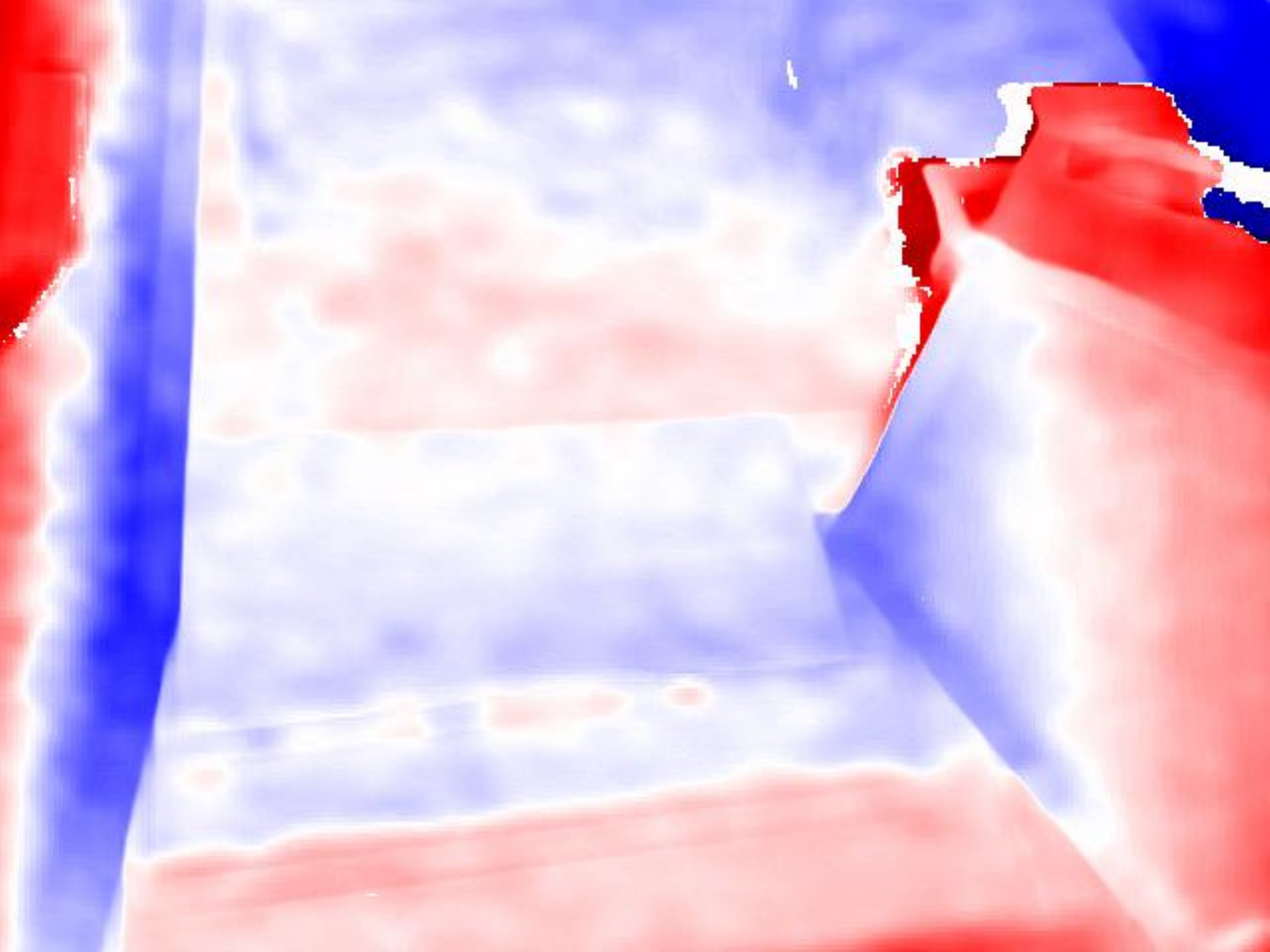}&
    \includegraphics[width=0.104\linewidth]{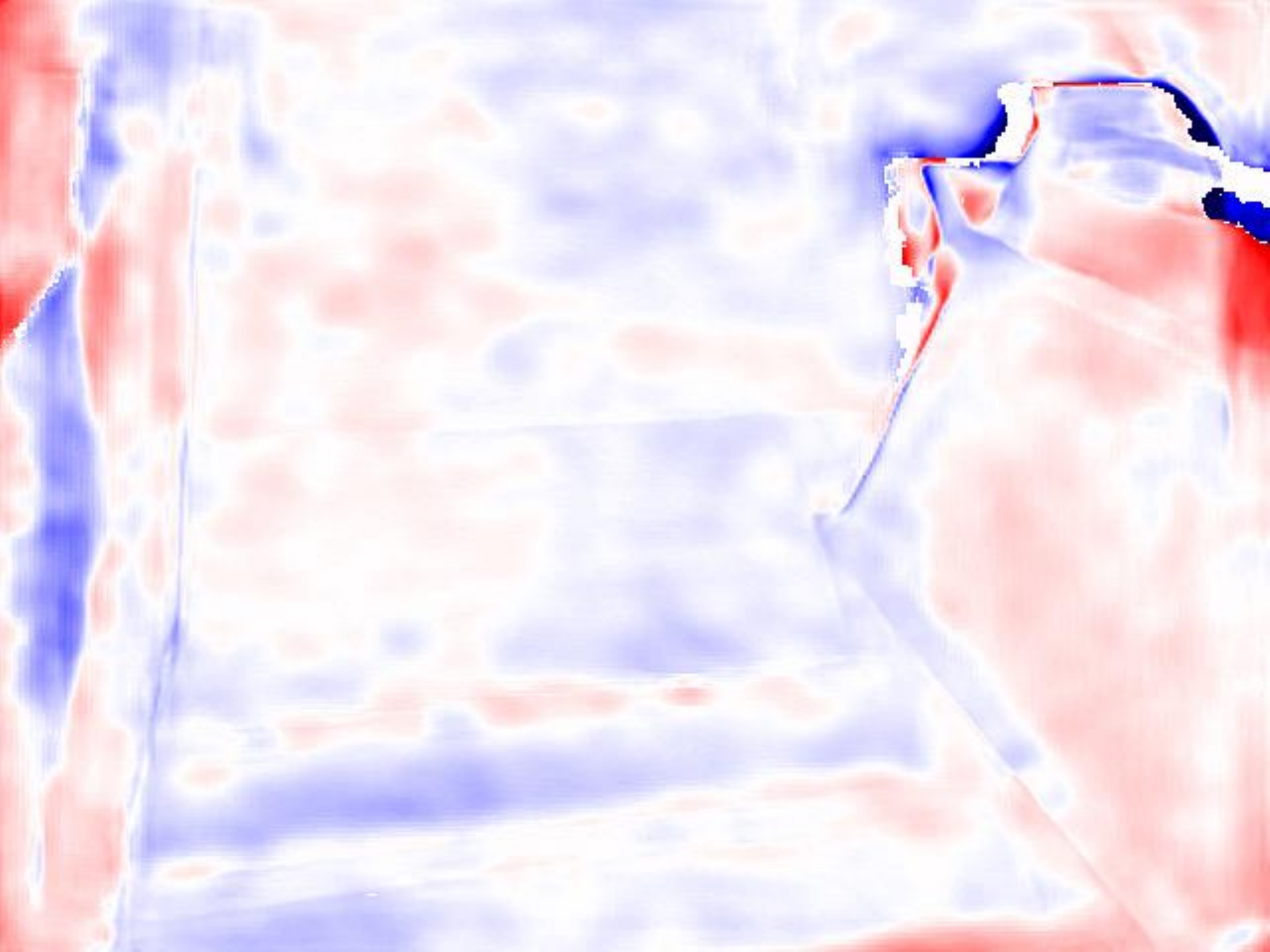}&
    \includegraphics[width=0.024\linewidth]{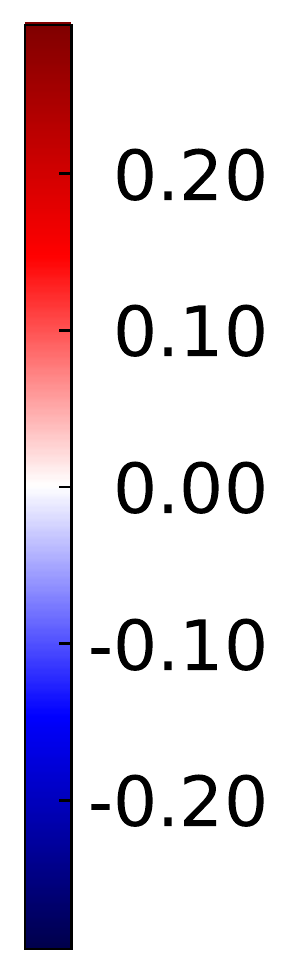}\\
  \end{tabular}
  \caption{Expanded visualization of our approach on samples from the VOID dataset (with density setting 150). In depth maps, brighter is closer and darker is farther. In error maps, red is positive inverse depth error (farther than ground truth) and blue is negative inverse depth error (closer than ground truth). Whiter regions in error maps indicate a reduction in metric depth error. The model used to generate these results was first pretrained on TartanAir and then trained on VOID, assuming a DPT-Hybrid depth estimator.}
  \label{fig:vis-void-expanded}
\end{figure*}

We provide visualizations for more samples from the VOID~\cite{Wong2020void} test set. In addition to depth and error maps, we visualize confidence maps pinpointing known sparse metric depth locations, as well as the scale map scaffolding input to SML and the scale map regressed by SML. Figure~\ref{fig:vis-void-expanded} shows this expanded visualization.

\begin{figure*}[p]
\centering
  \begin{tabular}{@{}l@{\hspace{0.5mm}}*{10}{c@{\hspace{0.5mm}}}@{}}
    &  &  &  & {\scriptsize GA+SML} & {\scriptsize GA+SML} & {\scriptsize GA+SML} & {\scriptsize GA+SML} & {\scriptsize GA+SML} & {\scriptsize GA+SML} & {\scriptsize GA+SML}\\
    & {\scriptsize RGB Image} & {\scriptsize Ground Truth} & {\scriptsize KBNet~\cite{Wong2021kbnet}} & {\scriptsize (DPT-BEiT-L)} & {\scriptsize (DPT-SwinV2-L)} & {\scriptsize (DPT-Large)} & {\scriptsize (DPT-Hybrid)} & {\scriptsize (DPT-SwinV2-T)} & {\scriptsize (DPT-LeViT)} & {\scriptsize (MiDaS-small)}\\
    \vspace{-0.75mm}
    \scriptsize a. &
    \includegraphics[width=0.096\linewidth]{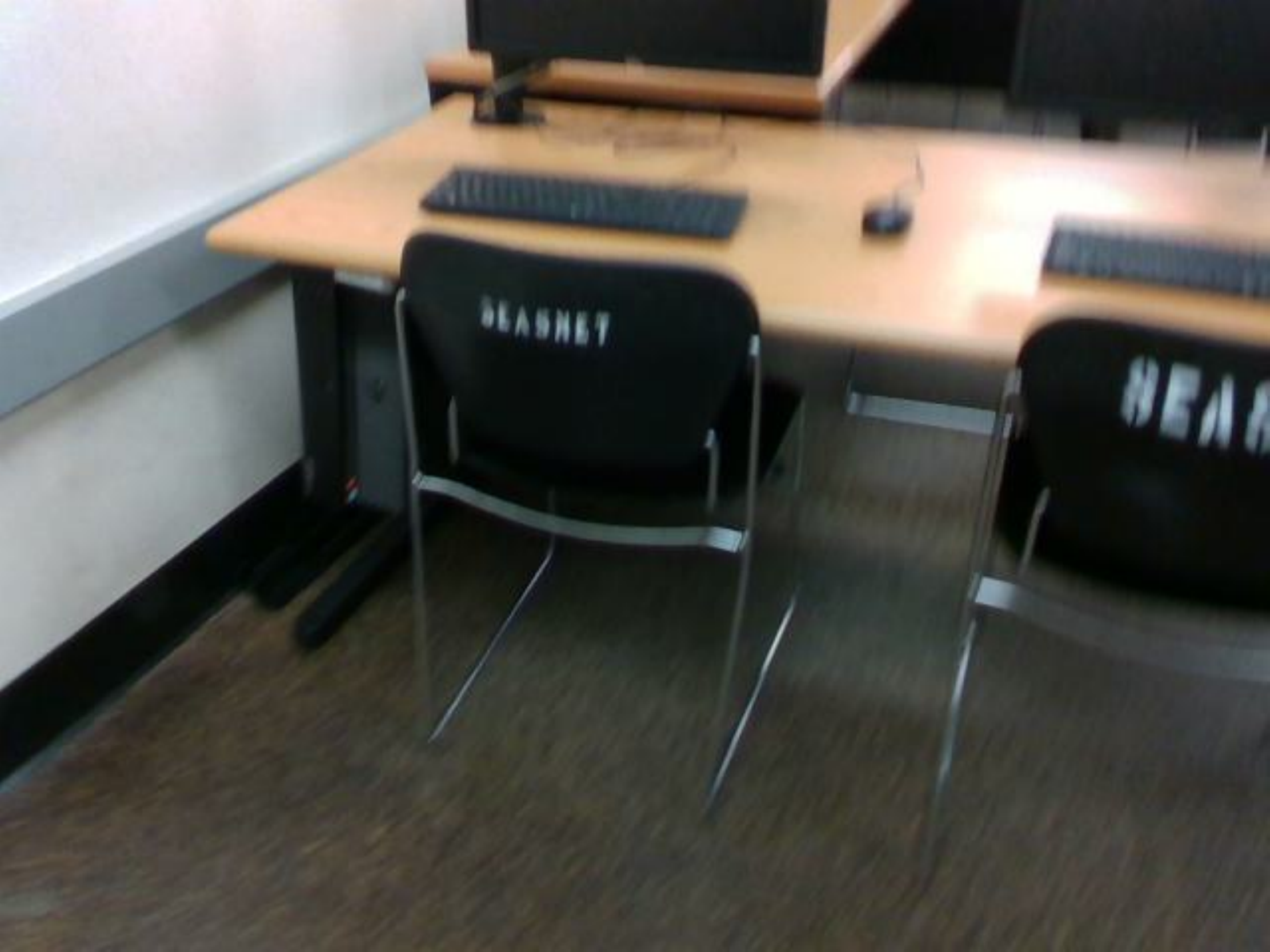}&
    \includegraphics[width=0.096\linewidth]{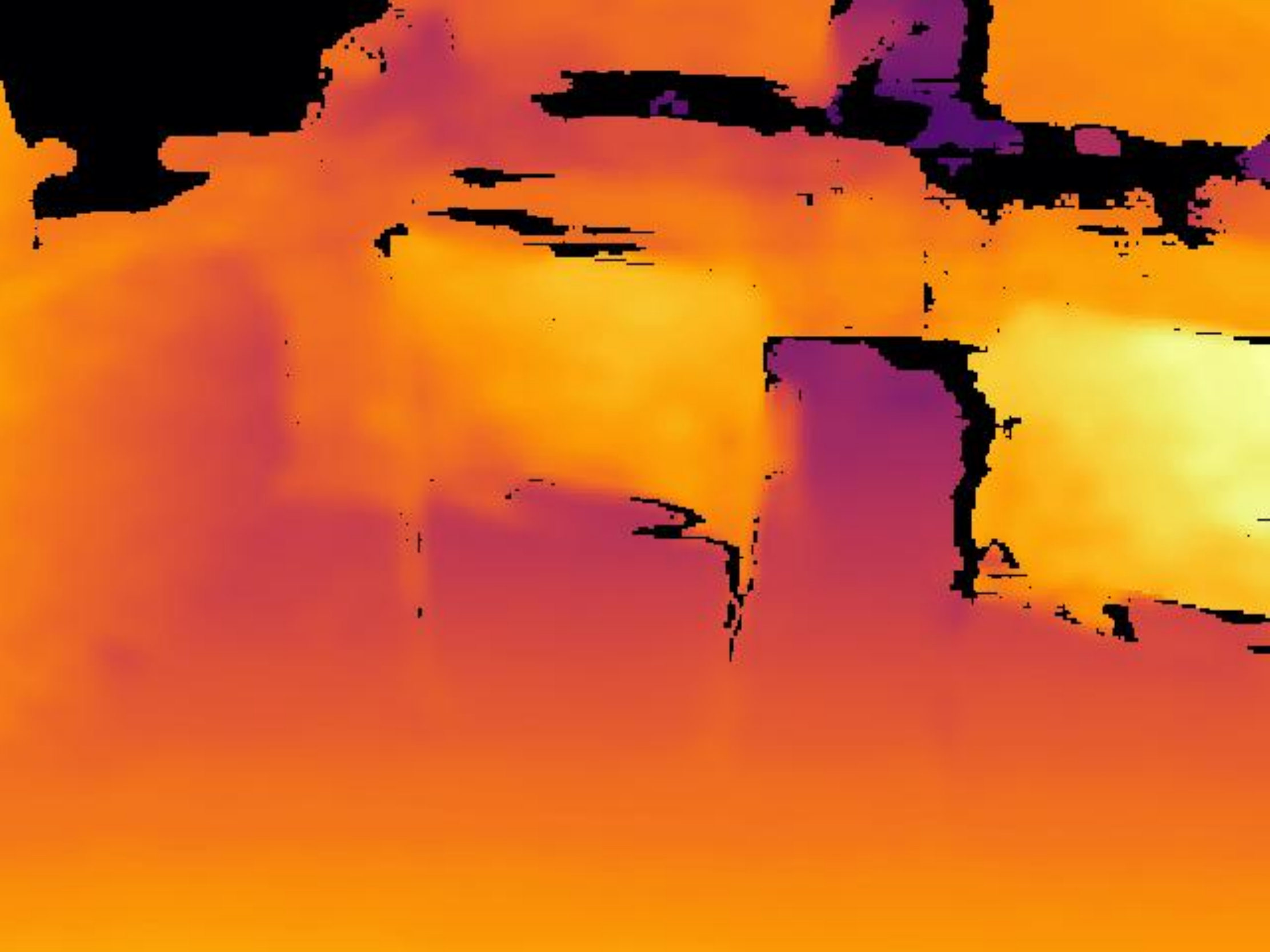}&
    \includegraphics[width=0.096\linewidth]{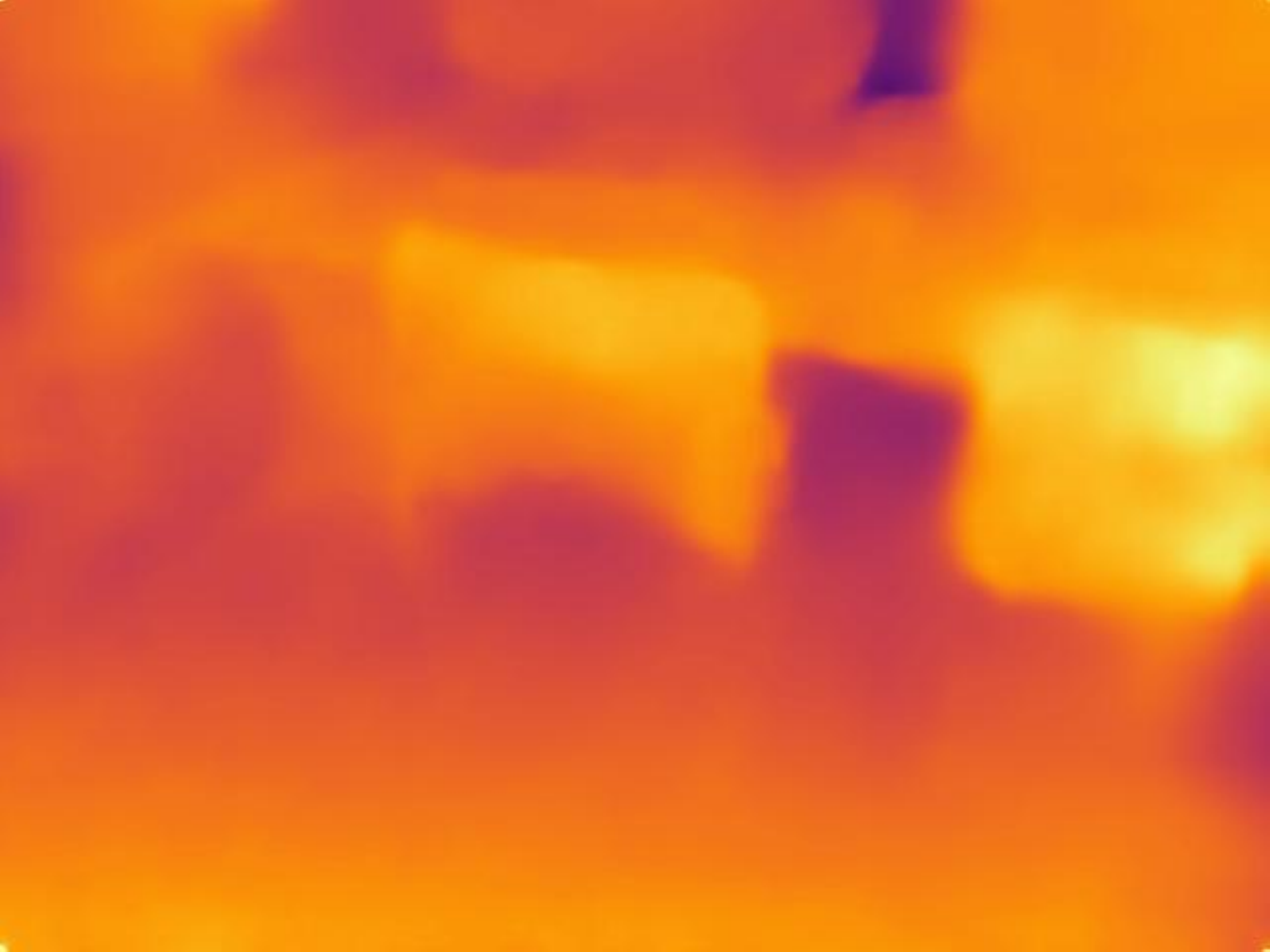}&
    \includegraphics[width=0.096\linewidth]{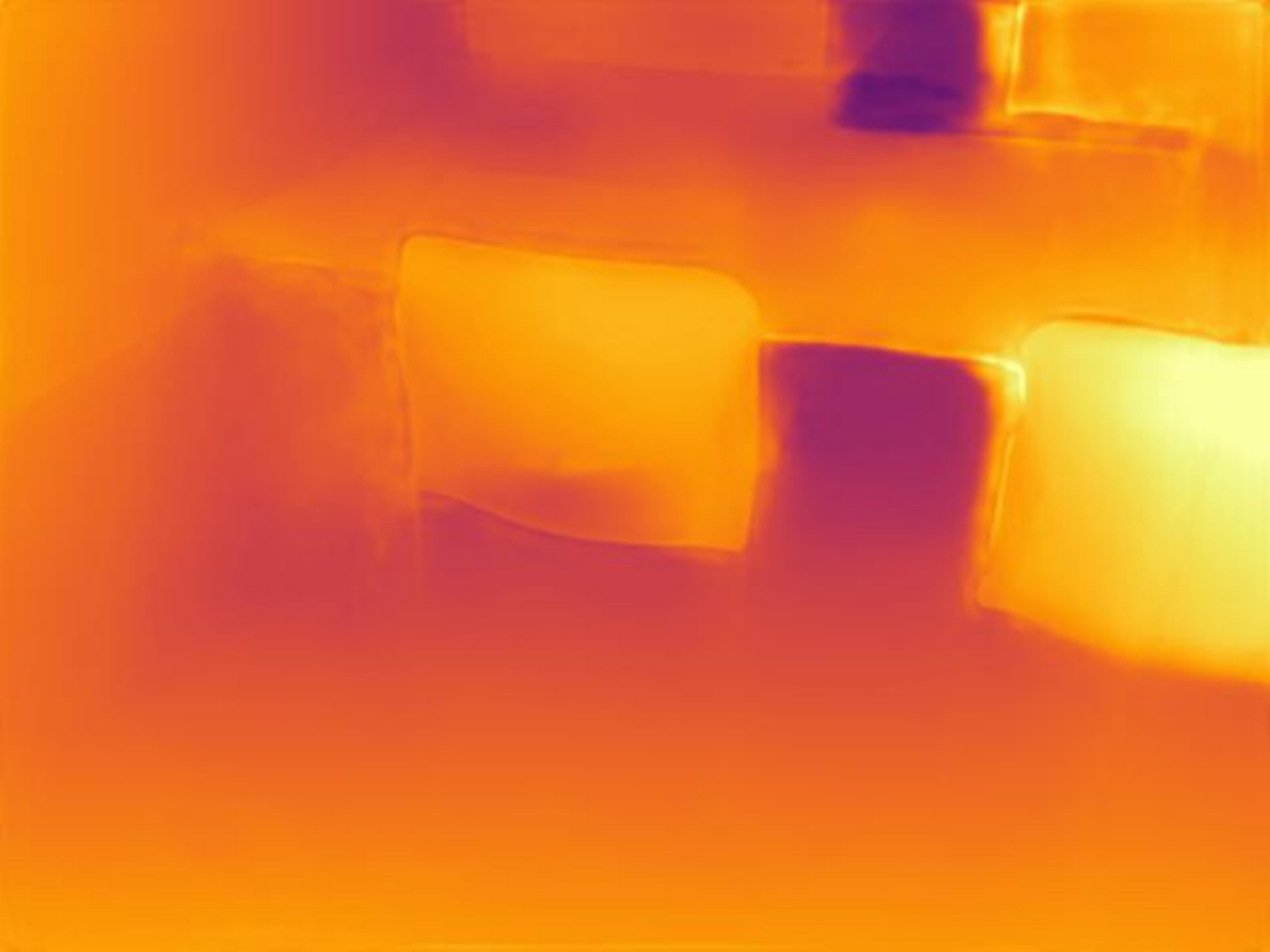}&
    \includegraphics[width=0.096\linewidth]{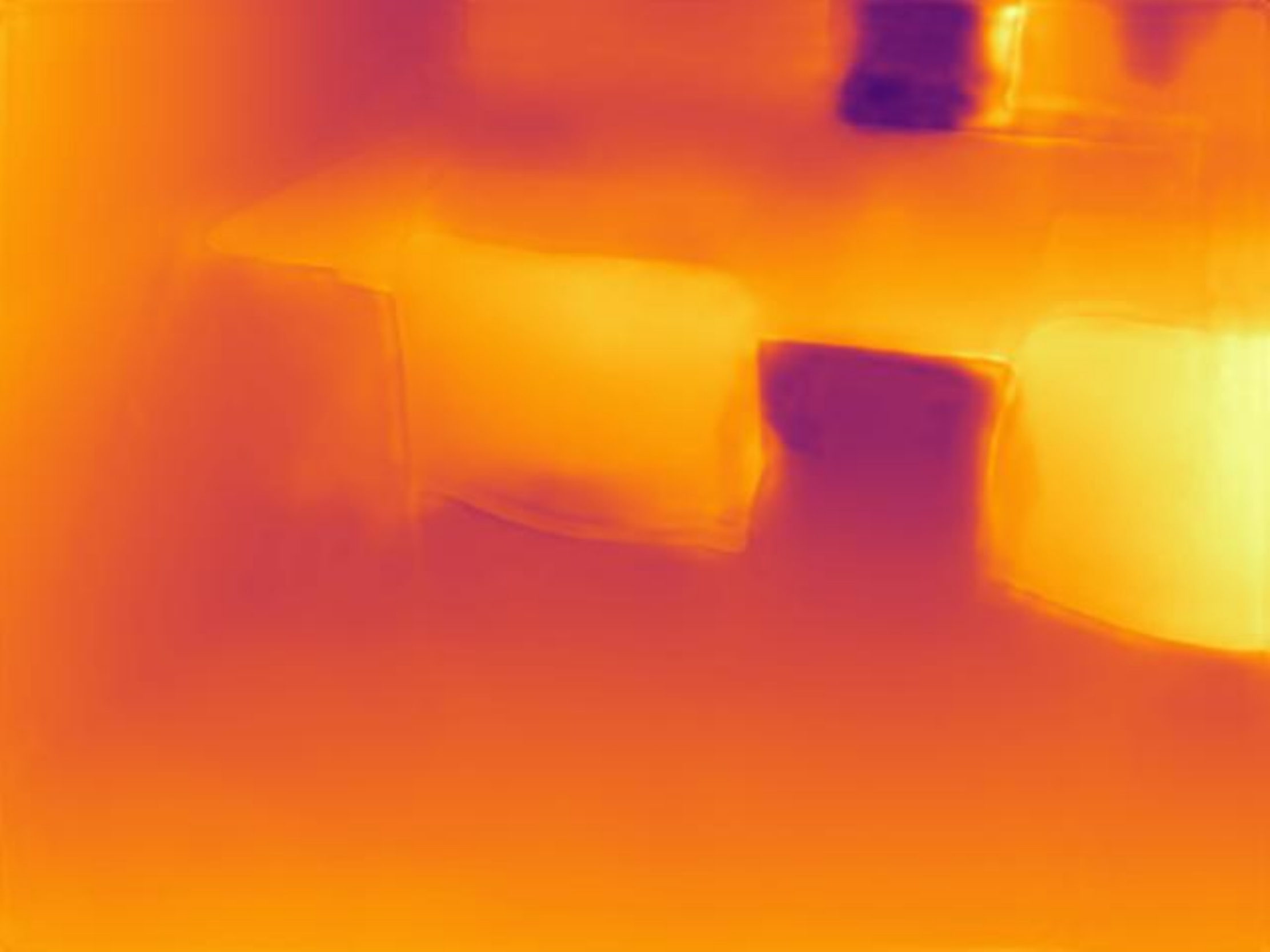}&
    \includegraphics[width=0.096\linewidth]{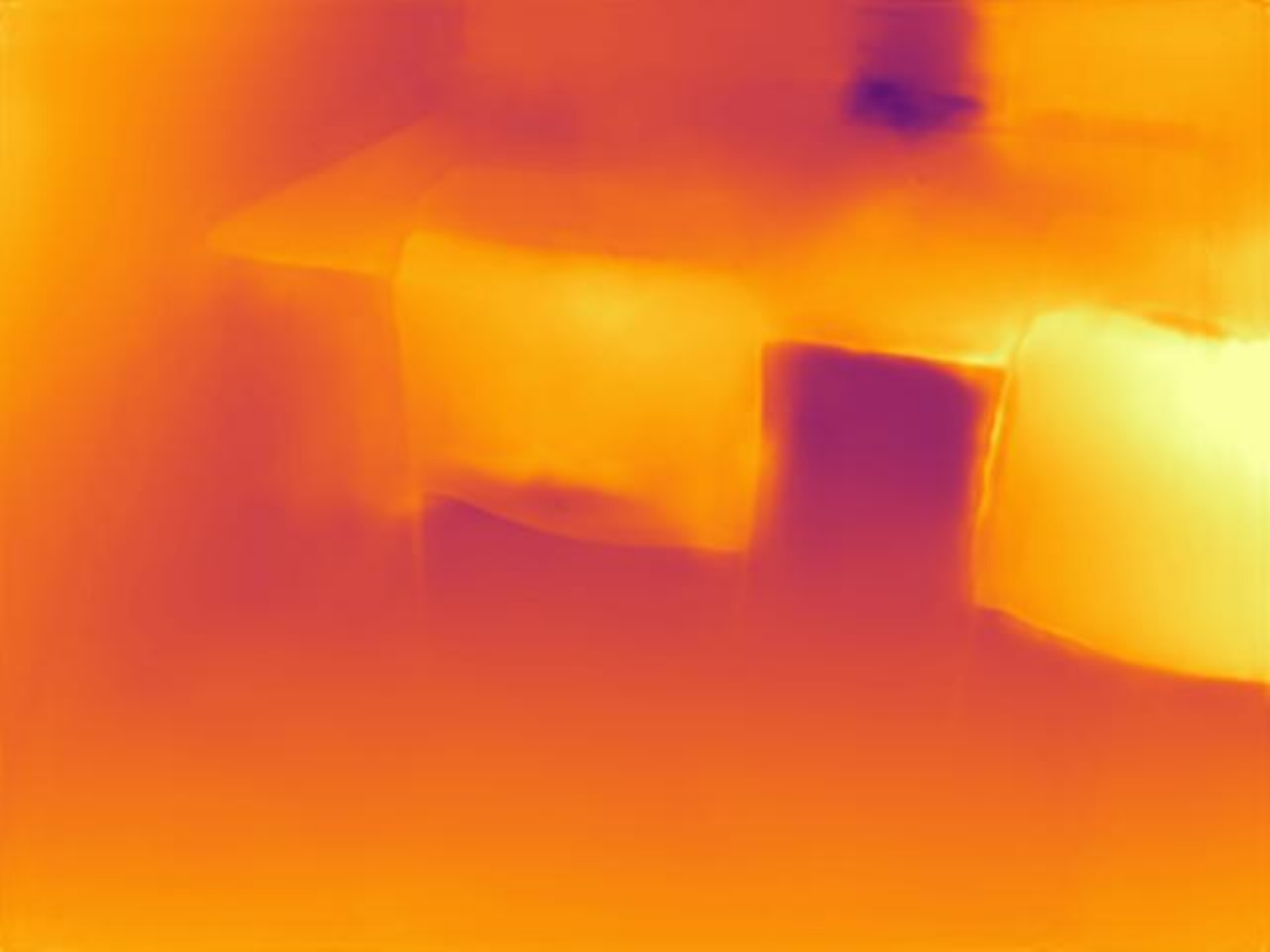}&
    \includegraphics[width=0.096\linewidth]{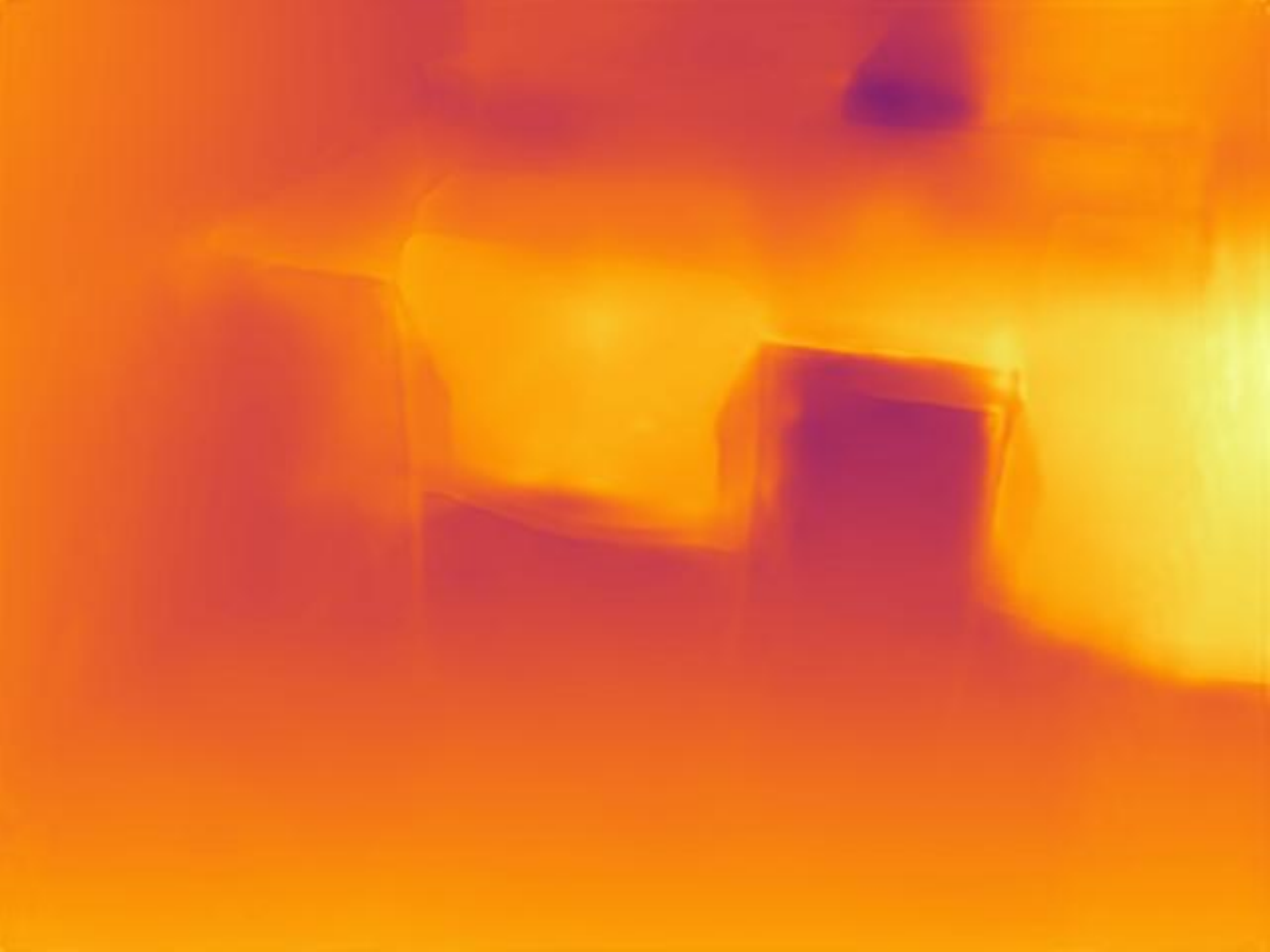}&
    \includegraphics[width=0.096\linewidth]{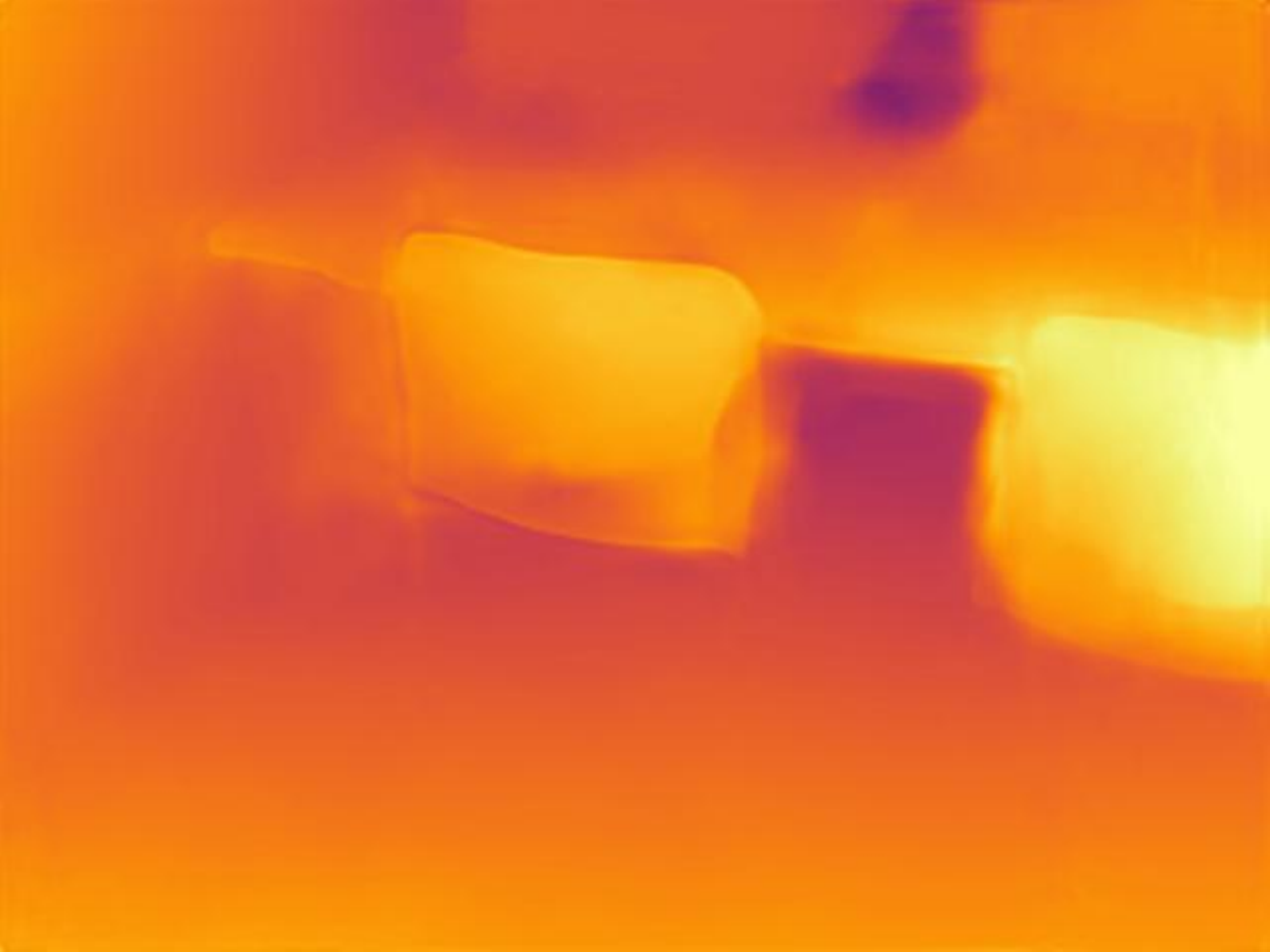}&
    \includegraphics[width=0.096\linewidth]{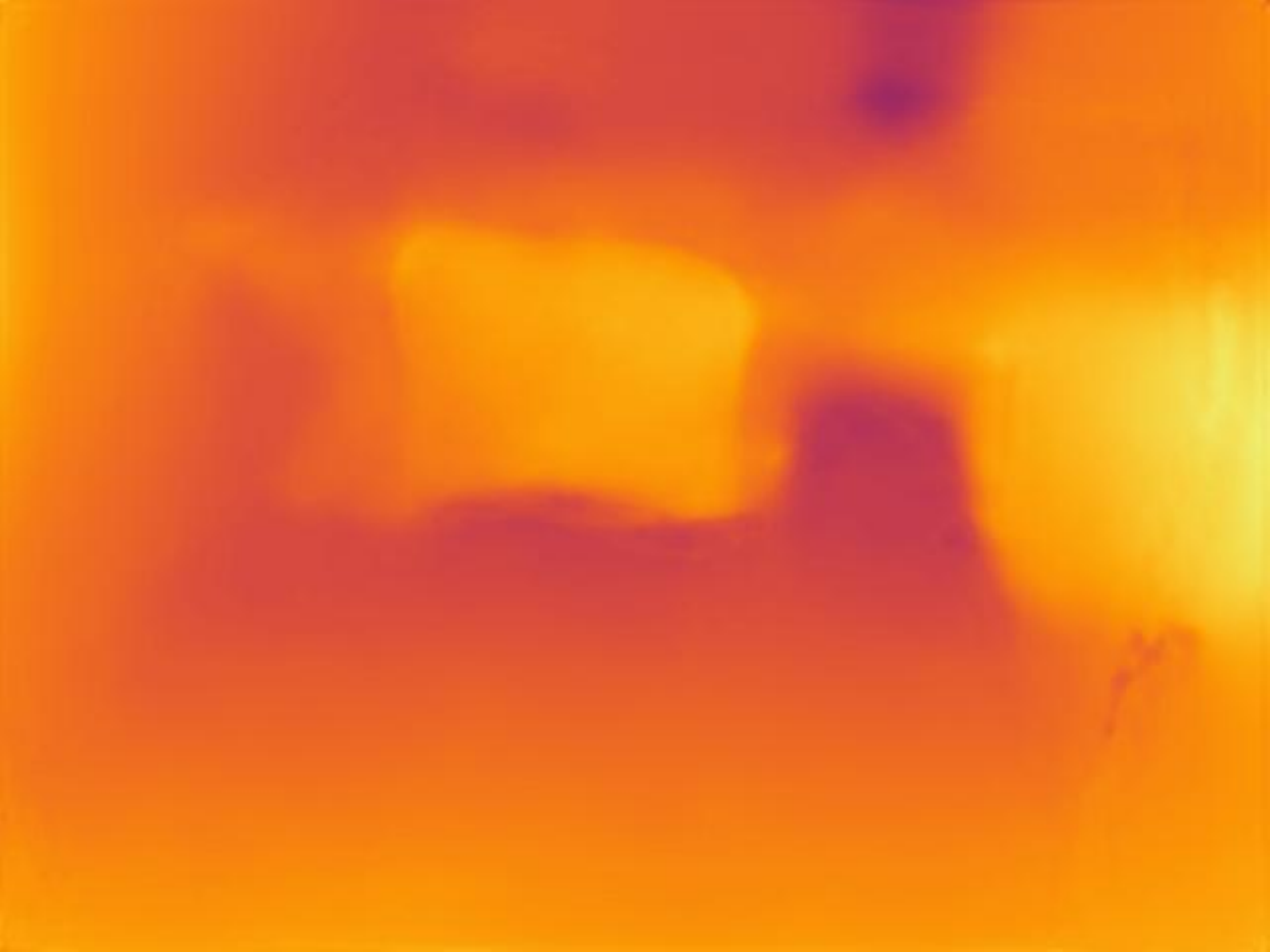}&
    \includegraphics[width=0.096\linewidth]{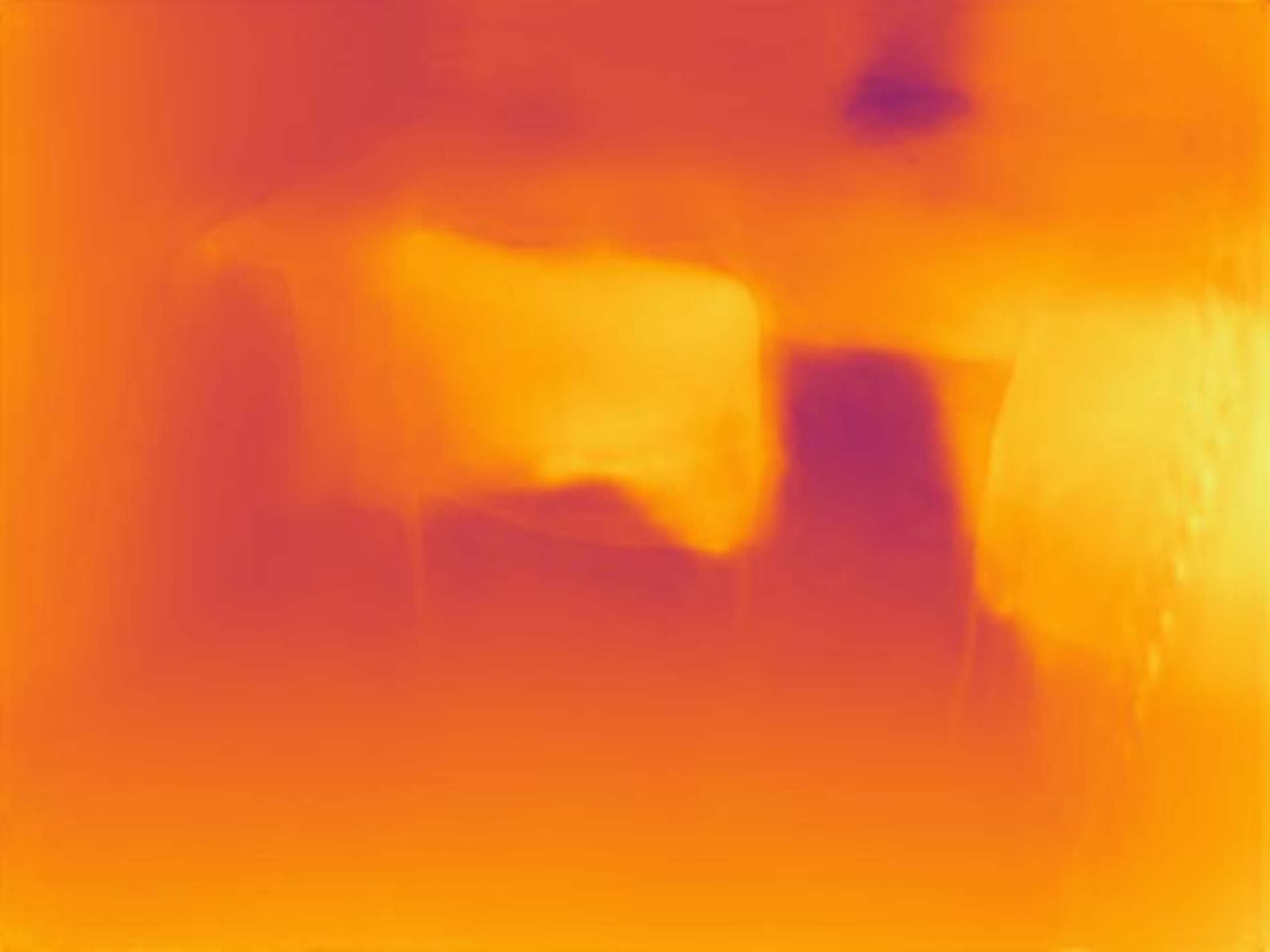}\\
    \vspace{-0.75mm}
    \scriptsize b. &
    \includegraphics[width=0.096\linewidth]{suppl/comparison_void_150/kbnet/image/0000000060.pdf}&
    \includegraphics[width=0.096\linewidth]{suppl/comparison_void_150/kbnet/ground_truth/0000000060.pdf}&
    \includegraphics[width=0.096\linewidth]{suppl/comparison_void_150/kbnet/output_depth/0000000060.pdf}&
    \includegraphics[width=0.096\linewidth]{suppl/comparison_void_150/dpt-beit-l/row_1_col_5.pdf}&
    \includegraphics[width=0.096\linewidth]{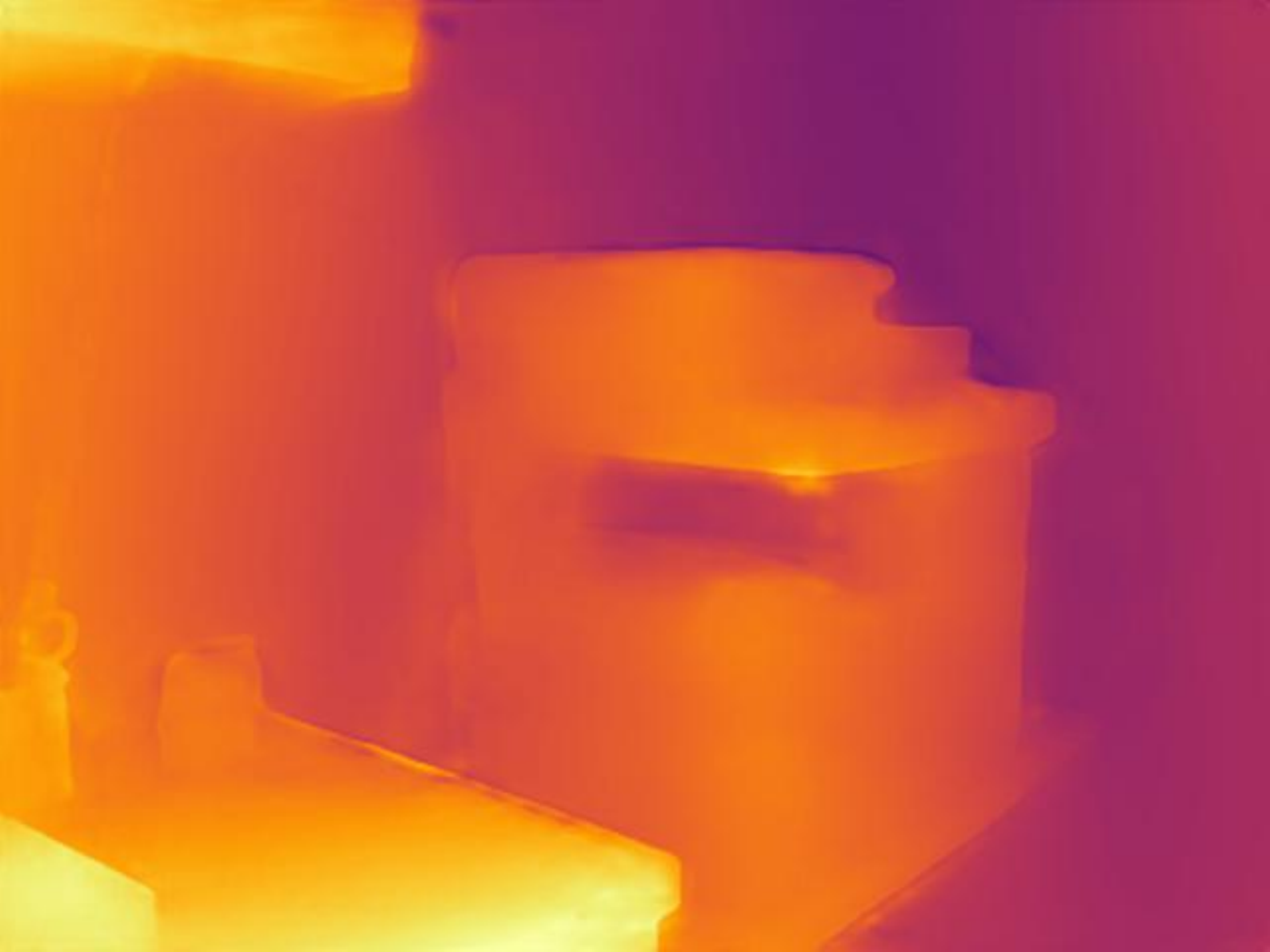}&
    \includegraphics[width=0.096\linewidth]{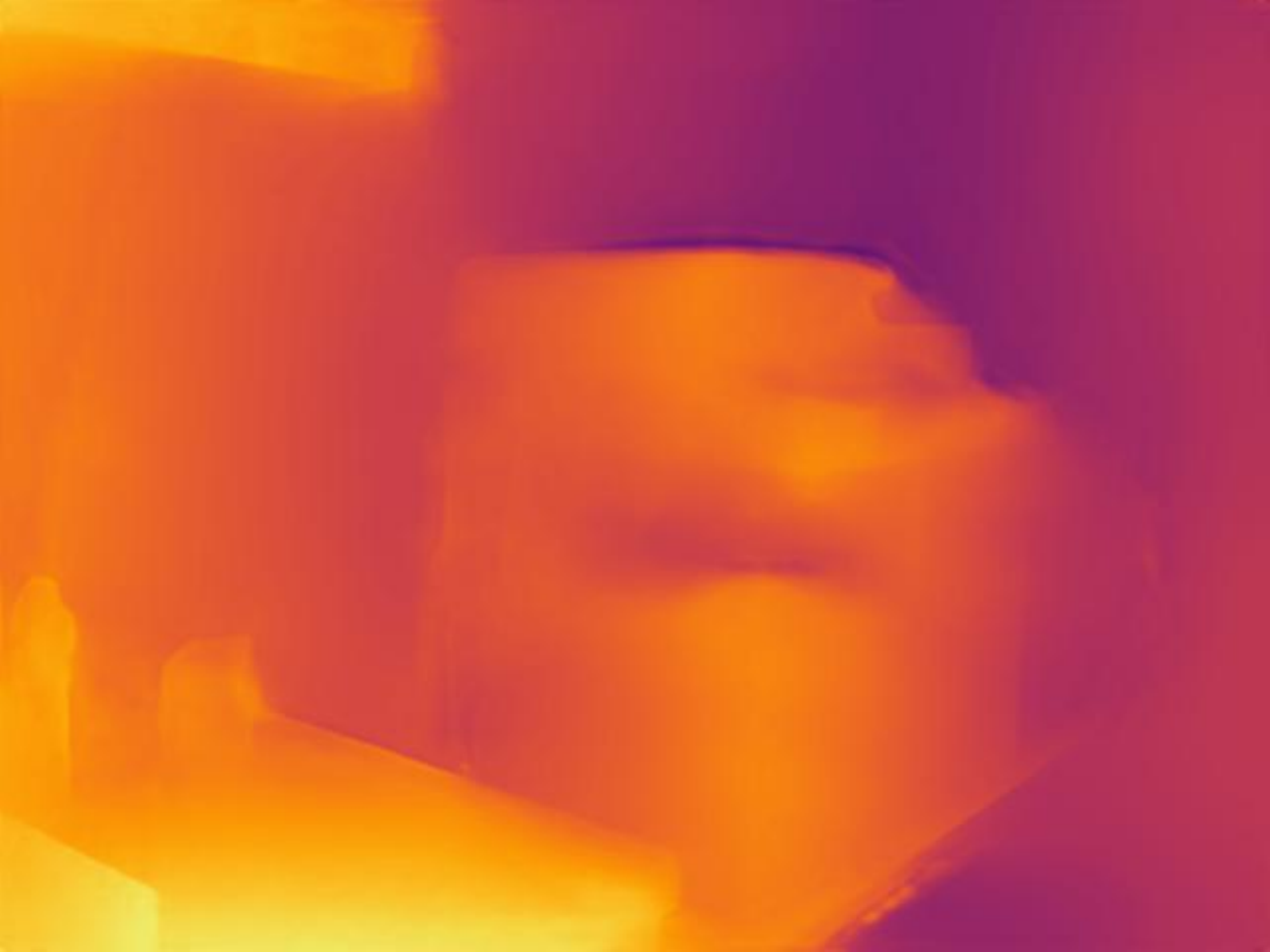}&
    \includegraphics[width=0.096\linewidth]{suppl/comparison_void_150/dpt-hybrid/row_1_col_5.pdf}&
    \includegraphics[width=0.096\linewidth]{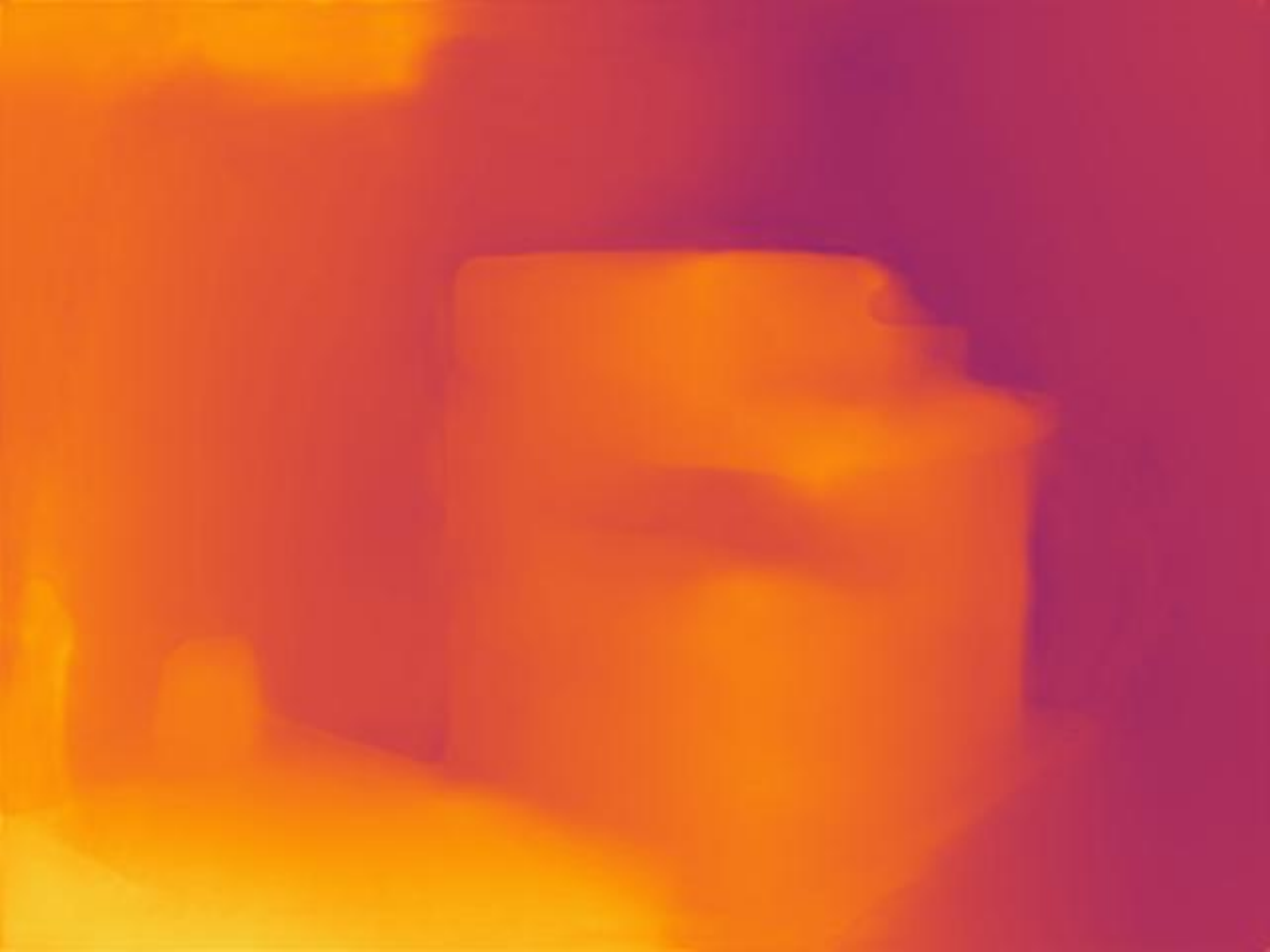}&
    \includegraphics[width=0.096\linewidth]{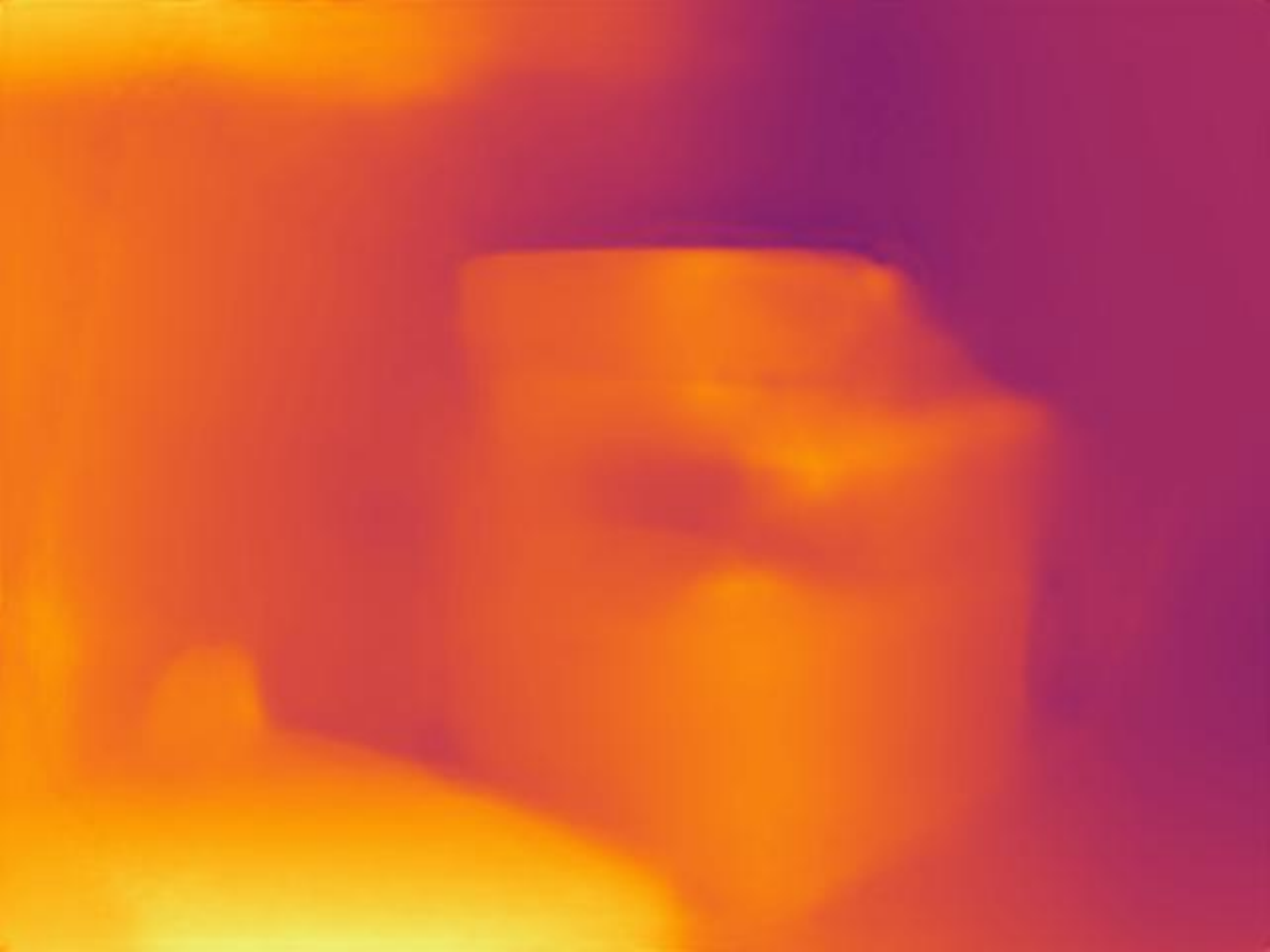}&
    \includegraphics[width=0.096\linewidth]{suppl/comparison_void_150/midas-small/row_1_col_5.pdf}\\
    \vspace{-0.75mm}
    \scriptsize c. &
    \includegraphics[width=0.096\linewidth]{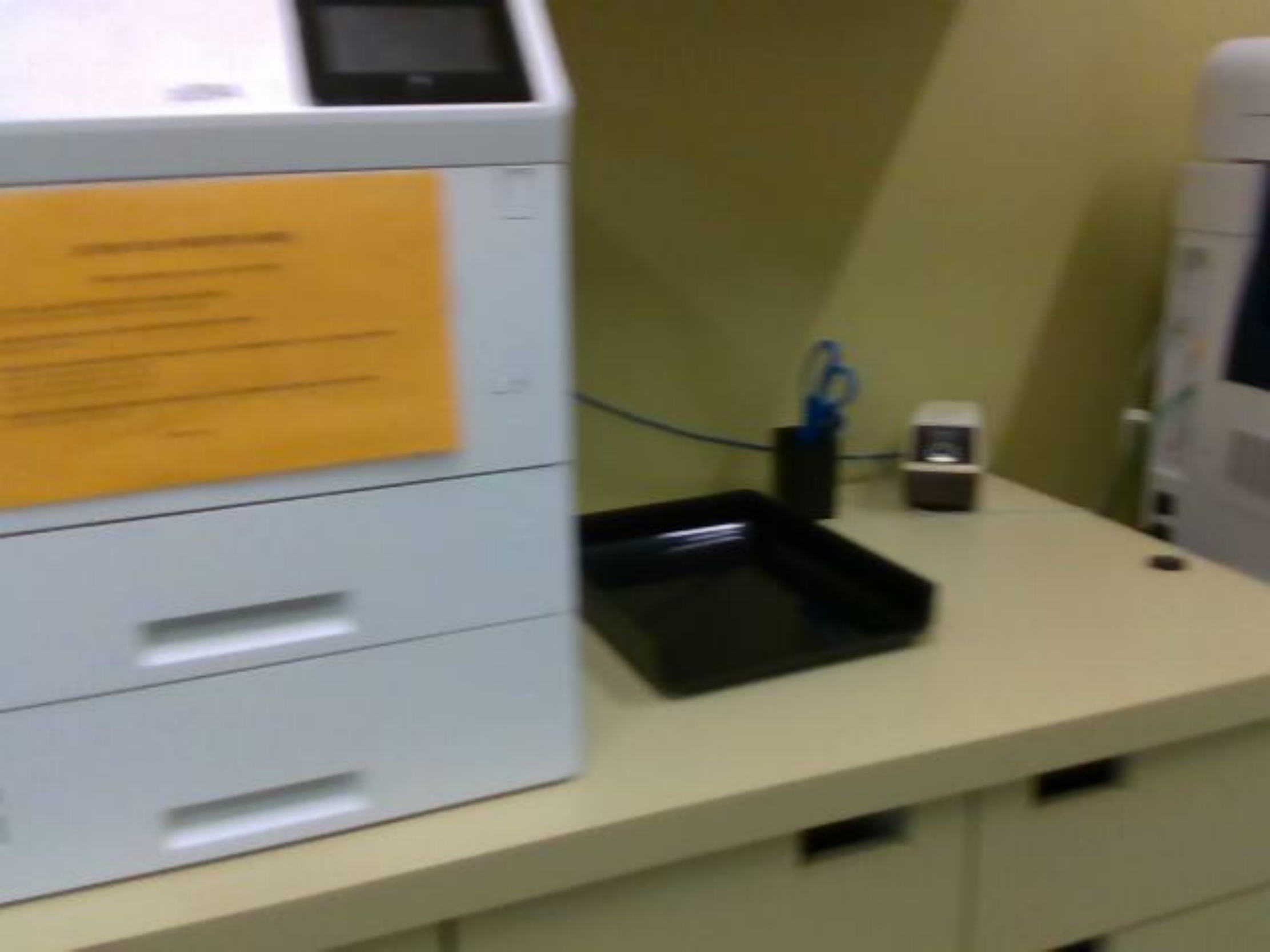}&
    \includegraphics[width=0.096\linewidth]{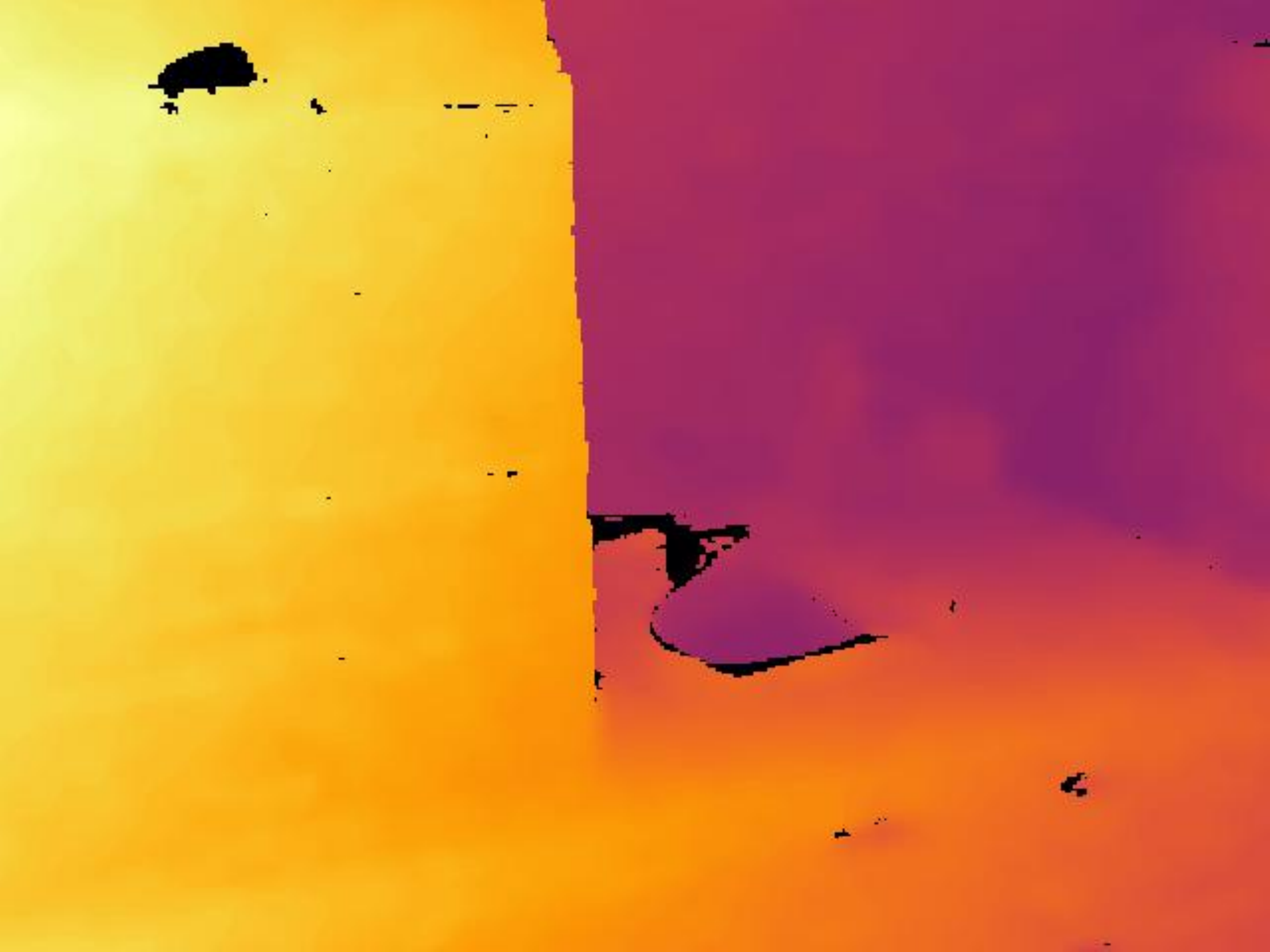}&
    \includegraphics[width=0.096\linewidth]{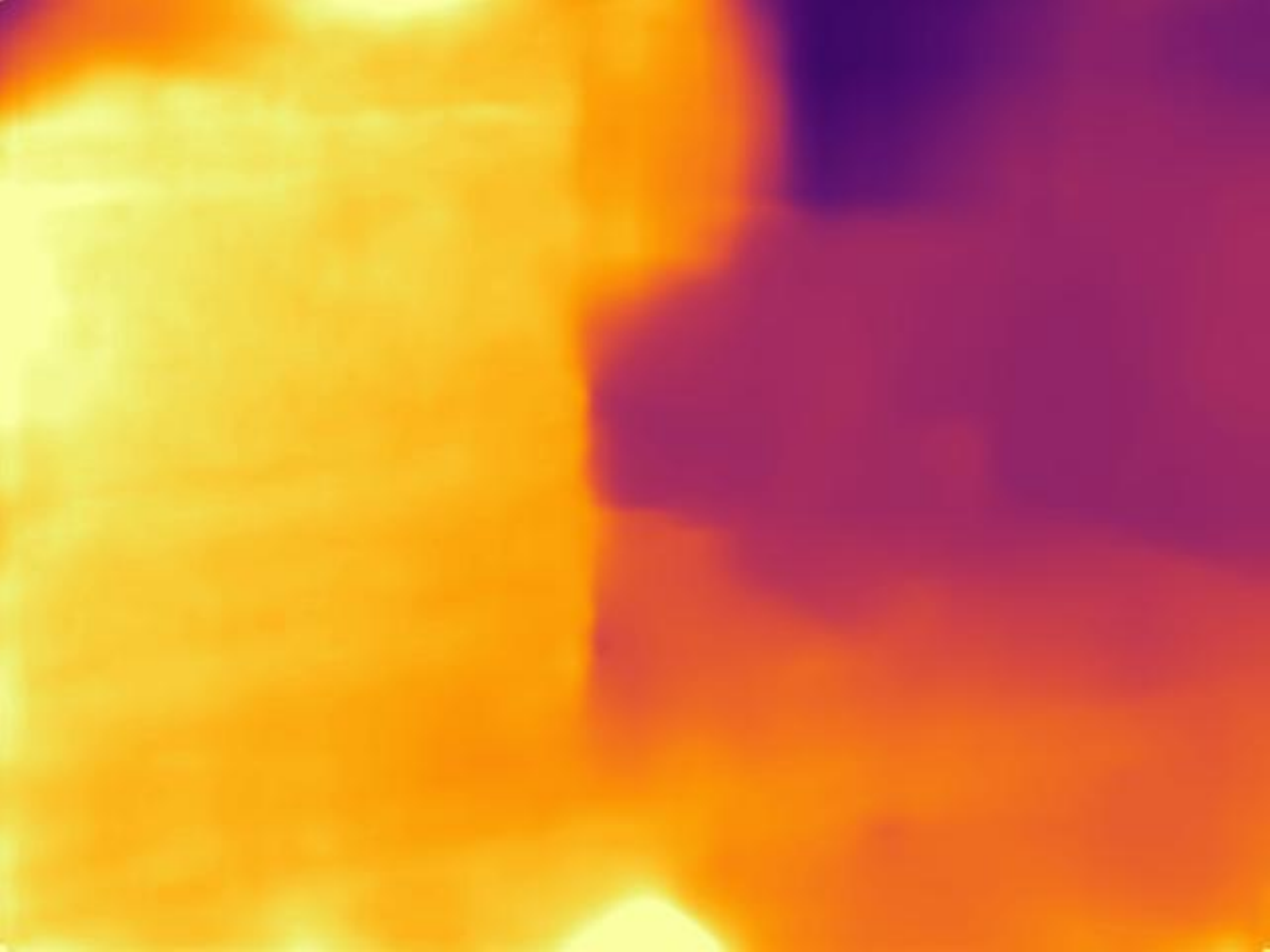}&
    \includegraphics[width=0.096\linewidth]{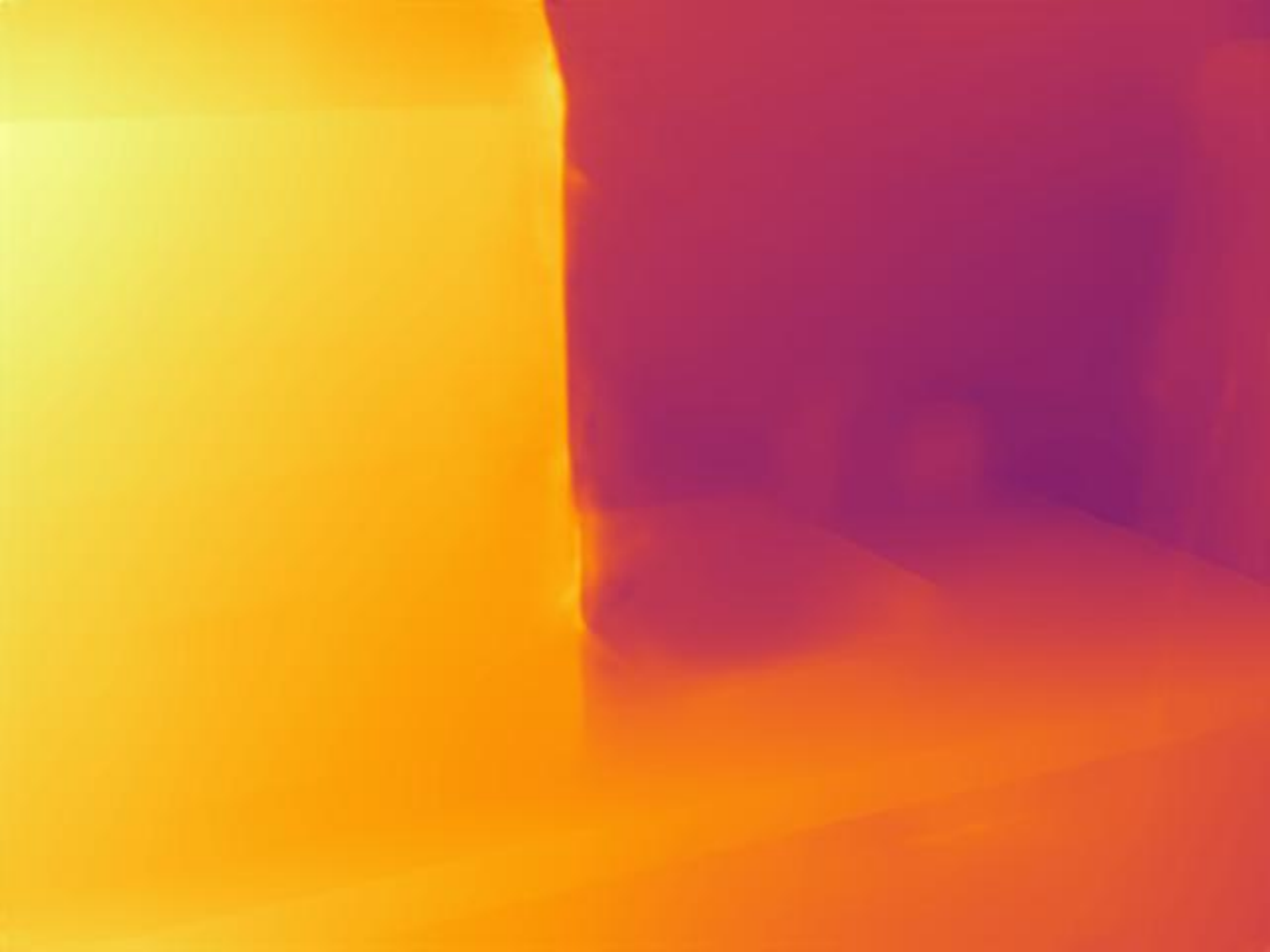}&
    \includegraphics[width=0.096\linewidth]{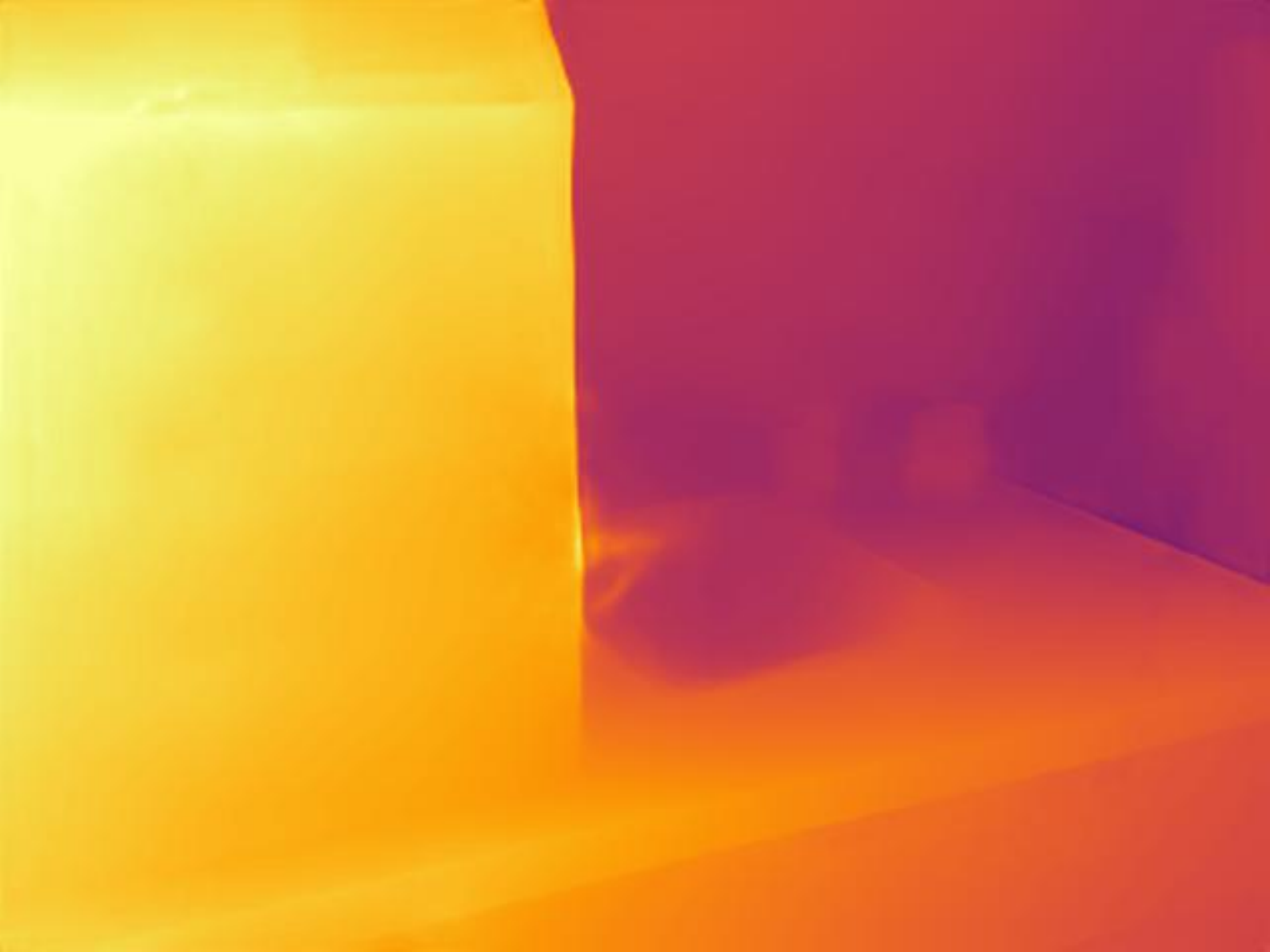}&
    \includegraphics[width=0.096\linewidth]{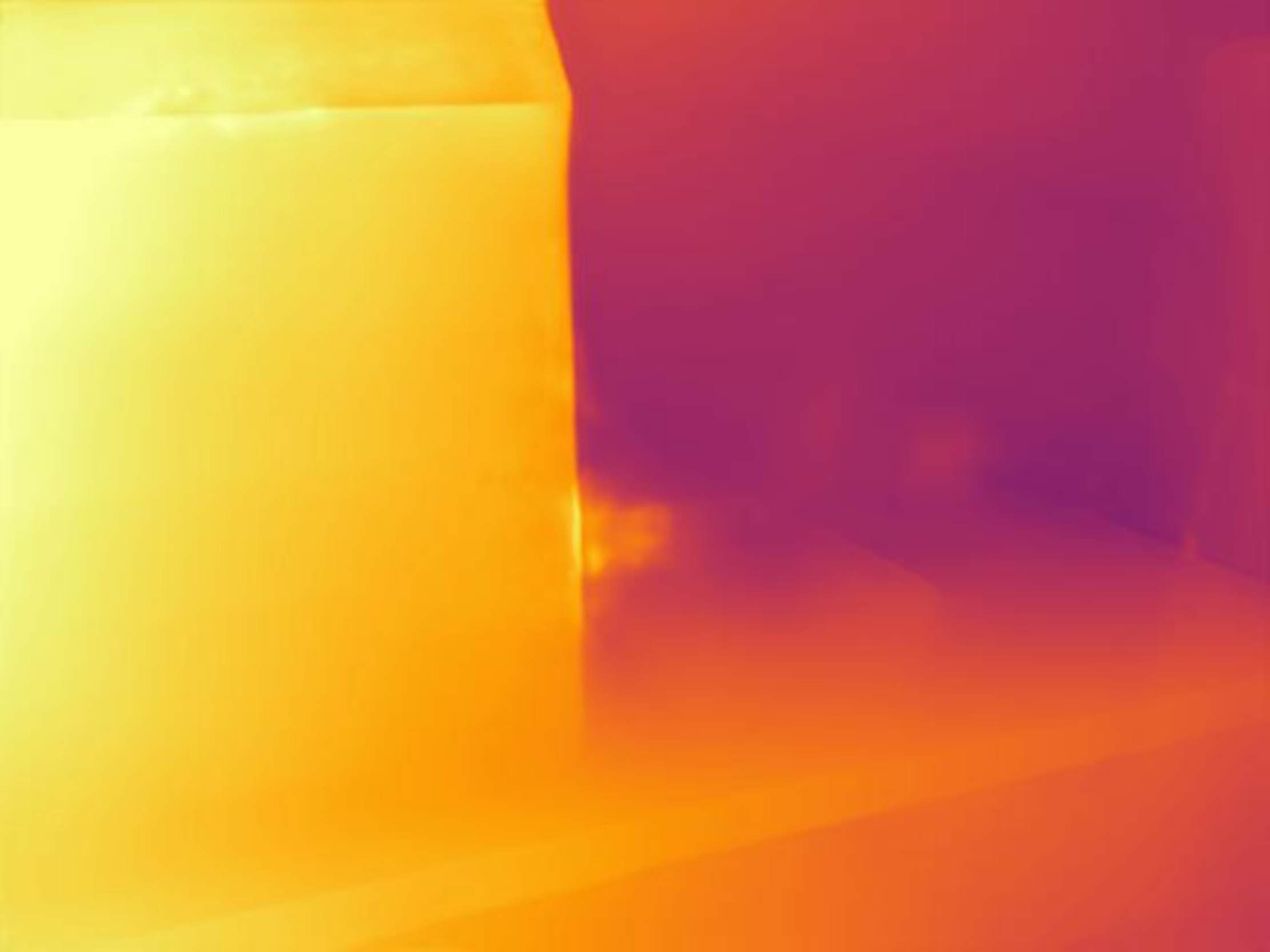}&
    \includegraphics[width=0.096\linewidth]{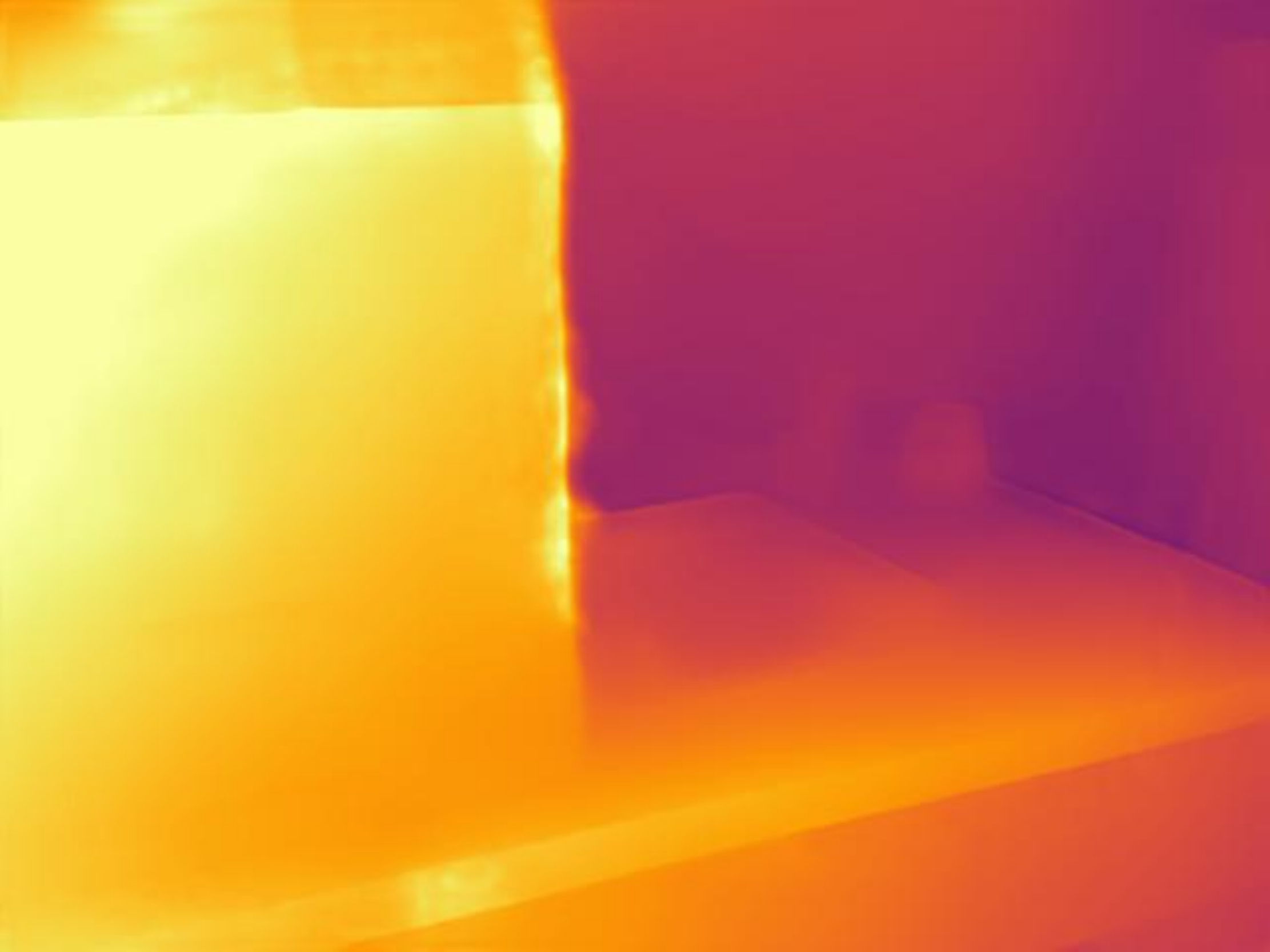}&
    \includegraphics[width=0.096\linewidth]{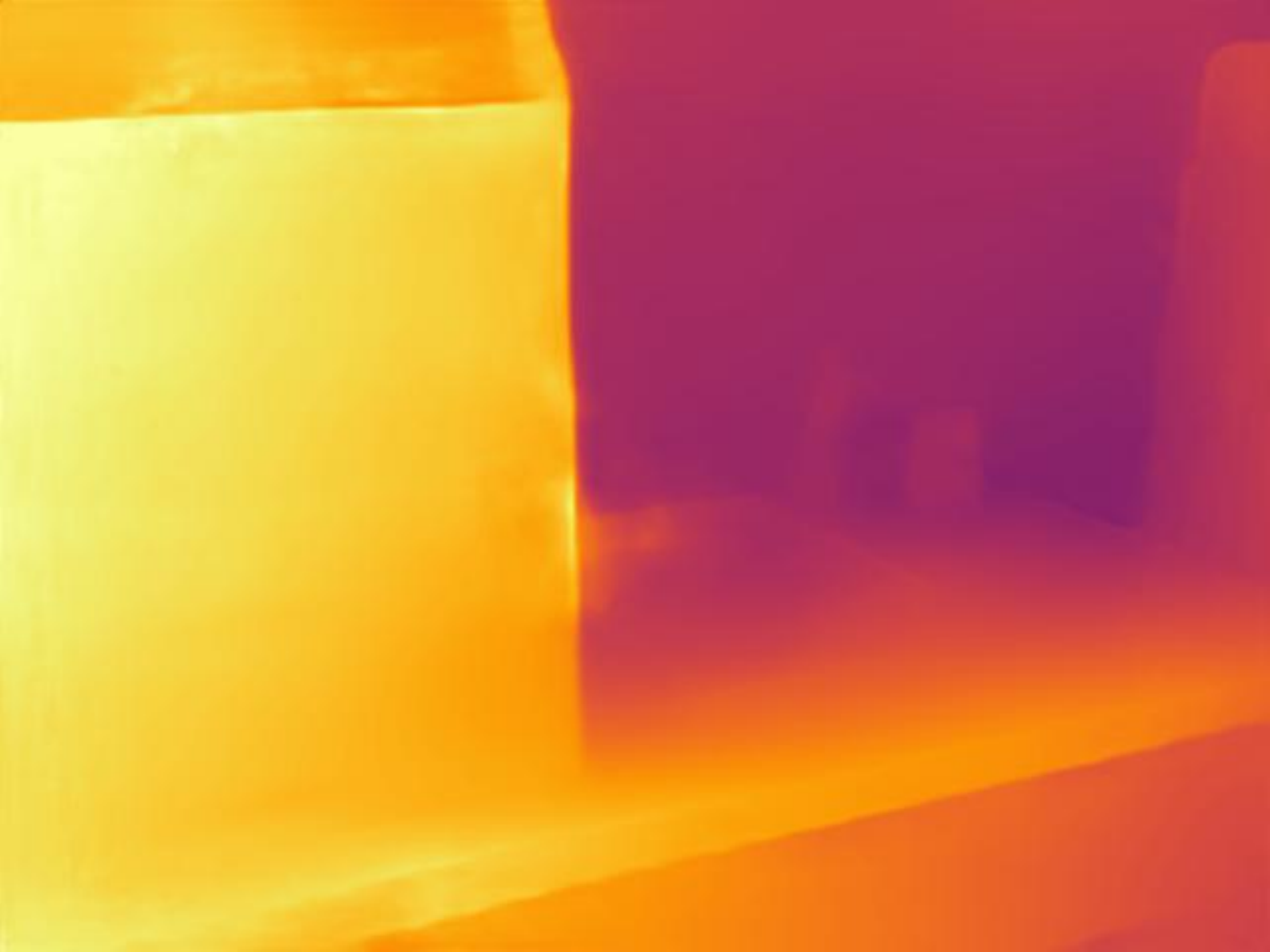}&
    \includegraphics[width=0.096\linewidth]{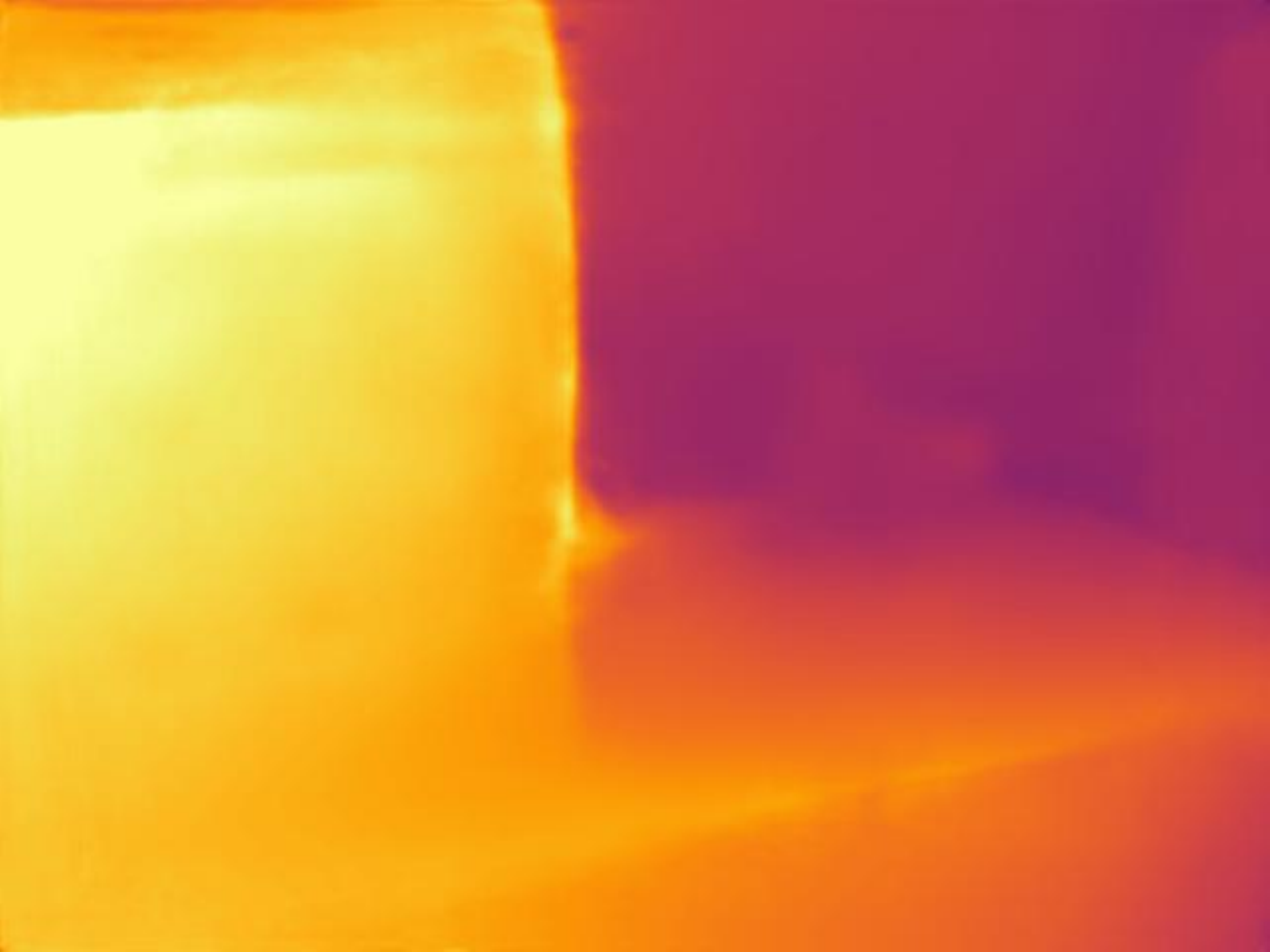}&
    \includegraphics[width=0.096\linewidth]{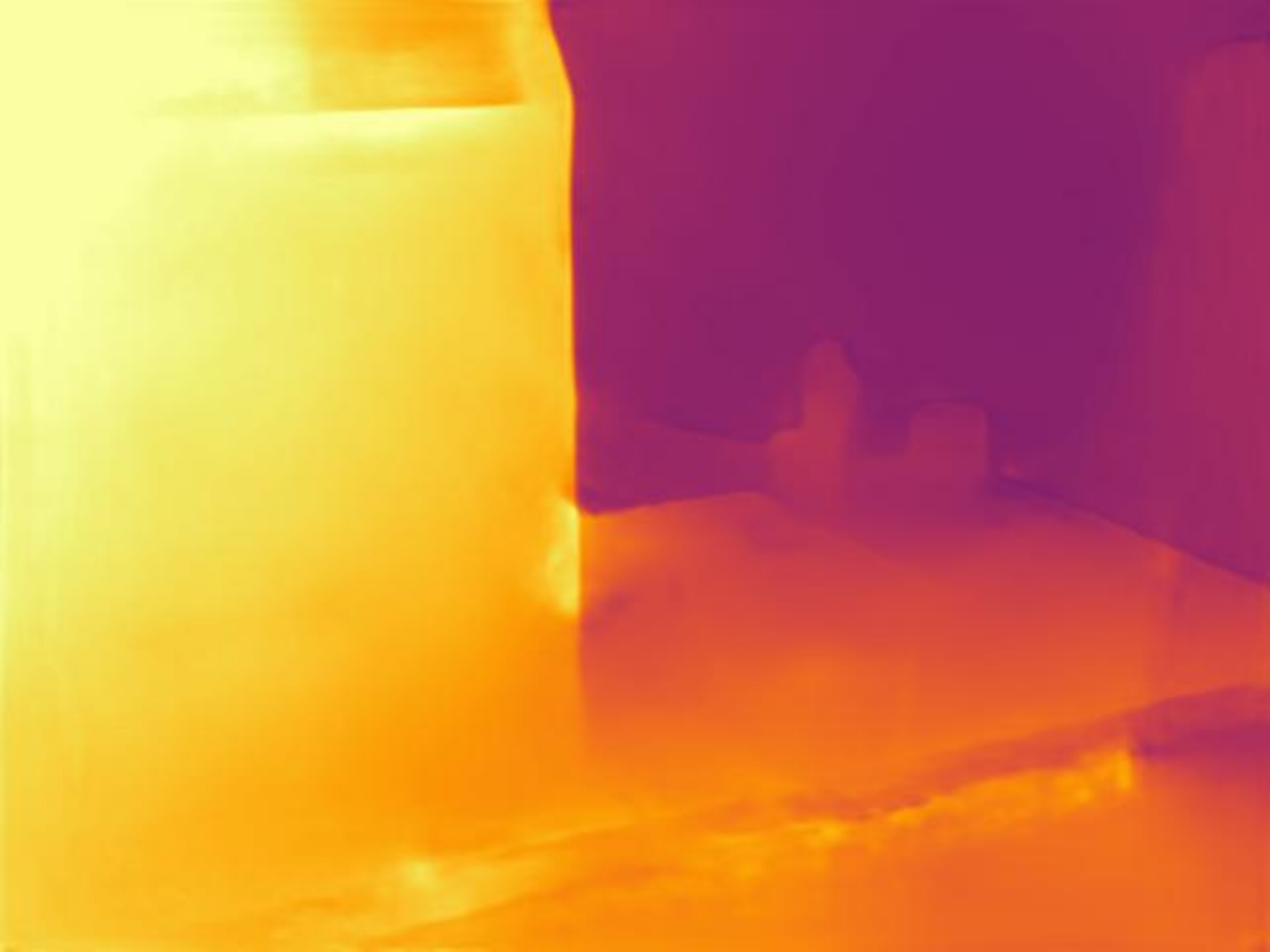}\\
    \vspace{-0.75mm}
    \scriptsize d. &
    \includegraphics[width=0.096\linewidth]{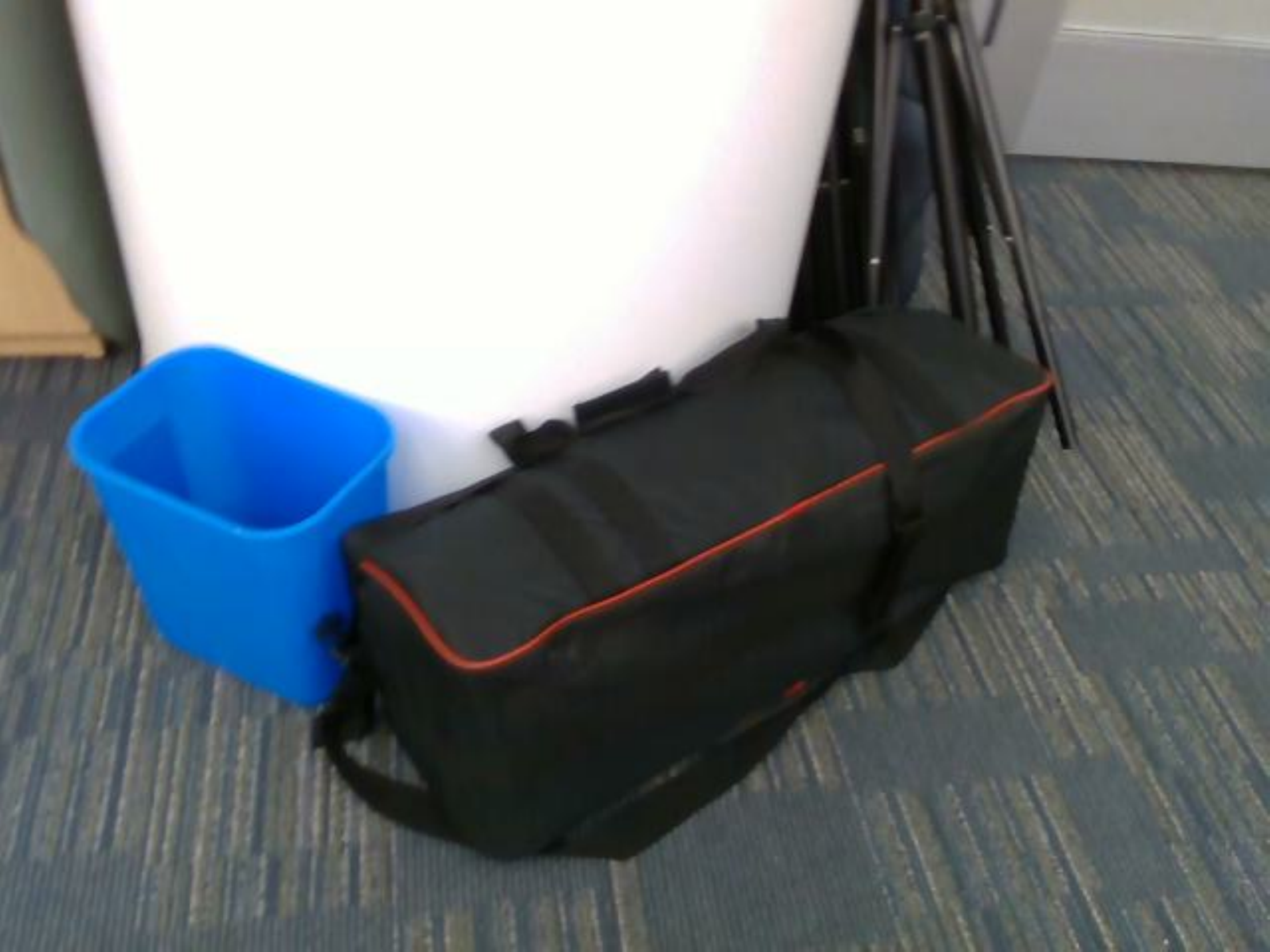}&
    \includegraphics[width=0.096\linewidth]{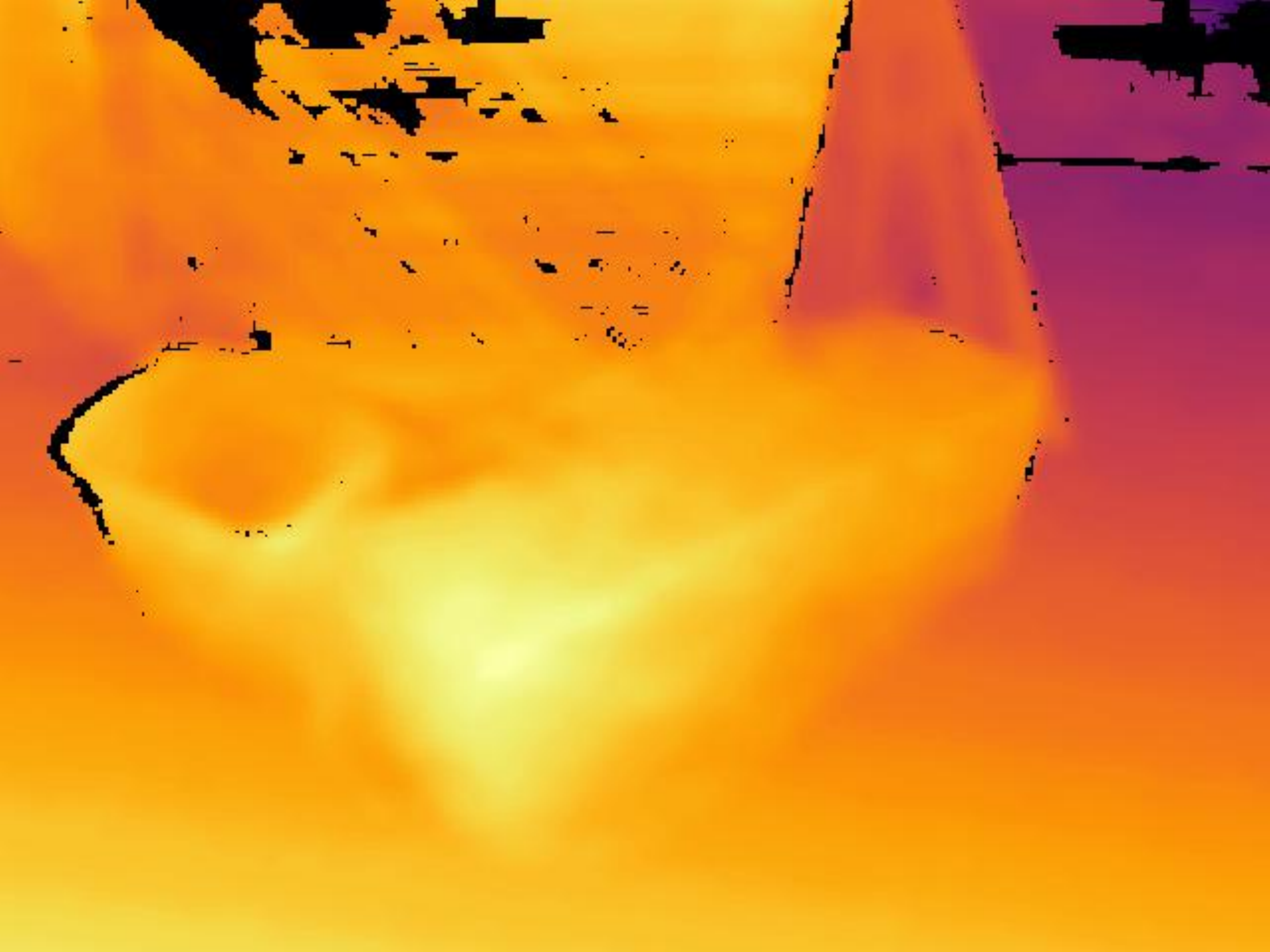}&
    \includegraphics[width=0.096\linewidth]{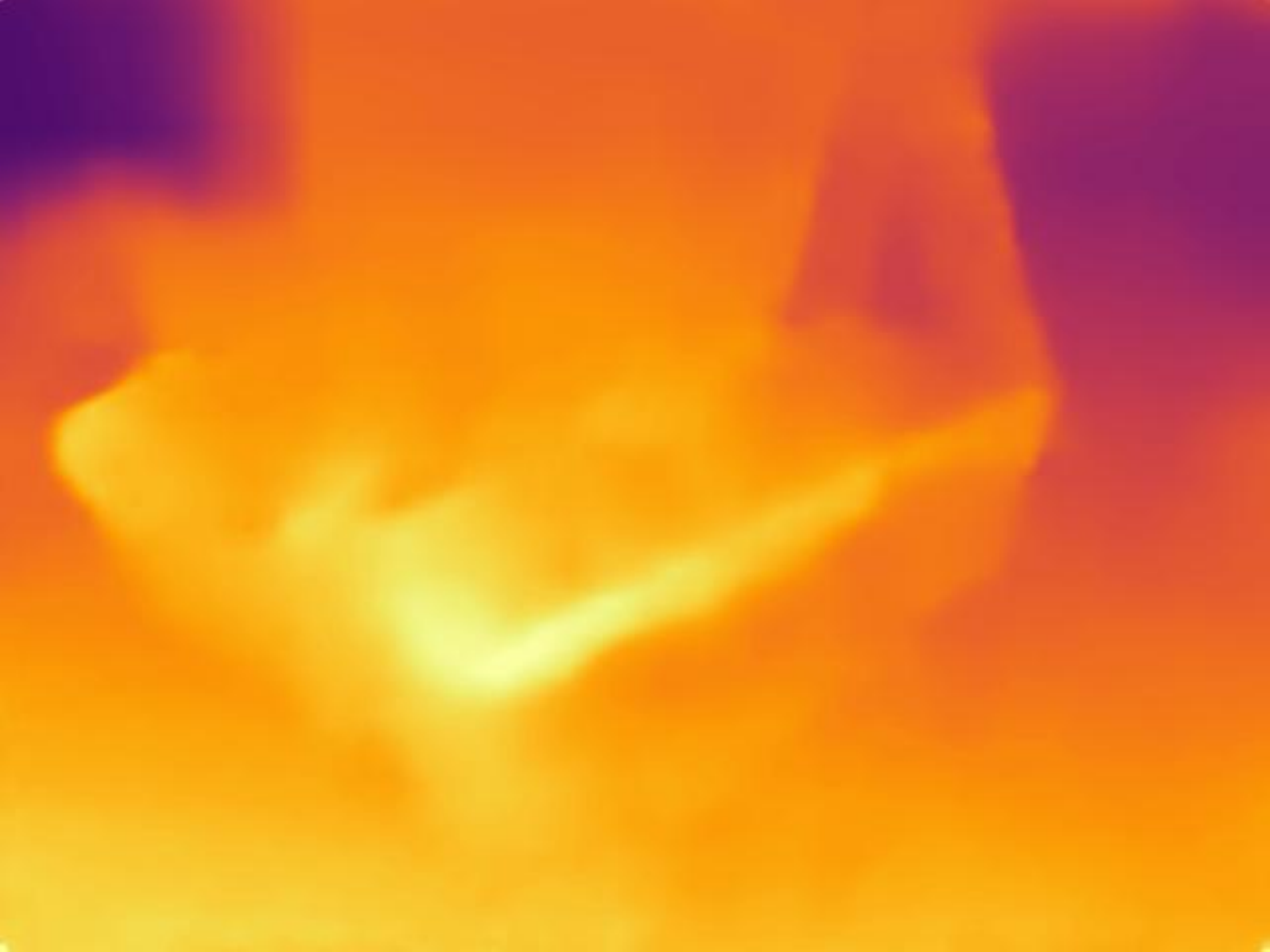}&
    \includegraphics[width=0.096\linewidth]{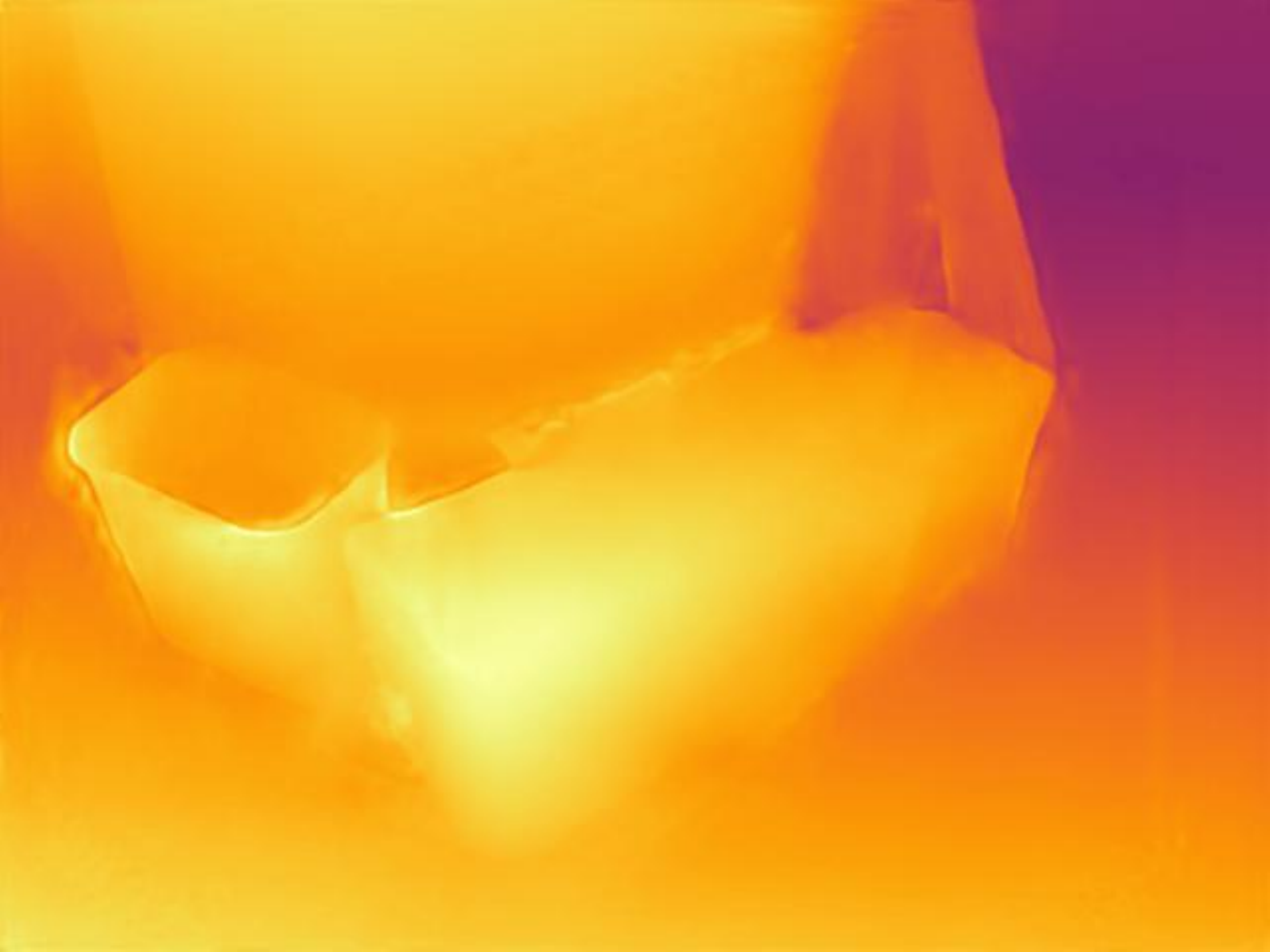}&
    \includegraphics[width=0.096\linewidth]{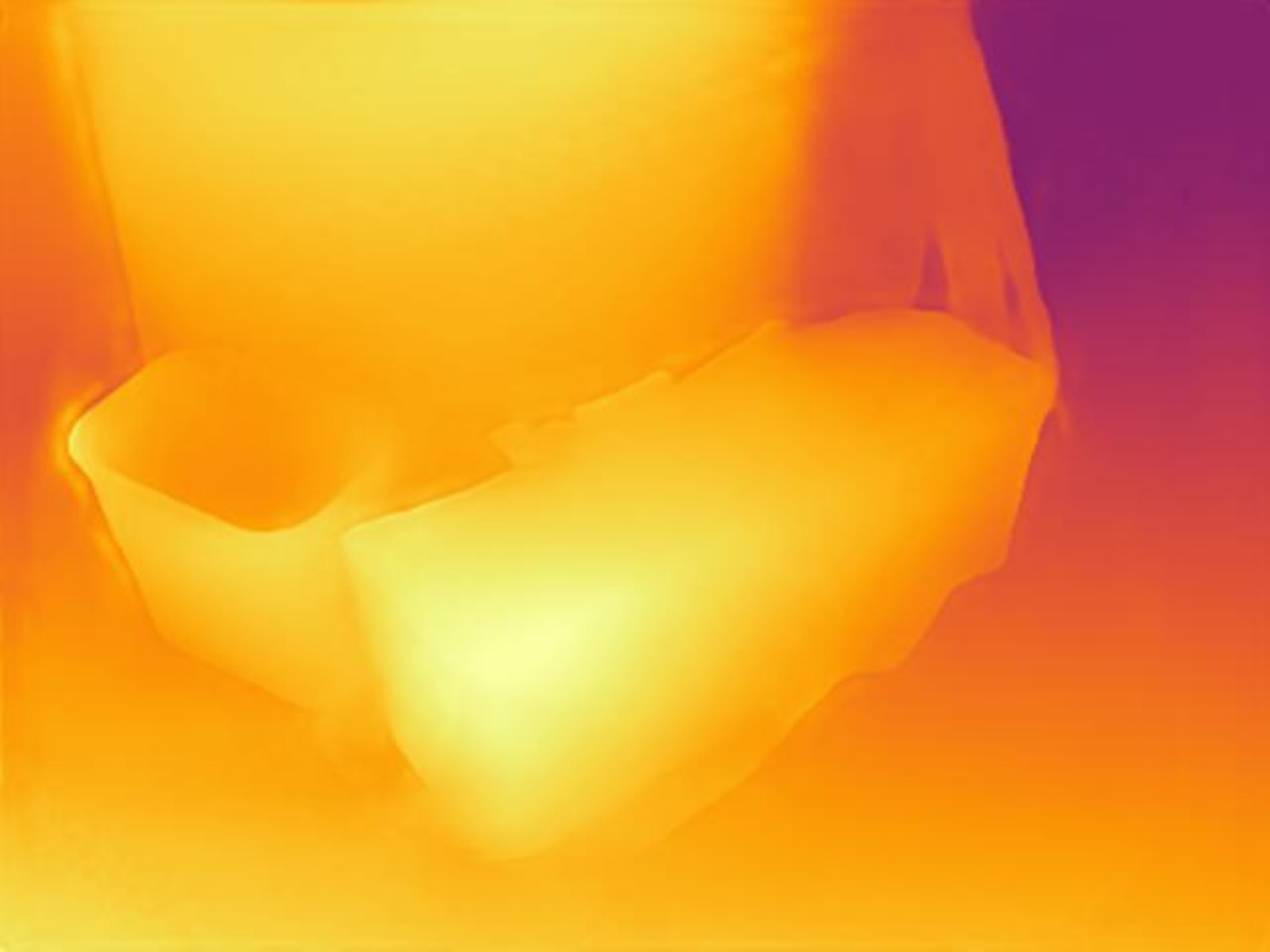}&
    \includegraphics[width=0.096\linewidth]{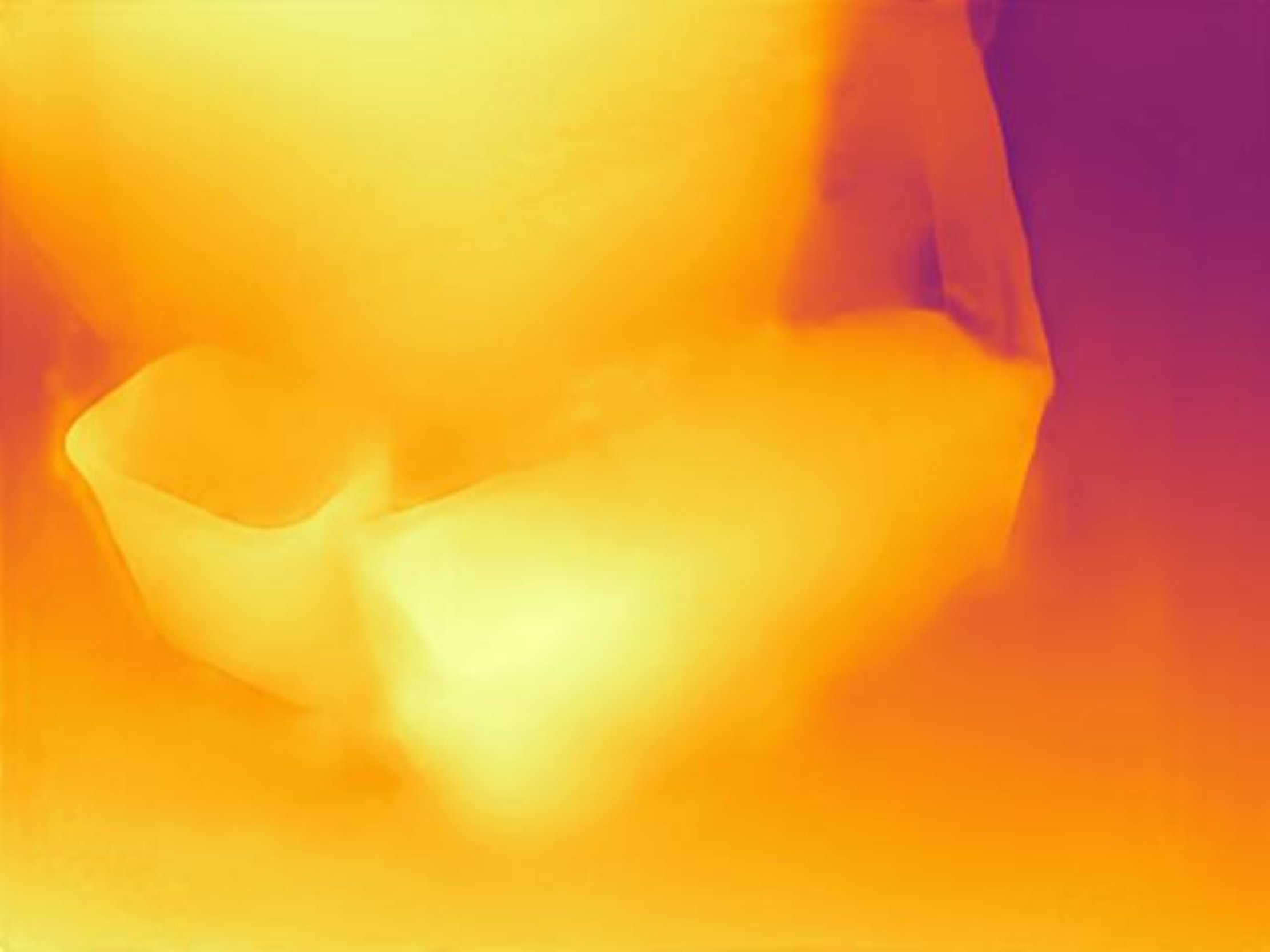}&
    \includegraphics[width=0.096\linewidth]{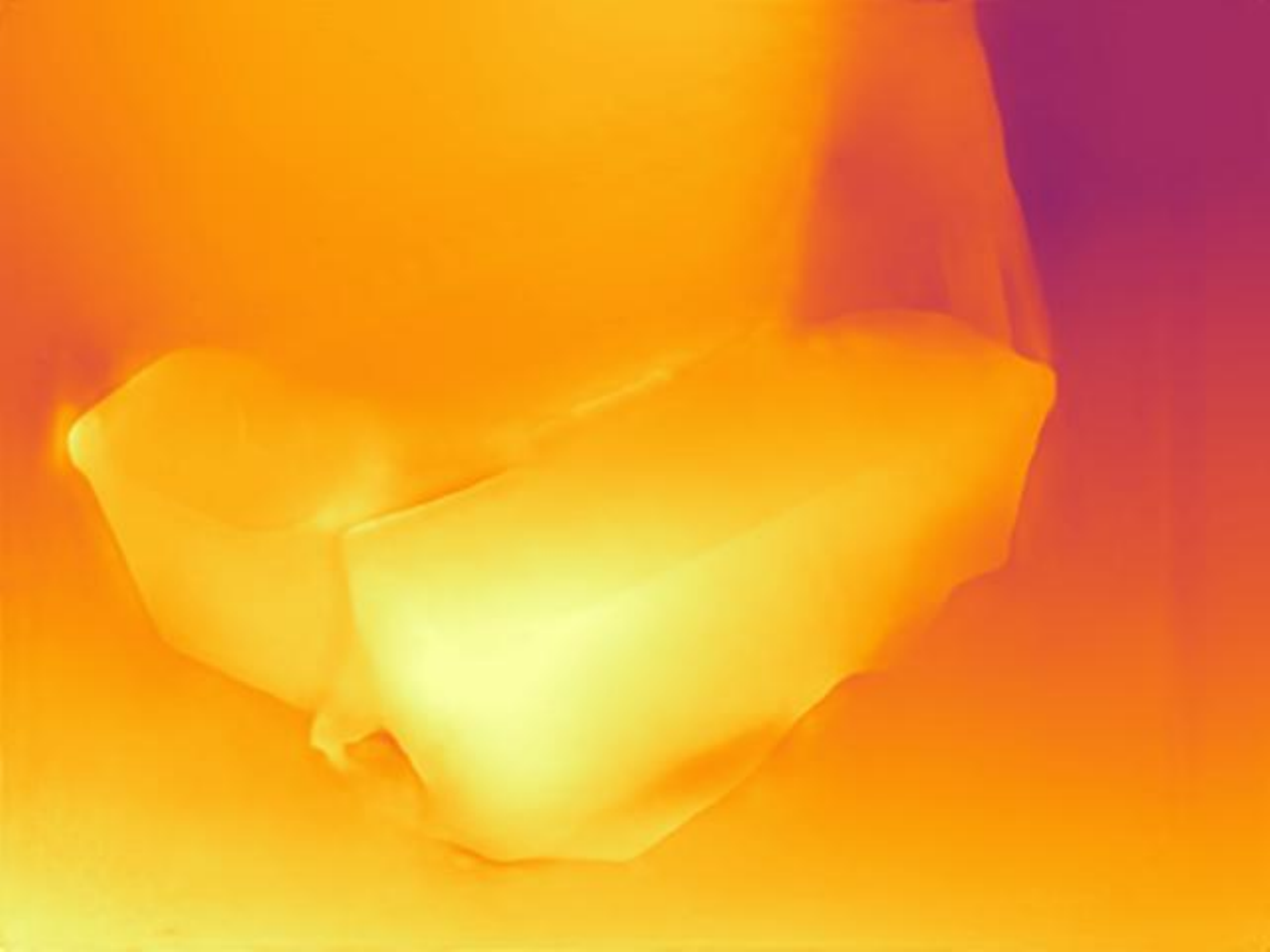}&
    \includegraphics[width=0.096\linewidth]{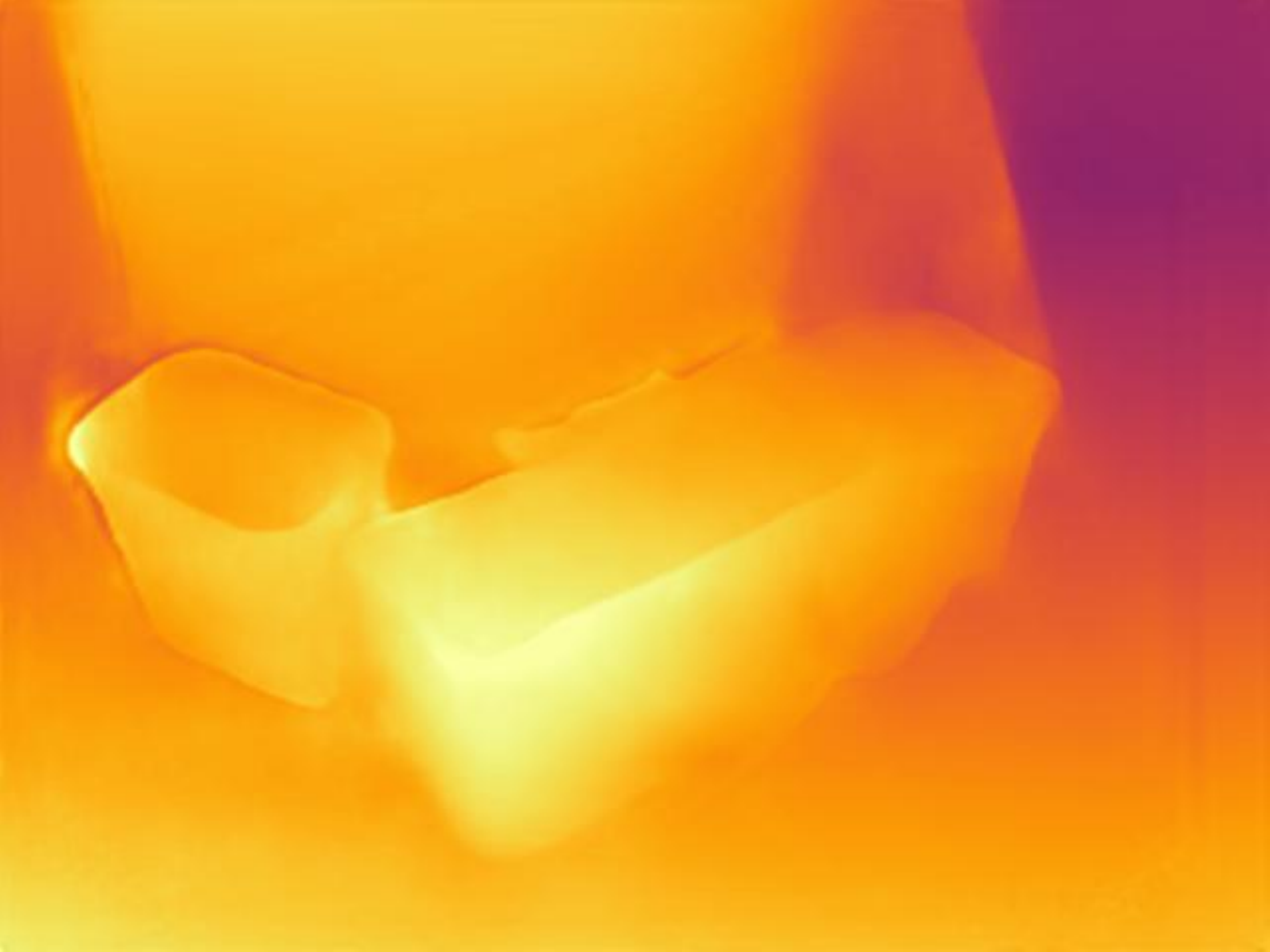}&
    \includegraphics[width=0.096\linewidth]{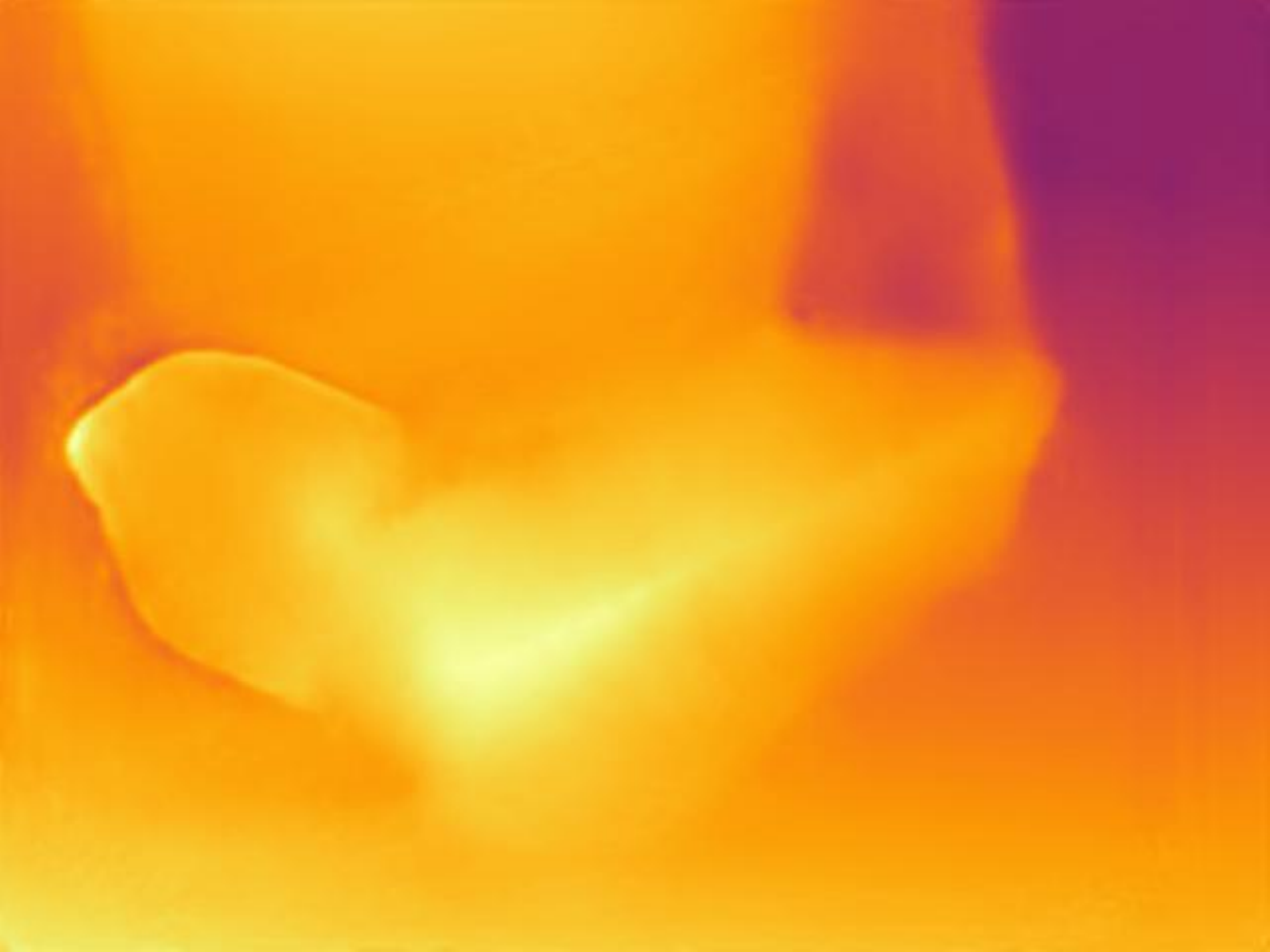}&
    \includegraphics[width=0.096\linewidth]{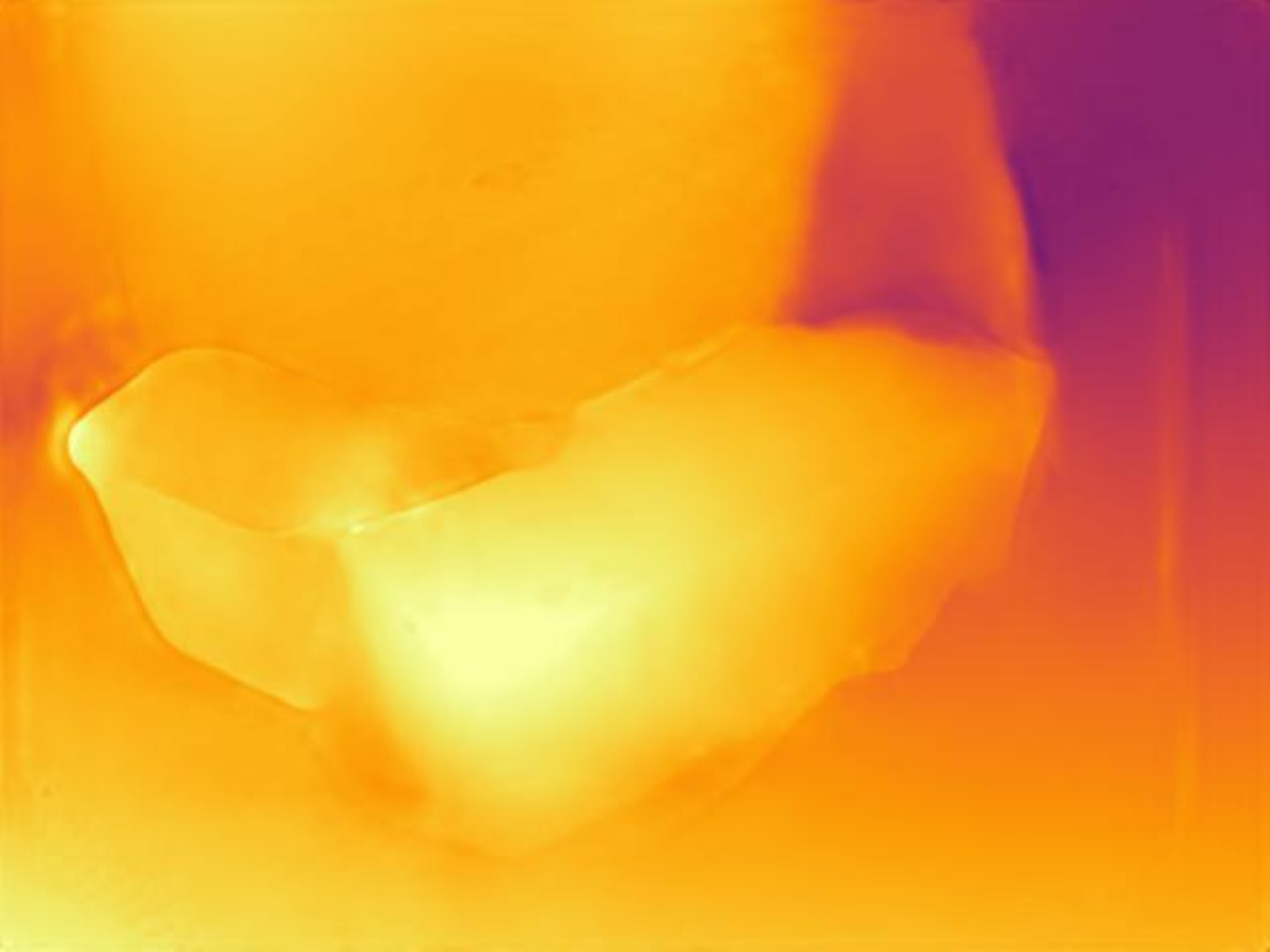}\\
    \vspace{-0.75mm}
    \scriptsize e. &
    \includegraphics[width=0.096\linewidth]{suppl/comparison_void_150/kbnet/image/0000000280.pdf}&
    \includegraphics[width=0.096\linewidth]{suppl/comparison_void_150/kbnet/ground_truth/0000000280.pdf}&
    \includegraphics[width=0.096\linewidth]{suppl/comparison_void_150/kbnet/output_depth/0000000280.pdf}&
    \includegraphics[width=0.096\linewidth]{suppl/comparison_void_150/dpt-beit-l/row_4_col_5.pdf}&
    \includegraphics[width=0.096\linewidth]{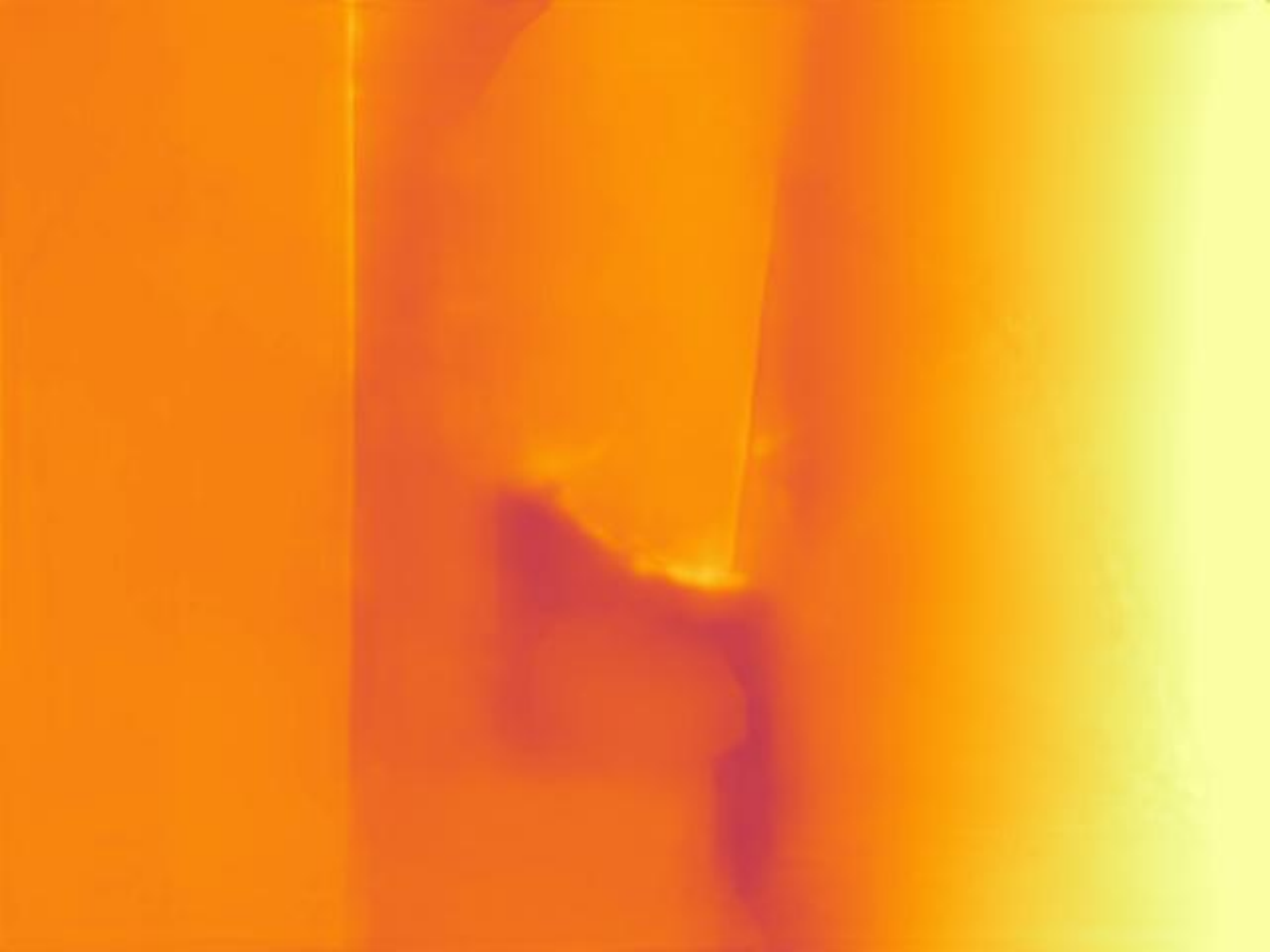}&
    \includegraphics[width=0.096\linewidth]{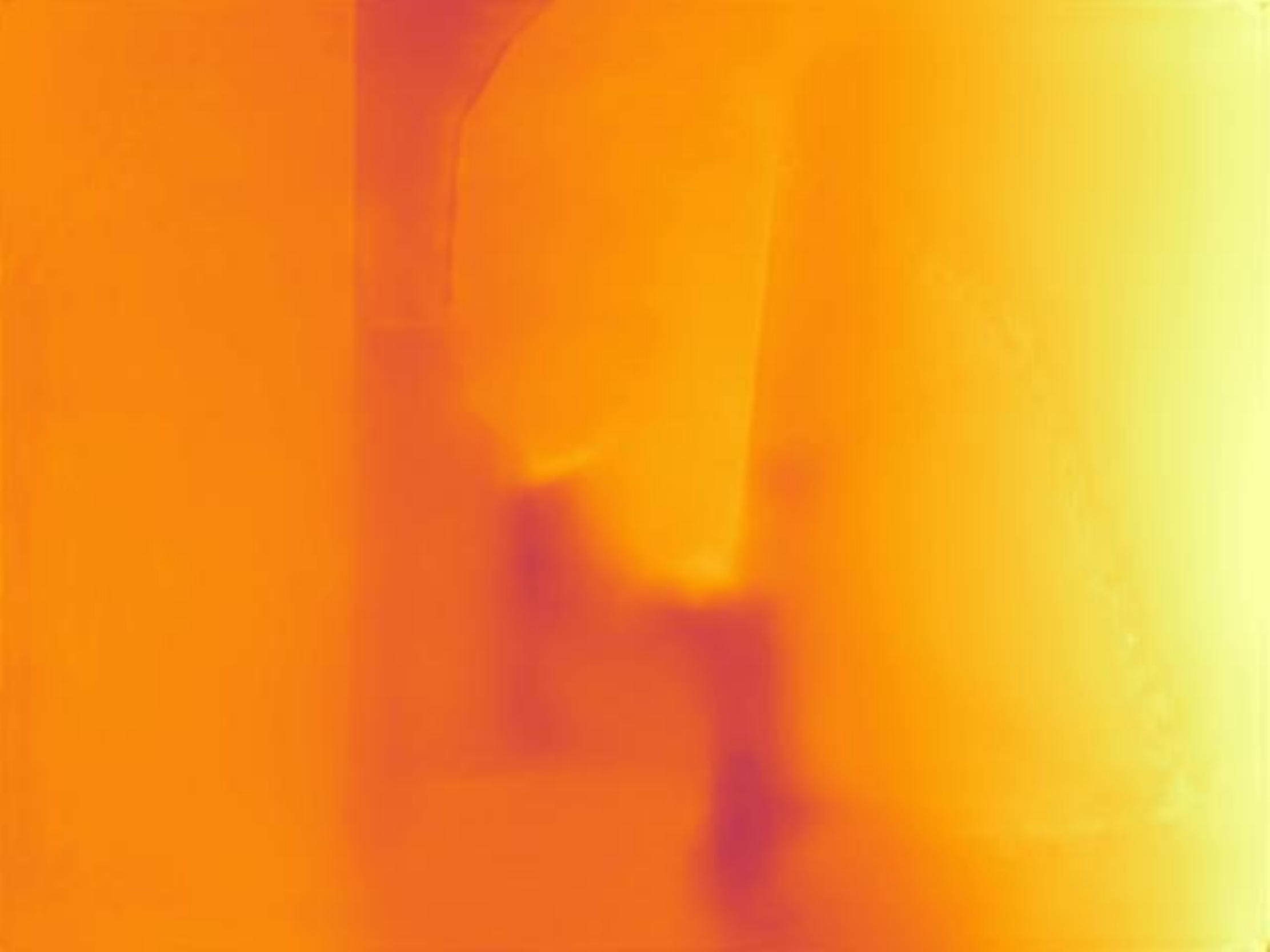}&
    \includegraphics[width=0.096\linewidth]{suppl/comparison_void_150/dpt-hybrid/row_4_col_5.pdf}&
    \includegraphics[width=0.096\linewidth]{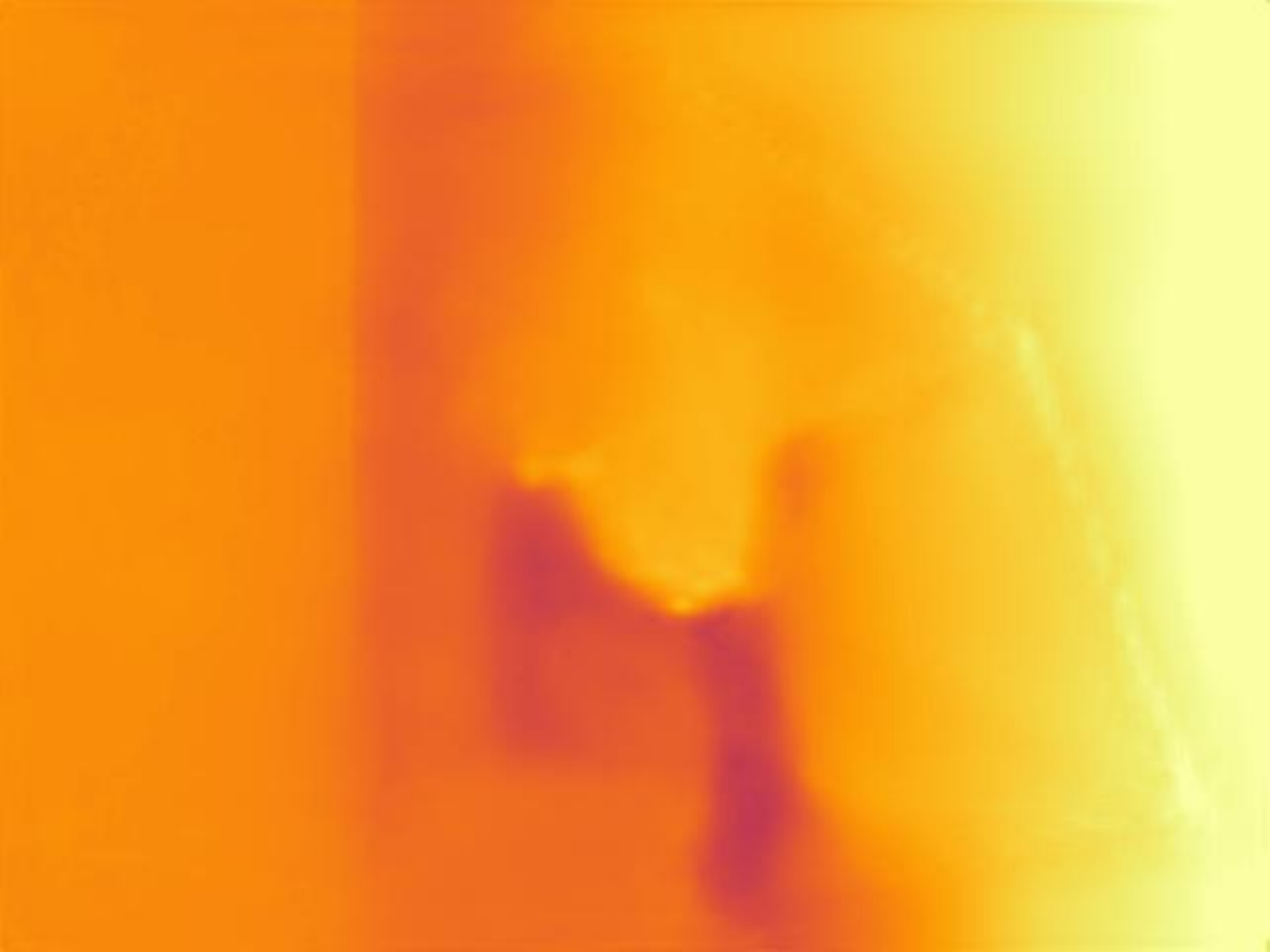}&
    \includegraphics[width=0.096\linewidth]{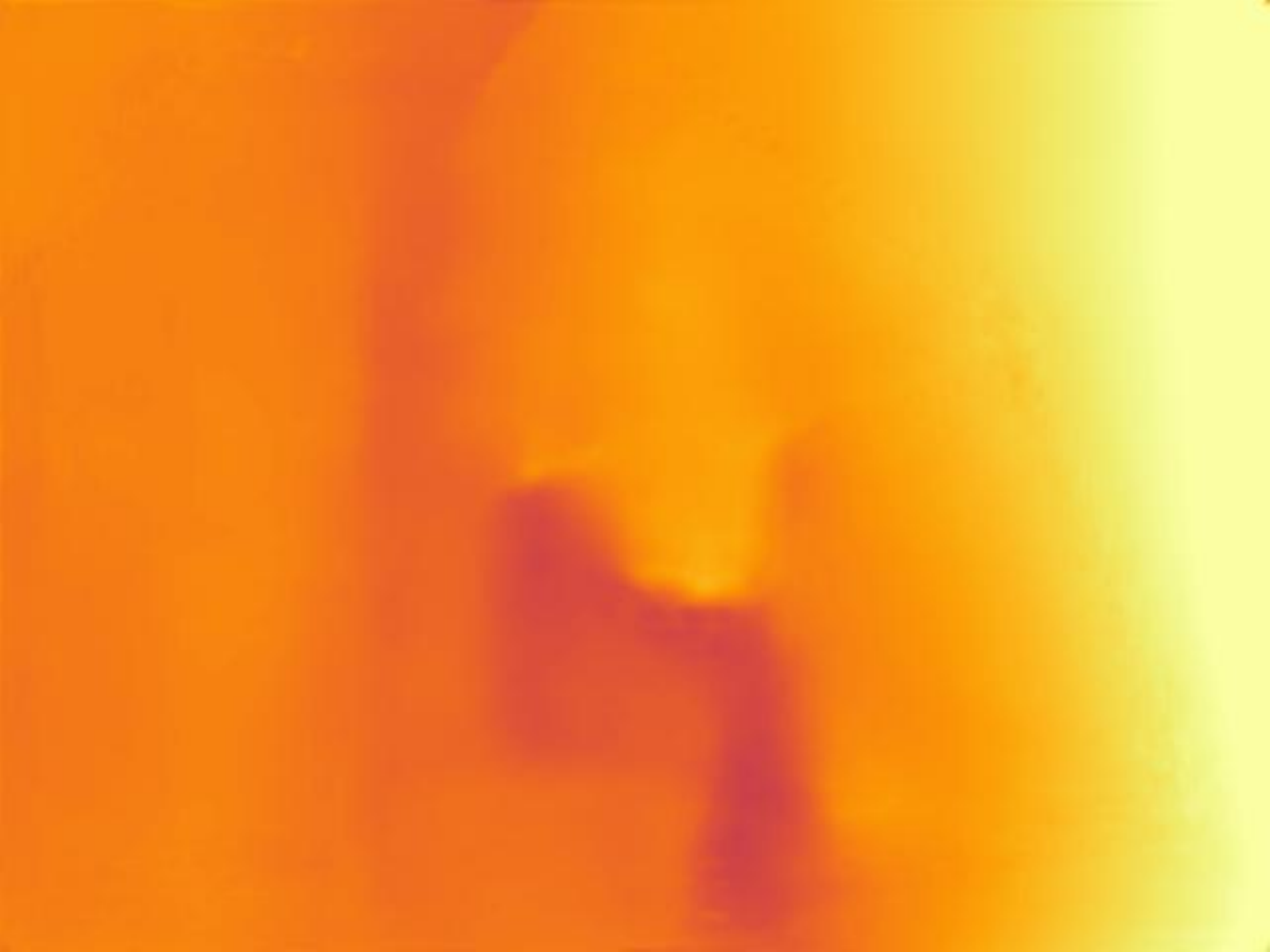}&
    \includegraphics[width=0.096\linewidth]{suppl/comparison_void_150/midas-small/row_4_col_5.pdf}\\
    \vspace{-0.75mm}
    \scriptsize f. &
    \includegraphics[width=0.096\linewidth]{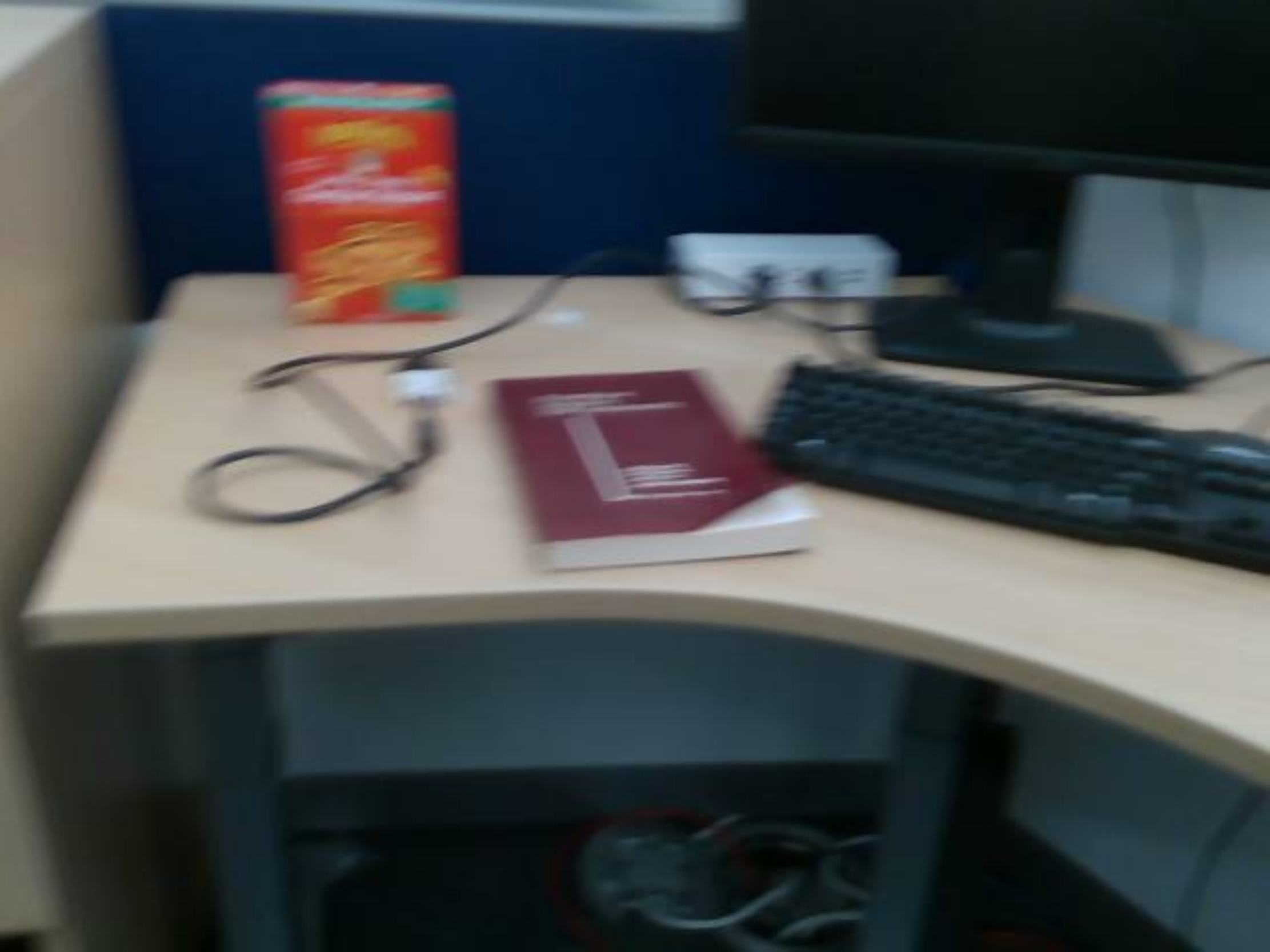}&
    \includegraphics[width=0.096\linewidth]{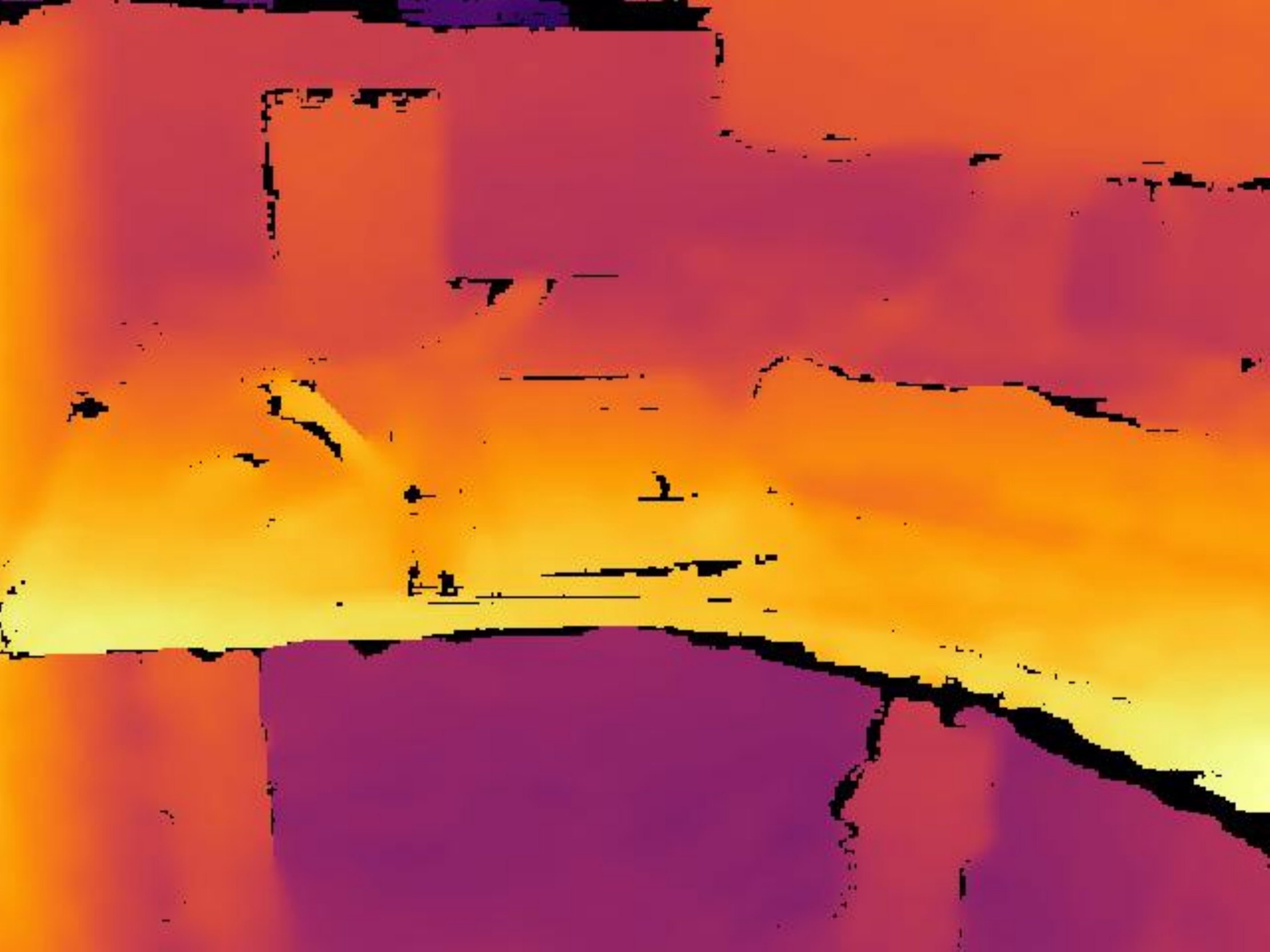}&
    \includegraphics[width=0.096\linewidth]{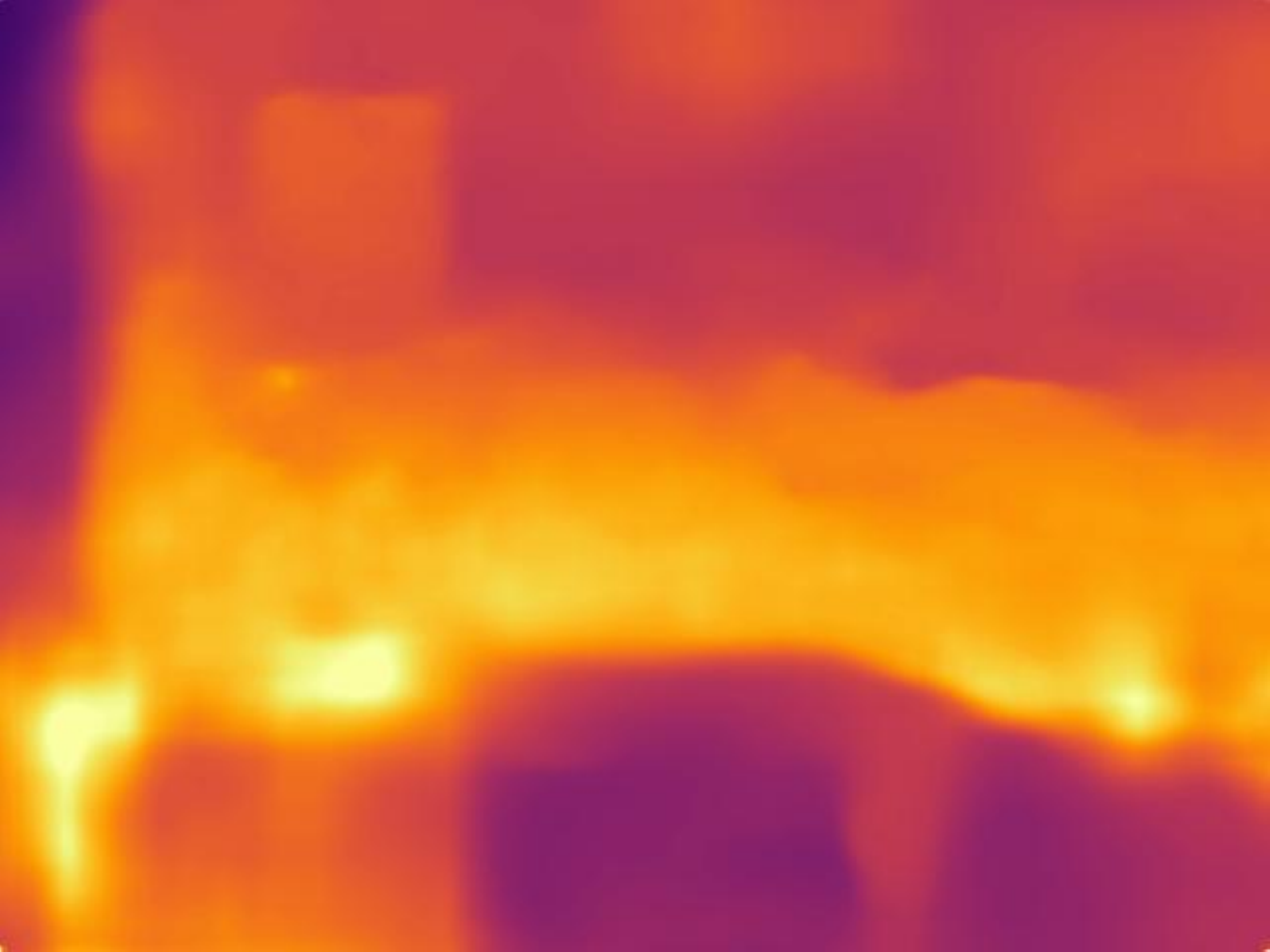}&
    \includegraphics[width=0.096\linewidth]{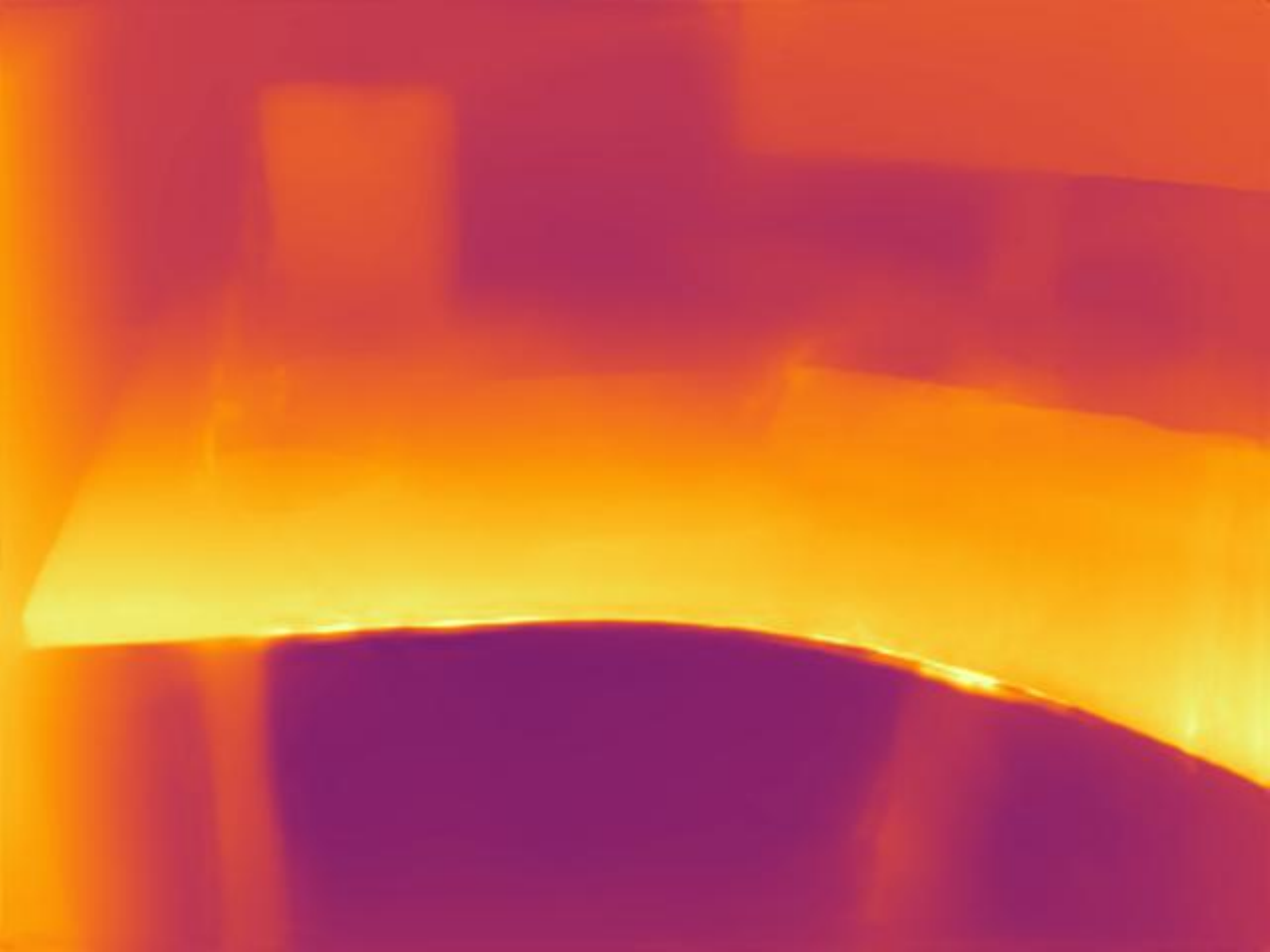}&
    \includegraphics[width=0.096\linewidth]{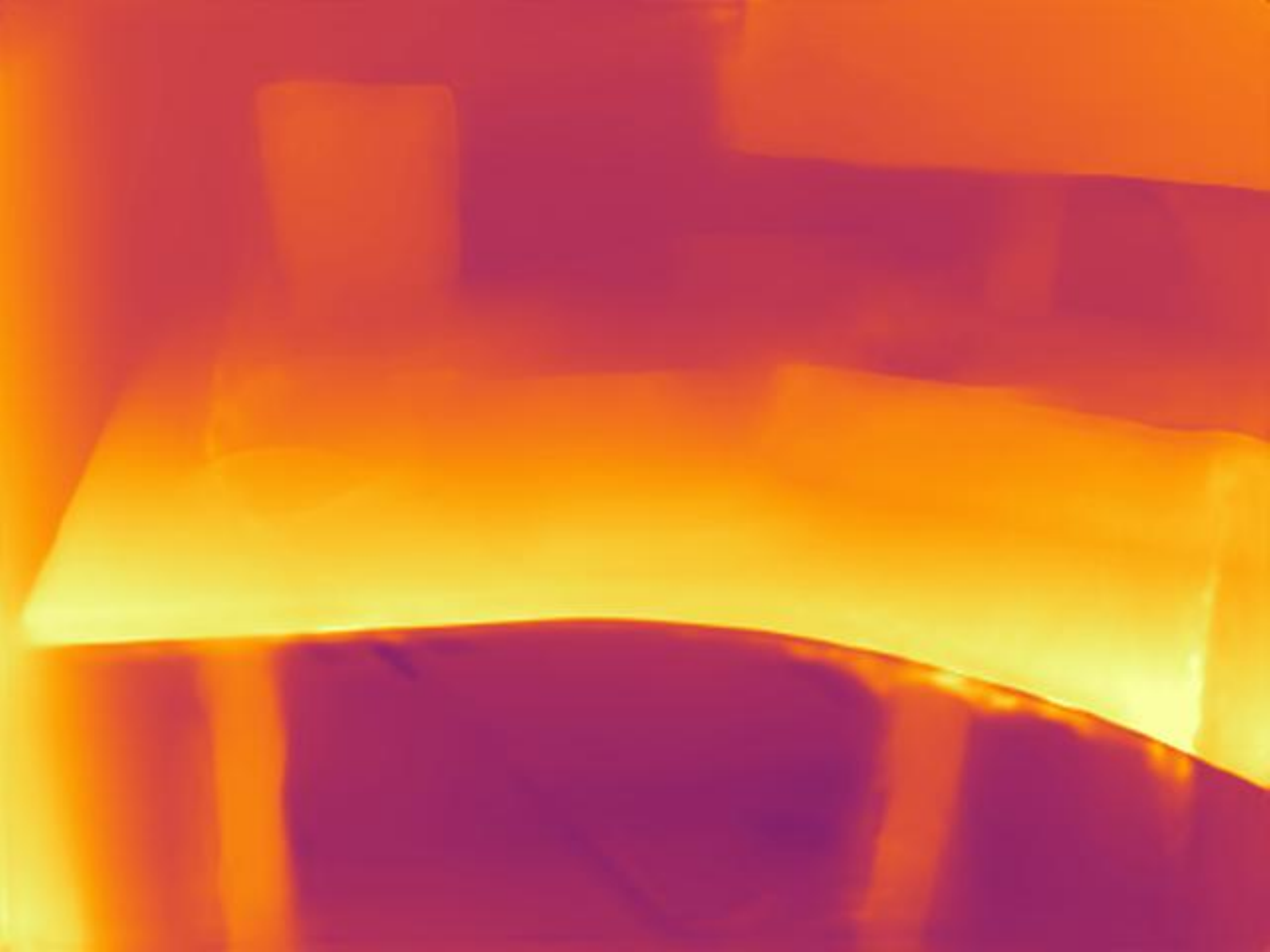}&
    \includegraphics[width=0.096\linewidth]{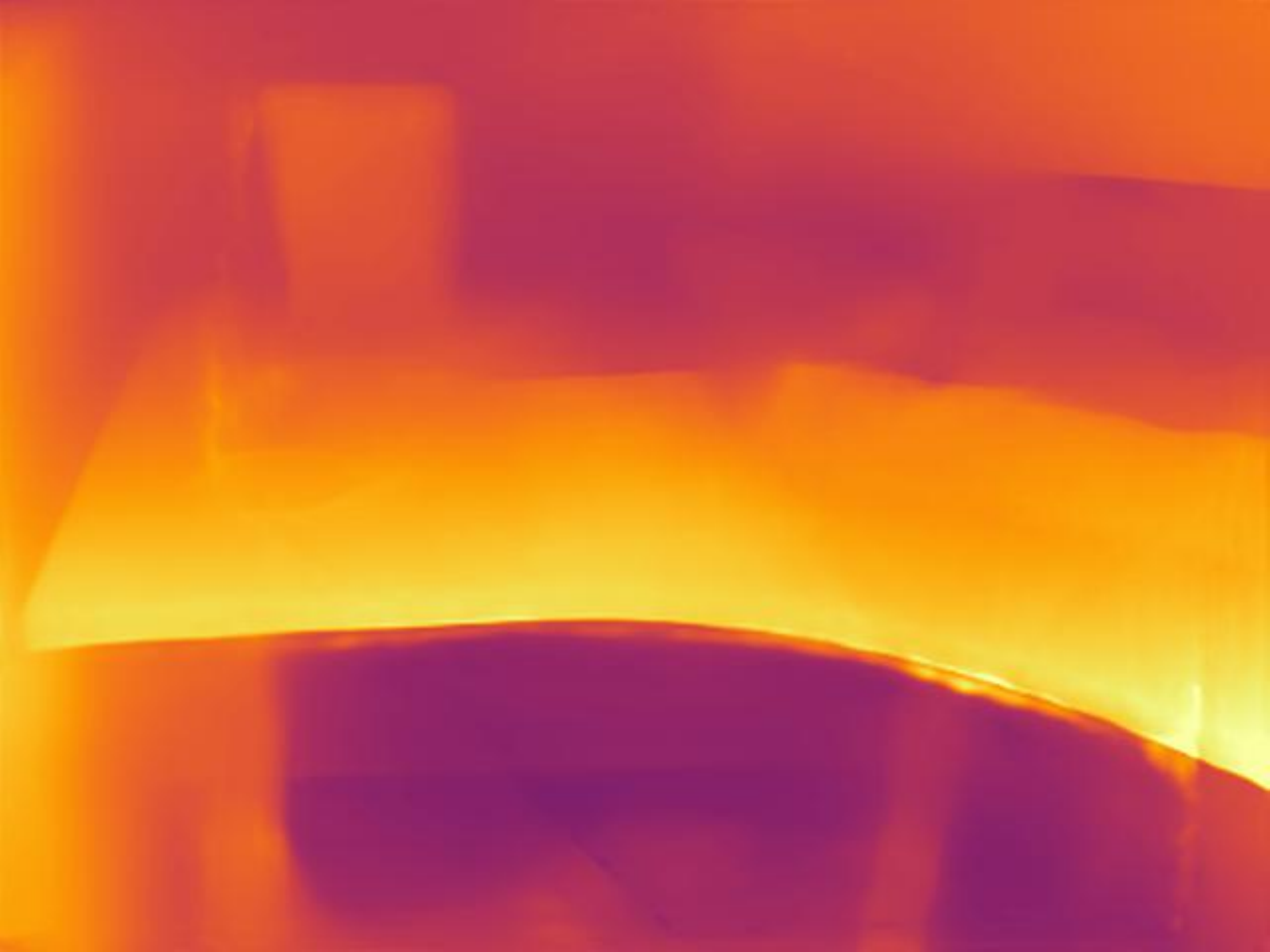}&
    \includegraphics[width=0.096\linewidth]{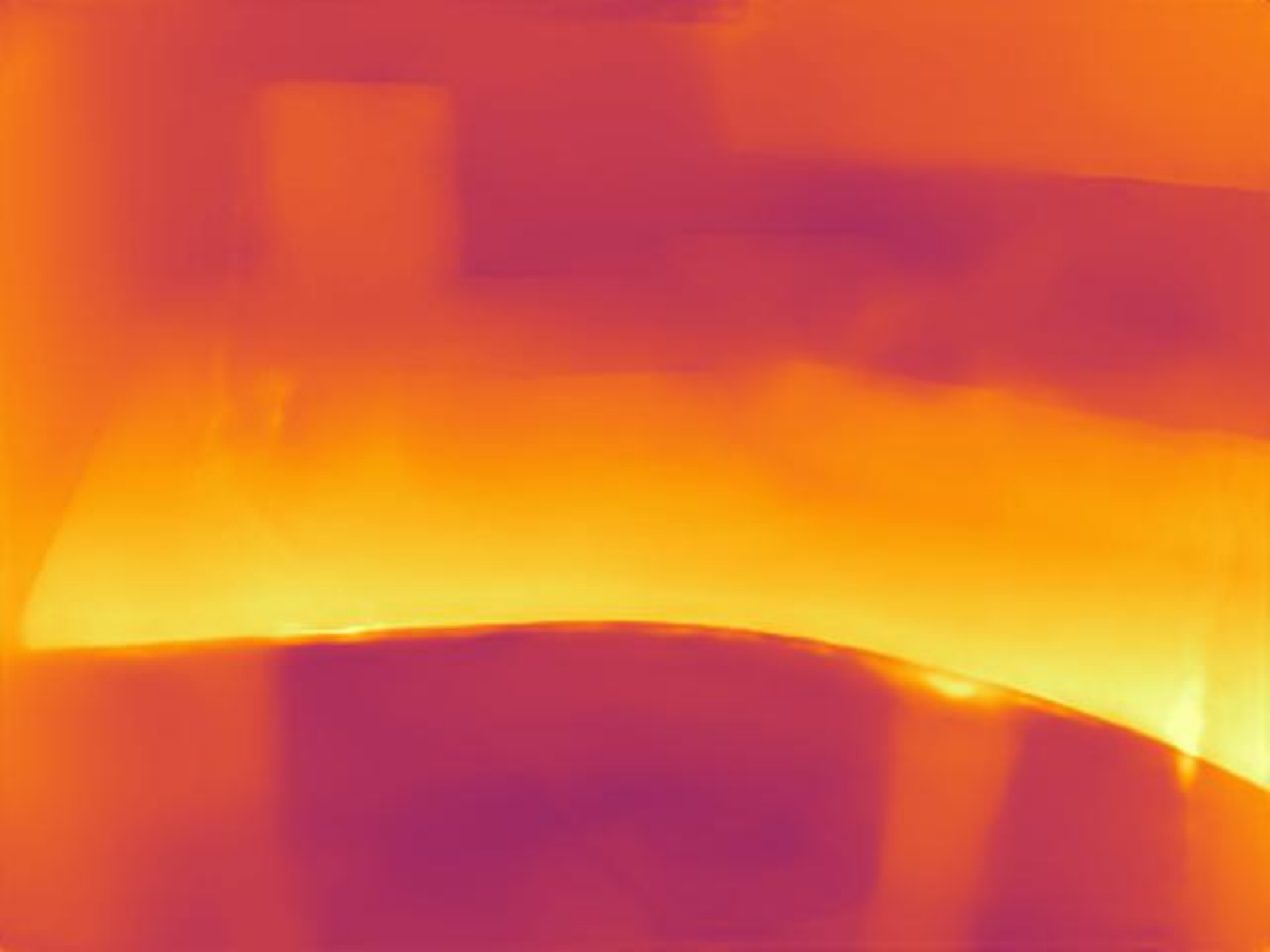}&
    \includegraphics[width=0.096\linewidth]{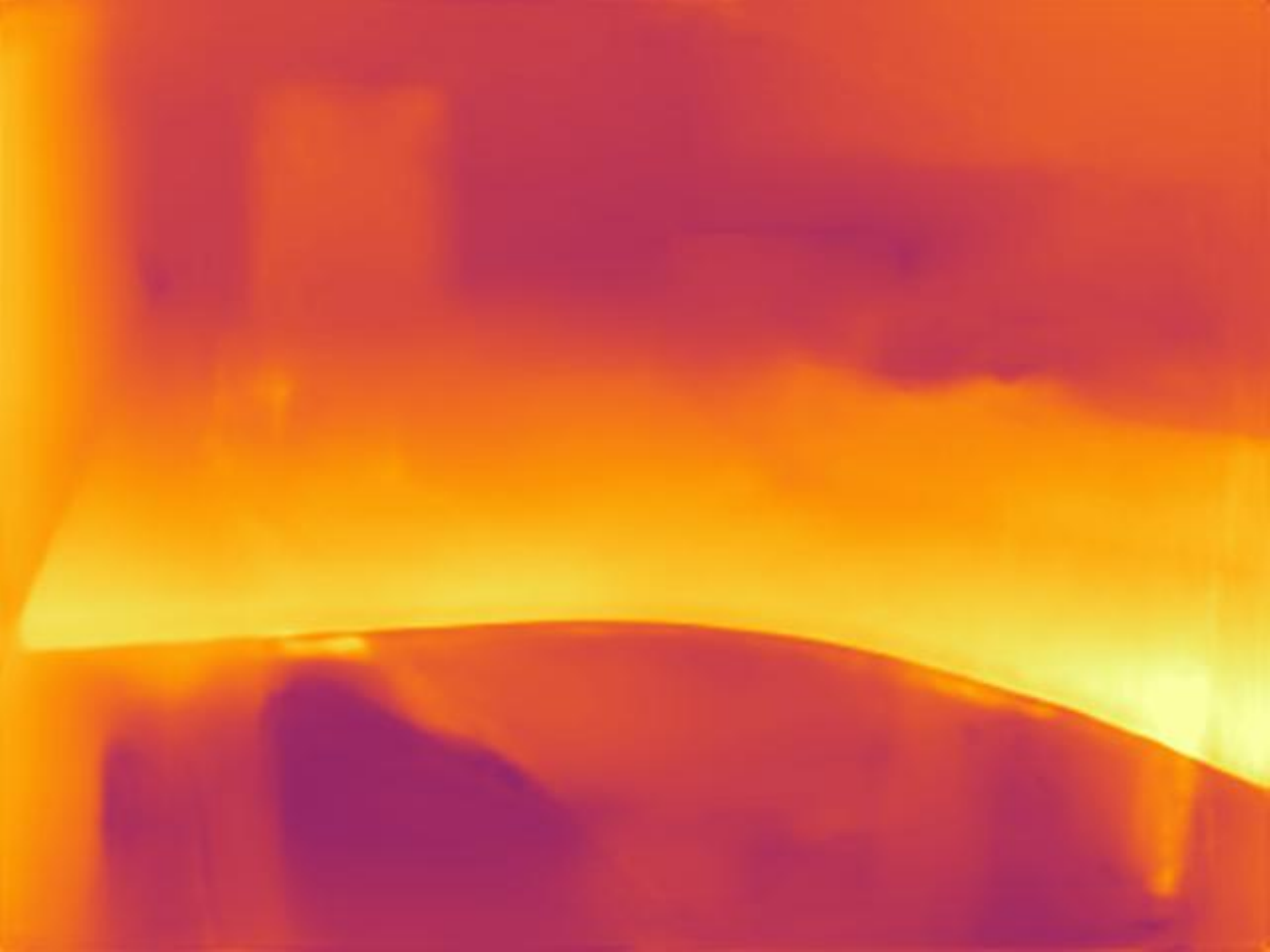}&
    \includegraphics[width=0.096\linewidth]{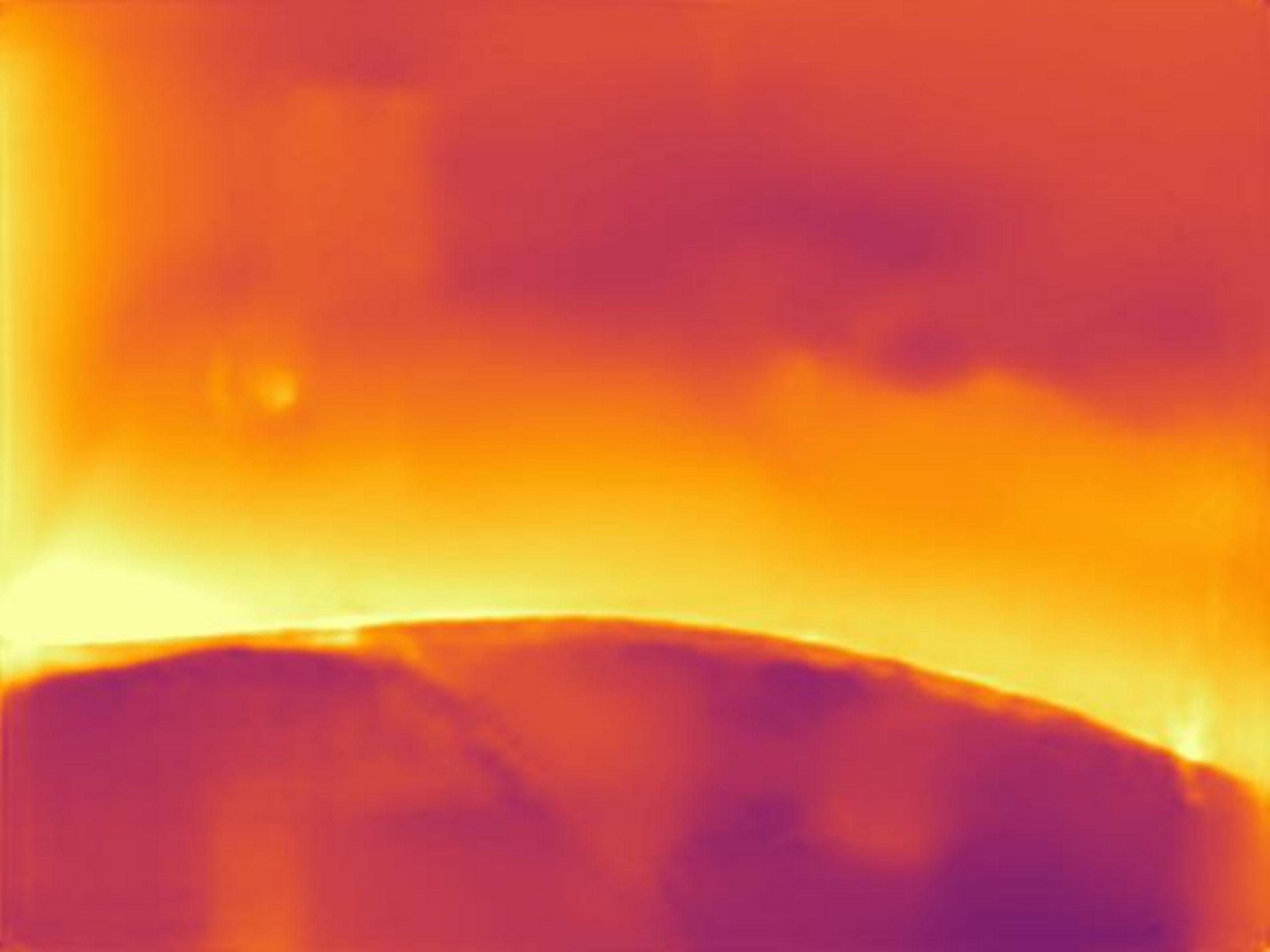}&
    \includegraphics[width=0.096\linewidth]{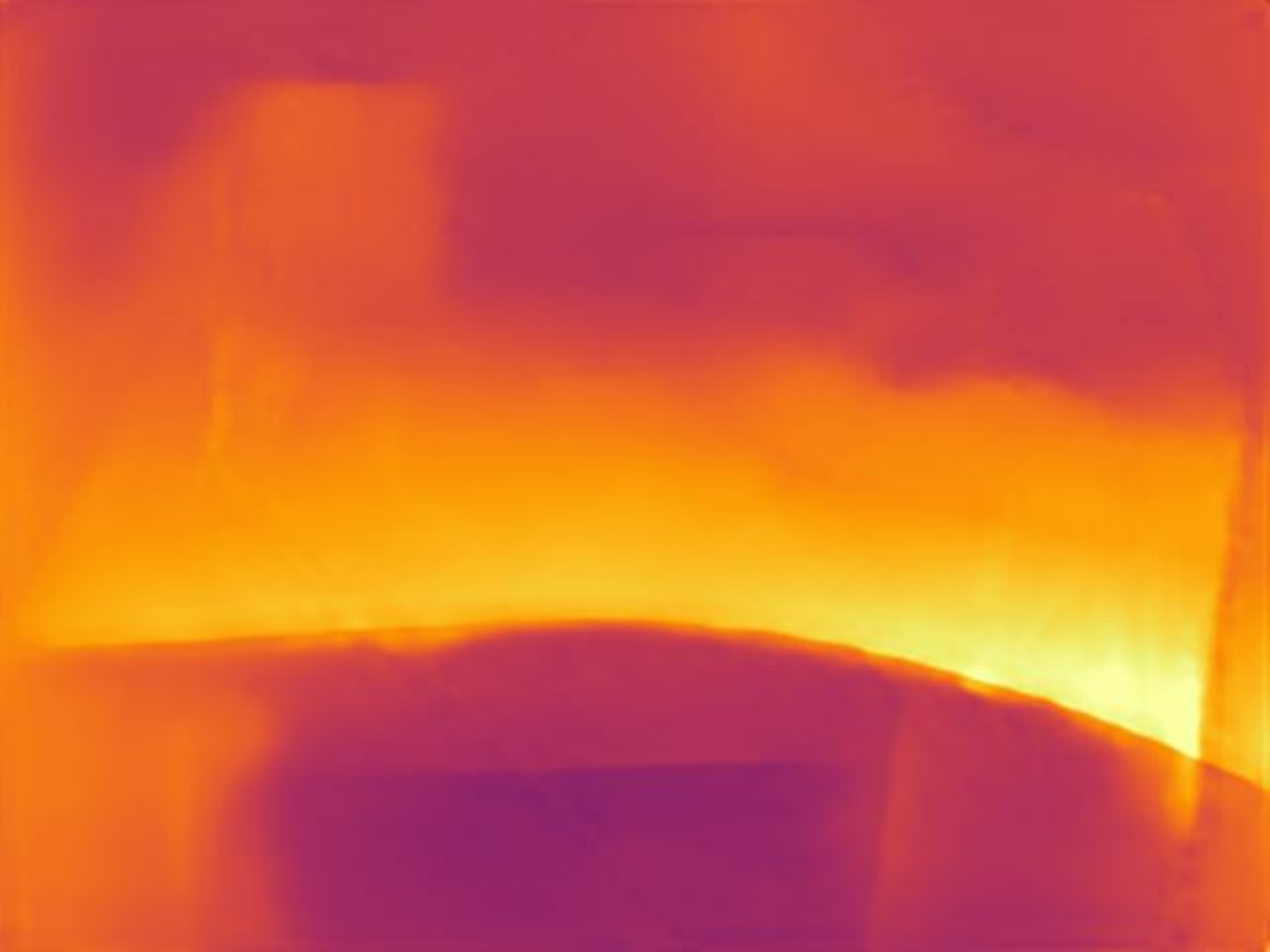}\\
    \vspace{-0.75mm}
    \scriptsize g. &
    \includegraphics[width=0.096\linewidth]{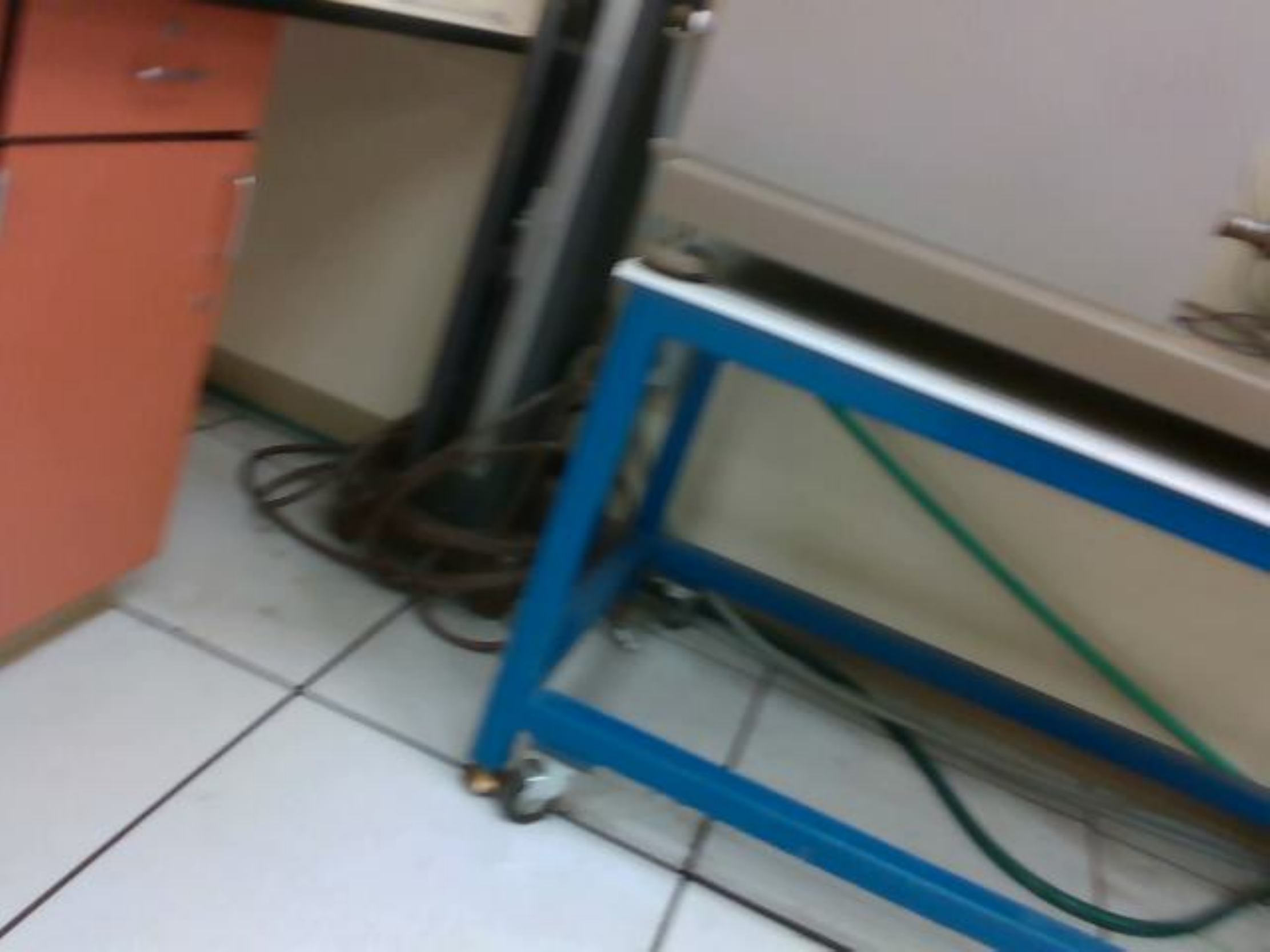}&
    \includegraphics[width=0.096\linewidth]{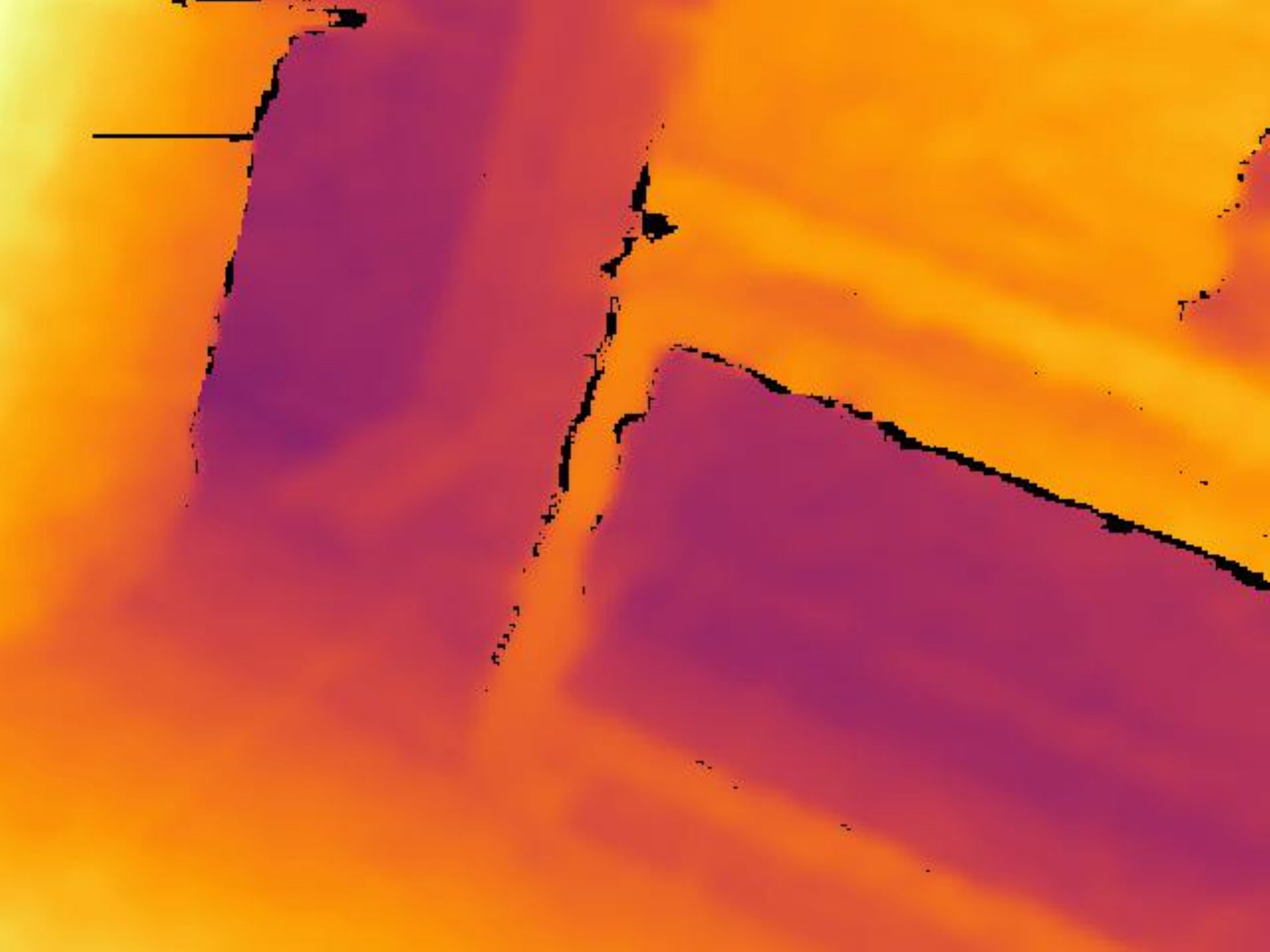}&
    \includegraphics[width=0.096\linewidth]{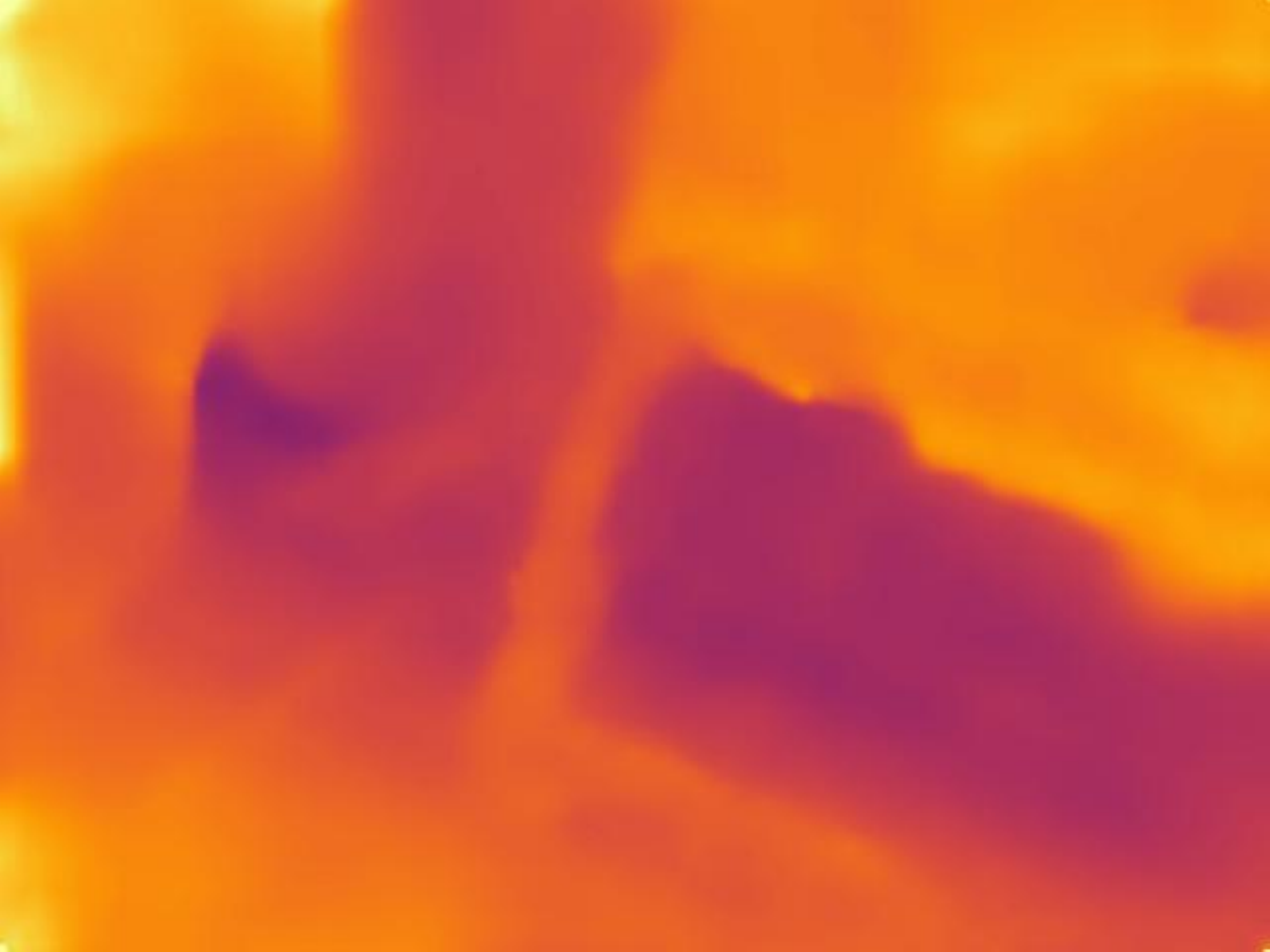}&
    \includegraphics[width=0.096\linewidth]{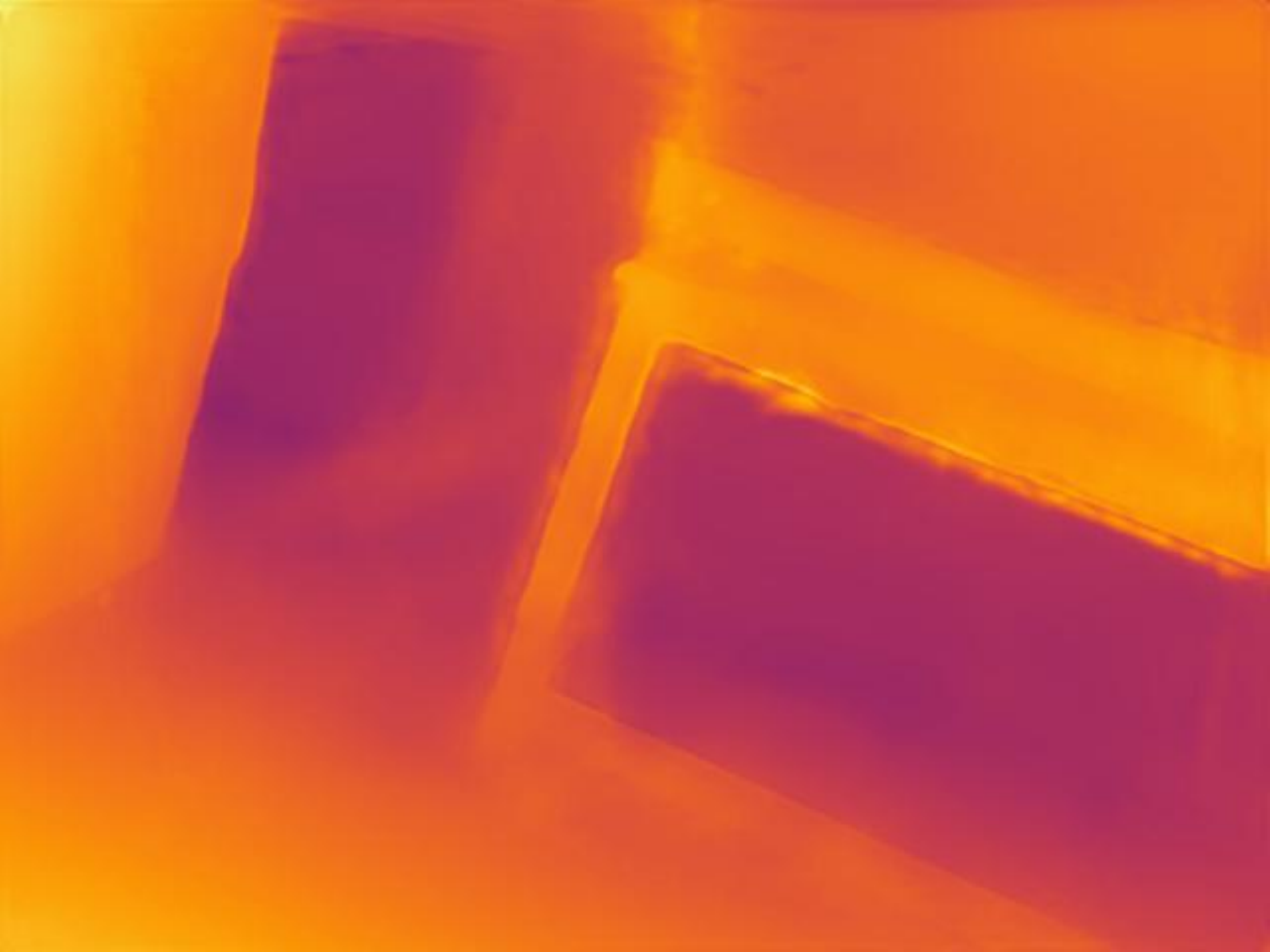}&
    \includegraphics[width=0.096\linewidth]{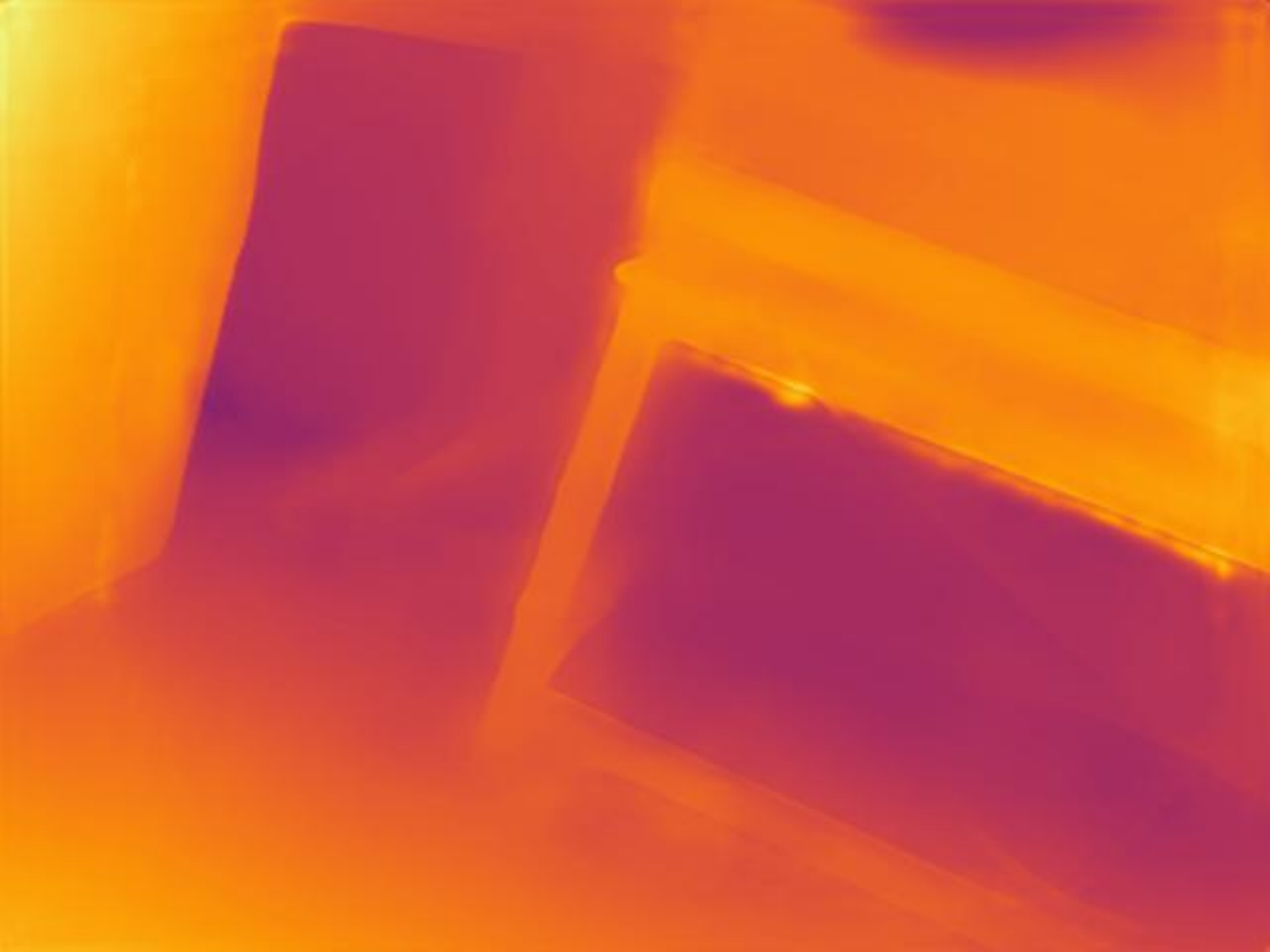}&
    \includegraphics[width=0.096\linewidth]{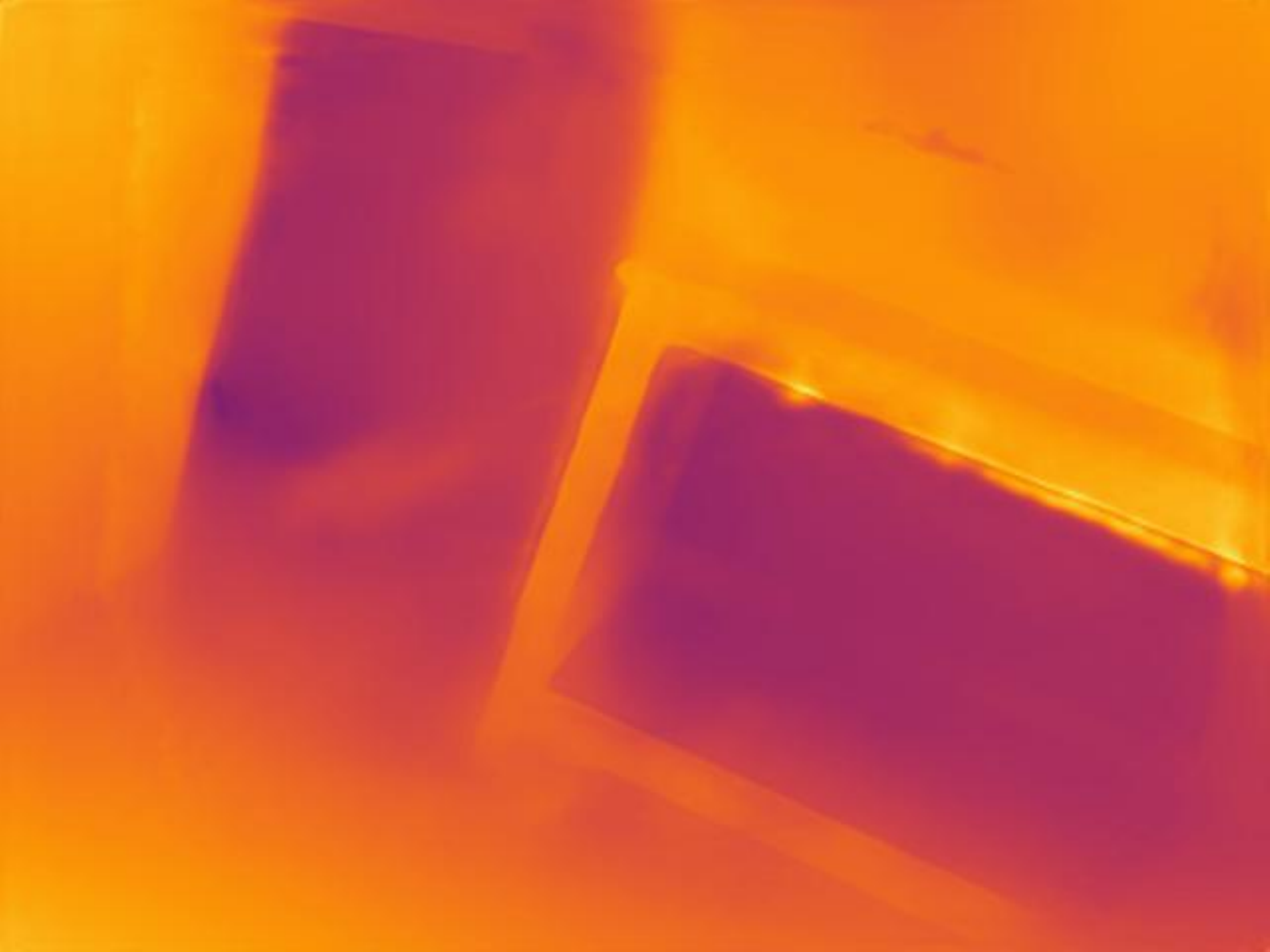}&
    \includegraphics[width=0.096\linewidth]{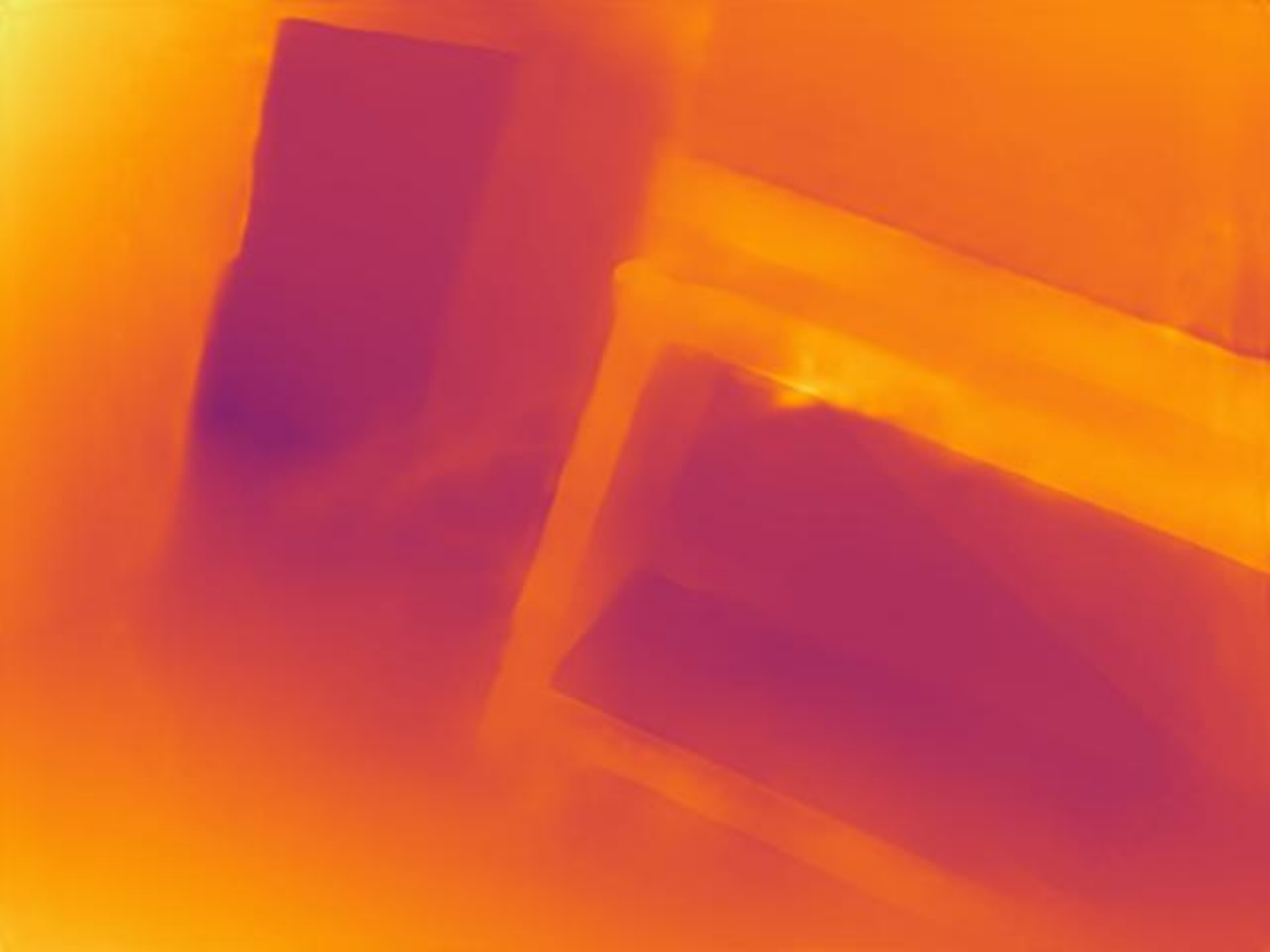}&
    \includegraphics[width=0.096\linewidth]{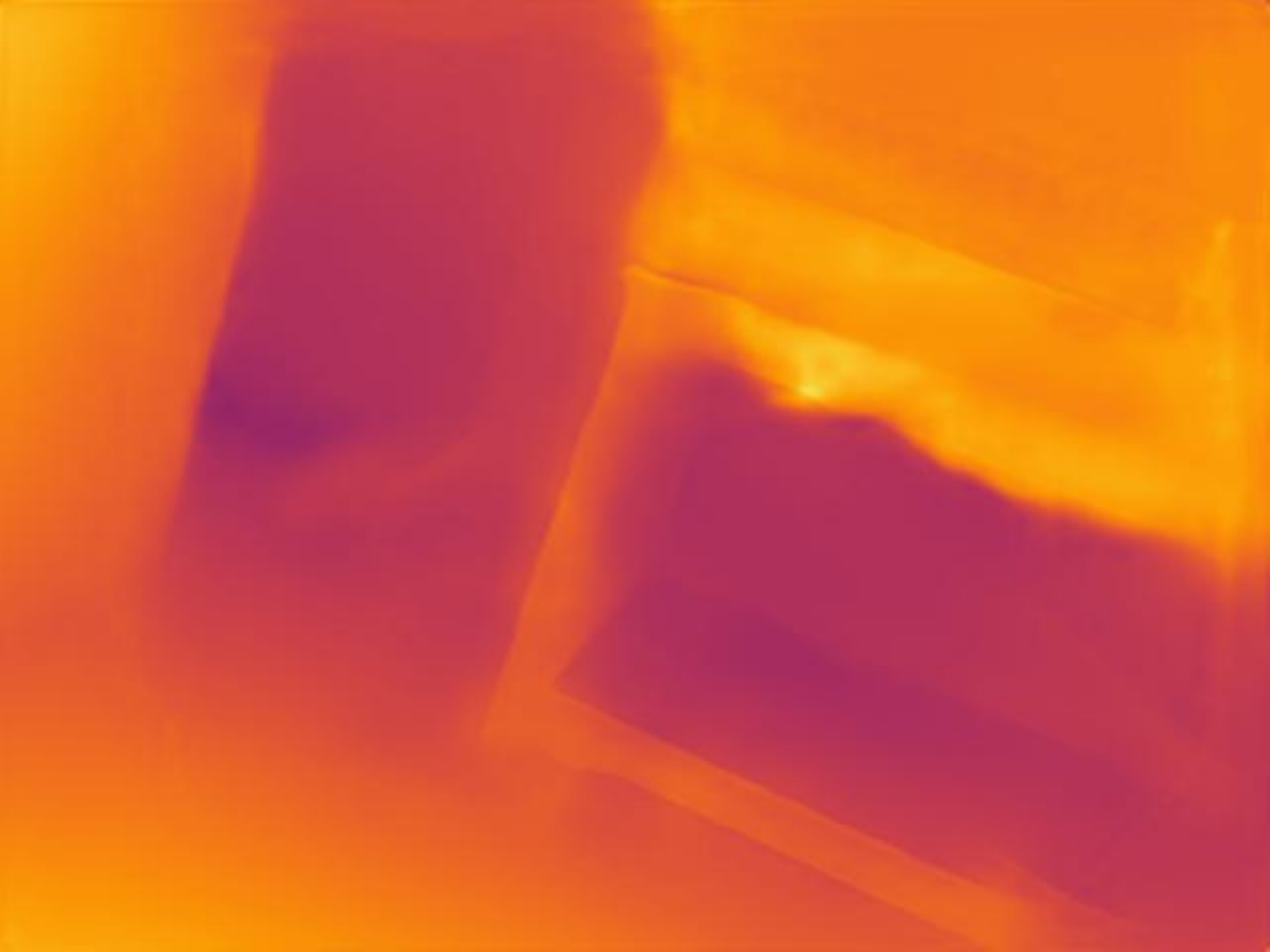}&
    \includegraphics[width=0.096\linewidth]{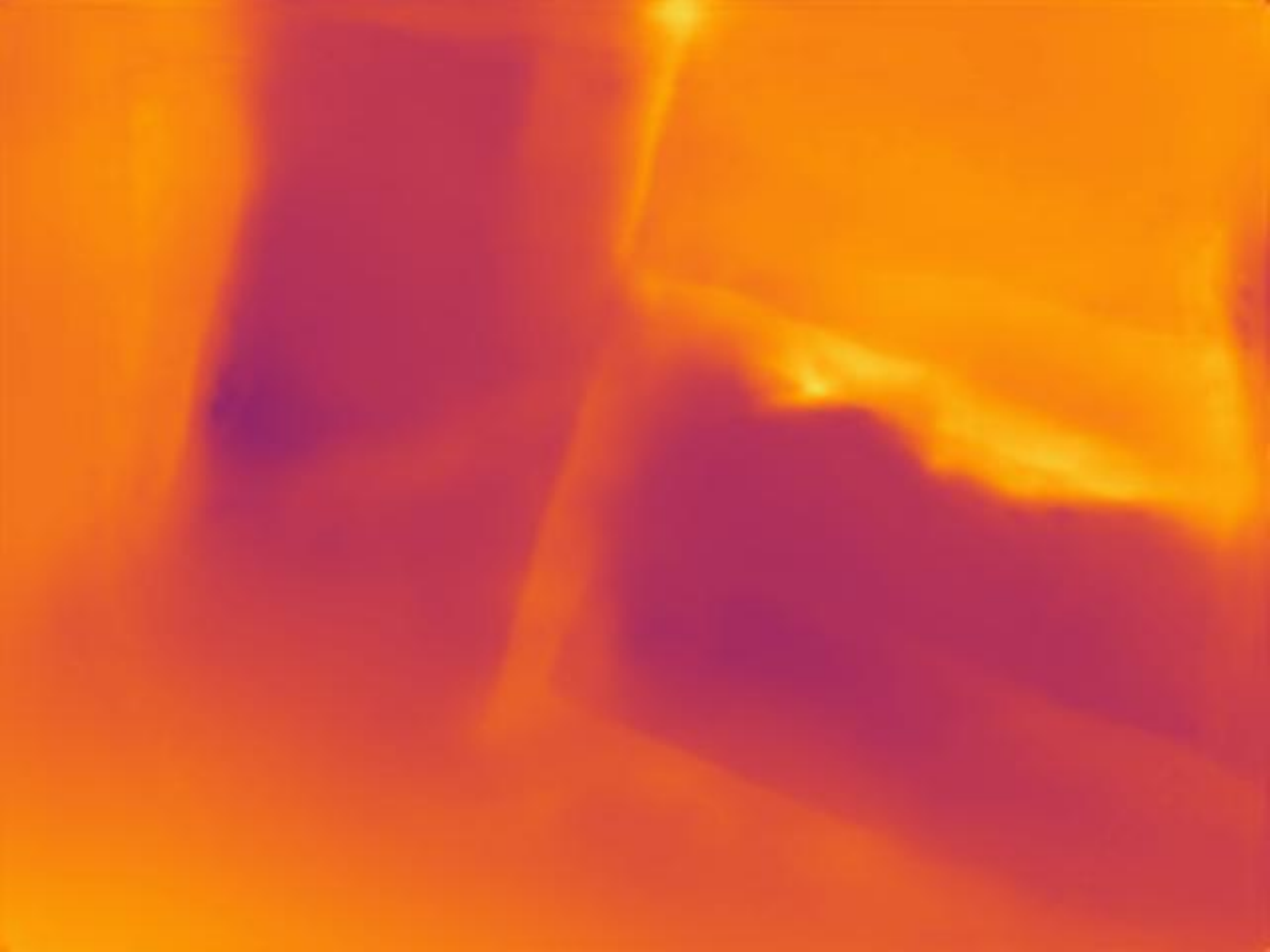}&
    \includegraphics[width=0.096\linewidth]{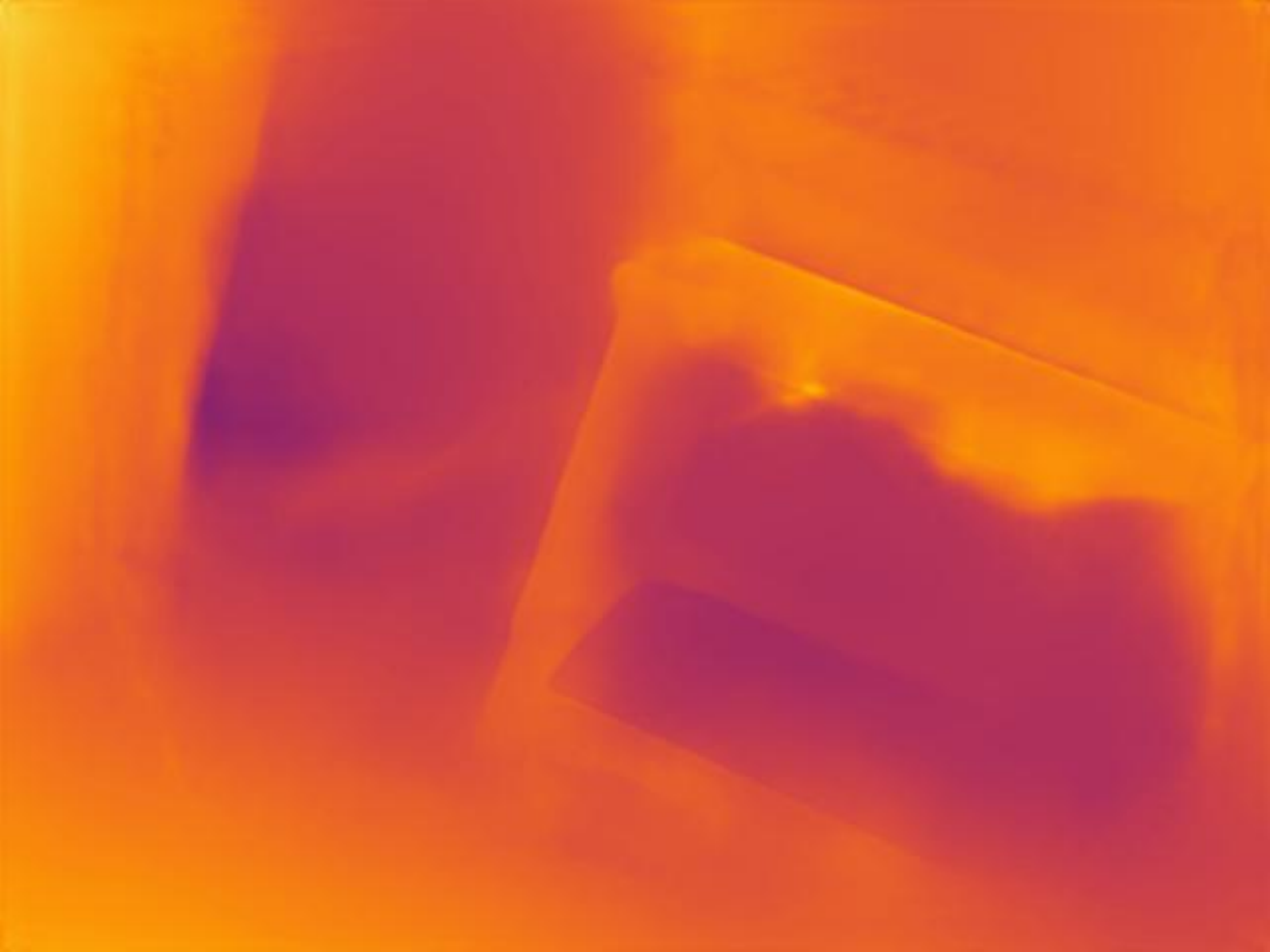}\\
    \vspace{-0.75mm}
    \scriptsize h. &
    \includegraphics[width=0.096\linewidth]{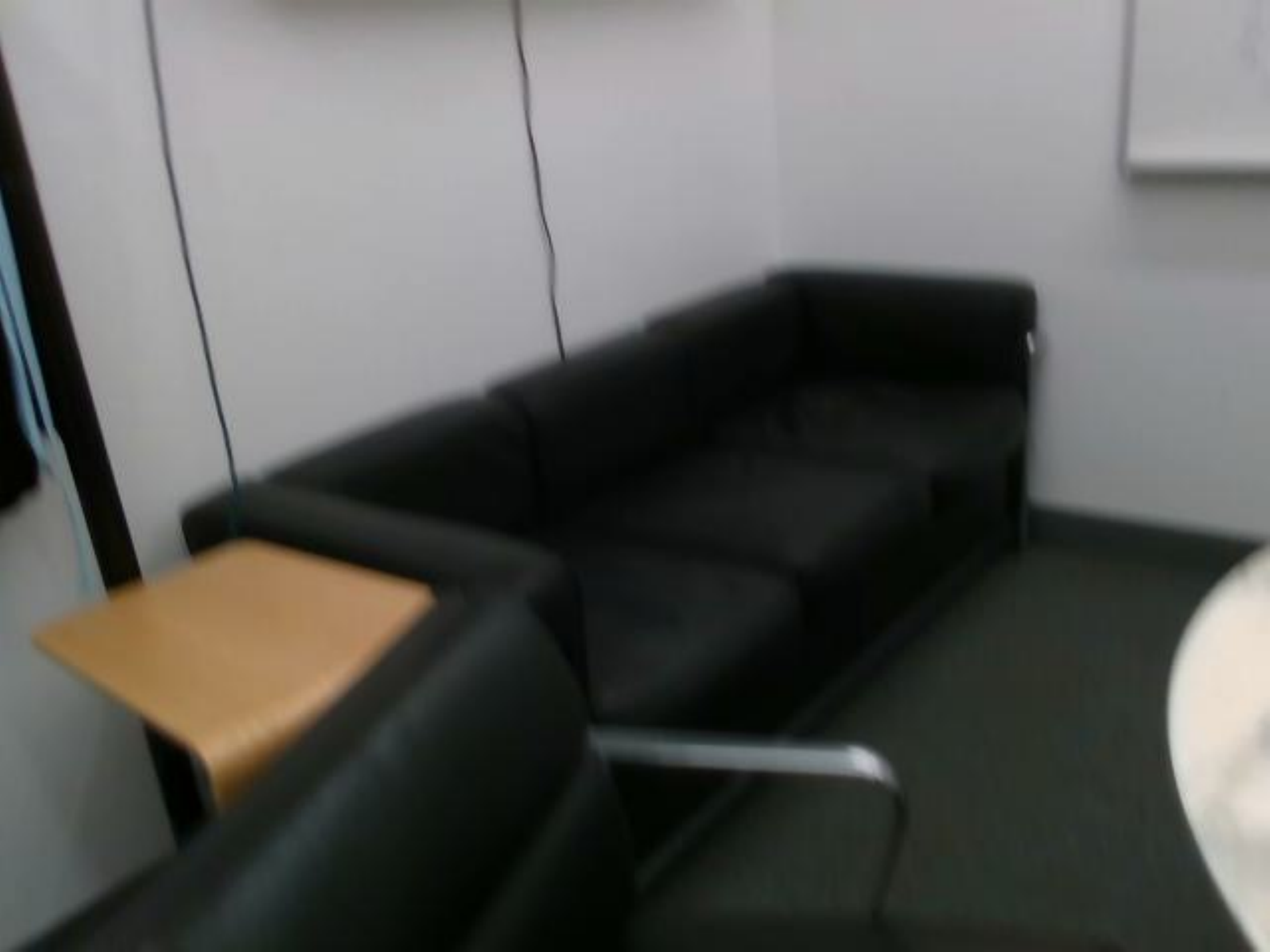}&
    \includegraphics[width=0.096\linewidth]{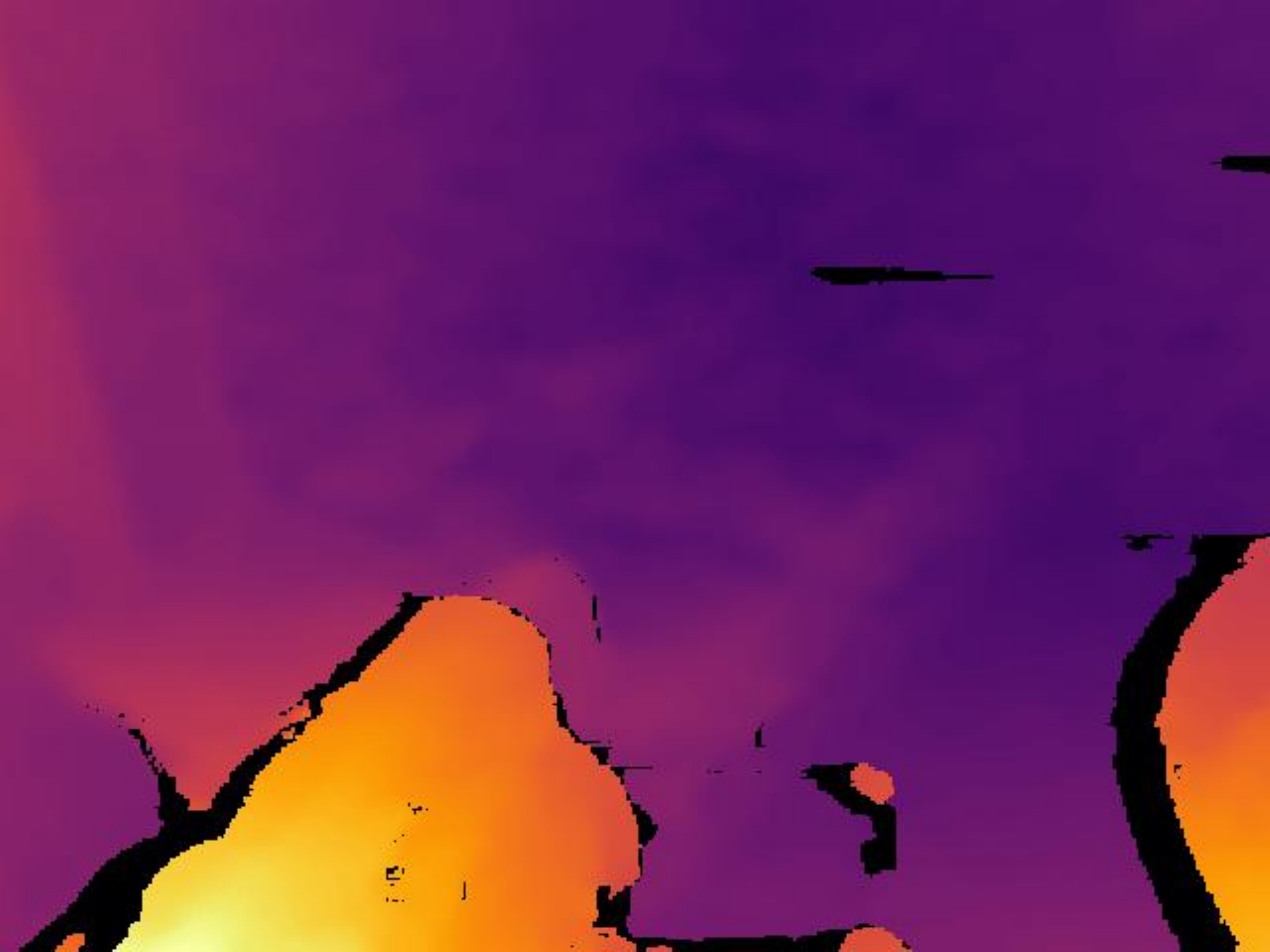}&
    \includegraphics[width=0.096\linewidth]{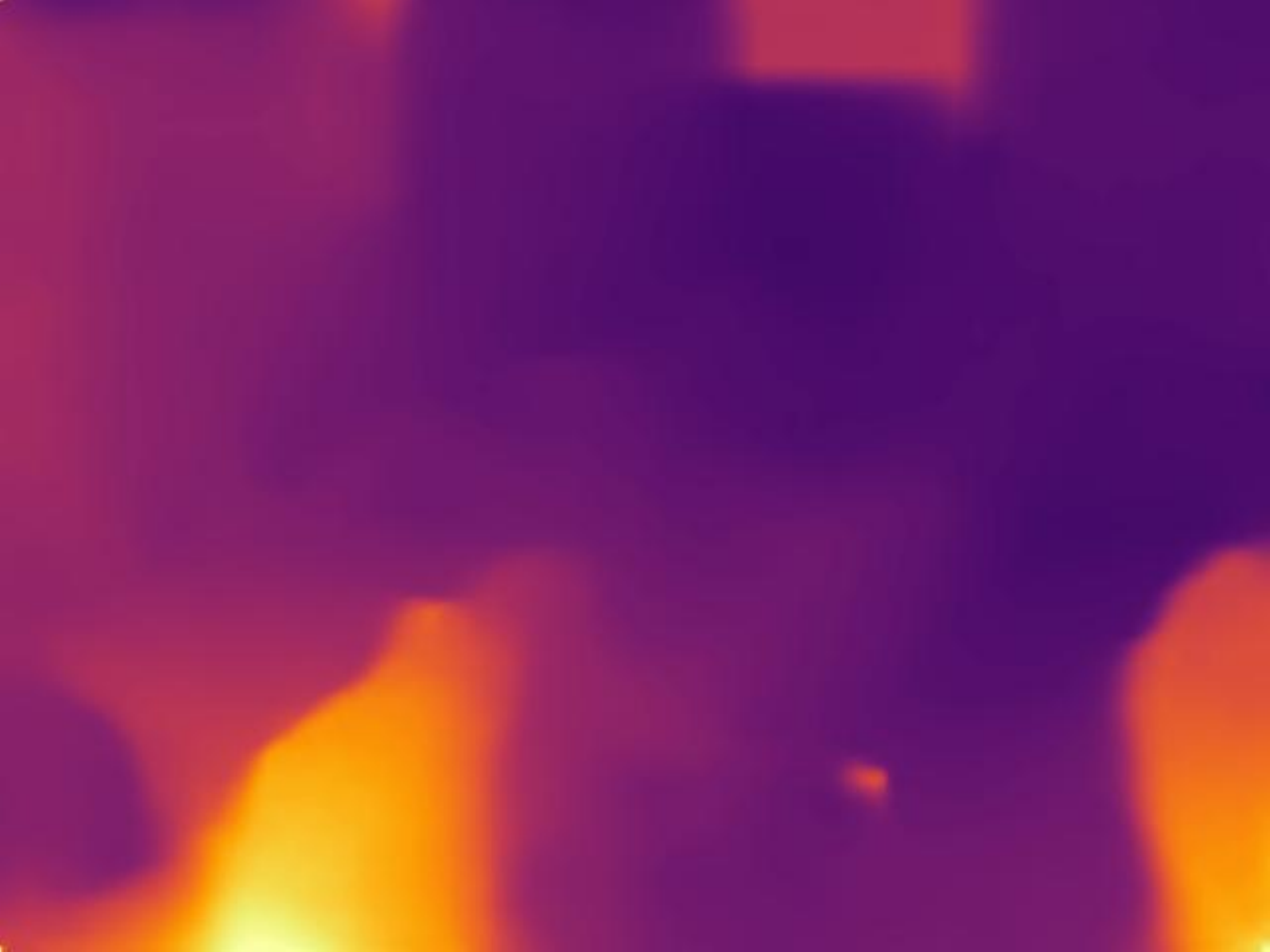}&
    \includegraphics[width=0.096\linewidth]{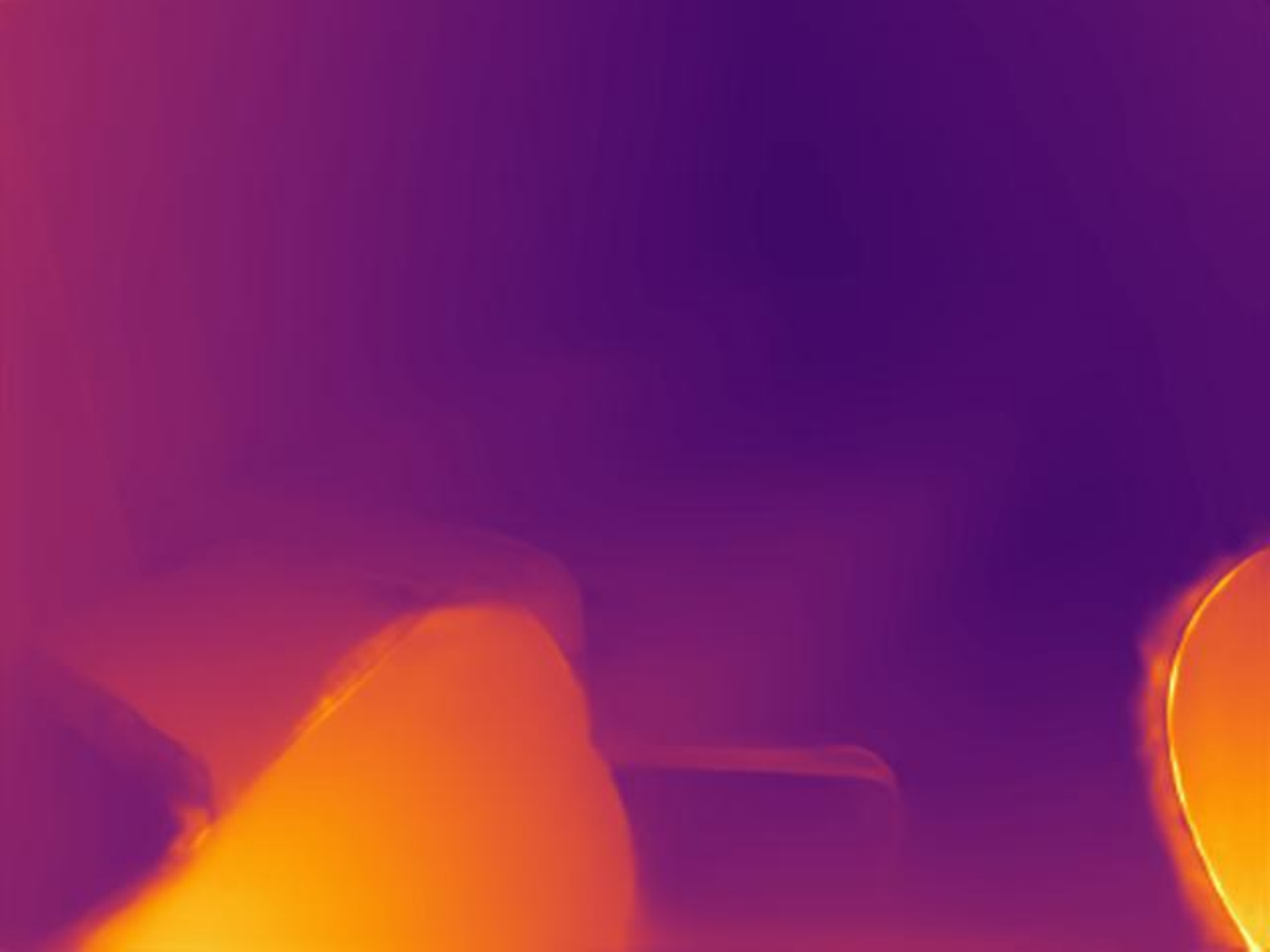}&
    \includegraphics[width=0.096\linewidth]{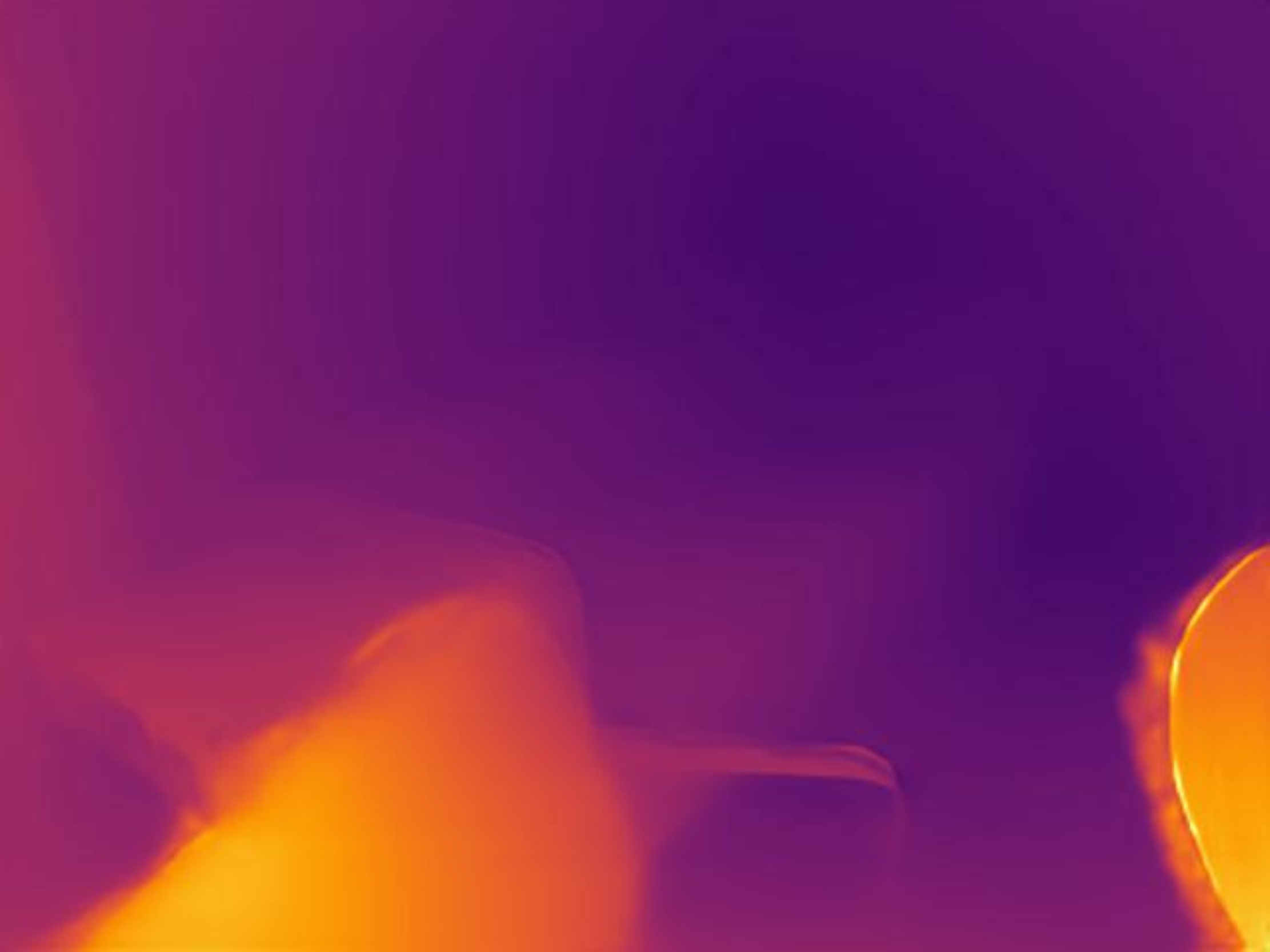}&
    \includegraphics[width=0.096\linewidth]{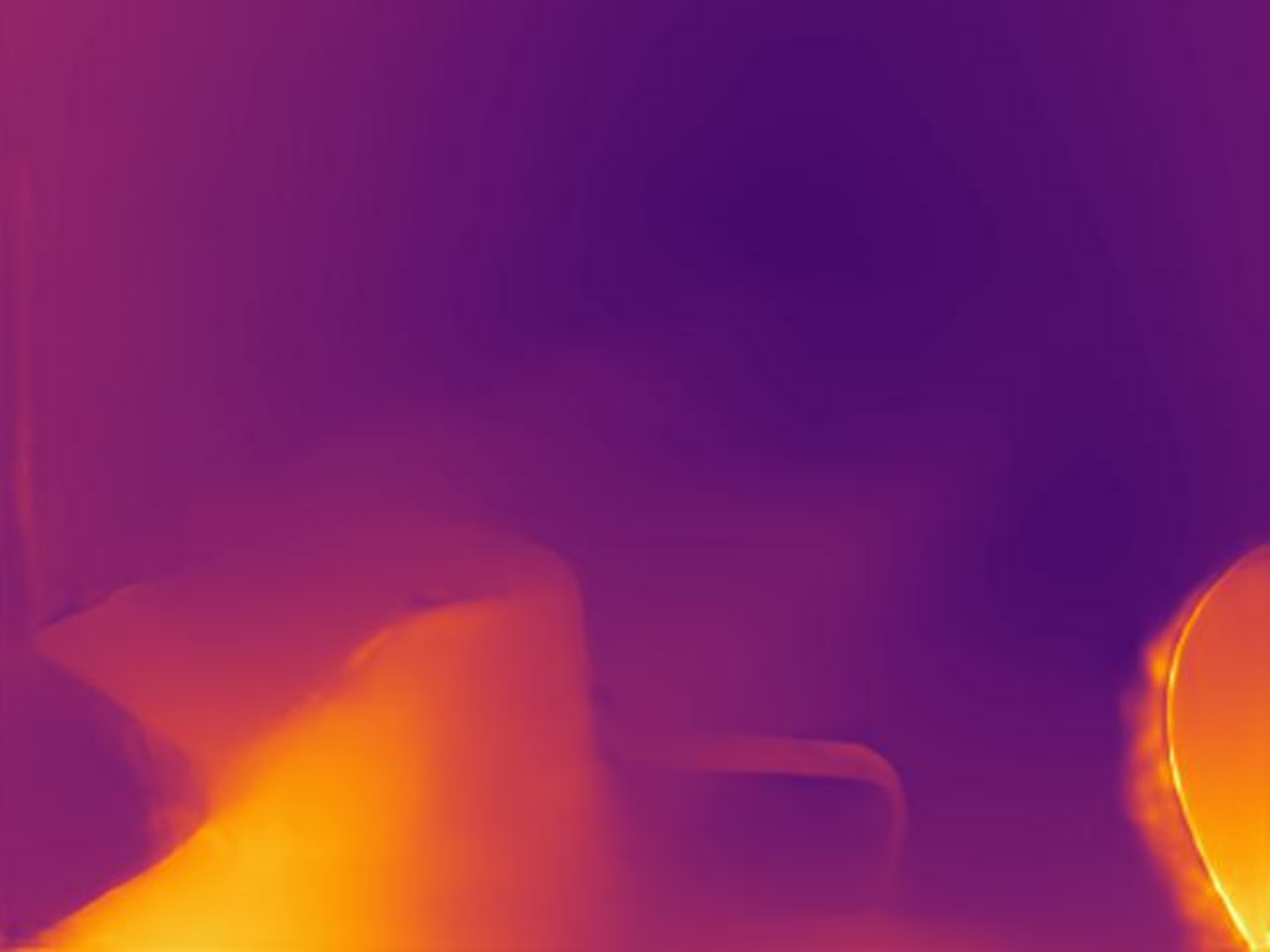}&
    \includegraphics[width=0.096\linewidth]{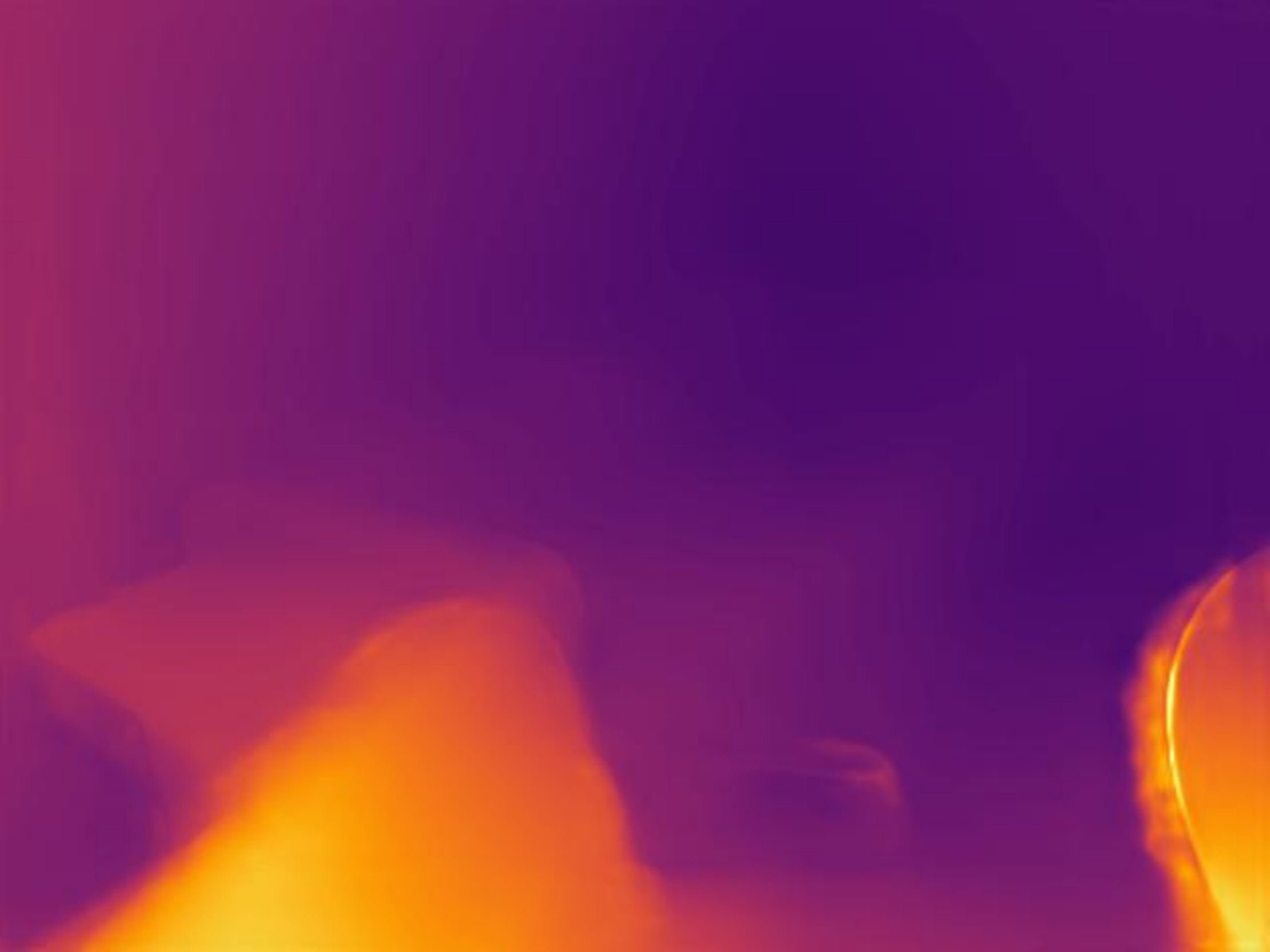}&
    \includegraphics[width=0.096\linewidth]{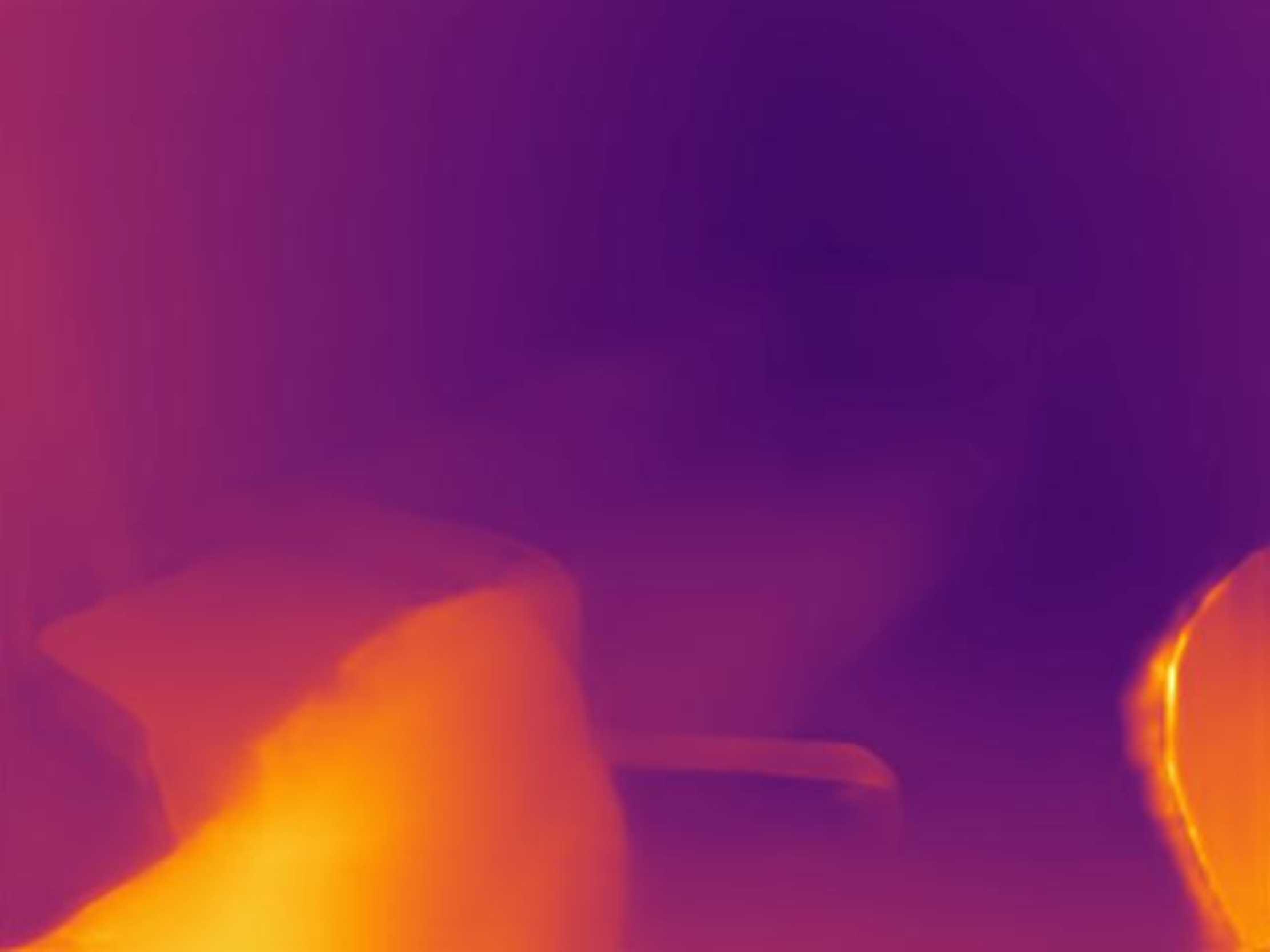}&
    \includegraphics[width=0.096\linewidth]{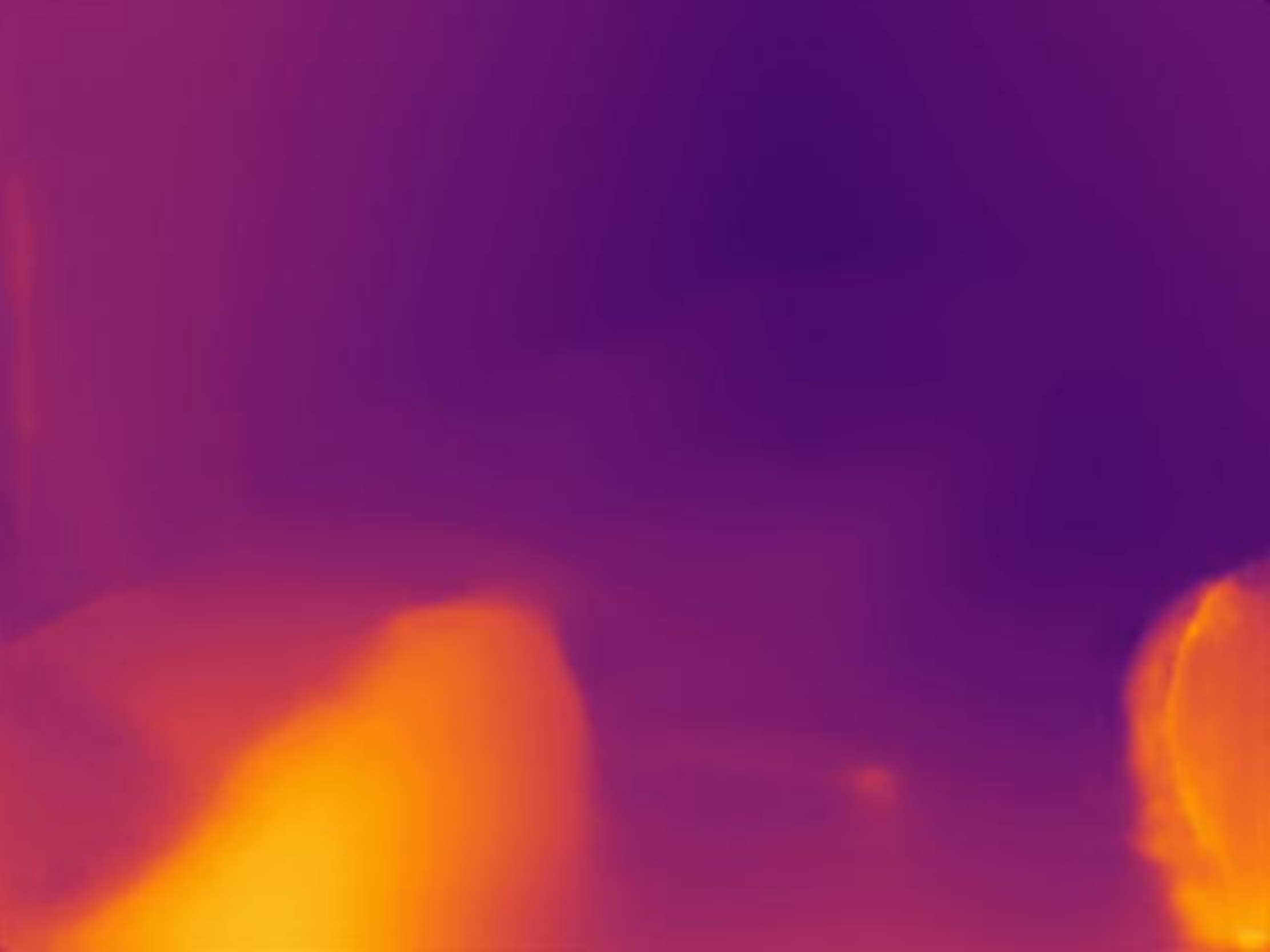}&
    \includegraphics[width=0.096\linewidth]{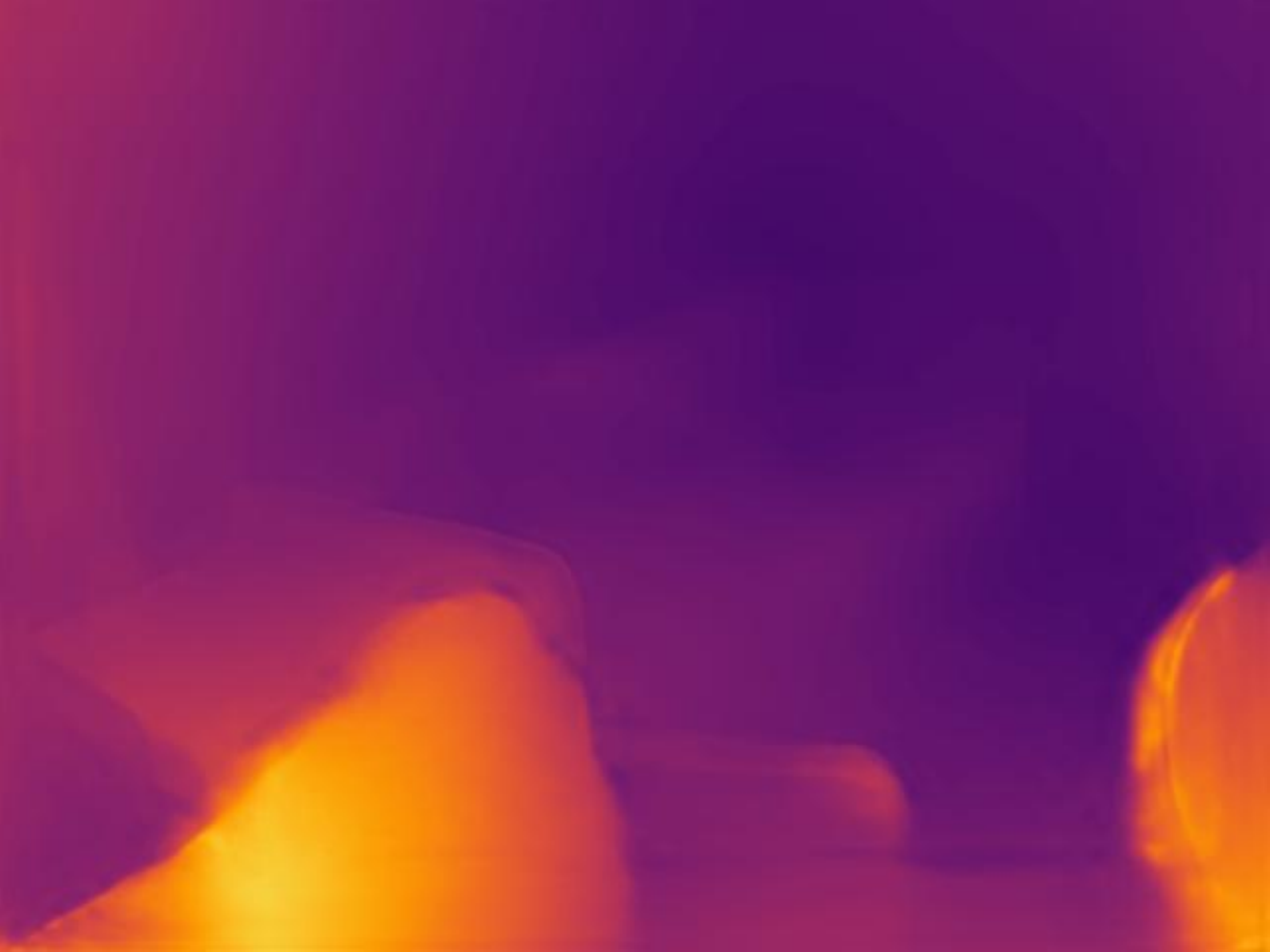}\\
    \vspace{-0.75mm}
    \scriptsize i. &
    \includegraphics[width=0.096\linewidth]{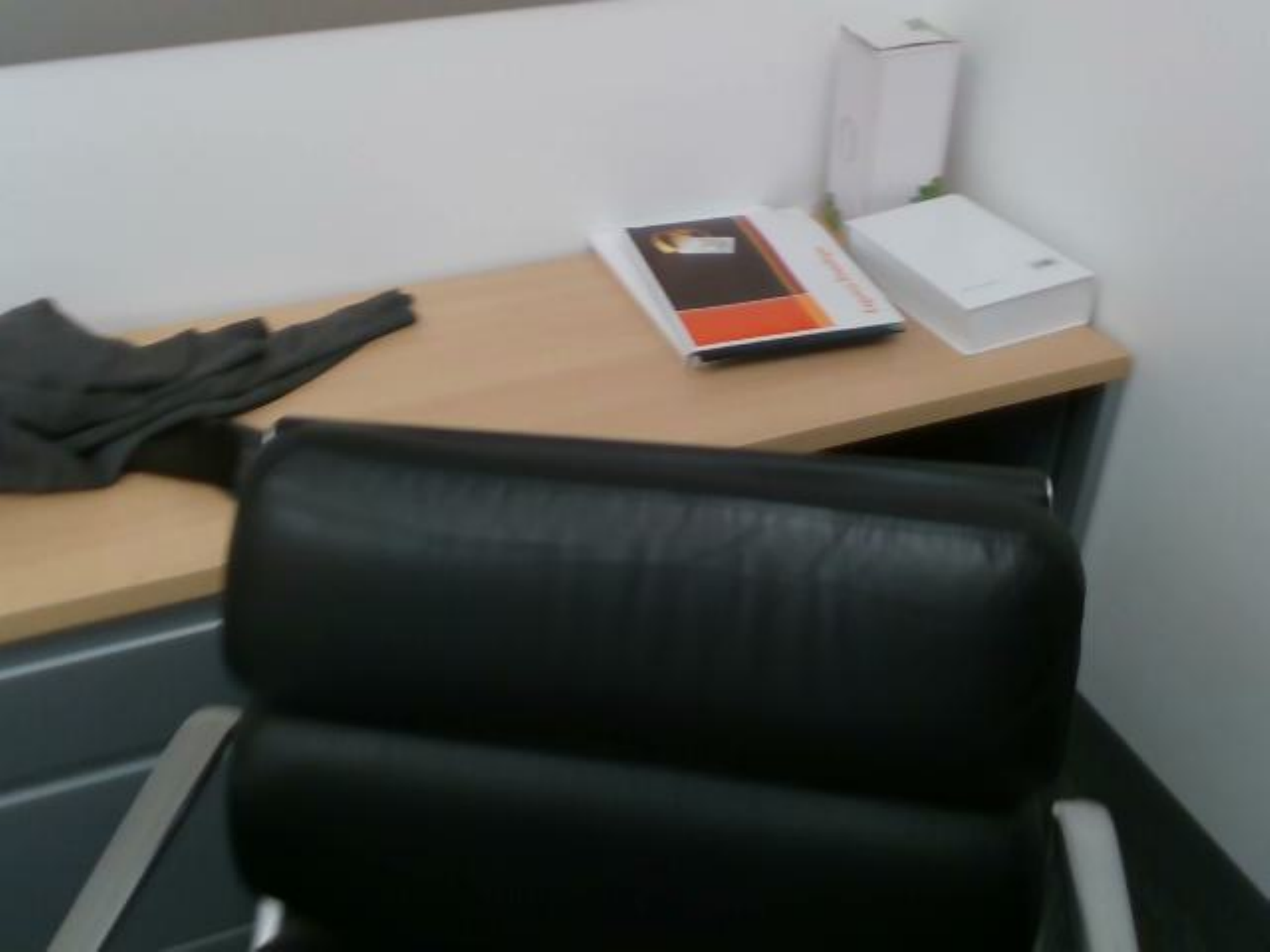}&
    \includegraphics[width=0.096\linewidth]{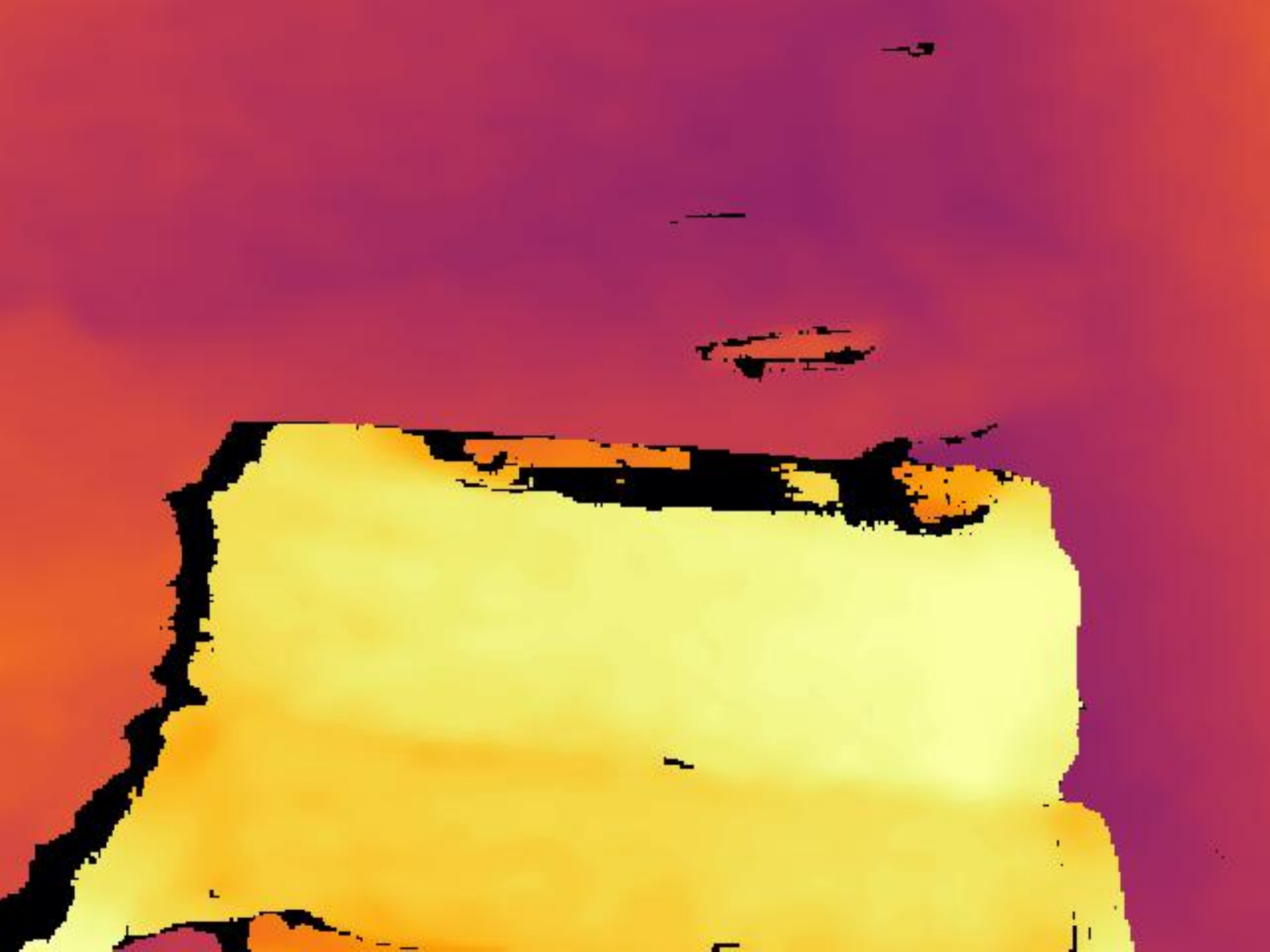}&
    \includegraphics[width=0.096\linewidth]{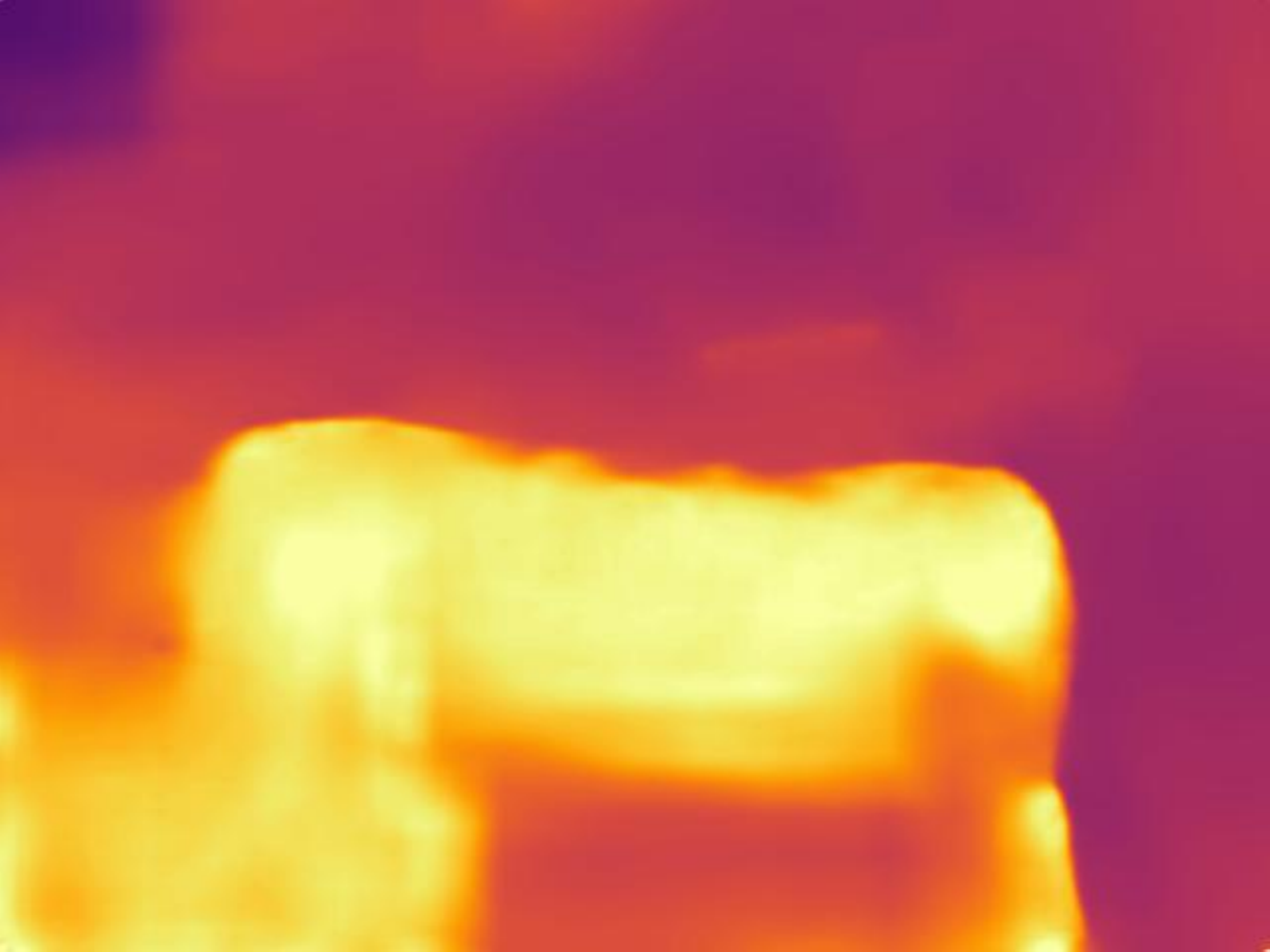}&
    \includegraphics[width=0.096\linewidth]{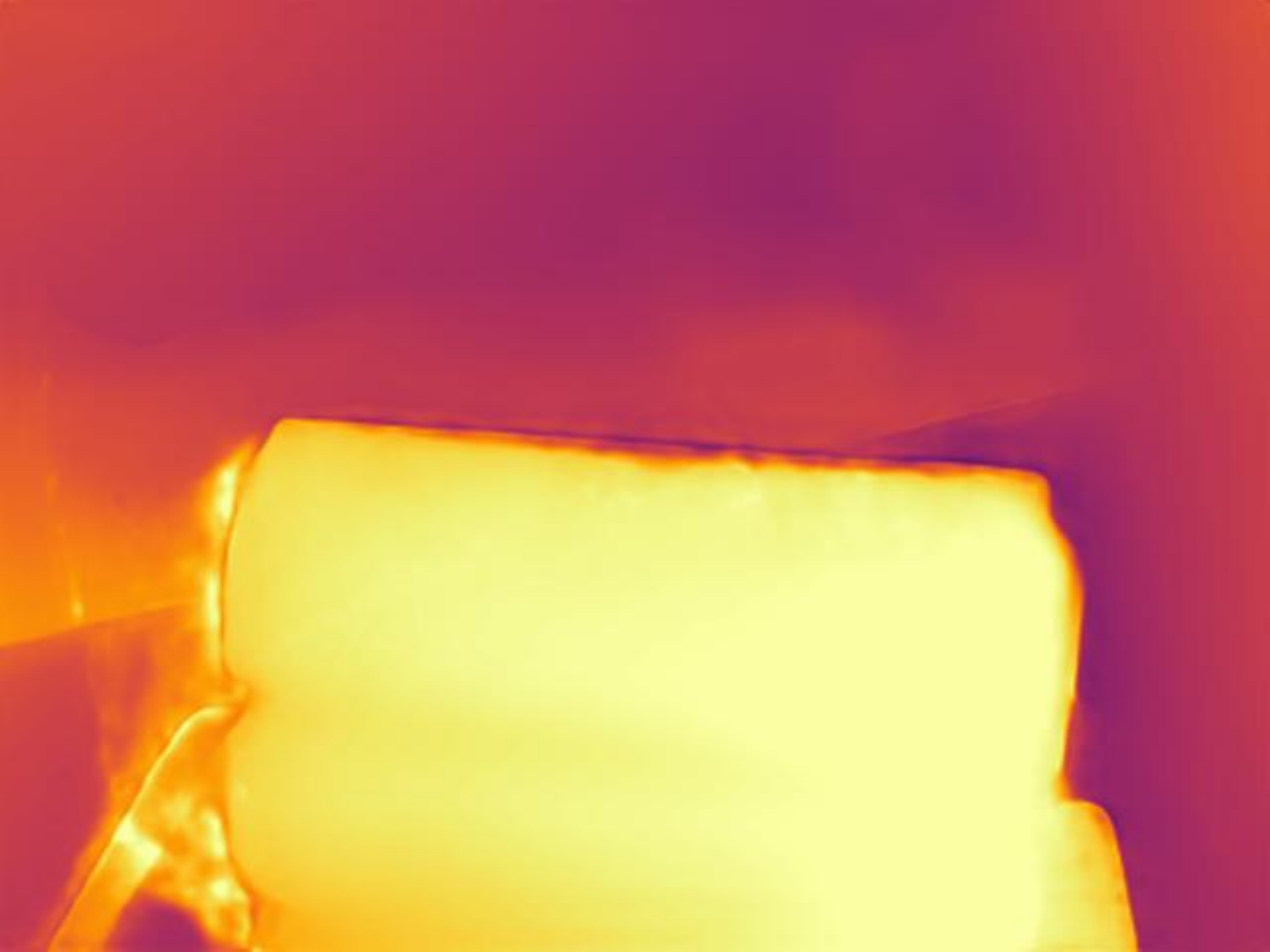}&
    \includegraphics[width=0.096\linewidth]{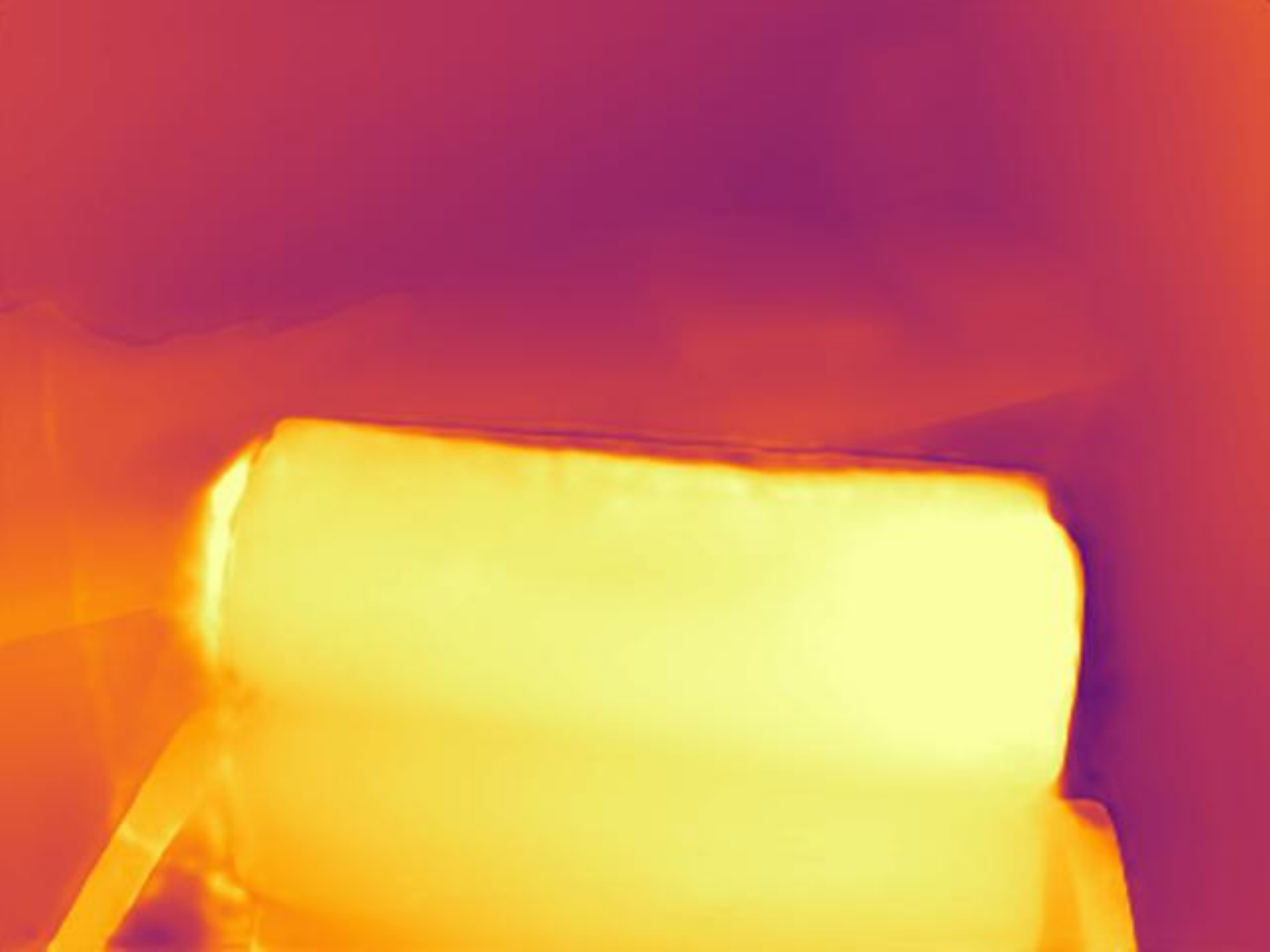}&
    \includegraphics[width=0.096\linewidth]{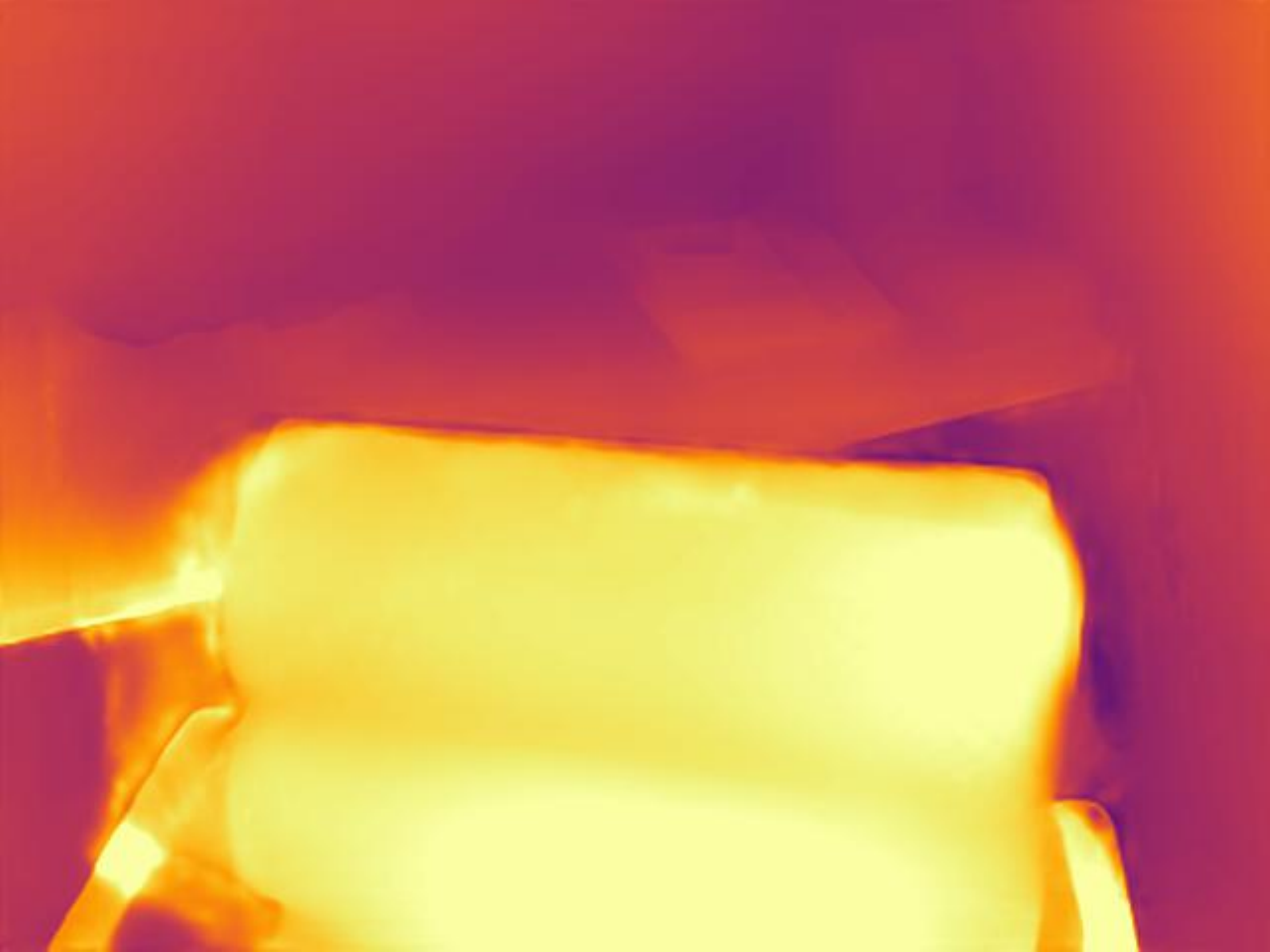}&
    \includegraphics[width=0.096\linewidth]{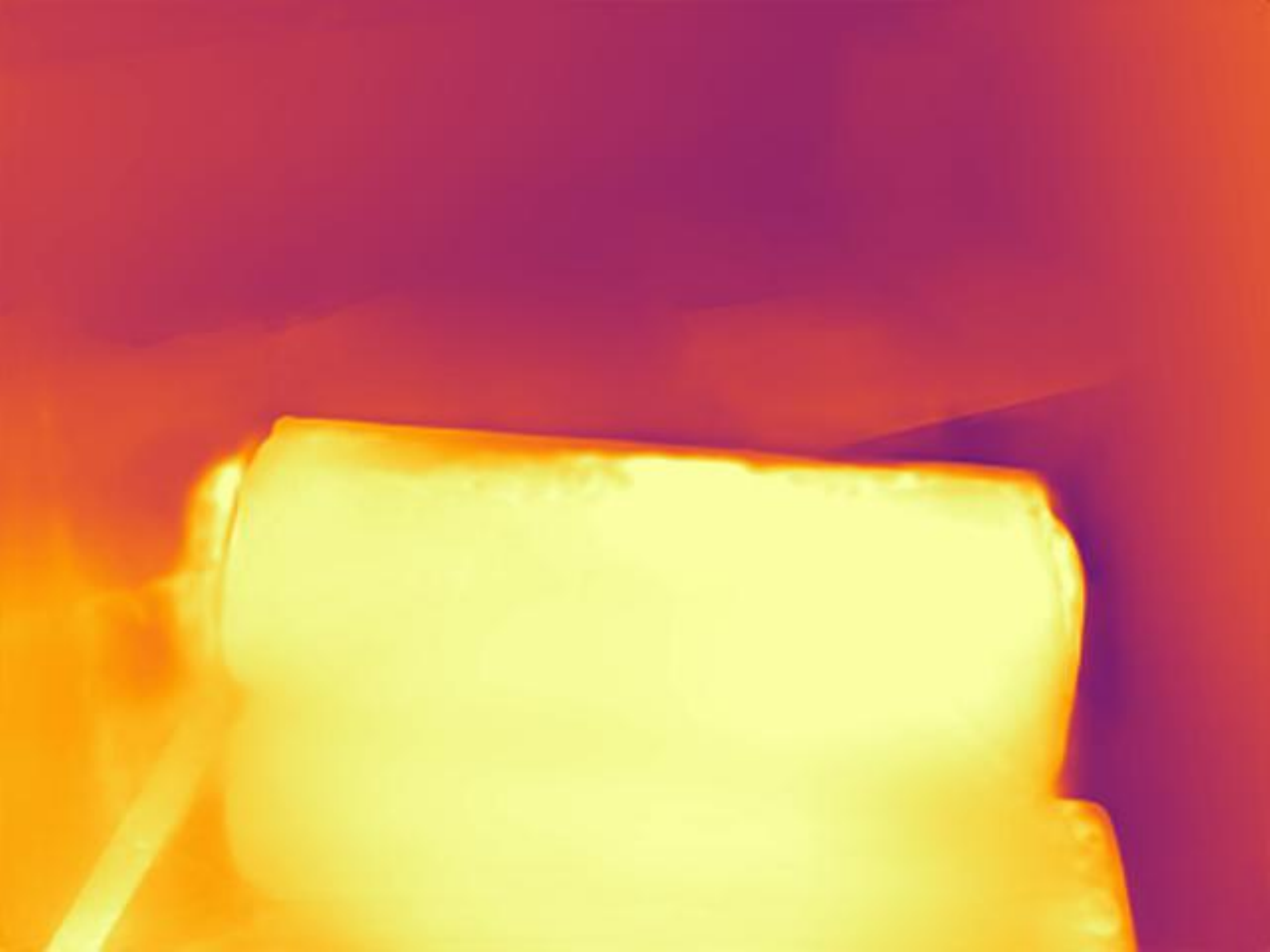}&
    \includegraphics[width=0.096\linewidth]{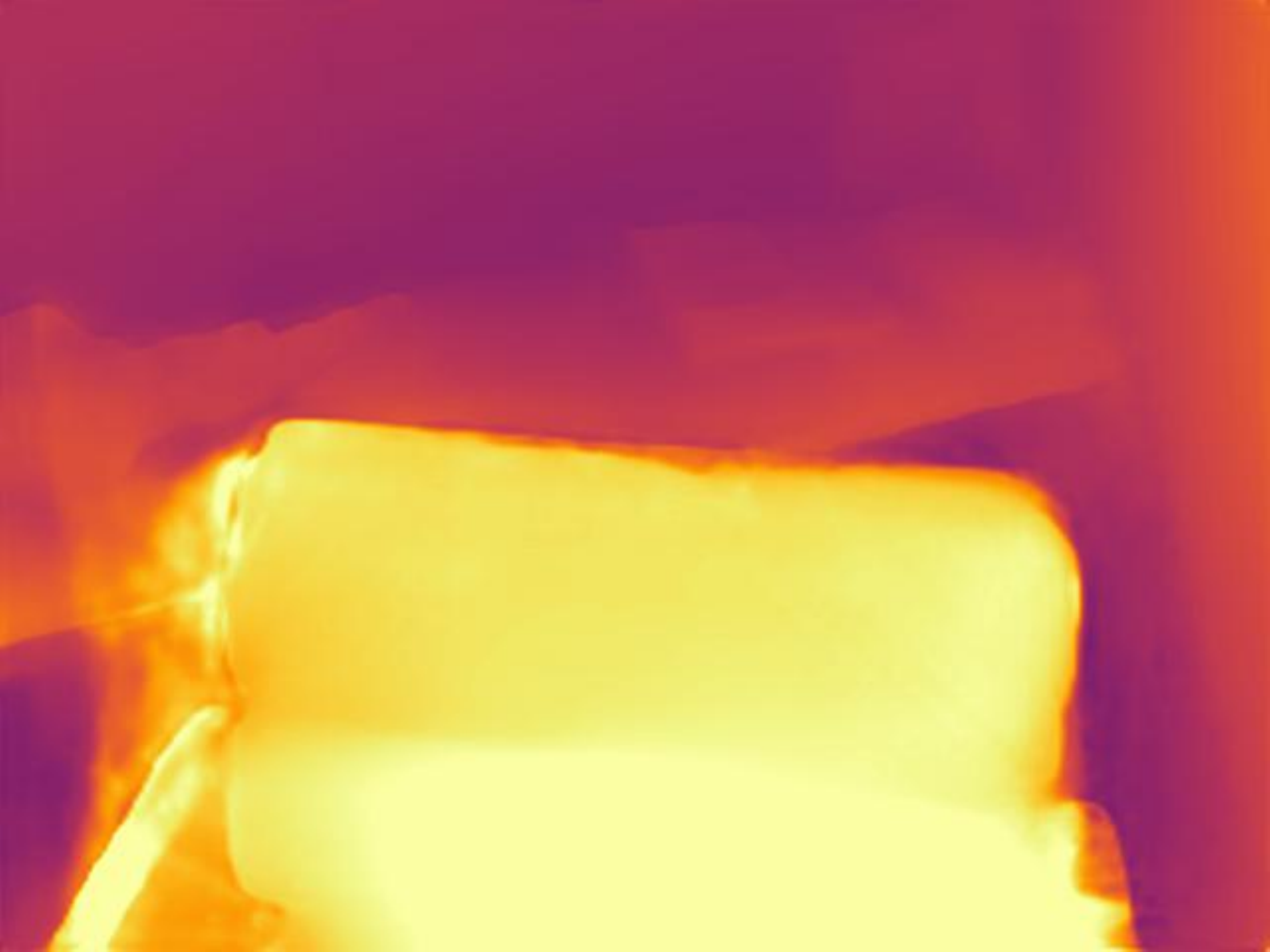}&
    \includegraphics[width=0.096\linewidth]{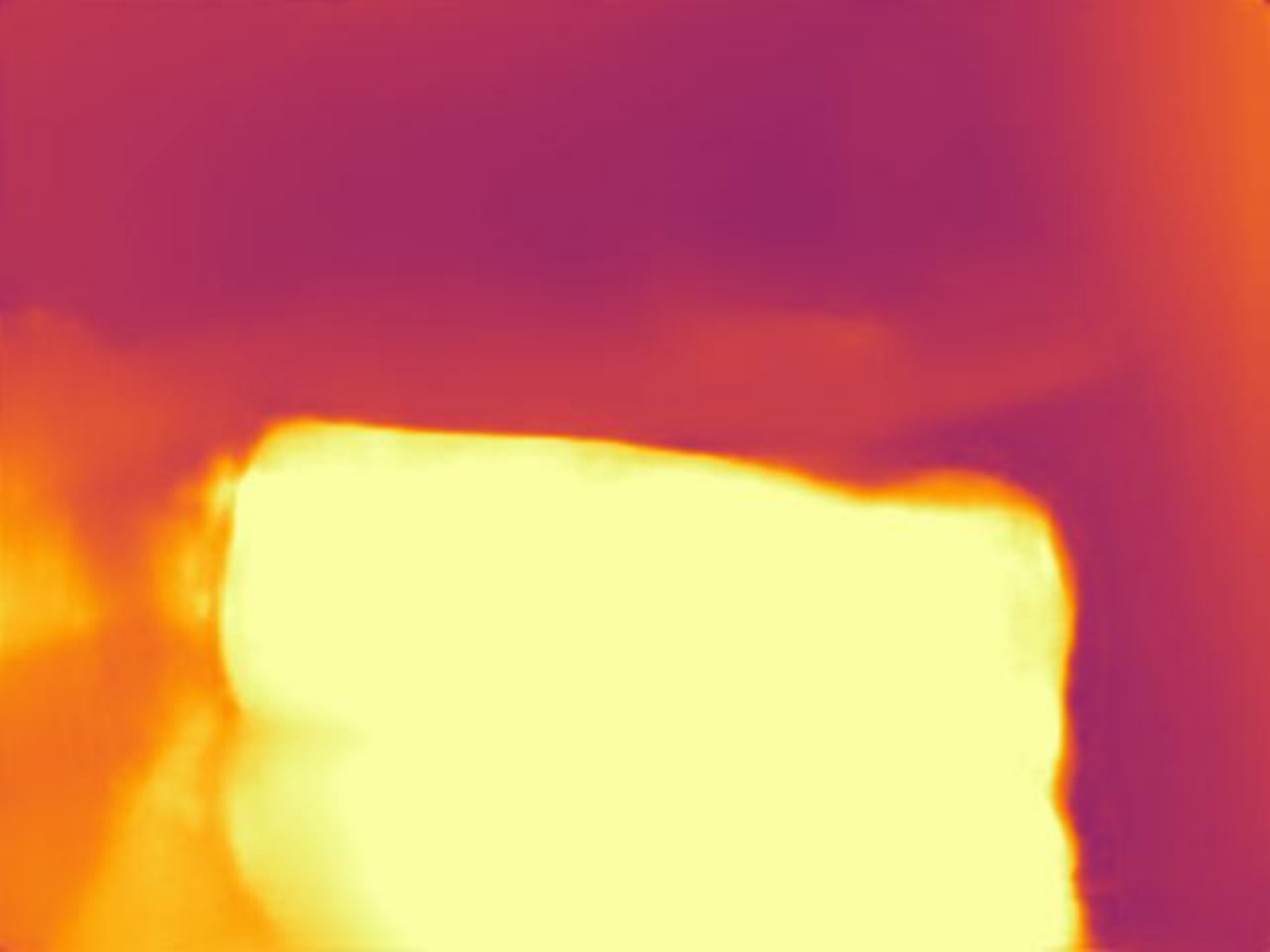}&
    \includegraphics[width=0.096\linewidth]{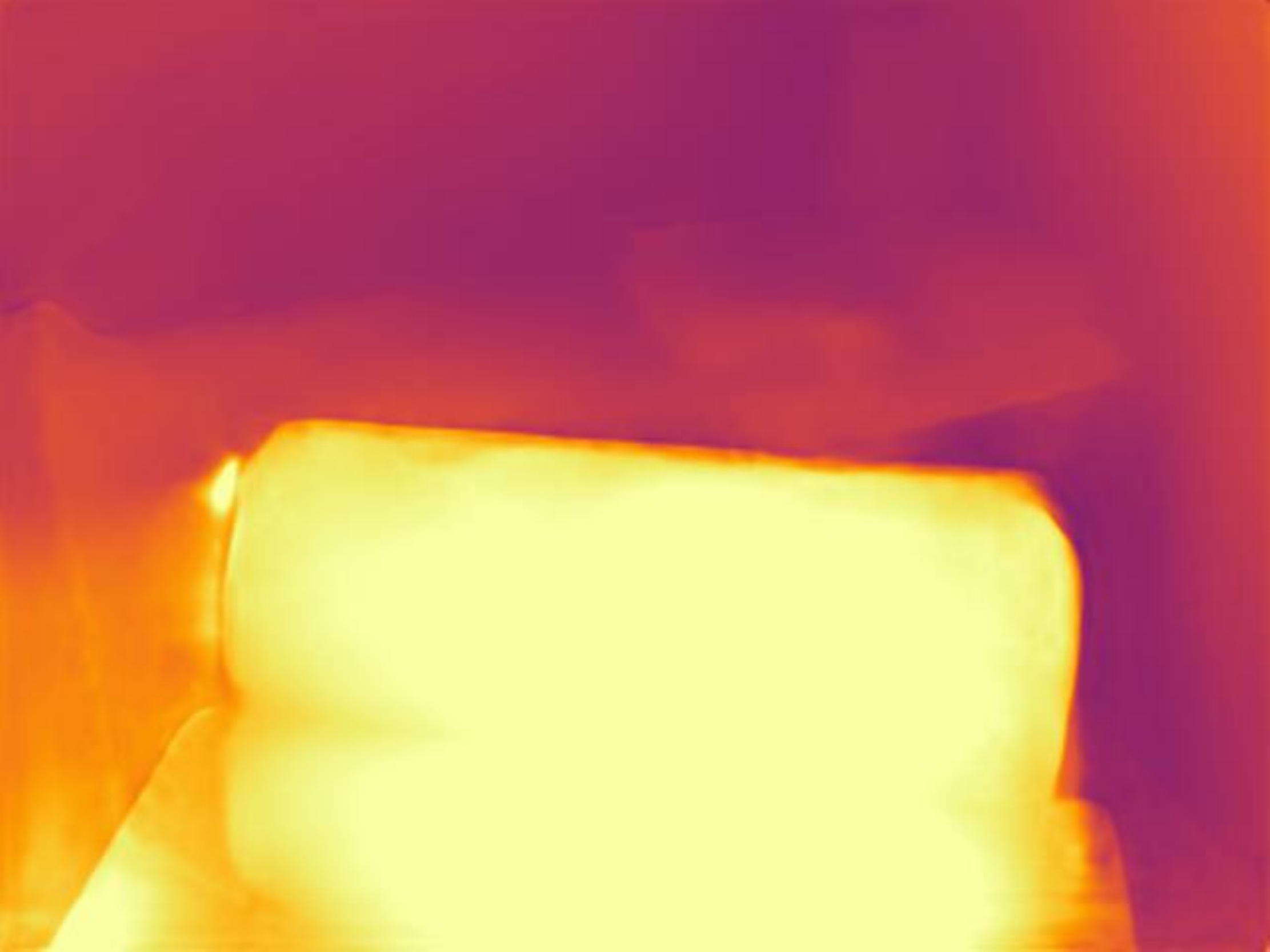}\\
    \vspace{-0.75mm}
    \scriptsize j. &
    \includegraphics[width=0.096\linewidth]{suppl/comparison_void_150/kbnet/image/0000000570.pdf}&
    \includegraphics[width=0.096\linewidth]{suppl/comparison_void_150/kbnet/ground_truth/0000000570.pdf}&
    \includegraphics[width=0.096\linewidth]{suppl/comparison_void_150/kbnet/output_depth/0000000570.pdf}&
    \includegraphics[width=0.096\linewidth]{suppl/comparison_void_150/dpt-beit-l/row_9_col_5.pdf}&
    \includegraphics[width=0.096\linewidth]{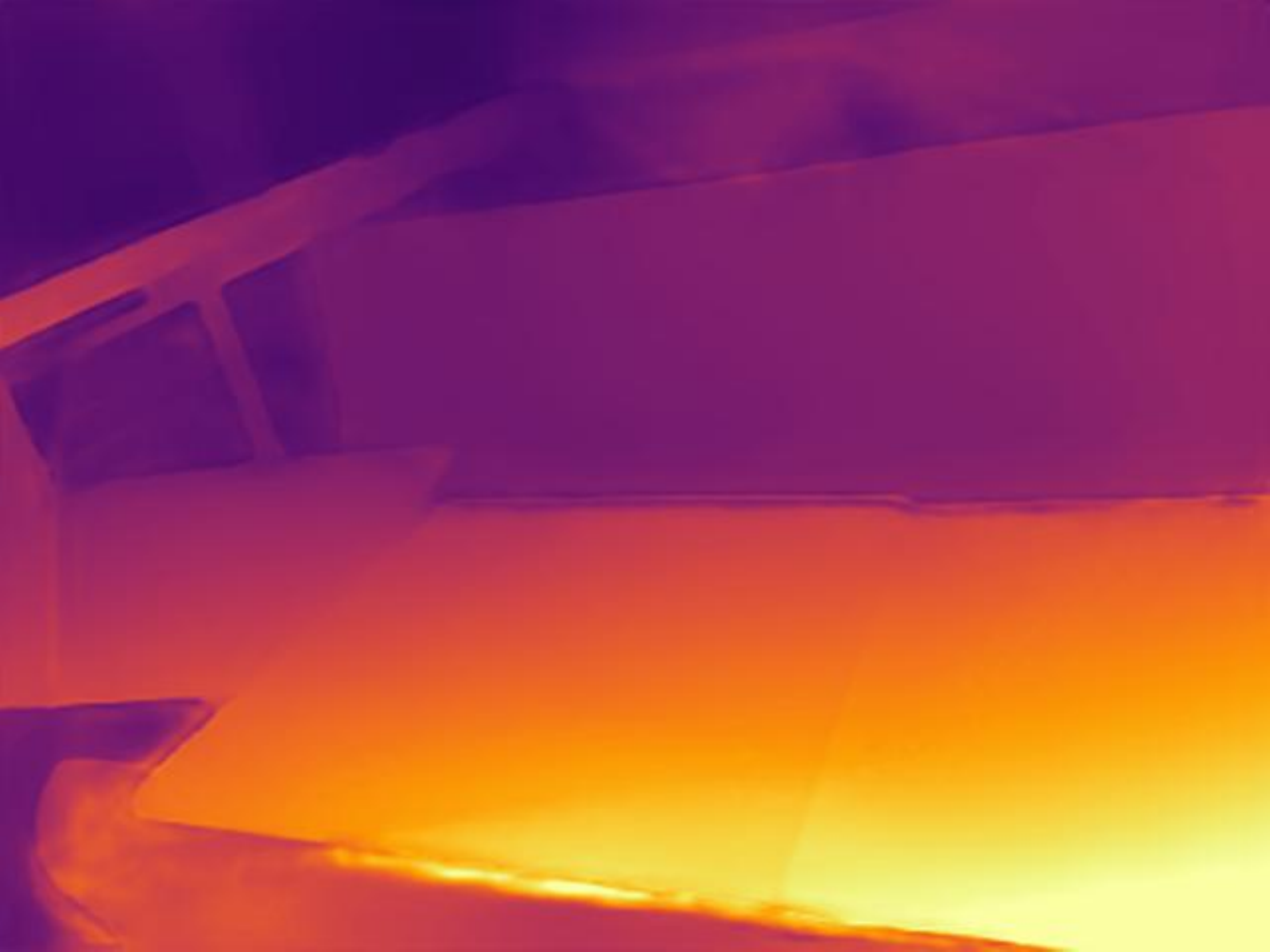}&
    \includegraphics[width=0.096\linewidth]{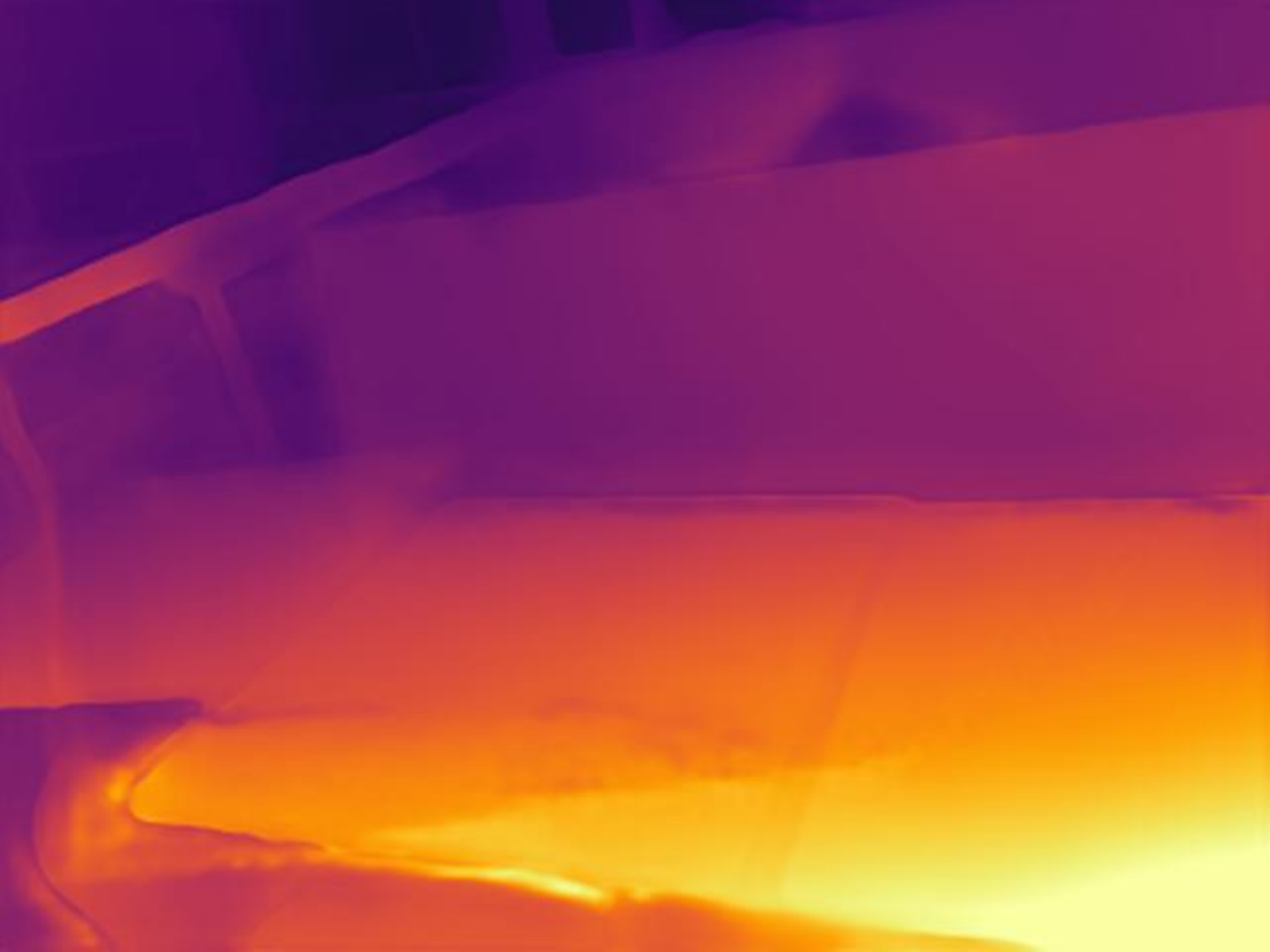}&
    \includegraphics[width=0.096\linewidth]{suppl/comparison_void_150/dpt-hybrid/row_9_col_5.pdf}&
    \includegraphics[width=0.096\linewidth]{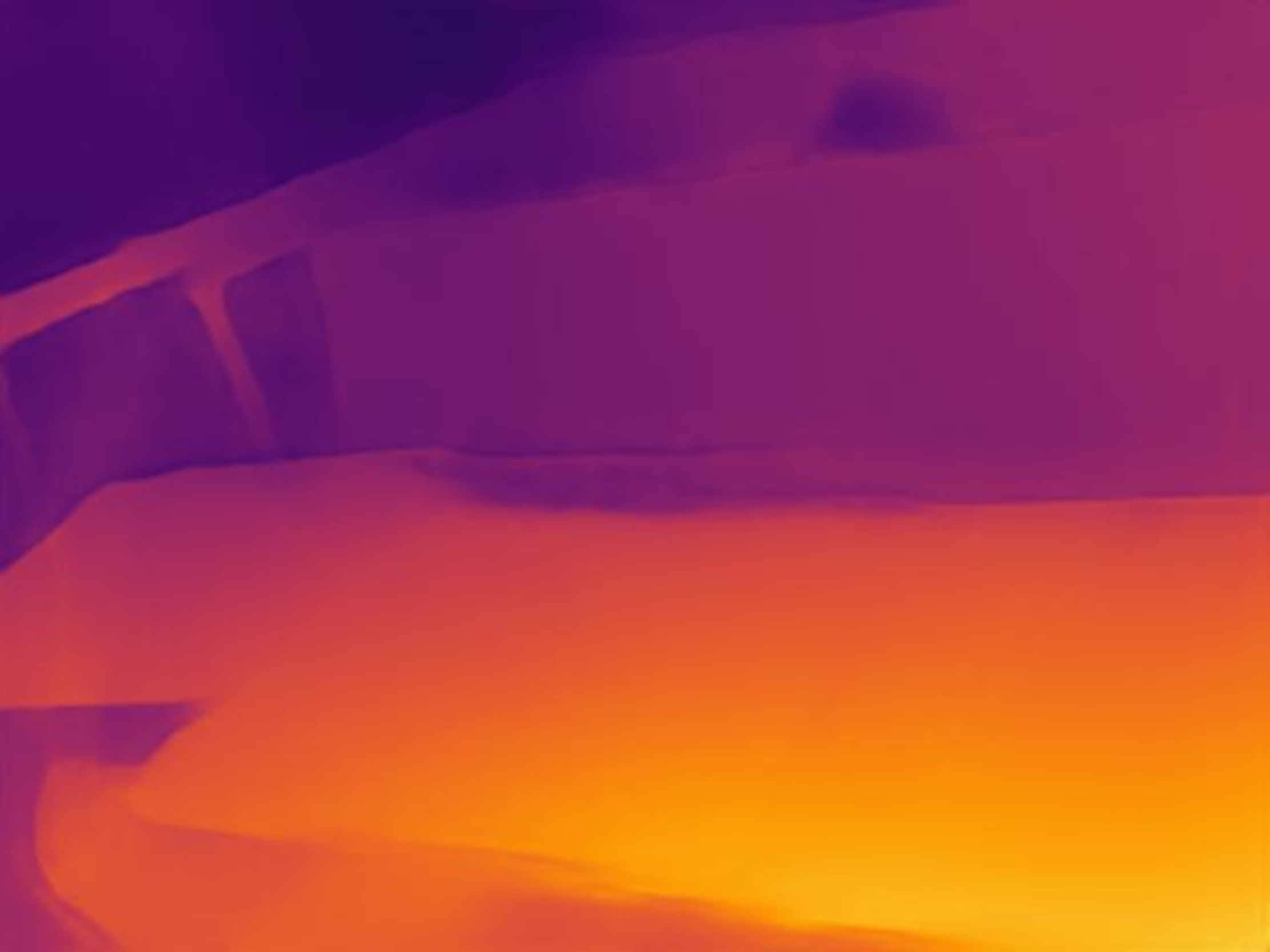}&
    \includegraphics[width=0.096\linewidth]{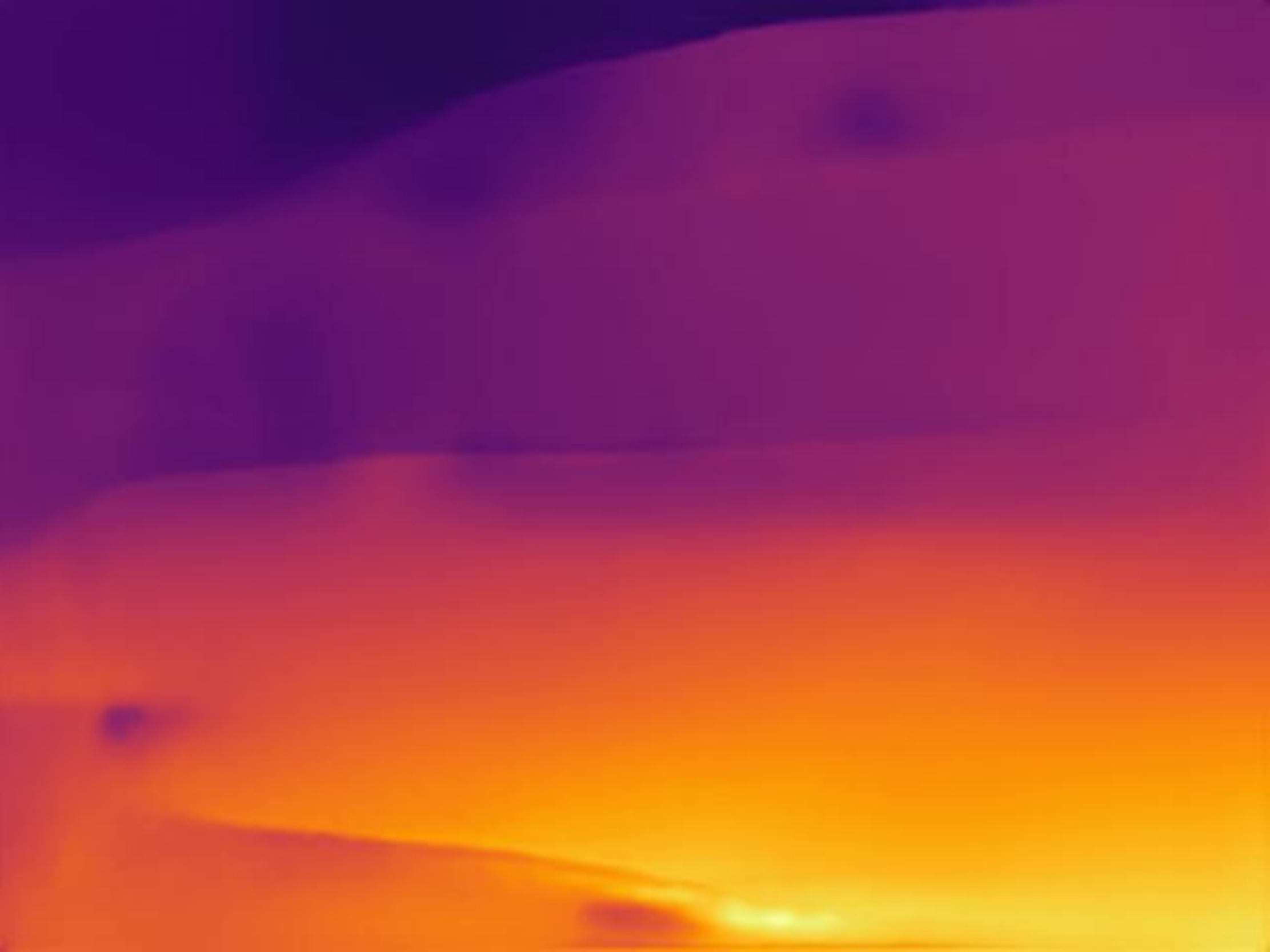}&
    \includegraphics[width=0.096\linewidth]{suppl/comparison_void_150/midas-small/row_9_col_5.pdf}\\
    \vspace{-0.75mm}
    \scriptsize k. &
    \includegraphics[width=0.096\linewidth]{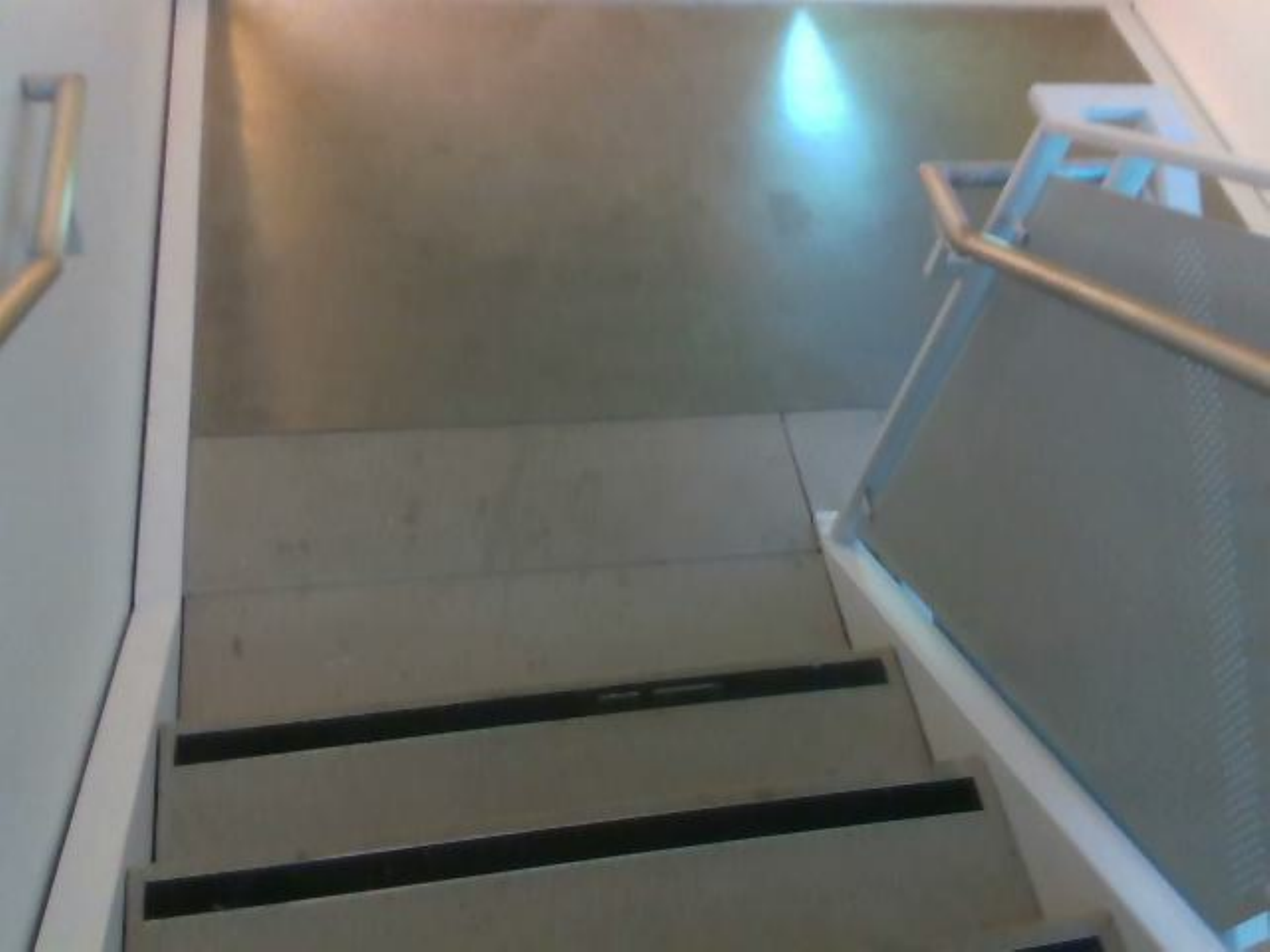}&
    \includegraphics[width=0.096\linewidth]{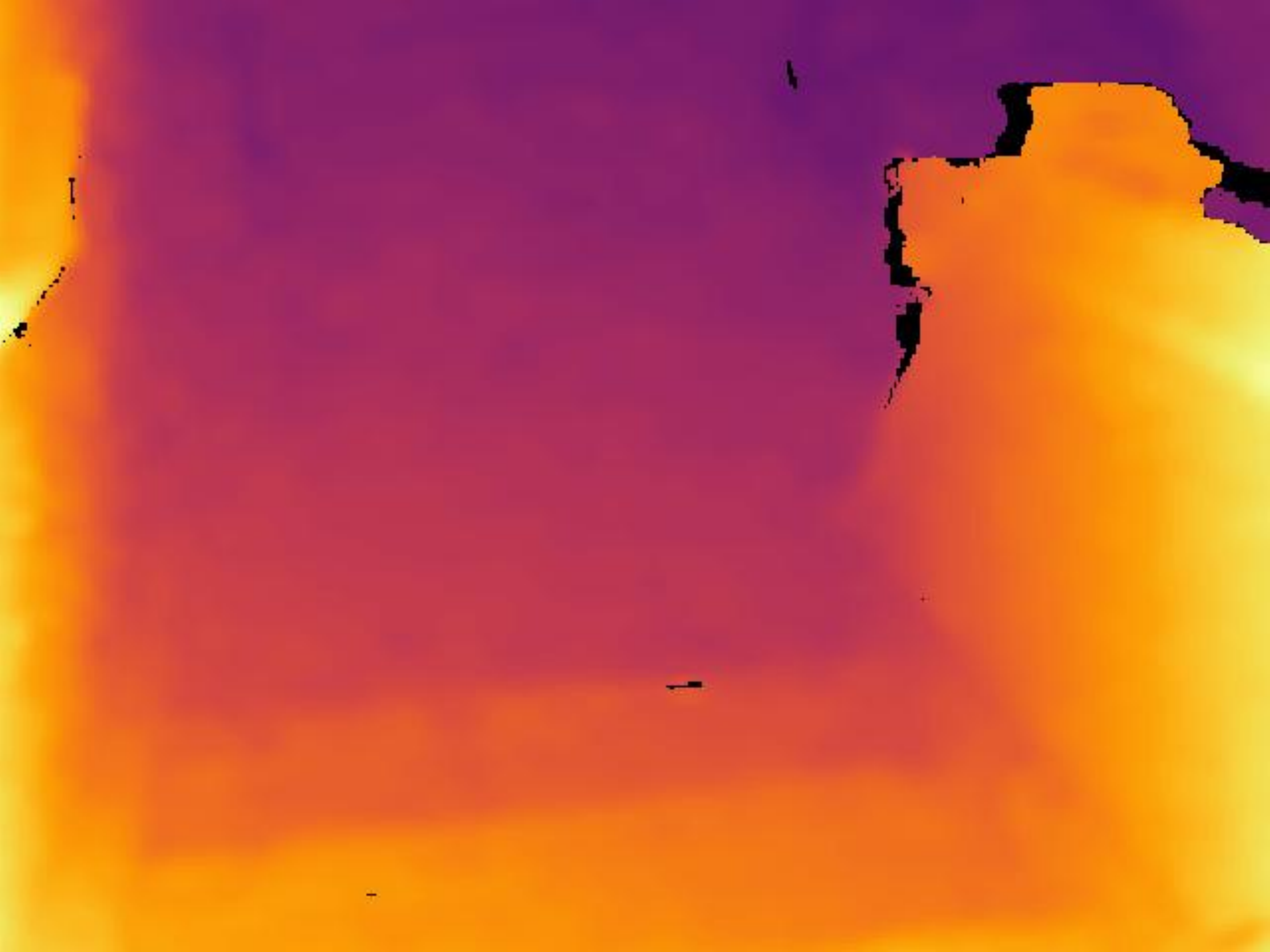}&
    \includegraphics[width=0.096\linewidth]{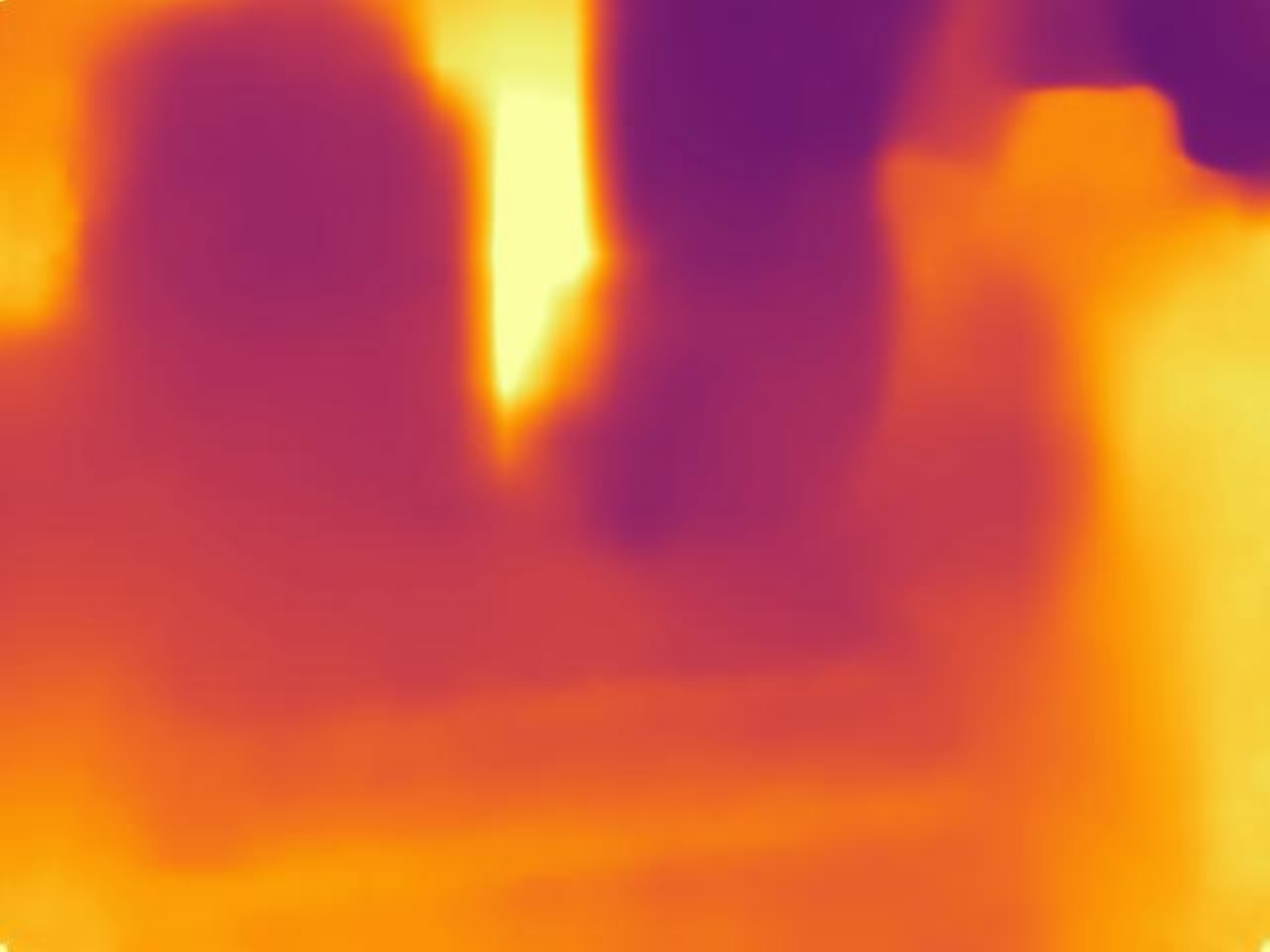}&
    \includegraphics[width=0.096\linewidth]{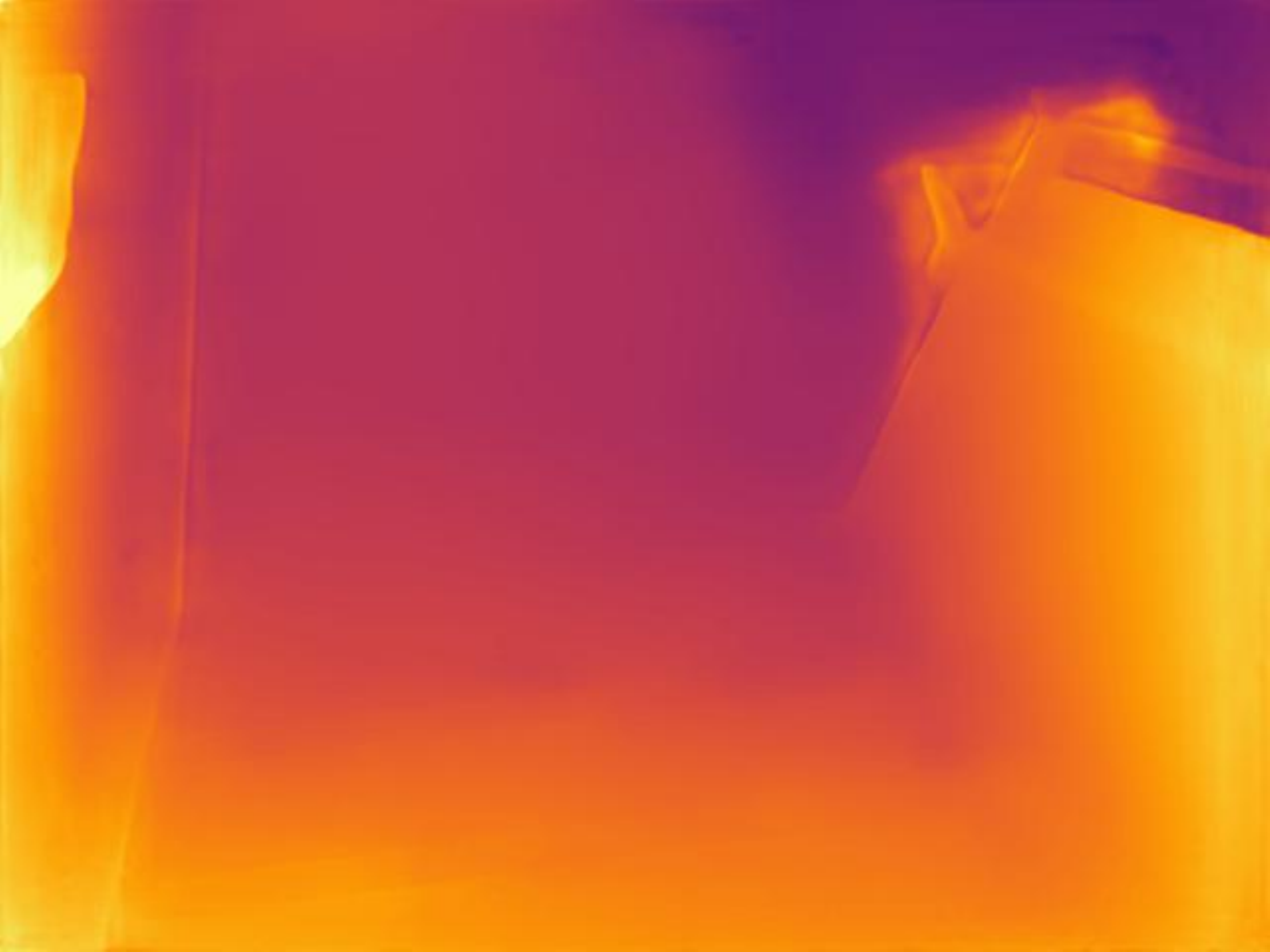}&
    \includegraphics[width=0.096\linewidth]{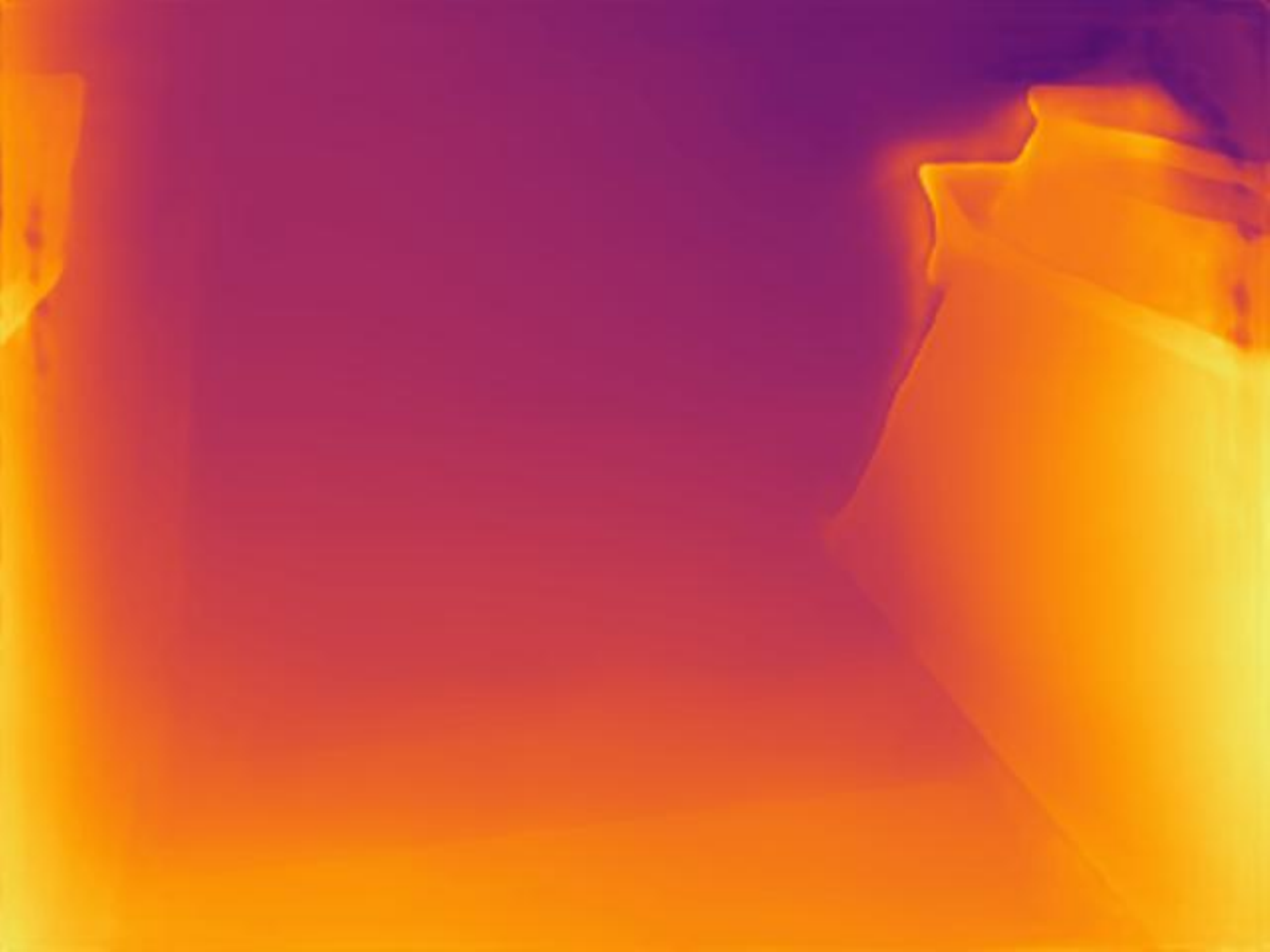}&
    \includegraphics[width=0.096\linewidth]{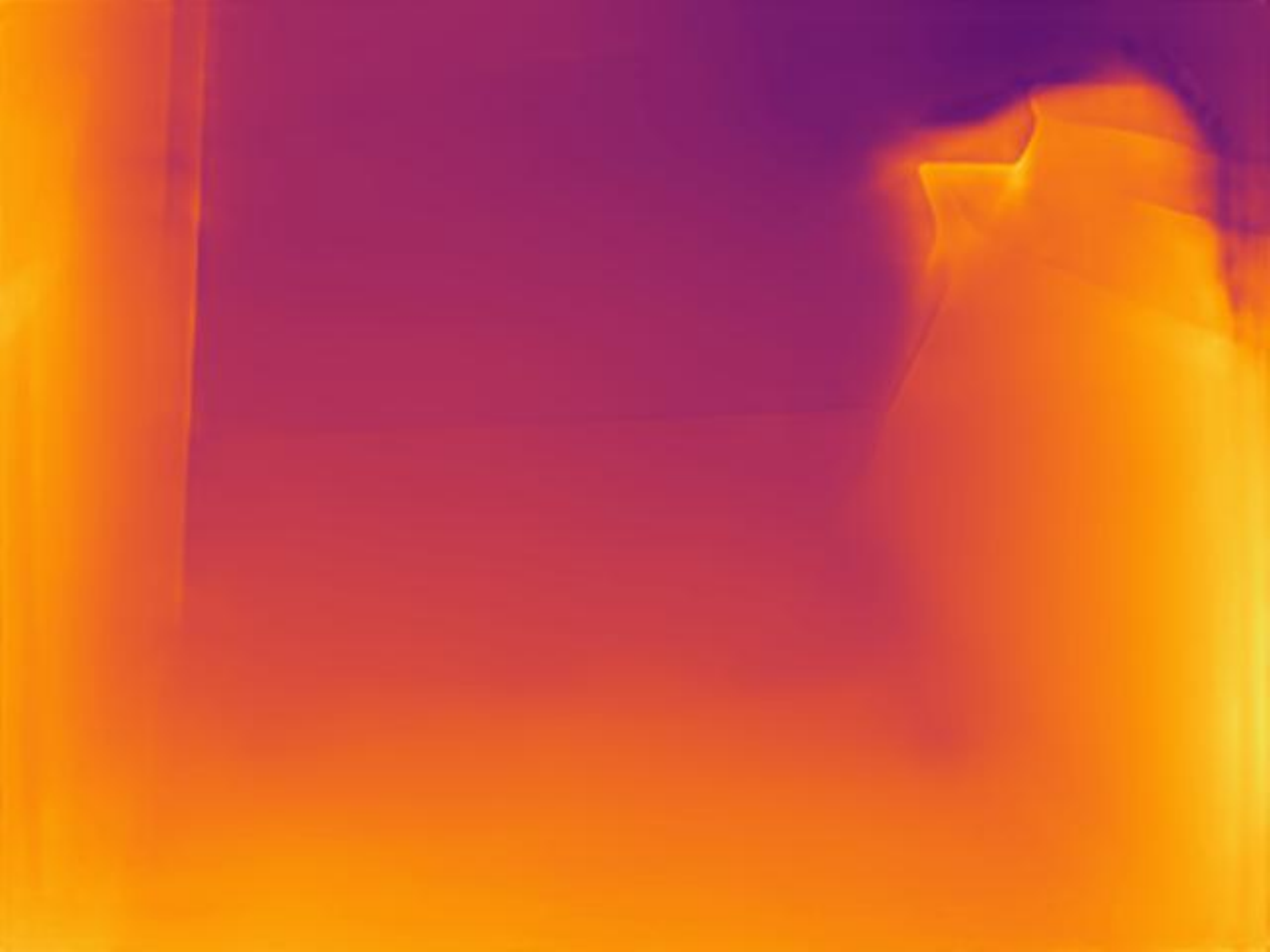}&
    \includegraphics[width=0.096\linewidth]{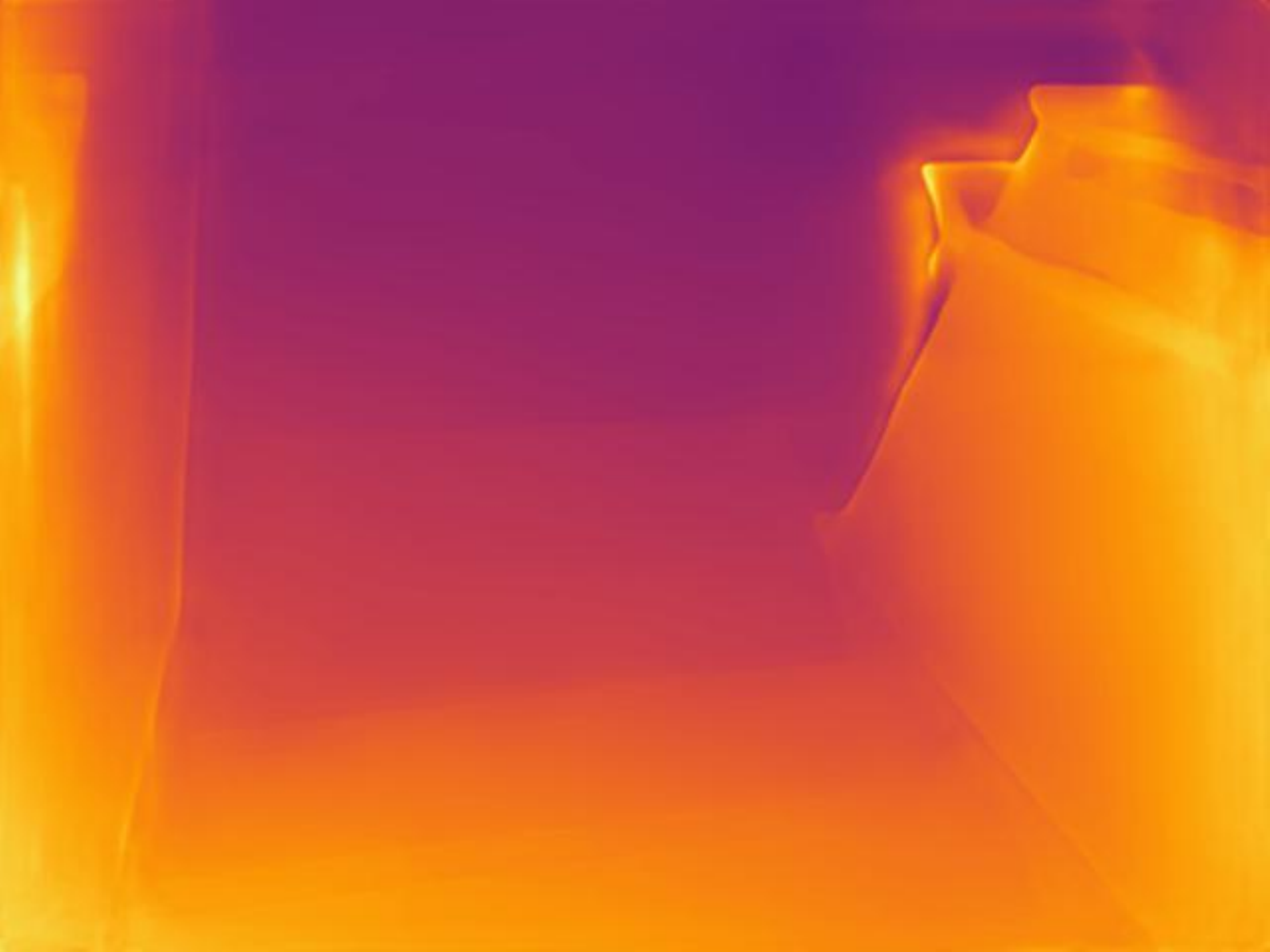}&
    \includegraphics[width=0.096\linewidth]{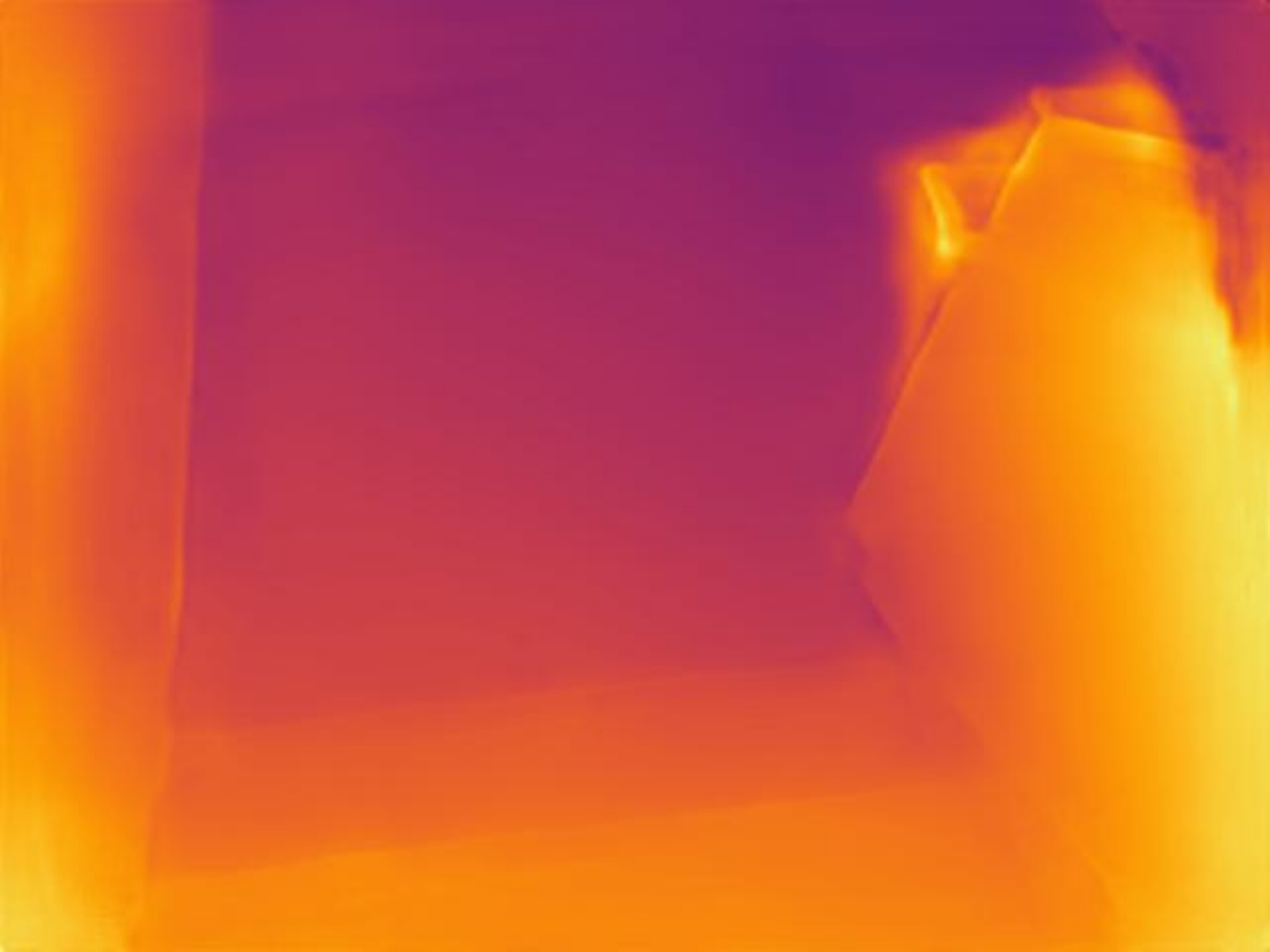}&
    \includegraphics[width=0.096\linewidth]{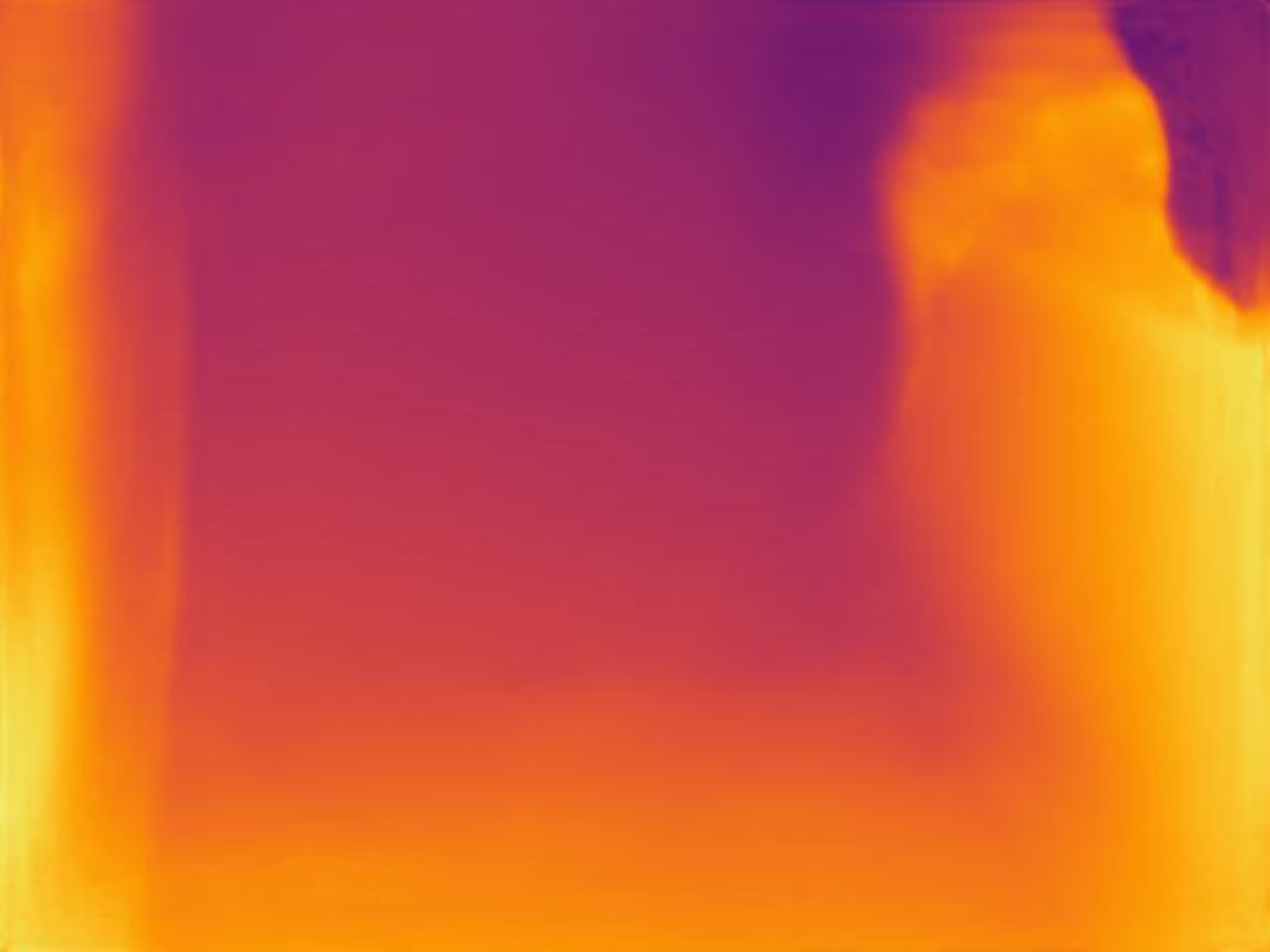}&
    \includegraphics[width=0.096\linewidth]{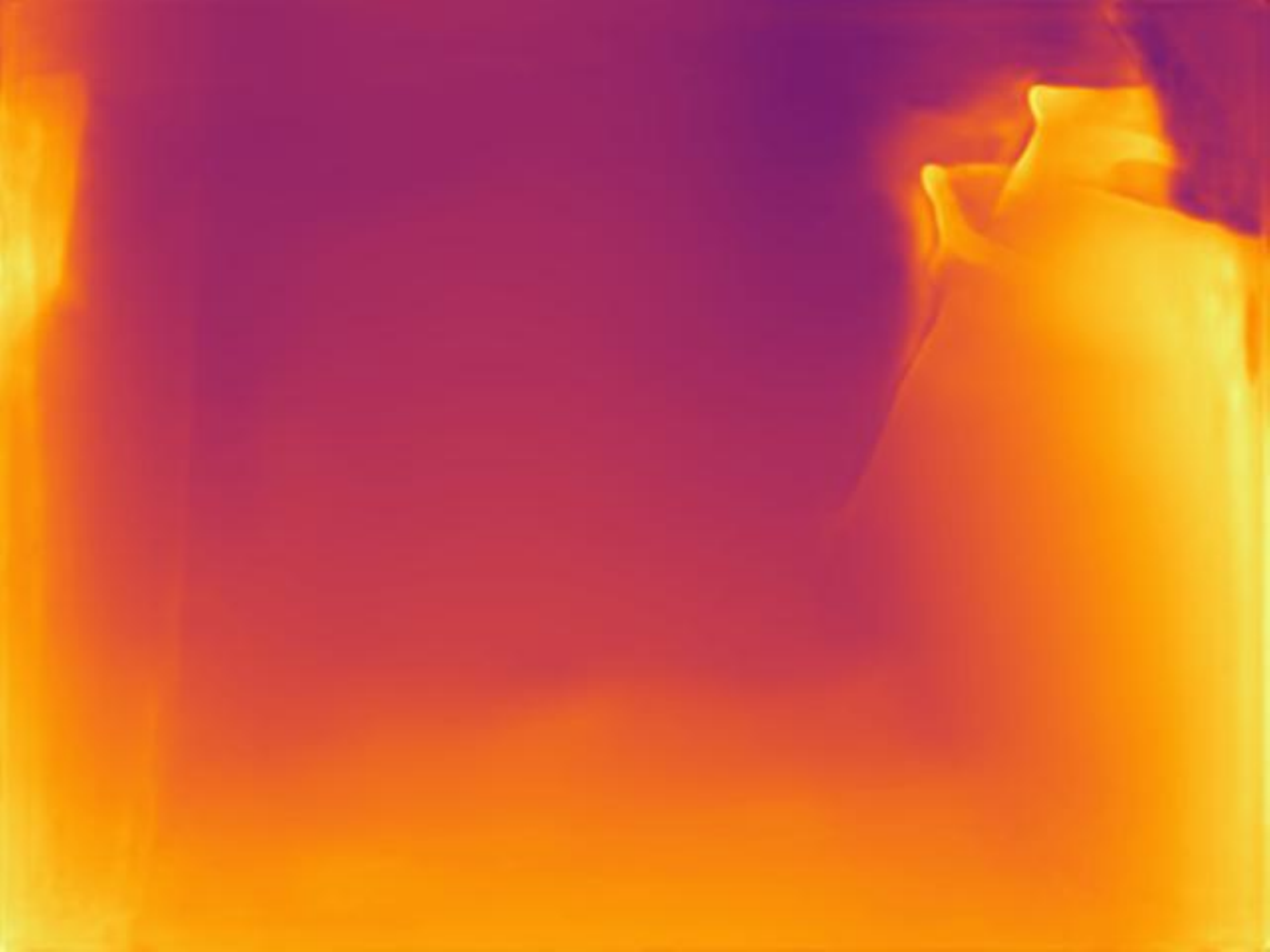}\\
    \vspace{-0.75mm}
  \end{tabular}
  \caption{Extension of Figure~\ref{fig:vis-void-expanded} to visualize depth output by our method GA+SML using different depth estimators from the MiDaS/DPT family of models. Prior state-of-the-art KBNet is also shown as a baseline. Our approach produces depth maps with improved sharpness and planar consistency.}
  \label{fig:vis-void-comparison-depth}
\end{figure*}

\mypara{Different depth estimators.} Our modular pipeline is agnostic to which monocular depth estimation model is used and can therefore benefit from continual improvement in these models. To demonstrate this, we substitute into our pipeline a variety of models from the MiDaS/DPT family of depth estimators. Table~\ref{tab:appendix_void_comparison} reports a more comprehensive comparison of our method GA+SML against related works VOICED~\cite{Wong2020void} and KBNet~\cite{Wong2021kbnet} on the VOID dataset. The two reported flavors of VOICED refer to whether pose used in that method is obtained via PoseNet (-P) or via SLAM (-S). It is concluded in \cite{Wong2020void} that leveraging pose from a visual-inertial SLAM system instead of a pose network leads to improved metrics. We note this out of relevance, since our approach too leverages information from VIO to improve metrics. Figure~\ref{fig:vis-void-comparison-depth} visualizes depth output by our pipeline with different depth estimators and compares against depth output by prior state-of-the-art KBNet. With powerful depth models like DPT-BEiT-Large and DPT-SwinV2-Large, our method achieves superior sharpness and planar consistency in metric depth maps.

\begin{table}
  \centering
  \footnotesize
  \caption{More comprehensive comparison on VOID. Lower is better for all metrics. All models trained directly on VOID.}
  \label{tab:appendix_void_comparison}
  \begin{tabular}{@{}
  l@{\hspace{1mm}}|
  l@{\hspace{2mm}}
  S[table-format=3.2]@{\hspace{2mm}} 
  S[table-format=3.2]@{\hspace{2mm}} 
  S[table-format=2.2]@{\hspace{2mm}} 
  S[table-format=3.2]@{\hspace{0mm}} @{}}
    \toprule
    & Method & {MAE} & {RMSE} & {iMAE} & {iRMSE} \\
    \midrule
    \multirow{10}{*}{\rot{150 points}} 
    & VOICED-P \cite{Wong2020void}      & 179.66 & 281.09 & 95.27 & 151.66 \\
    & VOICED-S \cite{Wong2020void}      & 174.04 & 253.14 & 87.39 & 126.30 \\
    & KBNet \cite{Wong2021kbnet}        & 131.54 & 263.54 & 66.84 & 128.29 \\
    & GA+SML (DPT-BEiT-Large)           & \bfseries 76.95 & \bfseries 142.85 & \bfseries 34.25 & \bfseries 57.13 \\
    & GA+SML (DPT-SwinV2-Large)         & \myuline{81.79} & \myuline{148.74} & \myuline{37.65} &  \myuline{61.11} \\
    & GA+SML (DPT-Large)                &  92.44 & 163.97 & 43.95 &  71.04 \\
    & GA+SML (DPT-Hybrid)               &  97.03 & 167.82 & 46.62 &  74.67 \\
    & GA+SML (DPT-SwinV2-Tiny)          & 100.25 & 174.16 & 47.15 &  75.10 \\
    & GA+SML (DPT-LeViT)                & 107.84 & 187.84 & 49.80 &  79.96 \\
    & GA+SML (MiDaS-small)              & 113.27 & 193.38 & 53.86 &  84.82 \\
    \midrule
    \multirow{10}{*}{\rot{500 points}} 
    & VOICED-P \cite{Wong2020void}      & 124.11 & 217.43 & 66.95 & 121.23 \\
    & VOICED-S \cite{Wong2020void}      & 118.01 & 195.32 & 59.29 & 101.72 \\
    & KBNet \cite{Wong2021kbnet}        &  77.70 & 172.49 & 38.87 &  85.59 \\
    & GA+SML (DPT-BEiT-Large)           & \bfseries 66.14 & \bfseries 126.44 & \bfseries 28.92 & \bfseries 49.85 \\
    & GA+SML (DPT-SwinV2-Large)         & \myuline{72.19} & \myuline{133.29} & \myuline{32.67} &  \myuline{54.49} \\
    & GA+SML (DPT-Large)                &  80.33 & 143.77 & 37.75 &  61.53 \\
    & GA+SML (DPT-Hybrid)               &  81.30 & 146.16 & 37.35 &  60.92 \\
    & GA+SML (DPT-SwinV2-Tiny)          &  86.11 & 153.83 & 39.47 &  63.83 \\
    & GA+SML (DPT-LeViT)                &  91.02 & 158.52 & 41.28 &  66.75 \\
    & GA+SML (MiDaS-small)              &  94.81 & 164.36 & 43.19 &  69.25 \\
    \midrule
    \multirow{10}{*}{\rot{1500 points}} 
    & VOICED-P \cite{Wong2020void}      &  85.05 & 169.79 & 48.92 & 104.02 \\
    & VOICED-S \cite{Wong2020void}      &  73.14 & 146.40 & 42.55 &  93.16 \\
    & KBNet \cite{Wong2021kbnet}        &  \bfseries 39.80 &  \bfseries 95.86 & \bfseries 21.16 &  49.72 \\
    & GA+SML (DPT-BEiT-Large)           &  \myuline{53.28} & \myuline{105.56} & \myuline{22.87} &  \bfseries 40.56 \\
    & GA+SML (DPT-SwinV2-Large)         &  59.90 & 110.94 & 26.90 &  \myuline{44.02} \\
    & GA+SML (DPT-Large)                &  64.89 & 116.72 & 31.02 &  49.98 \\
    & GA+SML (DPT-Hybrid)               &  69.03 & 123.62 & 31.91 &  51.44 \\
    & GA+SML (DPT-SwinV2-Tiny)          &  72.61 & 126.95 & 33.98 &  53.66 \\
    & GA+SML (DPT-LeViT)                &  71.62 & 126.47 & 33.42 &  53.22 \\
    & GA+SML (MiDaS-small)              &  77.89 & 135.27 & 36.70 &  57.55 \\
    \bottomrule
  \end{tabular}
  \vspace{-12pt}
\end{table}

\begin{table*}[t]
  \centering
  \footnotesize
  \caption{Runtime [ms] on Jetson platforms in MAX-N mode. All pipeline variants are tested with 150 sparse metric depth points.}
  \label{tab:appendix_performance}
  \begin{tabular}{@{}
    l@{\hspace{4mm}}|
    S[table-format=3.1]@{\hspace{2mm}}
    S[table-format=3.1]@{\hspace{2mm}}
    S[table-format=3.1]@{\hspace{2mm}}
    S[table-format=3.1]@{\hspace{2mm}}
    S[table-format=3.1]@{\hspace{2mm}}
    S[table-format=3.1]@{\hspace{2mm}}
    S[table-format=3.1]@{\hspace{2mm}}
    S[table-format=3.1]@{\hspace{2mm}} | 
    S[table-format=3.1]@{\hspace{0mm}}
    @{}}
    \toprule
    & \multicolumn{8}{c|}{\textit{On Jetson AGX Orin}} & \textit{On Jetson TX2} \\
    \midrule
    Depth predictor & {DPT-BEiT-L} & {DPT-SwinV2-L} & {DPT-L} & {DPT-H} & {DPT-SwinV2-T} & {DPT-LeViT} & {MiDaS-s} & {MiDaS-s-TRT} & {MiDaS-s-TRT} \\
    Inference resolution & {384$\times$384} & {384$\times$384} & {384$\times$384} & {384$\times$384} & {256$\times$256} & {224$\times$224} & {256$\times$256} & {256$\times$256} & {256$\times$256} \\
    \midrule
    Depth inference         & 144.8  & 70.0   & 38.7   & 53.9   & 36.2   & 31.4   & 29.2   & 1.5   & 5.1    \\
    D2H copy depth map      & 12.8   & 123.7  & 56.4   & 18.4   & 3.8    & 0.2    & 0.6    & 5.2   & 24.9   \\
    Global alignment        & 2.6    & 2.8    & 2.8    & 2.5    & 1.3    & 1.1    & 1.3    & 1.3   & 3.7    \\
    Scale map scaffolding   & 12.2   & 13.8   & 12.7   & 12.1   & 6.6    & 5.4    & 6.7    & 6.6   & 49.1   \\
    H2D copy SML inputs     & 3.3    & 3.6    & 3.5    & 3.3    & 2.3    & 2.2    & 2.4    & 2.2   & 4.2    \\
    SML-TRT inference       & 2.2    & 2.4    & 2.2    & 2.2    & 1.7    & 1.9    & 1.7    & 1.7   & 5.4    \\
    \midrule
    Total [ms]              & 177.9  & 216.4  & 116.2  & 92.5   & 51.9   & 42.2   & 41.9   & 18.5  & 92.2   \\
    (fps)                   & 5.6    & 4.6    & 8.6    & 10.8   & 19.3   & 23.7   & 23.9   & 53.9  & 10.8   \\ 
    \bottomrule
  \end{tabular}
\end{table*}

\begin{figure*}[p]
\centering
  \begin{tabular}{@{}l@{\hspace{0.5mm}}*{9}{c@{\hspace{0.5mm}}}c@{}}
    & {\scriptsize RGB Image} & {\scriptsize Sparse Depth} & {\scriptsize Scales Scaffold.} & {\scriptsize Regressed Scales} & {\scriptsize GA Depth} & {\scriptsize SML Depth} & {\scriptsize Ground Truth} & {\scriptsize GA Error} & {\scriptsize SML Error}\\
    \vspace{-0.75mm}
    \rot{\scriptsize VOID 150} &
    \includegraphics[width=0.104\linewidth]{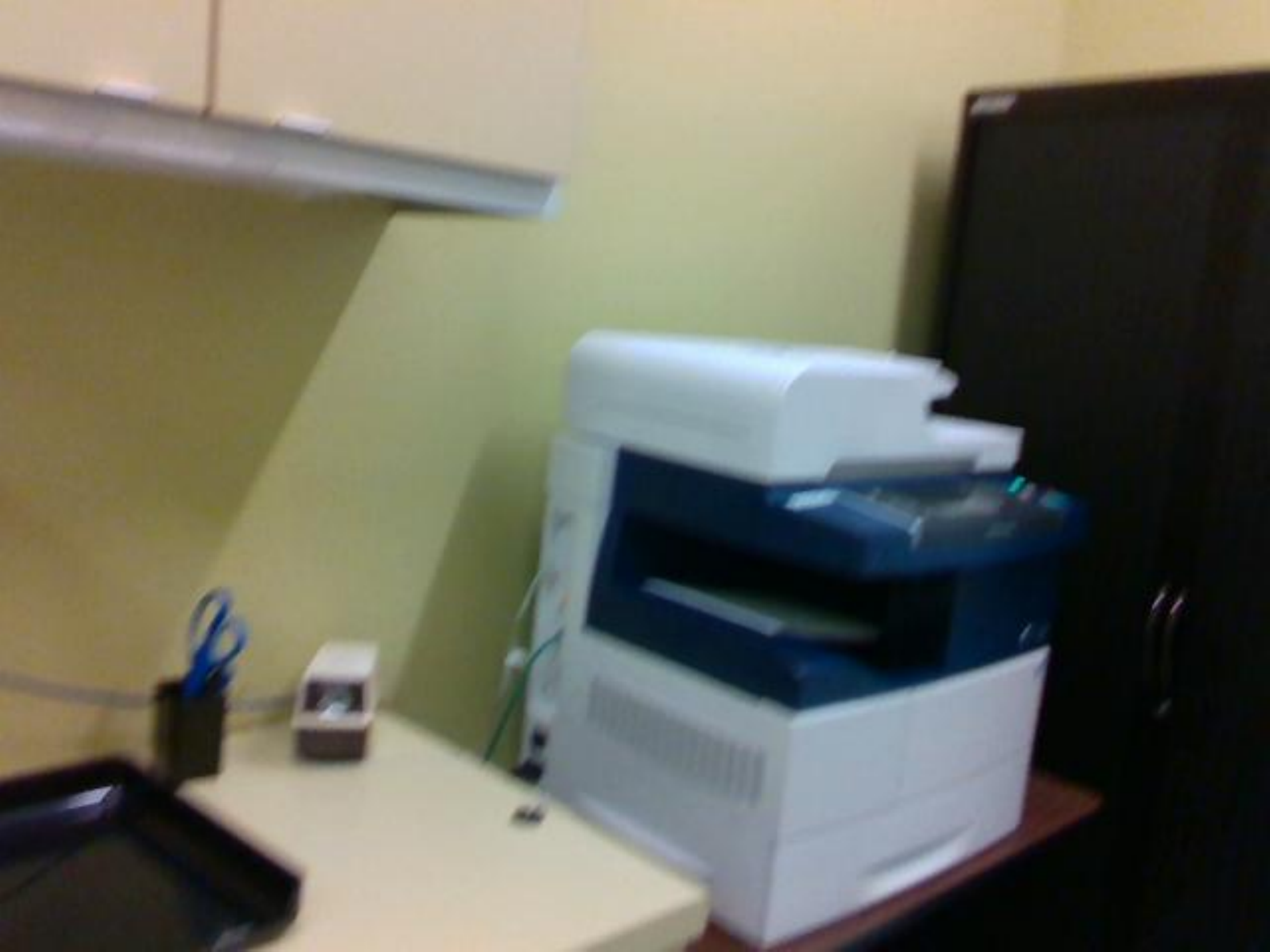}&
    \includegraphics[width=0.104\linewidth]{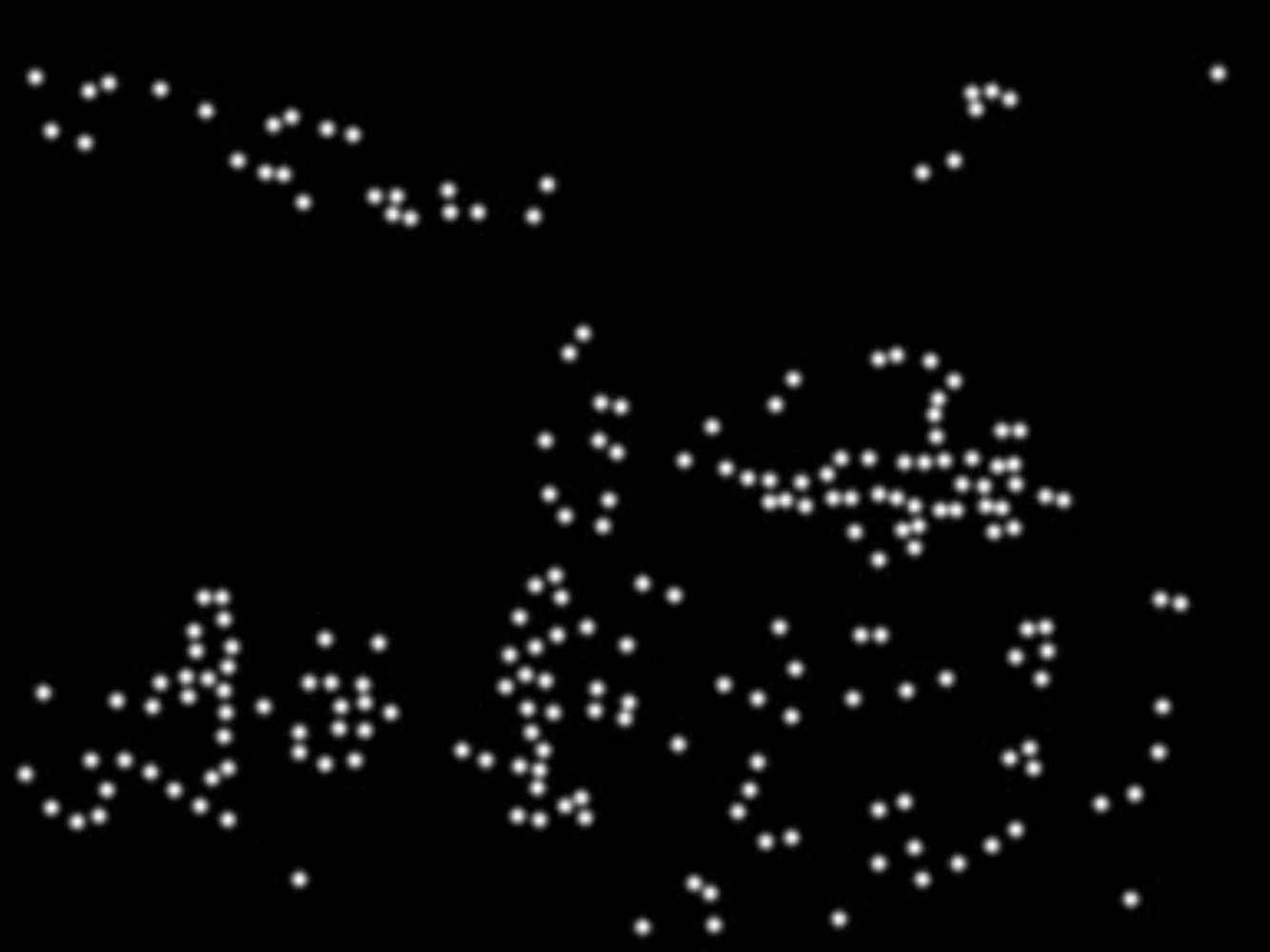}&
    \includegraphics[width=0.104\linewidth]{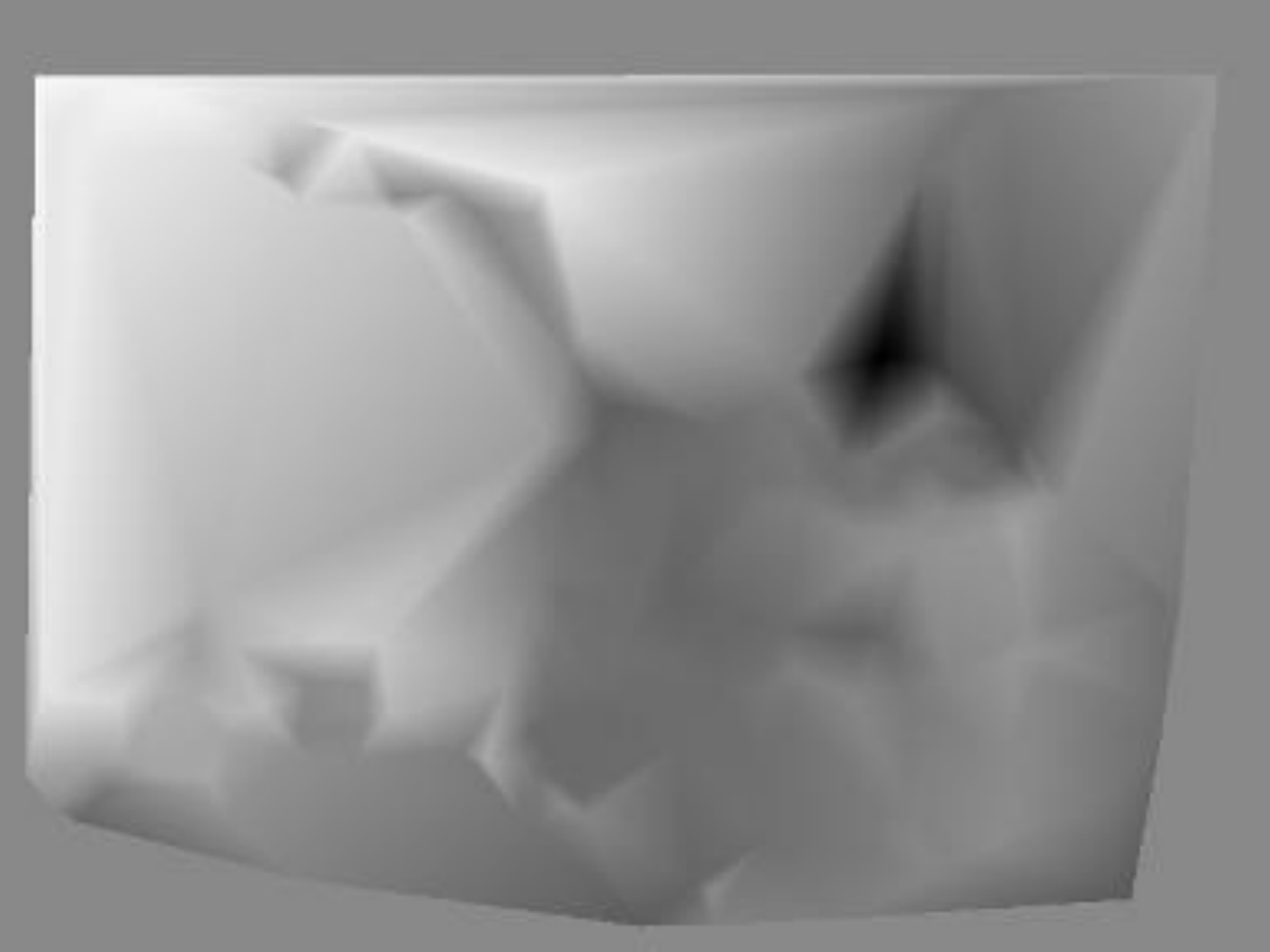}&
    \includegraphics[width=0.104\linewidth]{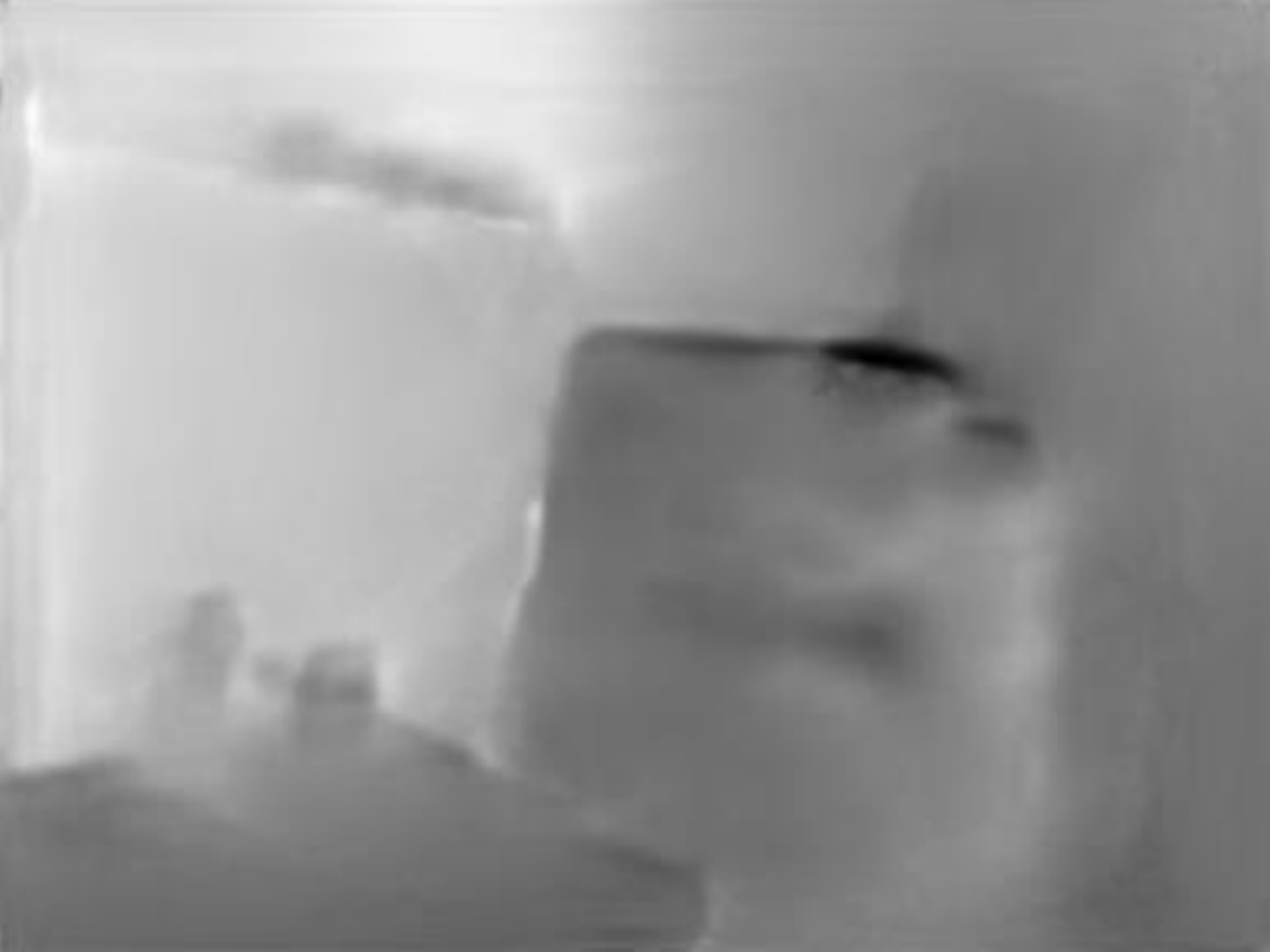}&
    \includegraphics[width=0.104\linewidth]{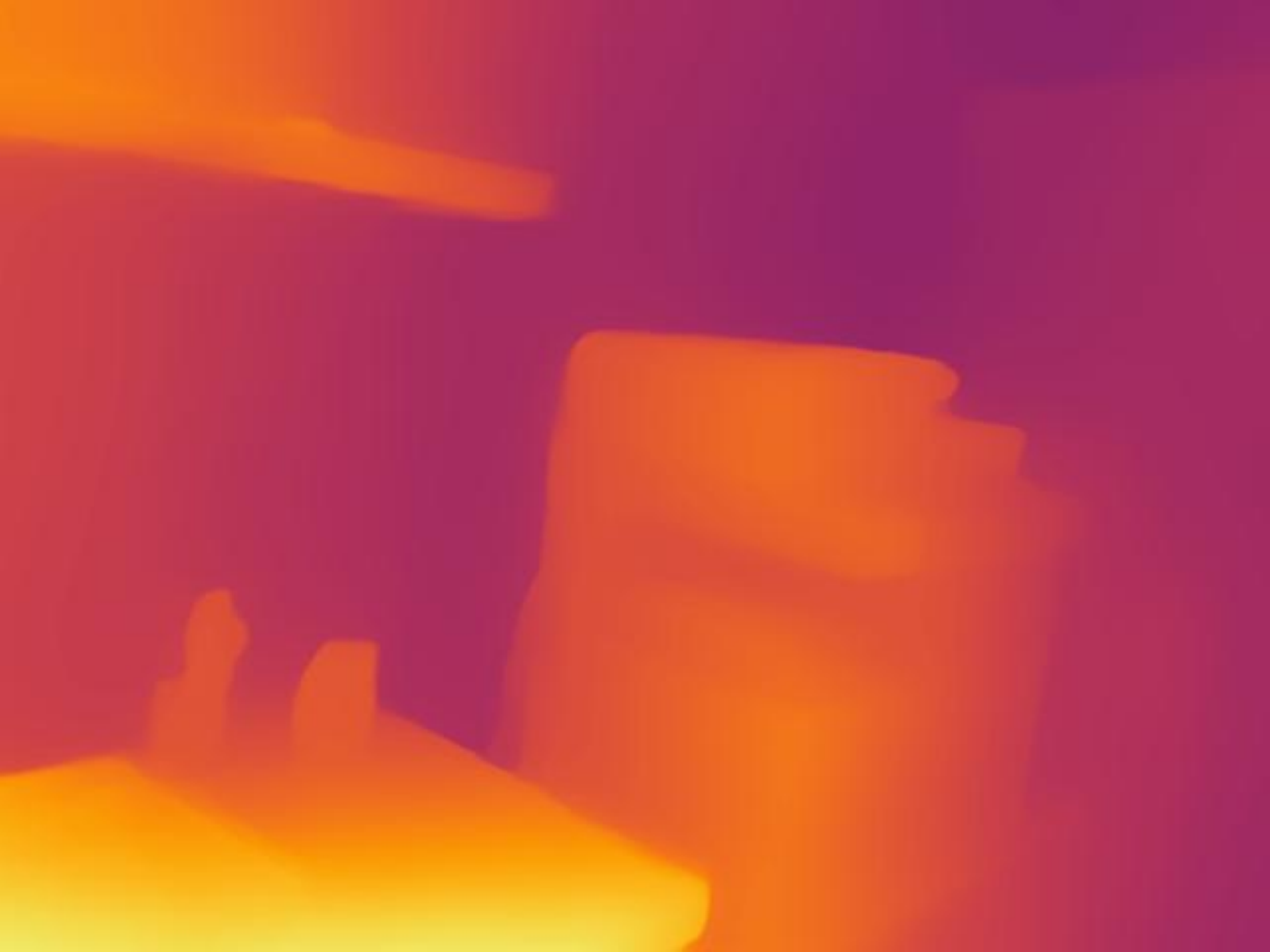}&
    \includegraphics[width=0.104\linewidth]{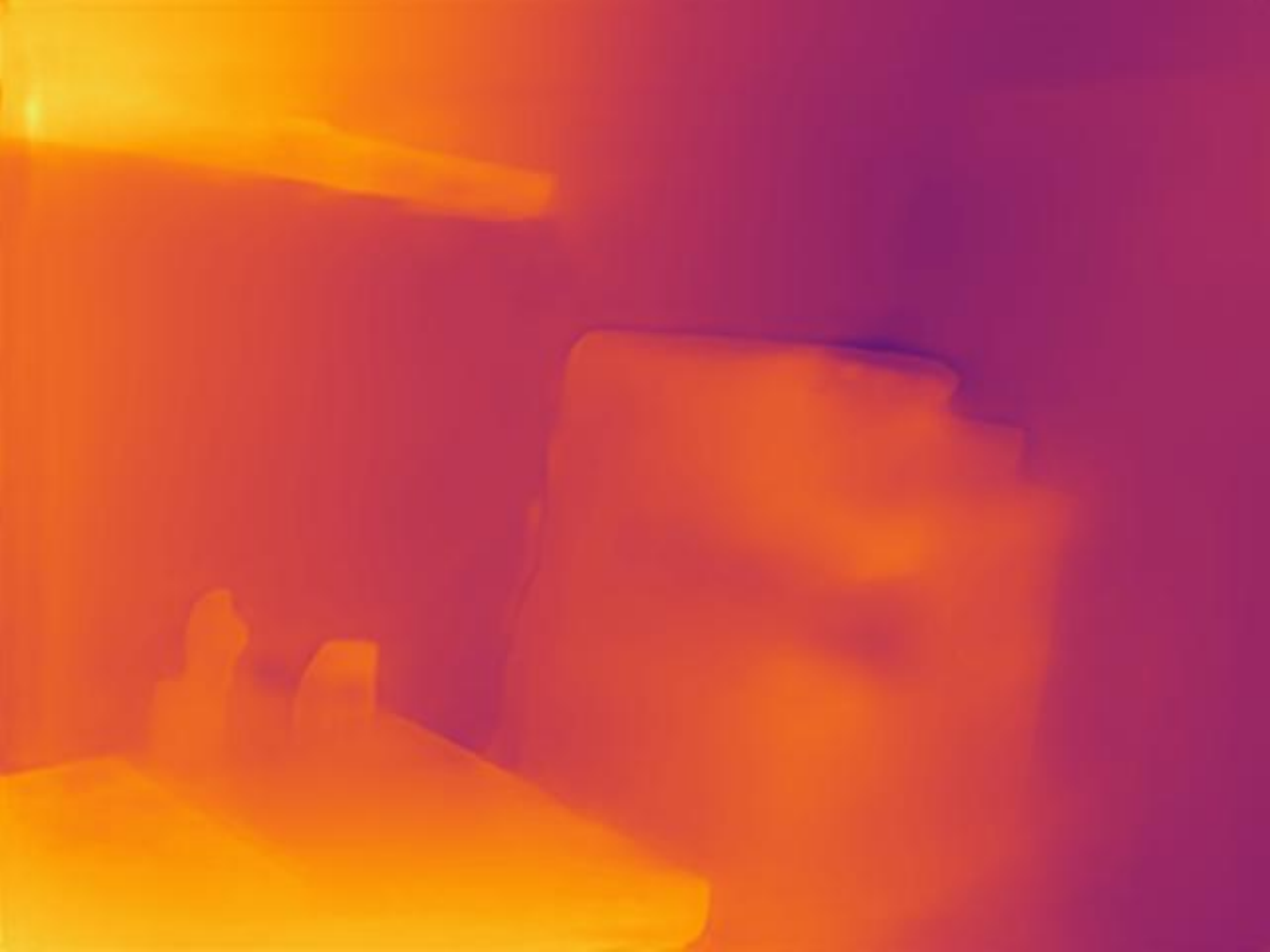}&
    \includegraphics[width=0.104\linewidth]{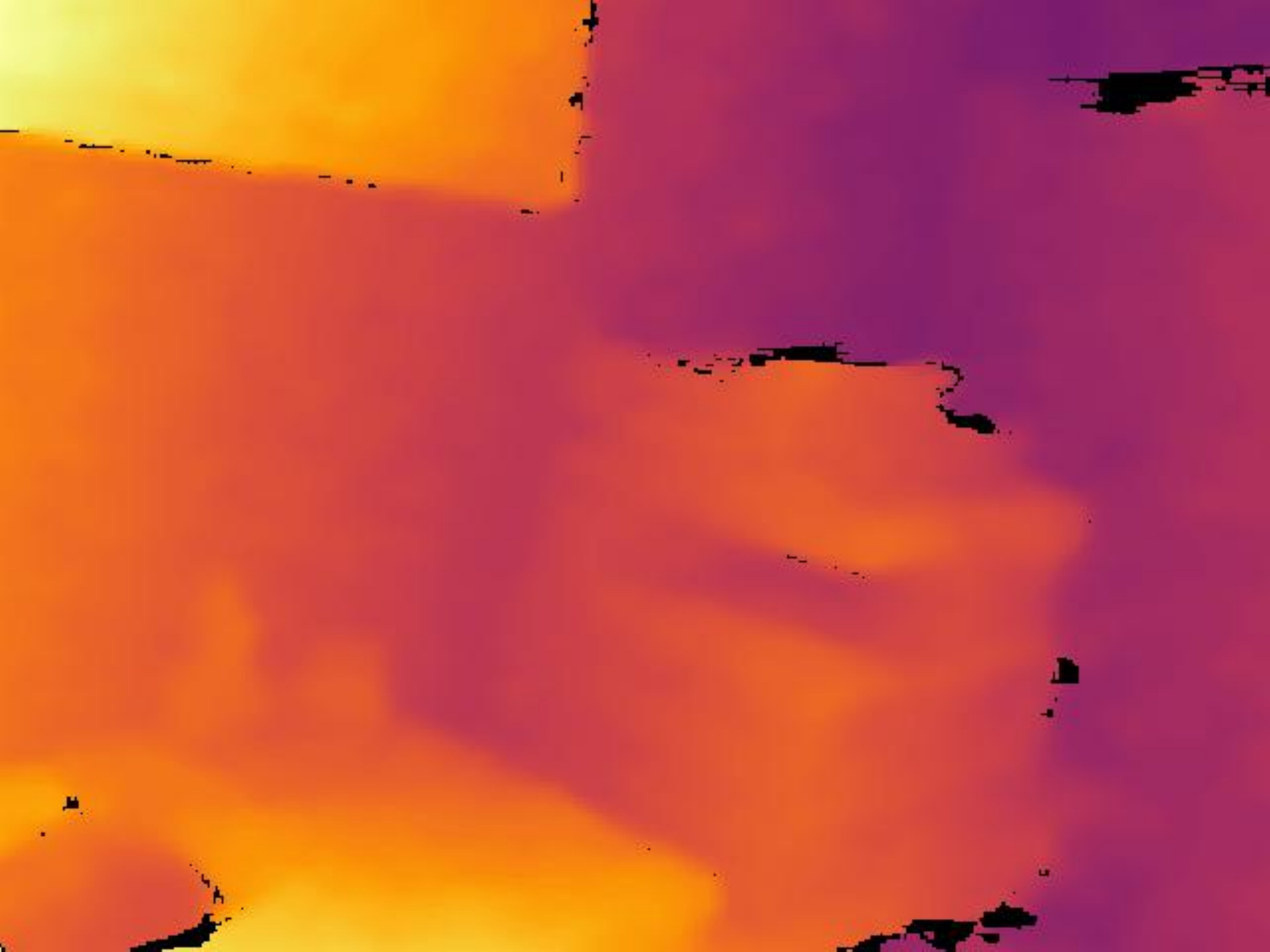}&
    \includegraphics[width=0.104\linewidth]{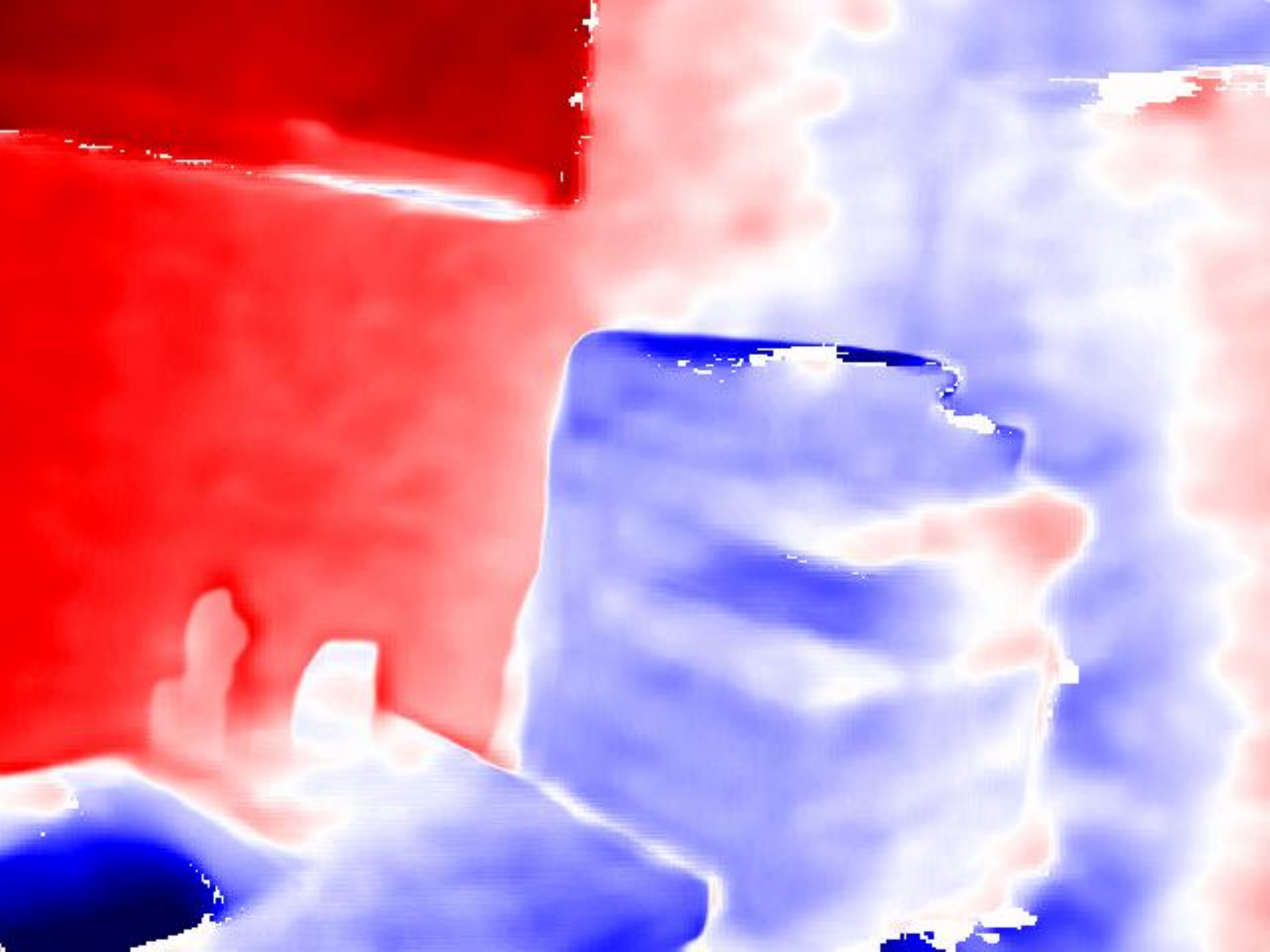}&
    \includegraphics[width=0.104\linewidth]{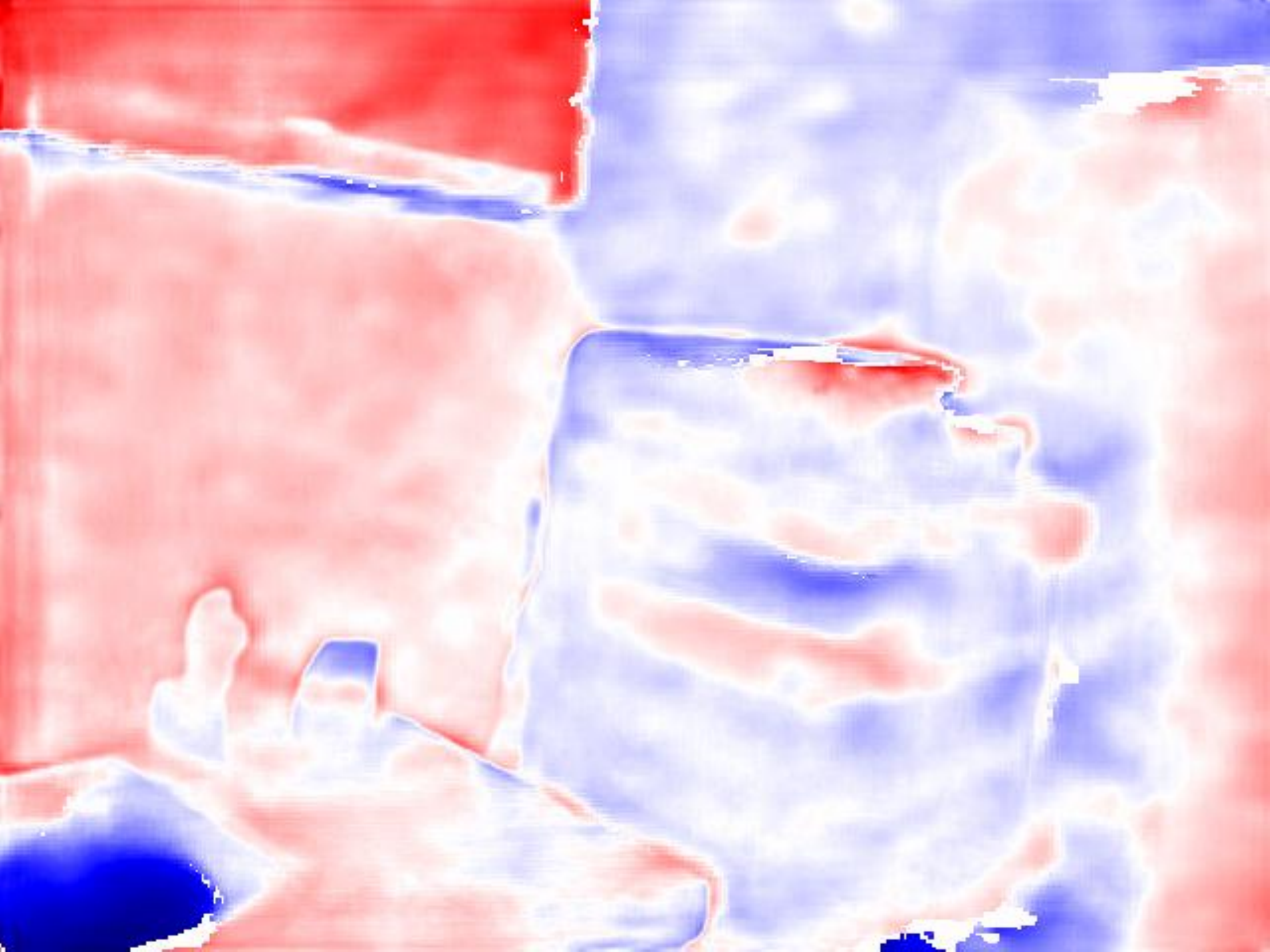}&
    \includegraphics[width=0.024\linewidth]{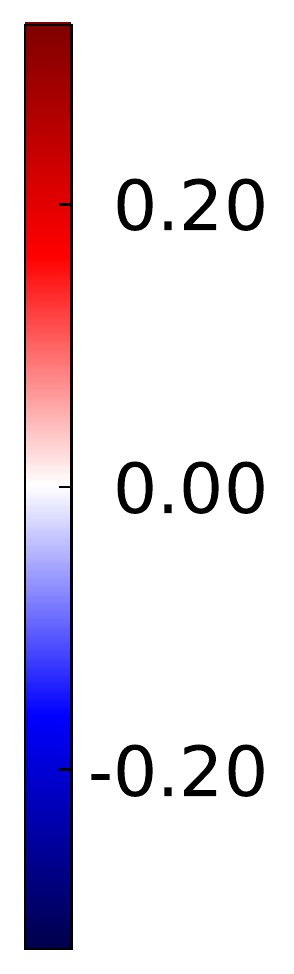}\\
    \vspace{-0.75mm}
    \rot{\scriptsize VOID 500} &
    \includegraphics[width=0.104\linewidth]{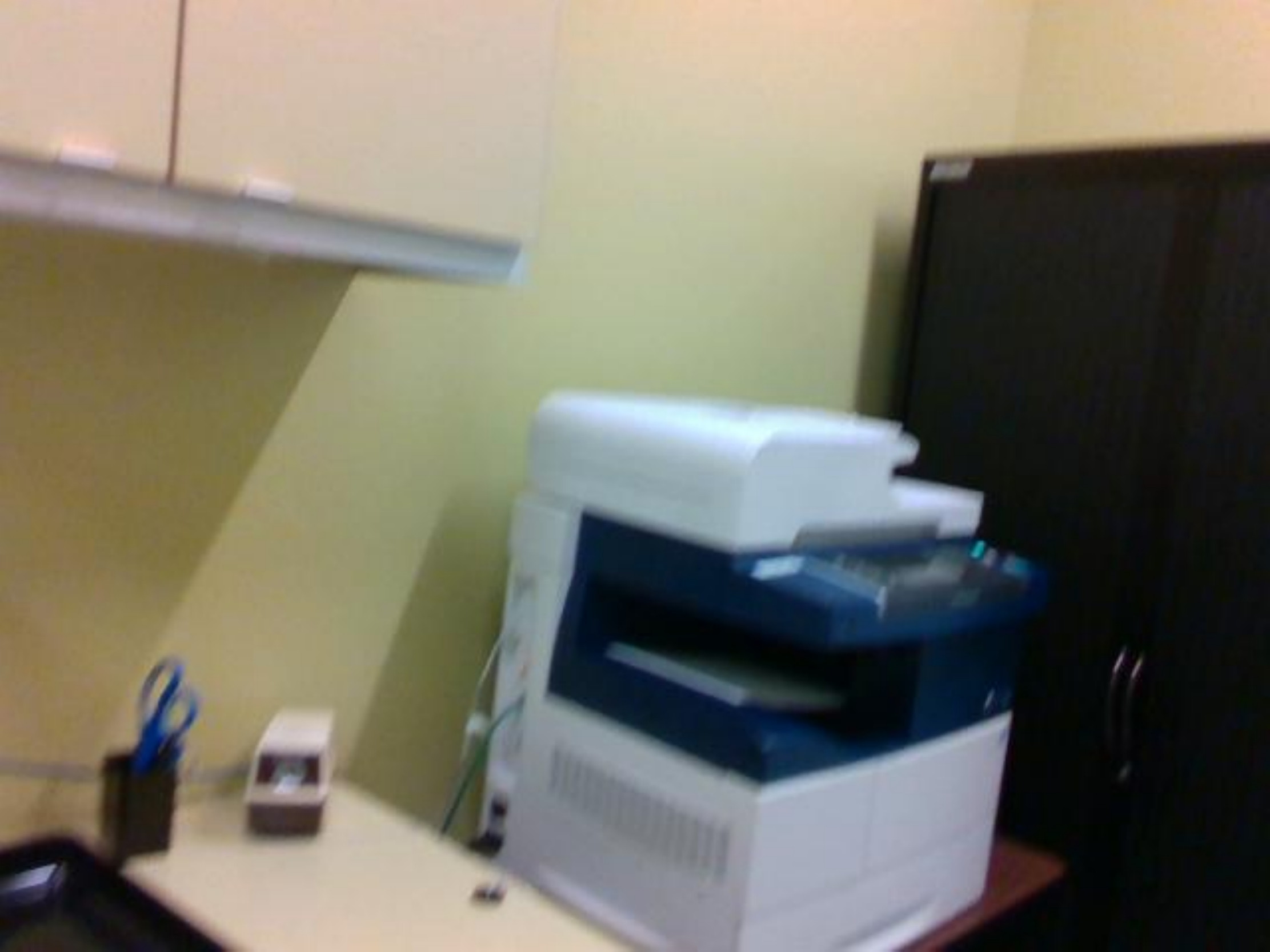}&
    \includegraphics[width=0.104\linewidth]{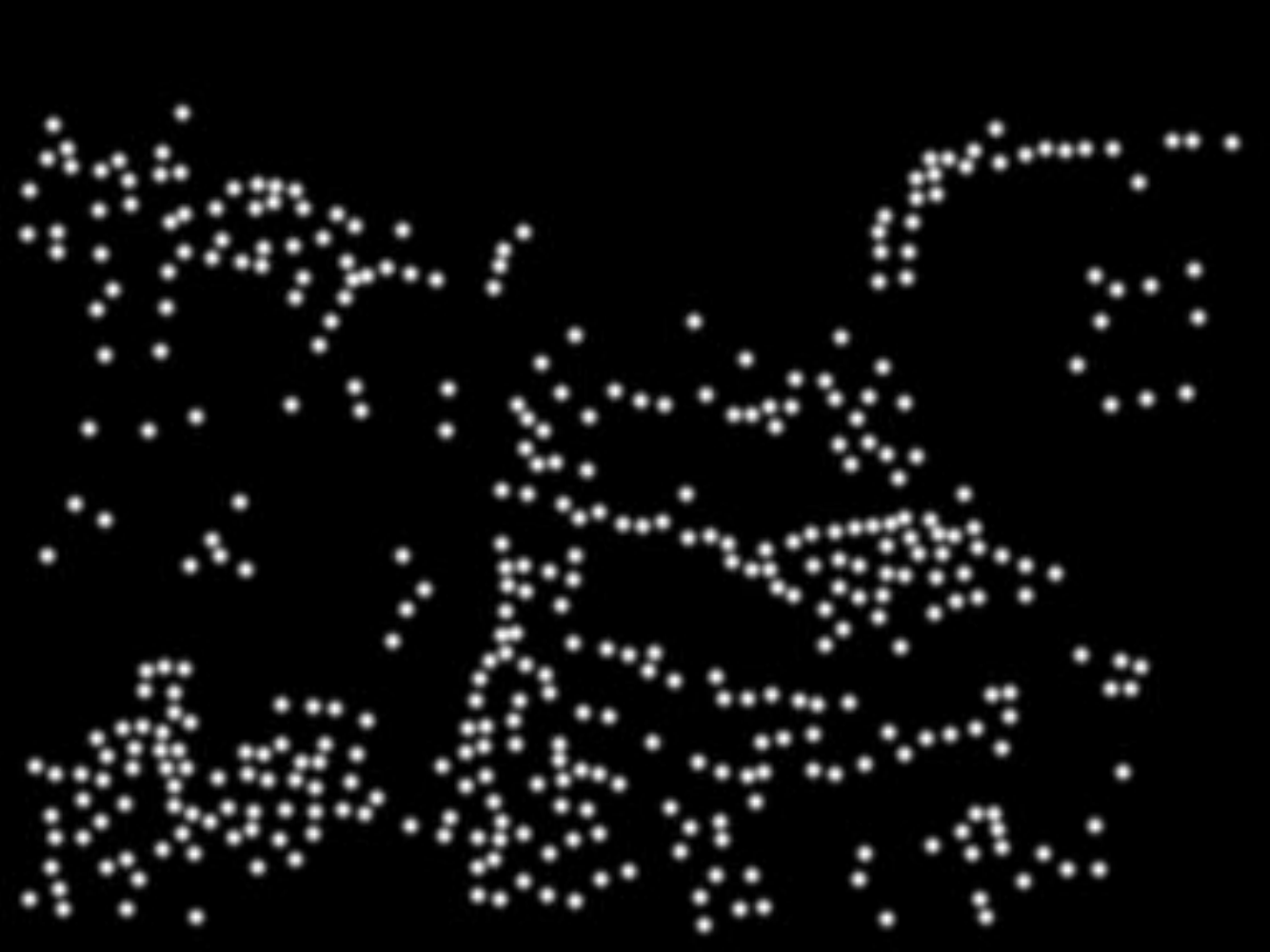}&
    \includegraphics[width=0.104\linewidth]{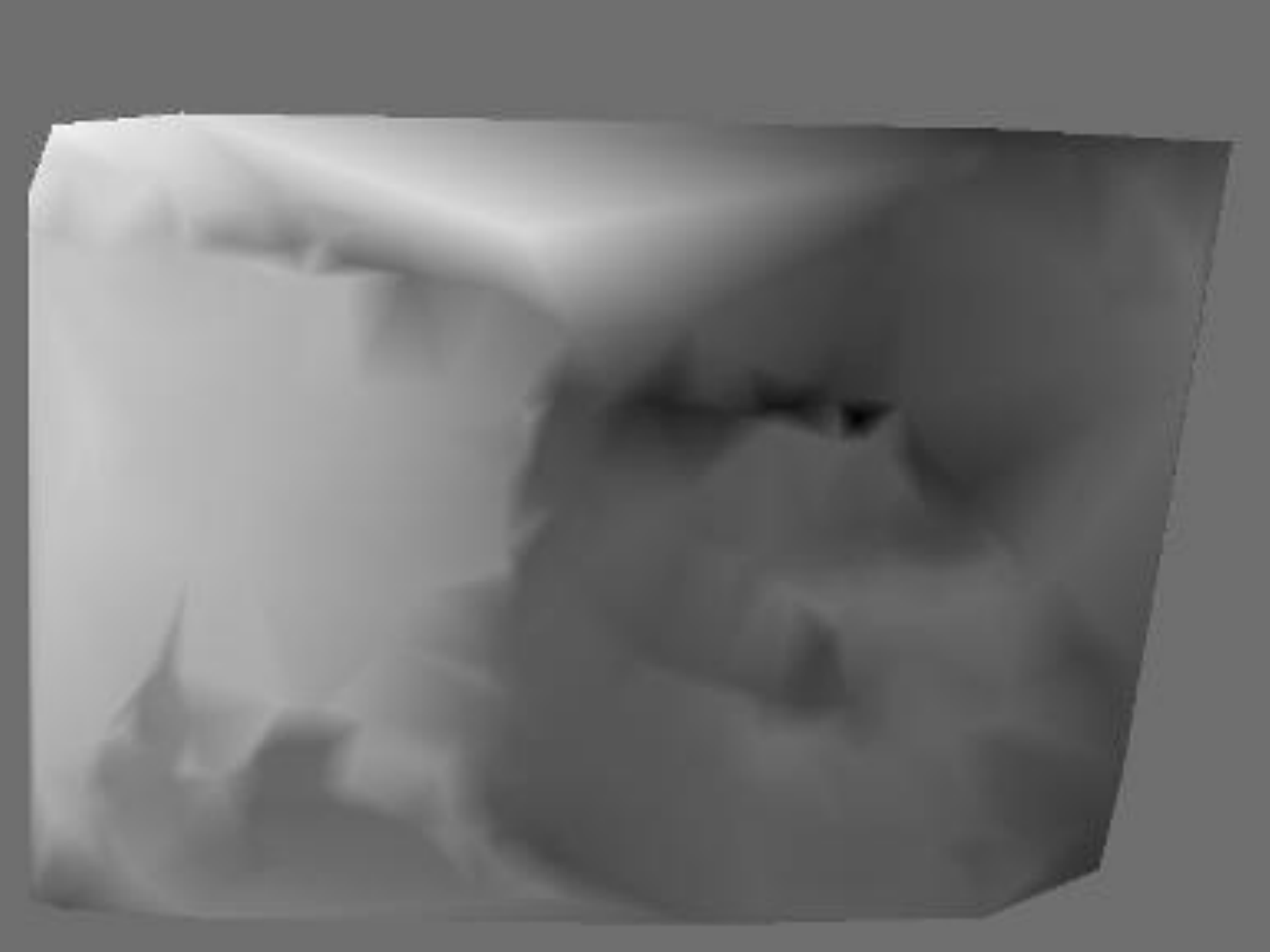}&
    \includegraphics[width=0.104\linewidth]{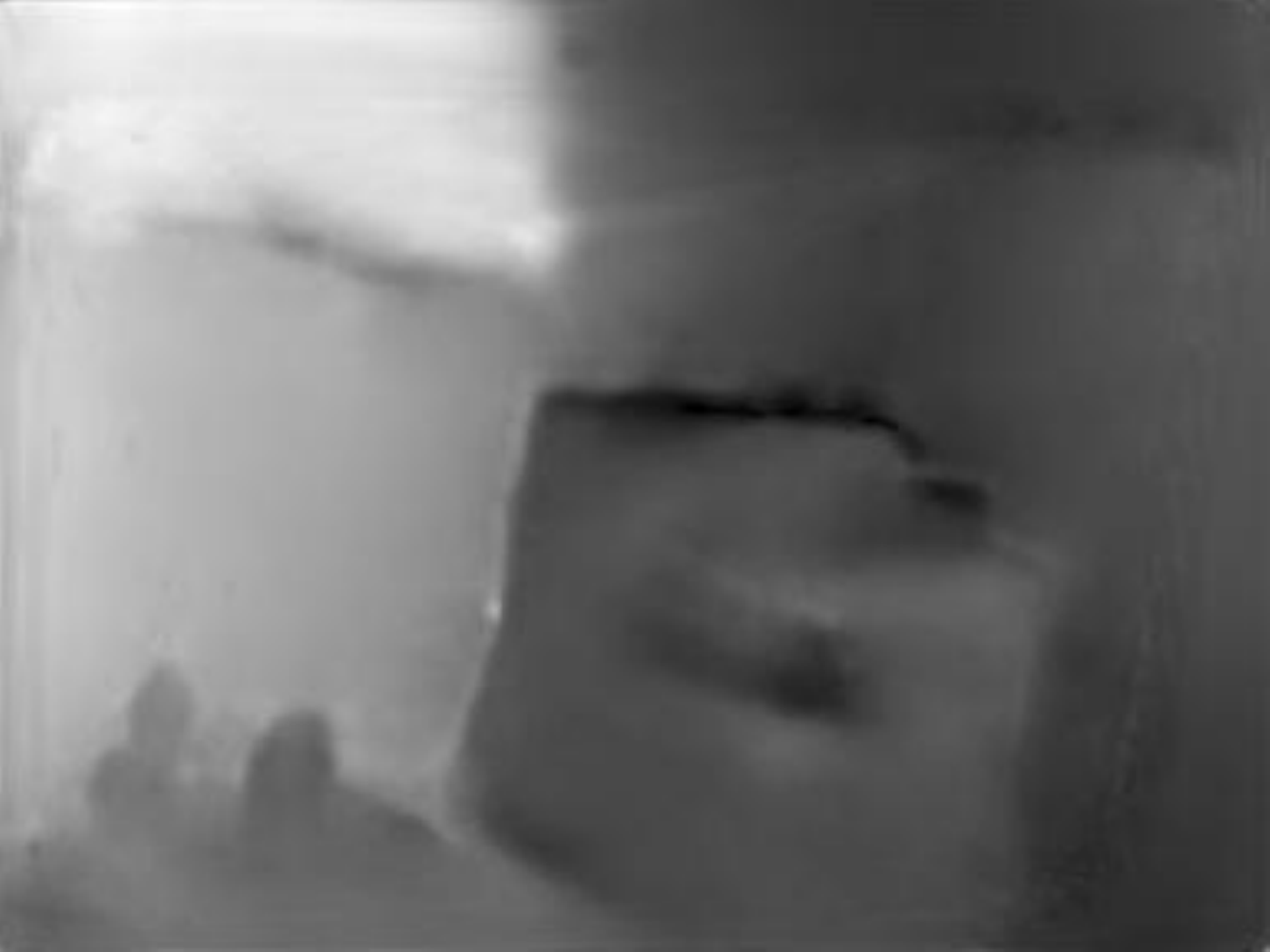}&
    \includegraphics[width=0.104\linewidth]{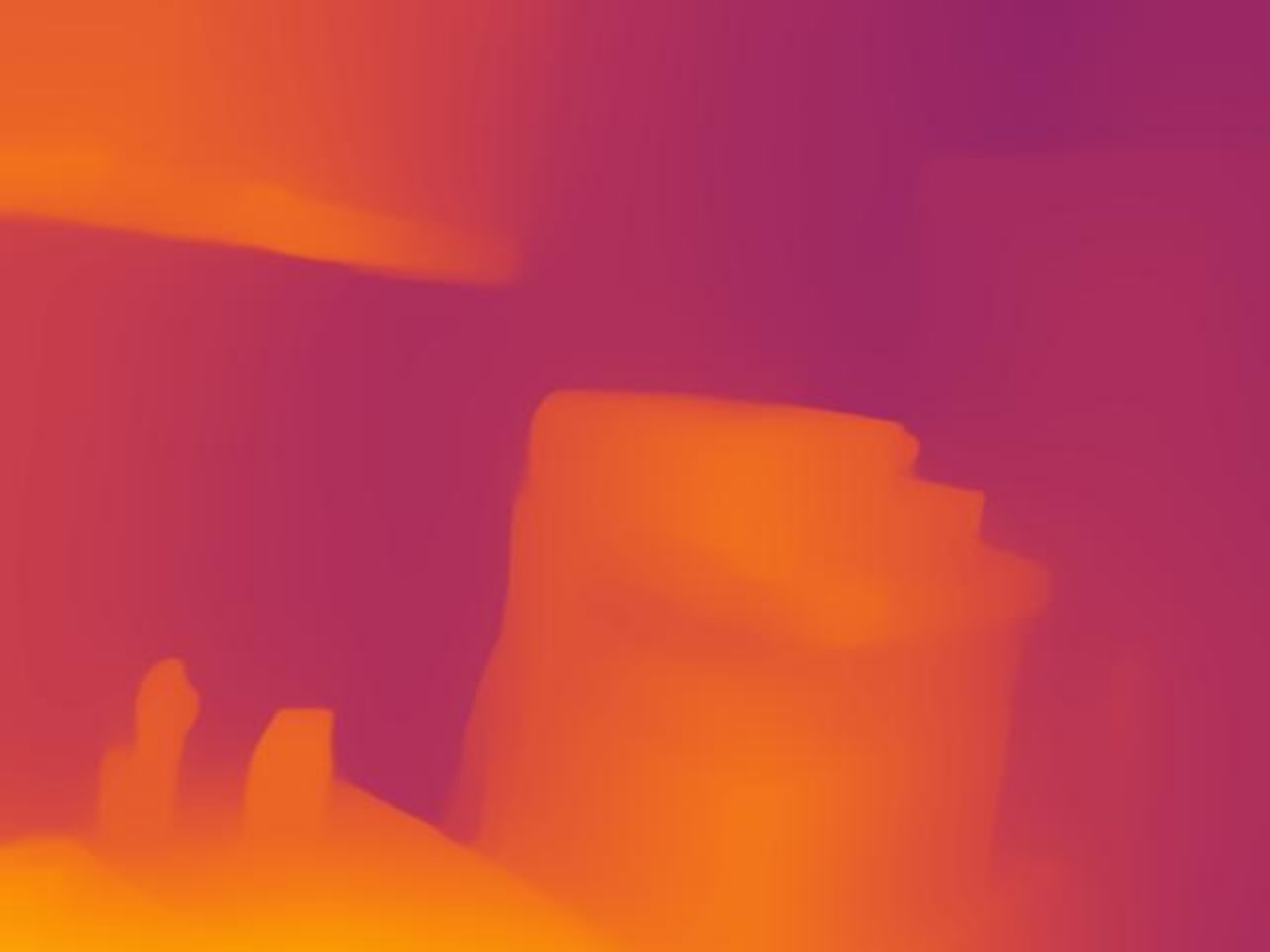}&
    \includegraphics[width=0.104\linewidth]{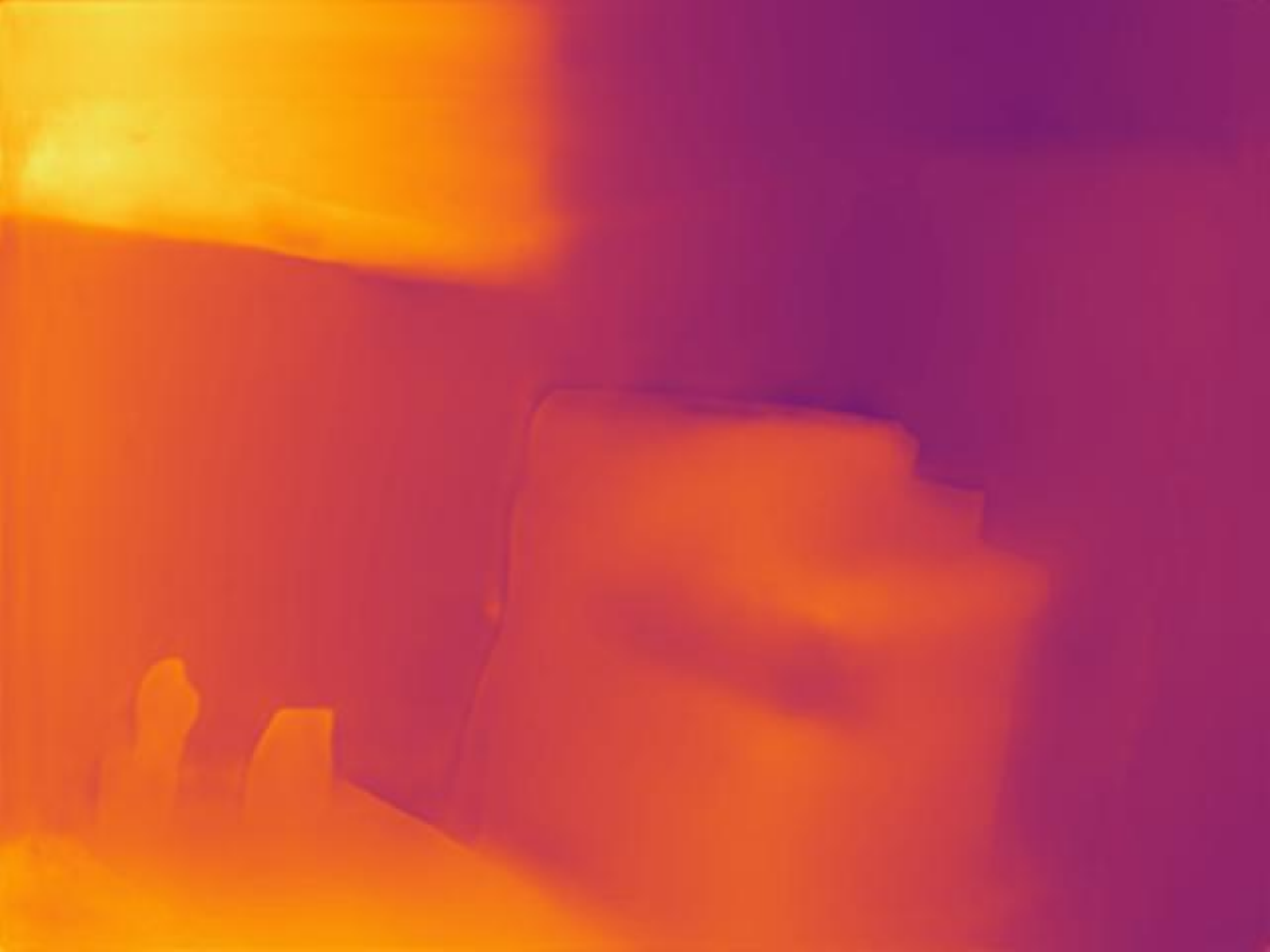}&
    \includegraphics[width=0.104\linewidth]{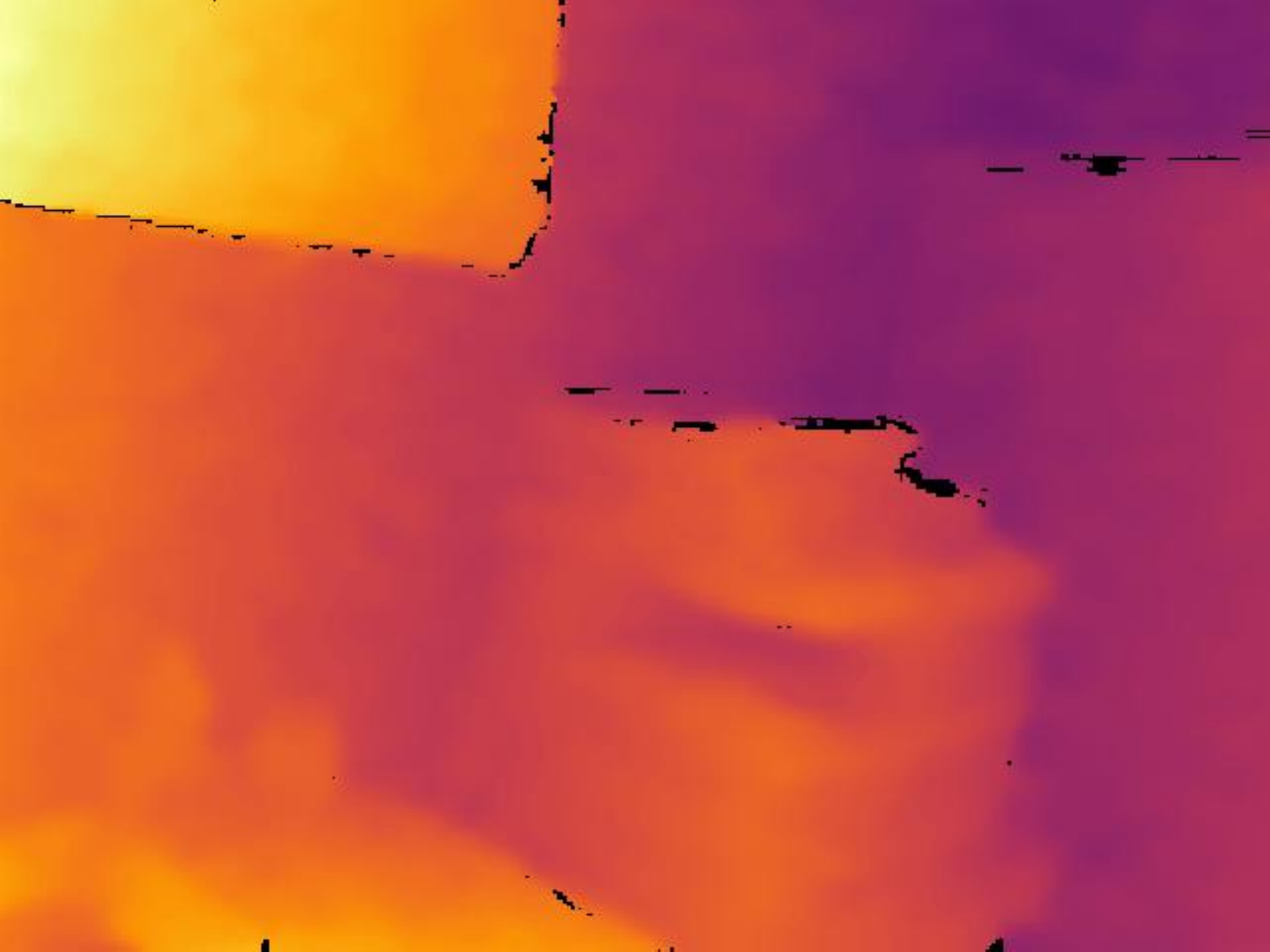}&
    \includegraphics[width=0.104\linewidth]{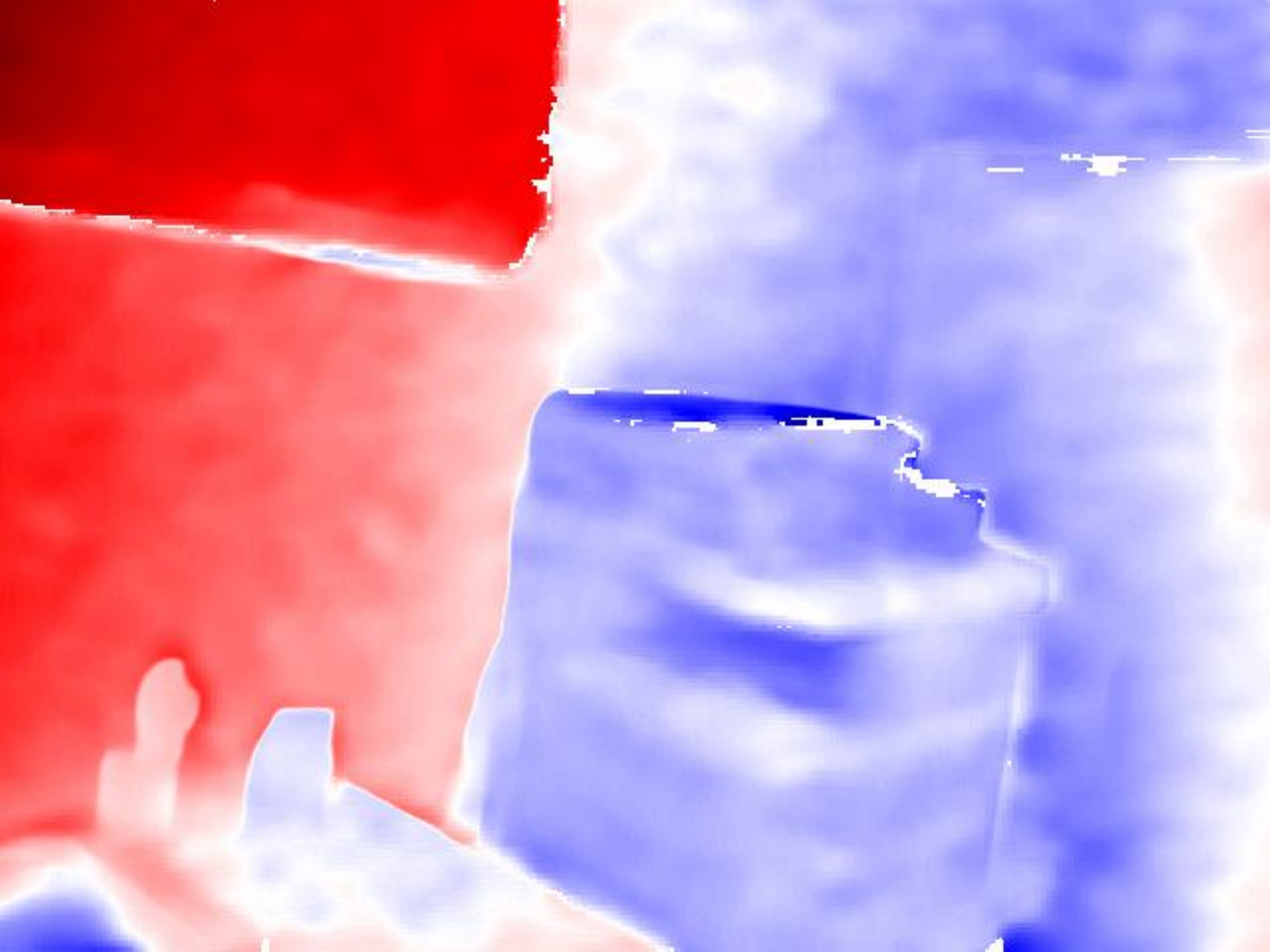}&
    \includegraphics[width=0.104\linewidth]{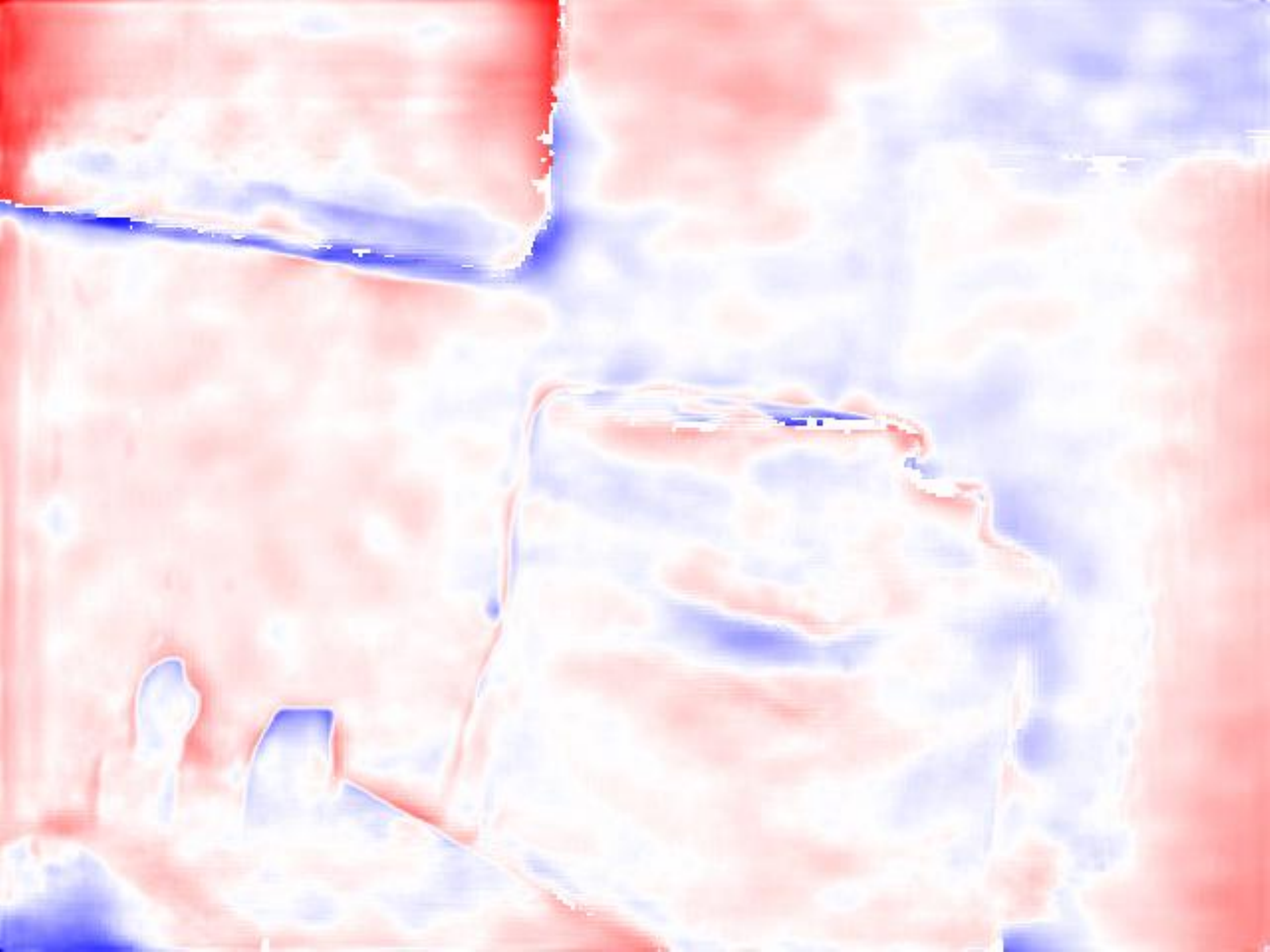}&
    \includegraphics[width=0.024\linewidth]{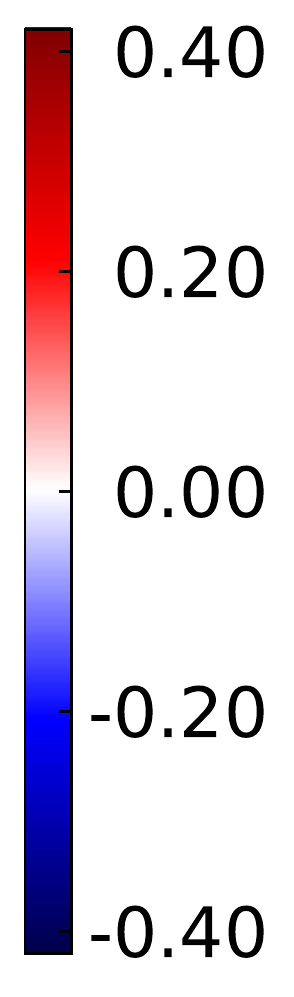}\\
    \vspace{-0.75mm}
    \rot{\scriptsize VOID 1500} &
    \includegraphics[width=0.104\linewidth]{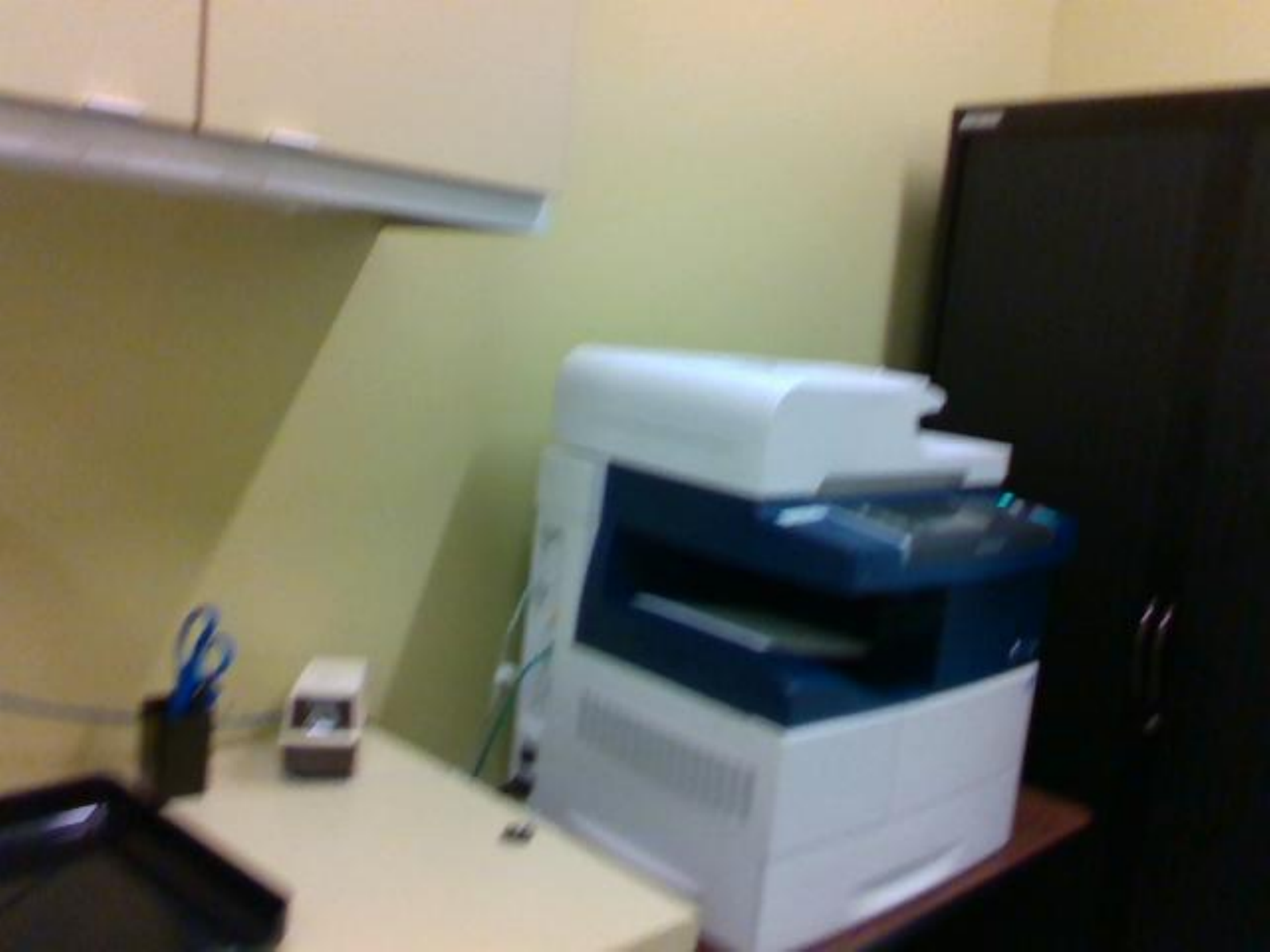}&
    \includegraphics[width=0.104\linewidth]{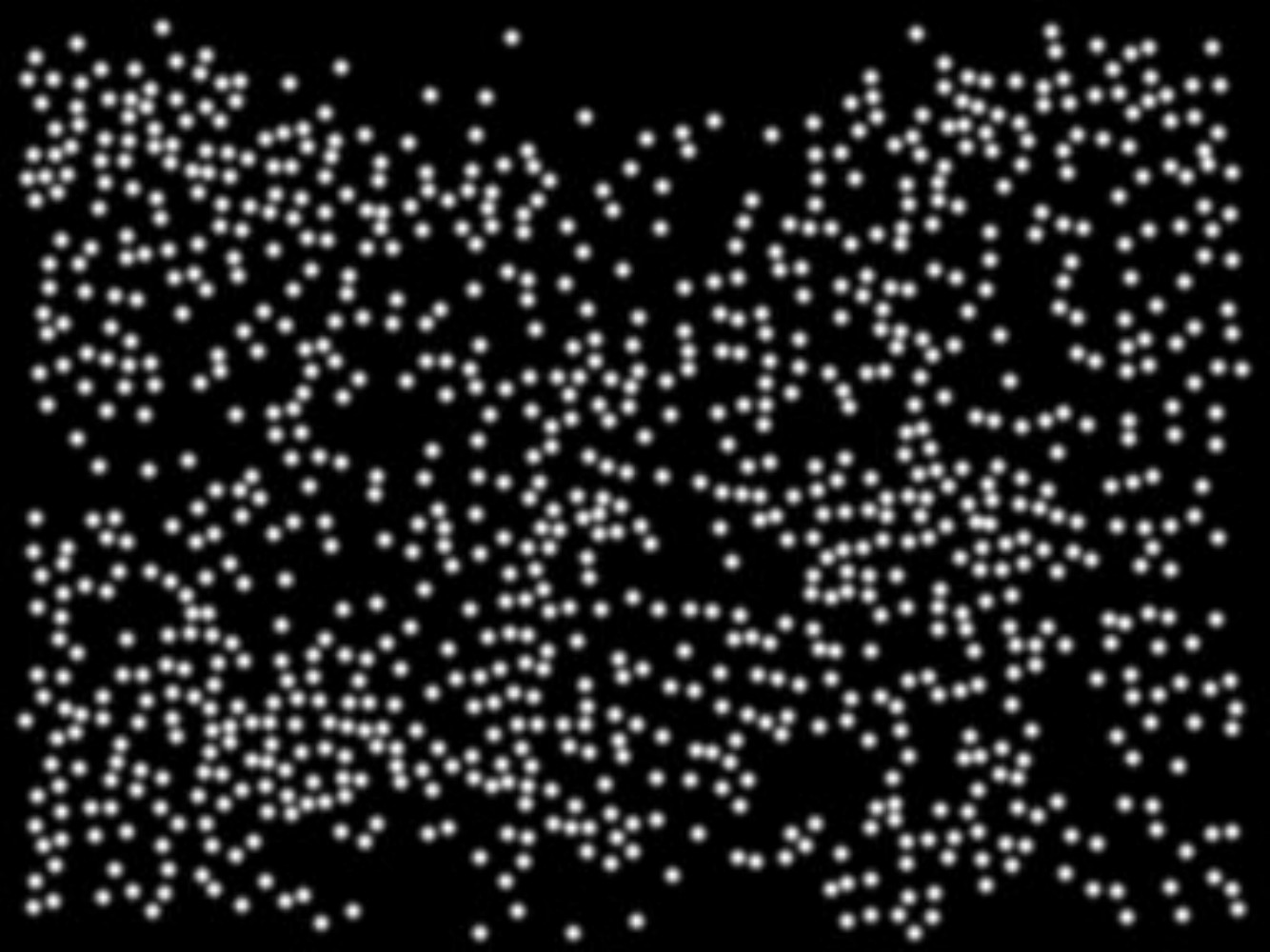}&
    \includegraphics[width=0.104\linewidth]{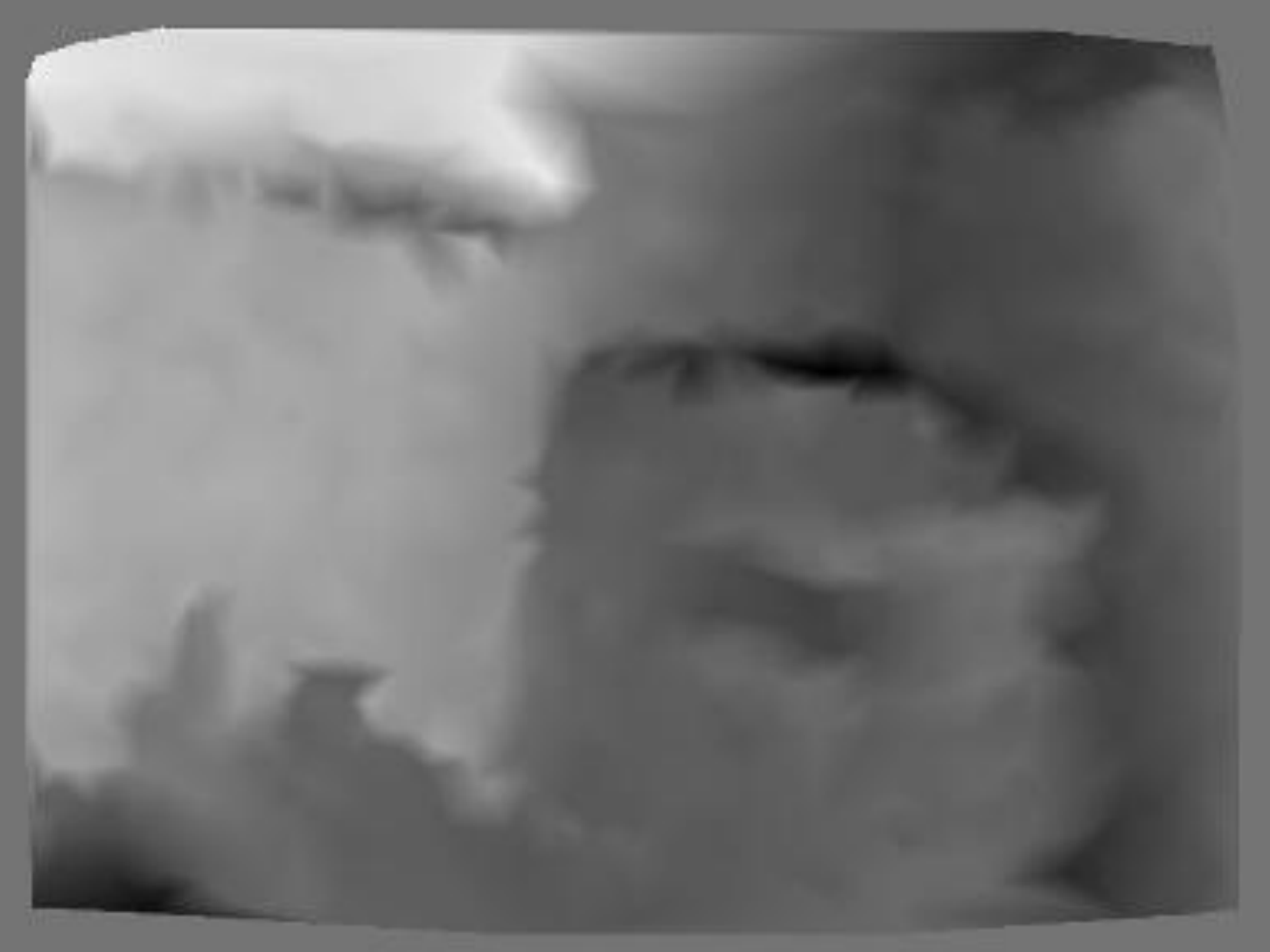}&
    \includegraphics[width=0.104\linewidth]{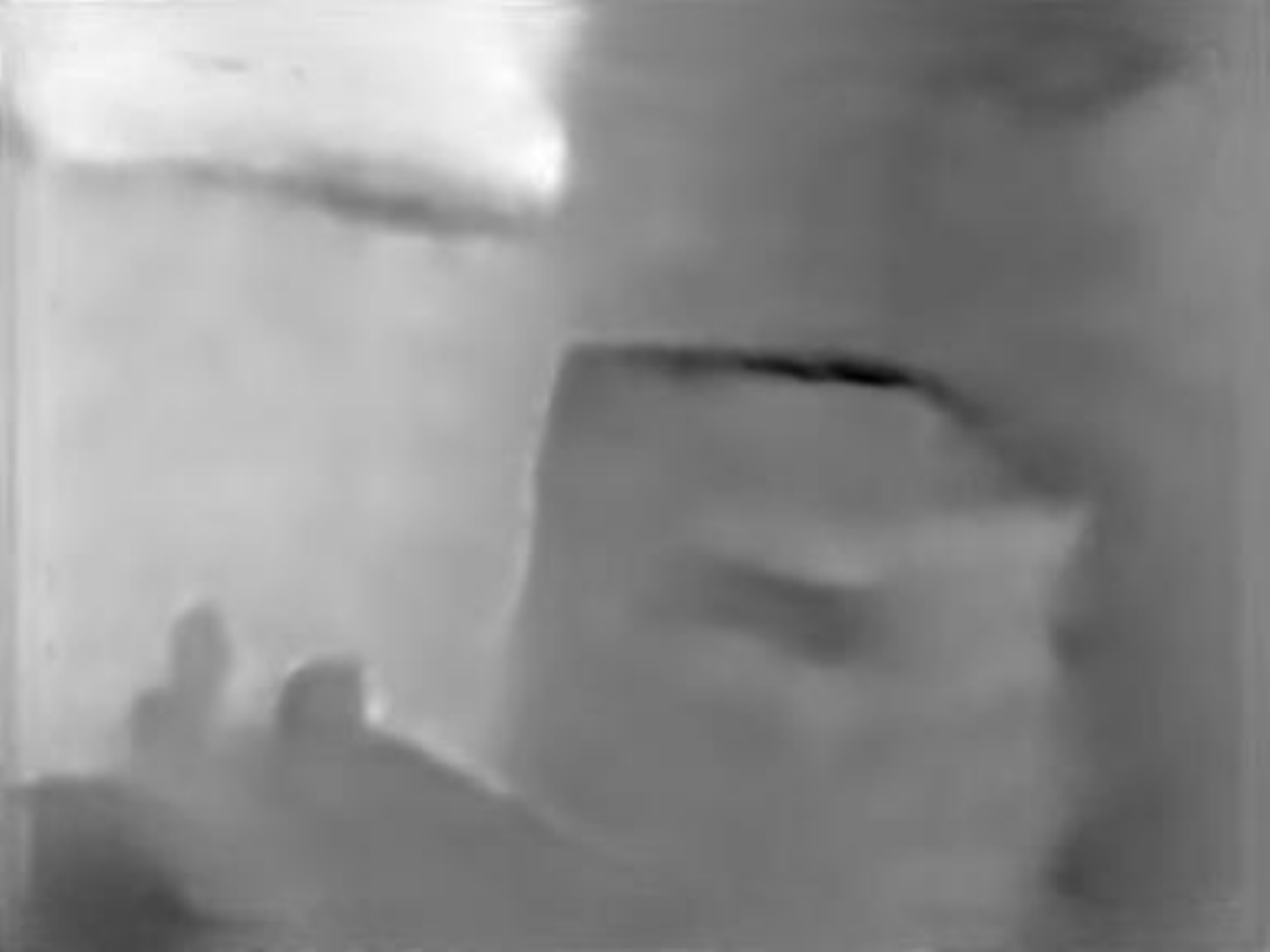}&
    \includegraphics[width=0.104\linewidth]{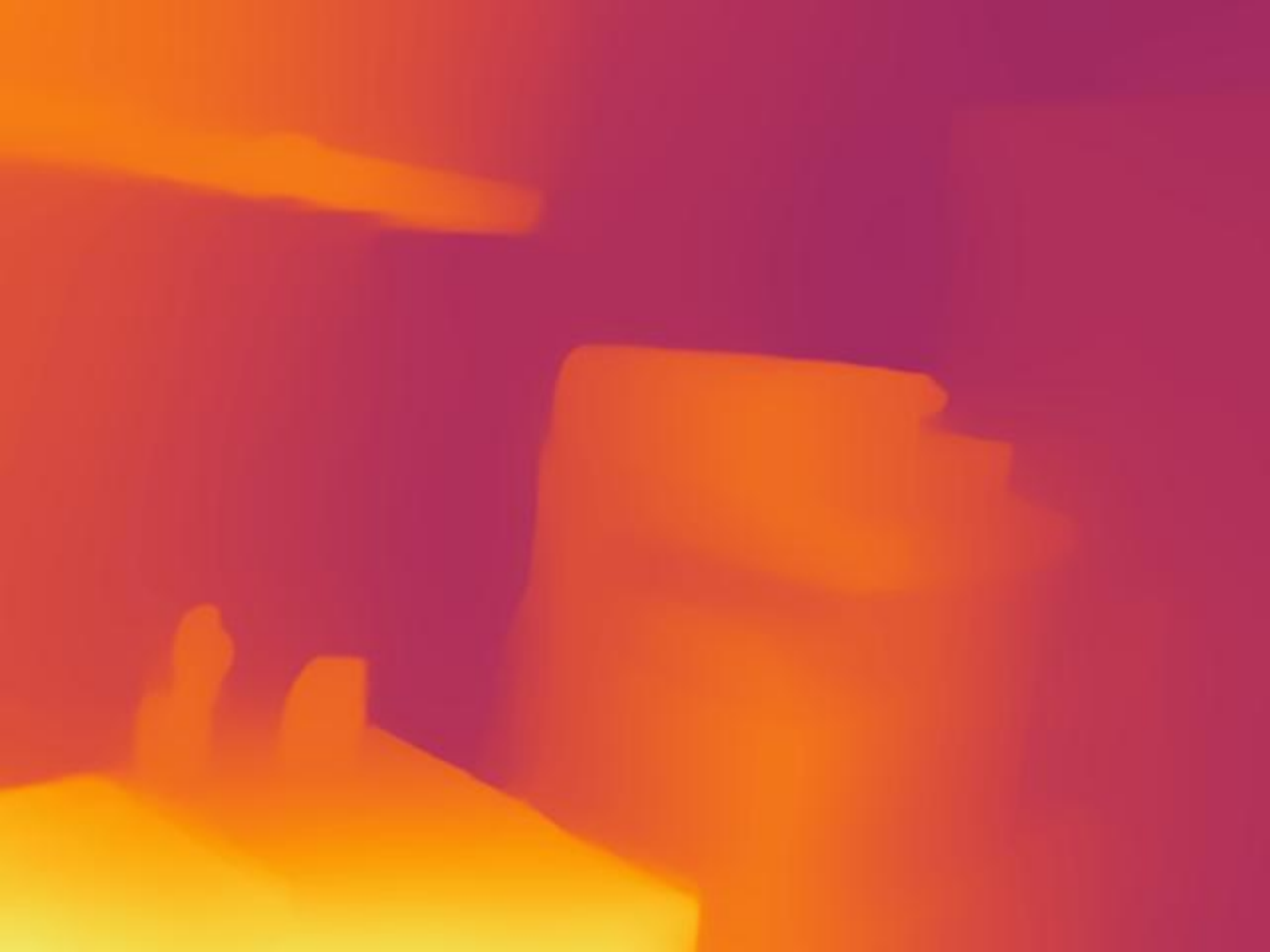}&
    \includegraphics[width=0.104\linewidth]{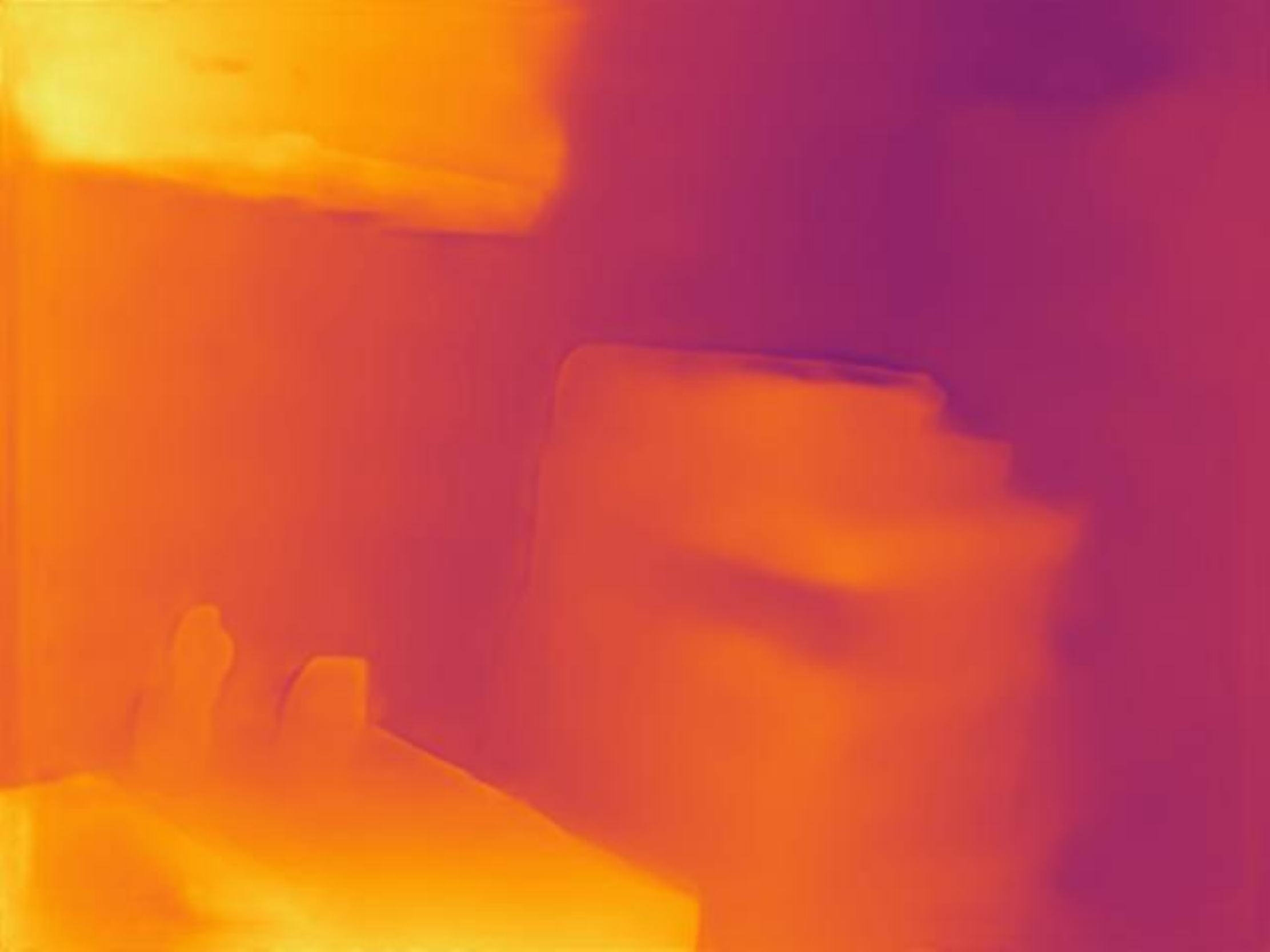}&
    \includegraphics[width=0.104\linewidth]{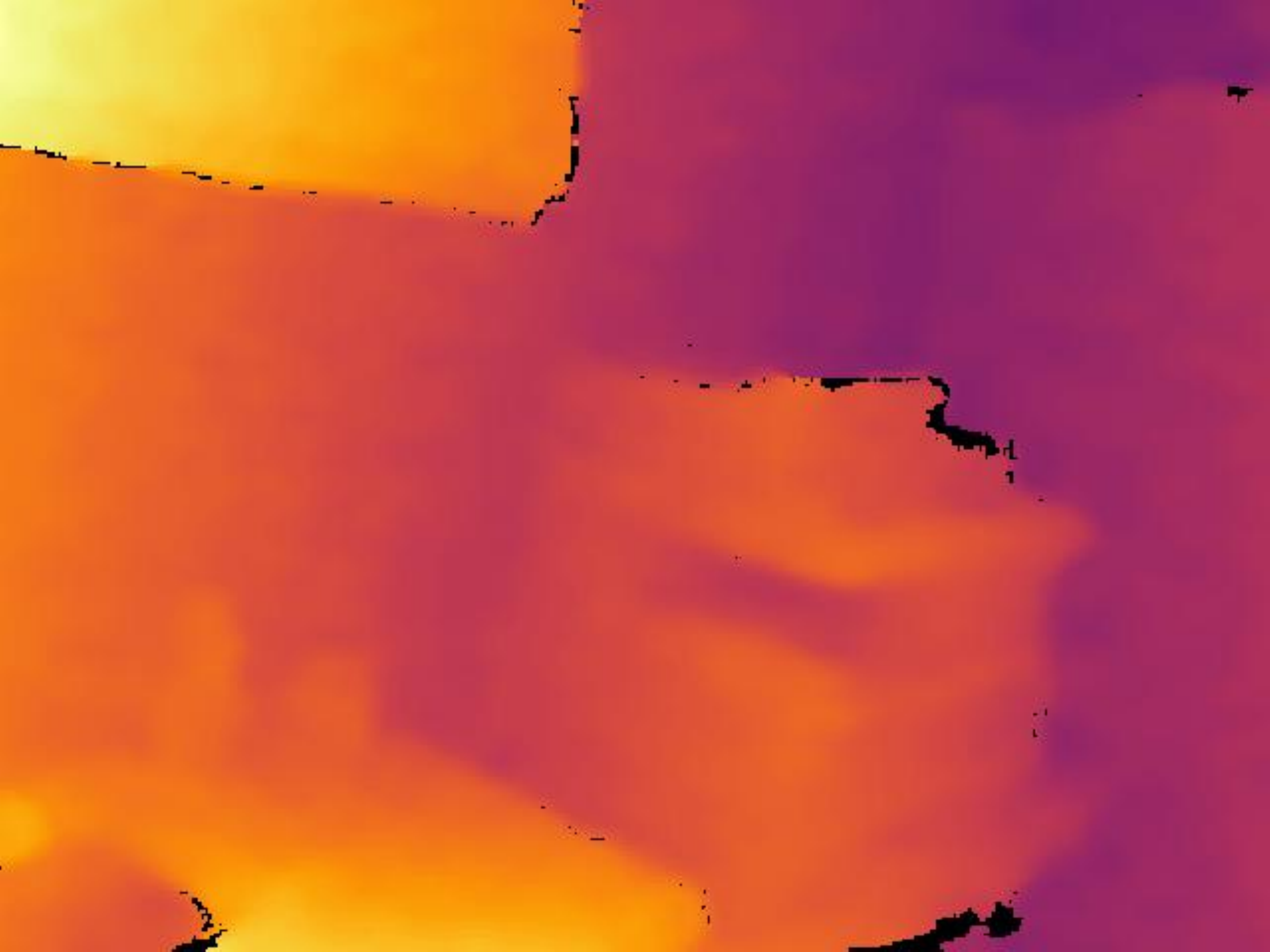}&
    \includegraphics[width=0.104\linewidth]{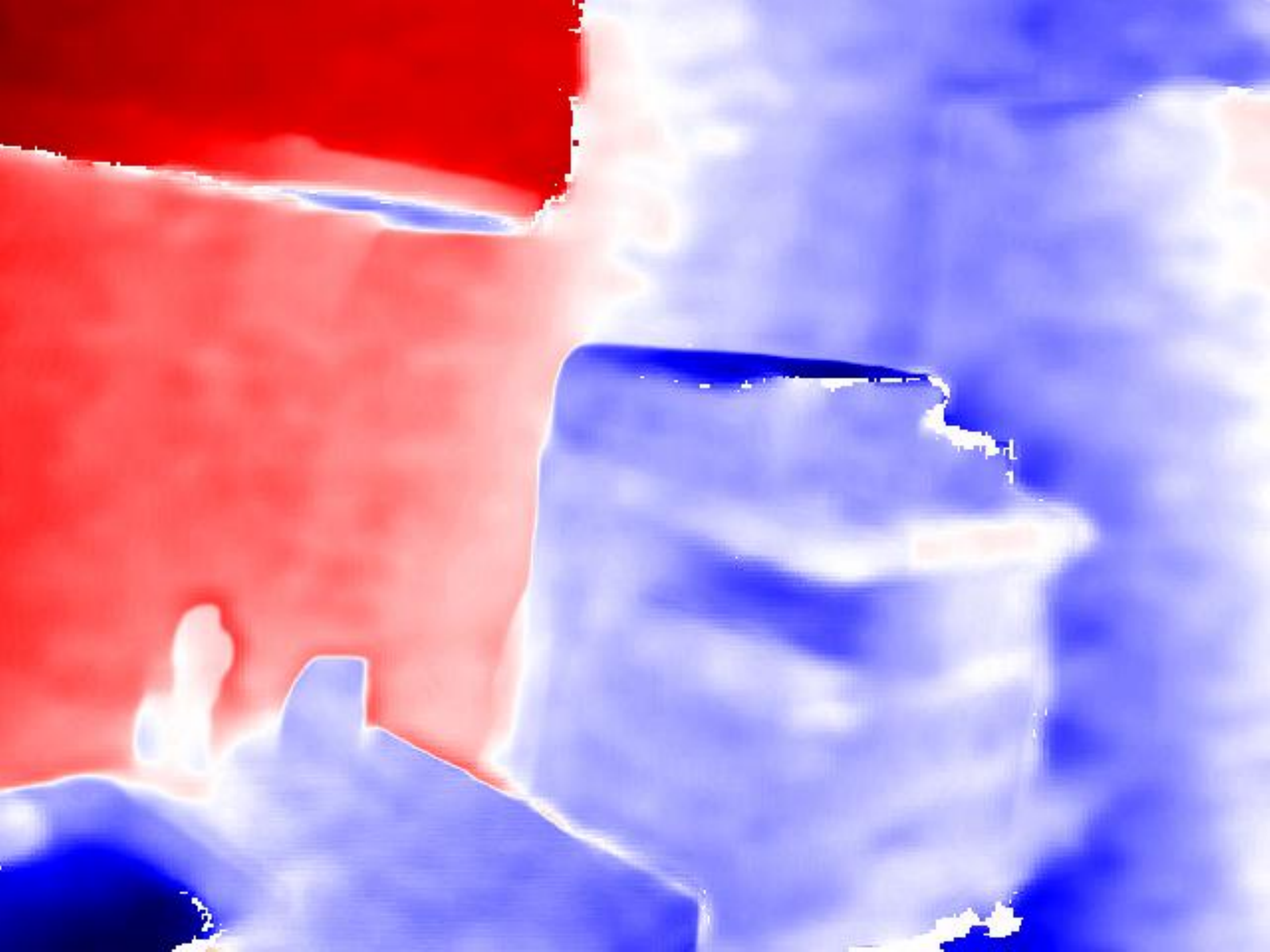}&
    \includegraphics[width=0.104\linewidth]{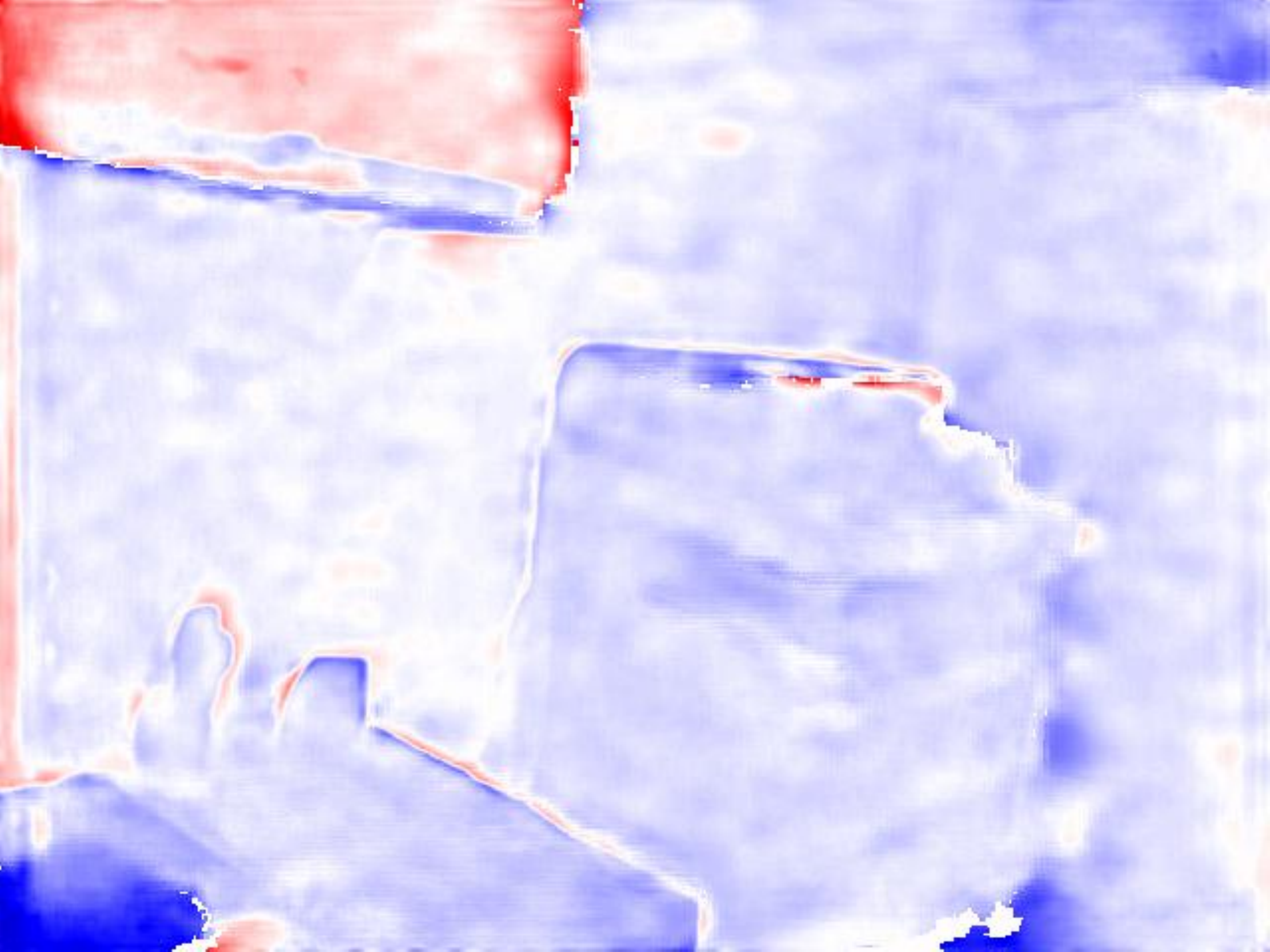}&
    \includegraphics[width=0.024\linewidth]{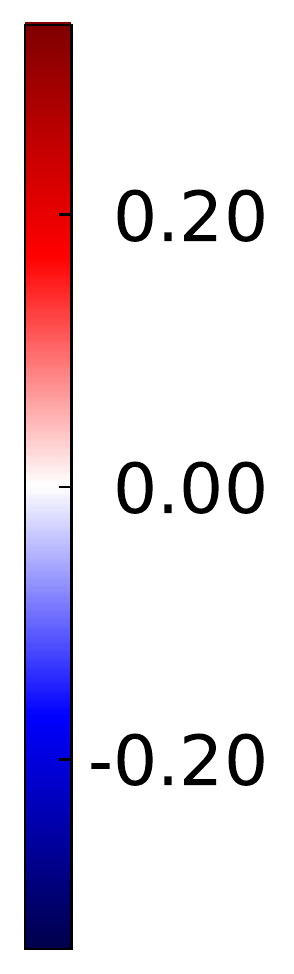}\\
    \multicolumn{9}{c}{} \\
    \vspace{-0.75mm}
    \rot{\scriptsize VOID 150} &
    \includegraphics[width=0.104\linewidth]{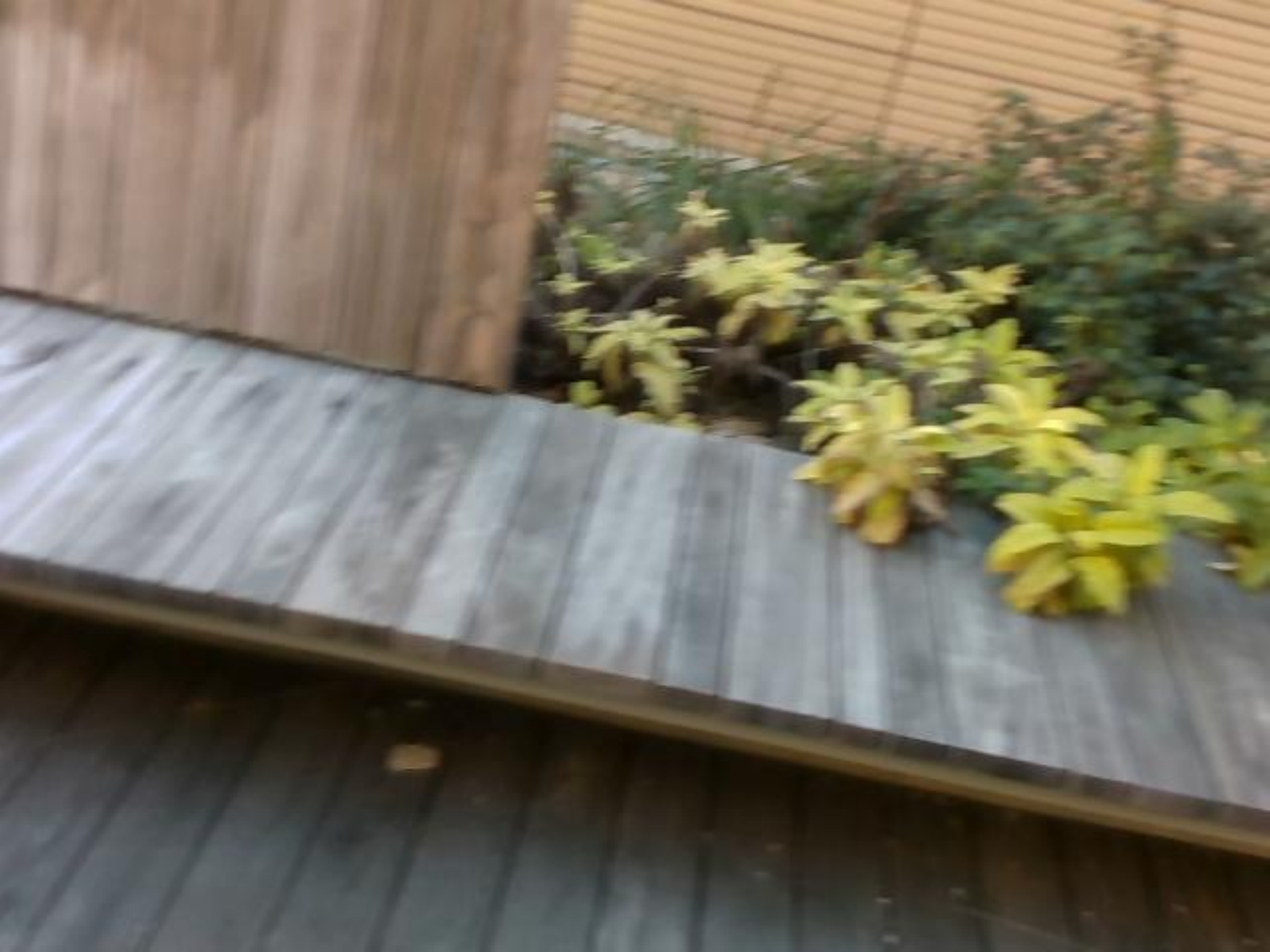}&
    \includegraphics[width=0.104\linewidth]{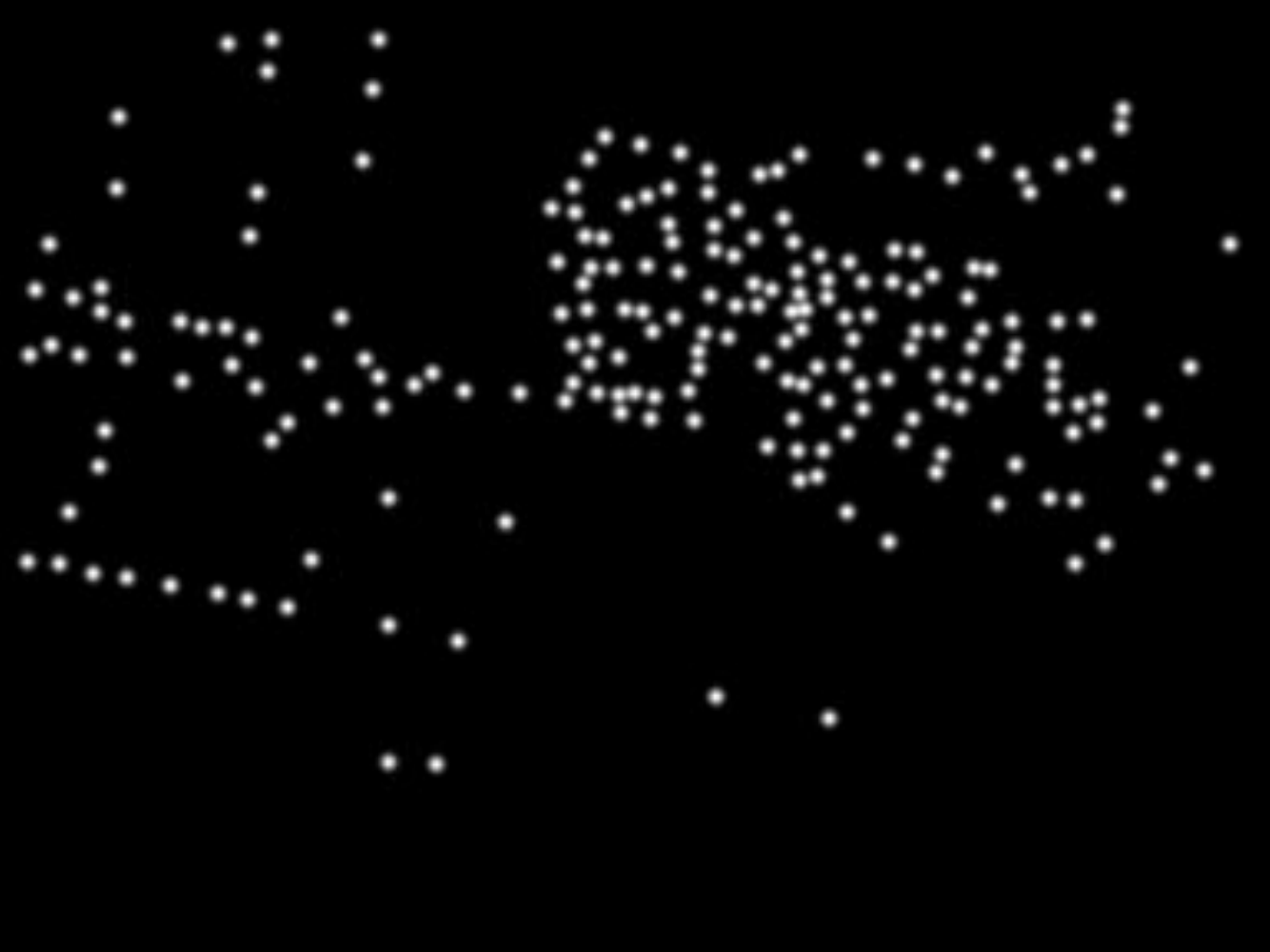}&
    \includegraphics[width=0.104\linewidth]{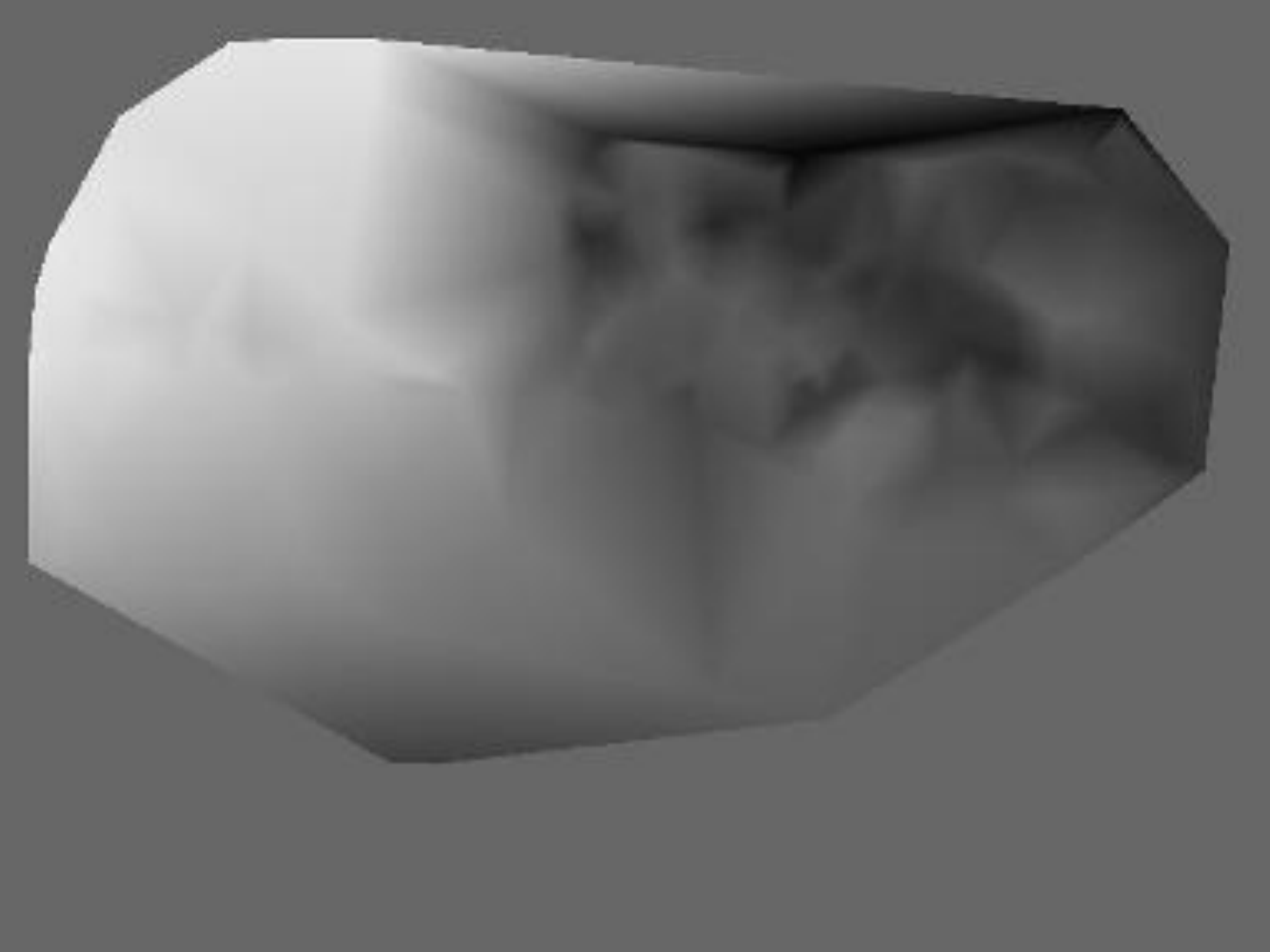}&
    \includegraphics[width=0.104\linewidth]{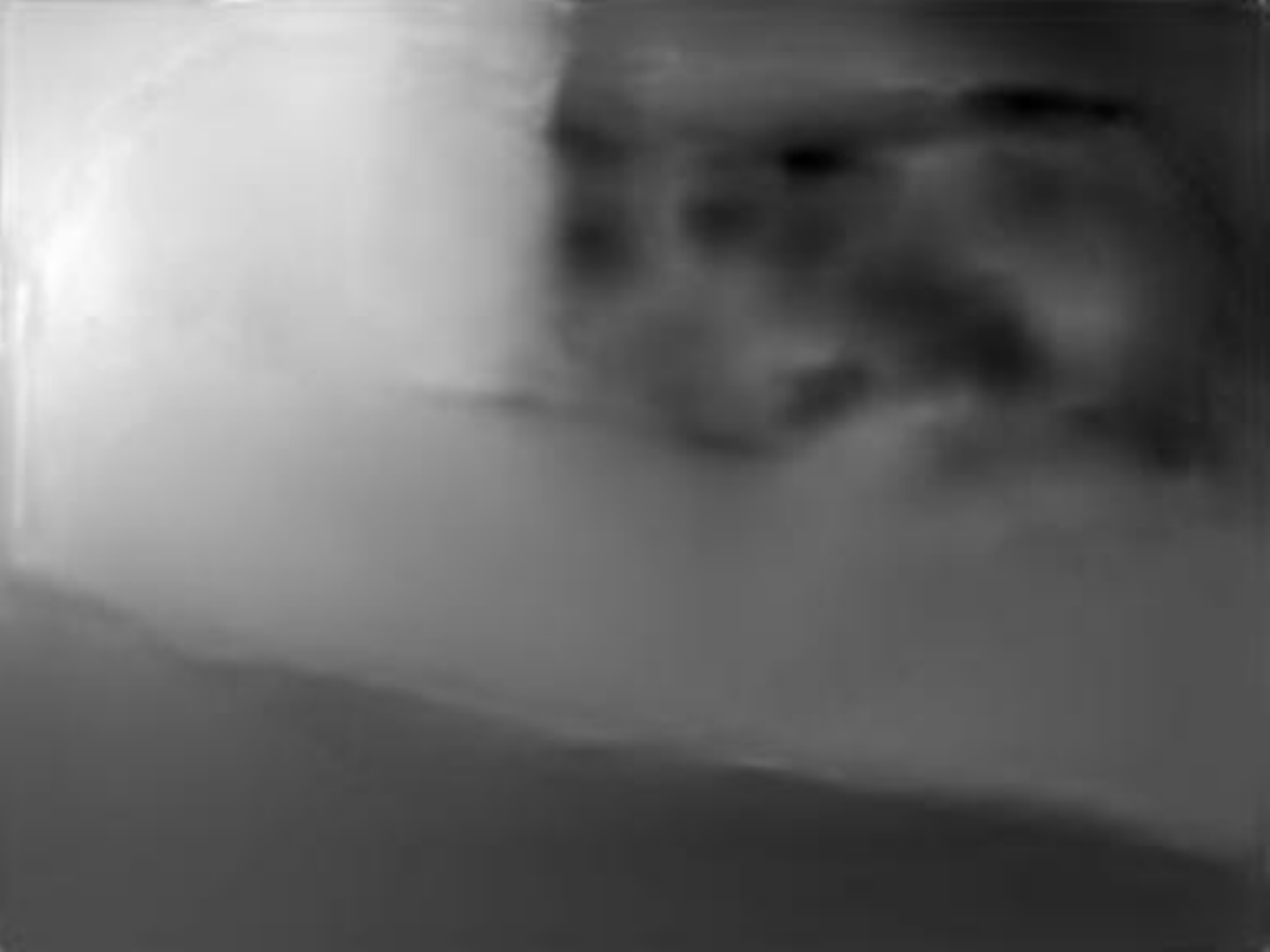}&
    \includegraphics[width=0.104\linewidth]{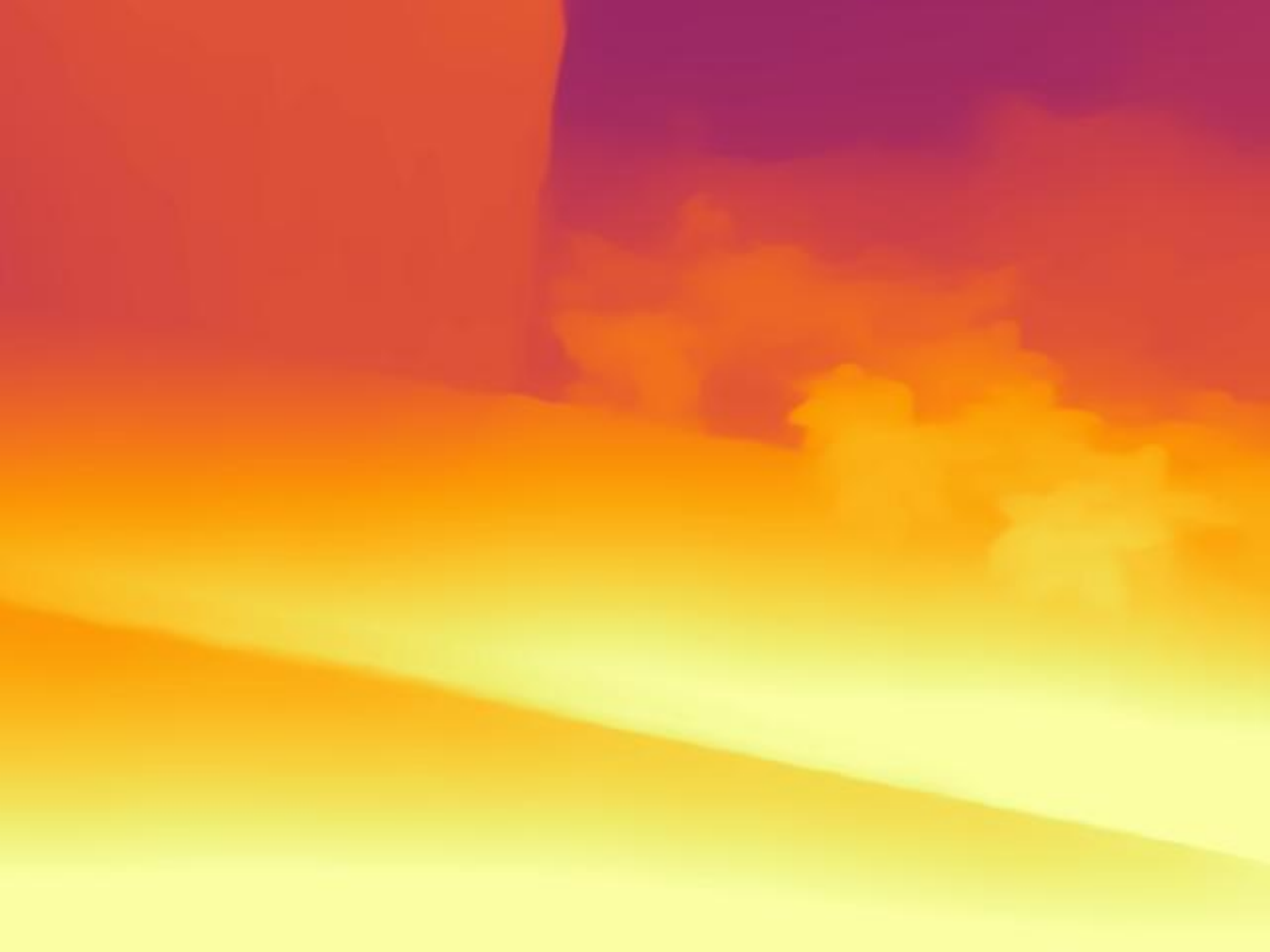}&
    \includegraphics[width=0.104\linewidth]{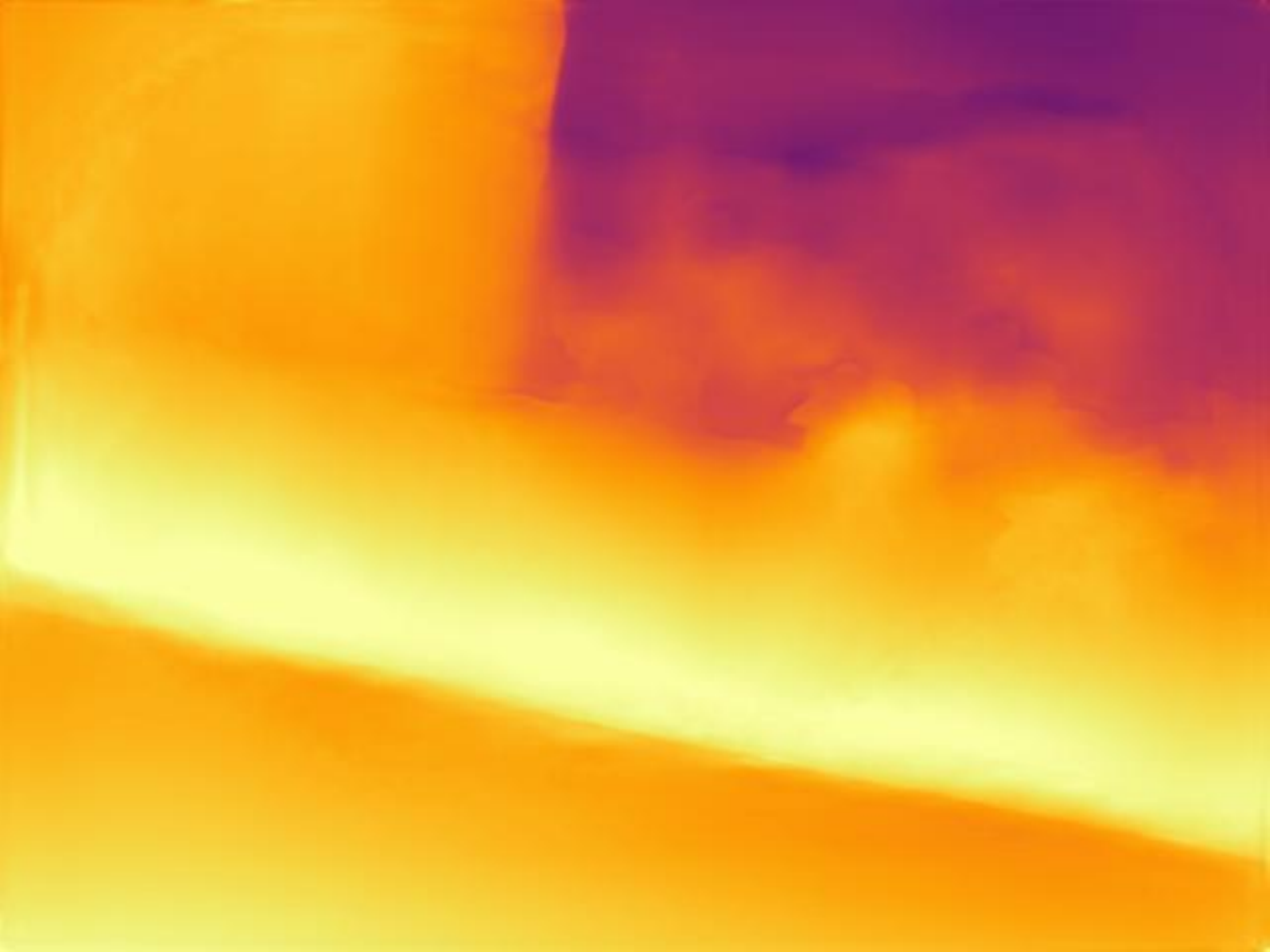}&
    \includegraphics[width=0.104\linewidth]{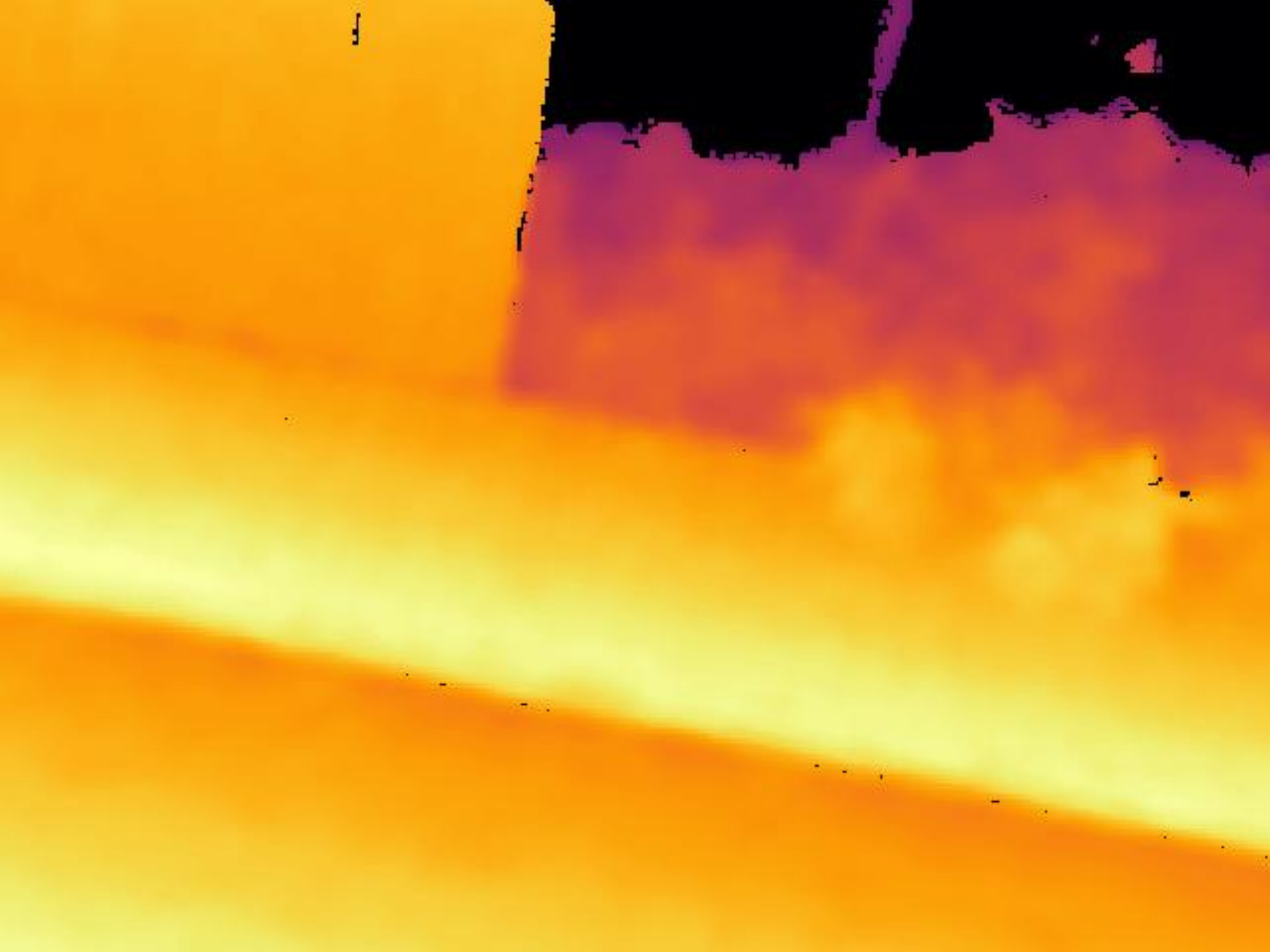}&
    \includegraphics[width=0.104\linewidth]{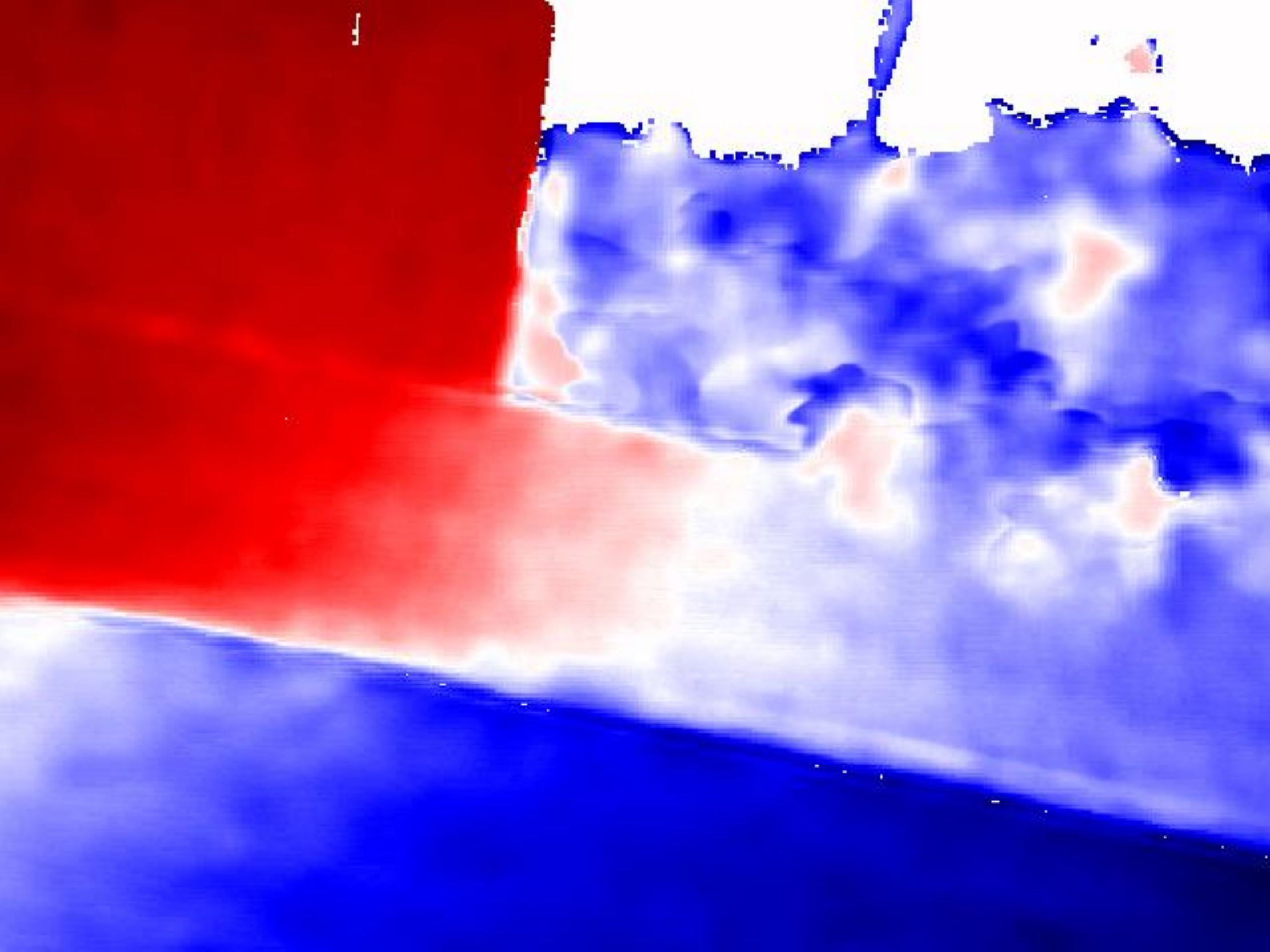}&
    \includegraphics[width=0.104\linewidth]{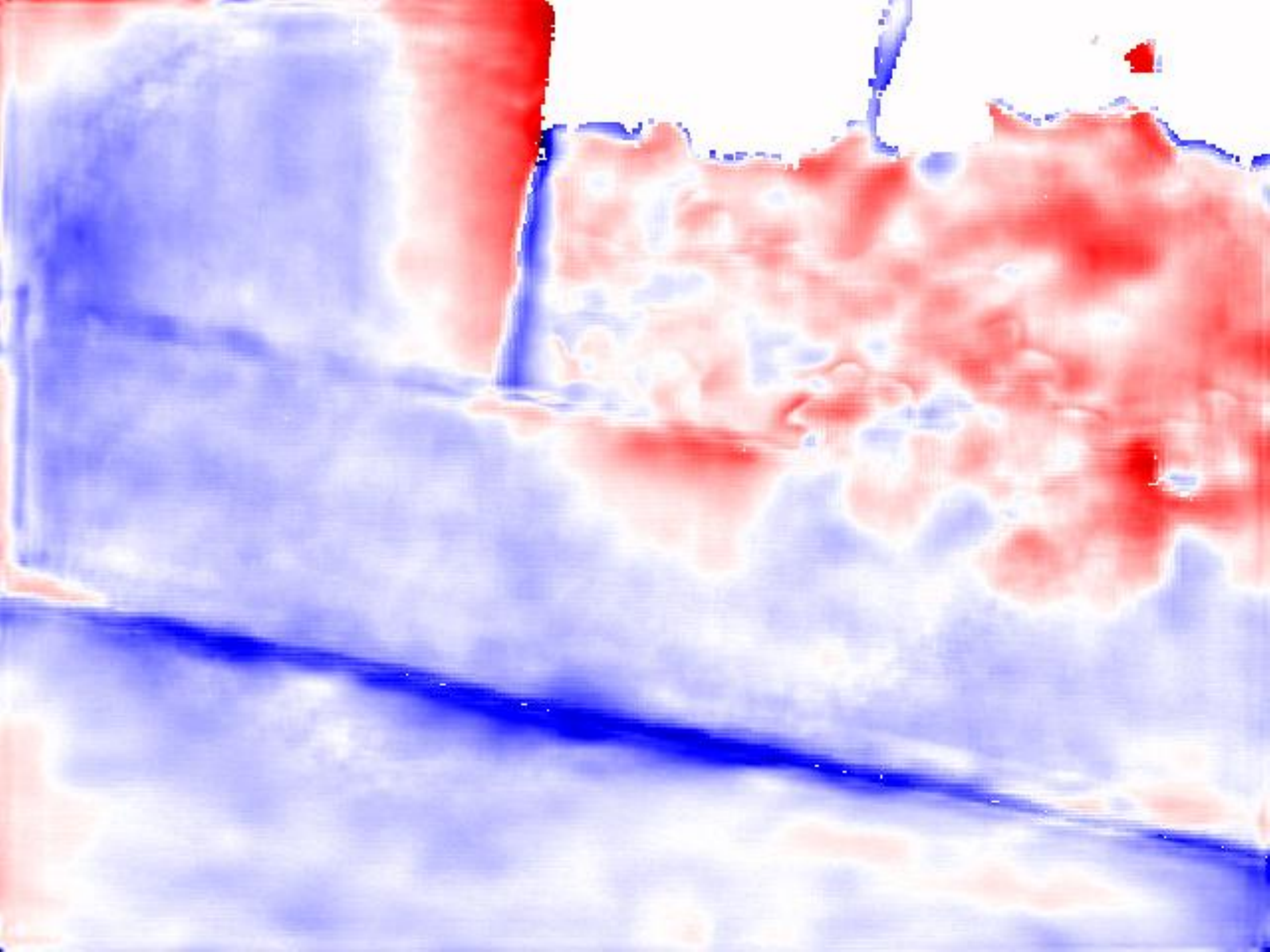}&
    \includegraphics[width=0.024\linewidth]{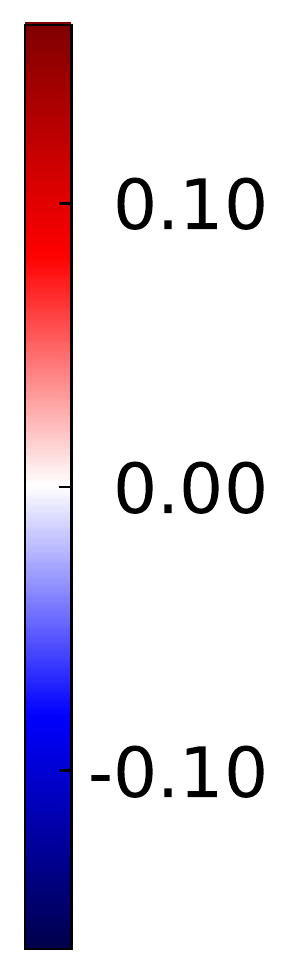}\\
    \vspace{-0.75mm}
    \rot{\scriptsize VOID 500} &
    \includegraphics[width=0.104\linewidth]{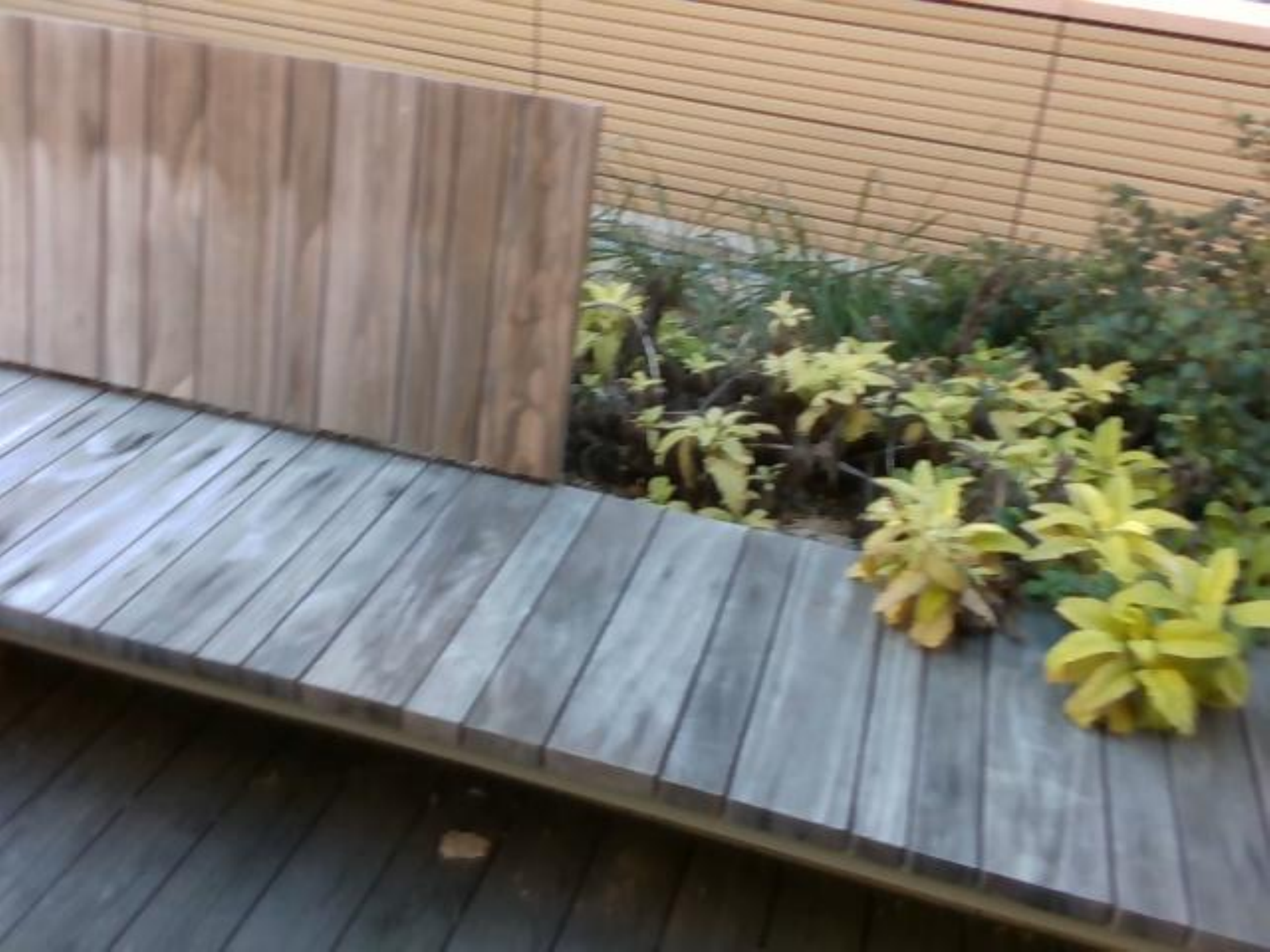}&
    \includegraphics[width=0.104\linewidth]{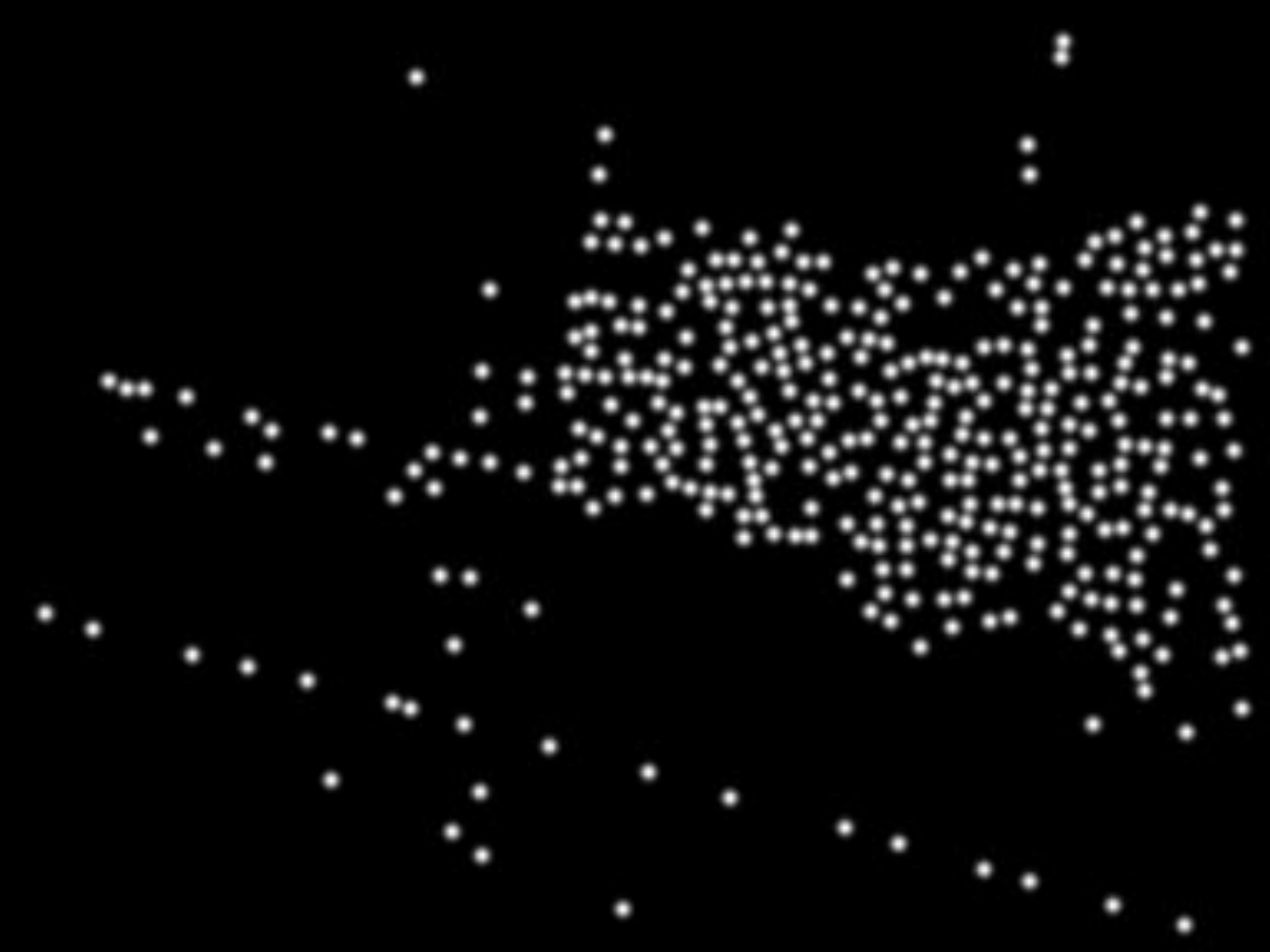}&
    \includegraphics[width=0.104\linewidth]{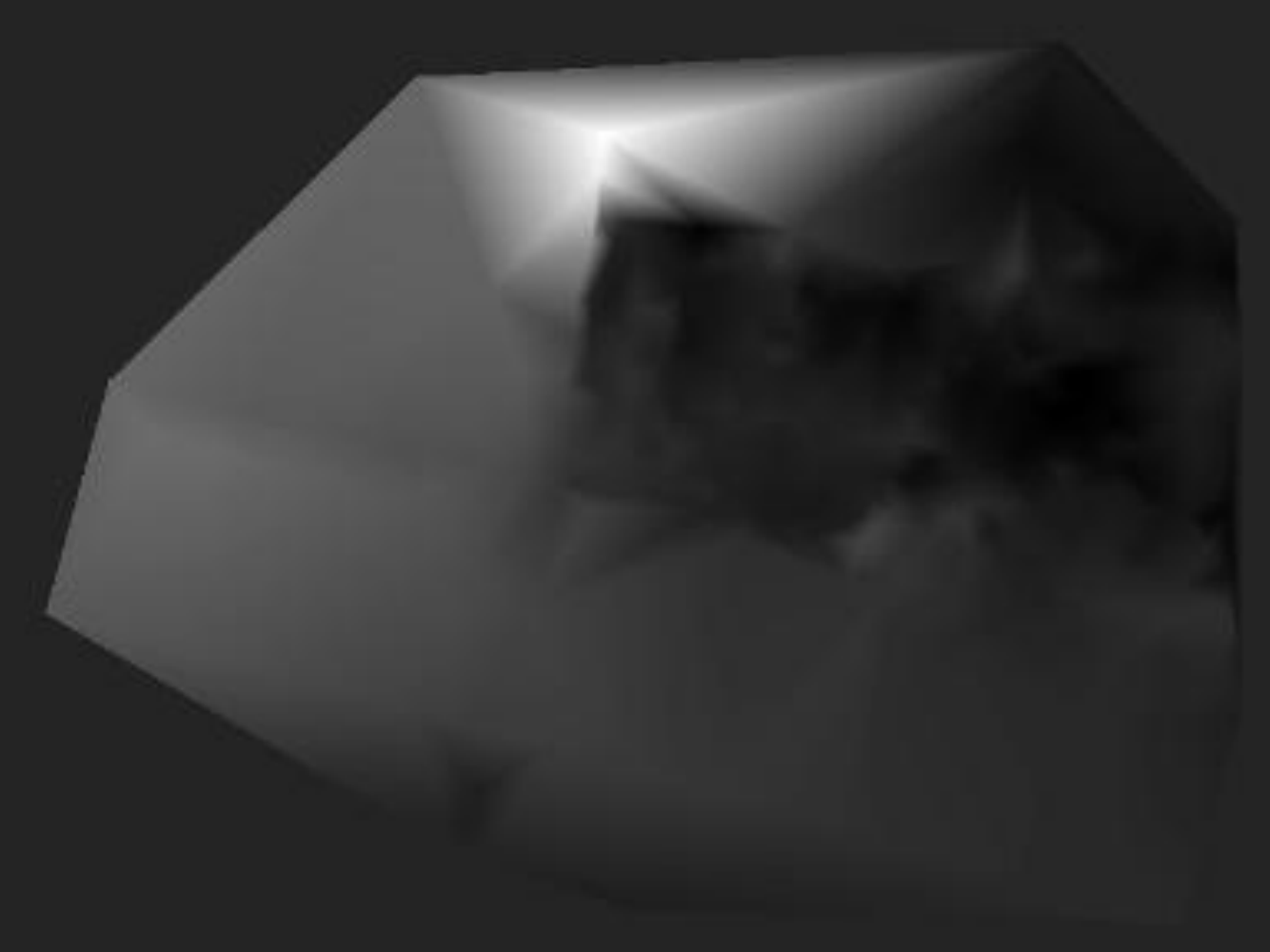}&
    \includegraphics[width=0.104\linewidth]{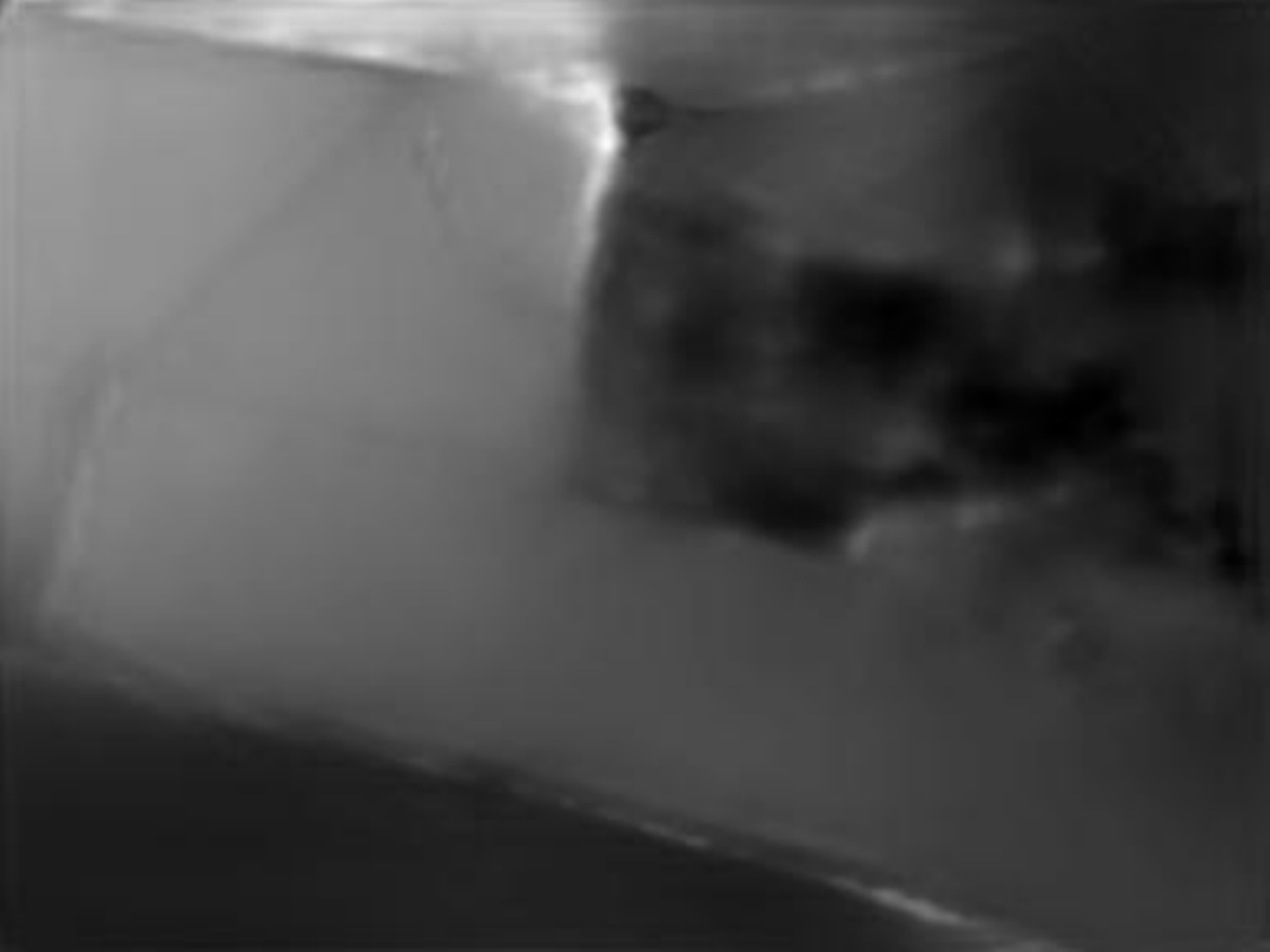}&
    \includegraphics[width=0.104\linewidth]{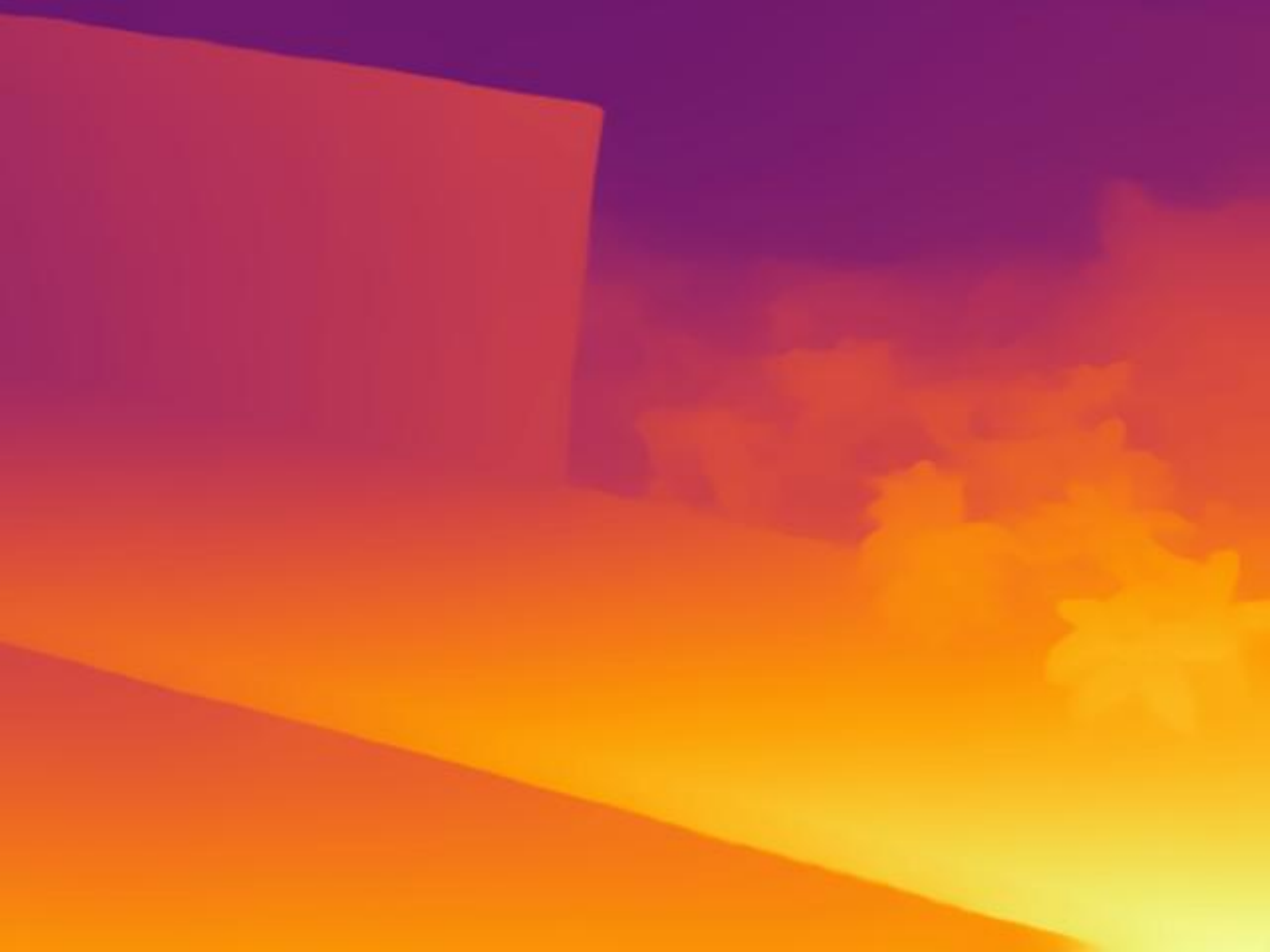}&
    \includegraphics[width=0.104\linewidth]{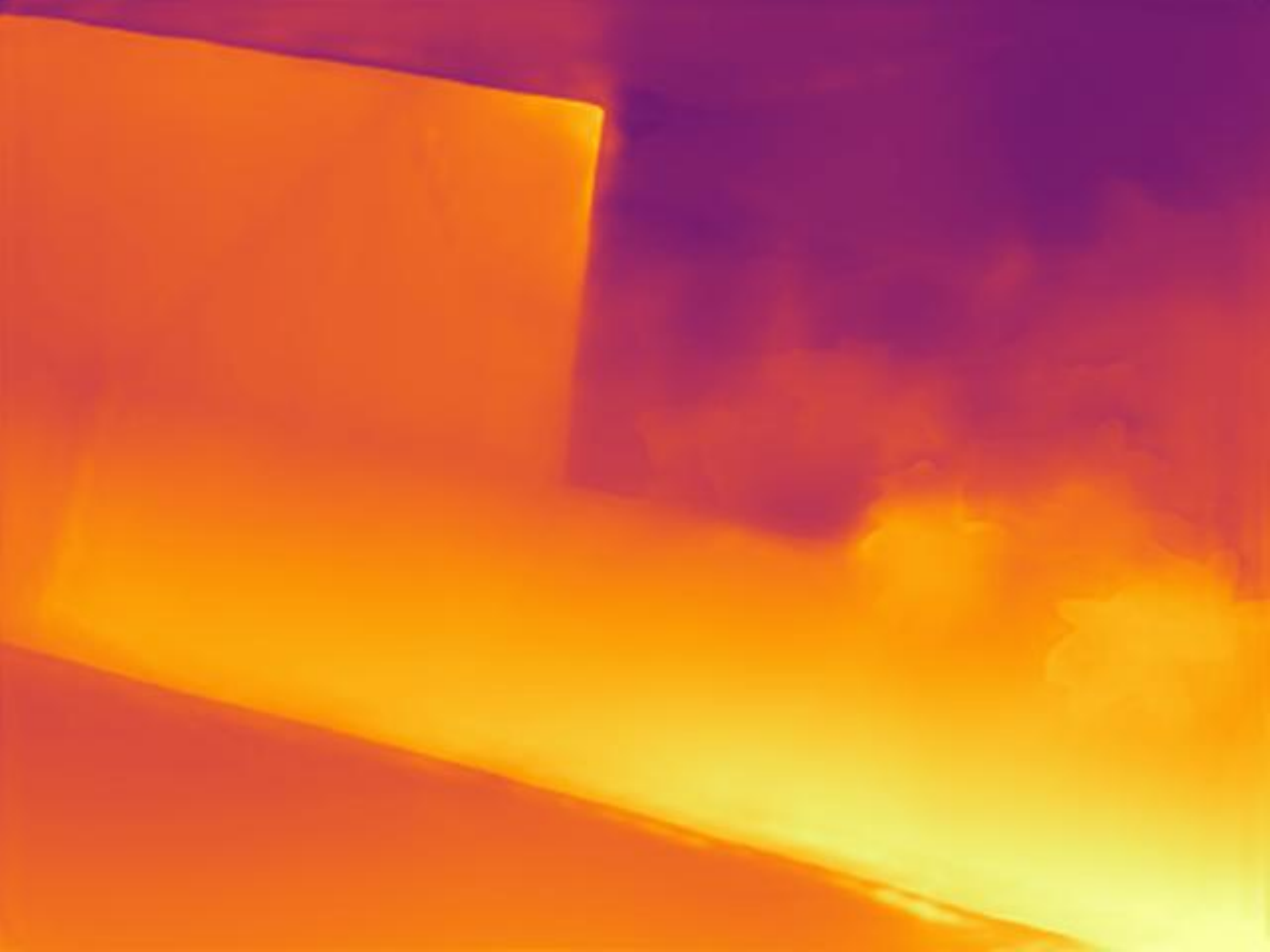}&
    \includegraphics[width=0.104\linewidth]{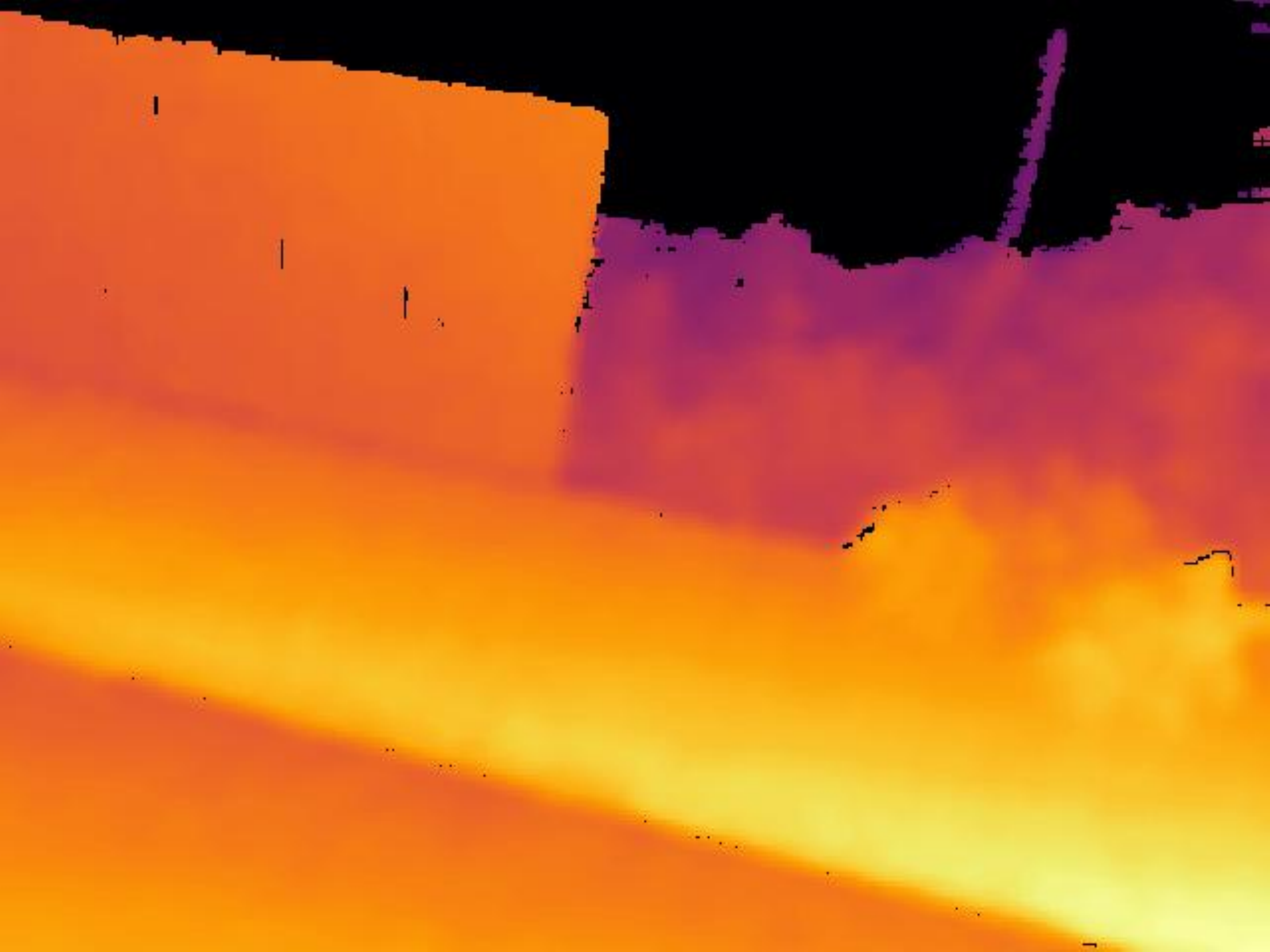}&
    \includegraphics[width=0.104\linewidth]{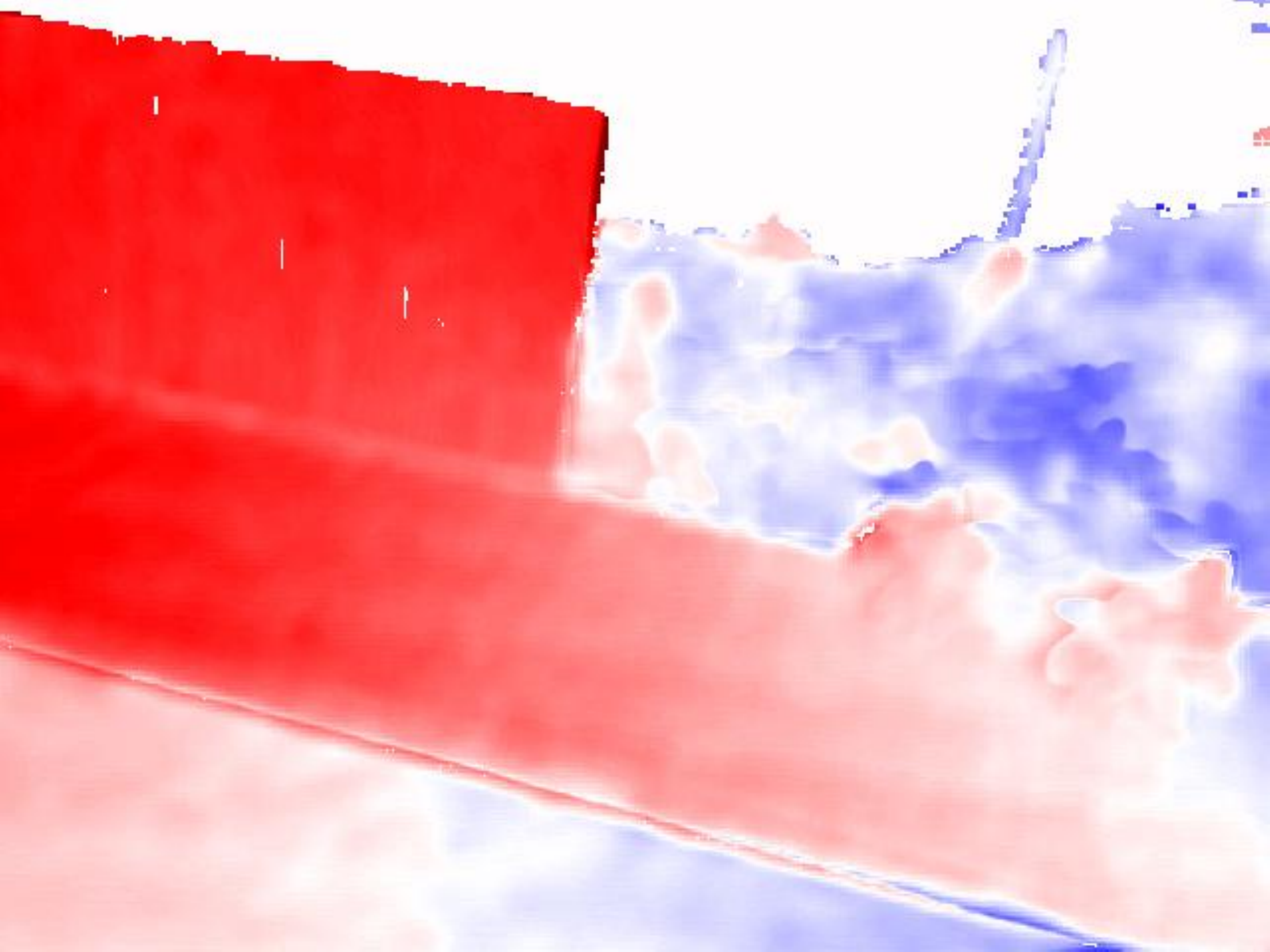}&
    \includegraphics[width=0.104\linewidth]{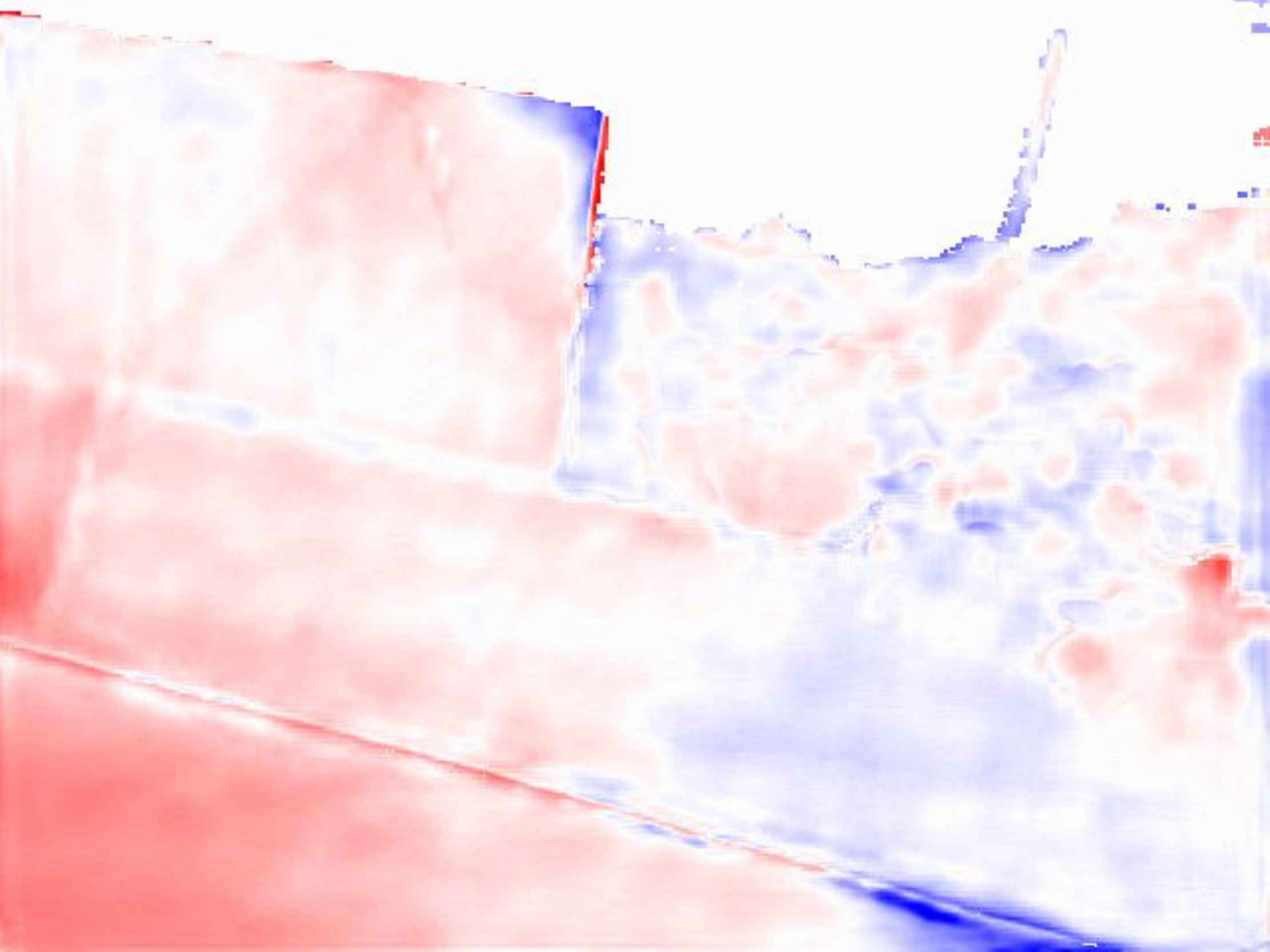}&
    \includegraphics[width=0.024\linewidth]{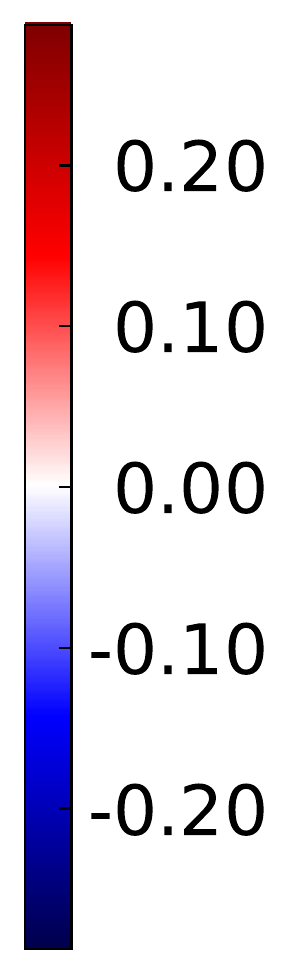}\\
    \vspace{-0.75mm}
    \rot{\scriptsize VOID 1500} &
    \includegraphics[width=0.104\linewidth]{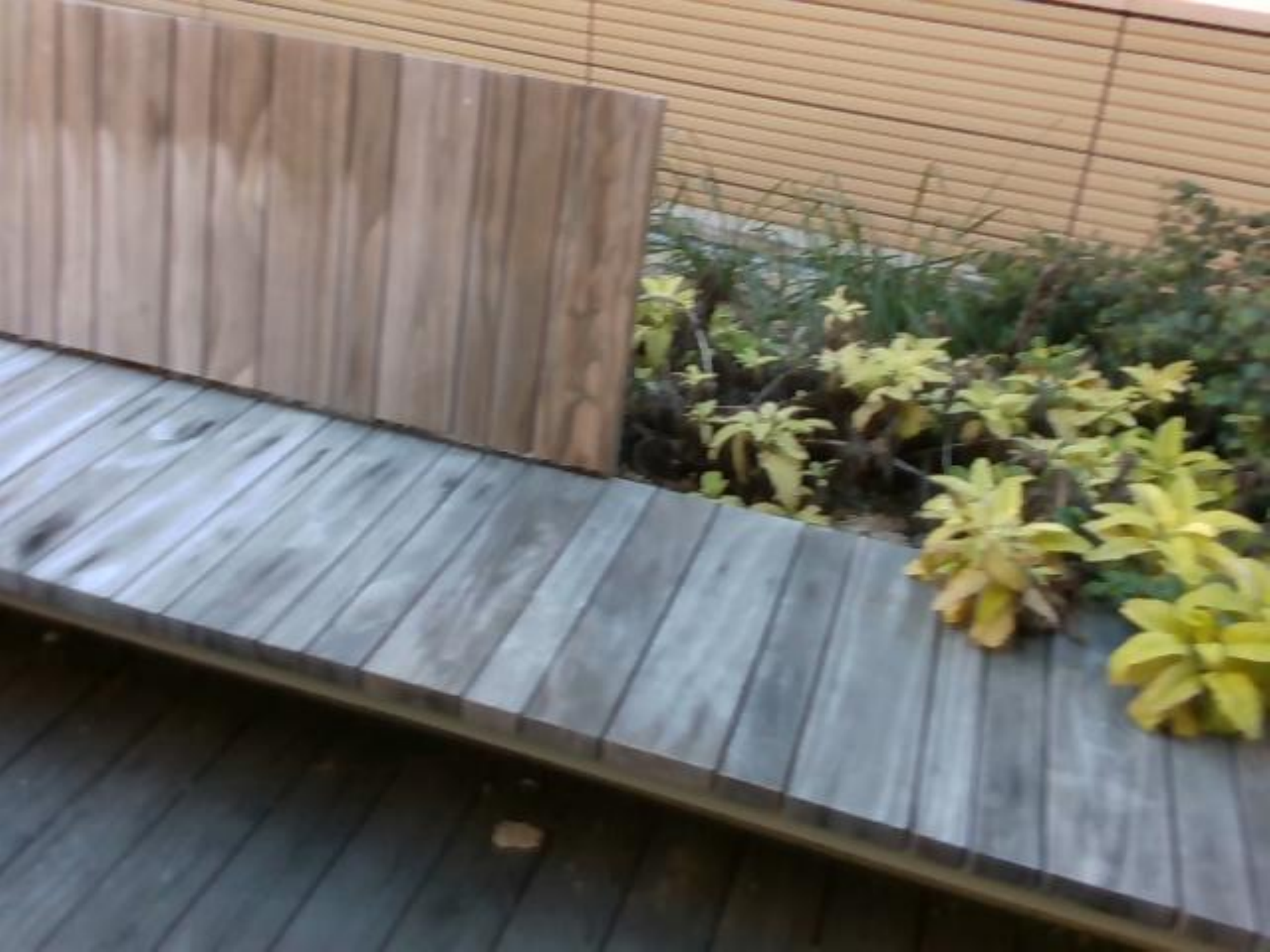}&
    \includegraphics[width=0.104\linewidth]{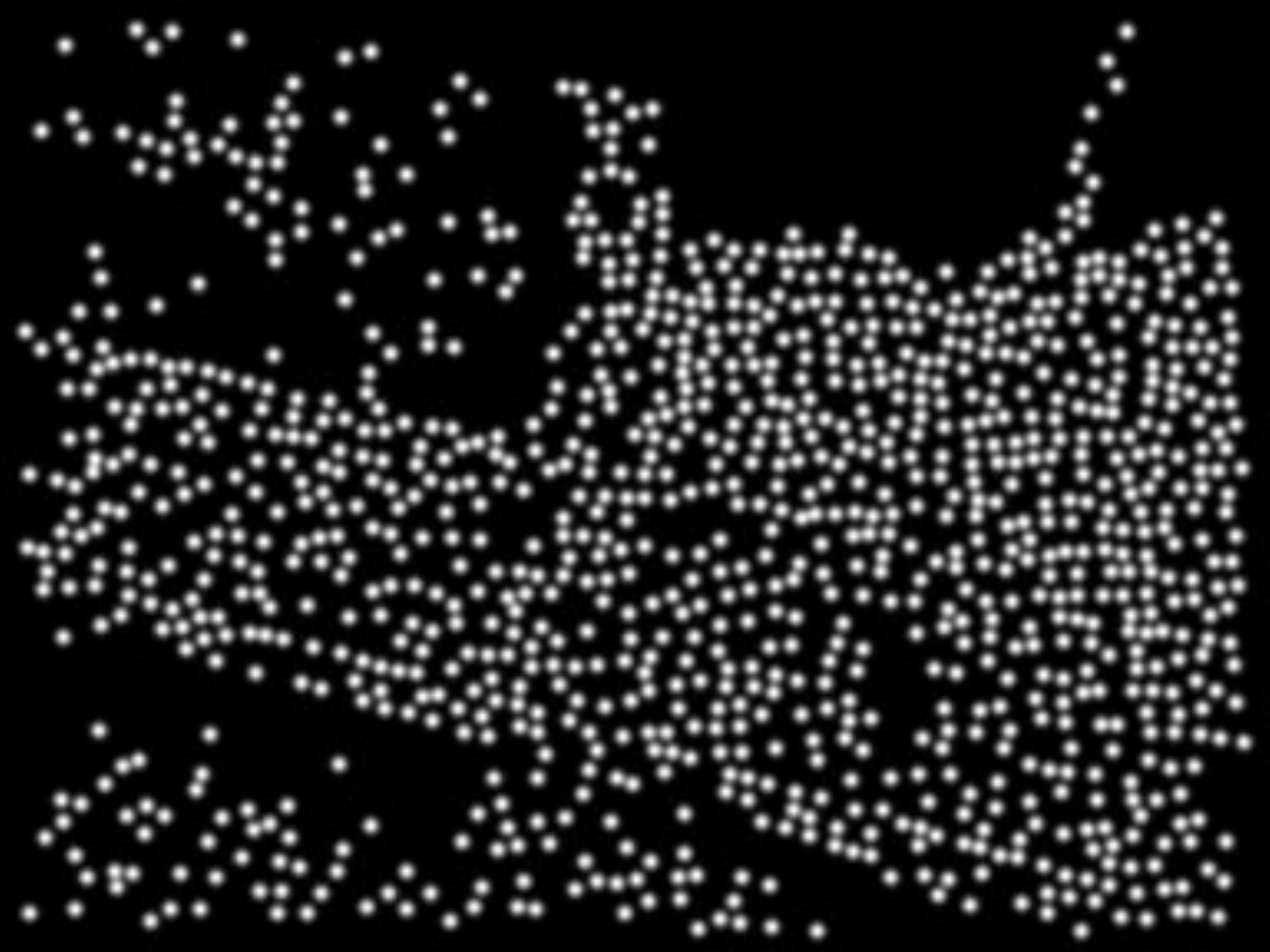}&
    \includegraphics[width=0.104\linewidth]{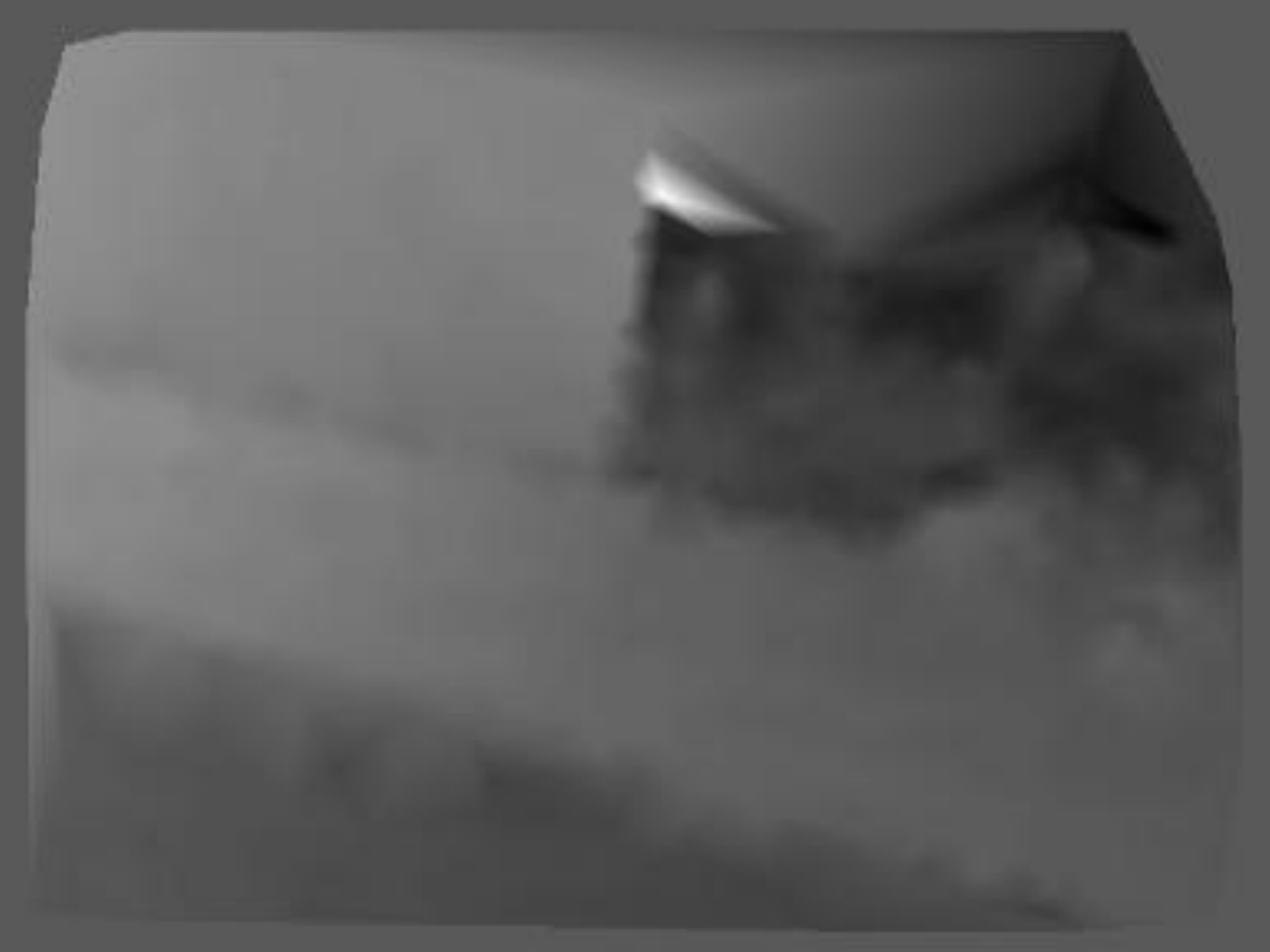}&
    \includegraphics[width=0.104\linewidth]{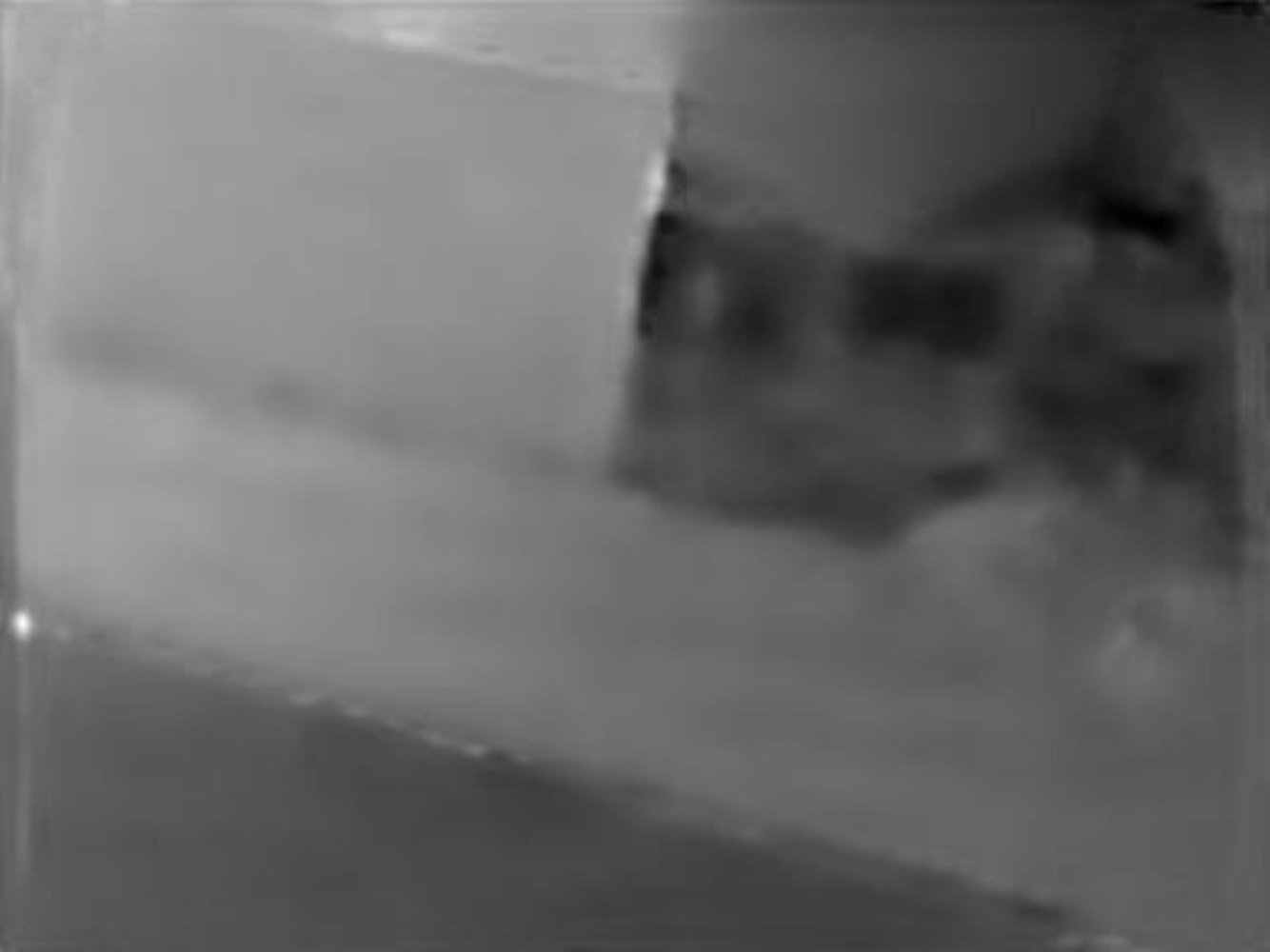}&
    \includegraphics[width=0.104\linewidth]{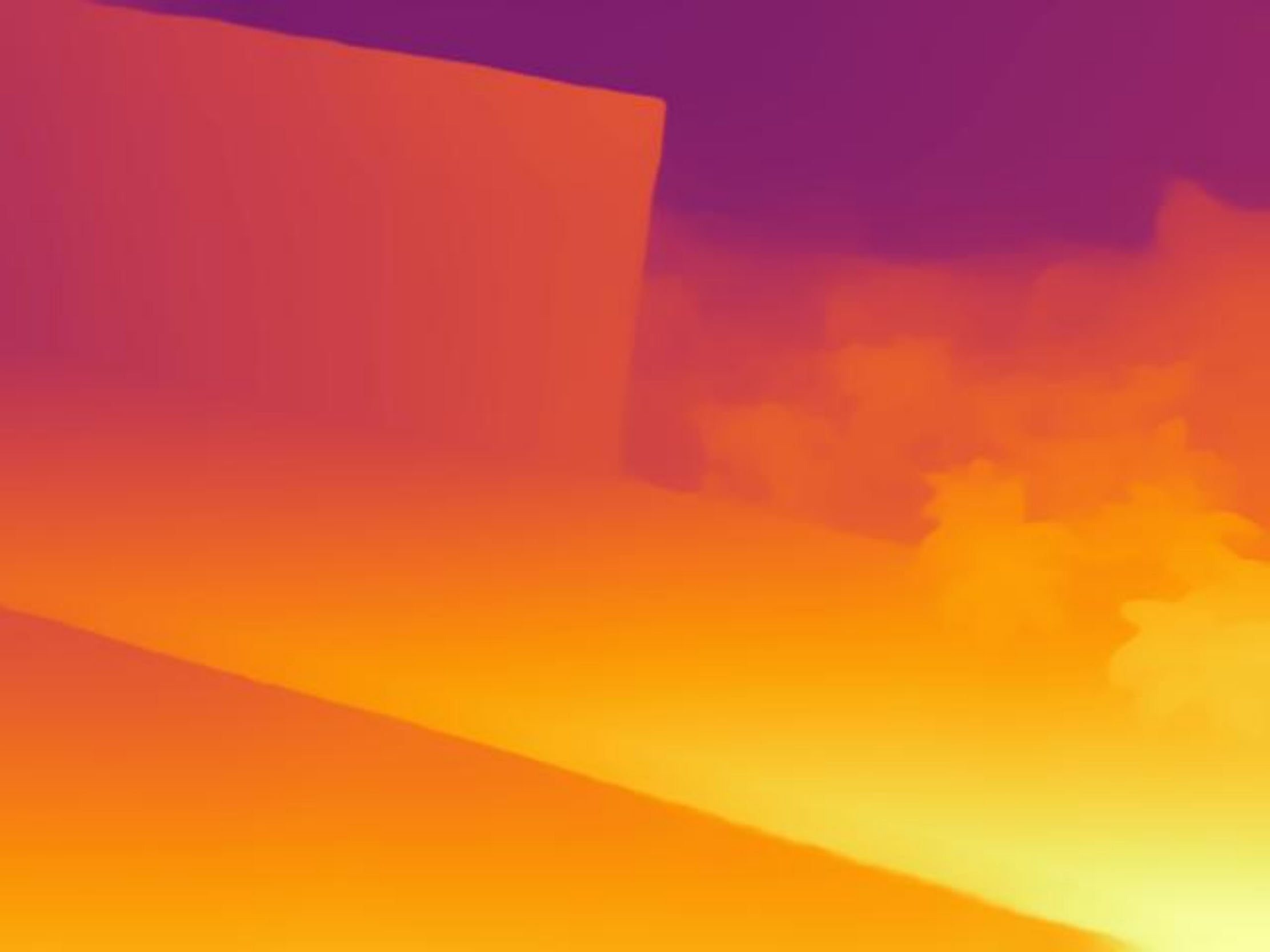}&
    \includegraphics[width=0.104\linewidth]{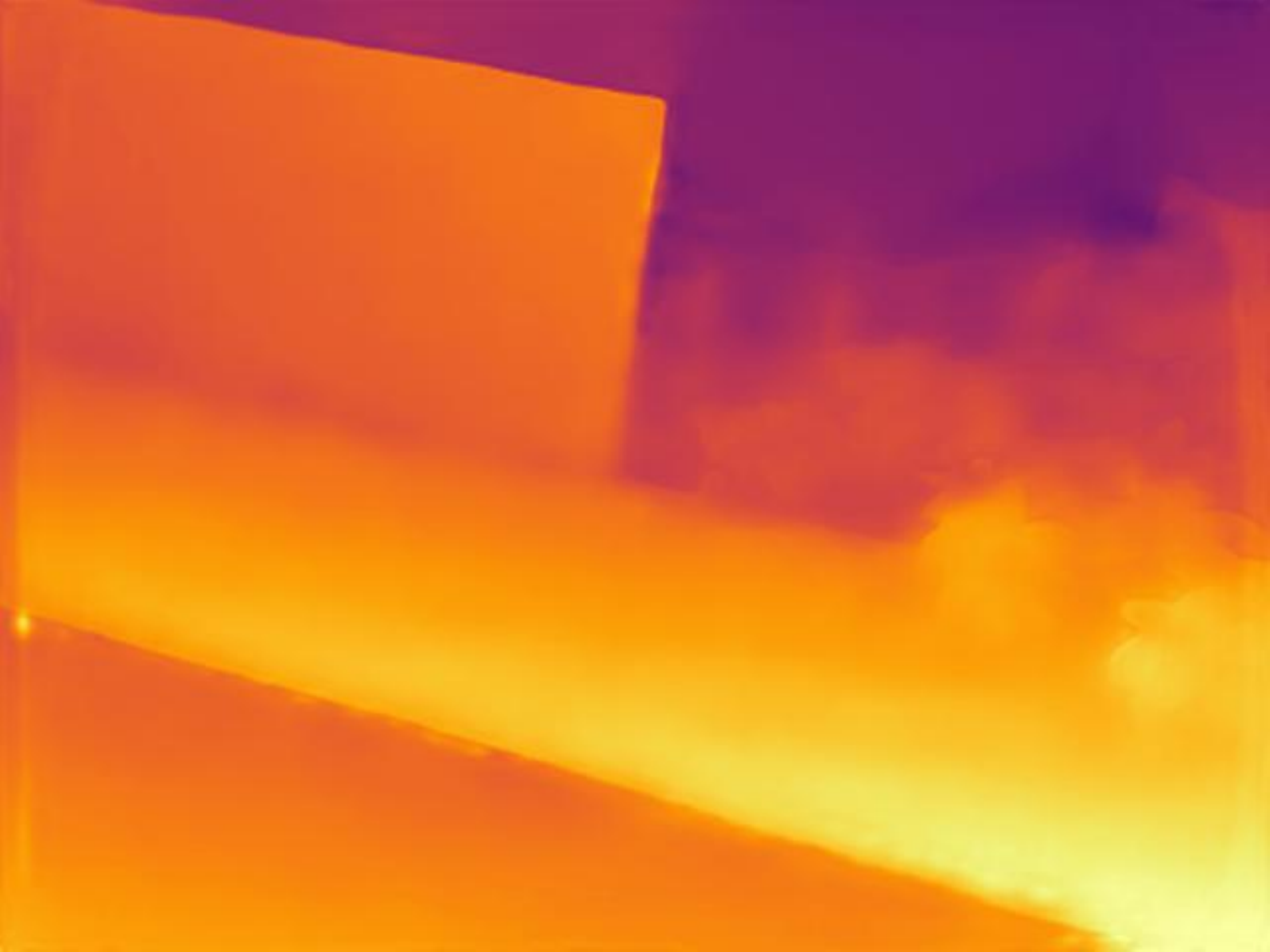}&
    \includegraphics[width=0.104\linewidth]{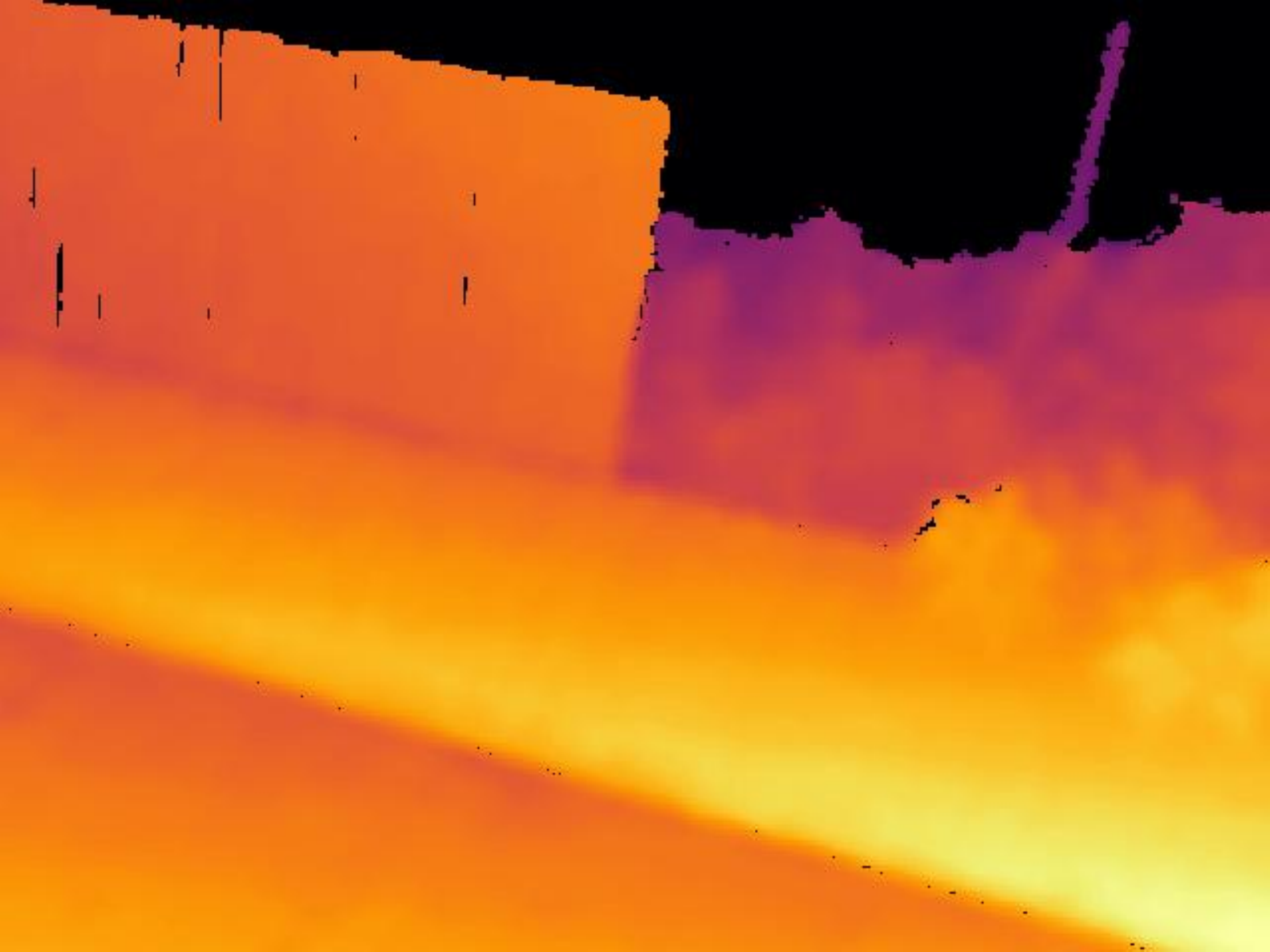}&
    \includegraphics[width=0.104\linewidth]{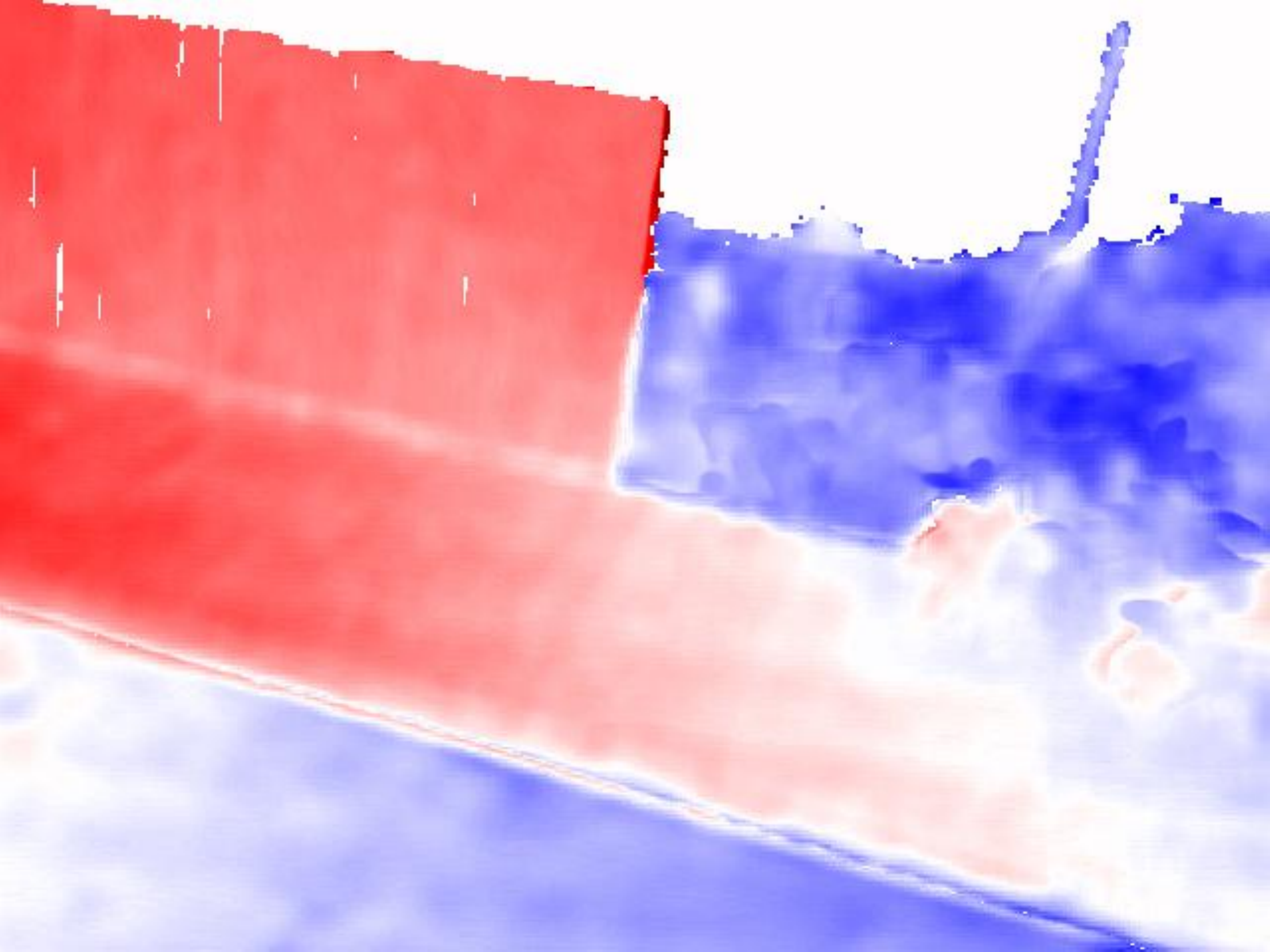}&
    \includegraphics[width=0.104\linewidth]{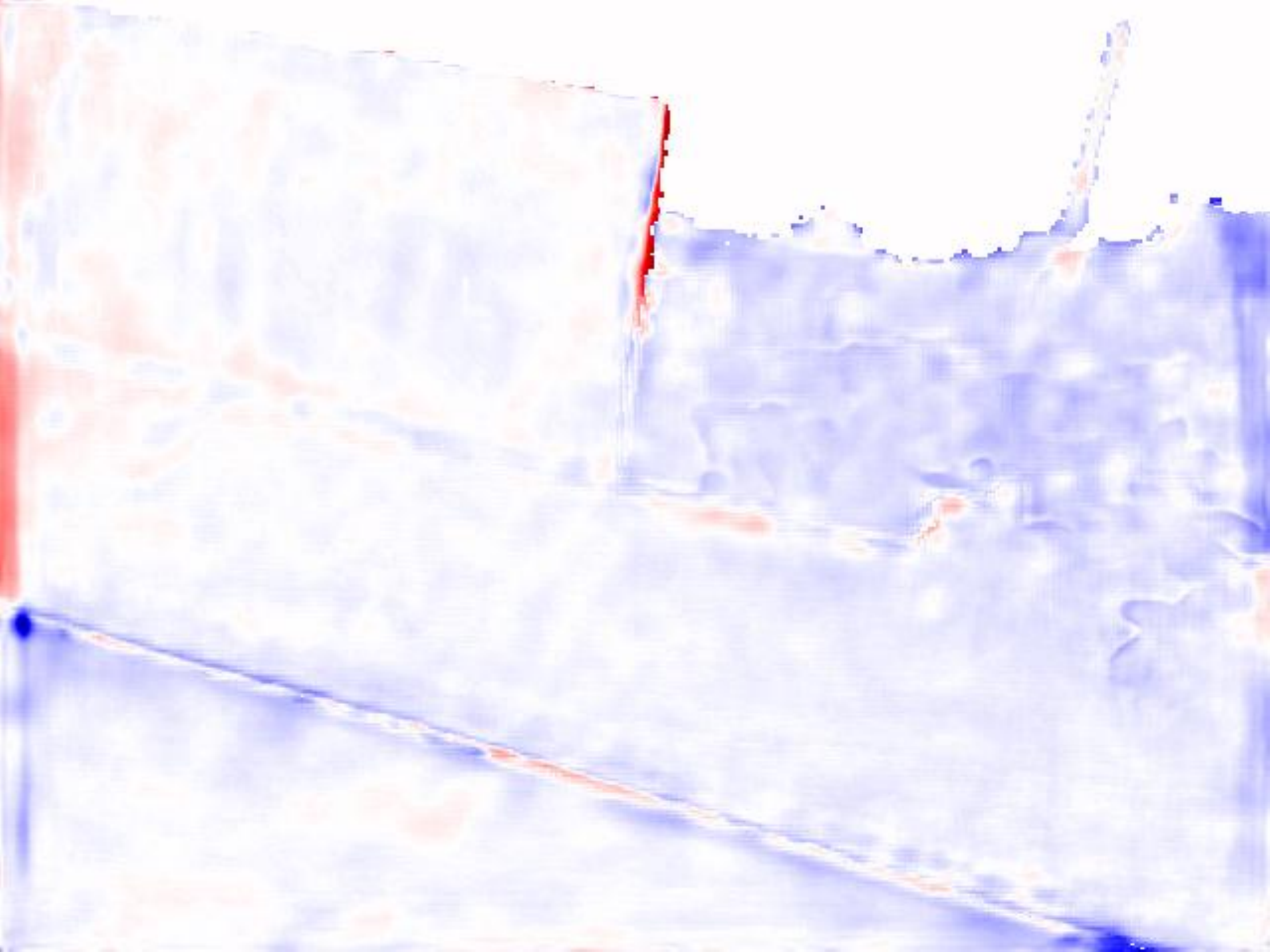}&
    \includegraphics[width=0.024\linewidth]{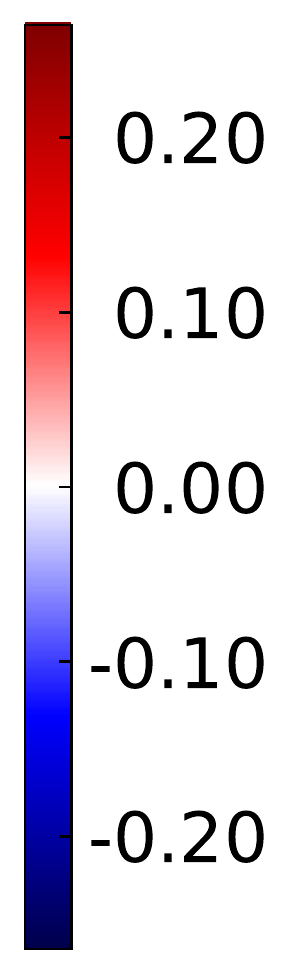}\\
    \multicolumn{9}{c}{} \\
    \vspace{-0.75mm}
    \rot{\scriptsize VOID 150} &
    \includegraphics[width=0.104\linewidth]{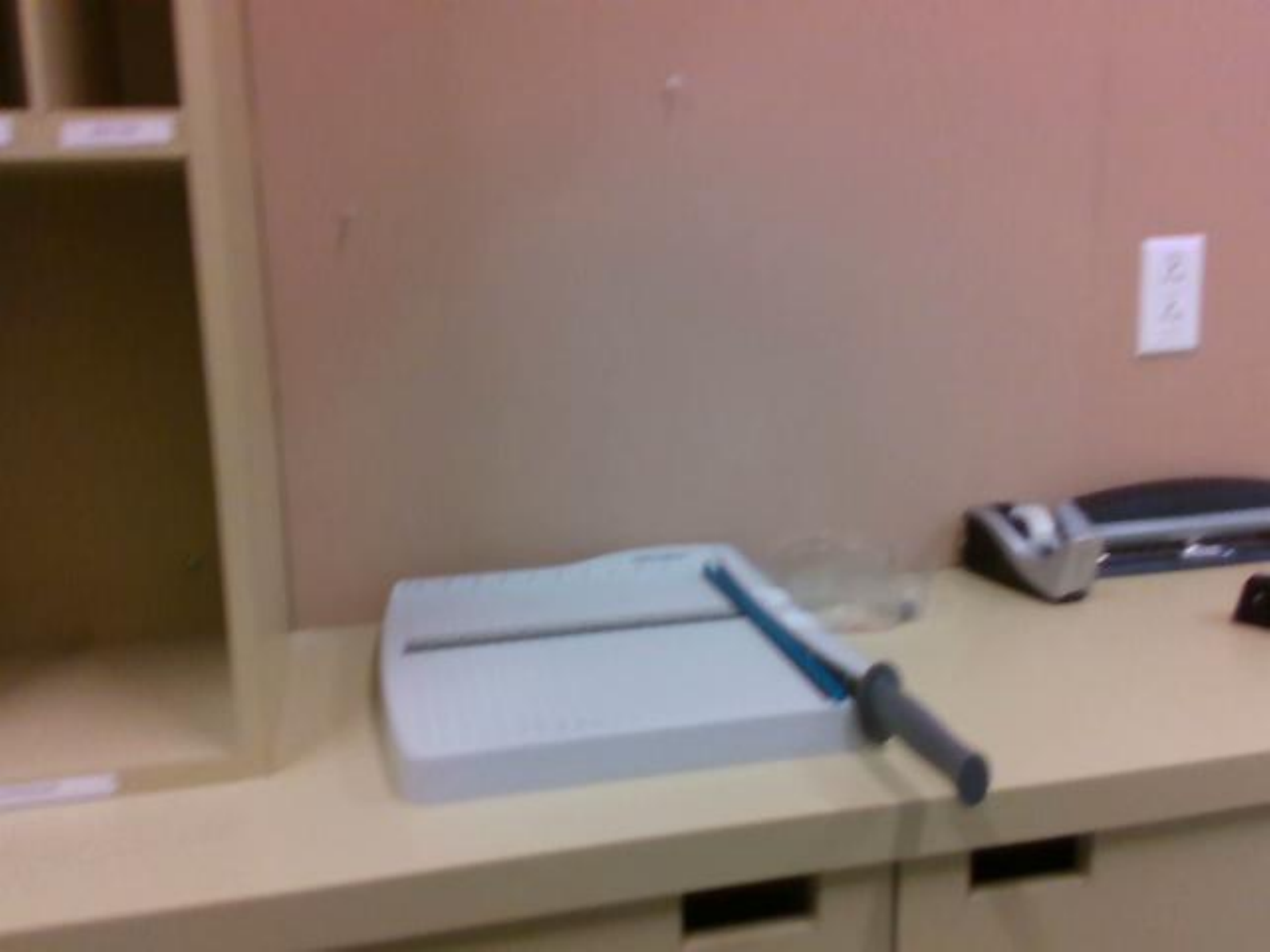}&
    \includegraphics[width=0.104\linewidth]{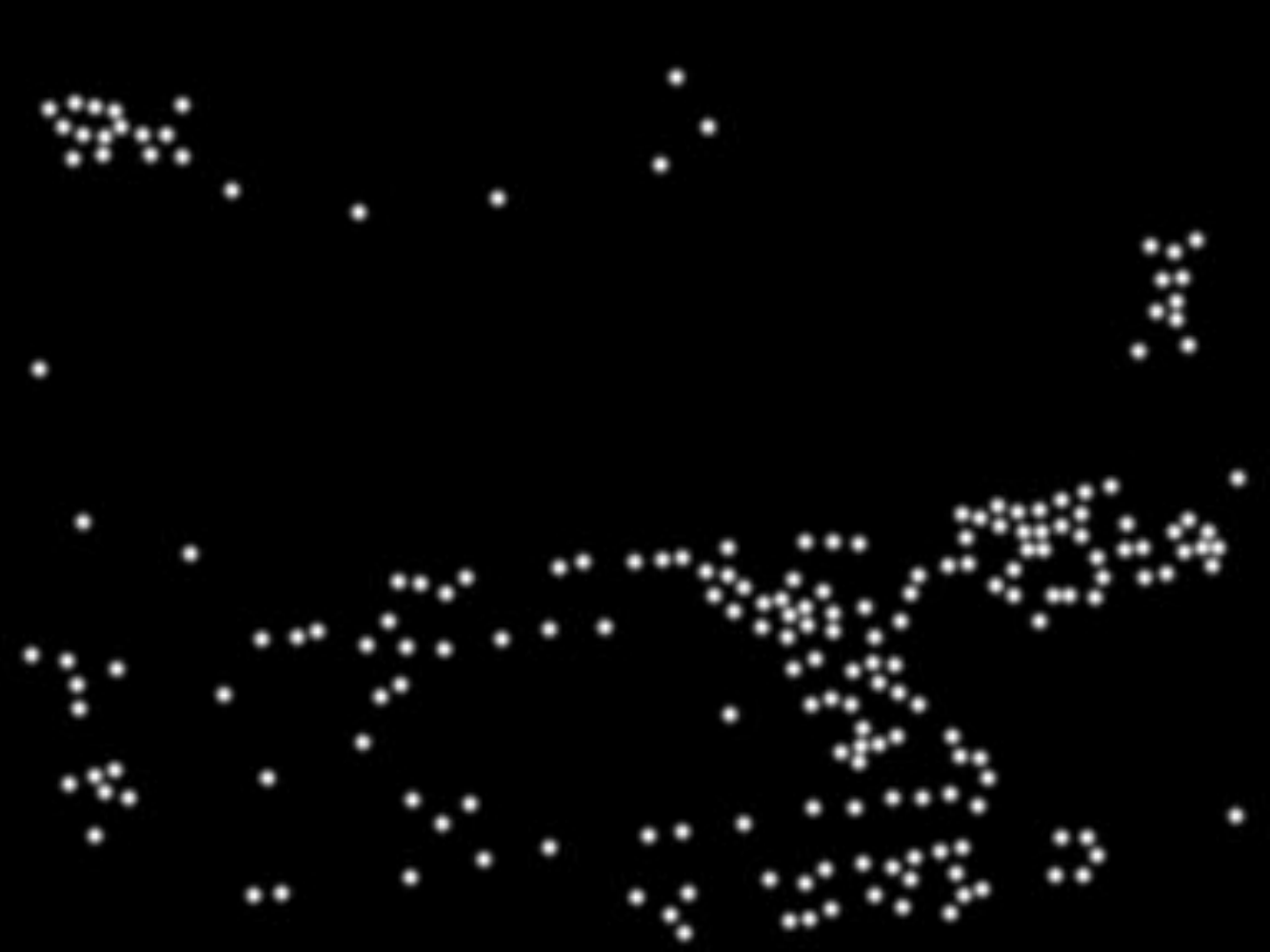}&
    \includegraphics[width=0.104\linewidth]{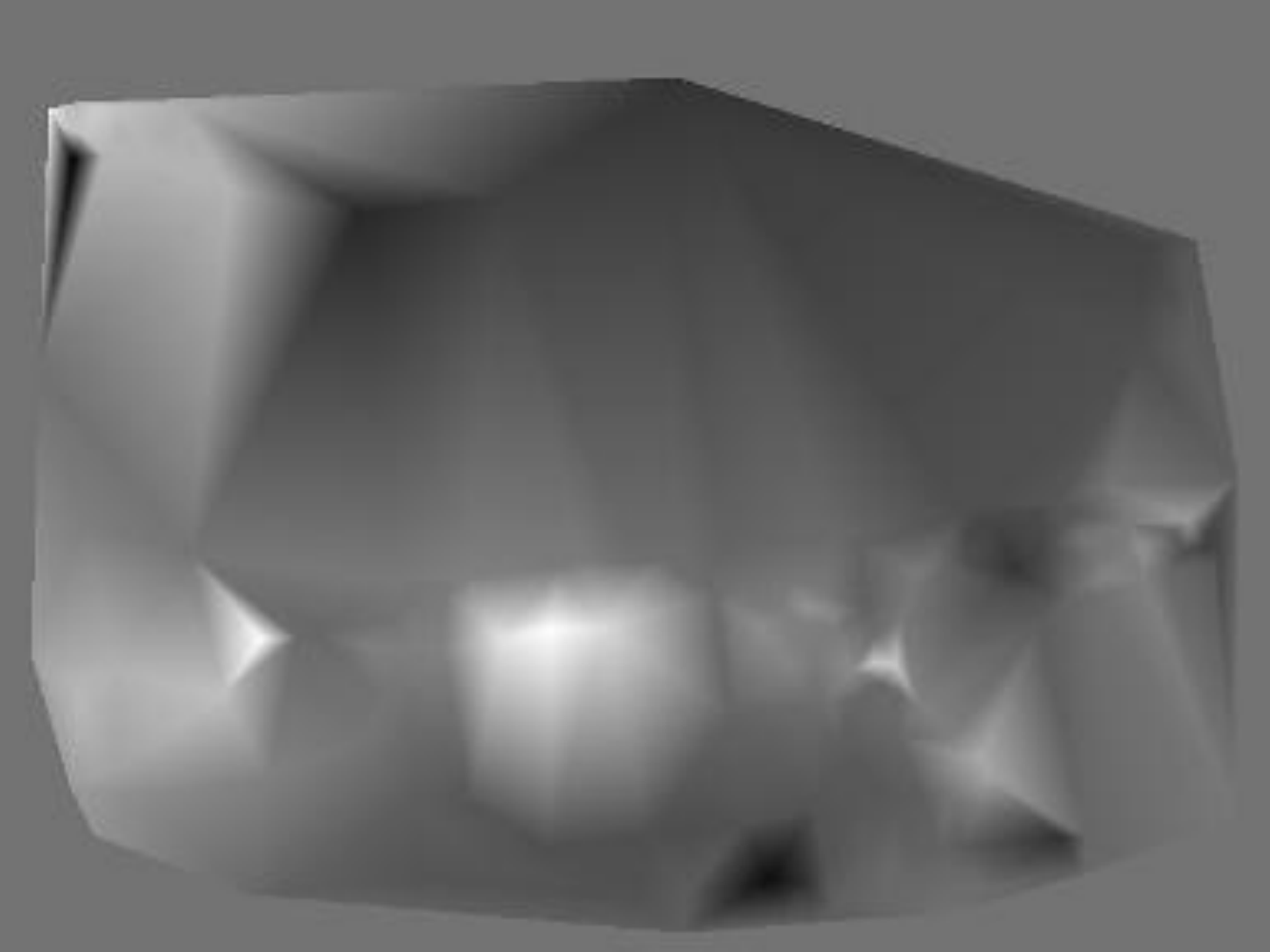}&
    \includegraphics[width=0.104\linewidth]{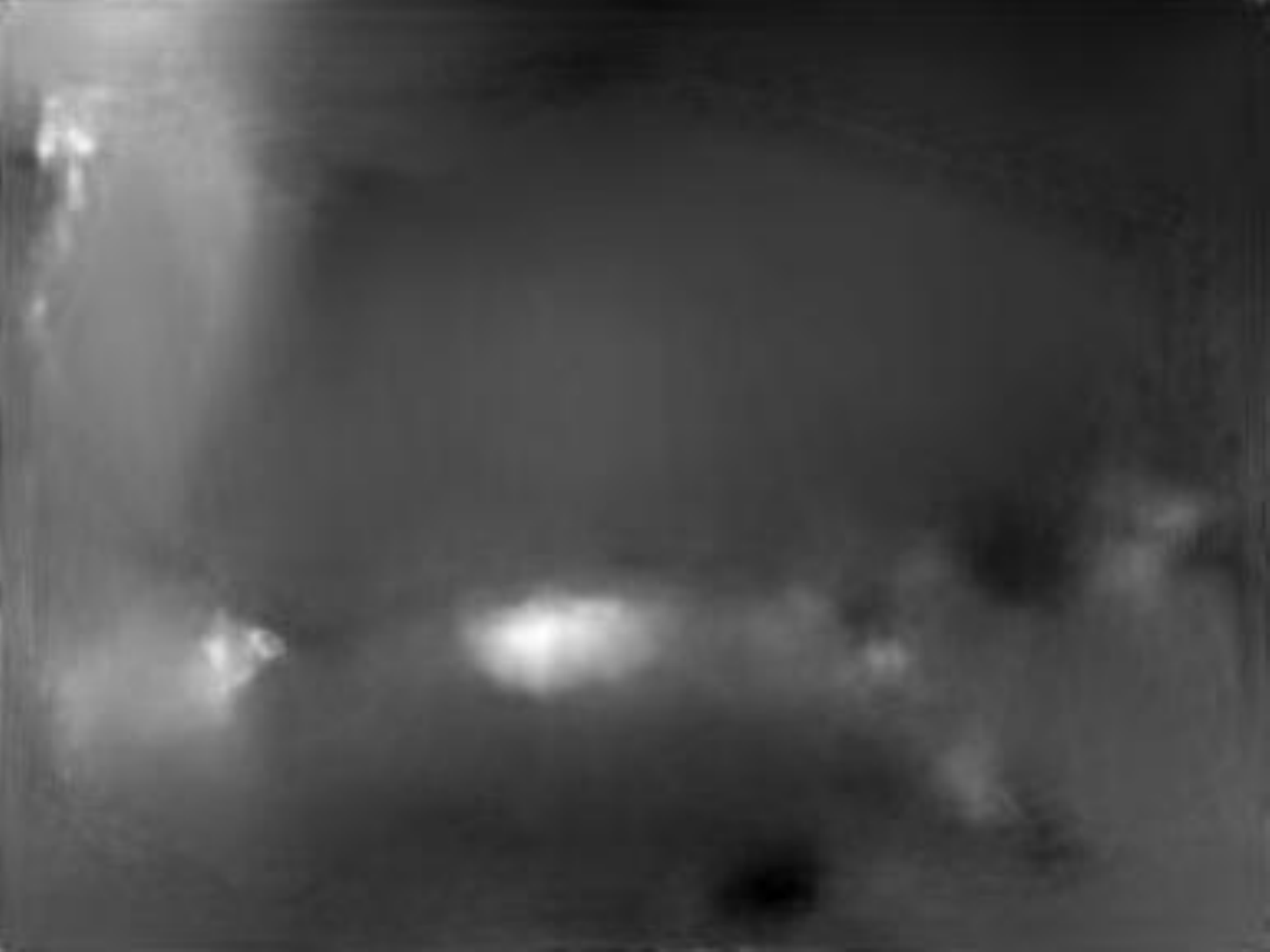}&
    \includegraphics[width=0.104\linewidth]{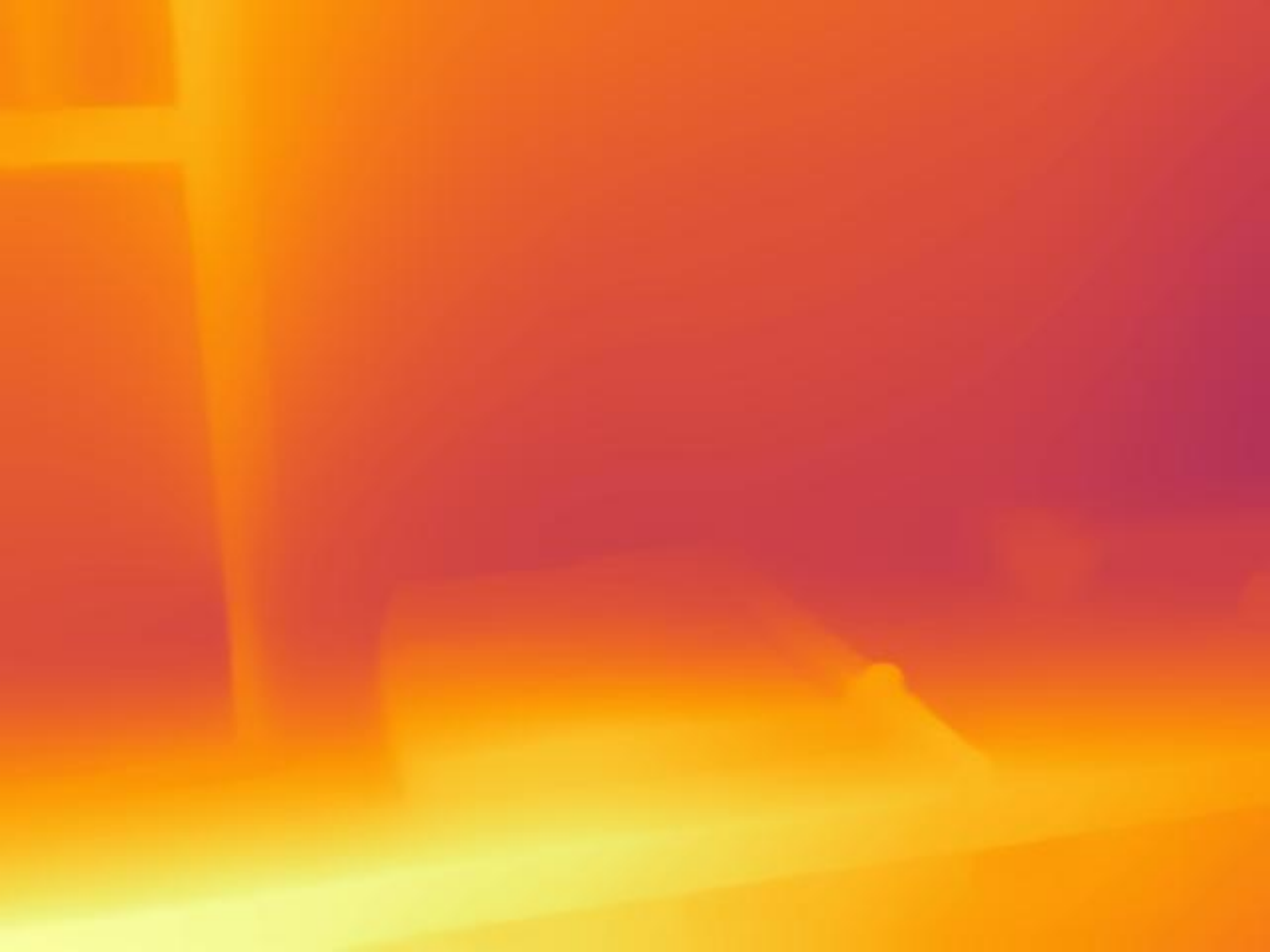}&
    \includegraphics[width=0.104\linewidth]{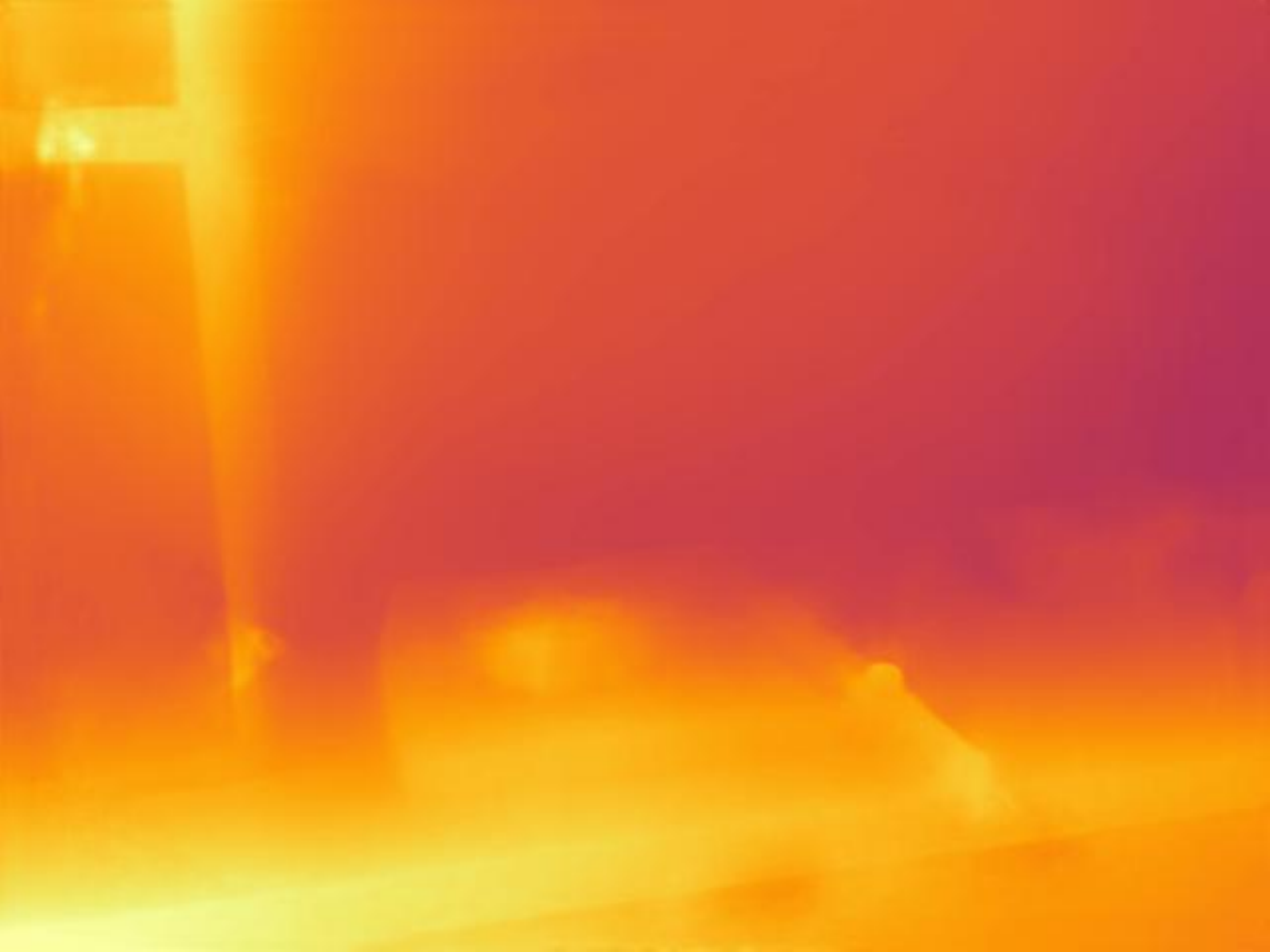}&
    \includegraphics[width=0.104\linewidth]{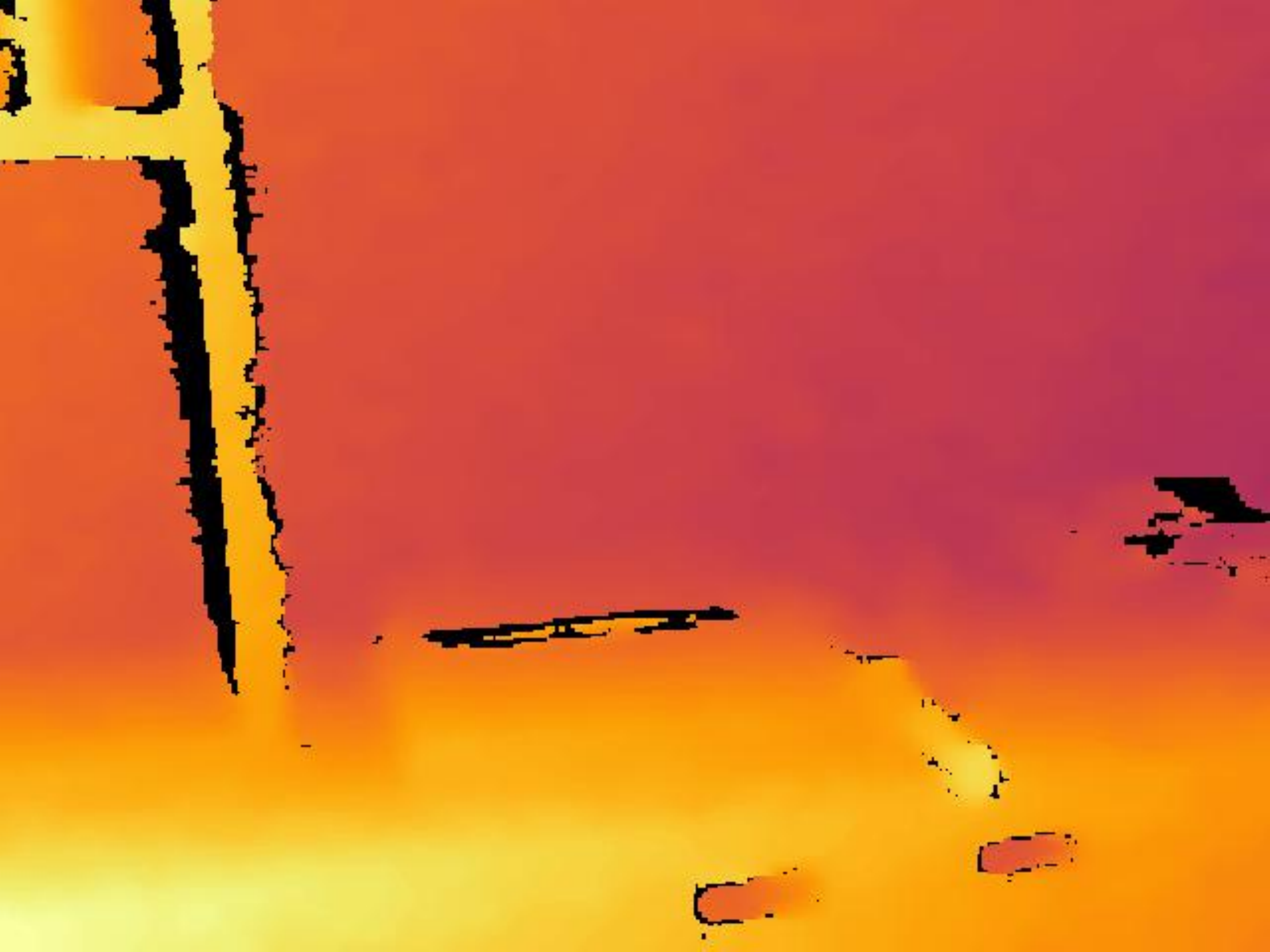}&
    \includegraphics[width=0.104\linewidth]{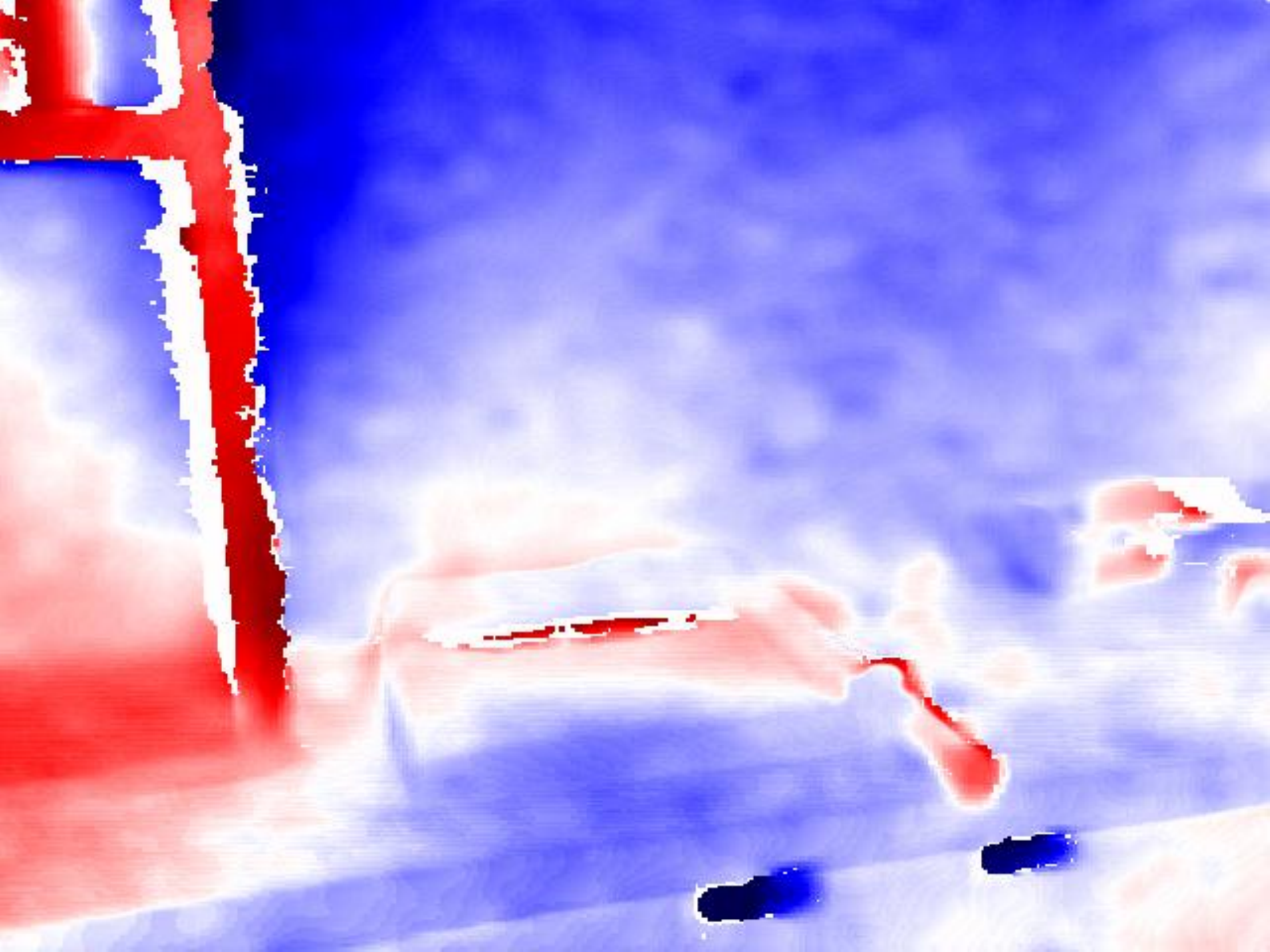}&
    \includegraphics[width=0.104\linewidth]{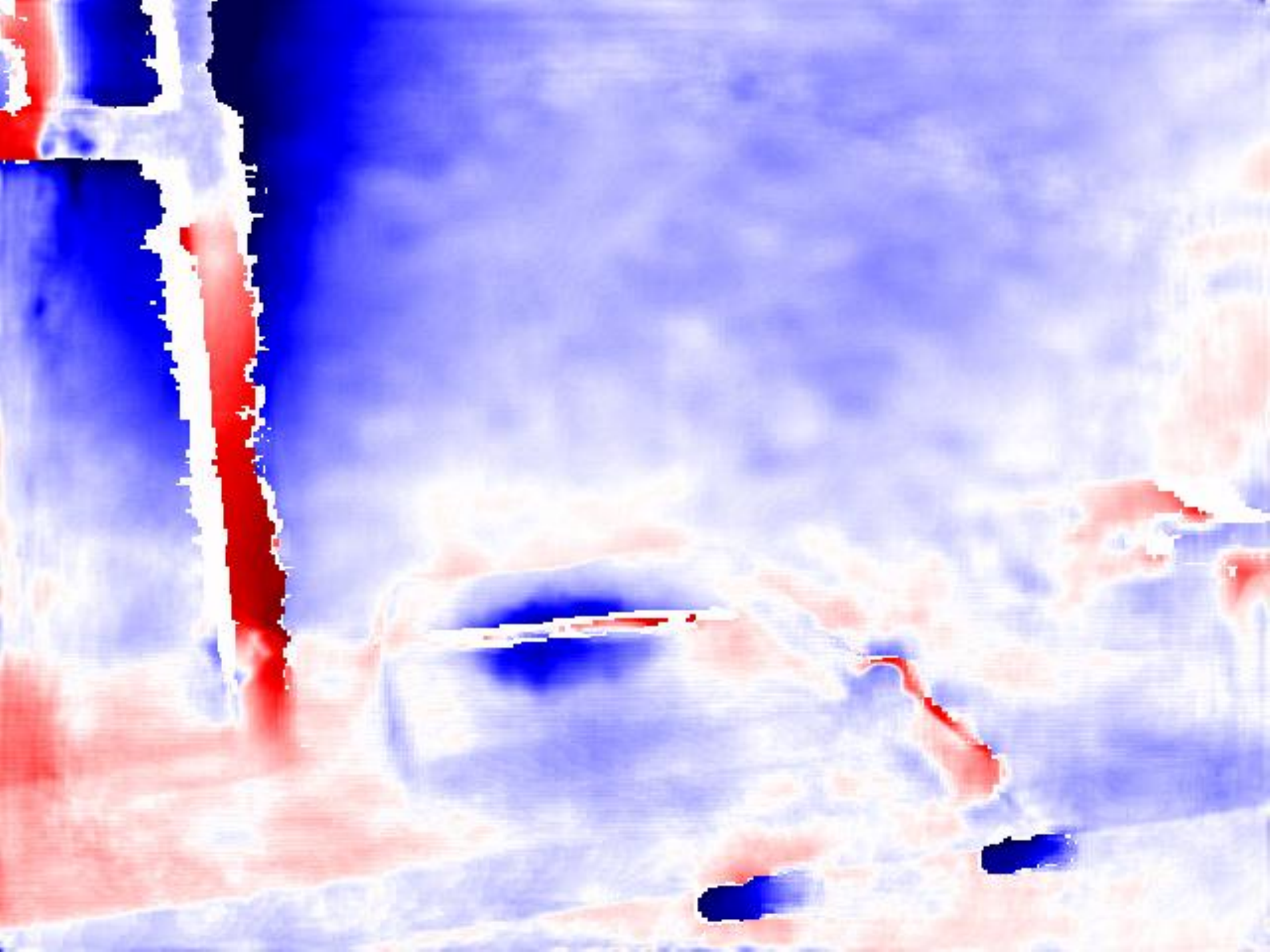}&
    \includegraphics[width=0.024\linewidth]{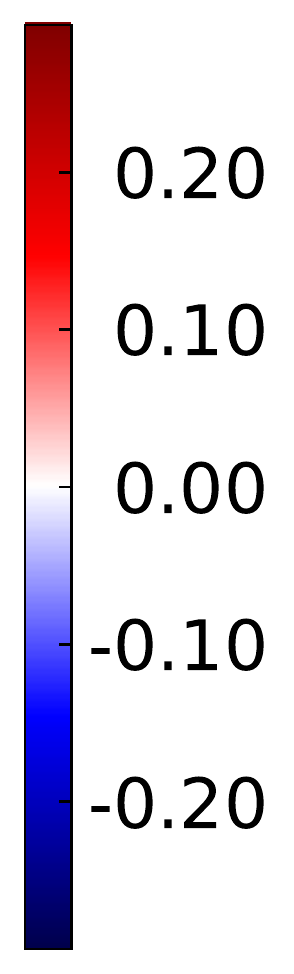}\\
    \vspace{-0.75mm}
    \rot{\scriptsize VOID 500} &
    \includegraphics[width=0.104\linewidth]{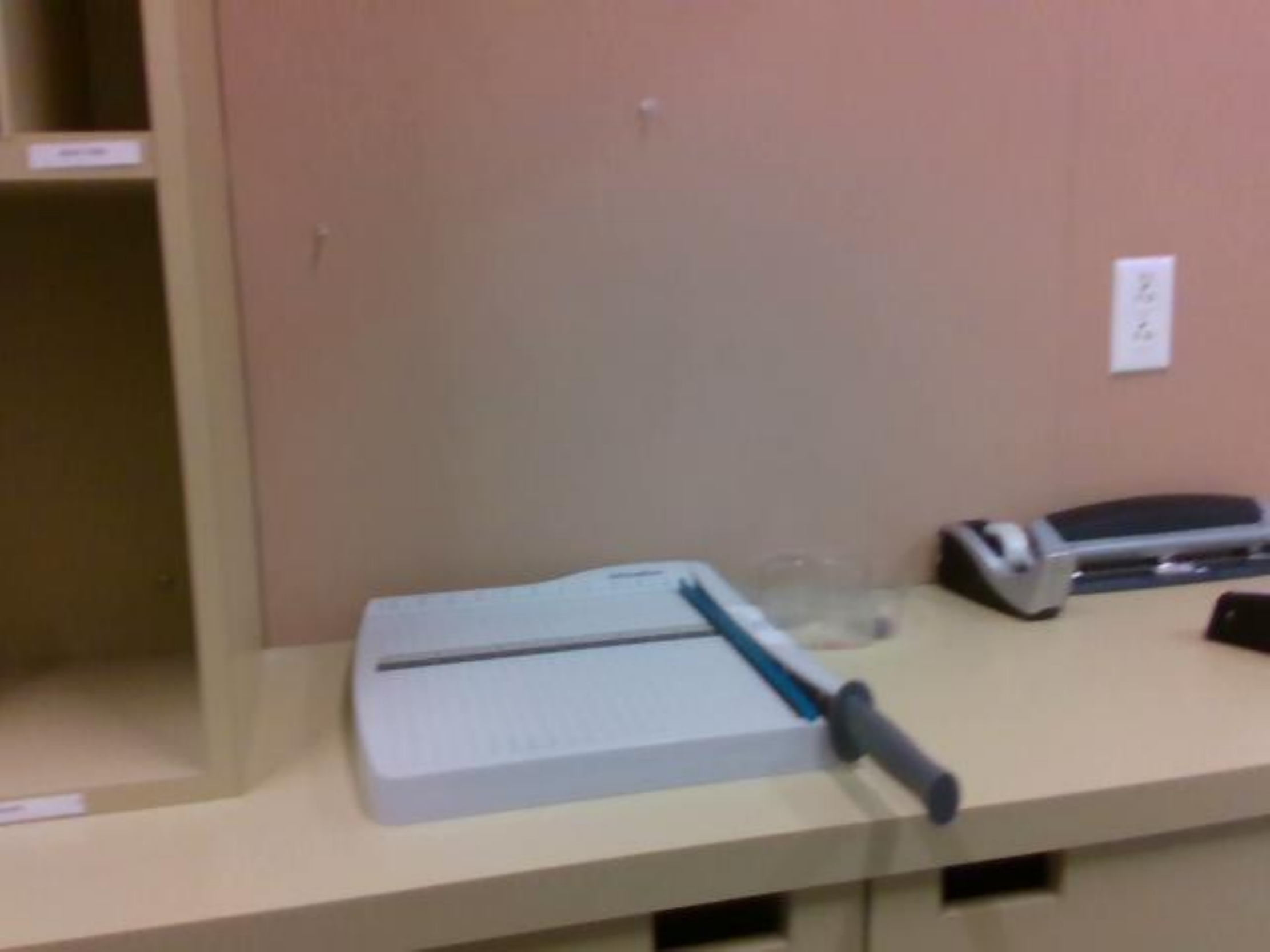}&
    \includegraphics[width=0.104\linewidth]{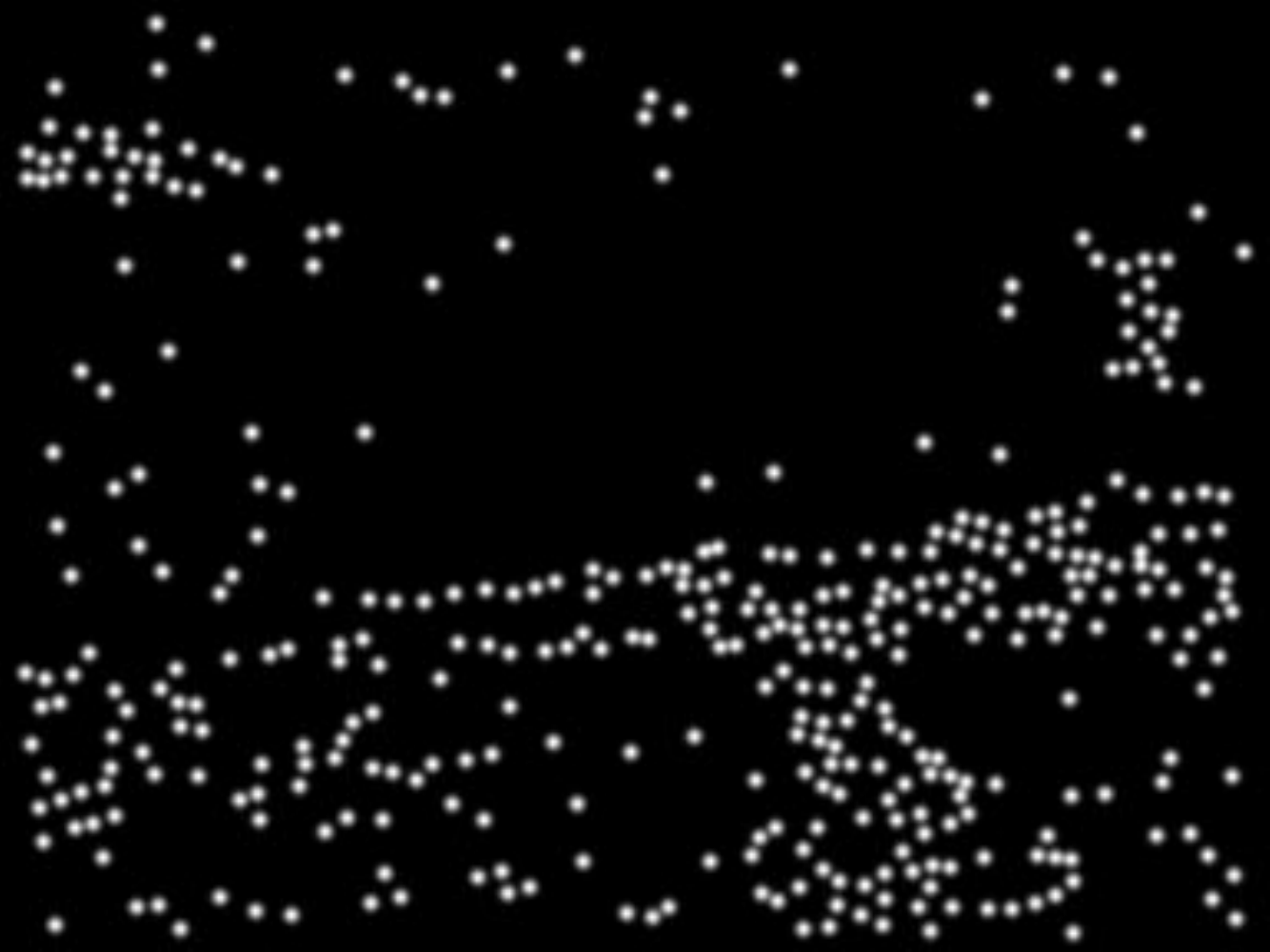}&
    \includegraphics[width=0.104\linewidth]{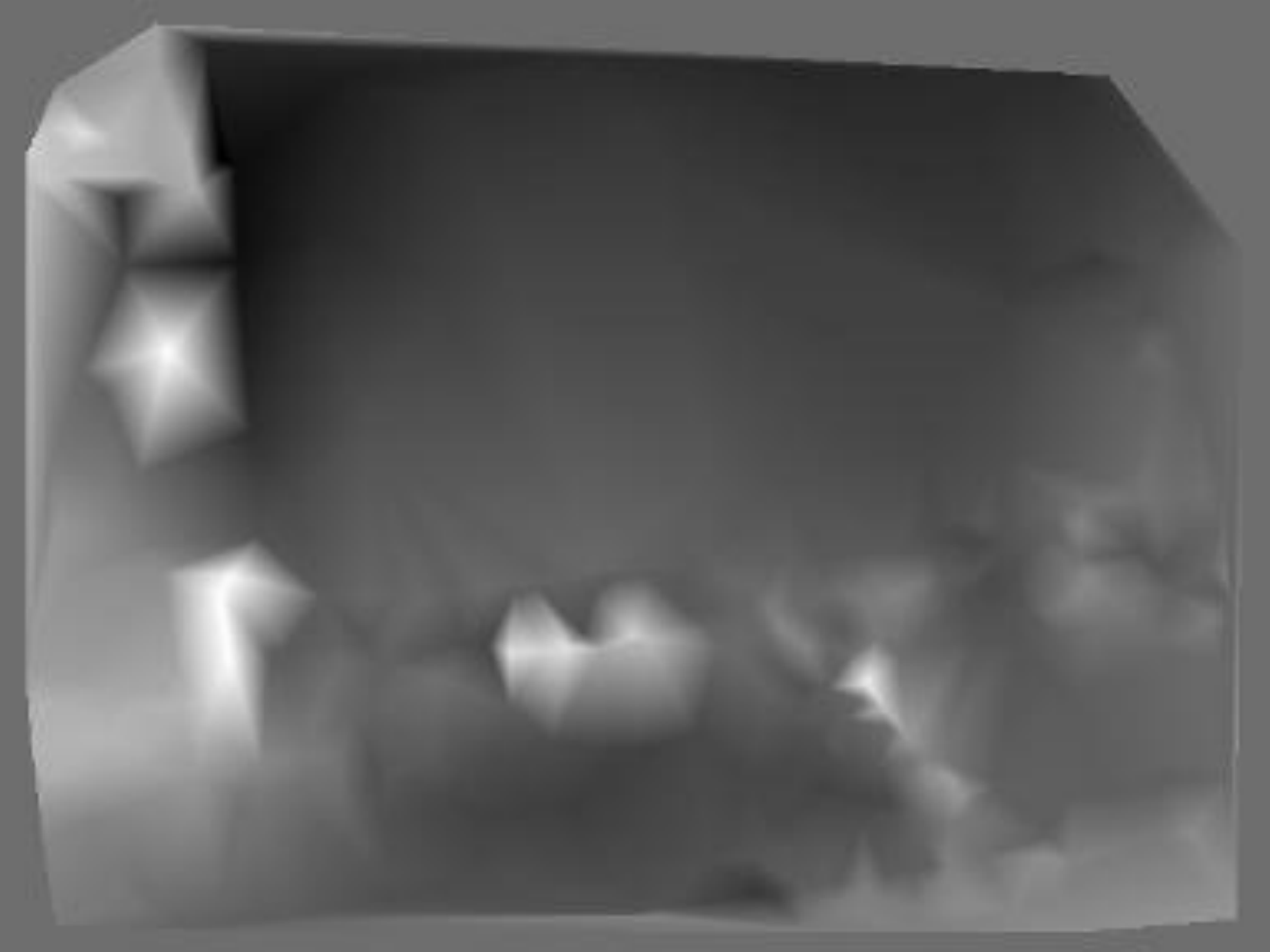}&
    \includegraphics[width=0.104\linewidth]{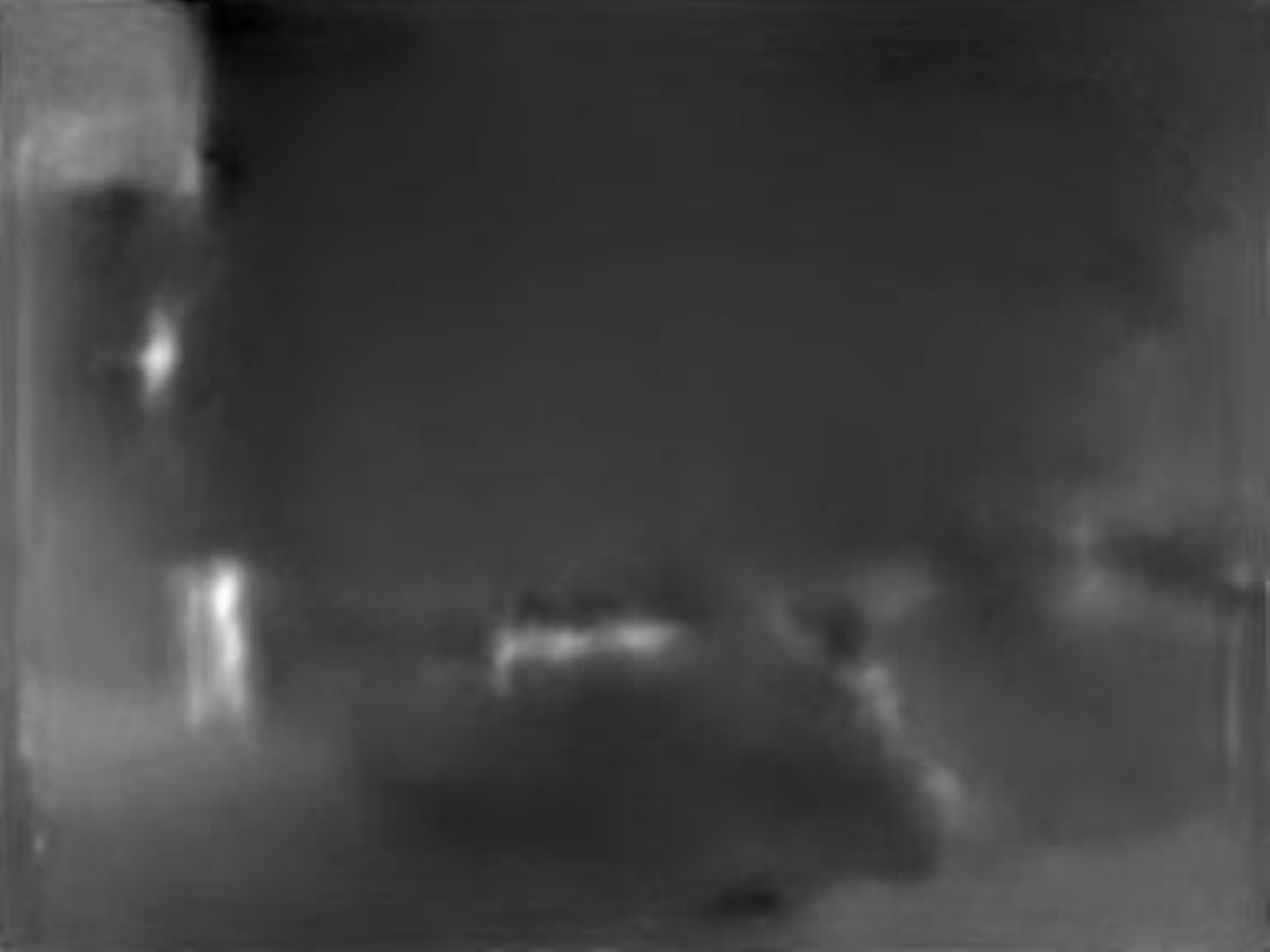}&
    \includegraphics[width=0.104\linewidth]{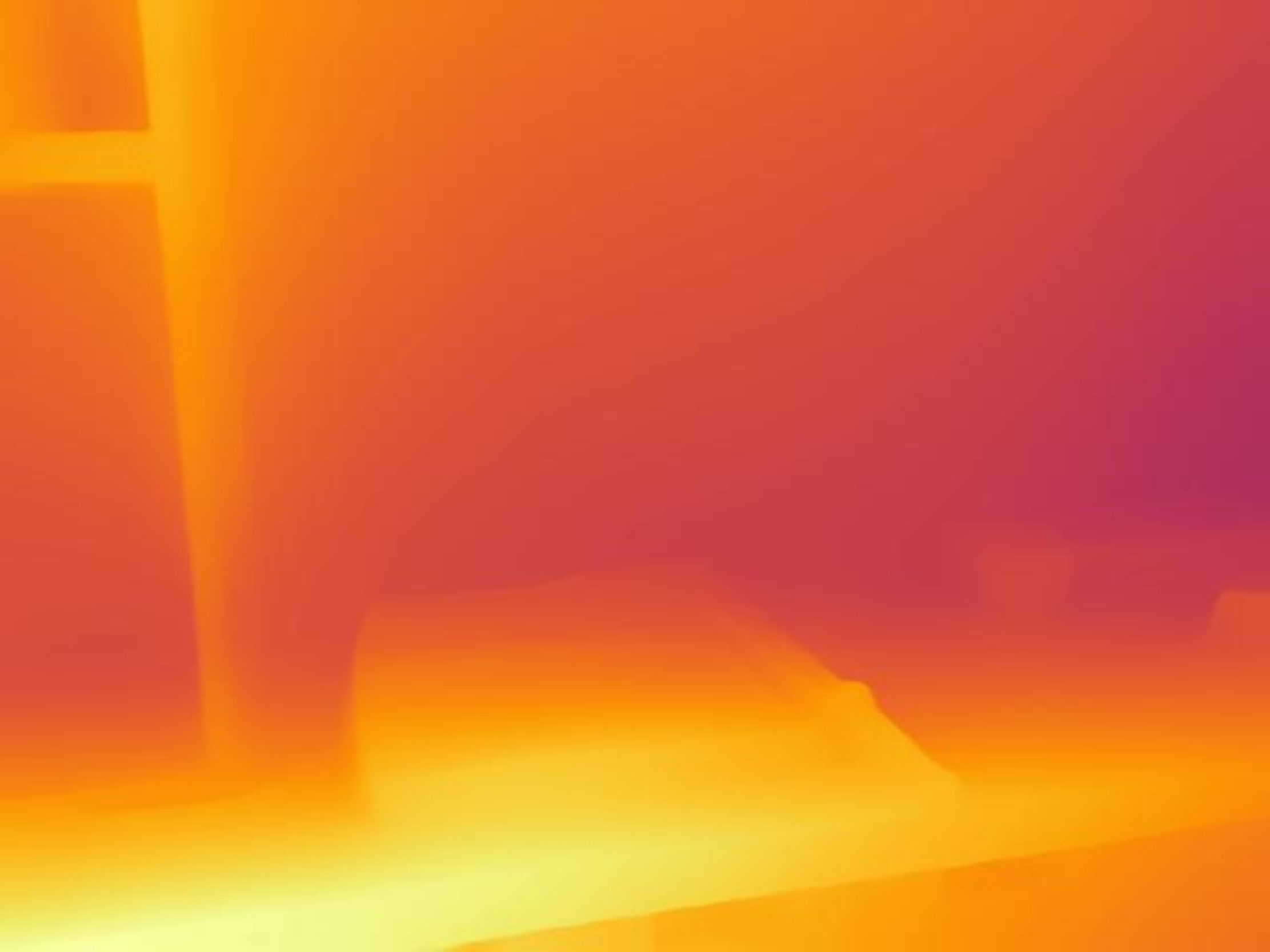}&
    \includegraphics[width=0.104\linewidth]{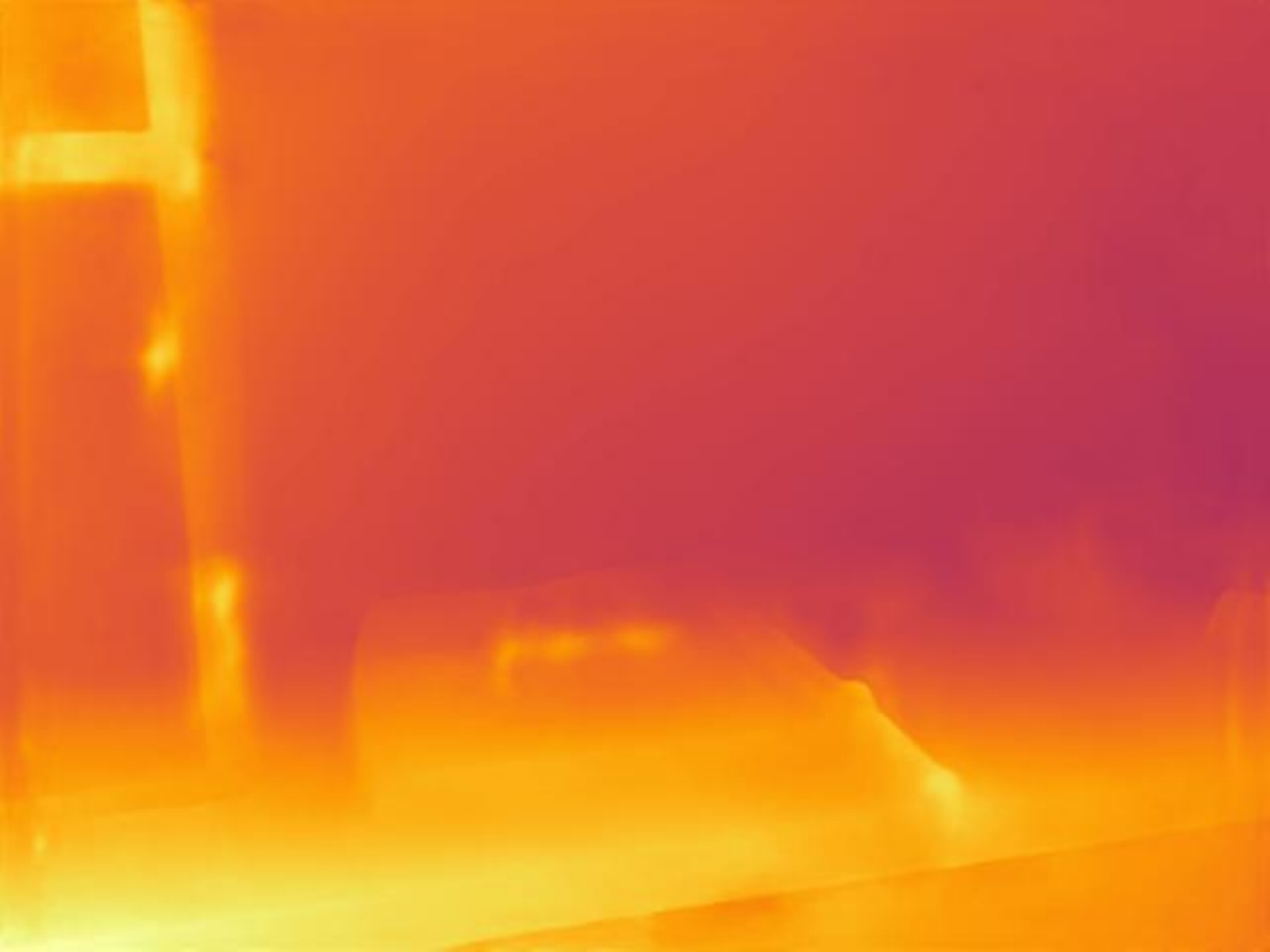}&
    \includegraphics[width=0.104\linewidth]{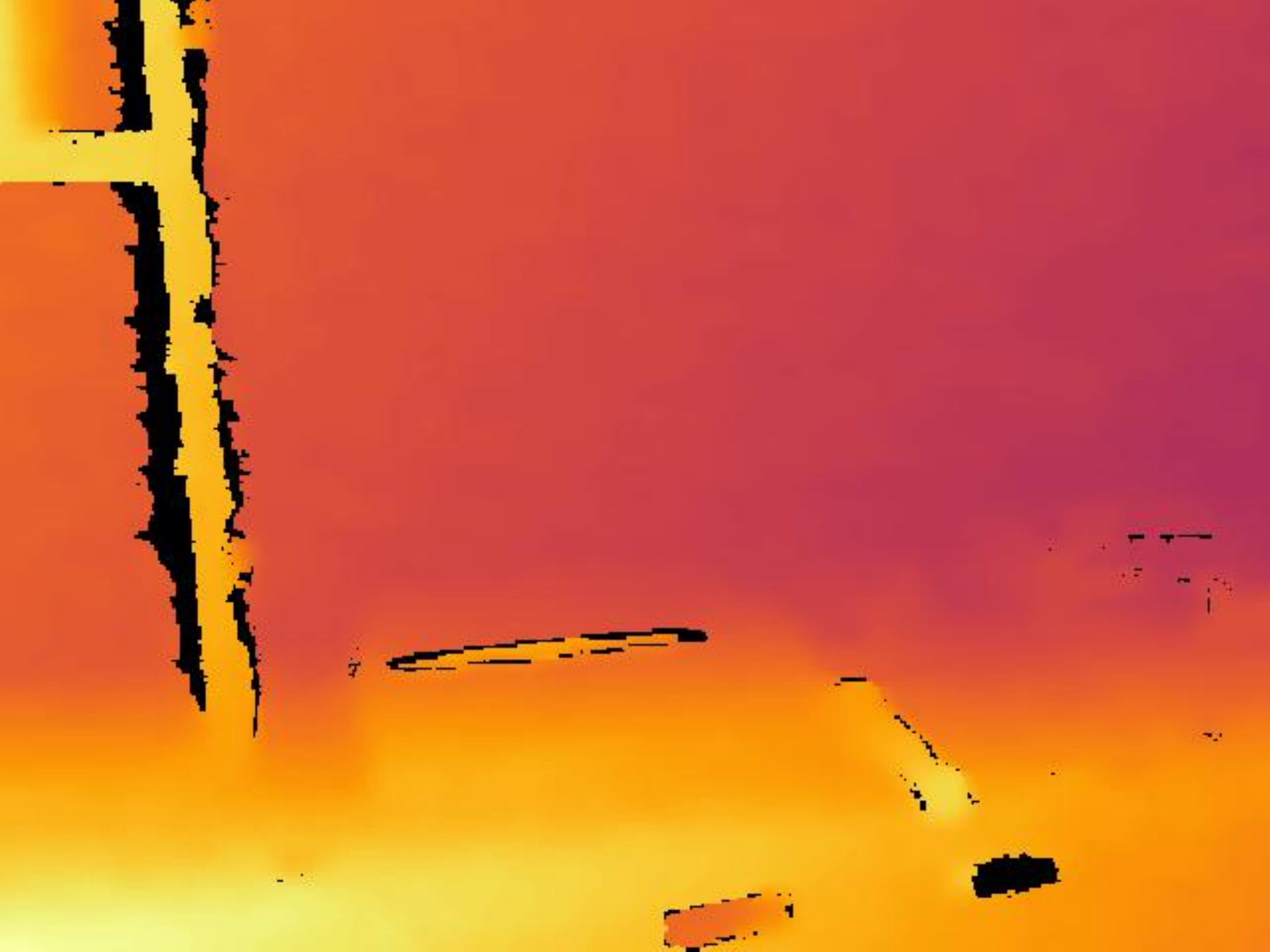}&
    \includegraphics[width=0.104\linewidth]{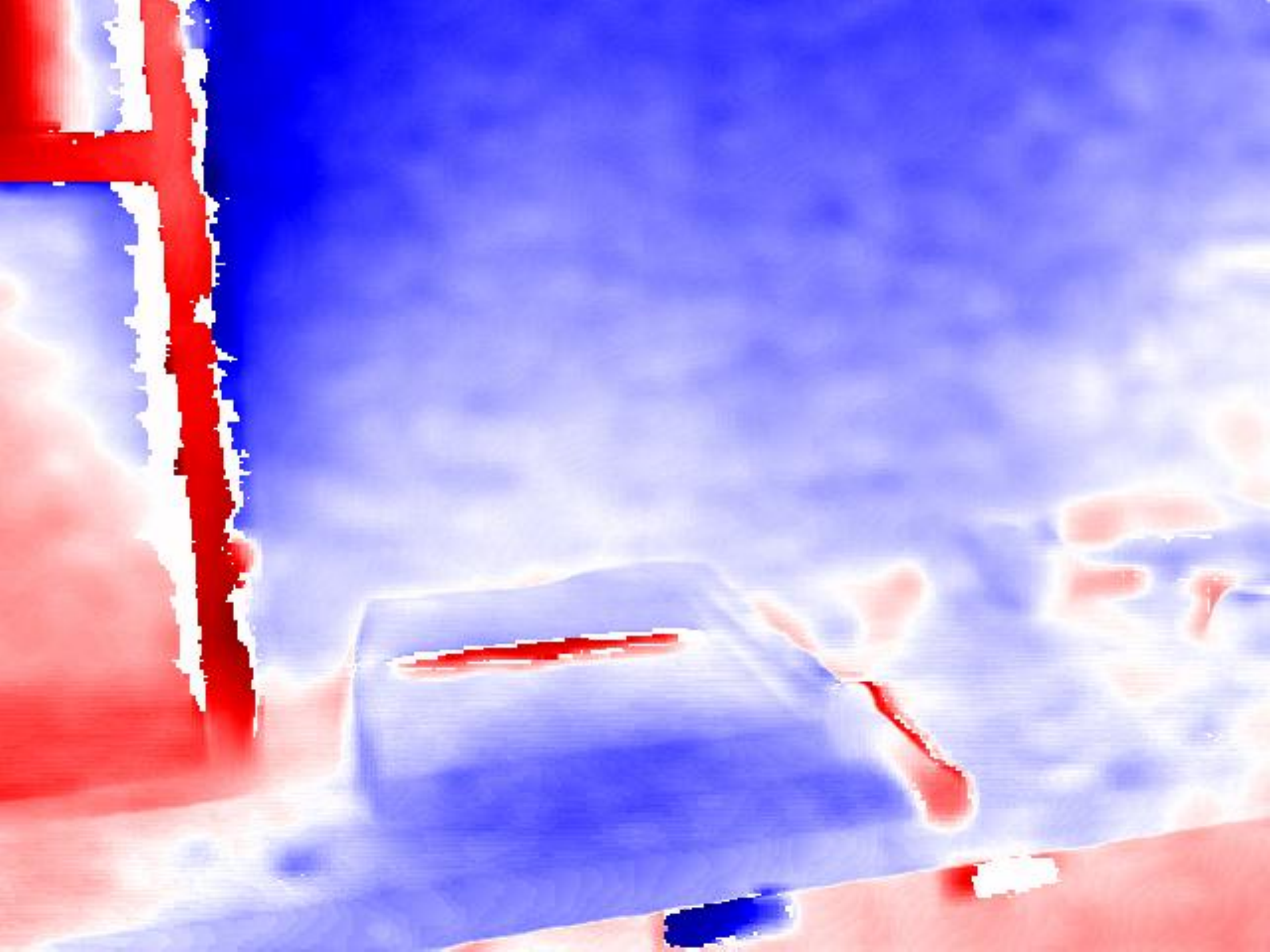}&
    \includegraphics[width=0.104\linewidth]{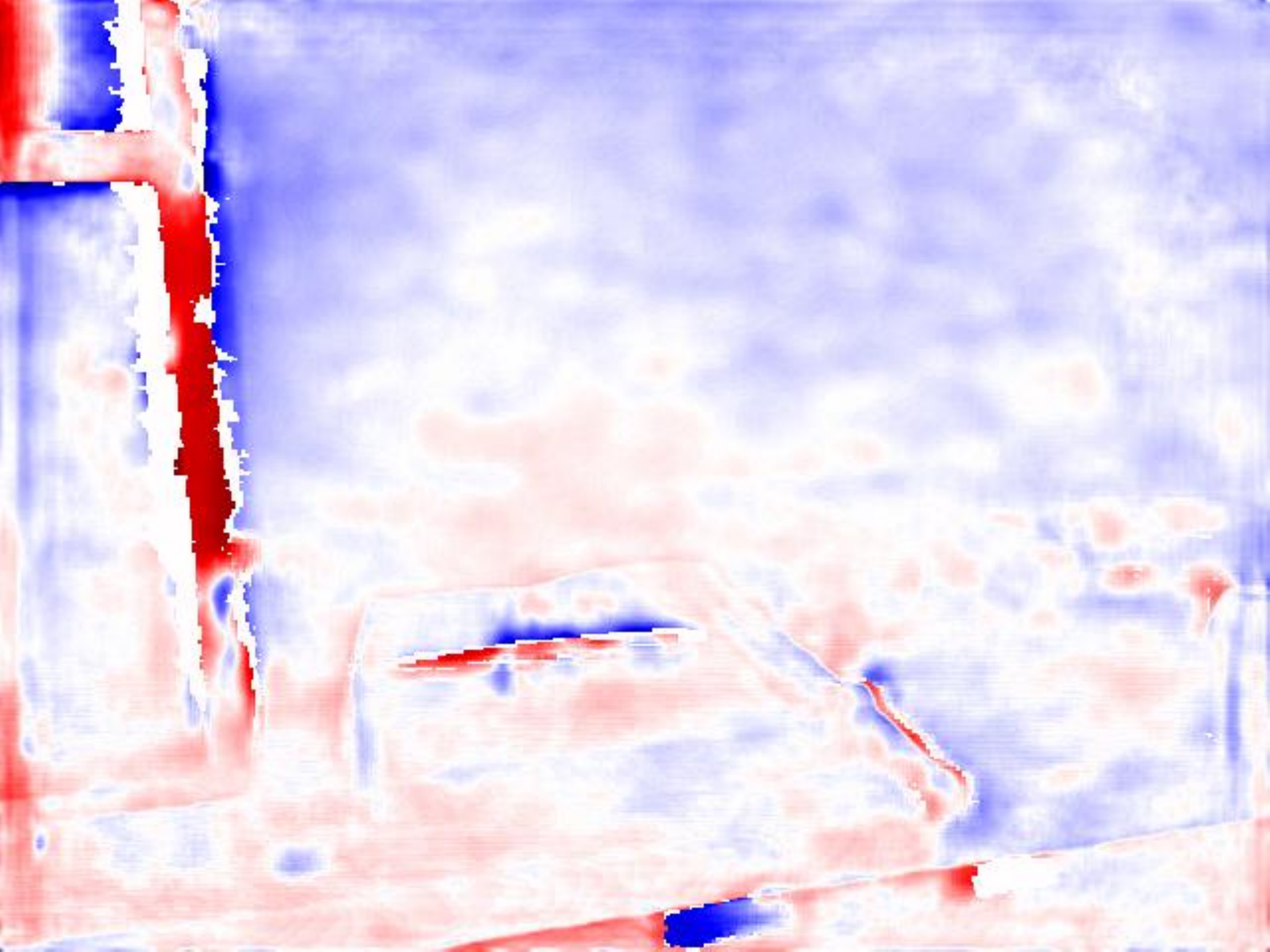}&
    \includegraphics[width=0.024\linewidth]{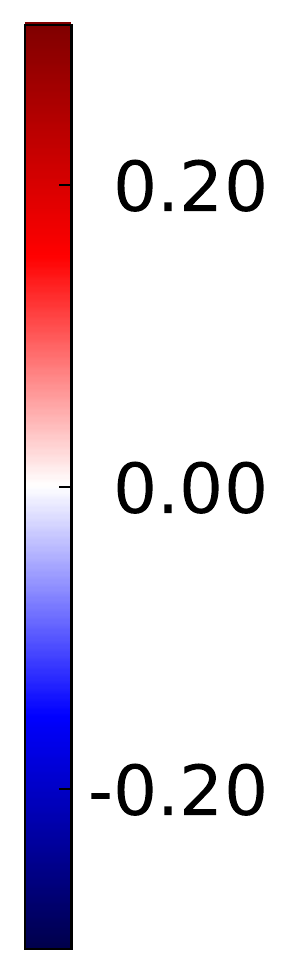}\\
    \vspace{-0.75mm}
    \rot{\scriptsize VOID 1500} &
    \includegraphics[width=0.104\linewidth]{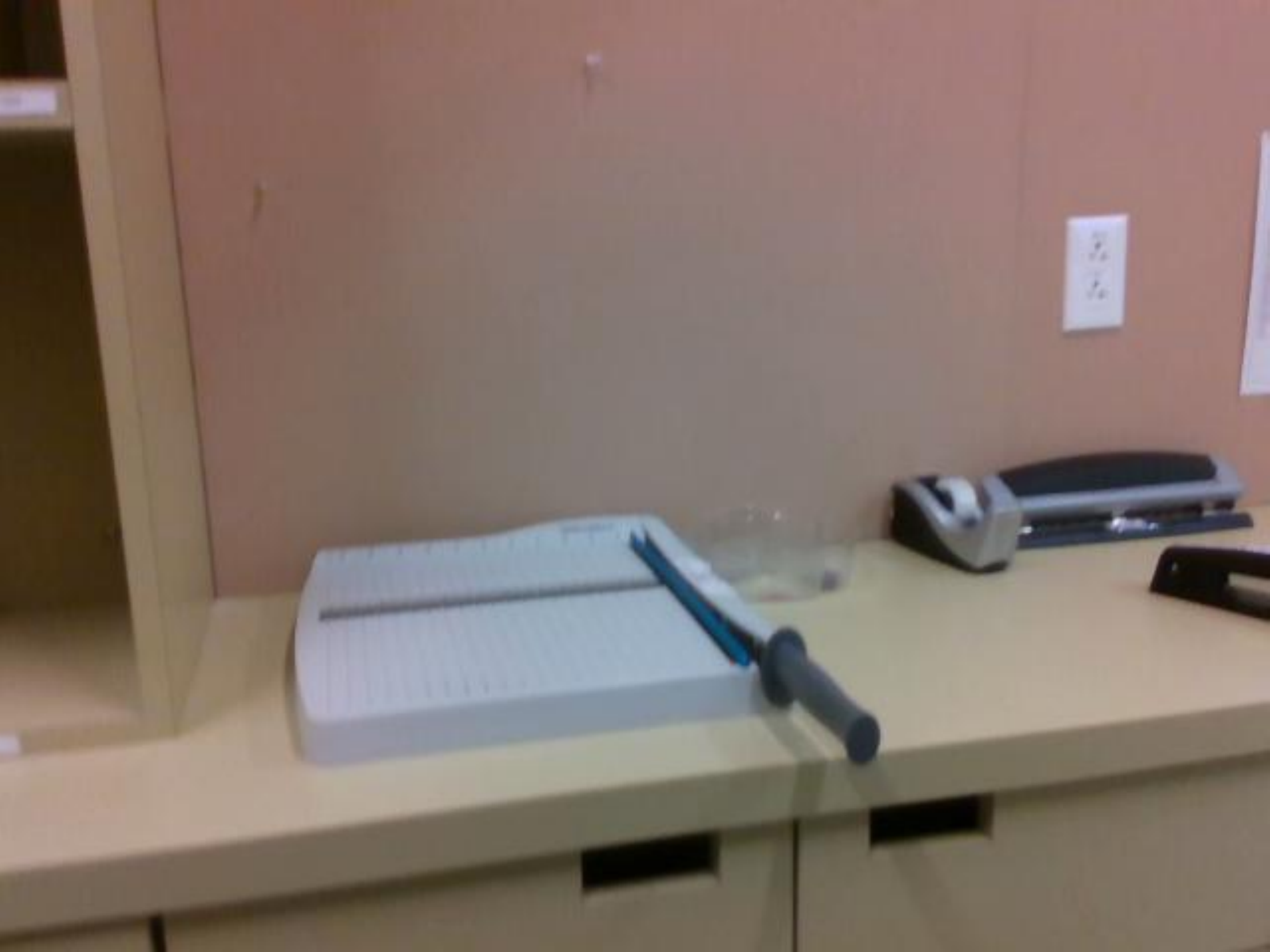}&
    \includegraphics[width=0.104\linewidth]{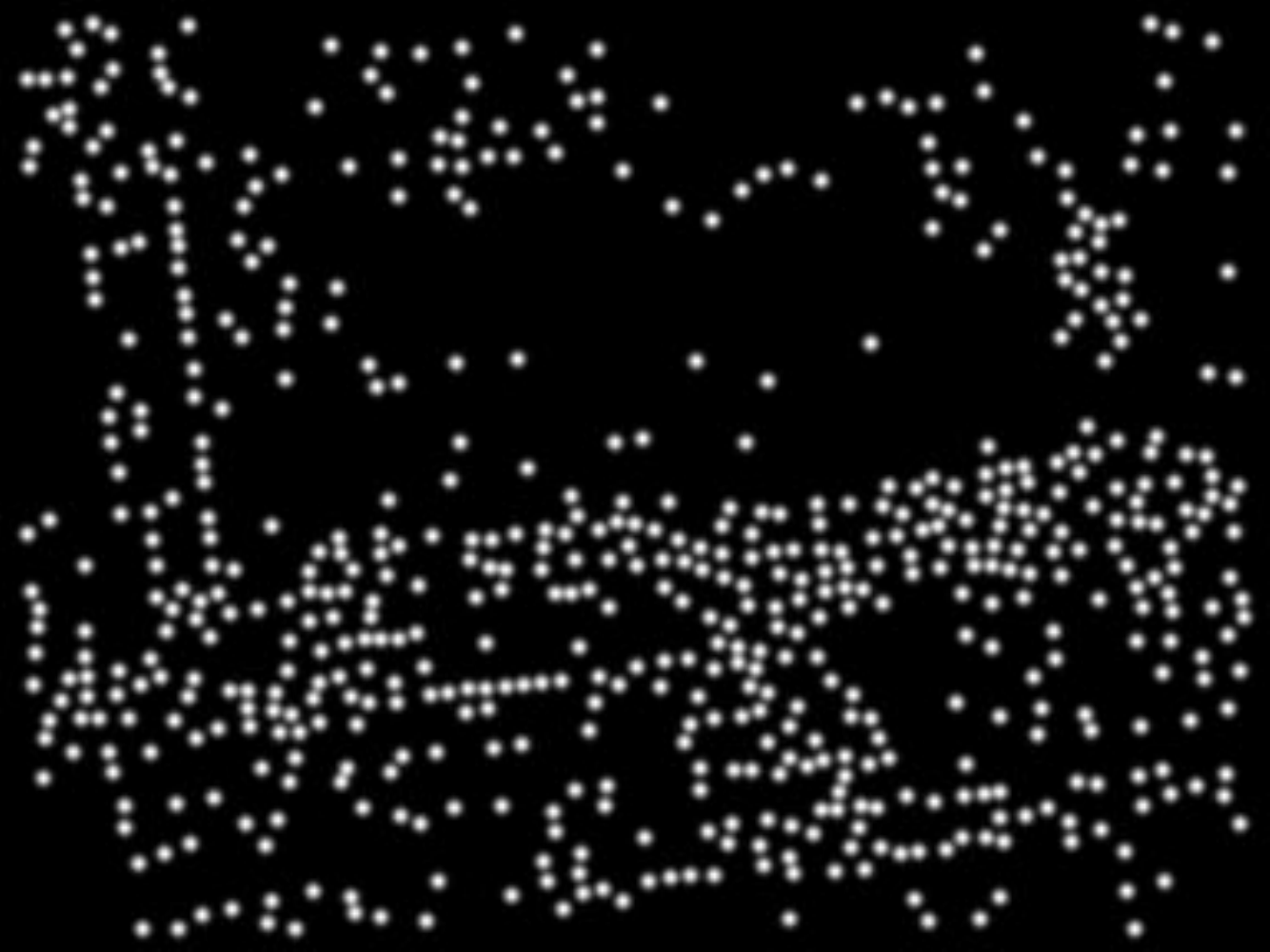}&
    \includegraphics[width=0.104\linewidth]{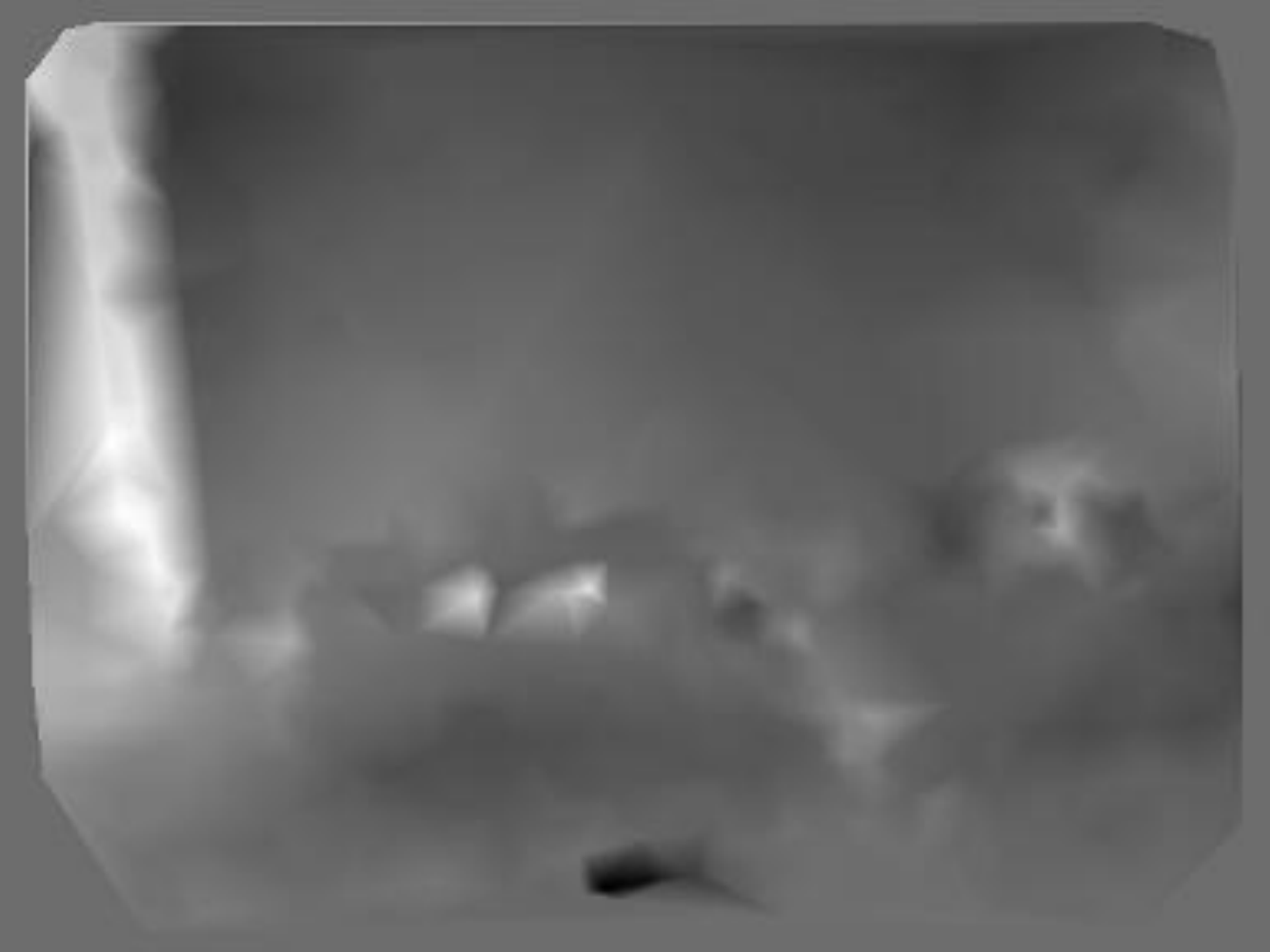}&
    \includegraphics[width=0.104\linewidth]{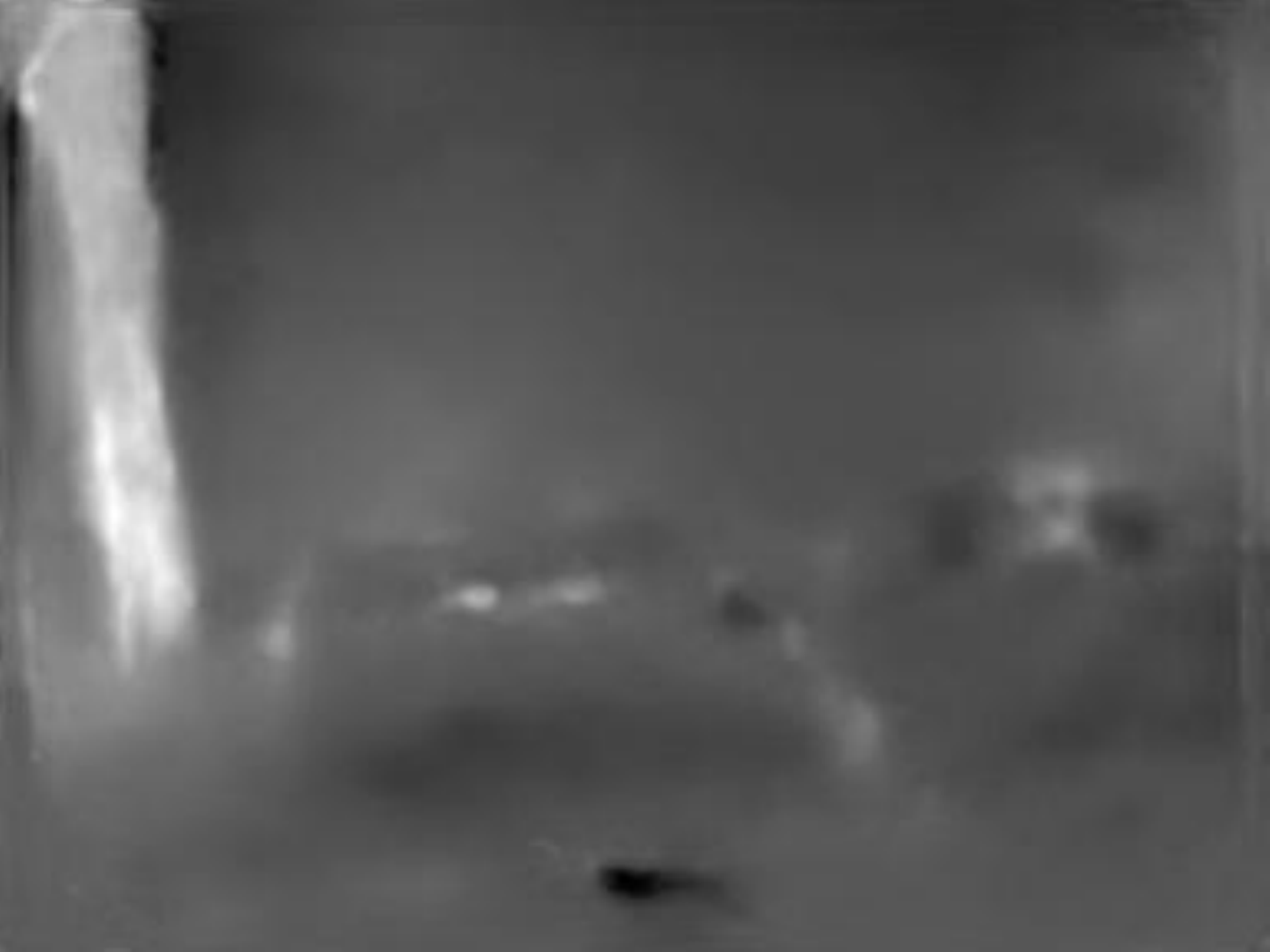}&
    \includegraphics[width=0.104\linewidth]{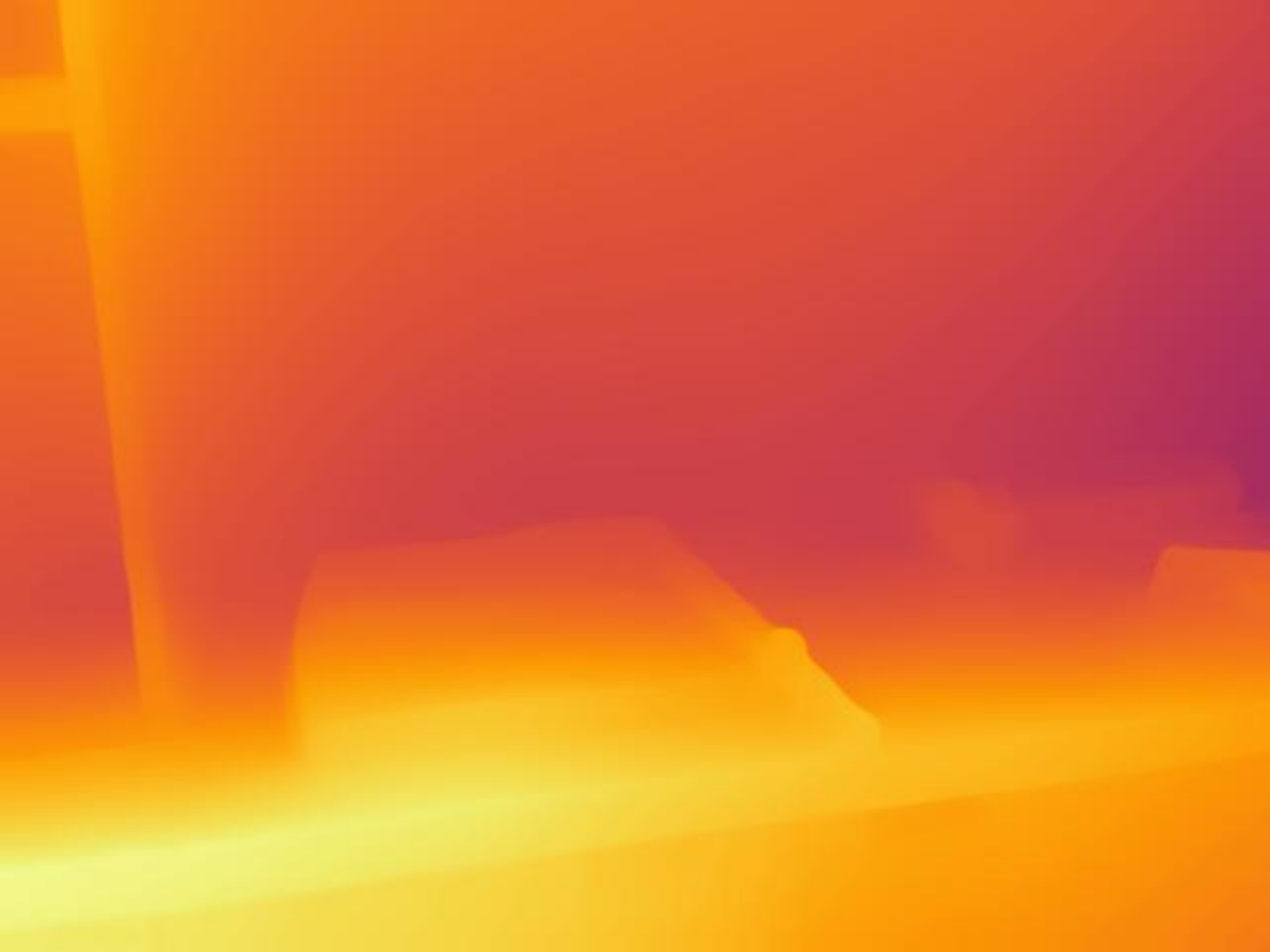}&
    \includegraphics[width=0.104\linewidth]{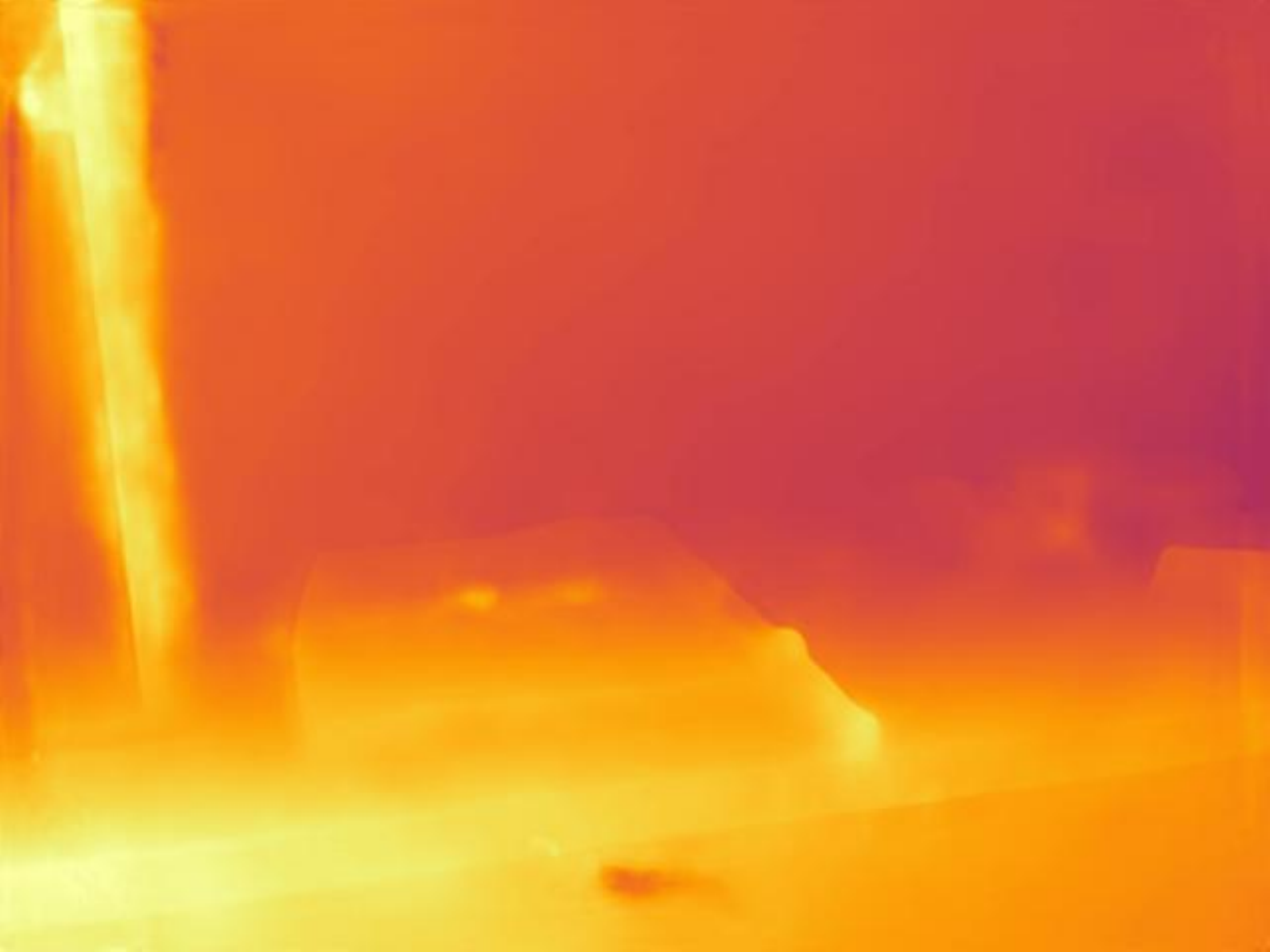}&
    \includegraphics[width=0.104\linewidth]{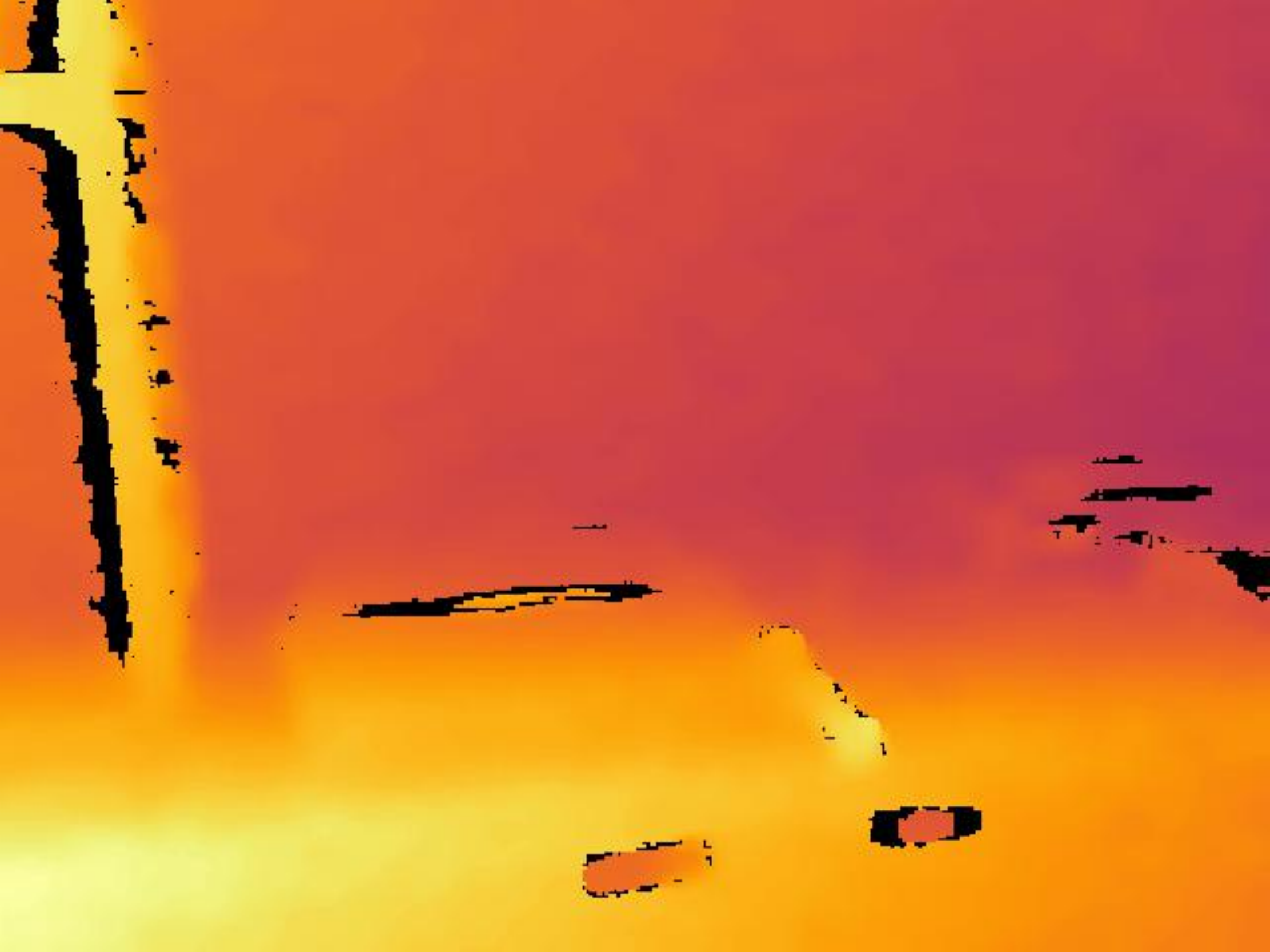}&
    \includegraphics[width=0.104\linewidth]{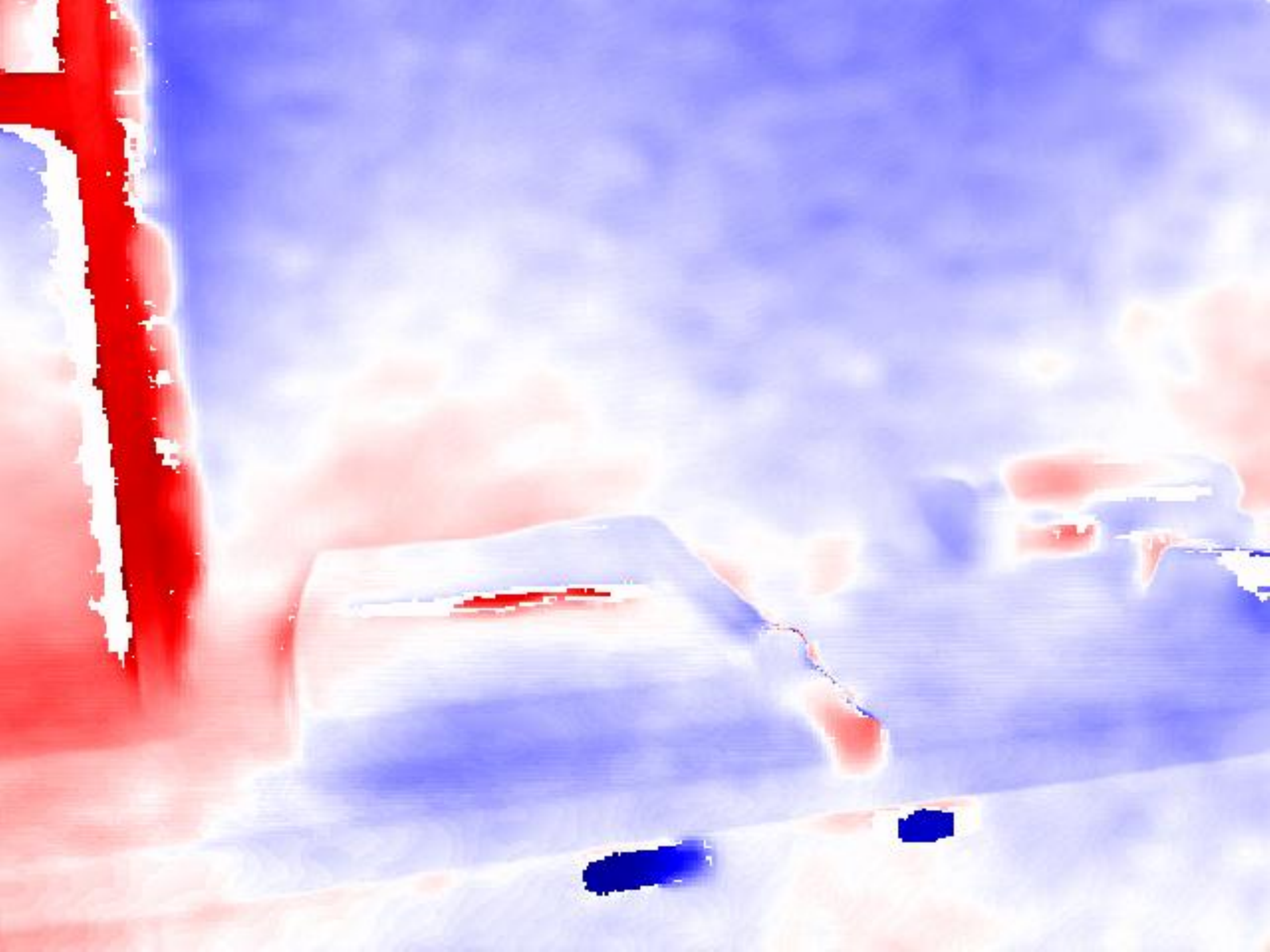}&
    \includegraphics[width=0.104\linewidth]{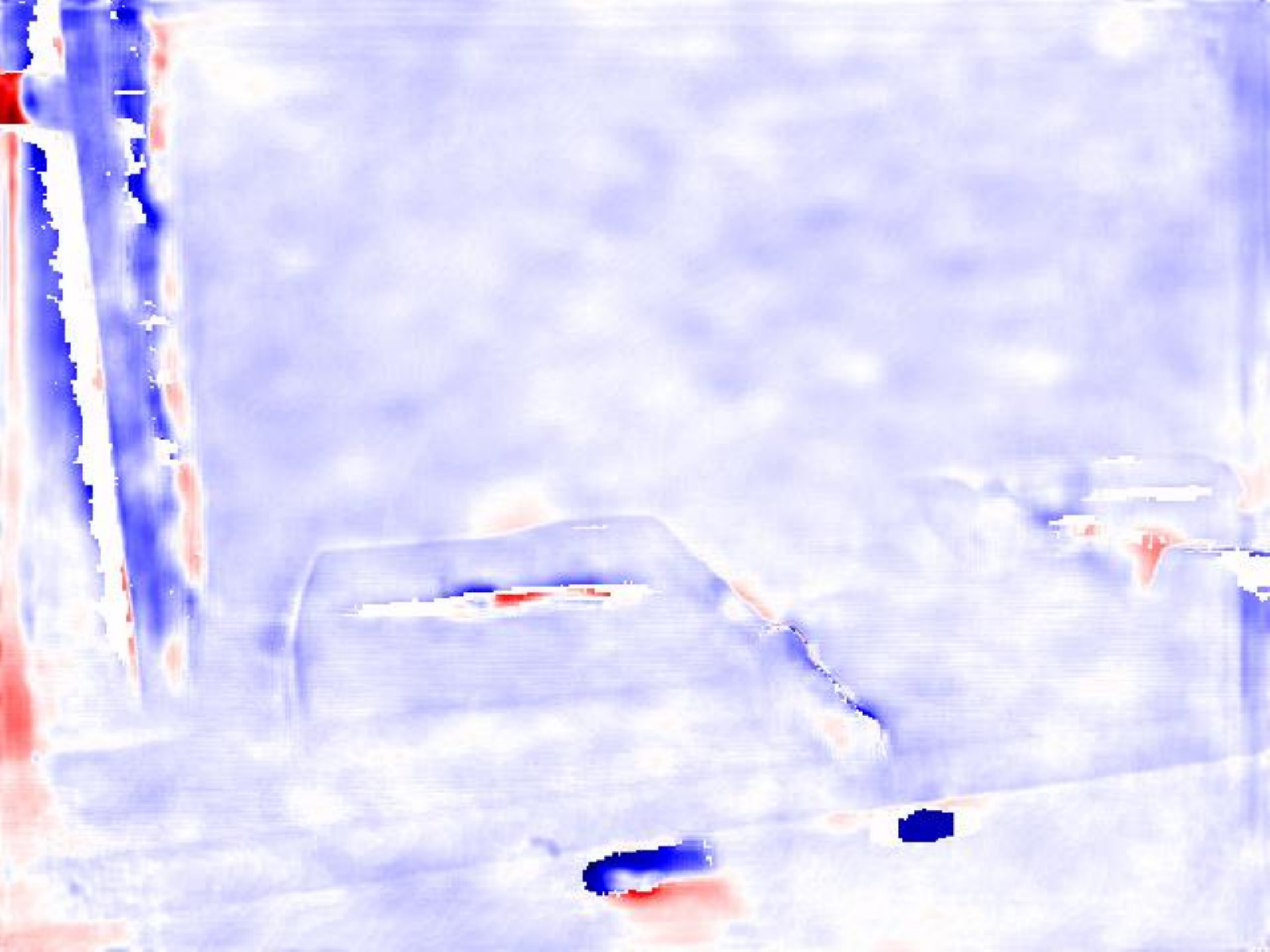}&
    \includegraphics[width=0.024\linewidth]{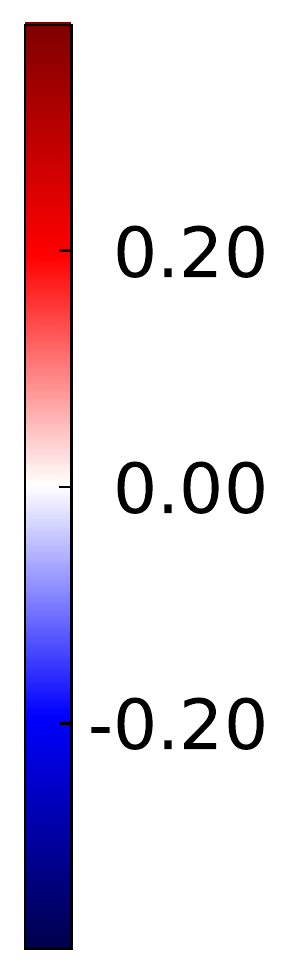}\\
    \multicolumn{9}{c}{} \\
  \end{tabular}
  \caption{Visualization of our approach on VOID samples with different densities of sparse depth points. Color coding of depth and error maps is the same as in Figure~\ref{fig:vis-void-expanded}. Samples within each group of three are not identical but were selected to match as closely as possible. The models used to generate these results were trained solely on VOID (without TartanAir pretraining) and assume a DPT-Hybrid depth estimator.}
  \label{fig:vis-void-densities}
\end{figure*}

\mypara{Different densities of sparse metric depth.} When comparing against related works \cite{Wong2020void} and \cite{Wong2021kbnet} that also benchmark on VOID, we perform experiments at all three densities available in the published dataset: 150, 500, and 1500 sparse metric depth points. Figure~\ref{fig:vis-void-densities} visualizes what sparsity at different densities looks like, along with how this impacts the interpolated scale map scaffolding and regression with SML. In general, a larger number of sparse depth points results in better global alignment of depth (as indicated by whiter regions in the GA error maps) as well as more fine-grained scale map scaffolding. SML performs better with the medium density of 500 relative to the low density of 150; the SML error maps look whiter and lighter for VOID with 500 samples compared to VOID with 150 samples. However, a high density of 1500 sparse points results in diminishing returns; the scale map scaffolding already contains significant detail and consequently, there is less for the SML to learn to refine. The regressed scale maps in this case look very similar to the input scaffolding. Our method therefore neither requires nor significantly benefits from high densities of sparse depth.

\begin{figure*}[p]
\centering
  \begin{tabular}{@{}l@{\hspace{0.5mm}}*{10}{c@{\hspace{0.5mm}}}@{}}
    &  &  &  & {\scriptsize GA+SML} & {\scriptsize GA+SML} & {\scriptsize GA+SML} & {\scriptsize GA+SML} & {\scriptsize GA+SML} & {\scriptsize GA+SML} & {\scriptsize GA+SML}\\
    & {\scriptsize RGB Image} & {\scriptsize Ground Truth} & {\scriptsize KBNet~\cite{Wong2021kbnet}} & {\scriptsize (DPT-BEiT-L)} & {\scriptsize (DPT-SwinV2-L)} & {\scriptsize (DPT-Large)} & {\scriptsize (DPT-Hybrid)} & {\scriptsize (DPT-SwinV2-T)} & {\scriptsize (DPT-LeViT)} & {\scriptsize (MiDaS-small)}\\
    \vspace{-0.75mm}
    \rot{\scriptsize VOID 150} &
    \includegraphics[width=0.096\linewidth]{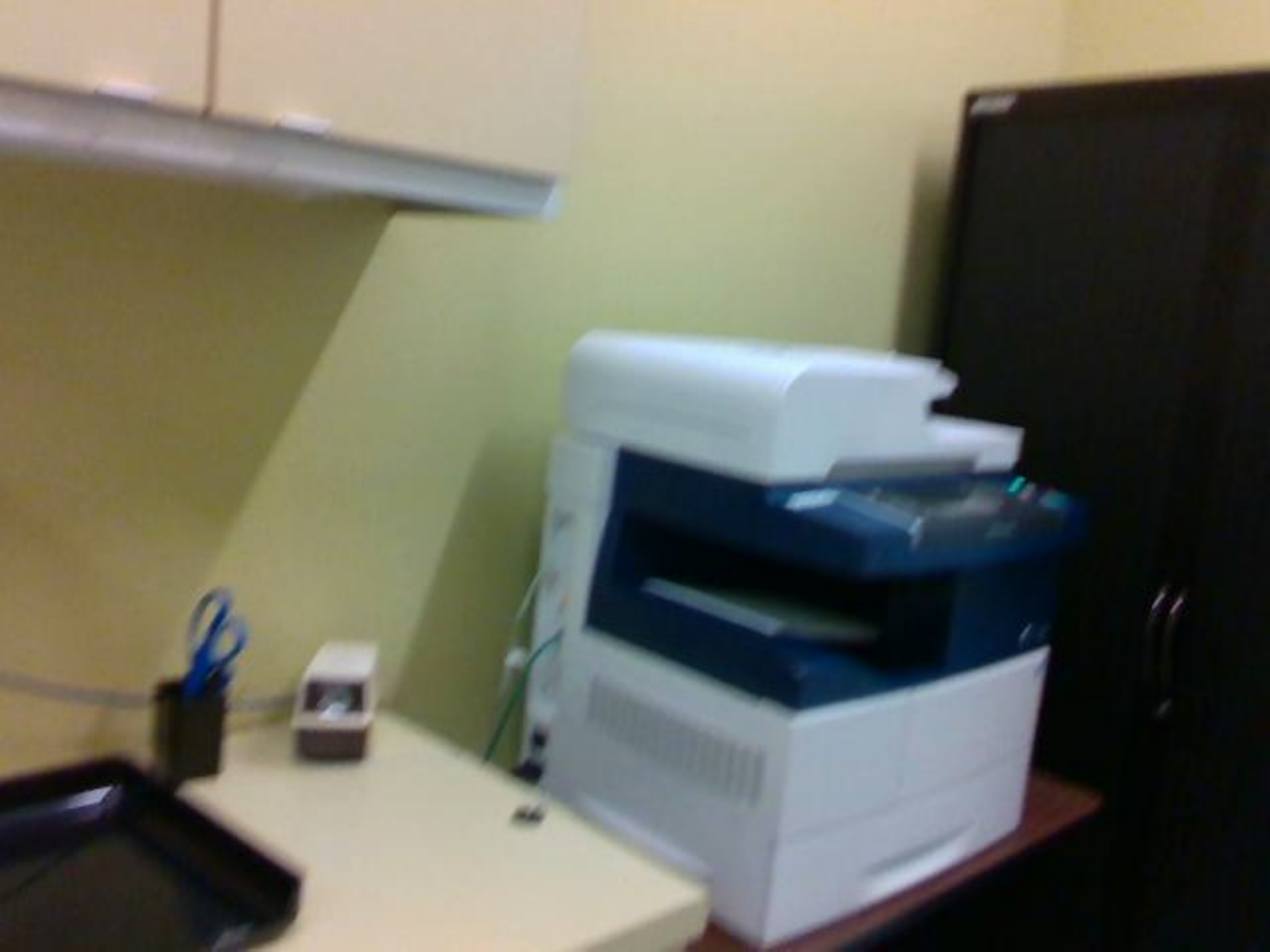}&
    \includegraphics[width=0.096\linewidth]{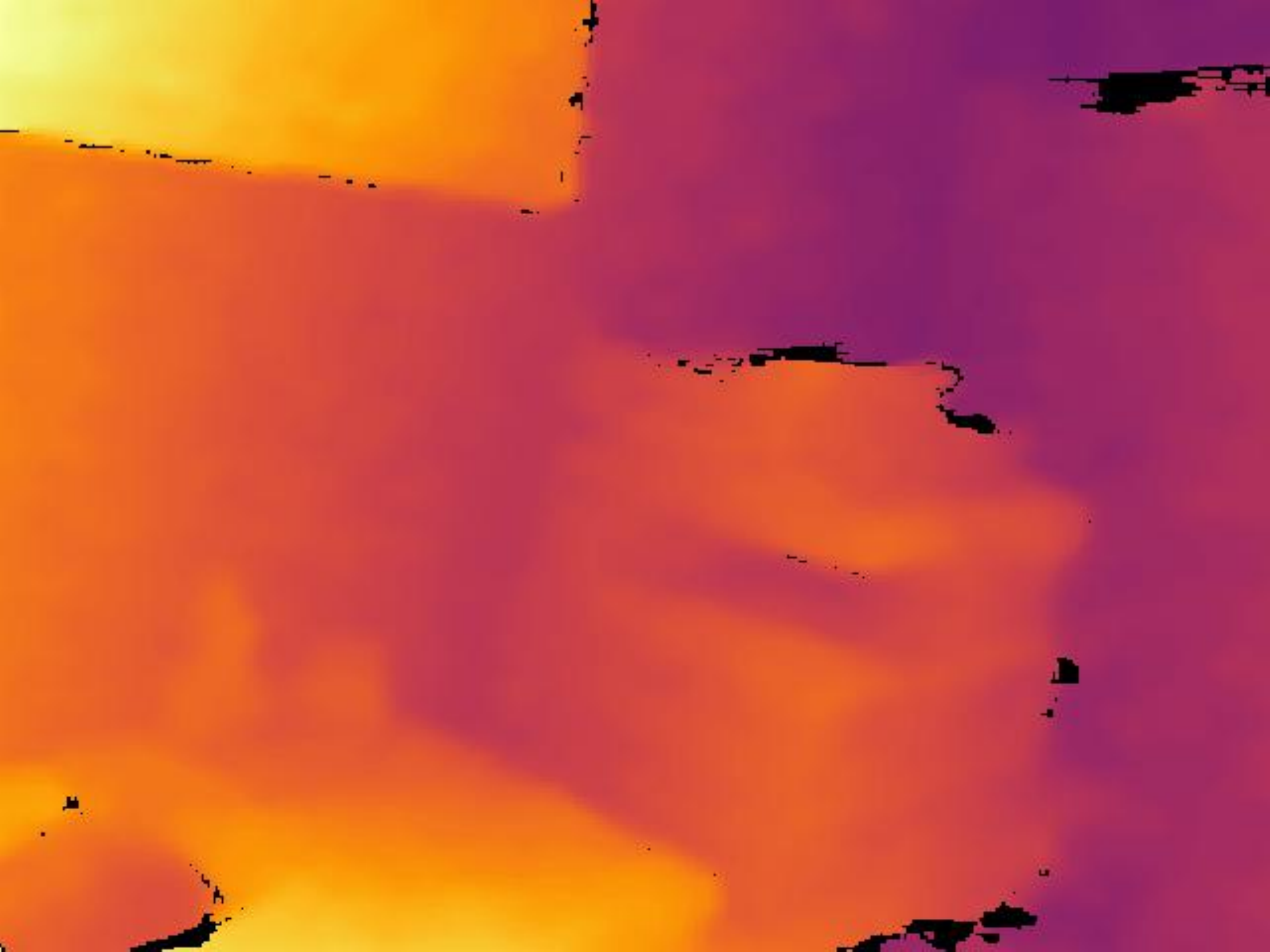}&
    \includegraphics[width=0.096\linewidth]{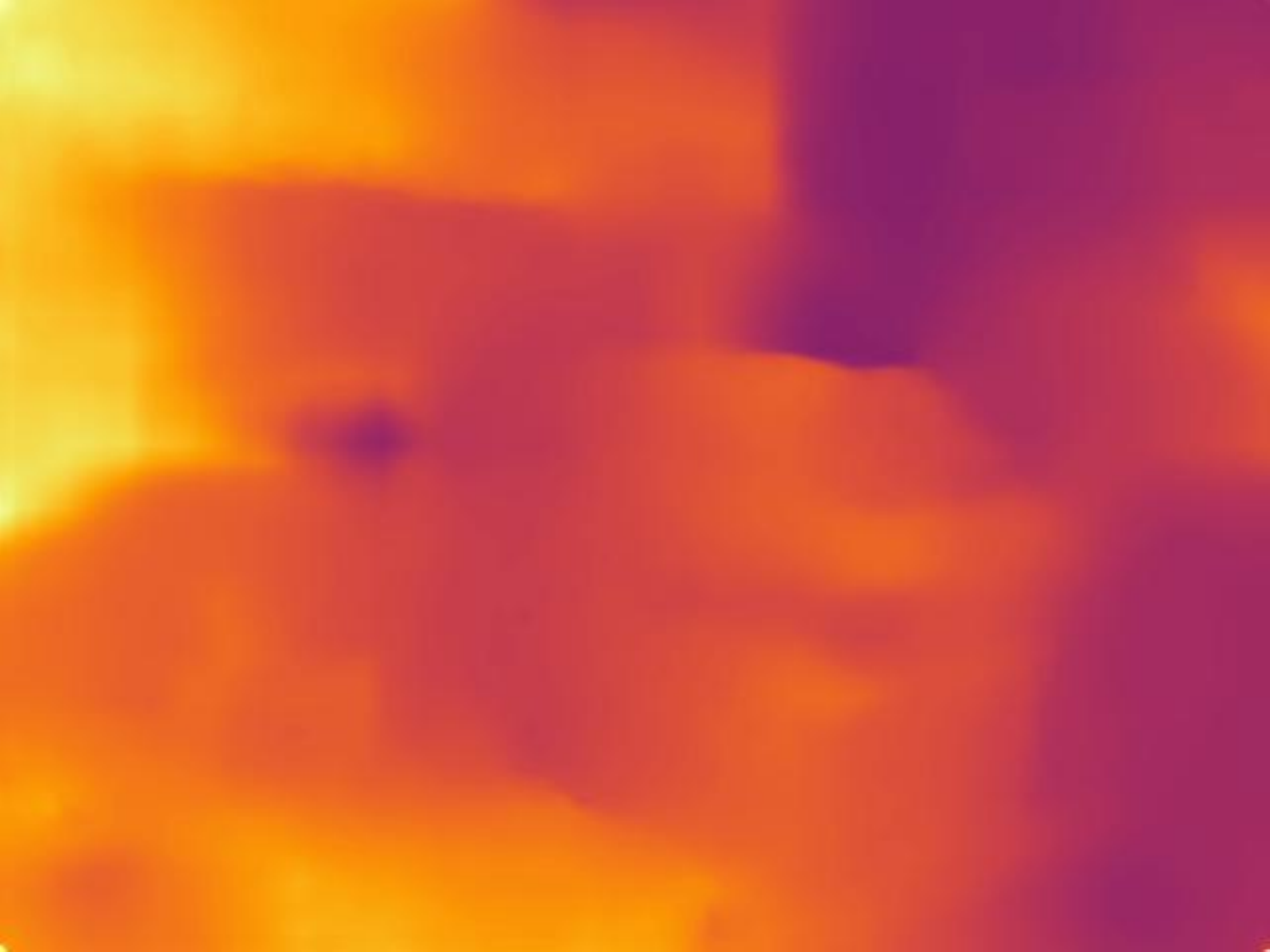}&
    \includegraphics[width=0.096\linewidth]{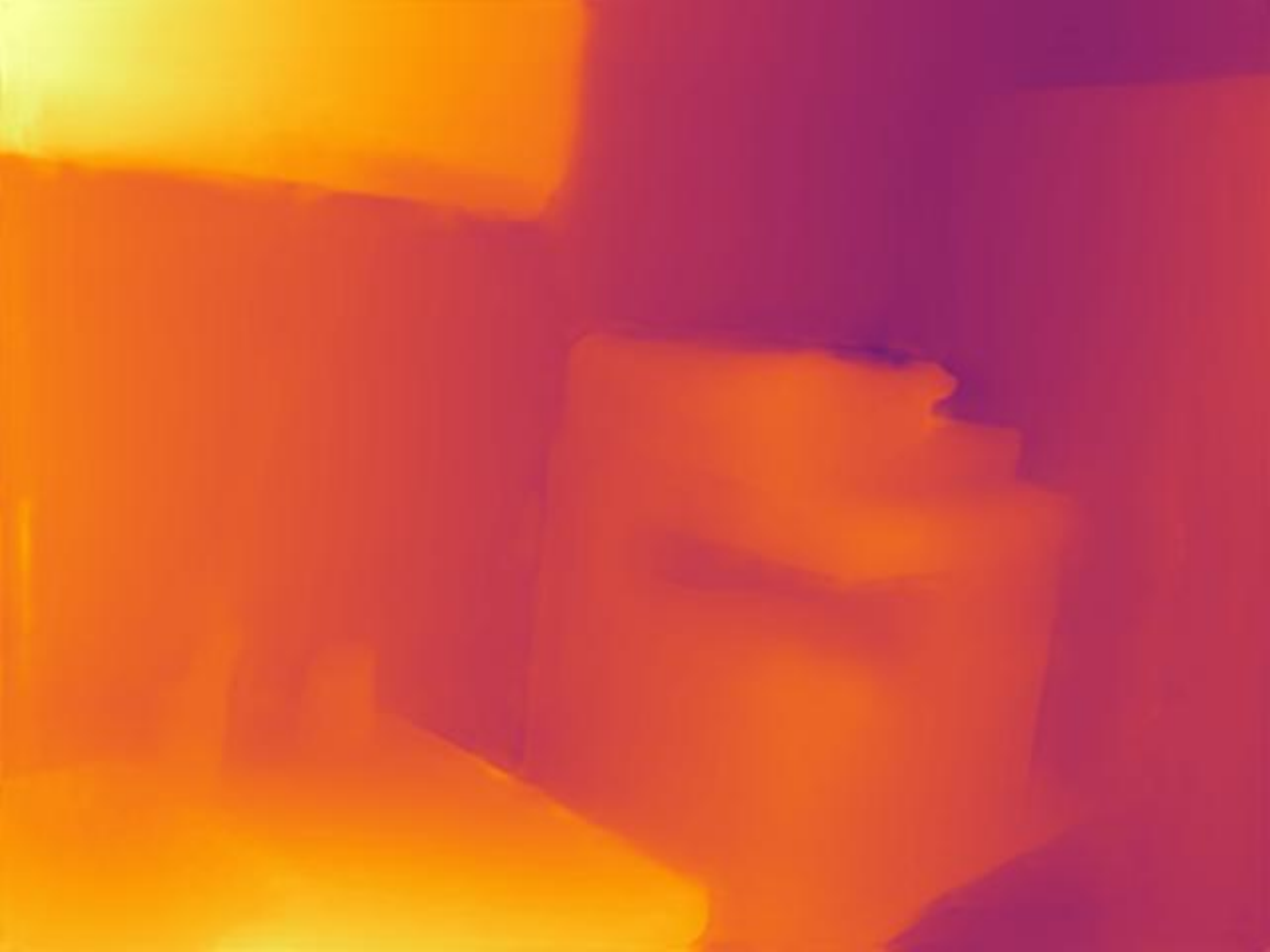}&
    \includegraphics[width=0.096\linewidth]{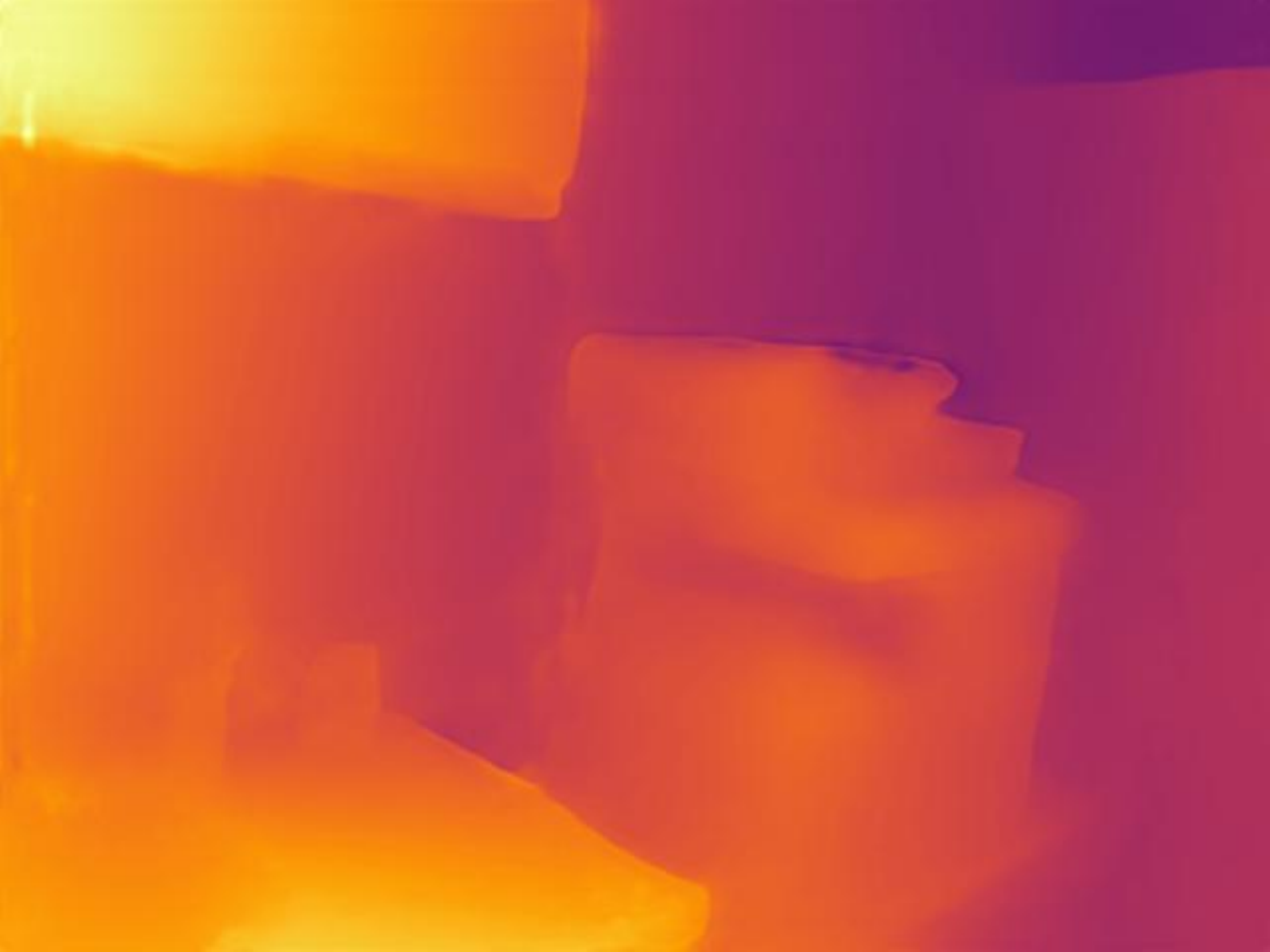}&
    \includegraphics[width=0.096\linewidth]{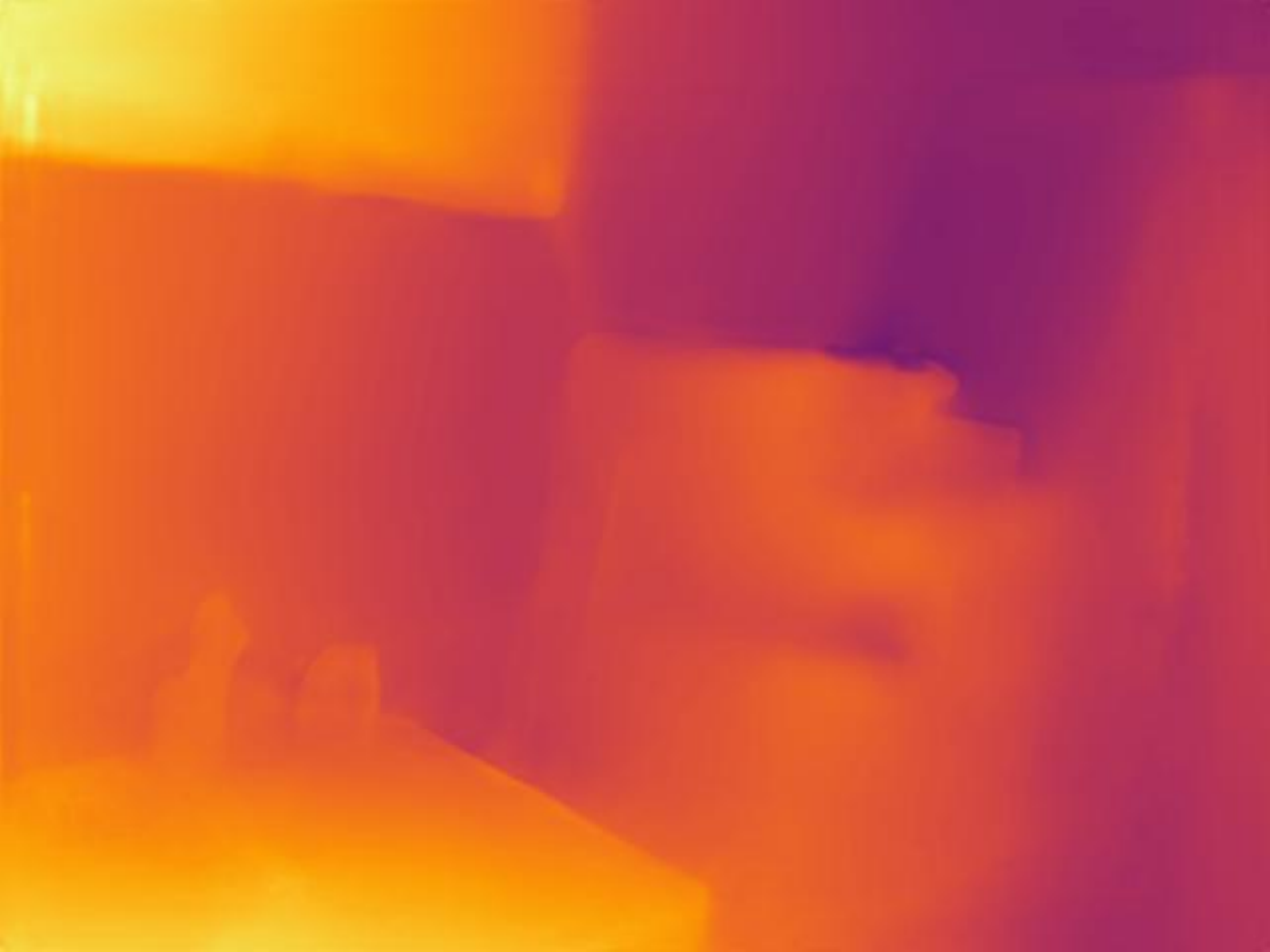}&
    \includegraphics[width=0.096\linewidth]{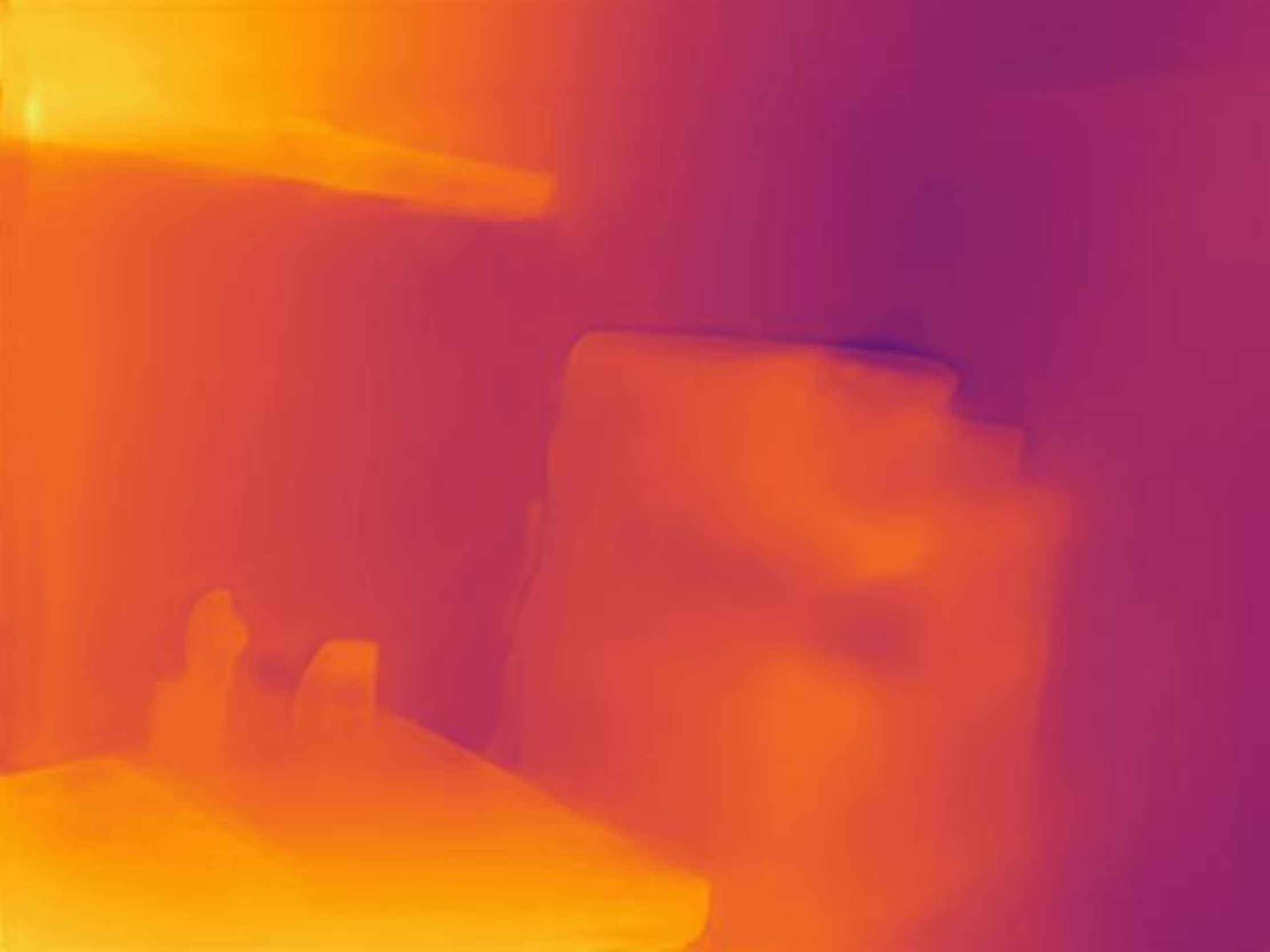}&
    \includegraphics[width=0.096\linewidth]{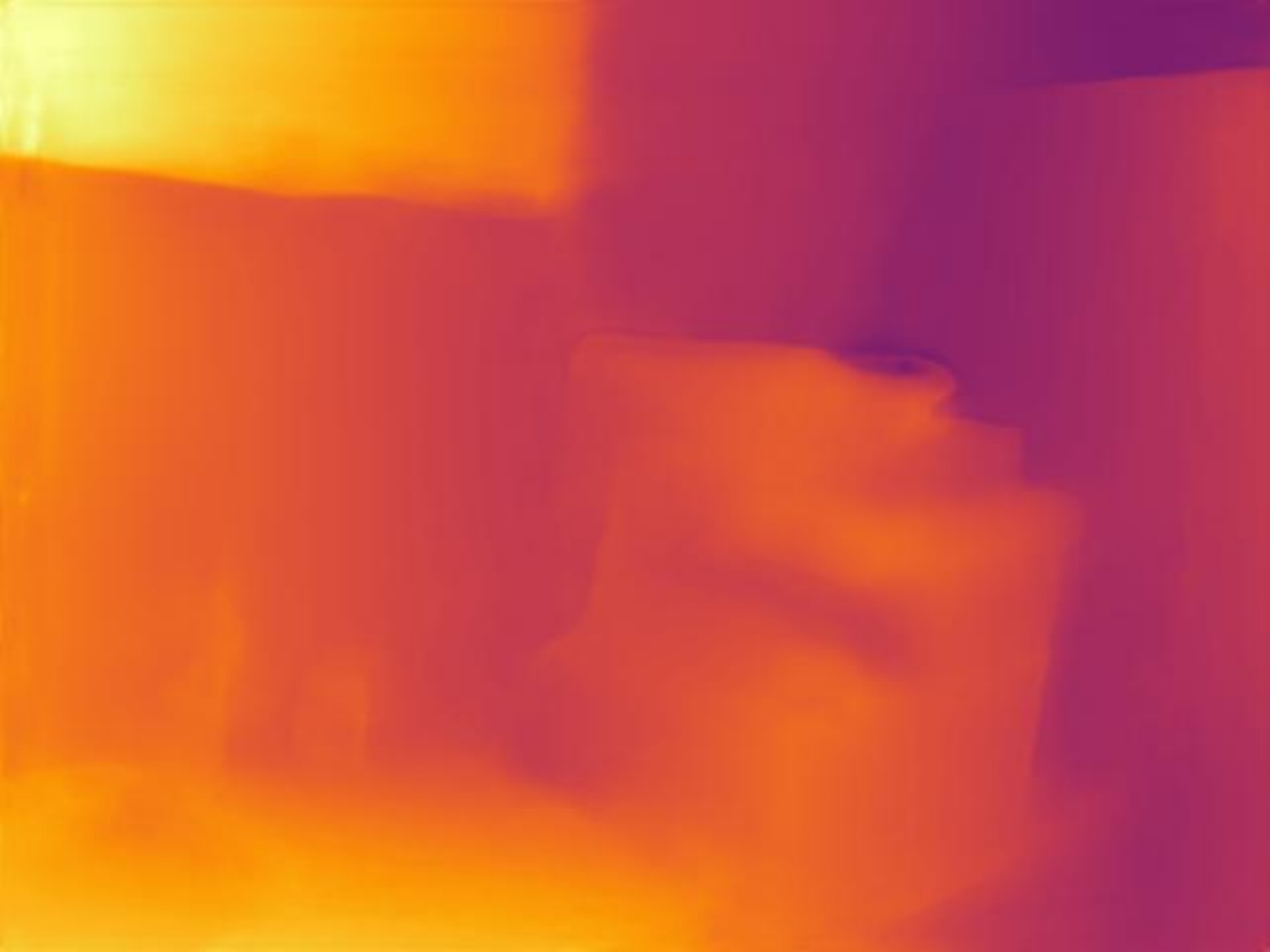}&
    \includegraphics[width=0.096\linewidth]{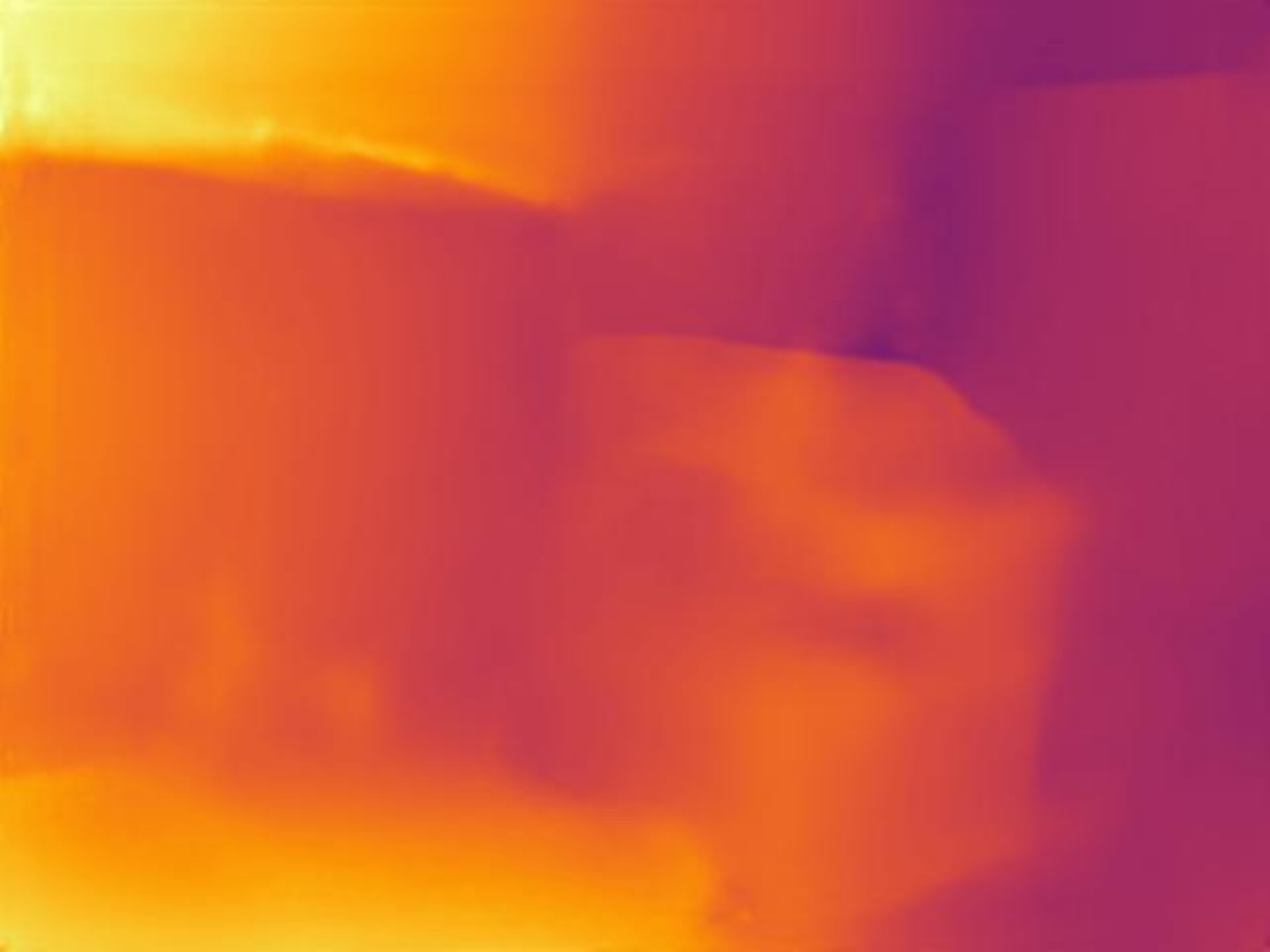}&
    \includegraphics[width=0.096\linewidth]{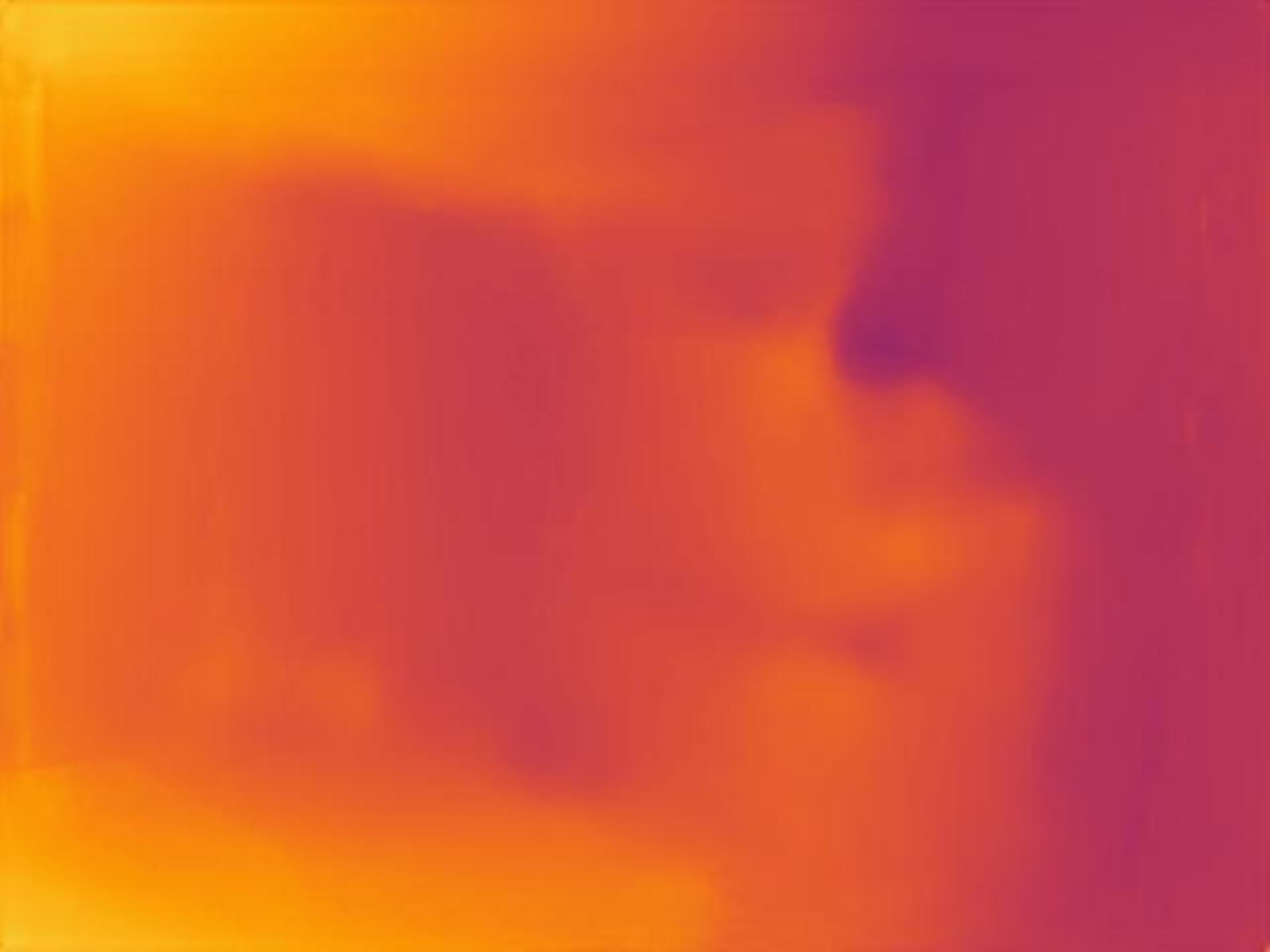}\\
    \vspace{-0.75mm}
    \rot{\scriptsize VOID 500} &
    \includegraphics[width=0.096\linewidth]{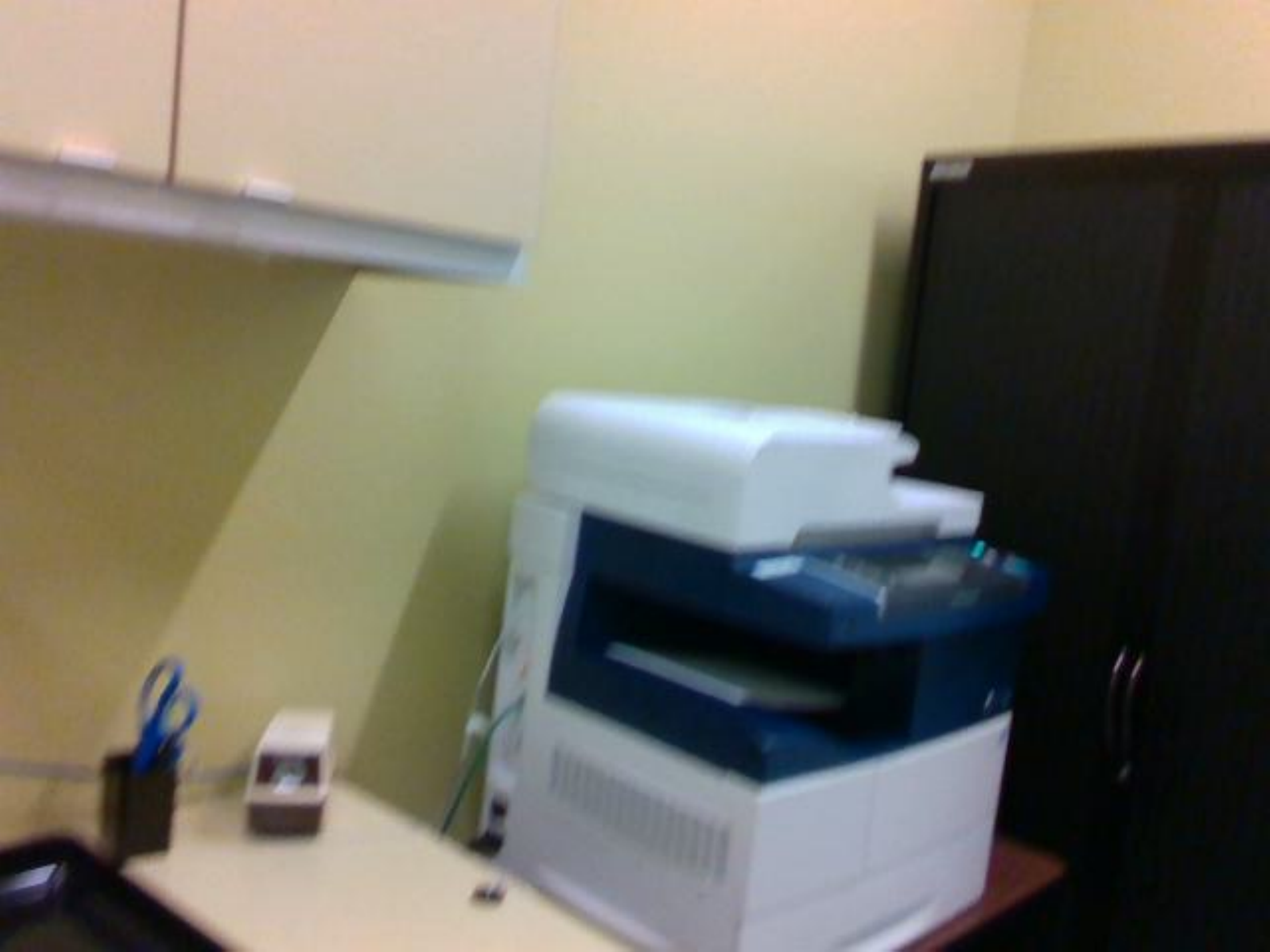}&
    \includegraphics[width=0.096\linewidth]{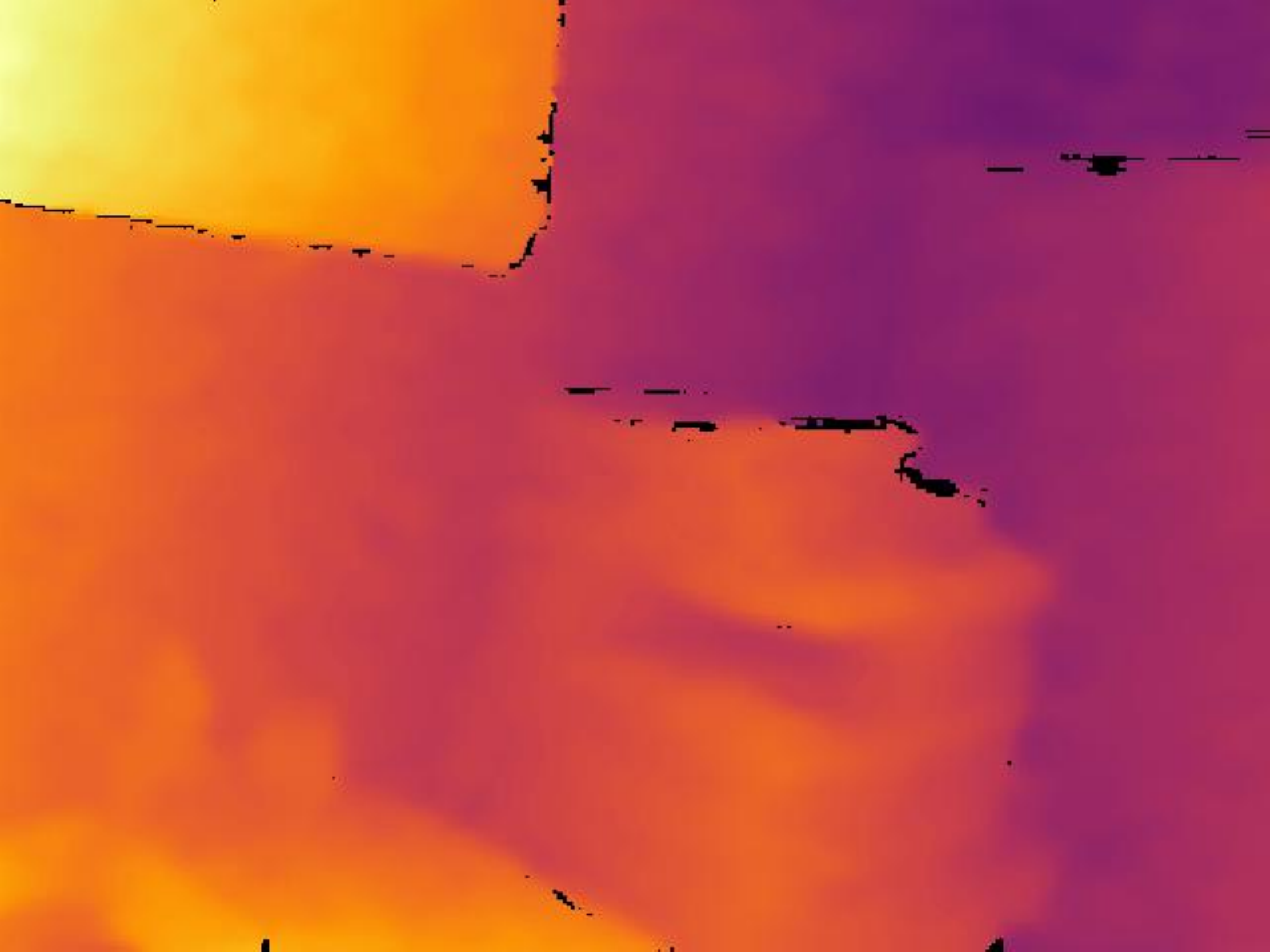}&
    \includegraphics[width=0.096\linewidth]{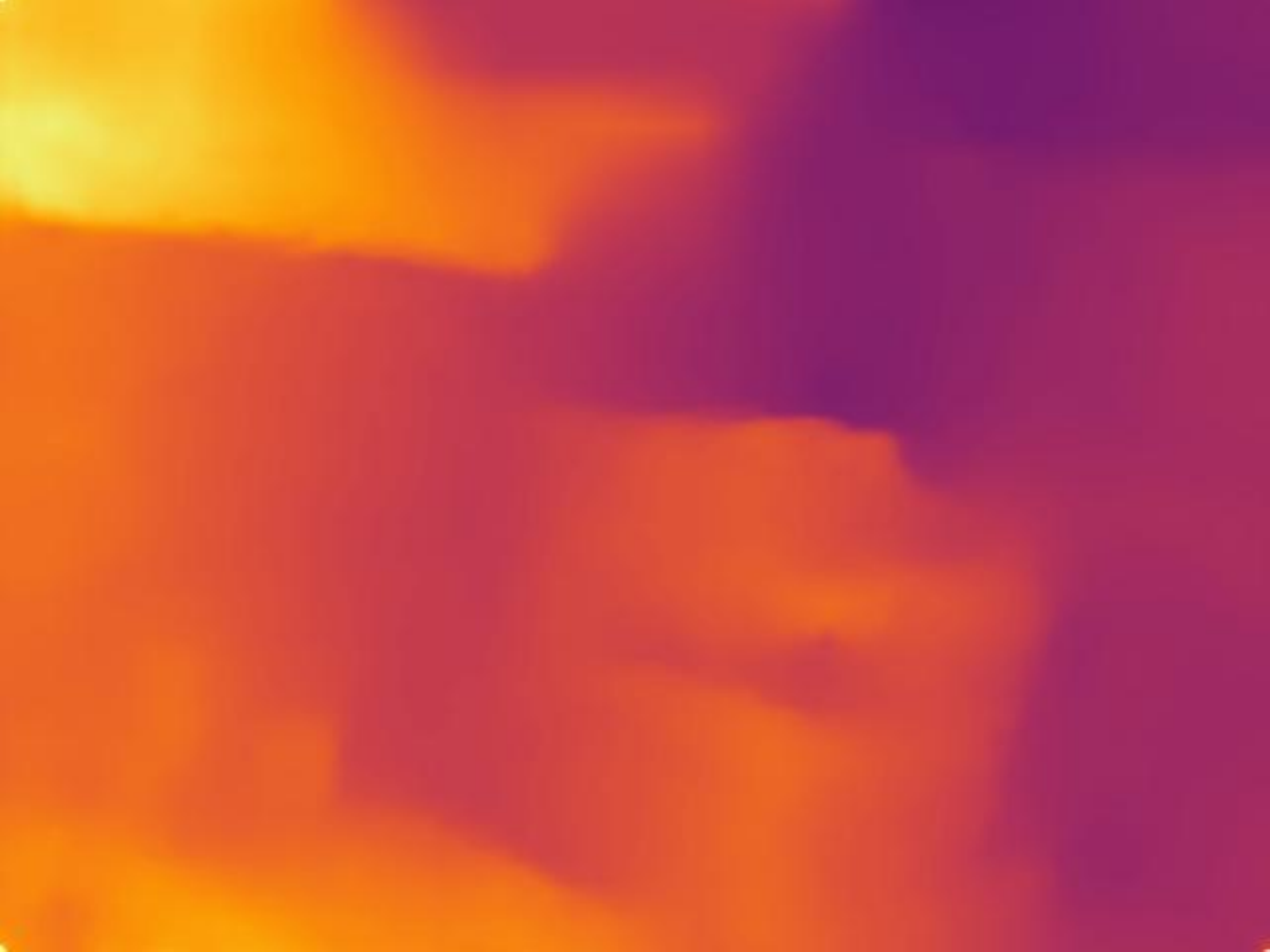}&
    \includegraphics[width=0.096\linewidth]{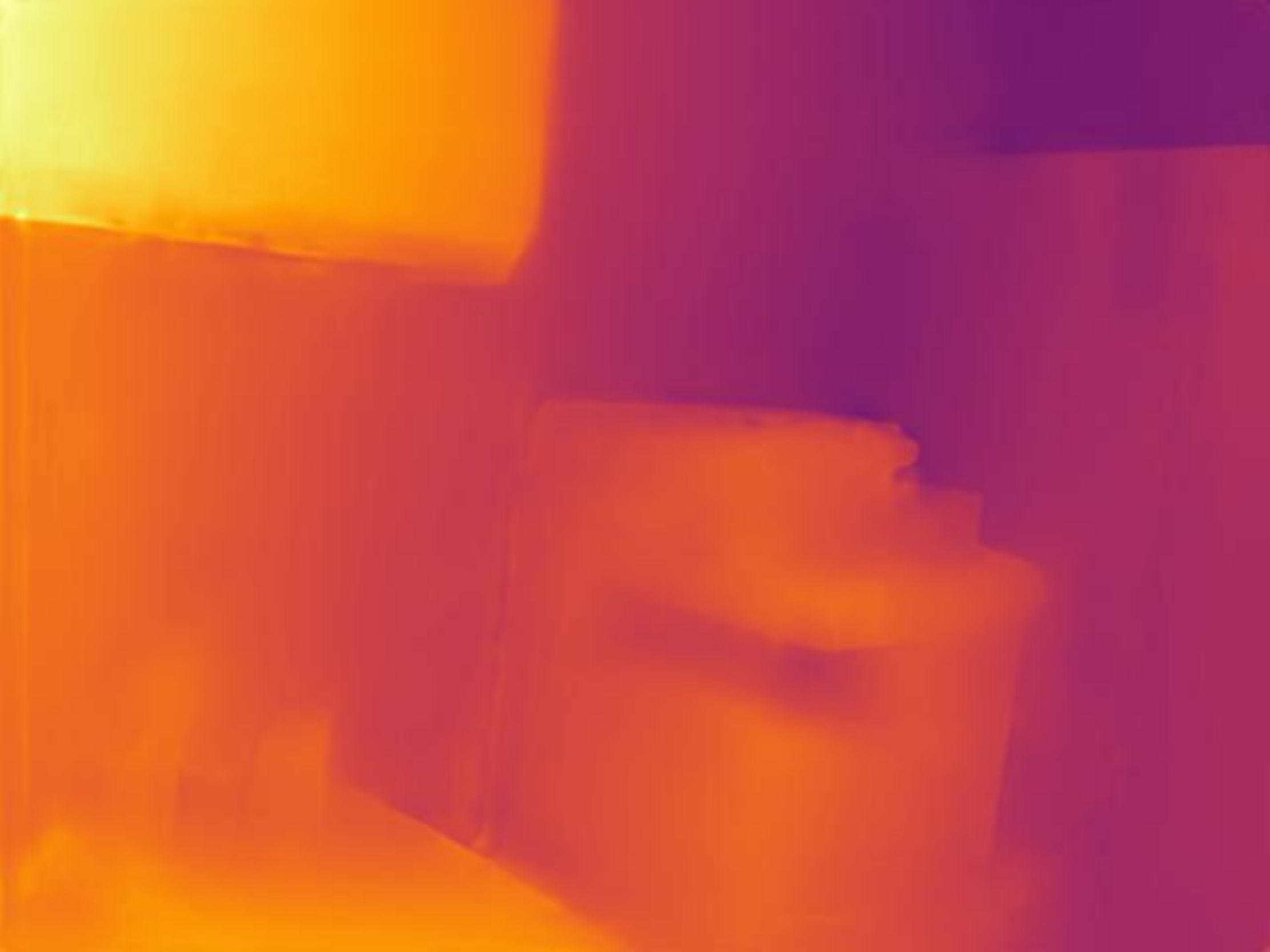}&
    \includegraphics[width=0.096\linewidth]{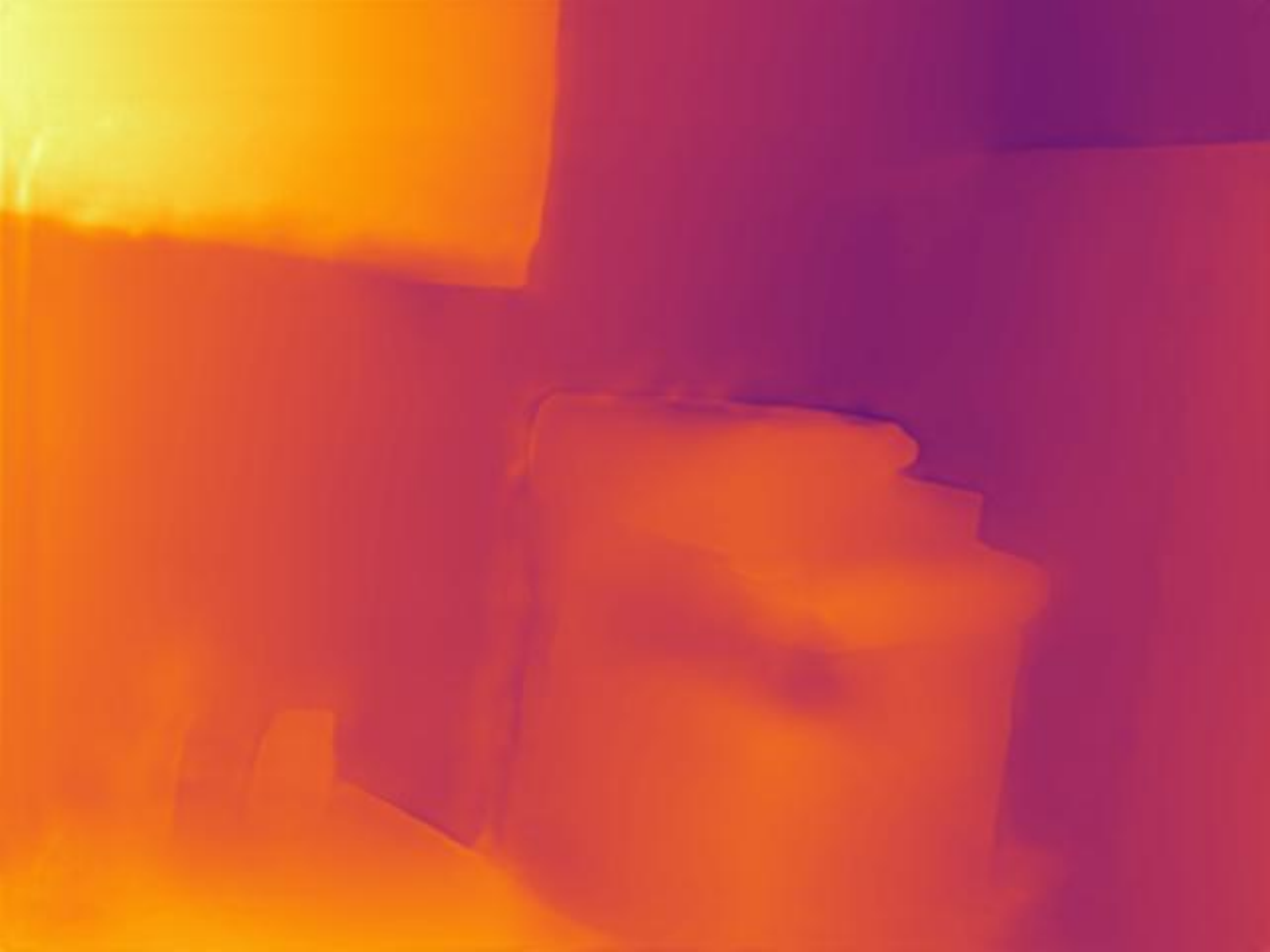}&
    \includegraphics[width=0.096\linewidth]{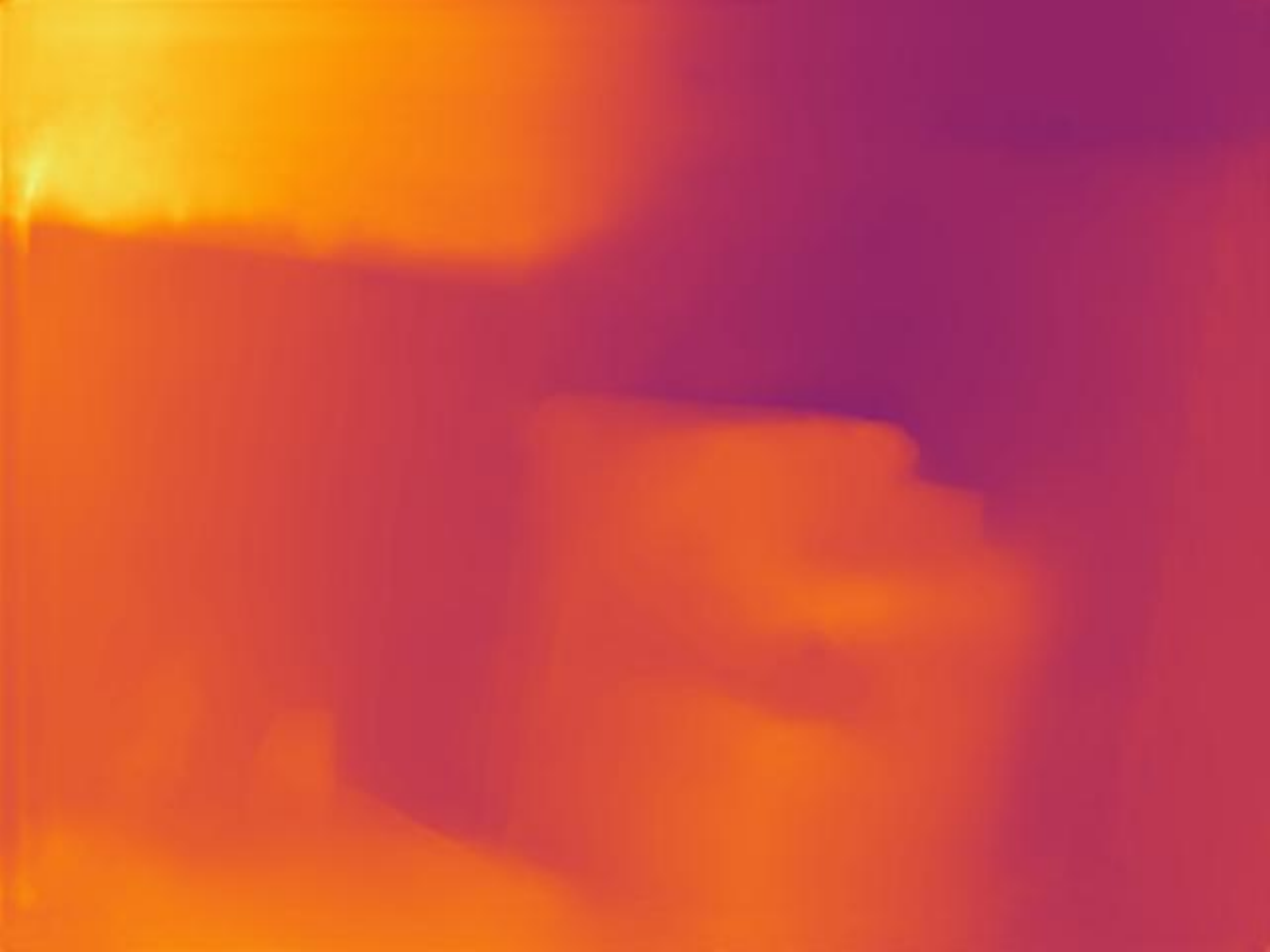}&
    \includegraphics[width=0.096\linewidth]{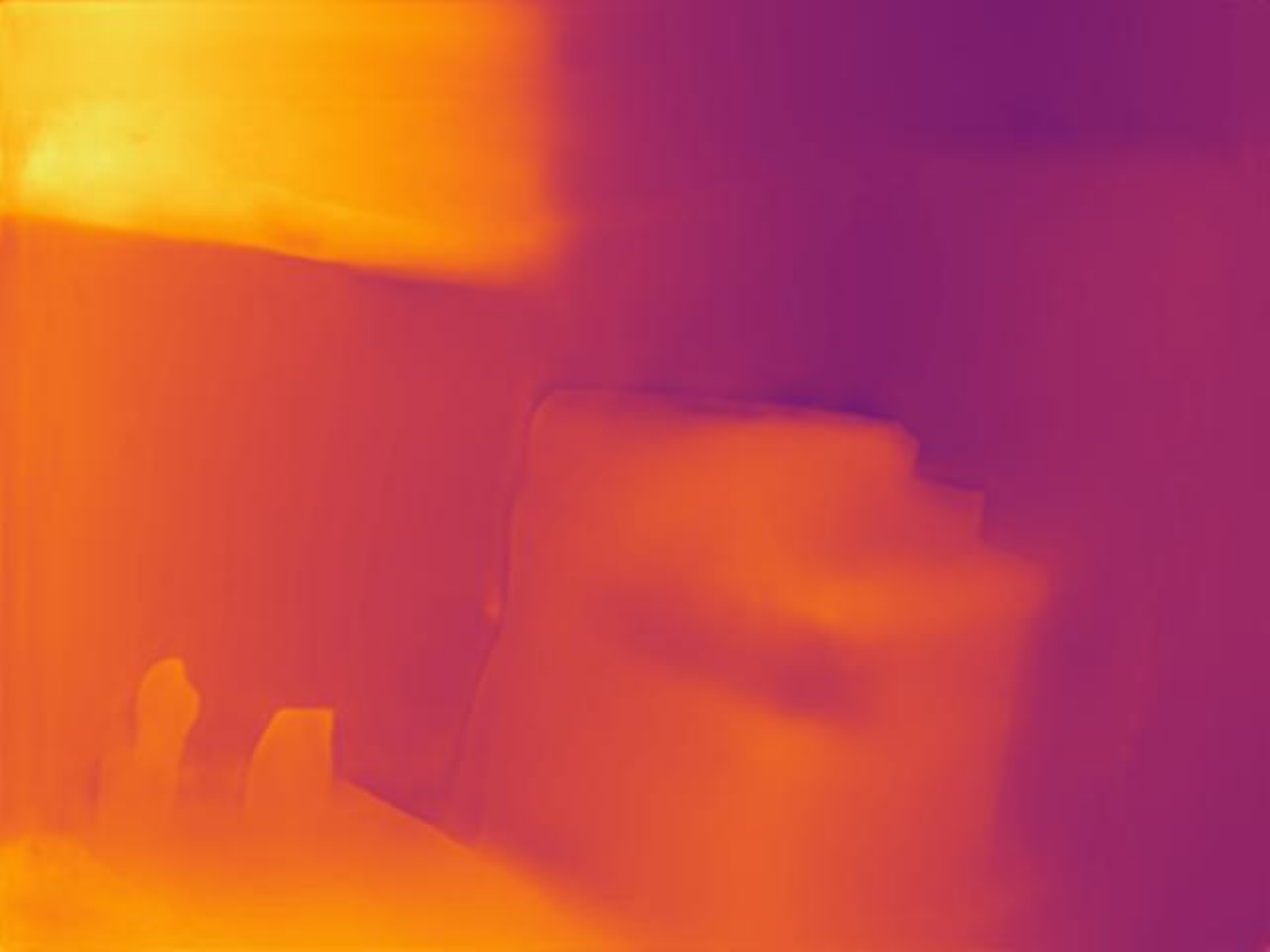}&
    \includegraphics[width=0.096\linewidth]{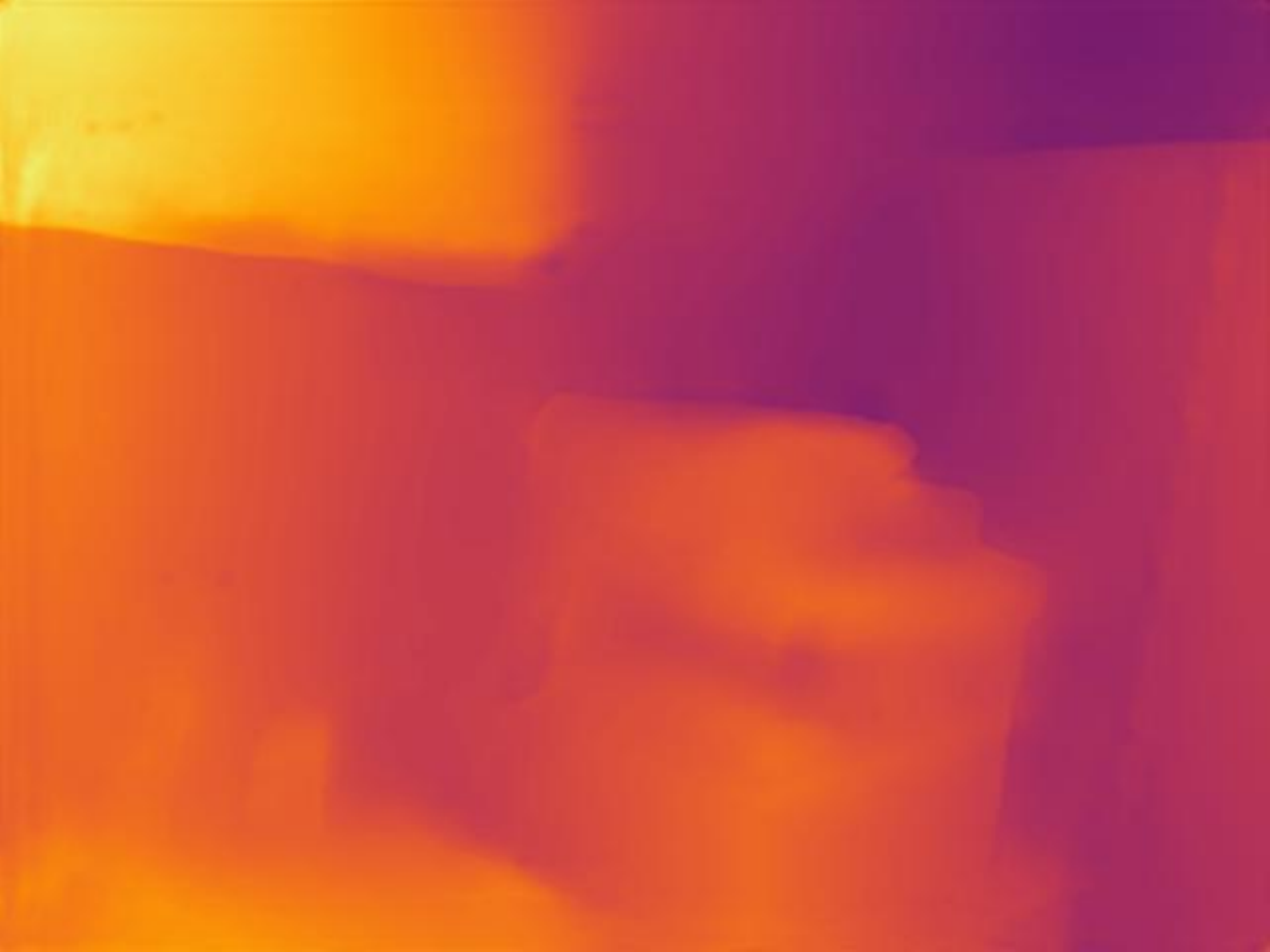}&
    \includegraphics[width=0.096\linewidth]{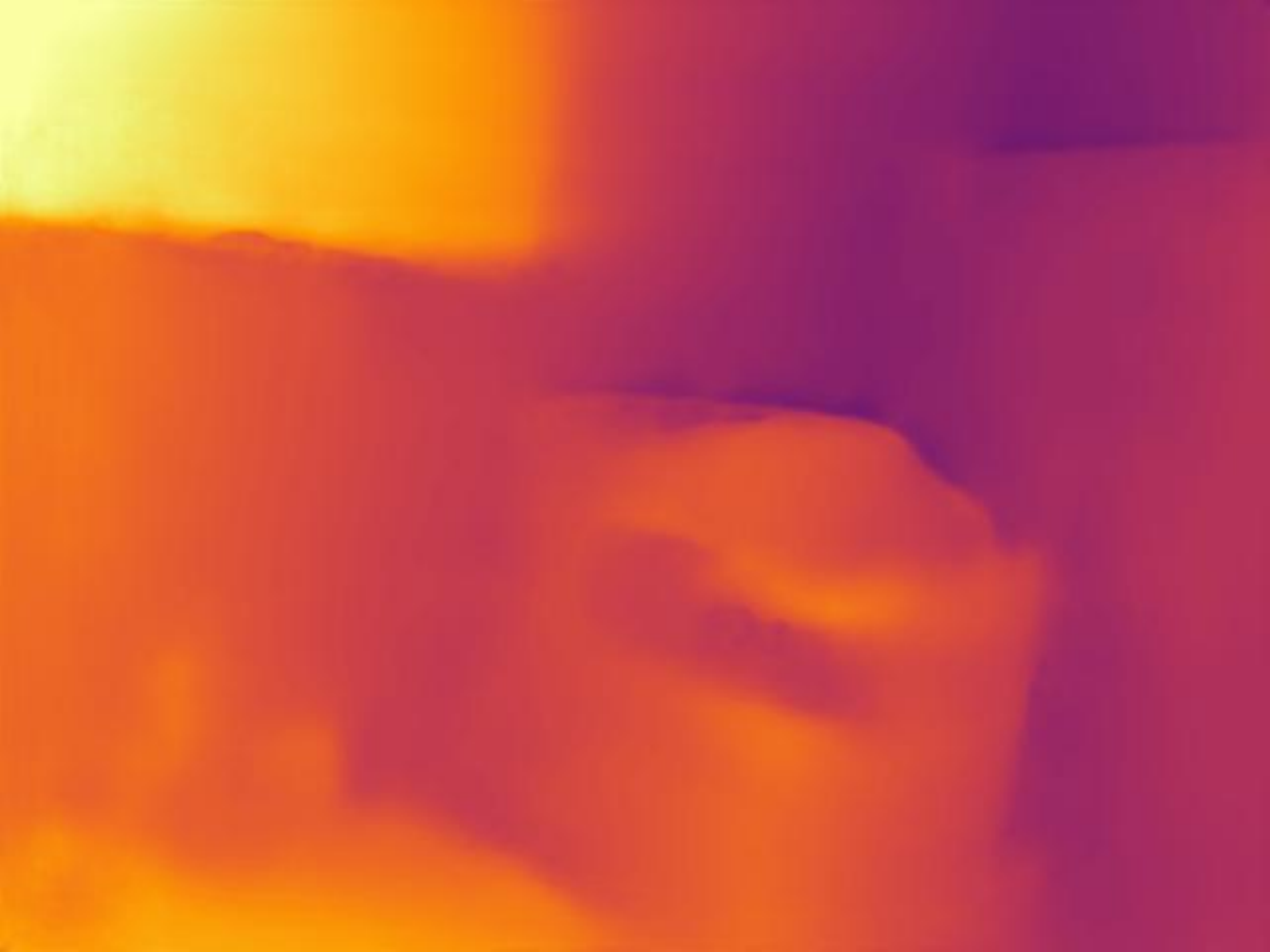}&
    \includegraphics[width=0.096\linewidth]{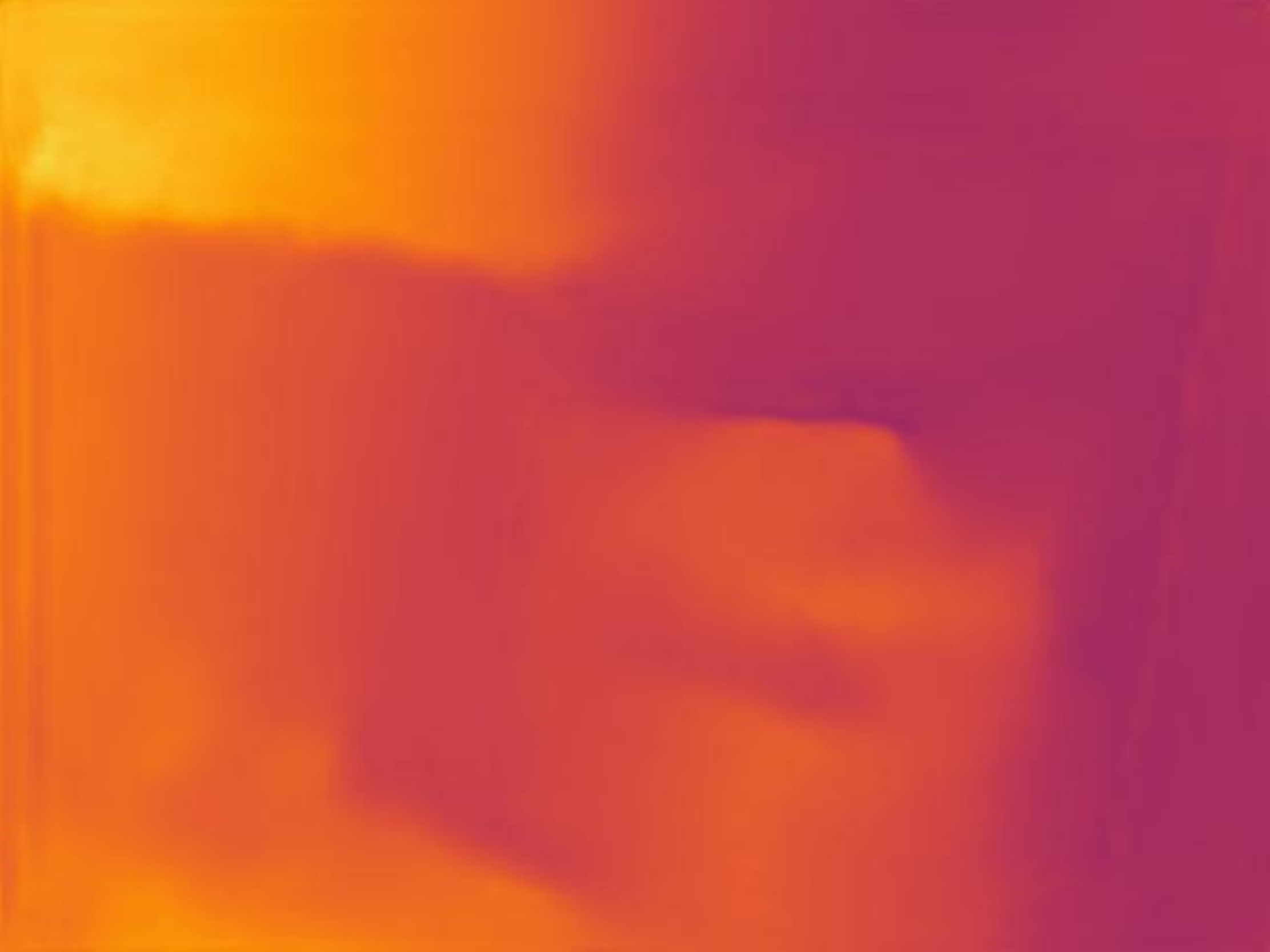}\\
    \vspace{-0.75mm}
    \rot{\scriptsize VOID 1500} &
    \includegraphics[width=0.096\linewidth]{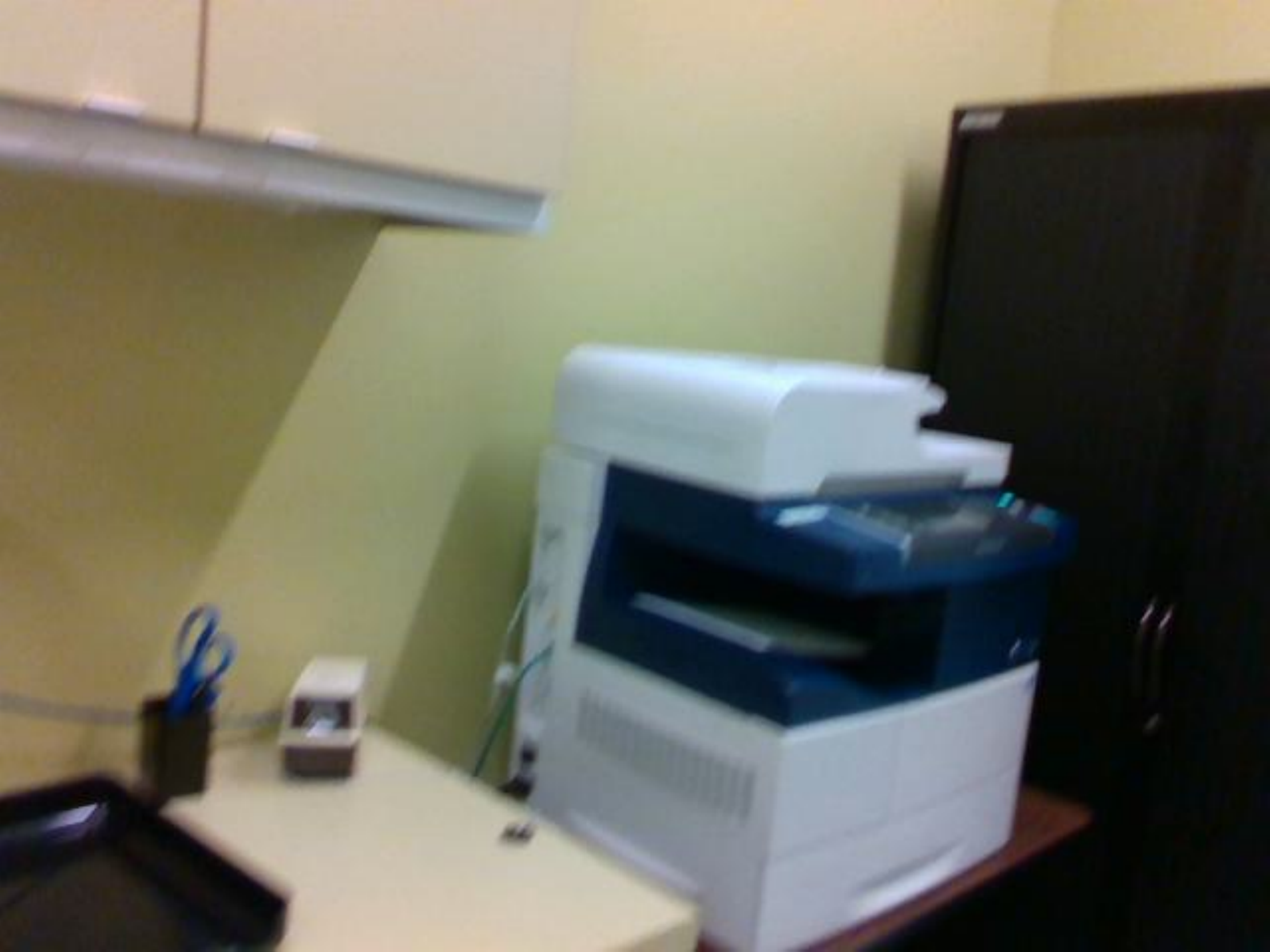}&
    \includegraphics[width=0.096\linewidth]{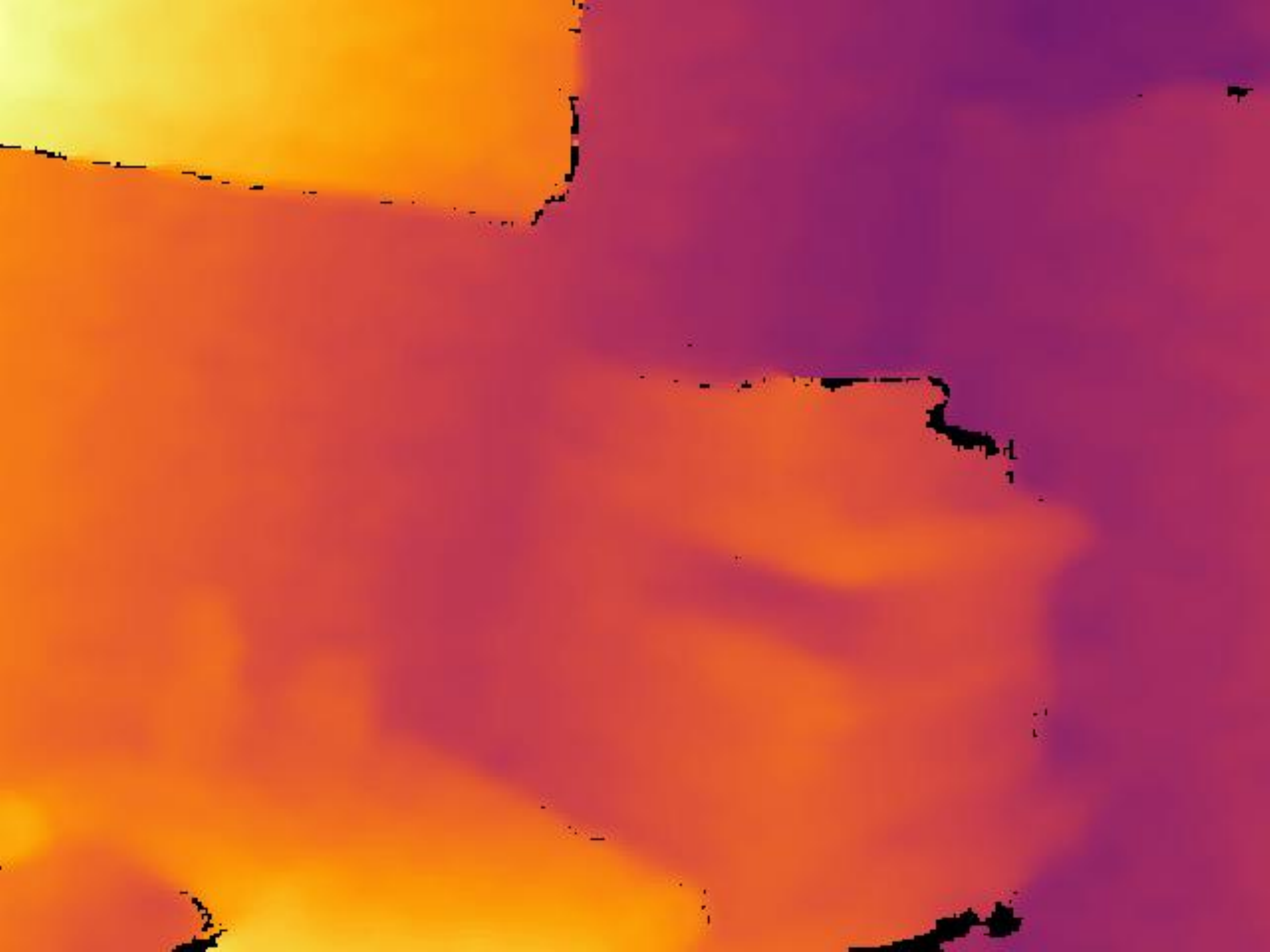}&
    \includegraphics[width=0.096\linewidth]{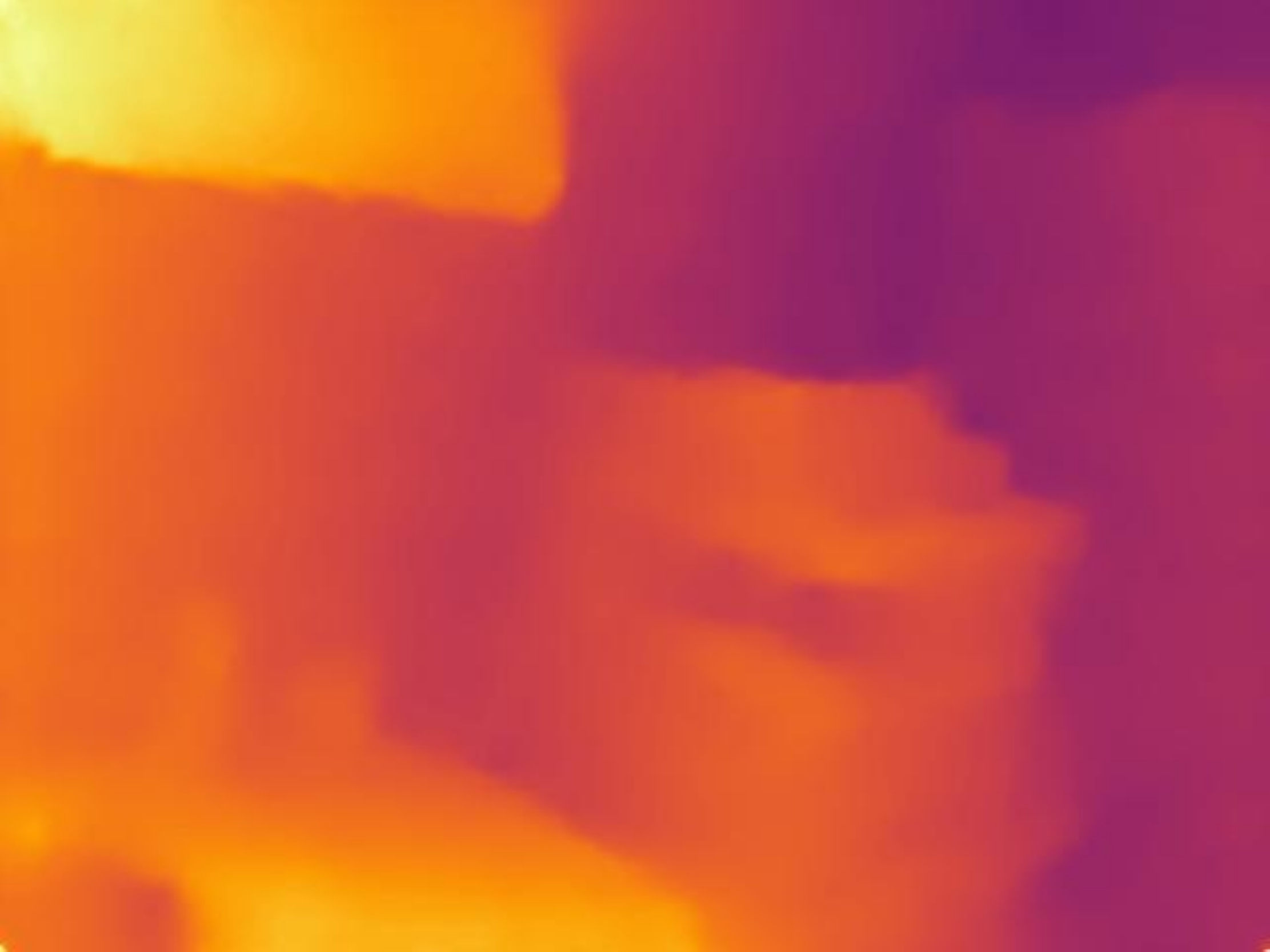}&
    \includegraphics[width=0.096\linewidth]{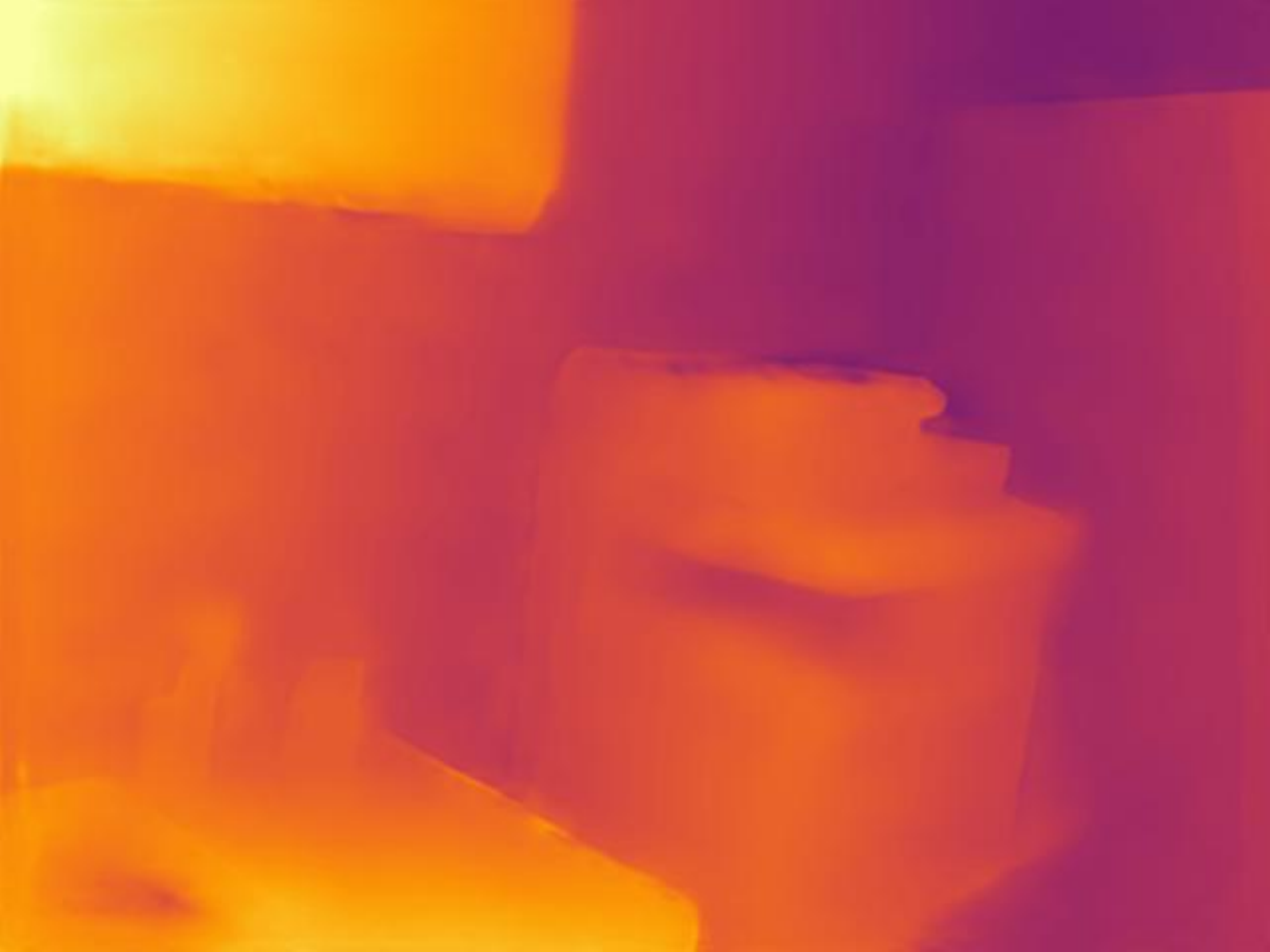}&
    \includegraphics[width=0.096\linewidth]{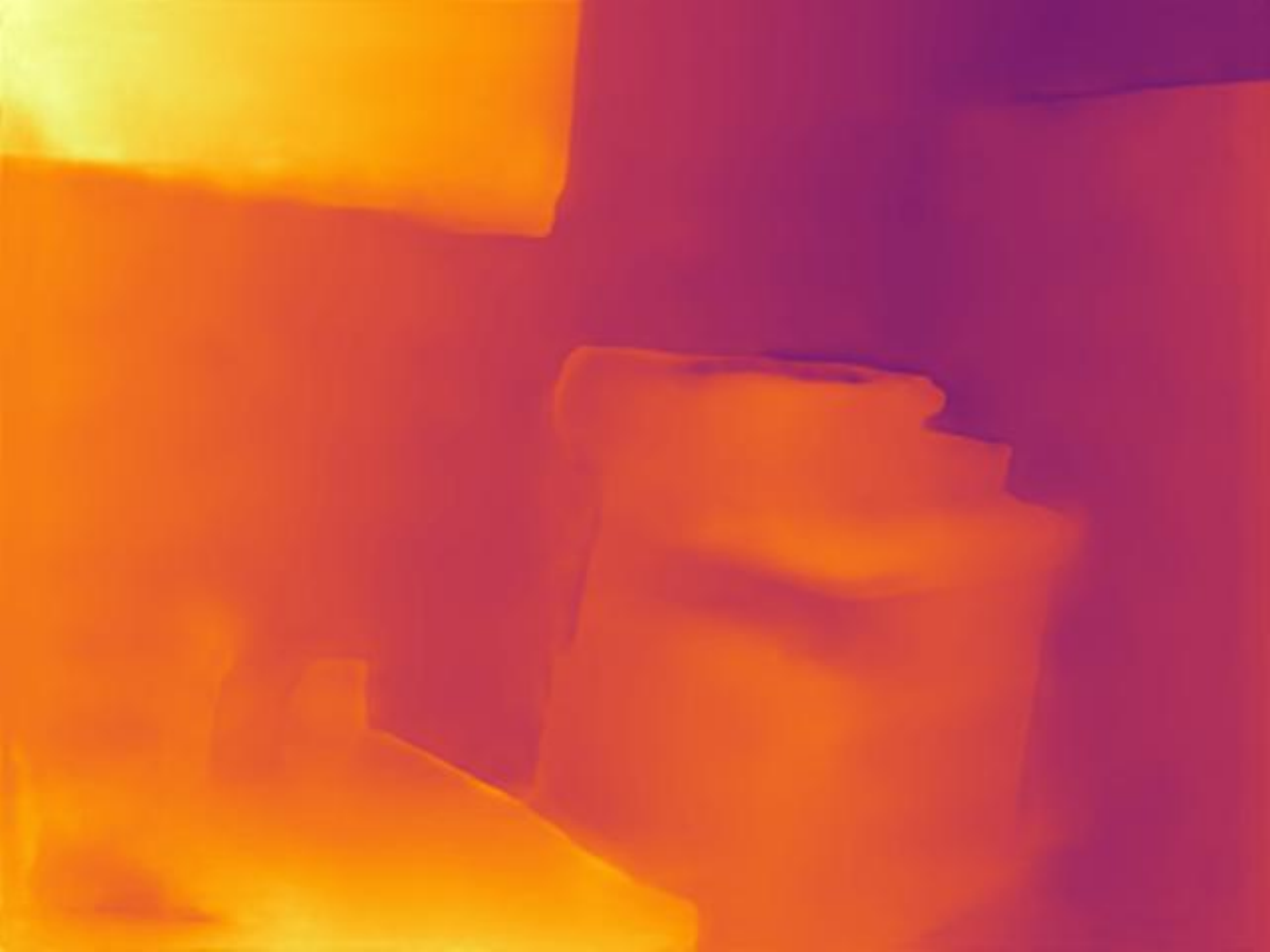}&
    \includegraphics[width=0.096\linewidth]{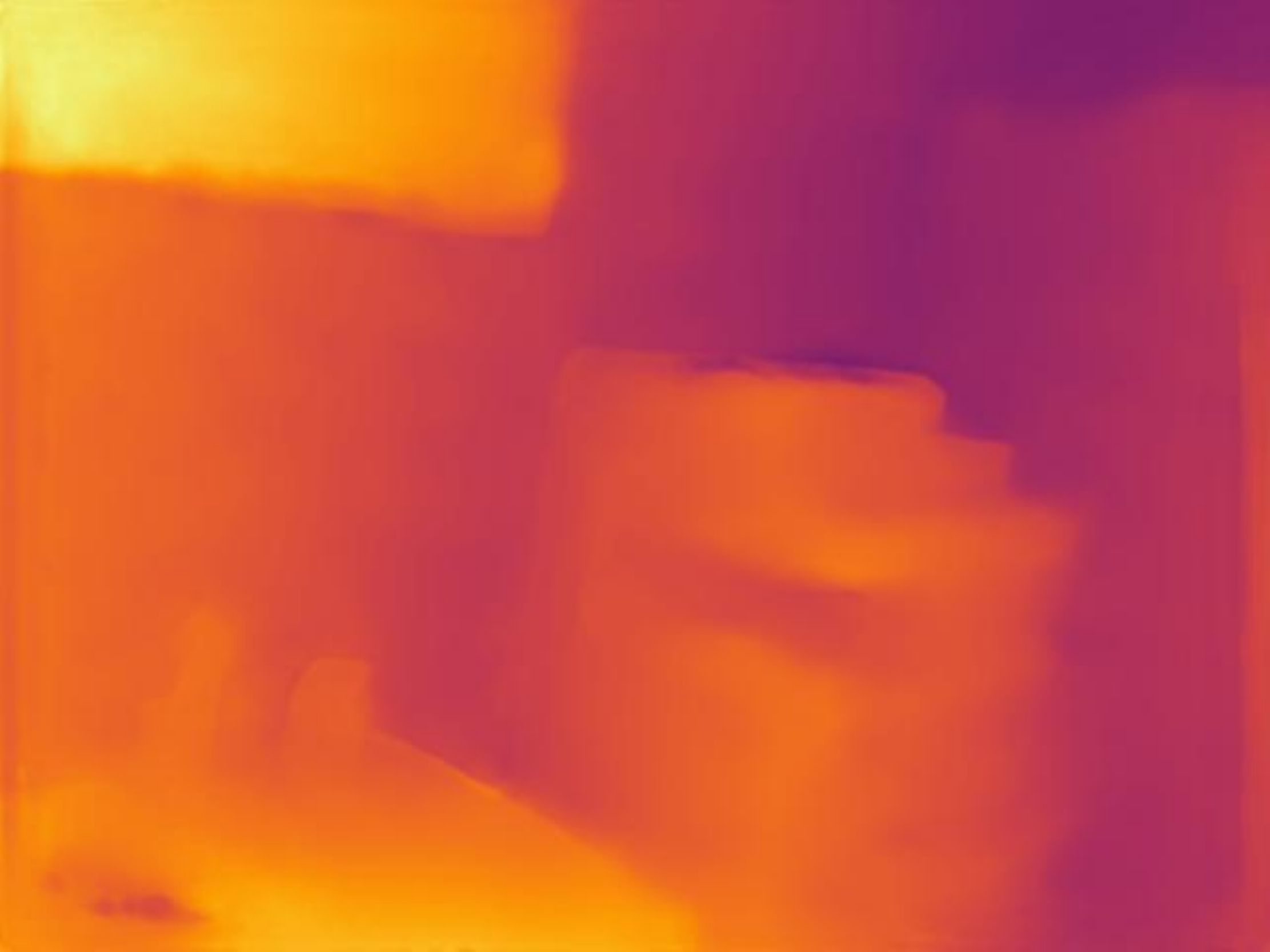}&
    \includegraphics[width=0.096\linewidth]{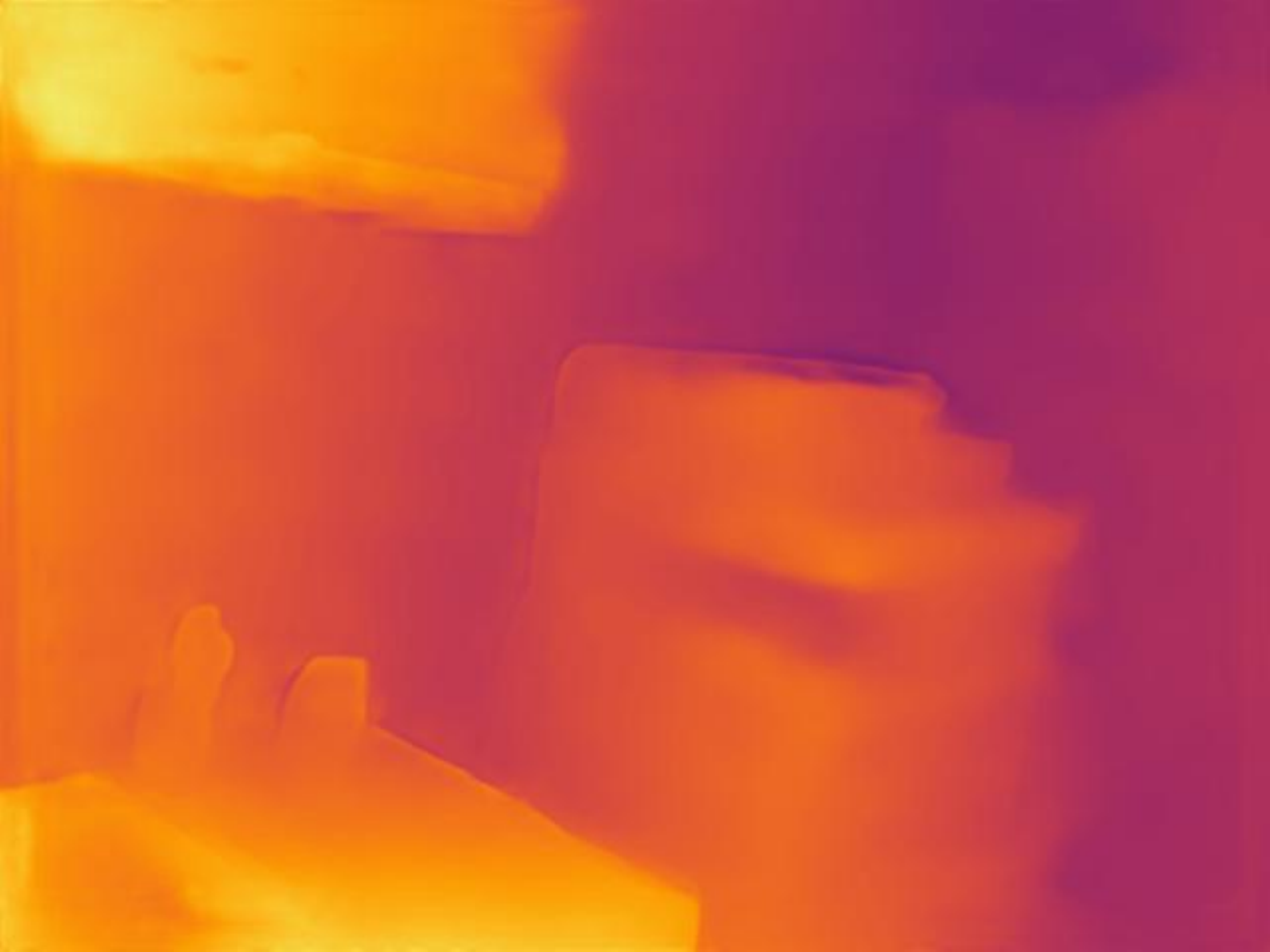}&
    \includegraphics[width=0.096\linewidth]{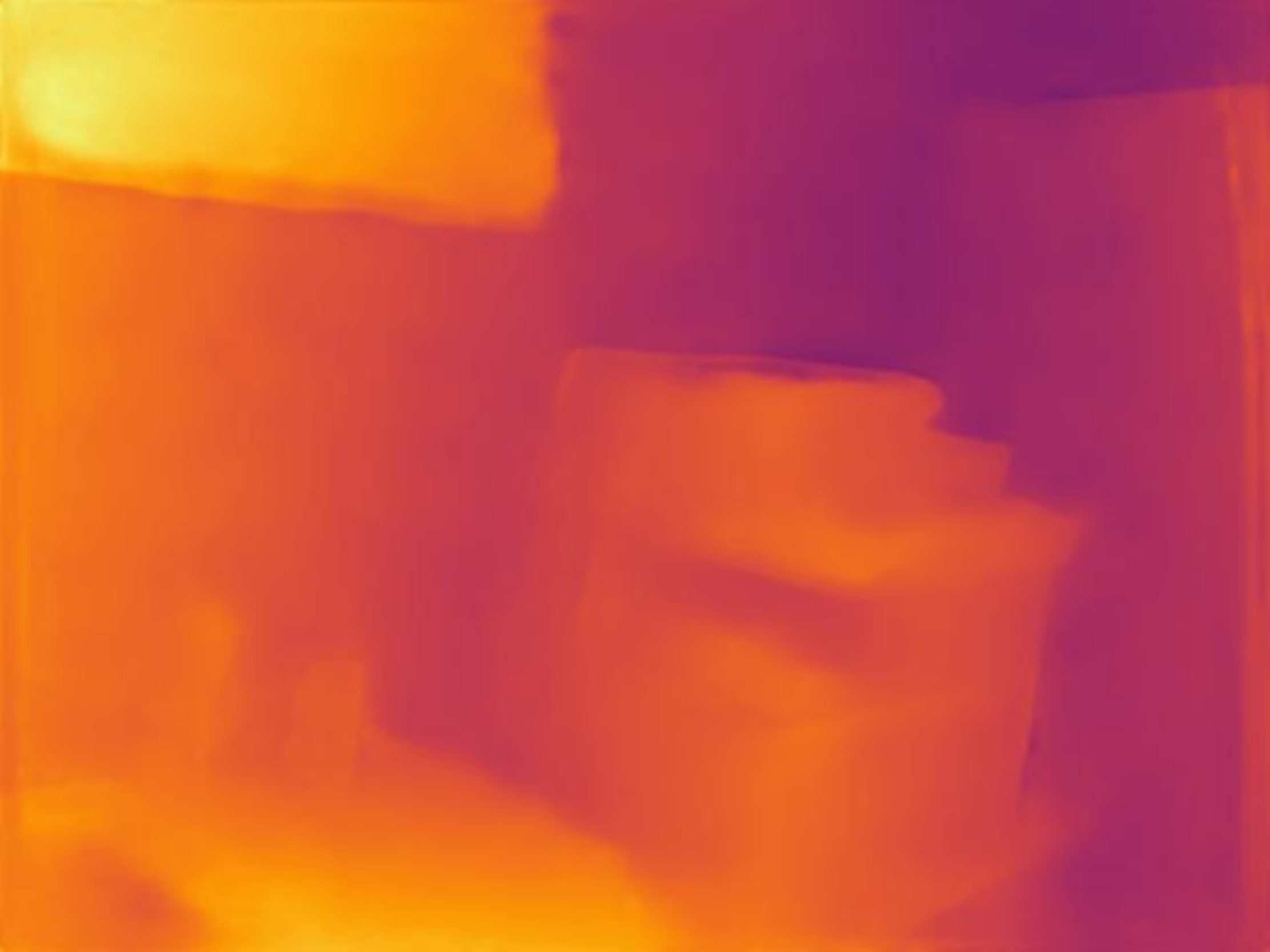}&
    \includegraphics[width=0.096\linewidth]{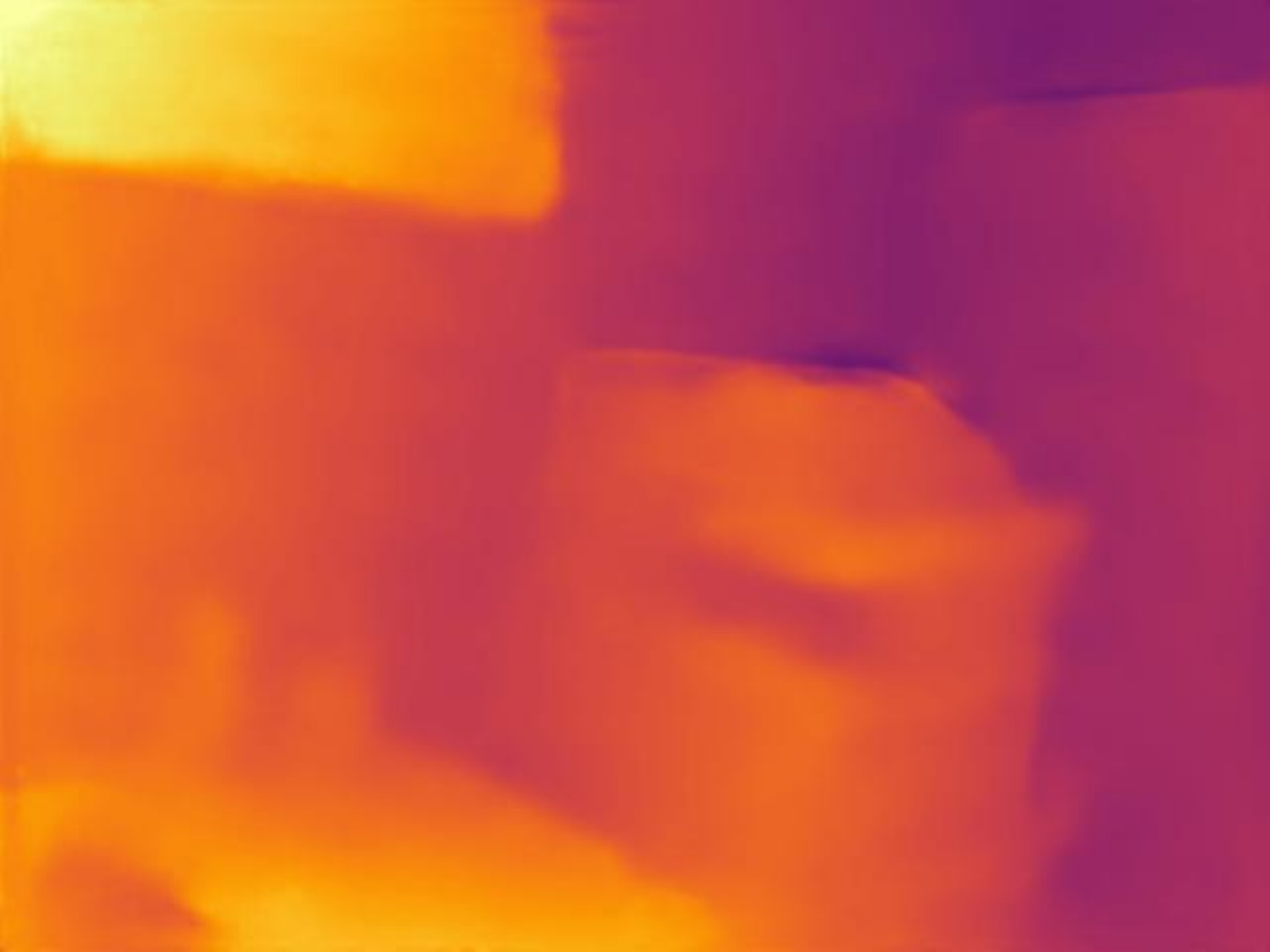}&
    \includegraphics[width=0.096\linewidth]{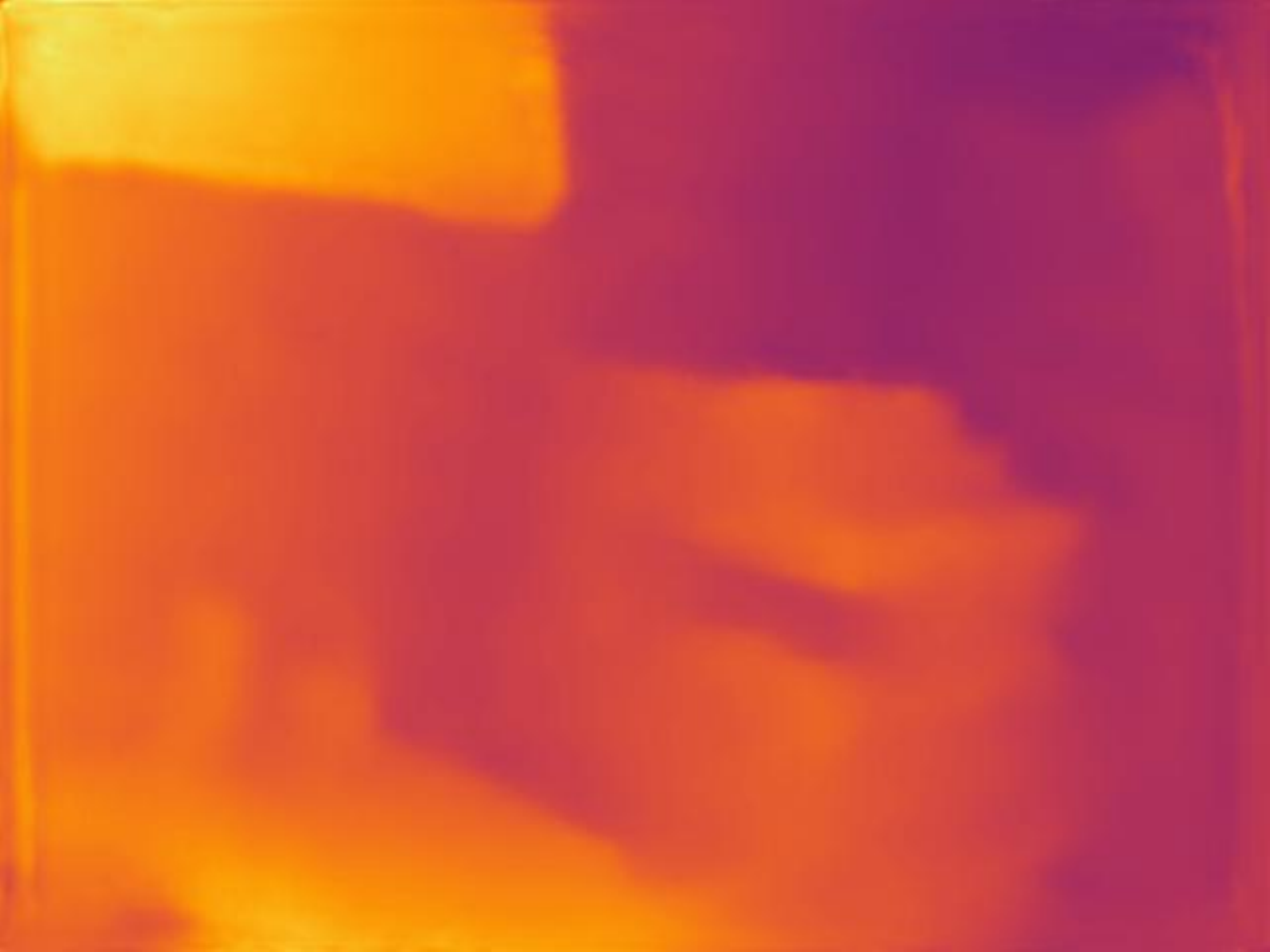}\\
    \multicolumn{9}{c}{} \\
    \vspace{-0.75mm}
    \rot{\scriptsize VOID 150} &
    \includegraphics[width=0.096\linewidth]{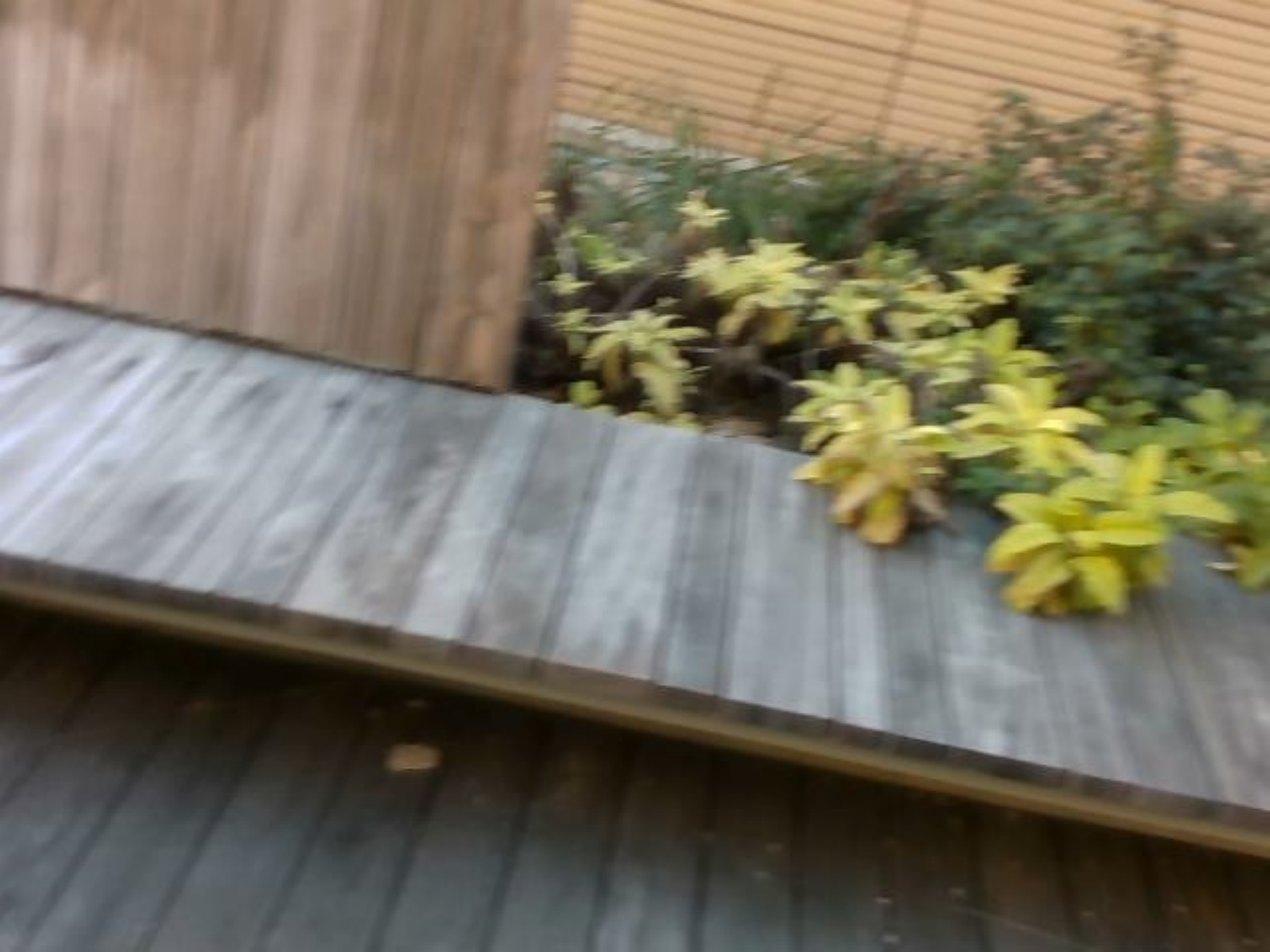}&
    \includegraphics[width=0.096\linewidth]{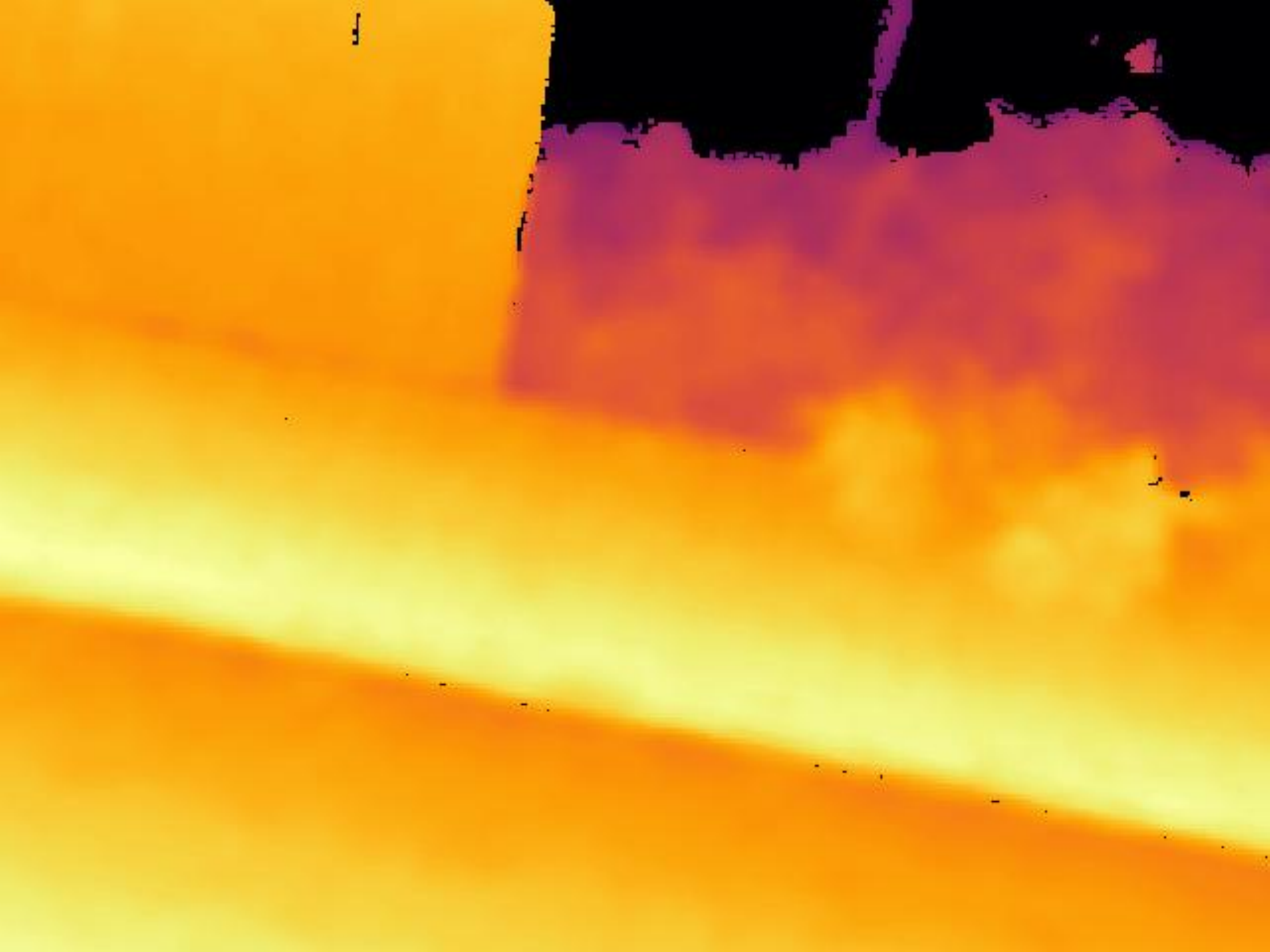}&
    \includegraphics[width=0.096\linewidth]{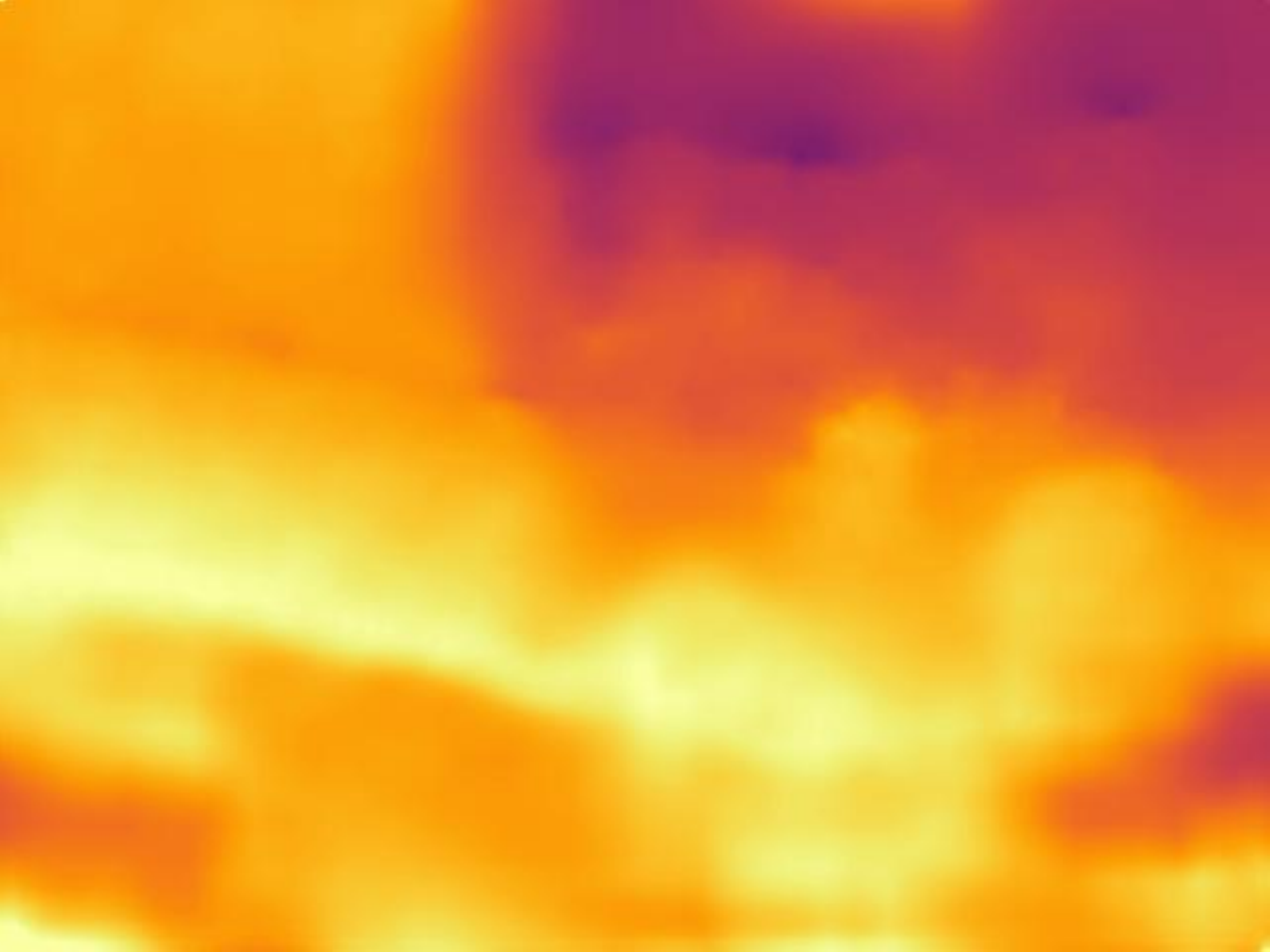}&
    \includegraphics[width=0.096\linewidth]{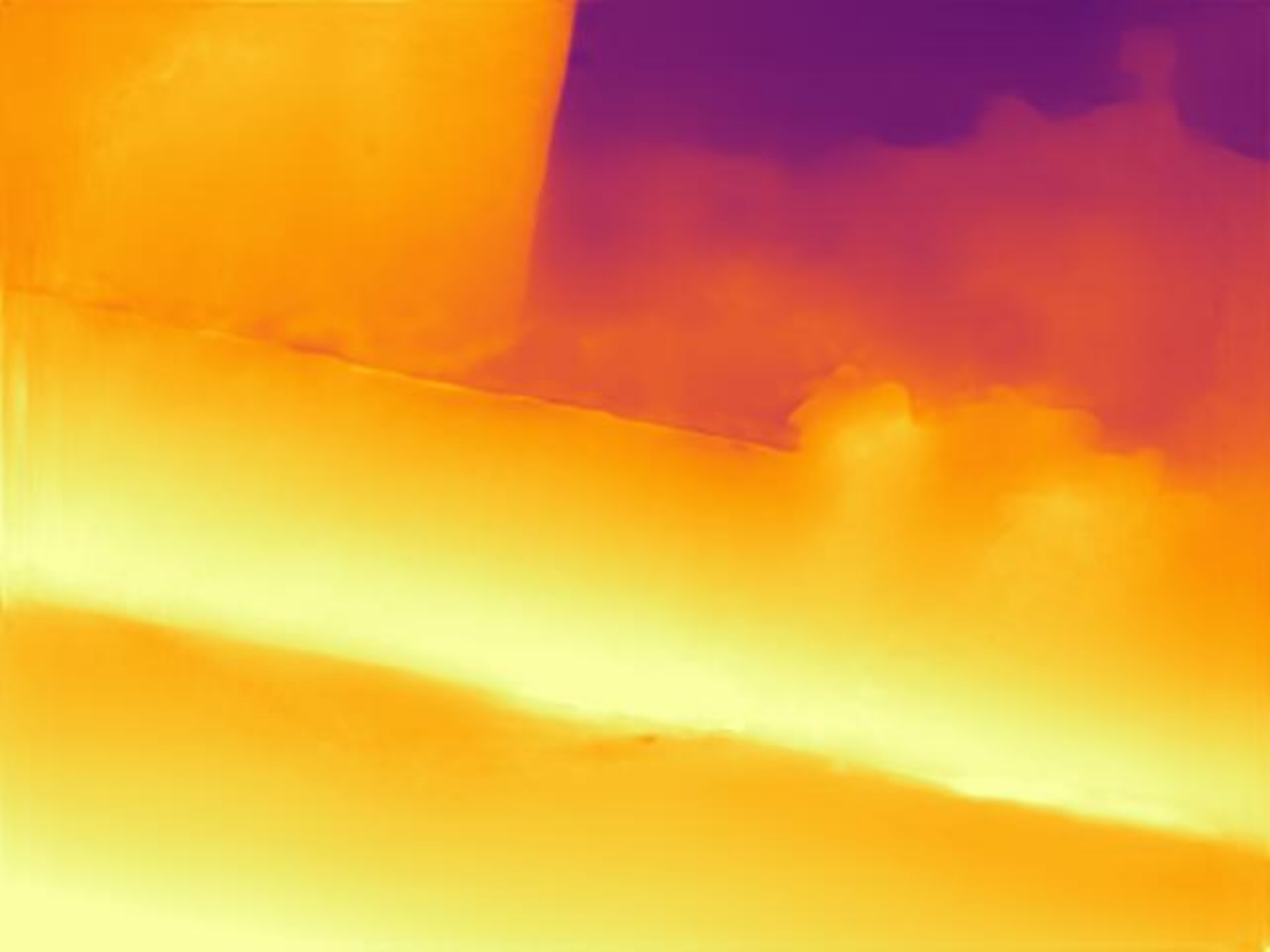}&
    \includegraphics[width=0.096\linewidth]{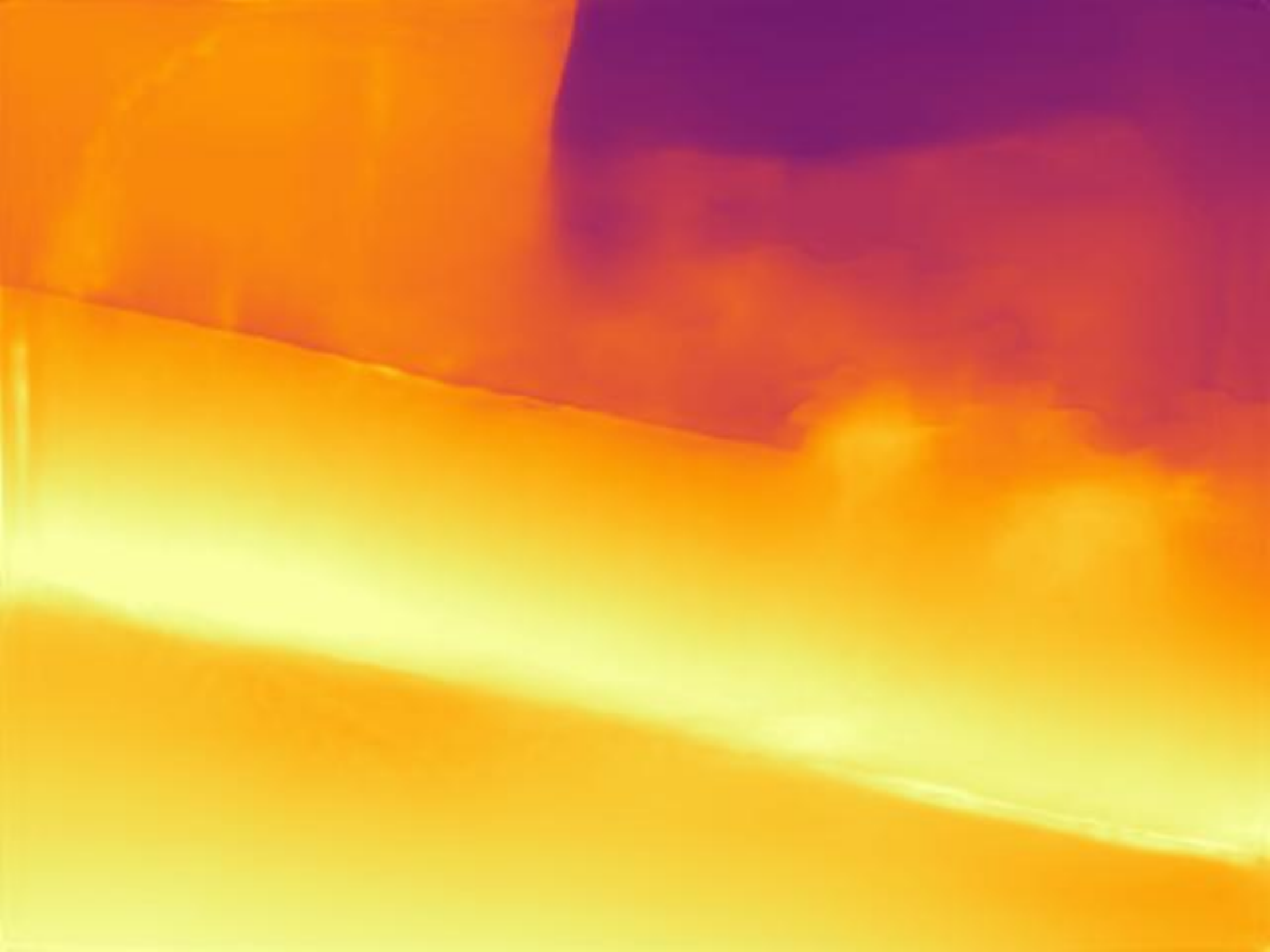}&
    \includegraphics[width=0.096\linewidth]{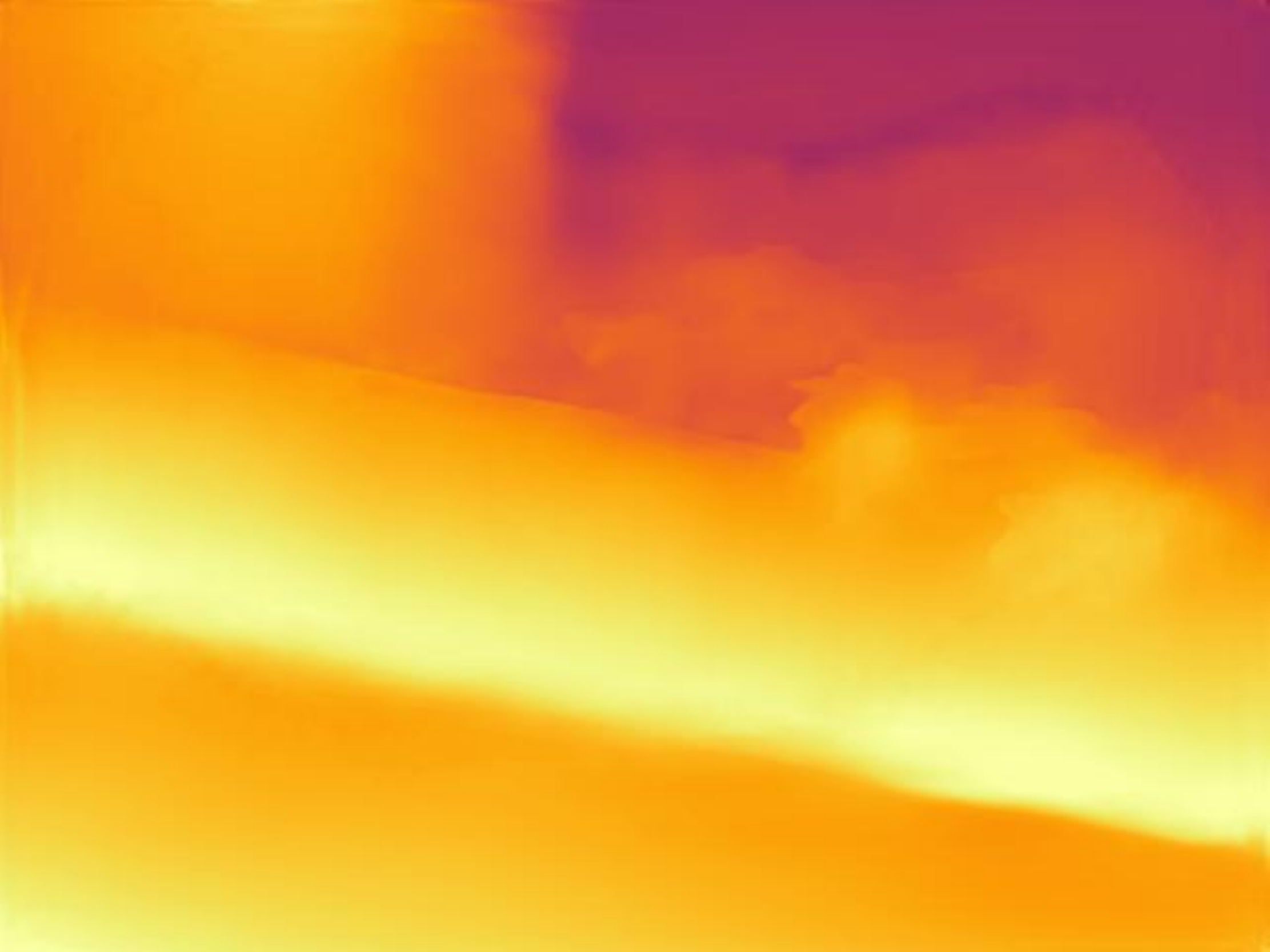}&
    \includegraphics[width=0.096\linewidth]{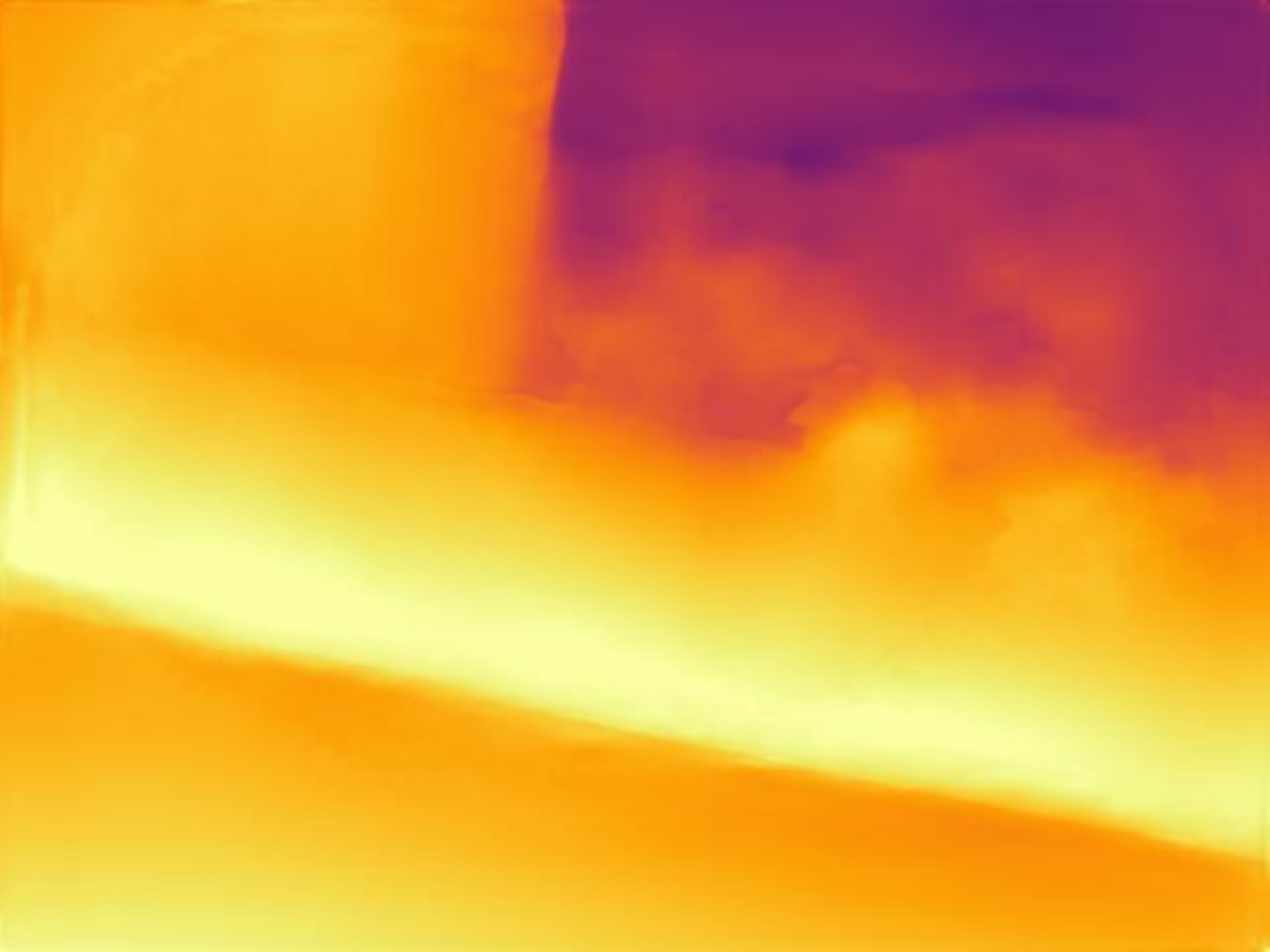}&
    \includegraphics[width=0.096\linewidth]{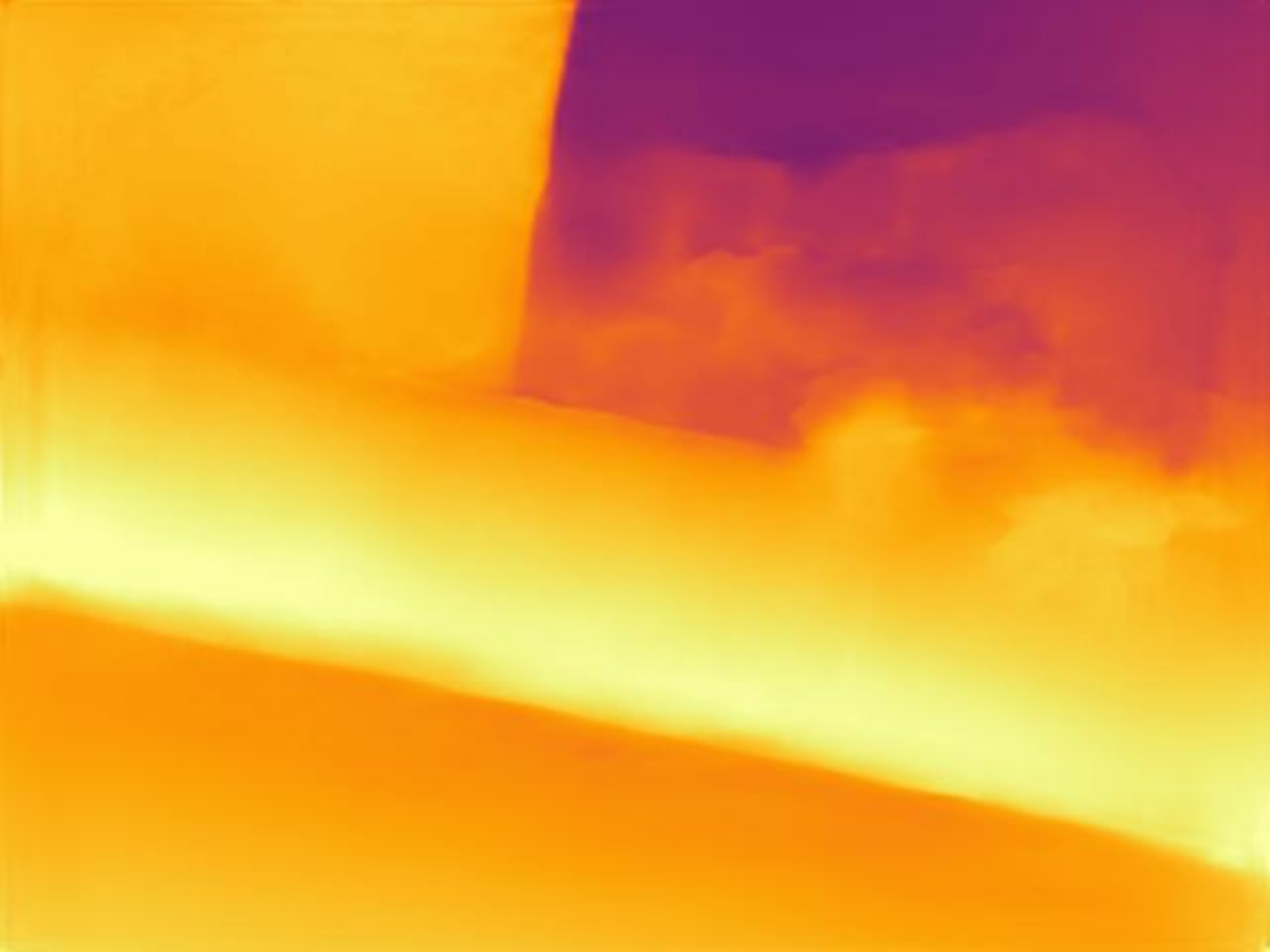}&
    \includegraphics[width=0.096\linewidth]{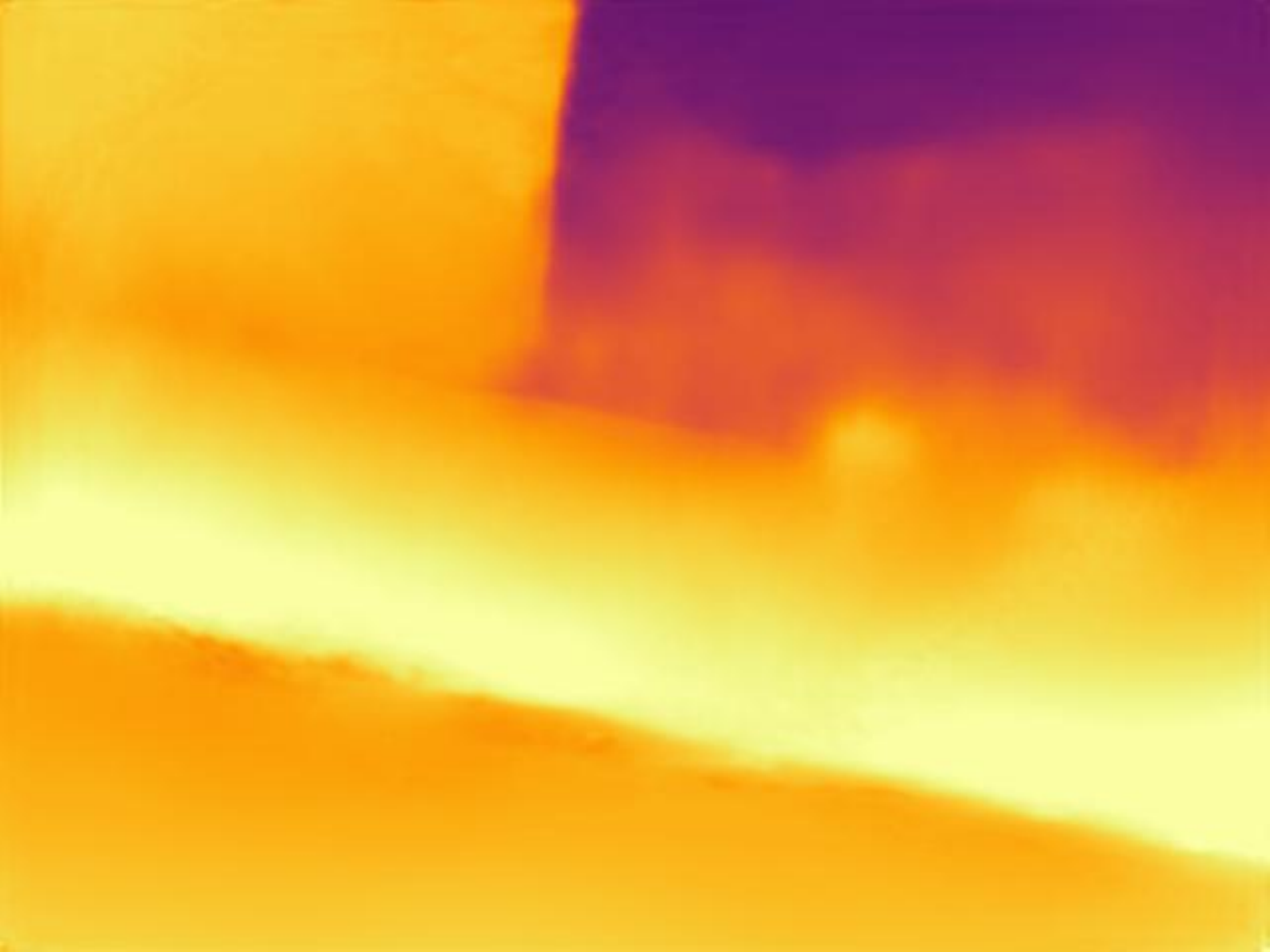}&
    \includegraphics[width=0.096\linewidth]{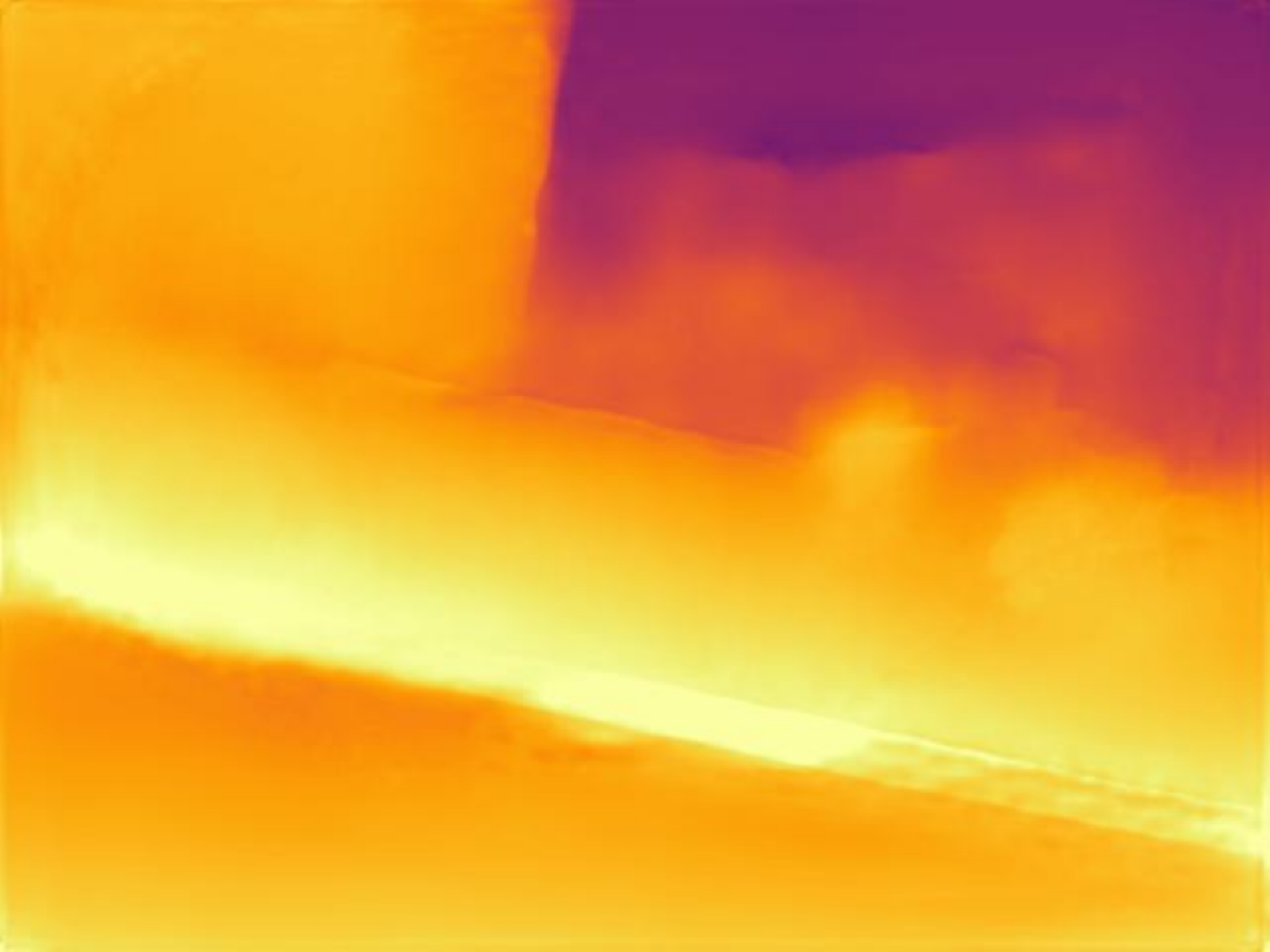}\\
    \vspace{-0.75mm}
    \rot{\scriptsize VOID 500} &
    \includegraphics[width=0.096\linewidth]{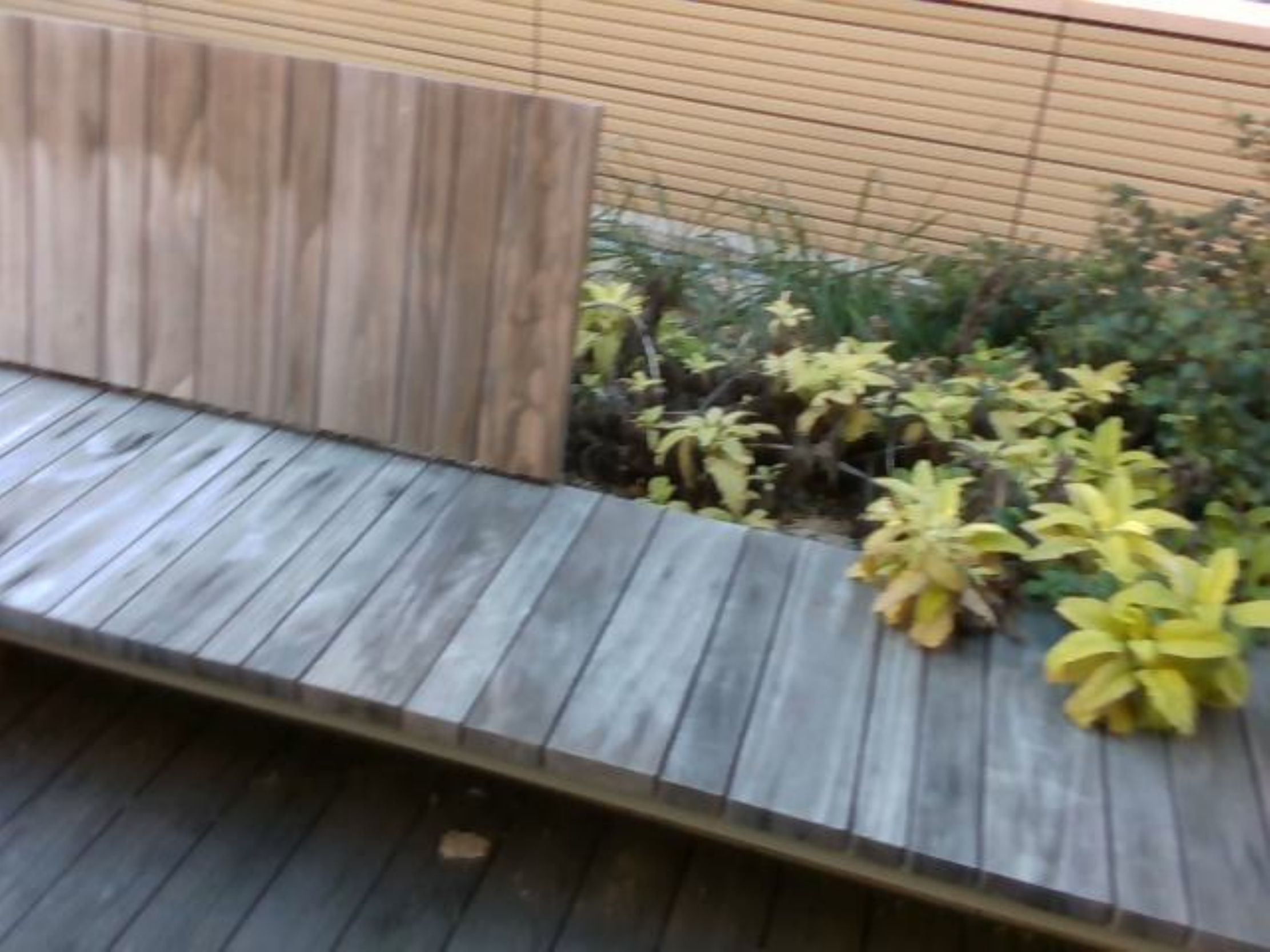}&
    \includegraphics[width=0.096\linewidth]{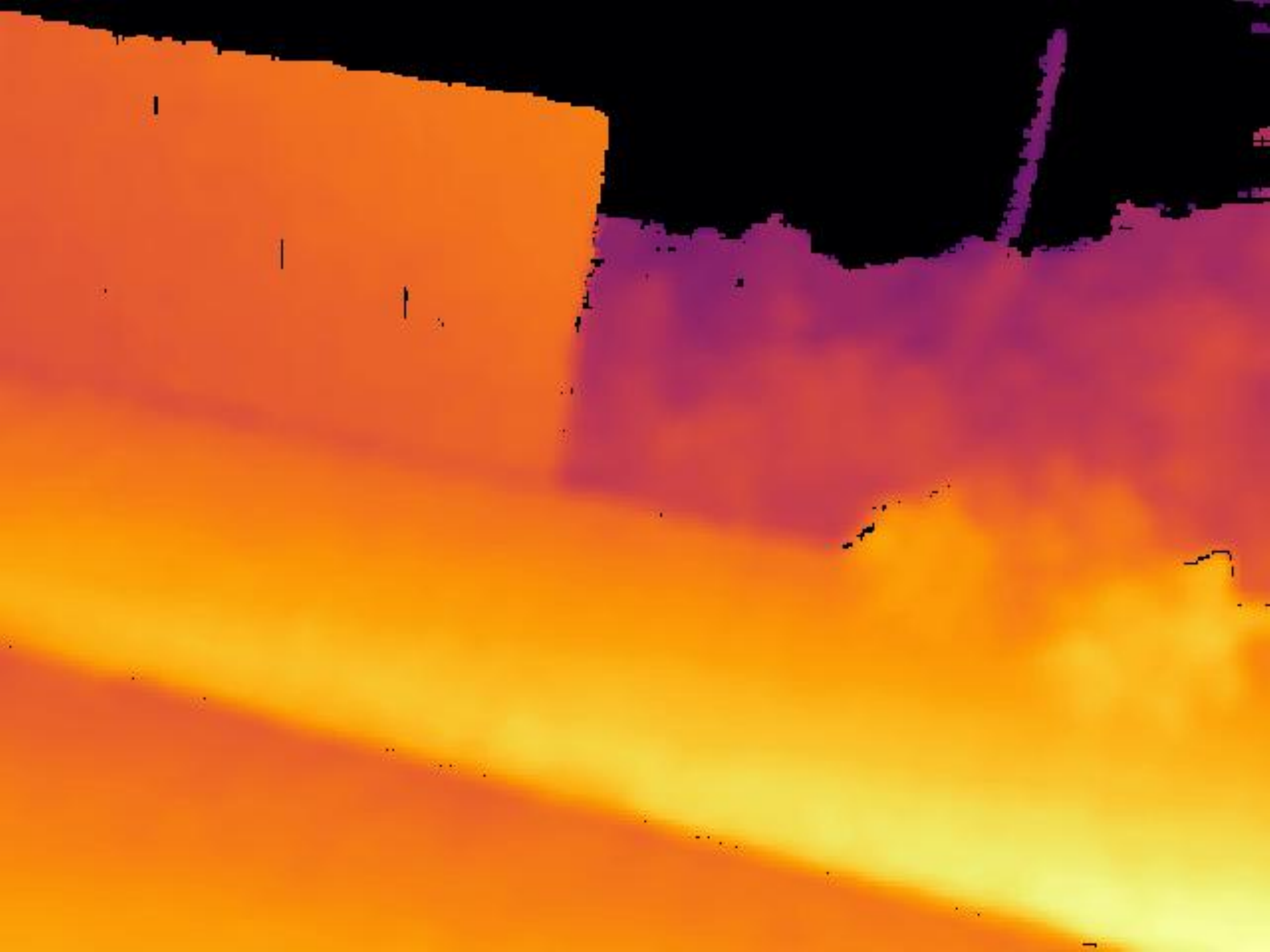}&
    \includegraphics[width=0.096\linewidth]{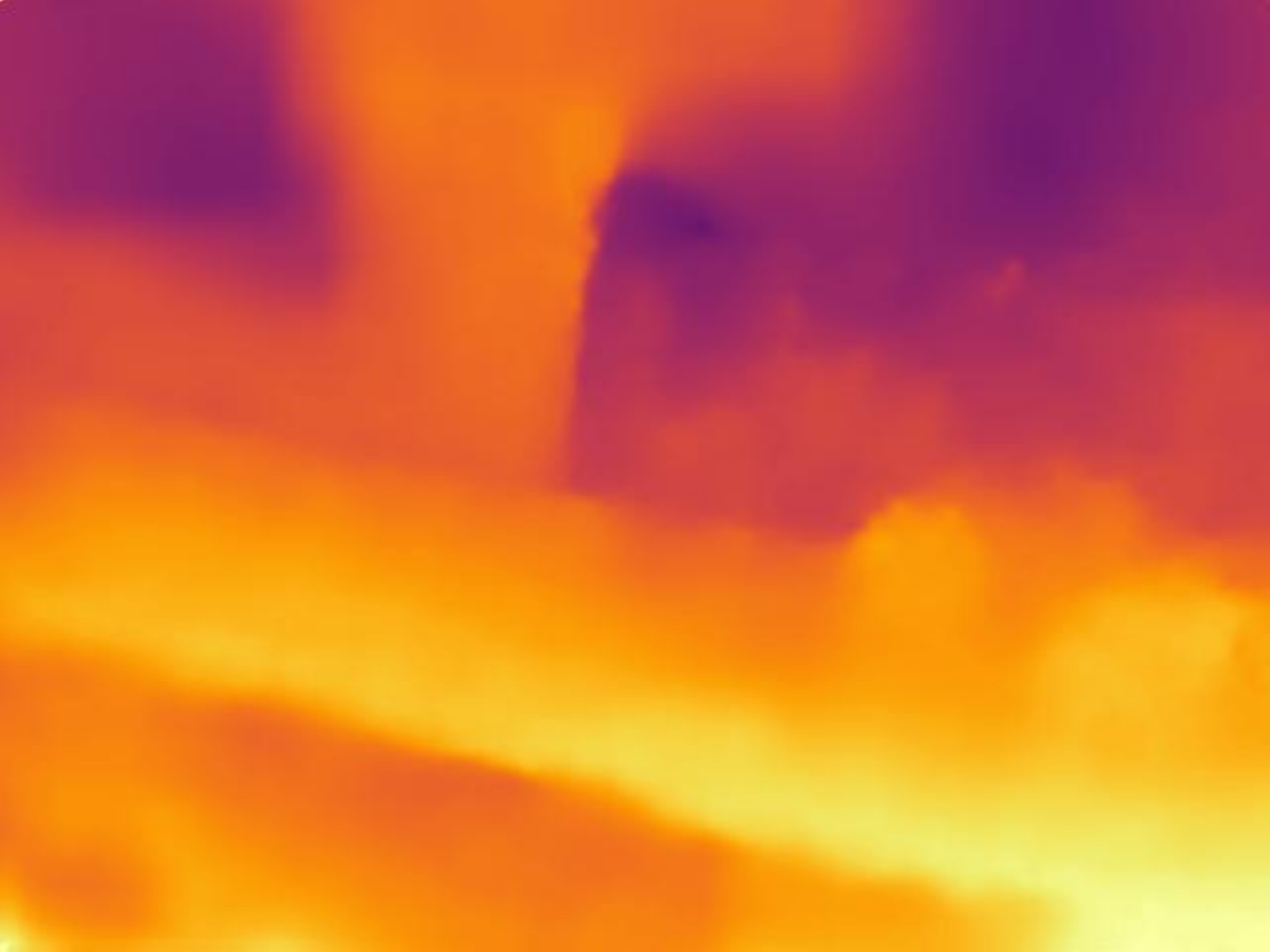}&
    \includegraphics[width=0.096\linewidth]{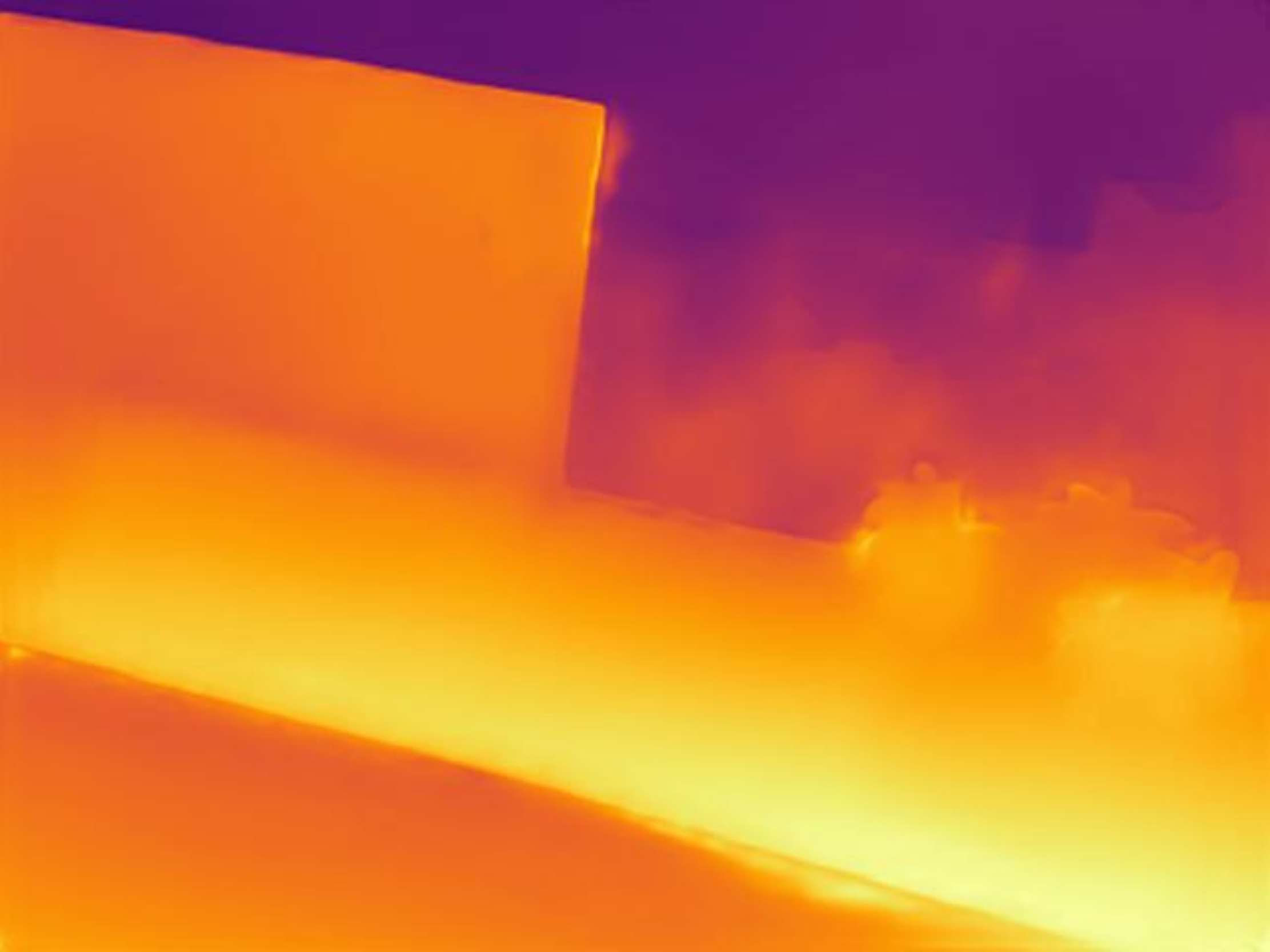}&
    \includegraphics[width=0.096\linewidth]{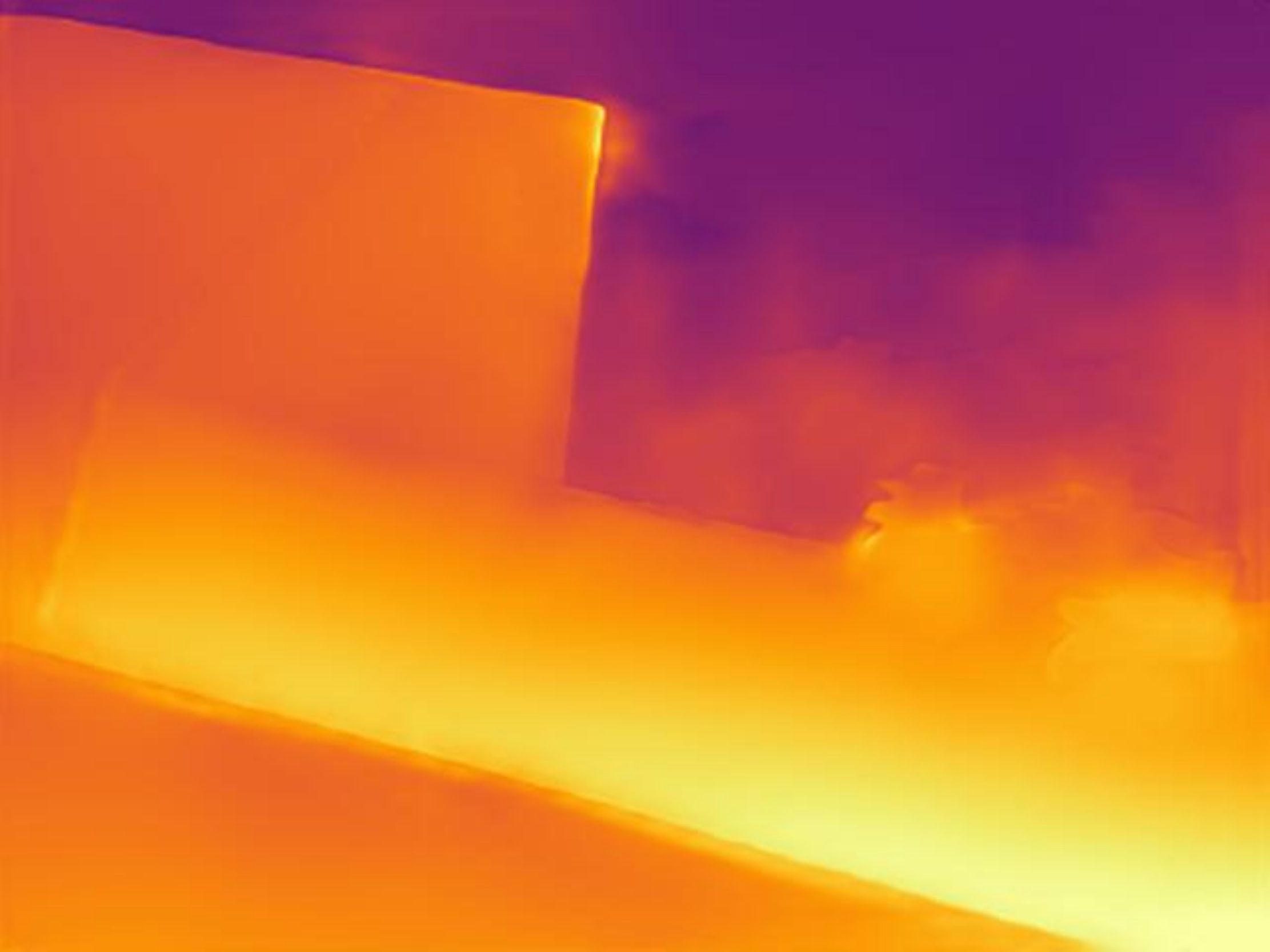}&
    \includegraphics[width=0.096\linewidth]{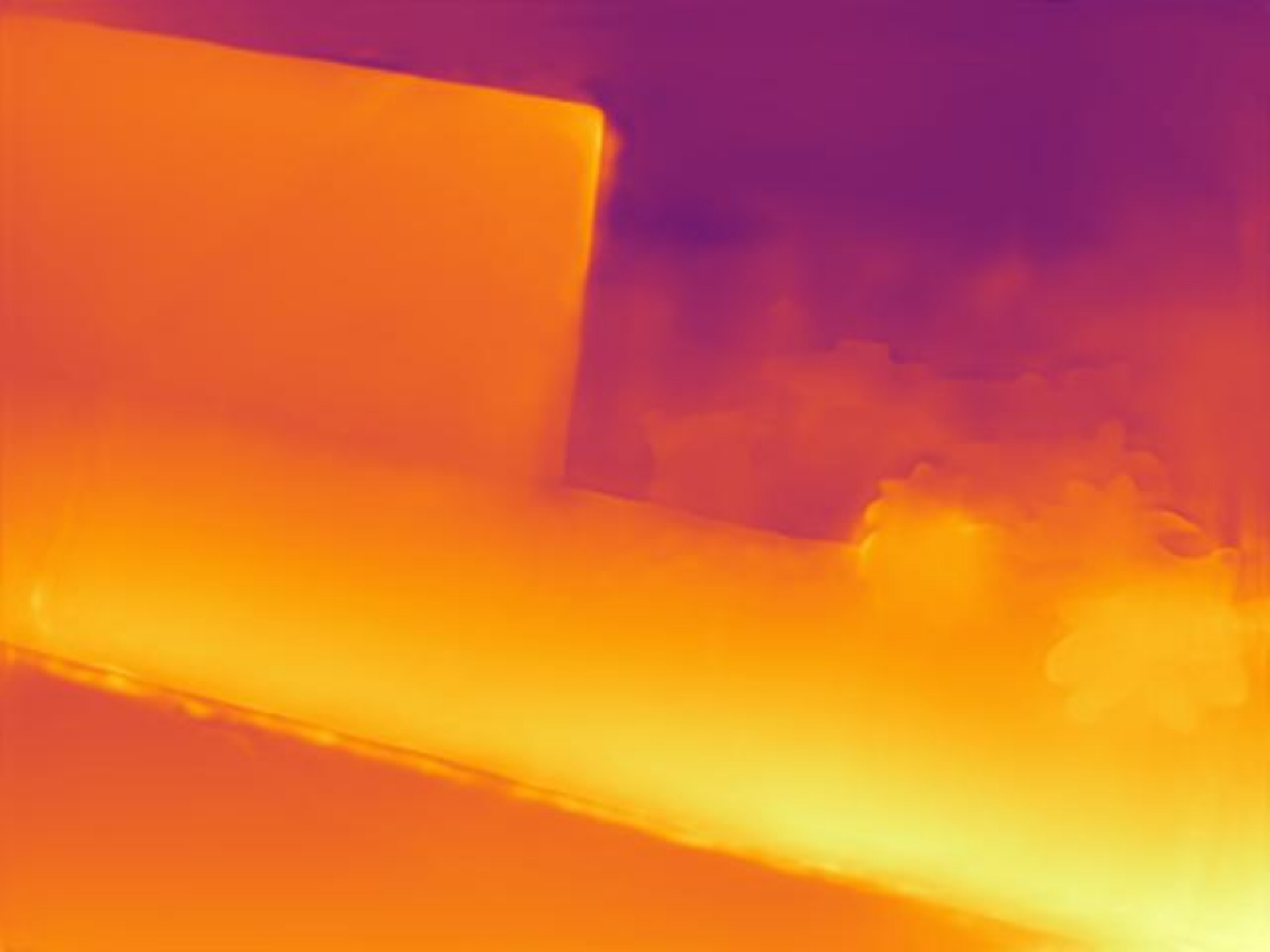}&
    \includegraphics[width=0.096\linewidth]{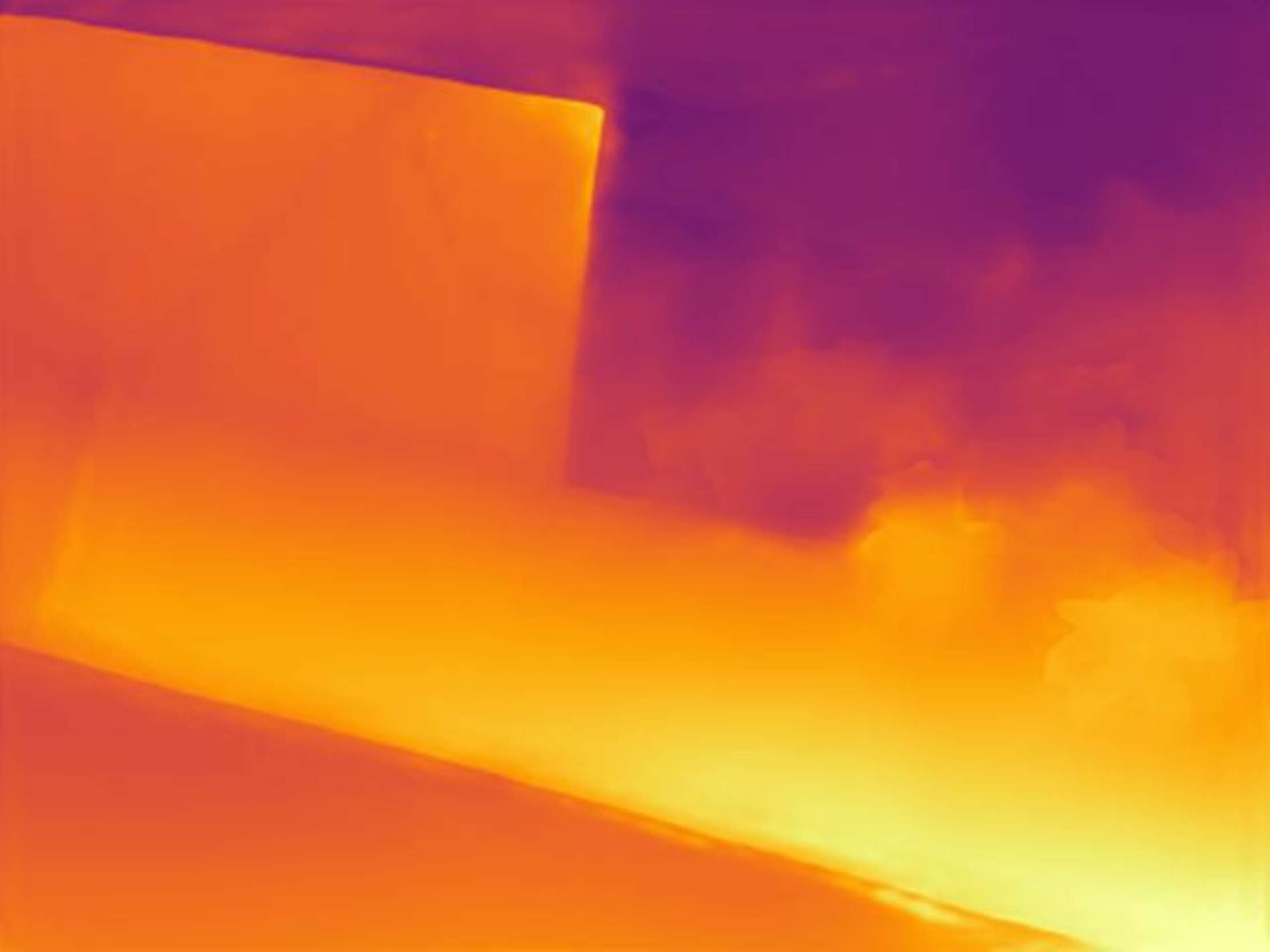}&
    \includegraphics[width=0.096\linewidth]{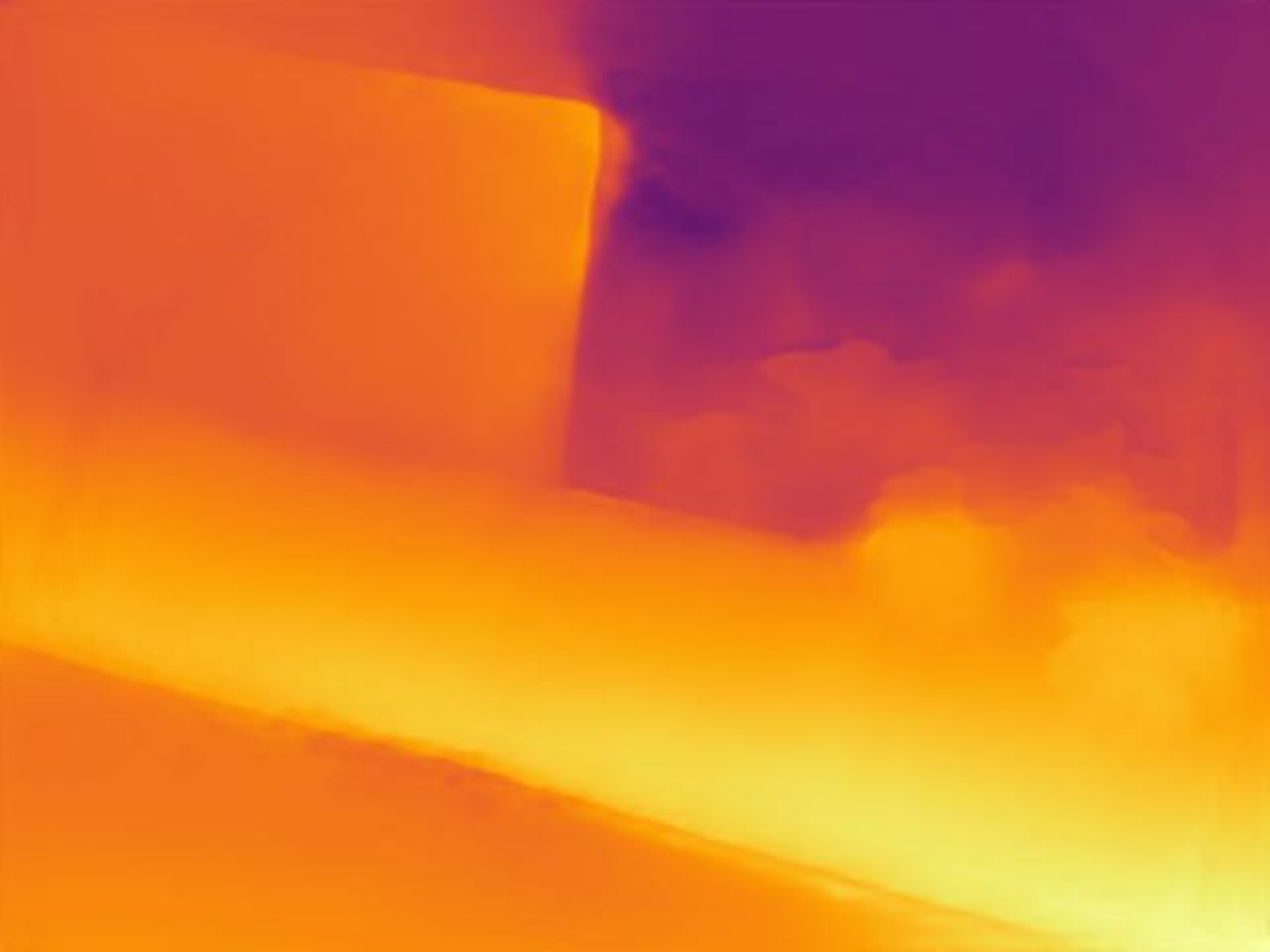}&
    \includegraphics[width=0.096\linewidth]{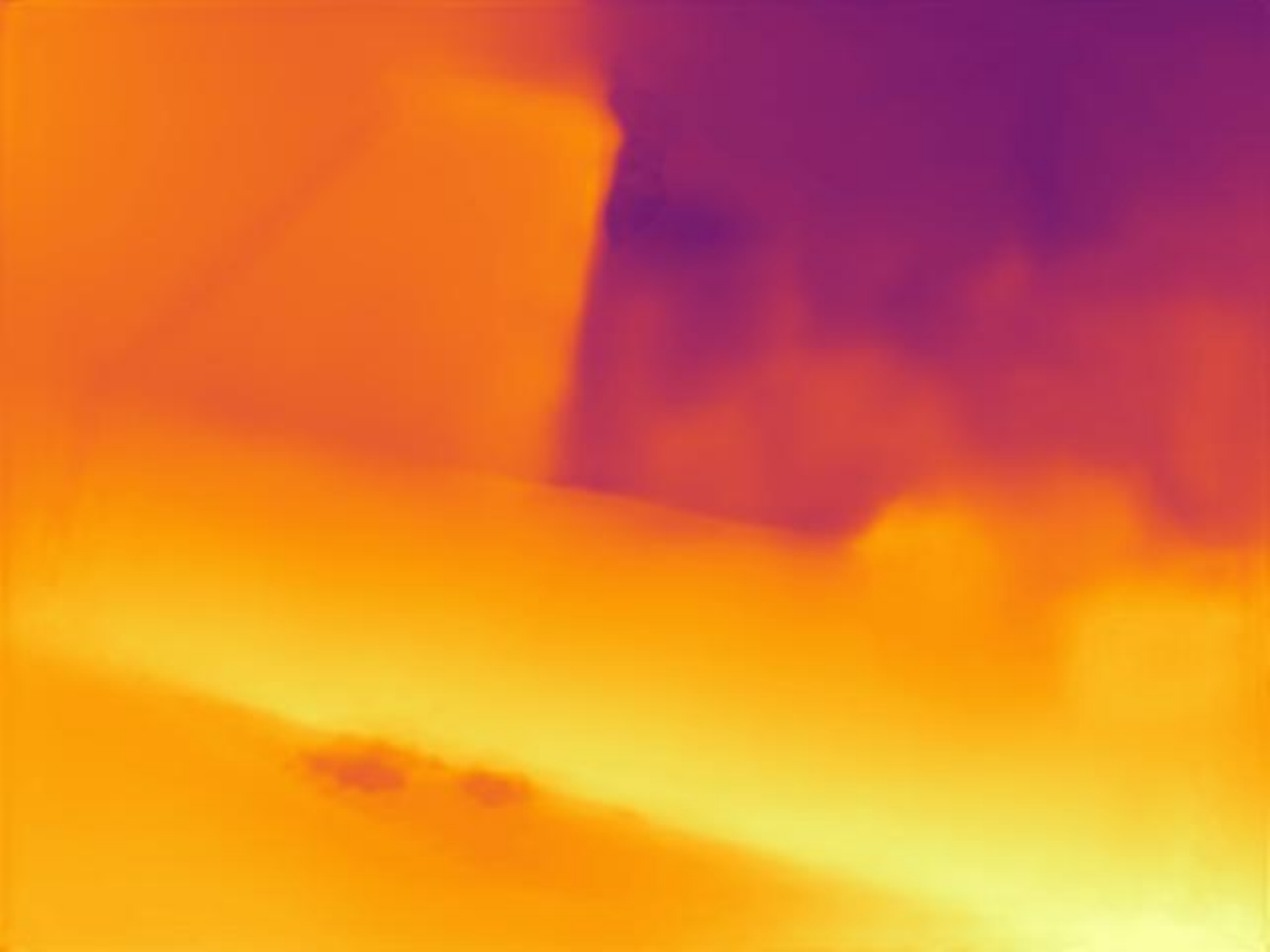}&
    \includegraphics[width=0.096\linewidth]{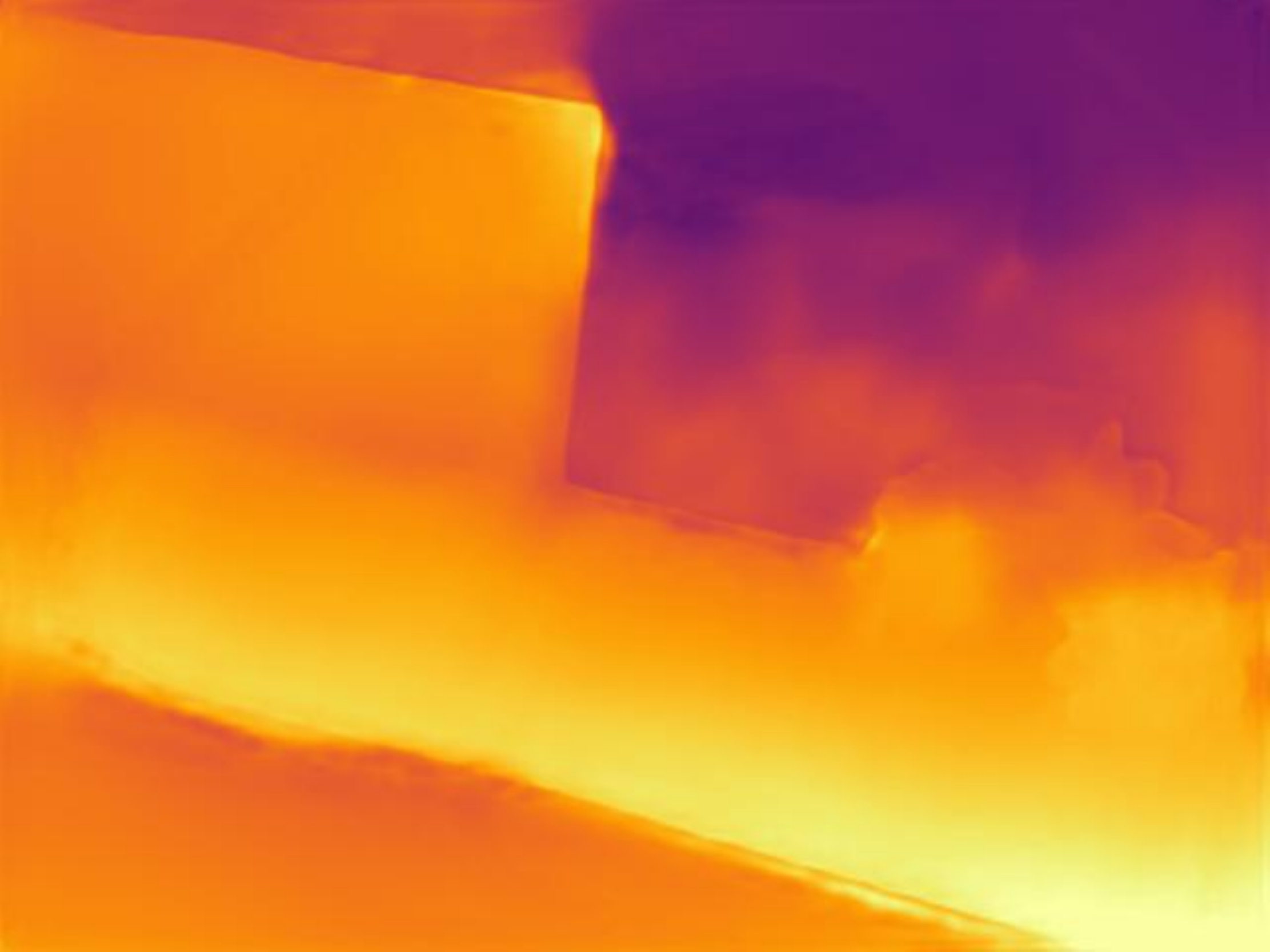}\\
    \vspace{-0.75mm}
    \rot{\scriptsize VOID 1500} &
    \includegraphics[width=0.096\linewidth]{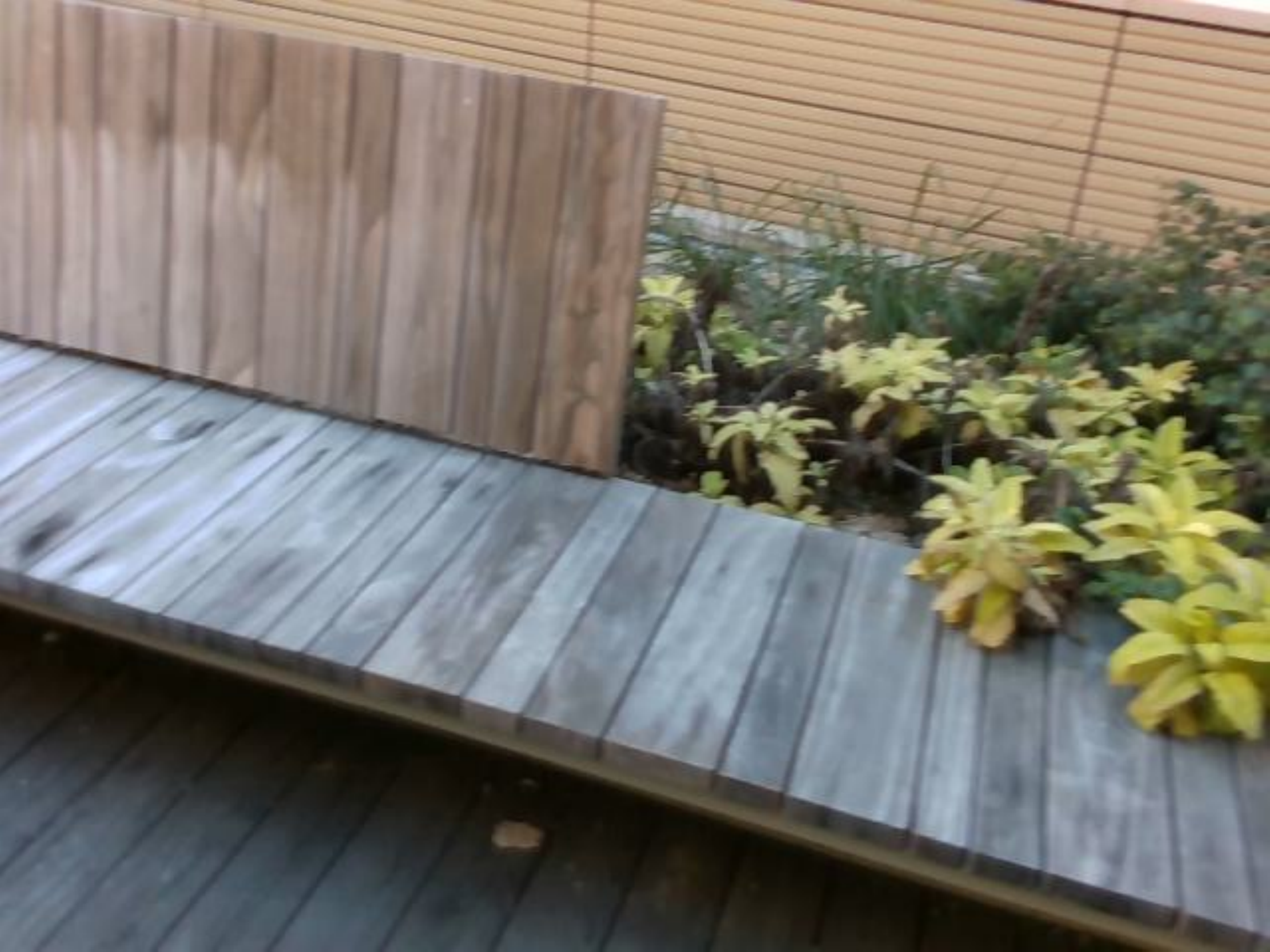}&
    \includegraphics[width=0.096\linewidth]{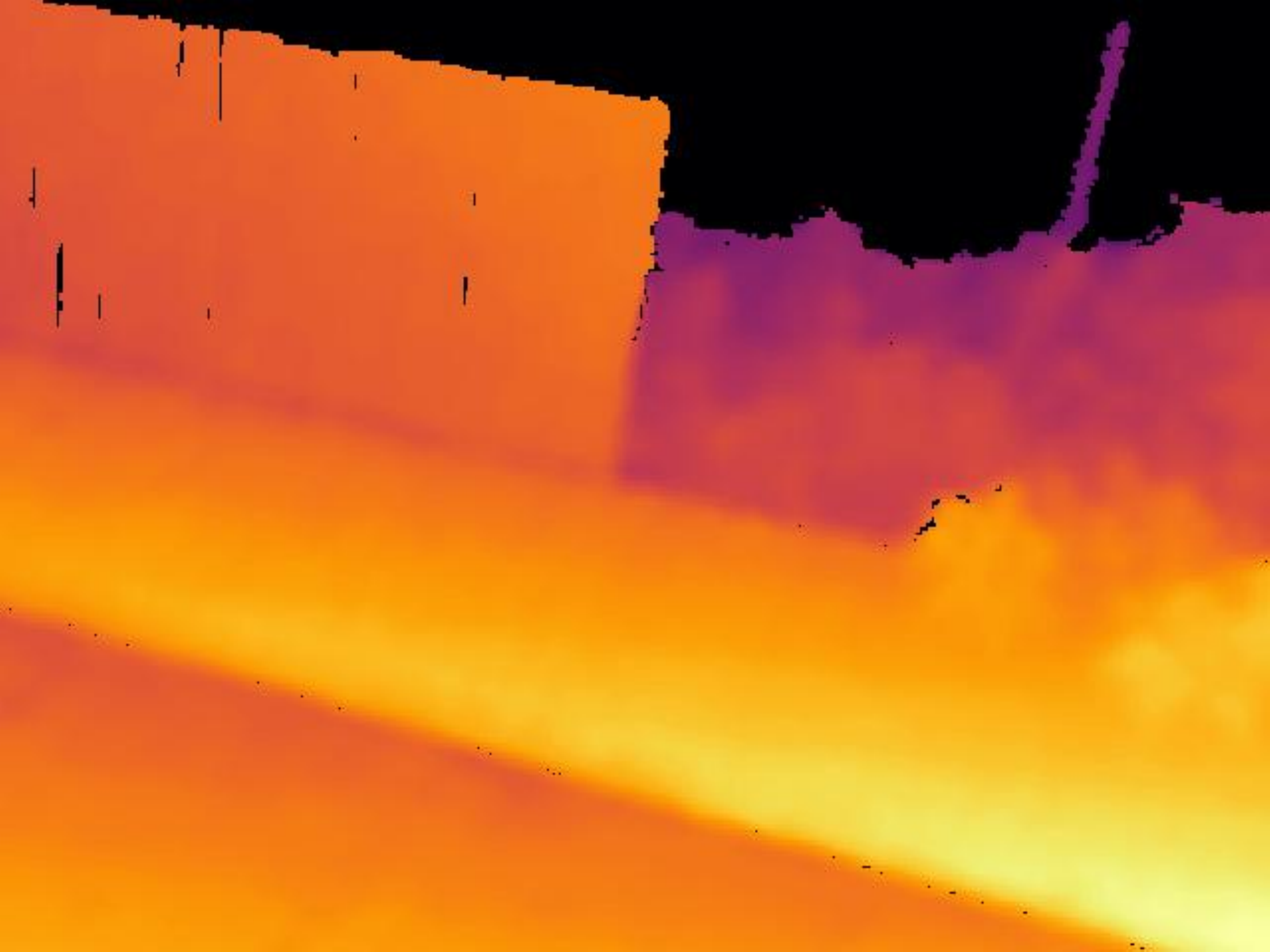}&
    \includegraphics[width=0.096\linewidth]{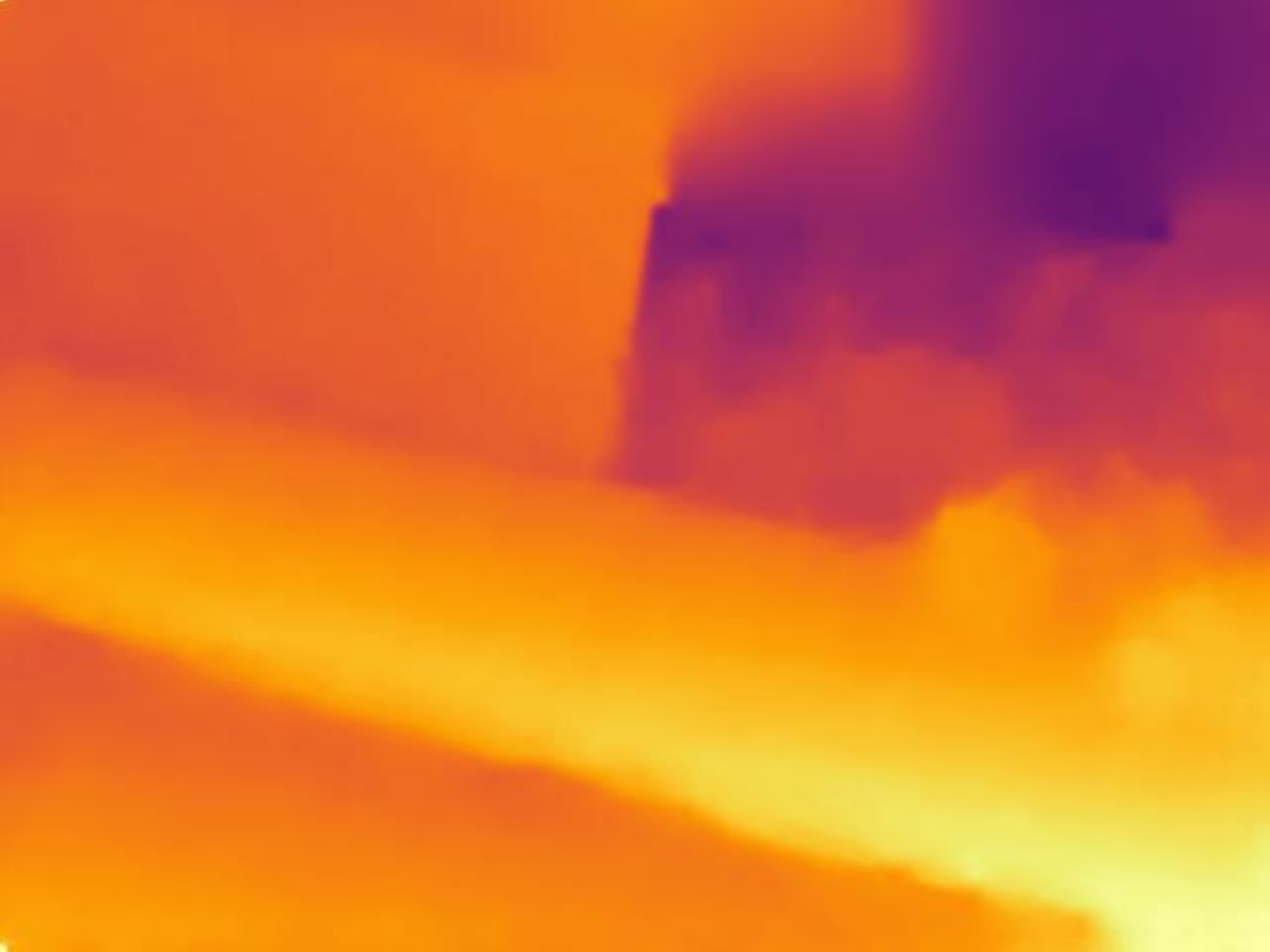}&
    \includegraphics[width=0.096\linewidth]{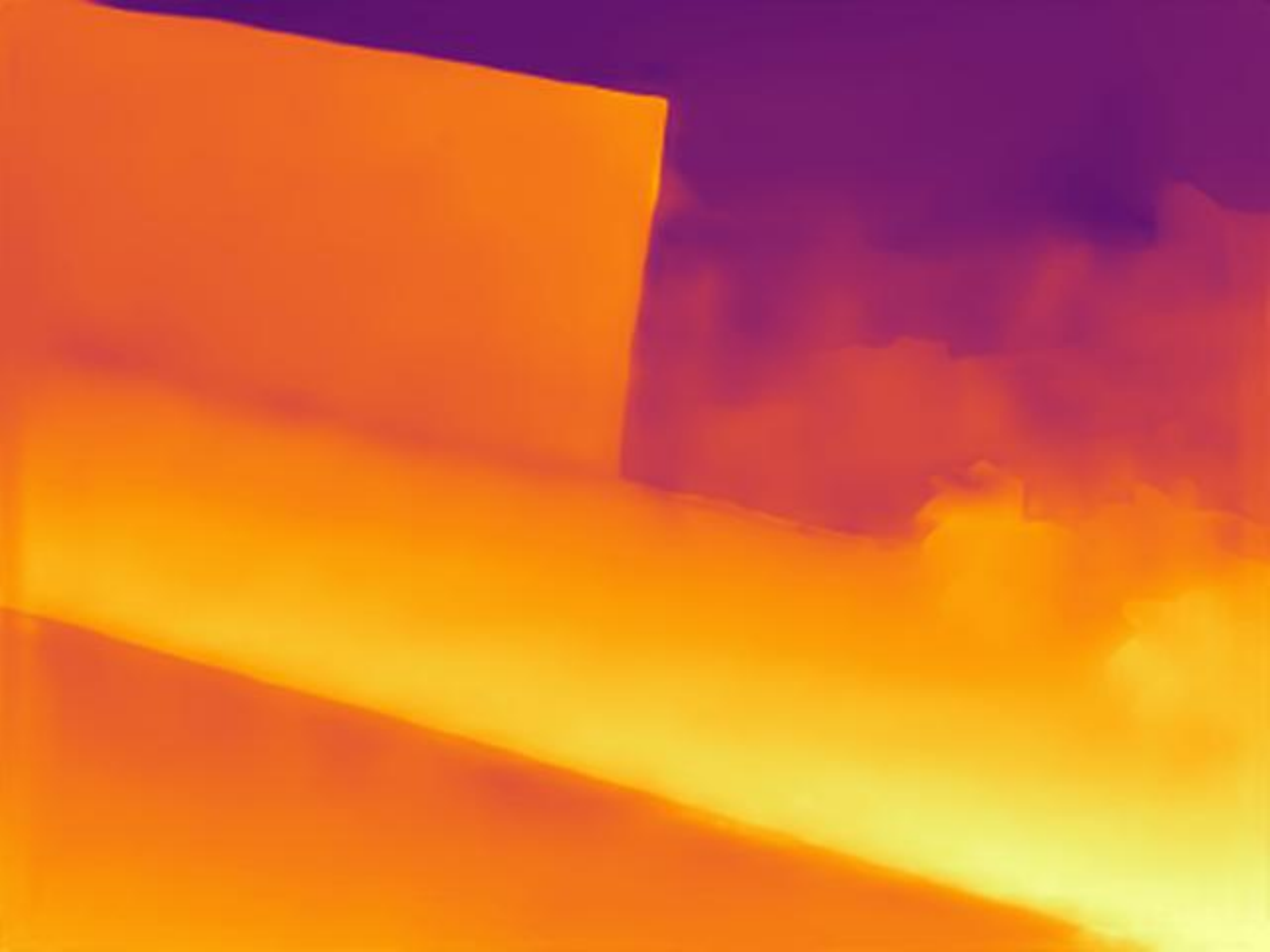}&
    \includegraphics[width=0.096\linewidth]{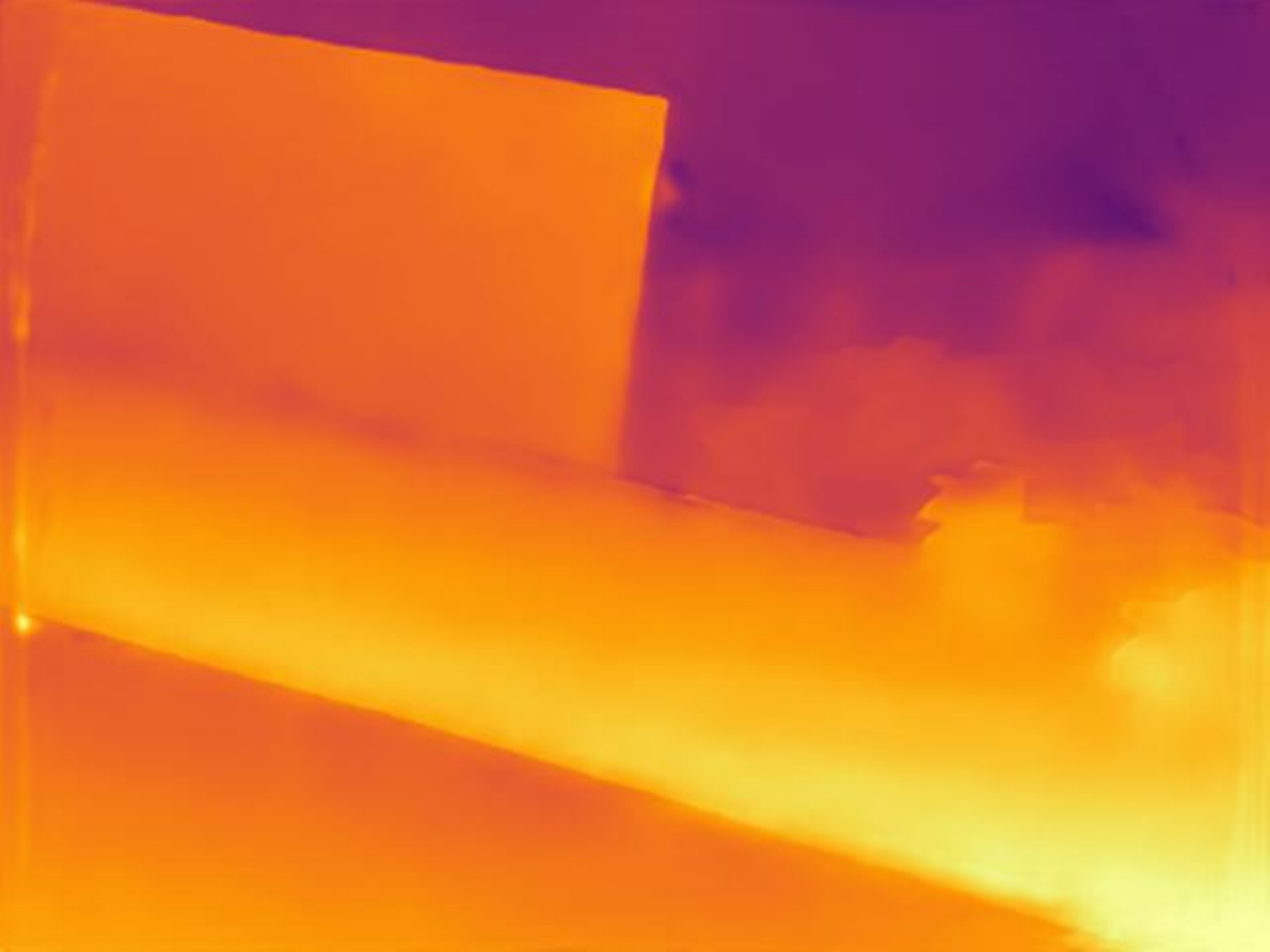}&
    \includegraphics[width=0.096\linewidth]{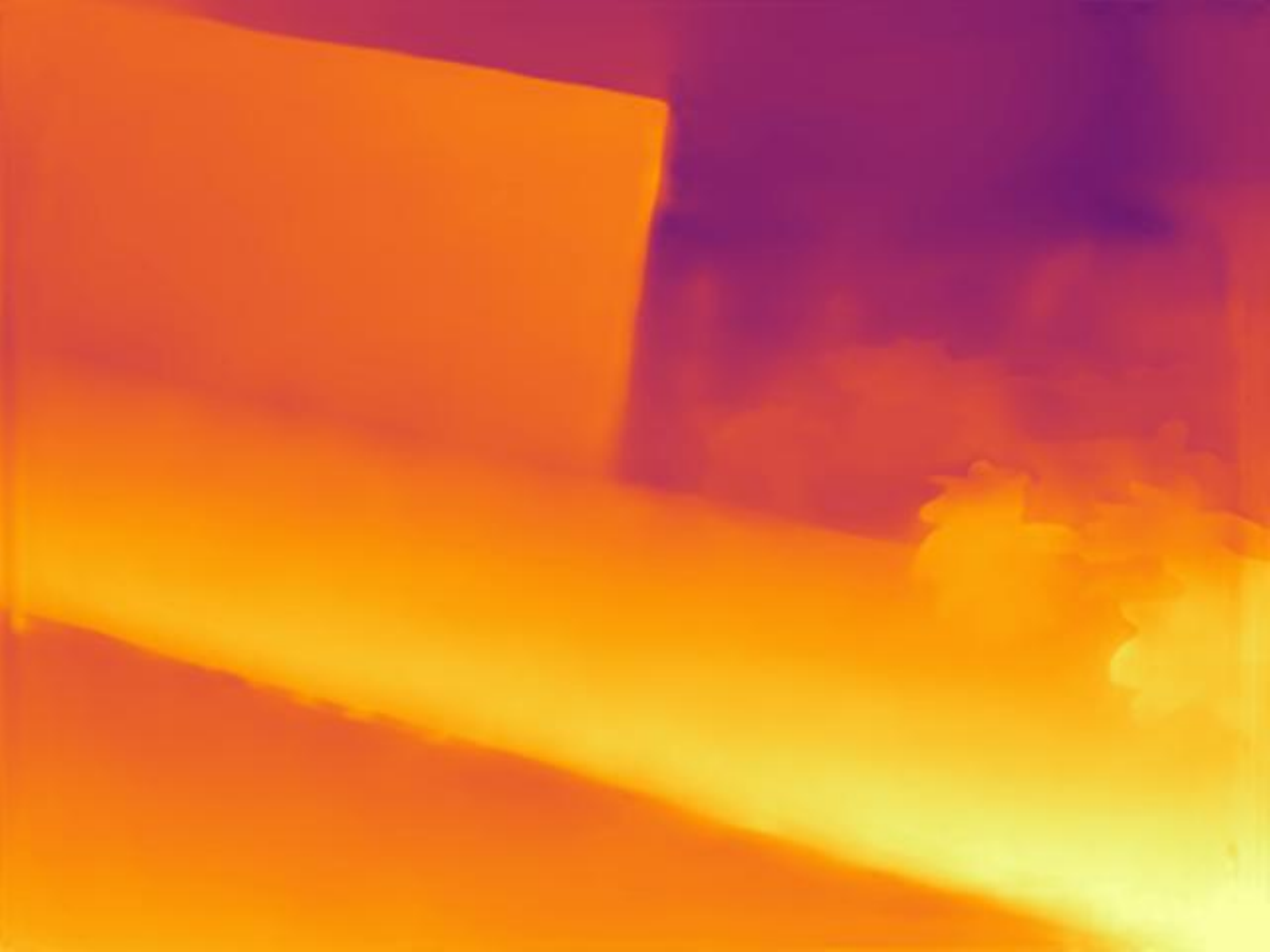}&
    \includegraphics[width=0.096\linewidth]{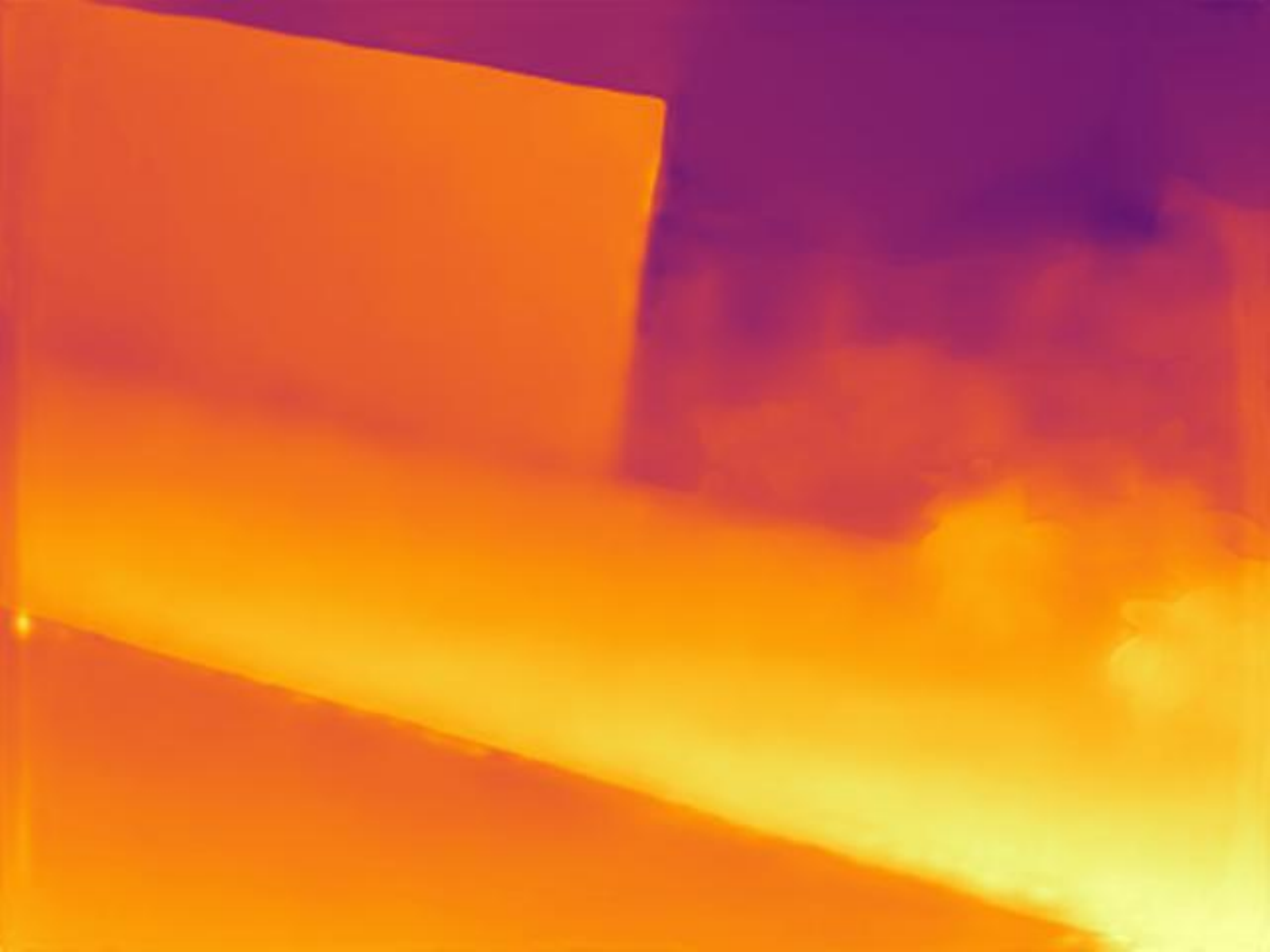}&
    \includegraphics[width=0.096\linewidth]{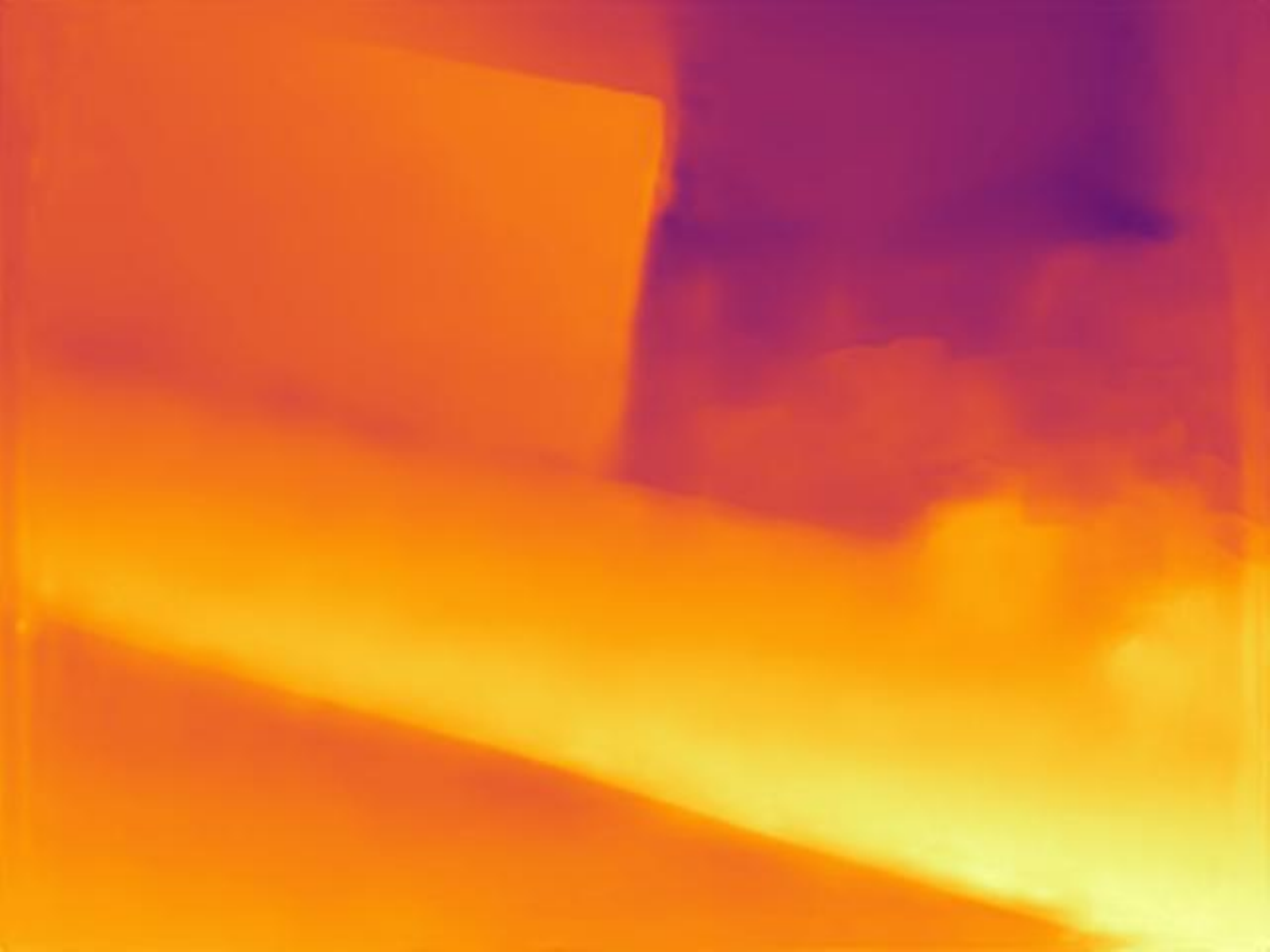}&
    \includegraphics[width=0.096\linewidth]{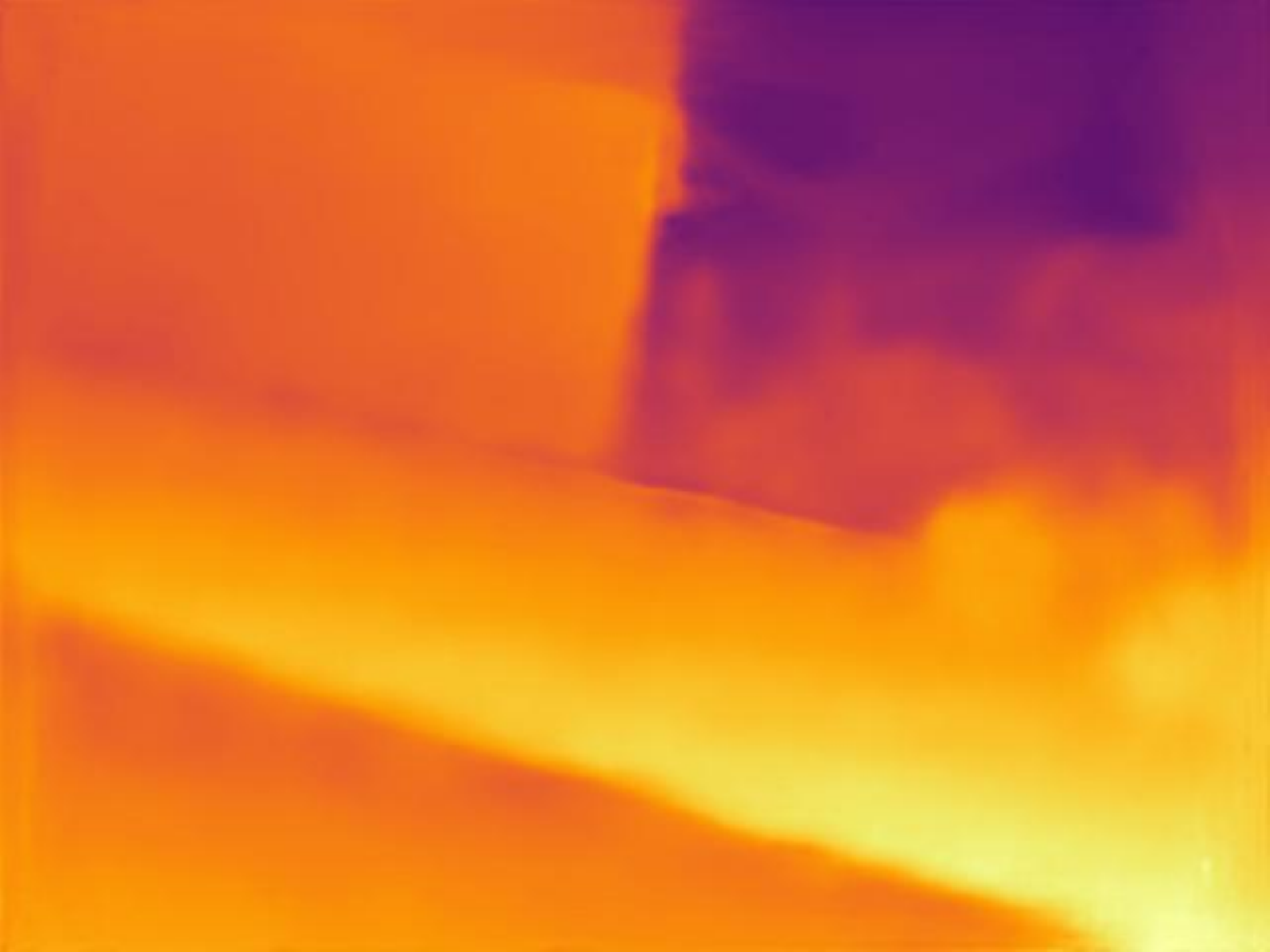}&
    \includegraphics[width=0.096\linewidth]{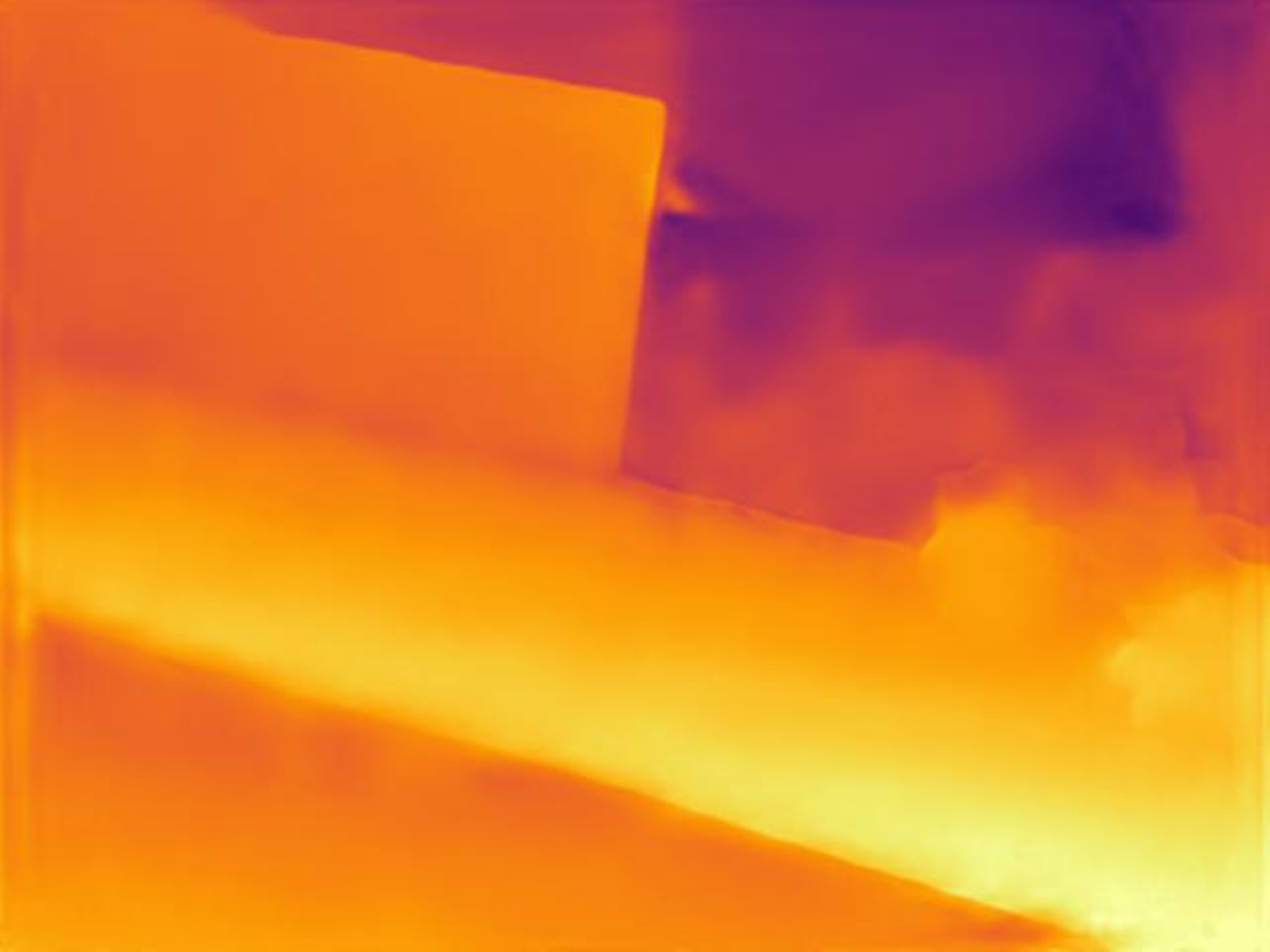}\\
    \multicolumn{9}{c}{} \\
    \vspace{-0.75mm}
    \rot{\scriptsize VOID 150} &
    \includegraphics[width=0.096\linewidth]{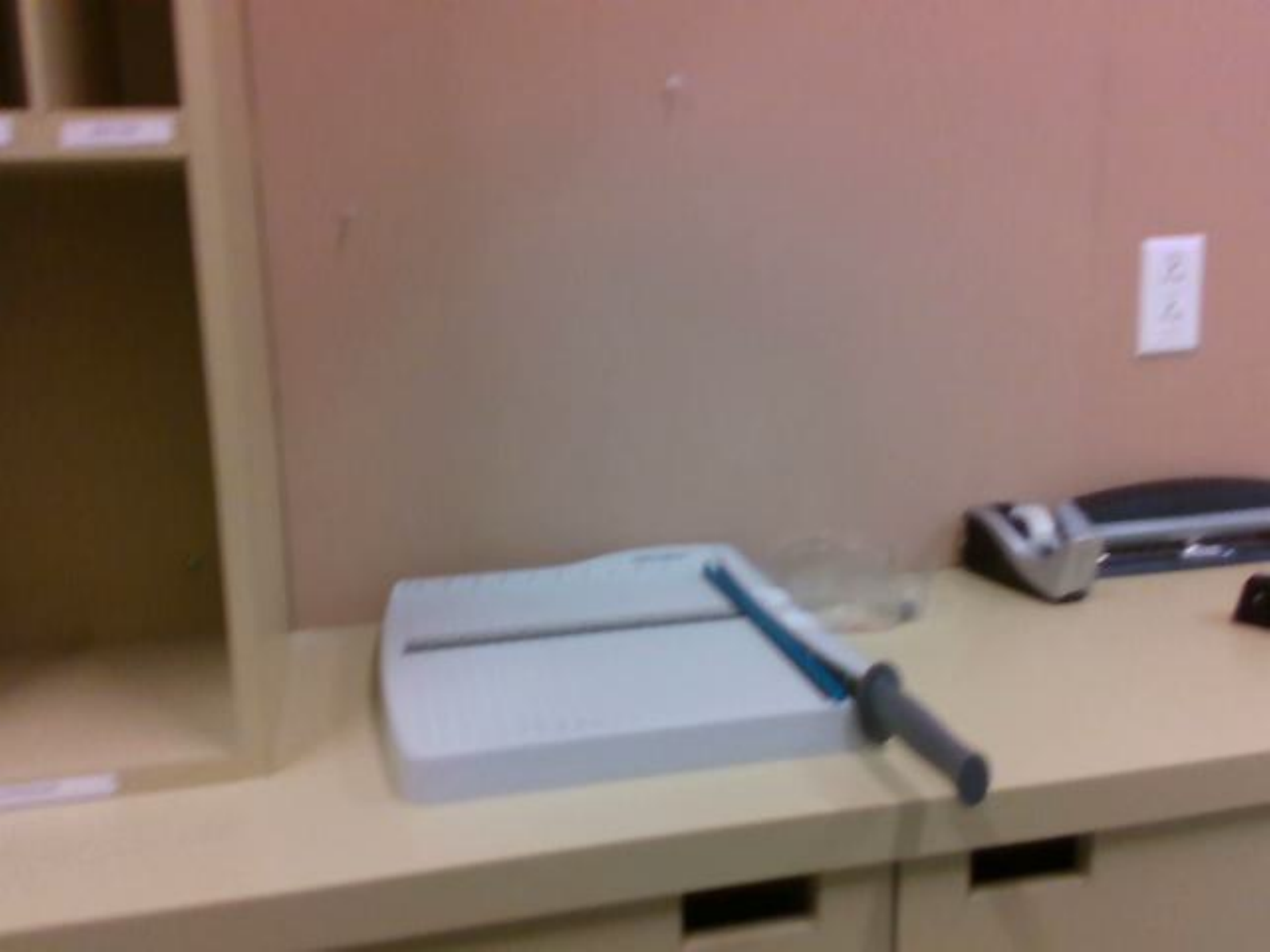}&
    \includegraphics[width=0.096\linewidth]{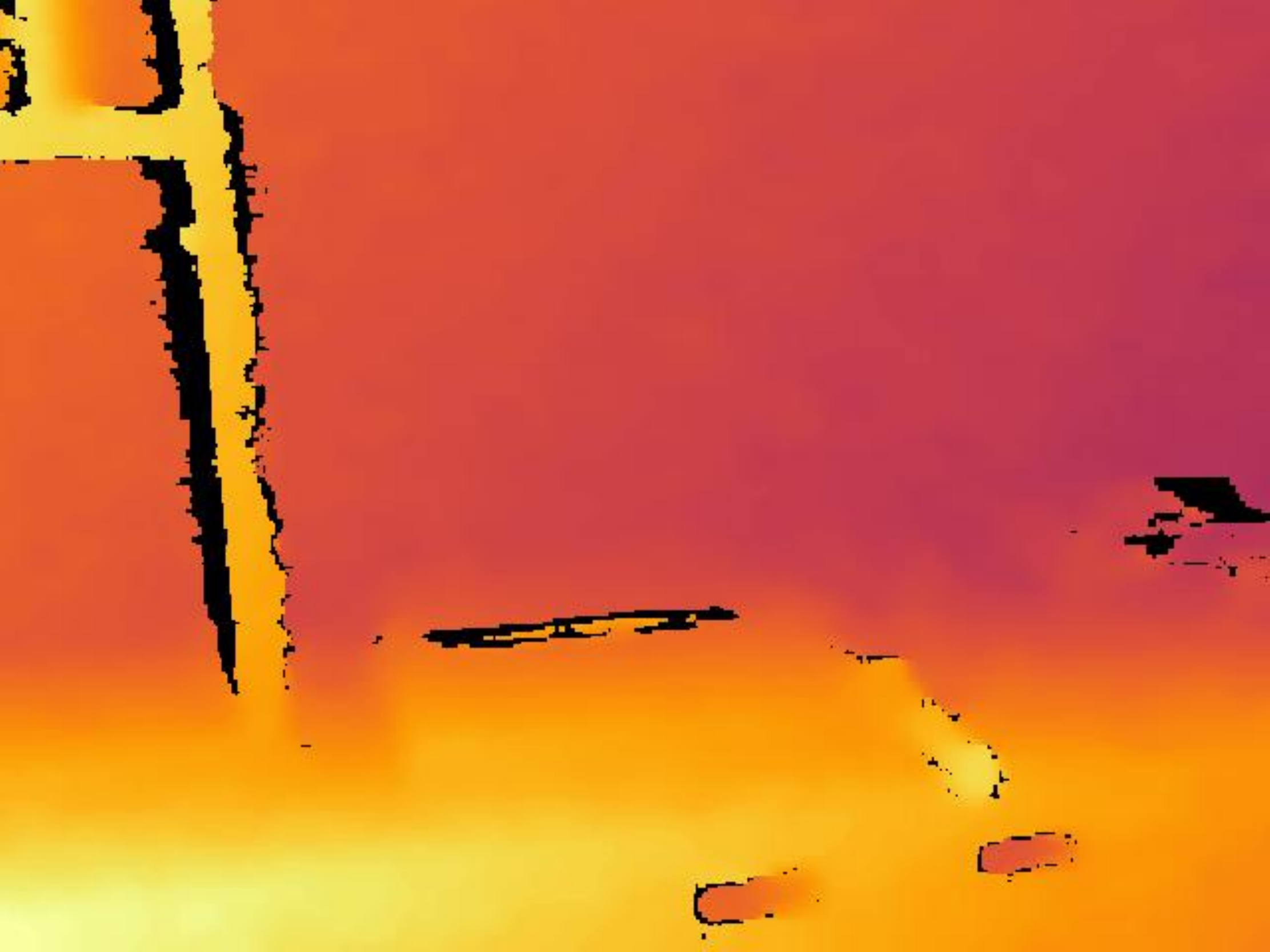}&
    \includegraphics[width=0.096\linewidth]{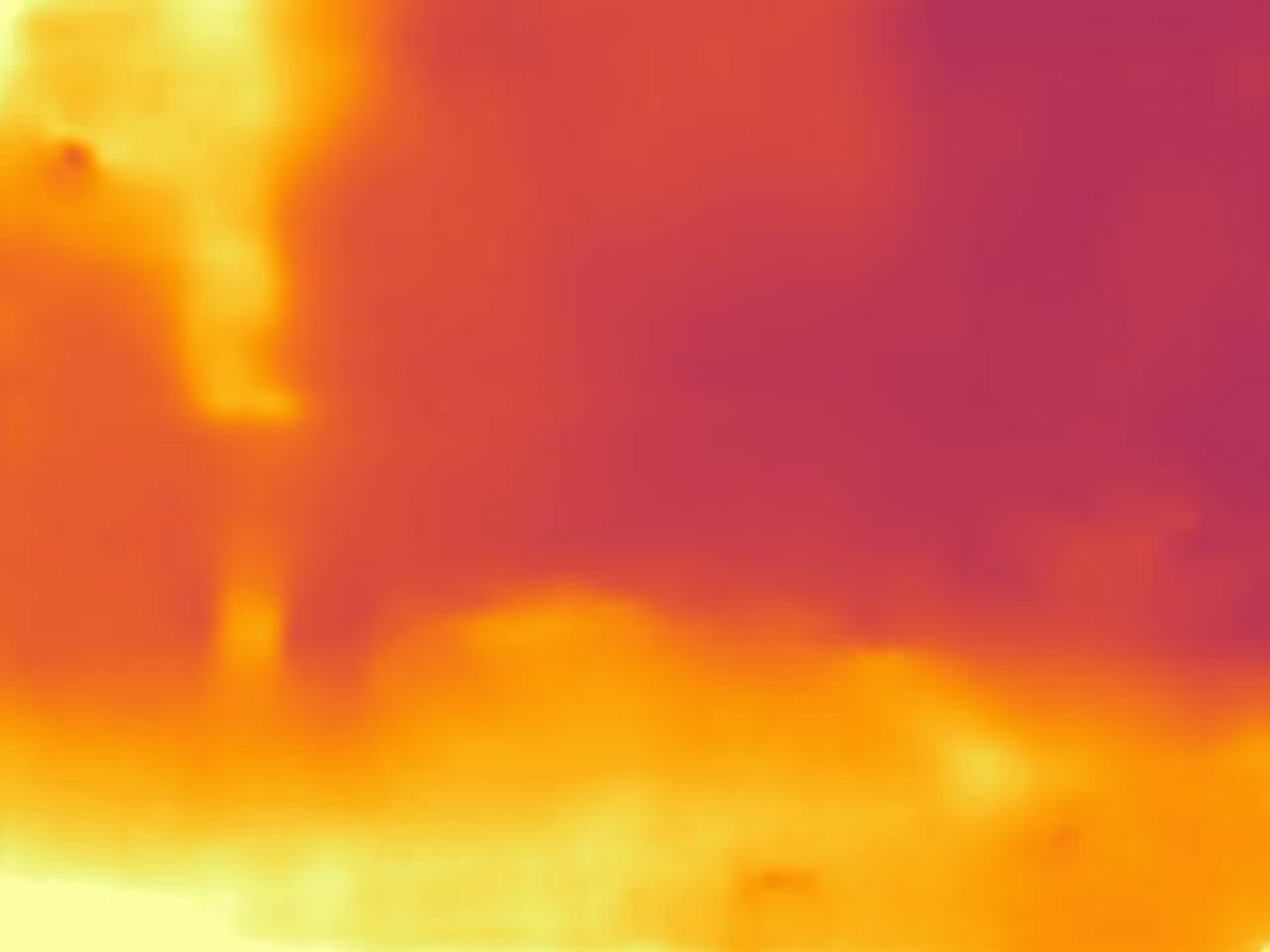}&
    \includegraphics[width=0.096\linewidth]{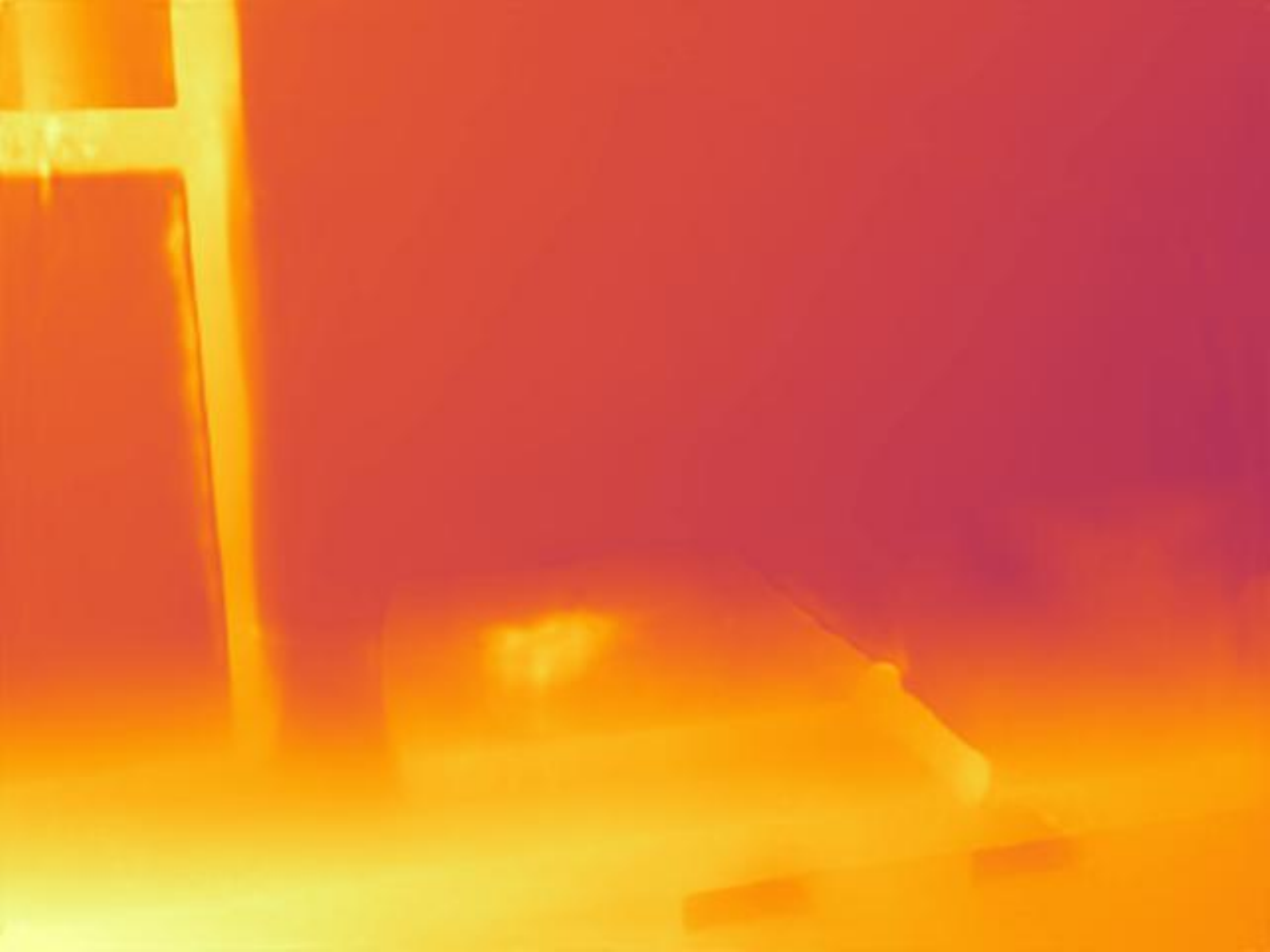}&
    \includegraphics[width=0.096\linewidth]{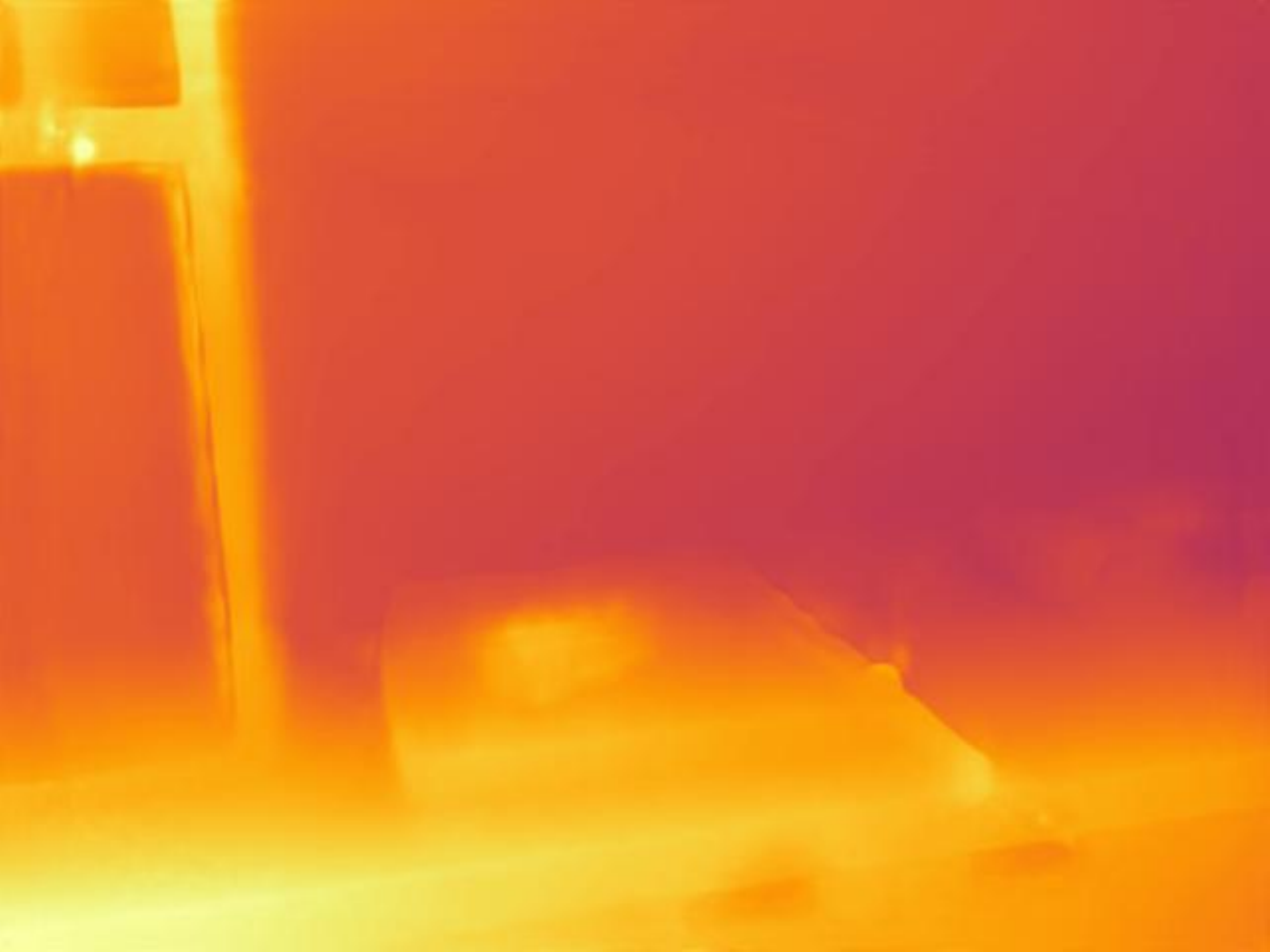}&
    \includegraphics[width=0.096\linewidth]{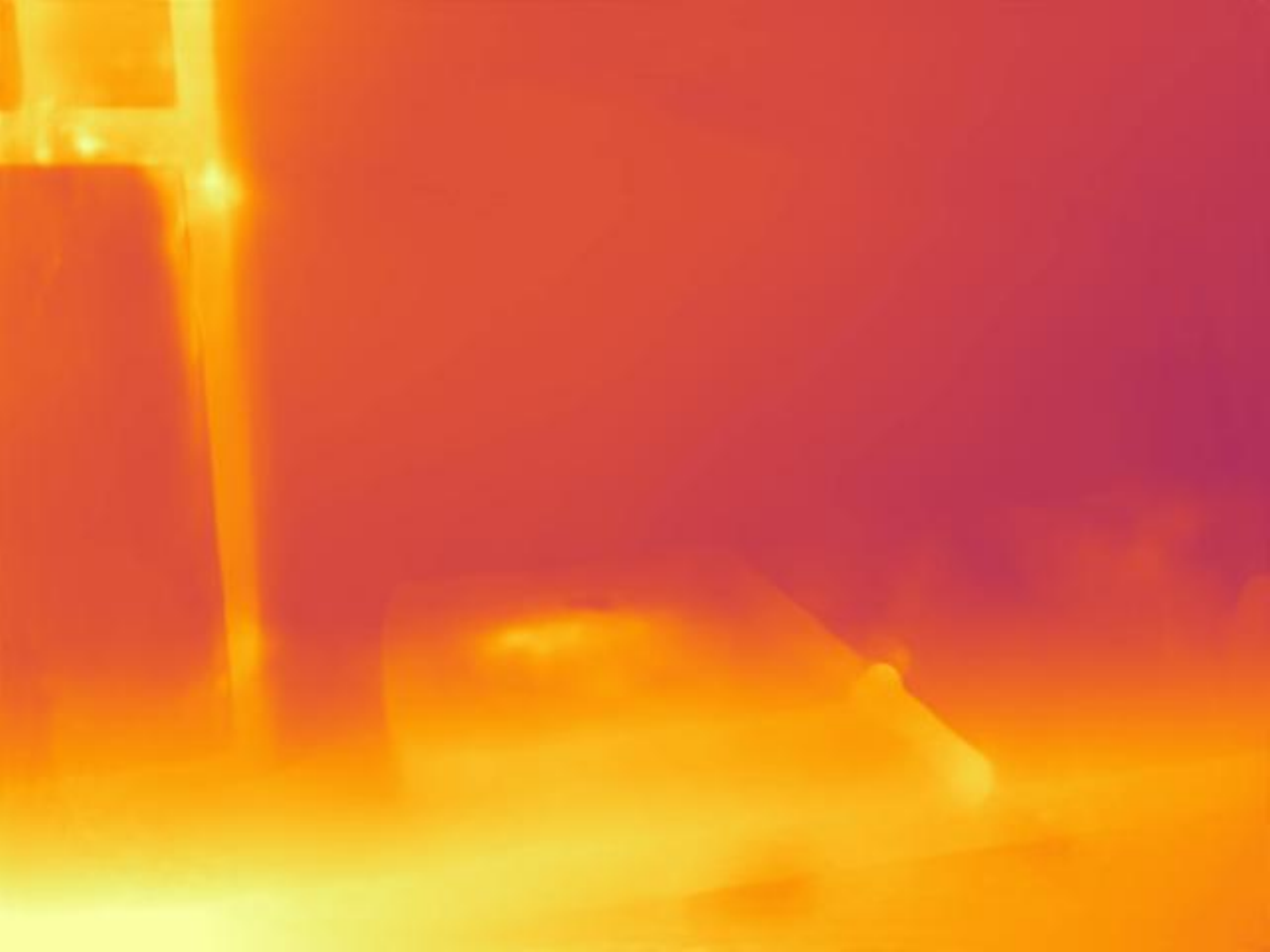}&
    \includegraphics[width=0.096\linewidth]{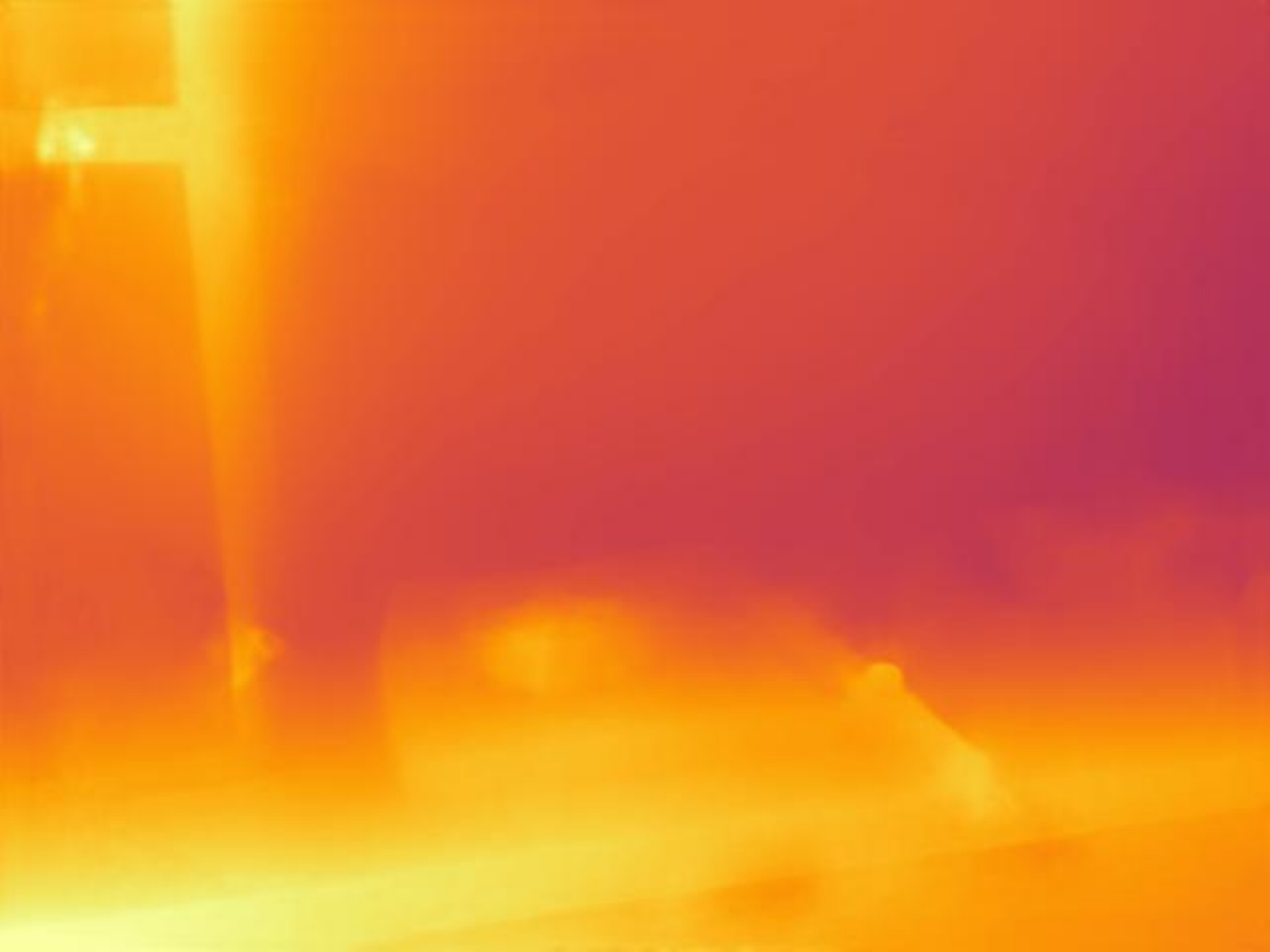}&
    \includegraphics[width=0.096\linewidth]{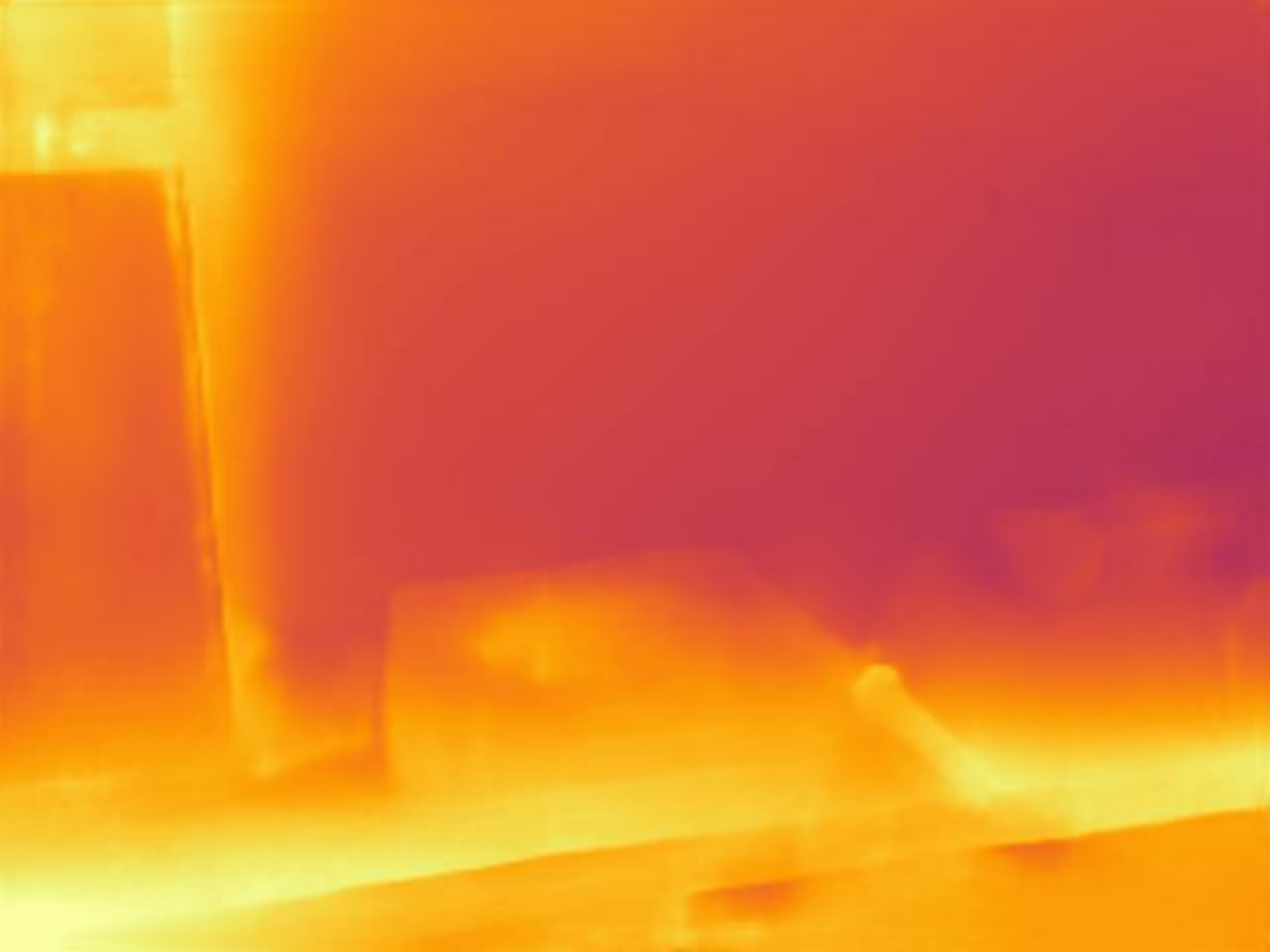}&
    \includegraphics[width=0.096\linewidth]{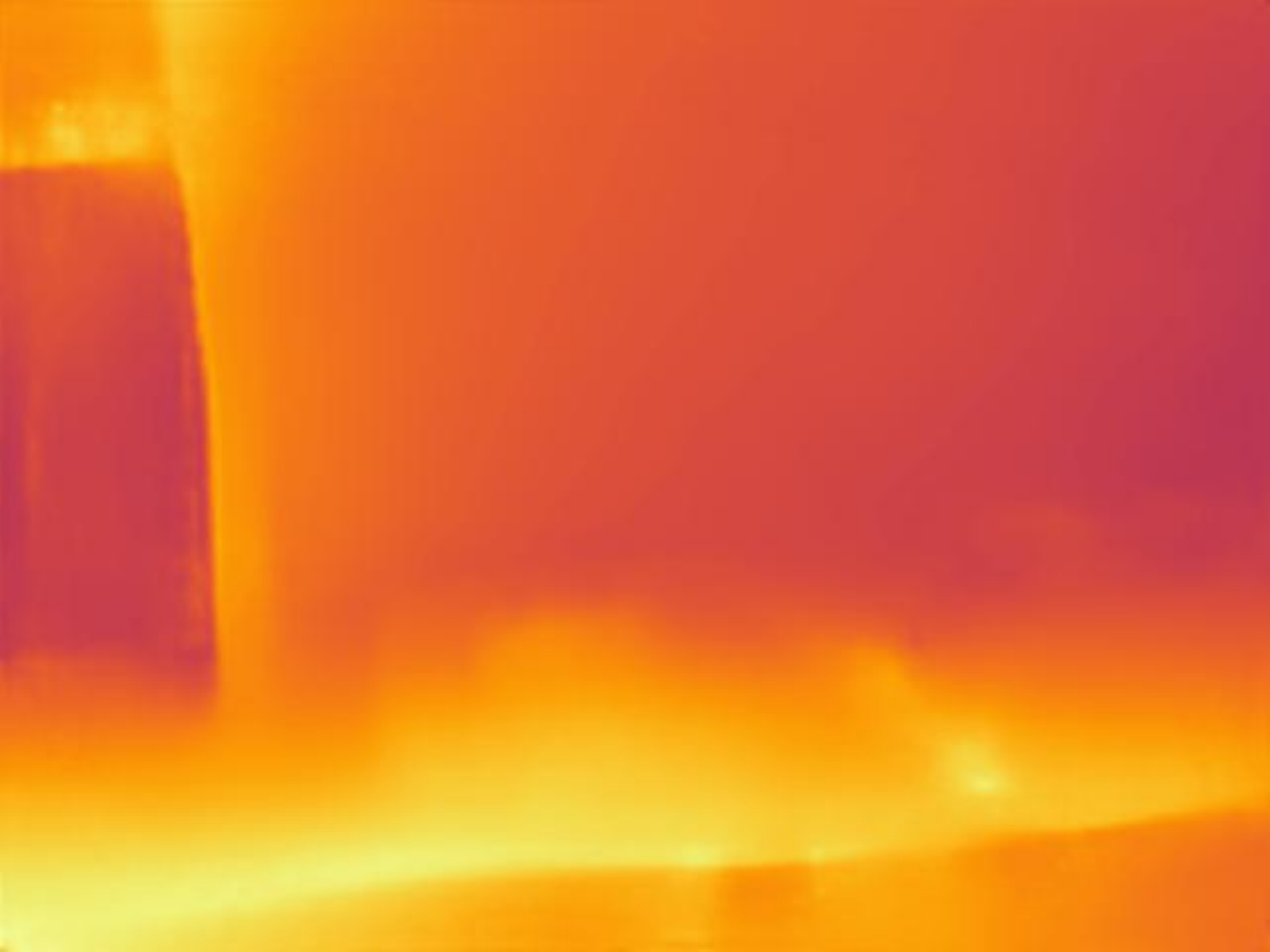}&
    \includegraphics[width=0.096\linewidth]{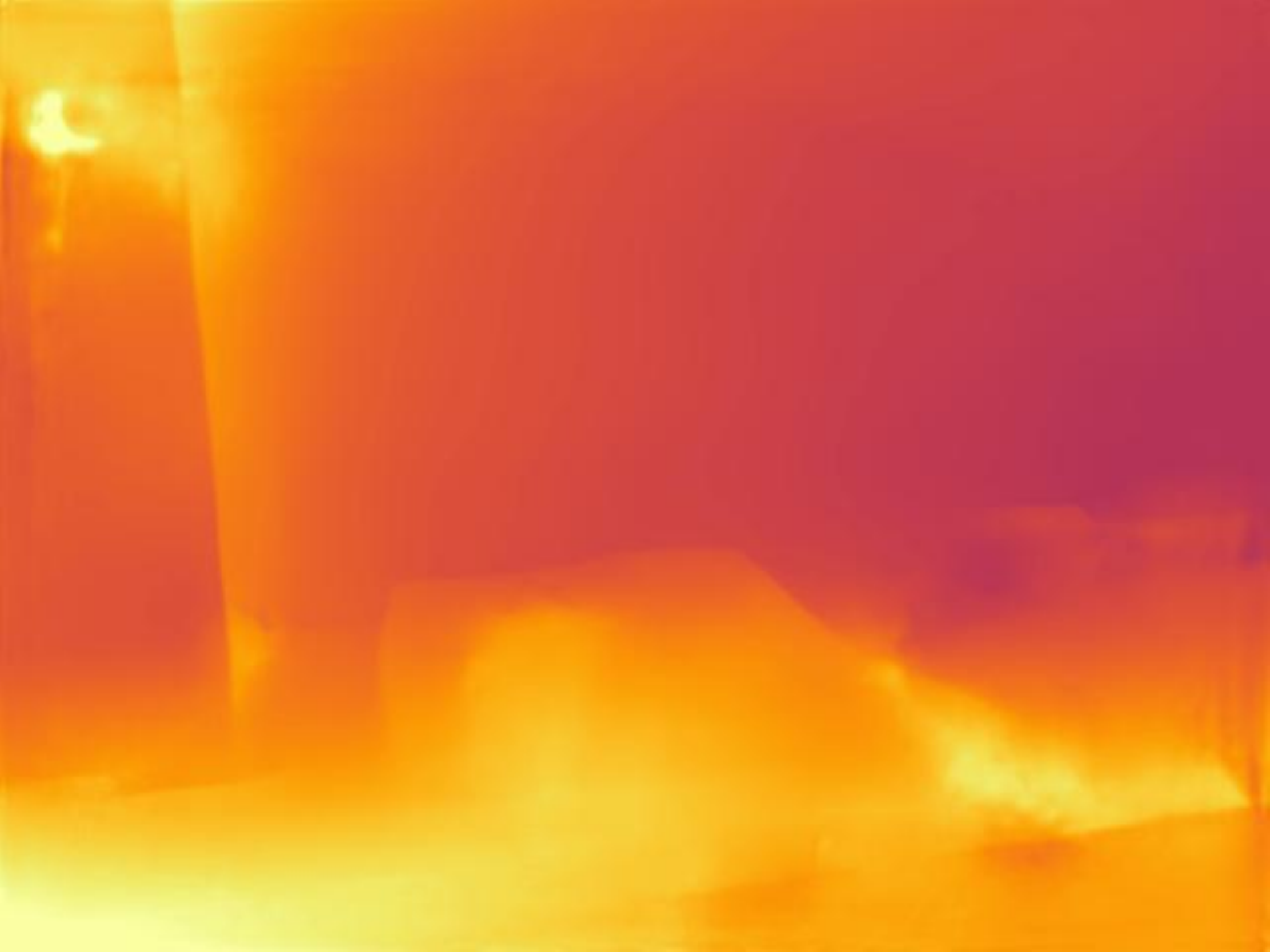}\\
    \vspace{-0.75mm}
    \rot{\scriptsize VOID 500} &
    \includegraphics[width=0.096\linewidth]{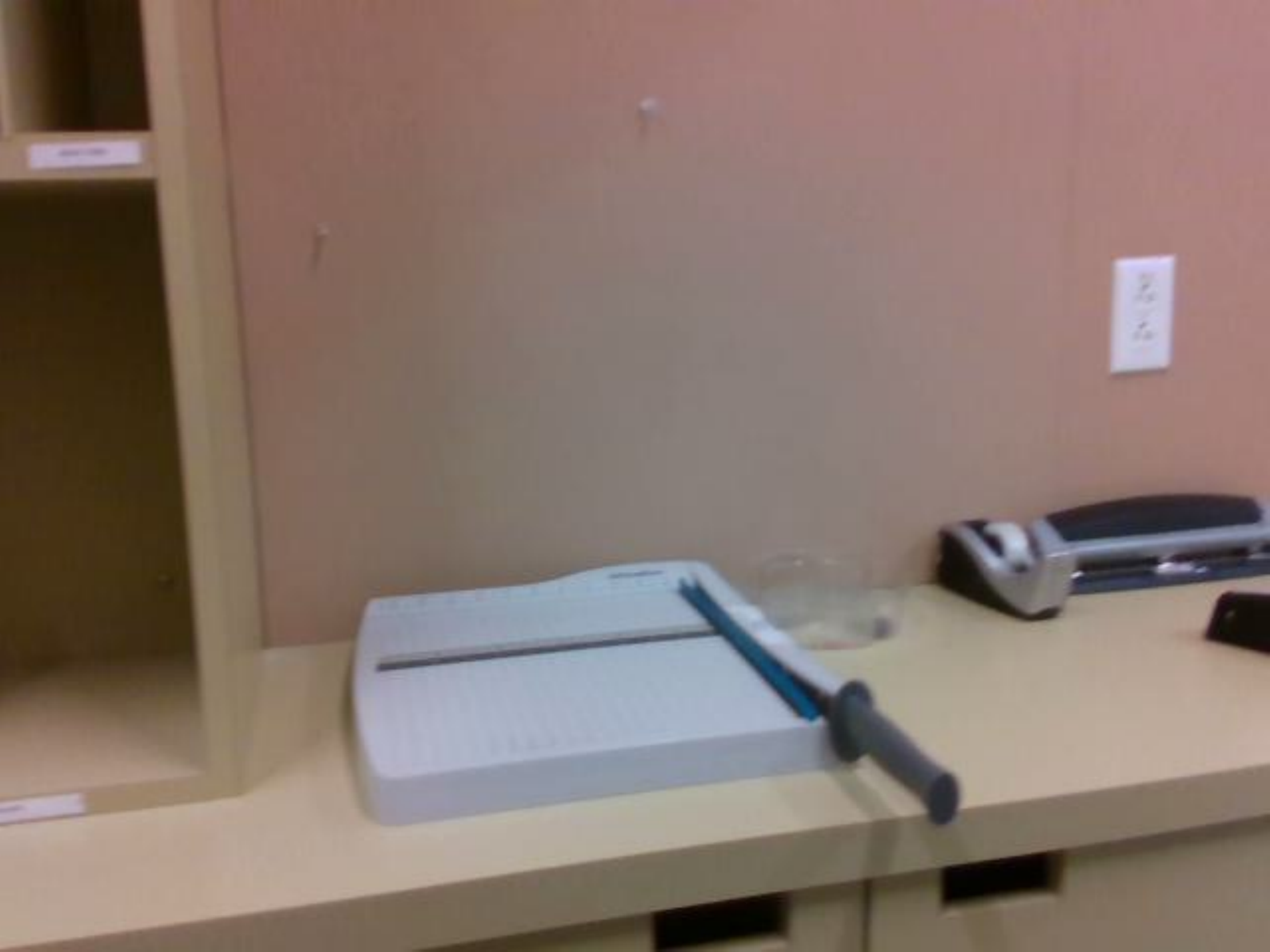}&
    \includegraphics[width=0.096\linewidth]{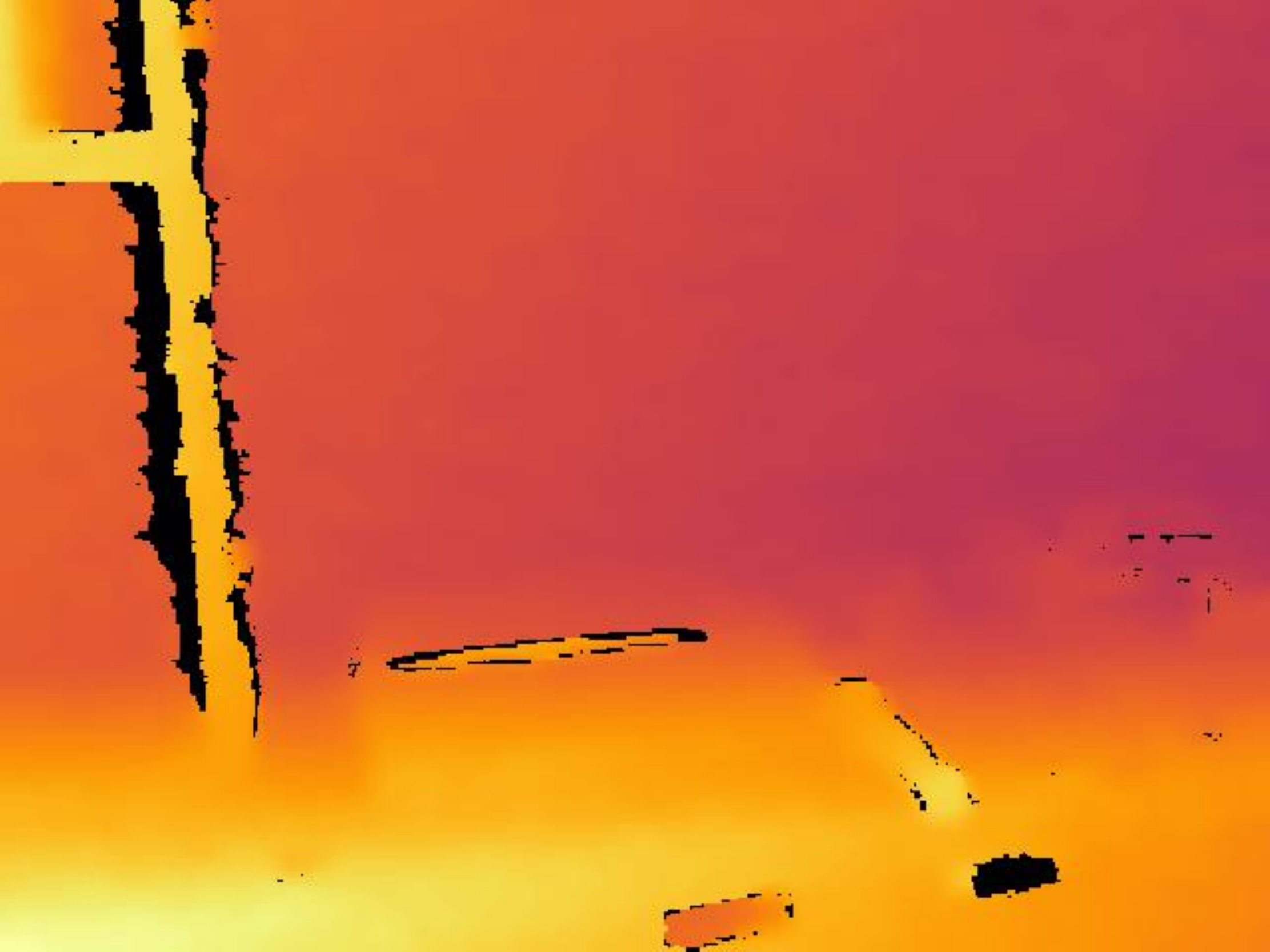}&
    \includegraphics[width=0.096\linewidth]{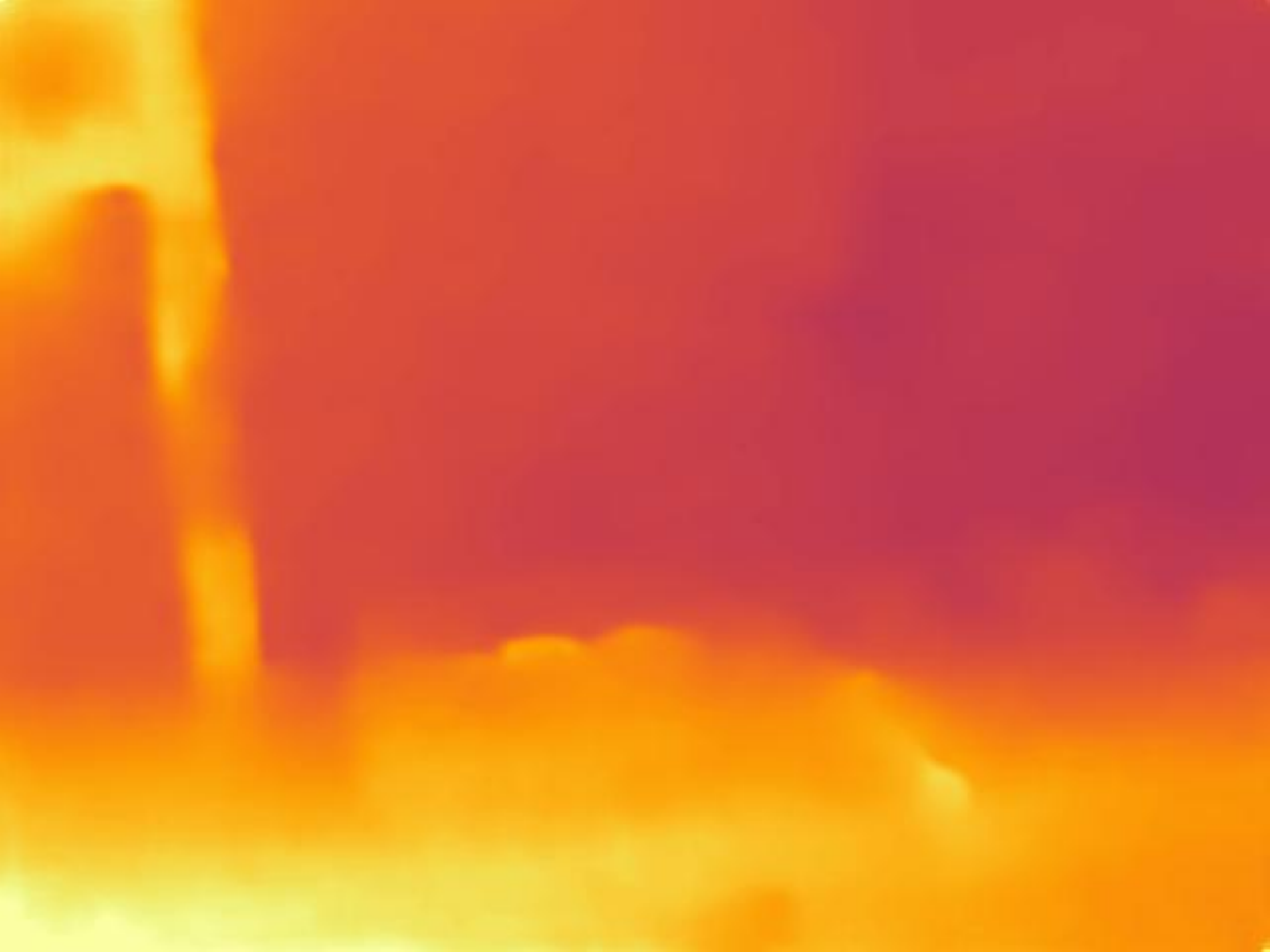}&
    \includegraphics[width=0.096\linewidth]{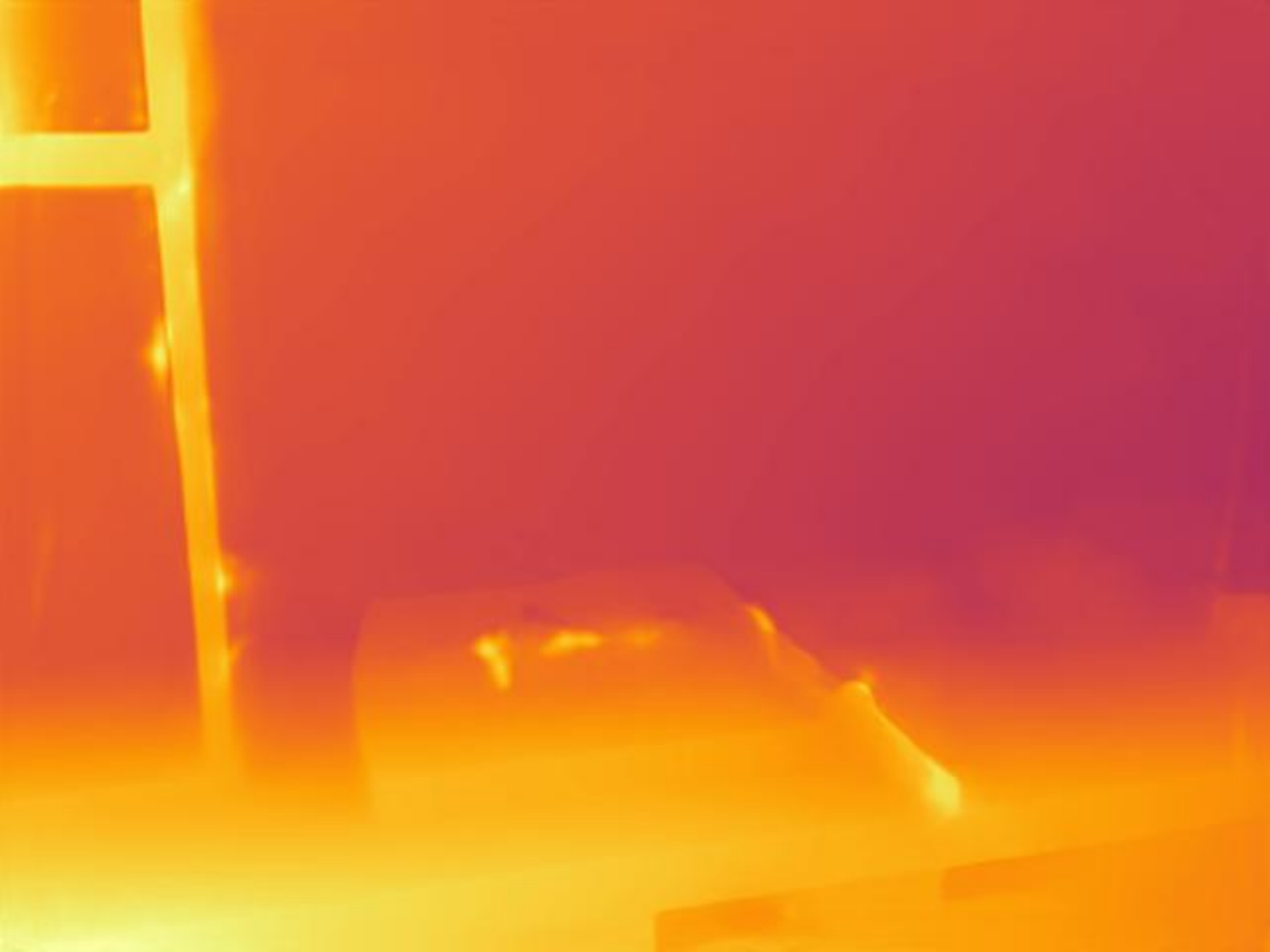}&
    \includegraphics[width=0.096\linewidth]{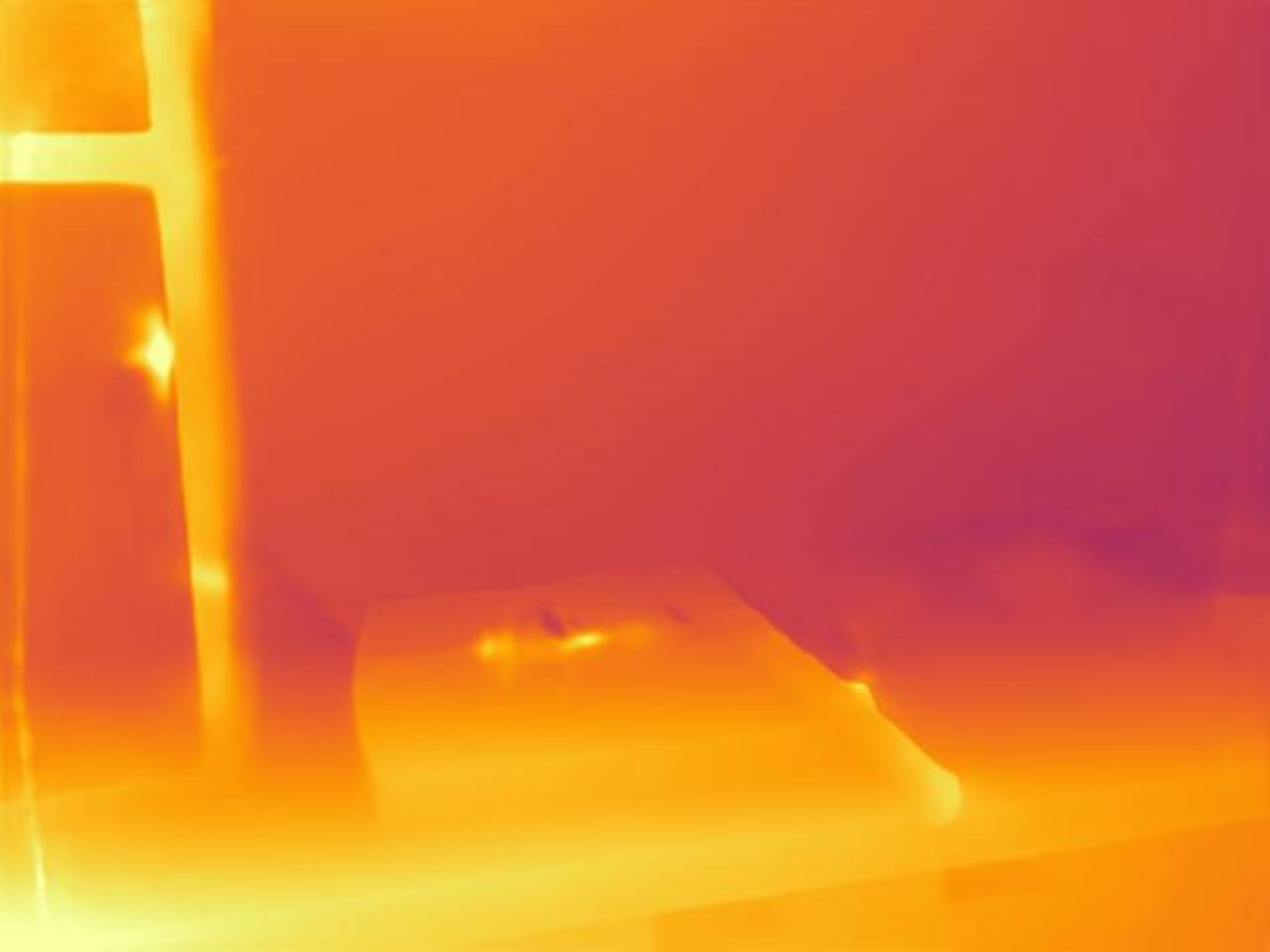}&
    \includegraphics[width=0.096\linewidth]{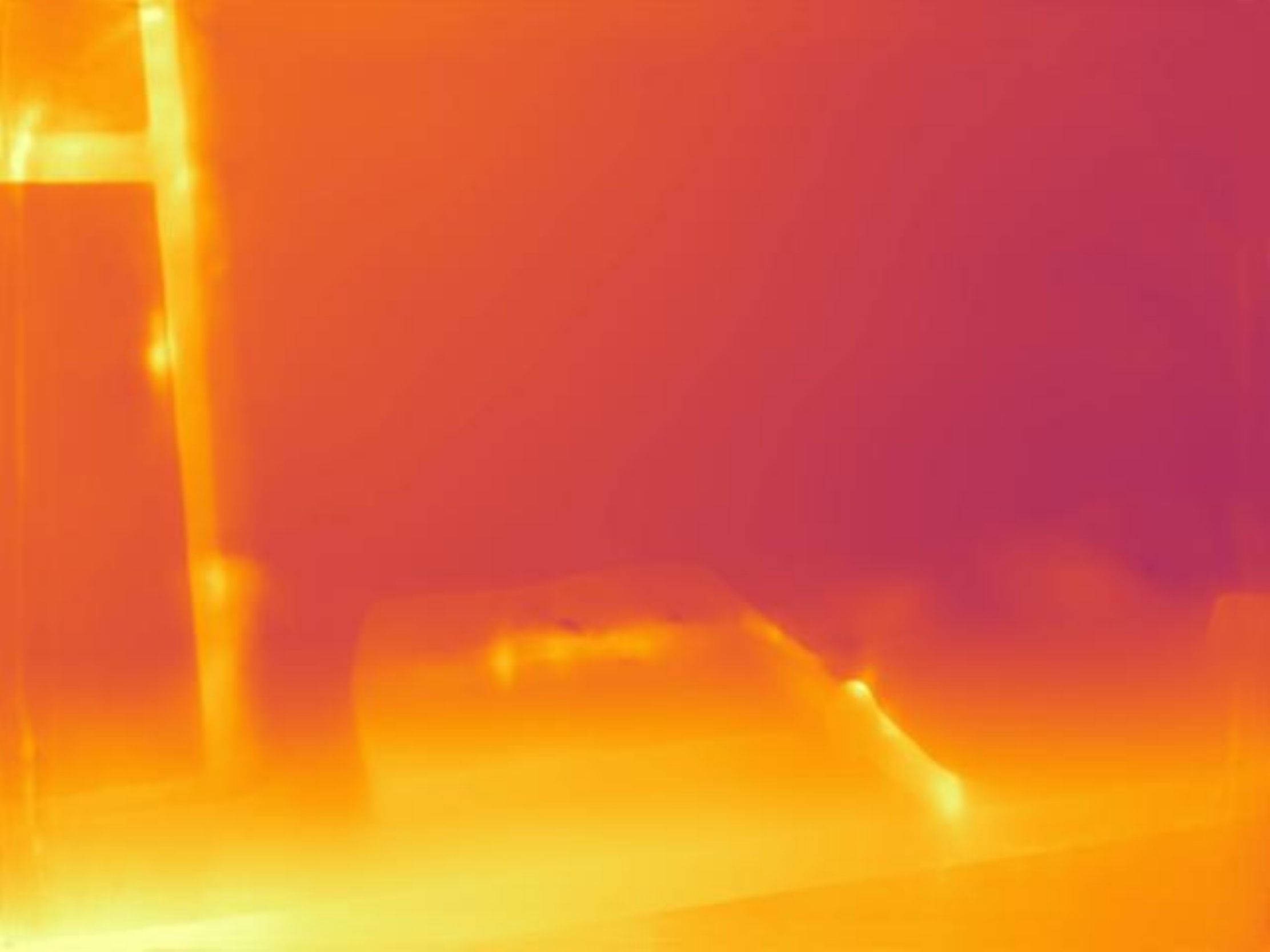}&
    \includegraphics[width=0.096\linewidth]{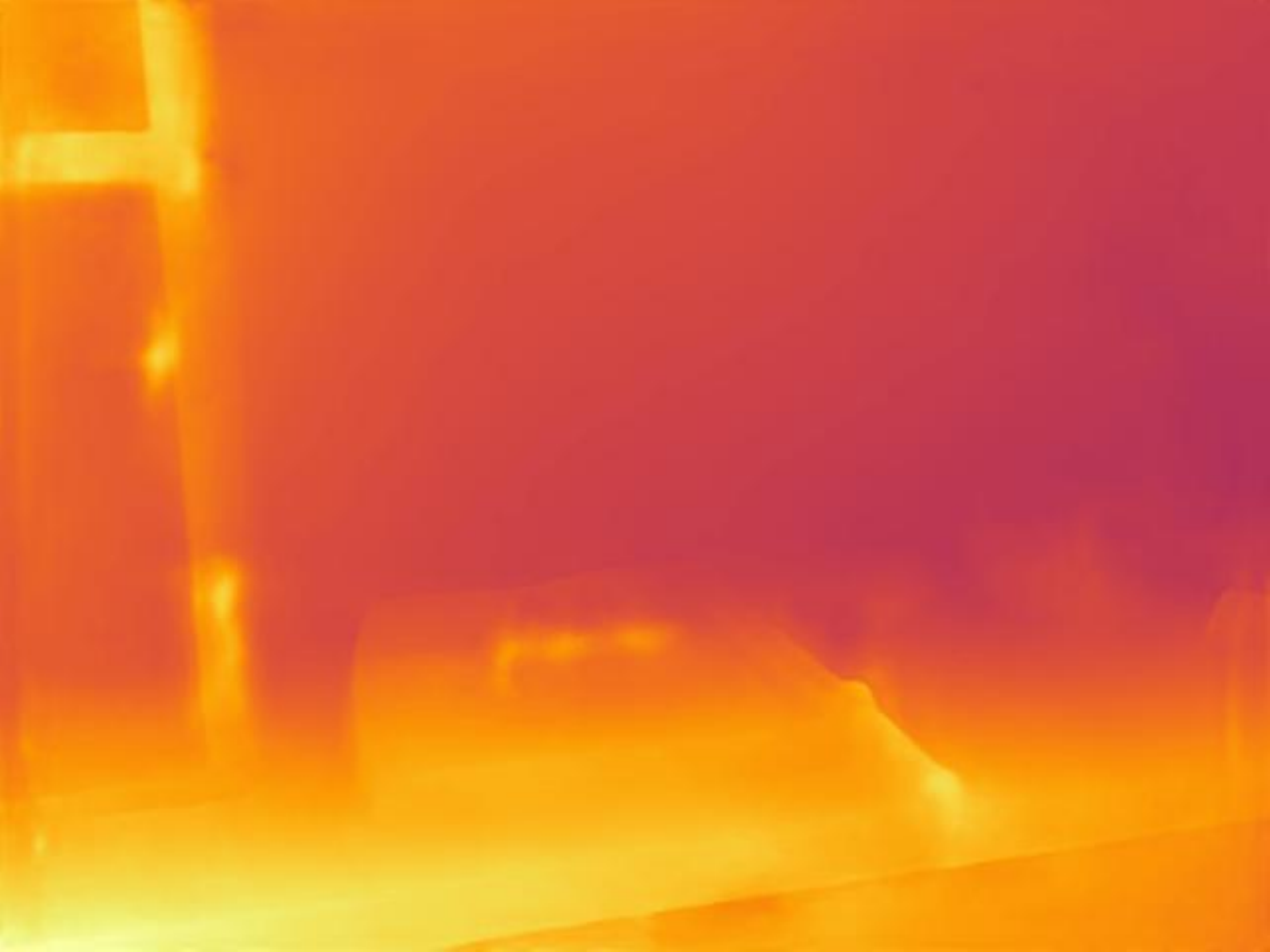}&
    \includegraphics[width=0.096\linewidth]{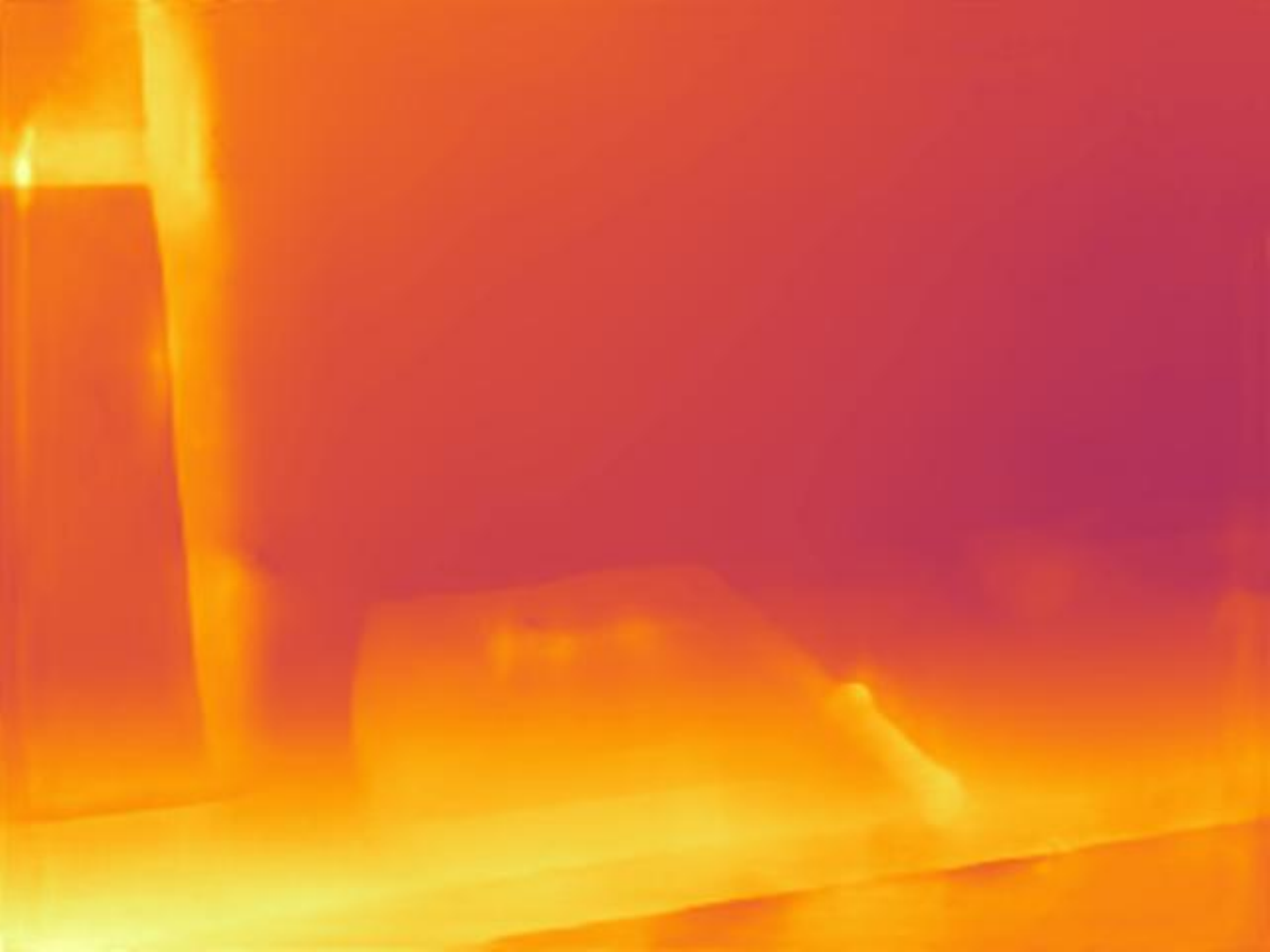}&
    \includegraphics[width=0.096\linewidth]{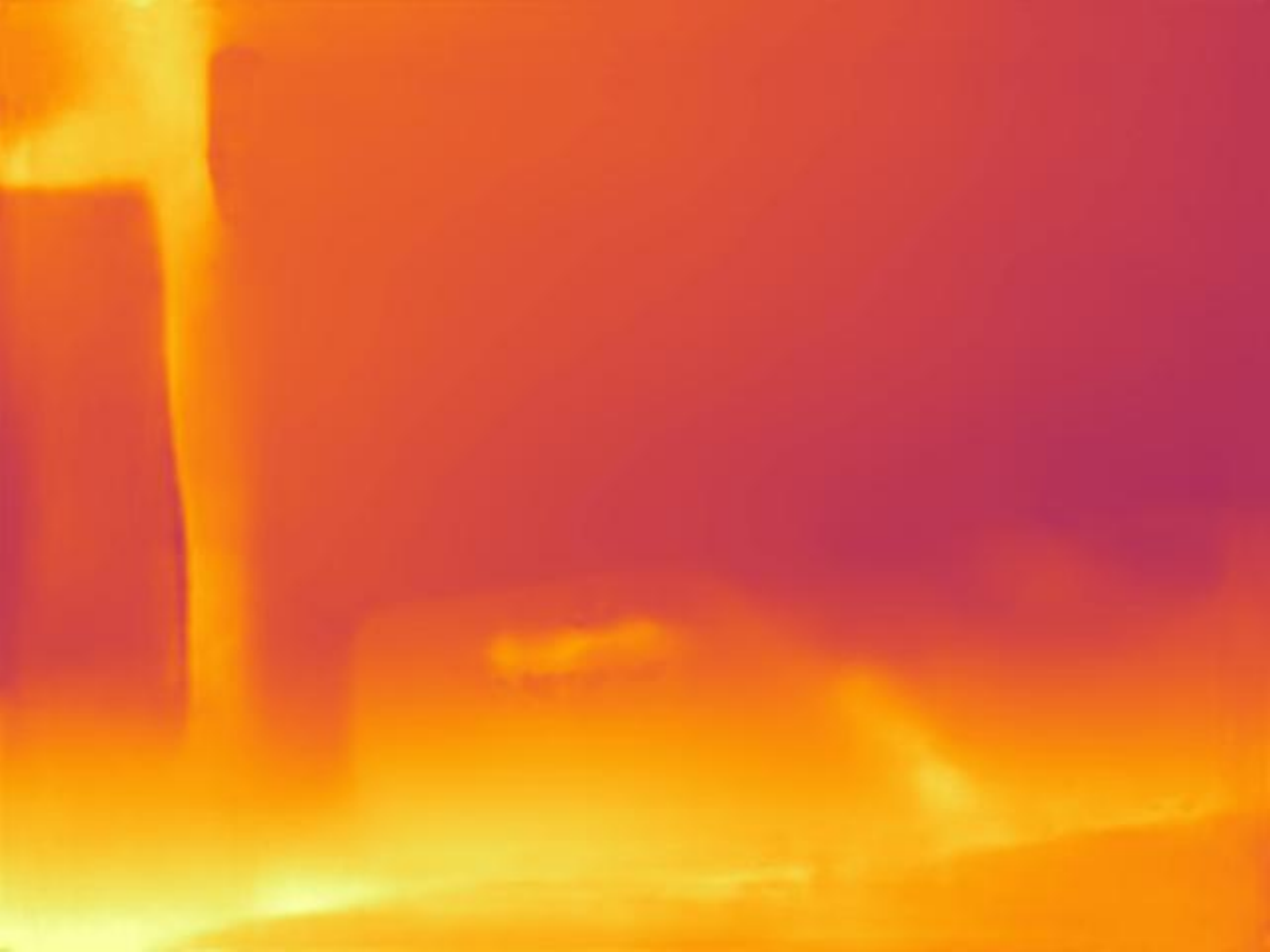}&
    \includegraphics[width=0.096\linewidth]{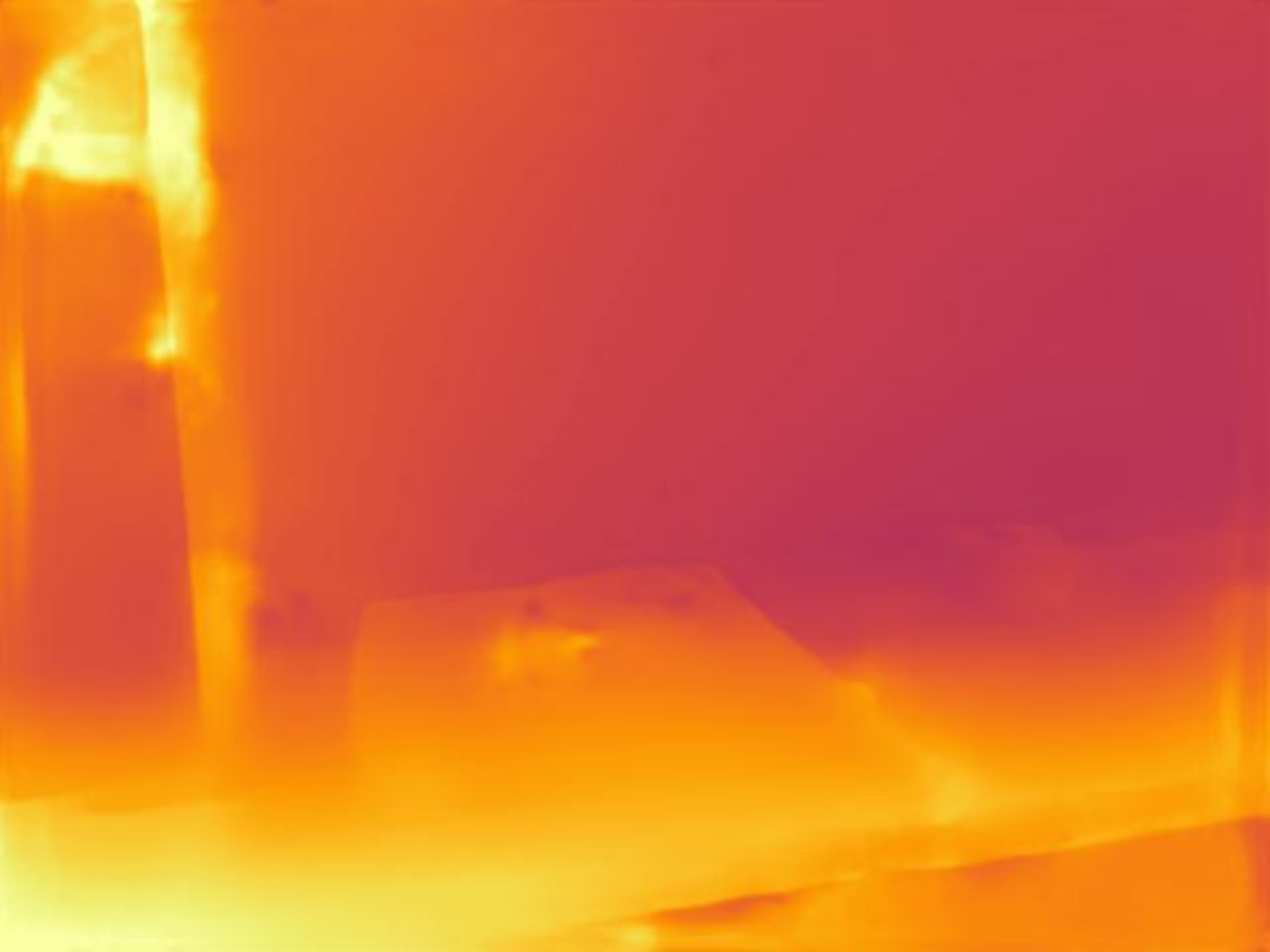}\\
    \vspace{-0.75mm}
    \rot{\scriptsize VOID 1500} &
    \includegraphics[width=0.096\linewidth]{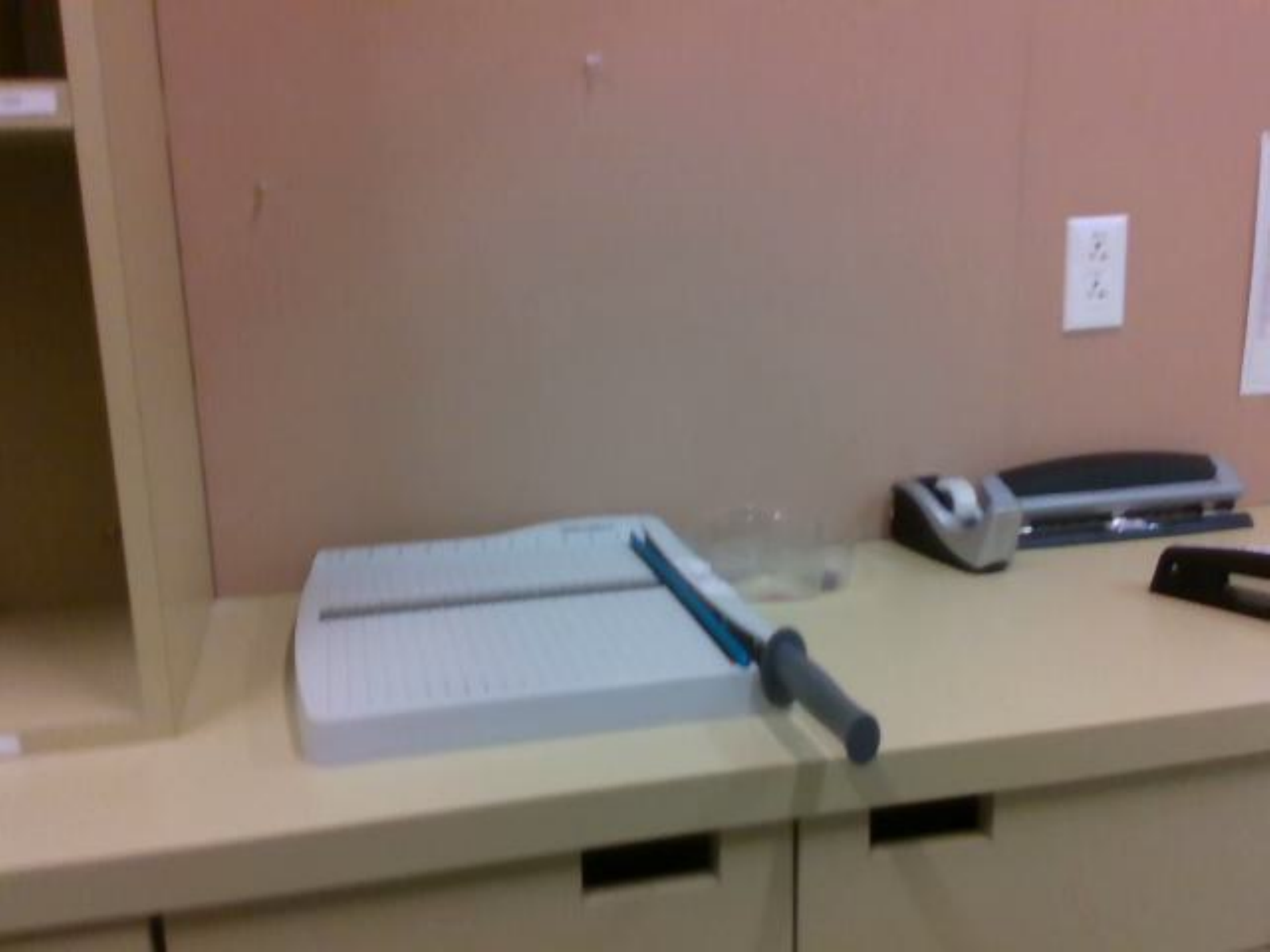}&
    \includegraphics[width=0.096\linewidth]{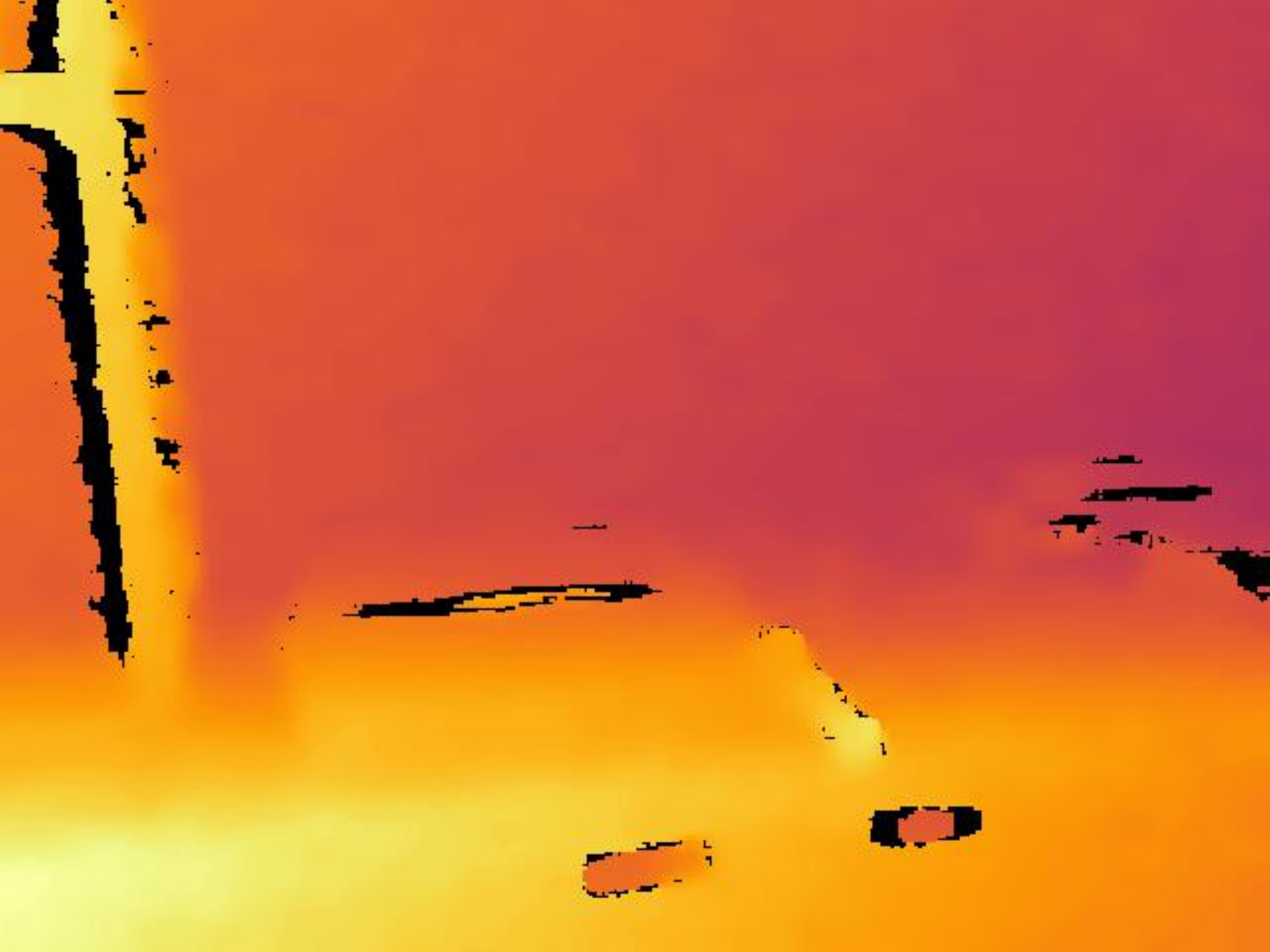}&
    \includegraphics[width=0.096\linewidth]{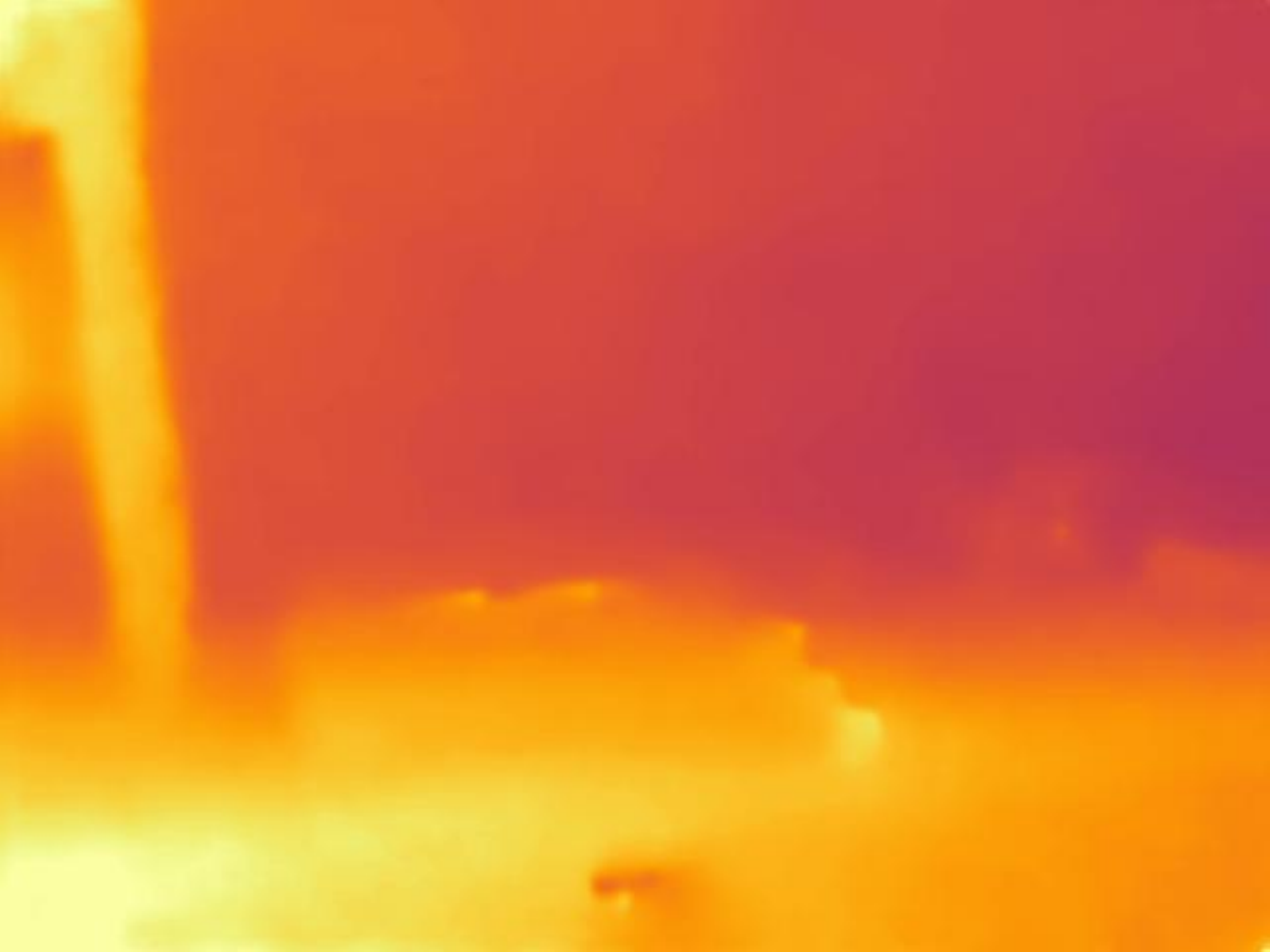}&
    \includegraphics[width=0.096\linewidth]{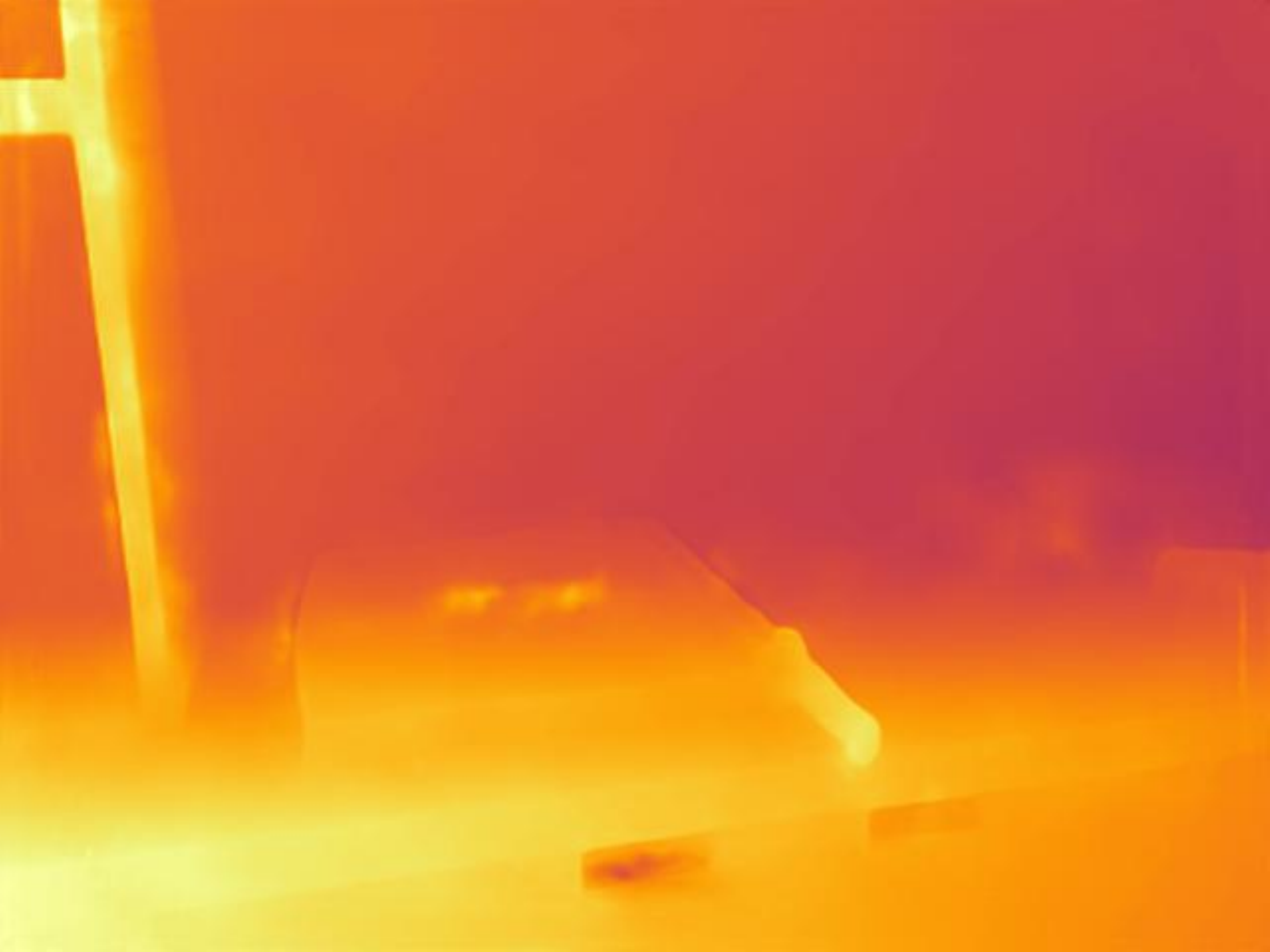}&
    \includegraphics[width=0.096\linewidth]{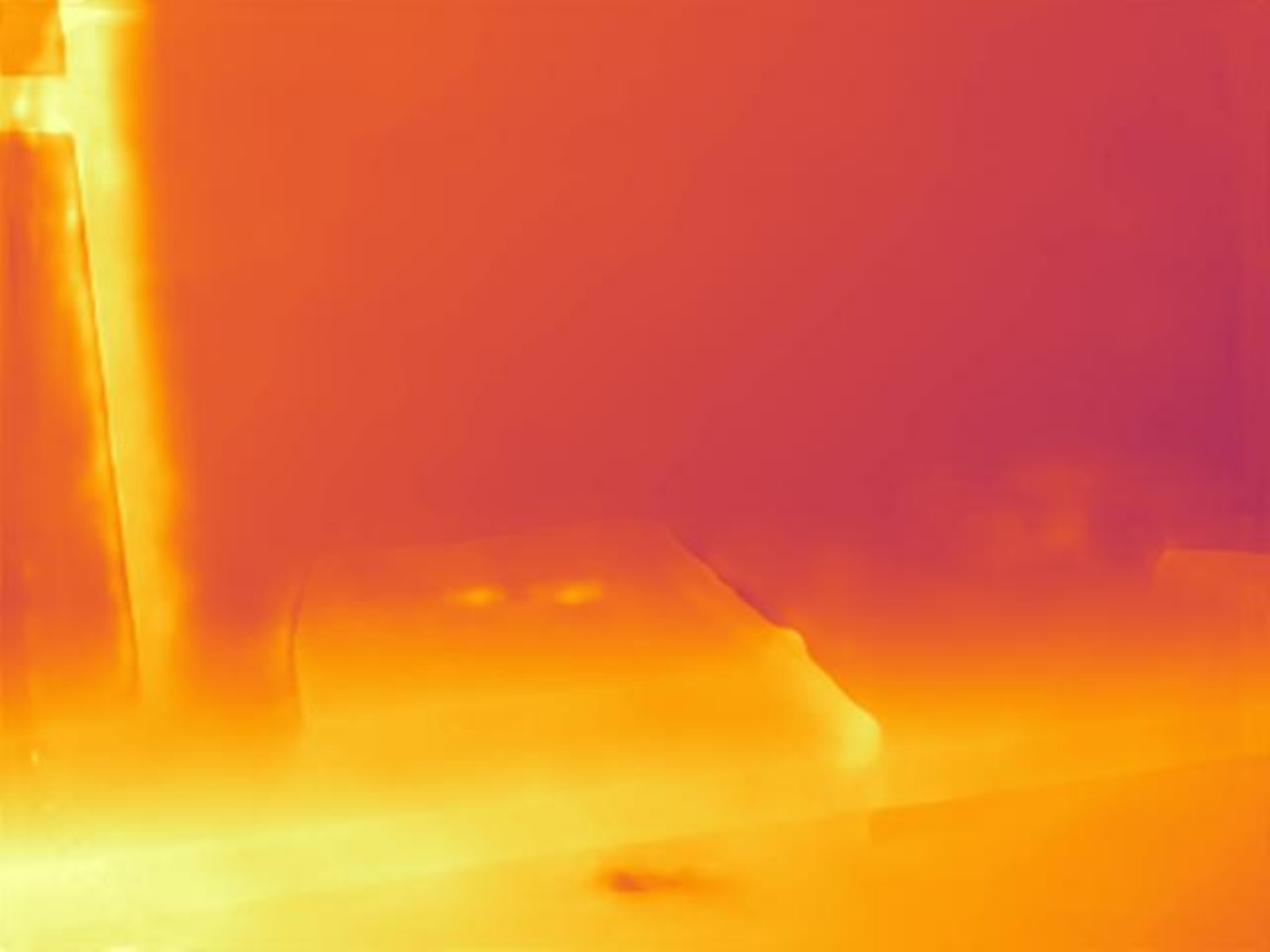}&
    \includegraphics[width=0.096\linewidth]{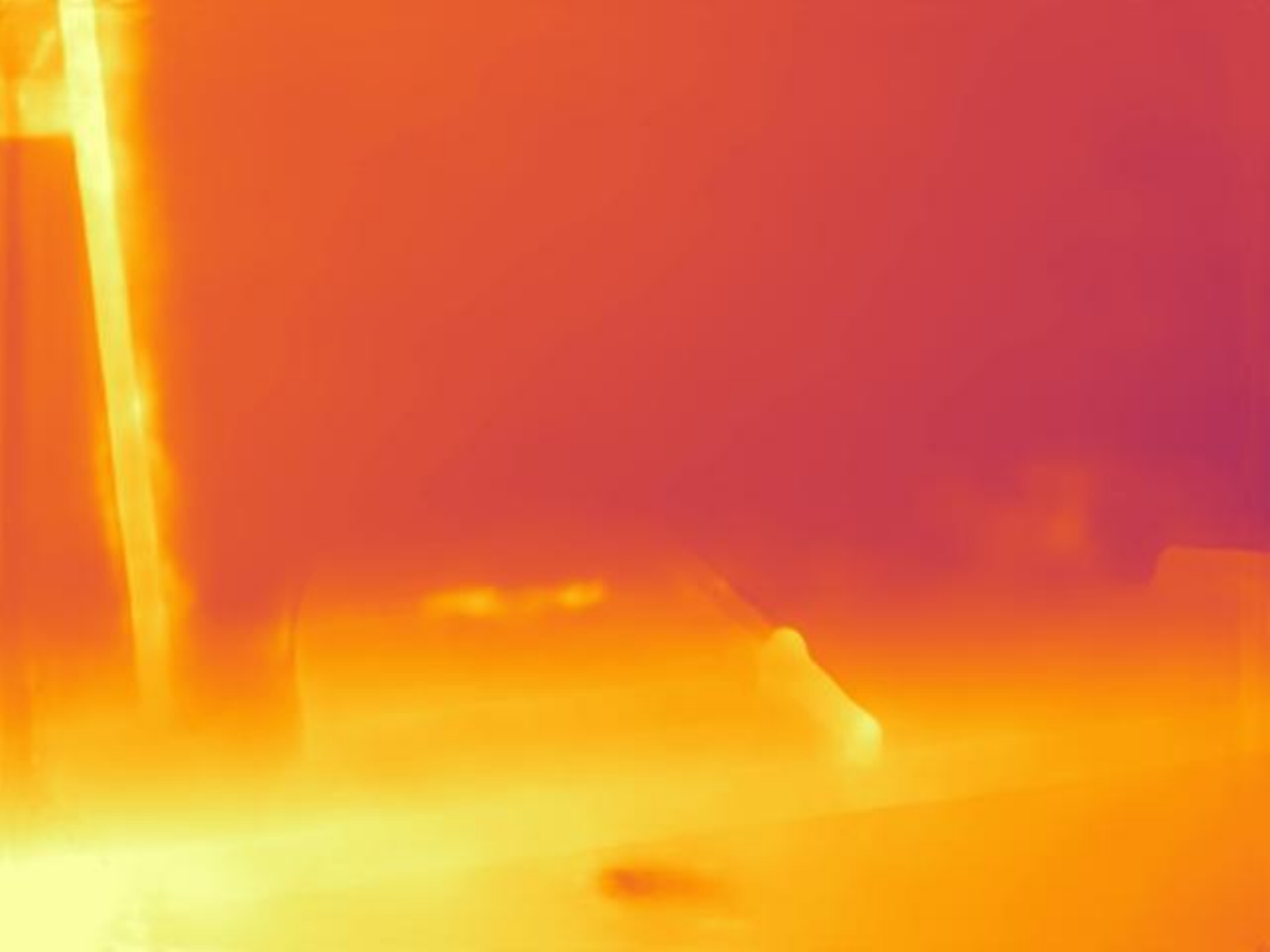}&
    \includegraphics[width=0.096\linewidth]{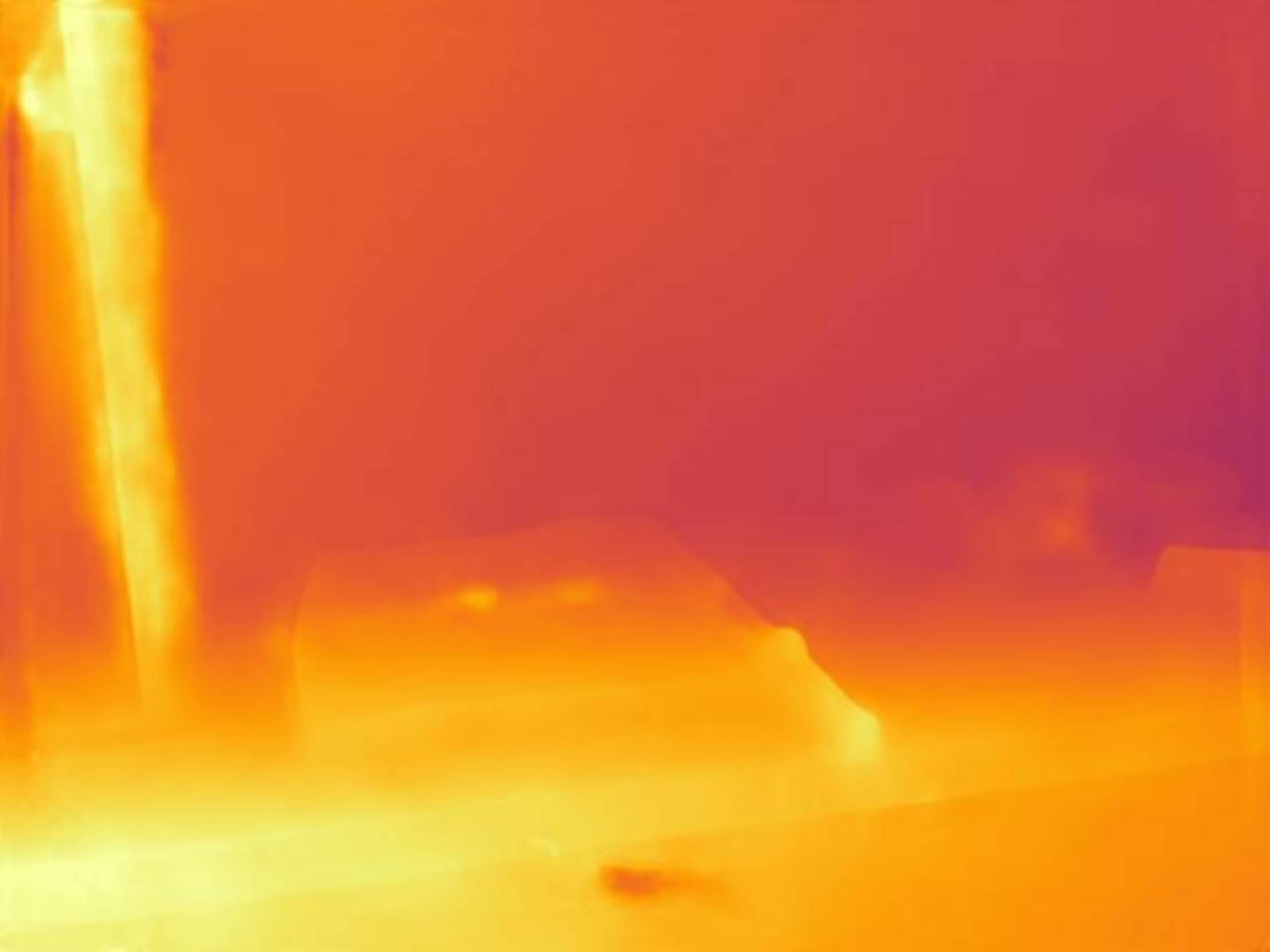}&
    \includegraphics[width=0.096\linewidth]{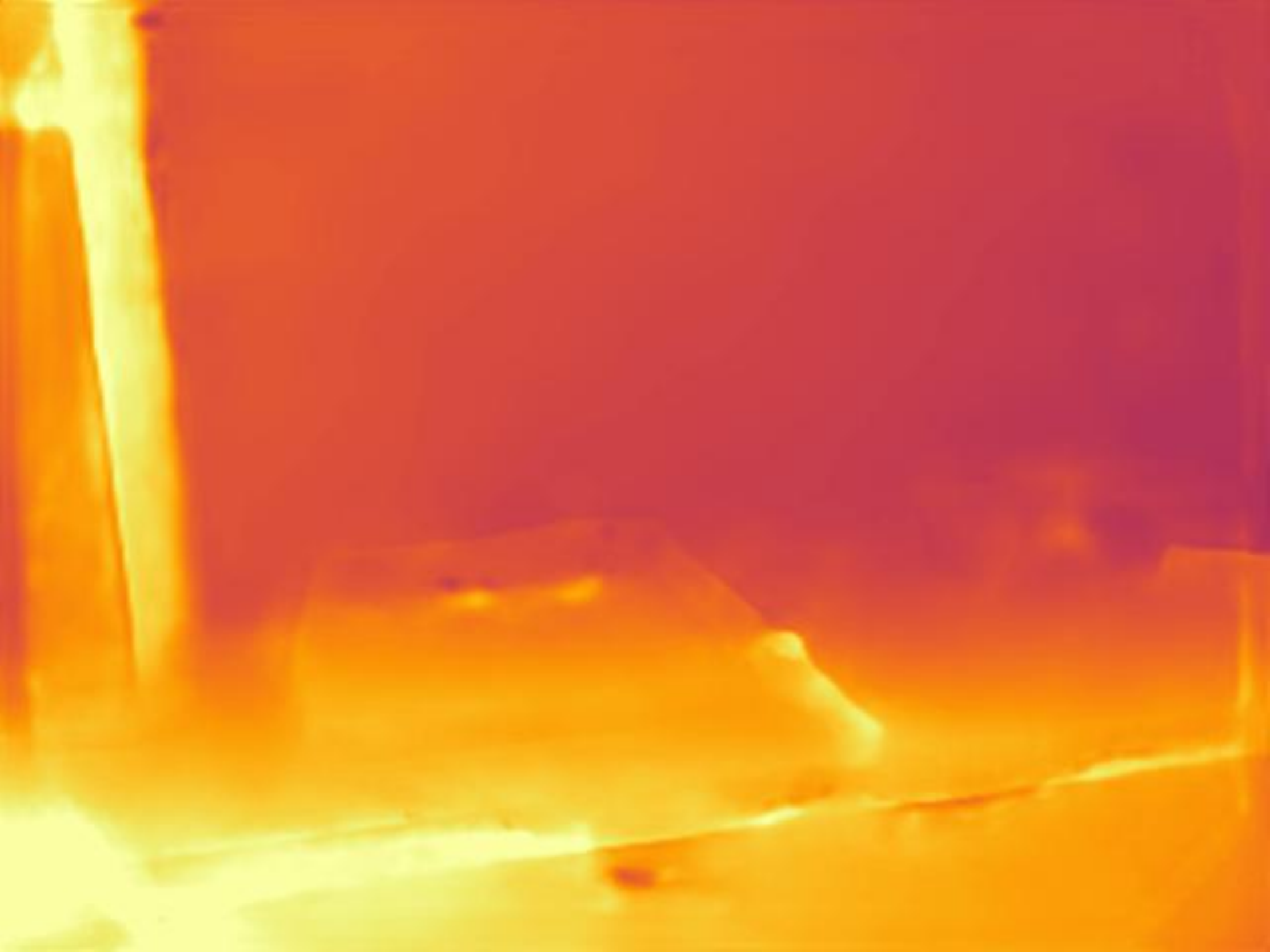}&
    \includegraphics[width=0.096\linewidth]{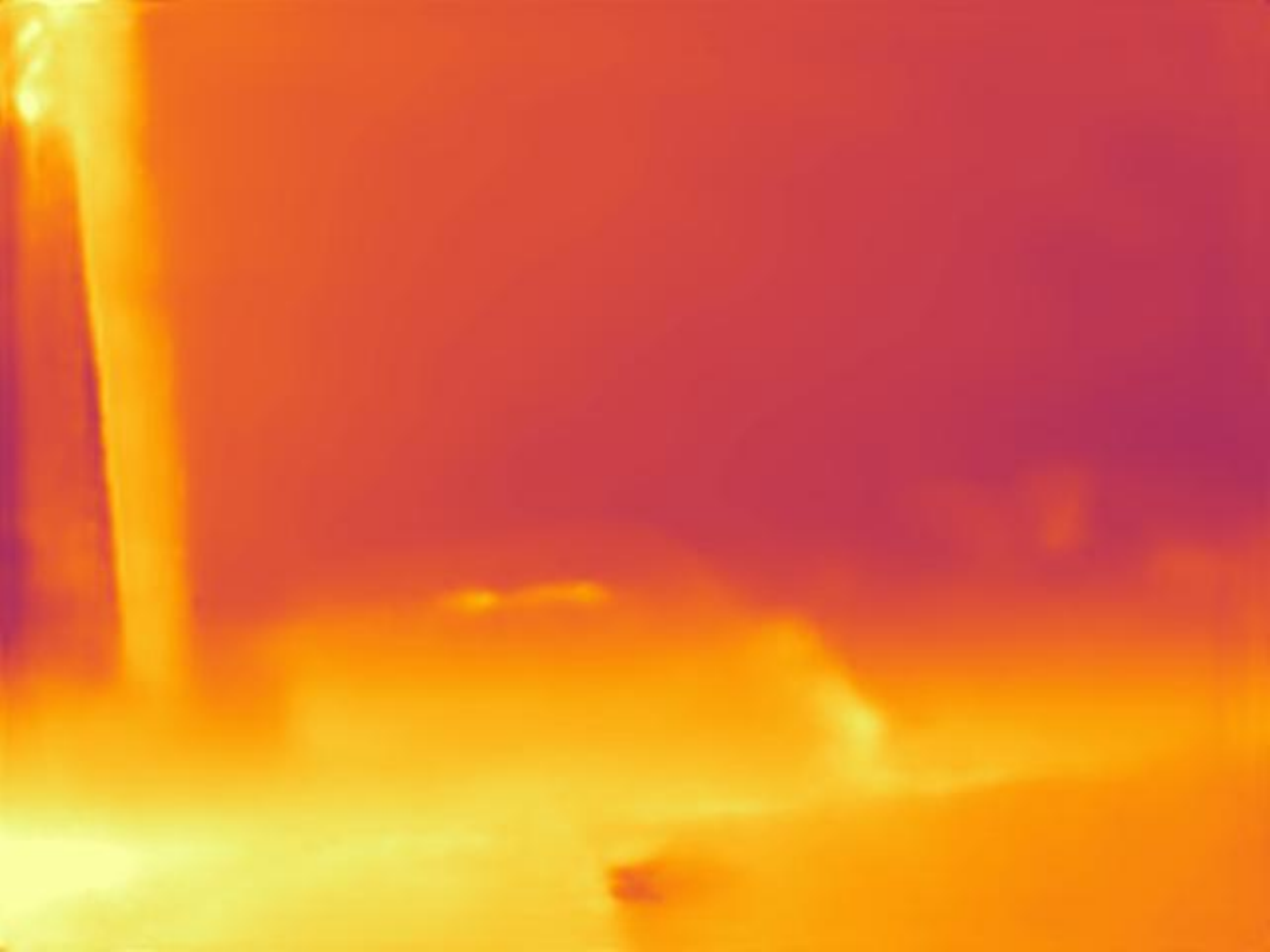}&
    \includegraphics[width=0.096\linewidth]{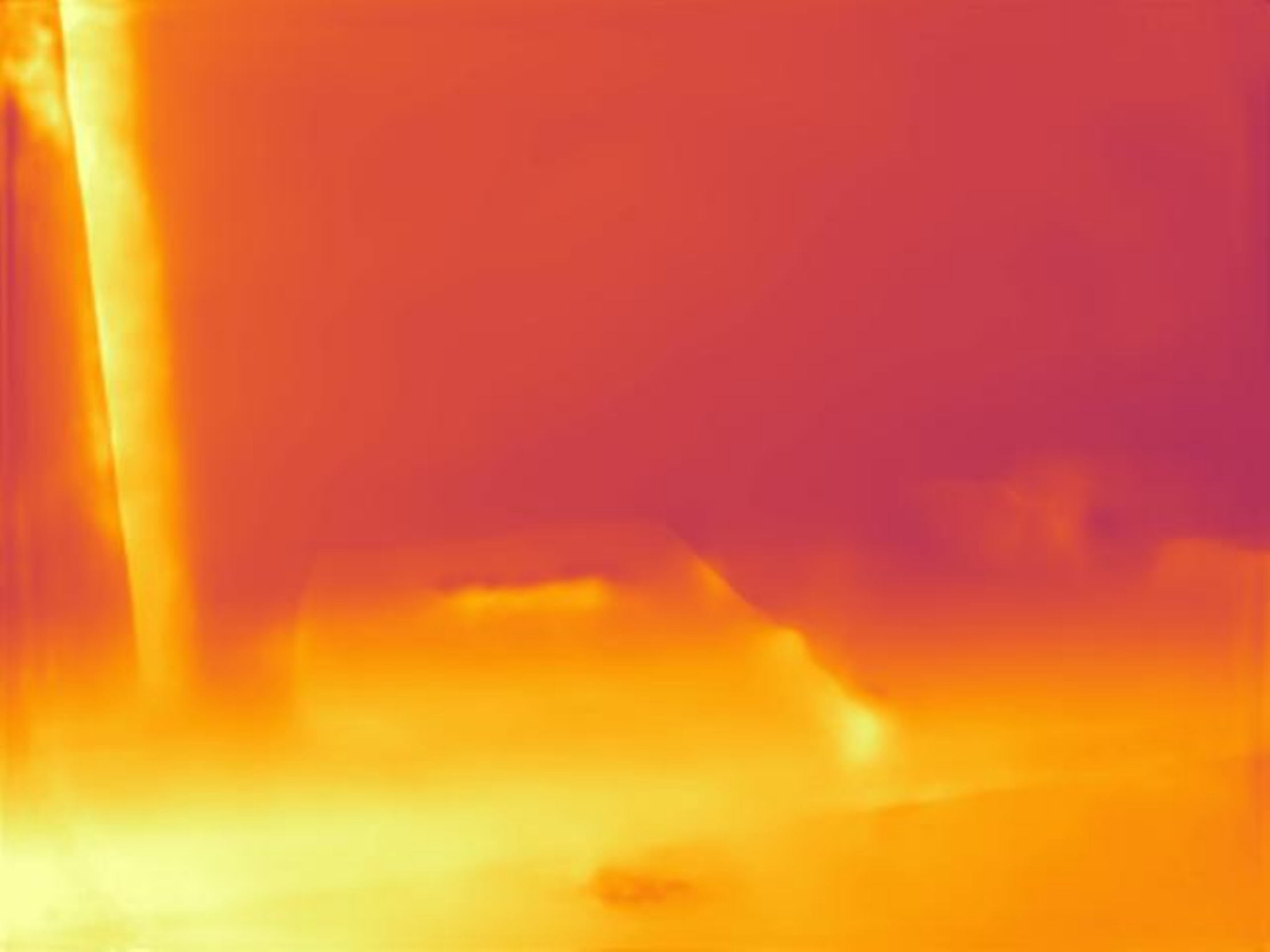}\\
    \multicolumn{9}{c}{} \\
  \end{tabular}
  \caption{Extension of Figure~\ref{fig:vis-void-densities} to visualize depth output by our method GA+SML using different depth estimators from the MiDaS/DPT family of models. Prior state-of-the-art KBNet is also shown as a baseline. Our approach produces depth maps with improved sharpness and planar consistency. Improvements are more significant at lower sparse depth densities, which is in accordance with our quantitative evaluation in Table~\ref{tab:appendix_void_comparison}.}
  \label{fig:vis-void-comparison-densities-depth}
  \vspace{50pt}
\end{figure*}

We visualize depth output by our pipeline with different depth estimators in Figure~\ref{fig:vis-void-comparison-densities-depth}. Our observation that our method yields sharper depth maps than KBNet continues to hold across different sparse depth densities. 

\begin{figure*}[p]
\centering
  \begin{tabular}{@{}l@{\hspace{0.5mm}}*{9}{c@{\hspace{0.5mm}}}c@{}}
    & {\scriptsize RGB Image} & {\scriptsize Sparse Depth} & {\scriptsize Scales Scaffold.} & {\scriptsize Regressed Scales} & {\scriptsize GA Depth} & {\scriptsize SML Depth} & {\scriptsize Ground Truth} & {\scriptsize GA Error} & {\scriptsize SML Error}\\
    \vspace{-0.75mm}
    \rot{\scriptsize VOID 150} & \includegraphics[width=0.104\linewidth]{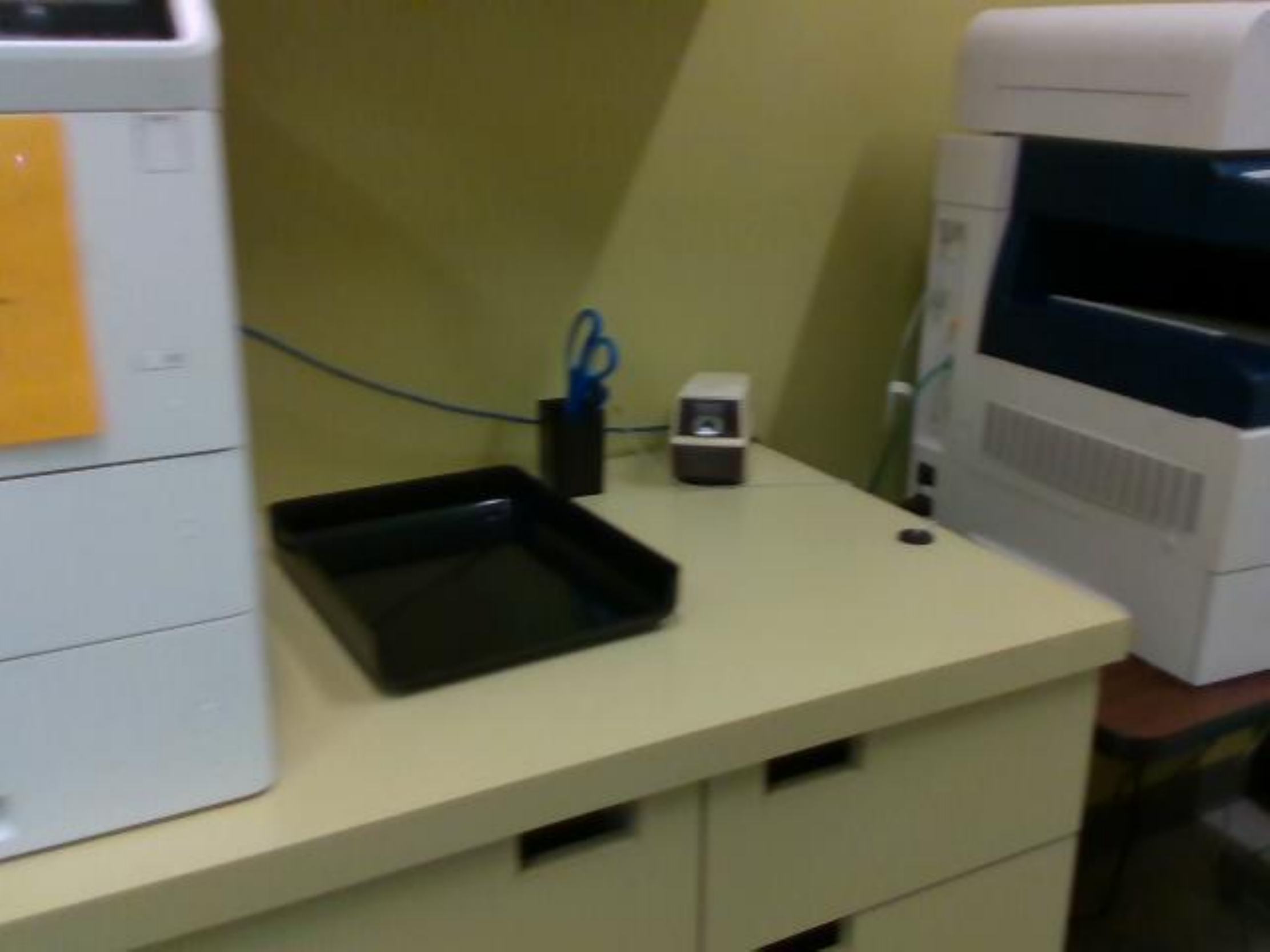}&
    \includegraphics[width=0.104\linewidth]{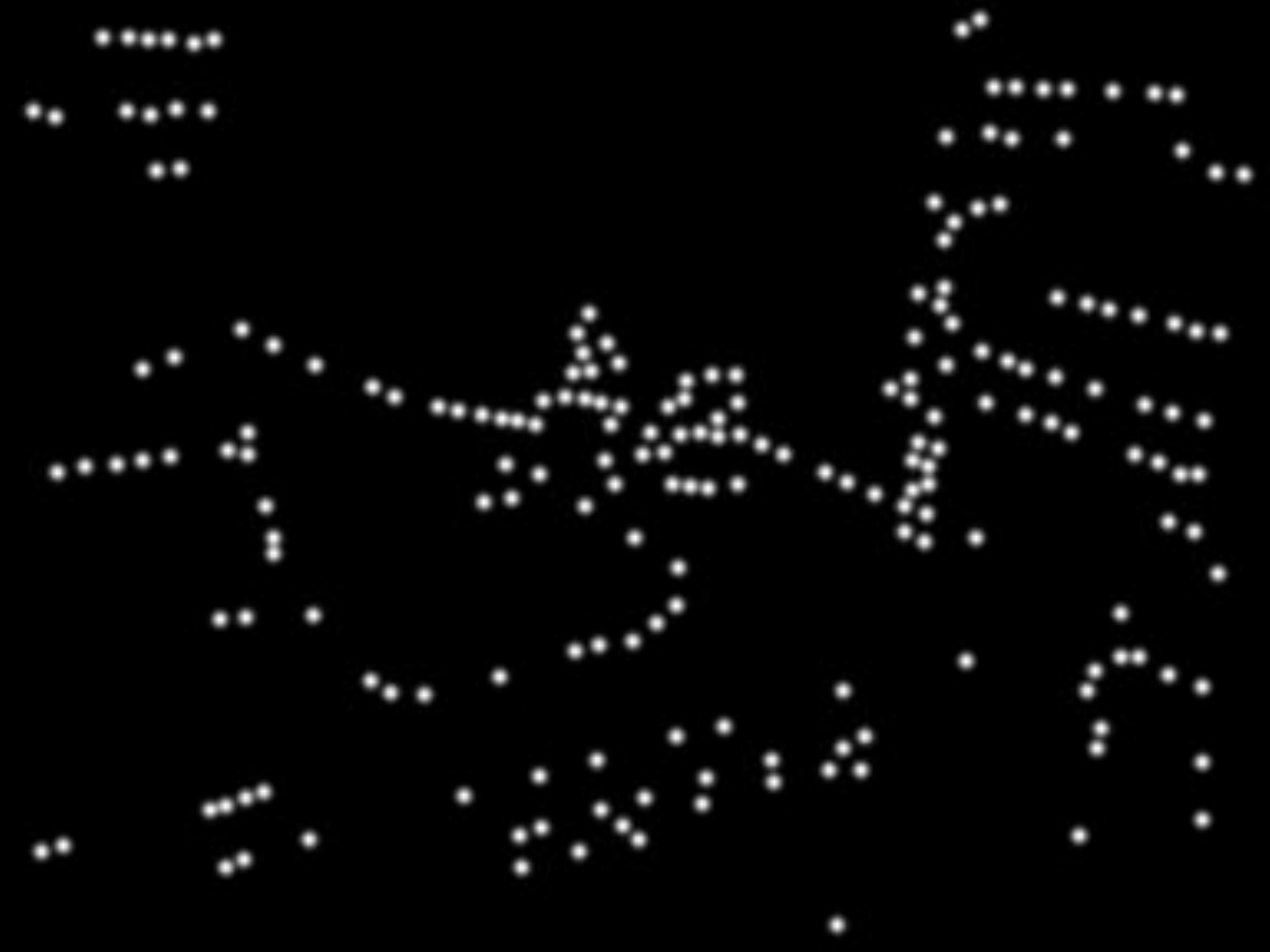}&
    \includegraphics[width=0.104\linewidth]{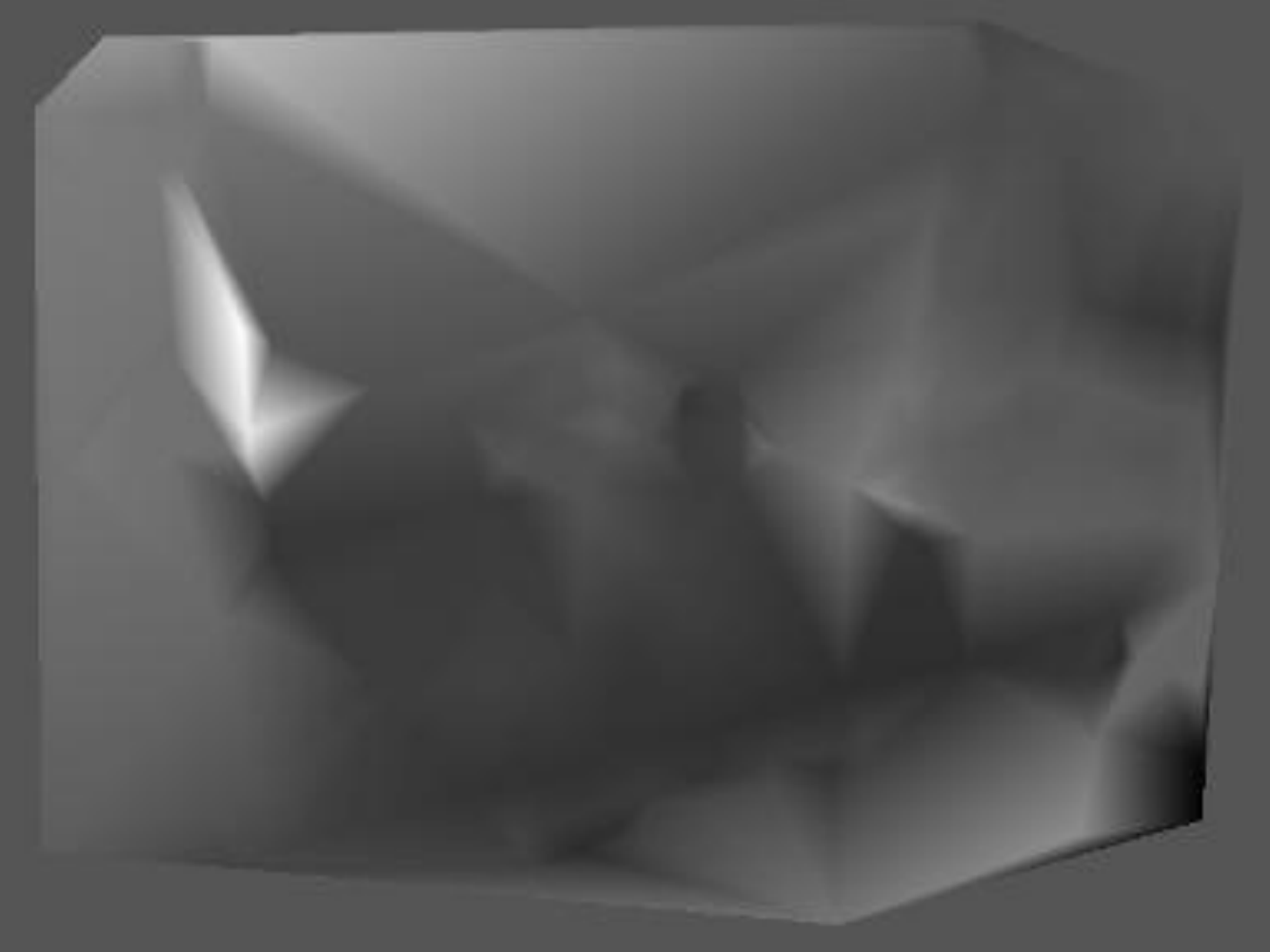}&
    \includegraphics[width=0.104\linewidth]{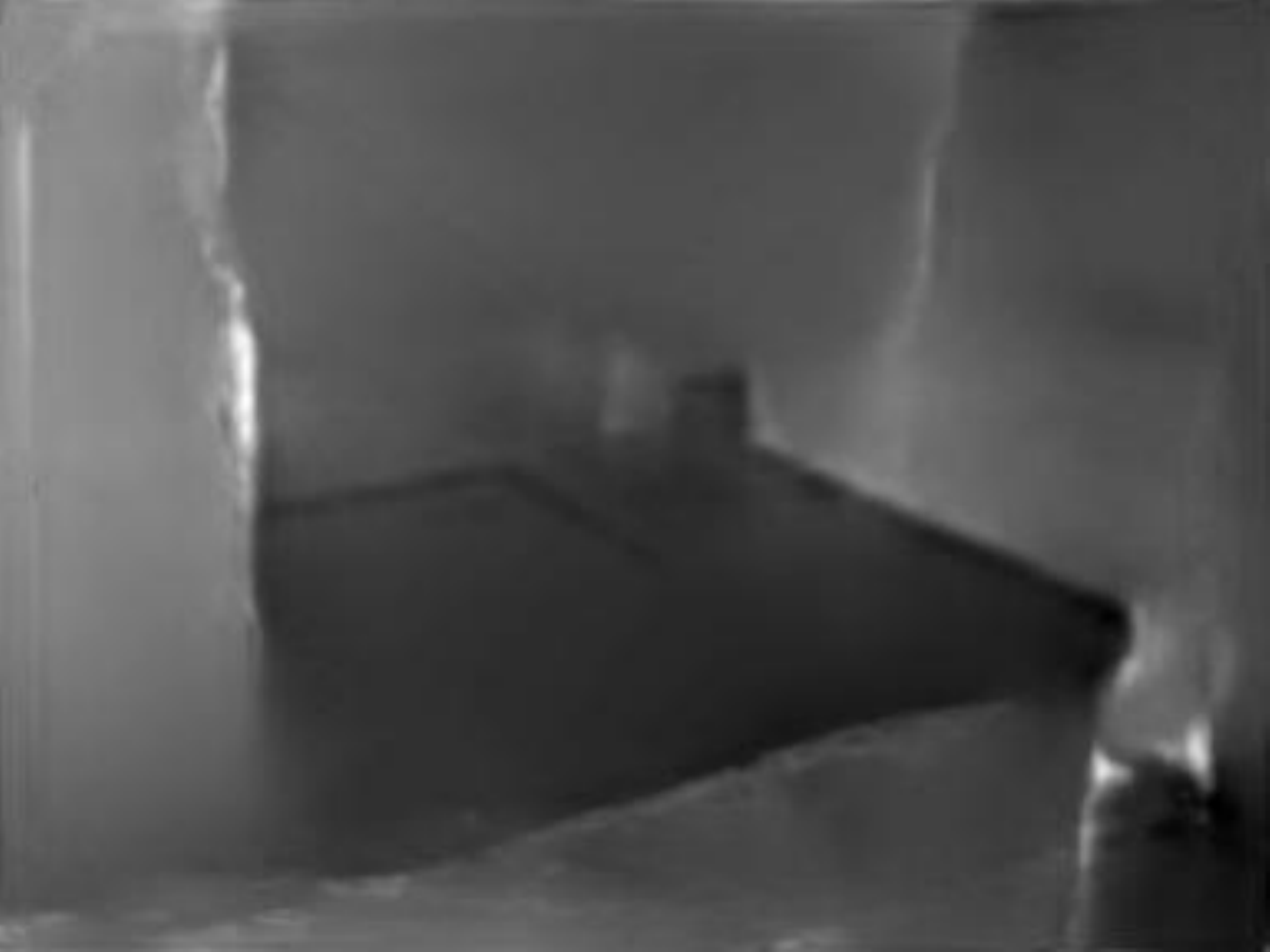}&
    \includegraphics[width=0.104\linewidth]{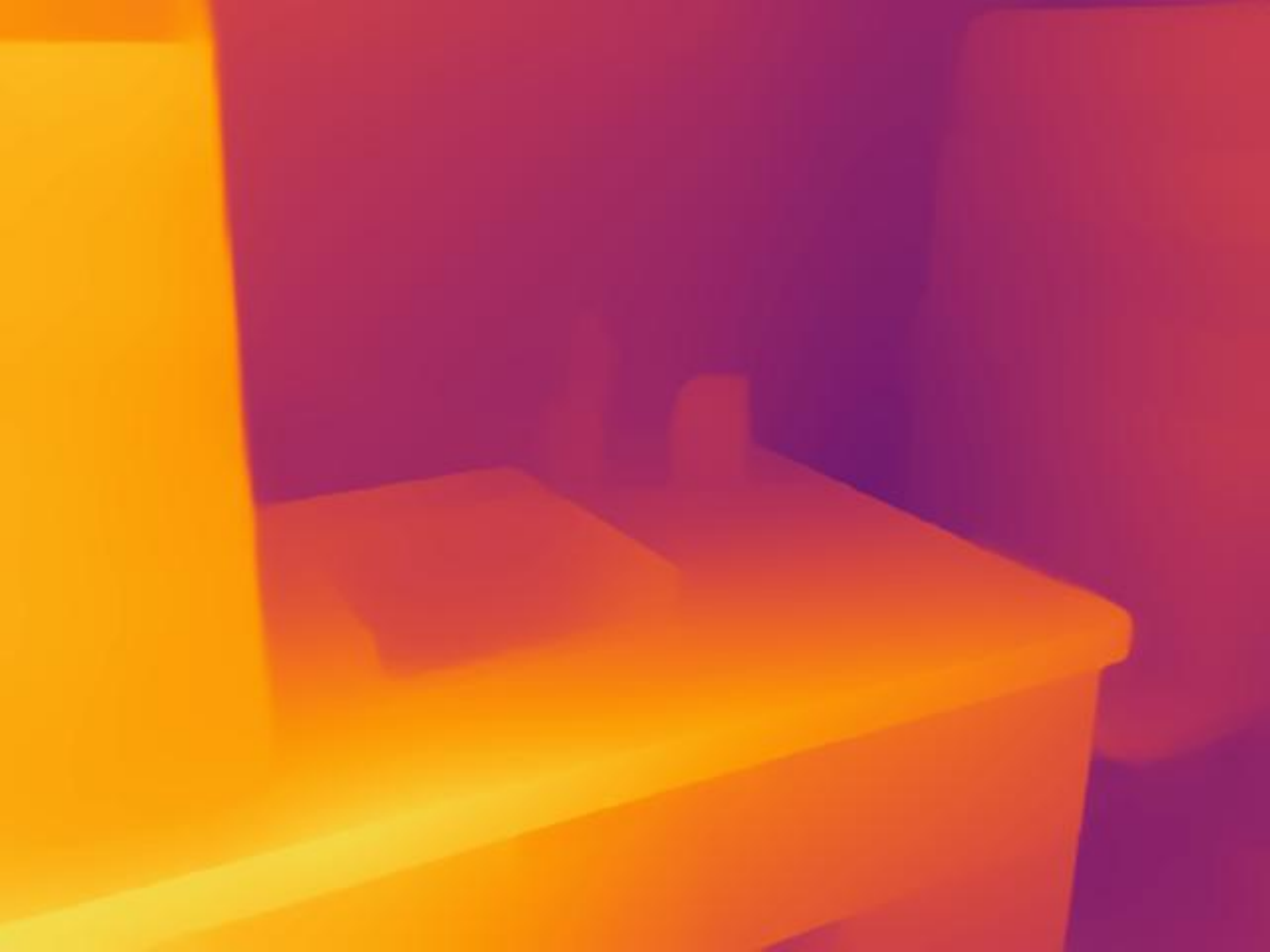}&
    \includegraphics[width=0.104\linewidth]{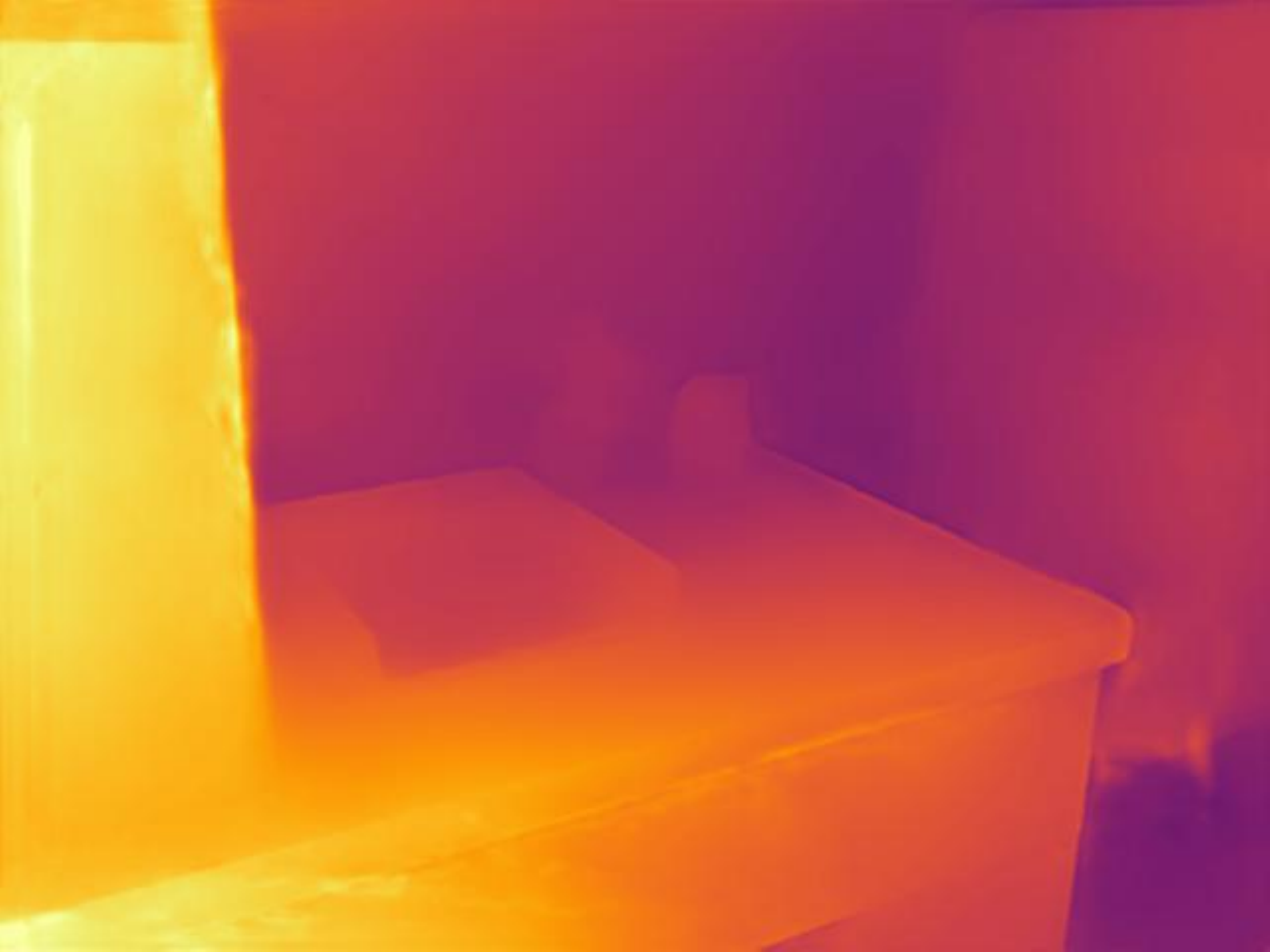}&
    \includegraphics[width=0.104\linewidth]{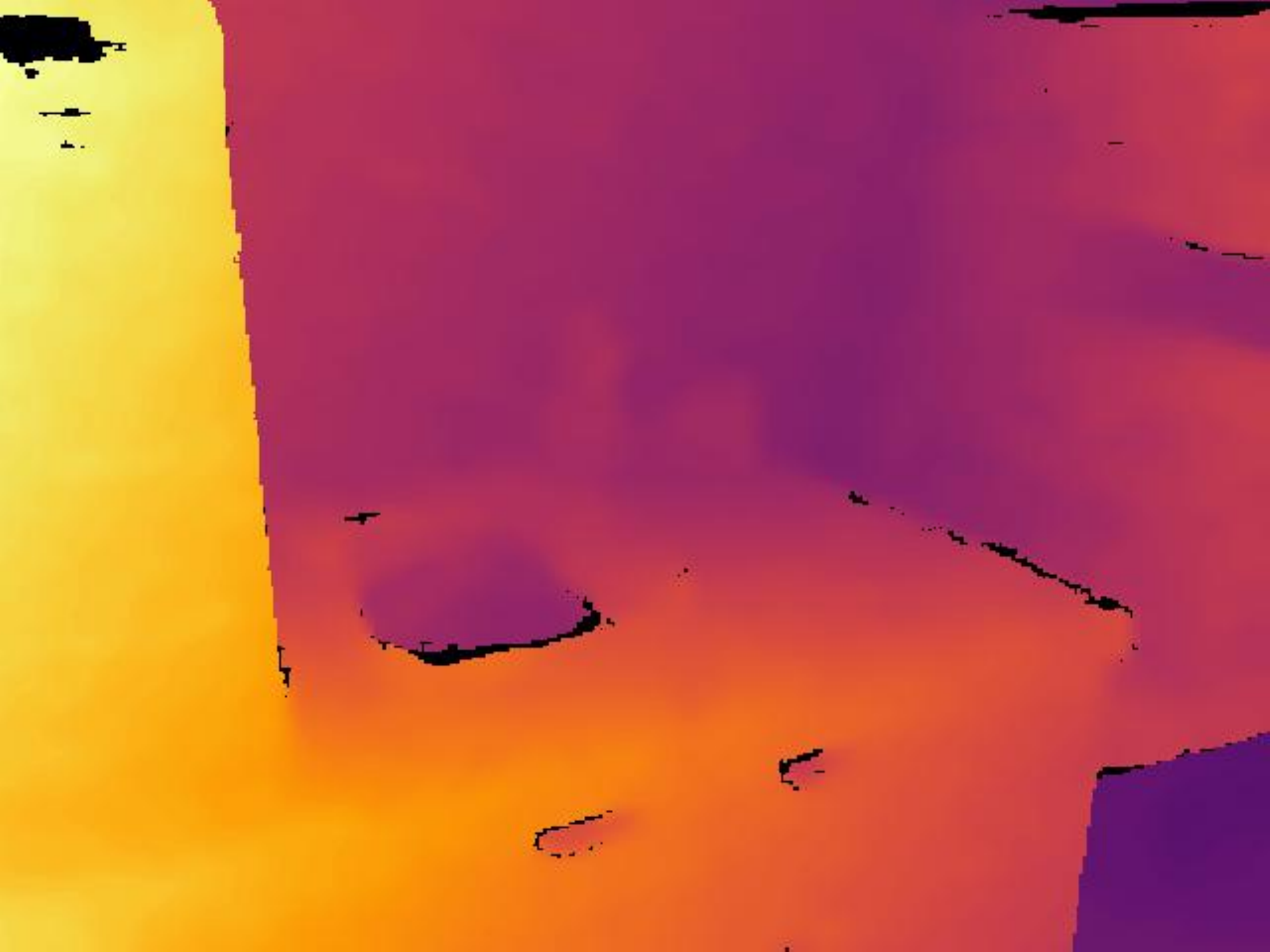}&
    \includegraphics[width=0.104\linewidth]{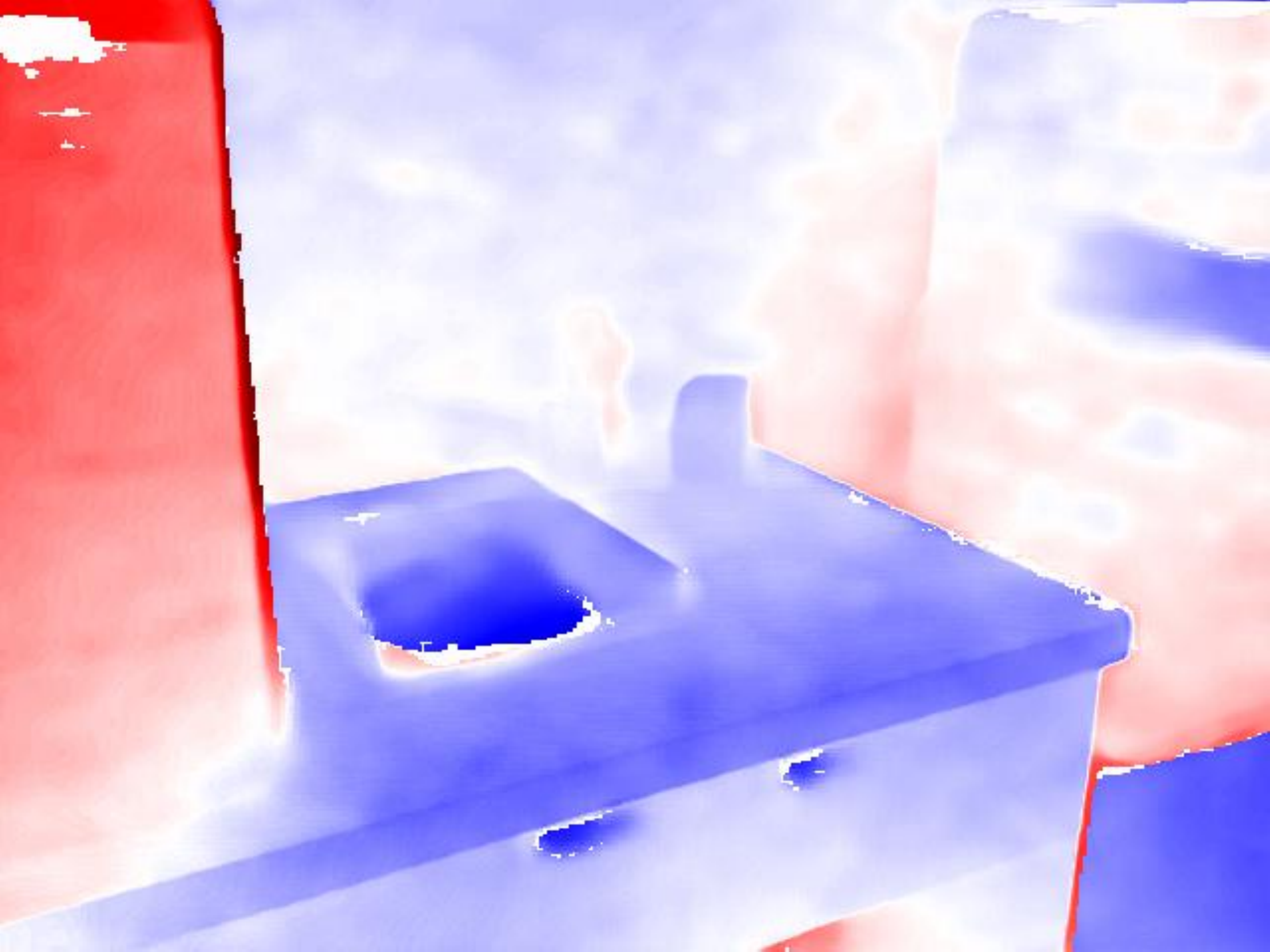}&
    \includegraphics[width=0.104\linewidth]{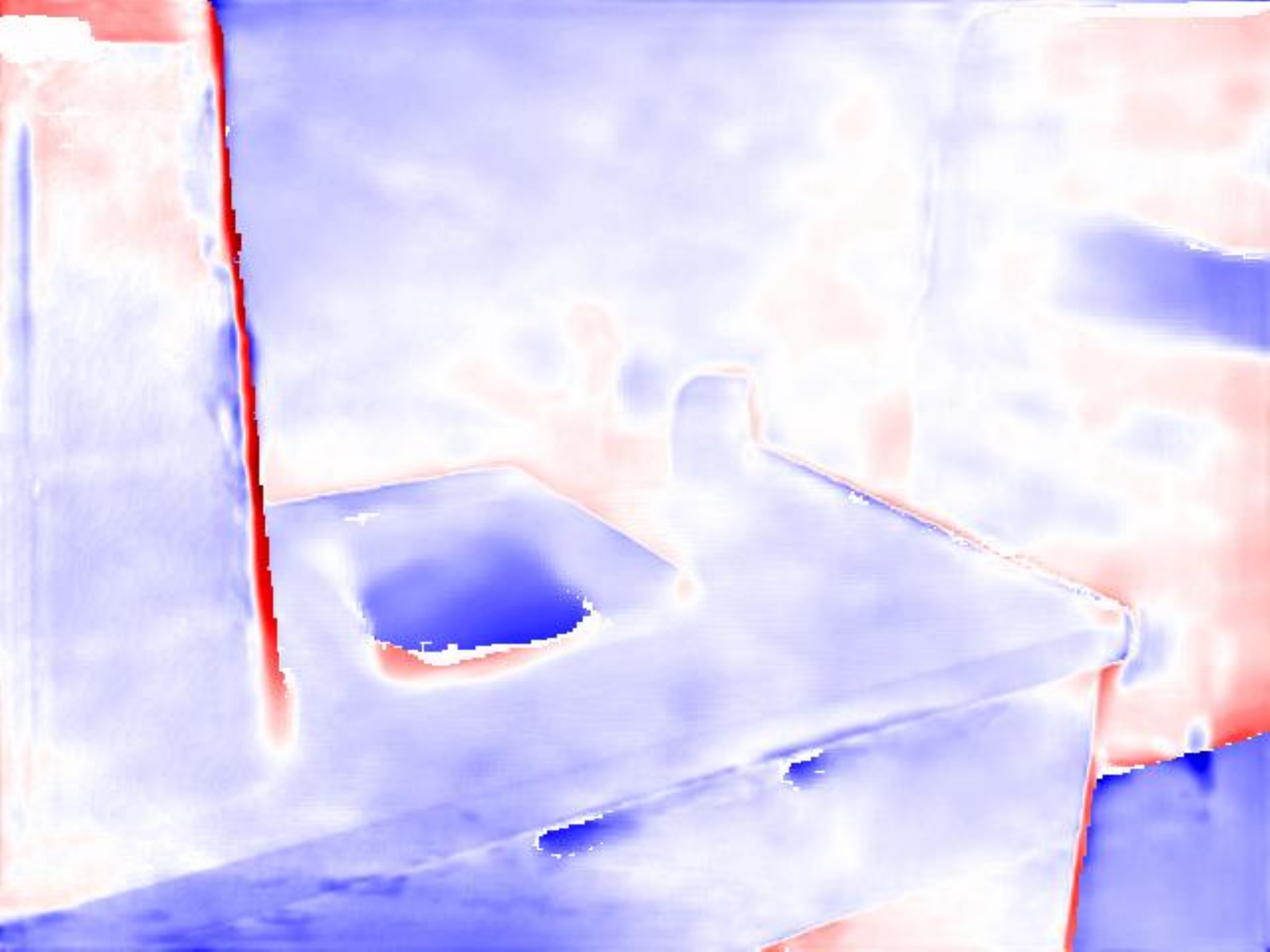}&
    \includegraphics[width=0.024\linewidth]{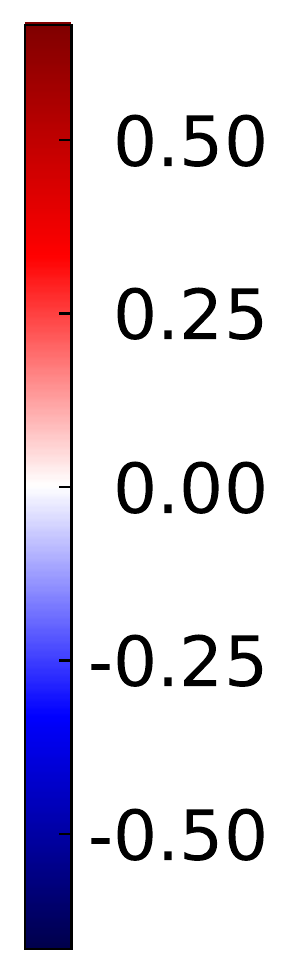}\\
    \vspace{-0.75mm}
    \rot{\scriptsize VINS 150} &
    \includegraphics[width=0.104\linewidth]{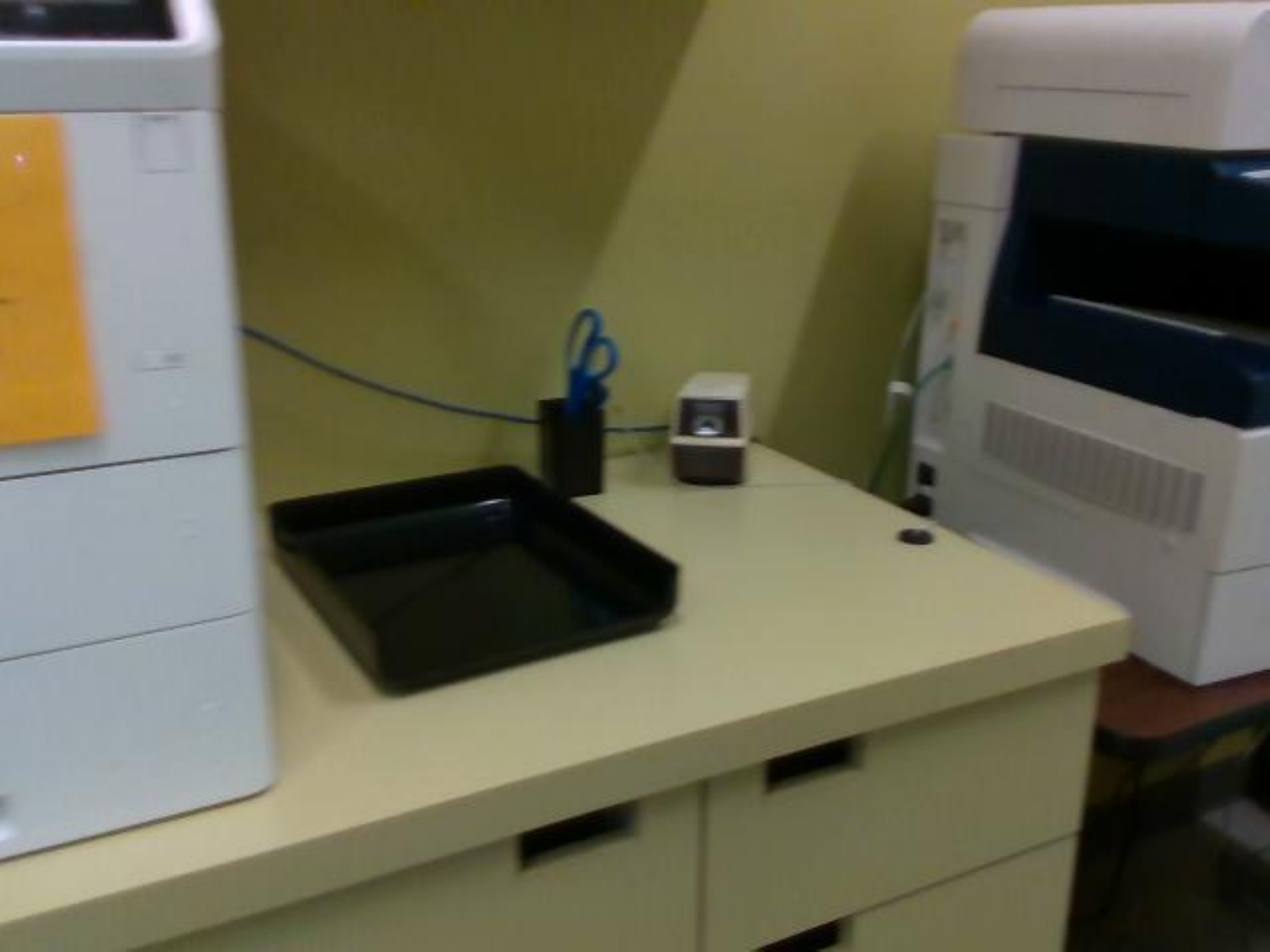}&
    \includegraphics[width=0.104\linewidth]{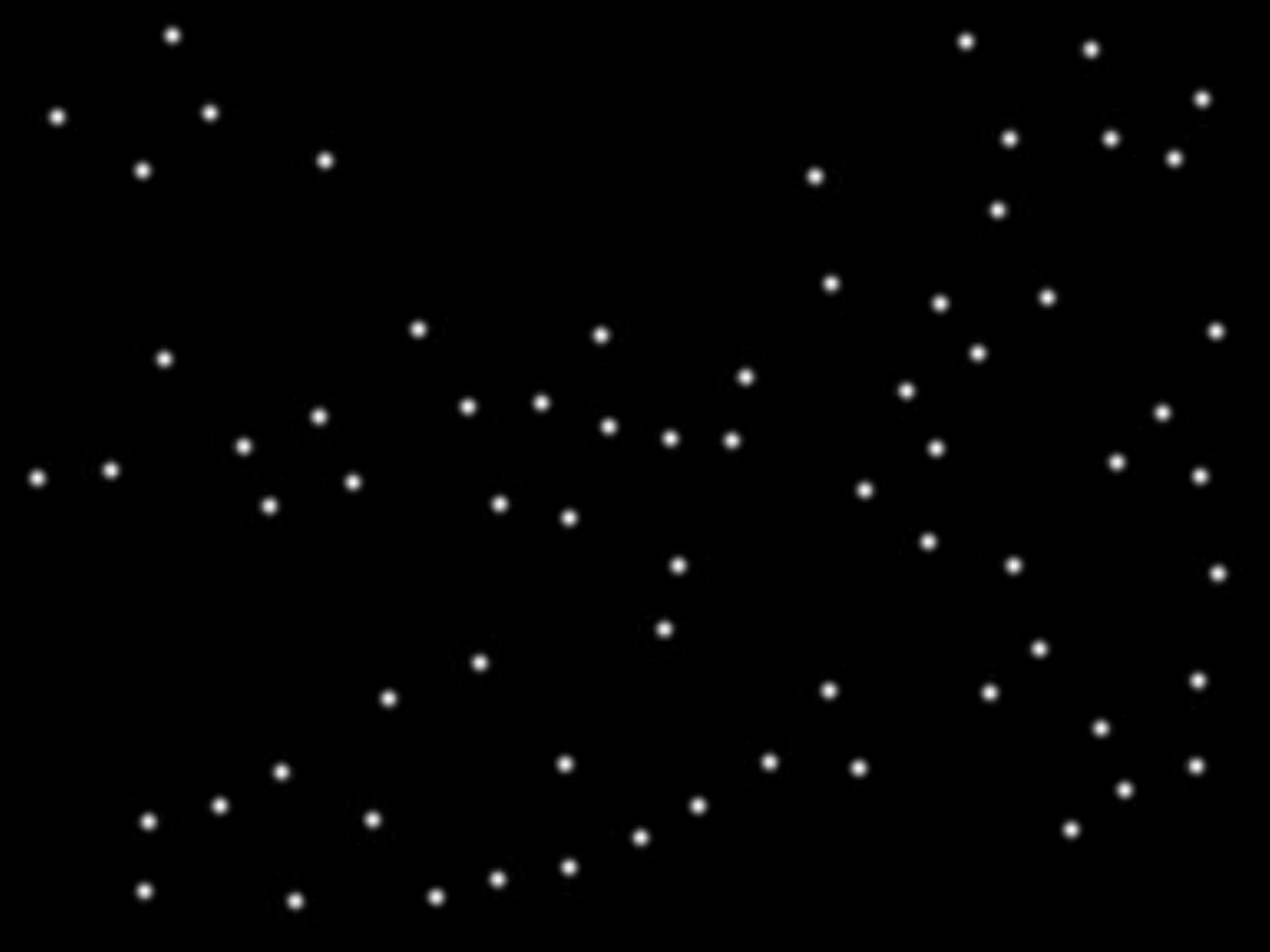}&
    \includegraphics[width=0.104\linewidth]{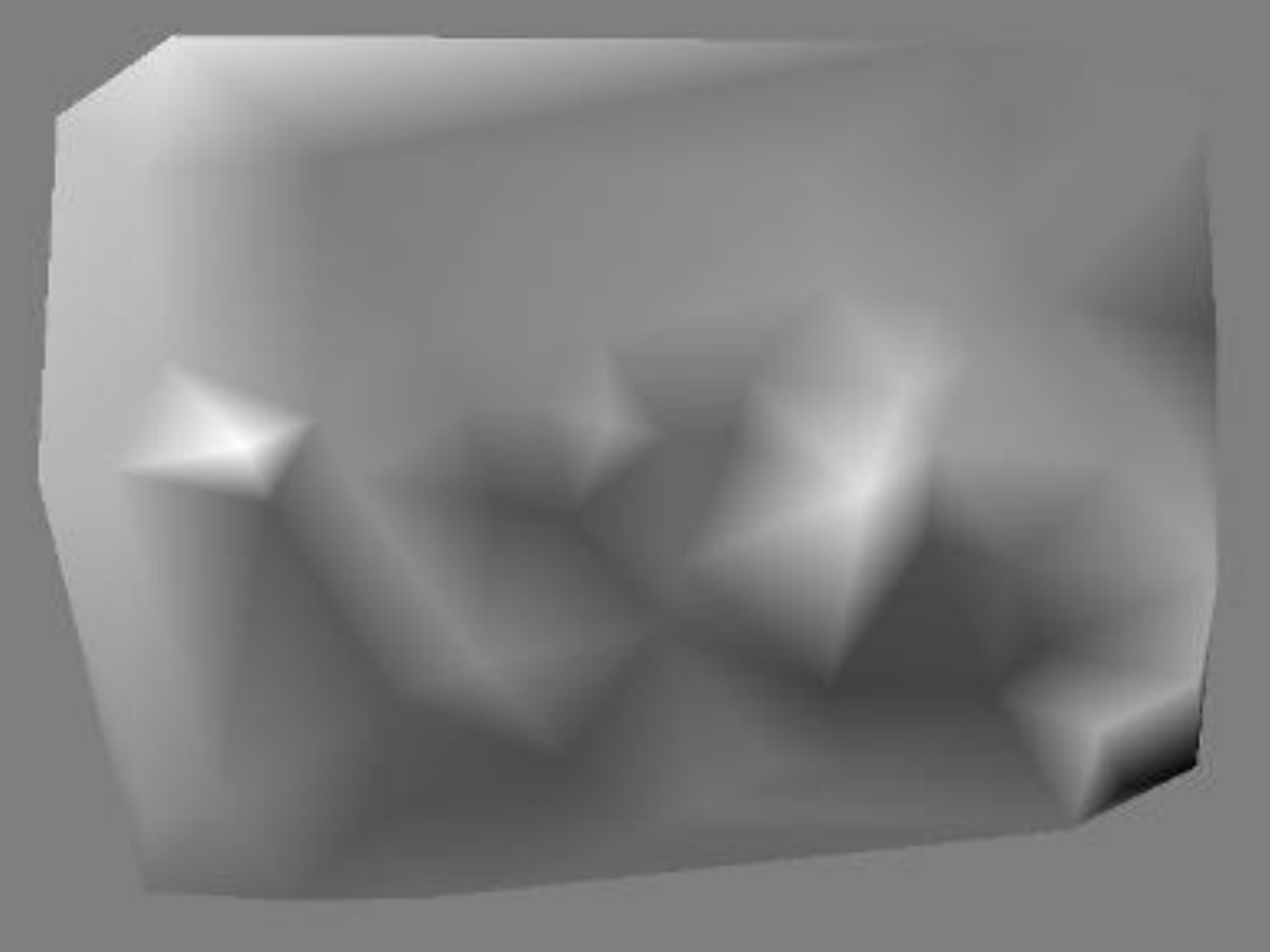}&
    \includegraphics[width=0.104\linewidth]{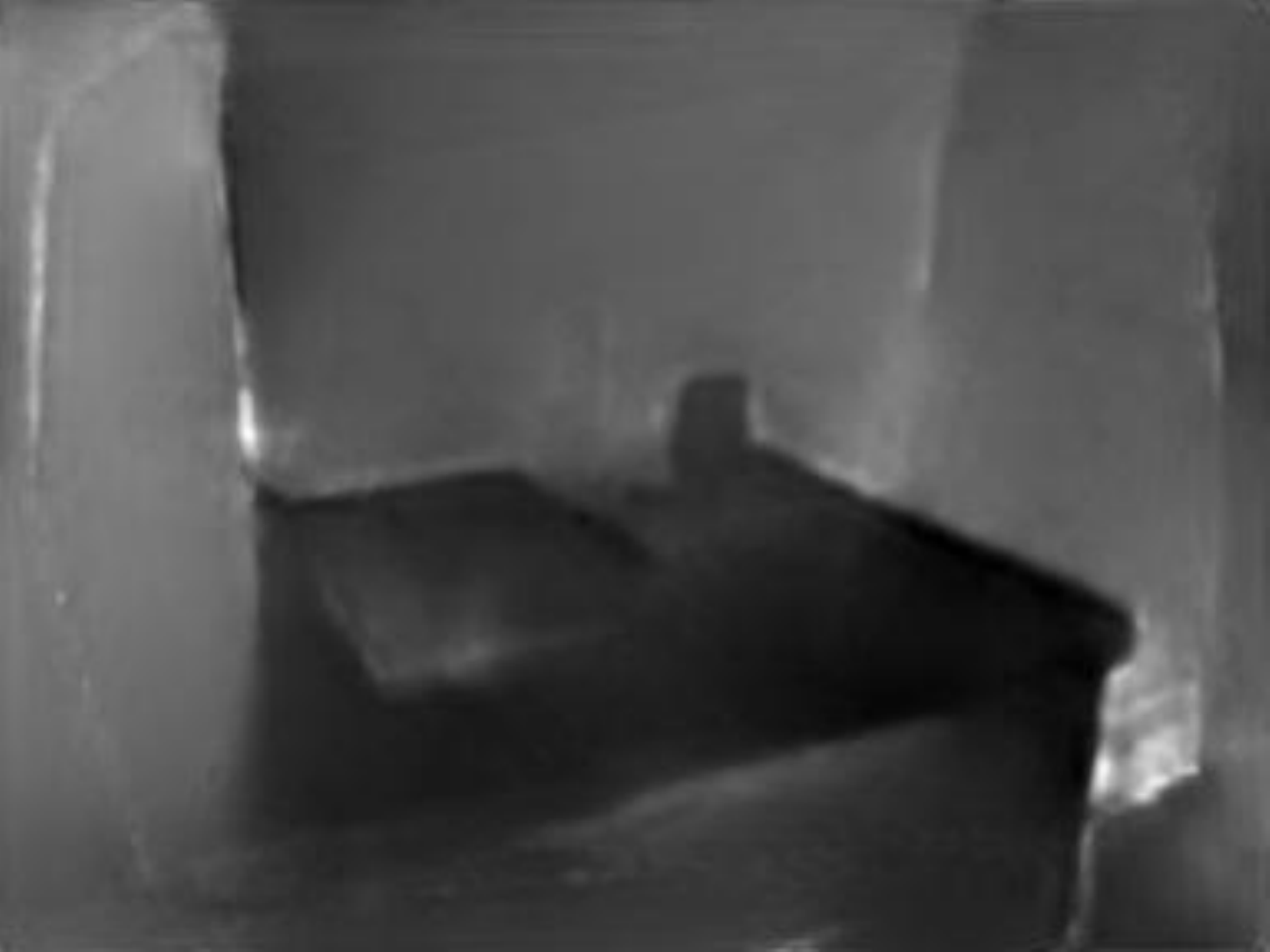}&
    \includegraphics[width=0.104\linewidth]{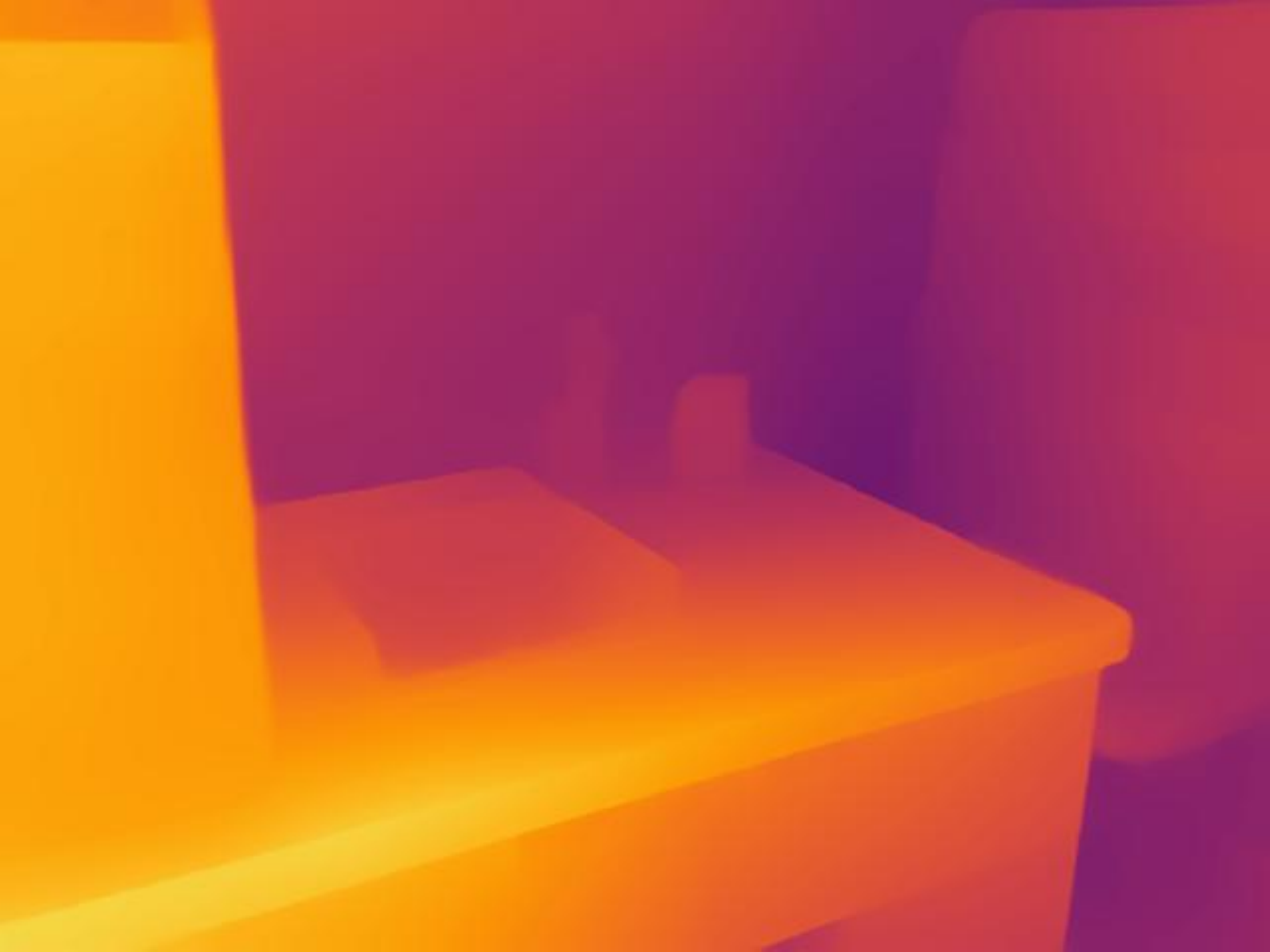}&
    \includegraphics[width=0.104\linewidth]{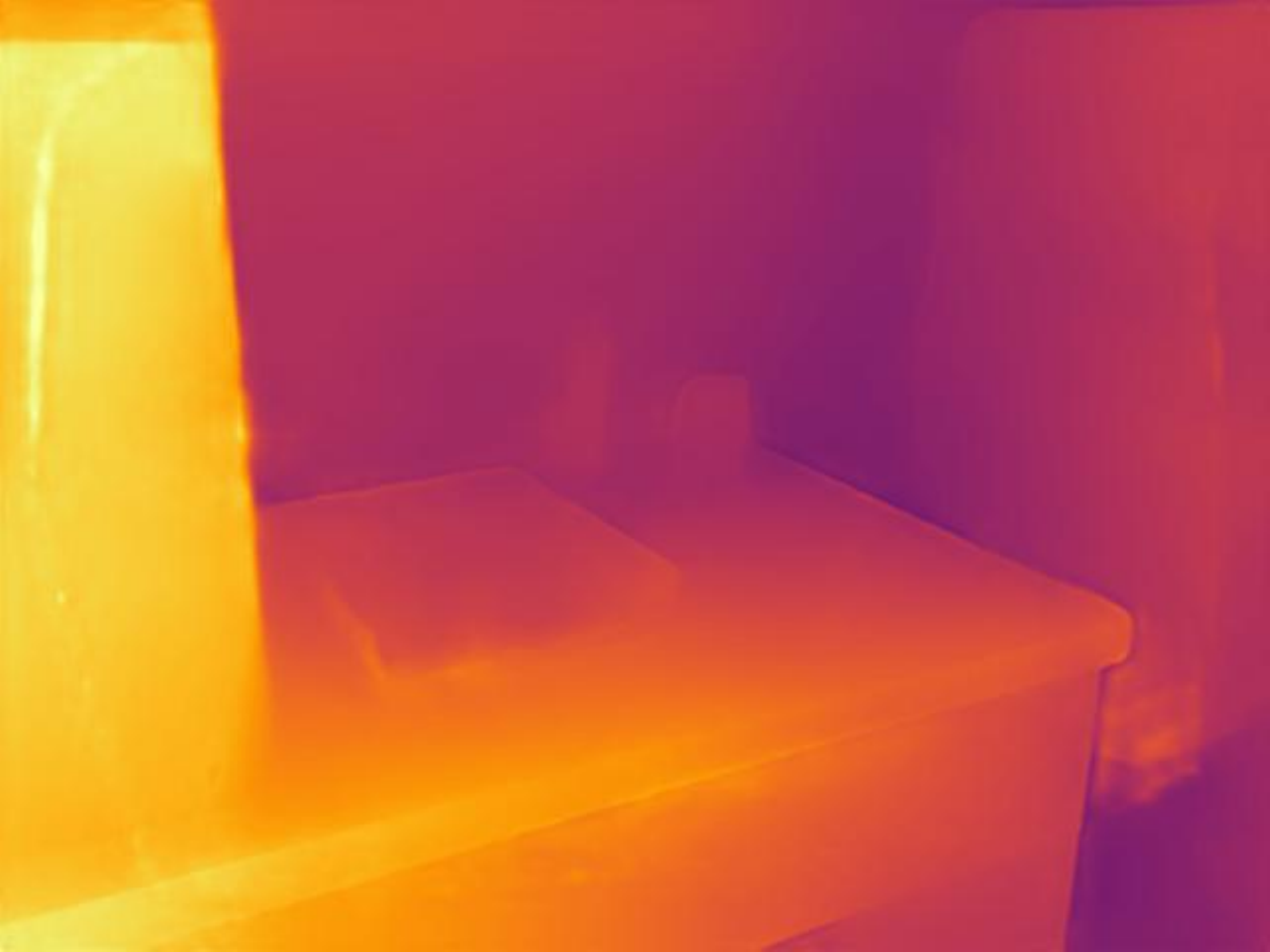}&
    \includegraphics[width=0.104\linewidth]{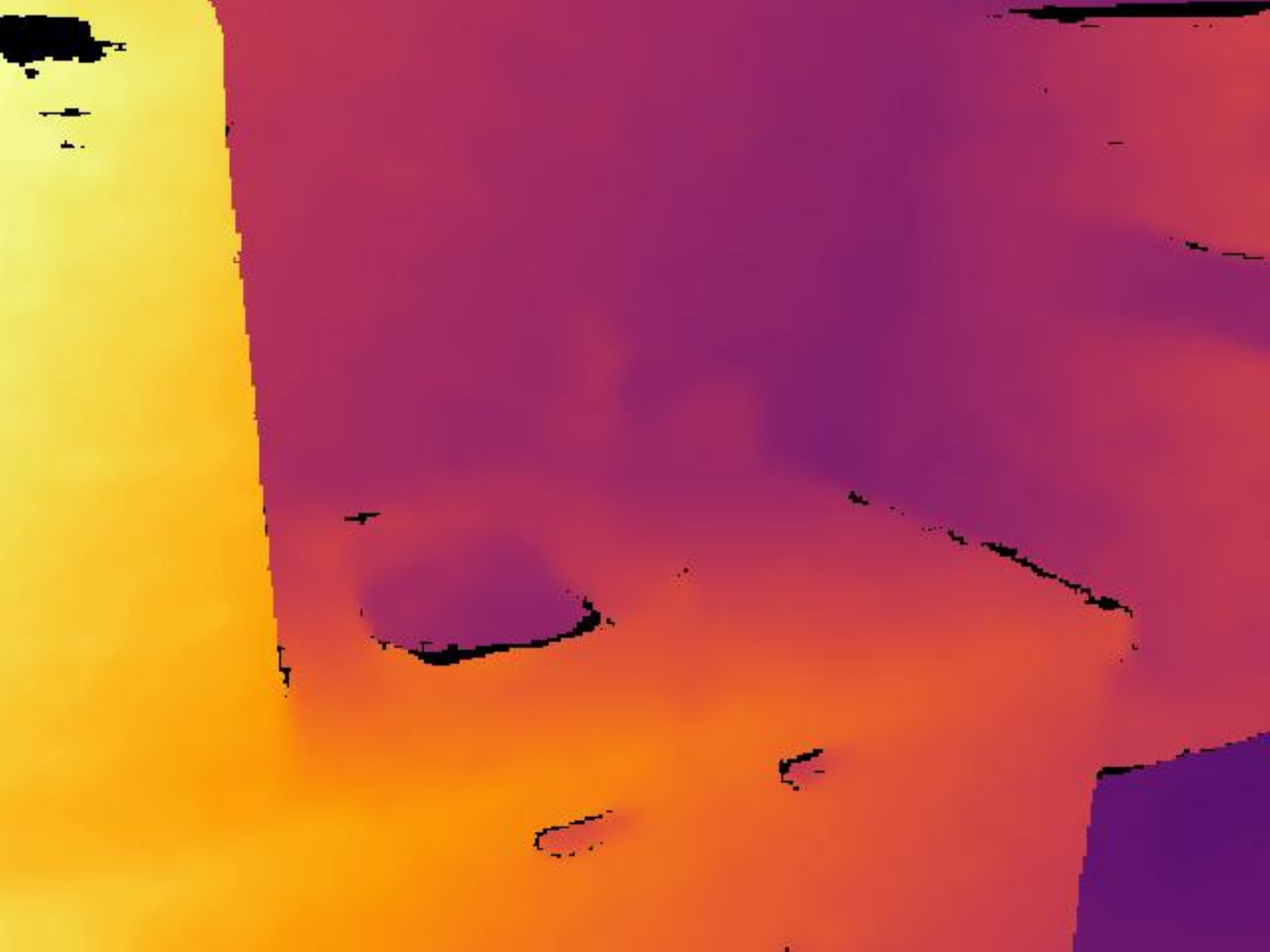}&
    \includegraphics[width=0.104\linewidth]{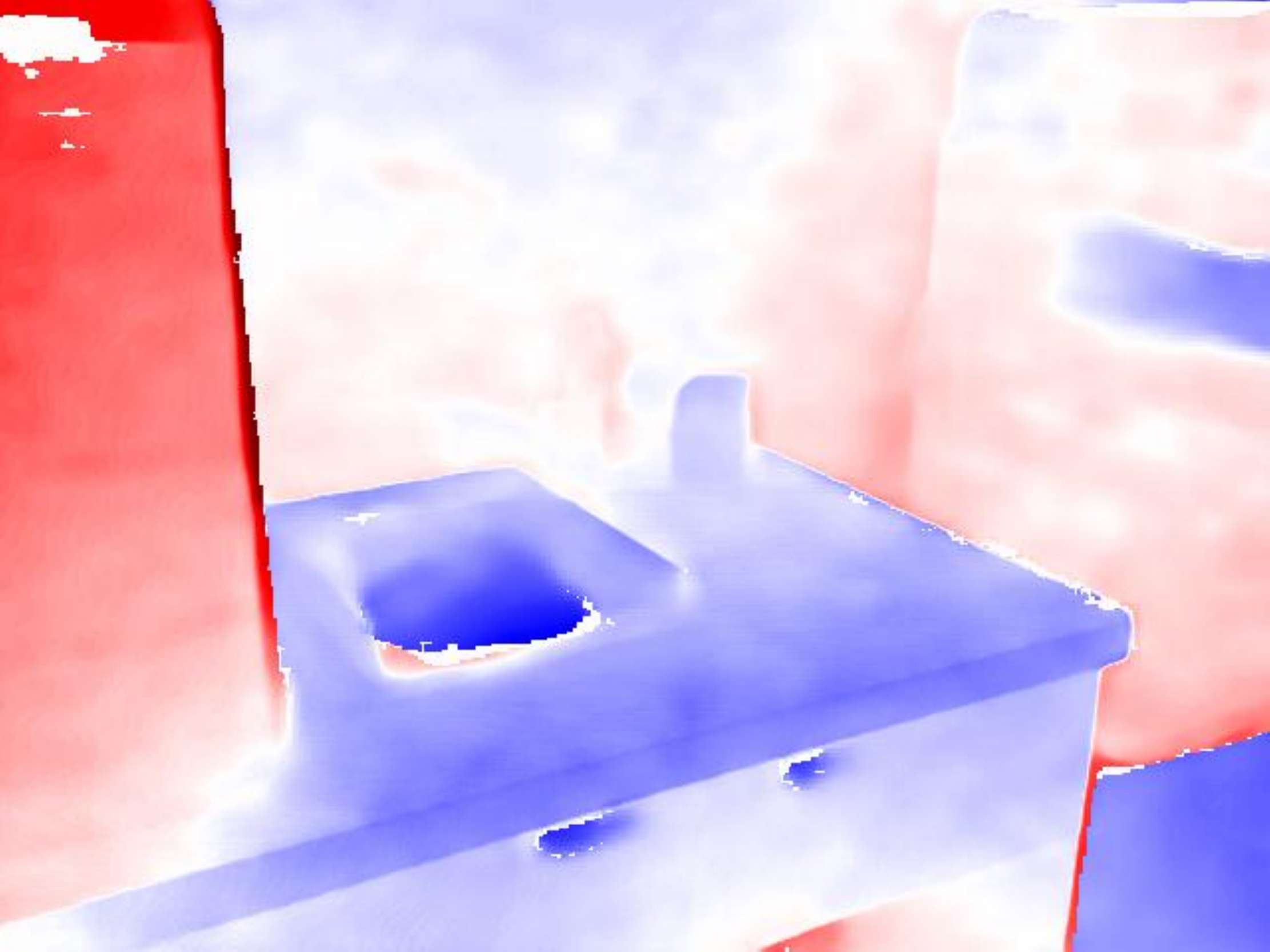}&
    \includegraphics[width=0.104\linewidth]{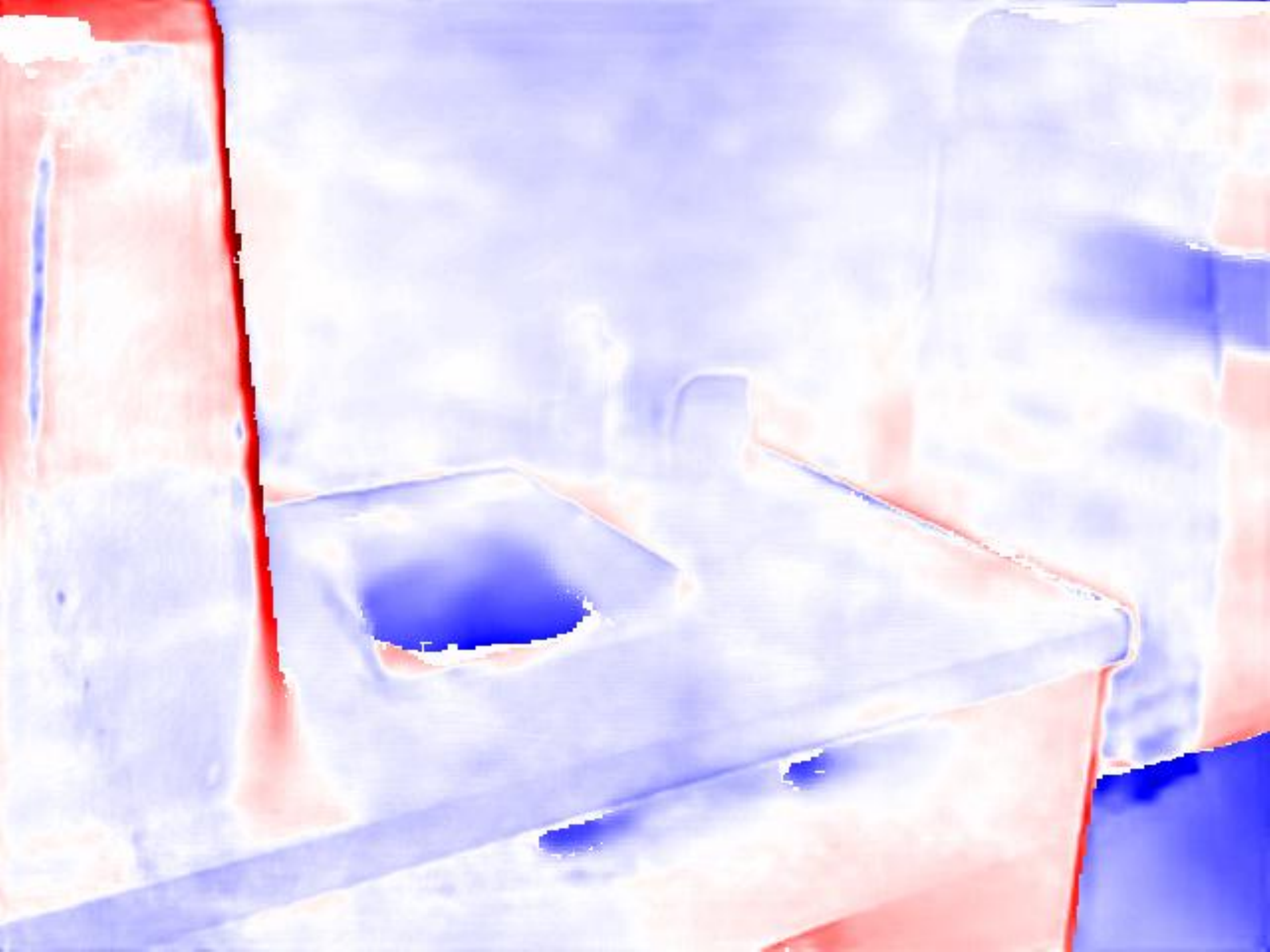}&
    \includegraphics[width=0.024\linewidth]{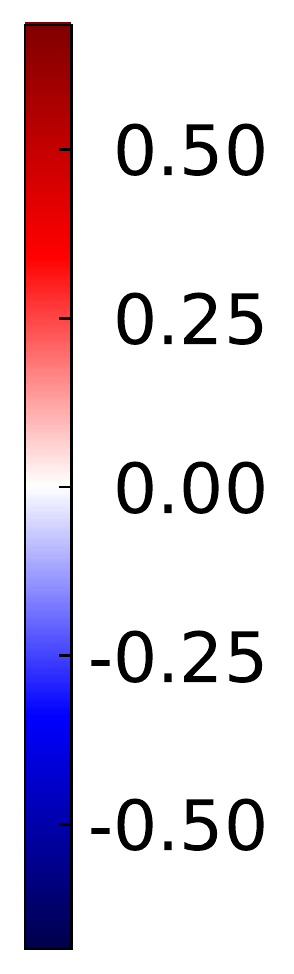}\\
    \vspace{-0.75mm}
    \rot{\scriptsize VINS 50} &
    \includegraphics[width=0.104\linewidth]{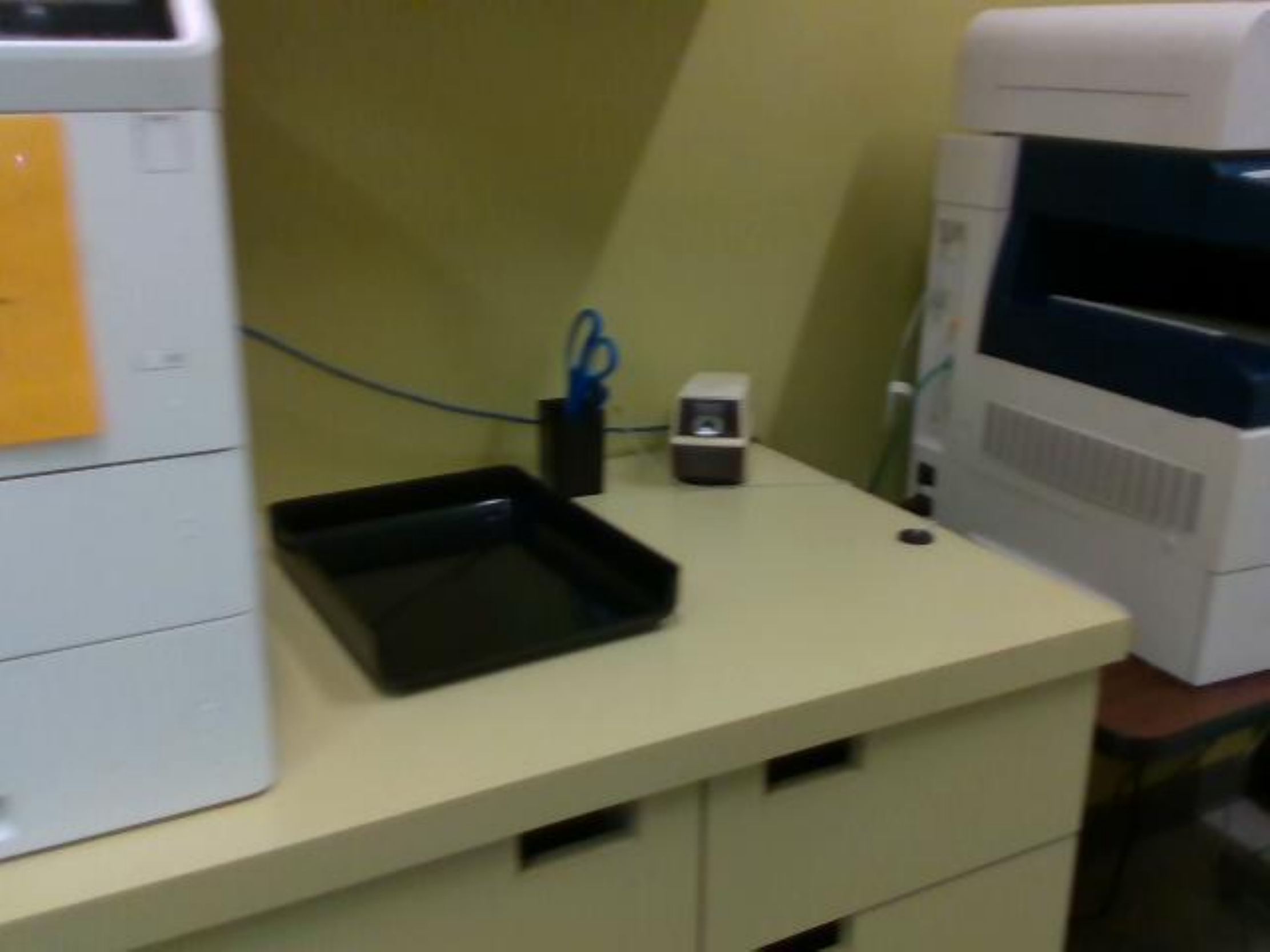}&
    \includegraphics[width=0.104\linewidth]{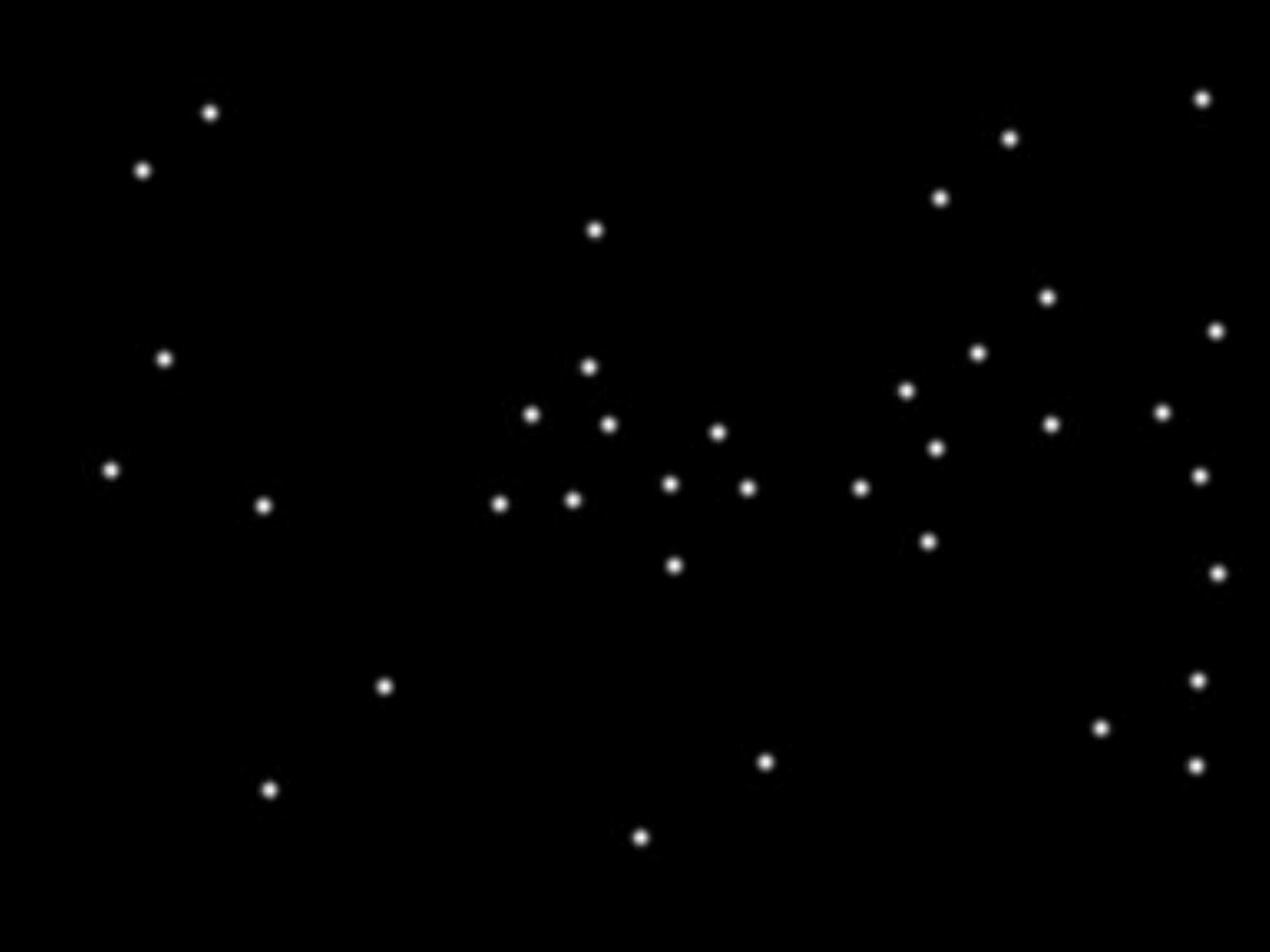}&
    \includegraphics[width=0.104\linewidth]{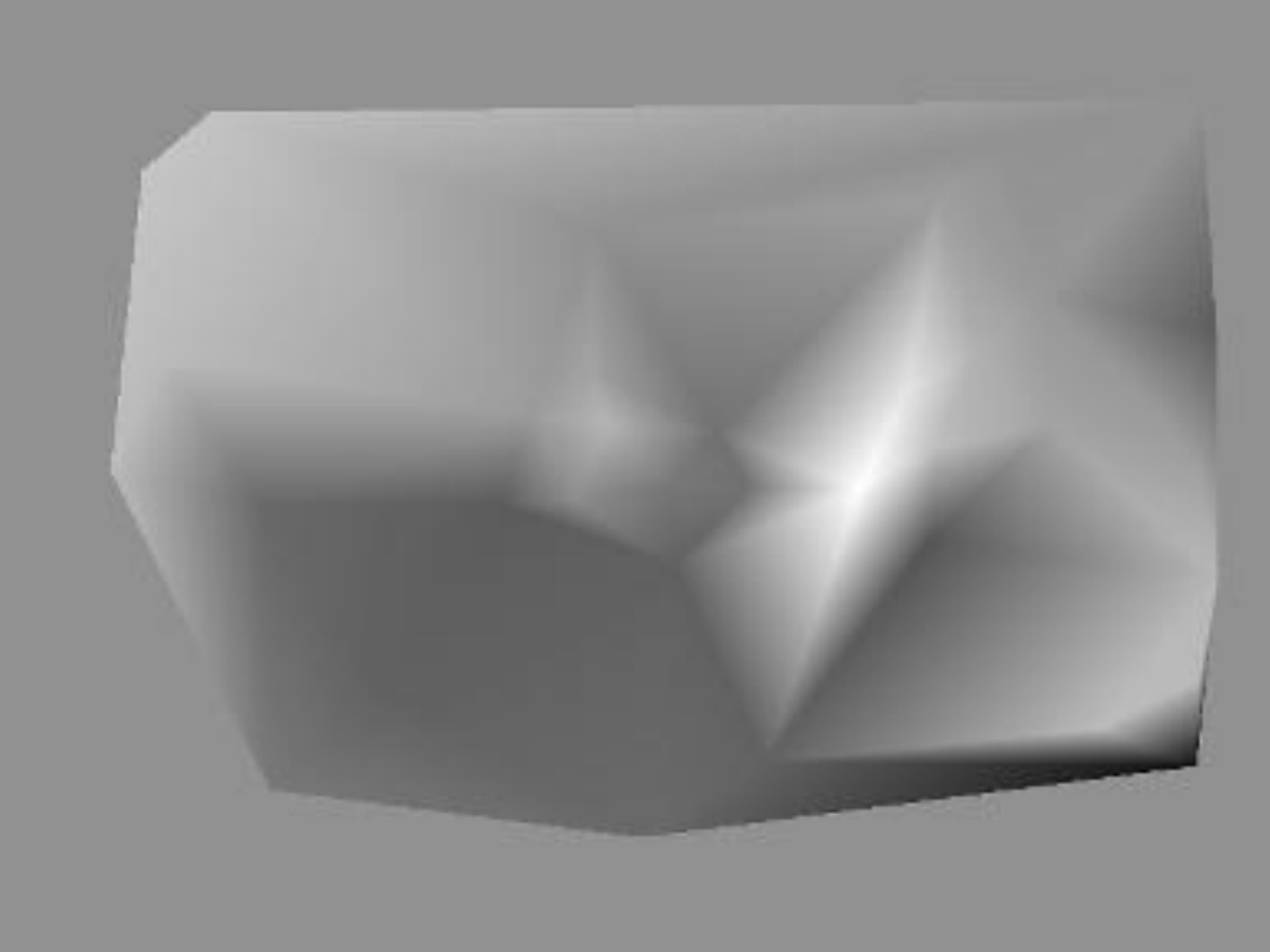}&
    \includegraphics[width=0.104\linewidth]{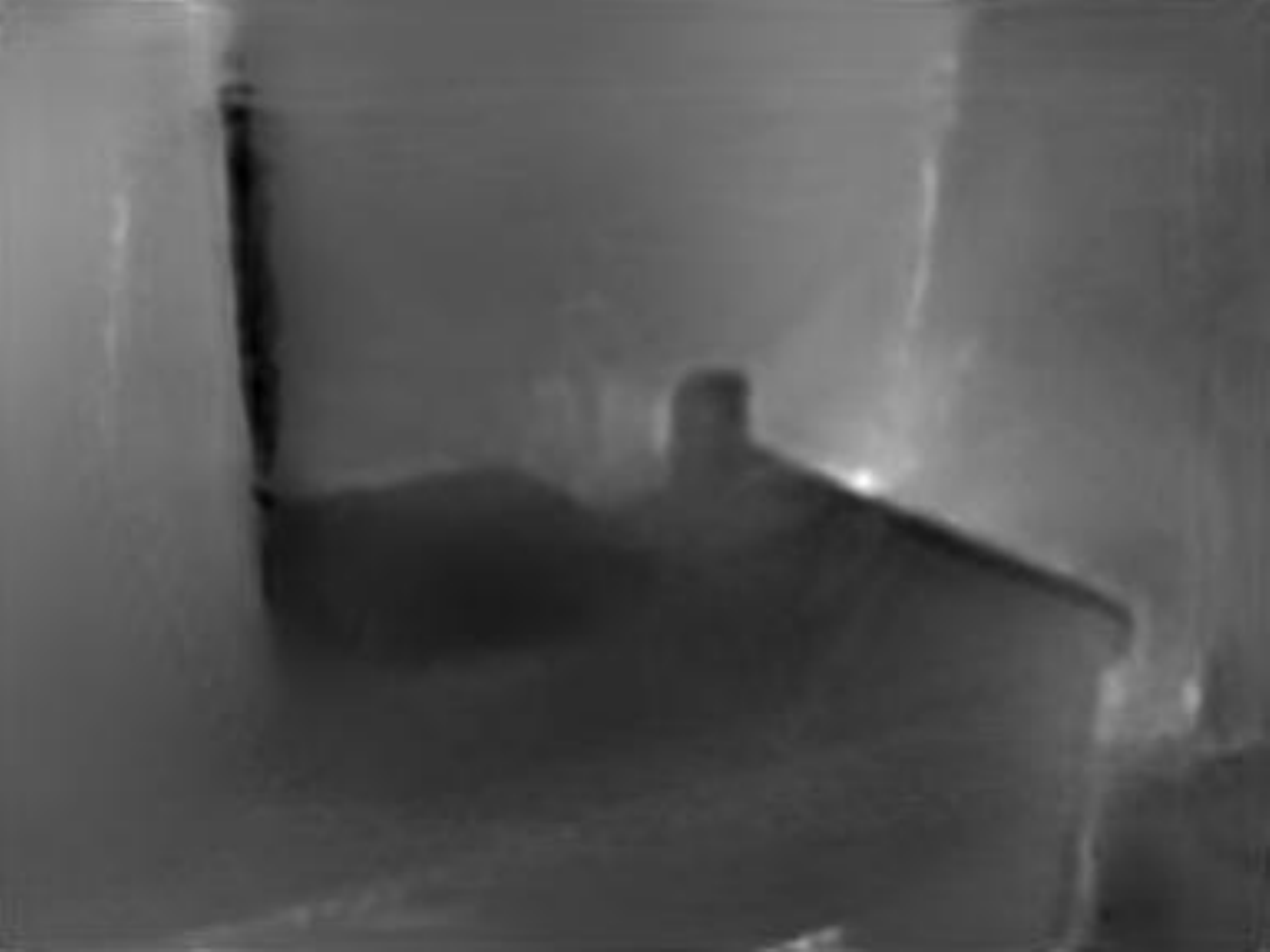}&
    \includegraphics[width=0.104\linewidth]{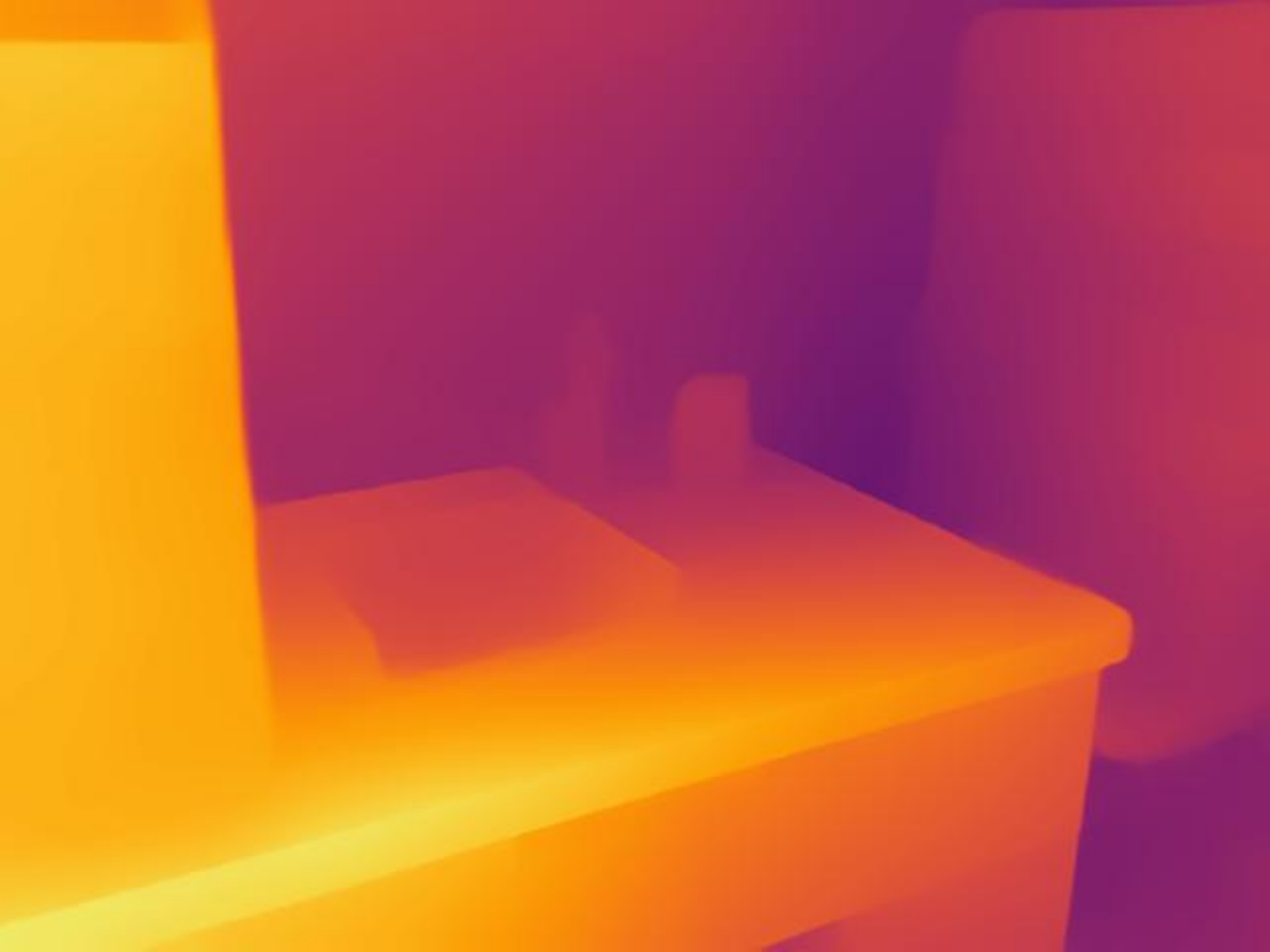}&
    \includegraphics[width=0.104\linewidth]{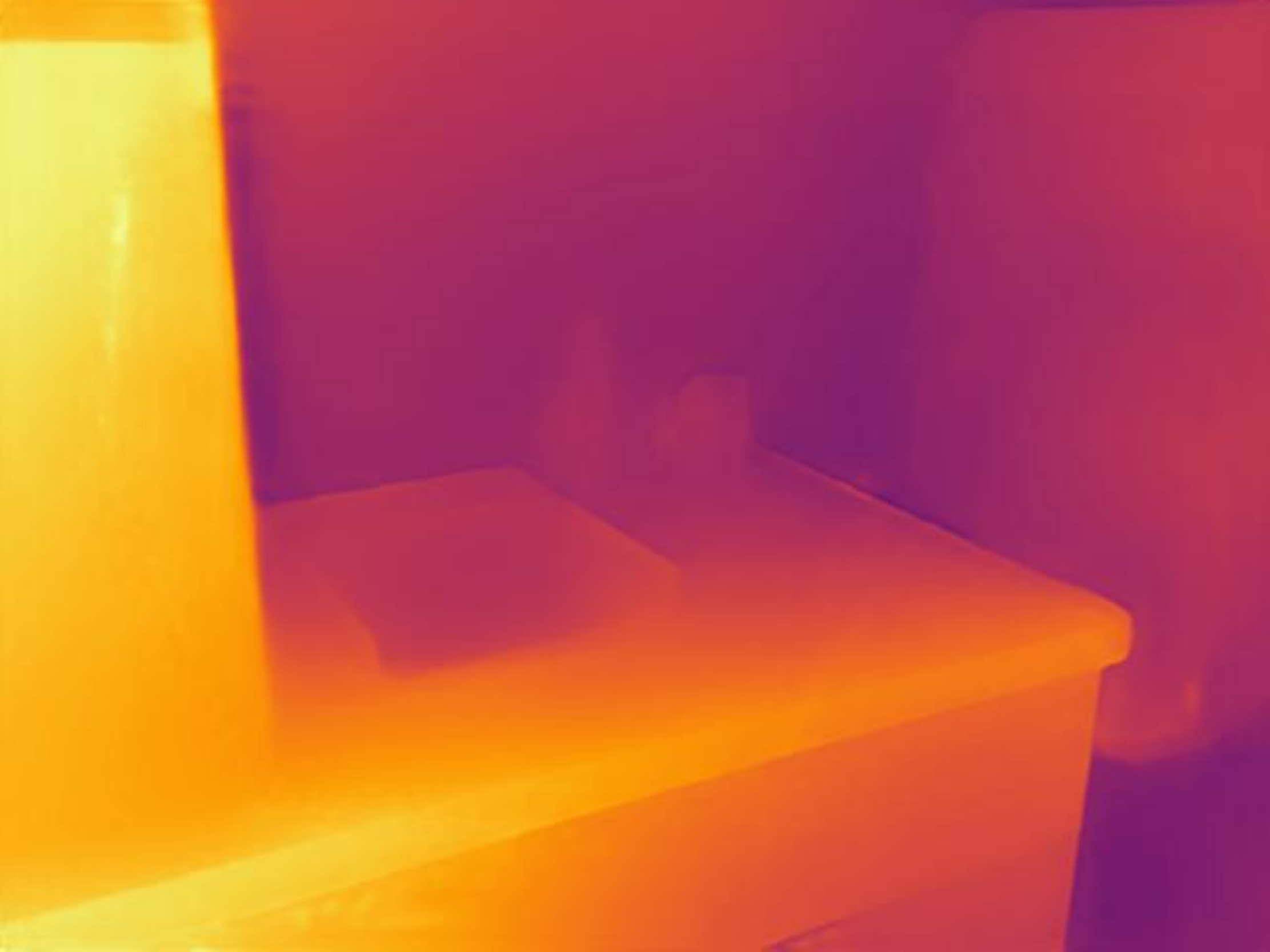}&
    \includegraphics[width=0.104\linewidth]{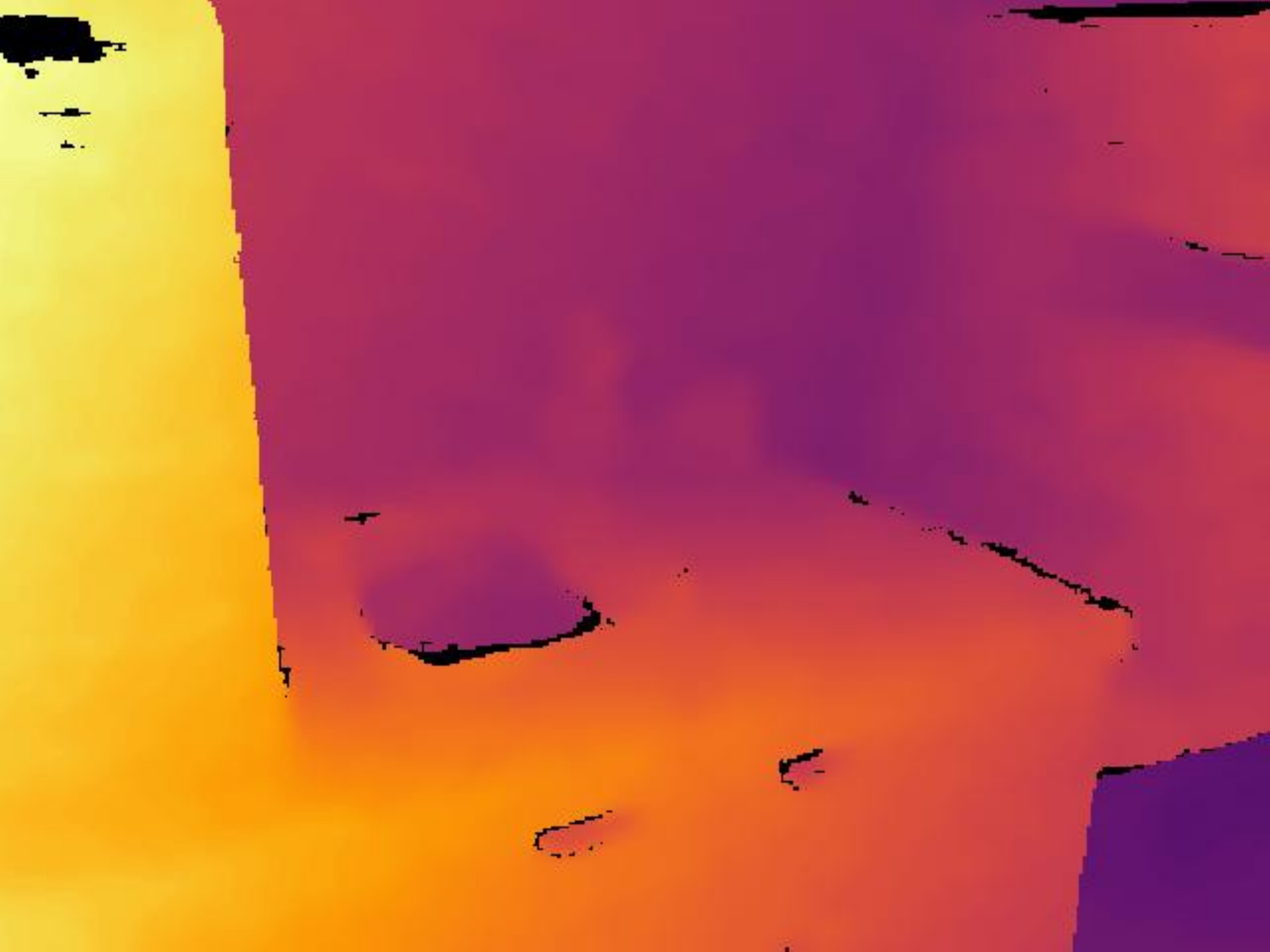}&
    \includegraphics[width=0.104\linewidth]{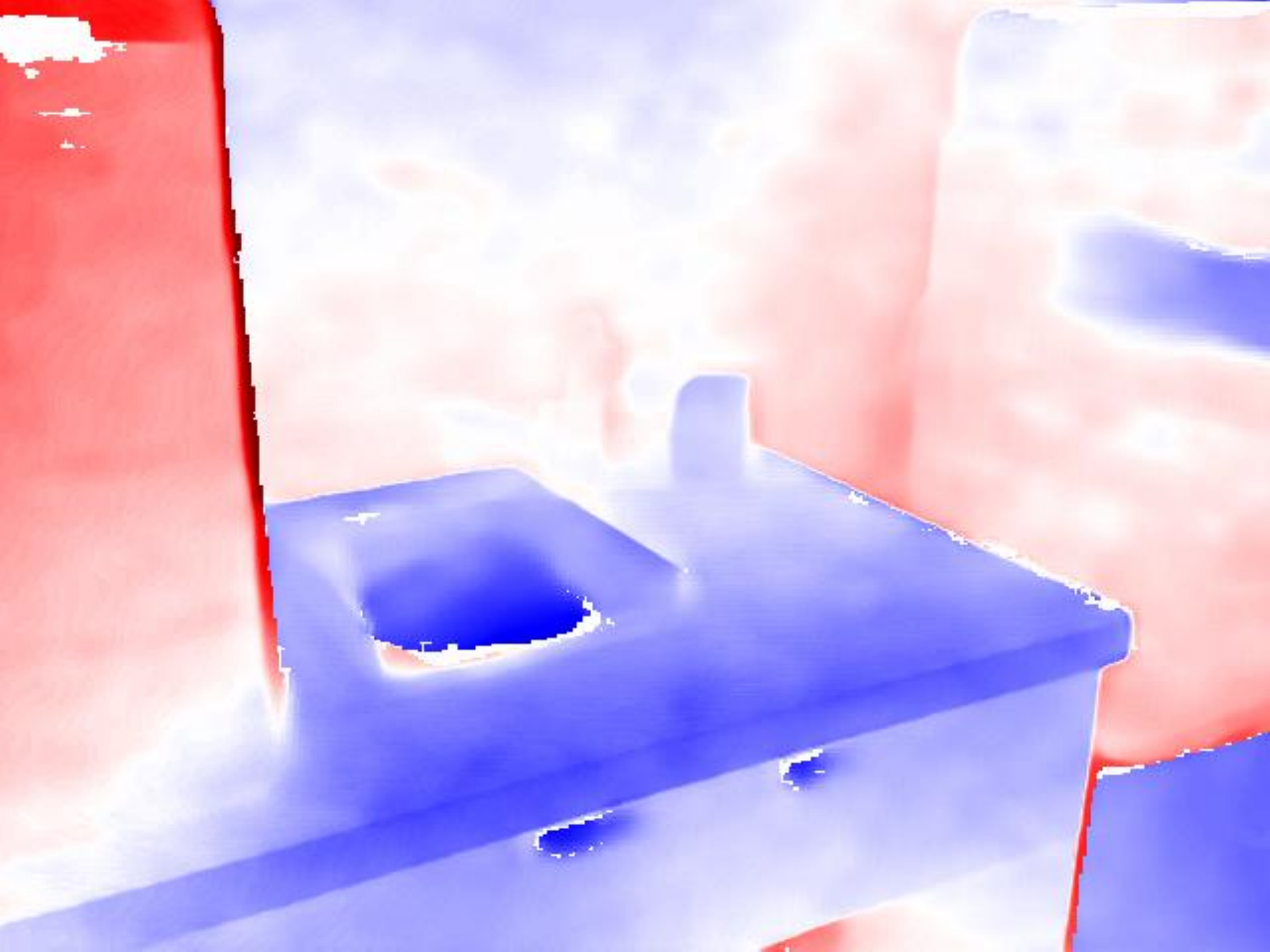}&
    \includegraphics[width=0.104\linewidth]{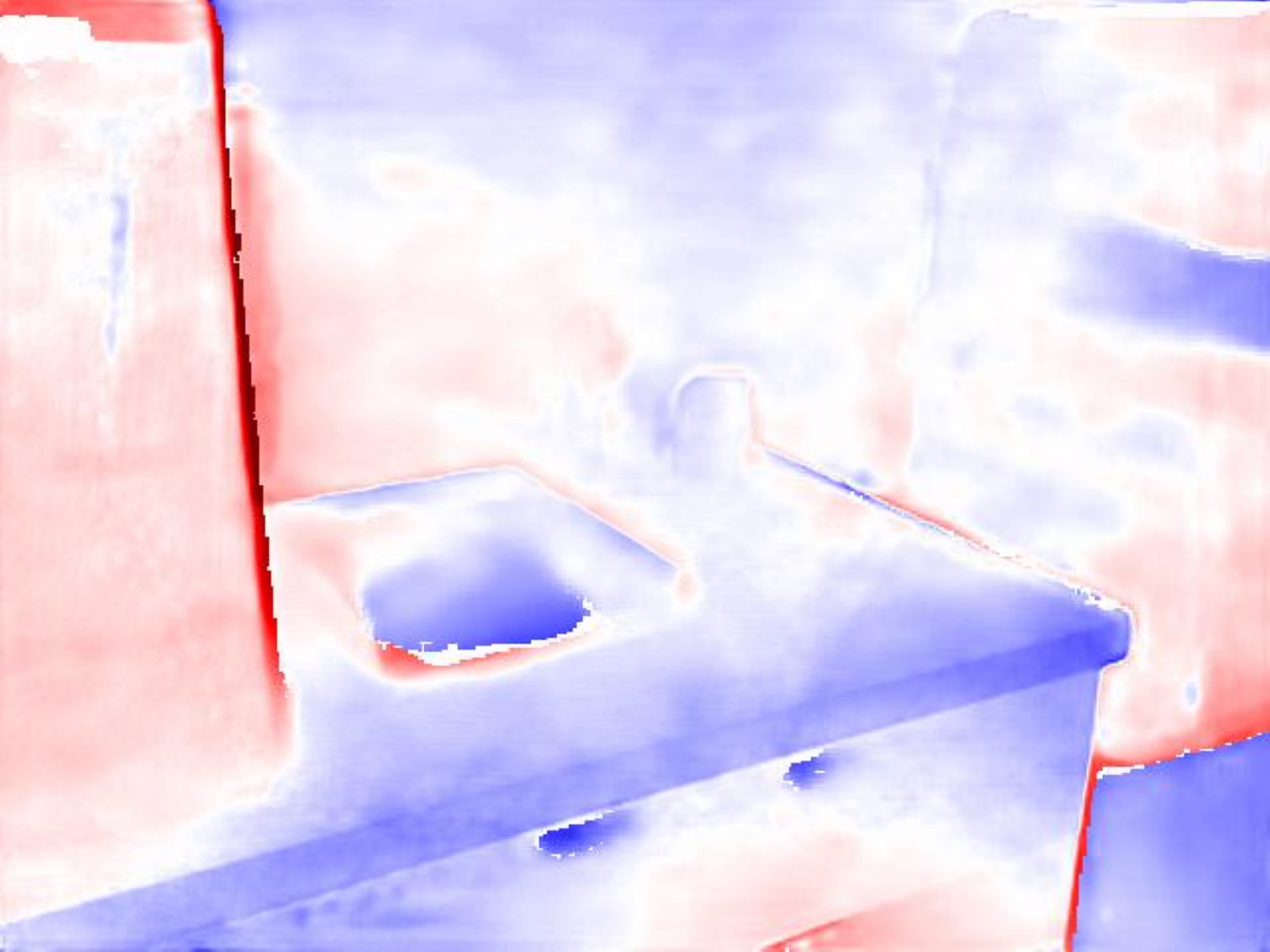}&
    \includegraphics[width=0.024\linewidth]{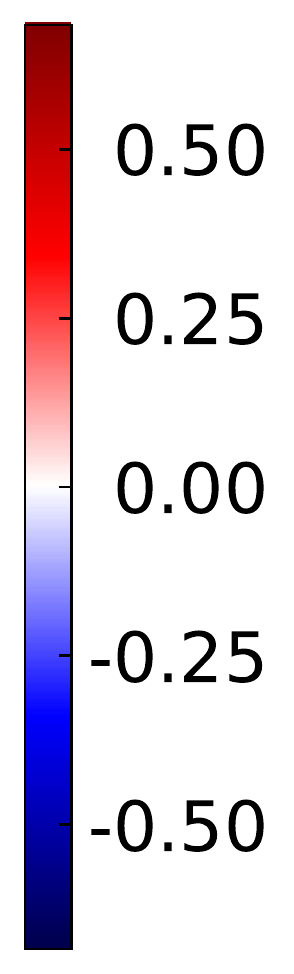}\\
    \multicolumn{9}{c}{} \\
    \vspace{-0.75mm}
    \rot{\scriptsize VOID 150} &
    \includegraphics[width=0.104\linewidth]{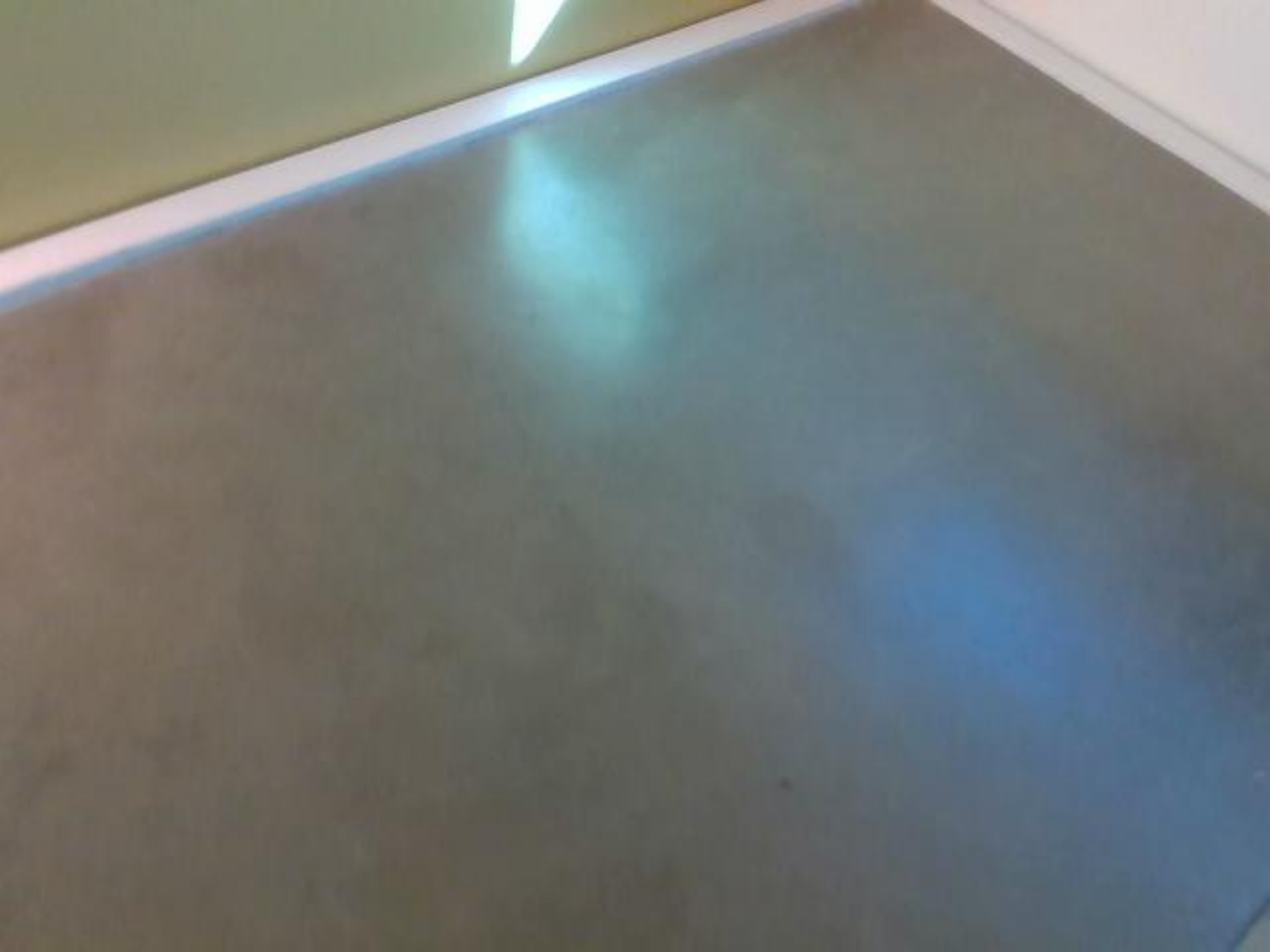}&
    \includegraphics[width=0.104\linewidth]{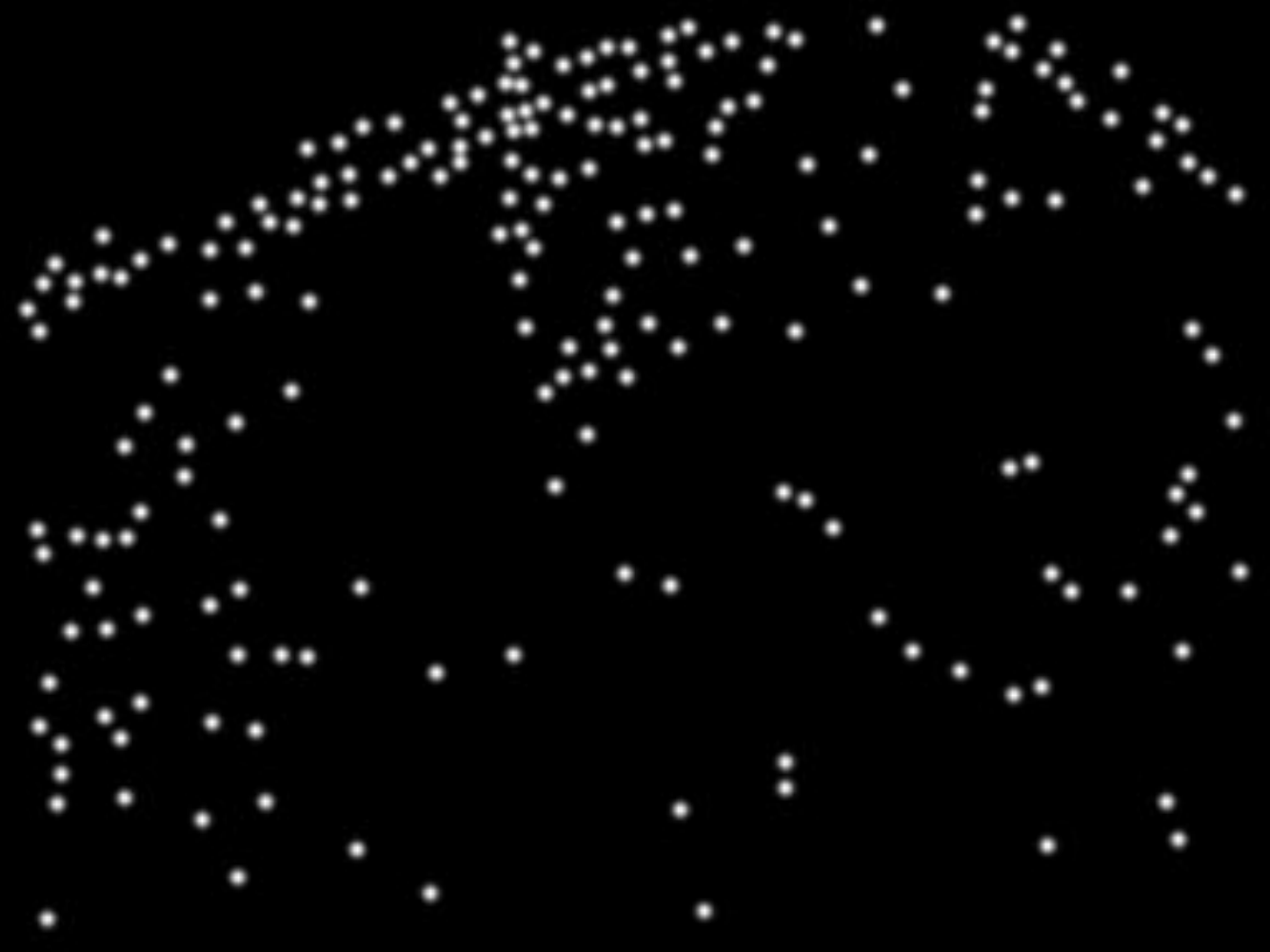}&
    \includegraphics[width=0.104\linewidth]{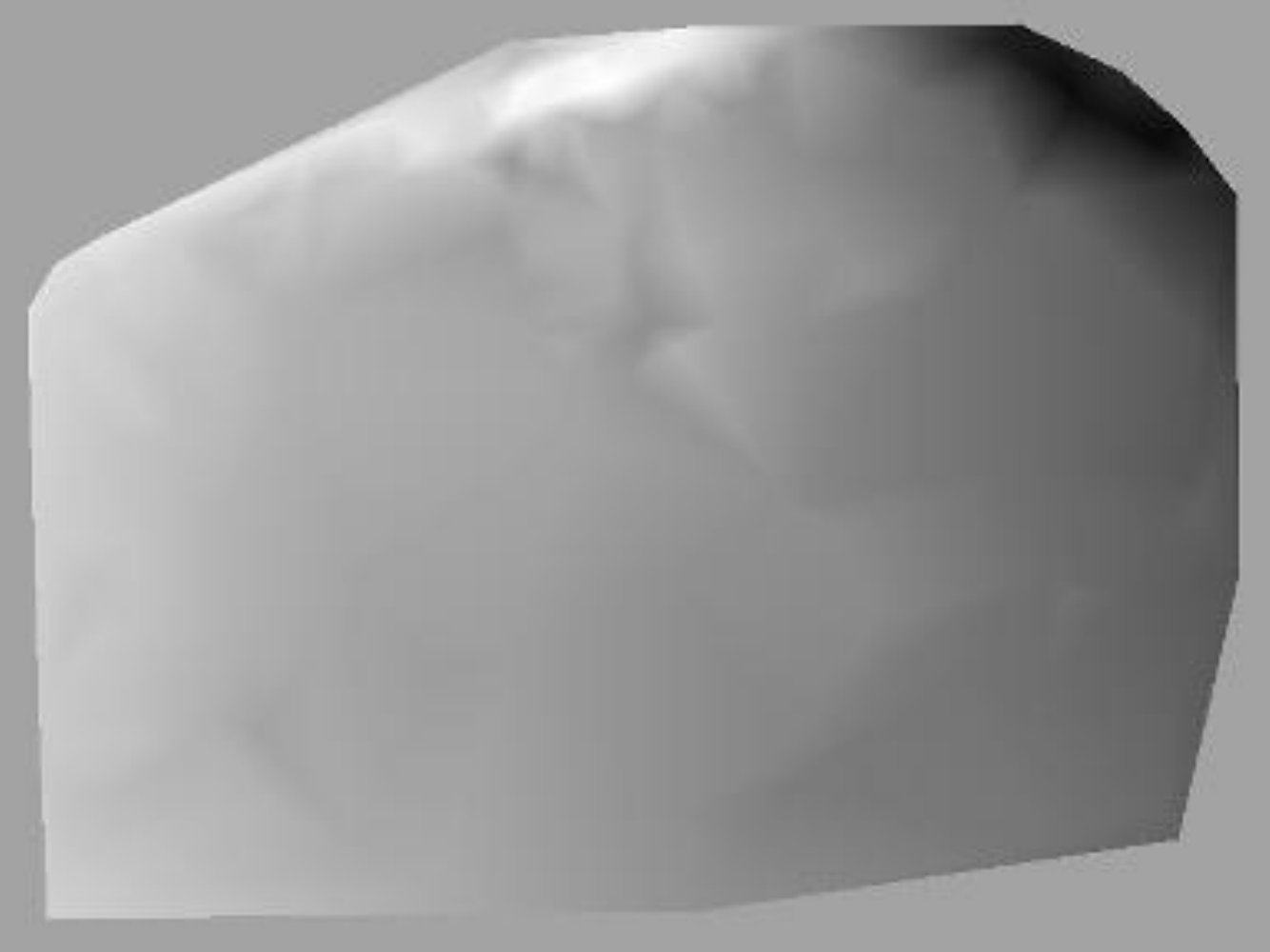}&
    \includegraphics[width=0.104\linewidth]{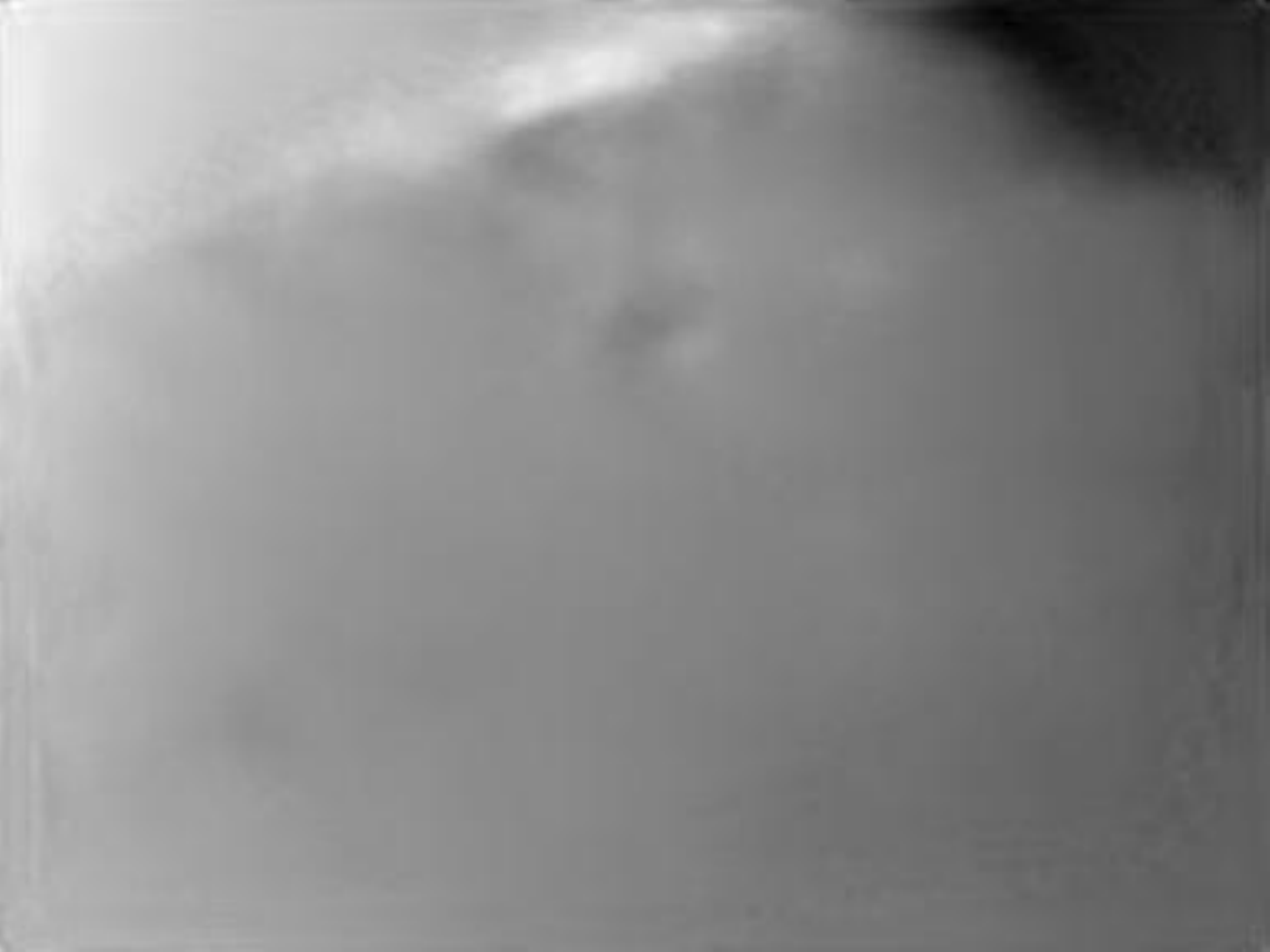}&
    \includegraphics[width=0.104\linewidth]{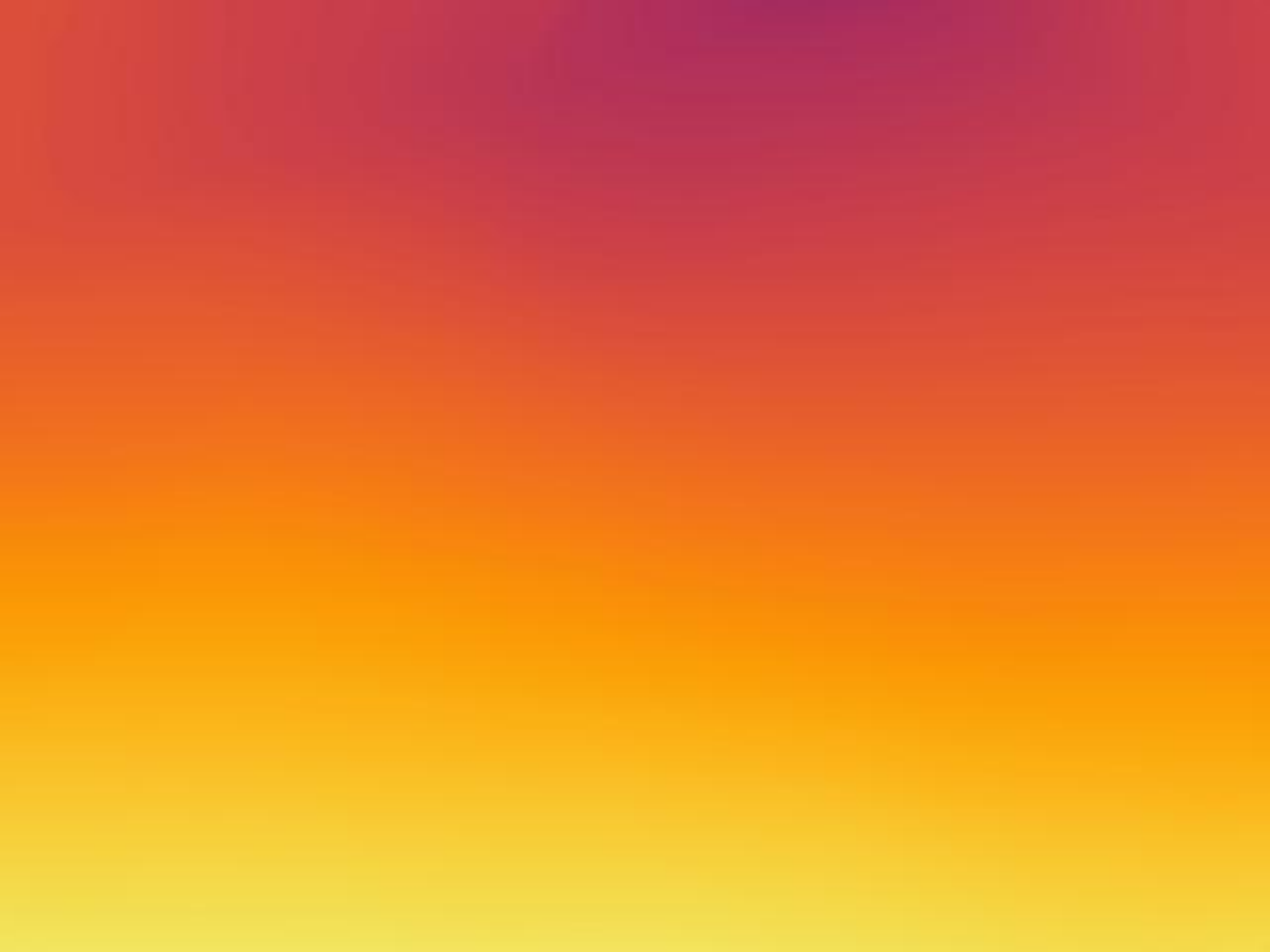}&
    \includegraphics[width=0.104\linewidth]{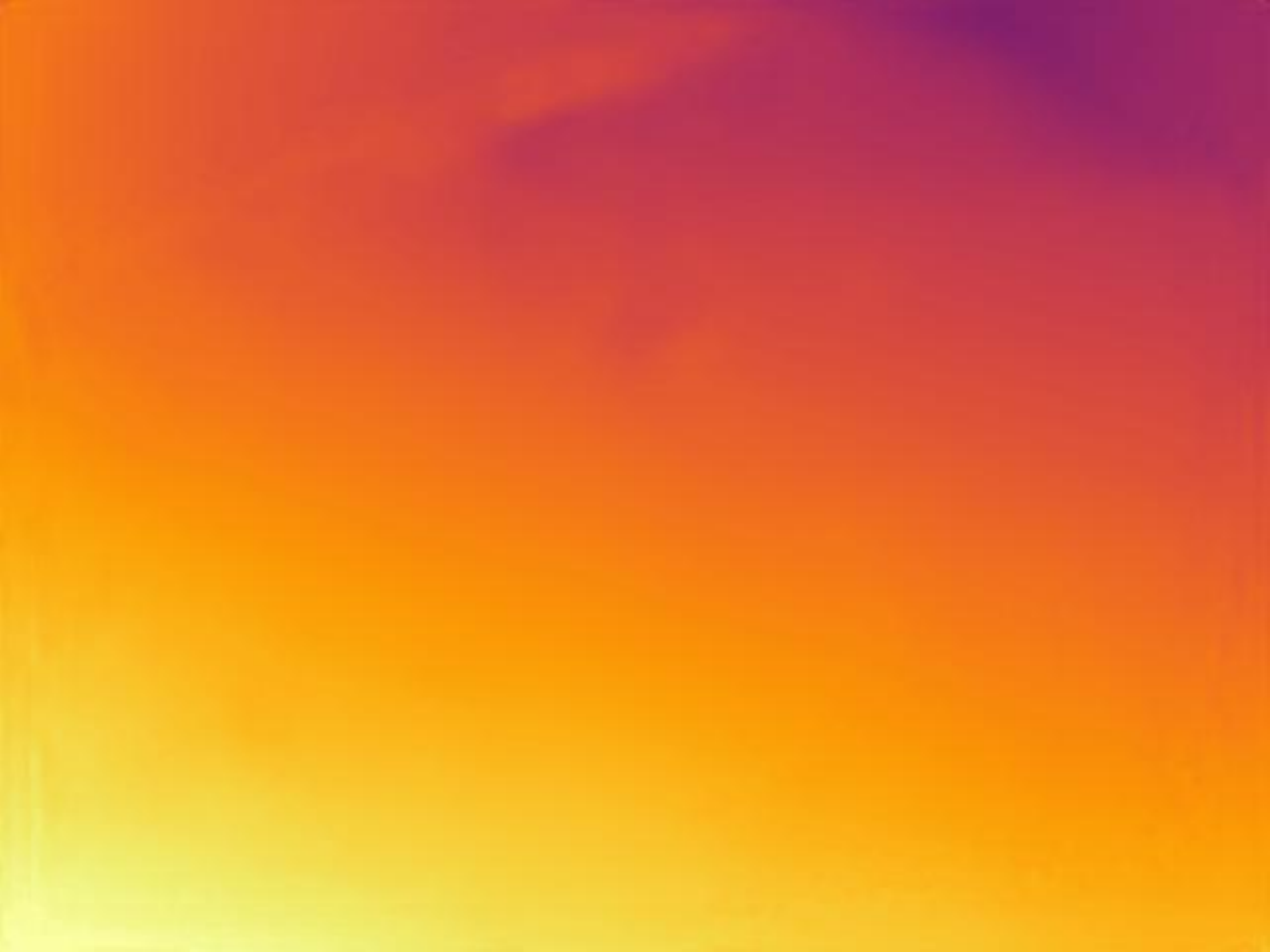}&
    \includegraphics[width=0.104\linewidth]{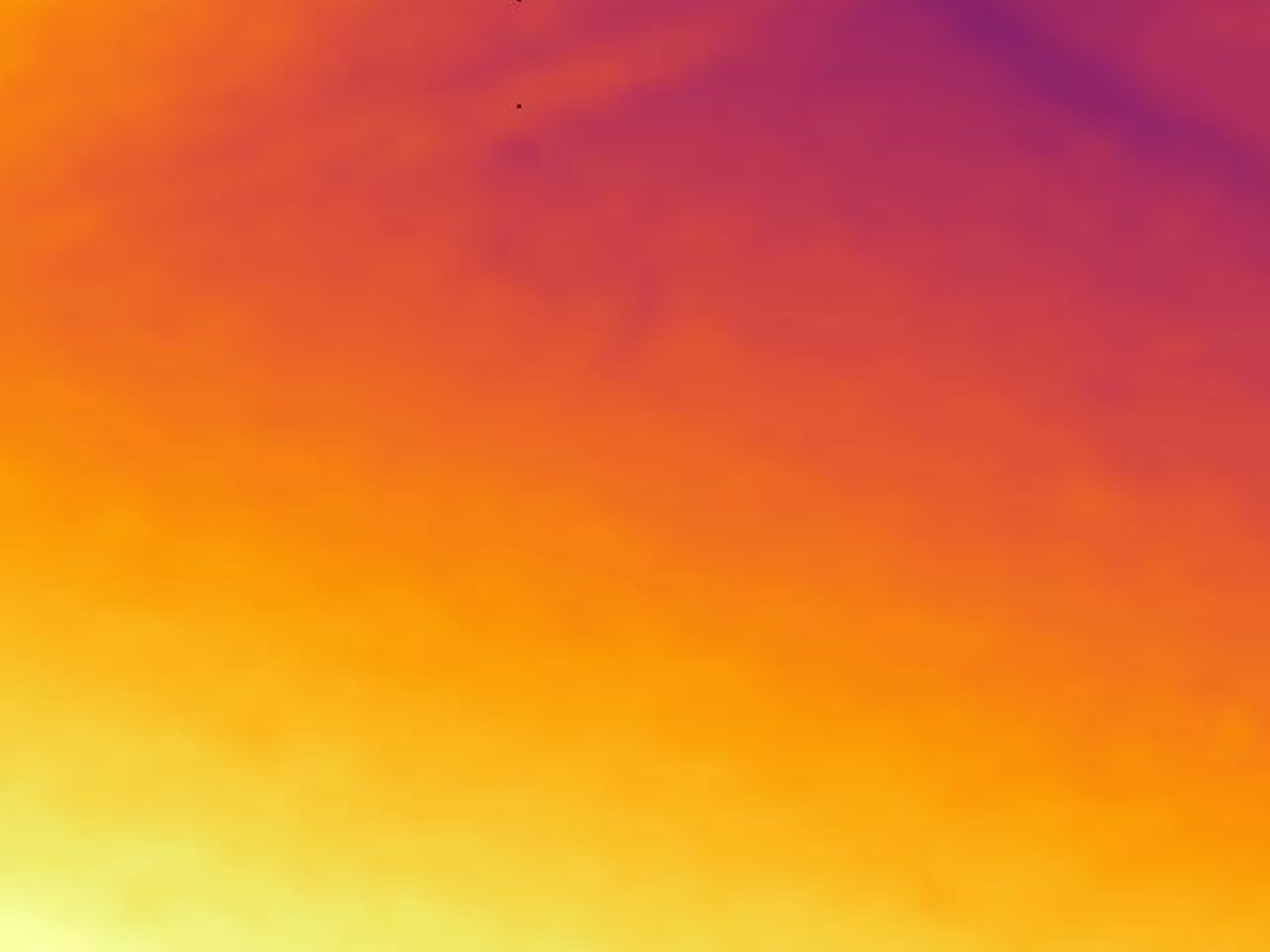}&
    \includegraphics[width=0.104\linewidth]{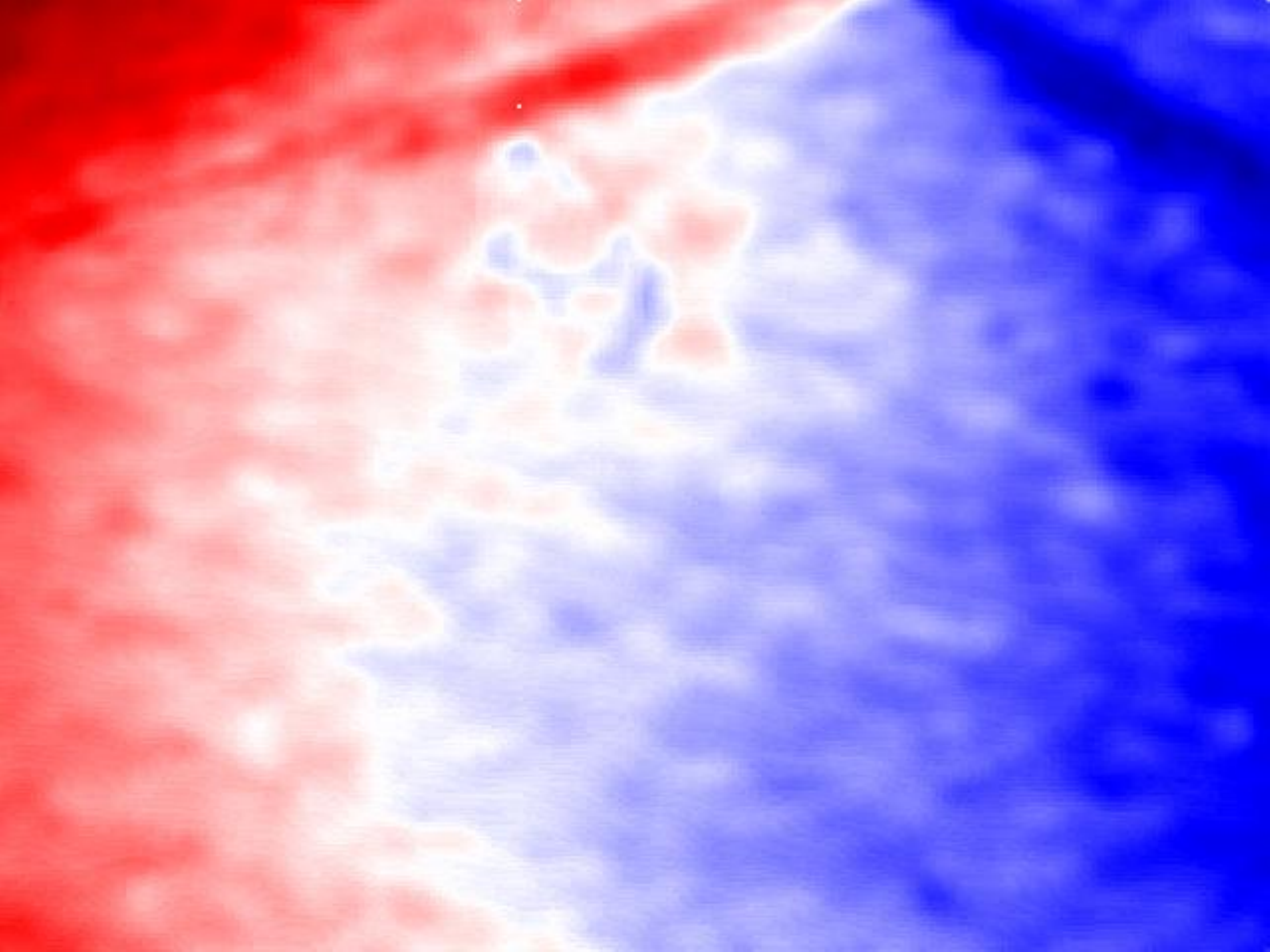}&
    \includegraphics[width=0.104\linewidth]{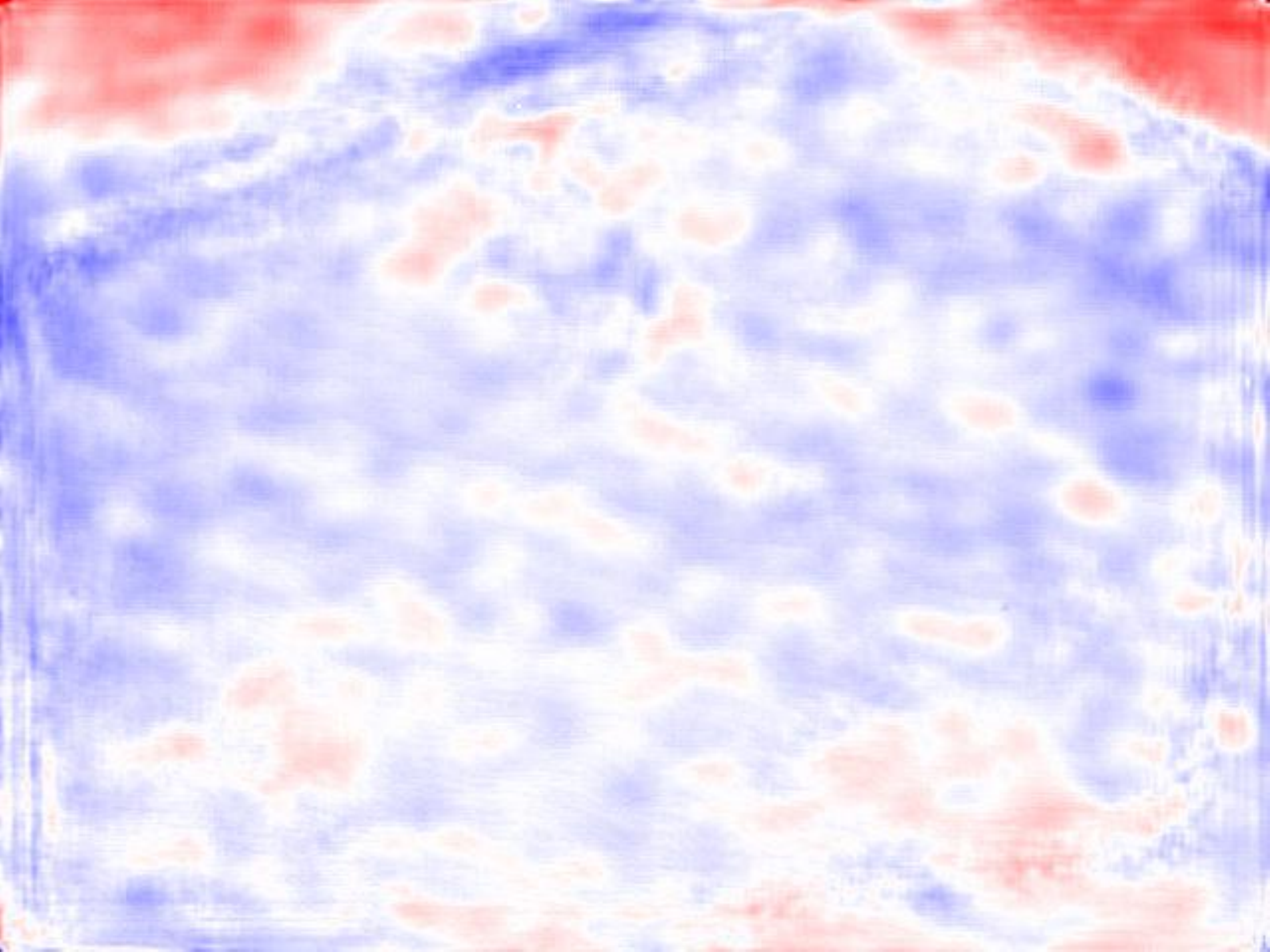}&
    \includegraphics[width=0.024\linewidth]{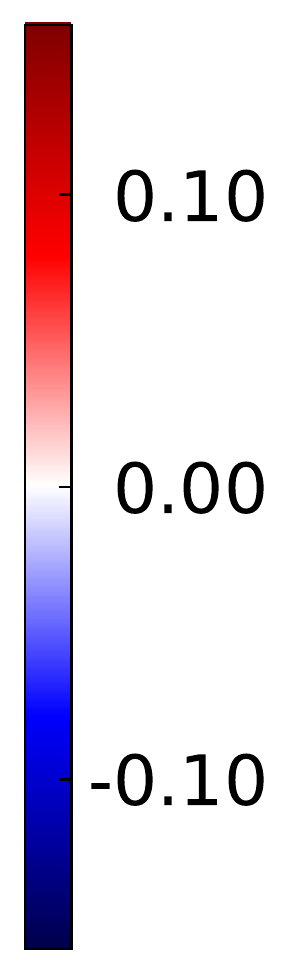}\\
    \vspace{-0.75mm}
    \rot{\scriptsize VINS 150} &
    \includegraphics[width=0.104\linewidth]{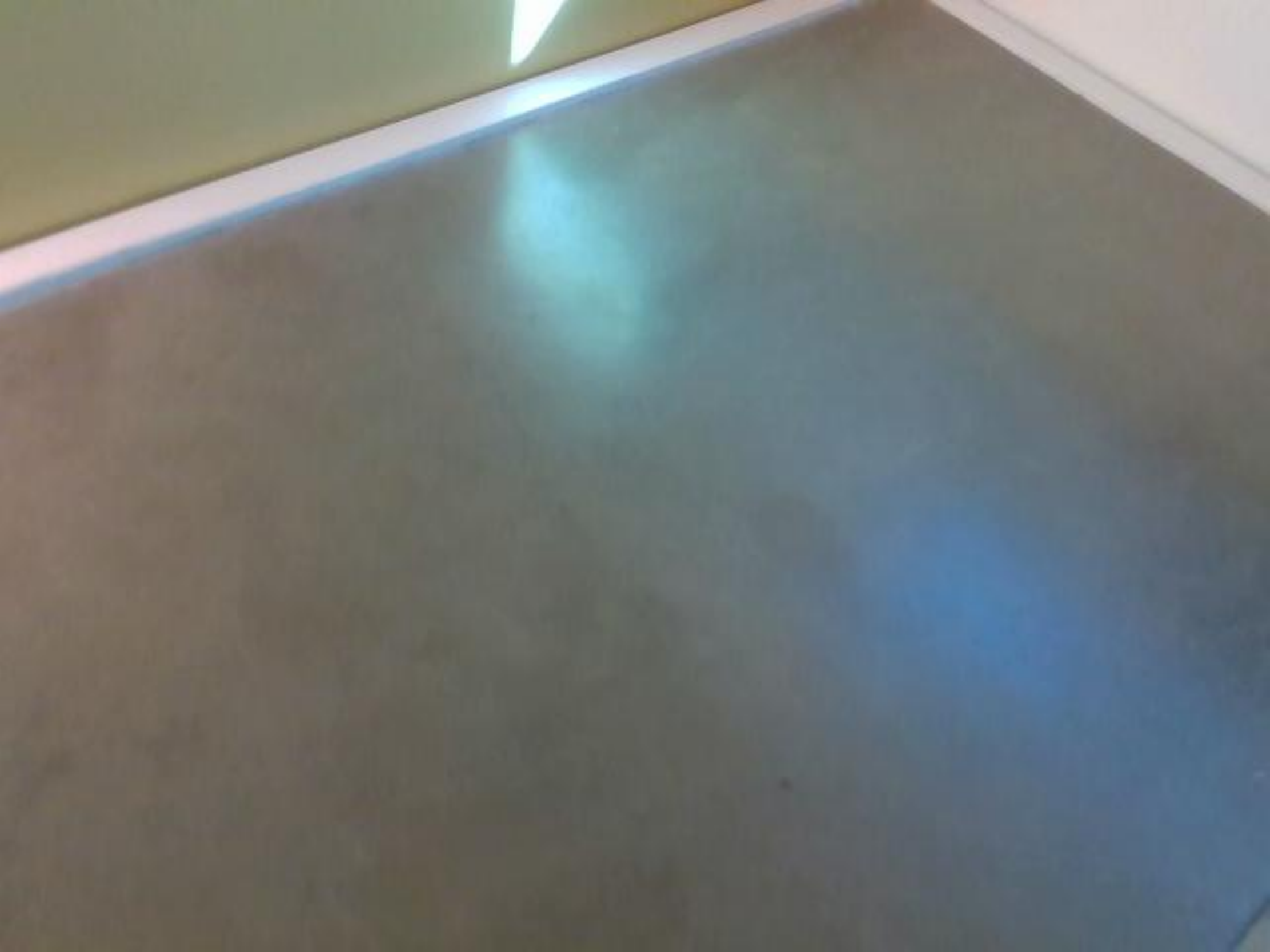}&
    \includegraphics[width=0.104\linewidth]{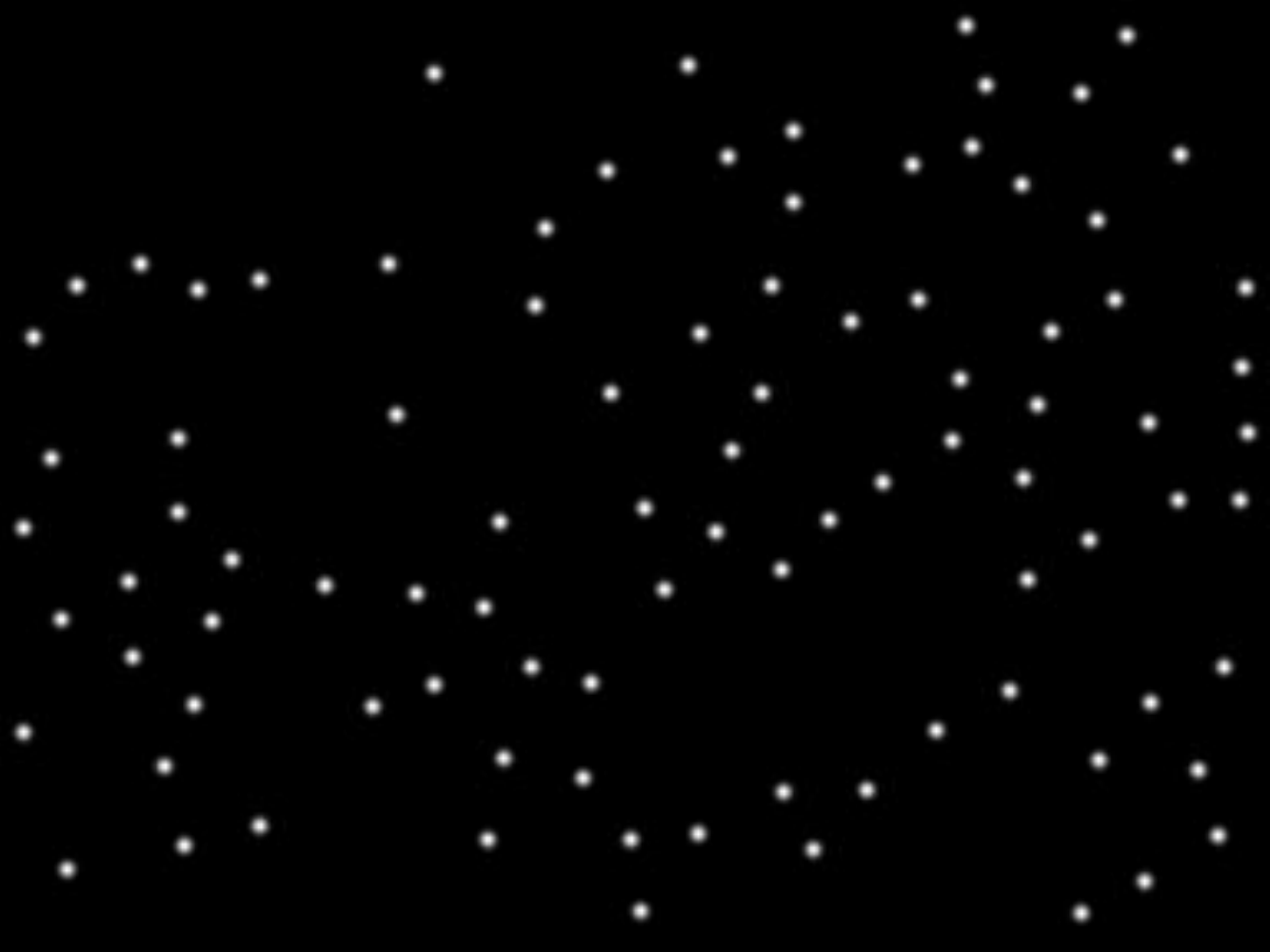}&
    \includegraphics[width=0.104\linewidth]{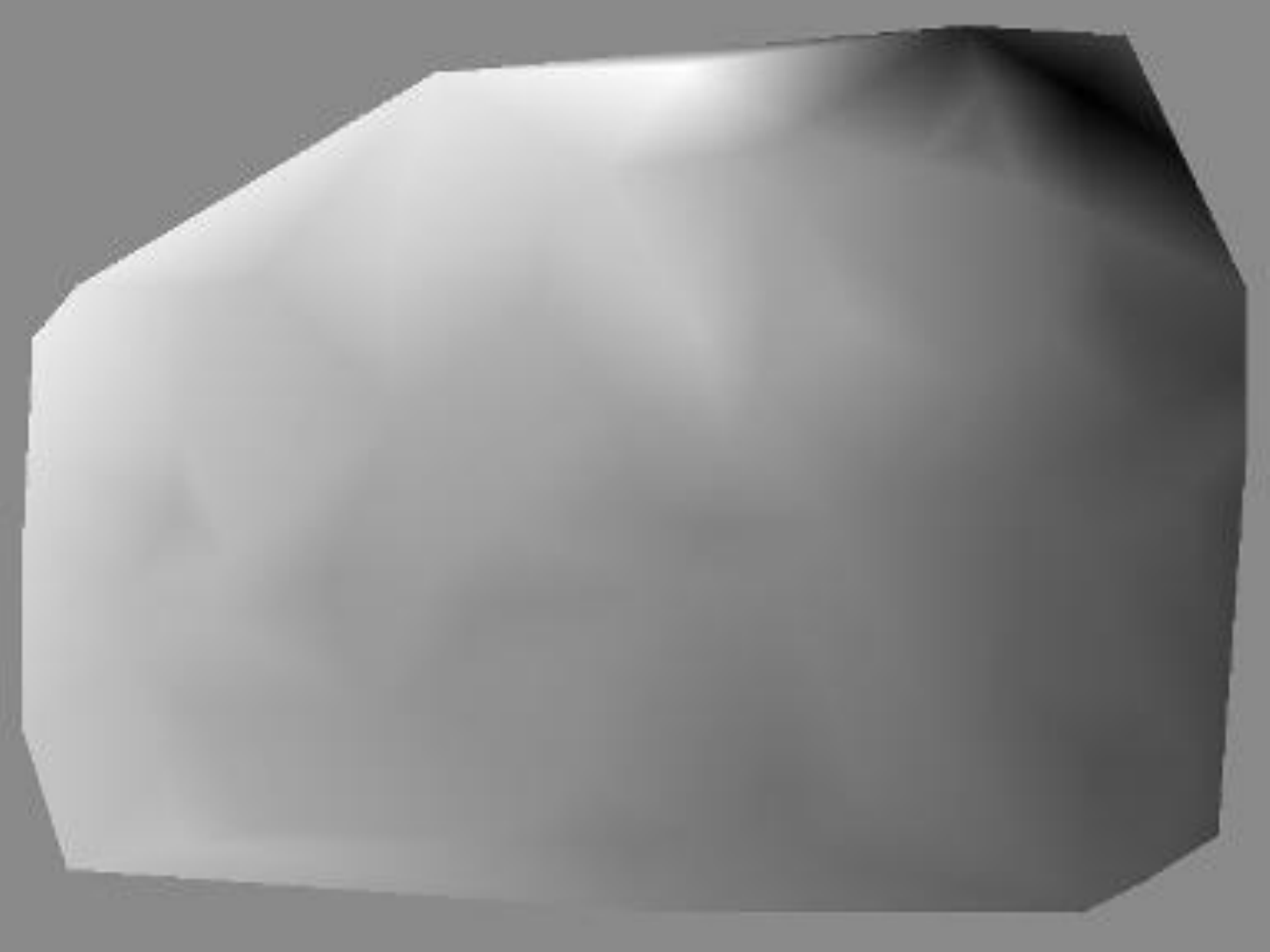}&
    \includegraphics[width=0.104\linewidth]{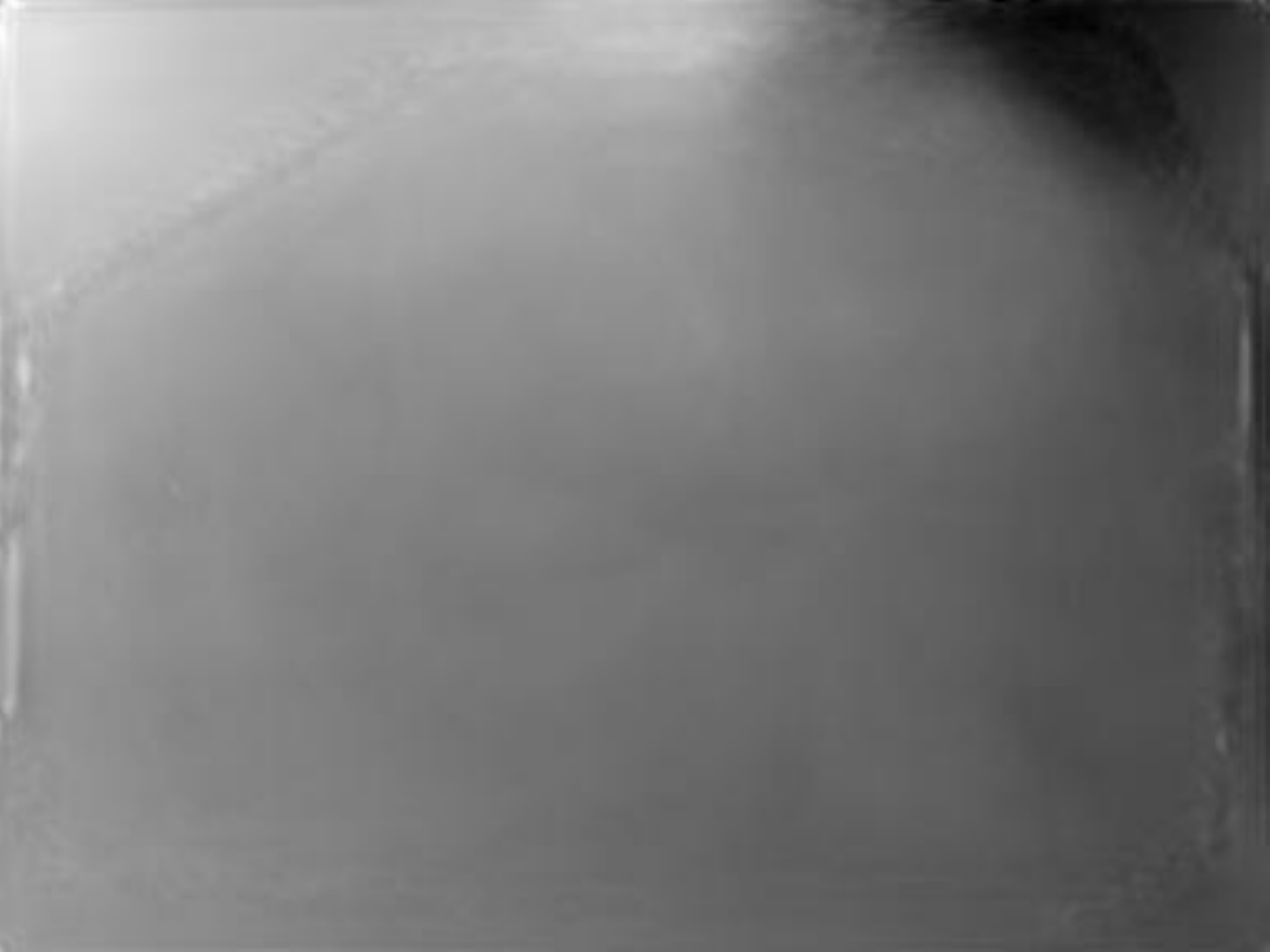}&
    \includegraphics[width=0.104\linewidth]{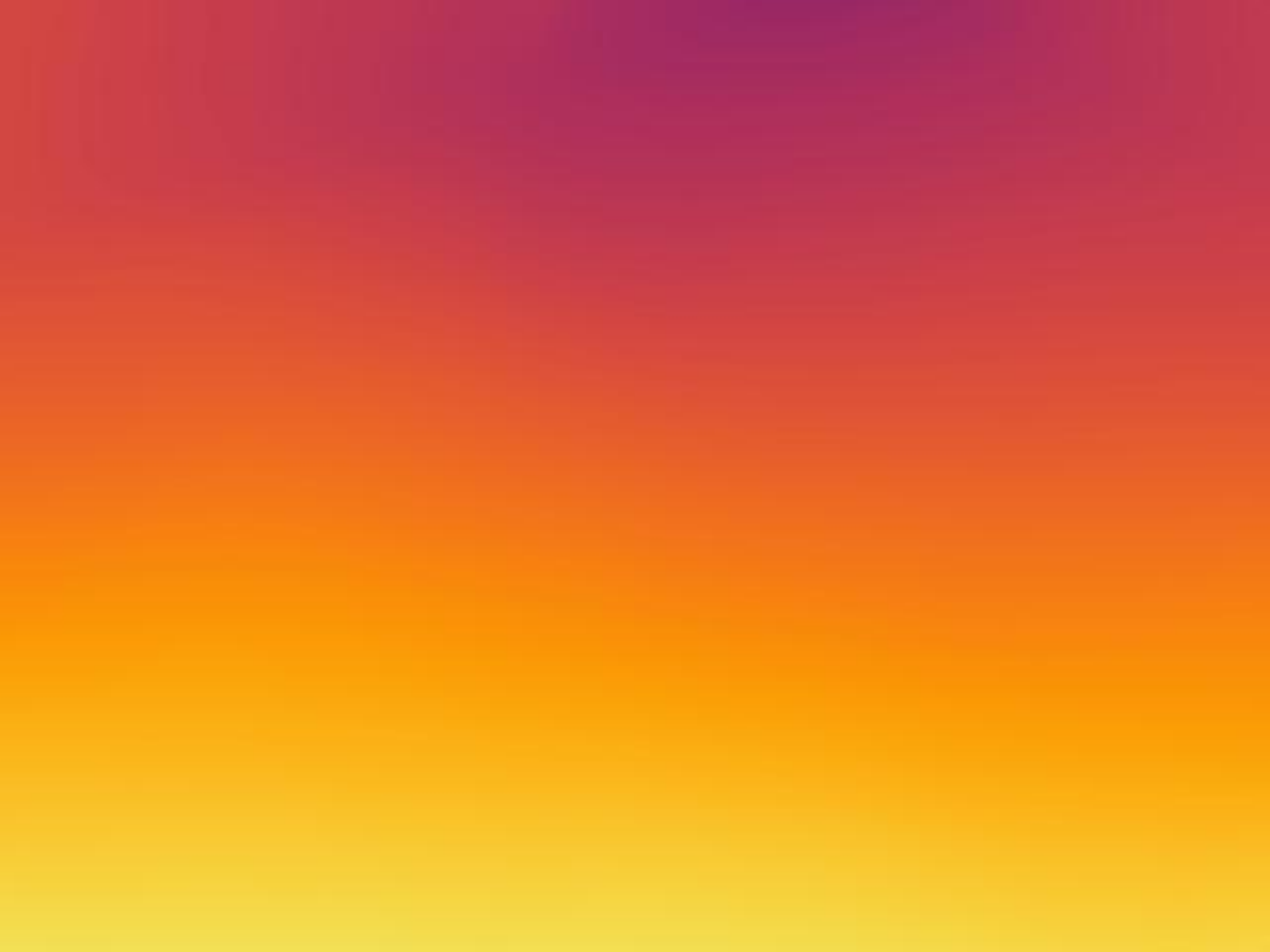}&
    \includegraphics[width=0.104\linewidth]{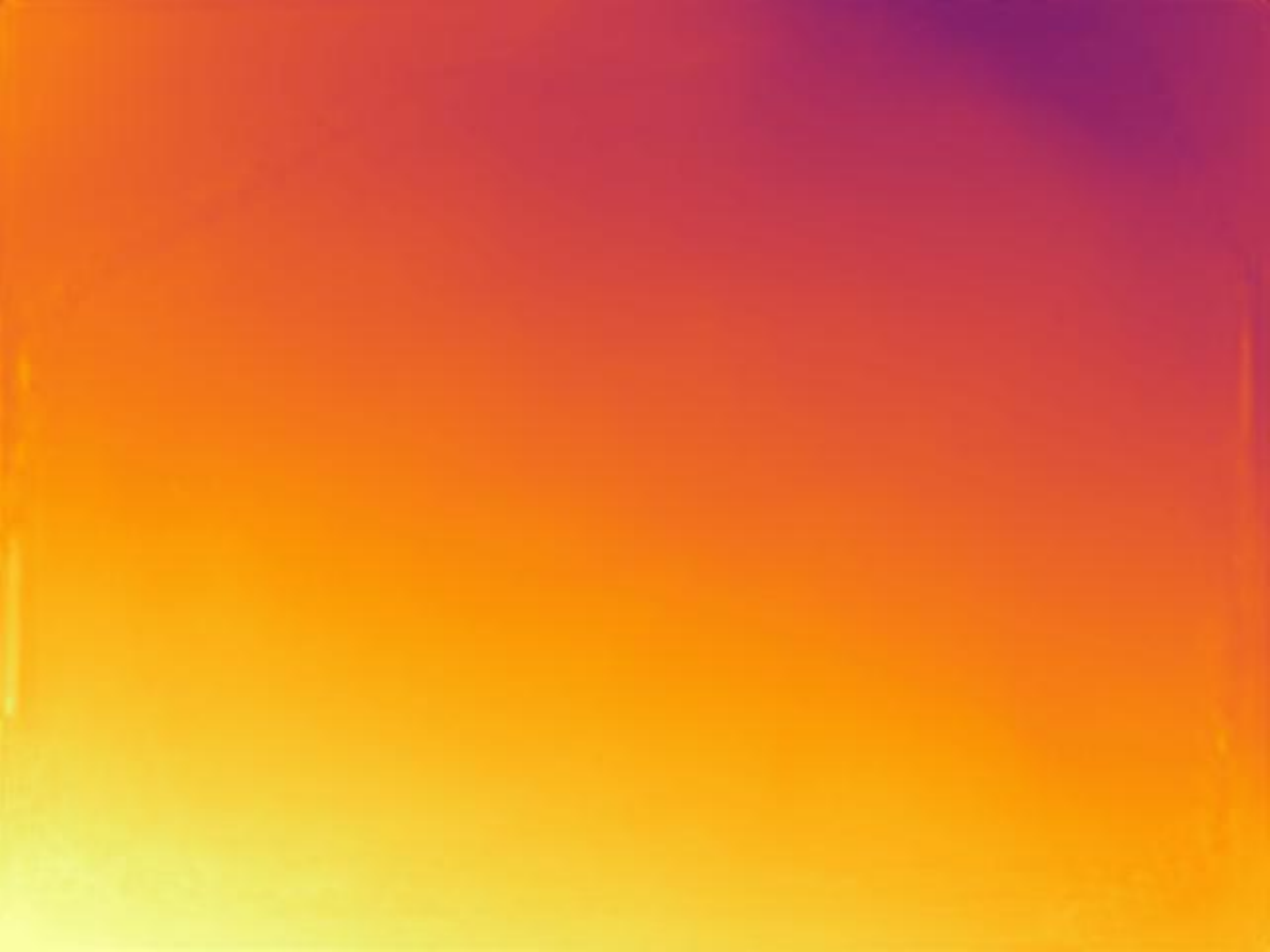}&
    \includegraphics[width=0.104\linewidth]{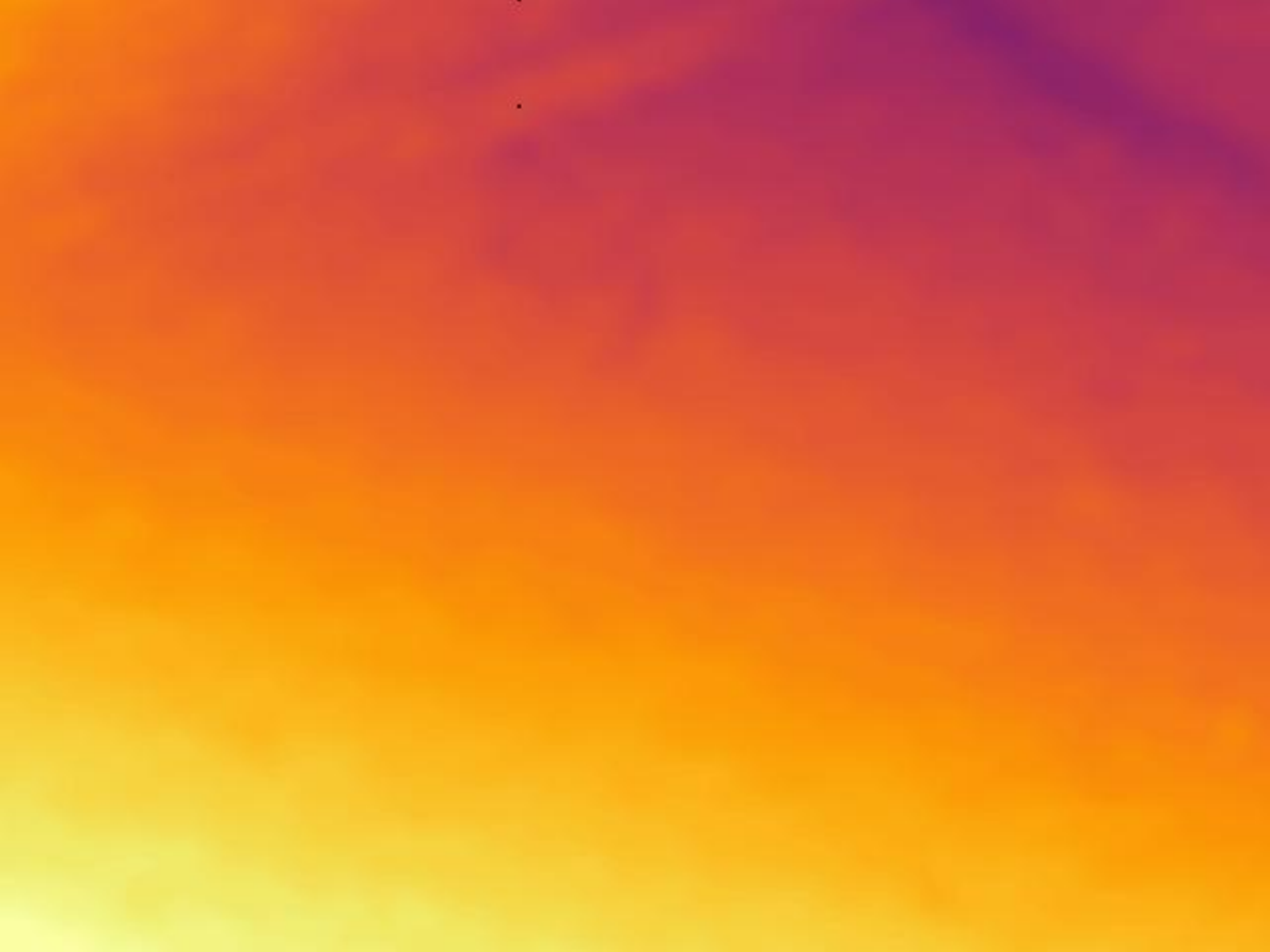}&
    \includegraphics[width=0.104\linewidth]{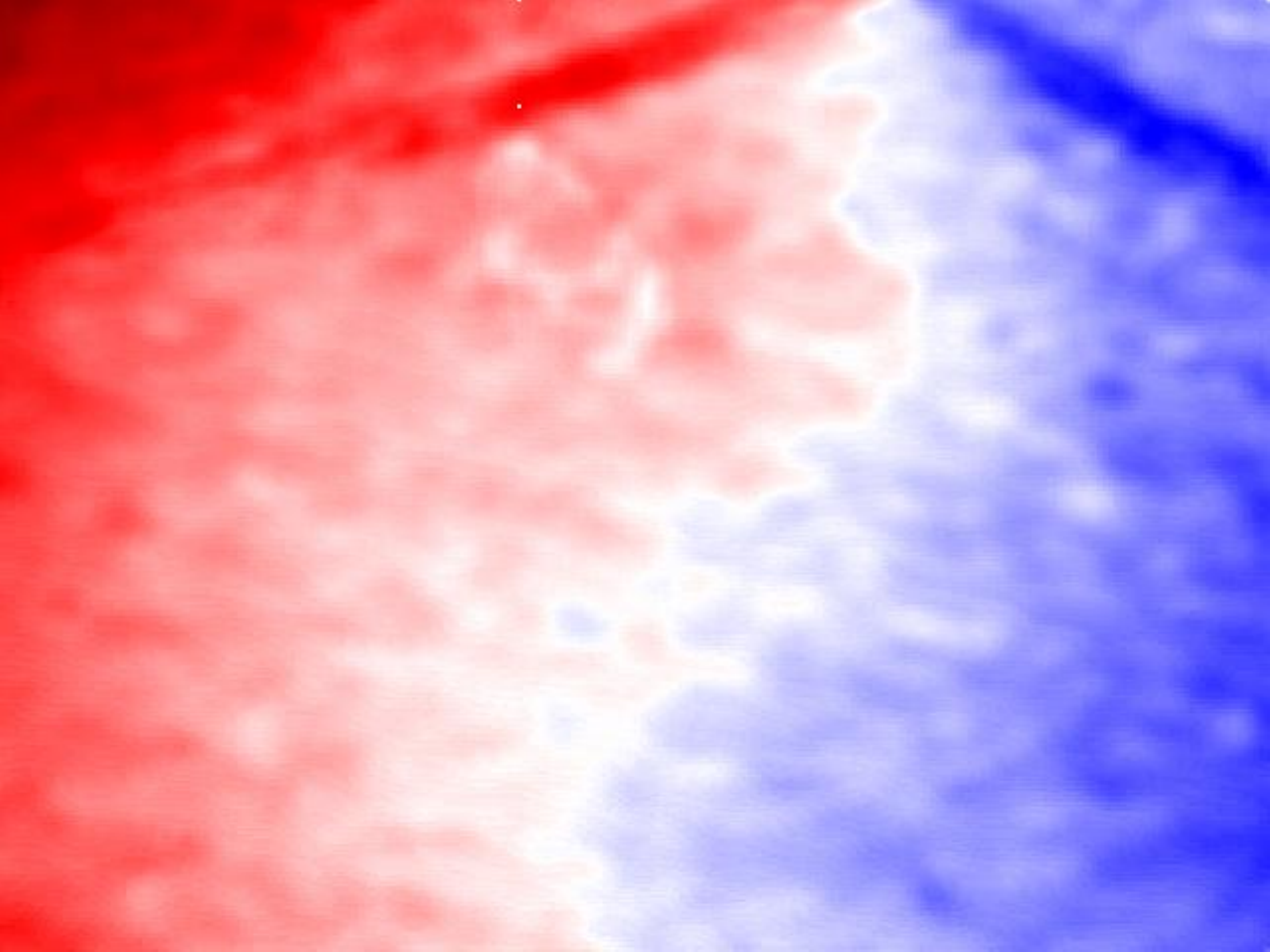}&
    \includegraphics[width=0.104\linewidth]{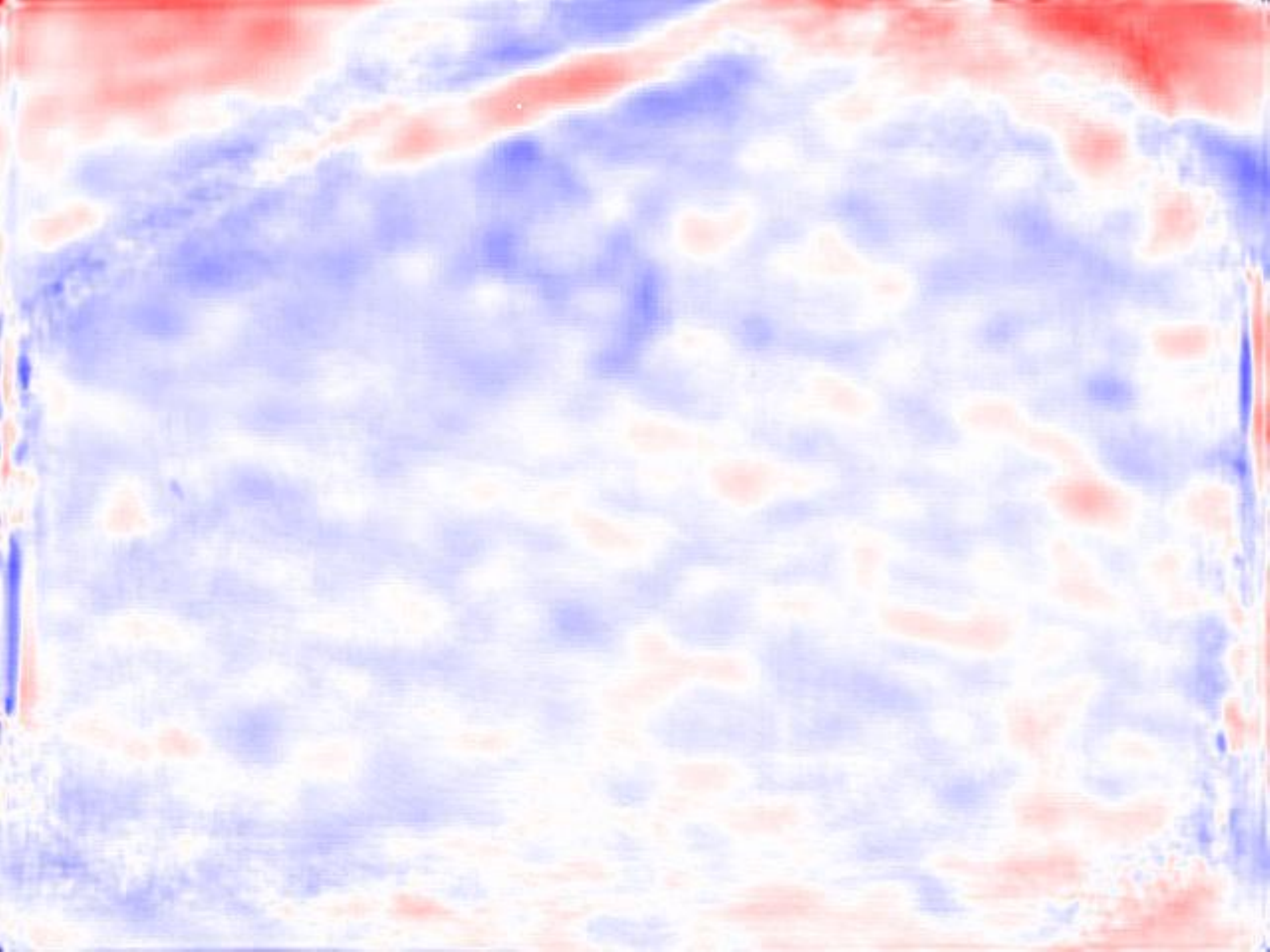}&
    \includegraphics[width=0.024\linewidth]{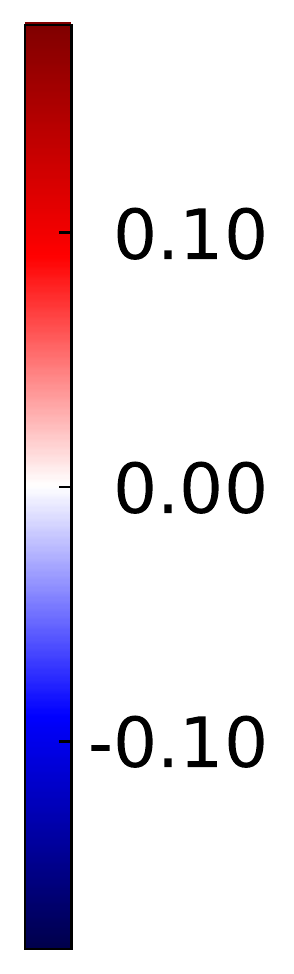}\\
    \vspace{-0.75mm}
    \rot{\scriptsize VINS 50} &
    \includegraphics[width=0.104\linewidth]{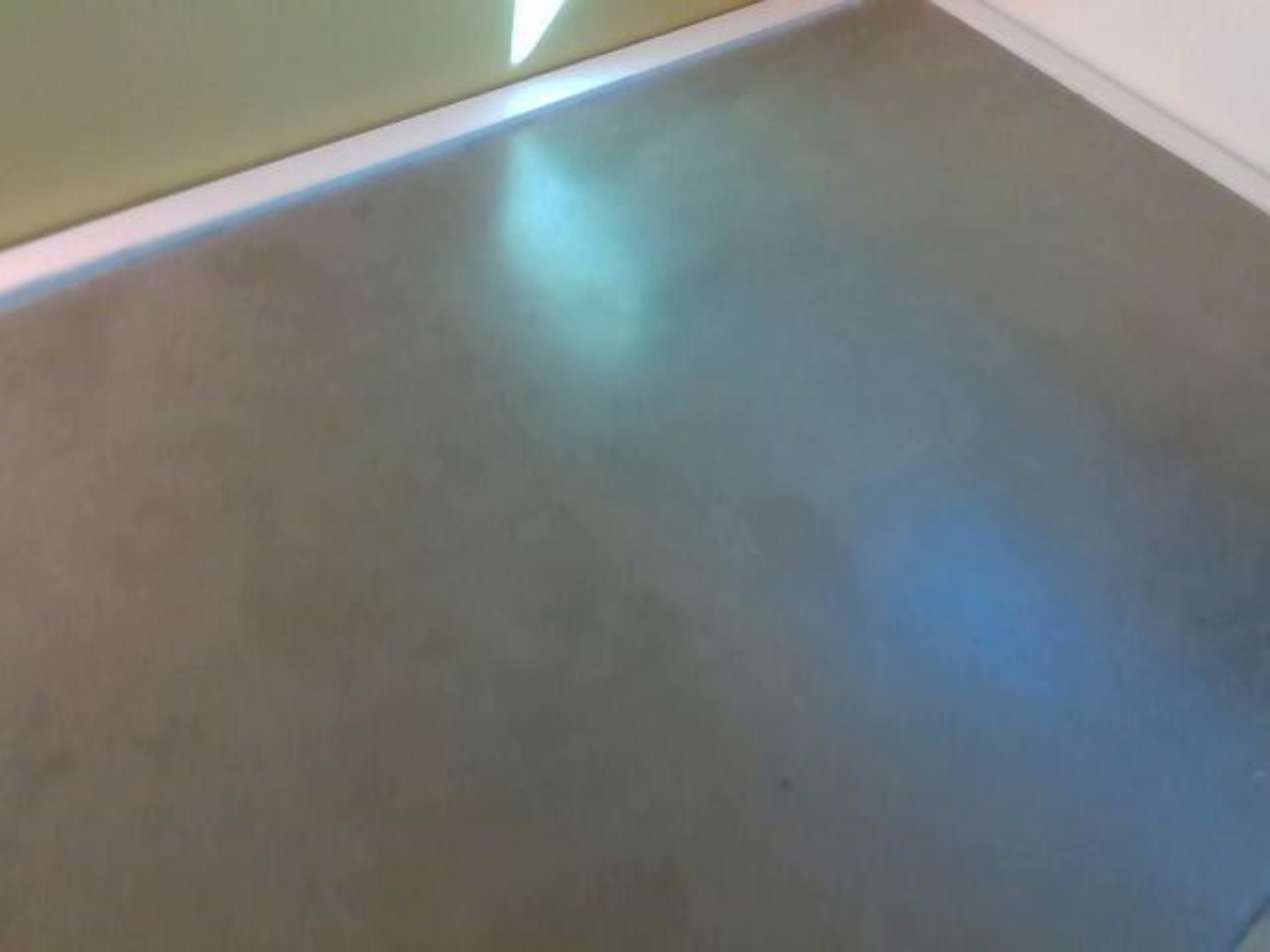}&
    \includegraphics[width=0.104\linewidth]{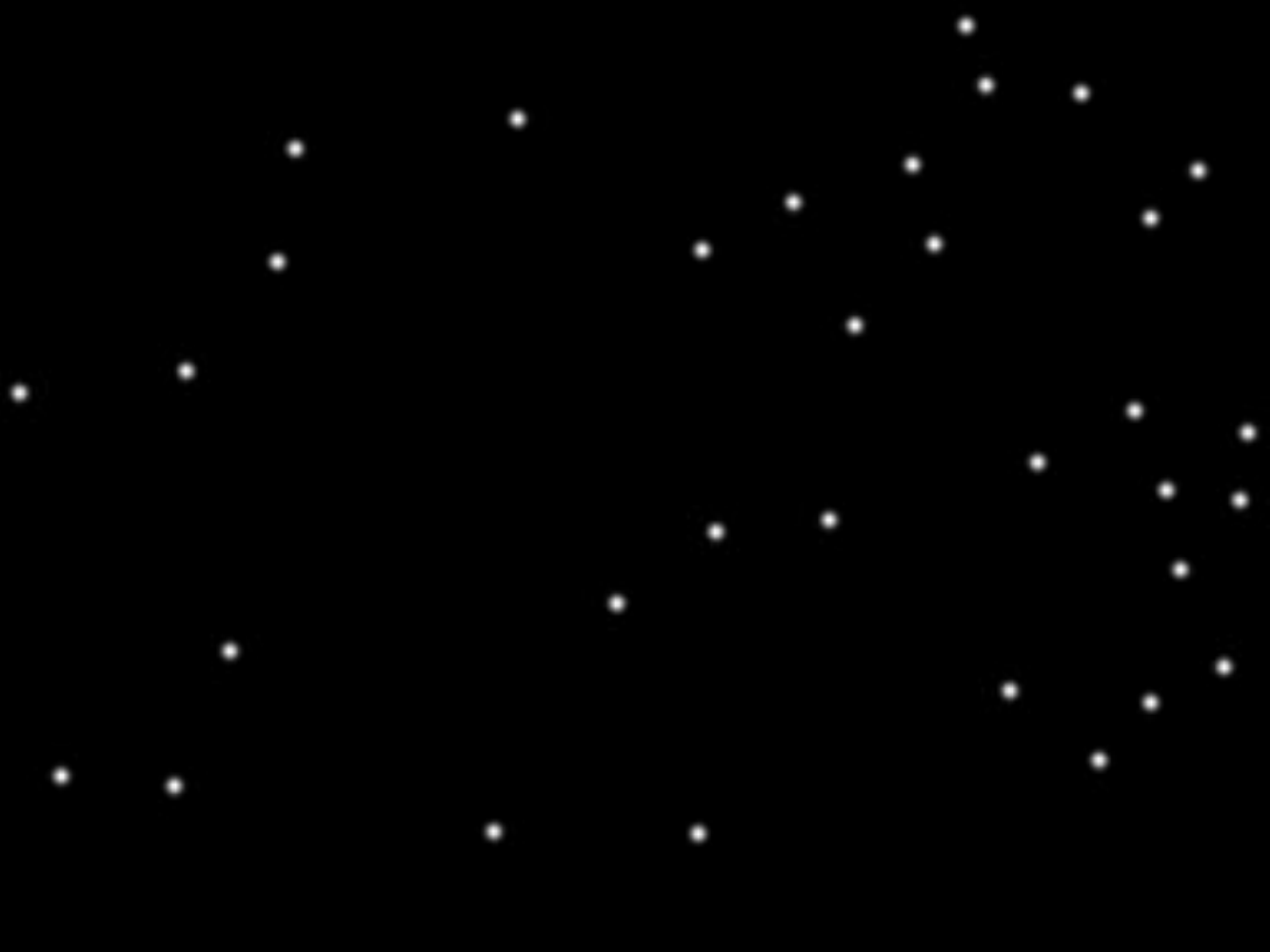}&
    \includegraphics[width=0.104\linewidth]{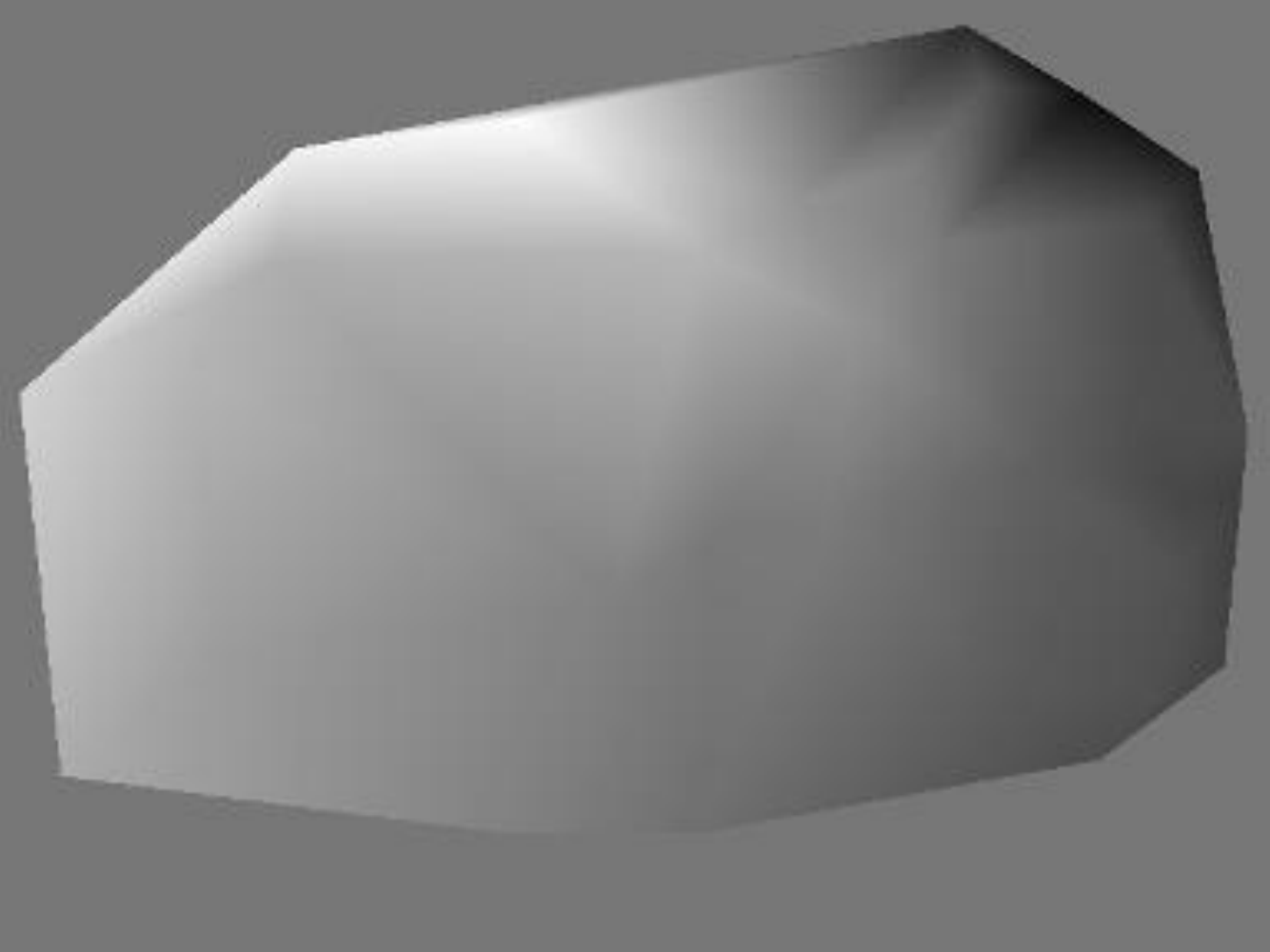}&
    \includegraphics[width=0.104\linewidth]{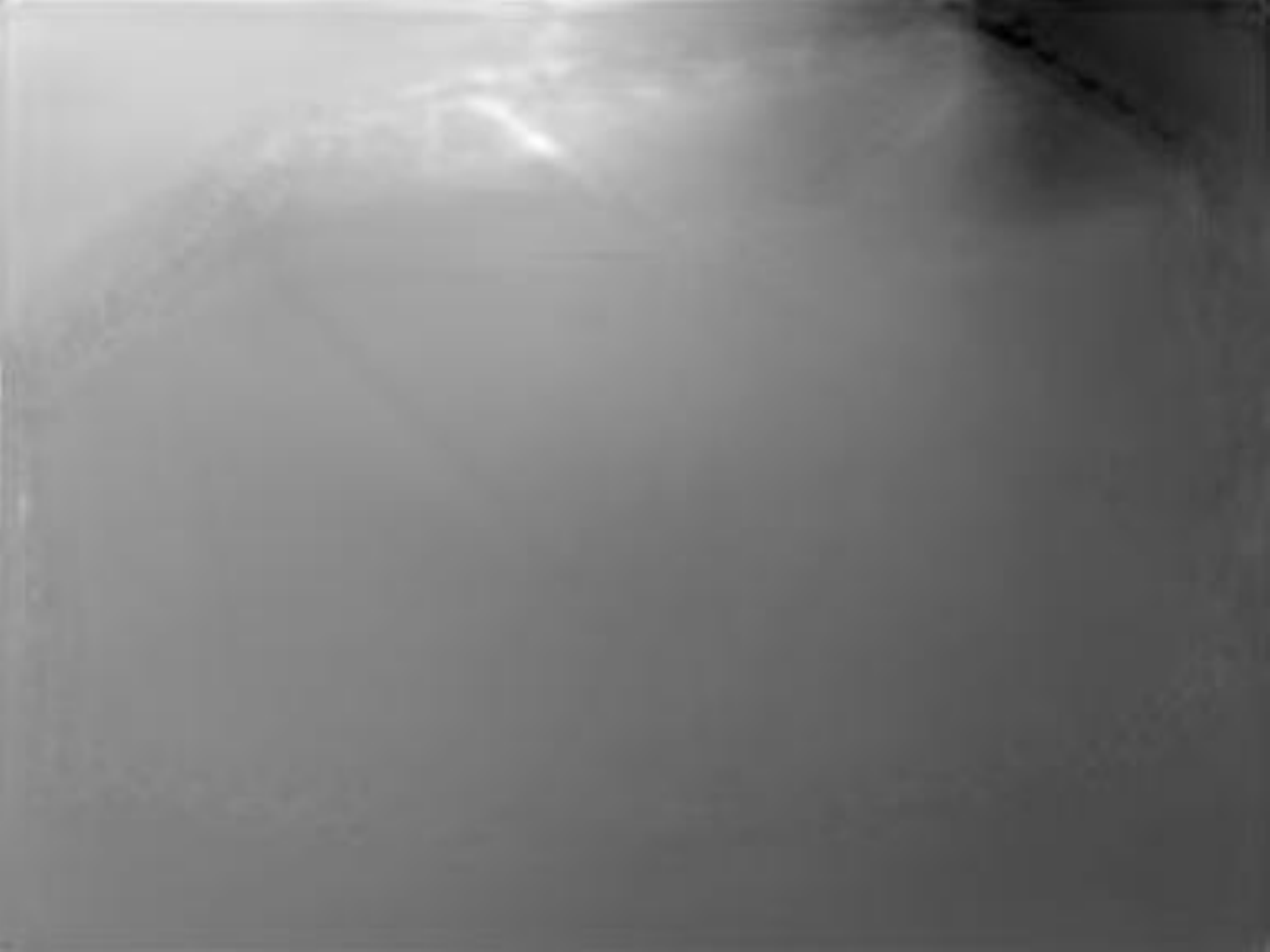}&
    \includegraphics[width=0.104\linewidth]{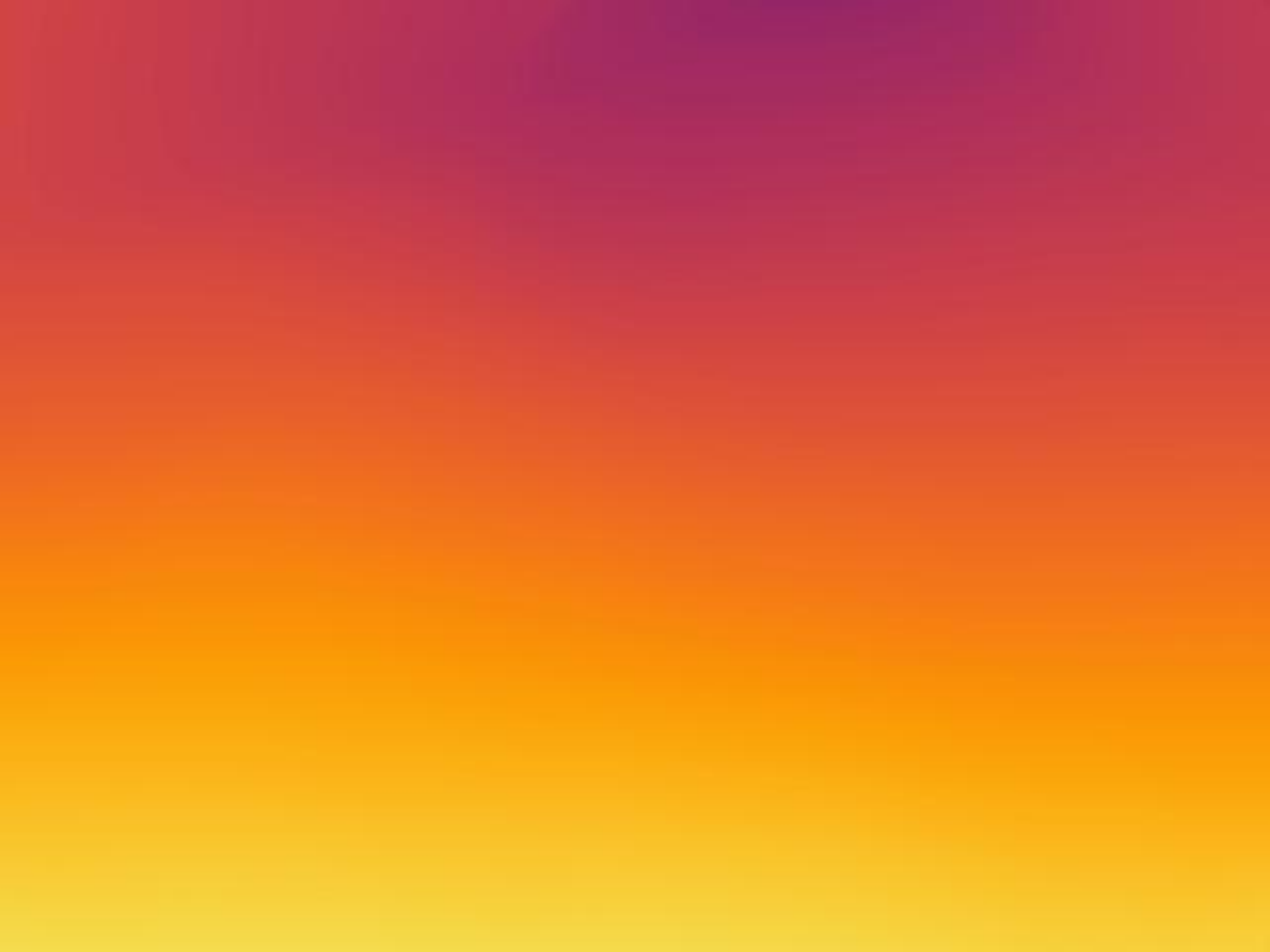}&
    \includegraphics[width=0.104\linewidth]{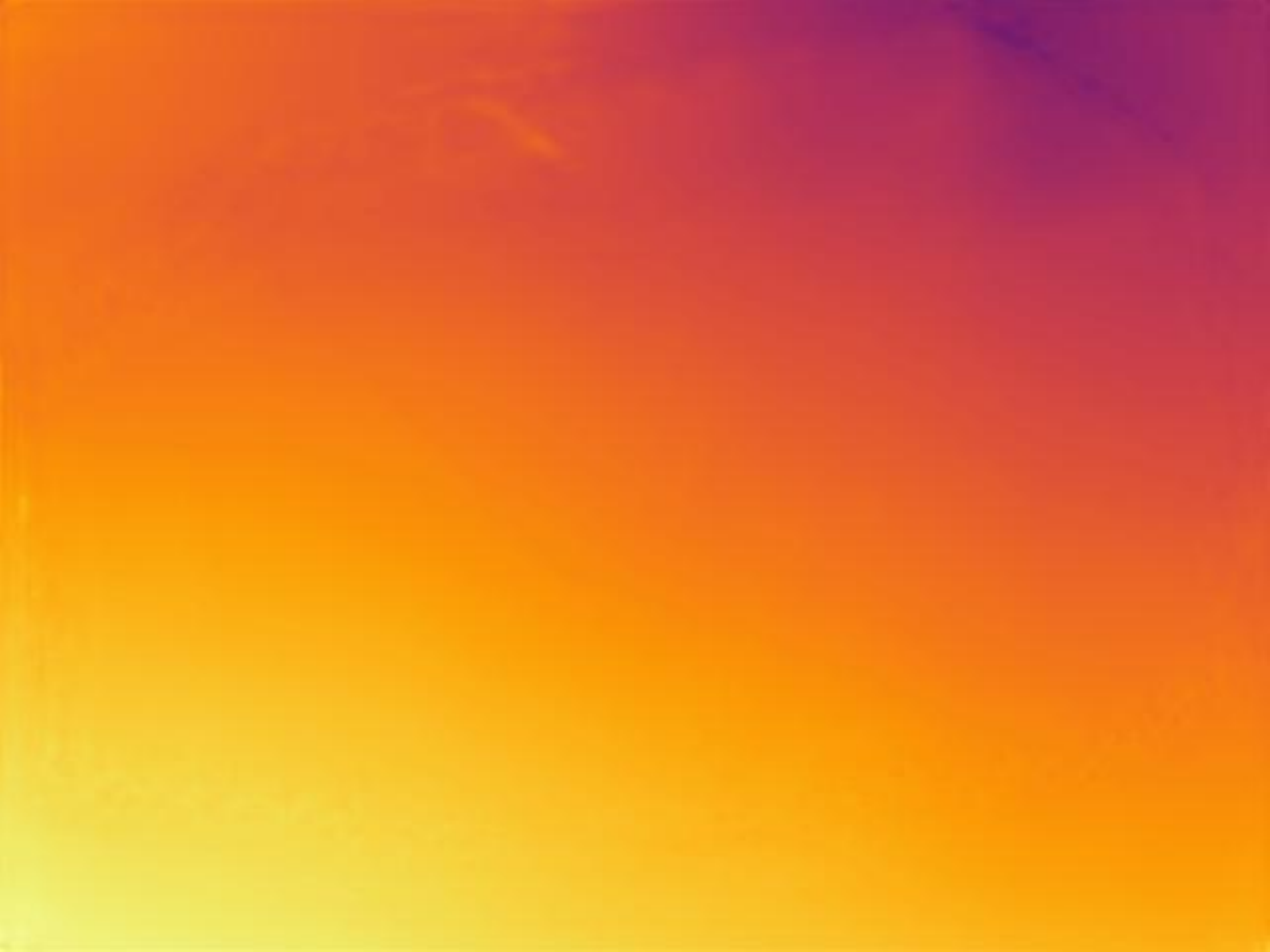}&
    \includegraphics[width=0.104\linewidth]{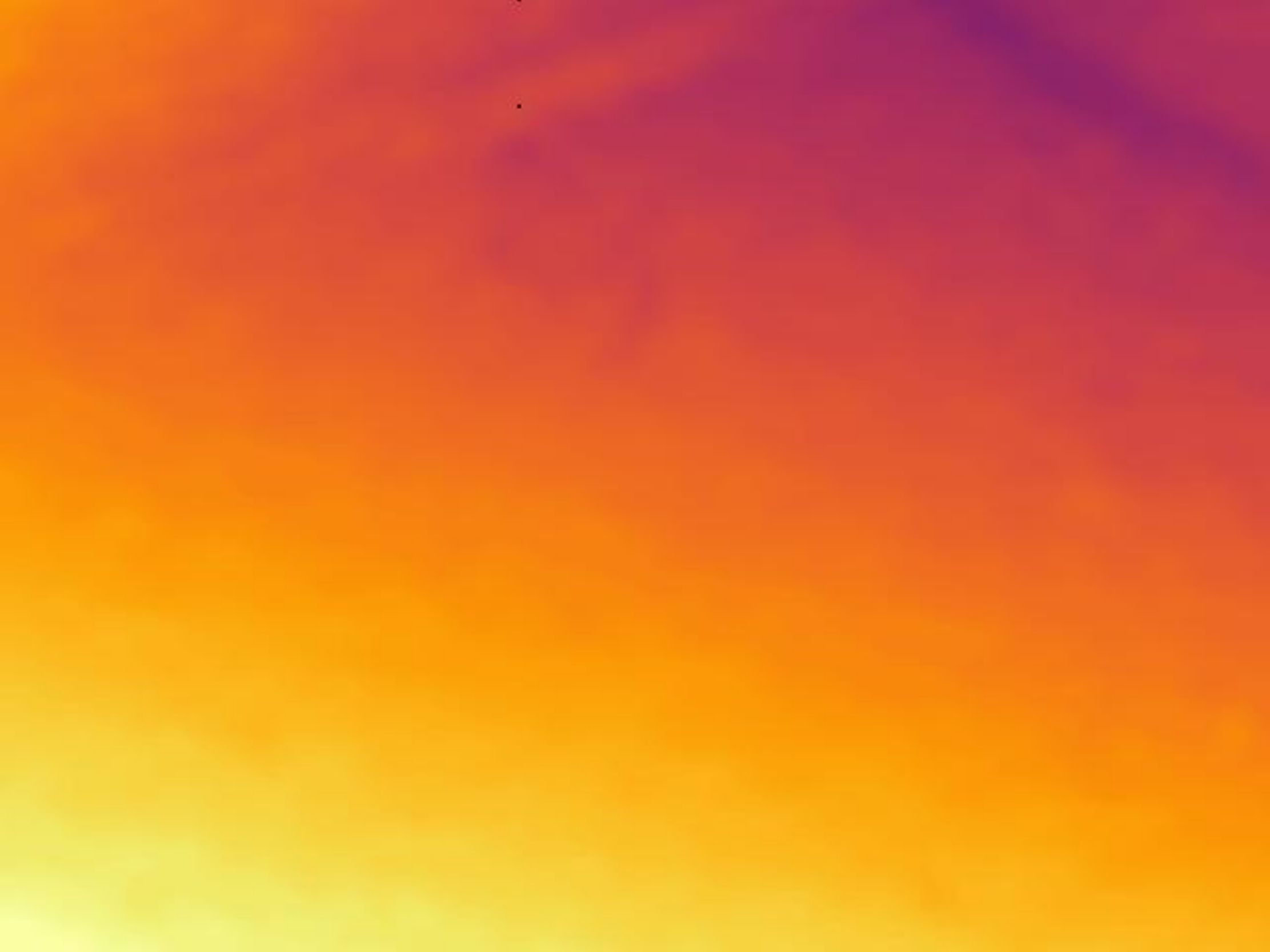}&
    \includegraphics[width=0.104\linewidth]{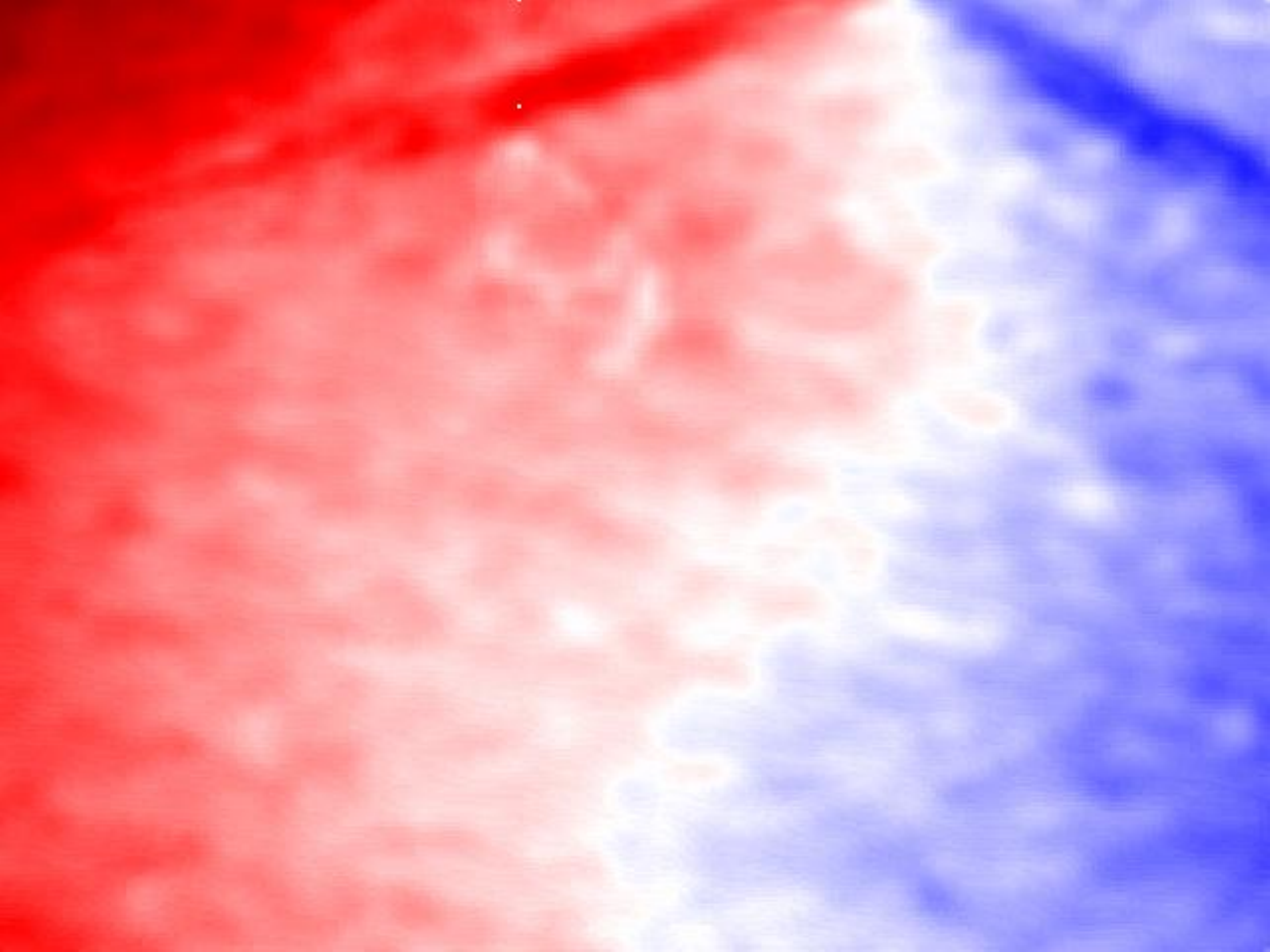}&
    \includegraphics[width=0.104\linewidth]{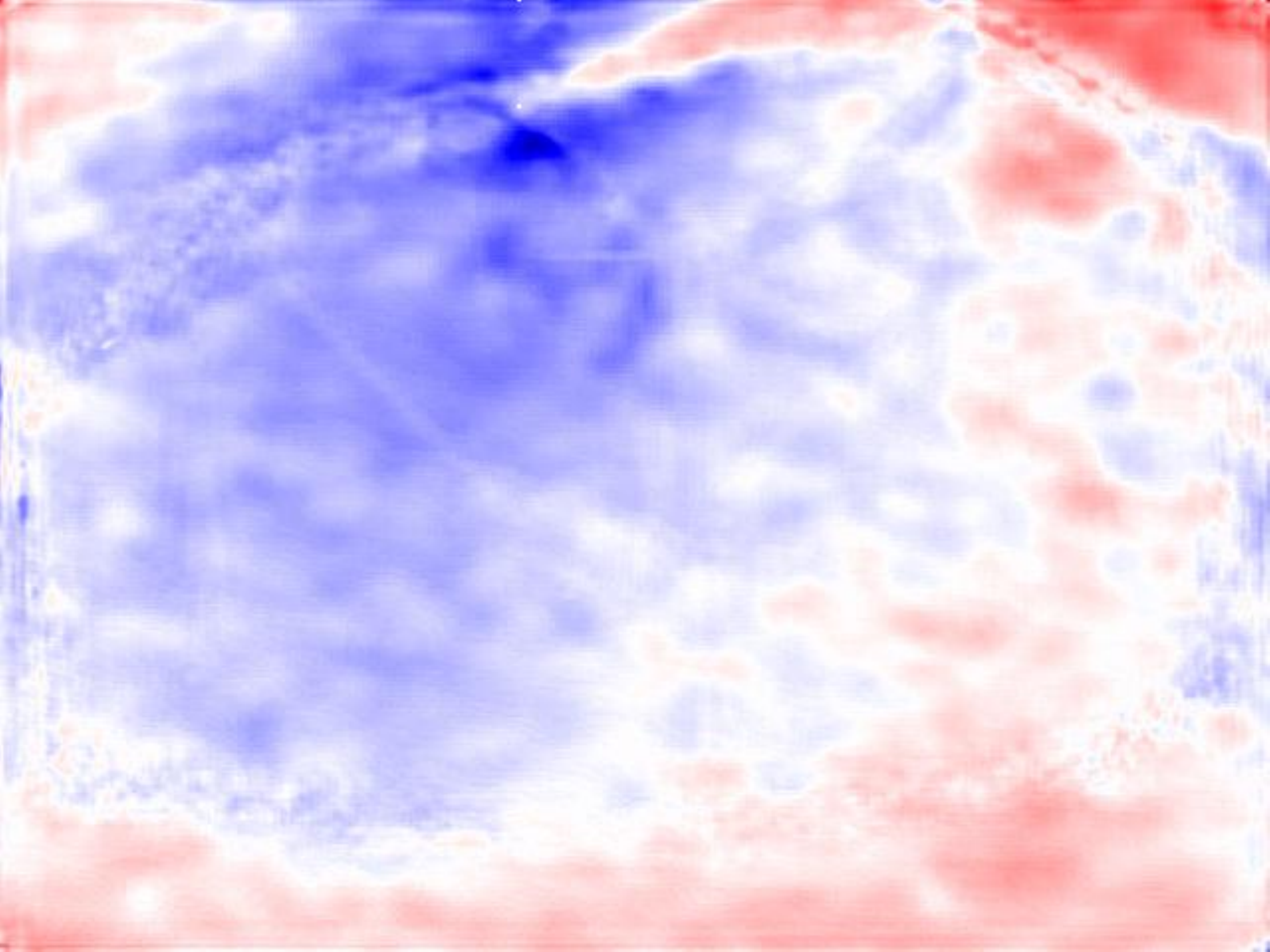}&
    \includegraphics[width=0.024\linewidth]{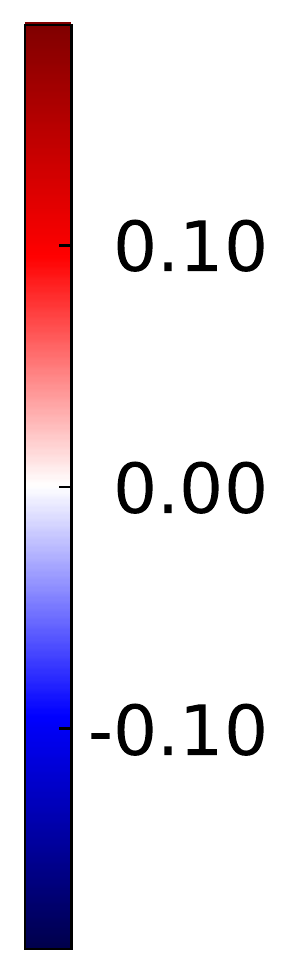}\\   
  \end{tabular}
  \caption{Visualization on two VOID samples to accompany Table~\ref{tab:vis_void_low_vins}. Sparse depth obtained using VINS-Mono is more spread out. This, together with constraining to a lower density of 50 points, leads to inferior scale map scaffolding. It becomes harder for SML to perform dense scale regression in regions originally devoid of sparse metric depth information, as seen with the VINS 50 samples here. Overall, our proposed method still succeeds in reducing metric depth error at very low densities of sparse points. The models used to generate these results were trained solely on VOID (without TartanAir pretraining) and assume a DPT-Hybrid depth estimator.} 
  \label{fig:vis_void_vins}
\end{figure*}

\mypara{Reducing sparse metric depth density.} We further investigate the performance of our pipeline at low numbers of sparse metric depth points. Still evaluating on VOID, we preprocess the VOID-150 dataset using a VINS-Mono frontend re-implementation~\cite{Lusk2018} (the same sparsifier that is used on TartanAir data in our ablation and pretraining experiments). The feature tracker ensures a minimum distance between tracked features, resulting in lower clustering and higher coverage of projected sparse points across the image area. We perform two experiments with VINS-Mono on VOID: tracking strictly up to 150 features (0.05\% density at VGA resolution), and tracking strictly up to 50 features (0.02\% density). Table~\ref{tab:vis_void_low_vins} lists the results of these sparsifier experiments, and Figure~\ref{fig:vis_void_vins} shows the corresponding visualization. For fair comparison against related works, all SML models evaluated for this study are trained directly on VOID without any pretraining on TartanAir. Global alignment (GA) performs similarly at low densities; there are sufficient points to determine appropriate global scale and shift estimates. Dense alignment with SML continues to improve metric depth accuracy over GA even at the very low density of 50 sparse depth points. However, SML learning is worse when using VINS-Mono as a sparsifier, since the larger spacing between tracked features results in inferior scale map scaffolding. As a result, the improvement margins achievable with SML are observed to be lower.

\begin{table*}[p]
  \footnotesize
  \setlength{\tabcolsep}{1pt} 
  \renewcommand{\arraystretch}{1} 
  \centering
  \caption{We investigate how our method performs at low densities of sparse metric depth points. We process VOID data with a VINS-Mono frontend~\cite{Lusk2018} and constrain to lower densities: $\leq$150 and $\leq50$ refer to tracking strictly up to 150 and 50 features, respectively. We report the resulting average number of sparse points per sample (avg PPS), aggregated over the full dataset.}
  \label{tab:vis_void_low_vins}
  \begin{tabular}{@{}
  l@{\hspace{0mm}}
  c@{\hspace{2mm}}
  c@{\hspace{4mm}}
  c@{\hspace{4mm}}
  *{2}{S[table-format=3.2]@{\hspace{1mm}}l@{\hspace{3mm}}}
  *{1}{S[table-format=2.2]@{\hspace{1mm}}l@{\hspace{3mm}}}
  *{1}{S[table-format=3.2]@{\hspace{1mm}}l@{\hspace{3mm}}}
  *{1}{S[table-format=1.3]@{\hspace{0mm}}l@{\hspace{0mm}}}
  @{}}
    \toprule
    Method & Sparsifier & Density & {Avg PPS} & {MAE} & & {RMSE} & & {iMAE} & & {iRMSE} & & {iAbsRel} & \\
    \midrule
    \multicolumn{14}{l}{\textit{evaluating over all 800 samples of VOID test set}} \\
    \midrule
    VOICED-S\cite{Wong2020void} & \multirow{2}{*}{XIVO\cite{Fei2019xivo}} & \multirow{2}{*}{150} & \multirow{2}{*}{$\sim$188} & 174.04 & & 253.14 & & 87.39 & & 126.30 & & {---} & \\
    KBNet\cite{Wong2021kbnet}   &                                         &                      & & 131.54 & & 263.54 & & 66.84 & & 128.29 & & {---} & \\
    \midrule
    GA Only & \multirow{2}{*}{XIVO\cite{Fei2019xivo}}   & \multirow{2}{*}{150} & \multirow{2}{*}{$\sim$188} & 165.33 &          & 243.11 &          & 75.74 &          & 106.37 &          & 0.103 &          \\	
    GA+SML  &                                           &                      & &  97.03	& \fg{-41} & 167.82	& \fg{-31} & 46.62 & \fg{-38} &	 74.67 & \fg{-30} &	0.063 & \fg{-39} \\
    \midrule
    \multicolumn{14}{l}{\textit{excluding first-in-sequence samples due to lack of tracking in initial sample $\implies$ evaluating over 792/800 samples of VOID test set}} \\
    \midrule
    GA Only & \multirow{2}{*}{XIVO\cite{Fei2019xivo}}   & \multirow{2}{*}{150} & \multirow{2}{*}{$\sim$188} & 164.98 &          & 242.38 &          & 75.91 &	      & 106.65 &          & 0.103 &          \\
    GA+SML  &                                           &                      & &  96.78 & \fg{-41} & 167.35 & \fg{-31} & 46.73 & \fg{-38} &  74.87 & \fg{-30} & 0.062 & \fg{-40} \\
    \midrule
    GA Only & \multirow{2}{*}{VINS-Mono\cite{Lusk2018}} & \multirow{2}{*}{$\leq$150} & \multirow{2}{*}{$\sim$104} & 167.11 &          & 248.68 & 	       & 76.64 &          & 108.18 &          & 0.104 &          \\
    GA+SML  &                                           &                      & & 138.42	& \fg{-17} & 223.08 & \fg{-10} & 67.39 & \fg{-12} &	 98.96 &  \fg{-9} &	0.092 & \fg{-12} \\
    \midrule
    GA Only & \multirow{2}{*}{VINS-Mono\cite{Lusk2018}} & \multirow{2}{*}{$\leq$50}  & \multirow{2}{*}{$\sim$44} & 180.29 &          & 264.95 &          & 81.94 &          & 114.72 &          & 0.111 &          \\
    GA+SML  &                                           &                      & & 148.65	& \fg{-18} & 236.71 & \fg{-11} & 70.23 & \fg{-14} & 103.93 &  \fg{-9} & 0.095 & \fg{-14} \\
    \bottomrule
  \end{tabular}
\end{table*}

\mypara{Benchmarking runtime.} We expand our performance evaluation to include more depth estimator models and report runtimes at native resolutions in Table~\ref{tab:appendix_performance}. Our primary target platform remains the Jetson AGX Orin board. There is an observed variance in the device-to-host copy overhead when moving the depth map between the GPU to host CPU for global alignment. Our pipeline achieves real-time performance with mobile-friendly depth models like DPT-LeViT and MiDaS-small. TensorRT (TRT) enables our pipeline to run at over 10 fps on the Jetson TX2 platform.
\section{Limitations}

When integrated into a real-world system, our approach would rely on successful VIO to produce sparse depth; hence, the limitations of our work are driven by limitations of visual-based odometry. In cases of rapid motion, poor lightning, or lack of sufficient texture, VIO will have difficulty tracking landmarks and output very few or even no points. Fewer sparse depth points will result in less reliable estimates for global scale and shift, and without any points, global alignment will fail. For scale map scaffolding, at least three non-colinear points are needed so that interpolation does not occur over a convex hull that is a point or line. Whenever fewer sparse depth points are provided, the resulting scale map scaffolding becomes an identity map and loses meaning. This hinders how well our SML network can regress dense scales. Similarly to how motion and depth estimation can be alternated to boost accuracy in both tasks \cite{Teed2020DeepV2D}, exploring conditions under which odometry-guided metric depth alignment could help recovery from odometry failure would be interesting future work.
\end{appendices}


\end{document}